\documentclass[letterpaper]{article} 
\usepackage{aaai2026}  
\usepackage{times}  
\usepackage{helvet}  
\usepackage{courier}  
\usepackage[hyphens]{url}  
\usepackage{graphicx} 
\usepackage{xcolor}
\usepackage{natbib}  
\usepackage{caption} 
\usepackage[ruled,vlined]{algorithm2e}
\usepackage{bm}
\usepackage{amsmath}
\usepackage{amssymb}
\usepackage{mathtools}
\usepackage{subcaption}
\usepackage{booktabs}
\usepackage{comment}

\extrafloats{400}
\nocopyright
\title{Forest vs Tree: The $(N, K)$ Trade-off in Reproducible ML Evaluation}
\author {
    Deepak Pandita\textsuperscript{\rm 1},
    Flip Korn\textsuperscript{\rm 2},
    Chris Welty\textsuperscript{\rm 2},
    Christopher M. Homan\textsuperscript{\rm 1}
}
\affiliations {
    \textsuperscript{\rm 1}Rochester Institute of Technology\\
    \textsuperscript{\rm 2}Google Research\\
    deepak@mail.rit.edu, flip@google.com, cawelty@gmail.com, cmh@cs.rit.edu
}

\begin{document}

\maketitle
\sloppy 
\newcommand{\fkcolor}[1]{{\color{purple}#1}} 
\newcommand{\fk}[1]{\fkcolor{\small\textbf{Flip: }\sf #1}}

\newcommand{\pv}{$p$-value}
\begin{abstract}

Reproducibility is a cornerstone of scientific validation and of the authority it confers on its results. Reproducibility in machine learning evaluations leads to greater trust, confidence, and value. However, the ground truth responses used in machine learning often necessarily come from humans, among whom disagreement is prevalent, and surprisingly little research has studied the impact of effectively ignoring disagreement in these responses, as is typically the case. One reason for the lack of research is that budgets for collecting human-annotated evaluation data are limited, and obtaining more samples from multiple raters for each example greatly increases the per-item annotation costs. We investigate the trade-off between the number of items ($N$) and the number of responses per item ($K$) needed for reliable machine learning evaluation. We analyze a diverse collection of categorical datasets for which multiple annotations per item exist, and simulated distributions fit to these datasets, to determine the optimal $(N, K)$ configuration, given a fixed budget ($N \times K$), for collecting evaluation data and reliably comparing the performance of machine learning models. Our findings show, first, that accounting for human disagreement may come with $N \times K$ at no more than 1000 (and often much lower) for every dataset tested on at least one metric. Moreover, this minimal $N \times K$ almost always occurred for $K > 10$. Furthermore, the nature of the tradeoff between $K$ and $N$, or if one even existed, depends on the evaluation metric, with metrics that are more sensitive to the full distribution of responses performing better at higher levels of $K$. Our methods can be used to help ML practitioners get more effective test data by finding the optimal metrics and number of items and annotations per item to collect to get the most reliability for their budget.
\end{abstract}

\begin{links}
    \link{Code}{https://github.com/google-research/vet}
\end{links}

\section{Introduction}

The scientific community, including the rapidly evolving fields of AI and NLP, is grappling with a pervasive reproducibility crisis \cite{baker_1500_2016, Gundersen_Kjensmo_2018, hutson_2018, mieskes-etal-2019-community, Gundersen_2020}. Researchers are increasingly unable to replicate the results of previous studies \cite{NEURIPS2019_c429429b}, thus undermining trust in experimental and empirical research. In machine learning, where data-driven research is essential for advancing knowledge, the comparison of models is central to determining the state-of-the-art for a given task. As such, ensuring the reproducibility of results through robust evaluation is critical. 

\begin{figure}[t]
    \centering
    \includegraphics[width=\columnwidth]{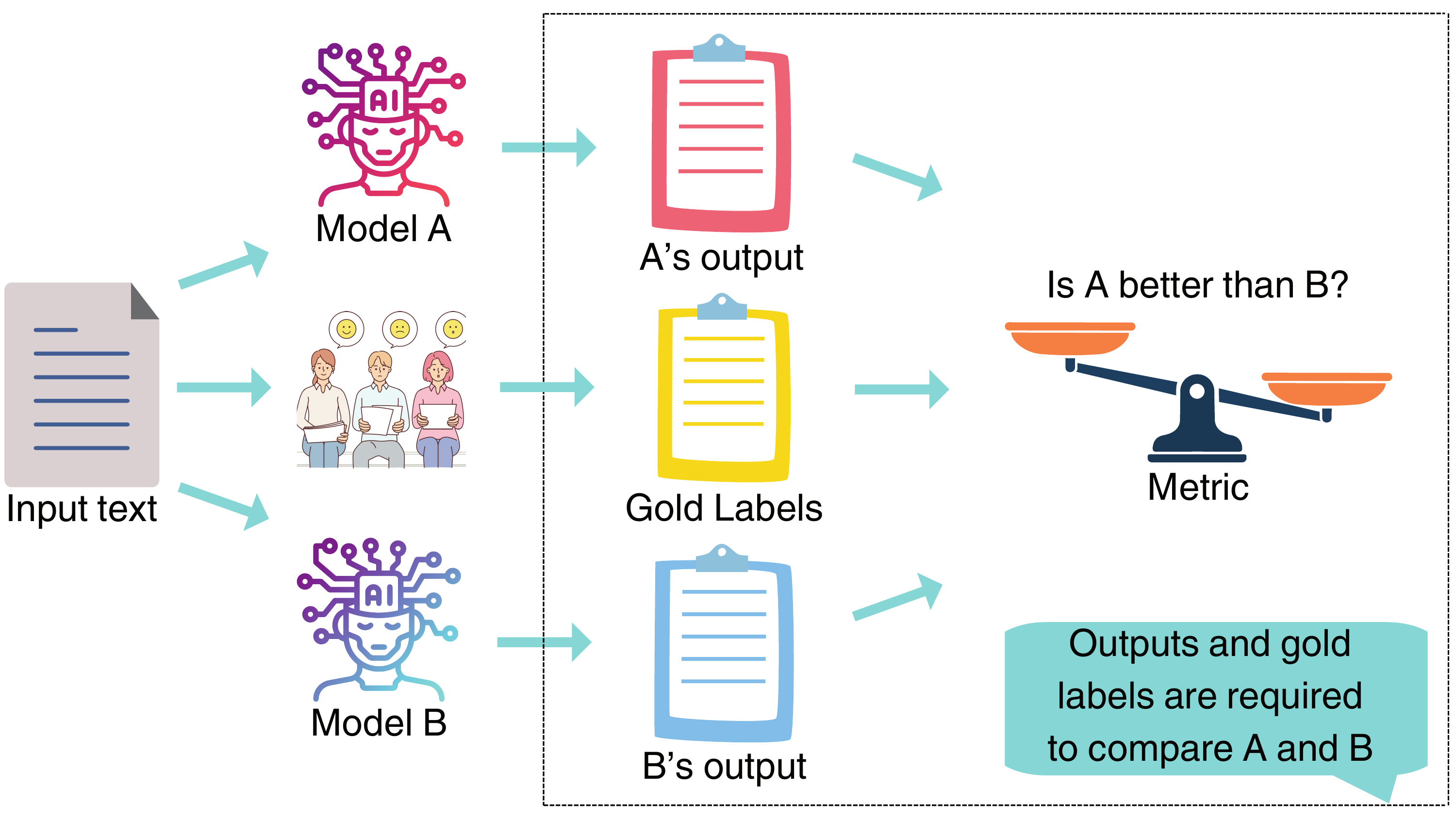}
    \caption{Model assessment process with ground truth.}
    \label{fig:fig_model_assessment}
\end{figure}

We focus here on an underlooked source of unreliability: failing to account for human disagreement and other sources of randomness in ML evaluation.
Conventional evaluation approaches treat disagreement, if at all, as nothing more than noise and may aggregate 3--5 labels per item---a number that comes from literature on machine learning \cite{snow2008cheap}, not machine learning evaluation---via plurality voting to represent consensus, overlooking disagreement, which is endemic in human responses, as anyone who has participated in a democratic process such as voting knows. 

Recent papers have advocated using and publishing disaggregated labels to account for human label variation \cite{basile-etal-2021-need, prabhakaran-etal-2021-releasing, plank-2022-problem, Cabitza_Campagner_Basile_2023}.
The field also faces a pervasive issue of inadequate statistical analysis; statistical significance is often misapplied, and reported outcomes are frequently unreliable \cite{sogaard-etal-2014-whats, dror-etal-2018-hitchhikers, van-der-lee-etal-2019-best}.

A crucial question, from a statistical perspective, is \emph{how much data needs to be collected to ensure statistically reliable testing}, via null hypothesis significance tests (NHSTs) and confidence intervals (CIs)? We are particularly interested in challenging the assumption (in a skeptical manner) that a small number of annotations per item is sufficient. It would seem that there should be a trade-off between the number of items $N$, i.e., how many trees are observed, and the number of annotations per item collected, i.e., the resolution by which each tree is observed. It would seem that the nature of the trade-off might depend on the metrics used, which depend on the performance expectations of the model{s} under consideration.
In this same vein, we would like to know whether the newer, but still uncommon approach of keeping disaggregated responses for each item has value from the perspective of comparing one ML model against another---the most basic way to evaluate ML models.
We investigate the following research questions:
\begin{description}
    \item[RQ1] What is the lowest number of total annotations needed $N\times K$ to ensure reasonably repeatable results in comparing two models?
    \item[RQ2] How does this this number $N\times K$ depend on:
    \begin{itemize}
    \item the distribution of responses found in five actual datasets with disaggregated annotations?
    \item the metric used?
    \item the number of categories?
    \item the statistical instrument (NHSTs vs. CIs)?
    \end{itemize}
    \item[RQ3] For fixed $N \times K$ (particularly the minimal ones found in RQ2), what is the smallest value of $K$ that ensures reasonably repeatable results, and how does this vary according to the same variables in RQ2?
\end{description}

To address these questions, we make the following contributions.

\begin{enumerate}
    \item   We believe this is the first paper to examine the optimization problem of allocating a human annotation budget to a sample of $N$ items, where each item is annotated by $K$ raters, such that the total budget $N \times K$ is fixed.

\item How many items, and how many annotations per item, to collect needs to be known \emph{before} the data is collected, but the answer should be based on realistic assumptions.
Towards this end, we apply a Bayesian approach to model existing datasets via simulation for arbitrary $N$ and $K$.
This enables a more robust way of modeling when the sample size is small (assuming accurate priors) and allows for maximum \emph{a posteriori} (MAP) fitting of data, versus maximum likelihood estimation (MLE)-based frequentist approaches, which provides regularization.
\item We extend an existing simulator to model categorical data and confidence intervals (along with NHSTs).
\item We report on a comprehensive set of experiments using five different real datasets as well as synthetic data, to demonstrate the impact on statistical significance and confidence of optimizing the trade-off between $N$ and $K$.
\end{enumerate}

Our findings show, first, that accounting for human disagreement may come with $N \times K$ at no more than 1000 (and often much lower) for every dataset tested on at least one metric. Moreover, this minimal $N \times K$ almost always occurred for $K > 10$.
Moreover, the nature of the tradeoff between $K$ and $N$---or if one even existed---depends on the evaluation metric, with metrics that are more sensitive to the full distribution of responses performing better at higher levels of $K$.

\section{Related Work}
\label{sec:rw}

The reproducibility crisis in AI and NLP, highlighted by numerous studies \cite{Gundersen_Kjensmo_2018, hutson_2018, mieskes-etal-2019-community, Gundersen_2020}, stems from various factors. A major contributor is the inherent non-deterministic nature of machine learning methods, algorithms, and implementations; even with shared code, multiple seemingly identical training runs of the same deep learning model can yield different models and test results, often due to factors like varying random seeds or hardware-specific operations \cite{pham_problems_2020}. Furthermore, a survey by \citet{pham_problems_2020} of 901 participants revealed that $84\%$ were either unaware or unsure about the variance stemming from different implementations. \citet{arvan-etal-2022-reproducibility-code} further underscored this challenge by achieving only a $25\%$ success rate in a reproducibility study of eight papers published in EMNLP 2021. These findings underscore the critical need to account for this inherent variance in machine learning evaluations, even when working with seemingly identical setups.

Beyond model-inherent variance, the human element in evaluation also introduces considerable variability. Human raters are frequently recruited to generate reference labels, commonly referred to as ``gold standards," for evaluating machine learning model performance. However, human disagreement is prevalent, especially in subjective tasks, leading to significant variance in responses \cite{basile-etal-2021-need, prabhakaran-etal-2021-releasing, uma2021learning, plank-2022-problem, Cabitza_Campagner_Basile_2023, homan2023intersectionality, weerasooriya-etal-2023-vicarious, prabhakaran2023framework, pandita-etal-2024-rater}. Typically, responses are aggregated via plurality voting to represent consensus, though recent work has shown the inadequacy of such aggregation for incorporating response variance \cite{barile2021toward, davani-etal-2022-dealing}. It is therefore not surprising that human evaluation studies also show a low degree of reproducibility \cite{belz_non-repeatable_2023}. The issue of response variance has been particularly explored in the context of conversational AI safety. For instance, \citet{homan2023intersectionality} utilized Bayesian multilevel models to understand the impact of rater demographics on safety ratings, while \citet{prabhakaran2023framework} proposed a framework to analyze diversity in safety ratings among rater subgroups. Further, \citet{NEURIPS2023_a74b697b} introduced a dedicated dataset to enable in-depth analysis and measurement of response variance in this domain.

\citet{wein-etal-2023-follow} proposed a framework and simulator using NHSTs to estimate the true \pv\ of model comparisons. The simulator considers item and response variance to sample a ``reference test set" of gold/human responses ($G$) and the responses of two models ($A$ and $B$). To construct responses for $G$, it produces $N$ items and $K$ responses per item using random variables. For each item, the mean and standard deviation are sampled from uniform distributions. Then, $K$ continuous responses are drawn using a normal distribution parameterized with the sampled mean and standard deviation. The responses for items in model $A$ are sampled using the same distribution as $G$, making model $A$ an ideal representation of $G$. Responses for items in model $B$ are sampled using the same mean as $G$ but with standard deviation perturbed by a small amount. Perturbation parameters for each item are randomly drawn at a given perturbation level from a uniform distribution. As the perturbation level increases, model $B$ increasingly differs from model $A$. NHSTs are used to estimate the \pv s\ for comparing model $A$ and model $B$ under different metrics and sampling methods. The data for the null hypothesis is generated assuming that responses for model $A$ and model $B$ are drawn from the same distribution. This is achieved by combining the responses from model $A$ and model $B$.

\citet{homanmany} utilized the simulator presented by \citet{wein-etal-2023-follow} to propose an evaluation framework tailored for foundation models.
Our approach builds on these prior works. We propose a framework for power analysis designed for evaluating machine learning models using NHSTs. Crucially, our framework accounts for both item and response variance, specifically under the assumption that the responses are nominal.

\section{Methods}

\begin{figure}[t]
    \centering
    \includegraphics[width=\columnwidth]{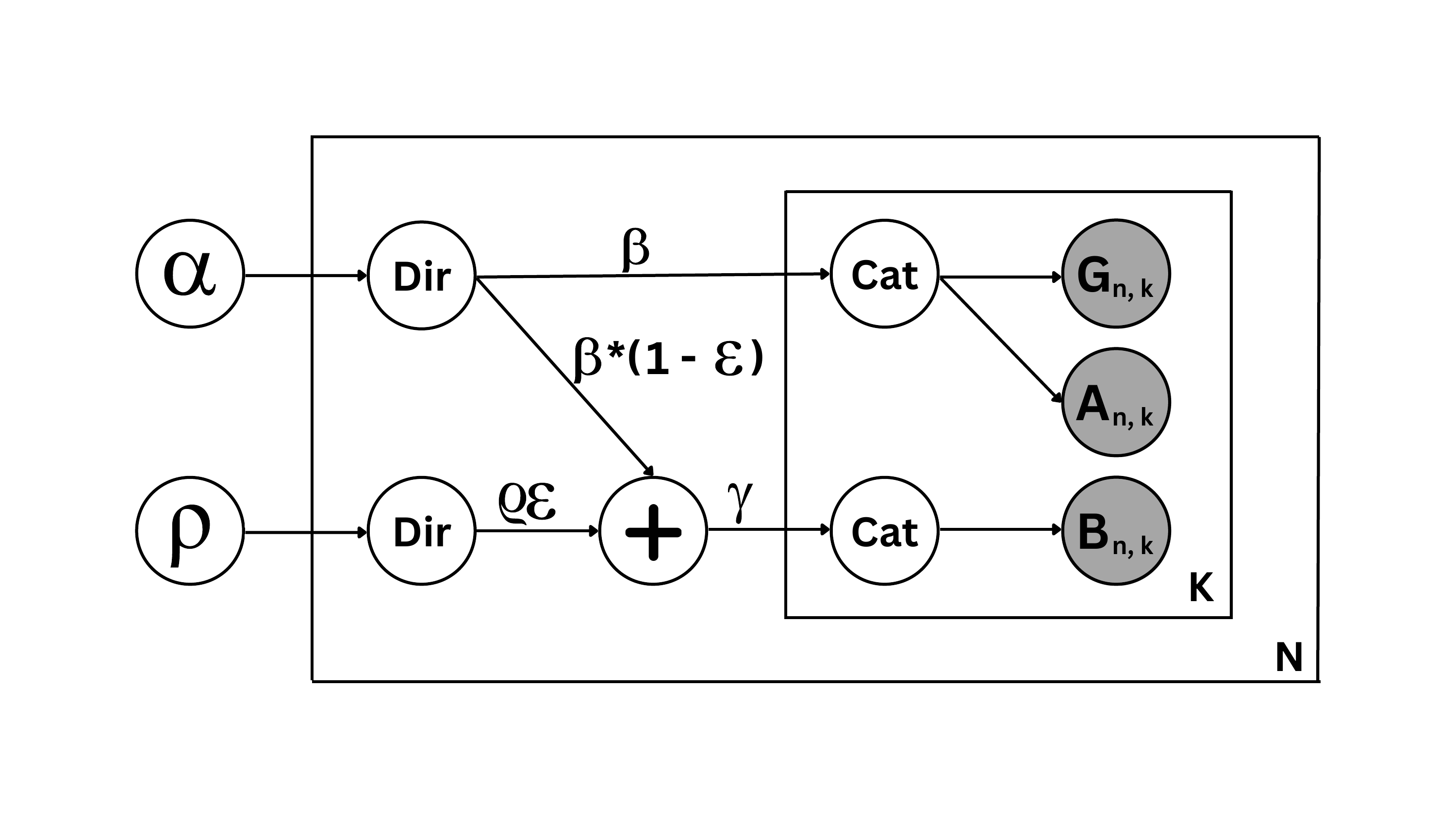}
    \caption{Plate notation for the simulator. Categorical parameters ($\beta$) and noise parameters ($\varrho$) are sampled from two Dirichlet distributions parameterized by $\alpha$ and $\rho$, respectively. Then, responses for $G$ and $A$ are produced by sampling from a categorical distribution parameterized by $\beta$. Responses for $B$ are produced by sampling from a categorical distribution parameterized by $\gamma$, where $\gamma$ is a convex combination of $\beta$ and $\varrho$ controlled by the perturbation parameter $\epsilon$.}
    \label{fig:fig_plate_notation}
\end{figure}

We use a simulator that generates the outputs for comparing two models, $A$ and $B$, against gold standard outputs $G$ (see Figure \ref{fig:fig_model_assessment} for an illustration). Our methods differ from earlier work \cite{wein-etal-2023-follow, homanmany} in that the responses produced by the simulator are categorical rather than continuous. The simulator produces the gold outputs $G$ and the outputs for model $A$ by sampling from a Dirichlet-categorical distribution, making $A$ an ideal model. For model $B$, the annotations are sampled after introducing noise in the model parameters controlled by the perturbation parameter $\epsilon$, making it slightly worse than model $A$. We then employ NHSTs to estimate the \pv\ for this comparison.

We use a Dirichlet-categorical distribution because it naturally models the probability of observing the categories, and the Dirichlet distribution is the conjugate prior for the categorical distribution, simplifying the calculations involved in Bayesian inference. When the prior is conjugate, the posterior distribution is also a Dirichlet distribution. Bayesian inference gives us the flexibility to incorporate prior information about response probabilities and produces more robust estimates, especially when the data is limited.

\subsection{Simulation Framework}
\label{sec:simulation_framework}

 We generate the gold outputs $G$ and the outputs for two models $A$ and $B$. Each sample consists of $N$ items and $K$ responses per item. The responses are discrete values chosen from $M$ categories. For each item $i \in N$, we sample categorical parameters ($\beta_i$) and noise parameters ($\varrho_i$) from two Dirichlet distributions parameterized by $\alpha$ and $\rho$, respectively. For $G$ and $A$, the response by each rater $j \in K$ is sampled from a categorical distribution parameterized by $\beta$. For model $B$, the response by each rater $j$ is sampled from a categorical distribution parameterized by $\gamma$, where $\gamma$ is a convex combination of $\beta$ and $\varrho$ controlled by the perturbation parameter $\epsilon$. This process is illustrated in Figure \ref{fig:fig_plate_notation} and is described in Algorithm \ref{alg:alt}.

\begin{algorithm}[t]
\SetKwInput{KwInput}{Input parameters} 
\caption{Simulations for $H_{alt}$}
\label{alg:alt}
\KwInput{$N, K, M, \alpha, \rho, \epsilon$}
\For{{$i=1$ \KwTo $N$}}{
\tcp{sample categorical parameters}
$\bm{\beta_i} = \beta_{i,1},...,\beta_{i,M} \sim Dir(\alpha_{1},...,\alpha_{M})$ \;
\tcp{sample noise parameters}
$\bm{\varrho_i} = \varrho_{i,1},...,\varrho_{i,M} \sim Dir(\rho_{1},...,\rho_{M})$ \;
\tcp{convex combination of categorical \& noise parameters}
$\bm{\gamma_i} = (1-\epsilon)*\bm{\beta_i} + \epsilon*\bm{\varrho_i}$ \;
\tcc{sample $j$’s response to $i$}
    \tcp{Gold}
    \For{{$j=1$ \KwTo $k$}}{
        $G_{i,j} = Cat(\beta_{i,1},...,\beta_{i,M})$ \;
    }
    \tcp{Model A}
    \For{{$j=1$ \KwTo $K$}}{
        $A_{i,j} = Cat(\beta_{i,1},...,\beta_{i,M})$ \;
    }
    \tcp{Model B}
    \For{{$j=1$ \KwTo $K$}}{
        $B_{i,j} = Cat(\gamma_{i,1},...,\gamma_{i,M})$ \;
    }
}
\end{algorithm}


\begin{algorithm}[t]
\SetKwInput{KwInput}{Input parameters} 
\caption{Simulations for $H_{null}$}
\label{alg:null}
\KwInput{$N, K, M, \alpha, \rho, \epsilon$}
\For{{$i=1$ \KwTo $N$}}{
\tcp{Use same steps as Algorithm \ref{alg:alt} for $\bm{\beta_i}$, $\bm{\varrho_i}$, $\bm{\gamma_i}$ and $G_{i,j}$}
    \tcp{Model A}
    \For{{$j=1$ \KwTo $K$}}{
        $x \sim Bernoulli(0.5)$ \;
        \If{$x==0$}{
            $A_{i,j} = Cat(\beta_{i,1},...,\beta_{i,M})$ \;
        }
        \Else{
            $A_{i,j} = Cat(\gamma_{i,1},...,\gamma_{i,M})$ \;
        }
    }
    \tcp{Model B}
    \For{{$j=1$ \KwTo $K$}}{
        $x \sim Bernoulli(0.5)$ \;
        \If{$x==0$}{
            $B_{i,j} = Cat(\gamma_{i,1},...,\gamma_{i,M})$ \;
        }
        \Else{
            $B_{i,j} = Cat(\beta_{i,1},...,\beta_{i,M})$ \;
        }
    }
}
\end{algorithm}

\subsection{Hypothesis Testing}
\label{hypothesis_testing}

For the alternative hypothesis ($H_{alt}$), we use the responses generated according to Algorithm \ref{alg:alt}.
For the null hypothesis ($H_{null}$), we generate data for two models $A$ and $B$, assuming they are drawn from the same distribution. The process is similar to the one in $H_{alt}$ except the response by each rater is sampled from a categorical distribution with parameters chosen uniformly at random from $\{\beta_i, \gamma_i\}$ for both models $A$ and $B$. This process is described in Algorithm \ref{alg:null}.

To compare $H_{alt}$ and $H_{null}$, we use a metric $\Gamma(A, B, G)$ for each pair of response samples $\{A, B\}$ and gold samples $G$ to obtain a score. $\Gamma(A, B, G) = \Gamma(A, G) - \Gamma(B, G)$, where larger is better and $\Gamma(A, B, G) = \Gamma(B, G) - \Gamma(A, G)$, where smaller is better. For each hypothesis, a distribution over metric scores $\Gamma^{H}$ is obtained by resampling.
We calculate a \pv\ for $\Gamma^{alt}$ \& $\Gamma^{null}$ by calculating the proportion of samples in the null distribution $\Gamma^{null}$ that exceed the scores in the alternative distribution $\Gamma^{alt}$.

\subsection{Confidence Interval Estimation}
\label{confidence_intervals}

We utilize the $\Gamma^{alt}$ bootstrap distribution to obtain $95\%$ confidence intervals around the mean by using the reverse percentile method (Algorithm \ref{alg:ci}).

\begin{algorithm}[t]
\SetKwInput{KwInput}{Input}
\caption{Calculate Confidence Interval (CI)}
\label{alg:ci}
\KwInput{$\Gamma^{alt}$}
$\hat{\Gamma} \leftarrow mean(\Gamma^{alt})$\;
$\Gamma^{alt}_{sorted} \leftarrow \text{sort}(\Gamma^{alt})$\;
\tcp{Choose 2.5th and 97.5th percentile (95\% CI)}
$\text{CI}_{lower} \leftarrow 2\hat{\Gamma} - \Gamma^{alt}_{sorted}[975]$\;
$\text{CI}_{upper} \leftarrow 2\hat{\Gamma} - \Gamma^{alt}_{sorted}[25]$\;
$\text{CI} \leftarrow [\text{CI}_{lower}, \text{CI}_{upper}]$\;
\end{algorithm}

\subsection{Metrics}

We choose a set of metrics that range from simple plurality agreement to a more nuanced comparison of full response distributions and head-to-head performance. Collectively, these metrics provide a comprehensive view of how well models $A$ and $B$ align with a gold standard $G$ when dealing with nominal data. We use the following metrics in our experiments:

\begin{itemize}
    \item \textbf{Accuracy}. Accuracy is the most commonly used metric to compare models against each other. First, take the plurality vote for all items in $A$, $B$, and $G$. Then compute the accuracy for $A$ and $B$ by comparing against $G$.
    
    Accuracy for $A$ against $G$:
    $$\Gamma_{Accuracy}(A, G) = \frac{1}{N} \sum_{i=1}^{N} \mathbb{I}(PV_A(i) = PV_G(i))$$
    where, $PV_X(i)$ is the plurality vote for item $i$ in set $X$ and $\mathbb{I}(\cdot)$ is the indicator function, which is 1 if the condition is true and 0 otherwise.
    
    \item \textbf{Total variation (TV)}. TV is related to Manhattan or L1 distance. It goes beyond the plurality vote and helps compare probability distributions for soft label evaluation. Compute the frequency of responses for all items in $A$, $B$, and $G$, normalize, and compute the mean Manhattan distance across all items in $A$ and $B$ against $G$.

    TV for $A$ against $G$:
    $$\Gamma_{TV}(A, G) = \frac{1}{N} \sum_{i=1}^{N} \sum_{m \in \mathcal{M}} \big|P_A(m | i) - P_G(m | i)\big|$$
    where, $P_X(m | i)$ be the normalized frequency (probability distribution) of response $m$ for item $i$ in set $X$. This means $\sum_{m} P_X(m | i) = 1$. $\mathcal{M}$ is the set of all possible responses.
    
    \item \textbf{Wins}. Wins is a meta-metric used for item-level comparison. We use TV as the base metric for Wins, but any other metric can be used. Calculate TV for all items in $A$ and $B$ against $G$, then count the wins of $A$ and $B$, i.e., the number of times $A$ has less TV than $B$ and vice-versa.
    
    Wins for $A$ over $B$:
    $$\Gamma_{Wins}(A > B) = \frac{1}{N} \sum_{i=1}^{N} \mathbb{I}(\text{TV}_A(i) < \text{TV}_B(i))$$
    where, $\text{TV}_A(i) = \sum_{m \in \mathcal{M}} |P_A(m | i) - P_G(m | i)|$
    and $\text{TV}_B(i) = \sum_{m \in \mathcal{M}} |P_B(m | i) - P_G(m | i)|$.
    
    \item \textbf{KL-Divergence (KL-Div)}. KL-Divergence is another frequently used metric for comparing probability distributions. Calculate the frequency of responses for all items in $A$, $B$, and $G$. Then, compute the mean KL-divergence across all items in $A$ and $B$ against $G$.
    
    KL-Divergence for $A$ against $G$:
    $$\Gamma_{KL}(A, G) = \frac{1}{N} \sum_{i=1}^{N} \sum_{m \in \mathcal{M}} P_G(m | i) \log\left(\frac{P_G(m | i)}{P_A(m | i)}\right)$$
    where, $P_X(m | i)$ be the normalized frequency (probability distribution) of response $m$ for item $i$ in set $X$.
    
\end{itemize}

\subsection{Fitting to Real-World Datasets}

We fit the prior parameters $\mathbf{\alpha}$ of our model to real-world datasets by computing the maximum a posteriori (MAP) estimate of the model. We use mean absolute bias (MAB) to measure goodness of fit:

$$\text{MAB} = \frac{1}{M} \sum_{m=1}^{M} \big|\theta_m - E[\hat{\theta}_m]\big|,$$

\noindent
where, $\theta_m$ is the percentage of category $m$ in the dataset and $\hat{\theta_m}=\alpha_m/\sum_m\alpha_m$ is the expected rate of category $m$. It indicates how much, on average, the predicted parameters deviate from the actual values. More details about the MAP estimation can be found in the extended version.

\section{Experiments}

\subsection{Datasets}

We use the following datasets, each comprising various categories and multiple responses per item, for our experiments.

\paragraph{Toxicity} Toxicity dataset \cite{kumar2021designing} consists of 107,620 social media comments labeled by 17,280 raters. (Number of categories $M=2$, $\alpha = [1.37,1.33]$)

\paragraph{DICES} Diversity in Conversational AI Evaluation for Safety 350 dataset \cite{NEURIPS2023_a74b697b} consists of 350 chatbot conversations rated for safety by 123 raters across 16 safety dimensions. ($M=3$, $\alpha = [5.22, 0.86, 2.75]$)

\paragraph{D3code} D3code \cite{davani_d3code_2024} is a large cross-cultural dataset comprising 4554 items, each labeled for offensiveness by 4309 raters from 21 countries and balanced across gender and age. ($M=2$, $\alpha = [6.08, 2.88]$)

\paragraph{Jobs} Jobs dataset \cite{liu-etal-2016-understanding} is a collection of 2000 job-related tweets labeled by 5 raters each. The raters answer 3 questions about each tweet, and the corresponding sets are denoted by JobsQ1/2/3. The categories in JobsQ1/2/3 represent the point of view of job-related information, employment status, and job transition events, respectively. We use \textbf{JobsQ1} and \textbf{JobsQ3} for our experiments. Here $M=5$, $\alpha =[1039.76, 38.24, 35.57, 310.29, 46.02]$  and  $M=12$,  $\alpha =[133.79, 834.51, 105.27, 3669.04, 206.80, 293.44, 585.58$, $1278.56, 1874.82,1838.49, 1576.10, 989.23]$, respectively.

\subsection{Experimental Setup}

We run experiments for hypothesis testing with different number of annotations ($N\times K$ = \{100, 250, 500, 1000, 2500, 5000, 10000, 25000, 50000\}) while ranging $K$ from 1 to 500 (in increments of 1 till 10, then 20, then in increments of 20 from 20 onwards)
for different metrics, and $\epsilon$ = \{0.1, 0.2, 0.3, 0.4\}. We use four metrics with four $\epsilon$, yielding 16 sets of 282 experiments for each dataset. 

\begin{table}[ht!]
\centering
\begin{tabular}{l|c|c|c|c|c}
 & Toxicity & DICES & D3code & JobsQ1 & JobsQ3 \\
\midrule
MAB & 0.0111 & 0.0231 & 0.0029 & 0.0537 & 0.0869 \\
\end{tabular}
\caption{Mean absolute bias for the parameters.}
\label{tab:dataset_bias}
\end{table}

For the real-world datasets, the value of parameter $\alpha$ is determined using the MAP estimate and $\rho = [1/M] \times M$ where $M$ is fixed for each dataset. To estimate the \pv s\, we repeat the sampling process 1000 times.
Our estimates of the MAB fit are shown in Table \ref{tab:dataset_bias}.

We also experiment using different prior distributions for parameter $\alpha$, \textbf{balanced} ($\alpha=[3] \times M$) and \textbf{unbalanced} ($\alpha=[10]+[3] \times M$) to simulate class imbalance, and varying the number of categories ($M$ = \{2, 3, 4, 5, 12\}).


\begin{table}[ht!]
\centering
\small
\begin{tabular}{l|c|cccc}
 &  & Accuracy & TV & Wins & KL-Div \\
\midrule
 & NK & 2500 & 1000 & 2500 & 1000 \\
Toxicity & \pv\ & 0.012 & 0.015 & 0.012 & 0.022 \\
(M=2) & K & 1 & 120 & 1 & 200 \\
 & $\Delta$ & 0.040 & 0.074 & 0.040 & 0.044 \\
\hline
& NK & 1000 & 500 & 1000 & 1000 \\
DICES & \pv\ & 0.036 & 0.017 & 0.028 & 0.020 \\
(M=3) & K & 1 & 80 & 20 & 300 \\
 & $\Delta$ & 0.055 & 0.063 & 0.346 & 0.082 \\
\hline
& NK & 2500 & 1000 & 2500 & 1000 \\
D3code & \pv\ & 0.037 & 0.020 & 0.024 & 0.022 \\
(M=2) & K & 2 & 140 & 60 & 100 \\
 & $\Delta$ & 0.034 & 0.072 & 0.413 & 0.036 \\
\hline
& NK & 250 & 250 & 250 & 250 \\
JobsQ1 & \pv\ & 0.035 & 0.015 & 0.036 & 0.035 \\
(M=5) & K & 1 & 40 & 1 & 1 \\
 & $\Delta$ & 0.104 & 0.050 & 0.104 & 2.864 \\
\hline
& NK & 500 & 250 & 500 & 500 \\
JobsQ3 & \pv\ & 0.047 & 0.014 & 0.038 & 0.030 \\
(M=12) & K & 100 & 240 & 80 & 500 \\
 & $\Delta$ & 0.595 & 0.024 & 0.868 & 0.182 \\
\midrule
\midrule
 & NK & 1000 & 500 & 1000 & 1000 \\
Unbalanced & \pv\ & 0.031 & 0.044 & 0.031 & 0.014 \\
(M=2) & K & 1 & 80 & 1 & 140 \\
 & $\Delta$ & 0.050 & 0.074 & 0.050 & 0.047 \\
\hline
& NK & 1000 & 500 & 1000 & 500 \\
Unbalanced & \pv\ & 0.039 & 0.023 & 0.040 & 0.031 \\
(M=3) & K & 2 & 100 & 40 & 100 \\
 & $\Delta$ & 0.061 & 0.061 & 0.473 & 0.068 \\
\hline
& NK & 1000 & 500 & 1000 & 500 \\
Unbalanced & \pv\ & 0.049 & 0.013 & 0.022 & 0.021 \\
(M=4) & K & 4 & 120 & 40 & 240 \\
 & $\Delta$ & 0.089 & 0.054 & 0.520 & 0.084 \\
\hline
& NK & 1000 & 500 & 1000 & 500 \\
Unbalanced & \pv\ & 0.045 & 0.009 & 0.015 & 0.010 \\
(M=5) & K & 10 & 100 & 40 & 240 \\
 & $\Delta$ & 0.138 & 0.043 & 0.545 & 0.098 \\
\hline
& NK & 1000 & 250 & 500 & 500 \\
Unbalanced & \pv\ & 0.027 & 0.014 & 0.042 & 0.004 \\
(M=12) & K & 80 & 240 & 60 & 460 \\
 & $\Delta$ & 0.436 & 0.023 & 0.763 & 0.156 \\
\end{tabular}
\caption{Minimum \pv, $K$, and corresponding effect size ($\Delta$) for lowest $NK$ with $p<0.05$ ($ \epsilon=0.3$).}
\label{tab:low_k_for_p_lte_05_nk_min_p}
\end{table}

\begin{table}[ht!]
\centering
\small
\begin{tabular}{l|c|cccc}
 &  & Accuracy & TV & Wins & KL-Div \\
\midrule
 & NK & 2500 & 1000 & 2500 & 1000 \\
Toxicity & ci-width & 0.050 & 0.067 & 0.050 & 0.085 \\
(M=2) & K & 1 & 5 & 1 & 100 \\
 & $\Delta$ & 0.117 & 0.063 & 0.728 & 0.042 \\
\hline
 & NK & 1000 & 500 & 1000 & 1000 \\
DICES & ci-width & 0.090 & 0.063 & 0.090 & 0.230 \\
(M=3) & K & 1 & 7 & 1 & 100 \\
 & $\Delta$ & 0.080 & 0.075 & 0.203 & 0.082 \\
\hline
& NK & 2500 & 1000 & 2500 & 1000 \\
D3code & ci-width & 0.054 & 0.067 & 0.054 & 0.068 \\
(M=2) & K & 1 & 7 & 1 & 100 \\
 & $\Delta$ & 0.116 & 0.066 & 0.669 & 0.036 \\
\hline
& NK & 250 & 250 & 250 & 250 \\
JobsQ1 & ci-width & 0.160 & 0.055 & 0.160 & 1.516 \\
(M=5) & K & 1 & 8 & 1 & 80 \\
 & $\Delta$ & 0.104 & 0.038 & 0.104 & 0.109 \\
\hline
& NK & 500 & 250 & 500 & 500 \\
JobsQ3 & ci-width & 0.086 & 0.020 & 0.086 & 1.322 \\
(M=12) & K & 1 & 1 & 1 & 100 \\
 & $\Delta$ & 0.614 & 0.003 & 0.941 & 1.613 \\
\midrule
\midrule
 & NK & 1000 & 500 & 1000 & 1000 \\
Unbalanced & ci-width & 0.070 & 0.093 & 0.079 & 0.083 \\
(M=2) & K & 2 & 7 & 1 & 80 \\
 & $\Delta$ & 0.068 & 0.086 & 0.154 & 0.045 \\
\hline
& NK & 1000 & 500 & 1000 & 500 \\
Unbalanced & ci-width & 0.079 & 0.063 & 0.079 & 0.145 \\
(M=3) & K & 1 & 10 & 1 & 100 \\
 & $\Delta$ & 0.111 & 0.028 & 0.180 & 0.028 \\
\hline
& NK & 1000 & 500 & 1000 & 500 \\
Unbalanced & ci-width & 0.081 & 0.048 & 0.081 & 0.218 \\
(M=4) & K & 1 & 6 & 1 & 100 \\
 & $\Delta$ & 0.125 & 0.058 & 0.191 & 0.243 \\
\hline
& NK & 1000 & 500 & 1000 & 500 \\
Unbalanced & ci-width & 0.070 & 0.039 & 0.070 & 0.467 \\
(M=5) & K & 1 & 7 & 1 & 100 \\
 & $\Delta$ & 0.128 & 0.049 & 0.187 & 0.515 \\
\hline
& NK & 1000 & 250 & 500 & 500 \\
Unbalanced & ci-width & 0.056 & 0.019 & 0.080 & 1.292 \\
(M=12) & K & 1 & 1 & 1 & 100 \\
 & $\Delta$ & 0.105 & 0.002 & 0.920 & 1.264 \\
\end{tabular}
\caption{Lowest CI-width with corresponding value of $K$ and effect size ($\Delta$) for lowest $NK$ with $p<0.05$ ($ \epsilon=0.3$).}
\label{tab:k_low_ci_nk}
\end{table}

\subsection{Results}

Tables \ref{tab:low_k_for_p_lte_05_nk_min_p} shows the results for minimum \pv, $K$, and corresponding effect size ($\Delta$) for lowest $NK$ with $p<0.05$ ($ \epsilon=0.3$). Table \ref{tab:k_low_ci_nk} shows the results for the lowest CI-width with the corresponding value of $K$ and effect size - $\Delta$ for the lowest $NK$ observed in Table \ref{tab:low_k_for_p_lte_05_nk_min_p}. Our results suggest that whether or not a tradeoff exists, and where it is, depends much more on the metric used than the data source, and that the metrics behave very differently. They show that the TV metric requires the smallest number of $N \times K$ overall, and that this comes with a small number of $K > 10$.

\begin{figure*}[ht!]
  \centering
  \begin{subfigure}[b]{0.24\linewidth}
    \centering
    \includegraphics[width=\linewidth]{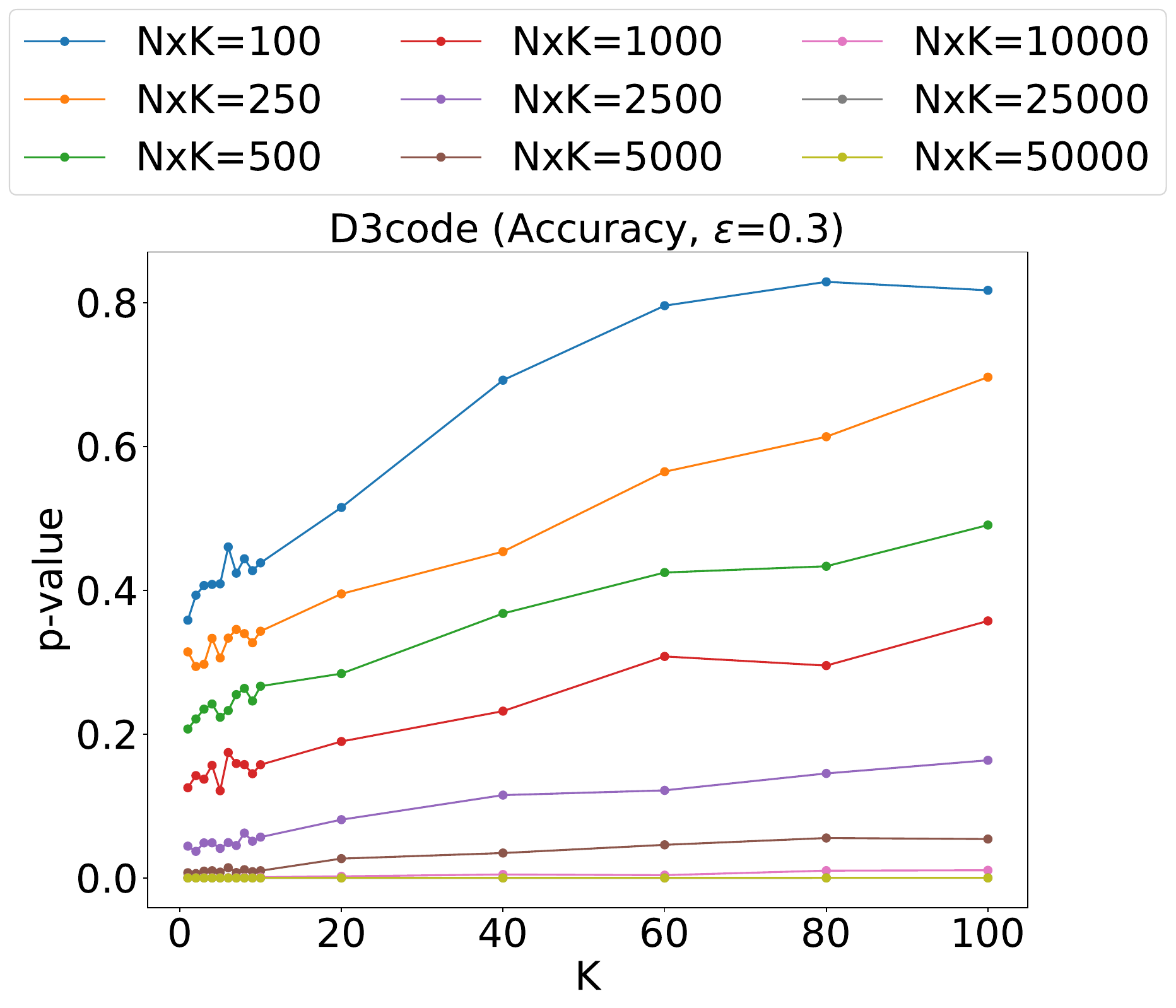}
    \caption{Accuracy}
    \label{fig:d3code_acc_e03_main}
  \end{subfigure} \hfill
  \begin{subfigure}[b]{0.24\linewidth}
    \centering
    \includegraphics[width=\linewidth]{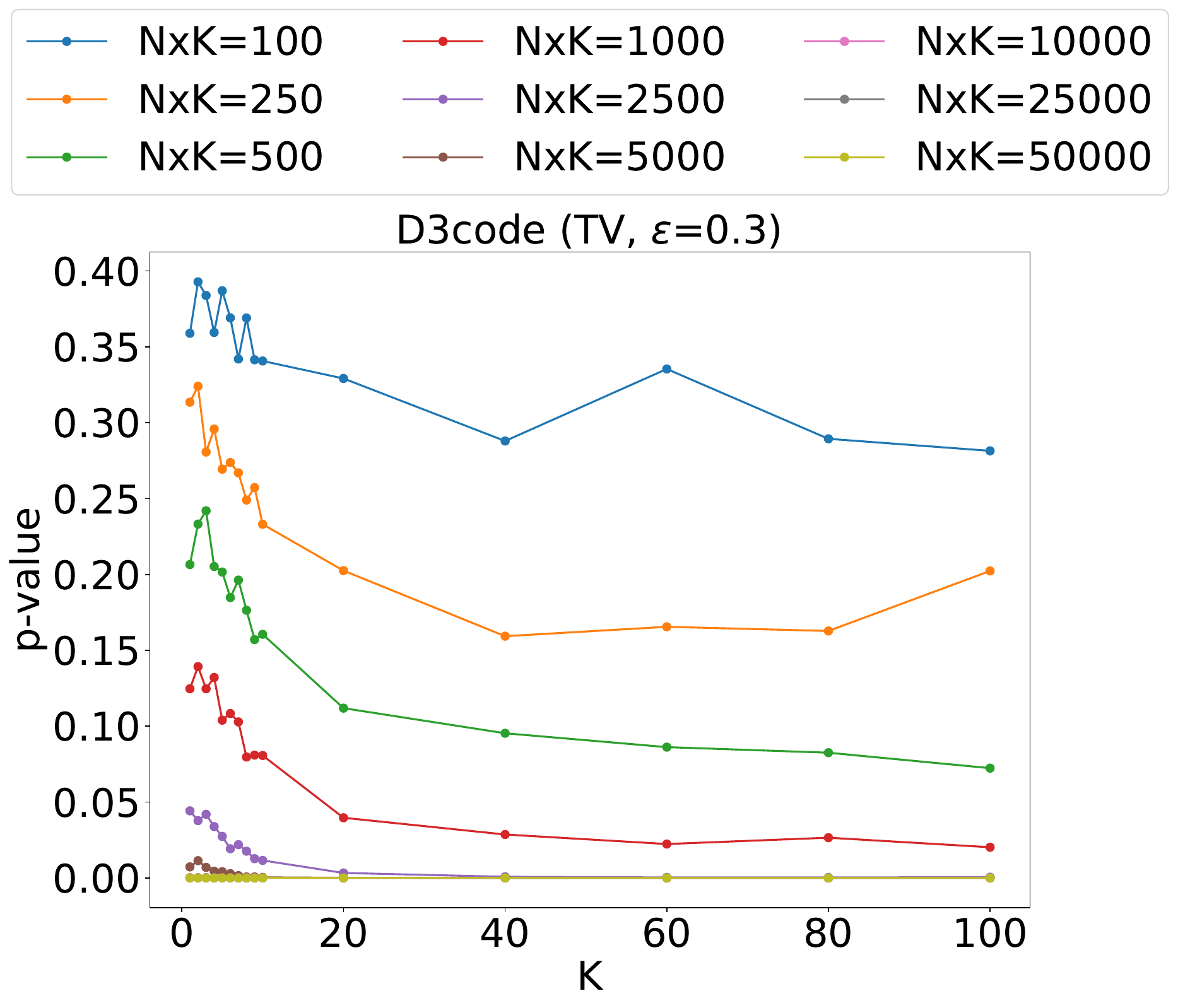}
    \caption{TV}
    \label{fig:d3code_tv_e03_main}
  \end{subfigure} \hfill
  \begin{subfigure}[b]{0.24\linewidth}
    \centering
    \includegraphics[width=\linewidth]{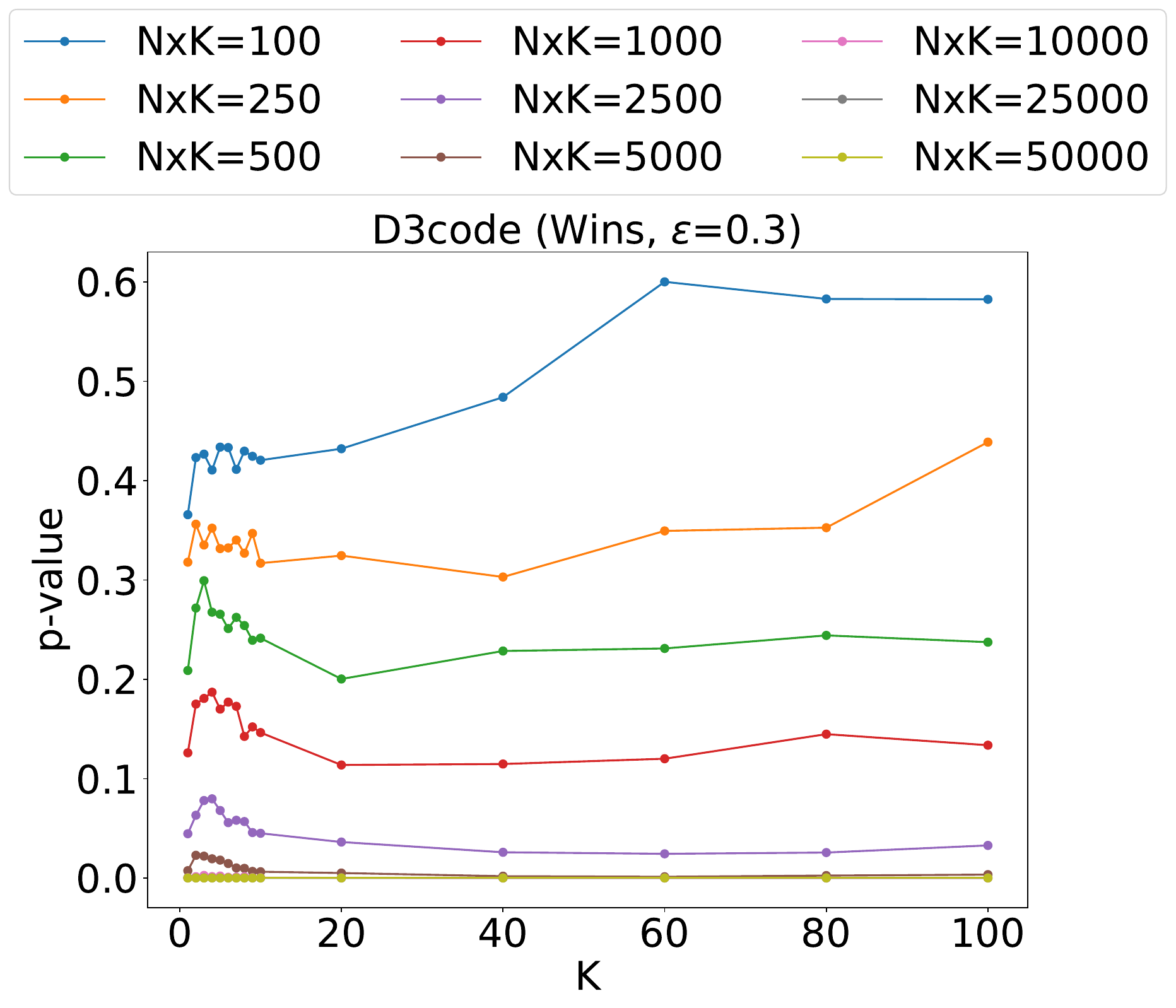}
    \caption{Wins}
    \label{fig:d3code_wins_e03_main}
  \end{subfigure} \hfill
  \begin{subfigure}[b]{0.24\linewidth}
    \centering
    \includegraphics[width=\linewidth]{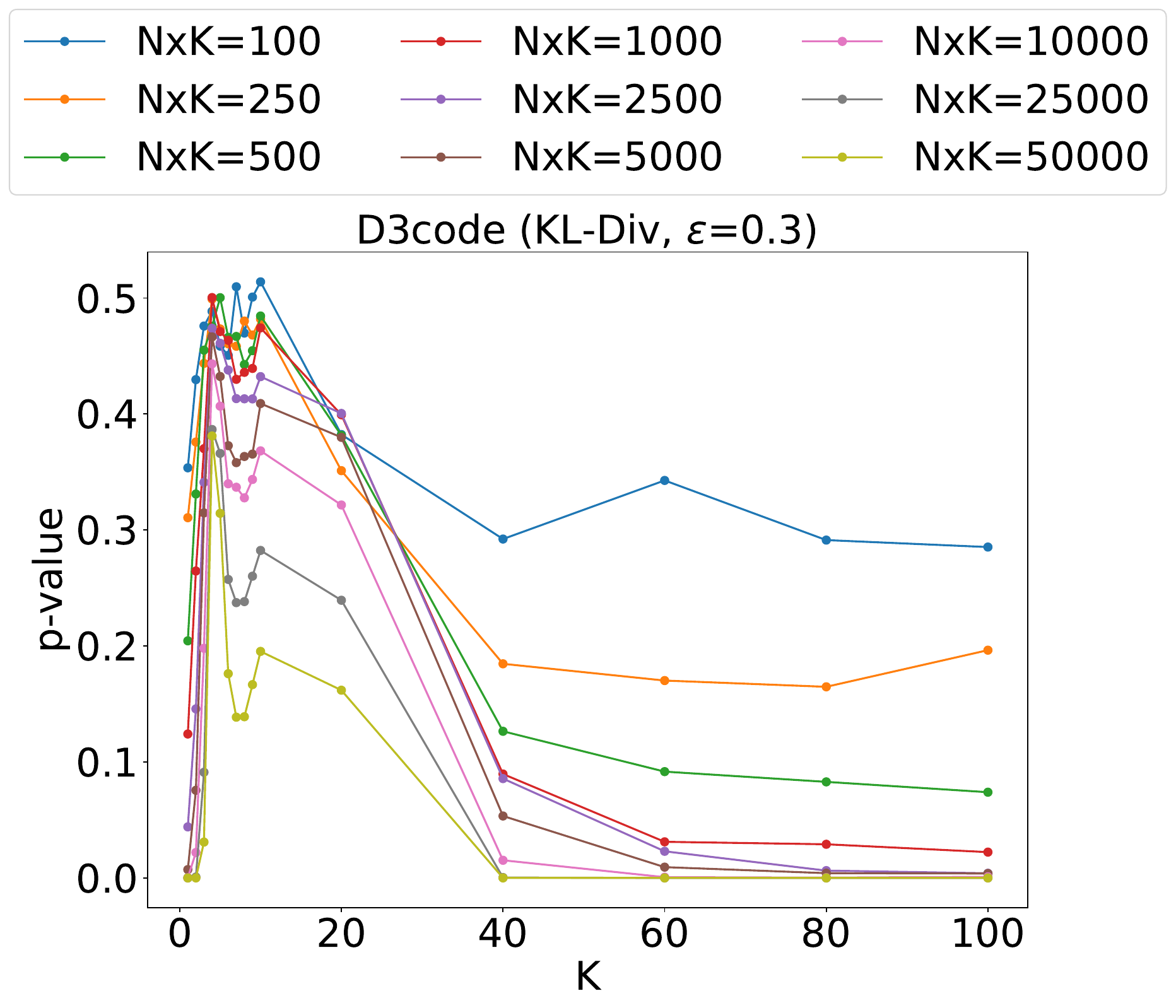}
    \caption{KL-Div}
    \label{fig:d3code_kl_e03_main}
  \end{subfigure}  
  \caption{\pv\ plots for D3code dataset, $\epsilon=0.3$.}
  \label{fig:d3code_p_vals_e03_main}
\end{figure*}

\begin{figure*}[ht!]
  \centering
  \begin{subfigure}[b]{0.24\linewidth}
    \centering
    \includegraphics[width=\linewidth]{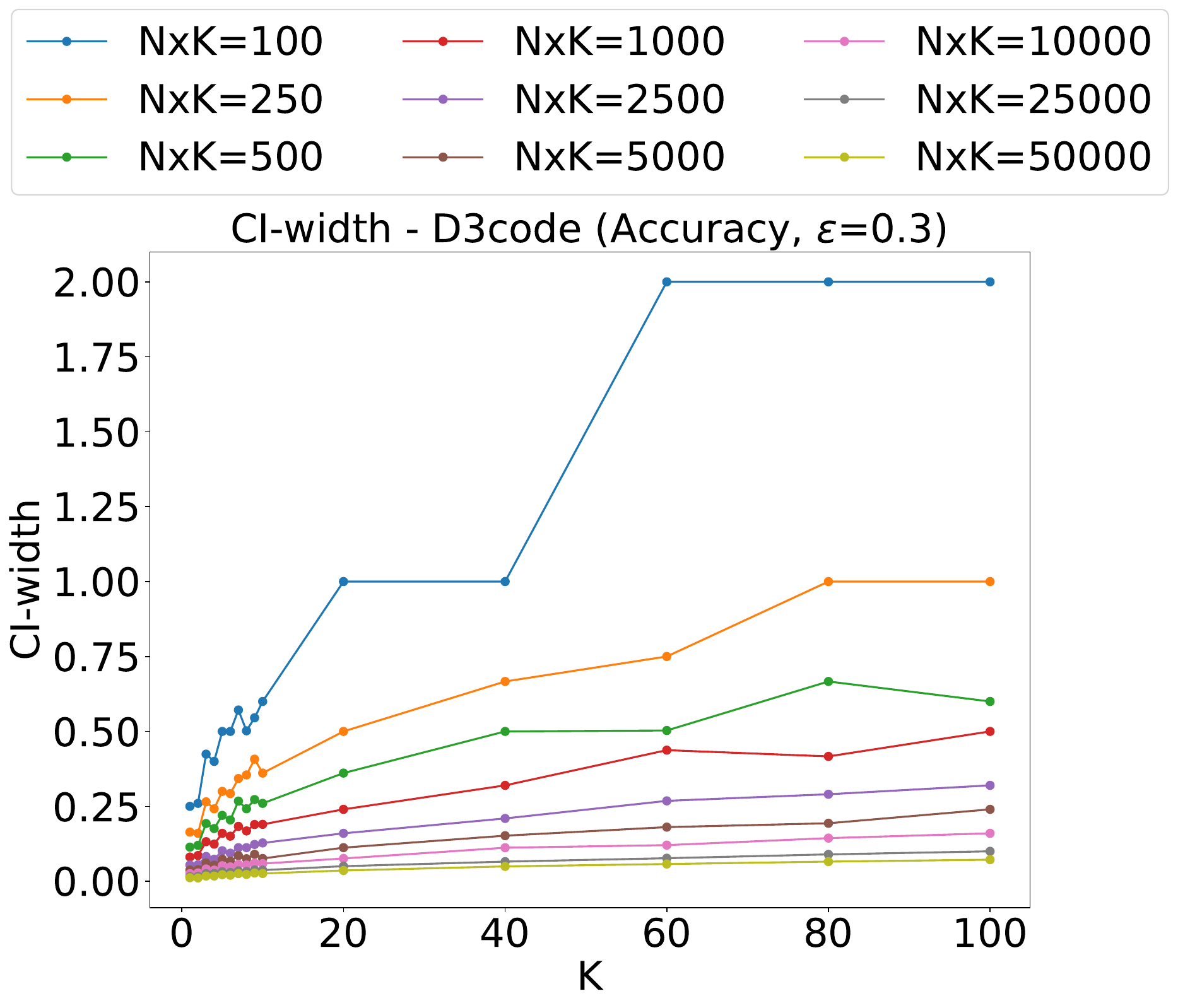}
    \caption{Accuracy}
    \label{fig:d3code_ci_acc_e03_main}
  \end{subfigure} \hfill
  \begin{subfigure}[b]{0.24\linewidth}
    \centering
    \includegraphics[width=\linewidth]{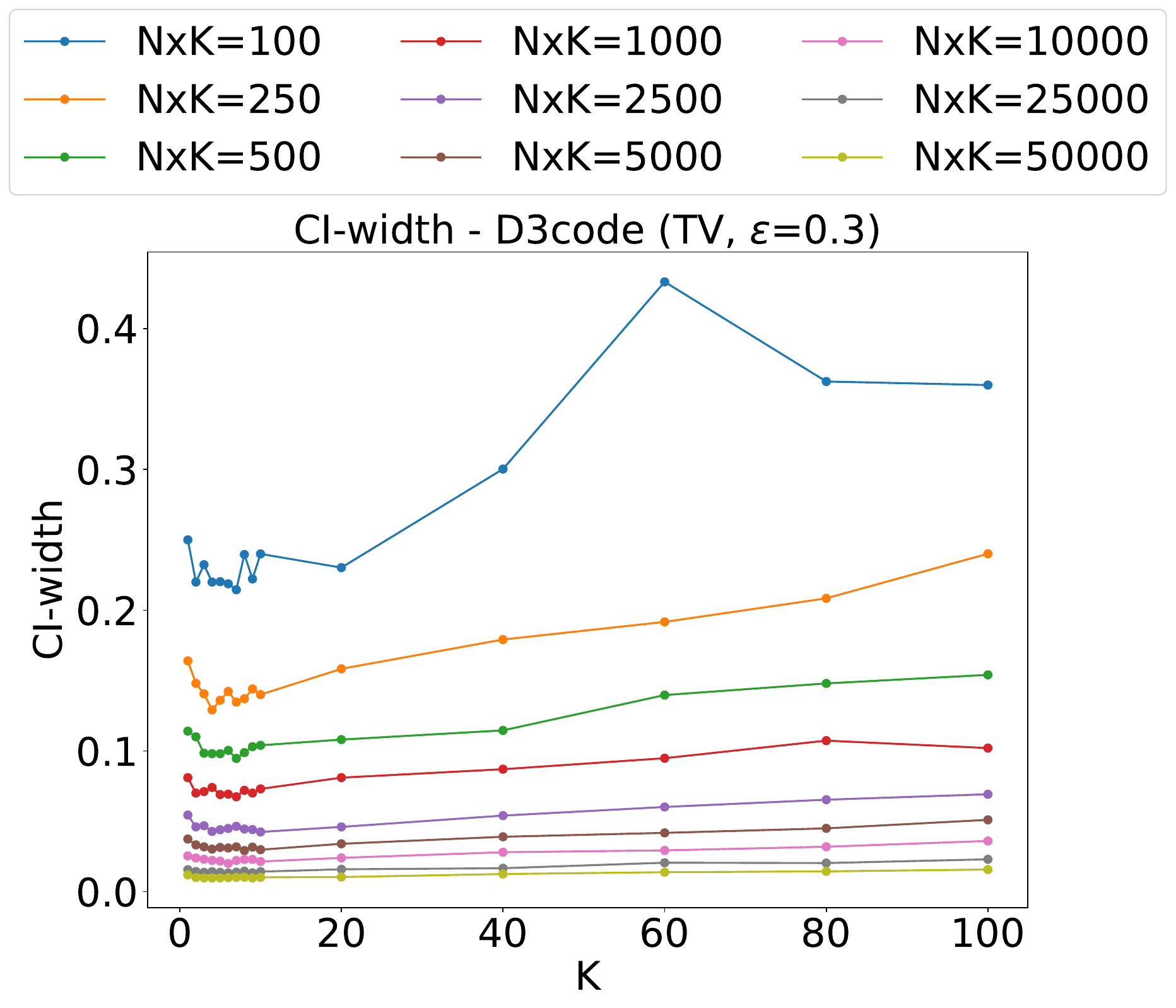}
    \caption{TV}
    \label{fig:d3code_ci_tv_e03_main}
  \end{subfigure} \hfill
  \begin{subfigure}[b]{0.24\linewidth}
    \centering
    \includegraphics[width=\linewidth]{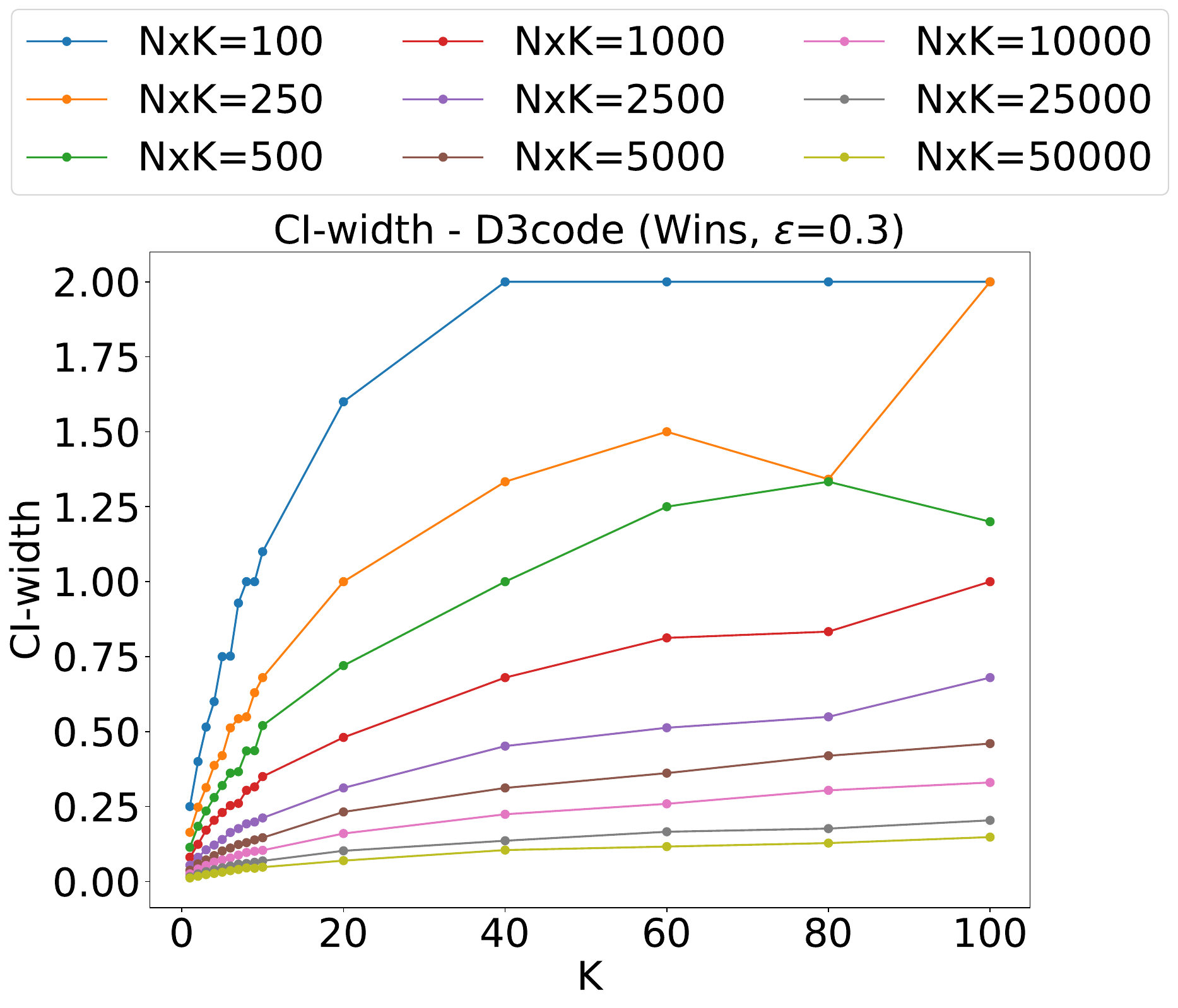}
    \caption{Wins}
    \label{fig:d3code_ci_wins_e03_main}
  \end{subfigure} \hfill
  \begin{subfigure}[b]{0.24\linewidth}
    \centering
    \includegraphics[width=\linewidth]{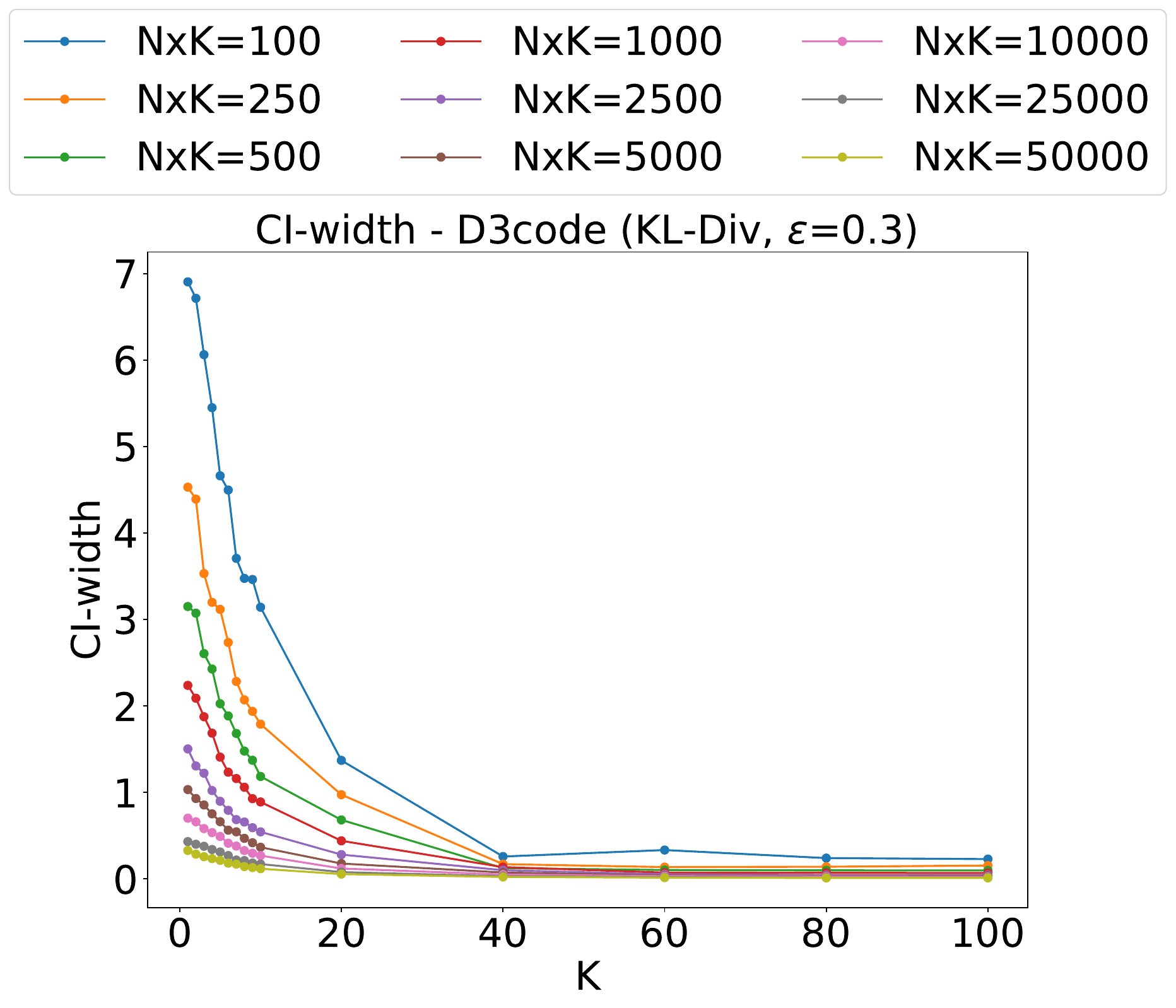}
    \caption{KL-Div}
    \label{fig:d3code_ci_kl_e03_main}
  \end{subfigure}
  \caption{CI-width plots for D3code dataset, $\epsilon=0.3$.}
  \label{fig:d3code_ci_e03_main}
\end{figure*}

\begin{figure*}[ht!]
  \centering
  \begin{subfigure}[b]{0.24\linewidth}
    \centering
    \includegraphics[width=\linewidth]{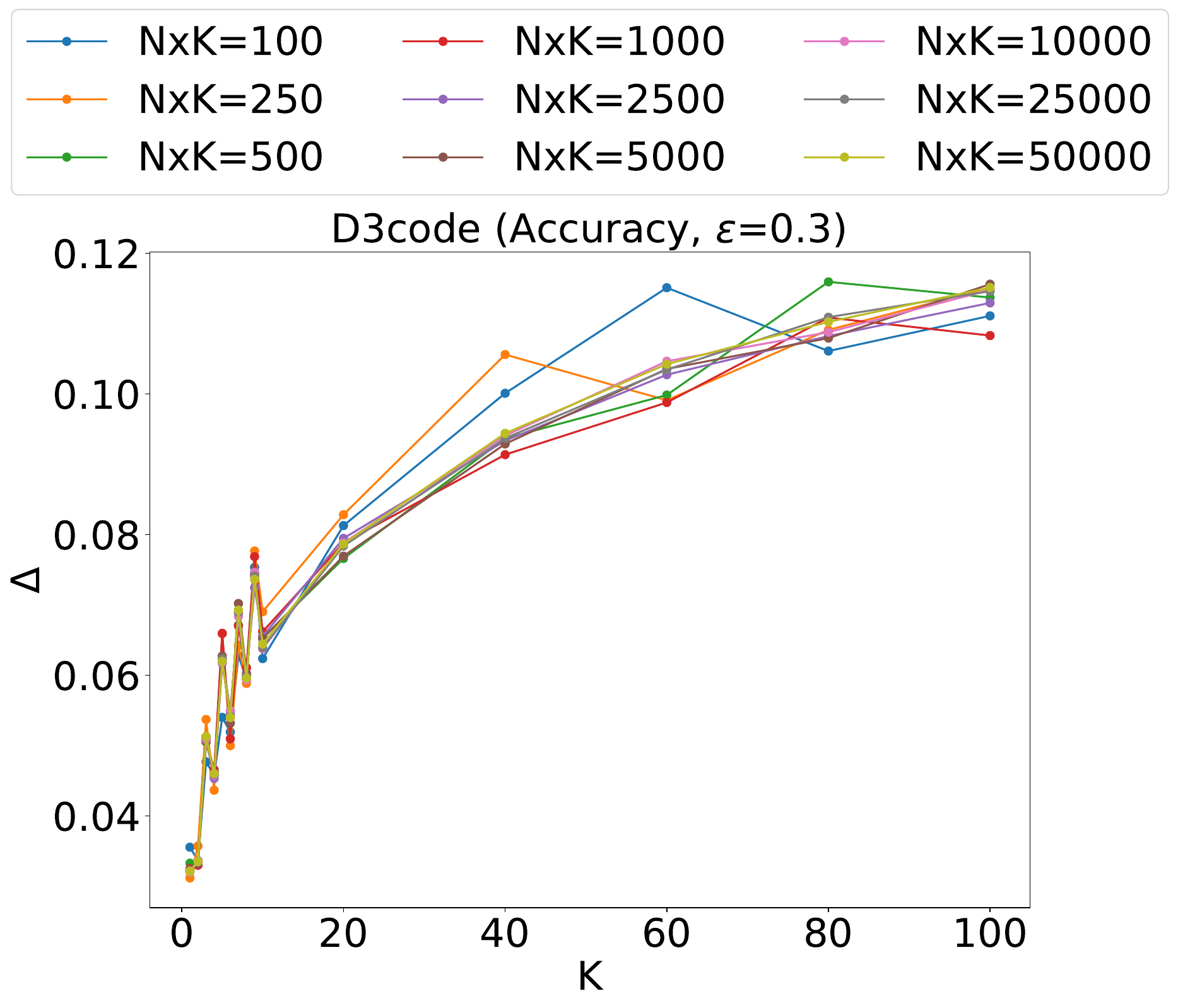}
    \caption{Accuracy}
    \label{fig:d3code_delta_acc_e03_main}
  \end{subfigure} \hfill
  \begin{subfigure}[b]{0.24\linewidth}
    \centering
    \includegraphics[width=\linewidth]{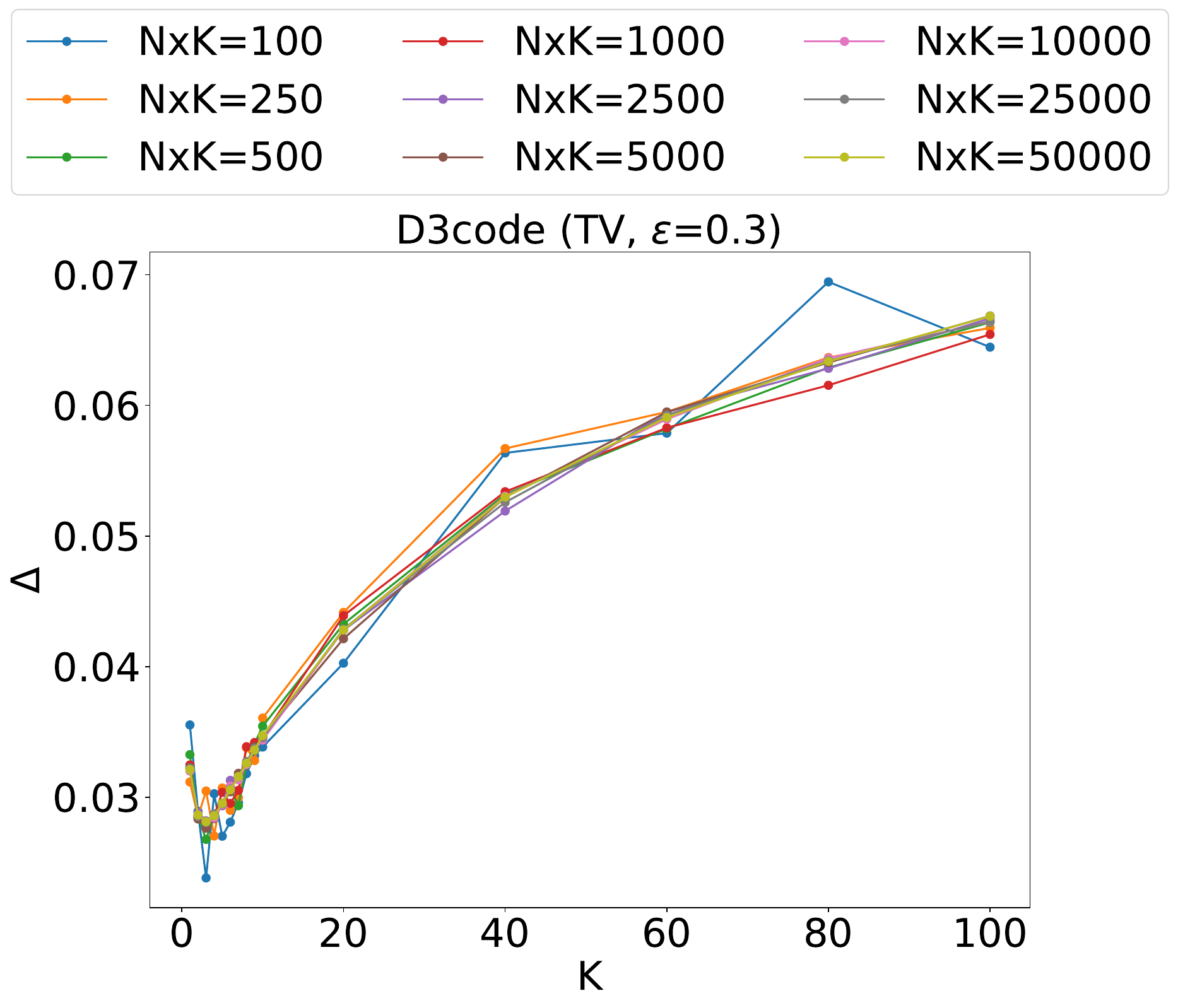}
    \caption{TV}
    \label{fig:d3code_delta_tv_e03_main}
  \end{subfigure} \hfill
  \begin{subfigure}[b]{0.24\linewidth}
    \centering
    \includegraphics[width=\linewidth]{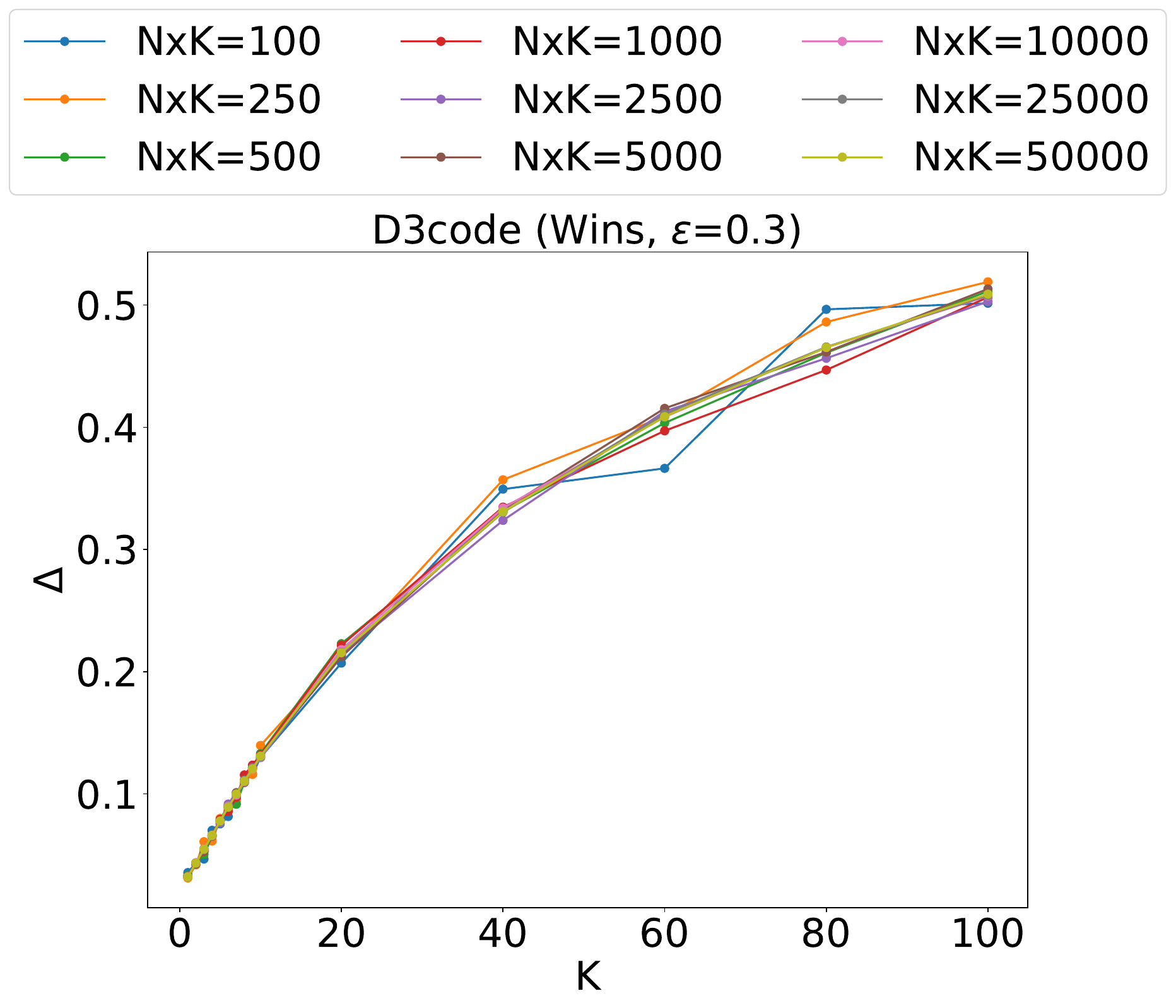}
    \caption{Wins}
    \label{fig:d3code_delta_wins_e03_main}
  \end{subfigure} \hfill
  \begin{subfigure}[b]{0.24\linewidth}
    \centering
    \includegraphics[width=\linewidth]{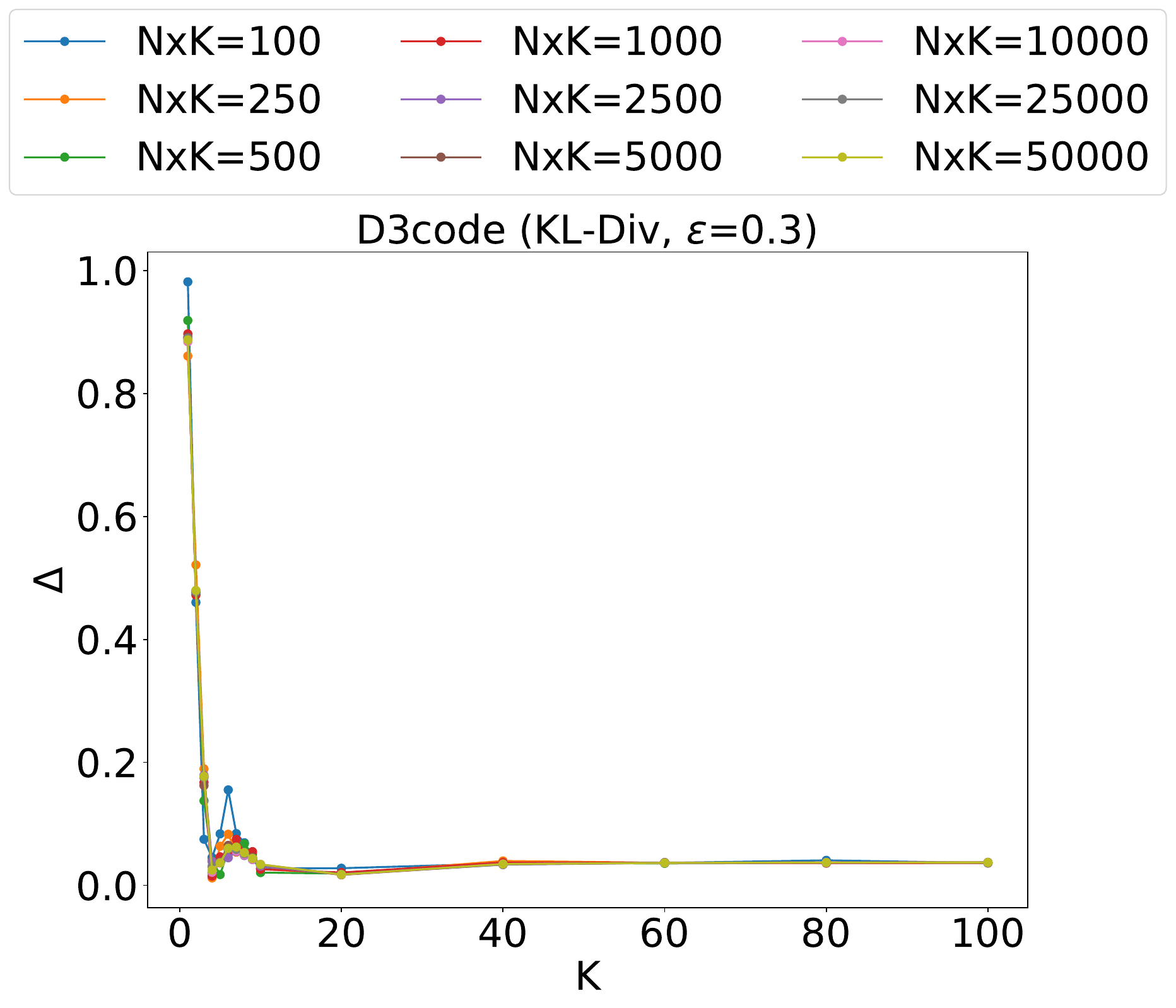}
    \caption{KL-Div}
    \label{fig:d3code_delta_kl_e03_main}
  \end{subfigure}
  \caption{Effect sizes ($\Delta$) for D3code dataset, $\epsilon=0.3$.}
  \label{fig:d3code_delta_e03_main}
\end{figure*}

Figures \ref{fig:d3code_p_vals_e03_main}--\ref{fig:d3code_delta_e03_main}
show results for \pv s, confidence intervals, and effect sizes
for $\epsilon = 0.3$ on the D3code dataset. Although there were exceptions, they exemplify many common observations found for other datasets (refer to the extended version).


\paragraph{Accuracy}
 We notice that the \pv\ increases as $K$ increases for all $N \times K$ in all datasets. The increase is sharp until $K=40$ for all $N \times K$ values, especially for lower values of $\epsilon$ (0.1, 0.2). Then, the \pv s\ start plateauing for higher $N \times K$ values, but continue to increase for lower $N \times K$ values. As $\epsilon$ increases, \pv s\ decrease as expected. Having more responses per item ($K$) does not seem to be helpful for accuracy. CI-width and effect sizes increase with increasing $K$ except for JobsQ1.

With some exceptions (particularly JobsQ1), \textbf{accuracy} yields the most reliable results (lowest scores) for both $p$-values and confidence intervals when $K = 1$ or very small $k$. For \textbf{total variation}, higher $K$ usually yields lower, more reliable $p$-values, with improvements occurring even for $K = 300$, but confidence intervals generally increase with $K$, and tend to have best results for $K \in [1, 10]$. \textbf{Wins} tends to have minimum \pv s\ for $K \in [25, 100]$), where confidence intervals improve with increasing $K$, with improvements occurring even for $K = 300$. \textbf{KL-divergence} is particularly interesting. While confidence intervals improve with increasing $K$, settling down at around $K=100$, $p$-values improve for $K \in [2,5]$ then get worse and then settle to lower numbers by around $k = 300$.

\paragraph{TV}
 For TV, \pv\ increases sharply with increasing $K$ till $K=10$, and then starts to decrease thereafter for all $N \times K$ for $\epsilon=0.1$. For the remaining values of $\epsilon$, the \pv s\ generally decrease with increasing $K$. We also notice an elbow plot emerging for total variation, suggesting an optimal value of $K$ for the datasets. As $\epsilon$ increases, \pv s\ decrease as expected. Having more responses per item ($K$) seems to be helpful for total variation and results in lower \pv s.\ CI-width and effect sizes increase as $K$ increases.

For simulations with balanced categories, \pv s\ monotonically decrease with increasing $K$ as $M$ gets larger. CI-width and effect sizes increase as $K$ increases. For simulations with unbalanced categories, \pv s\ decrease with increasing $K$ as $M$ gets larger, whereas CI-width and effect sizes increase as $K$ increases.

\paragraph{Wins}
 For Wins, \pv\ increases sharply with increasing $K$ till $K=10$, and then starts to plateau for all $N \times K$ for $\epsilon=0.1$. For the remaining values of $\epsilon$, \pv s\ start going back up as $K$ gets higher. As $\epsilon$ increases, \pv s\ decrease as expected. There seems to be an optimal $K$ for Dices, D3code, and JobsQ3 dataset. CI-width and effect sizes increase with increasing $K$.

For simulations with balanced categories, \pv s\ generally decrease with increasing $K$ and start to go back up for lower $N \times K$. CI-width and effect sizes increase as $K$ increases, with some exceptions. For simulations with unbalanced categories, \pv s\, CI-width, and effect sizes show similar trends as balanced categories.

\paragraph{KL-Divergence}
 We notice that \pv s\ exhibit double peaks till about $K=40$ for all datasets and for all $N \times K$. The \pv s\ continue to decrease for higher values of $K$. As $\epsilon$ increases, \pv s\ decrease as expected. CI-width and effect sizes generally decrease as $K$ increases, except for JobsQ3 with one initial peak.

For simulations with balanced categories, \pv s\ exhibit a single or double peak initially but settle down with higher $K$. CI-width and effect sizes decrease with increasing $K$; however start to have one peak as $M$ increases. For simulations with unbalanced categories, \pv s\, CI-width, and effect sizes show similar behavior to balanced categories.

\section{Discussion}

\begin{figure*}[ht!]
    \centering
    \includegraphics[width=\linewidth]{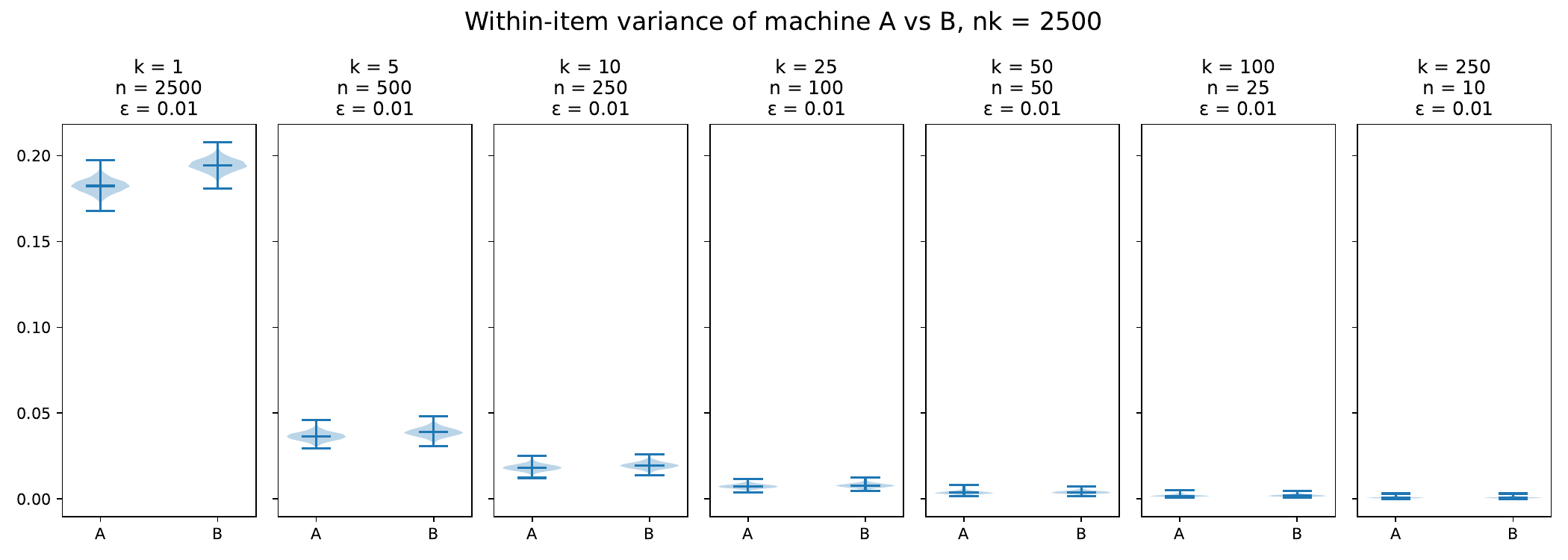}
    \caption{Distribution of scores for machines $A$ and $B$ for within-item variance estimates, for $N \times K = 2500$.}
    \label{fig:within-variances}
\end{figure*}

\begin{figure*}[ht!]
    \centering
    \begin{subfigure}[b]{\linewidth}
    \centering
    \includegraphics[width=0.825\linewidth]{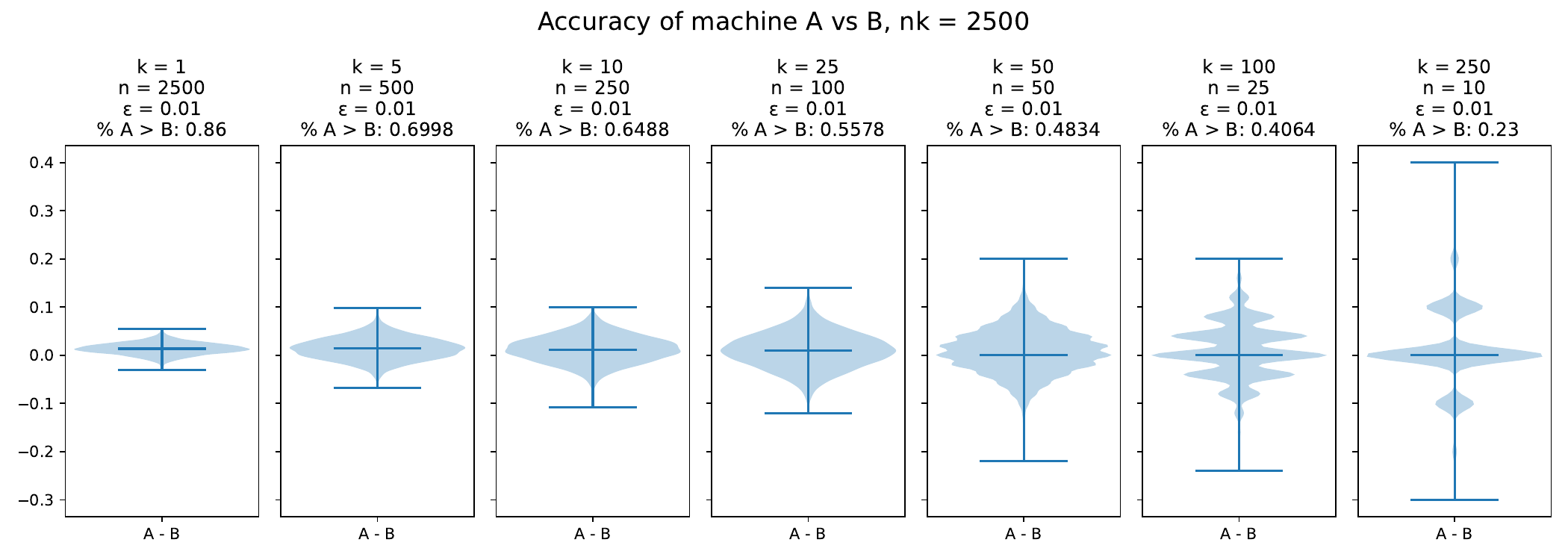}
    \end{subfigure} \hfill
    \begin{subfigure}[b]{\linewidth}
    \centering
    \includegraphics[width=0.825\linewidth]{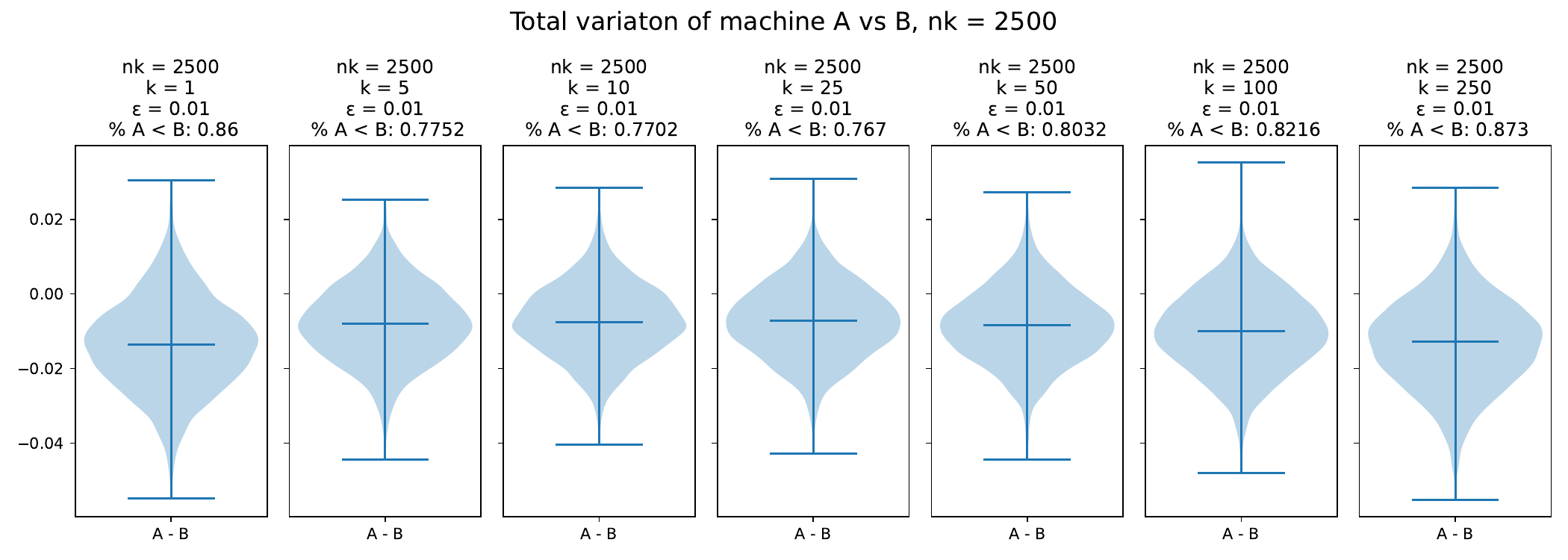}
    \end{subfigure} \hfill
    \begin{subfigure}[b]{\linewidth}
    \centering
    \includegraphics[width=0.825\linewidth]{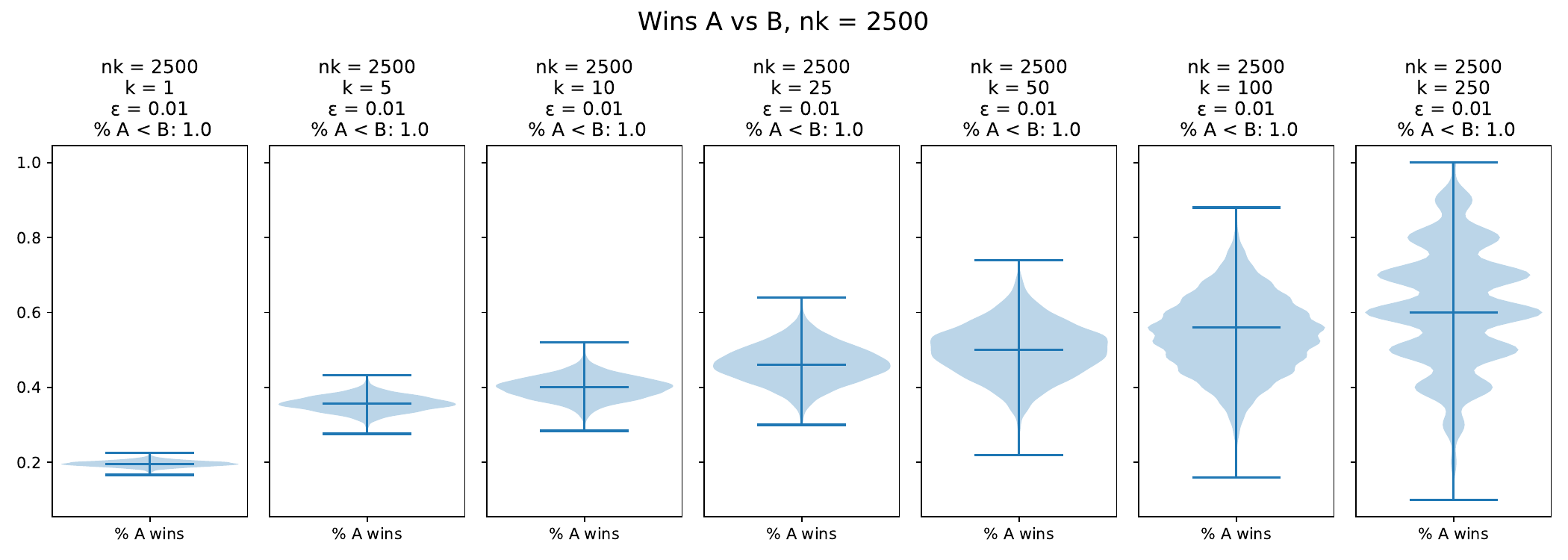}
    \end{subfigure} \hfill
    \begin{subfigure}[b]{\linewidth}
    \centering
    \includegraphics[width=0.825\linewidth]{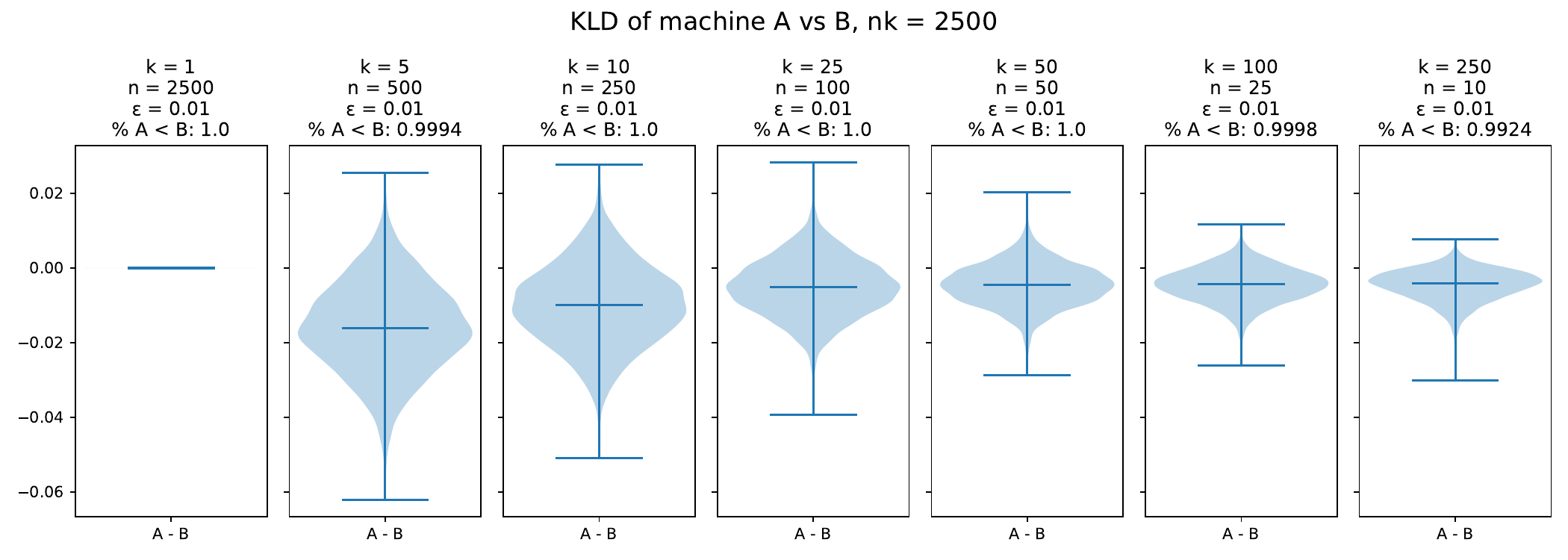}
    \end{subfigure} \hfill
    \caption{Distribution of scores differences for machines $A$ and $B$ against gold data over the Toxicity distribution, for $N \times K = 2500$, $\epsilon = 0.01$ for Accuracy and Wins.}
    \label{fig:score-variances}
\end{figure*}

Figure \ref{fig:within-variances} provides some insight into our results. It shows, using the distribution fitted to the JobsQ3 data for $N \times K = 2500$ as an example (but which is representative of other distributions). We see, first, that within-item variance drops precipitously from $K=1$ to $K=5$, and to nearly zero by $K=100$, by which point we can assume to have enough within-item samples for nearly any item-level metric. 

Looking at the metric scores, it is worth noting that raw scores generally improve as $k$ increases (though we do not show this here). Taking accuracy as an example, this is easy to see why: the larger $K$ is, the more likely it is that the most likely response is the most common one. So as long as the machine and gold have the same most frequent response, accuracy increases. However, accuracy increases for \emph{both} machines, and this, under our error model, allows for machine B to catch up. And as $N$ decreases, we would expect there to be more variance between individual samples.

Figure \ref{fig:score-variances} shows that for two of the metrics, accuracy and wins, their variance increases. Yet with wins, this increase occurs almost exclusively in a positive direction, whereas for accuracy (where p-values increase as $K$ increases), the variance is both positive and negative.

There are several limitations to our approach. Our simulator does not account for the machines using soft labels. We did not explore the impact of different noise models, and we see this as an important piece of the puzzle. We have not validated our results with real data collected based on the analysis of our simulation. This is because doing so would require us to collect multiple sets of data that have more annotations than are needed, and this is beyond our lab's budget.

\section{Conclusion}

In this work, we investigated the critical trade-off between the number of items ($N$) and the number of responses per item ($K$) for achieving reliable machine learning model evaluation under a fixed budget. Our findings demonstrate that increasing $K$ is often a more effective strategy for achieving reliable evaluation than increasing $N$. We discovered, across a diverse set of datasets, that accounting for the full human response distribution can be achieved with a surprisingly modest budget ($N \times K$) of 1000 or less, with $K>10$. Furthermore, we established that the relationship between $N$ and $K$ is heavily dependent on the chosen evaluation metric. Metrics that are more sensitive to the distributional nature of human responses benefit greatly from higher values of $K$. Our research provides a clear, data-driven methodology for ML practitioners to design more effective and budget-conscious evaluations. By moving beyond the single-truth paradigm and strategically collecting multiple responses, the field can build greater trust and confidence in model performance. Embracing human disagreement is not an expensive luxury but a cornerstone of robust and meaningful machine learning evaluation.



\bibliography{aaai2026, anthology, custom, references}

\newpage
\appendix


\section{Hypothesis testing}
To compare $H_{alt}$ and $H_{null}$, we use a metric $\Gamma(A, B, G)$ for each pair of response samples $\{A, B\}$ and gold samples $G$ to obtain a score. For each hypothesis, a distribution over metric scores $\Gamma^{H}$ is obtained by resampling.
We calculate a \pv\ based on (multisets) $\Gamma^{alt}$ and $\Gamma^{null}$ by using Algorithm \ref{alg:pval}. For our purposes, the \pv\ is the proportion of scores in the null distribution $\Gamma^{null}$ that exceed the scores in the alternative distribution $\Gamma^{alt}$.

\begin{algorithm}[h]
\SetKwInput{KwInput}{Input}
\caption{Calculate \pv\ }
\label{alg:pval}
\KwInput{$\Gamma^{alt}, \Gamma^{null}$}
$p \leftarrow 0$ \;
\For{{$score \in \Gamma^{alt}$}}{
    $p \leftarrow p + (|\Gamma^{null}>score|)/(|\Gamma^{null}|)$ \;
}
$p \leftarrow p/|\Gamma^{alt}|$ \;
\end{algorithm}

\section{MAP Estimate of the Model}
\label{sec:map_estimate}

\subsection{Model Definition}
For a given model, we have $n$ items, $k$ responses per item, and the responses are chosen from $m$ categories. For each item $i$, parameters $\bm{\beta_i}$ are sampled from a Dirichlet distribution parameterized by $m$ and $\alpha$. A response $x_{ij}$ is generated for each rater $j$ using a categorical distribution parameterized by $\bm{\beta_i}$. We also use a prior for the Dirichlet distribution.

\begin{displaymath}
    x_{ij} \sim Cat(\beta_{i1},...,\beta_{im})
\end{displaymath}
\begin{displaymath}
    \bm{\beta_i} = \beta_{i1},...,\beta_{im} \sim Dir(\alpha_1,...,\alpha_m)
\end{displaymath}
\begin{displaymath}
    \alpha = \alpha_1,...,\alpha_m
\end{displaymath}
\begin{displaymath}
    Prior = CD \left(\alpha \mid \nu, \eta \right)
\end{displaymath}

\subsection{Maximum
Likelihood Estimation}
\begin{displaymath}
\begin{split}
    L = & \prod_{i=1}^n \left\{ \frac{1}{B \left(\alpha \right)} \left(\prod_{m=1}^M \beta_{im}^{\alpha_m-1} \right) \prod_{j=1}^k \left(\prod_{m=1}^M \beta_{im}^{I \left\{x_{ij}=m \right\}}\right) \right\} \\
    &\text{where, } \frac{1}{B \left(\alpha \right)} = \frac{\Gamma (\sum_m \alpha_m)}{\prod\Gamma \left(\alpha_m \right)}
\end{split}
\end{displaymath}

\begin{displaymath}
    L = \prod_{i=1}^n \left\{ \frac{1}{B \left(\alpha \right)} \left(\prod_{m=1}^M \beta_{im}^{\alpha_m-1} \right) \prod_{m=1}^M \beta_{im}^{\sum_{j=1}^k I\left\{x_{ij}=m \right\}} \right\}
\end{displaymath}

\begin{displaymath}
    L = \prod_{i=1}^n\left\{ \frac{1}{B \left(\alpha \right)} \prod_{m=1}^M \beta_{im}^{\alpha_m - 1 + \sum_{j=1}^k I \left\{x_{ij}=m \right\}} \right\}
\end{displaymath}

\begin{equation}
\begin{split}
    \ln \left(L \right) = & \sum_{i=1}^n \left\{\ln\Gamma \left(\sum_{m=1}^M \alpha_m \right) - \sum_{m=1}^M \ln\Gamma \left(\alpha_m \right) + \right. \\
    & \left. \sum_{m=1}^M \left(\alpha_m - 1 + \sum_{j=1}^k I \left\{x_{ij}=m\right\}\right)\ln \beta_{im} \right\}
\end{split}
\label{eq:logl}
\end{equation}

Use Lagrange multipliers to enforce the constraint that $\sum_{m=1}^M \beta_{im} = 1$

\begin{displaymath}
    \mathcal{L} = \ln \left(L \right) + \sum_{i=1}^n \lambda_i \left(\sum_{m=1}^M \beta_{im} - 1 \right)
\end{displaymath}

\begin{equation}
    \frac{d \mathcal{L}}{d \lambda_i} = \sum_{m=1}^M \beta_{im} - 1
    \label{eq:d_li}
\end{equation}

Set the derivative to zero in Equation \ref{eq:d_li}

\begin{equation}
    \sum_{m=1}^M \beta_{im} = 1
    \label{eq:lagrc}
\end{equation}

\begin{equation}
    \frac{d \mathcal{L}}{d \beta_{im}} = \frac{1}{\beta_{im}} \left(\alpha_m - 1 + \sum_{j=1}^k I \left\{x_{ij}=m \}\right) \right) - \lambda_i
    \label{eq:d_pim}
\end{equation}

Set the derivative to zero in Equation \ref{eq:d_pim}

\begin{displaymath}
    \lambda_i = \frac{1}{\beta_{im}} \left(\alpha_m - 1 + \sum_{j=1}^k I \left\{x_{ij}=m \}\right) \right)
\end{displaymath}


\begin{displaymath}
    \beta_{im} = \frac{\left(\alpha_m - 1 + \sum_{j=1}^k I \left\{x_{ij}=m \}\right) \right)}{\lambda_i}
\end{displaymath}

Taking sum over all $m$ and using Equation \ref{eq:lagrc}

\begin{displaymath}
    1 = \sum_{m=1}^M \frac{\left(\alpha_m - 1 + \sum_{j=1}^k I \left\{x_{ij}=m \}\right) \right)}{\lambda_i}
\end{displaymath}

\begin{displaymath}
    \lambda_i = \sum_{m=1}^M \left(\alpha_m - 1 + \sum_{j=1}^k I \left\{x_{ij}=m \}\right) \right)
\end{displaymath}

\begin{equation}
    \beta_{im} = \frac{\alpha_m - 1 + \sum_{j=1}^k I \left\{x_{ij}=m \}\right)}{\sum_{m=1}^M \left(\alpha_m - 1 + \sum_{j=1}^k I \left\{x_{ij}=m \}\right) \right)}
    \label{eq:bim}
\end{equation}

\begin{displaymath}
\begin{split}
\frac{d \mathcal{L}}{d\alpha_m} = & \sum_{i=1}^n \left\{\Psi \left(\sum_{m=1}^M \alpha_m \right) - \Psi \left(\alpha_m \right) + \ln \beta_{im} \right\} \\
& \text{where, } \Psi \left(x \right) = \frac{d\ln\Gamma \left(x \right)}{dx}
\end{split}
\end{displaymath}

\begin{equation}
\begin{split}
\frac{d \mathcal{L}}{d\alpha_m} = & \enspace N\Psi \left(\sum_{m=1}^M \alpha_m \right) - N\Psi \left(\alpha_m \right) + N\ln \bar{\beta}_{m} \\
& \text{where, } \ln \bar{\beta}_{m} = \frac{1}{N}\sum_{i=1}^n\ln \beta_{im}
\end{split}
\label{eq:dl_am}
\end{equation}


Fixed-point iteration from \citet{minka2000estimating}:

\begin{equation}
     \Psi \left( \alpha_m^{new} \right) = \Psi \left( \sum_{m=1}^M \alpha_m^{old} \right) + \ln \bar{\beta}_{m}
     \label{eq:fpi}
\end{equation}

\subsection{Conjugate Prior of the Dirichlet Distribution}
Since the Dirichlet distribution belongs to the exponential family of probability distributions, it has a conjugate prior.

\begin{displaymath}
\begin{split}
    PR = & \enspace CD \left(\alpha \mid \nu, \eta \right)
        \propto \left(\frac{1}{B\left(\alpha\right)}\right)^\eta \exp\left(-\sum_{m=1}^M \nu_m \alpha_m\right)\\
        & \forall m \; v_m>0 \;and\; \eta>-1 \;\\
        & and \left(\eta \leq 0 \;or\; \sum_m \exp-\frac{v_m}{\eta}<1 \right)
\end{split}
\end{displaymath}

\begin{displaymath}
\begin{split}
    \ln \left(PR \right) = & \enspace \eta\left\{ \ln\Gamma\left(\sum_{m=1}^M \alpha_m \right) - \sum_{m=1}^M \ln\Gamma\left(\alpha_m \right) \right\}\\
    & - \sum_{m=1}^M \nu_k \alpha_m
\end{split}
\end{displaymath}

\begin{equation}
    \frac{d\ln \left(PR \right)}{d\alpha_m} = \eta\left\{ \Psi\left(\sum_{m=1}^M \alpha_m\right) - \Psi\left(\alpha_m\right) \right\} - \nu_m\\
\label{eq:dpr_am}
\end{equation}

\subsection{Posterior of the Model}

\begin{displaymath}
    Posterior \propto Prior*Likelihood
\end{displaymath}

\begin{displaymath}
   \frac{d\ln \left( PO \right)}{d\alpha_m} = \frac{d\ln \left(PR \right)}{d\alpha_m} + \frac{d \mathcal{L}}{d\alpha_m}
\end{displaymath}

Using Equation \ref{eq:dl_am} \& \ref{eq:dpr_am}

\begin{displaymath}
\begin{split}
   \frac{d\ln \left( PO \right)}{d\alpha_m} = \enspace & \eta\left\{ \Psi\left(\sum_{m=1}^M \alpha_m\right) - \Psi\left(\alpha_m\right) \right\} - \nu_m + \\
   & N \left\{ \Psi \left(\sum_{m=1}^M \alpha_m \right) - \Psi \left(\alpha_m \right) \right\} + N\ln \bar{\beta}_{m} \\
   & \text{where, } \ln \bar{\beta}_{m} = \frac{1}{N}\sum_{i=1}^n\ln \beta_{im}
\end{split}
\end{displaymath}

\begin{equation}
\begin{split}
   \frac{d \ln \left( PO \right)}{d \alpha_m} = &\left( \eta + N \right) \left\{ \Psi \left( \sum_{m=1}^M \alpha_m \right) - \Psi \left( \alpha_m \right) \right\}\\
   & + N\ln \bar{\beta}_{m} - \nu_m
\end{split}
\end{equation}

The Fixed-point iteration should follow from Equation \ref{eq:fpi} and \cite{minka2000estimating}:

\begin{equation}
     \Psi \left( \alpha_m^{new} \right) = \Psi \left( \sum_{m=1}^M \alpha_m^{old} \right) + \frac{1}{\left(\eta + N \right)}\left( N\ln \bar{\beta}_{m} - \nu_m \right)
     \label{eq:fpi2}
\end{equation}

\begin{displaymath}
   \frac{d\ln \left( PO \right)}{d\beta_{im}} = \frac{d\ln \left(PR \right)}{d\beta_{im}} + \frac{d \mathcal{L}}{d\beta_{im}}
\end{displaymath}

Derivative of prior w.r.t. $\beta_{im}$ is 0 and using Equation \ref{eq:d_pim} \& \ref{eq:bim}, we have:

\begin{displaymath}
    \beta_{im} = \frac{\alpha_m - 1 + \sum_{j=1}^k I \left\{x_{ij}=m \}\right)}{\sum_{m=1}^M \left(\alpha_m - 1 + \sum_{j=1}^k I \left\{x_{ij}=m \}\right) \right)}
\end{displaymath}

\subsection{Fitting to Real-World Datasets}

\begin{figure}[h]
  \centering
  \begin{subfigure}[b]{0.49\linewidth}
    \centering
    \includegraphics[width=\linewidth]{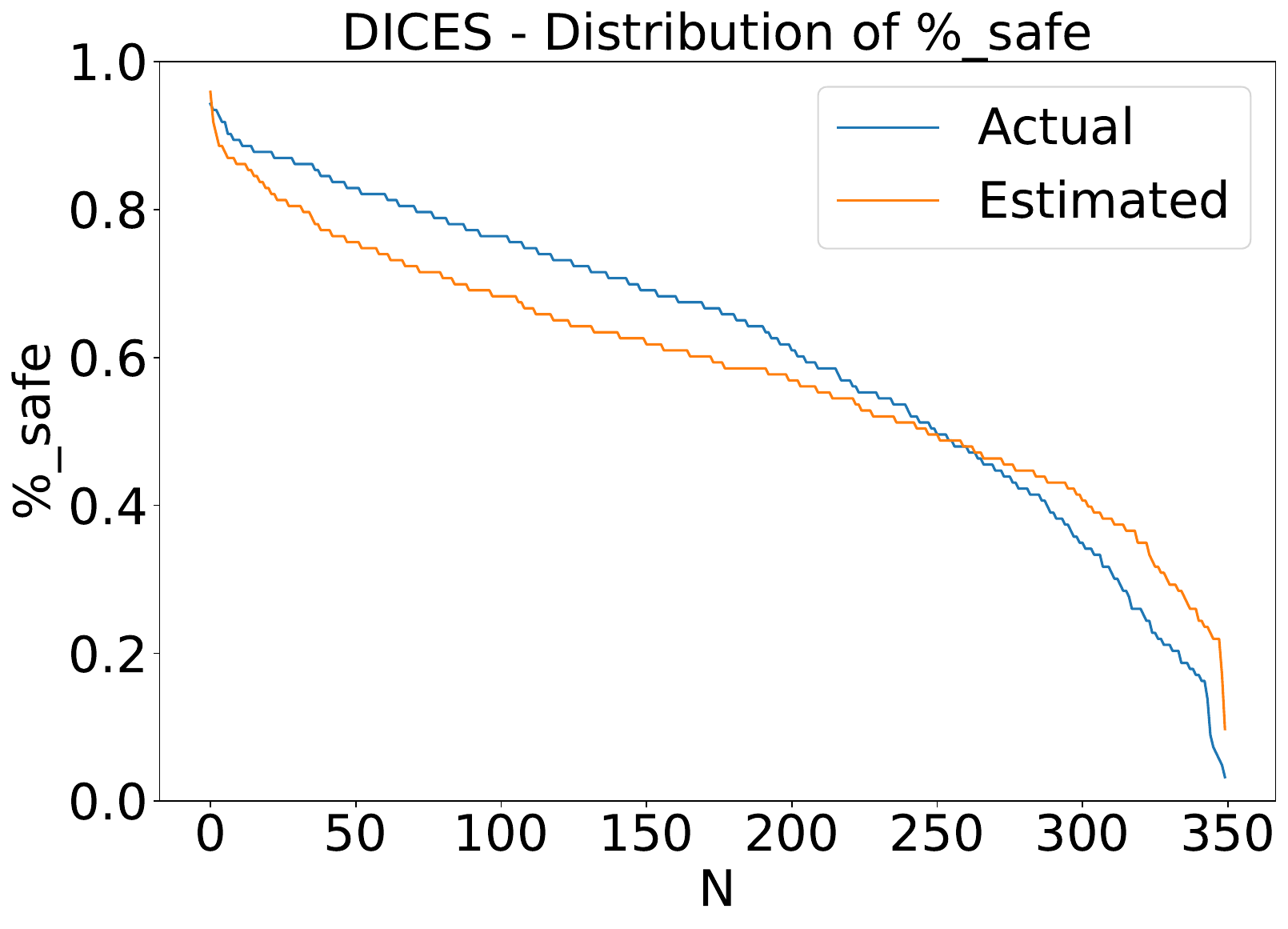}
    \label{fig:DICES_dist_per_safe}
  \end{subfigure} \hfill
  \begin{subfigure}[b]{0.49\linewidth}
    \centering
    \includegraphics[width=\linewidth]{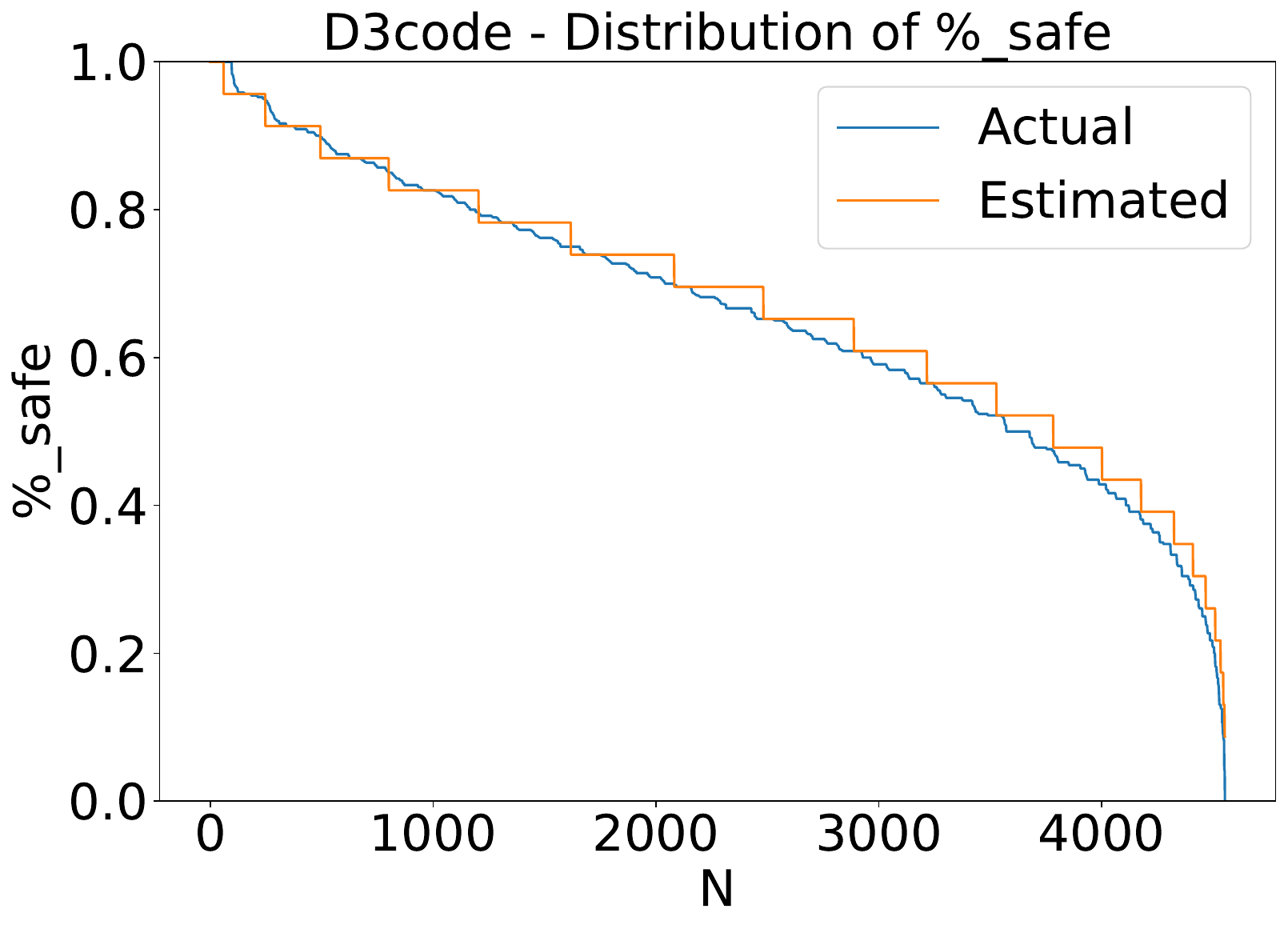}
    \label{fig:D3code_dist_per_safe}
  \end{subfigure} \hfill
  \caption{Distribution of percent safe responses using the actual data and estimated parameters demonstrating the goodness of model estimates.}
  \label{fig:dataset_fit}
\end{figure}

We fit the prior parameters $\mathbf{\alpha}$ of our model to real-world datasets by computing the maximum a posteriori (MAP) estimate of the model. Figure \ref{fig:dataset_fit} shows the distribution of percent safe responses in the DICES and D3code datasets using the actual data and estimated parameters to demonstrate the goodness of model estimates.

\section{Experiments}
\subsection{Experimental Setup}
We run experiments for hypothesis testing with different number of ratings ($N\times K$ = \{100, 250, 500, 1000, 2500, 5000, 10000, 25000, 50000\}) while ranging $K$ from 1 to 500 (in increments of 1 till 10, then 20, then in increments of 20 from 20 onwards)
for different metrics, and $\epsilon$ = \{0.1, 0.2, 0.3, 0.4\}. We use four metrics with four $\epsilon$, yielding 16 sets of 282 experiments for each dataset.

For the real-world datasets, the value of parameter $\alpha$ is determined using the MAP estimate and $\rho = [1/M] \times M$ where $M$ is fixed for each dataset. To estimate the \pv s\, we repeat the sampling process 1000 times.

We also run our experiments by choosing the parameter $\alpha$ using different prior distributions, \textbf{balanced} ($\alpha=[3] \times M$) and \textbf{unbalanced} ($\alpha=[10]+[3] \times M$) to simulate class imbalance, and varying the number of categories ($M$ = \{2, 3, 4, 5, 12\}).

\subsection{Implementation Details}
We implement our system by building on top of Google's VET library\footnote{https://github.com/google-research/vet}. The code has been supplied with the supplemental material and will be made publicly available later. Our experiments are conducted on either a local machine with a 16-core processor and 64GB RAM or a compute node with a 16-core processor and 40GB RAM. The experiments take anywhere between 12 and 18 hours to run for a single dataset.

\section{Results}

Tables \ref{tab:low_k_for_p_lte_05_nk_min_p_full} shows the results for minimum \pv, $K$, and corresponding effect size ($\Delta$) for lowest $N\times K$ with $p<=0.05$ ($ \epsilon=0.3$). Table \ref{tab:k_low_ci_nk_full} shows the results for the lowest CI-width with the corresponding value of $K$ and effect size - $\Delta$ for the lowest $N\times K$ observed in Table \ref{tab:low_k_for_p_lte_05_nk_min_p_full}.

\begin{table}[htbp!]
\centering
\small
\begin{tabular}{l|c|cccc}
 &  & Accuracy & TV & Wins & KL-Div \\
\midrule
 & NK & 2500 & 1000 & 2500 & 1000 \\
Toxicity & p-value & 0.012 & 0.015 & 0.012 & 0.022 \\
(M=2) & K & 1 & 120 & 1 & 200 \\
 & $\Delta$ & 0.040 & 0.074 & 0.040 & 0.044 \\
 \hline
 & NK & 1000 & 500 & 1000 & 1000 \\
DICES & p-value & 0.036 & 0.017 & 0.028 & 0.020 \\
(M=3) & K & 1 & 80 & 20 & 300 \\
 & $\Delta$ & 0.055 & 0.063 & 0.346 & 0.082 \\
 \hline
 & NK & 2500 & 1000 & 2500 & 1000 \\
D3code & p-value & 0.037 & 0.020 & 0.024 & 0.022 \\
(M=2) & K & 2 & 140 & 60 & 100 \\
 & $\Delta$ & 0.034 & 0.072 & 0.413 & 0.036 \\
 \hline
 & NK & 250 & 250 & 250 & 250 \\
JobsQ1 & p-value & 0.035 & 0.015 & 0.036 & 0.035 \\
(M=5) & K & 1 & 40 & 1 & 1 \\
 & $\Delta$ & 0.104 & 0.050 & 0.104 & 2.864 \\
 \hline
 & NK & 500 & 250 & 500 & 500 \\
JobsQ3 & p-value & 0.047 & 0.014 & 0.038 & 0.030 \\
(M=12) & K & 100 & 240 & 80 & 500 \\
 & $\Delta$ & 0.595 & 0.024 & 0.868 & 0.182 \\
 \midrule
 \midrule
 & NK & 5000 & 1000 & 2500 & 1000 \\
Balanced & p-value & 0.029 & 0.025 & 0.034 & 0.024 \\
(M=2) & K & 20 & 80 & 80 & 100 \\
 & $\Delta$ & 0.087 & 0.059 & 0.446 & 0.032 \\
 \hline
 & NK & 5000 & 500 & 2500 & 500 \\
Balanced & p-value & 0.010 & 0.038 & 0.005 & 0.039 \\
(M=3) & K & 40 & 100 & 60 & 100 \\
 & $\Delta$ & 0.159 & 0.053 & 0.510 & 0.054 \\
 \hline
 & NK & 2500 & 500 & 1000 & 500 \\
Balanced & p-value & 0.039 & 0.022 & 0.029 & 0.025 \\
(M=4) & K & 40 & 100 & 80 & 240 \\
 & $\Delta$ & 0.187 & 0.044 & 0.672 & 0.071 \\
 \hline
 & NK & 2500 & 500 & 1000 & 500 \\
Balanced & p-value & 0.028 & 0.012 & 0.025 & 0.016 \\
(M=5) & K & 60 & 240 & 40 & 220 \\
 & $\Delta$ & 0.247 & 0.046 & 0.509 & 0.087 \\
 \hline
 & NK & 2500 & 250 & 1000 & 250 \\
Balanced & p-value & 0.017 & 0.022 & 0.005 & 0.045 \\
(M=12) & K & 80 & 240 & 100 & 240 \\
 & $\Delta$ & 0.295 & 0.022 & 0.882 & 0.159 \\
  \midrule
 \midrule
 & NK & 1000 & 500 & 1000 & 1000 \\
Unbalanced & p-value & 0.031 & 0.044 & 0.031 & 0.014 \\
(M=2) & K & 1 & 80 & 1 & 140 \\
 & $\Delta$ & 0.050 & 0.074 & 0.050 & 0.047 \\
 \hline
 & NK & 1000 & 500 & 1000 & 500 \\
Unbalanced & p-value & 0.039 & 0.023 & 0.040 & 0.031 \\
(M=3) & K & 2 & 100 & 40 & 100 \\
 & $\Delta$ & 0.061 & 0.061 & 0.473 & 0.068 \\
 \hline
 & NK & 1000 & 500 & 1000 & 500 \\
Unbalanced & p-value & 0.049 & 0.013 & 0.022 & 0.021 \\
(M=4) & K & 4 & 120 & 40 & 240 \\
 & $\Delta$ & 0.089 & 0.054 & 0.520 & 0.084 \\
 \hline
 & NK & 1000 & 500 & 1000 & 500 \\
Unbalanced & p-value & 0.045 & 0.009 & 0.015 & 0.010 \\
(M=5) & K & 10 & 100 & 40 & 240 \\
 & $\Delta$ & 0.138 & 0.043 & 0.545 & 0.098 \\
 \hline
 & NK & 1000 & 250 & 500 & 500 \\
Unbalanced & p-value & 0.027 & 0.014 & 0.042 & 0.004 \\
(M=12) & K & 80 & 240 & 60 & 460 \\
 & $\Delta$ & 0.436 & 0.023 & 0.763 & 0.156 \\
\end{tabular}
\caption{Minimum \pv, $K$, and corresponding effect size ($\Delta$) for lowest $N\times K$ with $p<=0.05$ ($ \epsilon=0.3$).}
\label{tab:low_k_for_p_lte_05_nk_min_p_full}
\end{table}

\begin{table}[htbp!]
\centering
\small
\begin{tabular}{l|c|cccc}
 &  & Accuracy & TV & Wins & KL-Div \\
\midrule
 & NK & 2500 & 1000 & 2500 & 1000 \\
Toxicity & ci-width & 0.050 & 0.067 & 0.050 & 0.085 \\
(M=2) & K & 1 & 5 & 1 & 100 \\
 & $\Delta$ & 0.117 & 0.063 & 0.728 & 0.042 \\
 \hline
 & NK & 1000 & 500 & 1000 & 1000 \\
DICES & ci-width & 0.090 & 0.063 & 0.090 & 0.230 \\
(M=3) & K & 1 & 7 & 1 & 100 \\
 & $\Delta$ & 0.080 & 0.075 & 0.203 & 0.082 \\
 \hline
 & NK & 2500 & 1000 & 2500 & 1000 \\
D3code & ci-width & 0.054 & 0.067 & 0.054 & 0.068 \\
(M=2) & K & 1 & 7 & 1 & 100 \\
 & $\Delta$ & 0.116 & 0.066 & 0.669 & 0.036 \\
 \hline
 & NK & 250 & 250 & 250 & 250 \\
JobsQ1 & ci-width & 0.160 & 0.055 & 0.160 & 1.516 \\
(M=5) & K & 1 & 8 & 1 & 80 \\
 & $\Delta$ & 0.104 & 0.038 & 0.104 & 0.109 \\
 \hline
 & NK & 500 & 250 & 500 & 500 \\
JobsQ3 & ci-width & 0.086 & 0.020 & 0.086 & 1.322 \\
(M=12) & K & 1 & 1 & 1 & 100 \\
 & $\Delta$ & 0.614 & 0.003 & 0.941 & 1.613 \\
 \midrule
 \midrule
 & NK & 5000 & 1000 & 2500 & 1000 \\
Balanced & ci-width & 0.038 & 0.068 & 0.054 & 0.060 \\
(M=2) & K & 1 & 4 & 1 & 100 \\
 & $\Delta$ & 0.039 & 0.048 & 0.689 & 0.032 \\
 \hline
 & NK & 5000 & 500 & 2500 & 500 \\
Balanced & ci-width & 0.035 & 0.063 & 0.052 & 0.120 \\
(M=3) & K & 1 & 3 & 1 & 100 \\
 & $\Delta$ & 0.042 & 0.057 & 0.852 & 0.256 \\
 \hline
 & NK & 2500 & 500 & 1000 & 500 \\
Balanced & ci-width & 0.052 & 0.048 & 0.078 & 0.151 \\
(M=4) & K & 1 & 8 & 1 & 100 \\
 & $\Delta$ & 0.333 & 0.008 & 0.144 & 0.522 \\
 \hline
 & NK & 2500 & 500 & 1000 & 500 \\
Balanced & ci-width & 0.046 & 0.038 & 0.075 & 0.201 \\
(M=5) & K & 1 & 10 & 1 & 100 \\
 & $\Delta$ & 0.379 & 0.010 & 0.160 & 0.799 \\
 \hline
 & NK & 2500 & 250 & 1000 & 250 \\
Balanced & ci-width & 0.034 & 0.018 & 0.055 & 2.026 \\
(M=12) & K & 1 & 1 & 1 & 100 \\
 & $\Delta$ & 0.494 & 0.001 & 0.167 & 0.289 \\
 \midrule
 \midrule
 & NK & 1000 & 500 & 1000 & 1000 \\
Unbalanced & ci-width & 0.070 & 0.093 & 0.079 & 0.083 \\
(M=2) & K & 2 & 7 & 1 & 80 \\
 & $\Delta$ & 0.068 & 0.086 & 0.154 & 0.045 \\
 \hline
 & NK & 1000 & 500 & 1000 & 500 \\
Unbalanced & ci-width & 0.079 & 0.063 & 0.079 & 0.145 \\
(M=3) & K & 1 & 10 & 1 & 100 \\
 & $\Delta$ & 0.111 & 0.028 & 0.180 & 0.028 \\
 \hline
 & NK & 1000 & 500 & 1000 & 500 \\
Unbalanced & ci-width & 0.081 & 0.048 & 0.081 & 0.218 \\
(M=4) & K & 1 & 6 & 1 & 100 \\
 & $\Delta$ & 0.125 & 0.058 & 0.191 & 0.243 \\
 \hline
 & NK & 1000 & 500 & 1000 & 500 \\
Unbalanced & ci-width & 0.070 & 0.039 & 0.070 & 0.467 \\
(M=5) & K & 1 & 7 & 1 & 100 \\
 & $\Delta$ & 0.128 & 0.049 & 0.187 & 0.515 \\
 \hline
 & NK & 1000 & 250 & 500 & 500 \\
Unbalanced & ci-width & 0.056 & 0.019 & 0.080 & 1.292 \\
(M=12) & K & 1 & 1 & 1 & 100 \\
 & $\Delta$ & 0.105 & 0.002 & 0.920 & 1.264 \\
 \end{tabular}
\caption{Lowest CI-width with corresponding value of $K$ and effect size - $\Delta$ for lowest $N\times K$ with $p<=0.05$ ($ \epsilon=0.3$).}
\label{tab:k_low_ci_nk_full}
\end{table}

\subsection{Accuracy}

\subsubsection{Real-World Datasets}

\begin{figure*}
  \centering
  \begin{subfigure}[b]{0.24\linewidth}
    \centering
    \includegraphics[width=\linewidth]{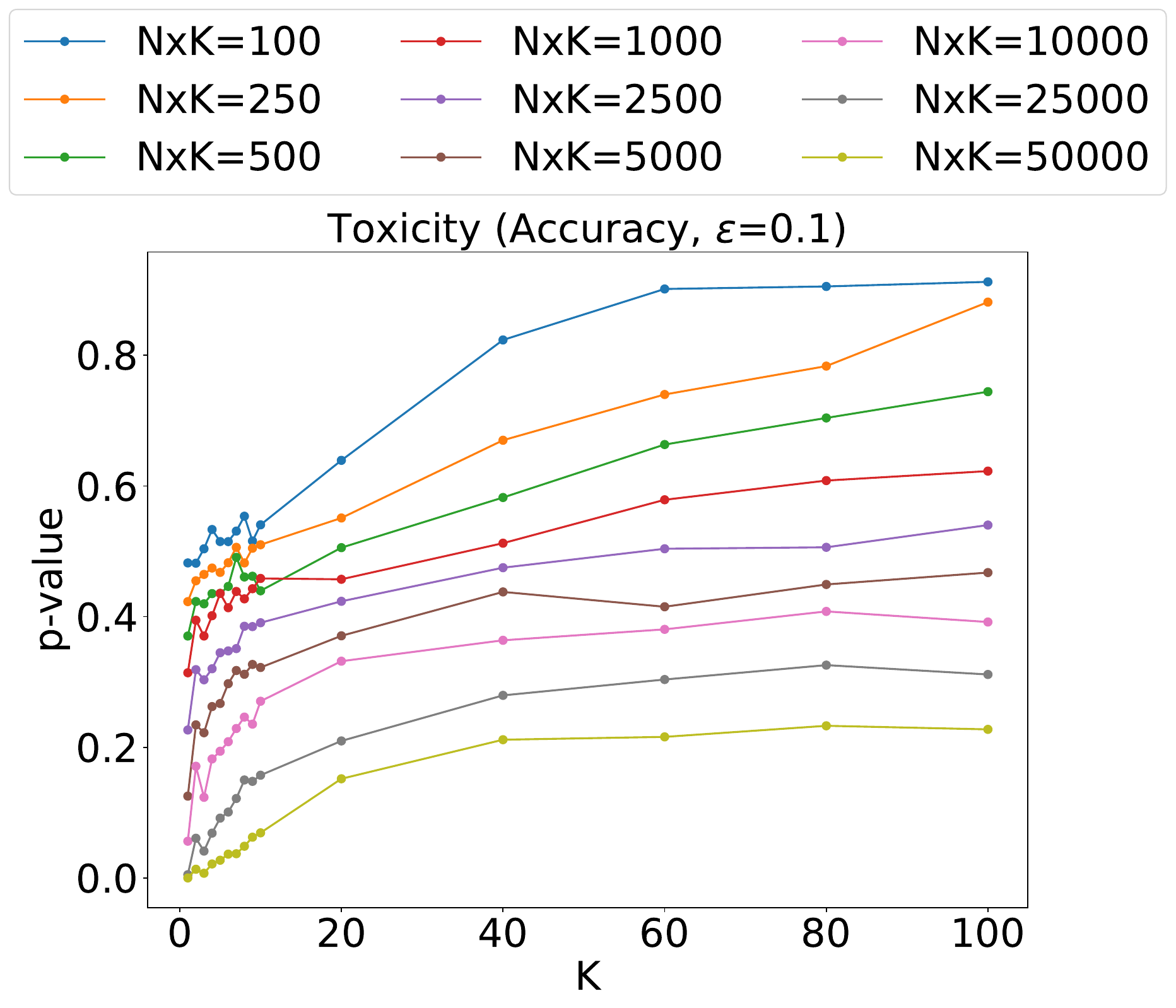}
    \caption{$\epsilon = 0.1$}
    \label{fig:toxicity_acc_e01}
  \end{subfigure} \hfill
  \begin{subfigure}[b]{0.24\linewidth}
    \centering
    \includegraphics[width=\linewidth]{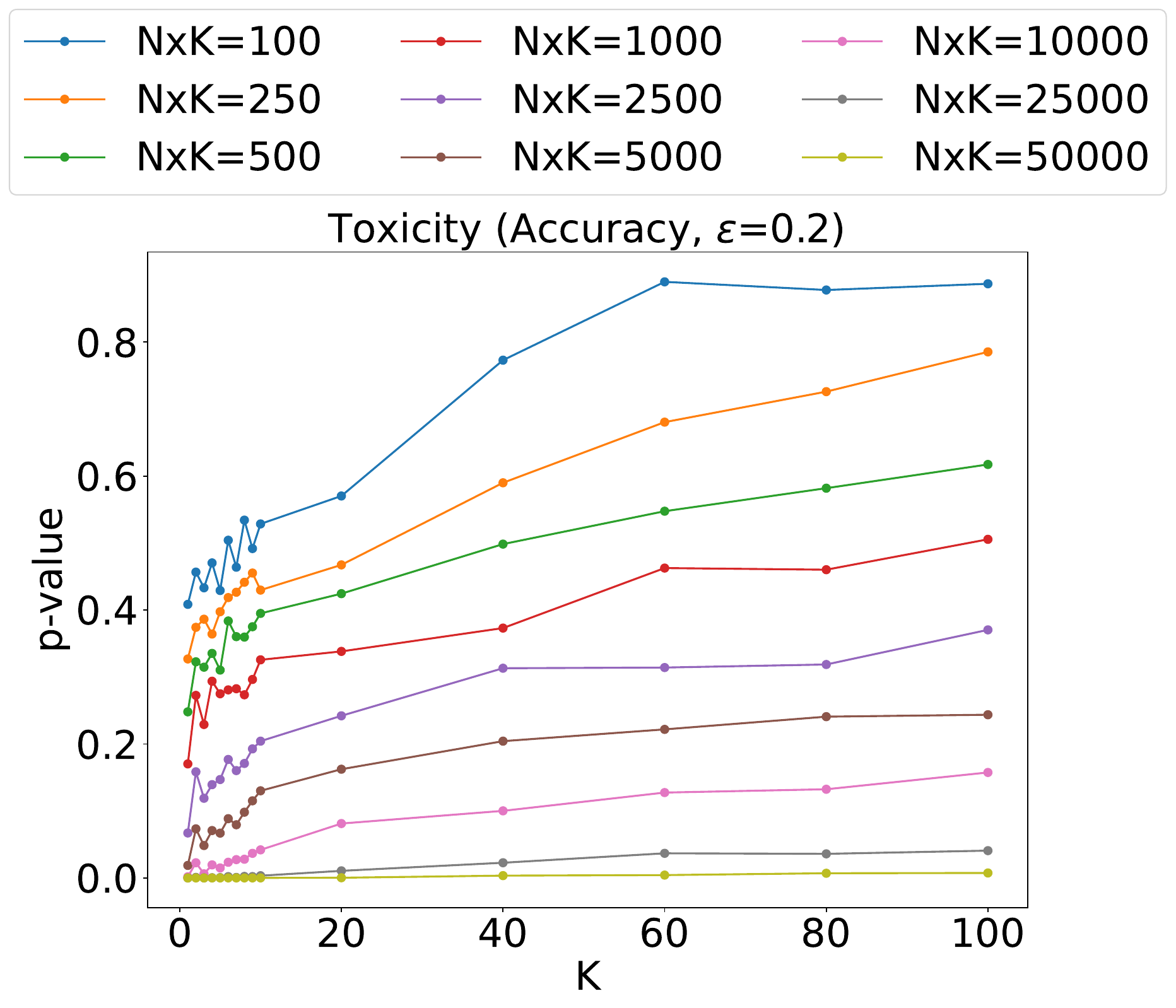}
    \caption{$\epsilon = 0.2$}
    \label{fig:toxicity_acc_e02}
  \end{subfigure} \hfill
  \begin{subfigure}[b]{0.24\linewidth}
    \centering
    \includegraphics[width=\linewidth]{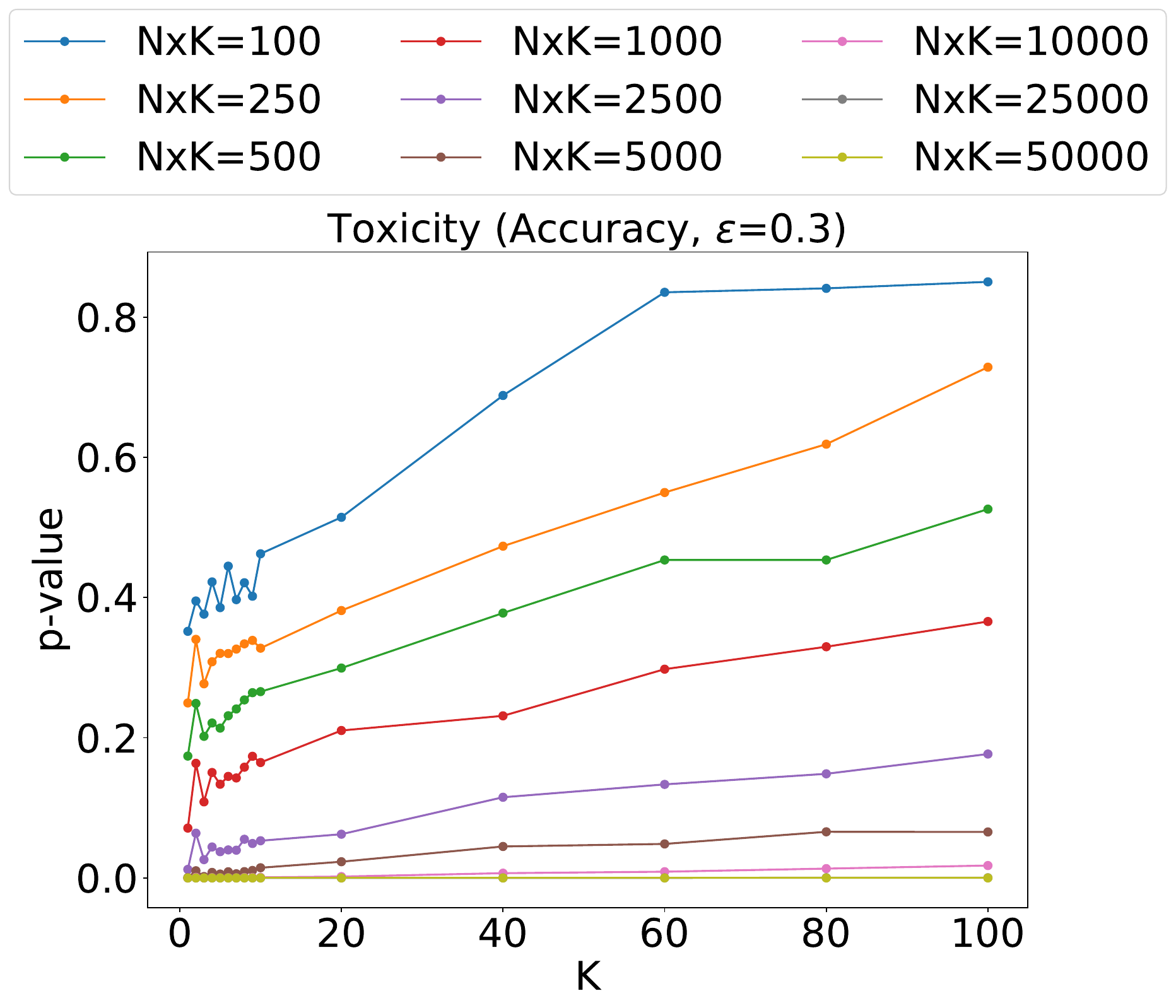}
    \caption{$\epsilon = 0.3$}
    \label{fig:toxicity_acc_e03}
  \end{subfigure} \hfill
  \begin{subfigure}[b]{0.24\linewidth}
    \centering
    \includegraphics[width=\linewidth]{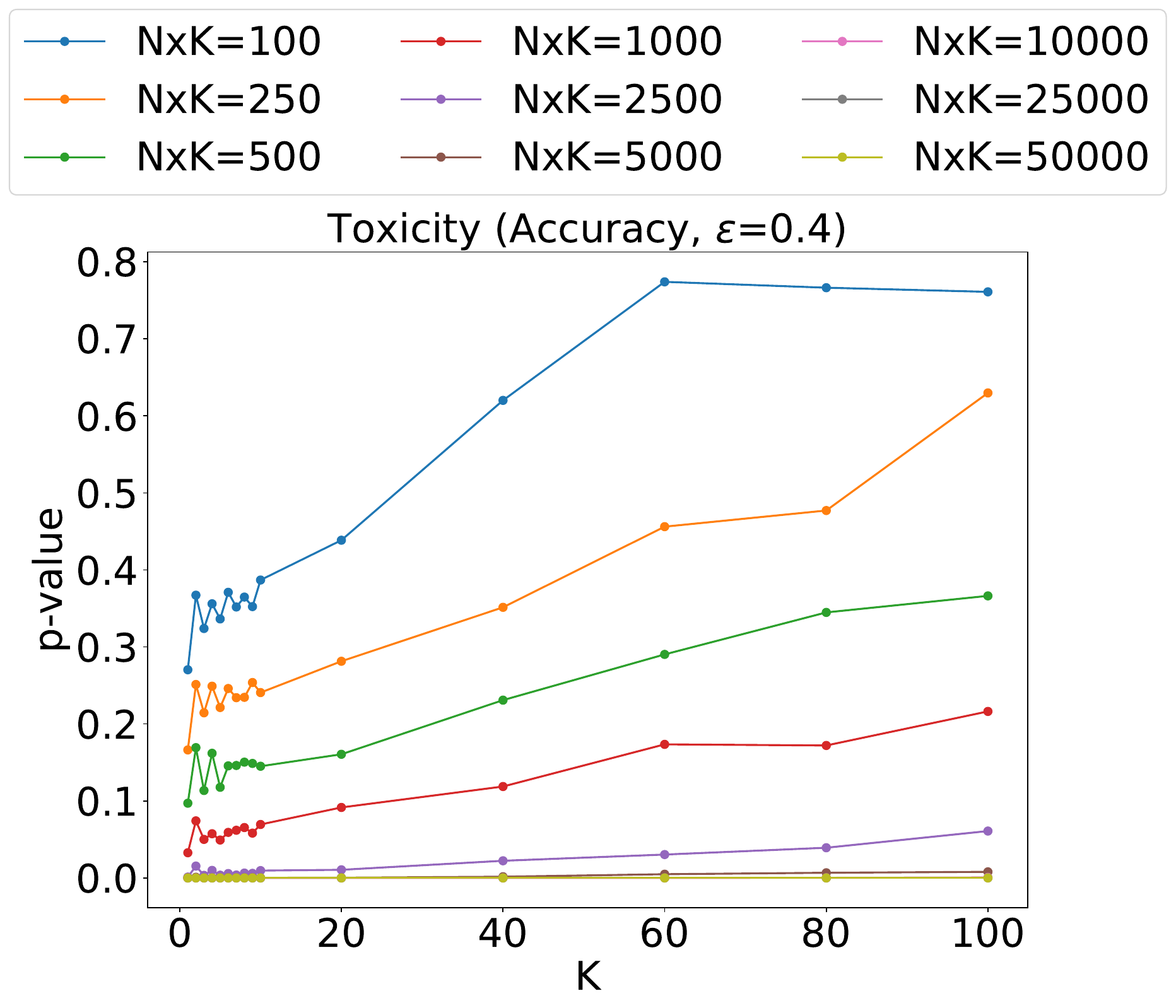}
    \caption{$\epsilon = 0.4$}
    \label{fig:toxicity_acc_e04}
  \end{subfigure}
  \caption{P-value plots for Toxicity dataset with Accuracy as the metric}
  \label{fig:toxicity_accuracy}
\end{figure*}

\begin{figure*}
  \centering
  \begin{subfigure}[b]{0.24\linewidth}
    \centering
    \includegraphics[width=\linewidth]{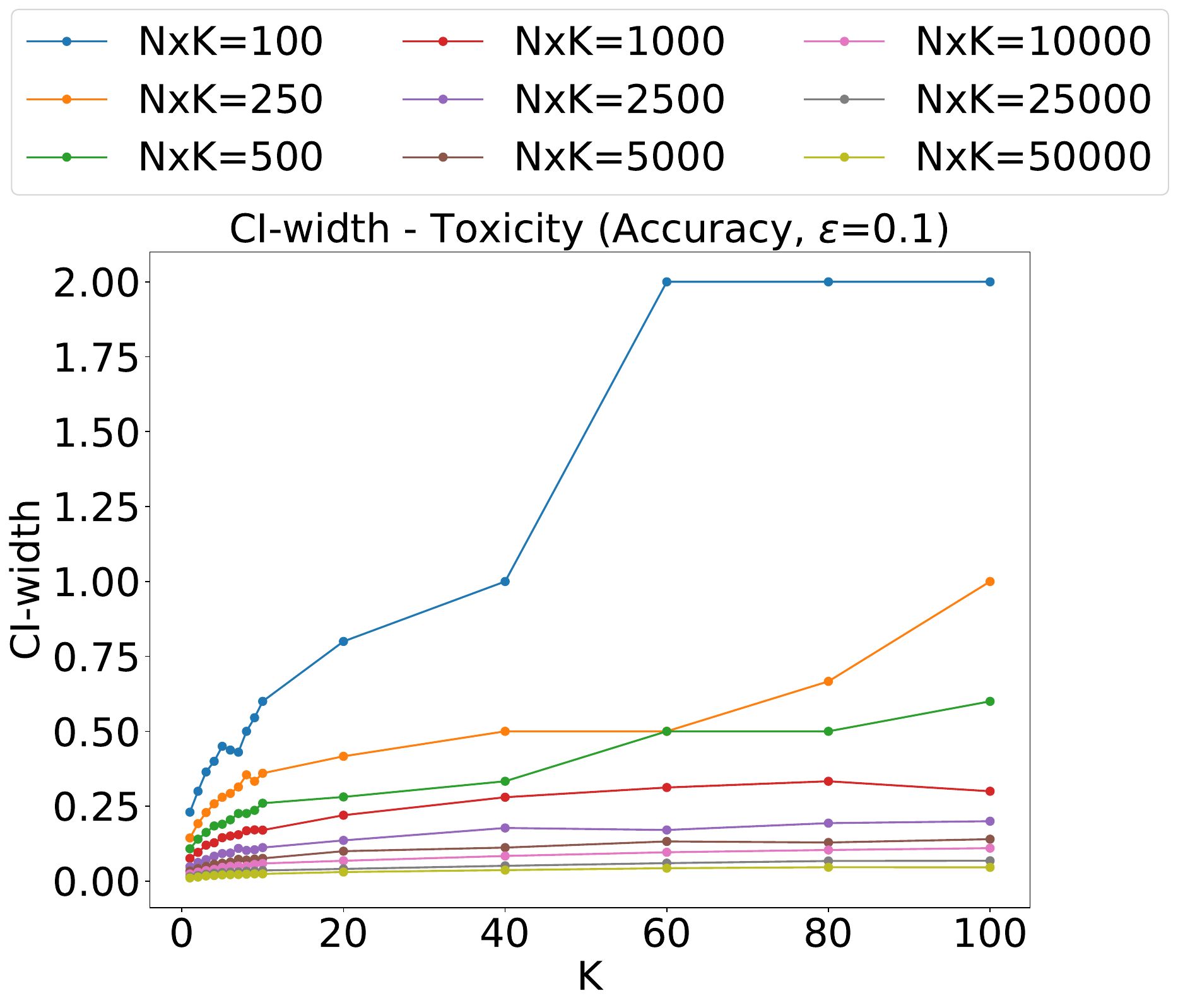}
    \caption{$\epsilon = 0.1$}
    \label{fig:toxicity_ci_acc_e01}
  \end{subfigure} \hfill
  \begin{subfigure}[b]{0.24\linewidth}
    \centering
    \includegraphics[width=\linewidth]{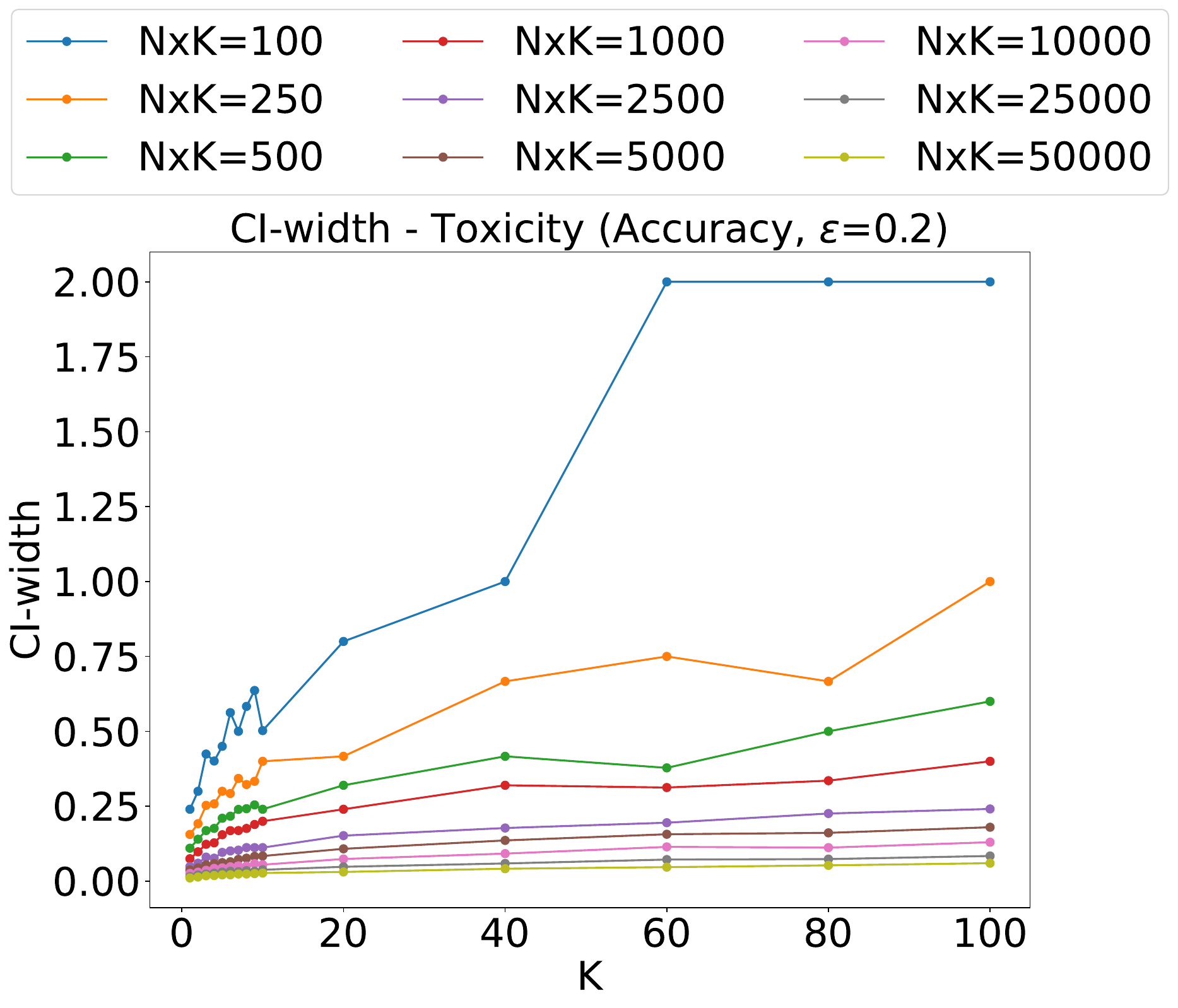}
    \caption{$\epsilon = 0.2$}
    \label{fig:toxicity_ci_acc_e02}
  \end{subfigure} \hfill
  \begin{subfigure}[b]{0.24\linewidth}
    \centering
    \includegraphics[width=\linewidth]{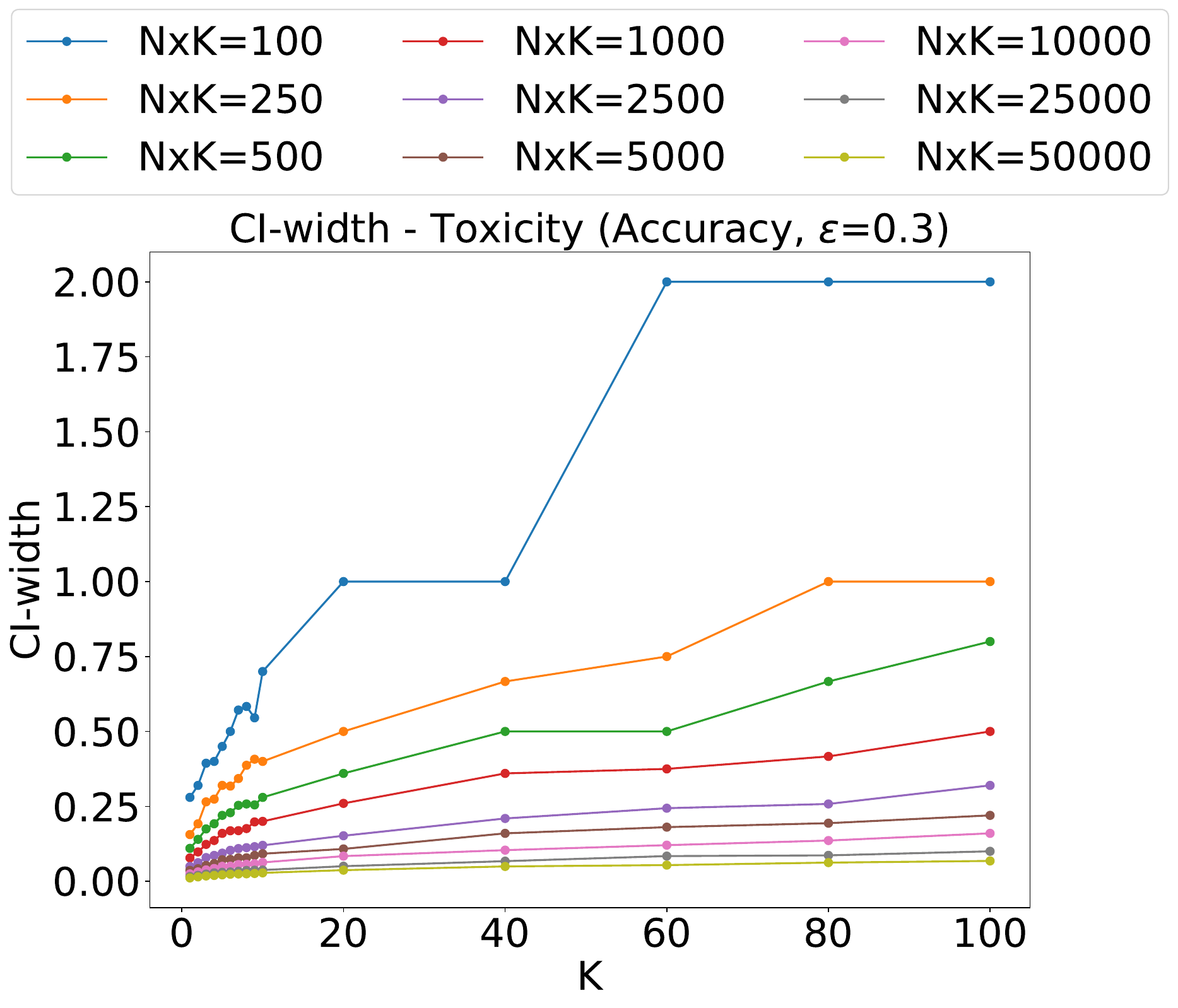}
    \caption{$\epsilon = 0.3$}
    \label{fig:toxicity_ci_acc_e03}
  \end{subfigure} \hfill
  \begin{subfigure}[b]{0.24\linewidth}
    \centering
    \includegraphics[width=\linewidth]{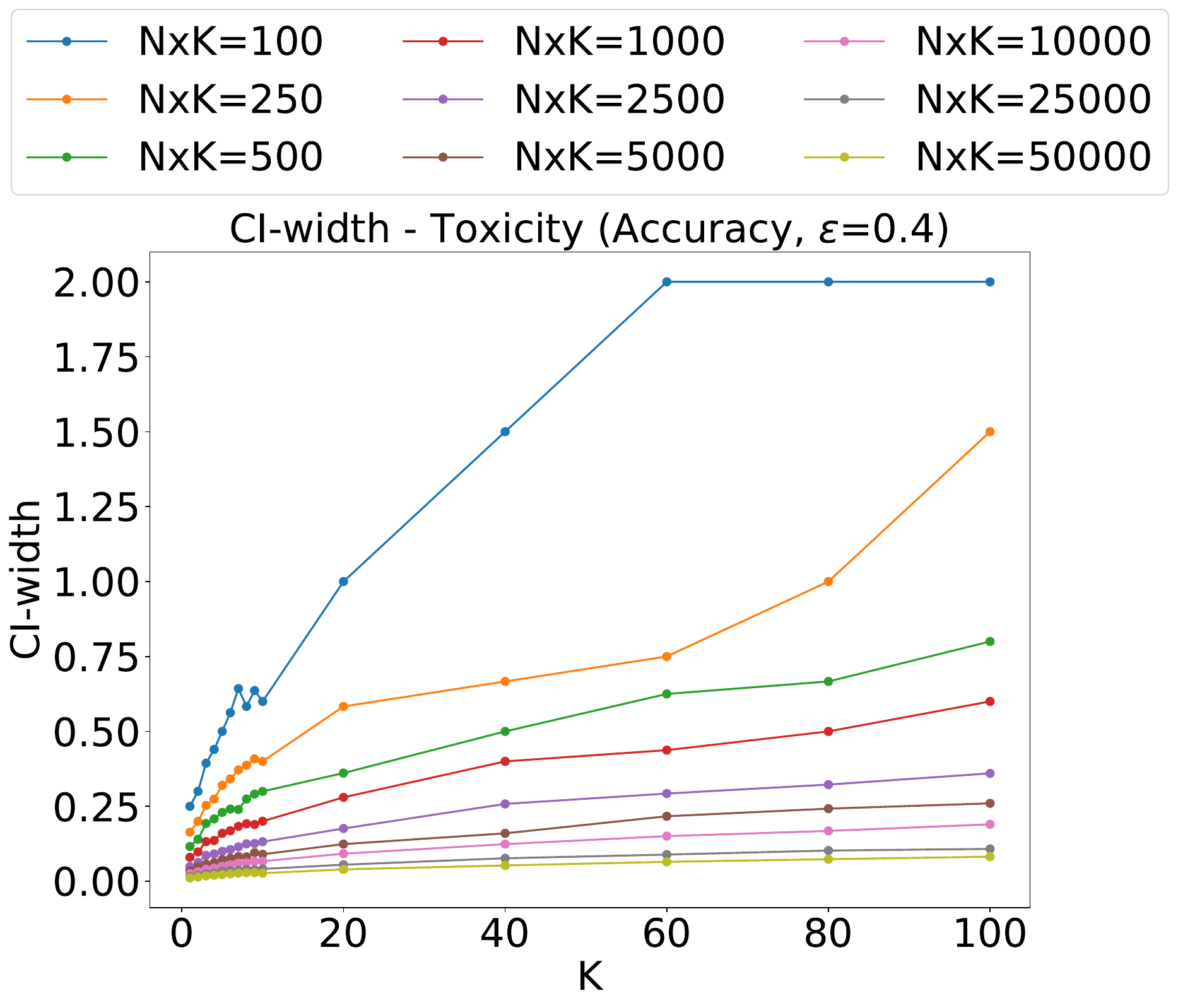}
    \caption{$\epsilon = 0.4$}
    \label{fig:toxicity_ci_acc_e04}
  \end{subfigure}
  \caption{CI-width plots for Toxicity dataset with Accuracy as the metric}
  \label{fig:toxicity_ci_accuracy}
\end{figure*}

\begin{figure*}
  \centering
  \begin{subfigure}[b]{0.24\linewidth}
    \centering
    \includegraphics[width=\linewidth]{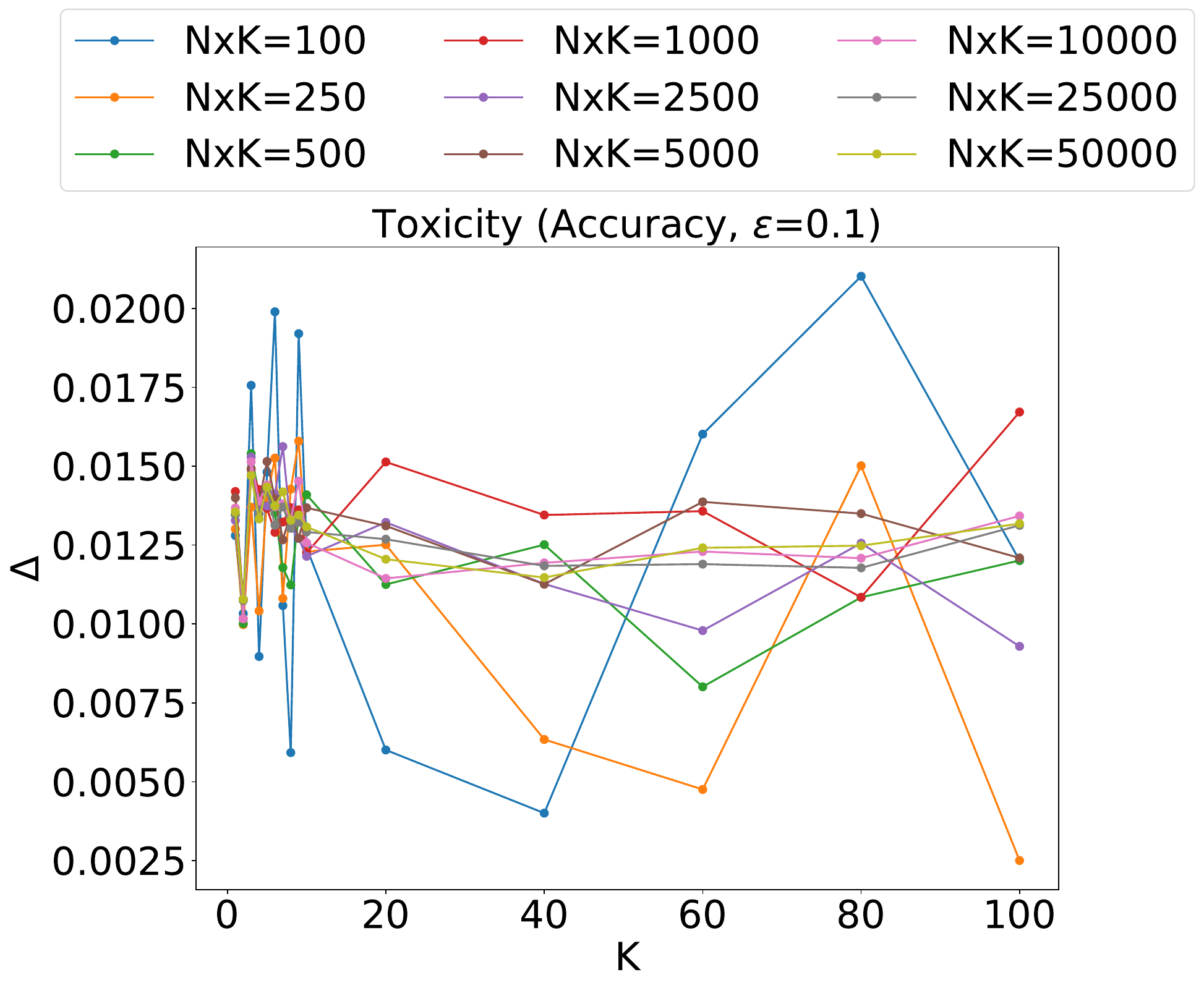}
    \caption{$\epsilon = 0.1$}
    \label{fig:toxicity_delta_acc_e01}
  \end{subfigure} \hfill
  \begin{subfigure}[b]{0.24\linewidth}
    \centering
    \includegraphics[width=\linewidth]{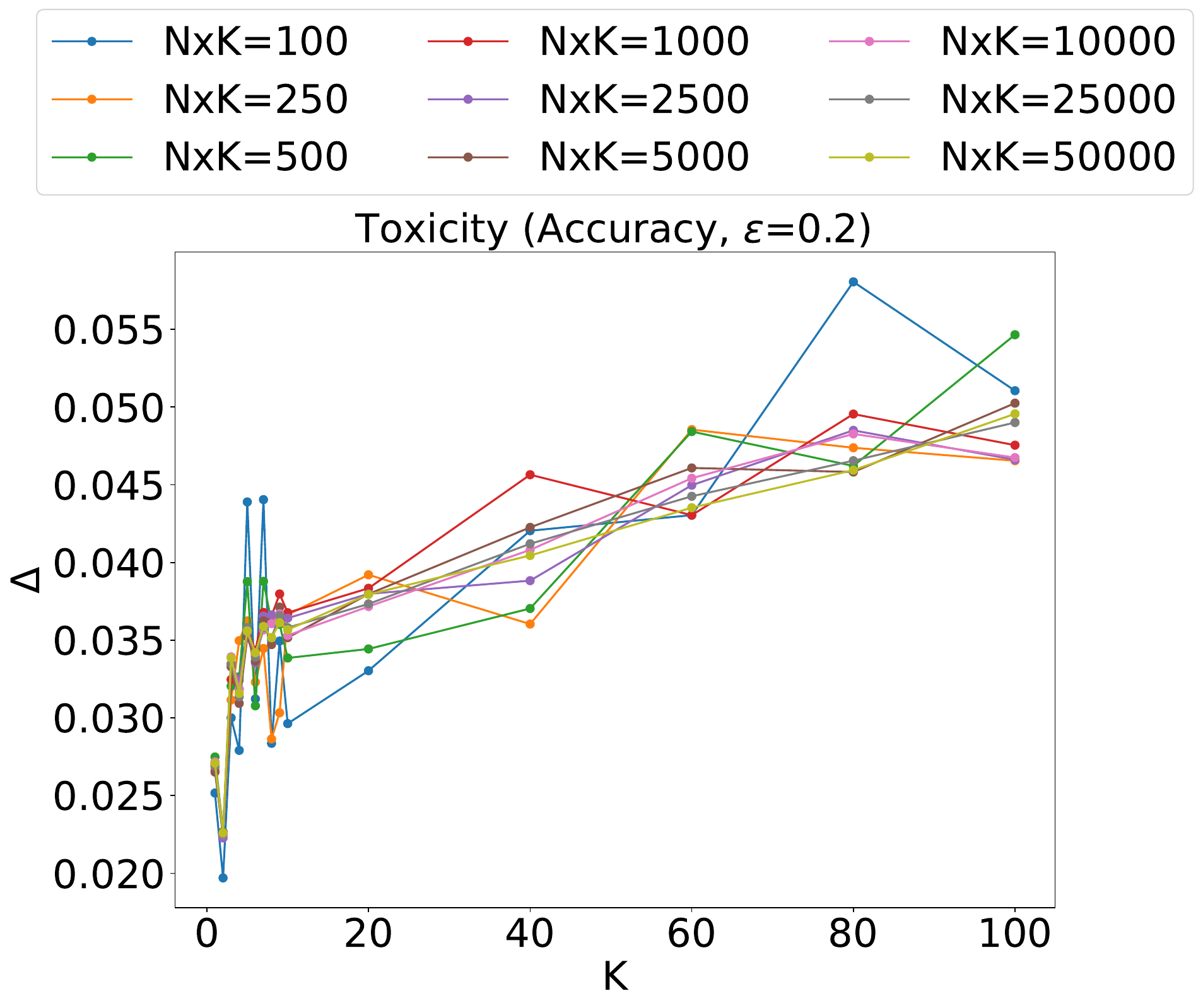}
    \caption{$\epsilon = 0.2$}
    \label{fig:toxicity_delta_acc_e02}
  \end{subfigure} \hfill
  \begin{subfigure}[b]{0.24\linewidth}
    \centering
    \includegraphics[width=\linewidth]{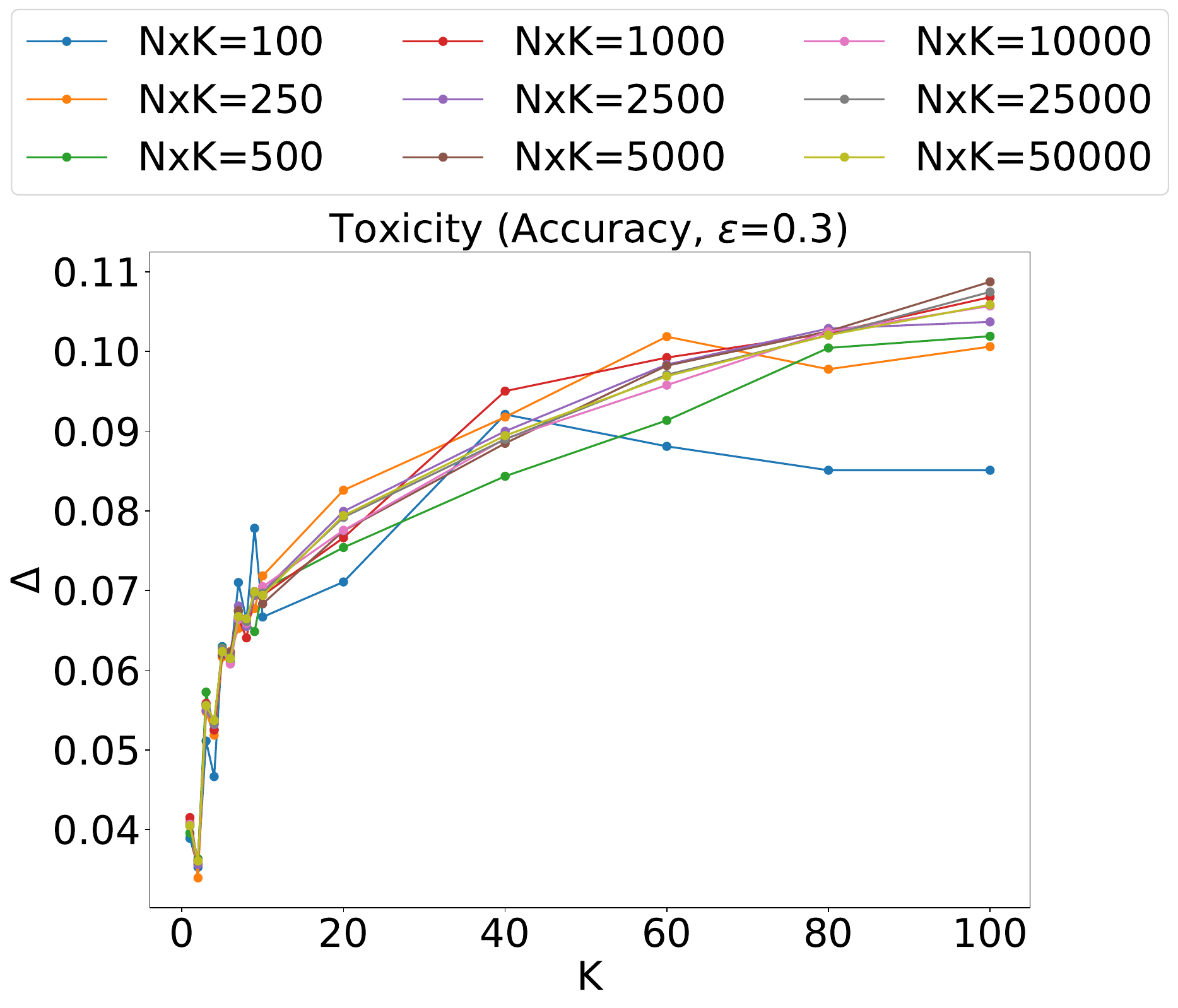}
    \caption{$\epsilon = 0.3$}
    \label{fig:toxicity_delta_acc_e03}
  \end{subfigure} \hfill
  \begin{subfigure}[b]{0.24\linewidth}
    \centering
    \includegraphics[width=\linewidth]{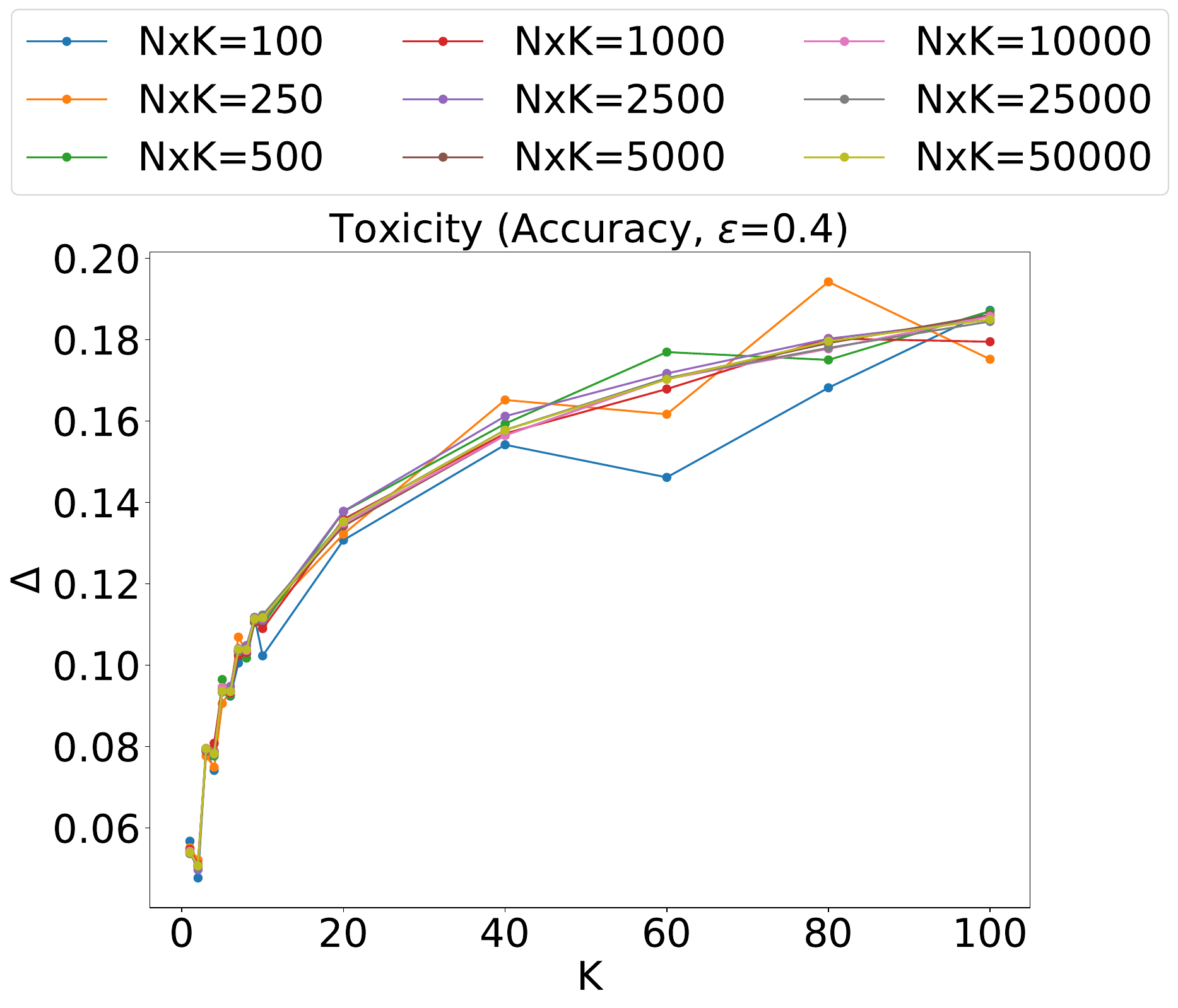}
    \caption{$\epsilon = 0.4$}
    \label{fig:toxicity_delta_acc_e04}
  \end{subfigure}
  \caption{Effect sizes ($\Delta$) for Toxicity dataset with Accuracy as the metric}
  \label{fig:toxicity_delta_accuracy}
\end{figure*}

\begin{figure*}
  \centering
  \begin{subfigure}[b]{0.24\linewidth}
    \centering
    \includegraphics[width=\linewidth]{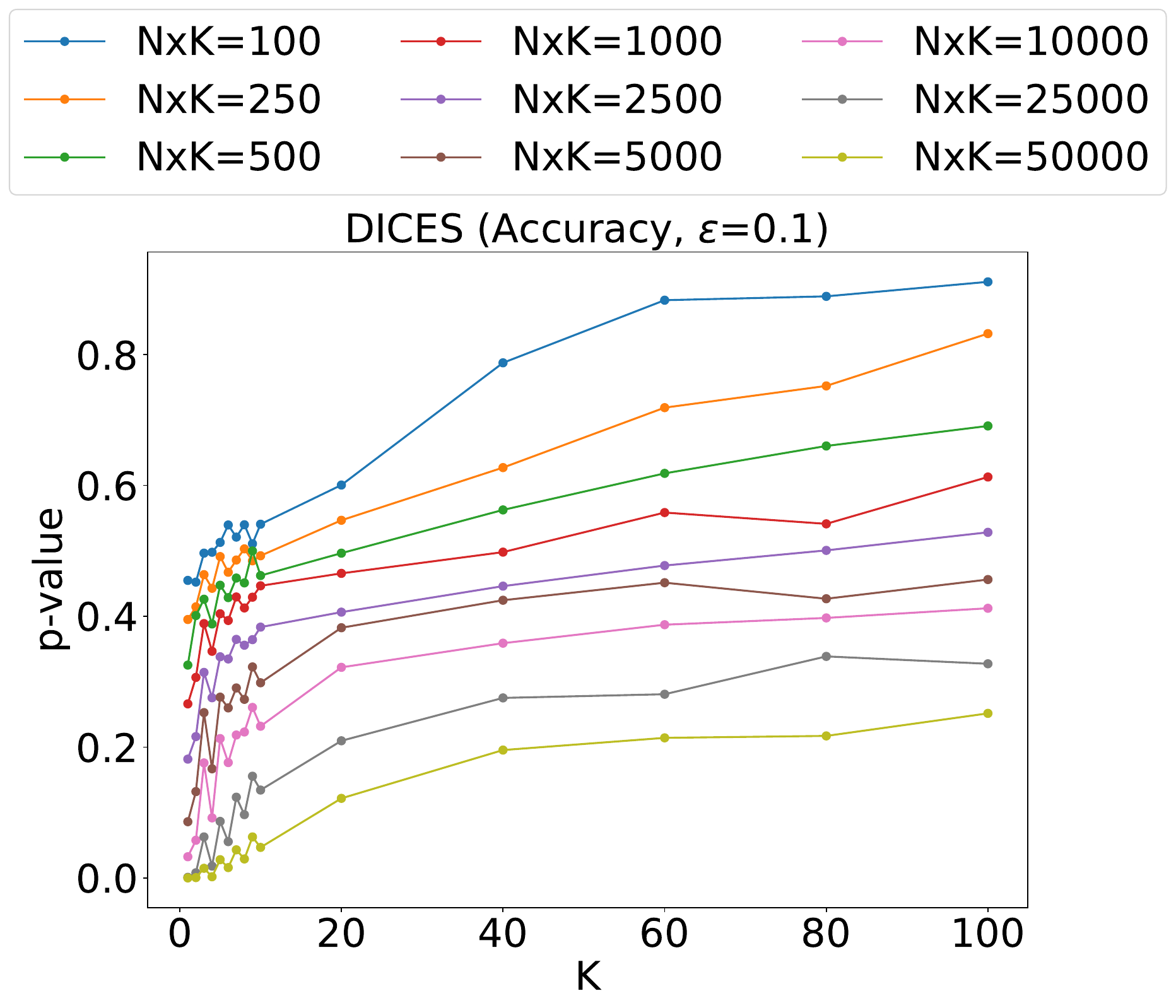}
    \caption{$\epsilon = 0.1$}
    \label{fig:dices_acc_e01}
  \end{subfigure} \hfill
  \begin{subfigure}[b]{0.24\linewidth}
    \centering
    \includegraphics[width=\linewidth]{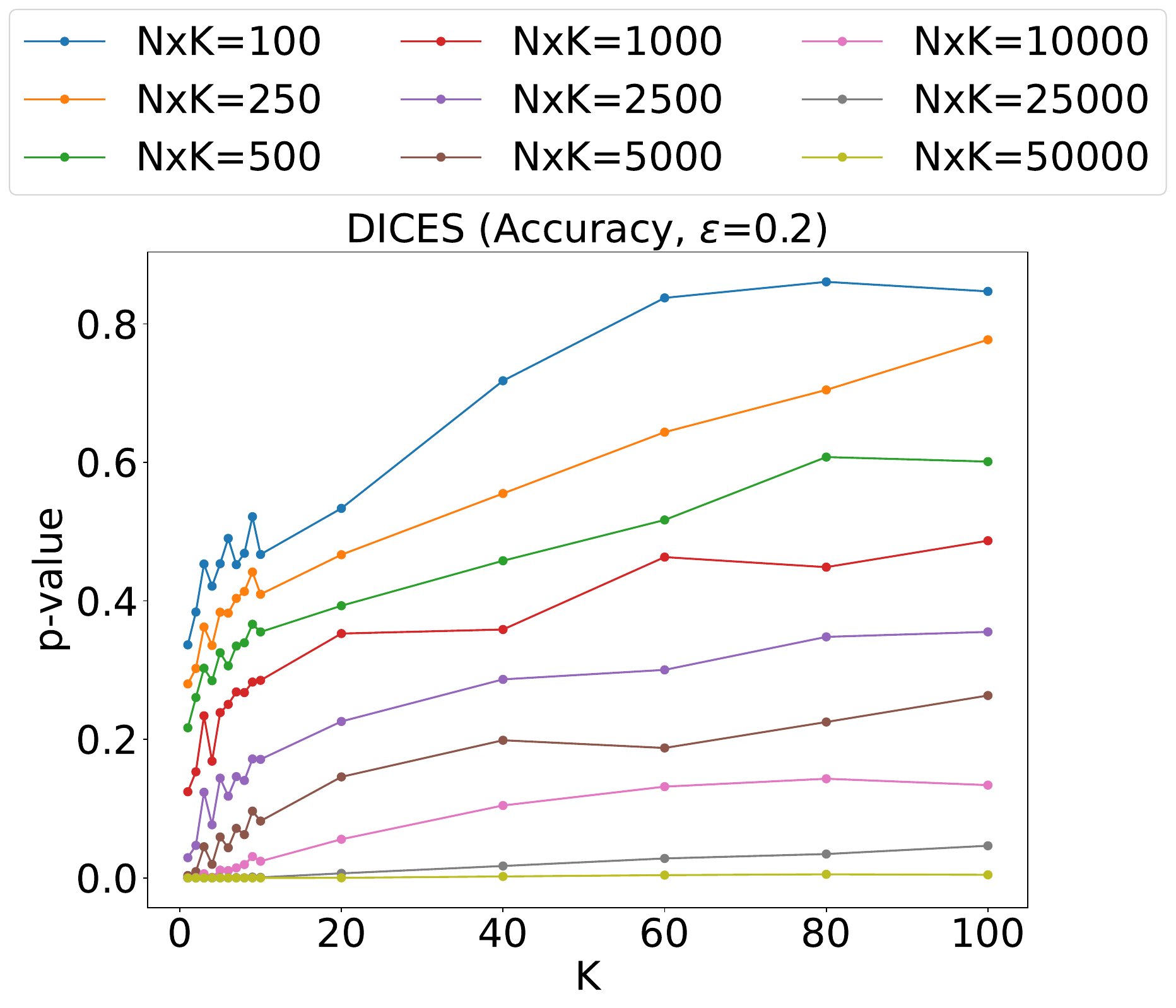}
    \caption{$\epsilon = 0.2$}
    \label{fig:dices_acc_e02}
  \end{subfigure} \hfill
  \begin{subfigure}[b]{0.24\linewidth}
    \centering
    \includegraphics[width=\linewidth]{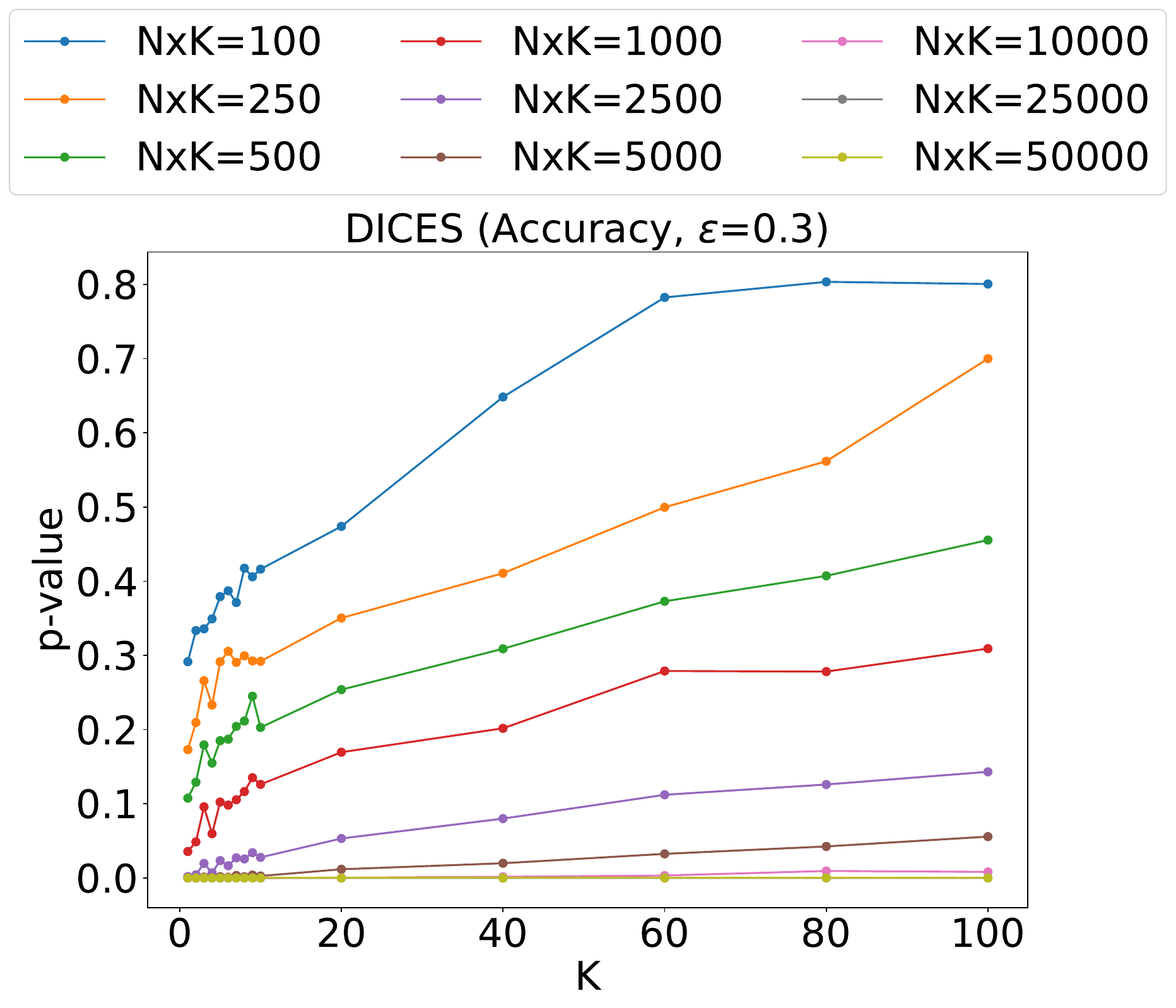}
    \caption{$\epsilon = 0.3$}
    \label{fig:dices_acc_e03}
  \end{subfigure} \hfill
  \begin{subfigure}[b]{0.24\linewidth}
    \centering
    \includegraphics[width=\linewidth]{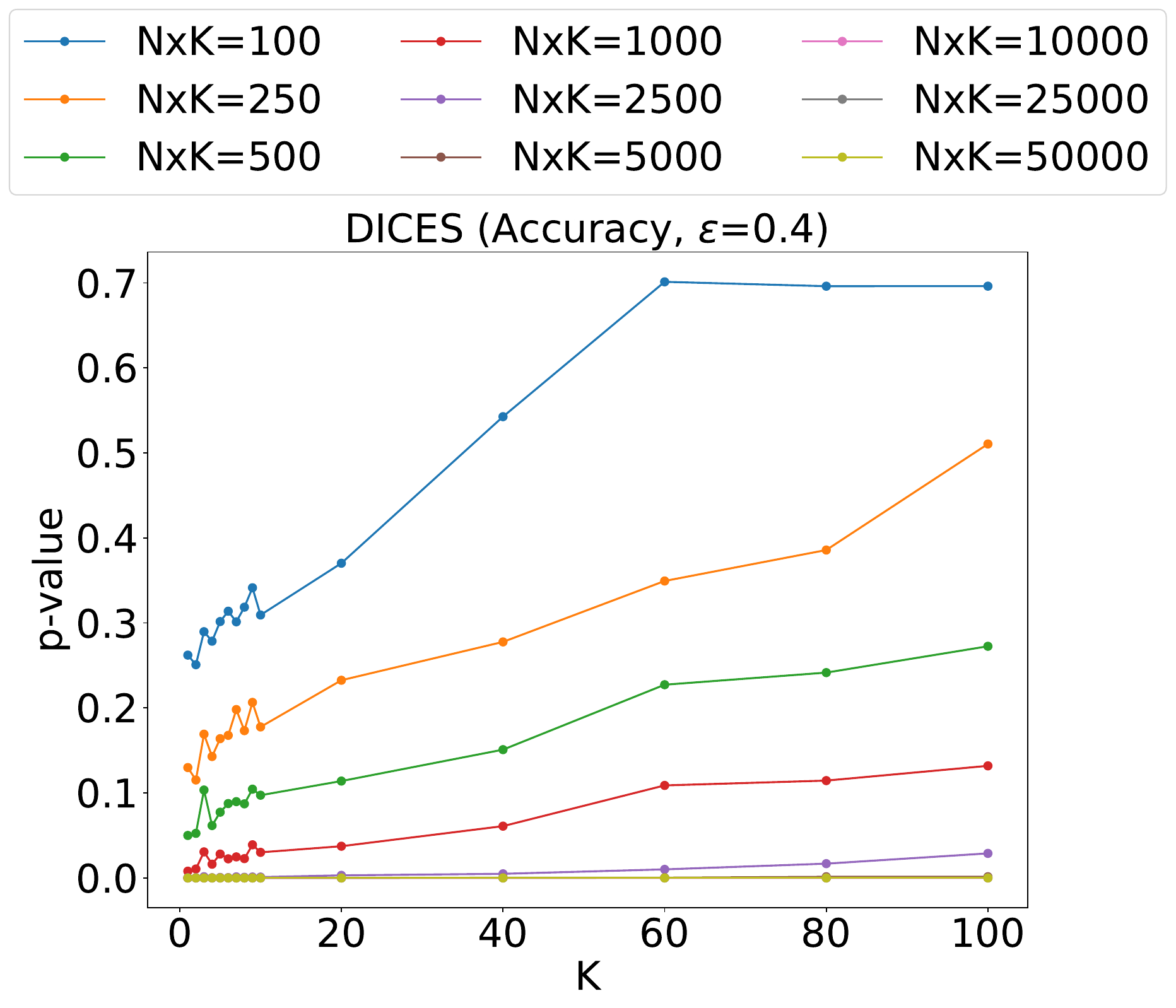}
    \caption{$\epsilon = 0.4$}
    \label{fig:dices_acc_e04}
  \end{subfigure}
  \caption{P-value plots for DICES dataset with Accuracy as the metric}
  \label{fig:dices_accuracy}
\end{figure*}

\begin{figure*}
  \centering
  \begin{subfigure}[b]{0.24\linewidth}
    \centering
    \includegraphics[width=\linewidth]{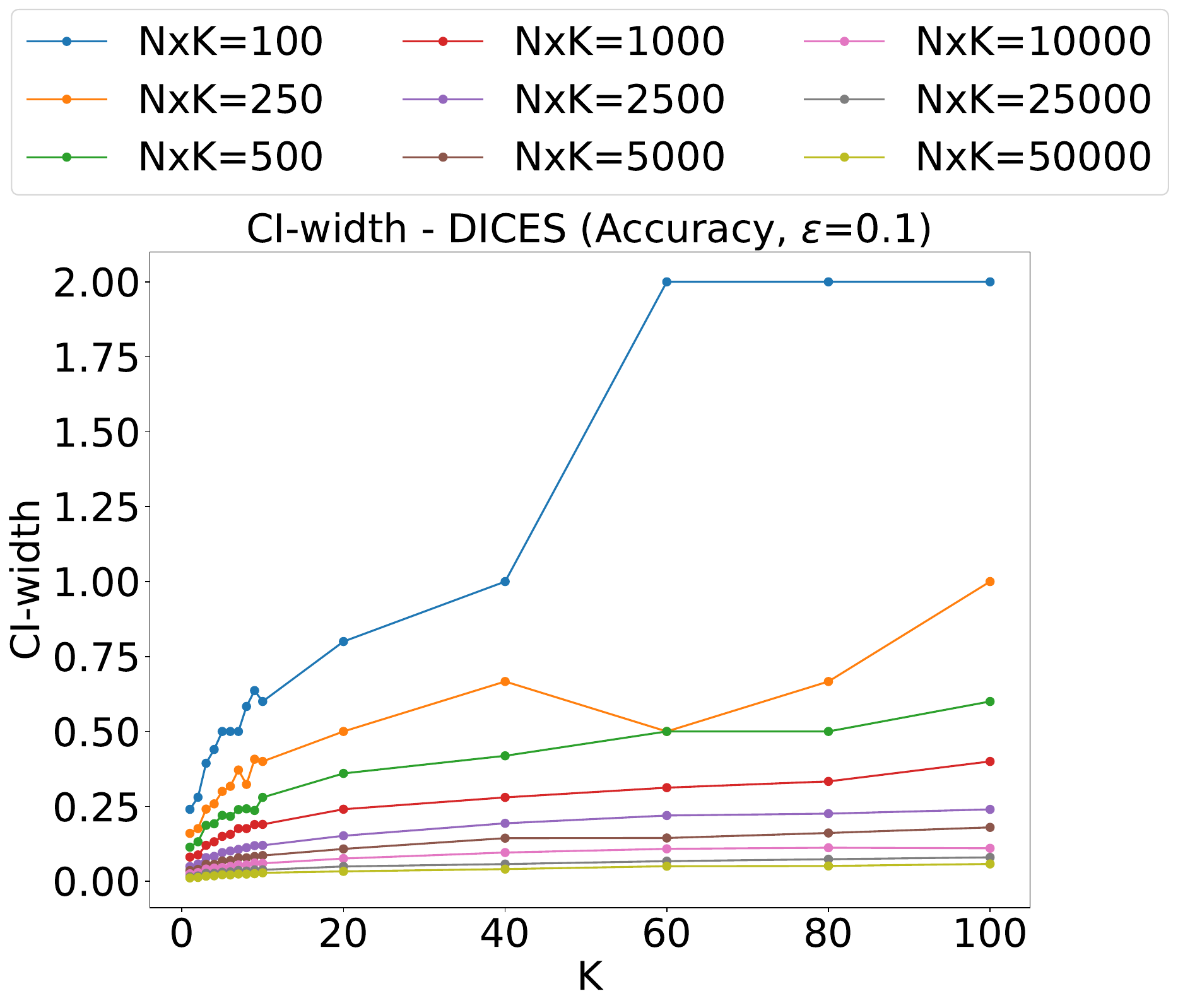}
    \caption{$\epsilon = 0.1$}
    \label{fig:dices_ci_acc_e01}
  \end{subfigure} \hfill
  \begin{subfigure}[b]{0.24\linewidth}
    \centering
    \includegraphics[width=\linewidth]{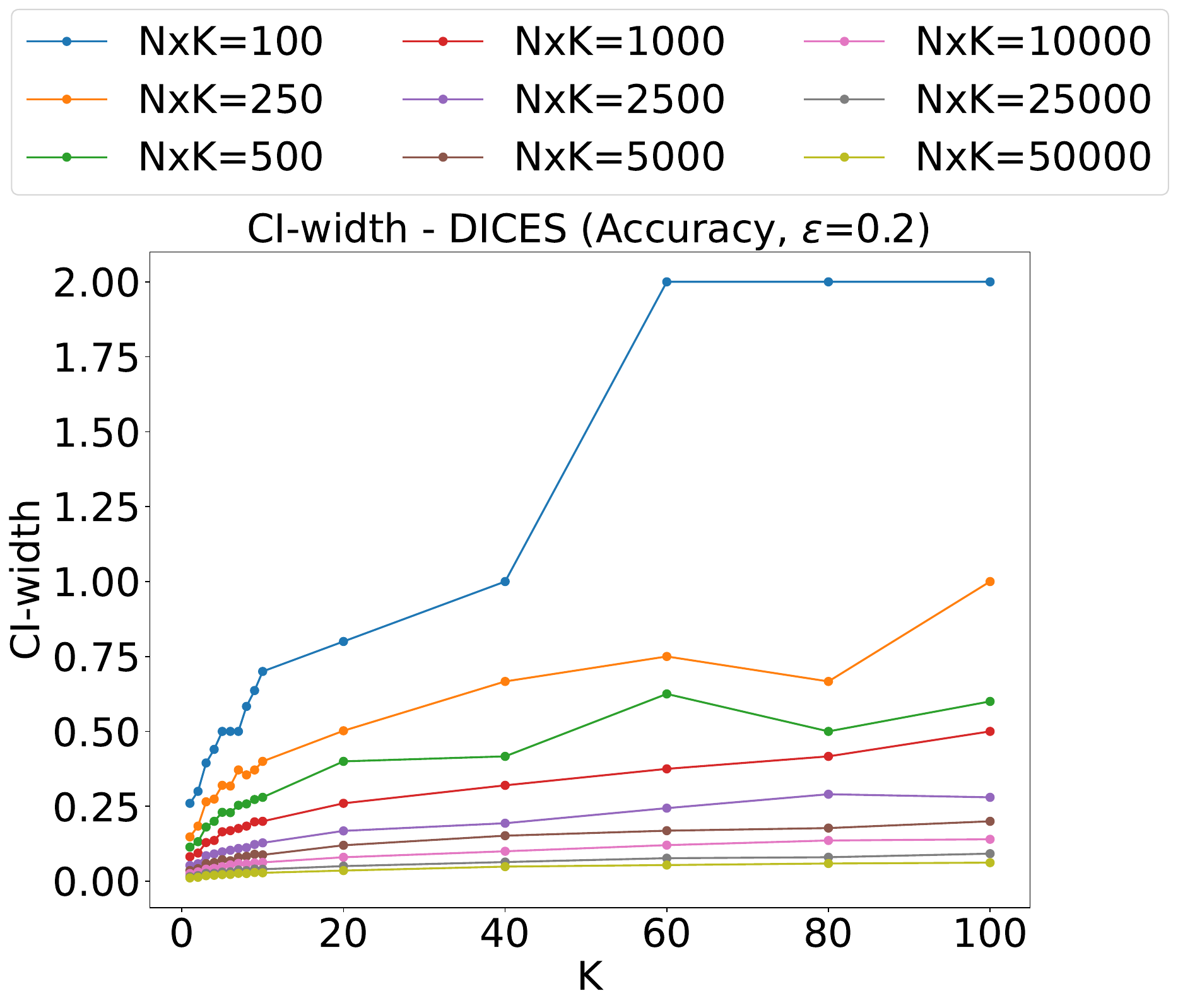}
    \caption{$\epsilon = 0.2$}
    \label{fig:dices_ci_acc_e02}
  \end{subfigure} \hfill
  \begin{subfigure}[b]{0.24\linewidth}
    \centering
    \includegraphics[width=\linewidth]{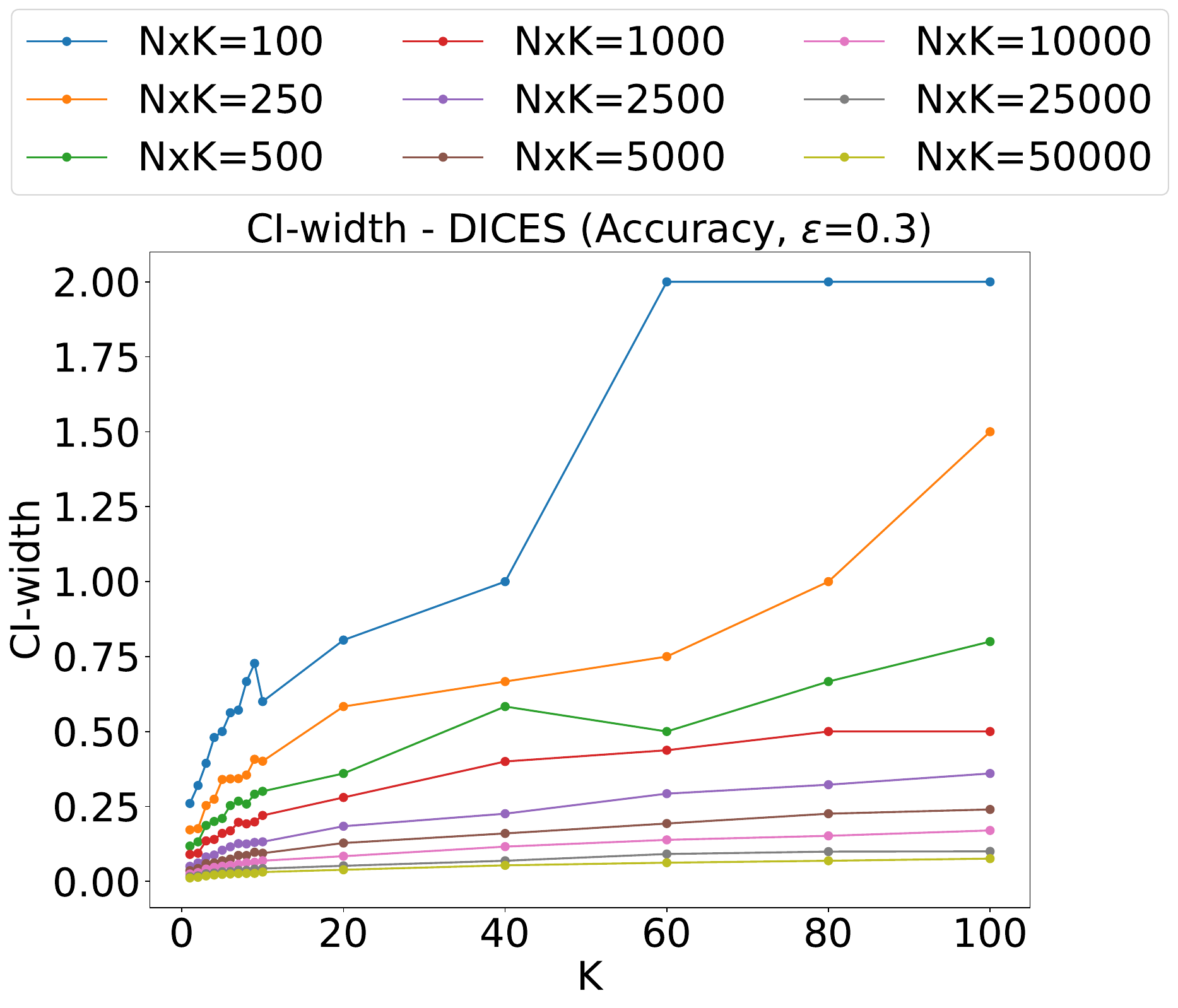}
    \caption{$\epsilon = 0.3$}
    \label{fig:dices_ci_acc_e03}
  \end{subfigure} \hfill
  \begin{subfigure}[b]{0.24\linewidth}
    \centering
    \includegraphics[width=\linewidth]{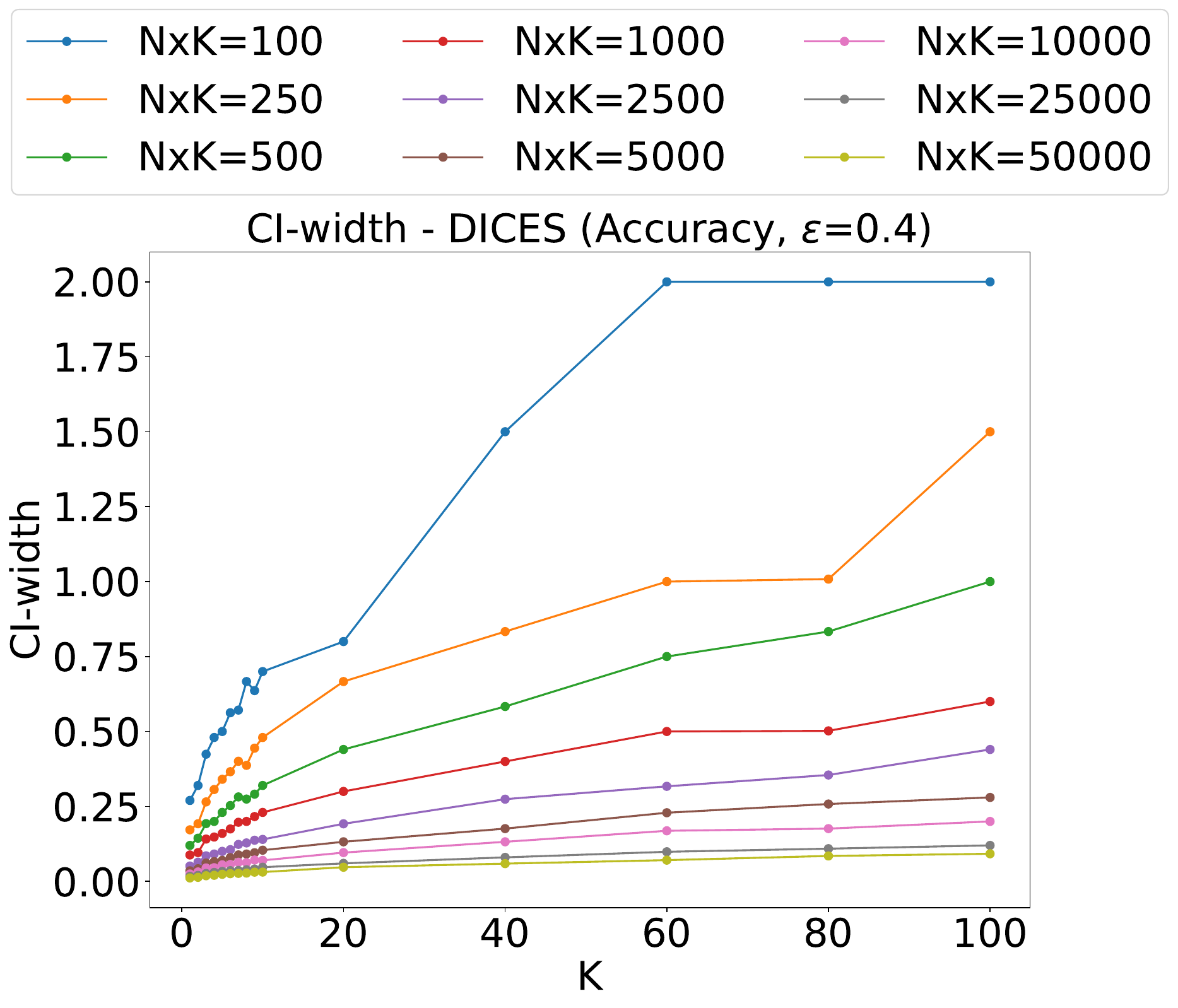}
    \caption{$\epsilon = 0.4$}
    \label{fig:dices_ci_acc_e04}
  \end{subfigure}
  \caption{CI-width plots for DICES dataset with Accuracy as the metric}
  \label{fig:dices_ci_accuracy}
\end{figure*}

\begin{figure*}
  \centering
  \begin{subfigure}[b]{0.24\linewidth}
    \centering
    \includegraphics[width=\linewidth]{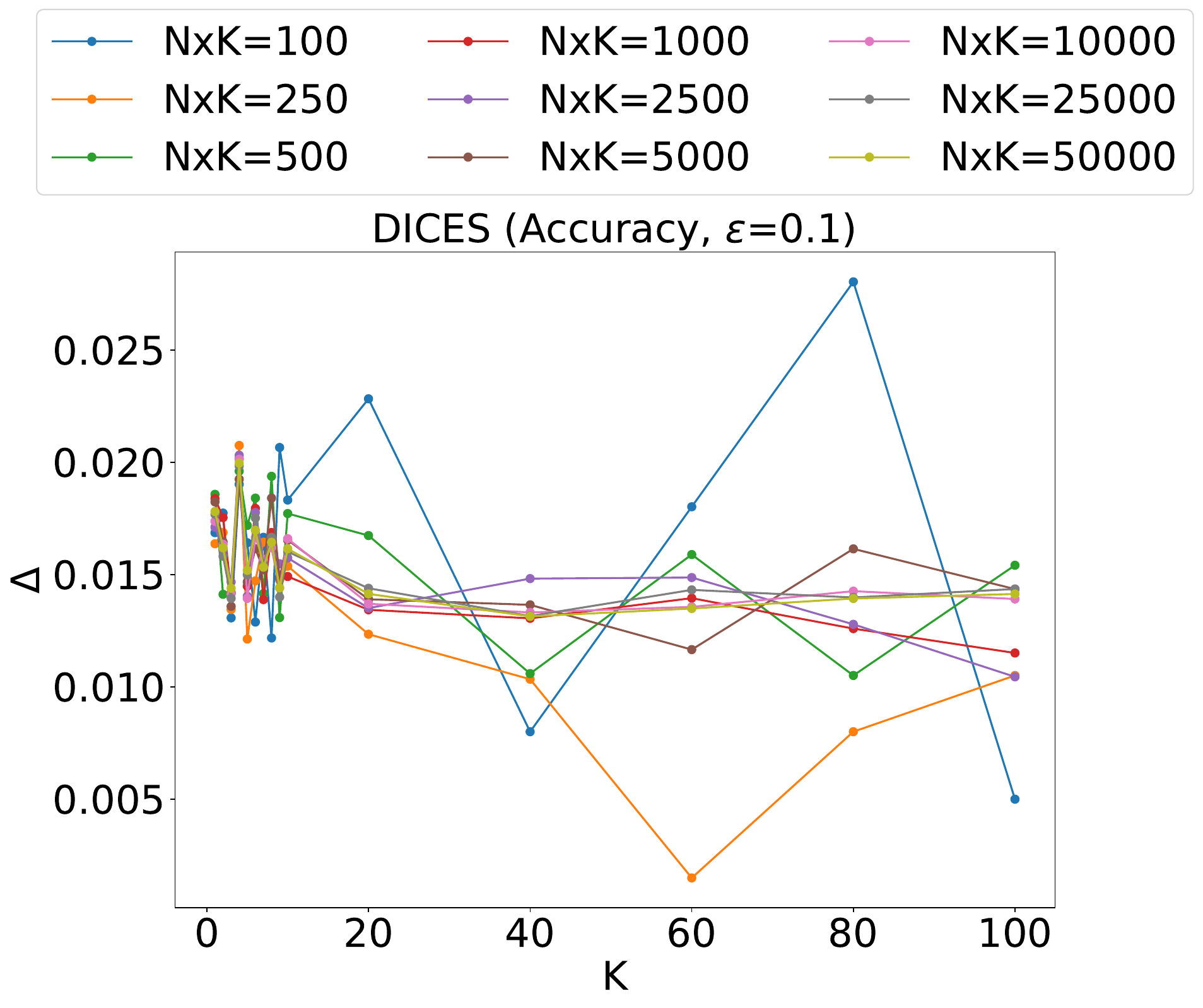}
    \caption{$\epsilon = 0.1$}
    \label{fig:dices_delta_acc_e01}
  \end{subfigure} \hfill
  \begin{subfigure}[b]{0.24\linewidth}
    \centering
    \includegraphics[width=\linewidth]{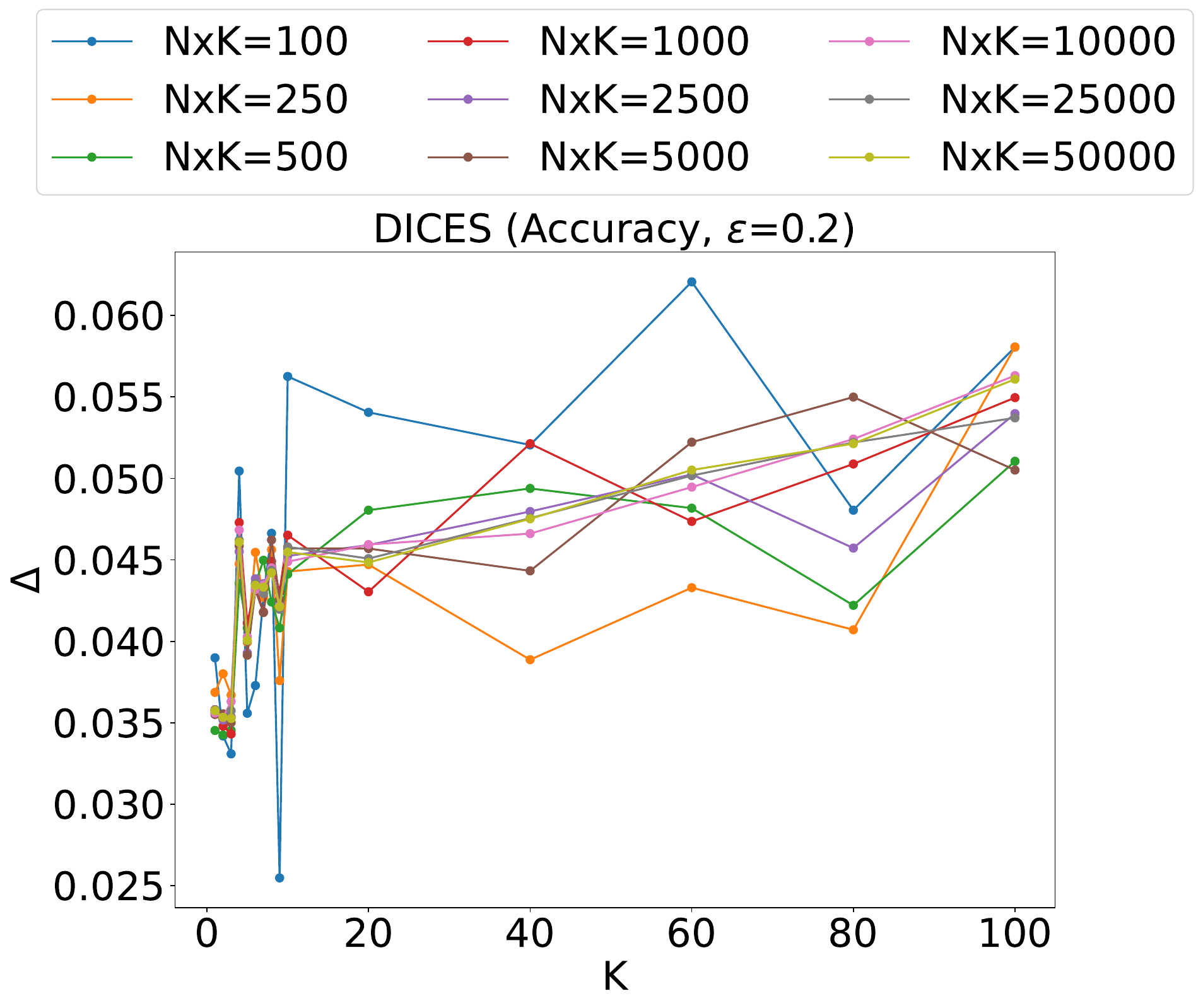}
    \caption{$\epsilon = 0.2$}
    \label{fig:dices_delta_acc_e02}
  \end{subfigure} \hfill
  \begin{subfigure}[b]{0.24\linewidth}
    \centering
    \includegraphics[width=\linewidth]{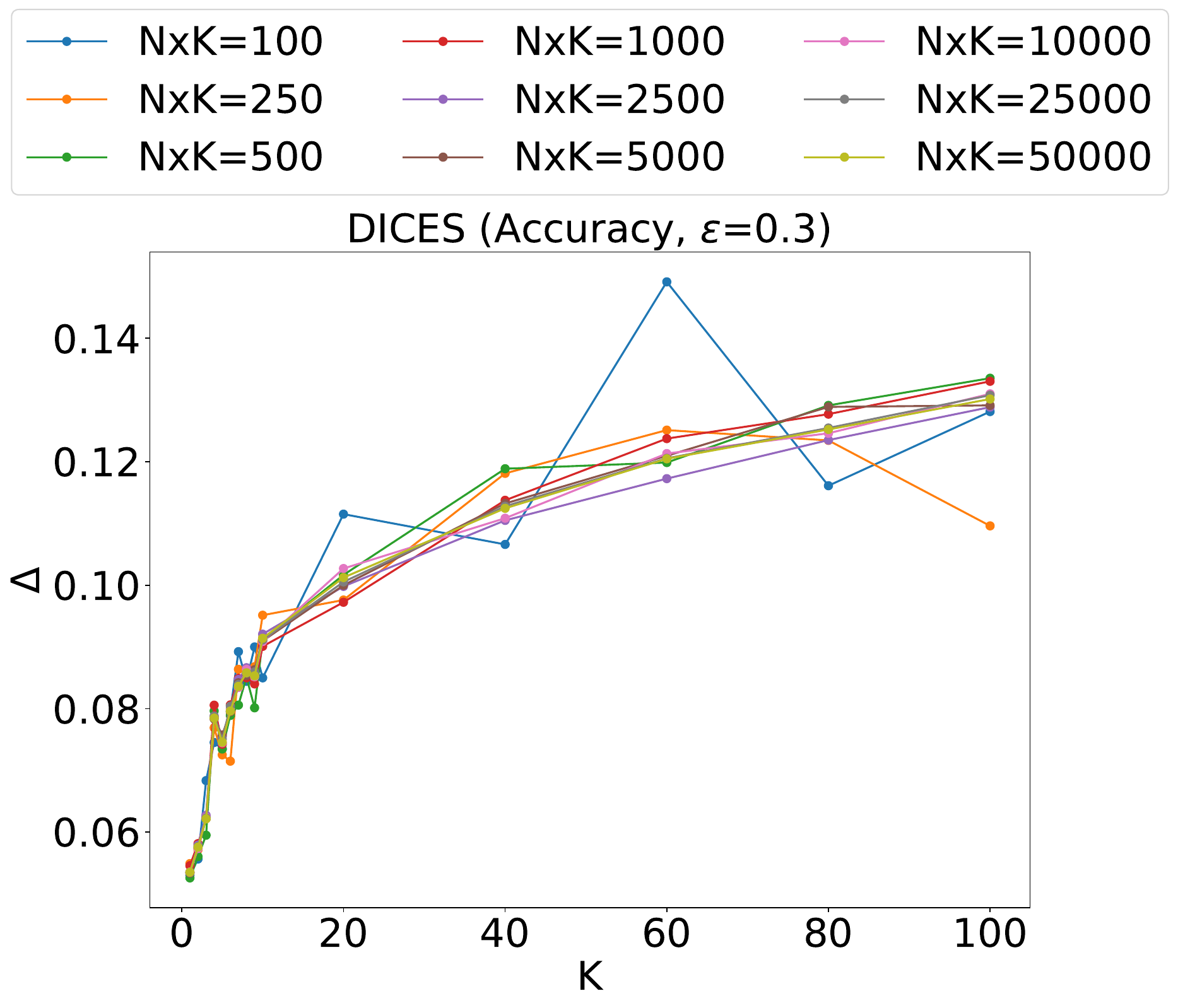}
    \caption{$\epsilon = 0.3$}
    \label{fig:dices_delta_acc_e03}
  \end{subfigure} \hfill
  \begin{subfigure}[b]{0.24\linewidth}
    \centering
    \includegraphics[width=\linewidth]{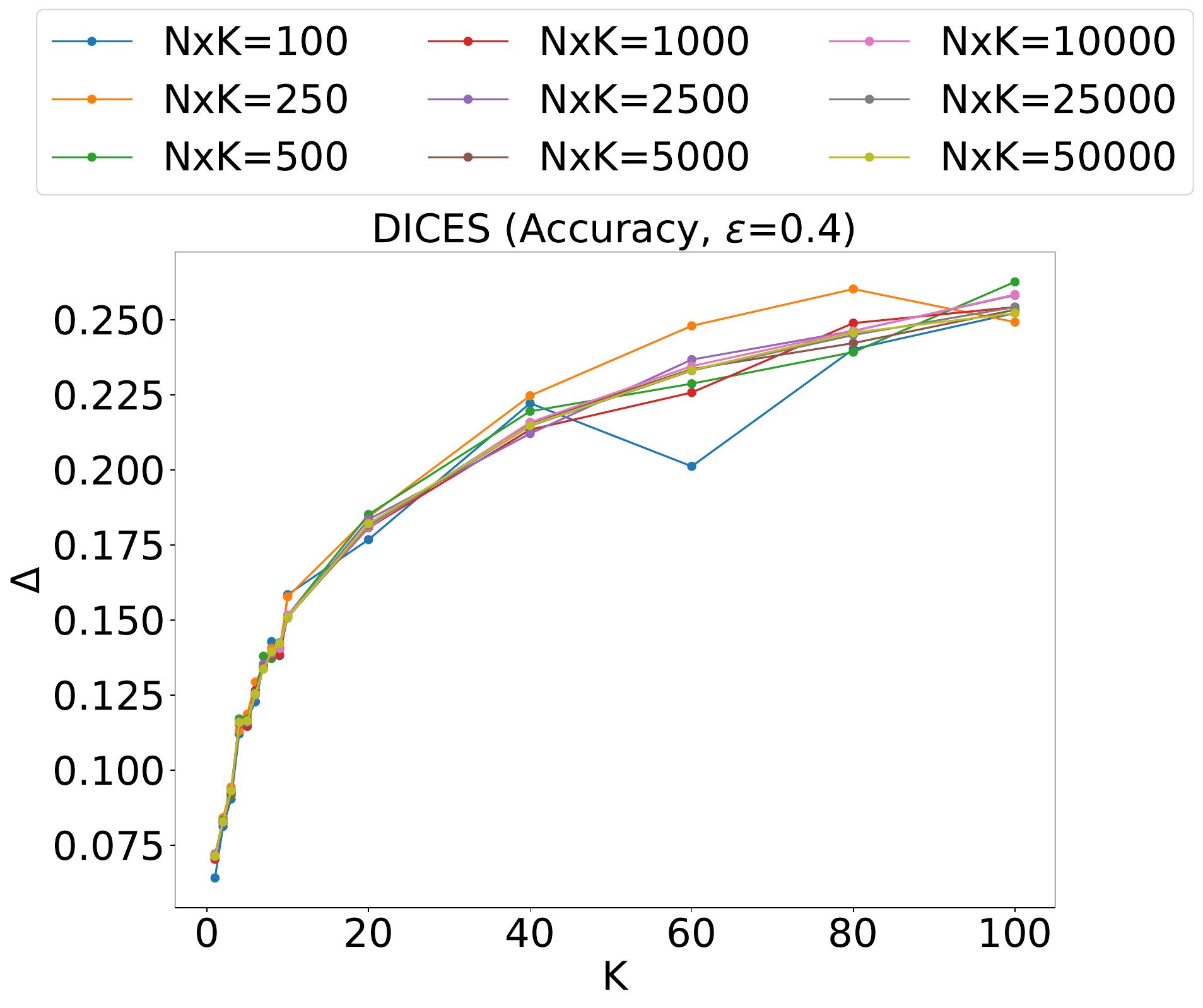}
    \caption{$\epsilon = 0.4$}
    \label{fig:dices_delta_acc_e04}
  \end{subfigure}
  \caption{Effect sizes ($\Delta$) for DICES dataset with Accuracy as the metric}
  \label{fig:dices_delta_accuracy}
\end{figure*}

\begin{figure*}
  \centering
  \begin{subfigure}[b]{0.24\linewidth}
    \centering
    \includegraphics[width=\linewidth]{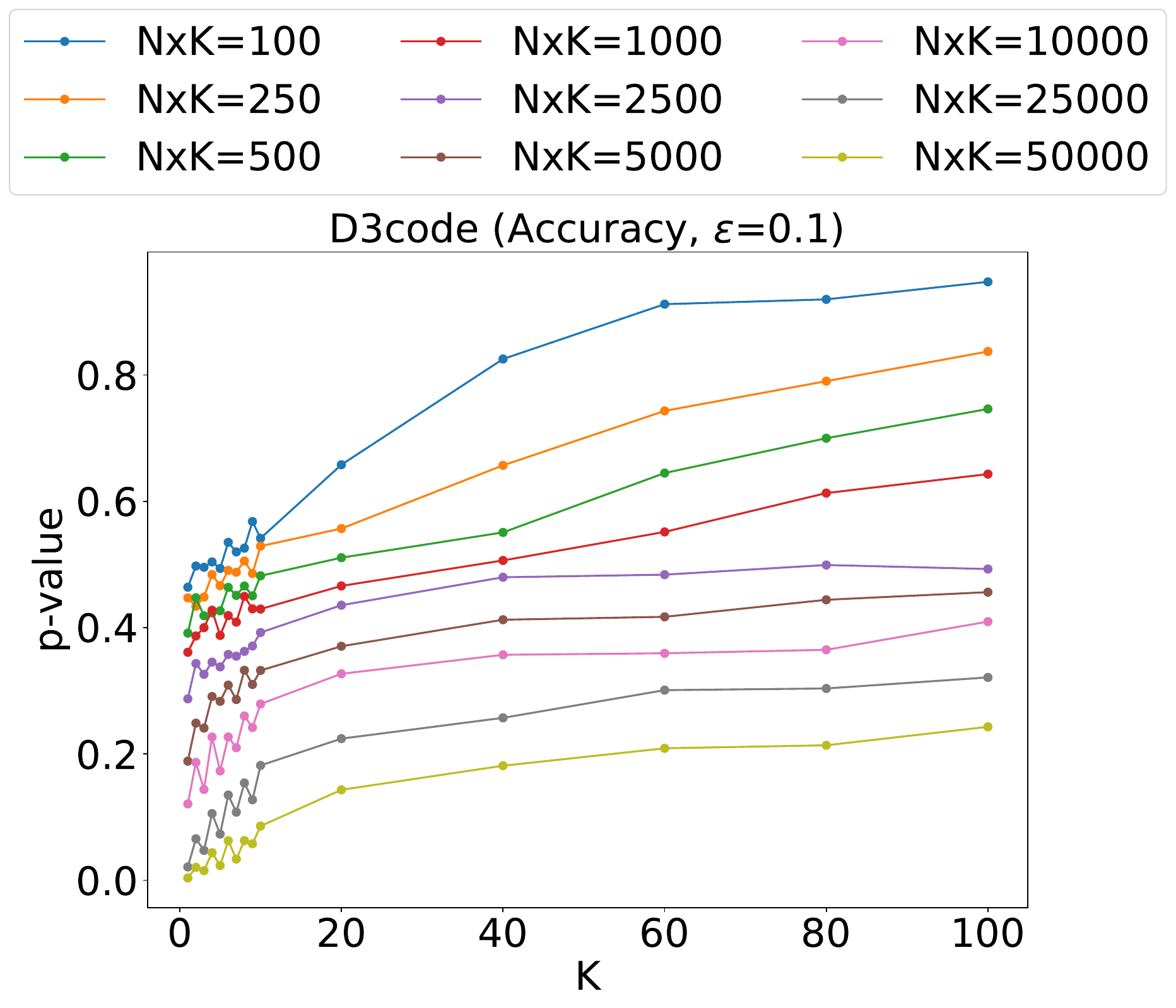}
    \caption{$\epsilon = 0.1$}
    \label{fig:d3code_acc_e01}
  \end{subfigure} \hfill
  \begin{subfigure}[b]{0.24\linewidth}
    \centering
    \includegraphics[width=\linewidth]{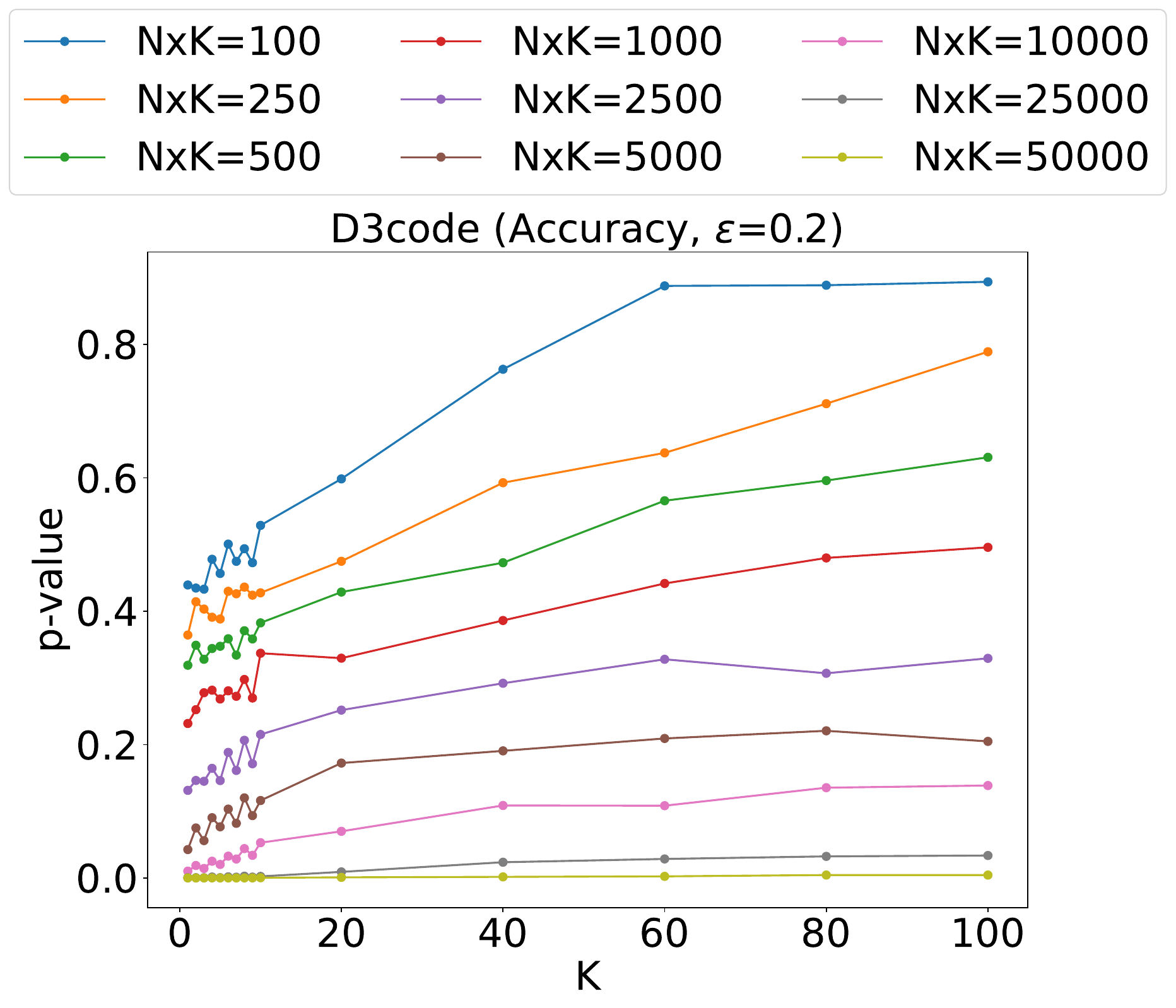}
    \caption{$\epsilon = 0.2$}
    \label{fig:d3code_acc_e02}
  \end{subfigure} \hfill
  \begin{subfigure}[b]{0.24\linewidth}
    \centering
    \includegraphics[width=\linewidth]{figures/K100/pvals_plots/D3code/D3code_p_vals_Accuracy_K_100_e_0.3.pdf}
    \caption{$\epsilon = 0.3$}
    \label{fig:d3code_acc_e03}
  \end{subfigure} \hfill
  \begin{subfigure}[b]{0.24\linewidth}
    \centering
    \includegraphics[width=\linewidth]{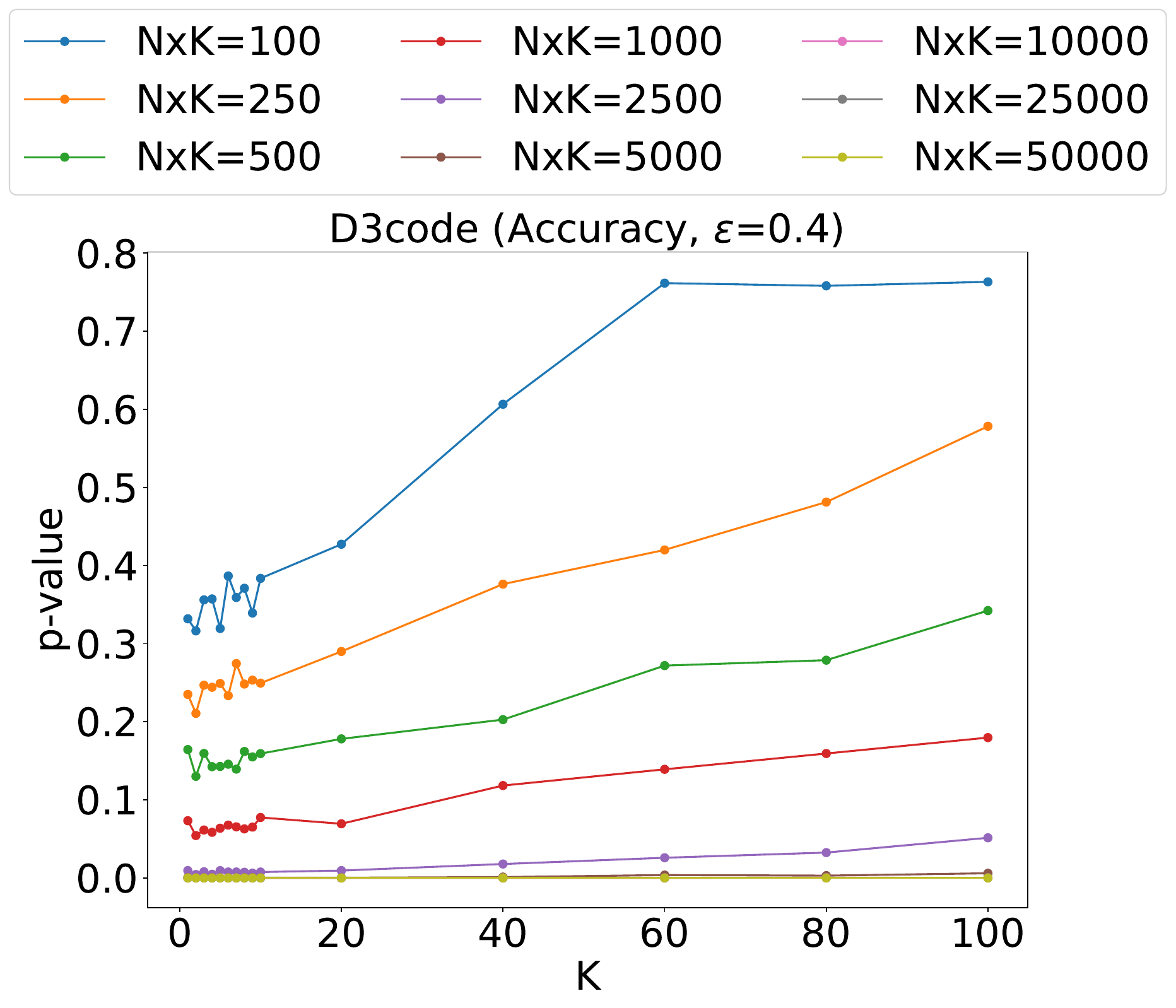}
    \caption{$\epsilon = 0.4$}
    \label{fig:d3code_acc_e04}
  \end{subfigure}
  \caption{P-value plots for D3code dataset with Accuracy as the metric}
  \label{fig:d3code_accuracy}
\end{figure*}

\begin{figure*}
  \centering
  \begin{subfigure}[b]{0.24\linewidth}
    \centering
    \includegraphics[width=\linewidth]{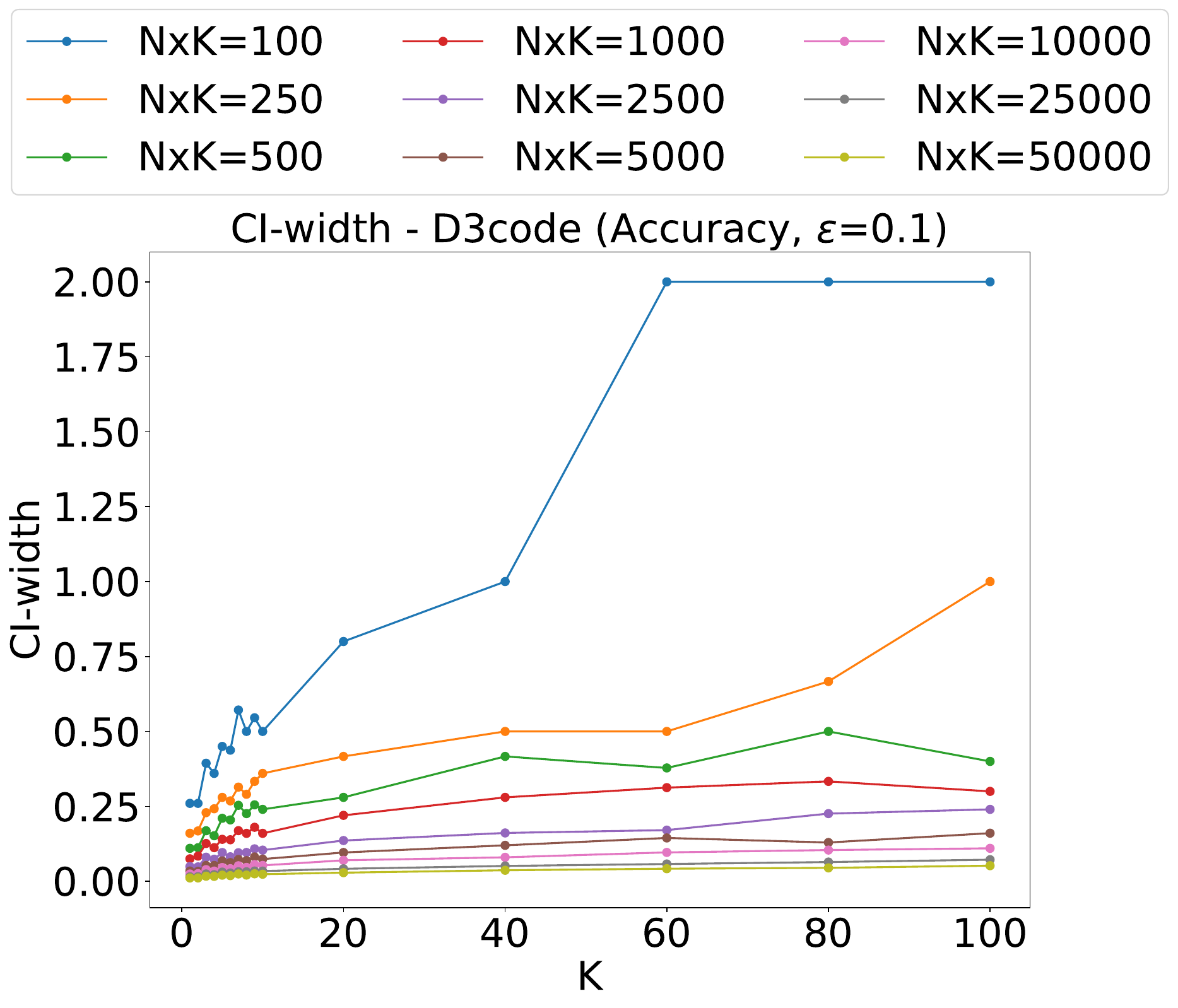}
    \caption{$\epsilon = 0.1$}
    \label{fig:d3code_ci_acc_e01}
  \end{subfigure} \hfill
  \begin{subfigure}[b]{0.24\linewidth}
    \centering
    \includegraphics[width=\linewidth]{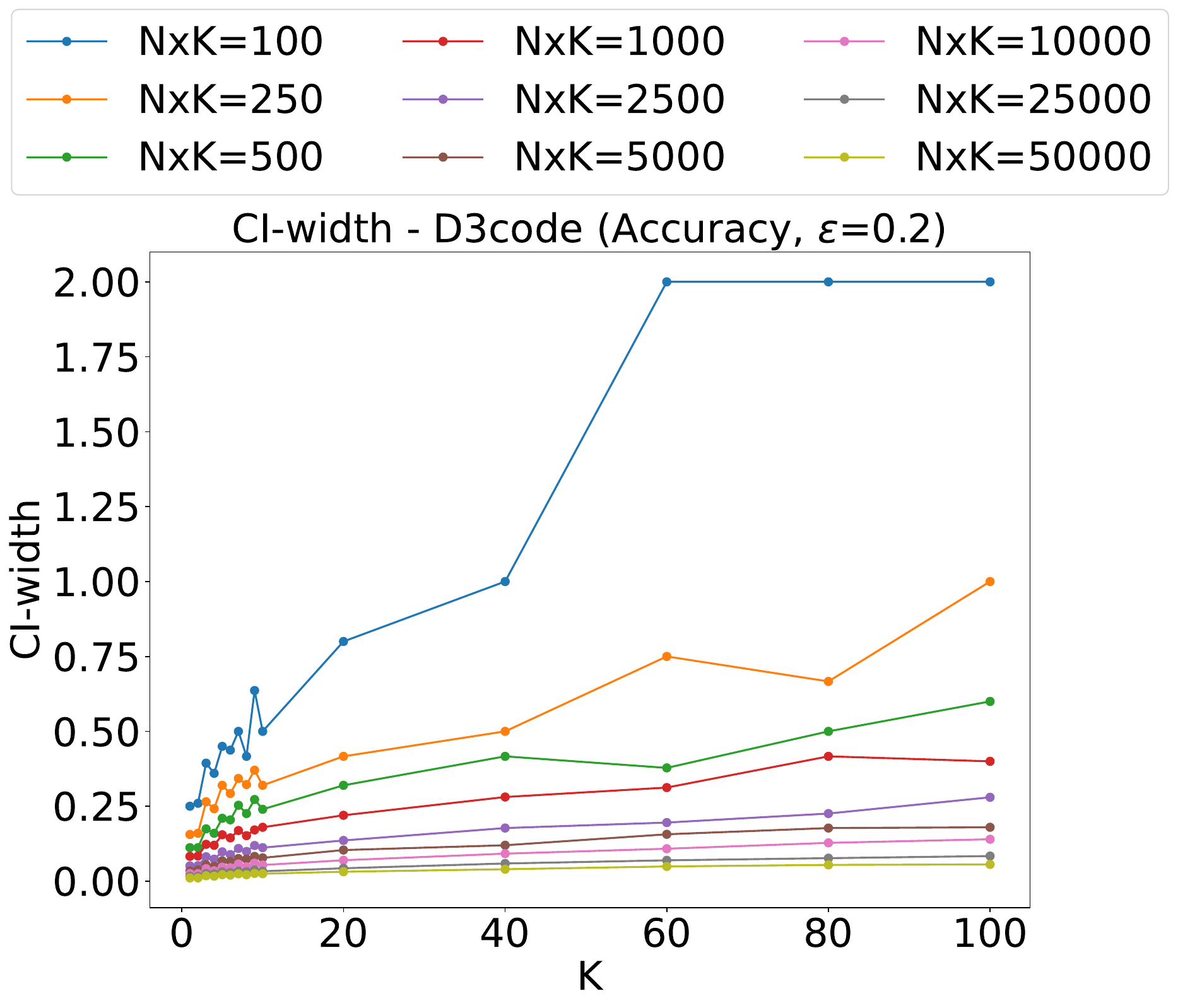}
    \caption{$\epsilon = 0.2$}
    \label{fig:d3code_ci_acc_e02}
  \end{subfigure} \hfill
  \begin{subfigure}[b]{0.24\linewidth}
    \centering
    \includegraphics[width=\linewidth]{figures/K100/ci_plots/D3code/D3code_CI_width_Accuracy_K_100_e_0.3.pdf}
    \caption{$\epsilon = 0.3$}
    \label{fig:d3code_ci_acc_e03}
  \end{subfigure} \hfill
  \begin{subfigure}[b]{0.24\linewidth}
    \centering
    \includegraphics[width=\linewidth]{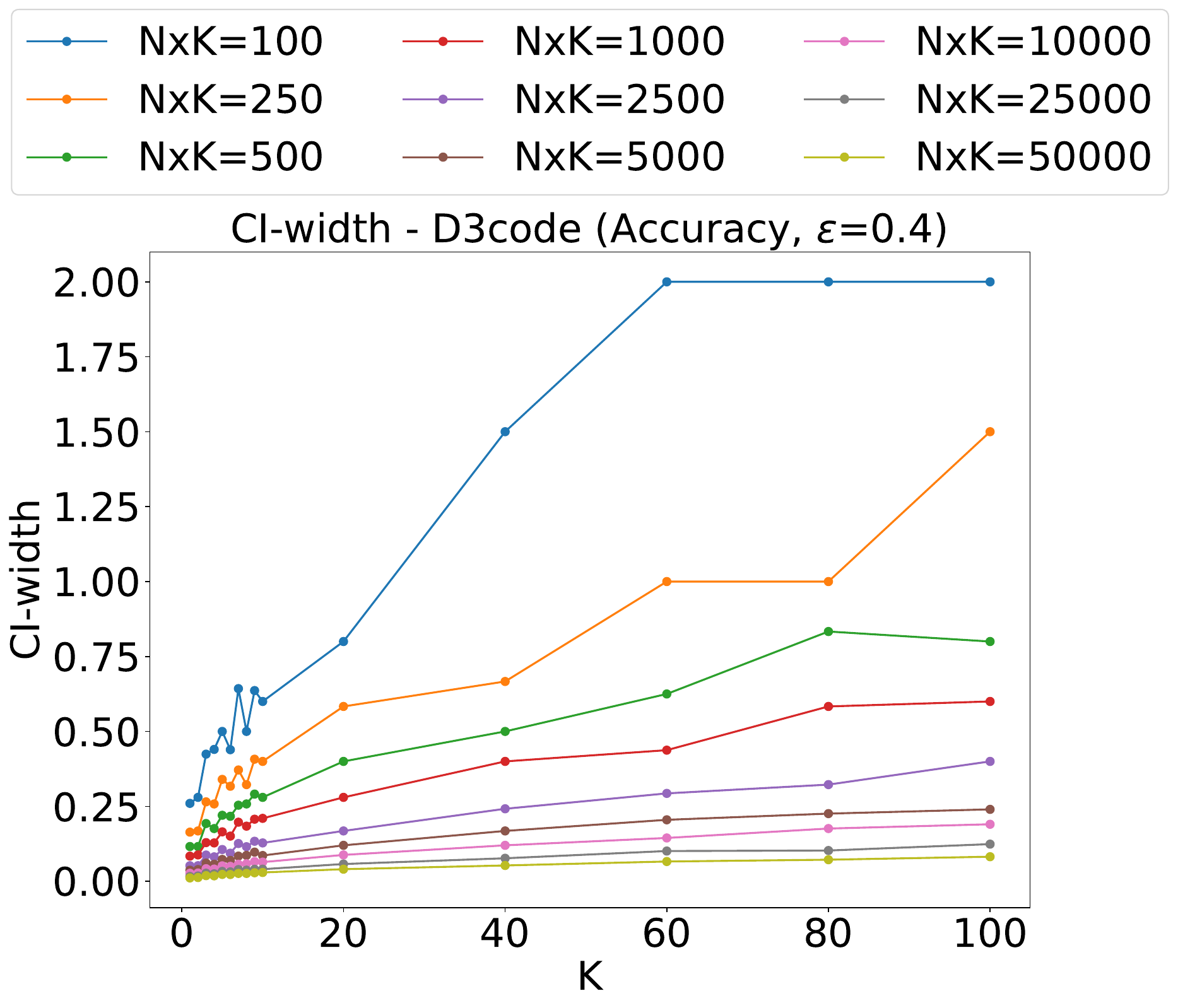}
    \caption{$\epsilon = 0.4$}
    \label{fig:d3code_ci_acc_e04}
  \end{subfigure}
  \caption{CI-width plots for D3code dataset with Accuracy as the metric}
  \label{fig:d3code_ci_accuracy}
\end{figure*}

\begin{figure*}
  \centering
  \begin{subfigure}[b]{0.24\linewidth}
    \centering
    \includegraphics[width=\linewidth]{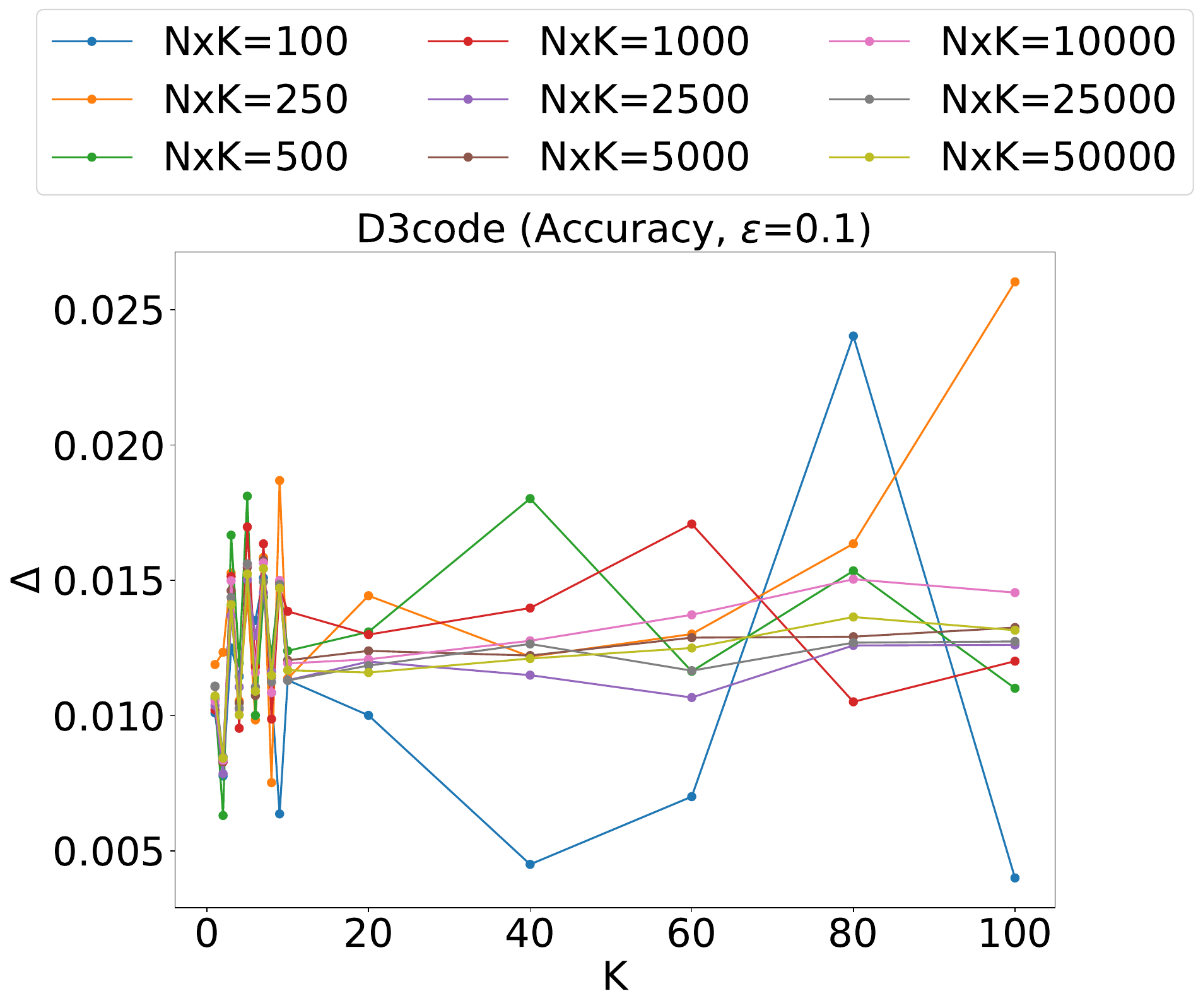}
    \caption{$\epsilon = 0.1$}
    \label{fig:d3code_delta_acc_e01}
  \end{subfigure} \hfill
  \begin{subfigure}[b]{0.24\linewidth}
    \centering
    \includegraphics[width=\linewidth]{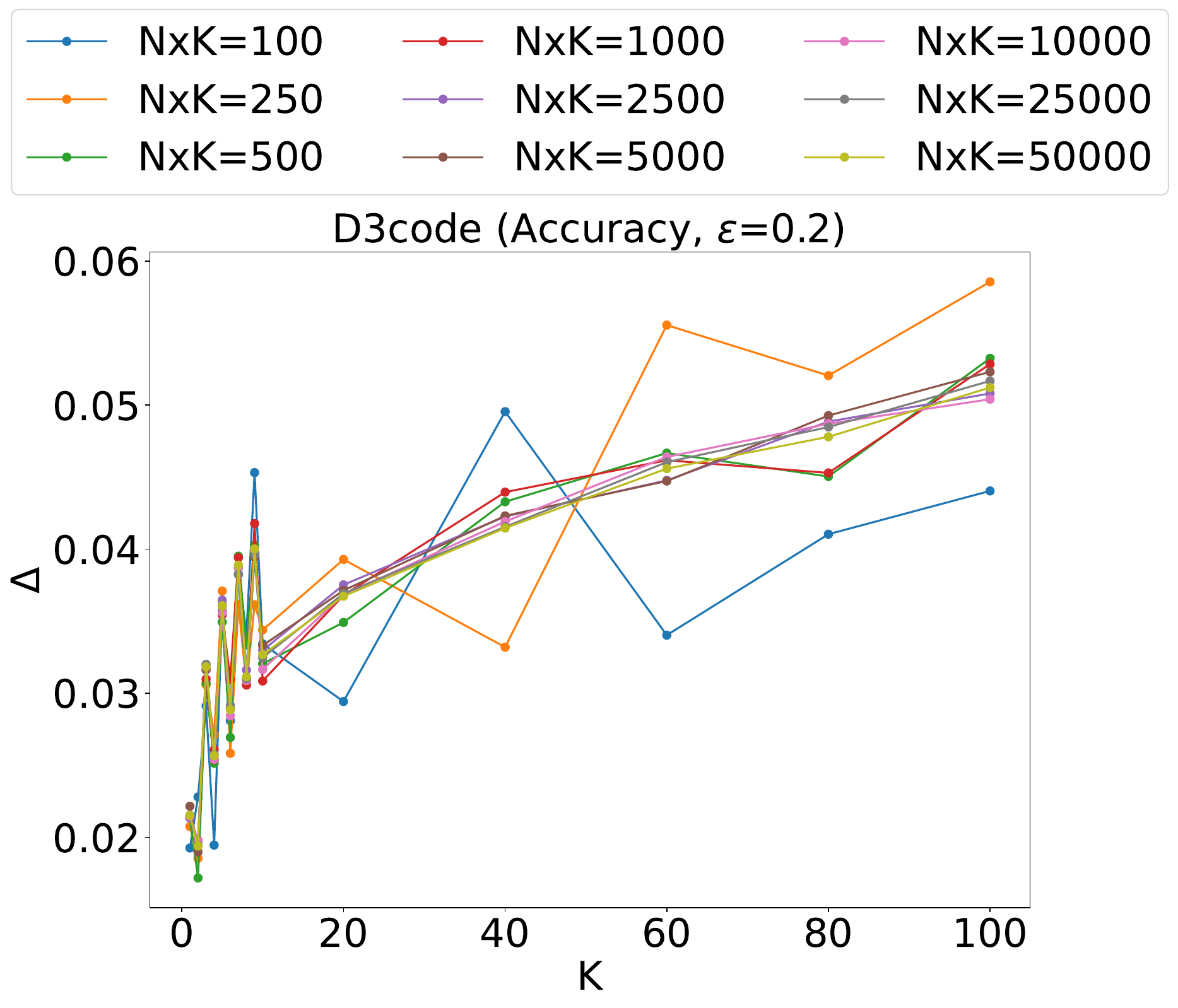}
    \caption{$\epsilon = 0.2$}
    \label{fig:d3code_delta_acc_e02}
  \end{subfigure} \hfill
  \begin{subfigure}[b]{0.24\linewidth}
    \centering
    \includegraphics[width=\linewidth]{figures/K100/delta_plots/D3code/D3code_delta_Accuracy_K_100_e_0.3.pdf}
    \caption{$\epsilon = 0.3$}
    \label{fig:d3code_delta_acc_e03}
  \end{subfigure} \hfill
  \begin{subfigure}[b]{0.24\linewidth}
    \centering
    \includegraphics[width=\linewidth]{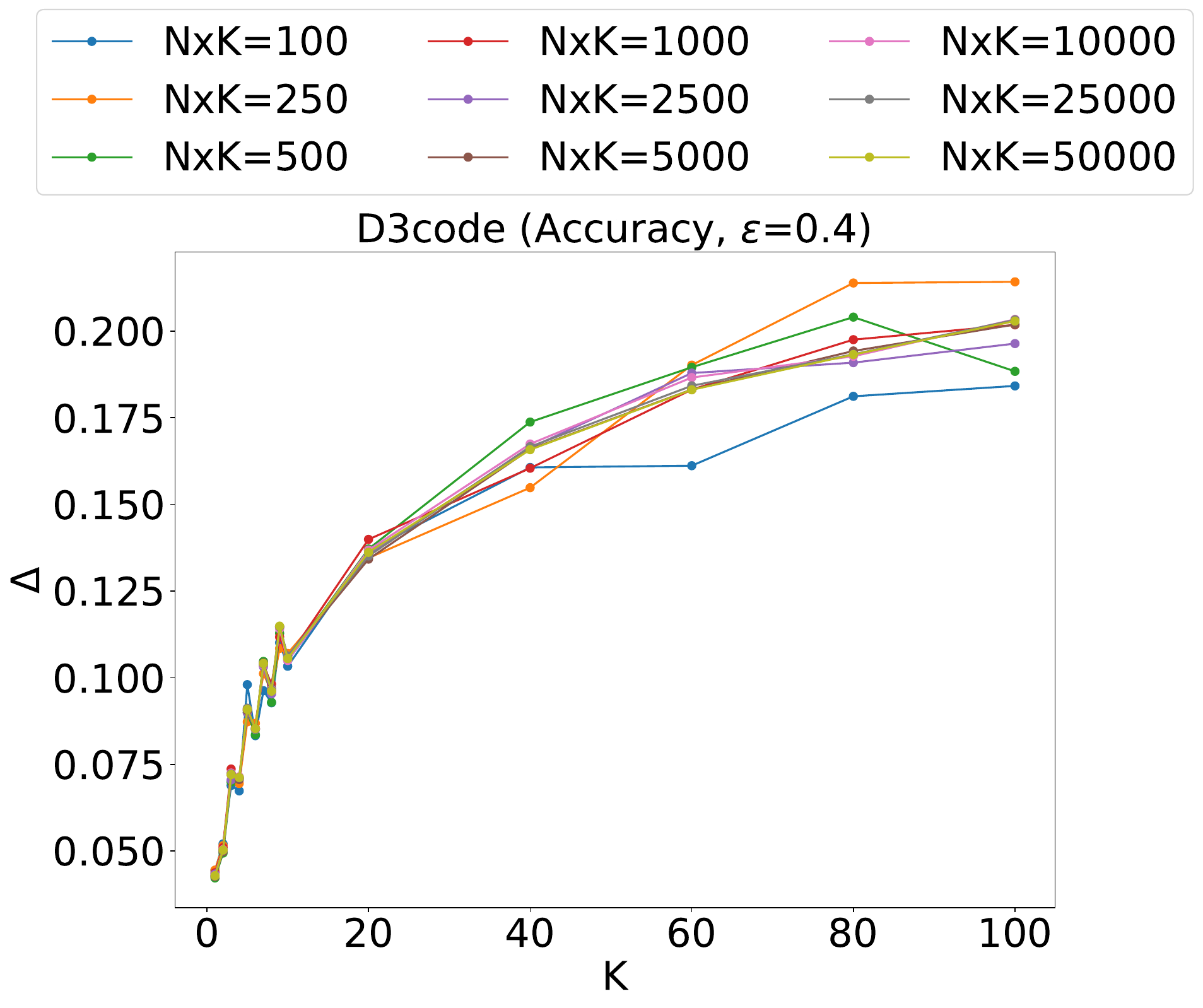}
    \caption{$\epsilon = 0.4$}
    \label{fig:d3code_delta_acc_e04}
  \end{subfigure}
  \caption{Effect sizes ($\Delta$) for D3code dataset with Accuracy as the metric}
  \label{fig:d3code_delta_accuracy}
\end{figure*}

\begin{figure*}
  \centering
  \begin{subfigure}[b]{0.24\linewidth}
    \centering
    \includegraphics[width=\linewidth]{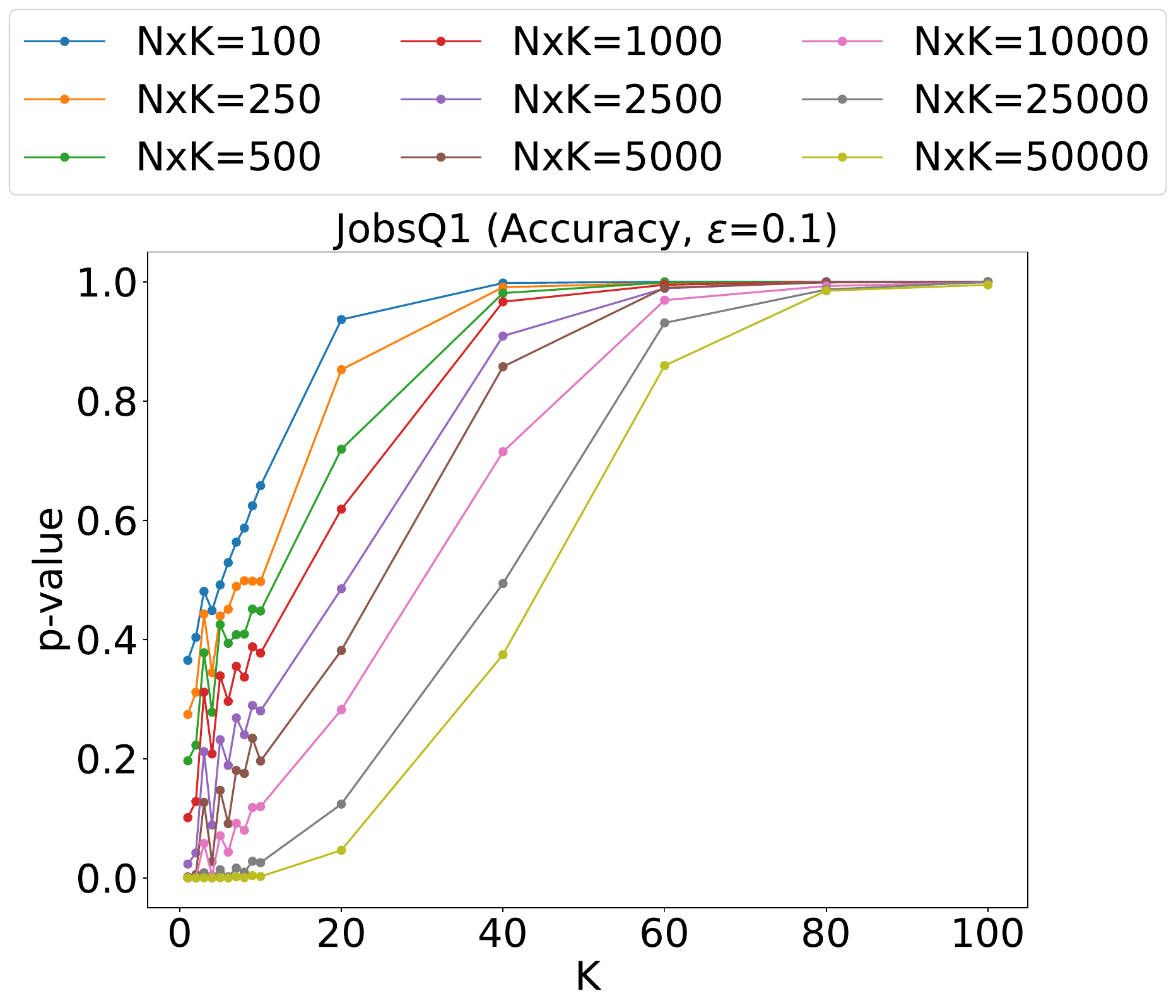}
    \caption{$\epsilon = 0.1$}
    \label{fig:jobsQ1_acc_e01}
  \end{subfigure} \hfill
  \begin{subfigure}[b]{0.24\linewidth}
    \centering
    \includegraphics[width=\linewidth]{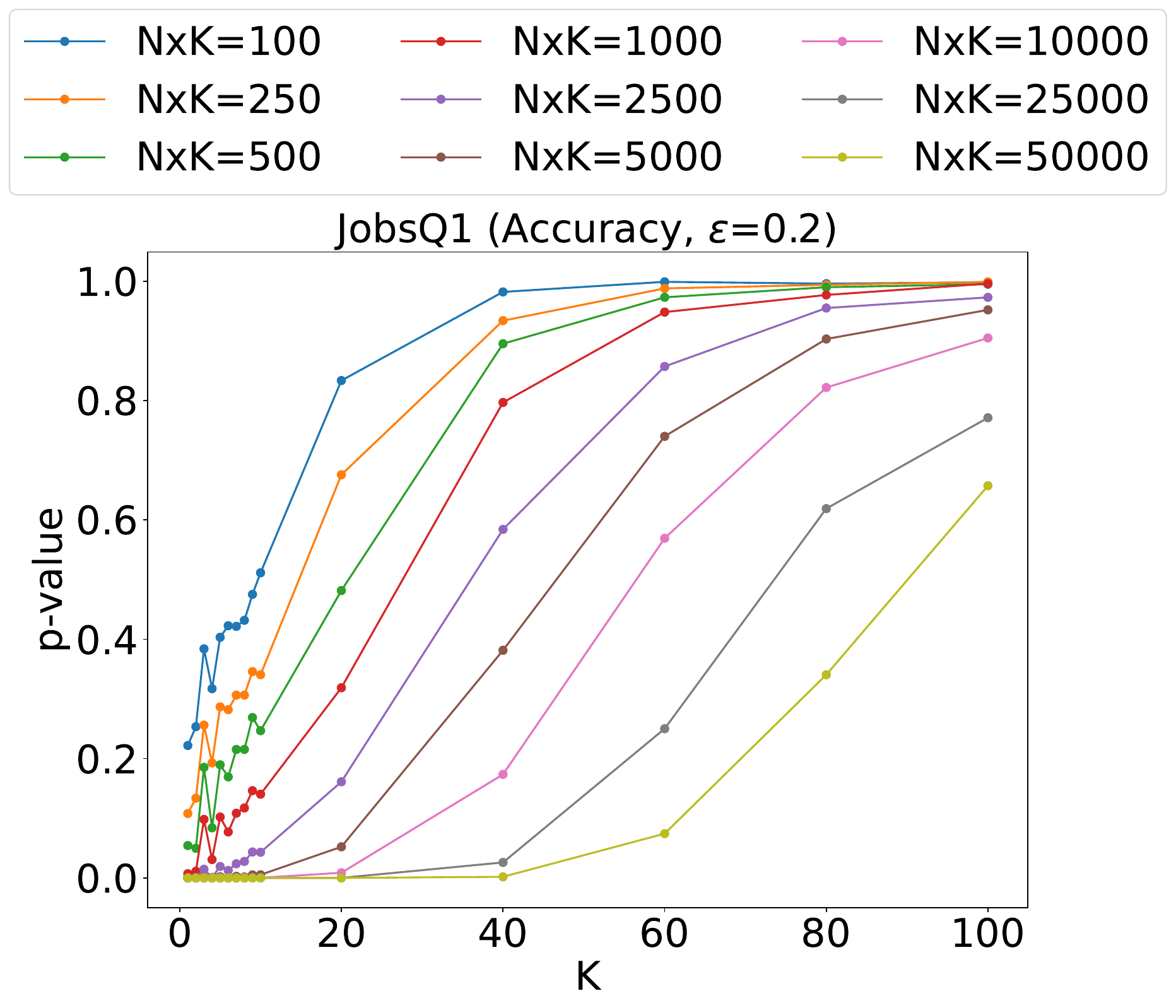}
    \caption{$\epsilon = 0.2$}
    \label{fig:jobsQ1_acc_e02}
  \end{subfigure} \hfill
  \begin{subfigure}[b]{0.24\linewidth}
    \centering
    \includegraphics[width=\linewidth]{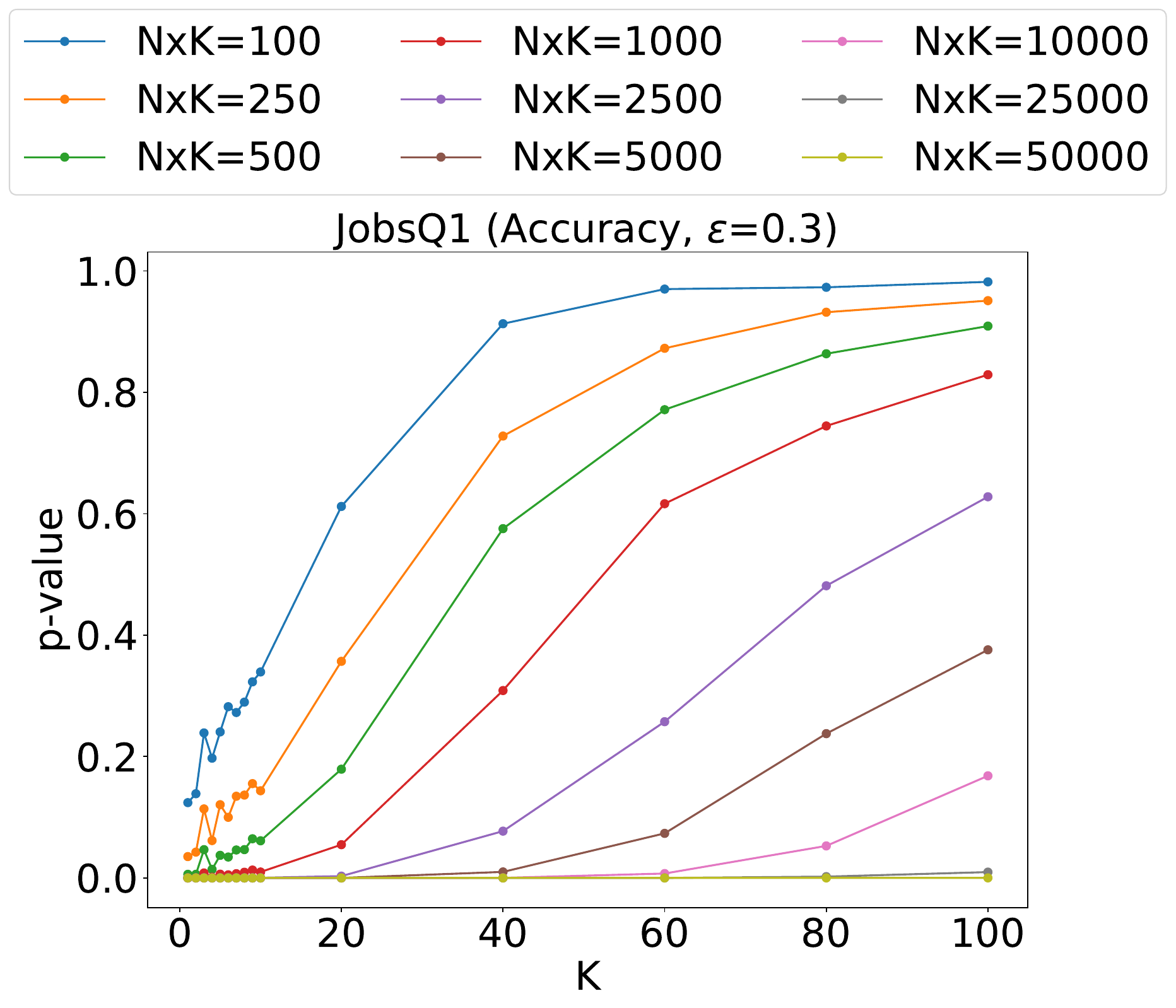}
    \caption{$\epsilon = 0.3$}
    \label{fig:jobsQ1_acc_e03}
  \end{subfigure} \hfill
  \begin{subfigure}[b]{0.24\linewidth}
    \centering
    \includegraphics[width=\linewidth]{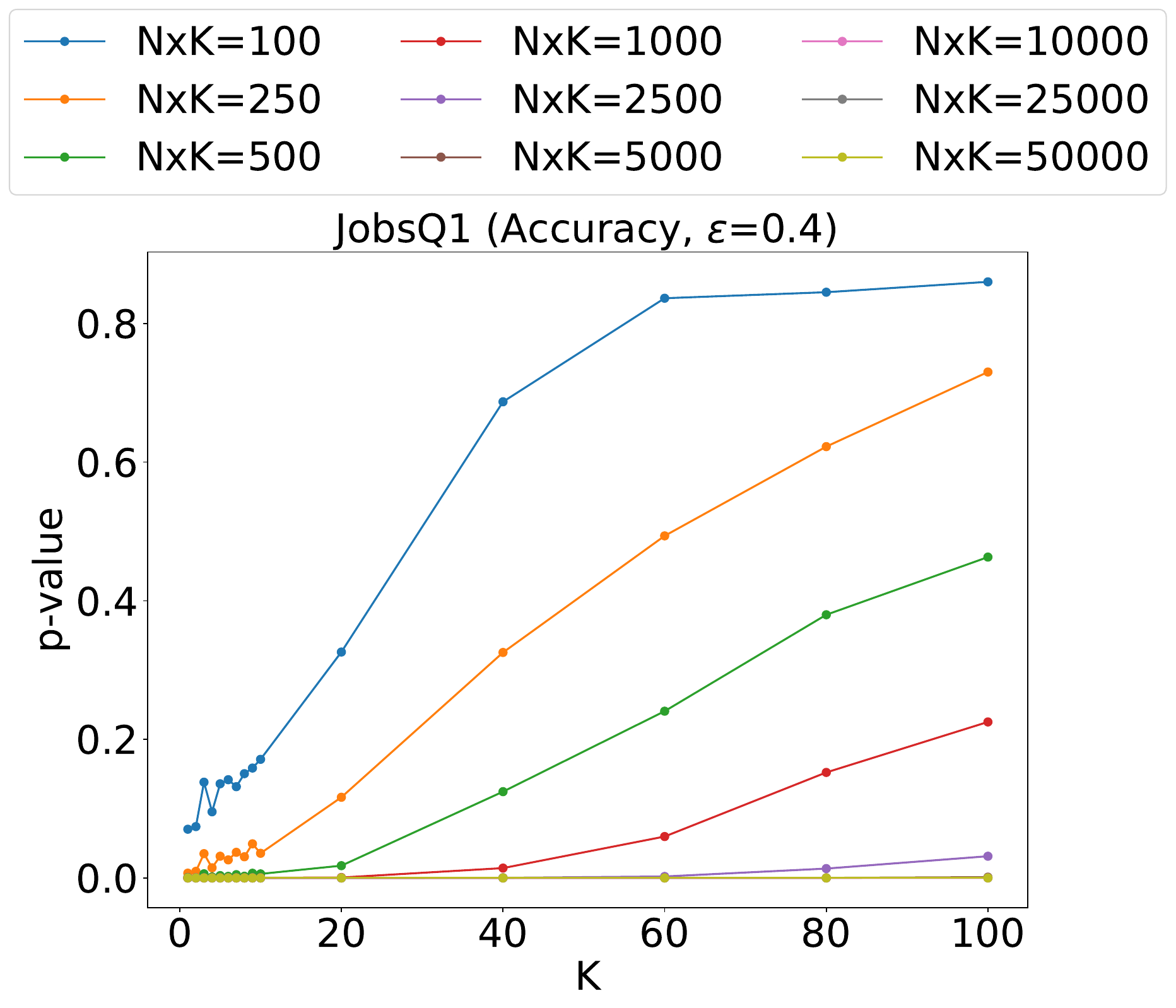}
    \caption{$\epsilon = 0.4$}
    \label{fig:jobsQ1_acc_e04}
  \end{subfigure}
  \caption{P-value plots for JobsQ1 dataset with Accuracy as the metric}
  \label{fig:jobsQ1_accuracy}
\end{figure*}

\begin{figure*}
  \centering
  \begin{subfigure}[b]{0.24\linewidth}
    \centering
    \includegraphics[width=\linewidth]{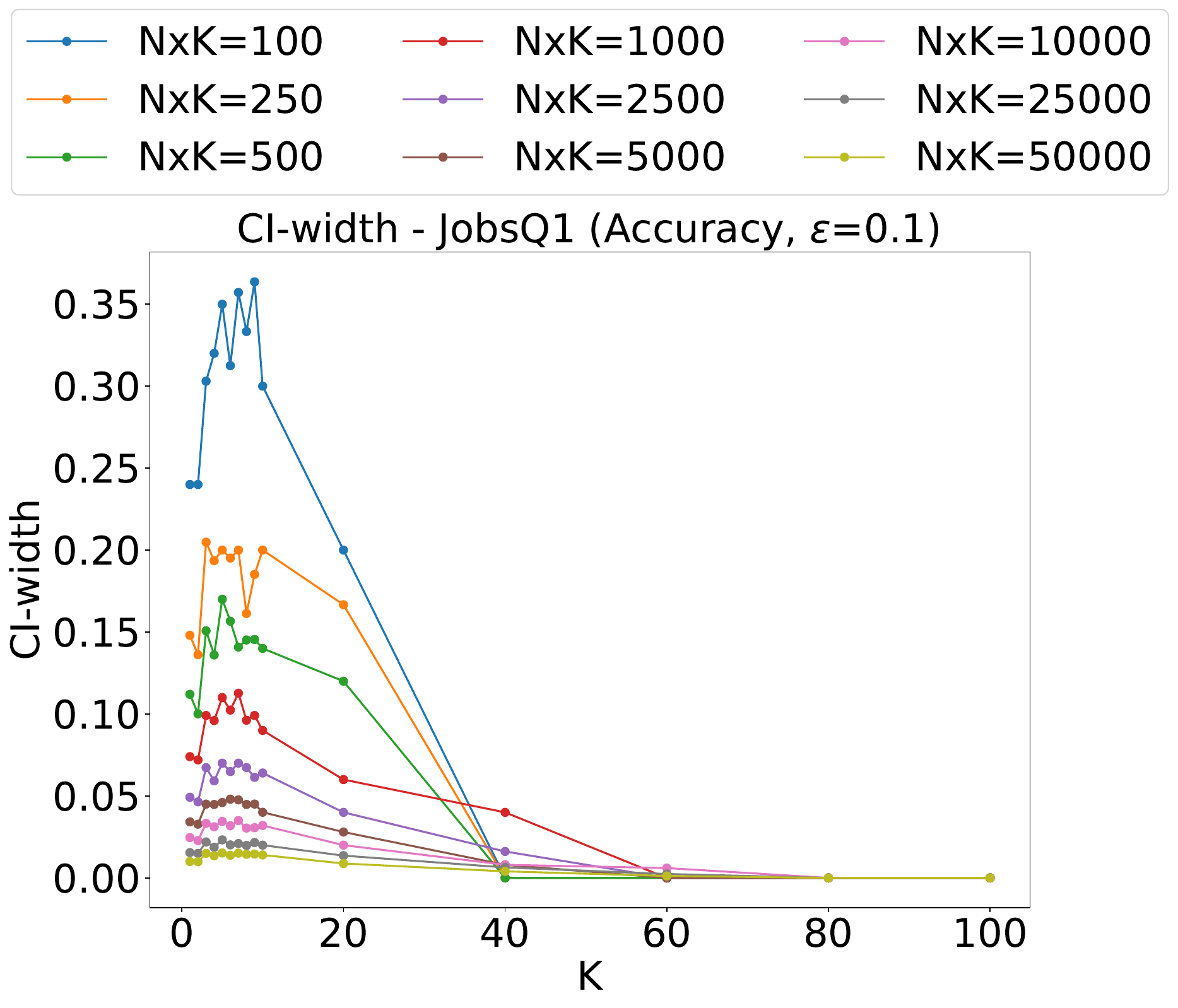}
    \caption{$\epsilon = 0.1$}
    \label{fig:jobsQ1_ci_acc_e01}
  \end{subfigure} \hfill
  \begin{subfigure}[b]{0.24\linewidth}
    \centering
    \includegraphics[width=\linewidth]{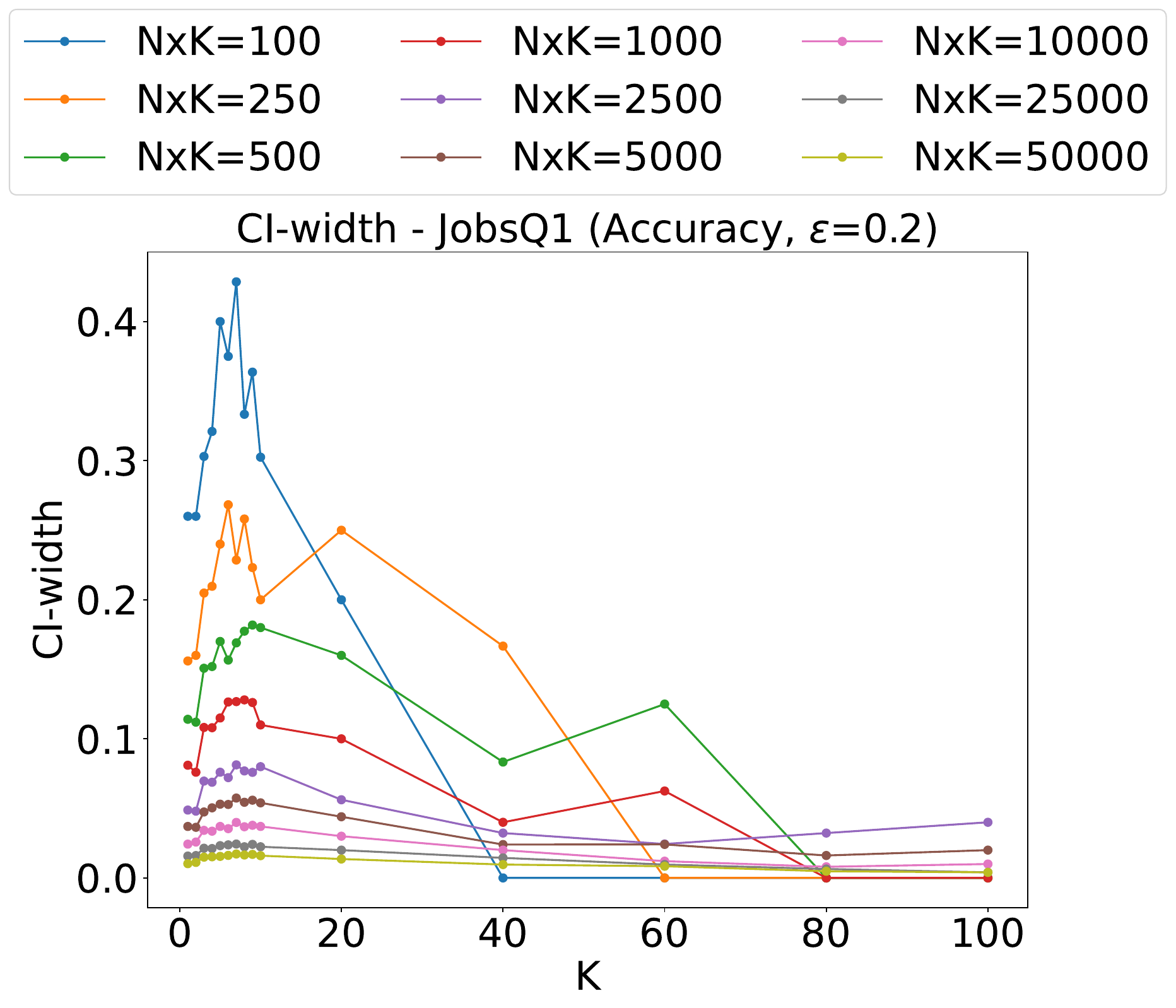}
    \caption{$\epsilon = 0.2$}
    \label{fig:jobsQ1_ci_acc_e02}
  \end{subfigure} \hfill
  \begin{subfigure}[b]{0.24\linewidth}
    \centering
    \includegraphics[width=\linewidth]{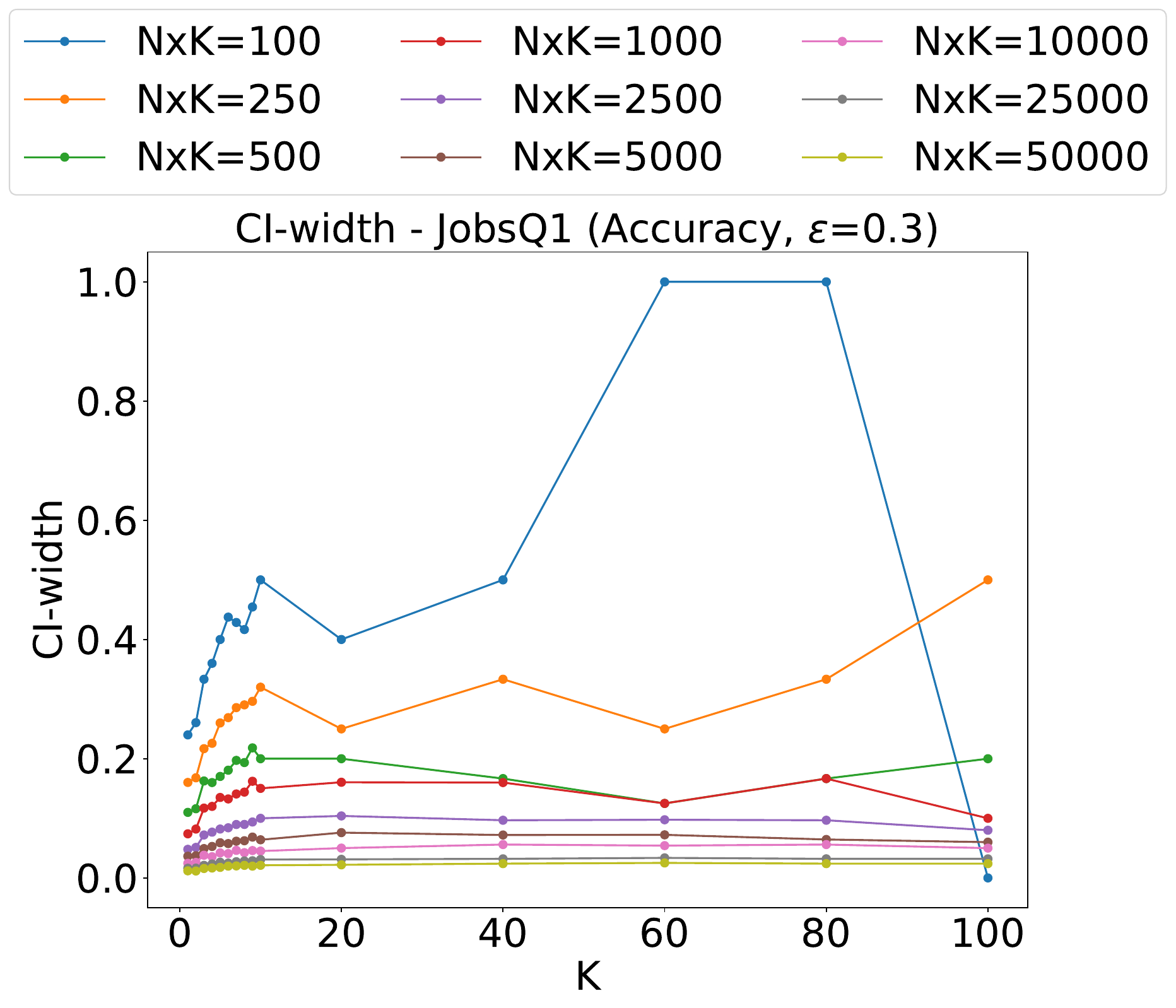}
    \caption{$\epsilon = 0.3$}
    \label{fig:jobsQ1_ci_acc_e03}
  \end{subfigure} \hfill
  \begin{subfigure}[b]{0.24\linewidth}
    \centering
    \includegraphics[width=\linewidth]{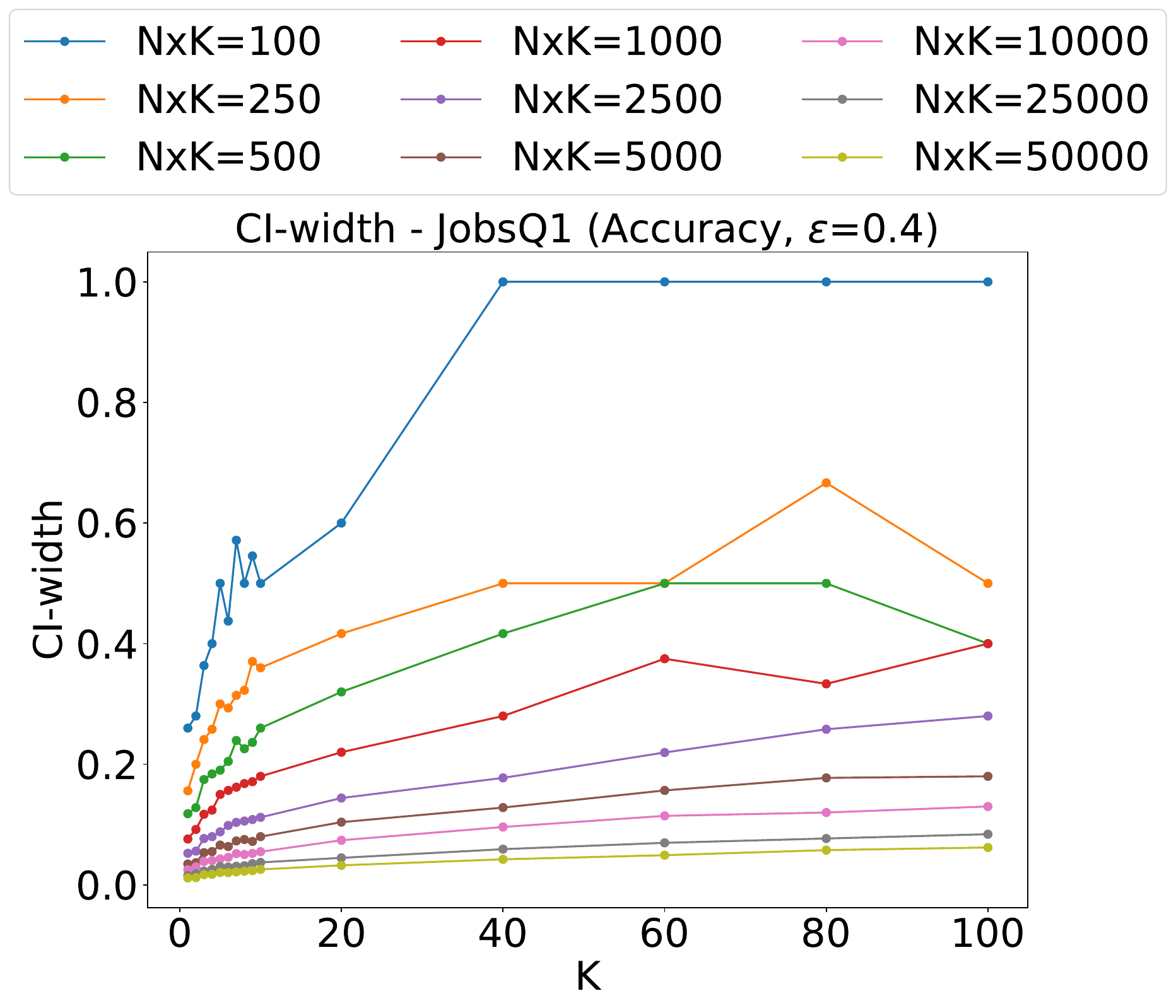}
    \caption{$\epsilon = 0.4$}
    \label{fig:jobsQ1_ci_acc_e04}
  \end{subfigure}
  \caption{CI-width plots for JobsQ1 dataset with Accuracy as the metric}
  \label{fig:jobsQ1_ci_accuracy}
\end{figure*}

\begin{figure*}
  \centering
  \begin{subfigure}[b]{0.24\linewidth}
    \centering
    \includegraphics[width=\linewidth]{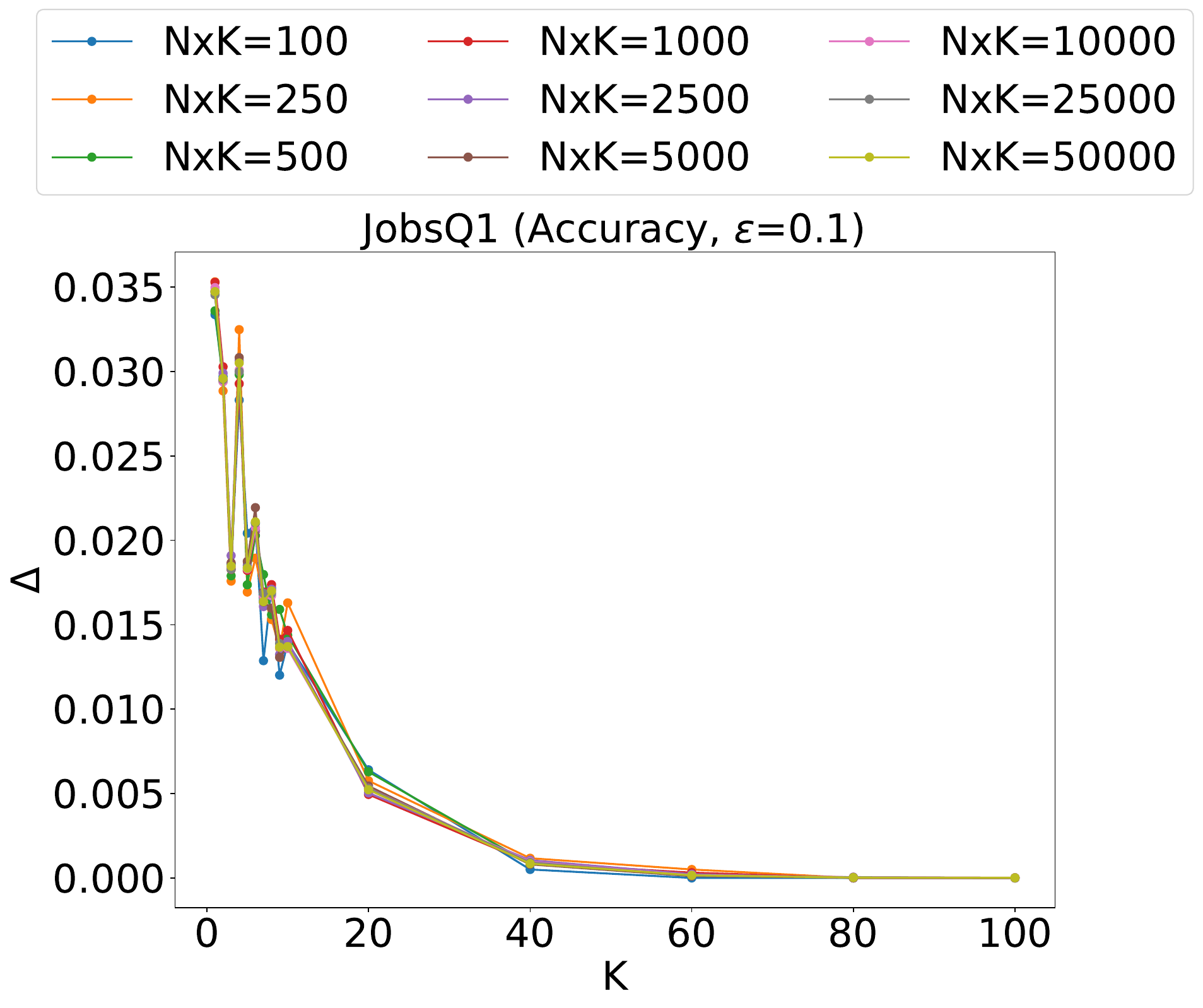}
    \caption{$\epsilon = 0.1$}
    \label{fig:jobsQ1_delta_acc_e01}
  \end{subfigure} \hfill
  \begin{subfigure}[b]{0.24\linewidth}
    \centering
    \includegraphics[width=\linewidth]{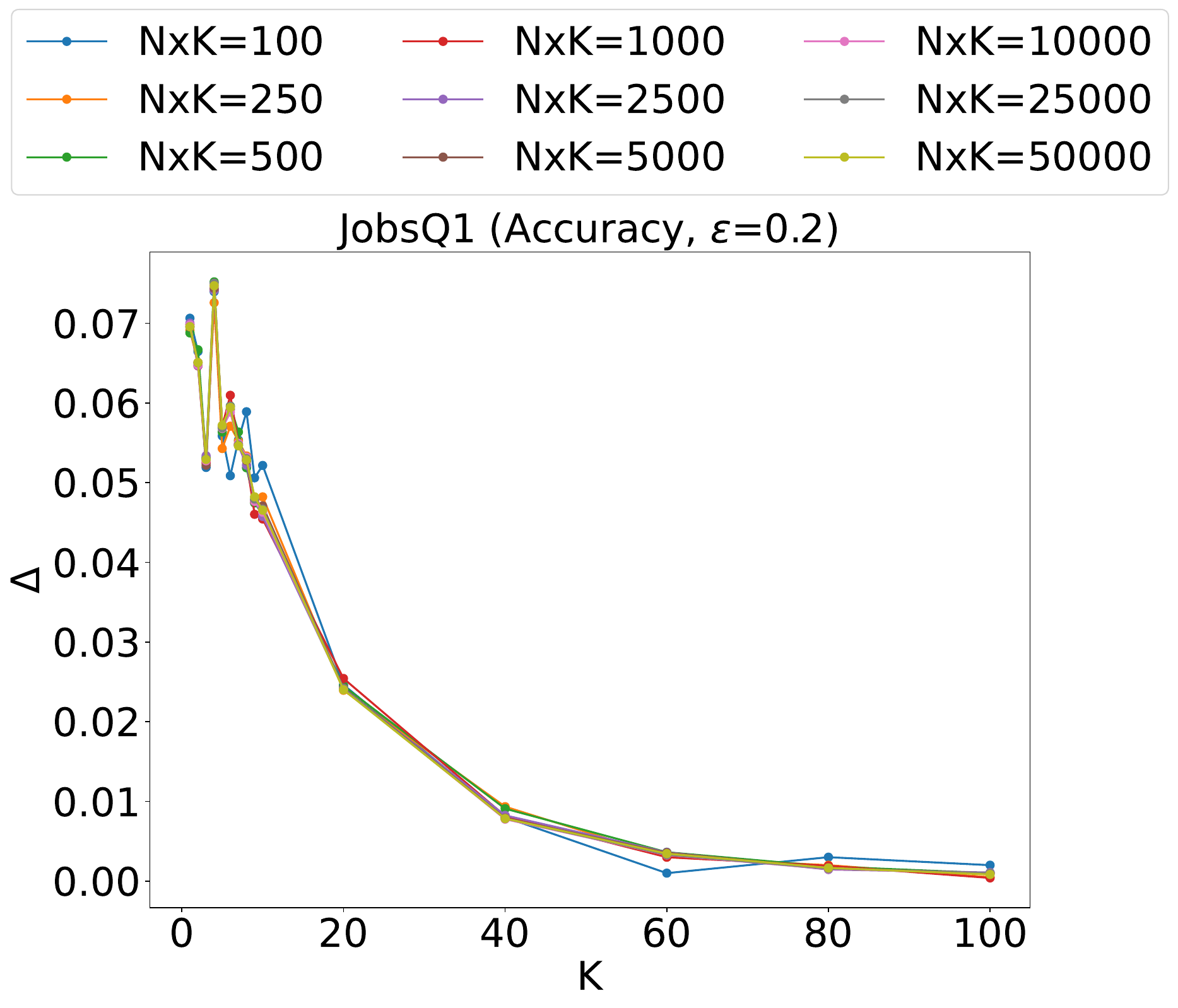}
    \caption{$\epsilon = 0.2$}
    \label{fig:jobsQ1_delta_acc_e02}
  \end{subfigure} \hfill
  \begin{subfigure}[b]{0.24\linewidth}
    \centering
    \includegraphics[width=\linewidth]{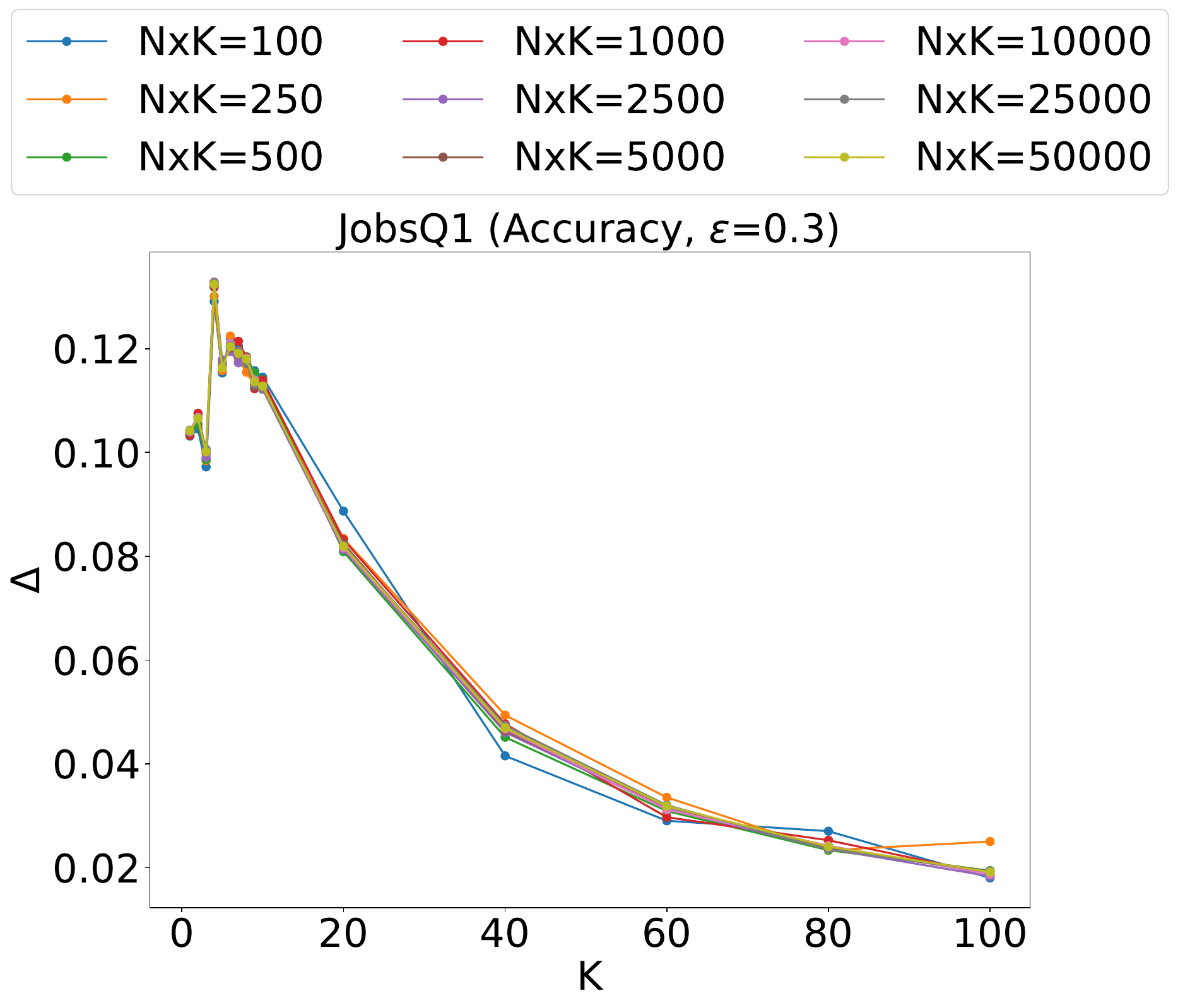}
    \caption{$\epsilon = 0.3$}
    \label{fig:jobsQ1_delta_acc_e03}
  \end{subfigure} \hfill
  \begin{subfigure}[b]{0.24\linewidth}
    \centering
    \includegraphics[width=\linewidth]{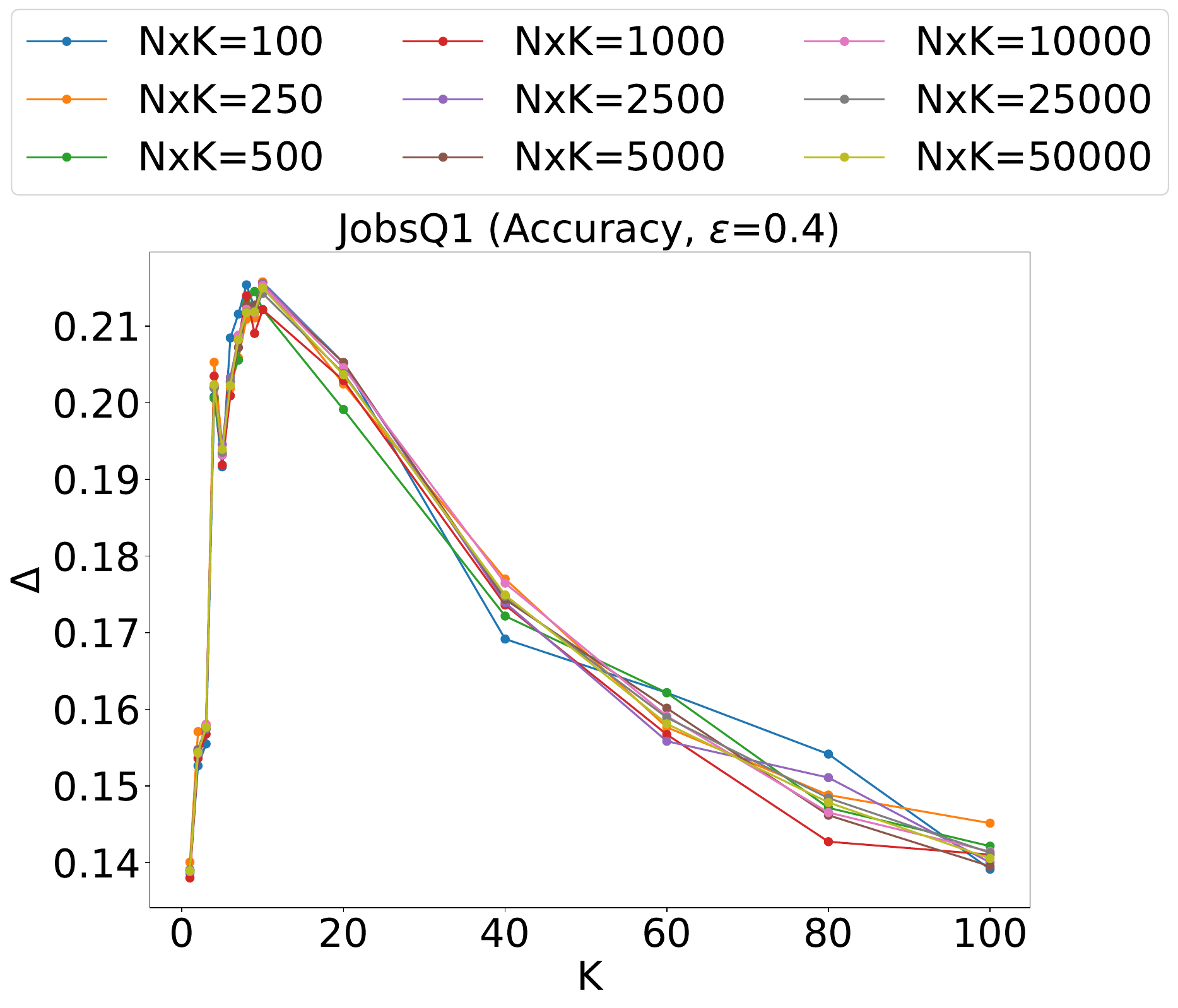}
    \caption{$\epsilon = 0.4$}
    \label{fig:jobsQ1_delta_acc_e04}
  \end{subfigure}
  \caption{Effect sizes ($\Delta$) for JobsQ1 dataset with Accuracy as the metric}
  \label{fig:jobsQ1_delta_accuracy}
\end{figure*}

\begin{figure*}
  \centering
  \begin{subfigure}[b]{0.24\linewidth}
    \centering
    \includegraphics[width=\linewidth]{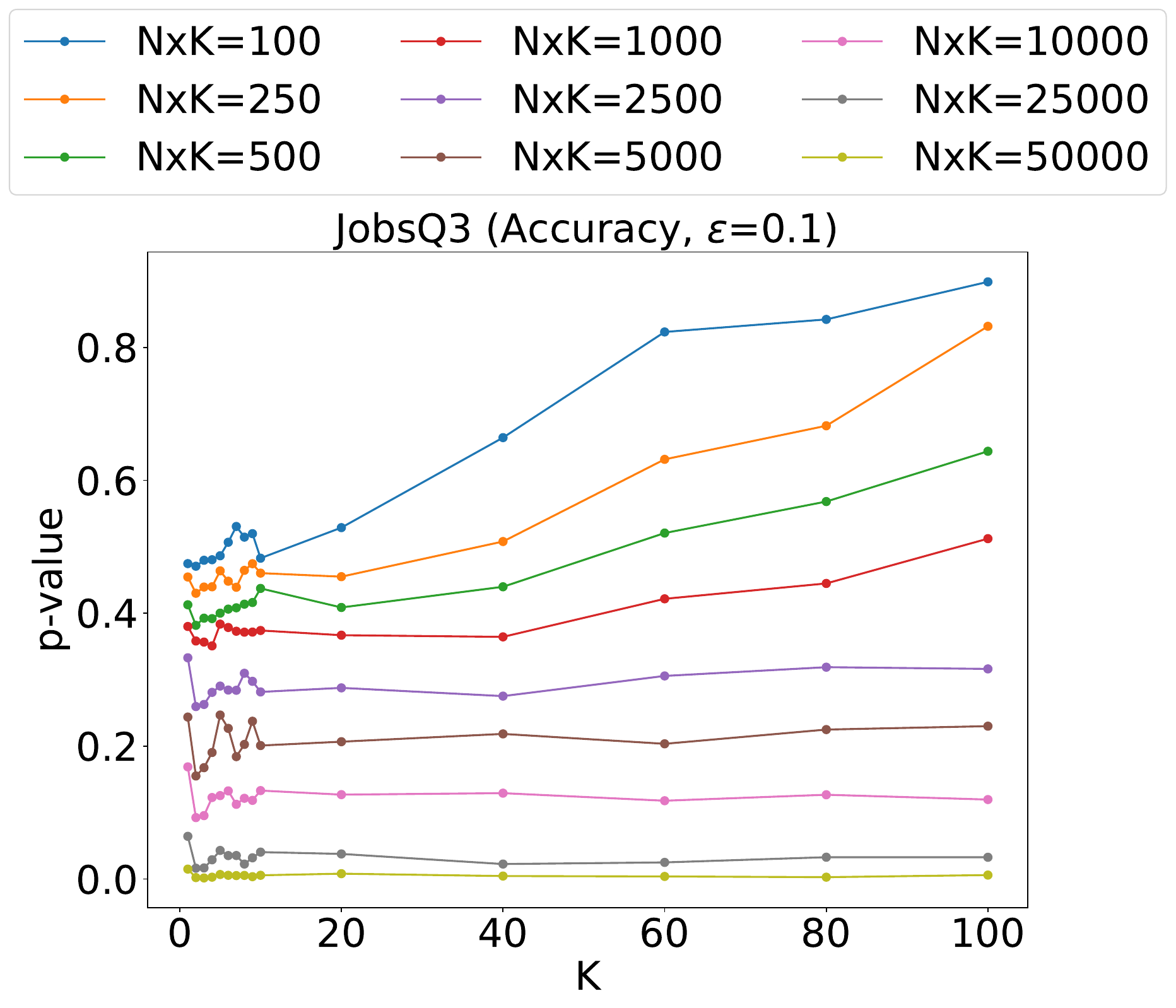}
    \caption{$\epsilon = 0.1$}
    \label{fig:jobsQ3_acc_e01}
  \end{subfigure} \hfill
  \begin{subfigure}[b]{0.24\linewidth}
    \centering
    \includegraphics[width=\linewidth]{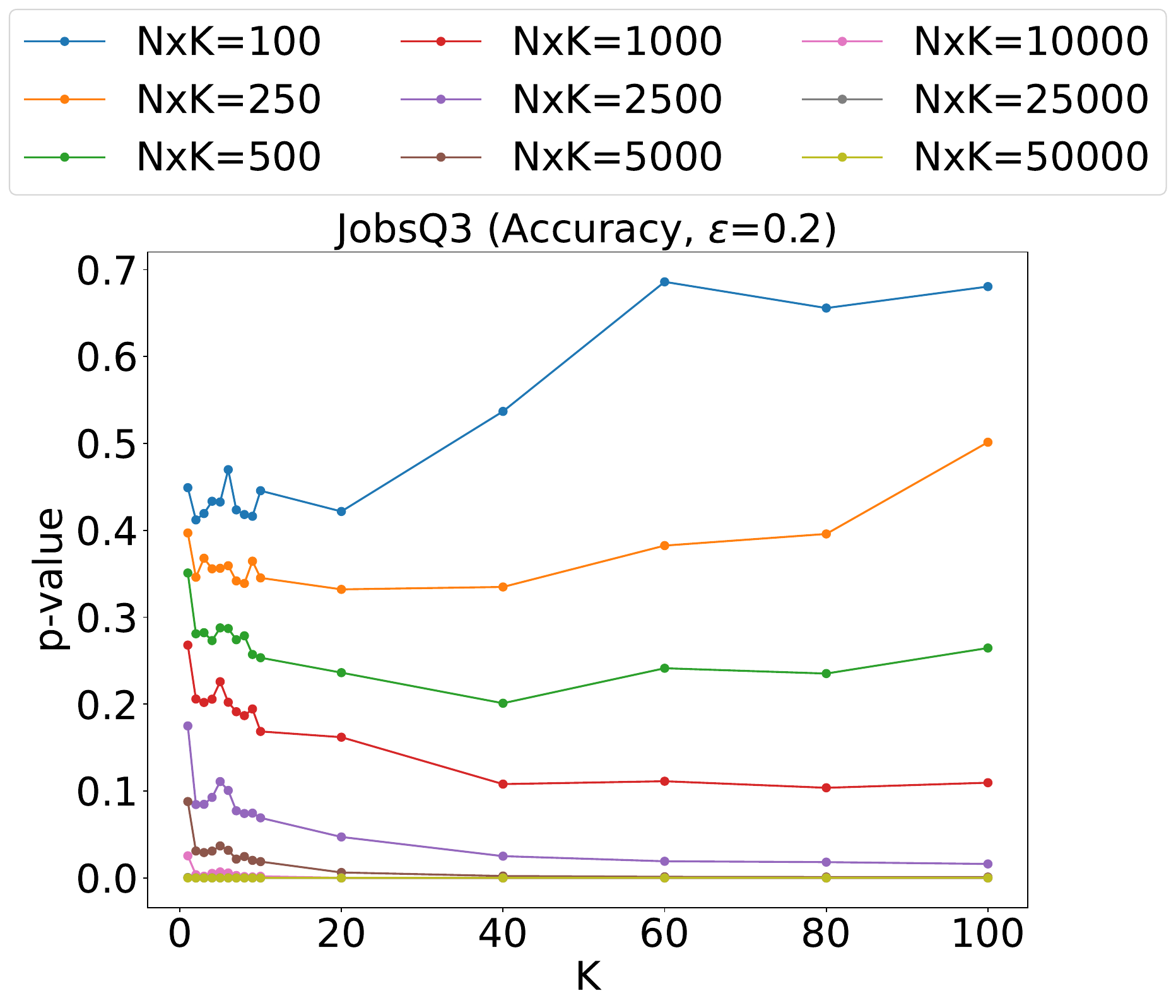}
    \caption{$\epsilon = 0.2$}
    \label{fig:jobsQ3_acc_e02}
  \end{subfigure} \hfill
  \begin{subfigure}[b]{0.24\linewidth}
    \centering
    \includegraphics[width=\linewidth]{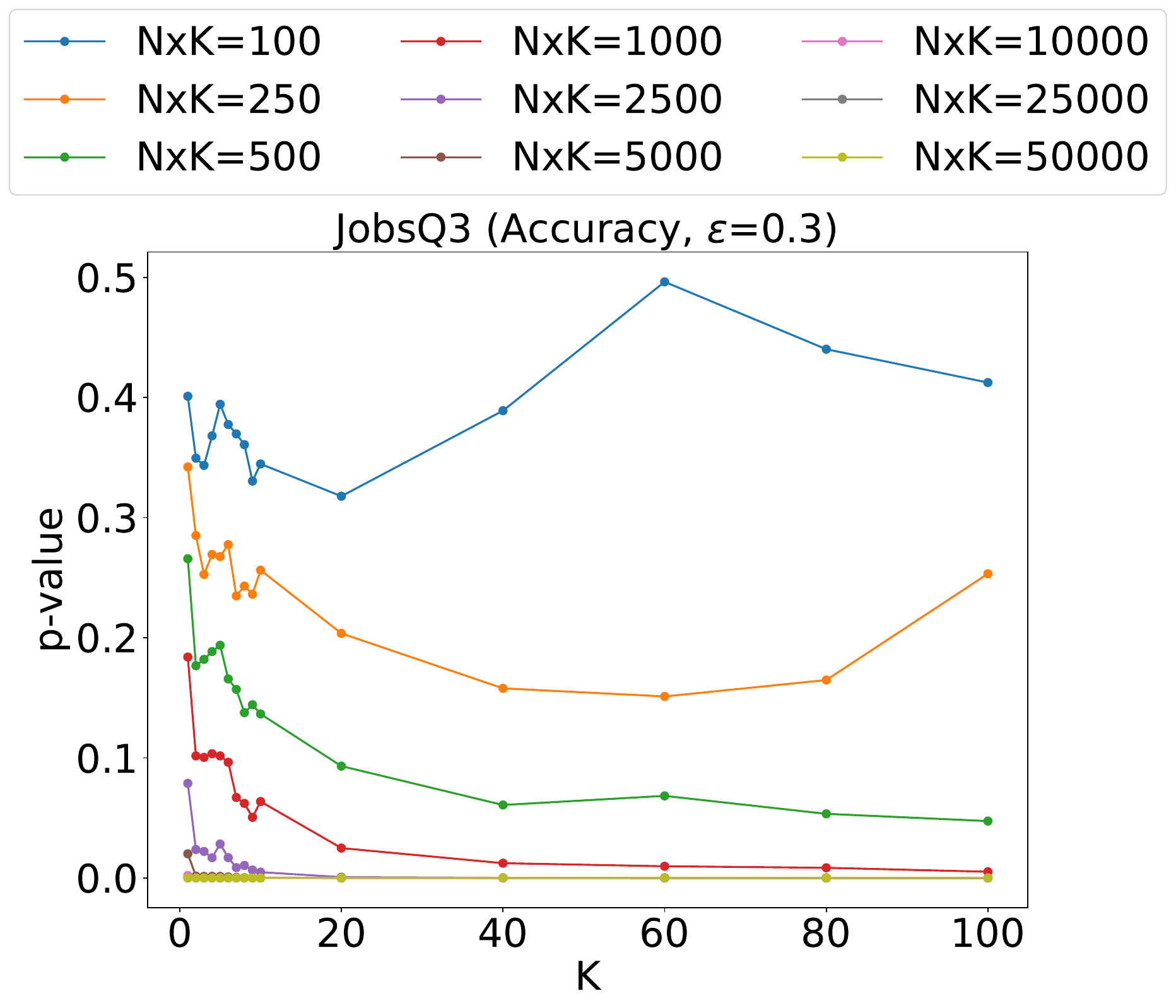}
    \caption{$\epsilon = 0.3$}
    \label{fig:jobsQ3_acc_e03}
  \end{subfigure} \hfill
  \begin{subfigure}[b]{0.24\linewidth}
    \centering
    \includegraphics[width=\linewidth]{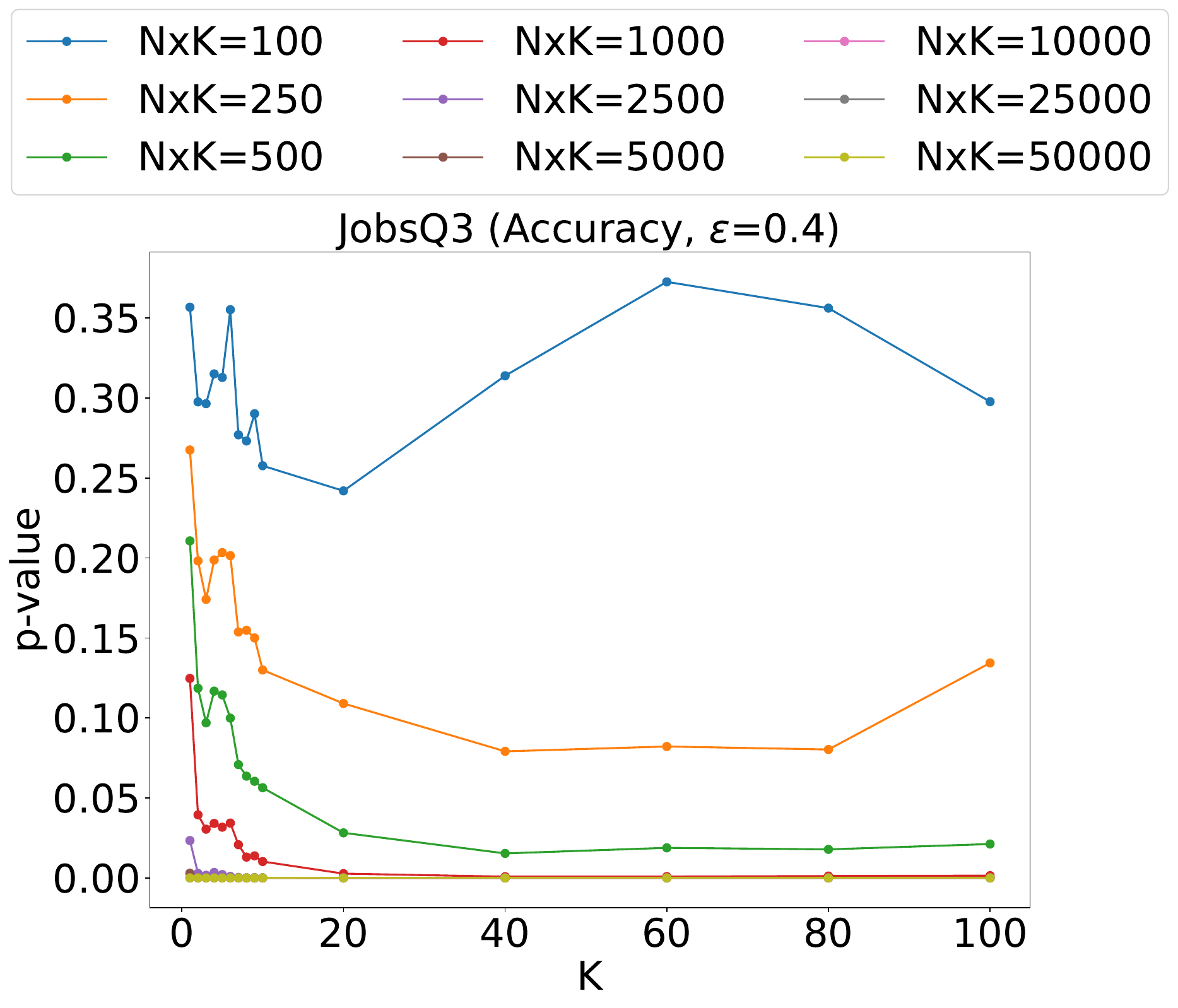}
    \caption{$\epsilon = 0.4$}
    \label{fig:jobsQ3_acc_e04}
  \end{subfigure}
  \caption{P-value plots for JobsQ3 dataset with Accuracy as the metric}
  \label{fig:jobsQ3_accuracy}
\end{figure*}

\begin{figure*}
  \centering
  \begin{subfigure}[b]{0.24\linewidth}
    \centering
    \includegraphics[width=\linewidth]{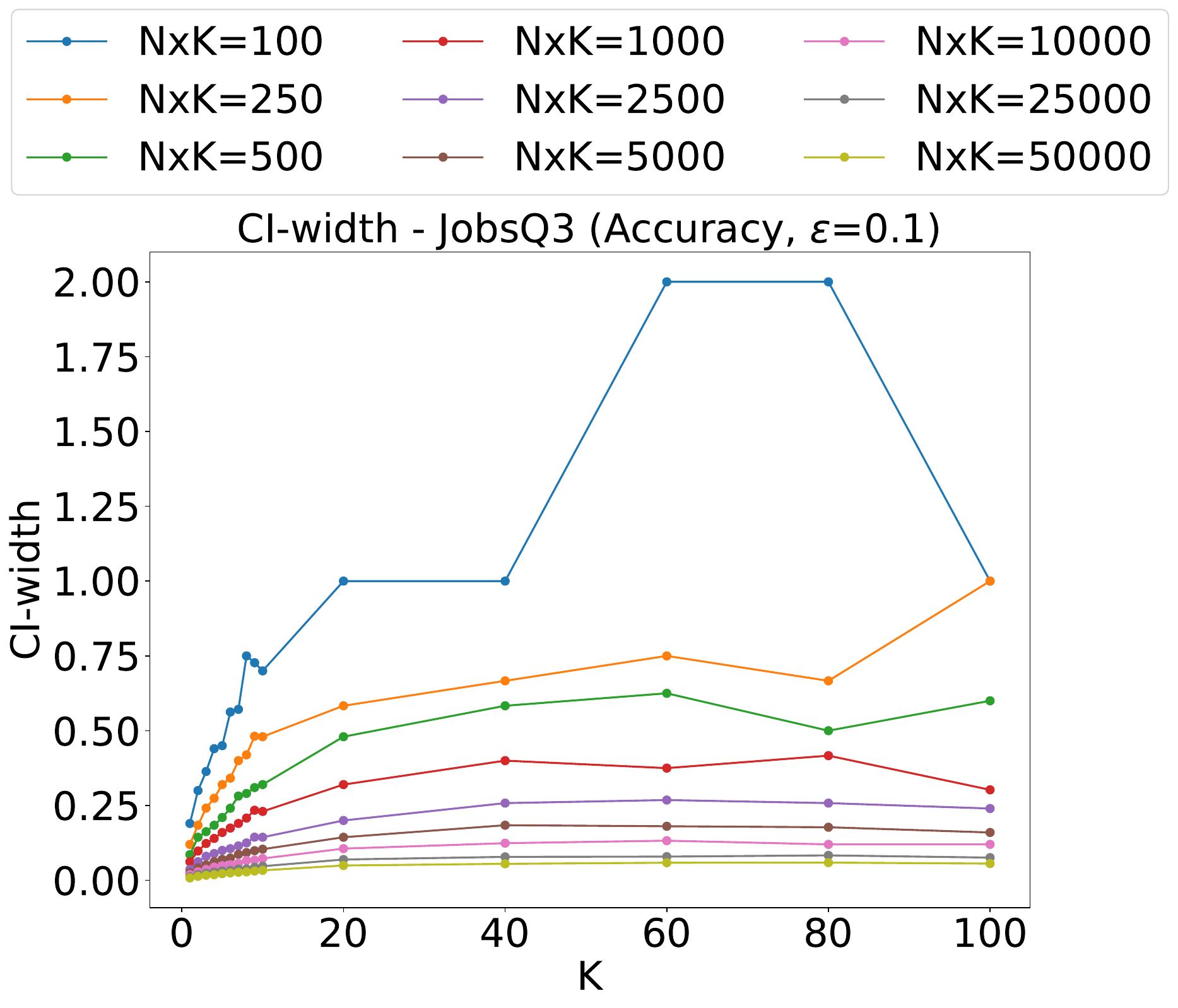}
    \caption{$\epsilon = 0.1$}
    \label{fig:jobsQ3_ci_acc_e01}
  \end{subfigure} \hfill
  \begin{subfigure}[b]{0.24\linewidth}
    \centering
    \includegraphics[width=\linewidth]{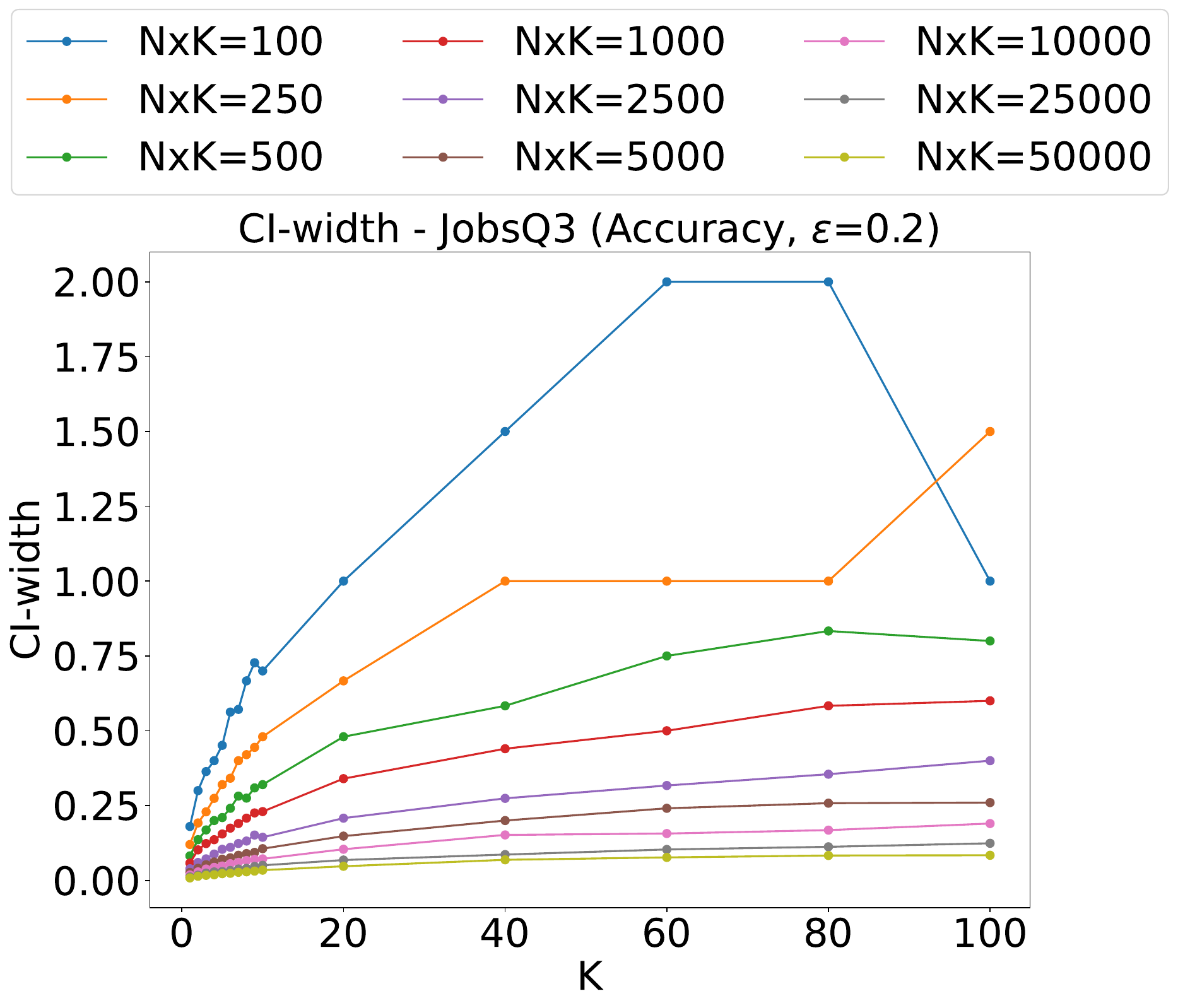}
    \caption{$\epsilon = 0.2$}
    \label{fig:jobsQ3_ci_acc_e02}
  \end{subfigure} \hfill
  \begin{subfigure}[b]{0.24\linewidth}
    \centering
    \includegraphics[width=\linewidth]{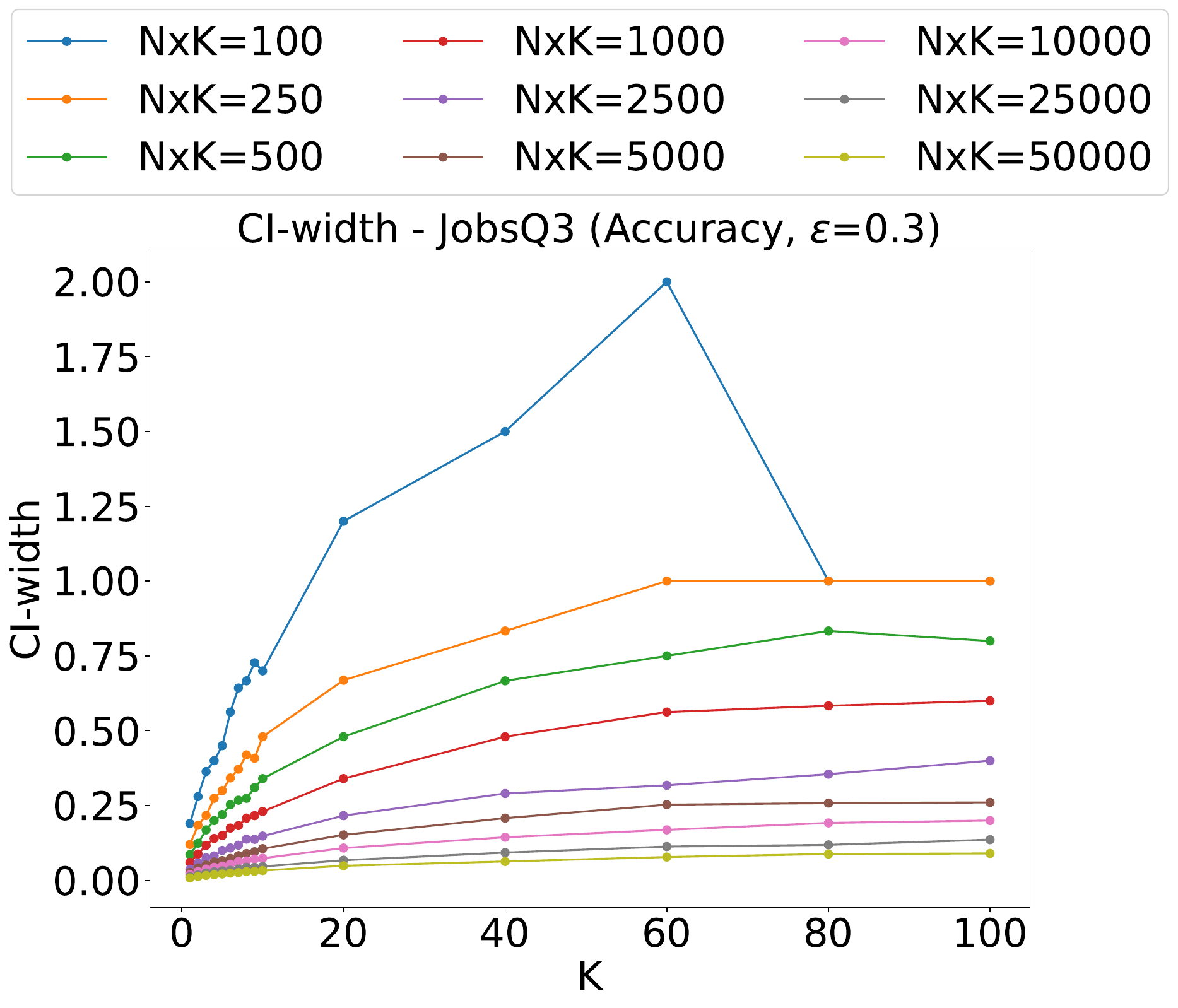}
    \caption{$\epsilon = 0.3$}
    \label{fig:jobsQ3_ci_acc_e03}
  \end{subfigure} \hfill
  \begin{subfigure}[b]{0.24\linewidth}
    \centering
    \includegraphics[width=\linewidth]{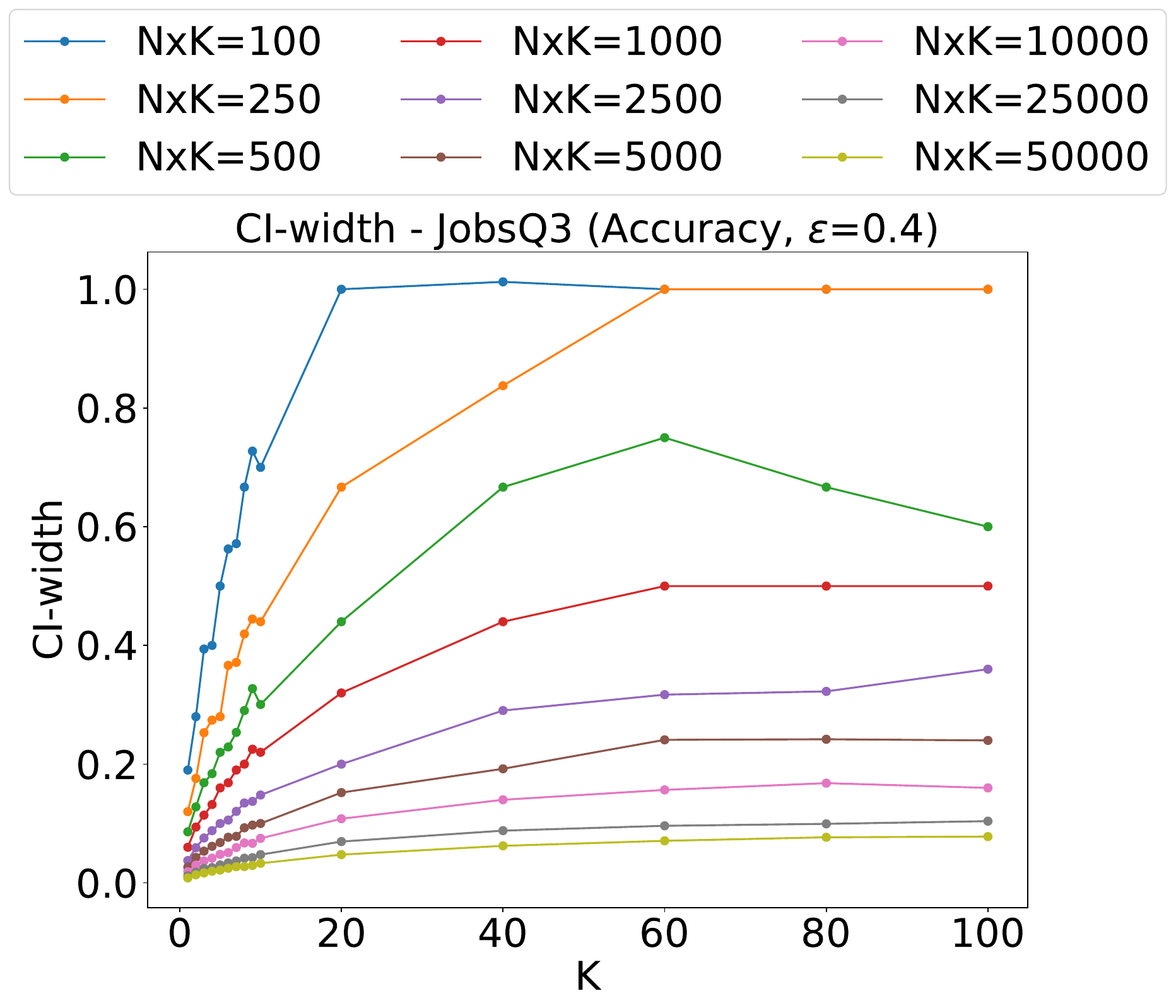}
    \caption{$\epsilon = 0.4$}
    \label{fig:jobsQ3_ci_acc_e04}
  \end{subfigure}
  \caption{CI-width plots for JobsQ3 dataset with Accuracy as the metric}
  \label{fig:jobsQ3_ci_accuracy}
\end{figure*}

\begin{figure*}
  \centering
  \begin{subfigure}[b]{0.24\linewidth}
    \centering
    \includegraphics[width=\linewidth]{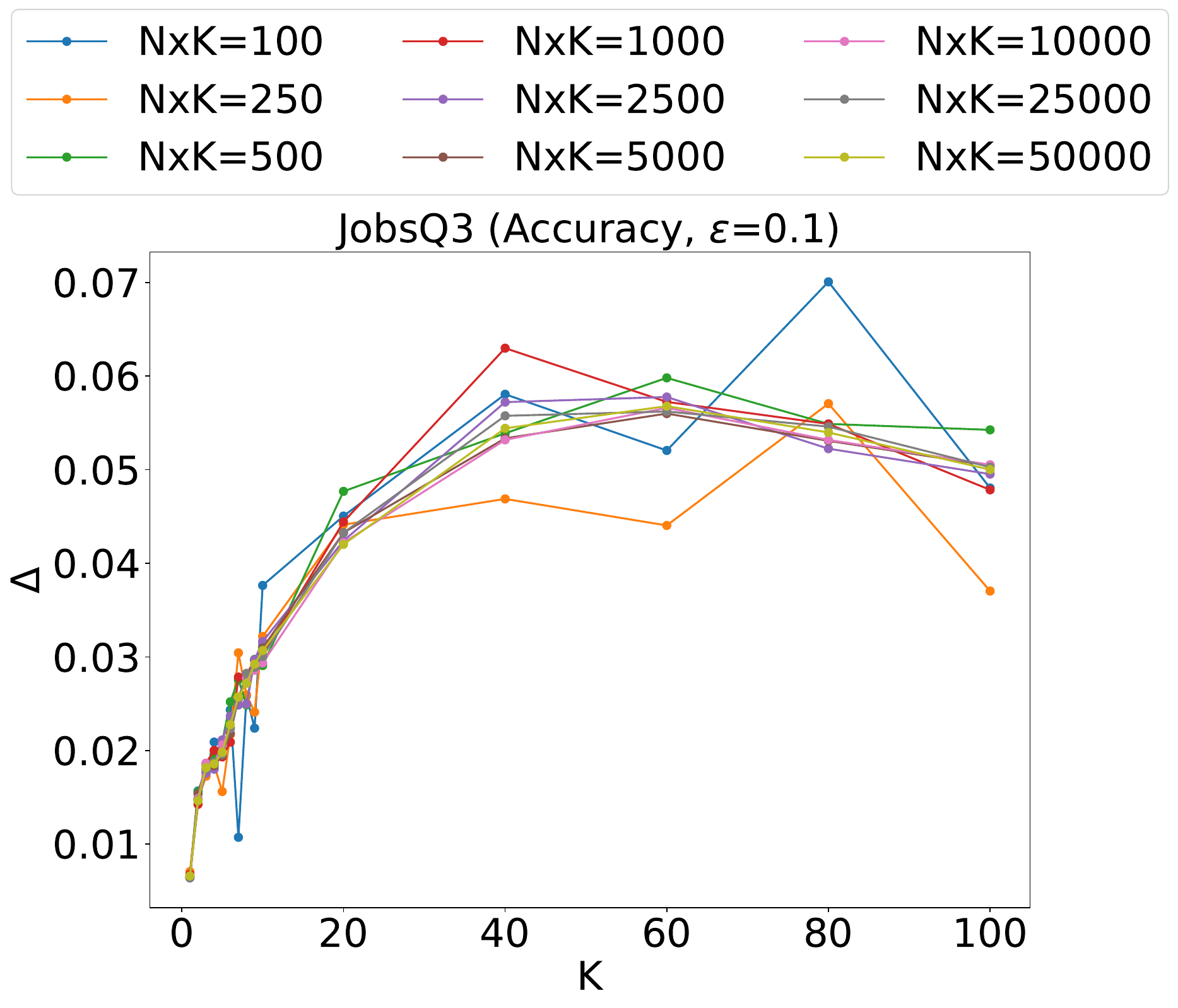}
    \caption{$\epsilon = 0.1$}
    \label{fig:jobsQ3_delta_acc_e01}
  \end{subfigure} \hfill
  \begin{subfigure}[b]{0.24\linewidth}
    \centering
    \includegraphics[width=\linewidth]{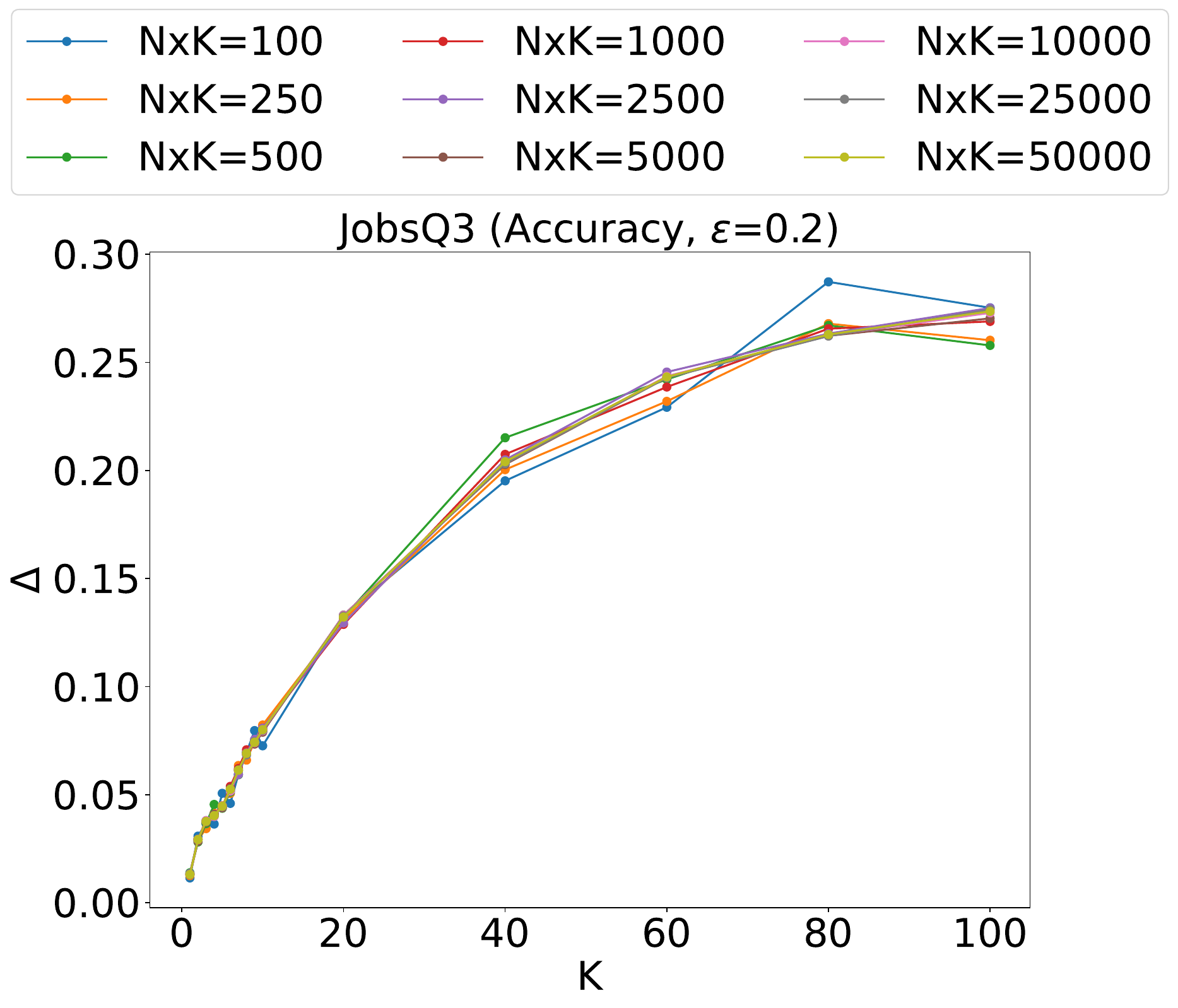}
    \caption{$\epsilon = 0.2$}
    \label{fig:jobsQ3_delta_acc_e02}
  \end{subfigure} \hfill
  \begin{subfigure}[b]{0.24\linewidth}
    \centering
    \includegraphics[width=\linewidth]{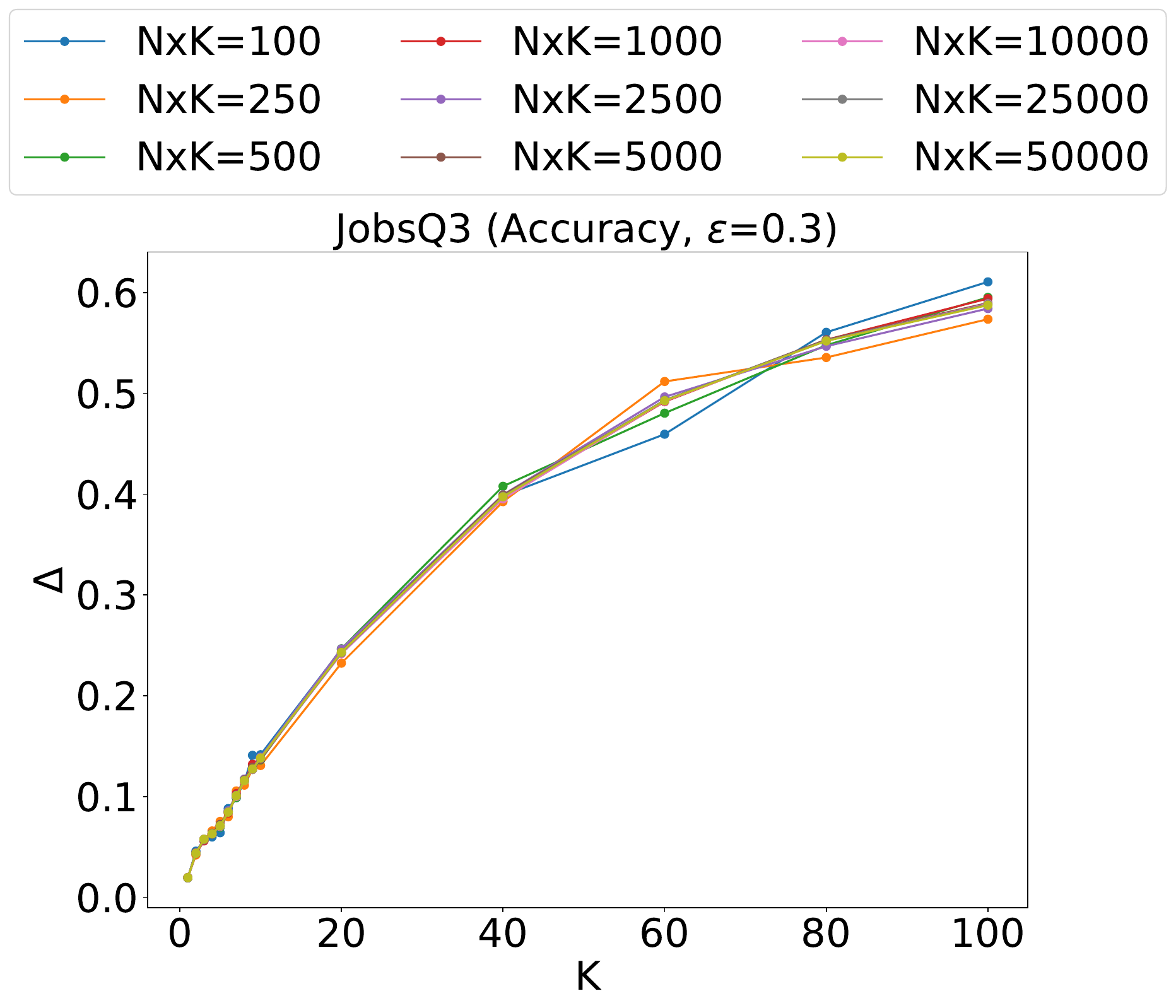}
    \caption{$\epsilon = 0.3$}
    \label{fig:jobsQ3_delta_acc_e03}
  \end{subfigure} \hfill
  \begin{subfigure}[b]{0.24\linewidth}
    \centering
    \includegraphics[width=\linewidth]{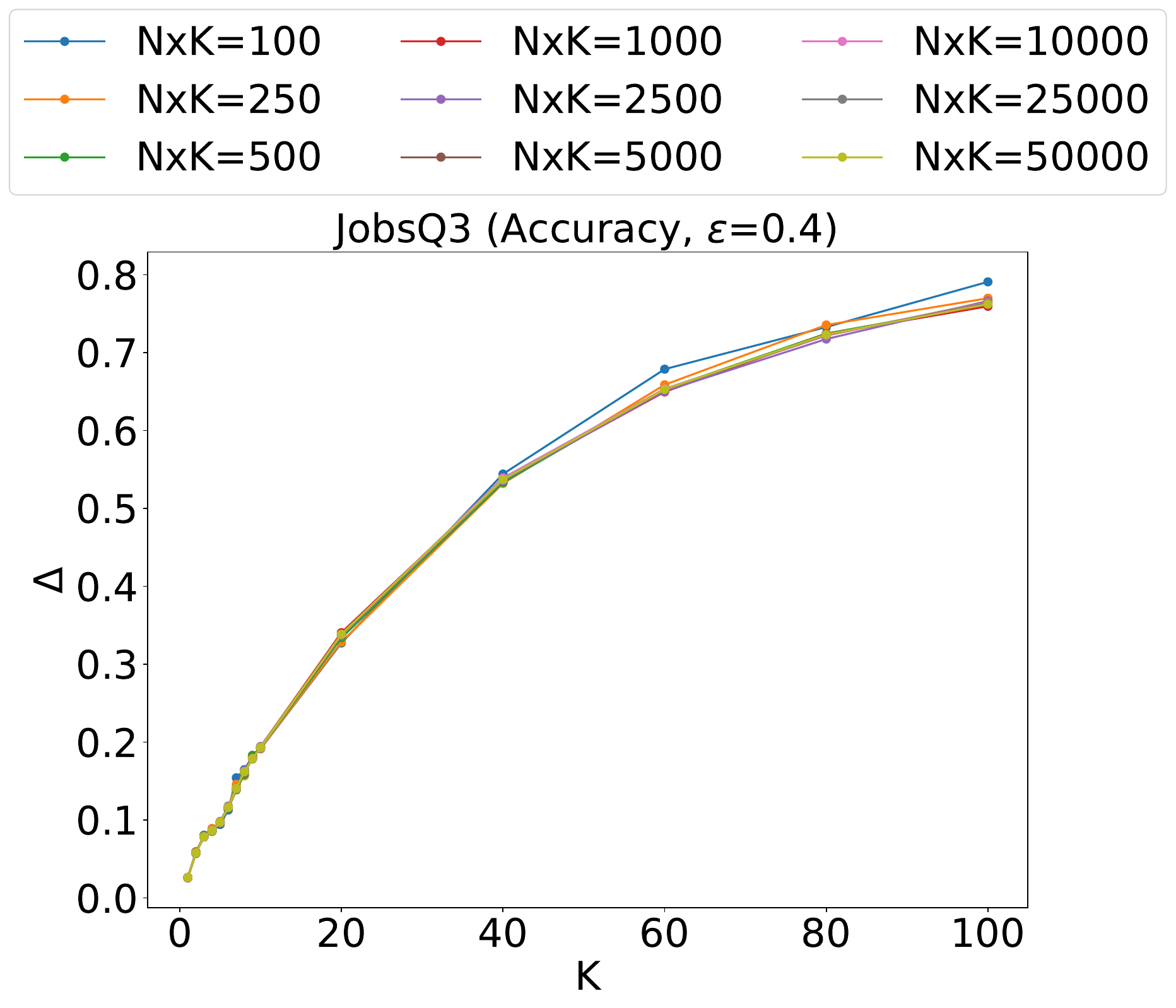}
    \caption{$\epsilon = 0.4$}
    \label{fig:jobsQ3_delta_acc_e04}
  \end{subfigure}
  \caption{Effect sizes ($\Delta$) for JobsQ3 dataset with Accuracy as the metric}
  \label{fig:jobsQ3_delta_accuracy}
\end{figure*}

Figures \ref{fig:toxicity_accuracy}--\ref{fig:jobsQ3_delta_accuracy} show results for \pv s, confidence intervals, and effect sizes for all real-world datasets.
 We notice that the \pv\ increases as $K$ increases for all $N \times K$ in all datasets. The increase is sharp until $K=40$ for all $N \times K$ values, especially for lower values of $\epsilon$ (0.1, 0.2). Then, the \pv s\ start plateauing for higher $N \times K$ values, but continue to increase for lower $N \times K$ values. As $\epsilon$ increases, \pv s\ decrease as expected. Having more responses per item ($K$) does not seem to be helpful for accuracy. CI-width and effect sizes increase with increasing $K$ except for JobsQ1.

\subsubsection{Artificial Distribution with Balanced Priors}

\begin{figure*}
  \centering
  \begin{subfigure}[b]{0.24\linewidth}
    \centering
    \includegraphics[width=\linewidth]{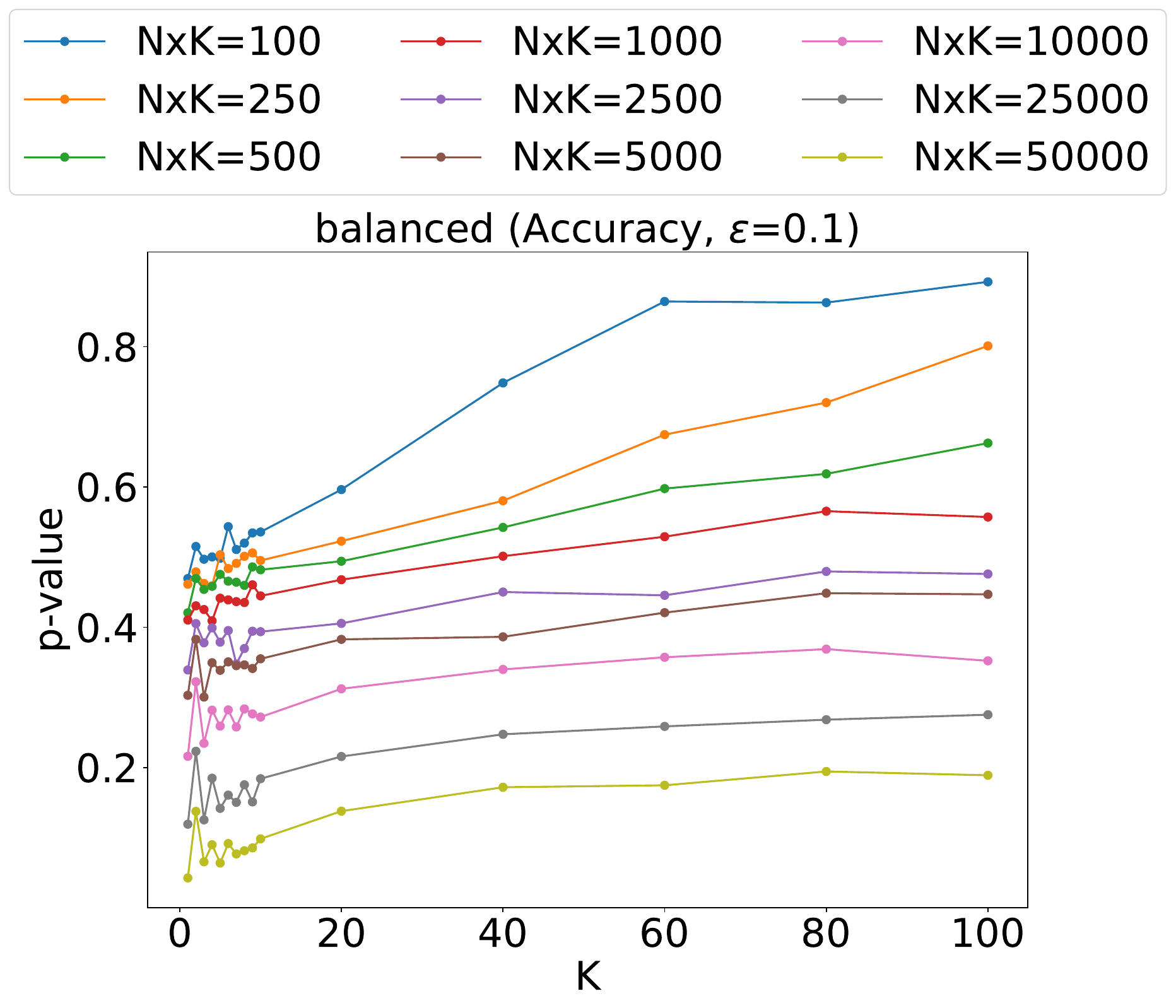}
    \caption{$\epsilon = 0.1$}
    \label{fig:uniform_accuracy_cat2_e01}
  \end{subfigure} \hfill
  \begin{subfigure}[b]{0.24\linewidth}
    \centering
    \includegraphics[width=\linewidth]{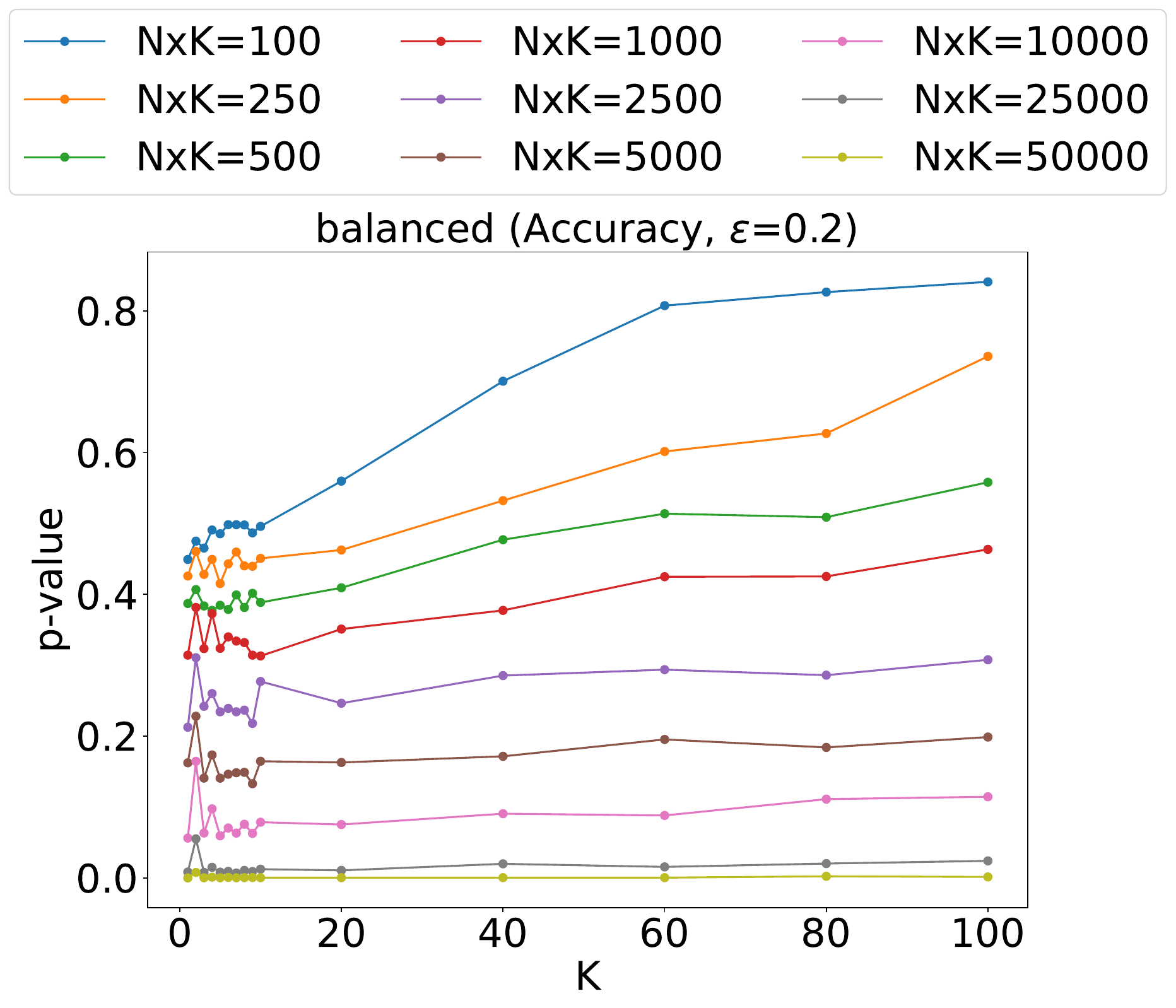}
    \caption{$\epsilon = 0.2$}
    \label{fig:uniform_accuracy_cat2_e02}
  \end{subfigure} \hfill
  \begin{subfigure}[b]{0.24\linewidth}
    \centering
    \includegraphics[width=\linewidth]{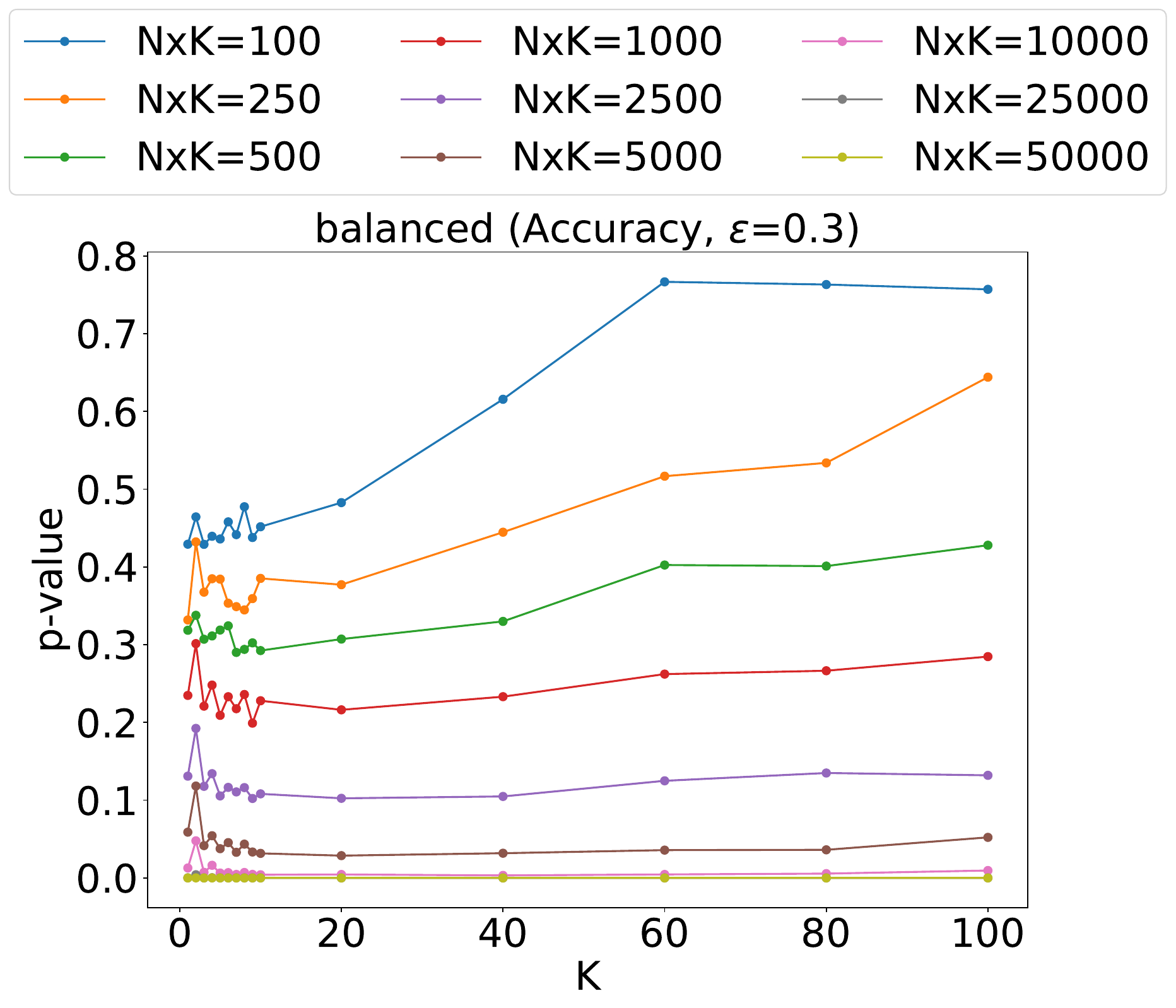}
    \caption{$\epsilon = 0.3$}
    \label{fig:uniform_accuracy_cat2_e03}
  \end{subfigure} \hfill
  \begin{subfigure}[b]{0.24\linewidth}
    \centering
    \includegraphics[width=\linewidth]{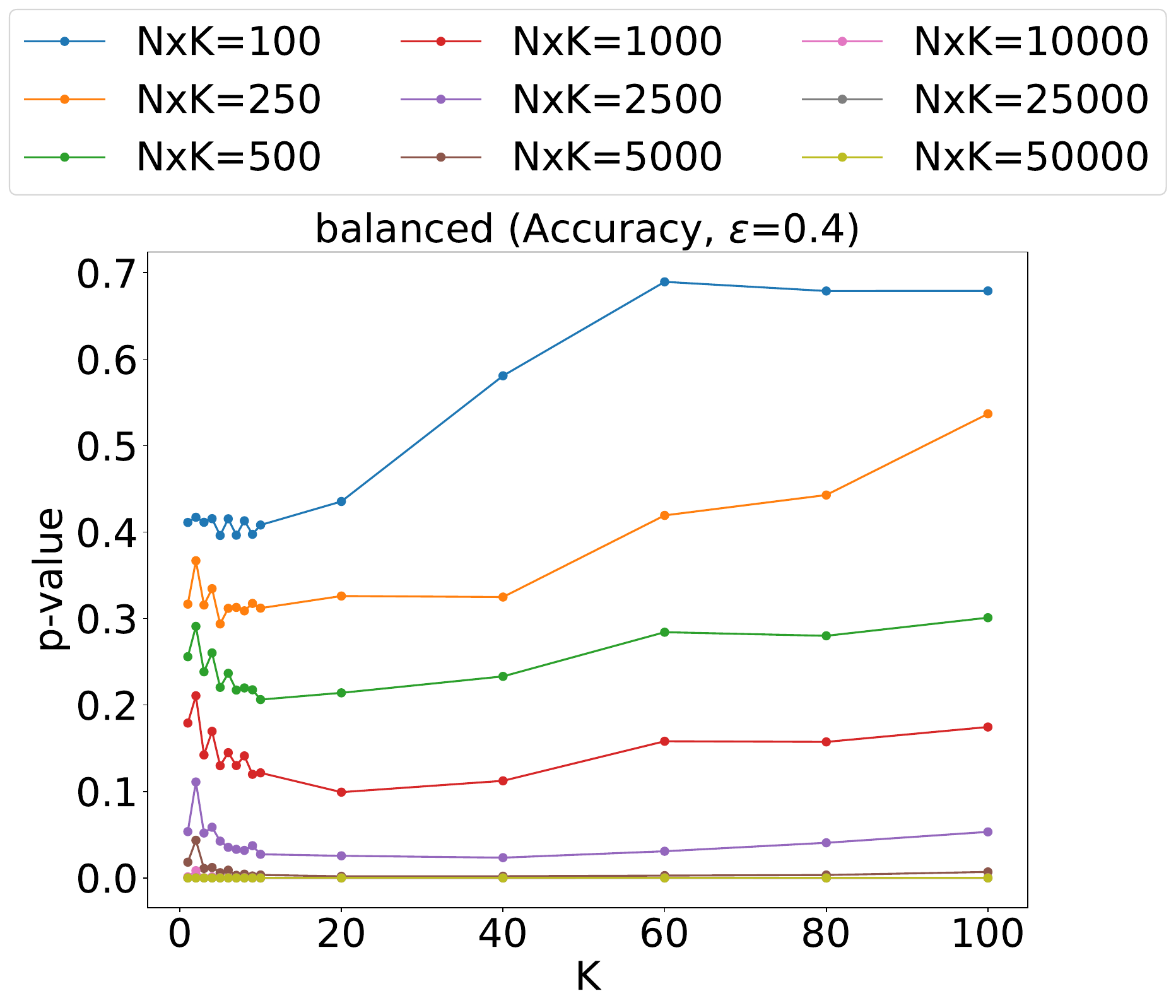}
    \caption{$\epsilon = 0.4$}
    \label{fig:uniform_accuracy_cat2_e04}
  \end{subfigure}
  \caption{P-value plots for balanced aplhas with Accuracy as the metric ($M=2$)}
  \label{fig:uniform_accuracy_cat2}
\end{figure*}

\begin{figure*}
  \centering
  \begin{subfigure}[b]{0.24\linewidth}
    \centering
    \includegraphics[width=\linewidth]{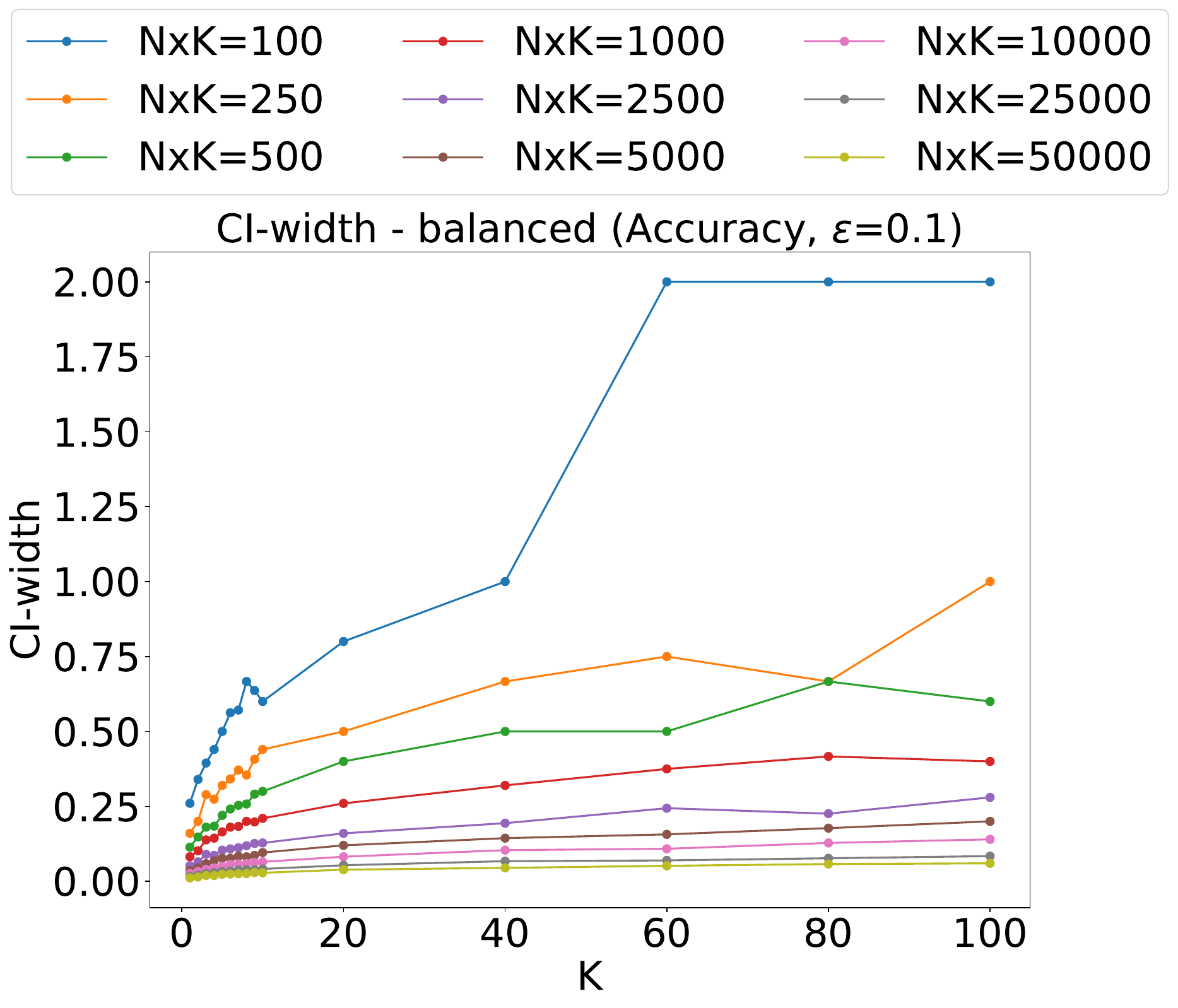}
    \caption{$\epsilon = 0.1$}
    \label{fig:uniform_ci_accuracy_cat2_e01}
  \end{subfigure} \hfill
  \begin{subfigure}[b]{0.24\linewidth}
    \centering
    \includegraphics[width=\linewidth]{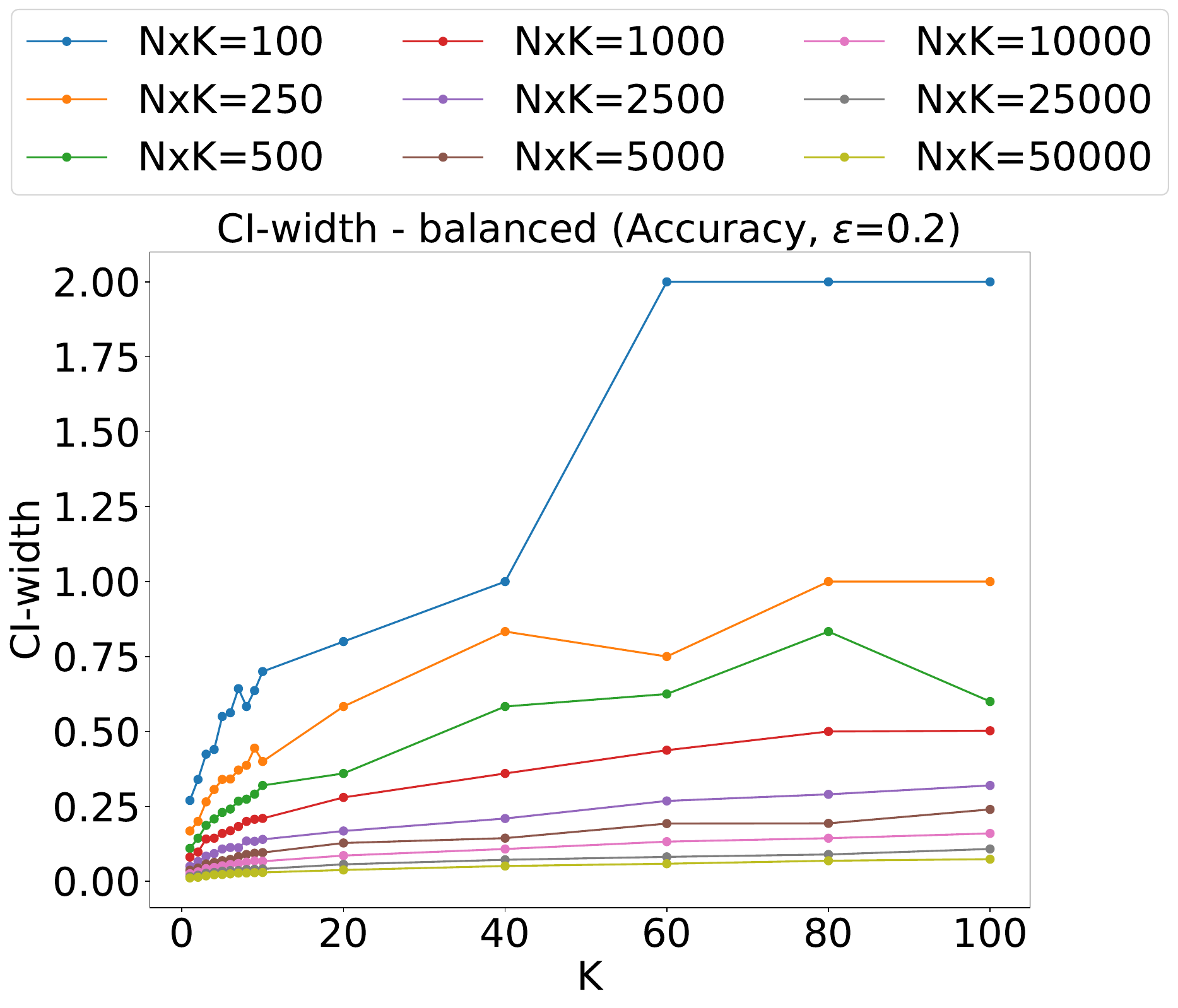}
    \caption{$\epsilon = 0.2$}
    \label{fig:uniform_ci_accuracy_cat2_e02}
  \end{subfigure} \hfill
  \begin{subfigure}[b]{0.24\linewidth}
    \centering
    \includegraphics[width=\linewidth]{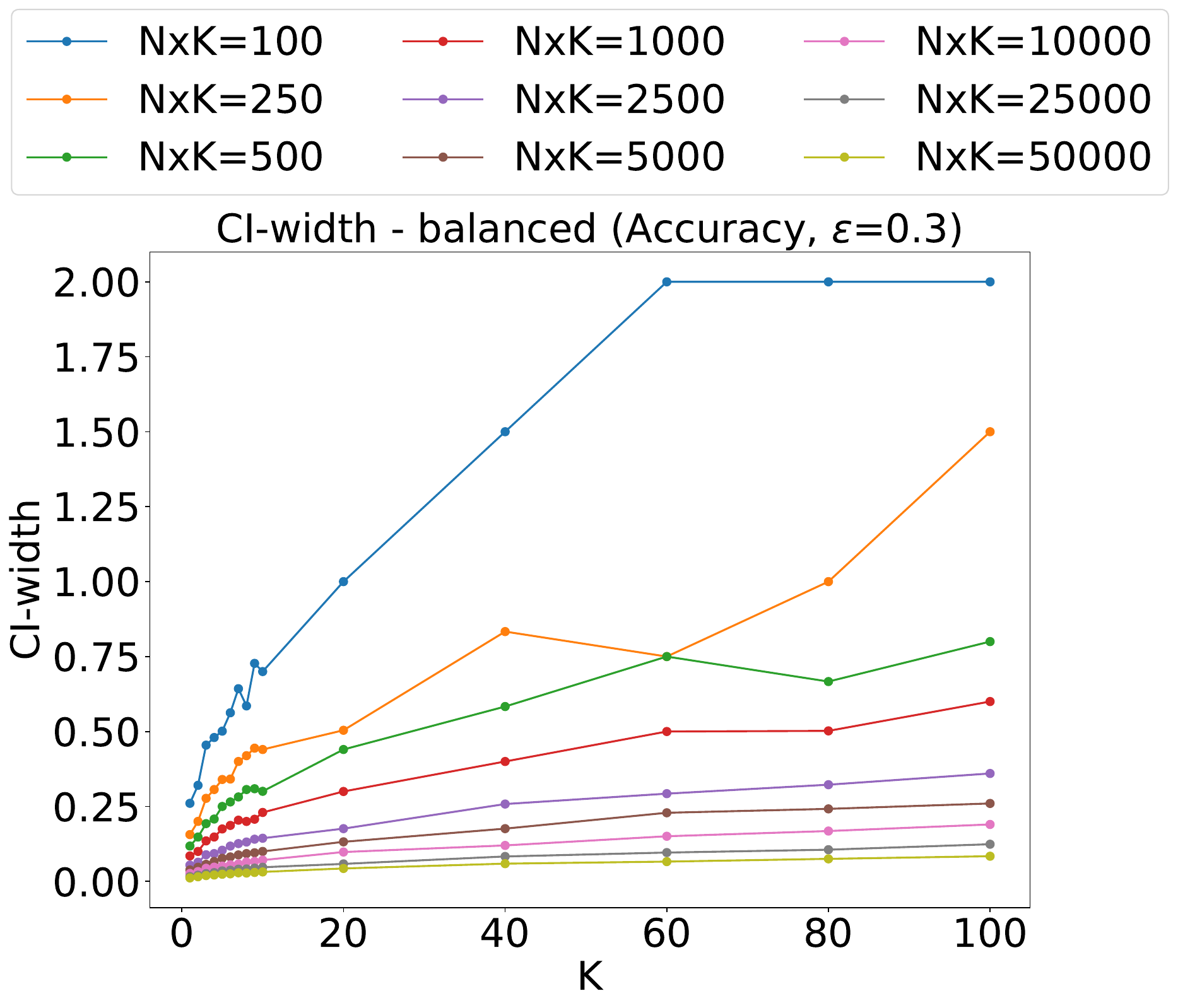}
    \caption{$\epsilon = 0.3$}
    \label{fig:uniform_ci_accuracy_cat2_e03}
  \end{subfigure} \hfill
  \begin{subfigure}[b]{0.24\linewidth}
    \centering
    \includegraphics[width=\linewidth]{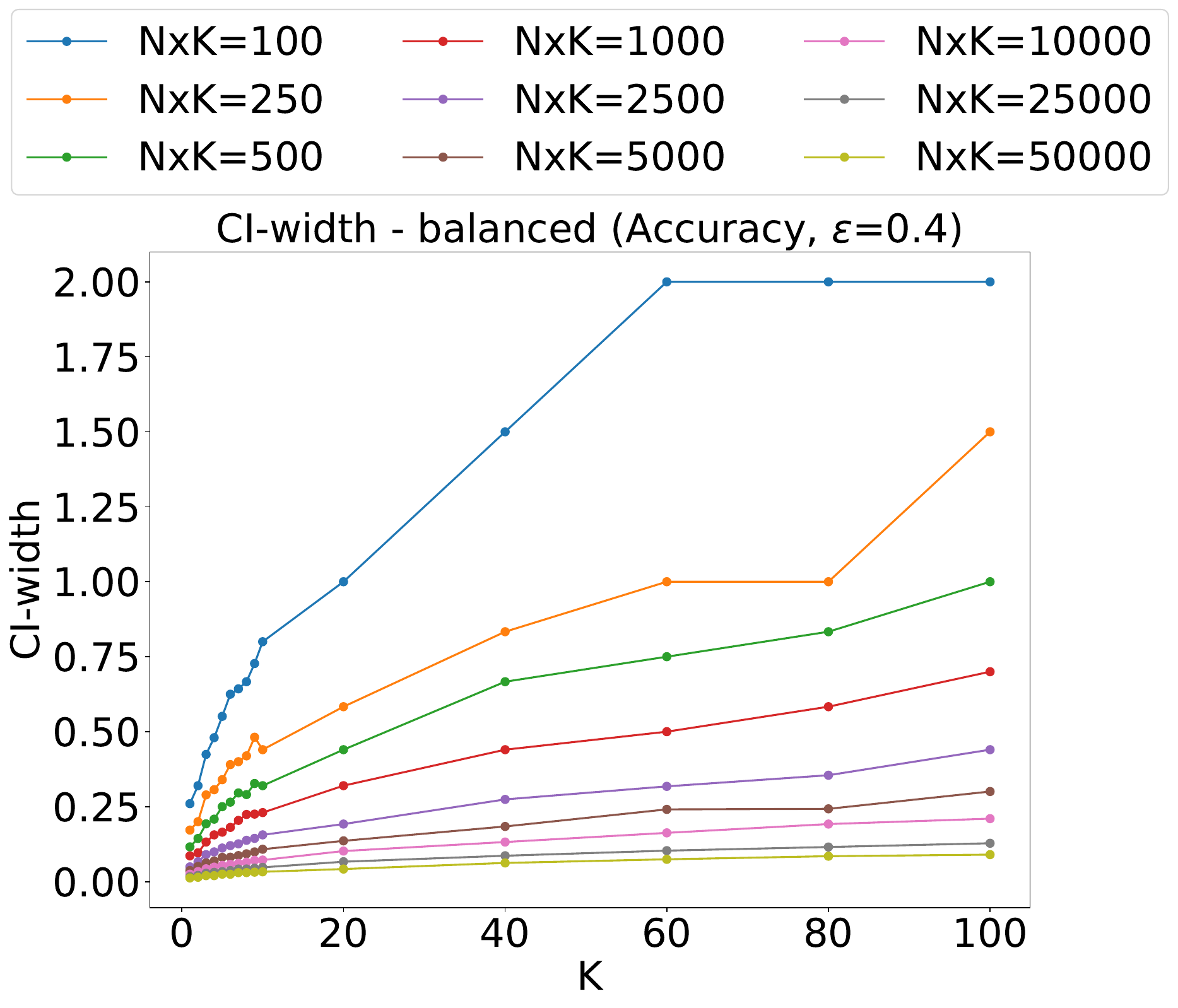}
    \caption{$\epsilon = 0.4$}
    \label{fig:uniform_ci_accuracy_cat2_e04}
  \end{subfigure}
  \caption{CI-width plots for balanced alphas with Accuracy as the metric ($M=2$)}
  \label{fig:uniform_ci_accuracy_cat2}
\end{figure*}

\begin{figure*}
  \centering
  \begin{subfigure}[b]{0.24\linewidth}
    \centering
    \includegraphics[width=\linewidth]{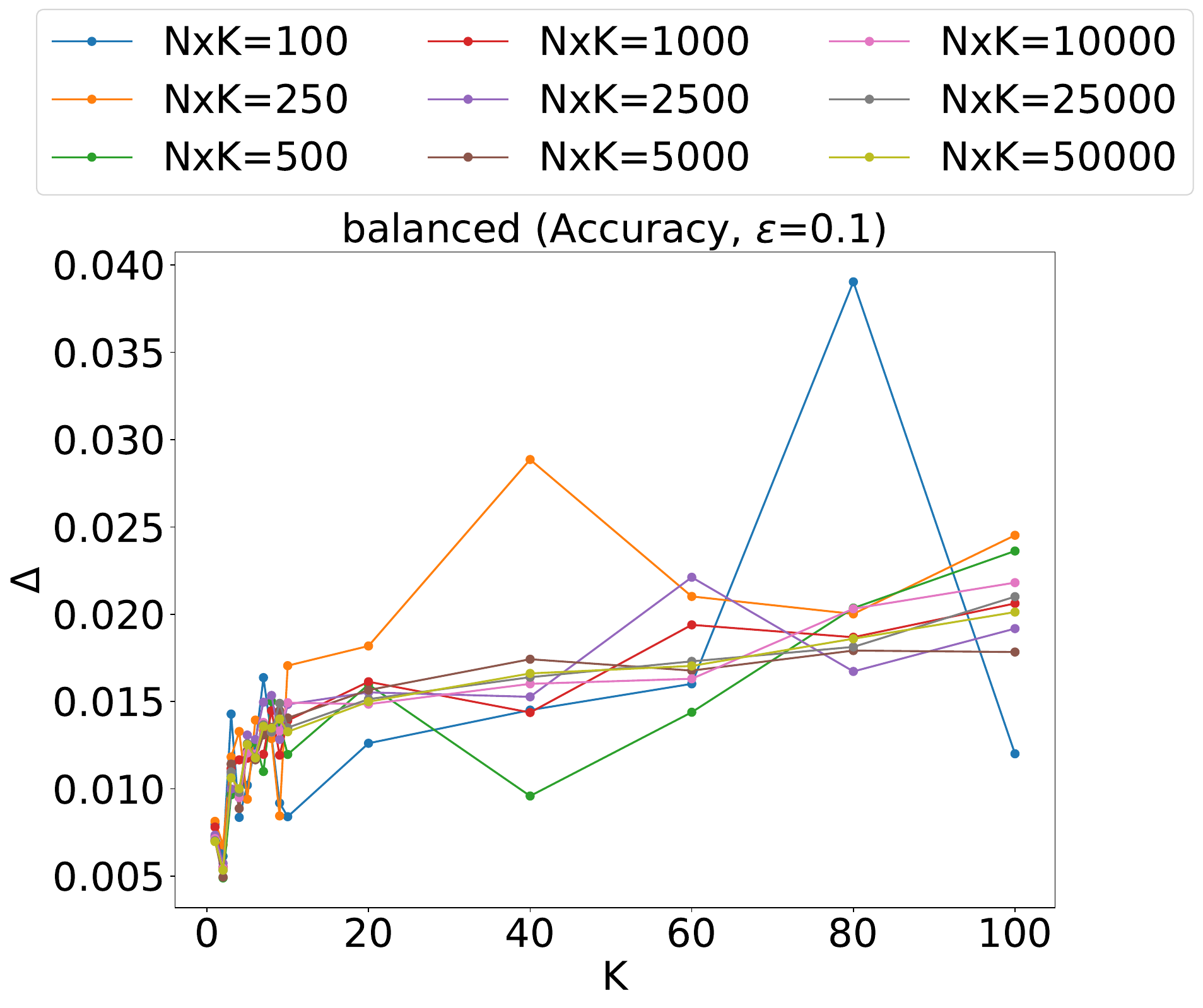}
    \caption{$\epsilon = 0.1$}
    \label{fig:uniform_delta_accuracy_cat2_e01}
  \end{subfigure} \hfill
  \begin{subfigure}[b]{0.24\linewidth}
    \centering
    \includegraphics[width=\linewidth]{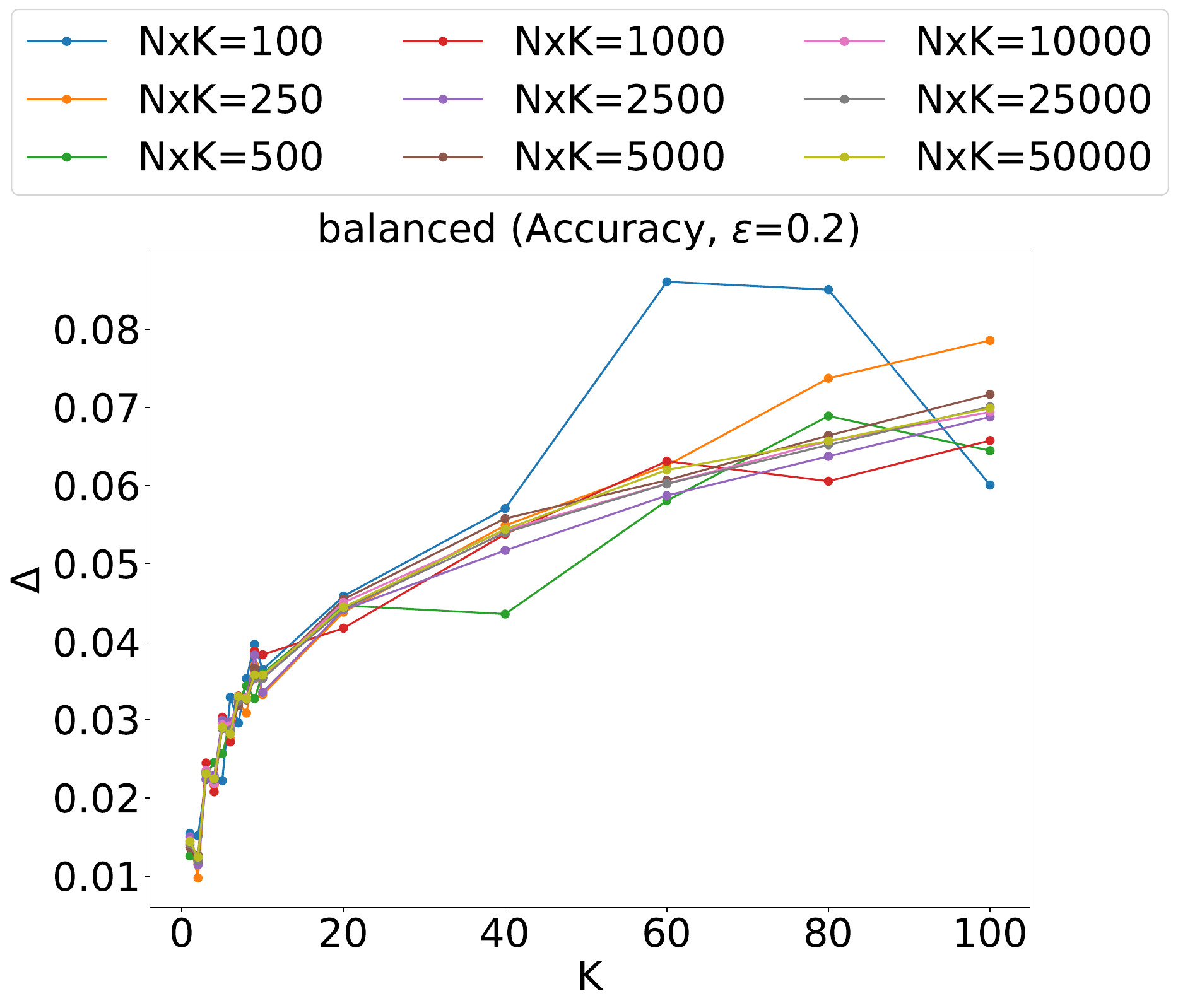}
    \caption{$\epsilon = 0.2$}
    \label{fig:uniform_delta_accuracy_cat2_e02}
  \end{subfigure} \hfill
  \begin{subfigure}[b]{0.24\linewidth}
    \centering
    \includegraphics[width=\linewidth]{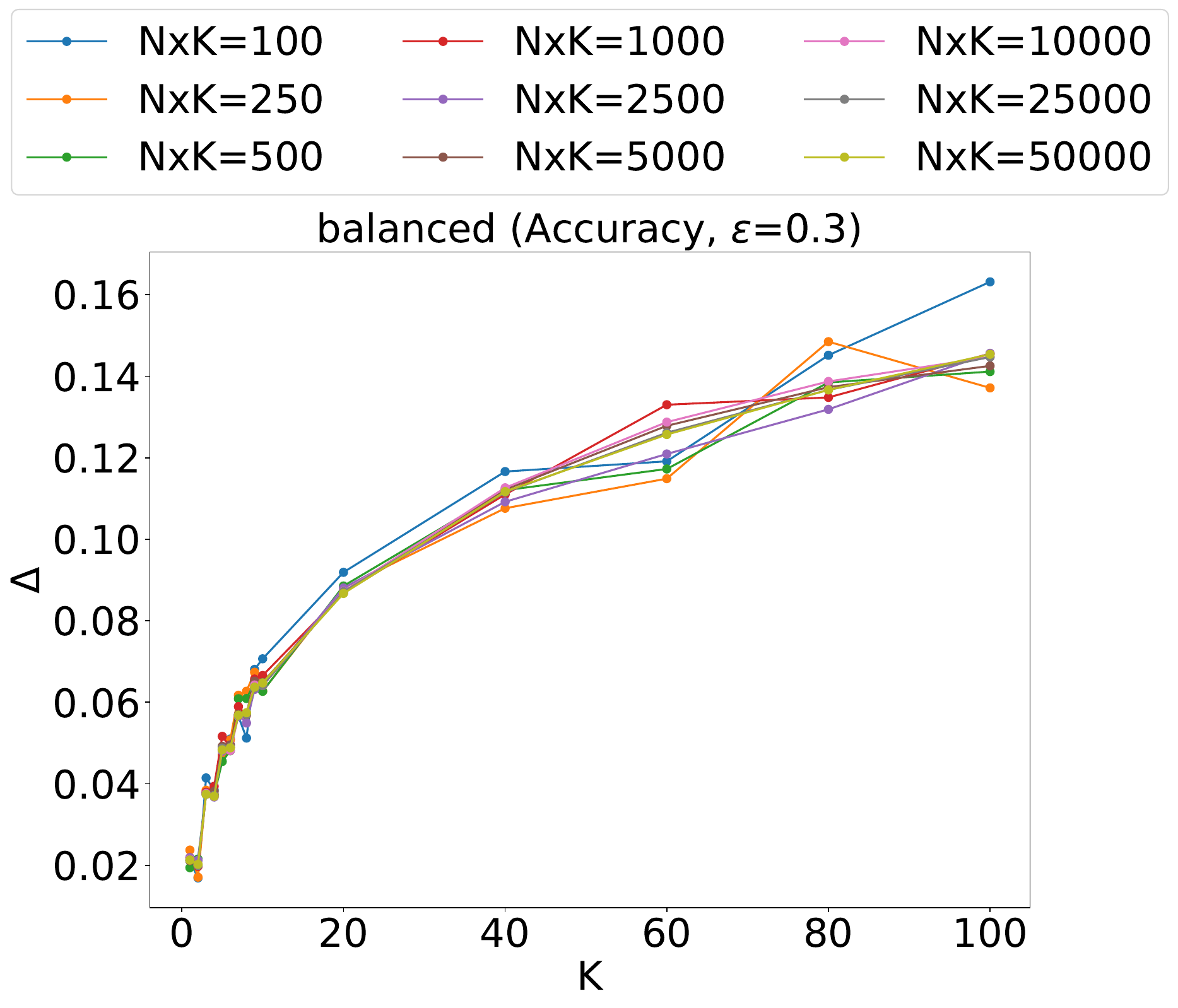}
    \caption{$\epsilon = 0.3$}
    \label{fig:uniform_delta_accuracy_cat2_e03}
  \end{subfigure} \hfill
  \begin{subfigure}[b]{0.24\linewidth}
    \centering
    \includegraphics[width=\linewidth]{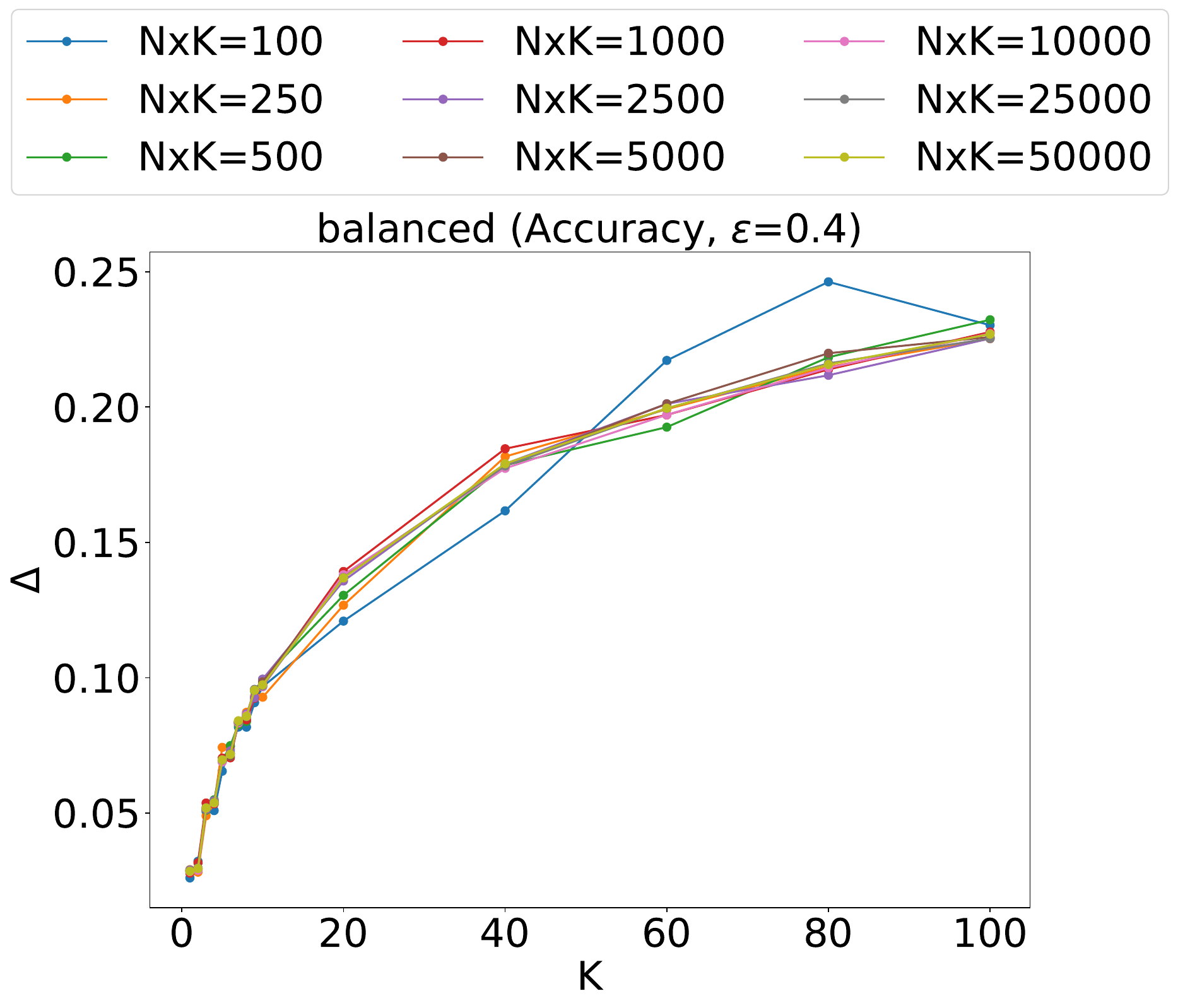}
    \caption{$\epsilon = 0.4$}
    \label{fig:uniform_delta_accuracy_cat2_e04}
  \end{subfigure}
  \caption{Effect sizes ($\Delta$) for balanced alphas with Accuracy as the metric ($M=2$)}
  \label{fig:uniform_delta_accuracy_cat2}
\end{figure*}

\begin{figure*}
  \centering
  \begin{subfigure}[b]{0.24\linewidth}
    \centering
    \includegraphics[width=\linewidth]{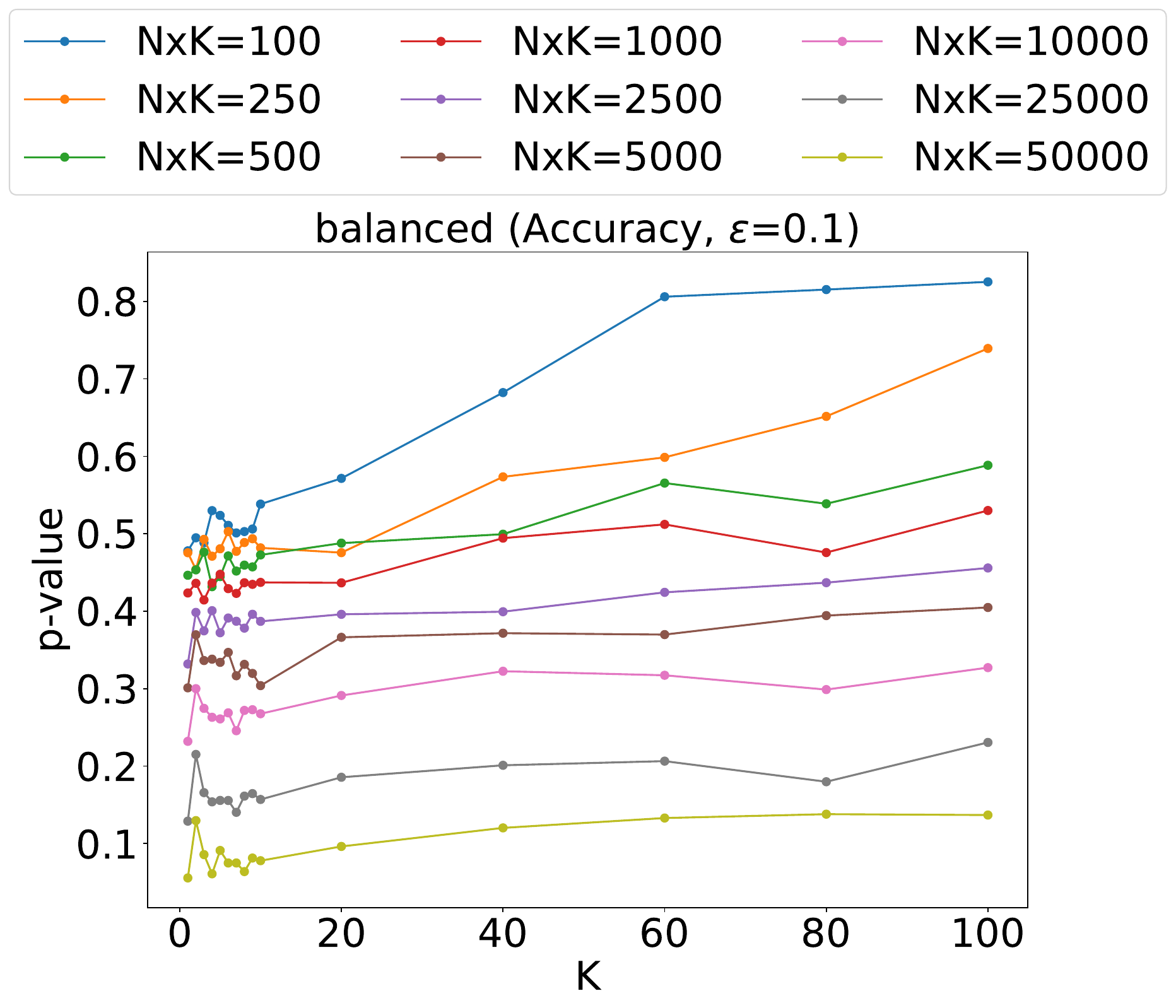}
    \caption{$\epsilon = 0.1$}
    \label{fig:uniform_accuracy_cat3_e01}
  \end{subfigure} \hfill
  \begin{subfigure}[b]{0.24\linewidth}
    \centering
    \includegraphics[width=\linewidth]{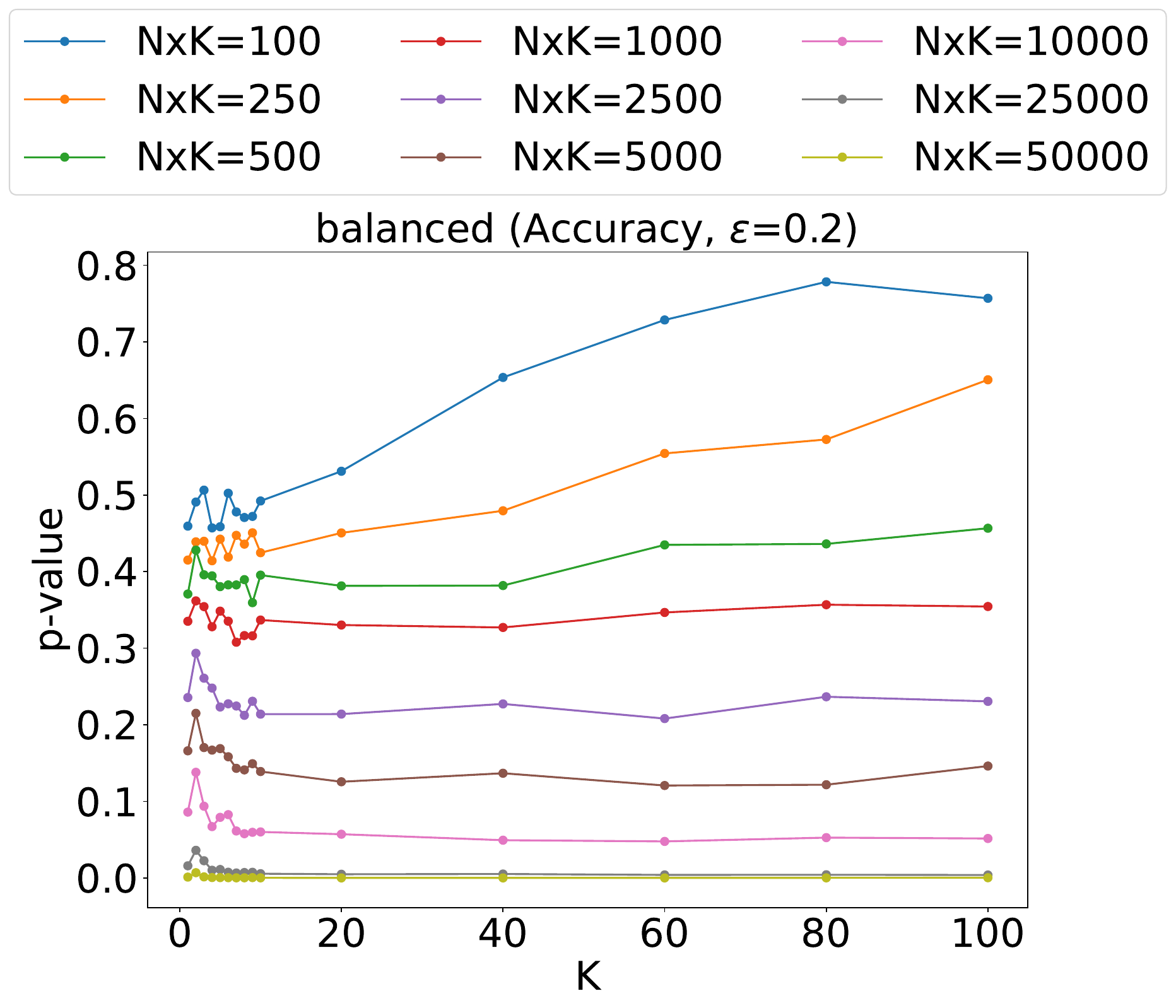}
    \caption{$\epsilon = 0.2$}
    \label{fig:uniform_accuracy_cat3_e02}
  \end{subfigure} \hfill
  \begin{subfigure}[b]{0.24\linewidth}
    \centering
    \includegraphics[width=\linewidth]{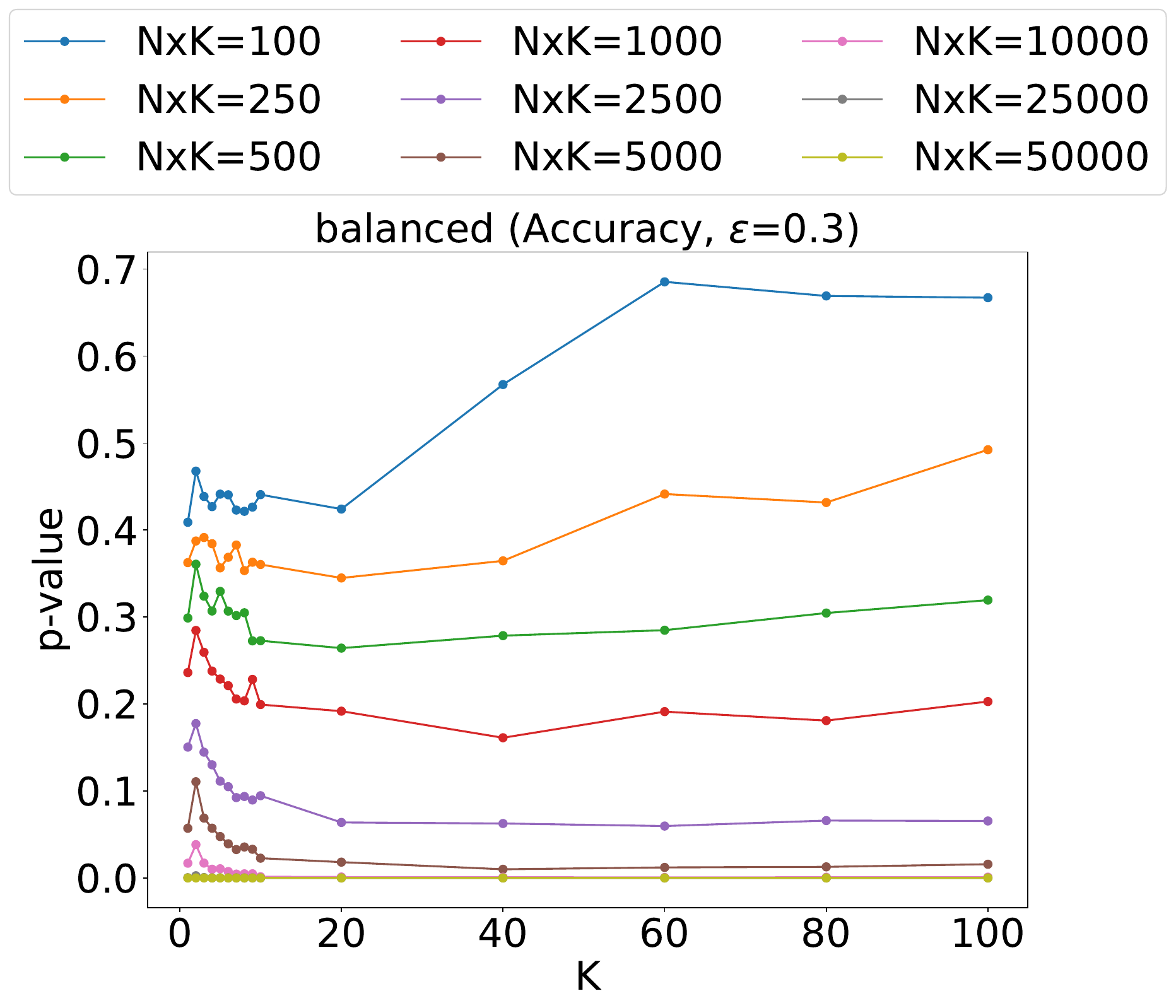}
    \caption{$\epsilon = 0.3$}
    \label{fig:uniform_accuracy_cat3_e03}
  \end{subfigure} \hfill
  \begin{subfigure}[b]{0.24\linewidth}
    \centering
    \includegraphics[width=\linewidth]{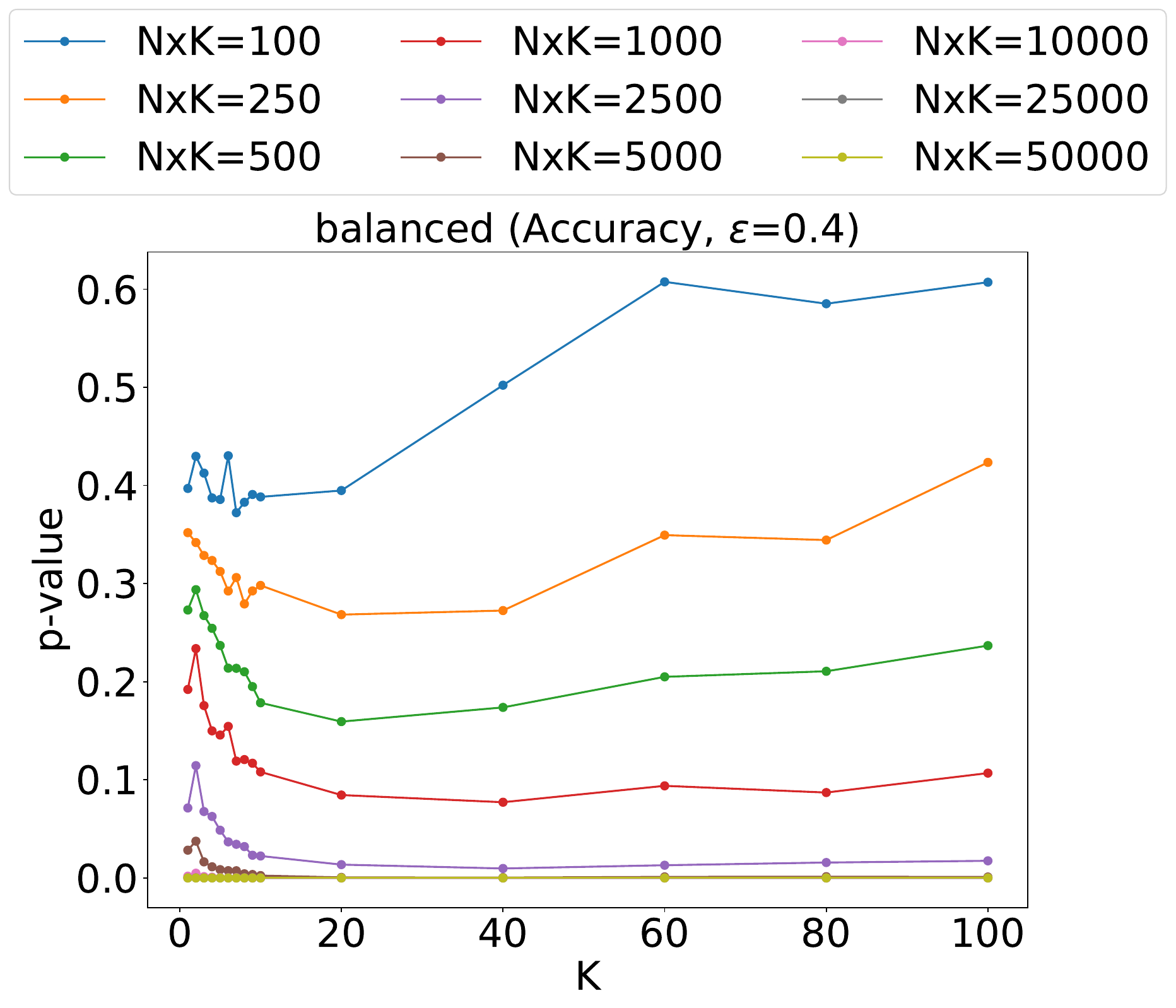}
    \caption{$\epsilon = 0.4$}
    \label{fig:uniform_accuracy_cat3_e04}
  \end{subfigure}
  \caption{P-value plots for balanced aplhas with Accuracy as the metric ($M=3$)}
  \label{fig:uniform_accuracy_cat3}
\end{figure*}

\begin{figure*}
  \centering
  \begin{subfigure}[b]{0.24\linewidth}
    \centering
    \includegraphics[width=\linewidth]{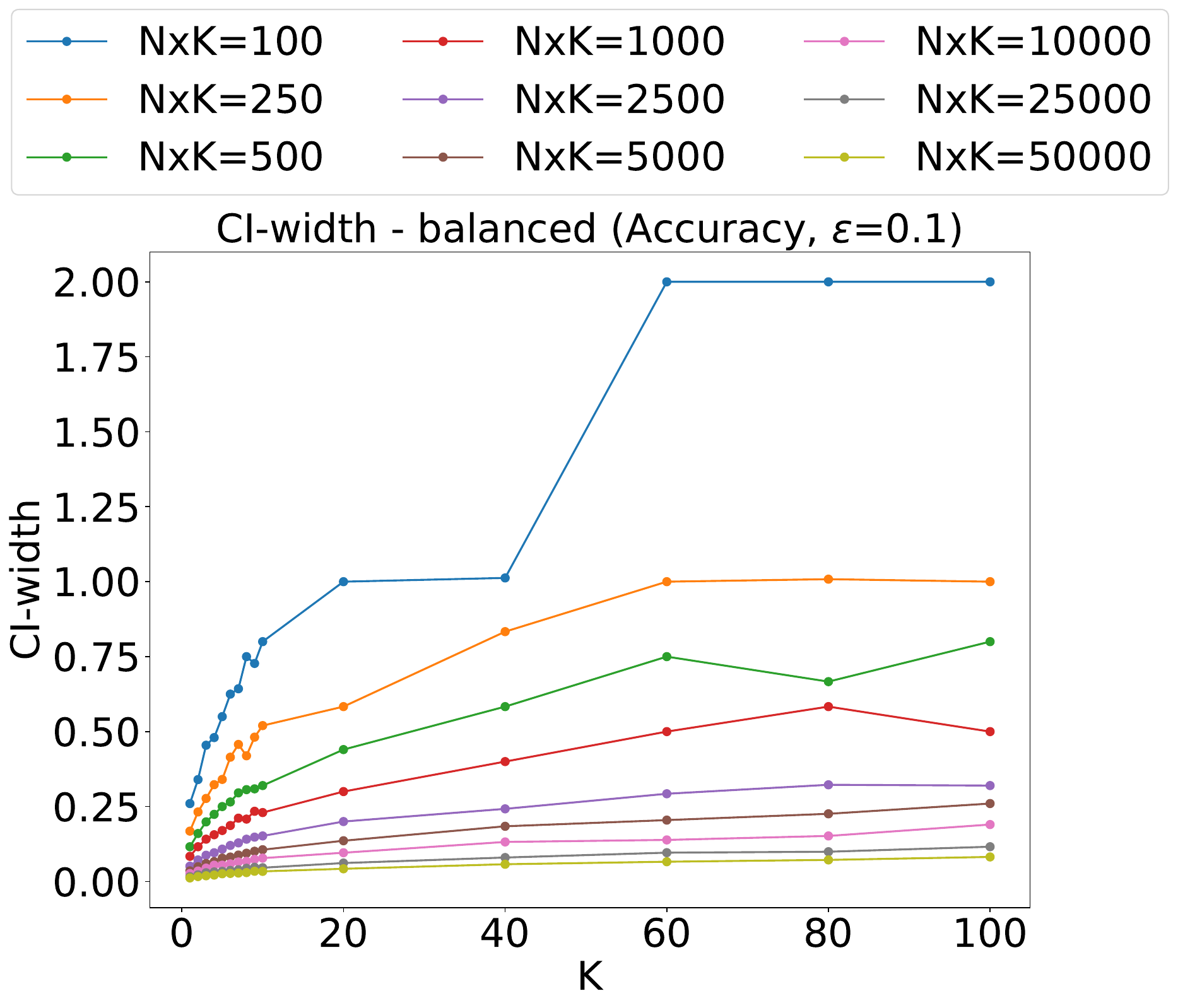}
    \caption{$\epsilon = 0.1$}
    \label{fig:uniform_ci_accuracy_cat3_e01}
  \end{subfigure} \hfill
  \begin{subfigure}[b]{0.24\linewidth}
    \centering
    \includegraphics[width=\linewidth]{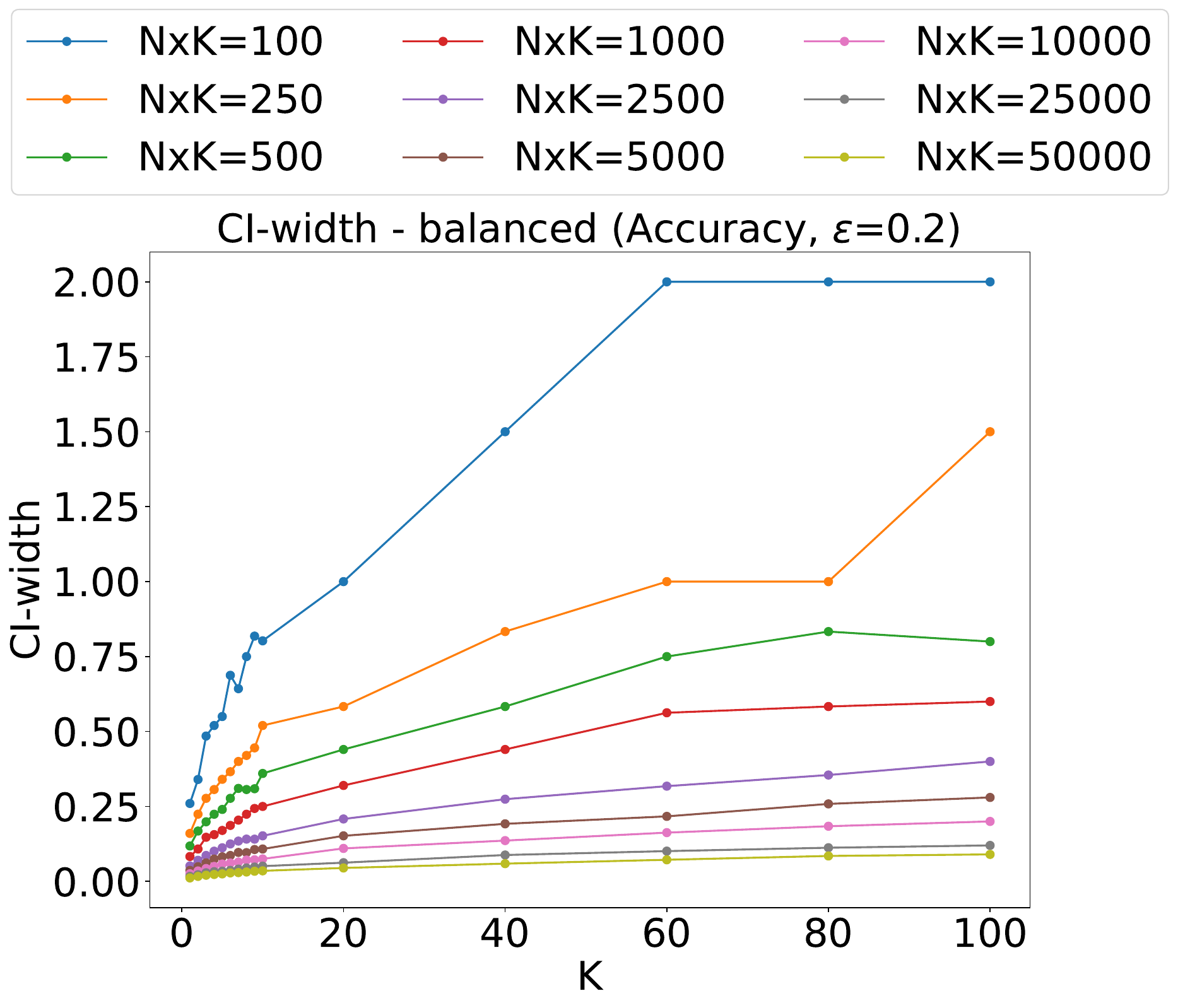}
    \caption{$\epsilon = 0.2$}
    \label{fig:uniform_ci_accuracy_cat3_e02}
  \end{subfigure} \hfill
  \begin{subfigure}[b]{0.24\linewidth}
    \centering
    \includegraphics[width=\linewidth]{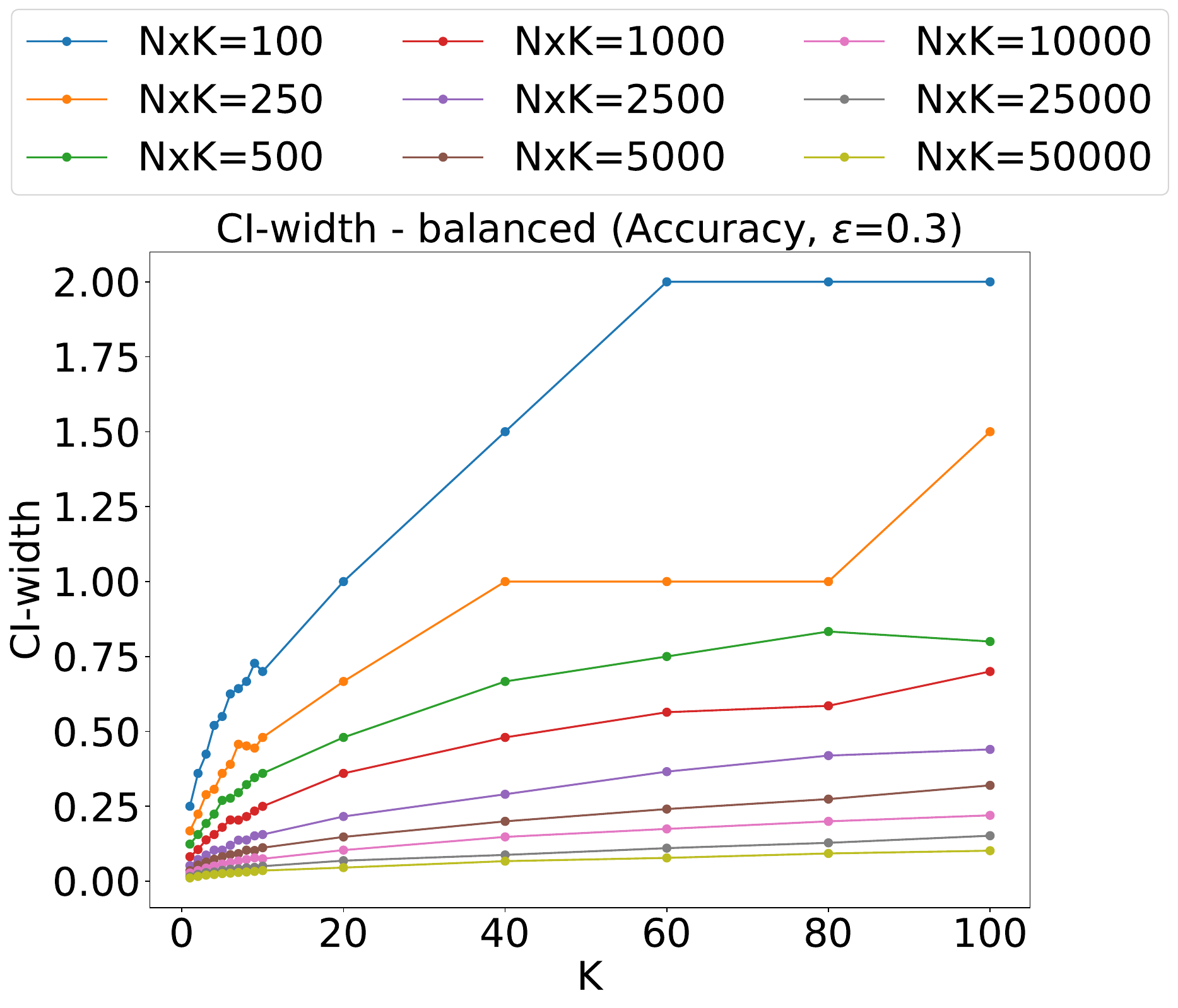}
    \caption{$\epsilon = 0.3$}
    \label{fig:uniform_ci_accuracy_cat3_e03}
  \end{subfigure} \hfill
  \begin{subfigure}[b]{0.24\linewidth}
    \centering
    \includegraphics[width=\linewidth]{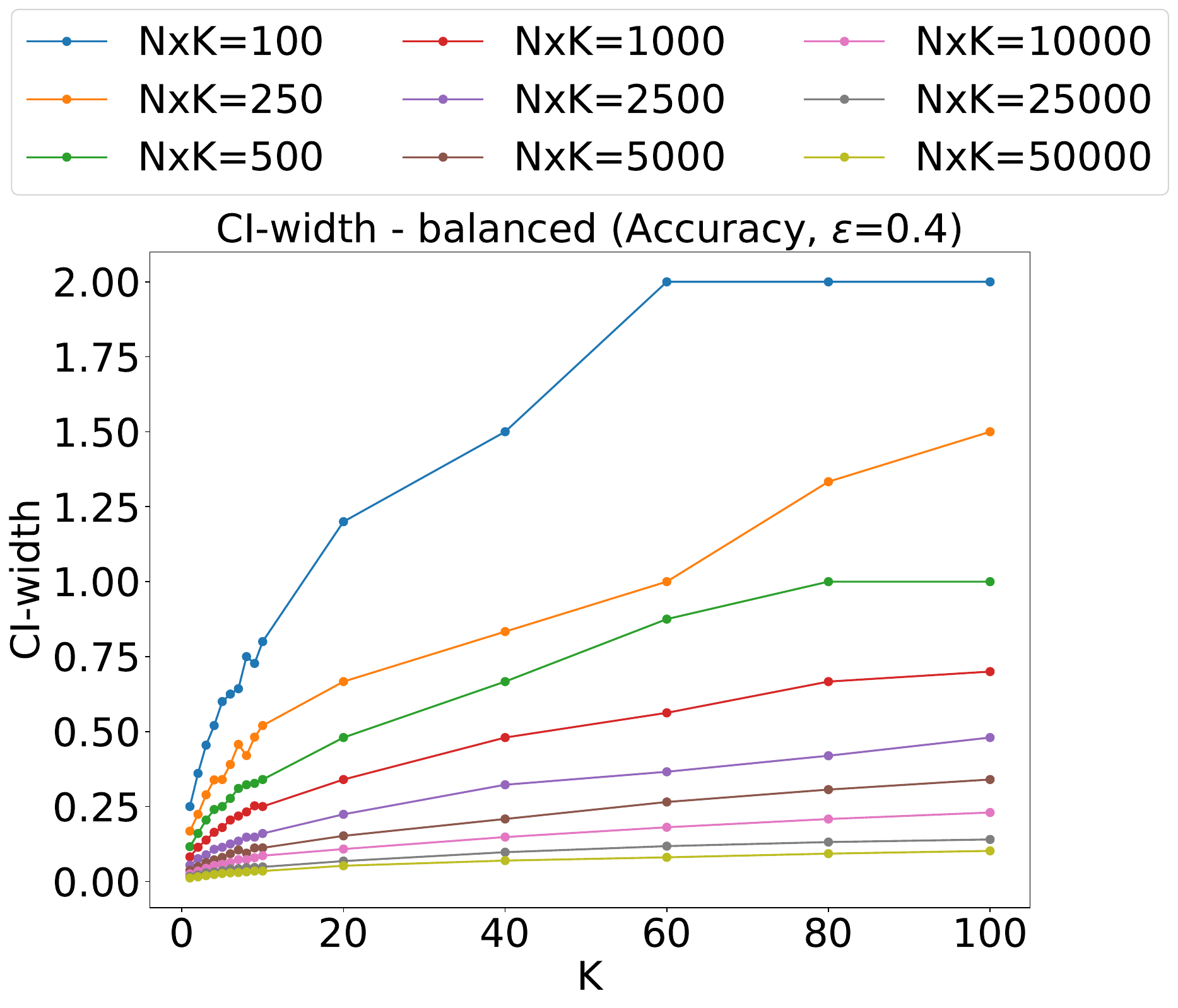}
    \caption{$\epsilon = 0.4$}
    \label{fig:uniform_ci_accuracy_cat3_e04}
  \end{subfigure}
  \caption{CI-width plots for balanced alphas with Accuracy as the metric ($M=3$)}
  \label{fig:uniform_ci_accuracy_cat3}
\end{figure*}

\begin{figure*}
  \centering
  \begin{subfigure}[b]{0.24\linewidth}
    \centering
    \includegraphics[width=\linewidth]{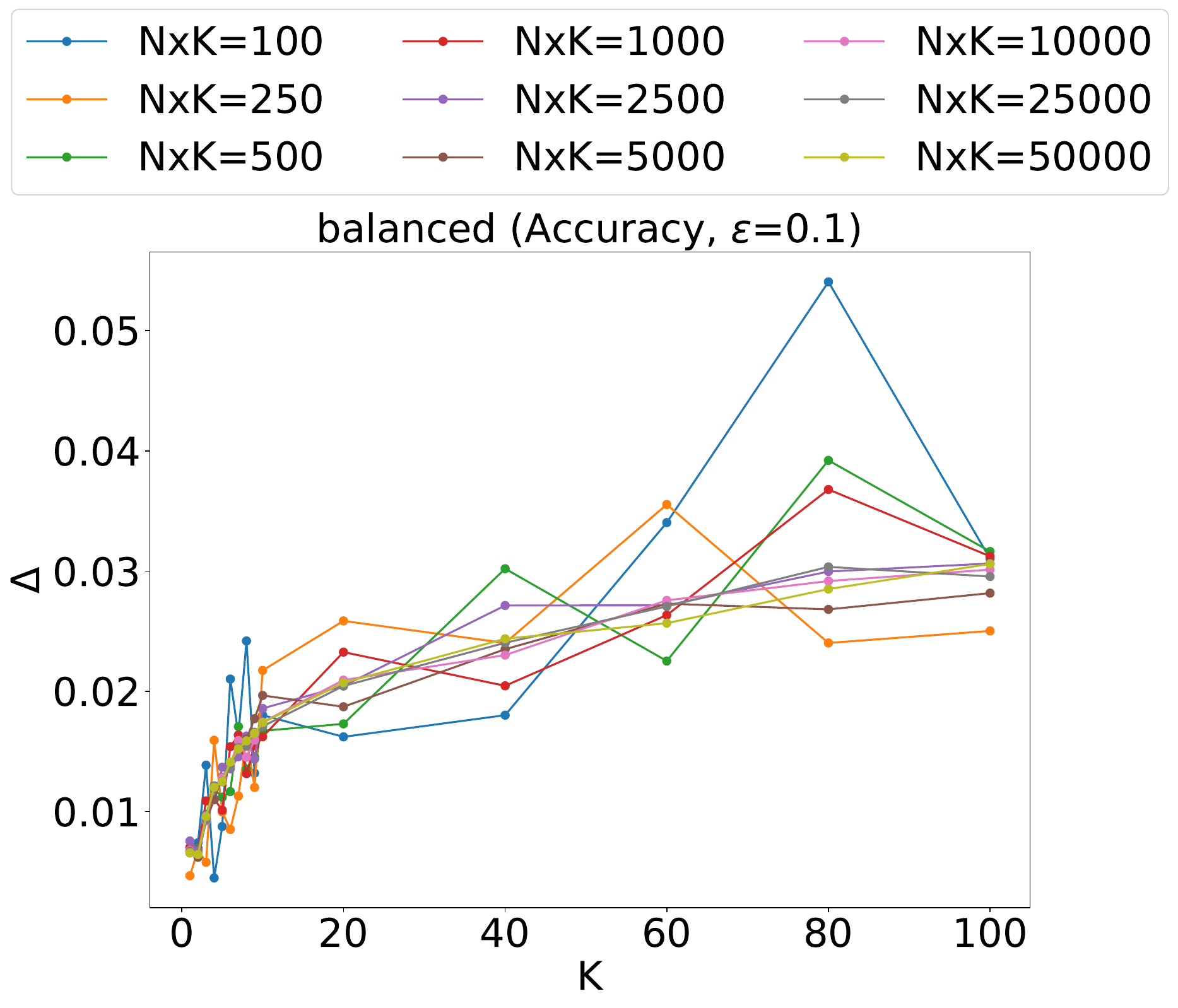}
    \caption{$\epsilon = 0.1$}
    \label{fig:uniform_delta_accuracy_cat3_e01}
  \end{subfigure} \hfill
  \begin{subfigure}[b]{0.24\linewidth}
    \centering
    \includegraphics[width=\linewidth]{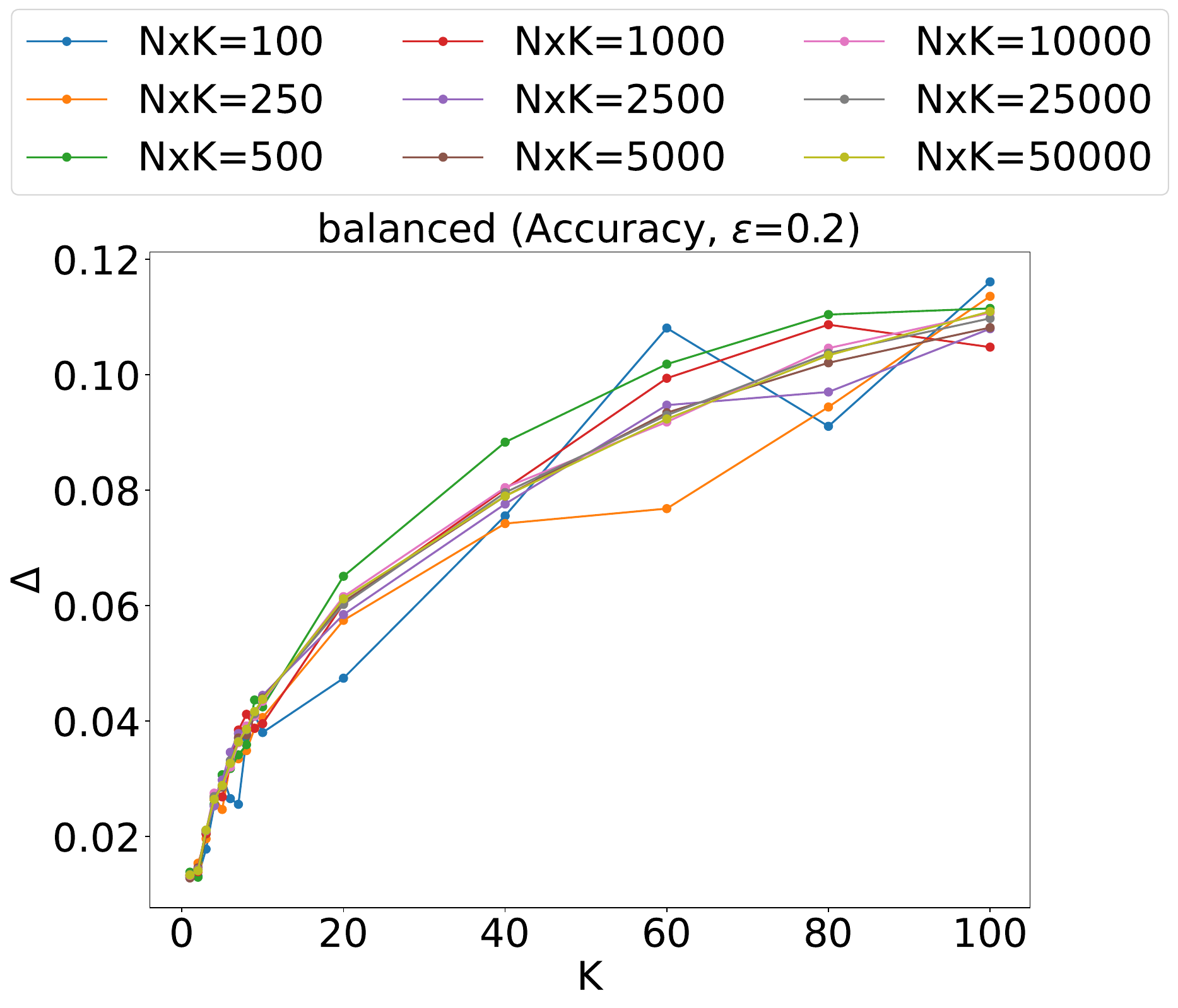}
    \caption{$\epsilon = 0.2$}
    \label{fig:uniform_delta_accuracy_cat3_e02}
  \end{subfigure} \hfill
  \begin{subfigure}[b]{0.24\linewidth}
    \centering
    \includegraphics[width=\linewidth]{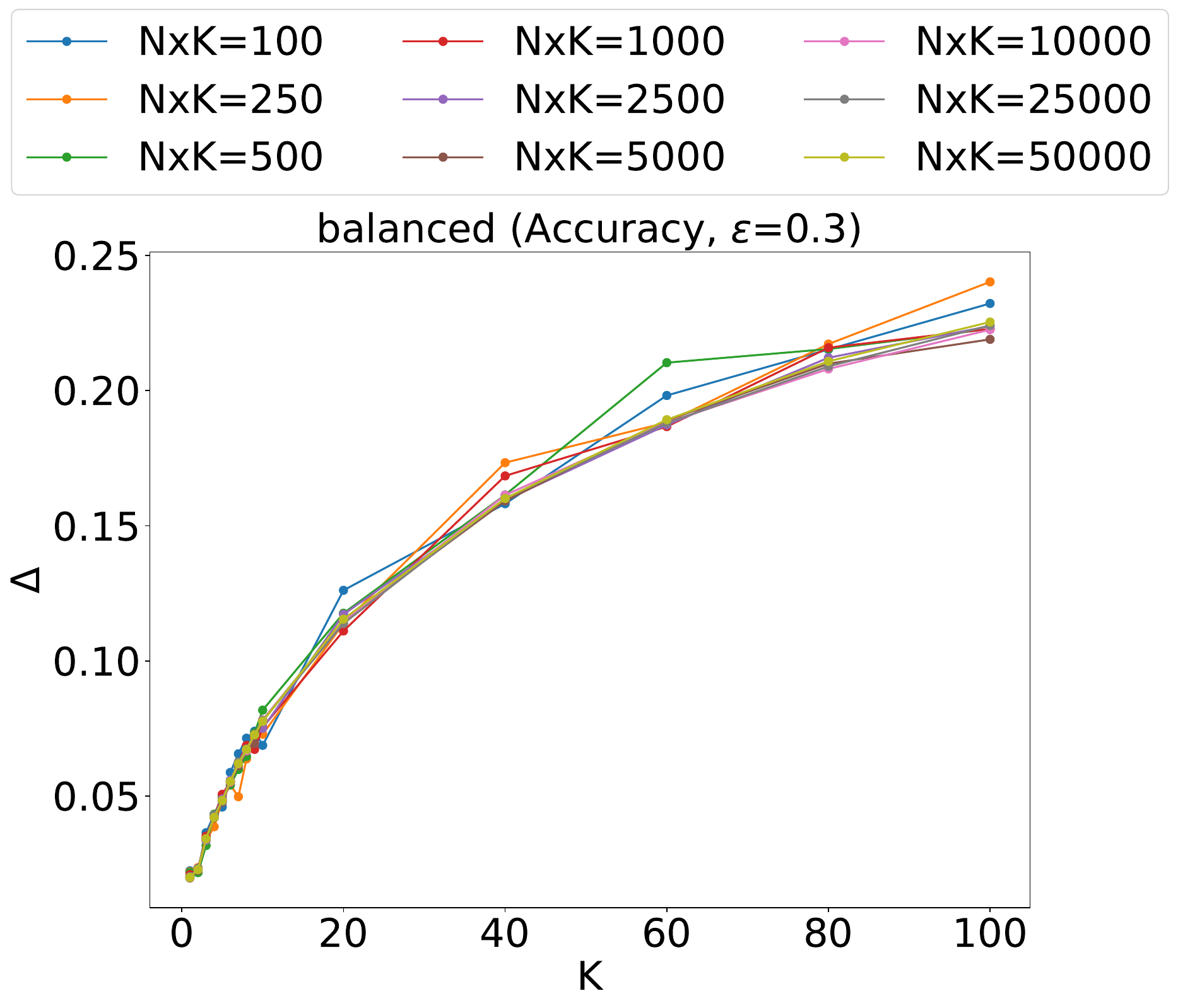}
    \caption{$\epsilon = 0.3$}
    \label{fig:uniform_delta_accuracy_cat3_e03}
  \end{subfigure} \hfill
  \begin{subfigure}[b]{0.24\linewidth}
    \centering
    \includegraphics[width=\linewidth]{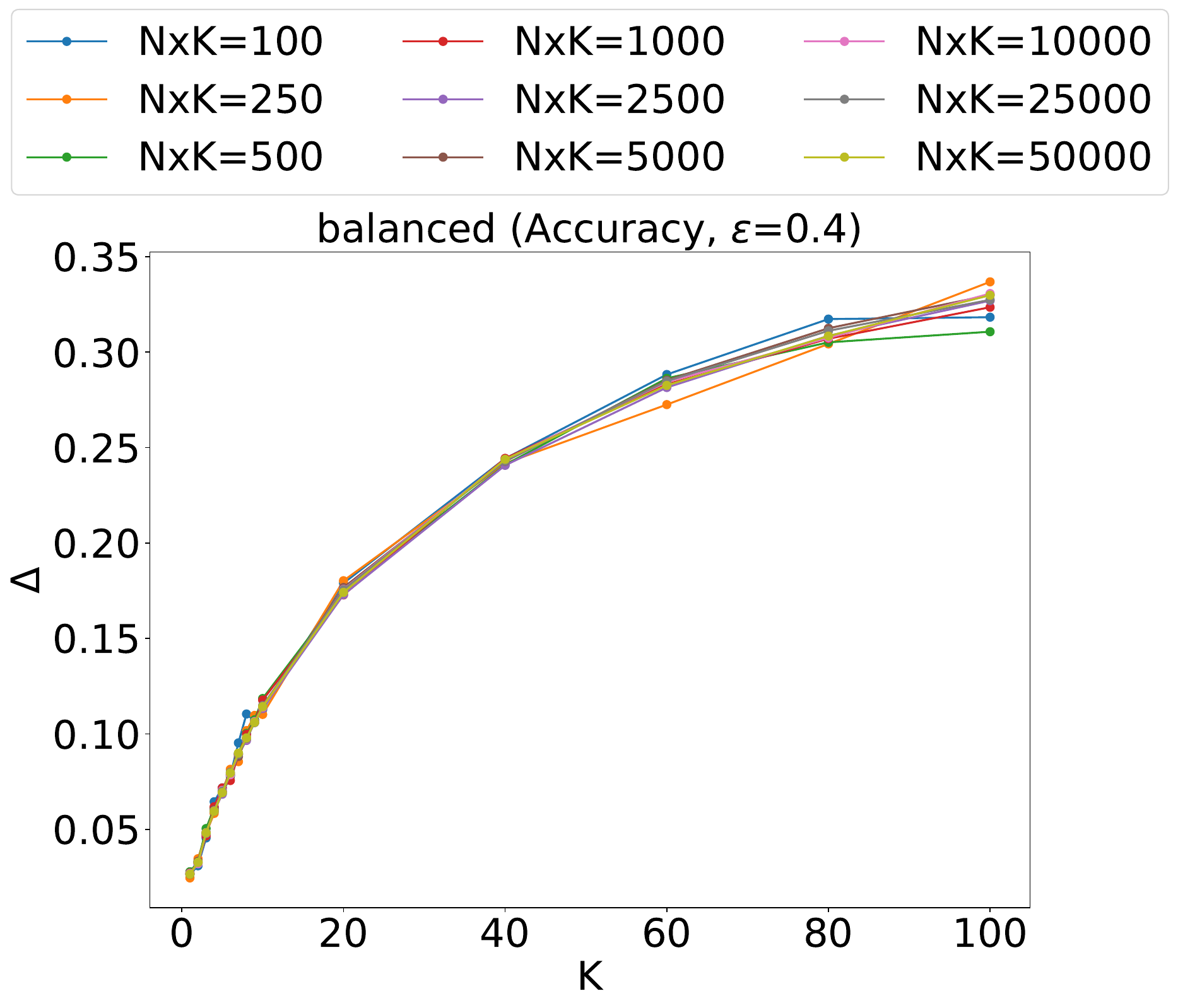}
    \caption{$\epsilon = 0.4$}
    \label{fig:uniform_delta_accuracy_cat3_e04}
  \end{subfigure}
  \caption{Effect sizes ($\Delta$) for balanced alphas with Accuracy as the metric ($M=3$)}
  \label{fig:uniform_delta_accuracy_cat3}
\end{figure*}

\begin{figure*}
  \centering
  \begin{subfigure}[b]{0.24\linewidth}
    \centering
    \includegraphics[width=\linewidth]{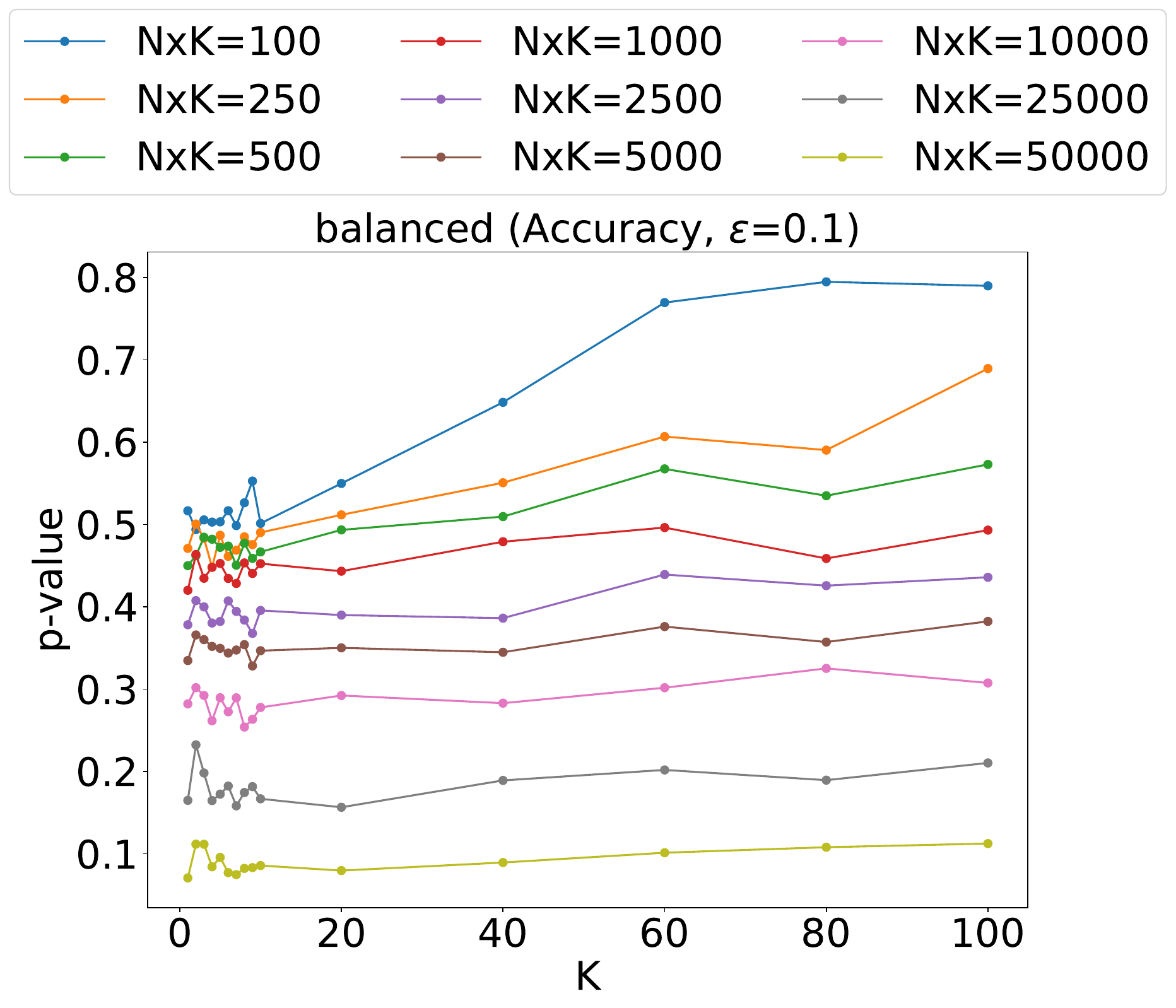}
    \caption{$\epsilon = 0.1$}
    \label{fig:uniform_accuracy_cat4_e01}
  \end{subfigure} \hfill
  \begin{subfigure}[b]{0.24\linewidth}
    \centering
    \includegraphics[width=\linewidth]{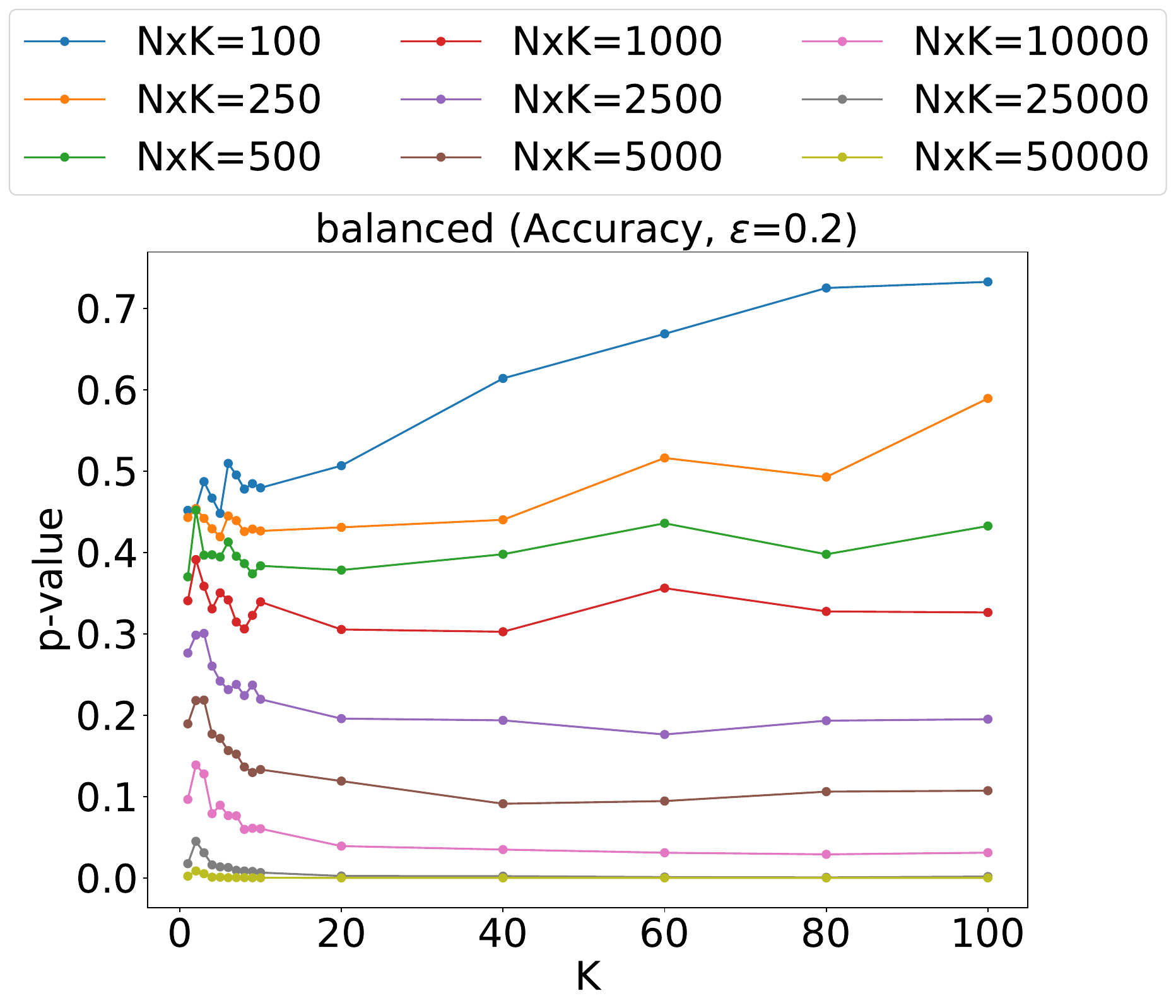}
    \caption{$\epsilon = 0.2$}
    \label{fig:uniform_accuracy_cat4_e02}
  \end{subfigure} \hfill
  \begin{subfigure}[b]{0.24\linewidth}
    \centering
    \includegraphics[width=\linewidth]{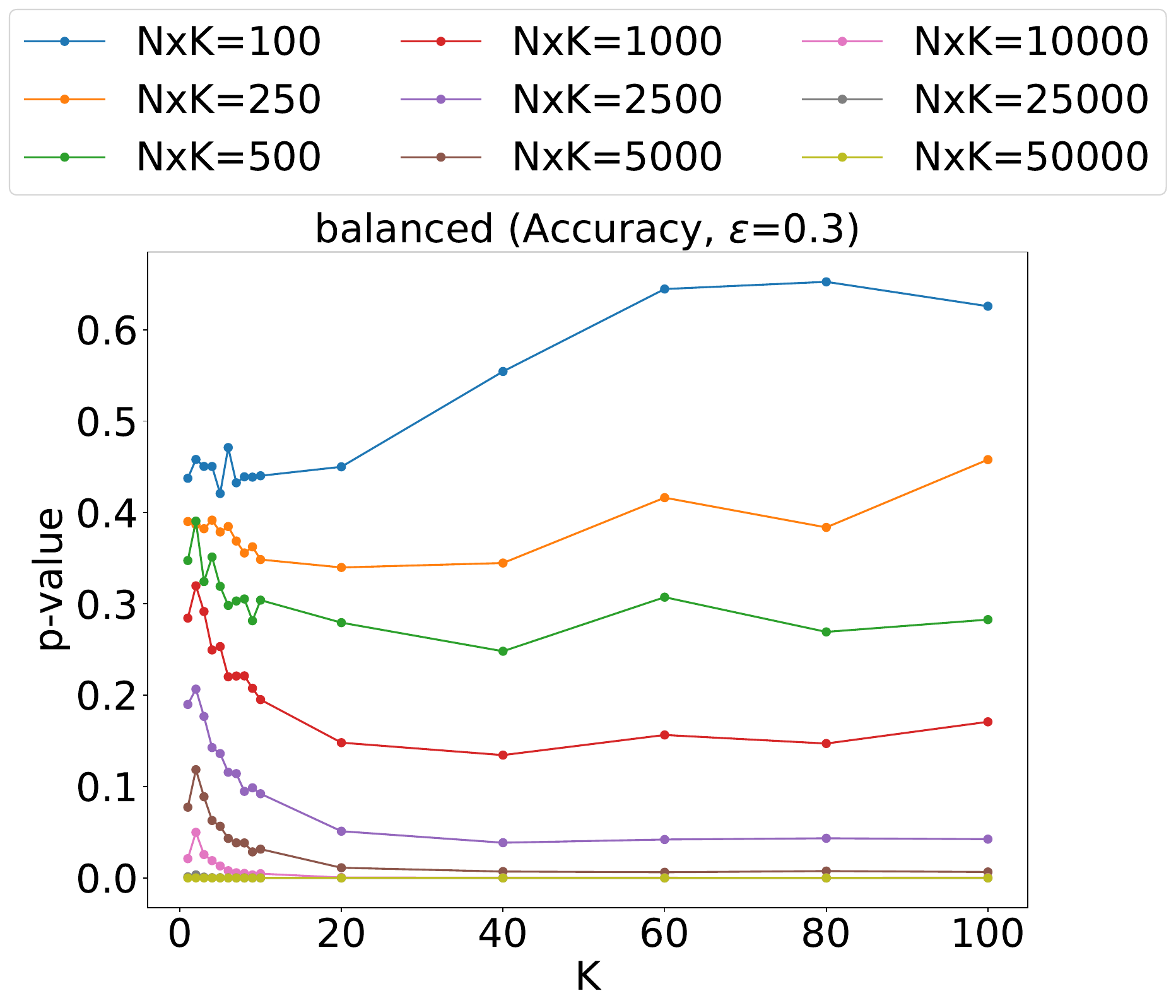}
    \caption{$\epsilon = 0.3$}
    \label{fig:uniform_accuracy_cat4_e03}
  \end{subfigure} \hfill
  \begin{subfigure}[b]{0.24\linewidth}
    \centering
    \includegraphics[width=\linewidth]{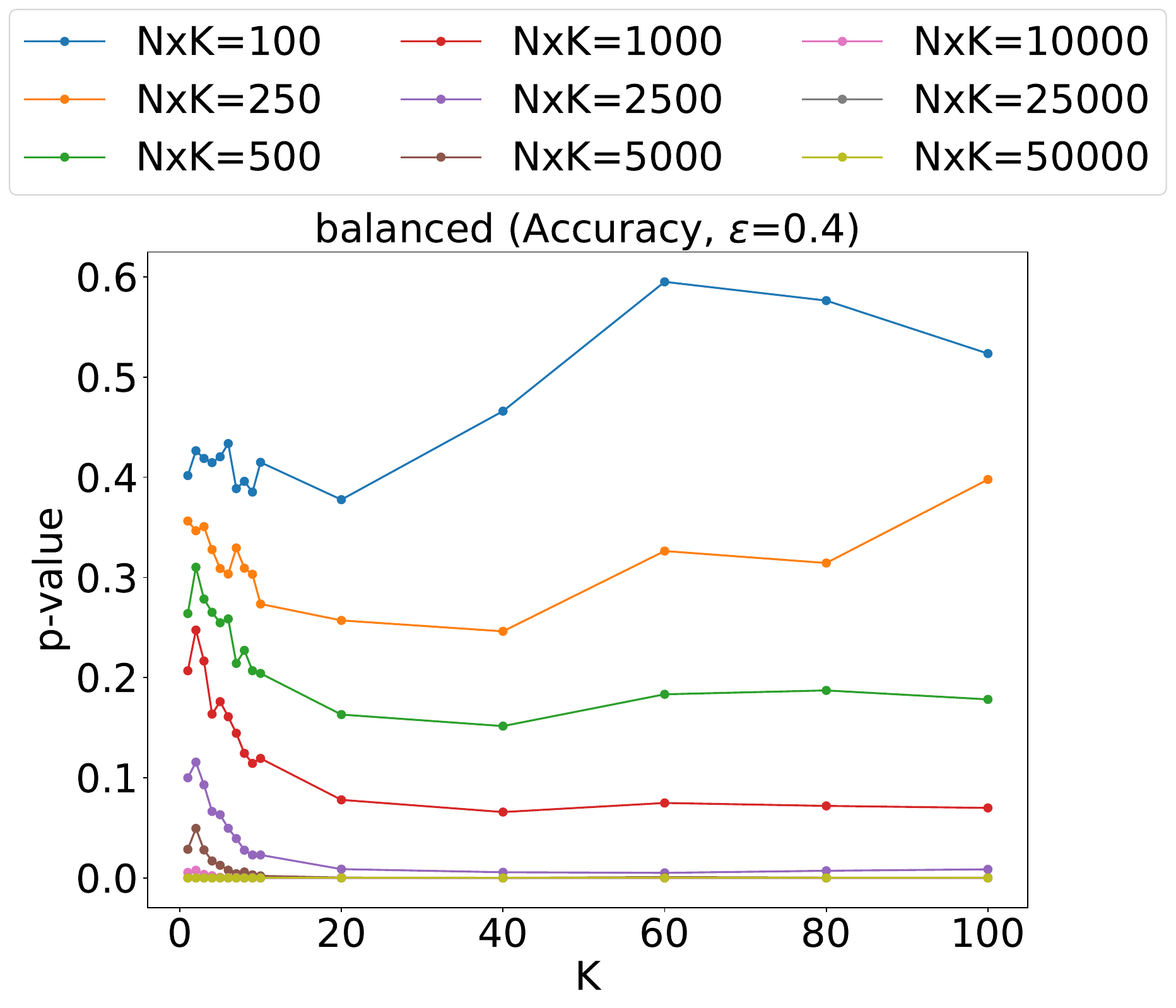}
    \caption{$\epsilon = 0.4$}
    \label{fig:uniform_accuracy_cat4_e04}
  \end{subfigure}
  \caption{P-value plots for balanced aplhas with Accuracy as the metric ($M=4$)}
  \label{fig:uniform_accuracy_cat4}
\end{figure*}

\begin{figure*}
  \centering
  \begin{subfigure}[b]{0.24\linewidth}
    \centering
    \includegraphics[width=\linewidth]{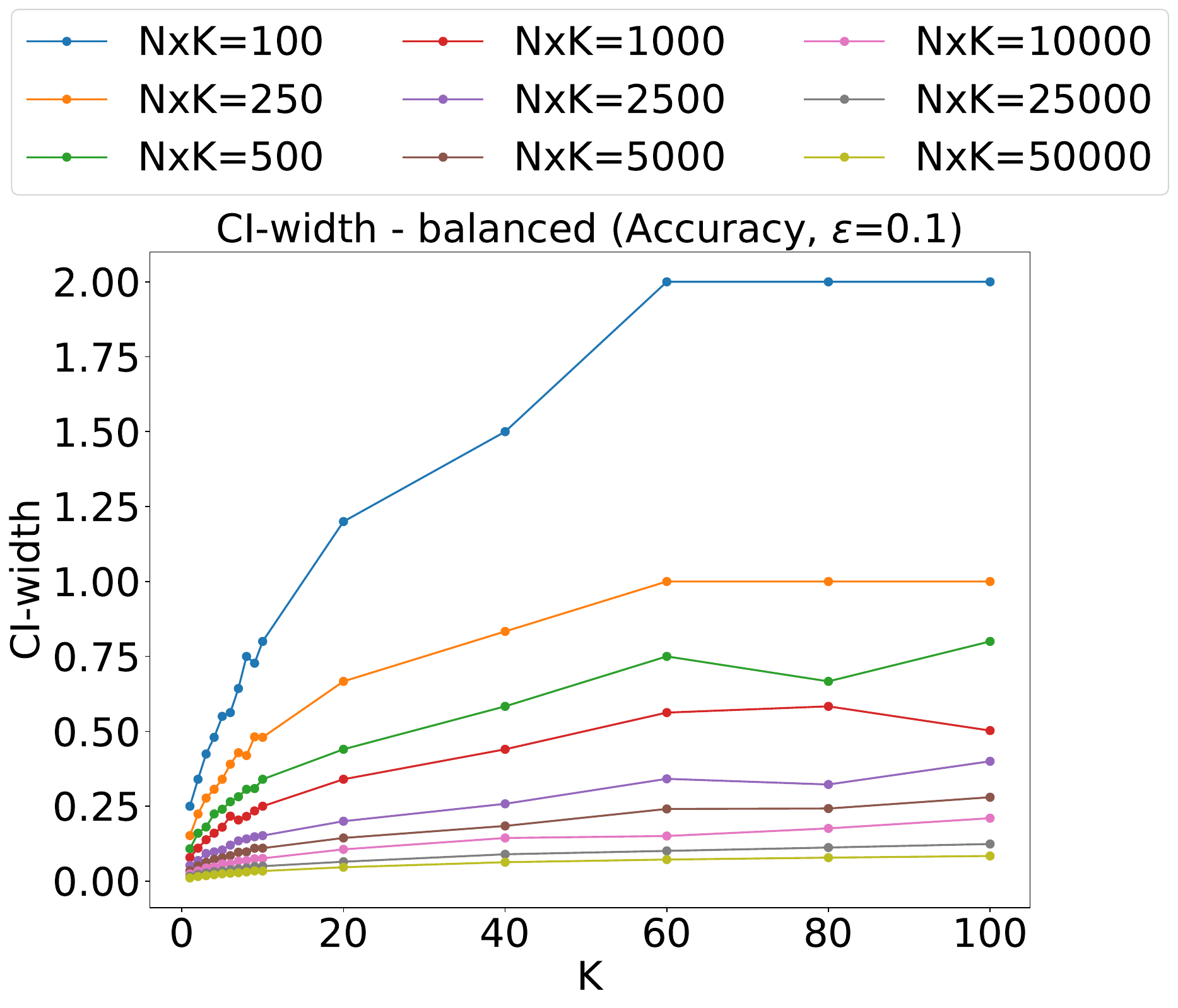}
    \caption{$\epsilon = 0.1$}
    \label{fig:uniform_ci_accuracy_cat4_e01}
  \end{subfigure} \hfill
  \begin{subfigure}[b]{0.24\linewidth}
    \centering
    \includegraphics[width=\linewidth]{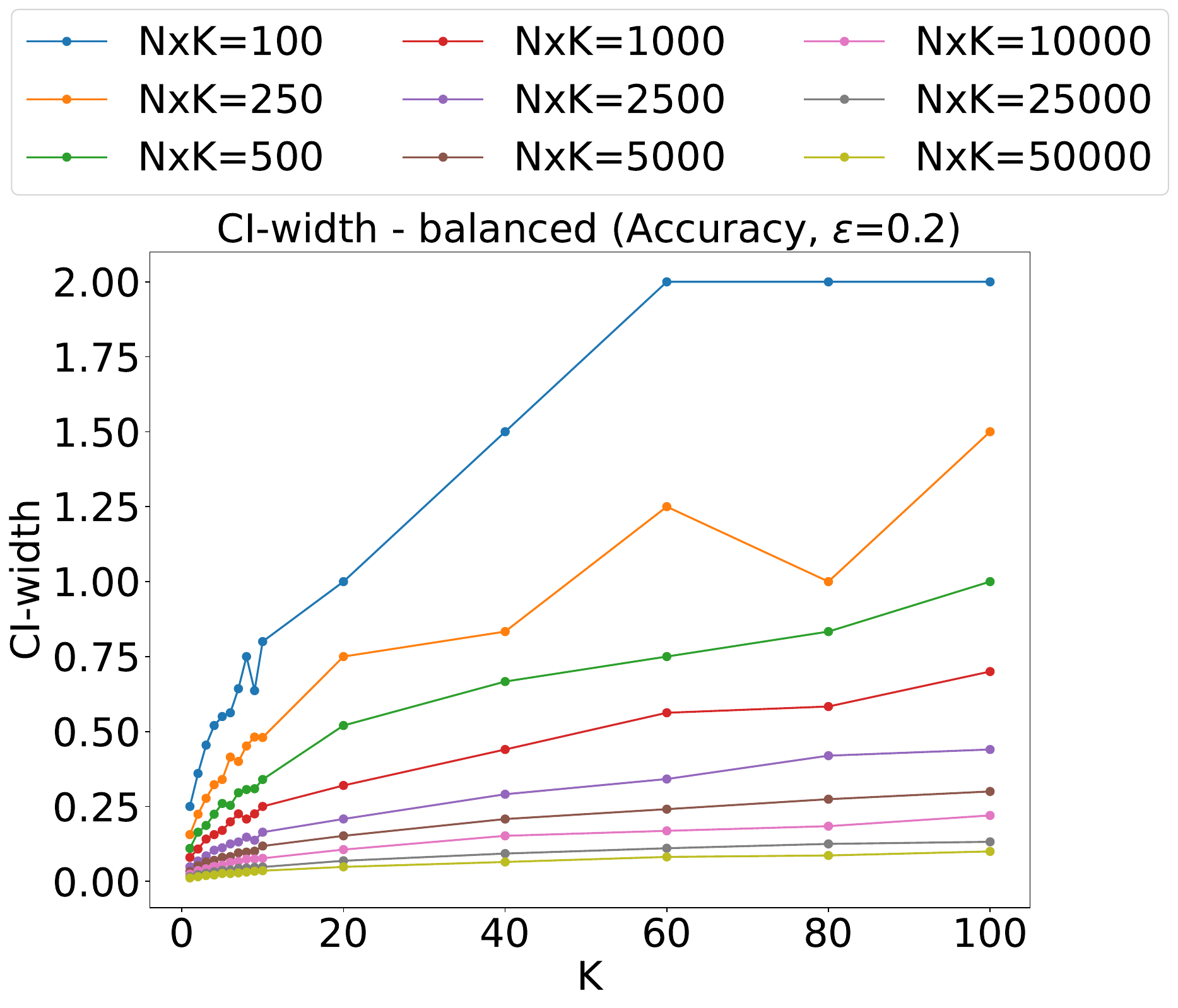}
    \caption{$\epsilon = 0.2$}
    \label{fig:uniform_ci_accuracy_cat4_e02}
  \end{subfigure} \hfill
  \begin{subfigure}[b]{0.24\linewidth}
    \centering
    \includegraphics[width=\linewidth]{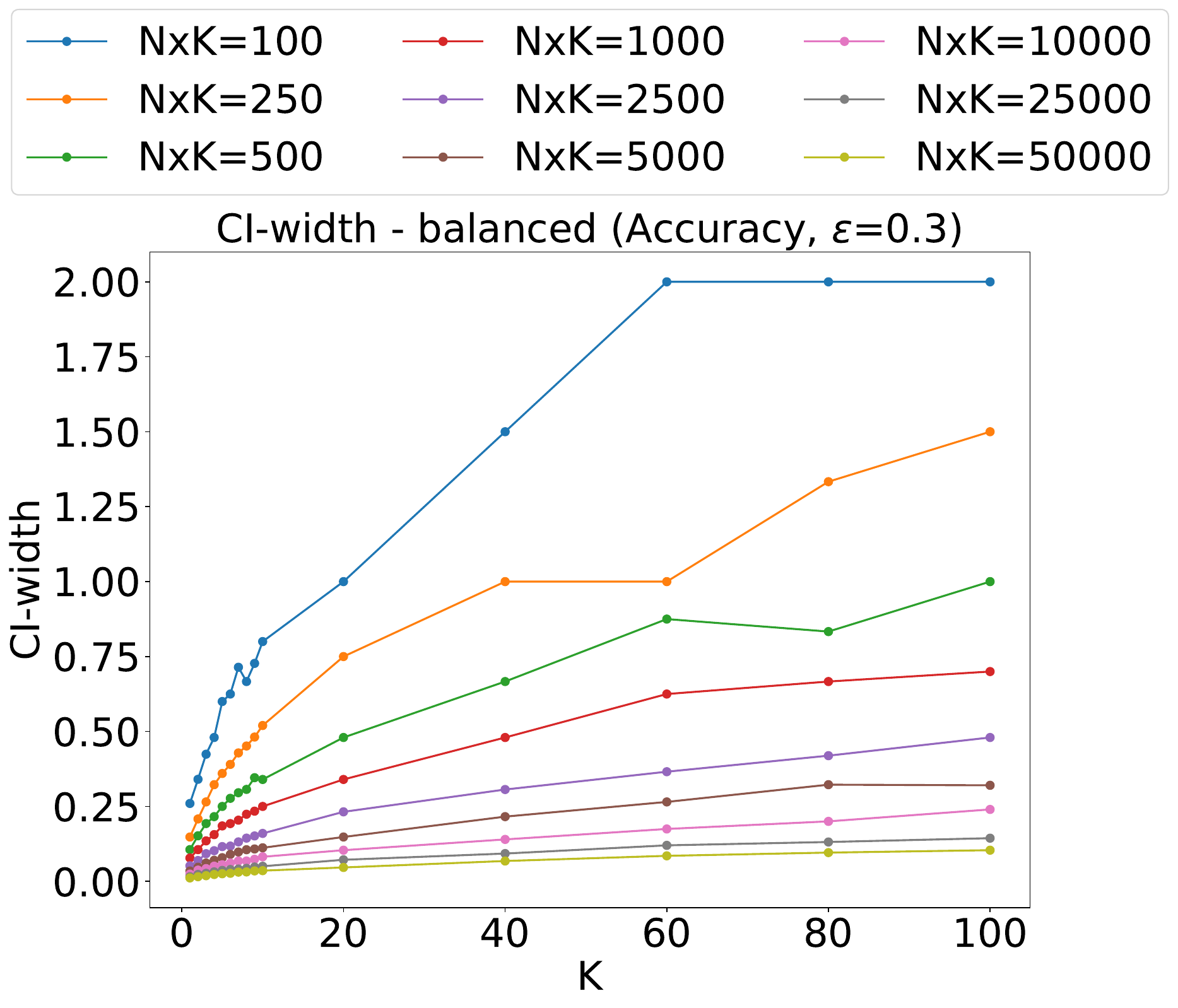}
    \caption{$\epsilon = 0.3$}
    \label{fig:uniform_ci_accuracy_cat4_e03}
  \end{subfigure} \hfill
  \begin{subfigure}[b]{0.24\linewidth}
    \centering
    \includegraphics[width=\linewidth]{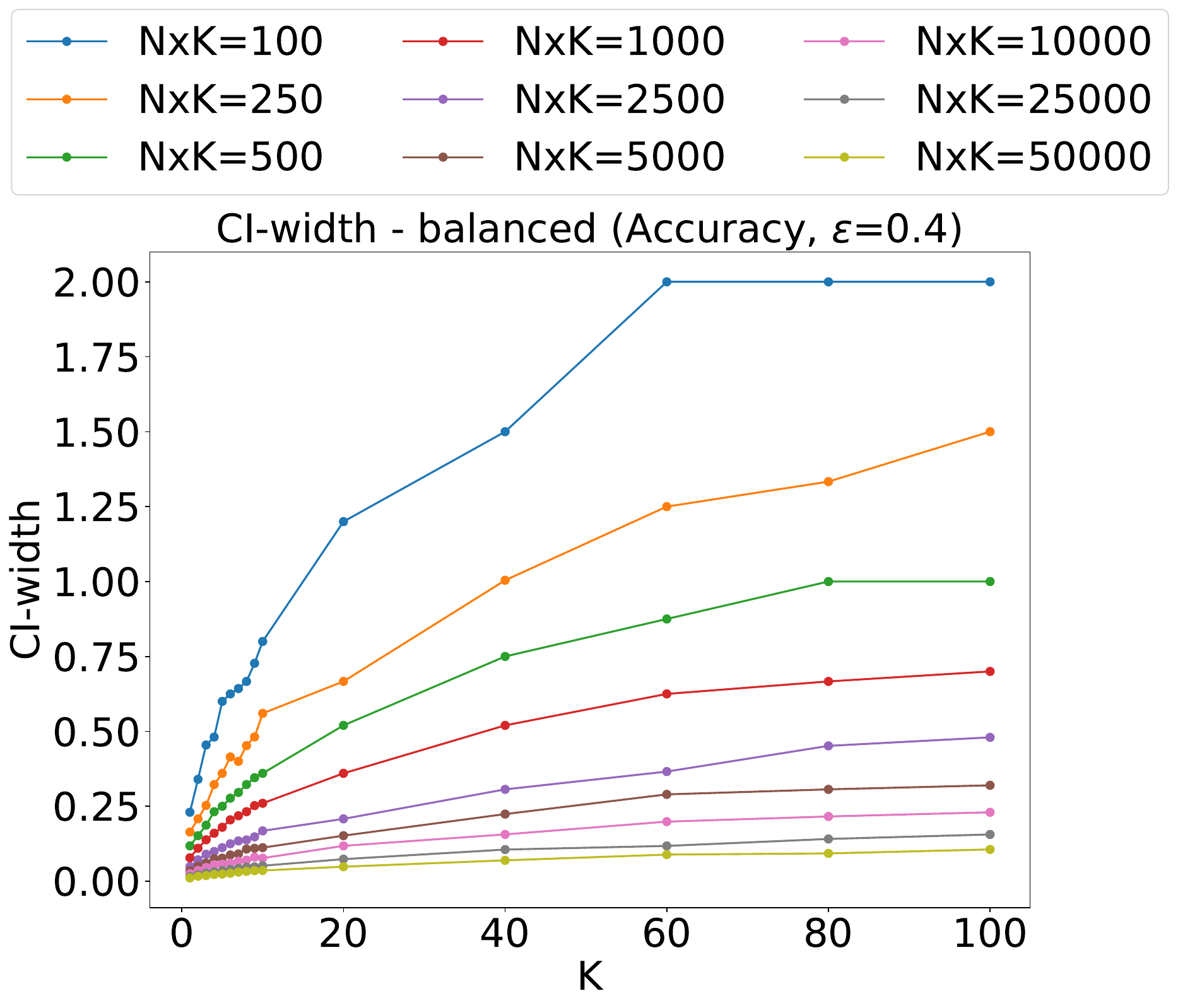}
    \caption{$\epsilon = 0.4$}
    \label{fig:uniform_ci_accuracy_cat4_e04}
  \end{subfigure}
  \caption{CI-width plots for balanced alphas with Accuracy as the metric ($M=4$)}
  \label{fig:uniform_ci_accuracy_cat4}
\end{figure*}

\begin{figure*}
  \centering
  \begin{subfigure}[b]{0.24\linewidth}
    \centering
    \includegraphics[width=\linewidth]{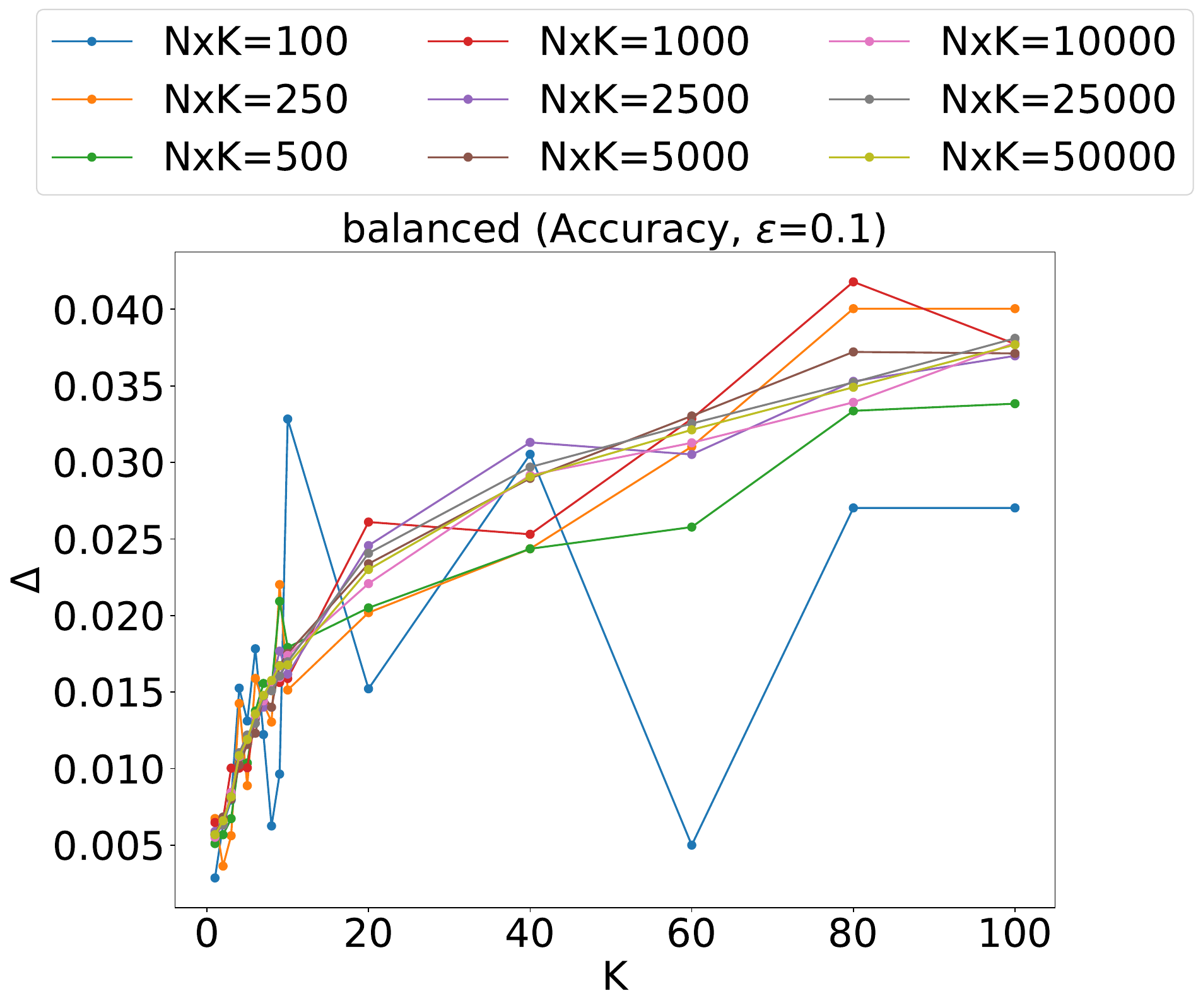}
    \caption{$\epsilon = 0.1$}
    \label{fig:uniform_delta_accuracy_cat4_e01}
  \end{subfigure} \hfill
  \begin{subfigure}[b]{0.24\linewidth}
    \centering
    \includegraphics[width=\linewidth]{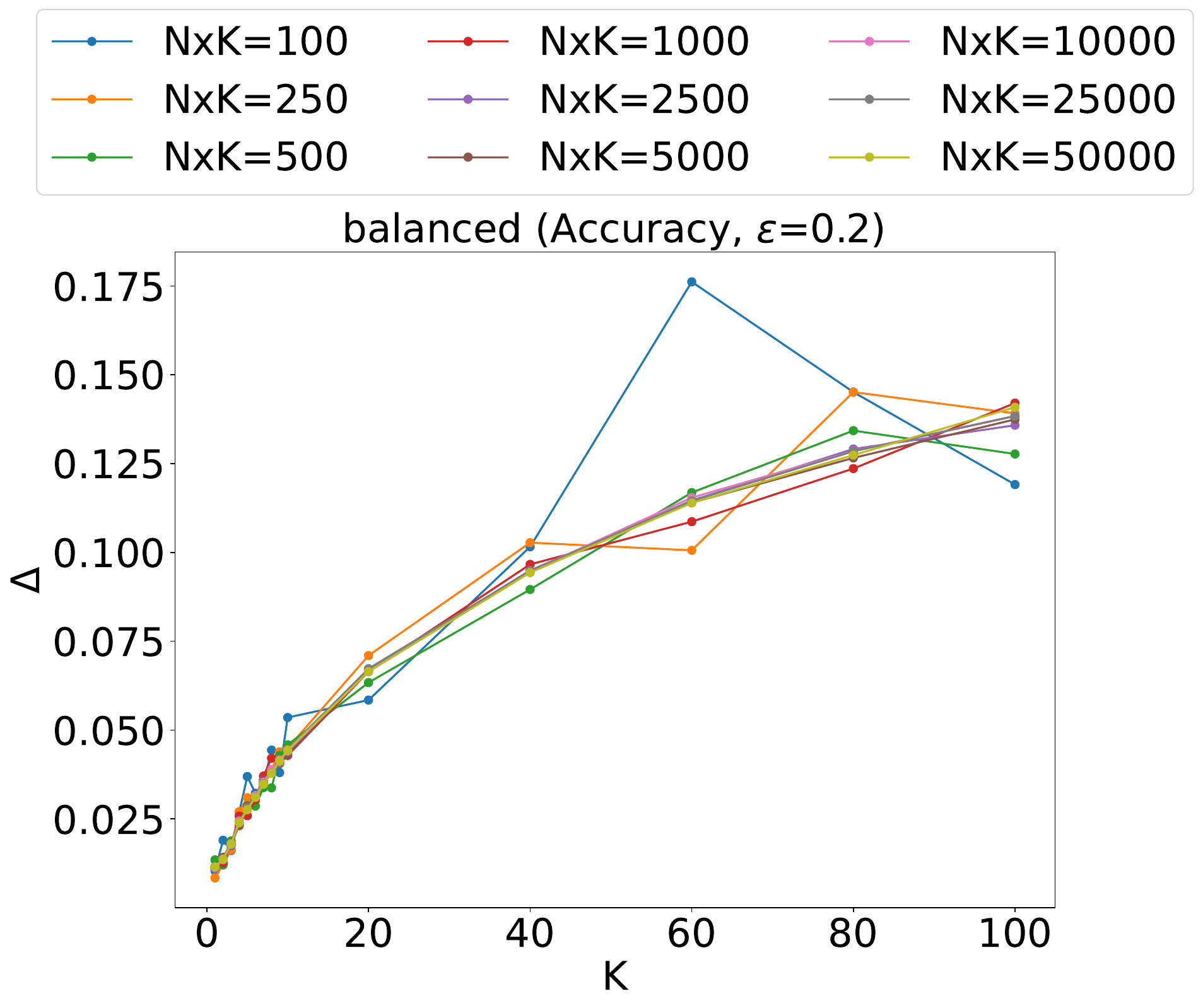}
    \caption{$\epsilon = 0.2$}
    \label{fig:uniform_delta_accuracy_cat4_e02}
  \end{subfigure} \hfill
  \begin{subfigure}[b]{0.24\linewidth}
    \centering
    \includegraphics[width=\linewidth]{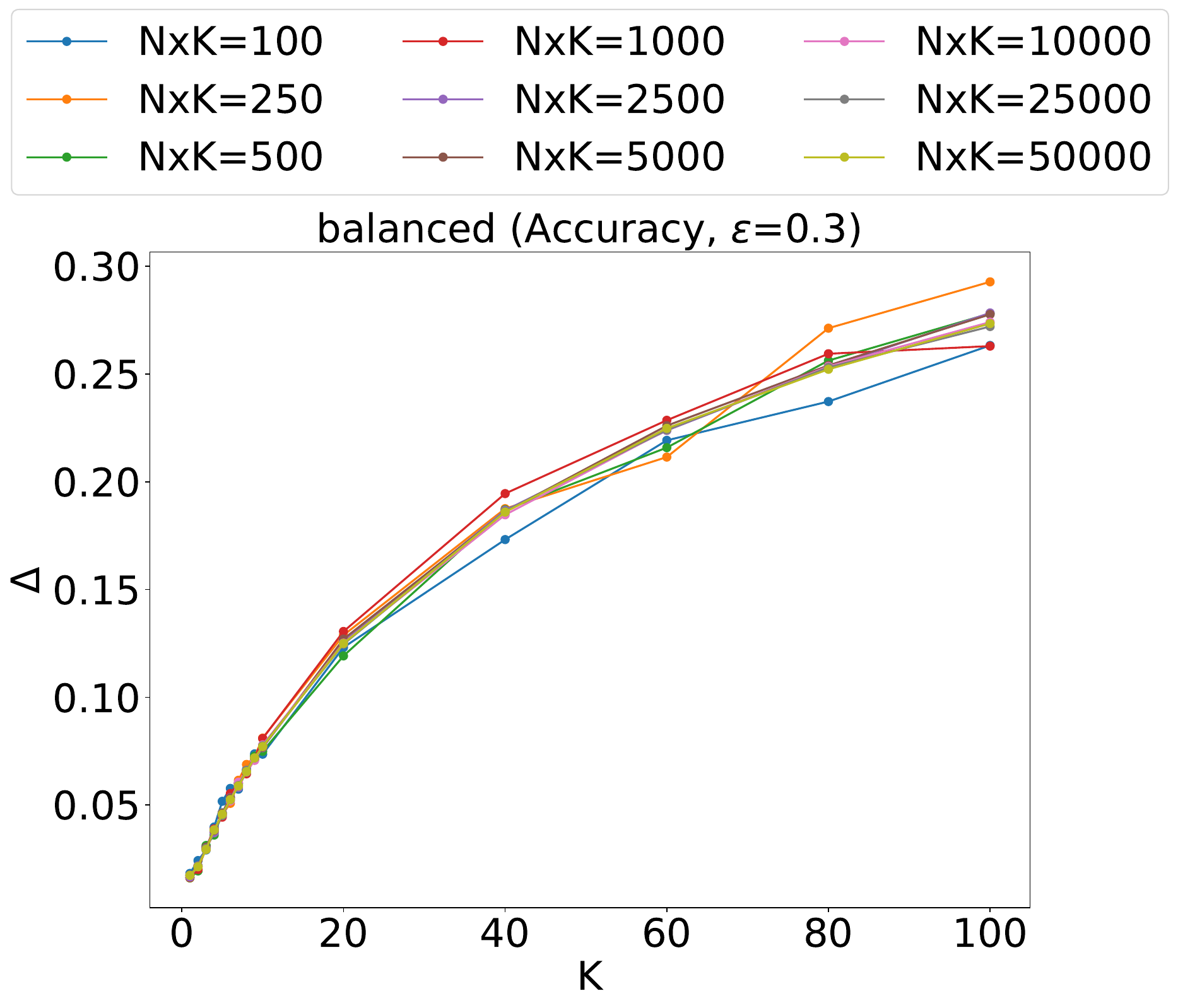}
    \caption{$\epsilon = 0.3$}
    \label{fig:uniform_delta_accuracy_cat4_e03}
  \end{subfigure} \hfill
  \begin{subfigure}[b]{0.24\linewidth}
    \centering
    \includegraphics[width=\linewidth]{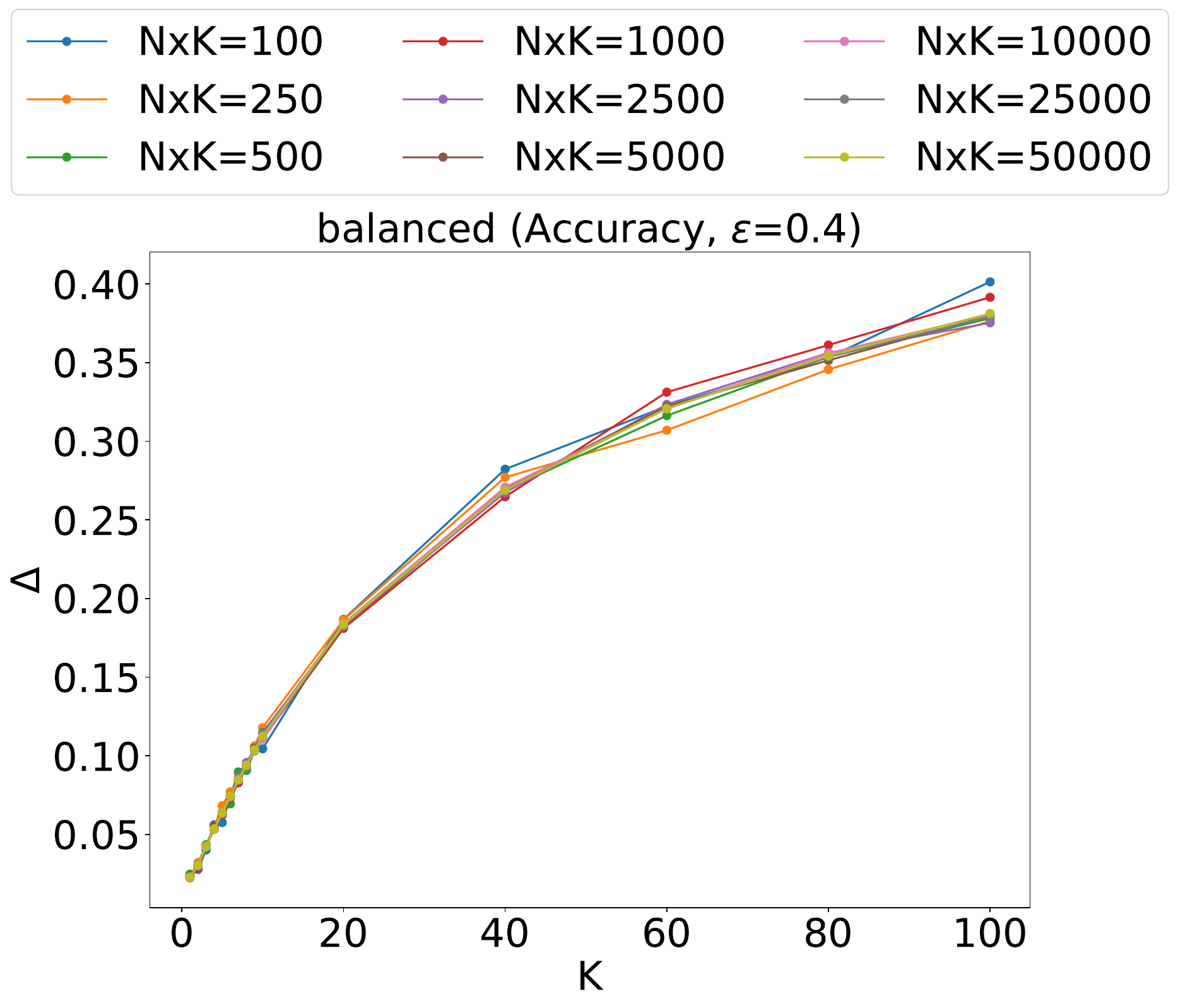}
    \caption{$\epsilon = 0.4$}
    \label{fig:uniform_delta_accuracy_cat4_e04}
  \end{subfigure}
  \caption{Effect sizes ($\Delta$) for balanced alphas with Accuracy as the metric ($M=4$)}
  \label{fig:uniform_delta_accuracy_cat4}
\end{figure*}

\begin{figure*}
  \centering
  \begin{subfigure}[b]{0.24\linewidth}
    \centering
    \includegraphics[width=\linewidth]{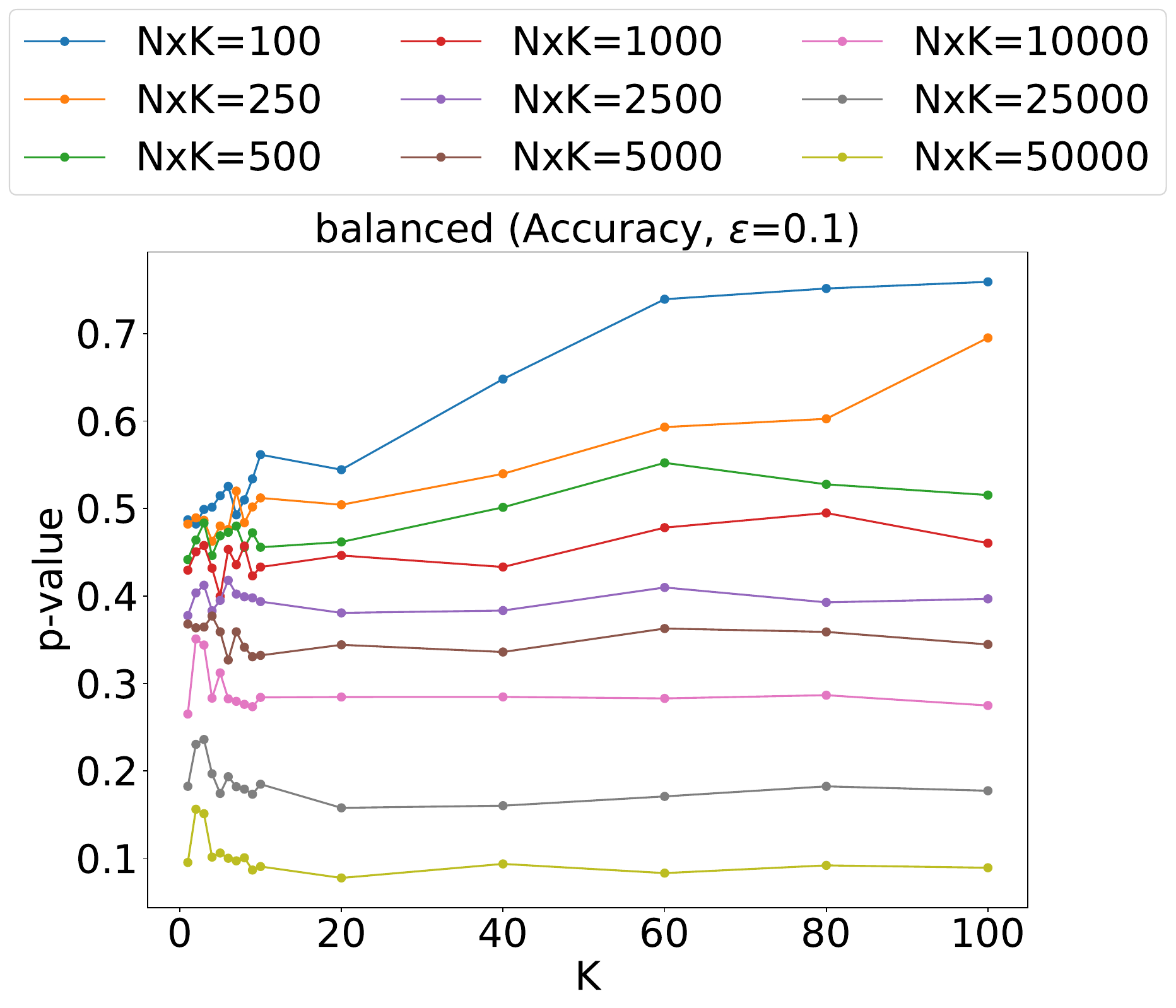}
    \caption{$\epsilon = 0.1$}
    \label{fig:uniform_accuracy_cat5_e01}
  \end{subfigure} \hfill
  \begin{subfigure}[b]{0.24\linewidth}
    \centering
    \includegraphics[width=\linewidth]{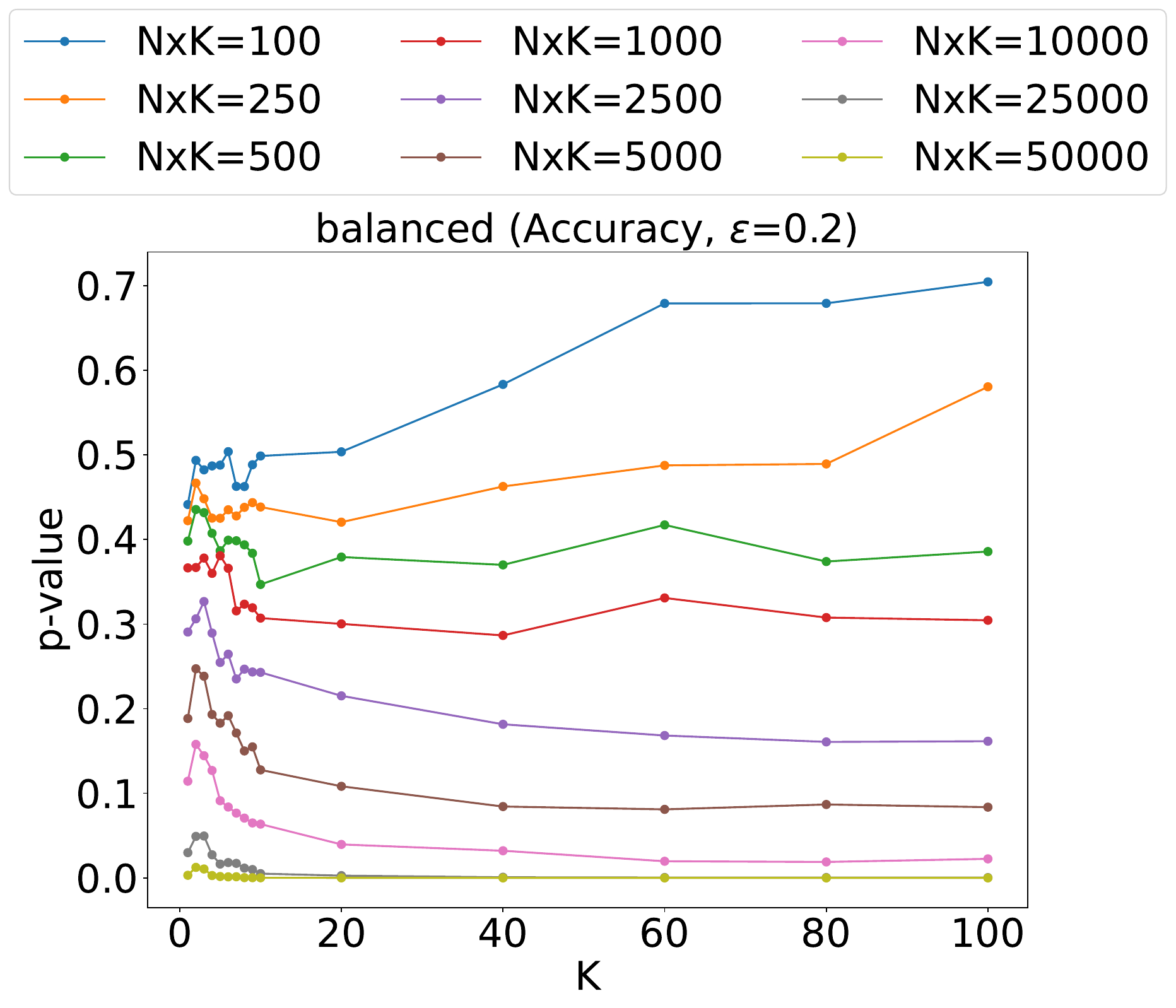}
    \caption{$\epsilon = 0.2$}
    \label{fig:uniform_accuracy_cat5_e02}
  \end{subfigure} \hfill
  \begin{subfigure}[b]{0.24\linewidth}
    \centering
    \includegraphics[width=\linewidth]{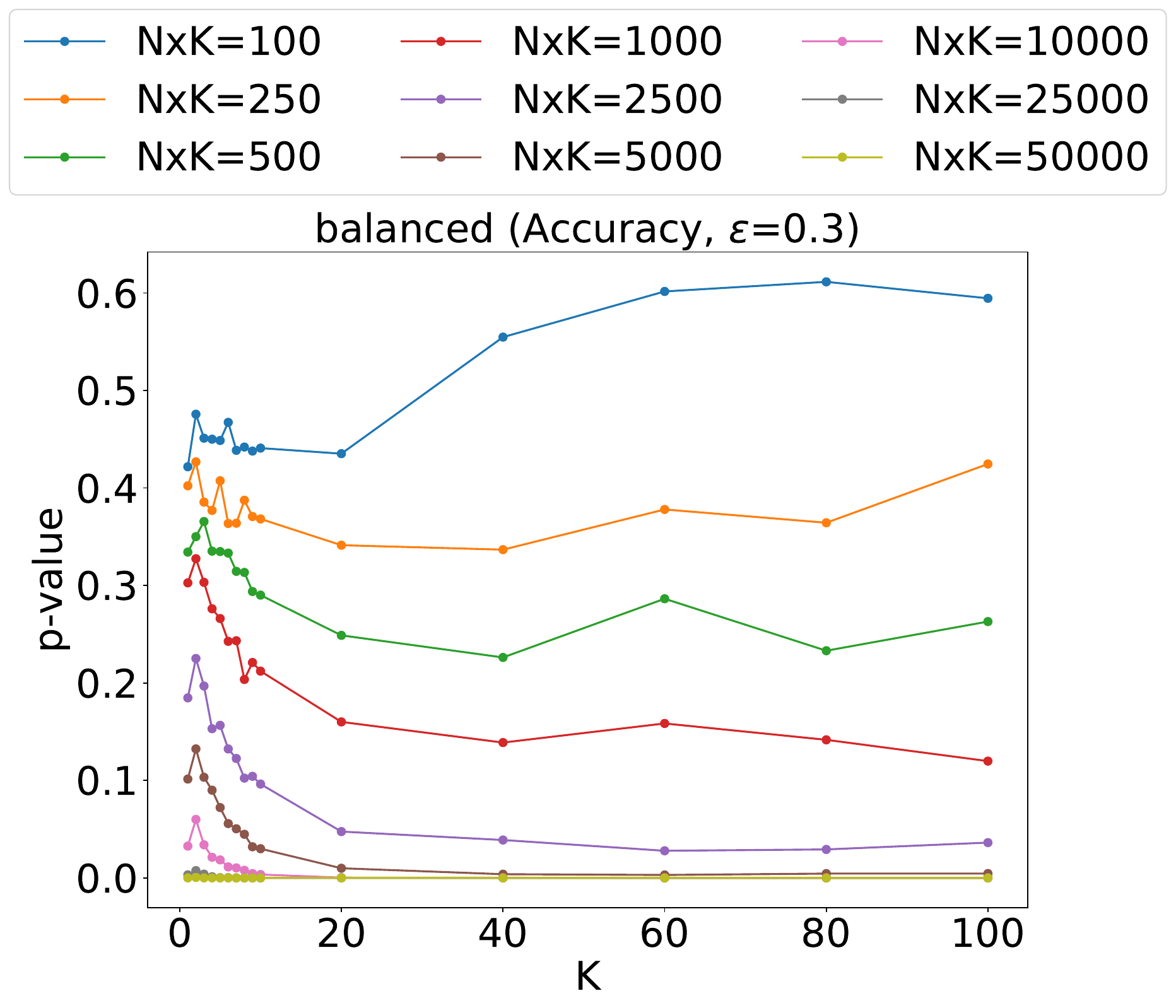}
    \caption{$\epsilon = 0.3$}
    \label{fig:uniform_accuracy_cat5_e03}
  \end{subfigure} \hfill
  \begin{subfigure}[b]{0.24\linewidth}
    \centering
    \includegraphics[width=\linewidth]{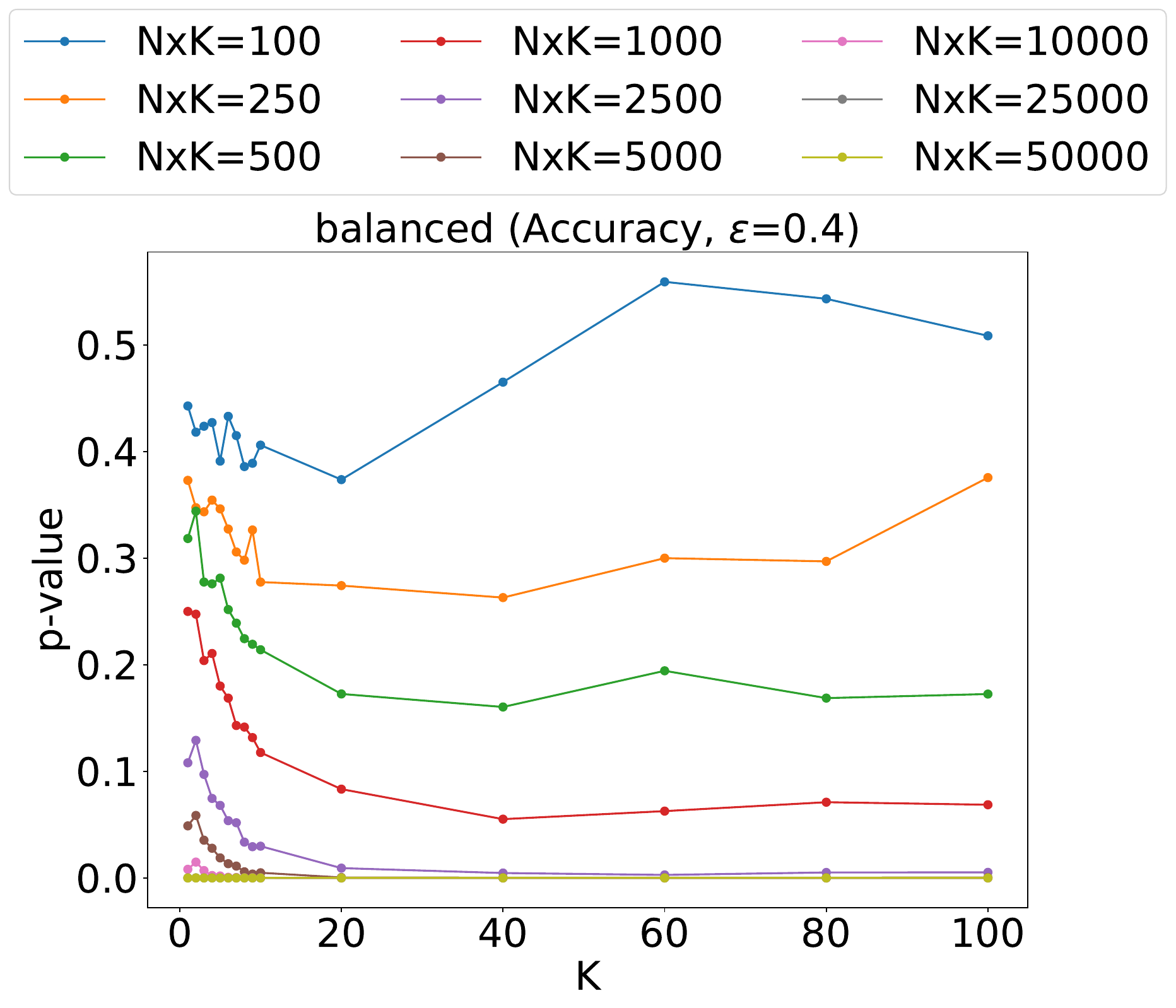}
    \caption{$\epsilon = 0.4$}
    \label{fig:uniform_accuracy_cat5_e04}
  \end{subfigure}
  \caption{P-value plots for balanced aplhas with Accuracy as the metric ($M=5$)}
  \label{fig:uniform_accuracy_cat5}
\end{figure*}

\begin{figure*}
  \centering
  \begin{subfigure}[b]{0.24\linewidth}
    \centering
    \includegraphics[width=\linewidth]{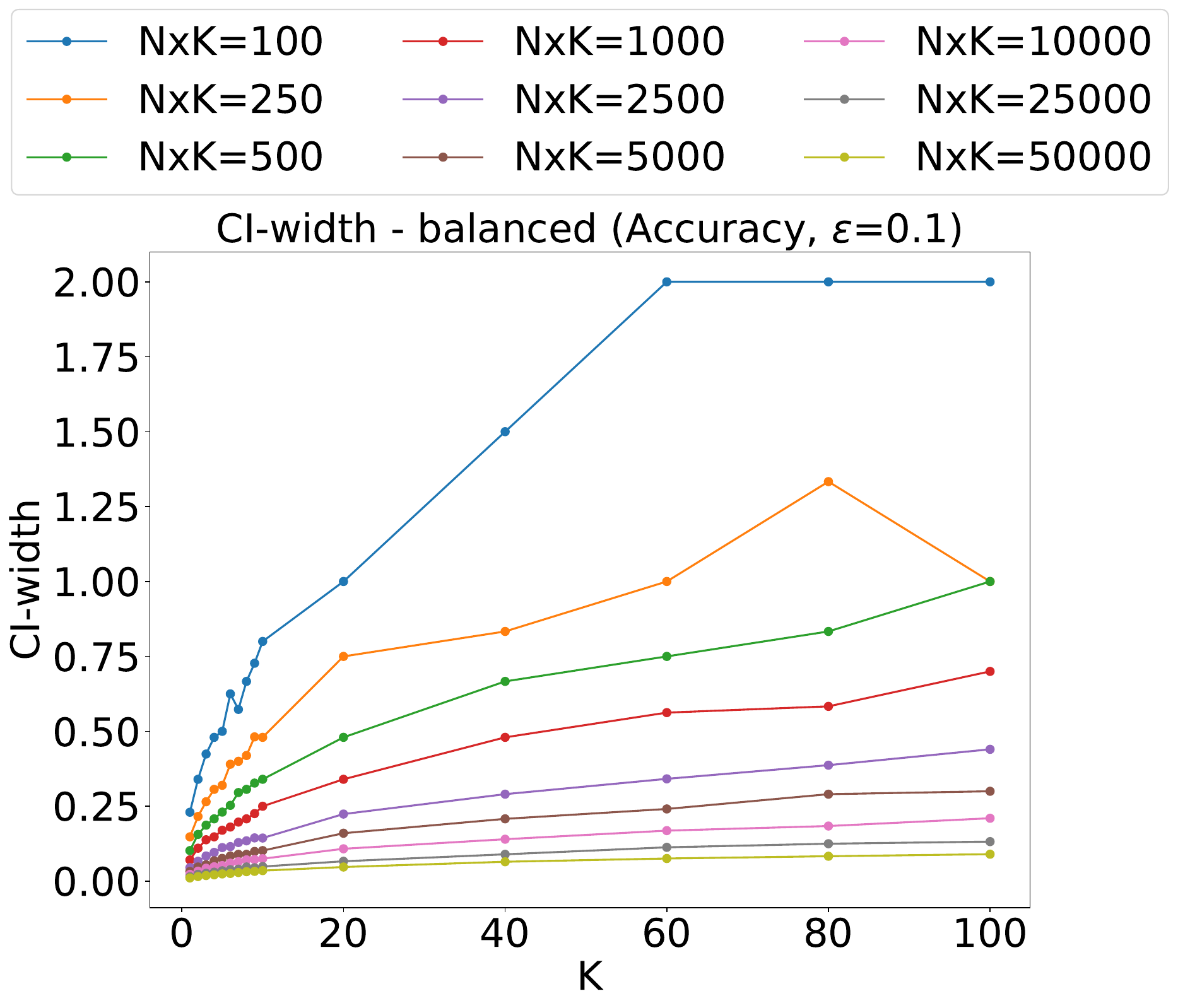}
    \caption{$\epsilon = 0.1$}
    \label{fig:uniform_ci_accuracy_cat5_e01}
  \end{subfigure} \hfill
  \begin{subfigure}[b]{0.24\linewidth}
    \centering
    \includegraphics[width=\linewidth]{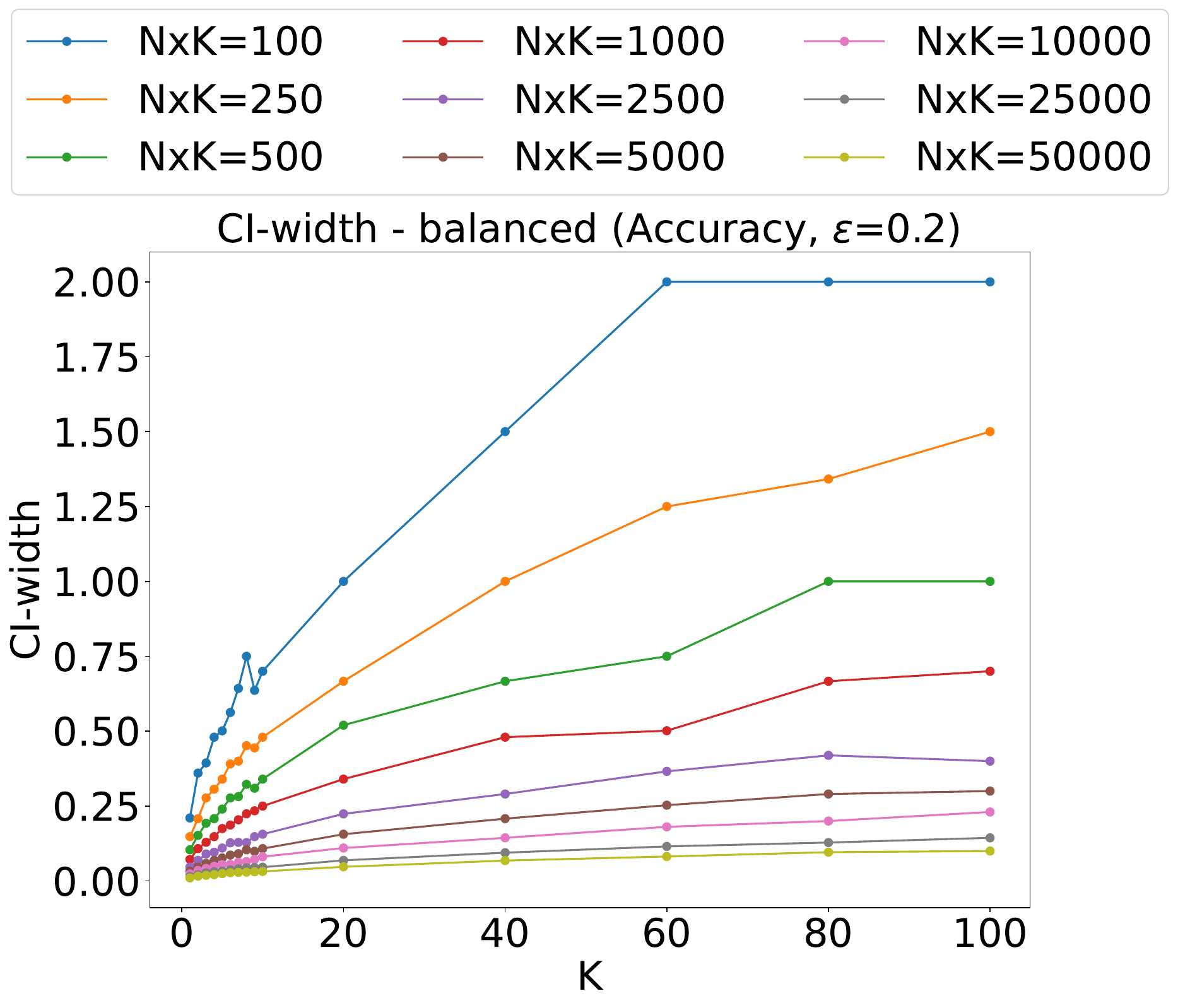}
    \caption{$\epsilon = 0.2$}
    \label{fig:uniform_ci_accuracy_cat5_e02}
  \end{subfigure} \hfill
  \begin{subfigure}[b]{0.24\linewidth}
    \centering
    \includegraphics[width=\linewidth]{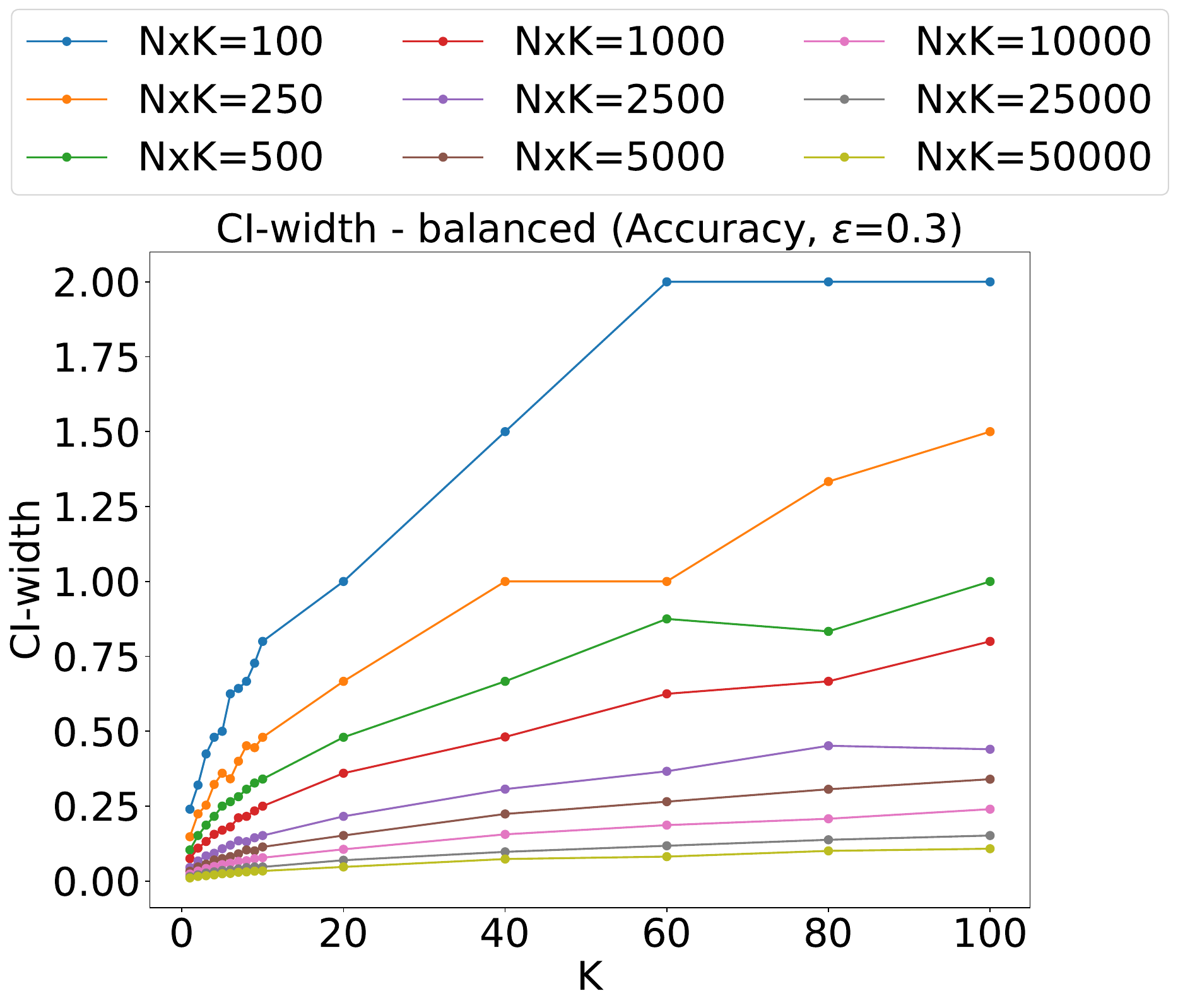}
    \caption{$\epsilon = 0.3$}
    \label{fig:uniform_ci_accuracy_cat5_e03}
  \end{subfigure} \hfill
  \begin{subfigure}[b]{0.24\linewidth}
    \centering
    \includegraphics[width=\linewidth]{figures/K100/ci_plots/artificial/balanced/cat4/balanced_CI_width_Accuracy_K_100_e_0.4.pdf}
    \caption{$\epsilon = 0.4$}
    \label{fig:uniform_ci_accuracy_cat5_e04}
  \end{subfigure}
  \caption{CI-width plots for balanced alphas with Accuracy as the metric ($M=5$)}
  \label{fig:uniform_ci_accuracy_cat5}
\end{figure*}

\begin{figure*}
  \centering
  \begin{subfigure}[b]{0.24\linewidth}
    \centering
    \includegraphics[width=\linewidth]{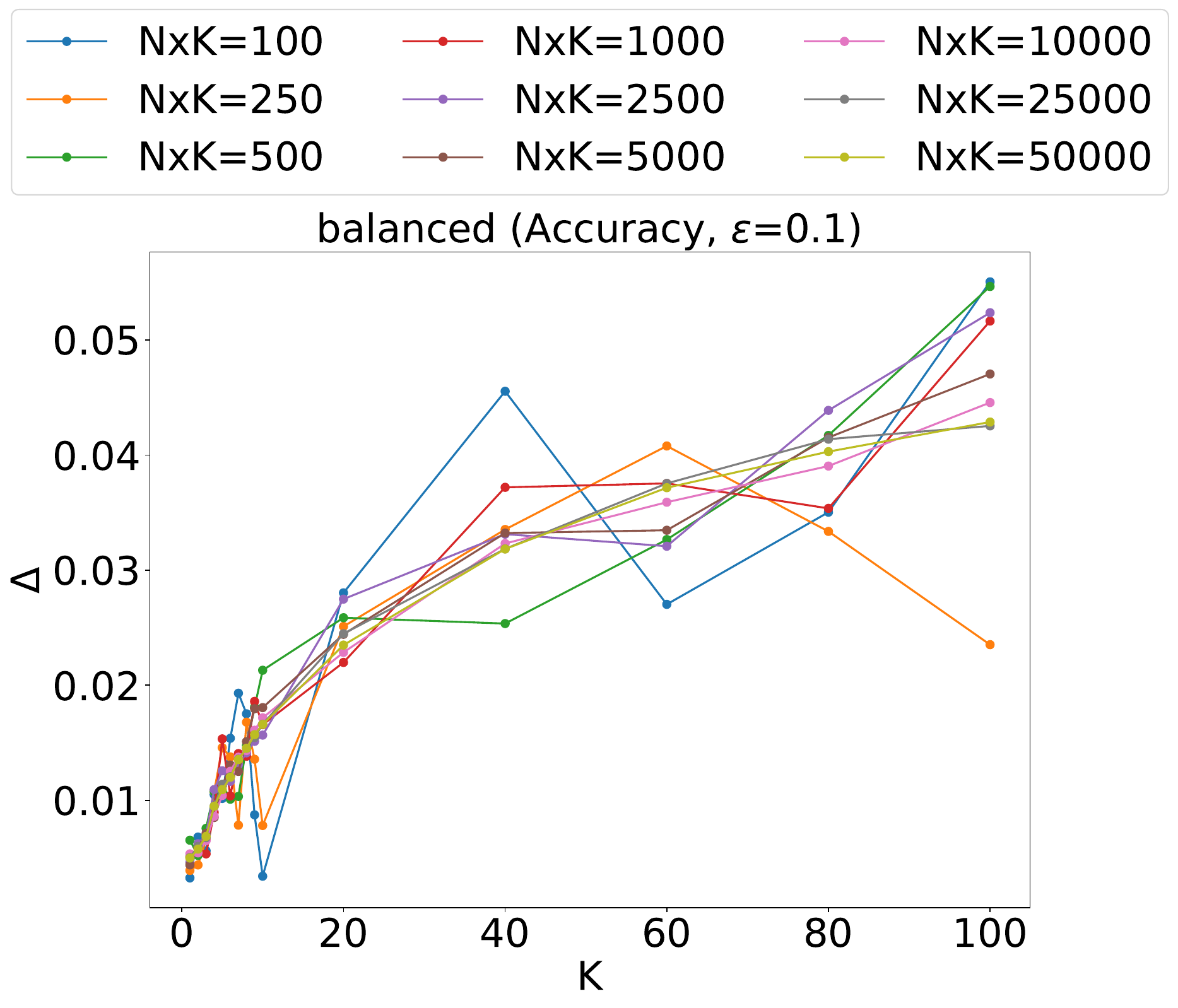}
    \caption{$\epsilon = 0.1$}
    \label{fig:uniform_delta_accuracy_cat5_e01}
  \end{subfigure} \hfill
  \begin{subfigure}[b]{0.24\linewidth}
    \centering
    \includegraphics[width=\linewidth]{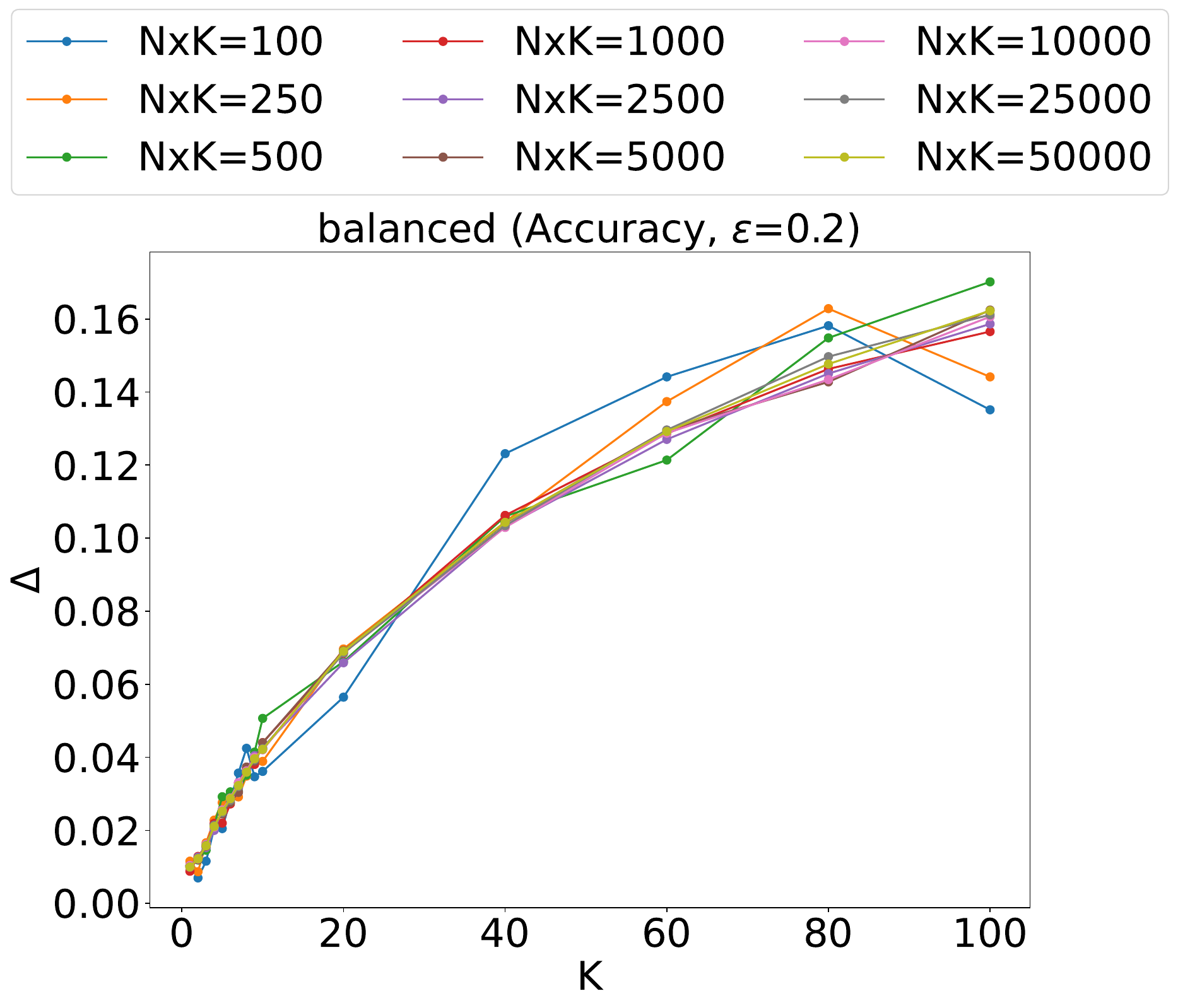}
    \caption{$\epsilon = 0.2$}
    \label{fig:uniform_delta_accuracy_cat5_e02}
  \end{subfigure} \hfill
  \begin{subfigure}[b]{0.24\linewidth}
    \centering
    \includegraphics[width=\linewidth]{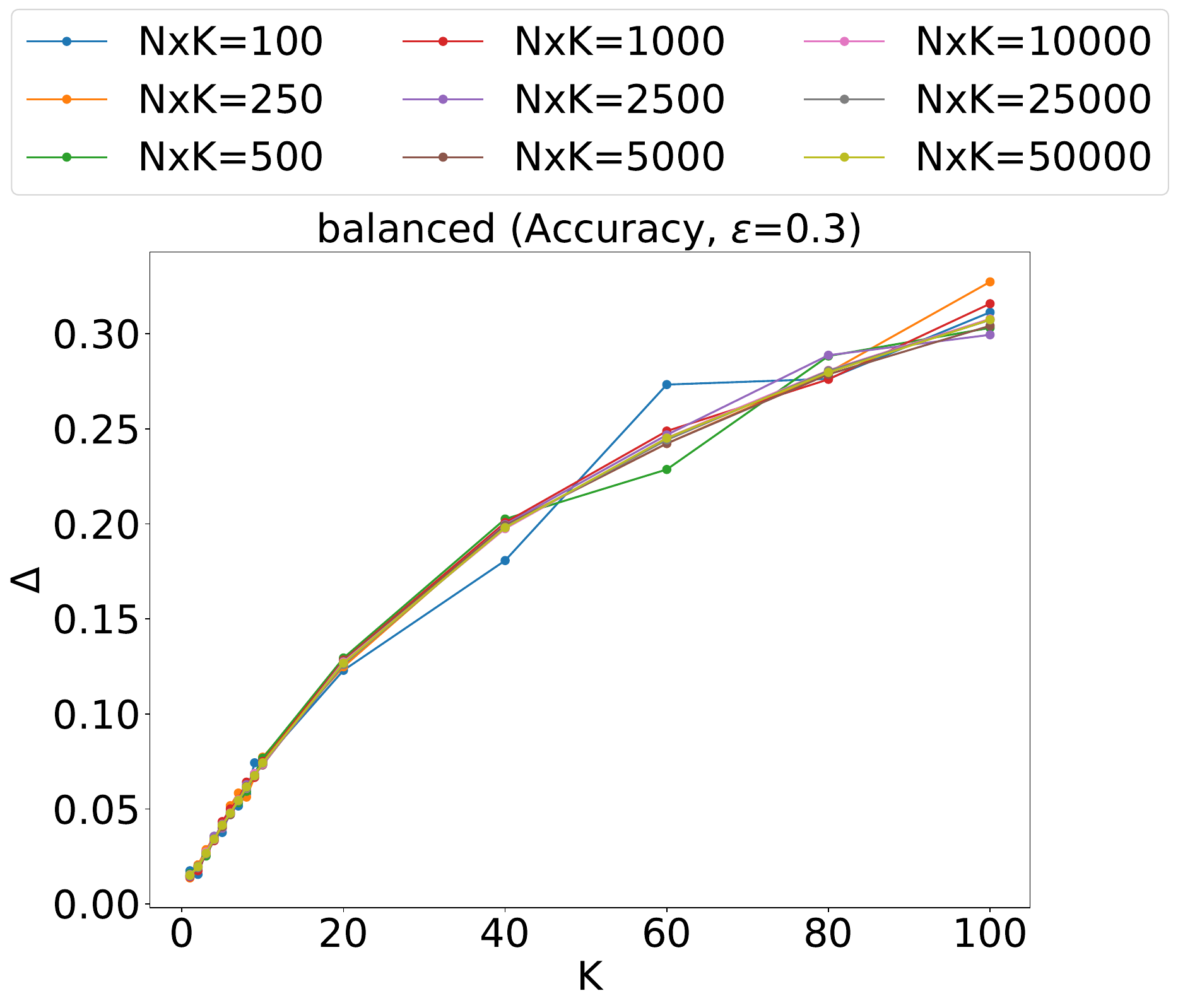}
    \caption{$\epsilon = 0.3$}
    \label{fig:uniform_delta_accuracy_cat5_e03}
  \end{subfigure} \hfill
  \begin{subfigure}[b]{0.24\linewidth}
    \centering
    \includegraphics[width=\linewidth]{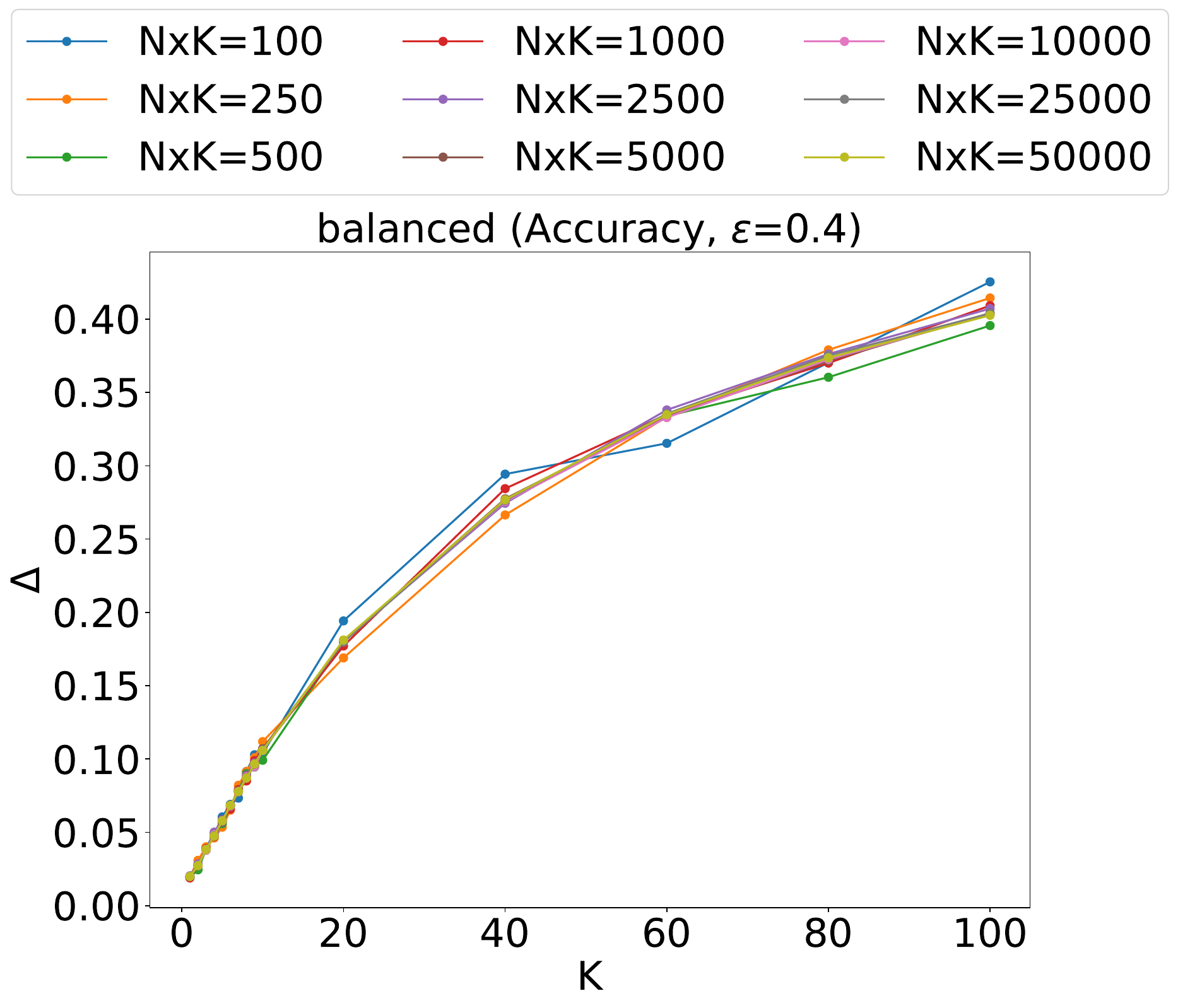}
    \caption{$\epsilon = 0.4$}
    \label{fig:uniform_delta_accuracy_cat5_e04}
  \end{subfigure}
  \caption{Effect sizes ($\Delta$) for balanced alphas with Accuracy as the metric ($M=5$)}
  \label{fig:uniform_delta_accuracy_cat5}
\end{figure*}

\begin{figure*}
  \centering
  \begin{subfigure}[b]{0.24\linewidth}
    \centering
    \includegraphics[width=\linewidth]{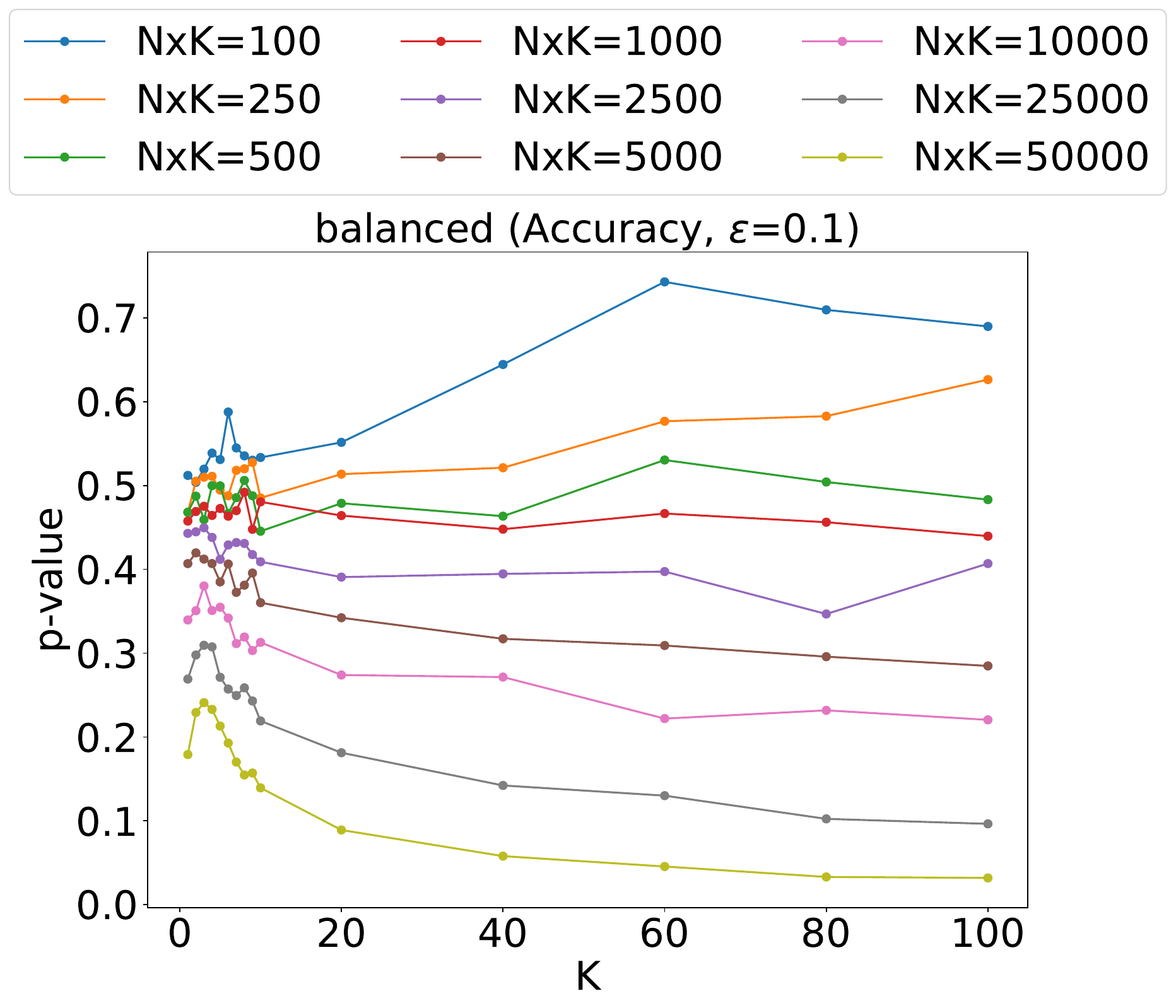}
    \caption{$\epsilon = 0.1$}
    \label{fig:uniform_accuracy_cat12_e01}
  \end{subfigure} \hfill
  \begin{subfigure}[b]{0.24\linewidth}
    \centering
    \includegraphics[width=\linewidth]{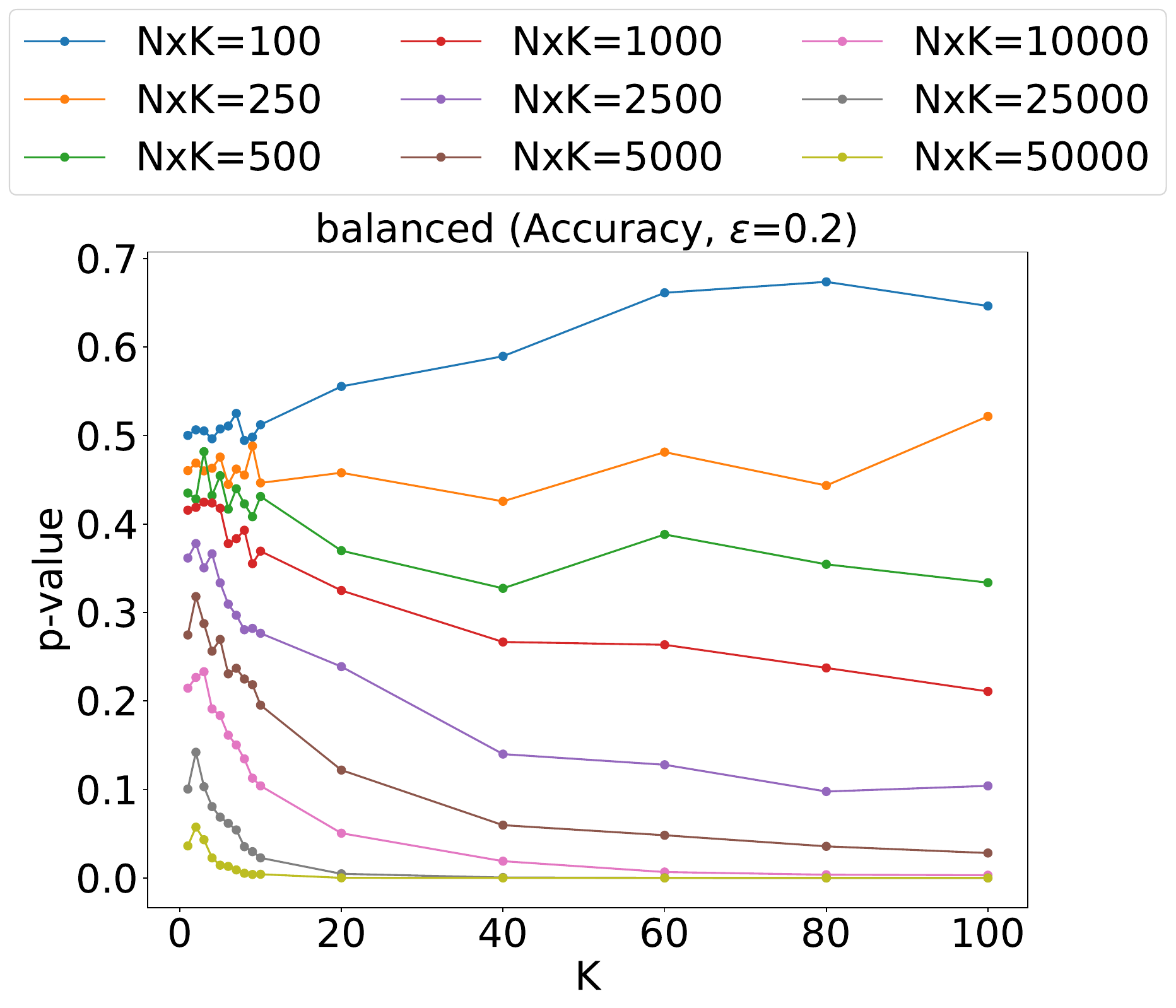}
    \caption{$\epsilon = 0.2$}
    \label{fig:uniform_accuracy_cat12_e02}
  \end{subfigure} \hfill
  \begin{subfigure}[b]{0.24\linewidth}
    \centering
    \includegraphics[width=\linewidth]{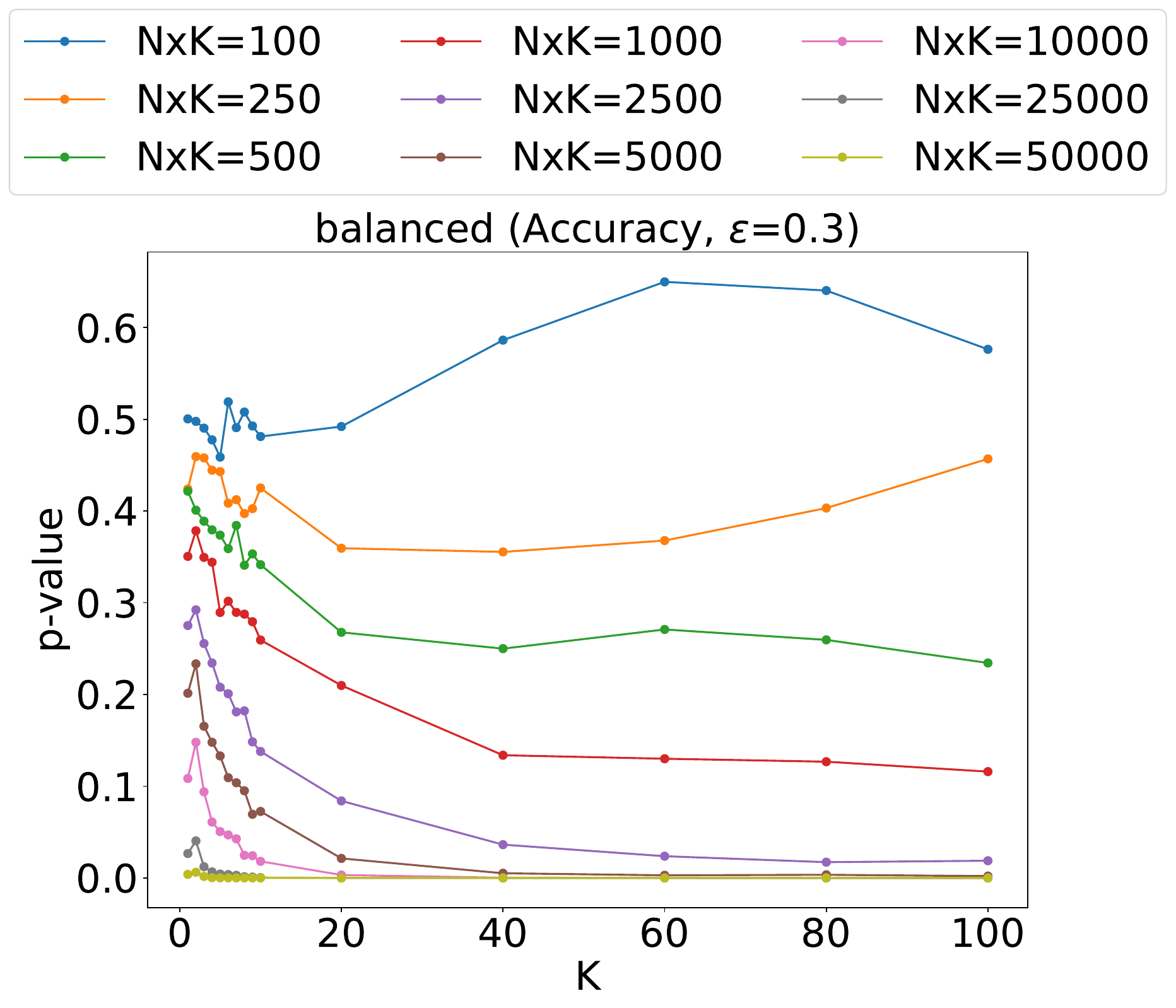}
    \caption{$\epsilon = 0.3$}
    \label{fig:uniform_accuracy_cat12_e03}
  \end{subfigure} \hfill
  \begin{subfigure}[b]{0.24\linewidth}
    \centering
    \includegraphics[width=\linewidth]{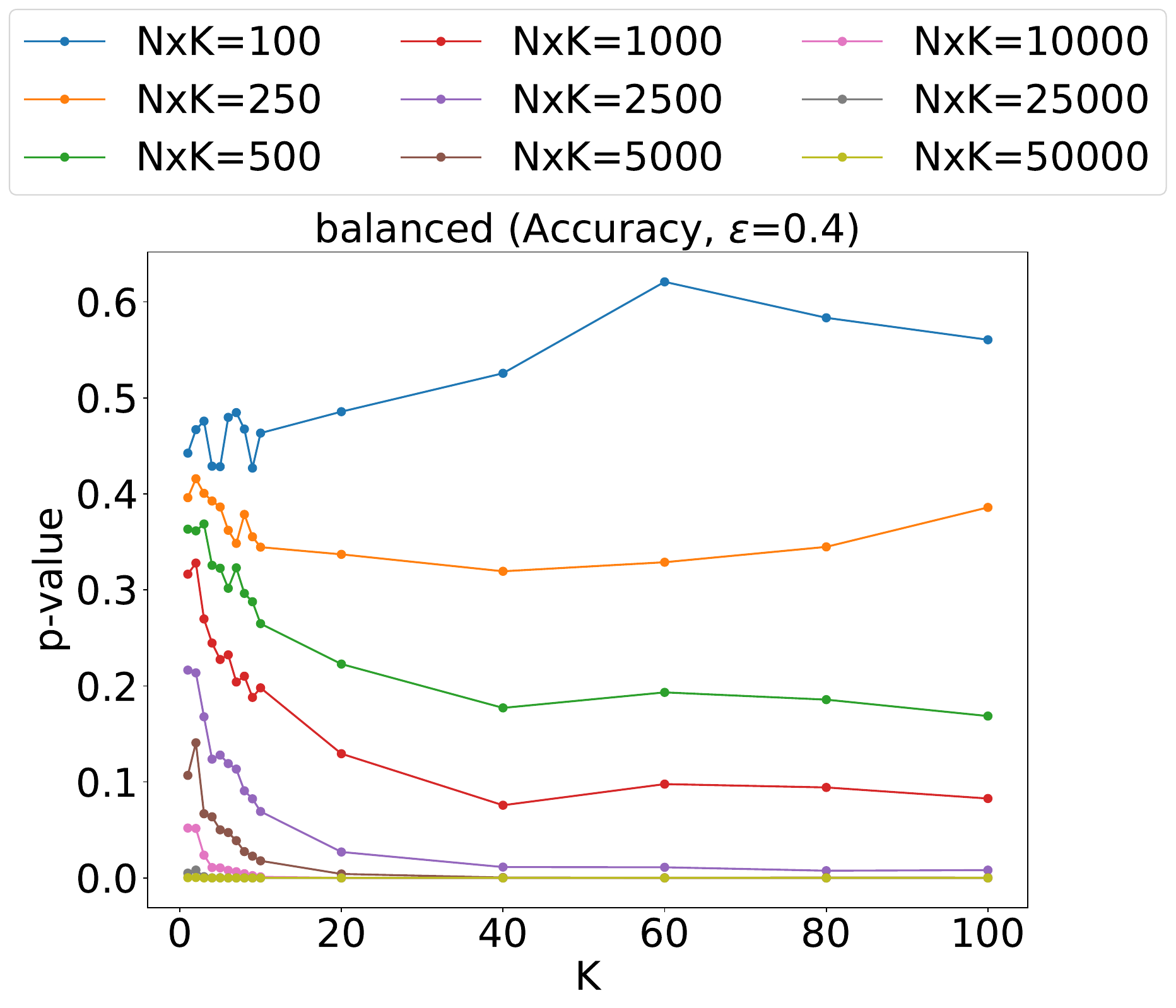}
    \caption{$\epsilon = 0.4$}
    \label{fig:uniform_accuracy_cat12_e04}
  \end{subfigure}
  \caption{P-value plots for balanced aplhas with Accuracy as the metric ($M=12$)}
  \label{fig:uniform_accuracy_cat12}
\end{figure*}

\begin{figure*}
  \centering
  \begin{subfigure}[b]{0.24\linewidth}
    \centering
    \includegraphics[width=\linewidth]{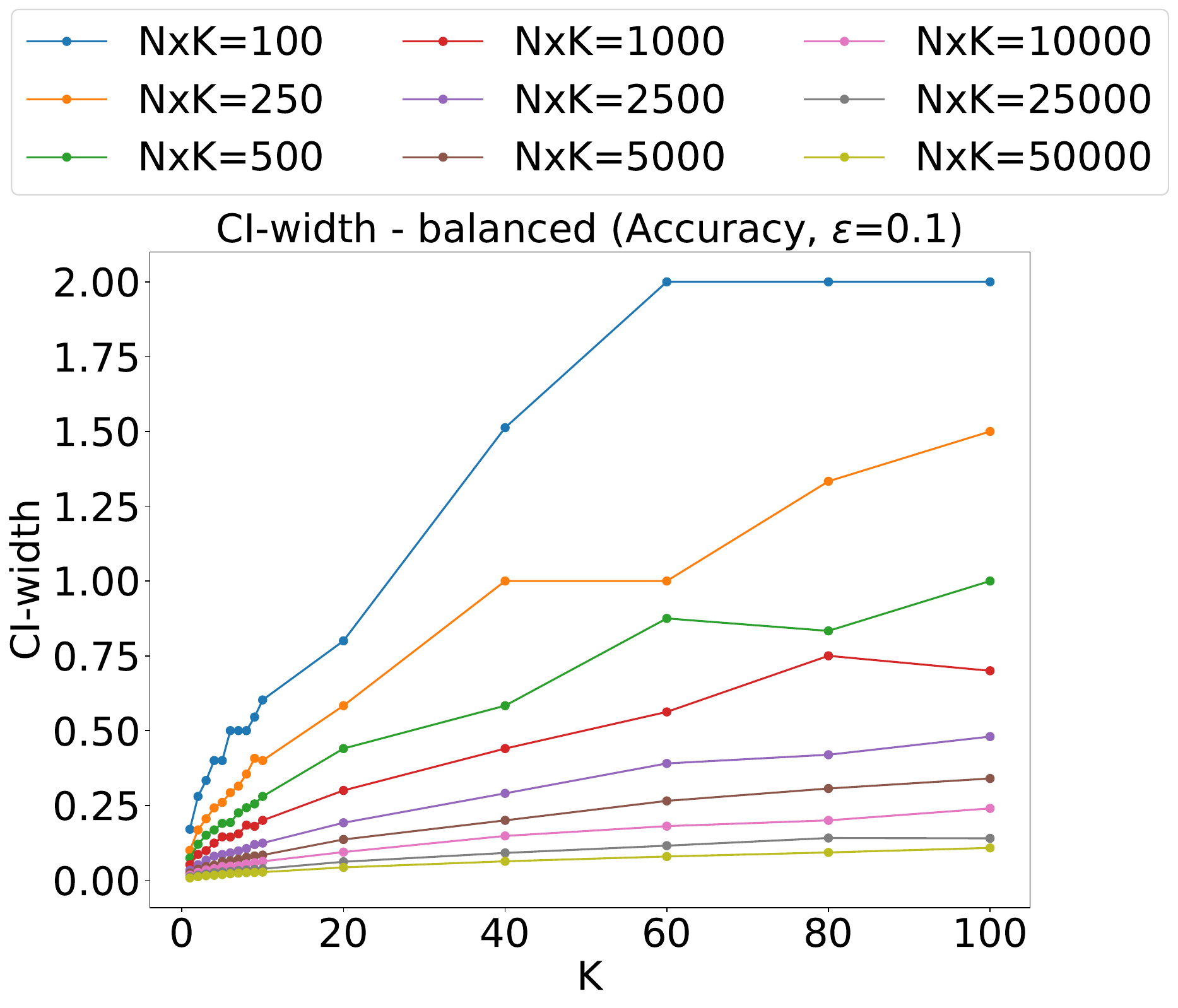}
    \caption{$\epsilon = 0.1$}
    \label{fig:uniform_ci_accuracy_cat12_e01}
  \end{subfigure} \hfill
  \begin{subfigure}[b]{0.24\linewidth}
    \centering
    \includegraphics[width=\linewidth]{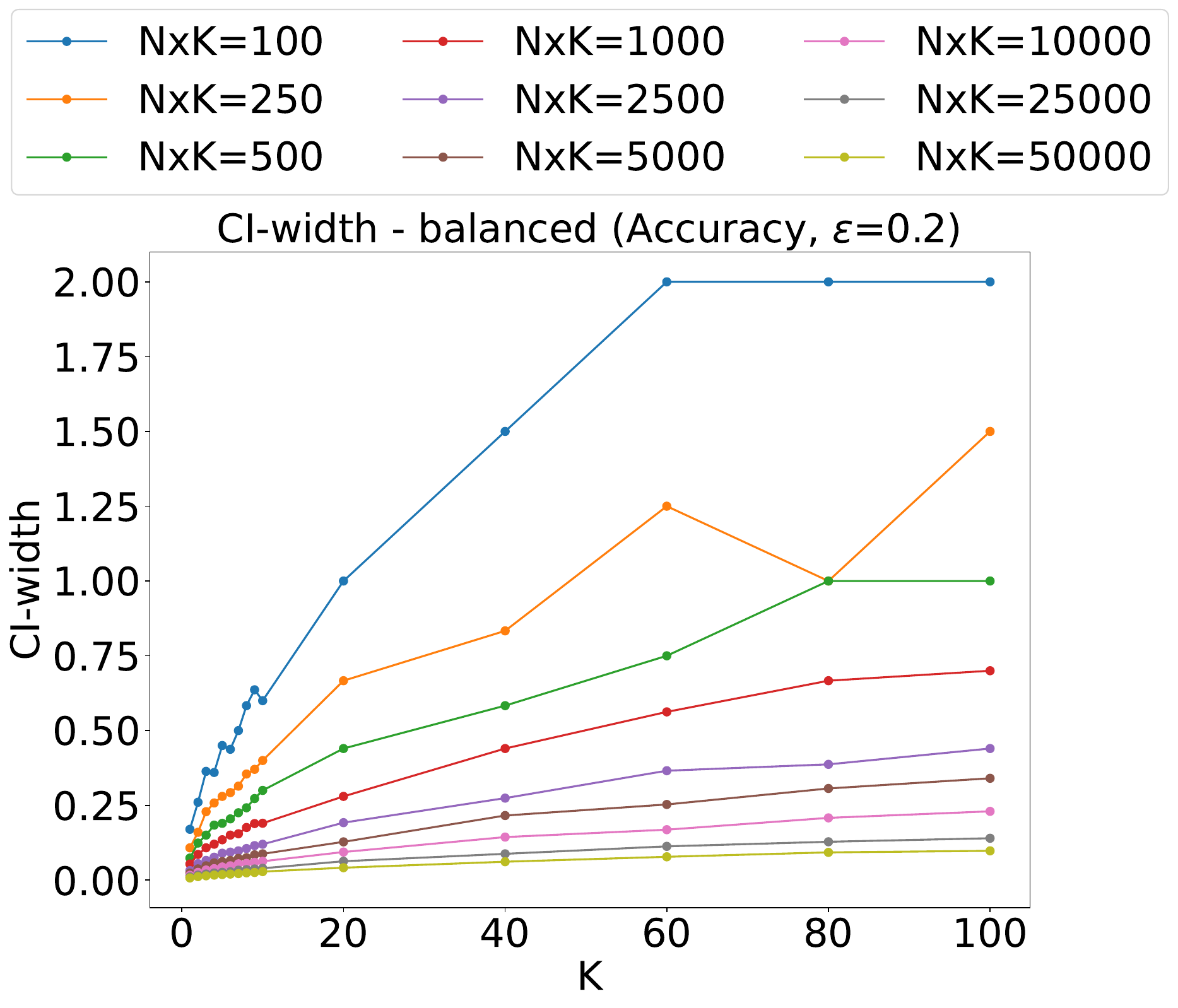}
    \caption{$\epsilon = 0.2$}
    \label{fig:uniform_ci_accuracy_cat12_e02}
  \end{subfigure} \hfill
  \begin{subfigure}[b]{0.24\linewidth}
    \centering
    \includegraphics[width=\linewidth]{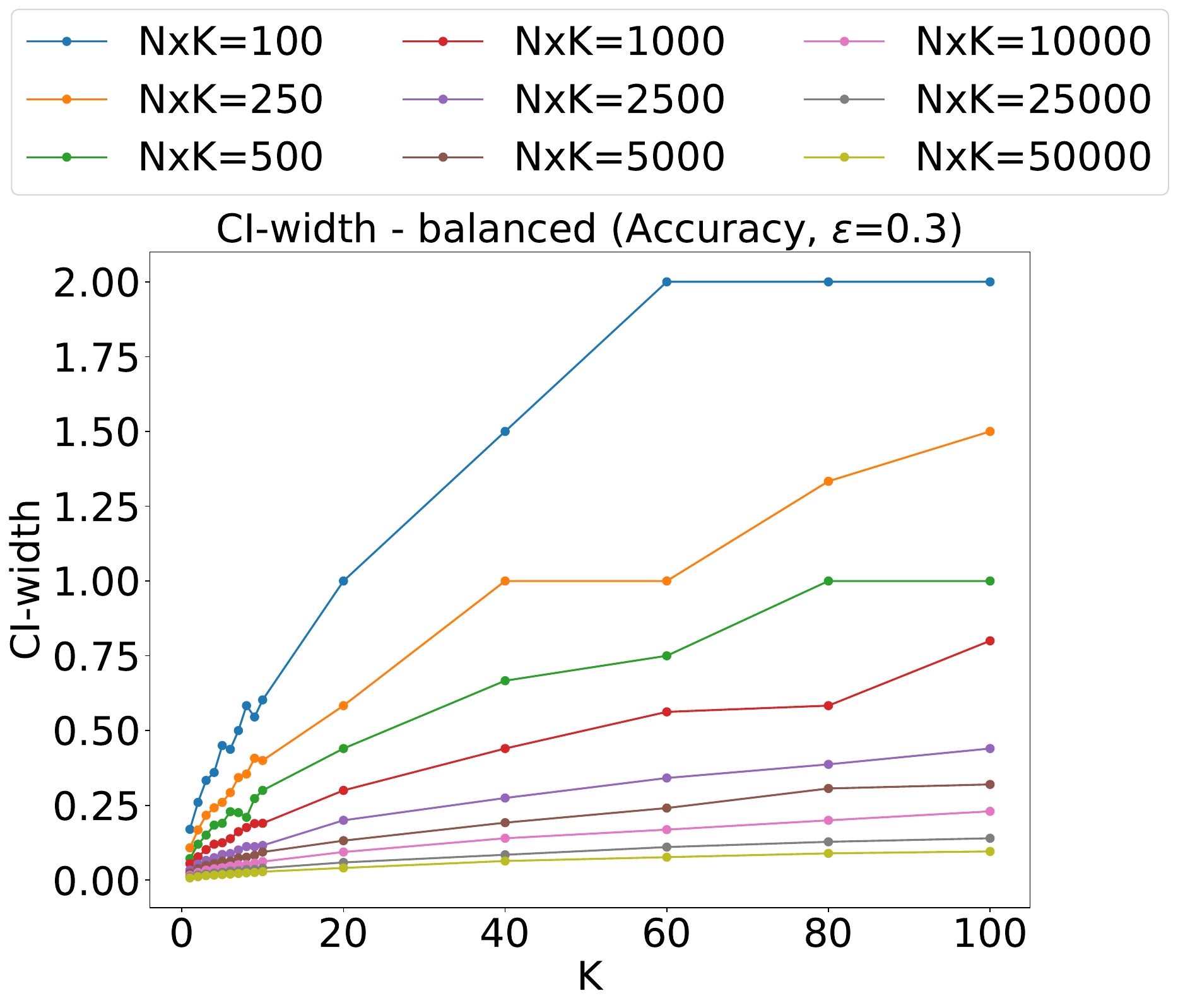}
    \caption{$\epsilon = 0.3$}
    \label{fig:uniform_ci_accuracy_cat12_e03}
  \end{subfigure} \hfill
  \begin{subfigure}[b]{0.24\linewidth}
    \centering
    \includegraphics[width=\linewidth]{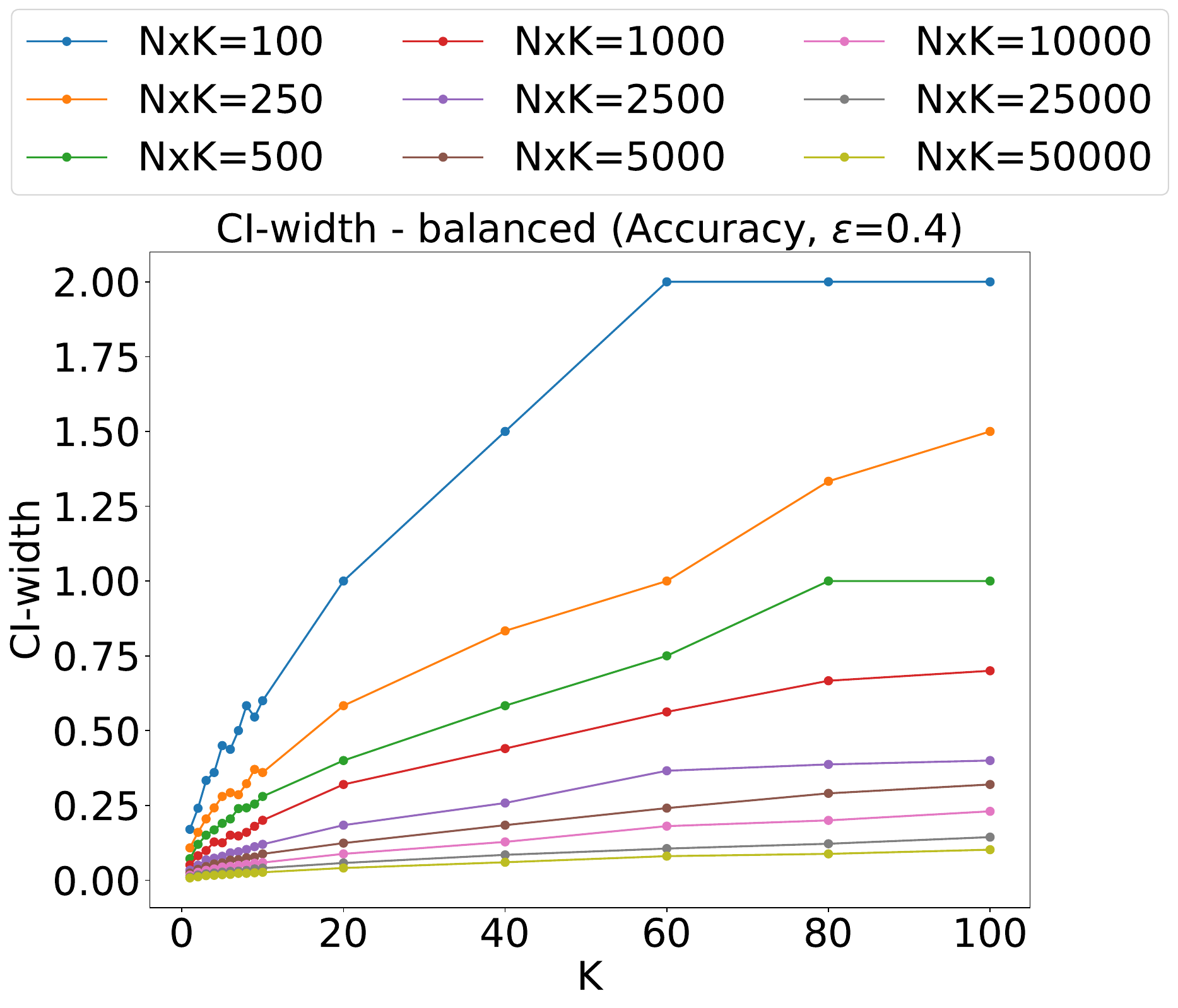}
    \caption{$\epsilon = 0.4$}
    \label{fig:uniform_ci_accuracy_cat12_e04}
  \end{subfigure}
  \caption{CI-width plots for balanced alphas with Accuracy as the metric ($M=12$)}
  \label{fig:uniform_ci_accuracy_cat12}
\end{figure*}

\begin{figure*}
  \centering
  \begin{subfigure}[b]{0.24\linewidth}
    \centering
    \includegraphics[width=\linewidth]{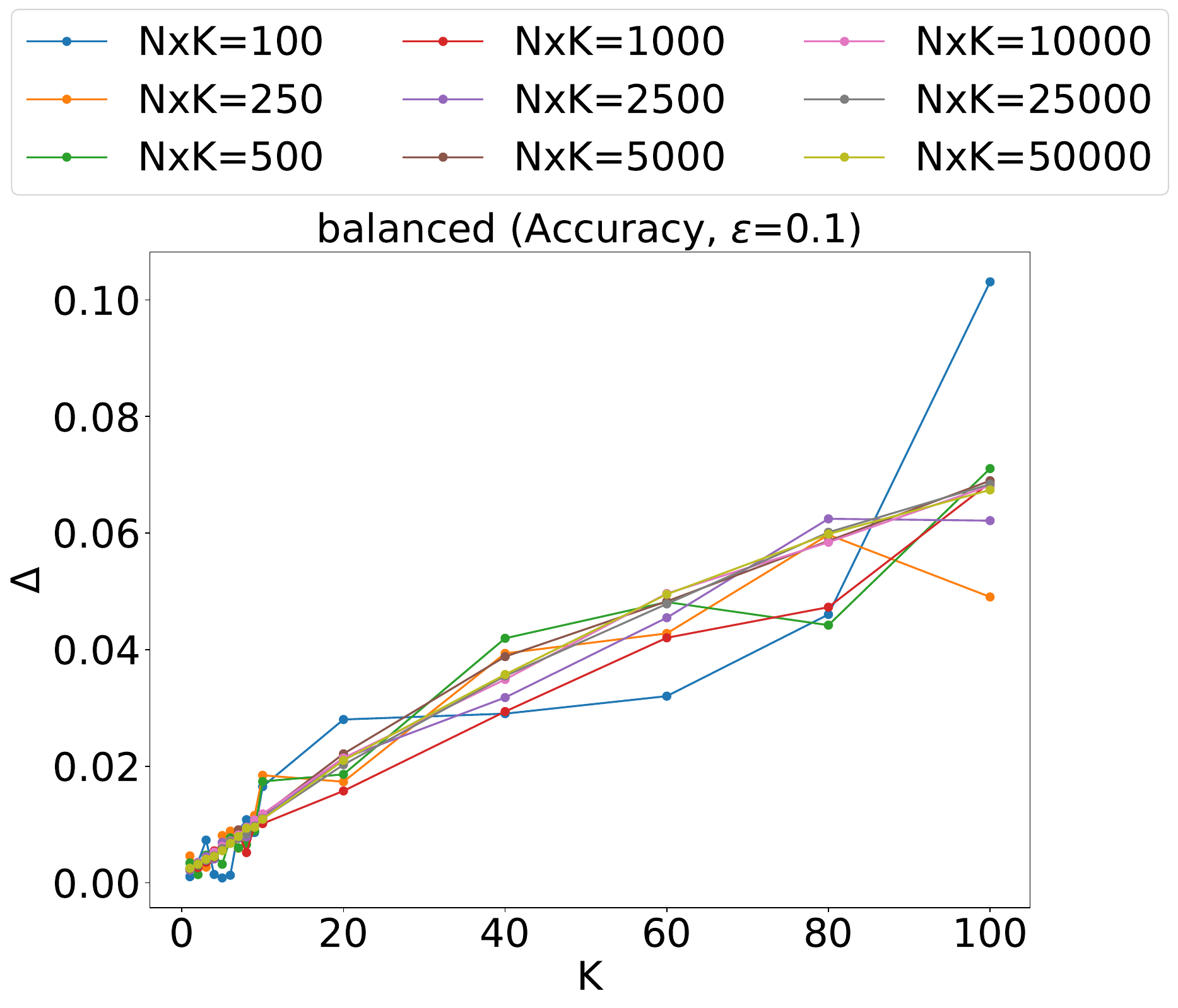}
    \caption{$\epsilon = 0.1$}
    \label{fig:uniform_delta_accuracy_cat12_e01}
  \end{subfigure} \hfill
  \begin{subfigure}[b]{0.24\linewidth}
    \centering
    \includegraphics[width=\linewidth]{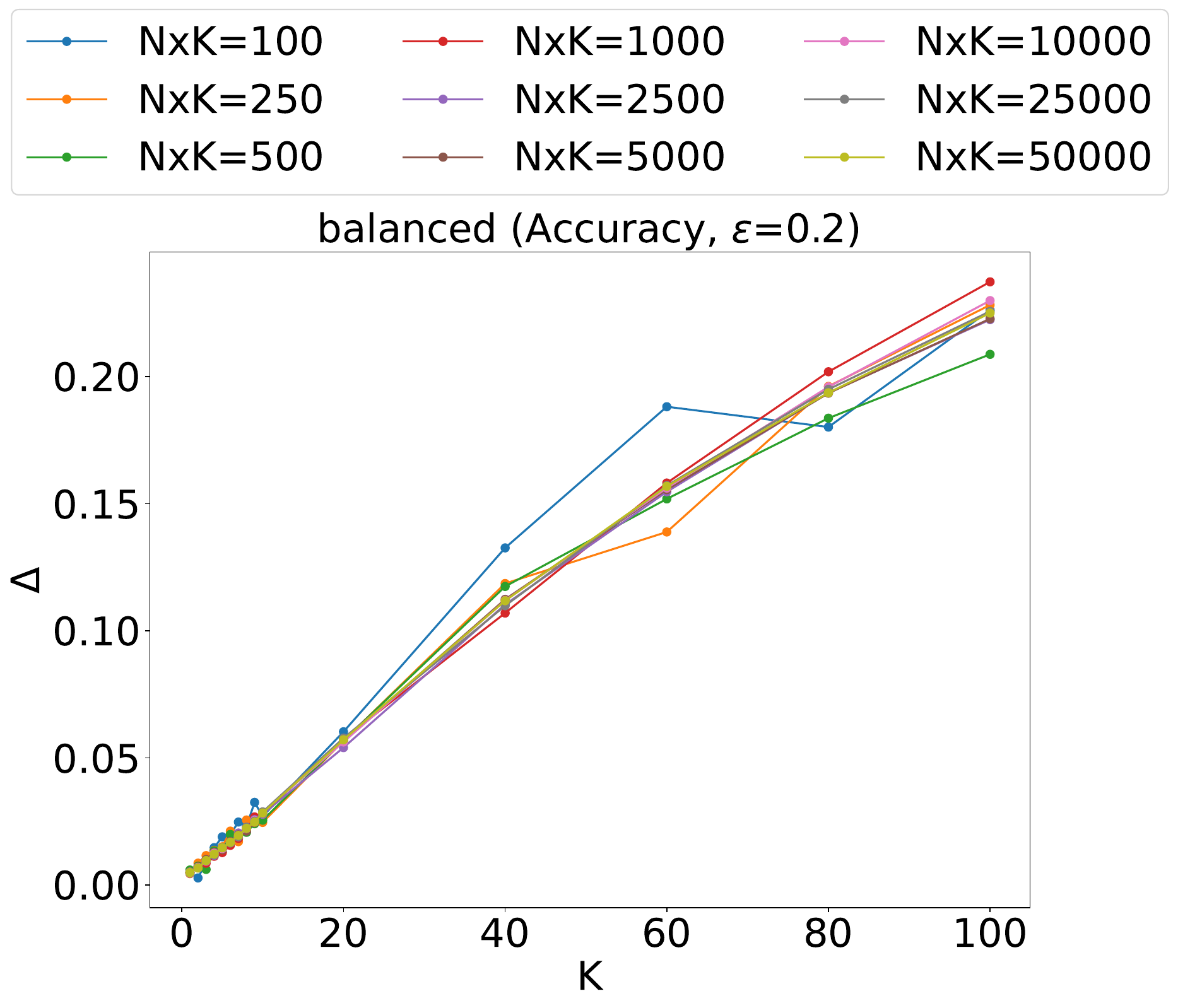}
    \caption{$\epsilon = 0.2$}
    \label{fig:uniform_delta_accuracy_cat12_e02}
  \end{subfigure} \hfill
  \begin{subfigure}[b]{0.24\linewidth}
    \centering
    \includegraphics[width=\linewidth]{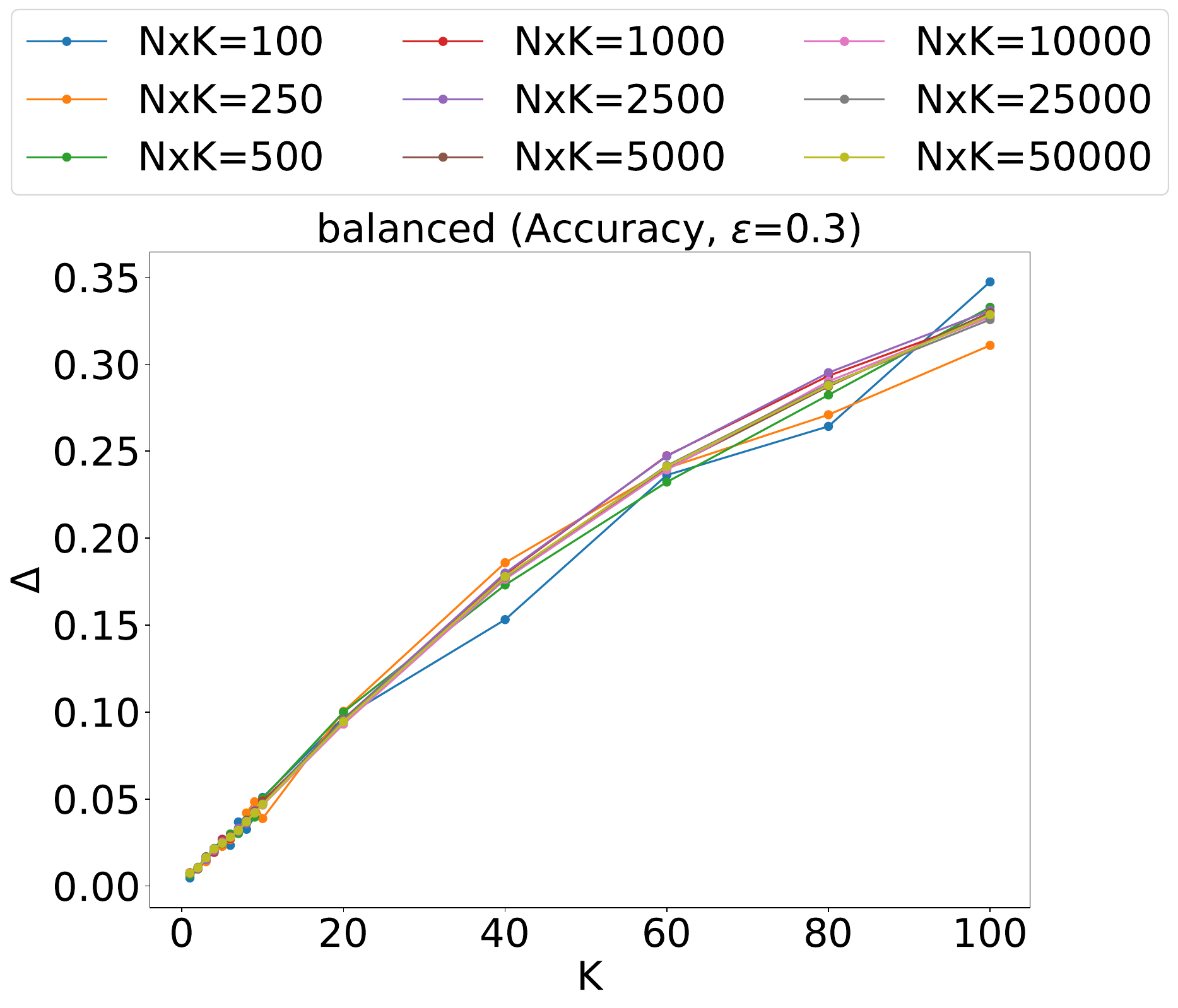}
    \caption{$\epsilon = 0.3$}
    \label{fig:uniform_delta_accuracy_cat12_e03}
  \end{subfigure} \hfill
  \begin{subfigure}[b]{0.24\linewidth}
    \centering
    \includegraphics[width=\linewidth]{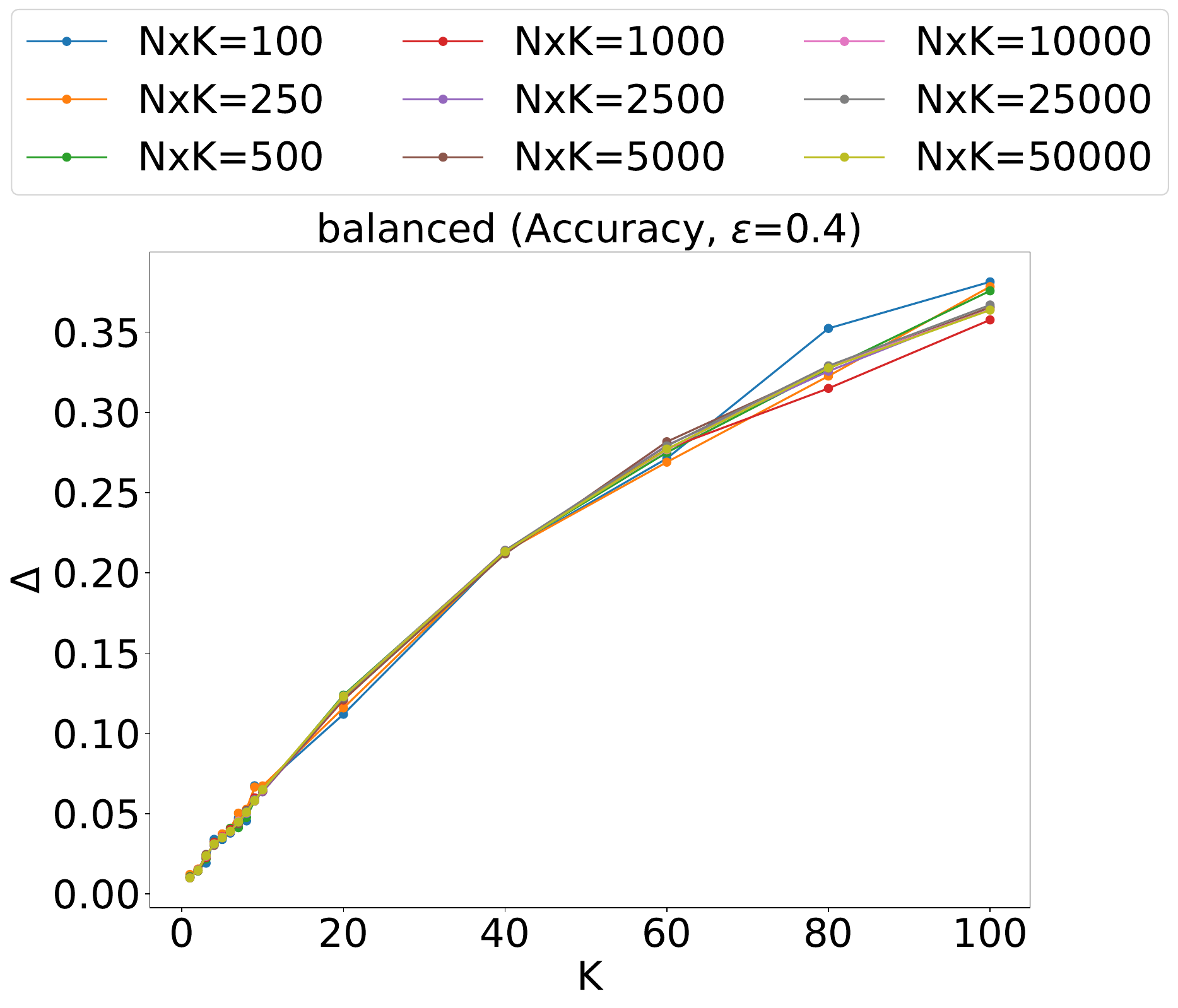}
    \caption{$\epsilon = 0.4$}
    \label{fig:uniform_delta_accuracy_cat12_e04}
  \end{subfigure}
  \caption{Effect sizes ($\Delta$) for balanced alphas with Accuracy as the metric ($M=12$)}
  \label{fig:uniform_delta_accuracy_cat12}
\end{figure*}

Figures \ref{fig:uniform_accuracy_cat2}--\ref{fig:uniform_delta_accuracy_cat12} show results for \pv s, confidence intervals, and effect sizes for balanced priors on parameter $\alpha$ with different values of $\epsilon$ and increasing $K$ for various $N \times K$. For simulations with balanced categories, \pv s\ decrease with increasing $K$ as $M$ gets larger. CI-width and effect sizes increase as $K$ increases.

\subsubsection{Artificial Distribution with Unbalanced Priors}

\begin{figure*}
  \centering
  \begin{subfigure}[b]{0.24\linewidth}
    \centering
    \includegraphics[width=\linewidth]{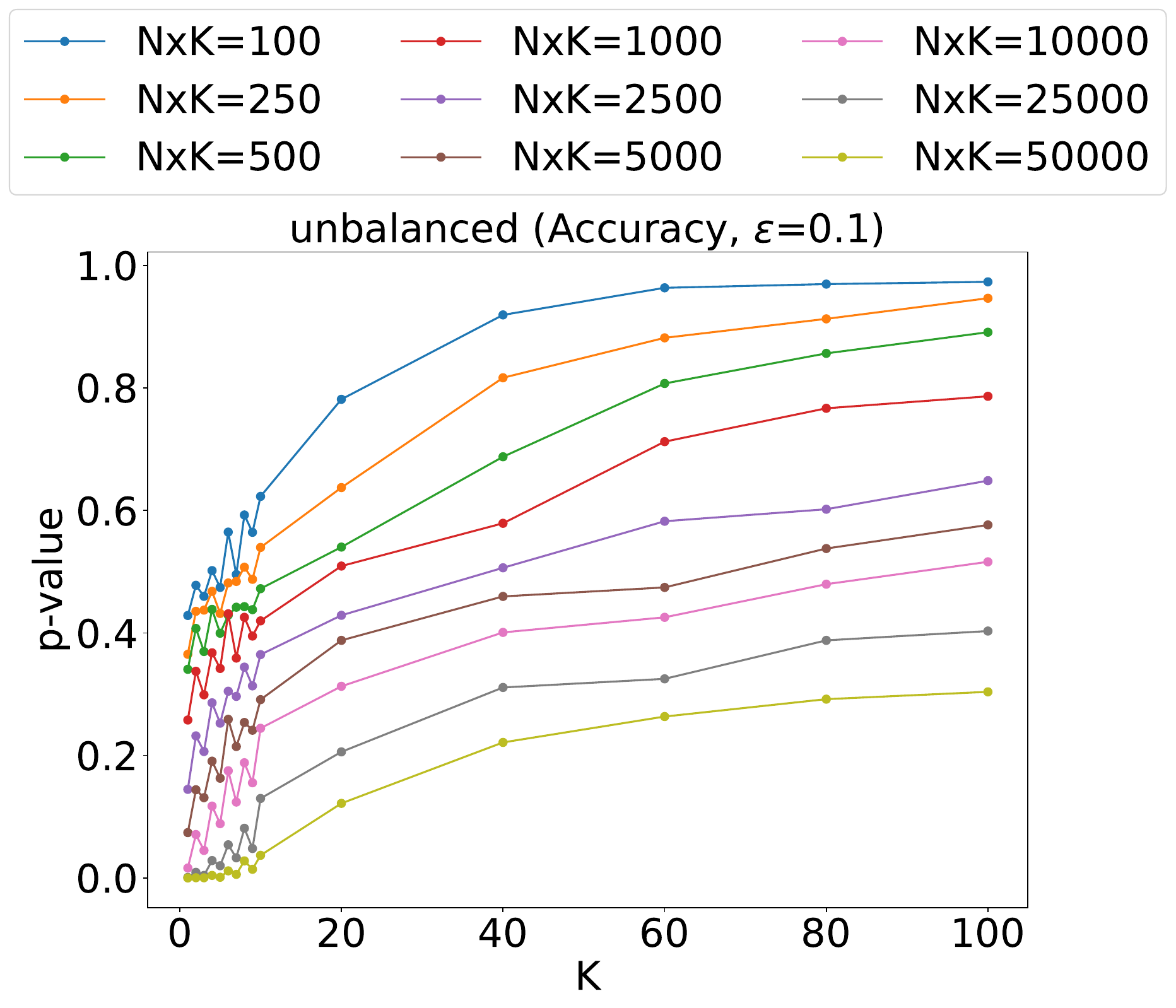}
    \caption{$\epsilon = 0.1$}
    \label{fig:gamma_accuracy_cat2_e01}
  \end{subfigure} \hfill
  \begin{subfigure}[b]{0.24\linewidth}
    \centering
    \includegraphics[width=\linewidth]{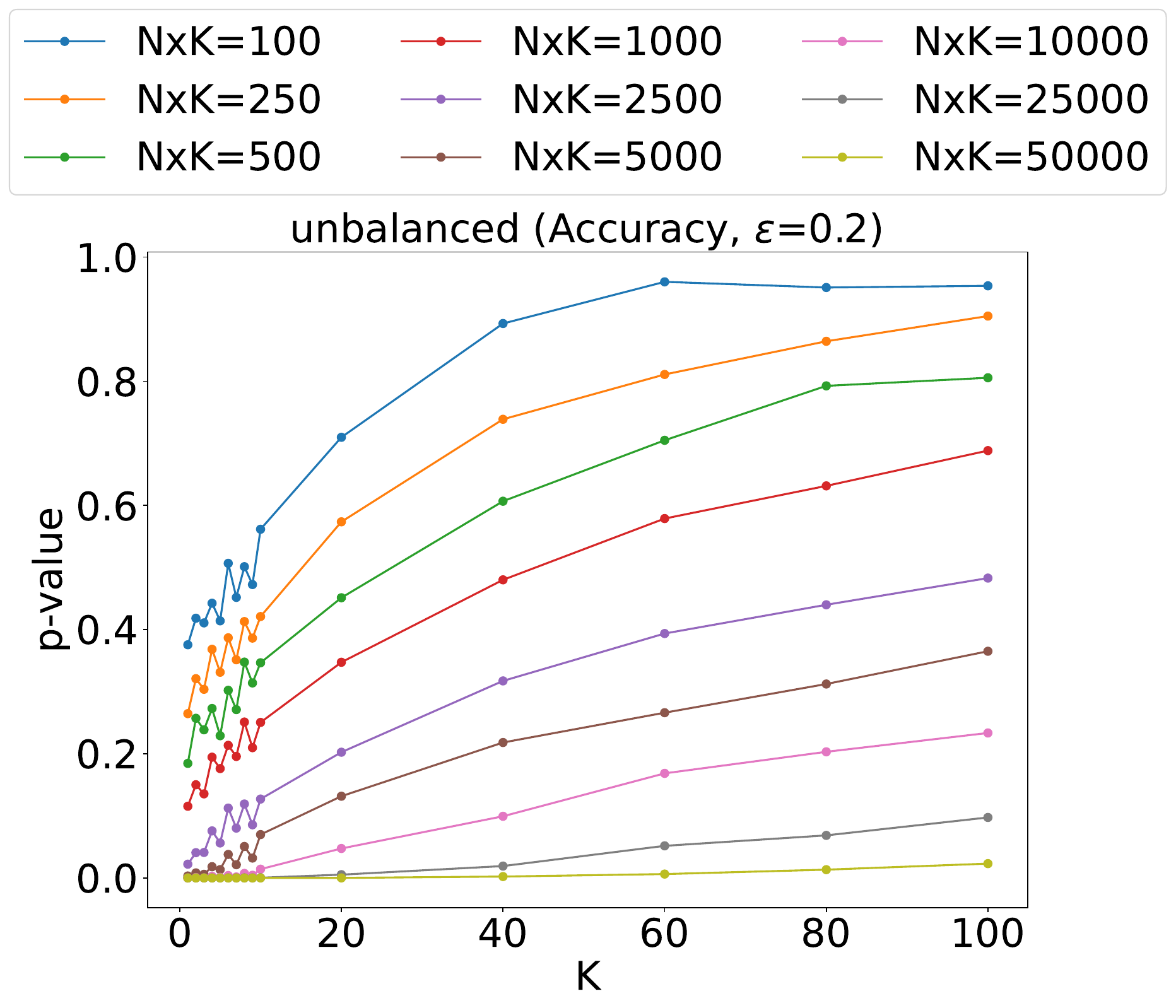}
    \caption{$\epsilon = 0.2$}
    \label{fig:gamma_accuracy_cat2_e02}
  \end{subfigure} \hfill
  \begin{subfigure}[b]{0.24\linewidth}
    \centering
    \includegraphics[width=\linewidth]{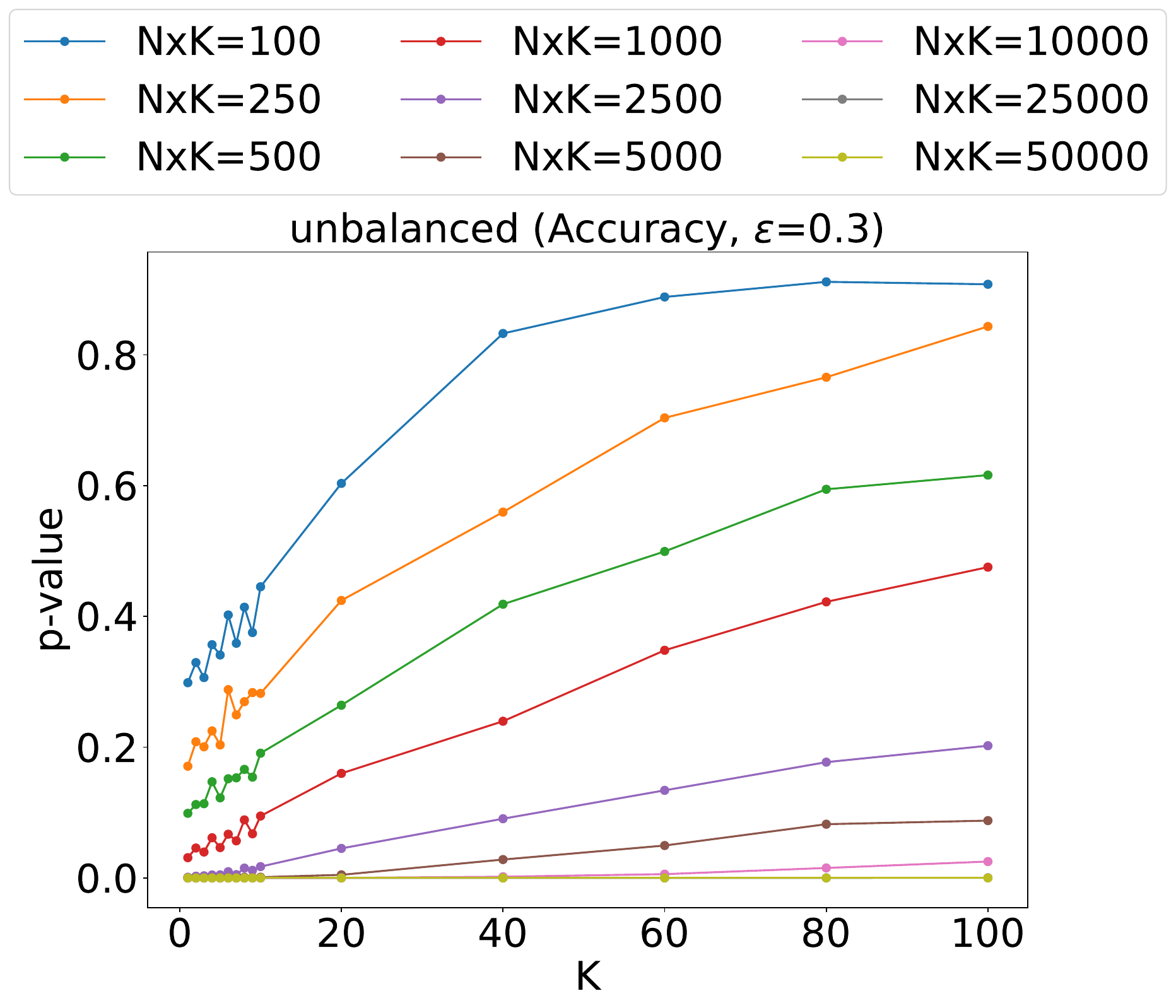}
    \caption{$\epsilon = 0.3$}
    \label{fig:gamma_accuracy_cat2_e03}
  \end{subfigure} \hfill
  \begin{subfigure}[b]{0.24\linewidth}
    \centering
    \includegraphics[width=\linewidth]{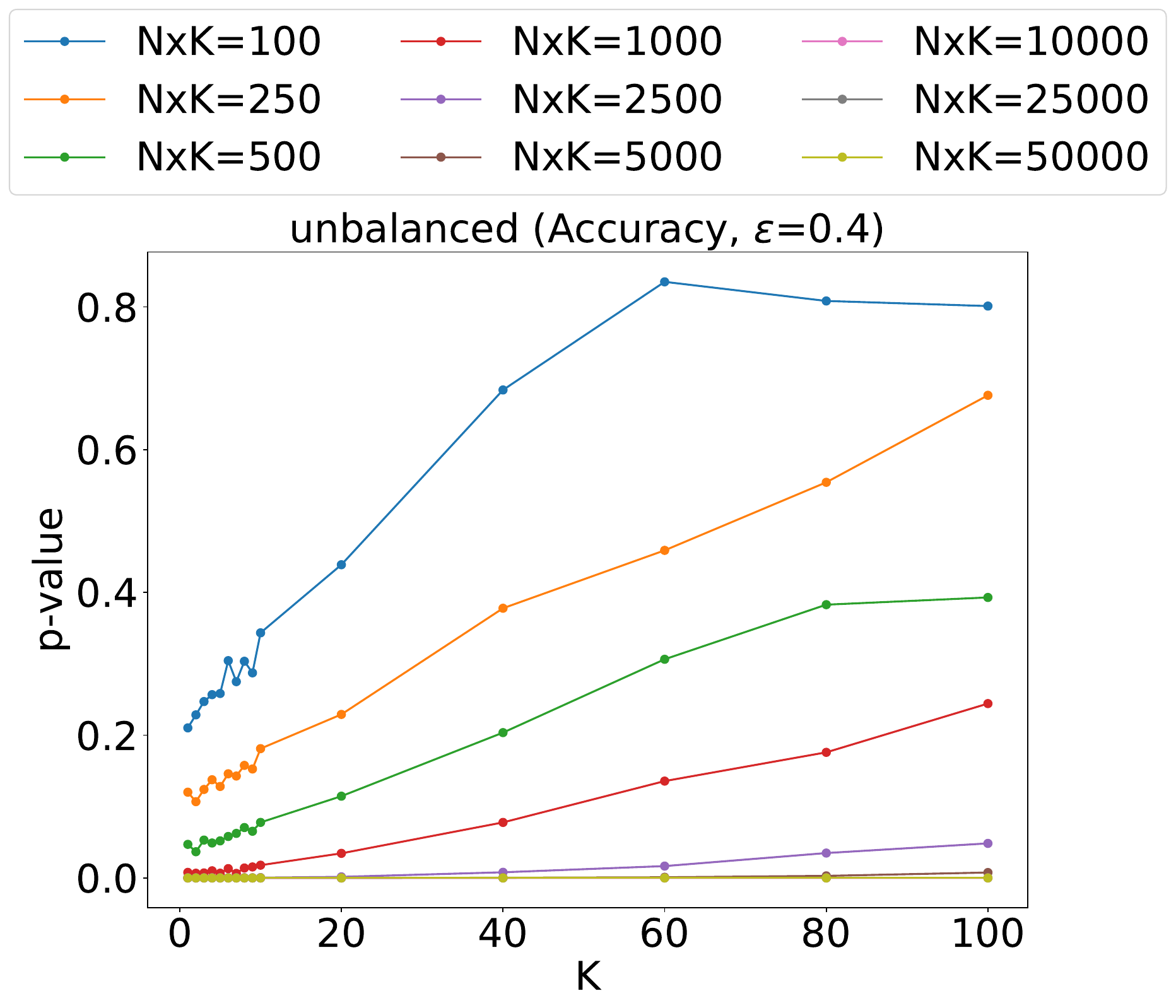}
    \caption{$\epsilon = 0.4$}
    \label{fig:gamma_accuracy_cat2_e04}
  \end{subfigure}
  \caption{P-value plots for unbalanced alphas with Accuracy as the metric ($M=2$)}
  \label{fig:gamma_accuracy_cat2}
\end{figure*}

\begin{figure*}
  \centering
  \begin{subfigure}[b]{0.24\linewidth}
    \centering
    \includegraphics[width=\linewidth]{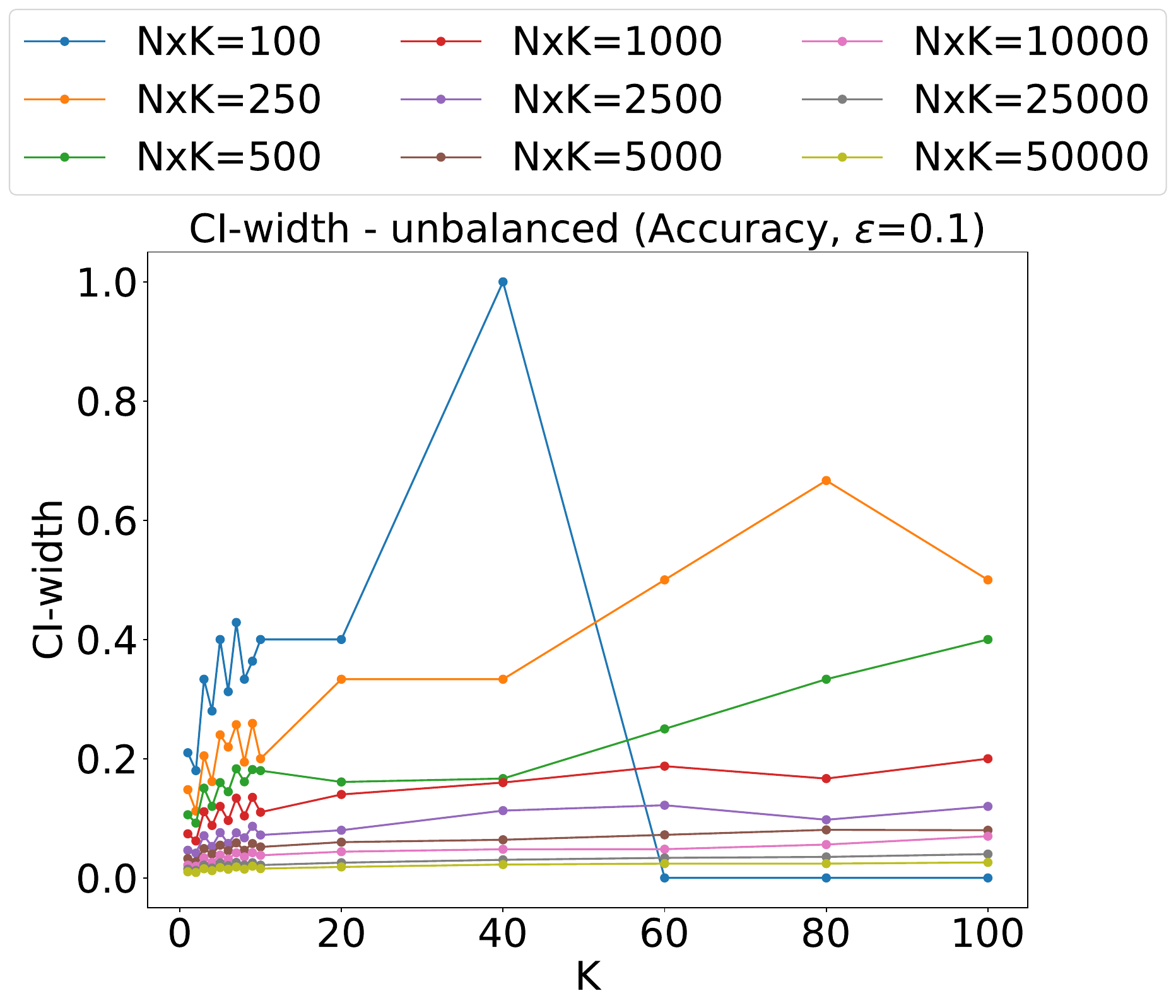}
    \caption{$\epsilon = 0.1$}
    \label{fig:gamma_ci_accuracy_cat2_e01}
  \end{subfigure} \hfill
  \begin{subfigure}[b]{0.24\linewidth}
    \centering
    \includegraphics[width=\linewidth]{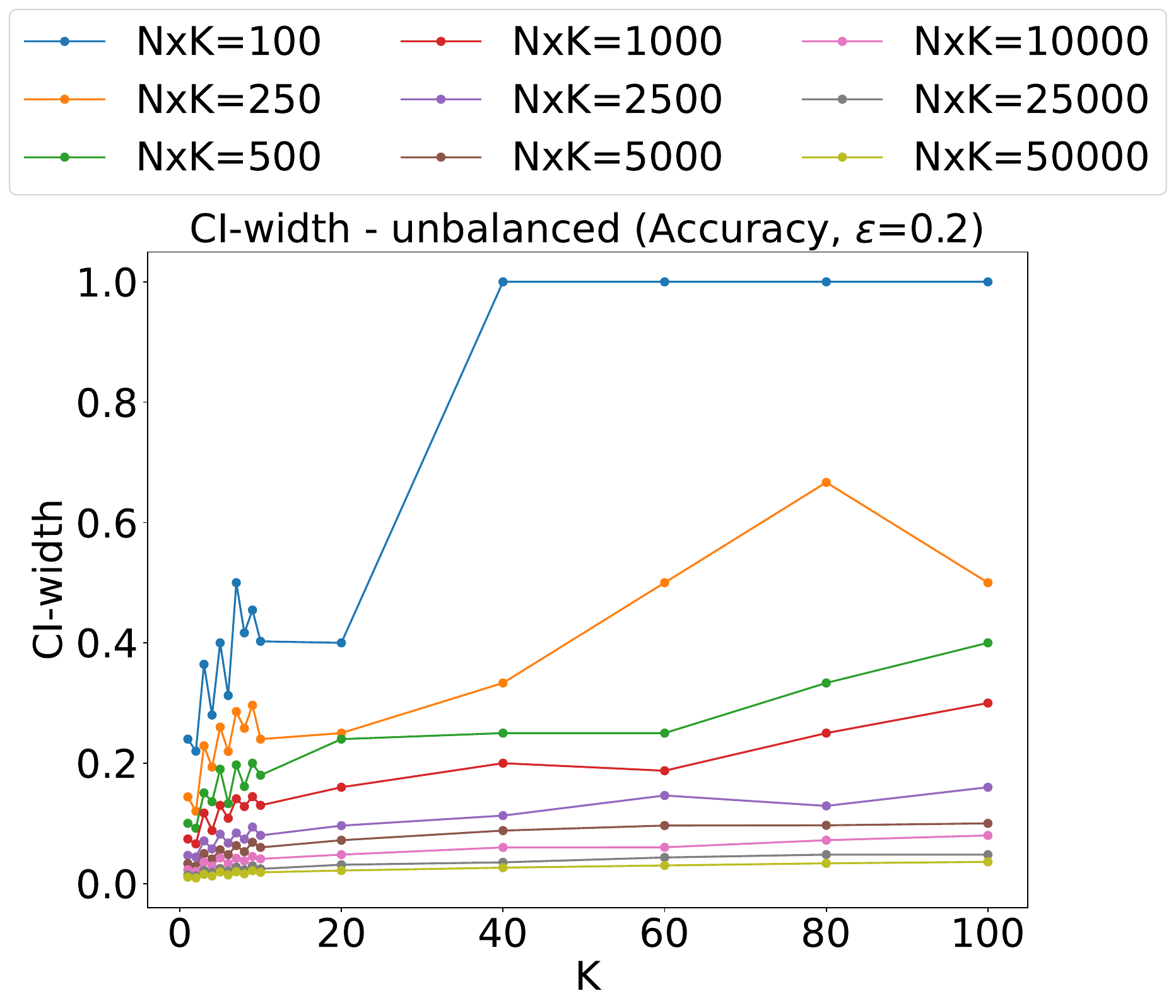}
    \caption{$\epsilon = 0.2$}
    \label{fig:gamma_ci_accuracy_cat2_e02}
  \end{subfigure} \hfill
  \begin{subfigure}[b]{0.24\linewidth}
    \centering
    \includegraphics[width=\linewidth]{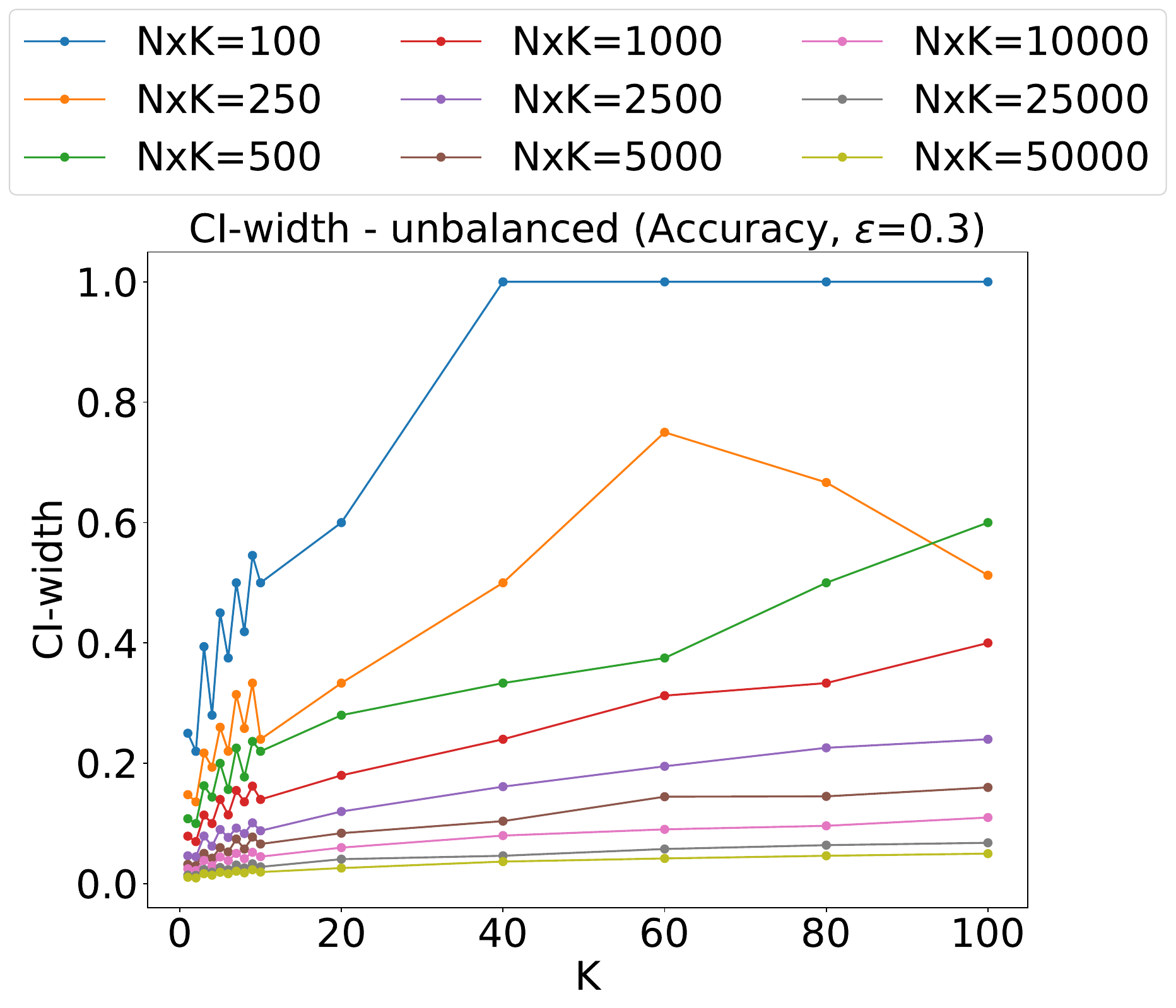}
    \caption{$\epsilon = 0.3$}
    \label{fig:gamma_ci_accuracy_cat2_e03}
  \end{subfigure} \hfill
  \begin{subfigure}[b]{0.24\linewidth}
    \centering
    \includegraphics[width=\linewidth]{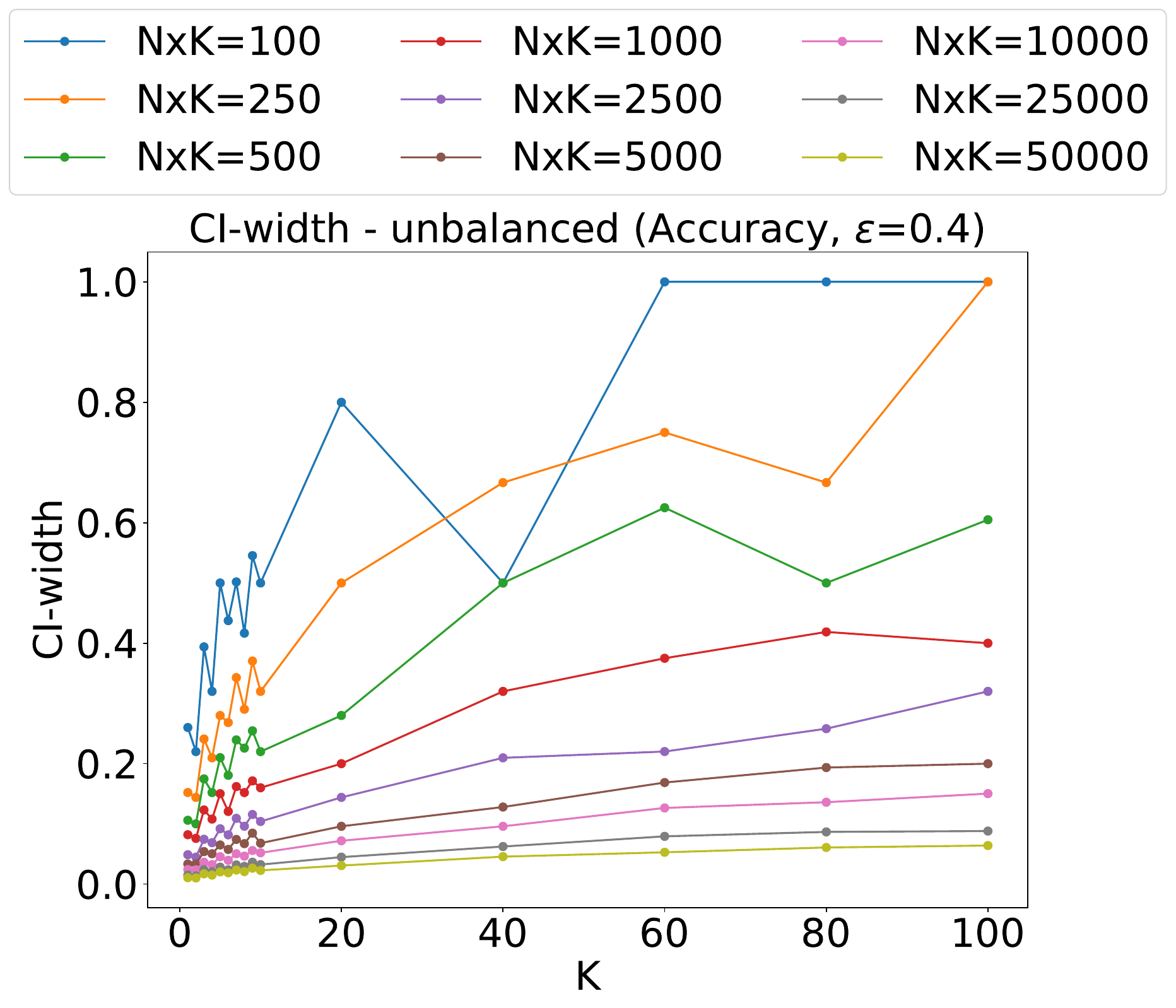}
    \caption{$\epsilon = 0.4$}
    \label{fig:gamma_ci_accuracy_cat2_e04}
  \end{subfigure}
  \caption{CI-width plots for unbalanced alphas with Accuracy as the metric ($M=2$)}
  \label{fig:gamma_ci_accuracy_cat2}
\end{figure*}

\begin{figure*}
  \centering
  \begin{subfigure}[b]{0.24\linewidth}
    \centering
    \includegraphics[width=\linewidth]{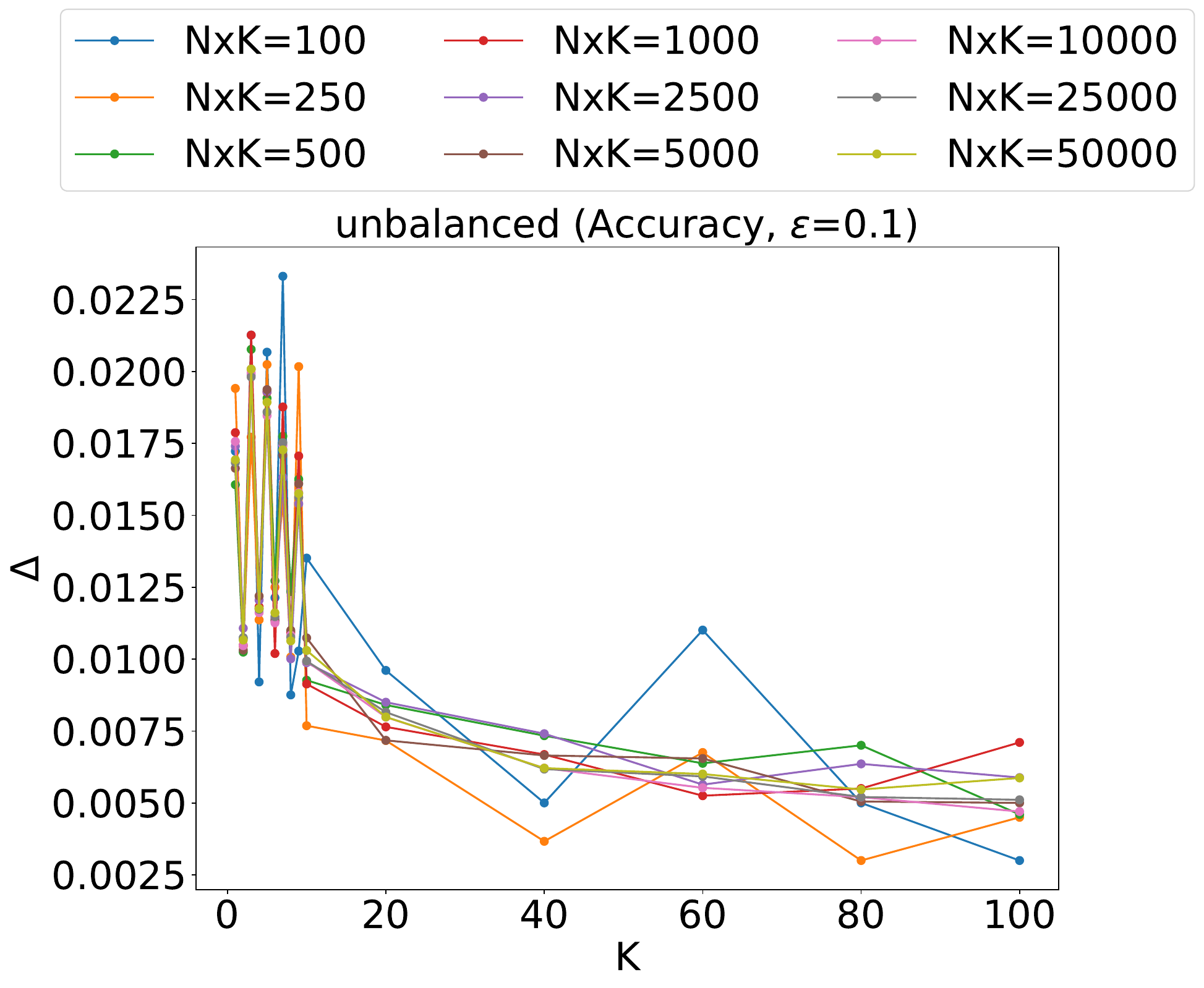}
    \caption{$\epsilon = 0.1$}
    \label{fig:gamma_delta_accuracy_cat2_e01}
  \end{subfigure} \hfill
  \begin{subfigure}[b]{0.24\linewidth}
    \centering
    \includegraphics[width=\linewidth]{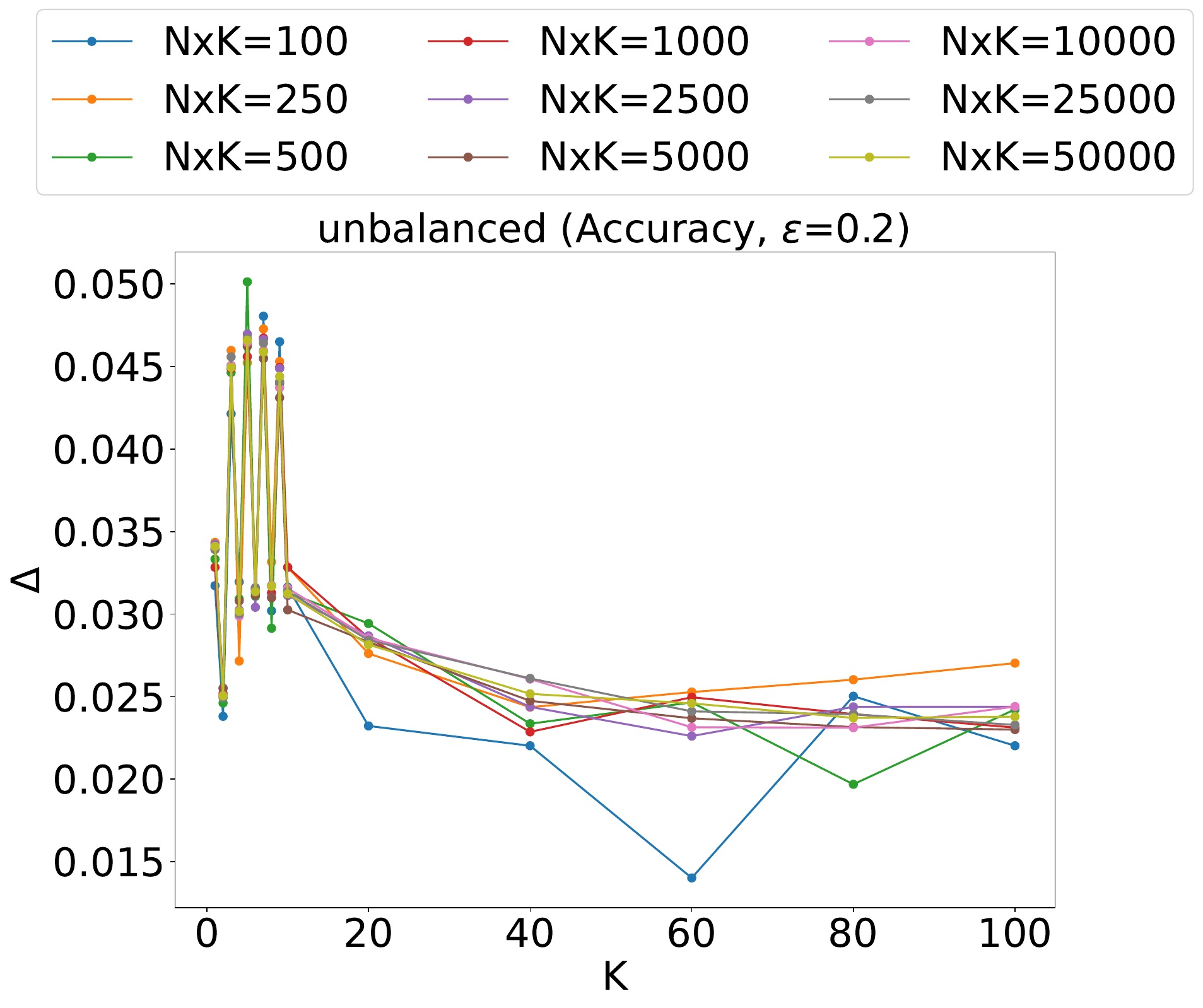}
    \caption{$\epsilon = 0.2$}
    \label{fig:gamma_delta_accuracy_cat2_e02}
  \end{subfigure} \hfill
  \begin{subfigure}[b]{0.24\linewidth}
    \centering
    \includegraphics[width=\linewidth]{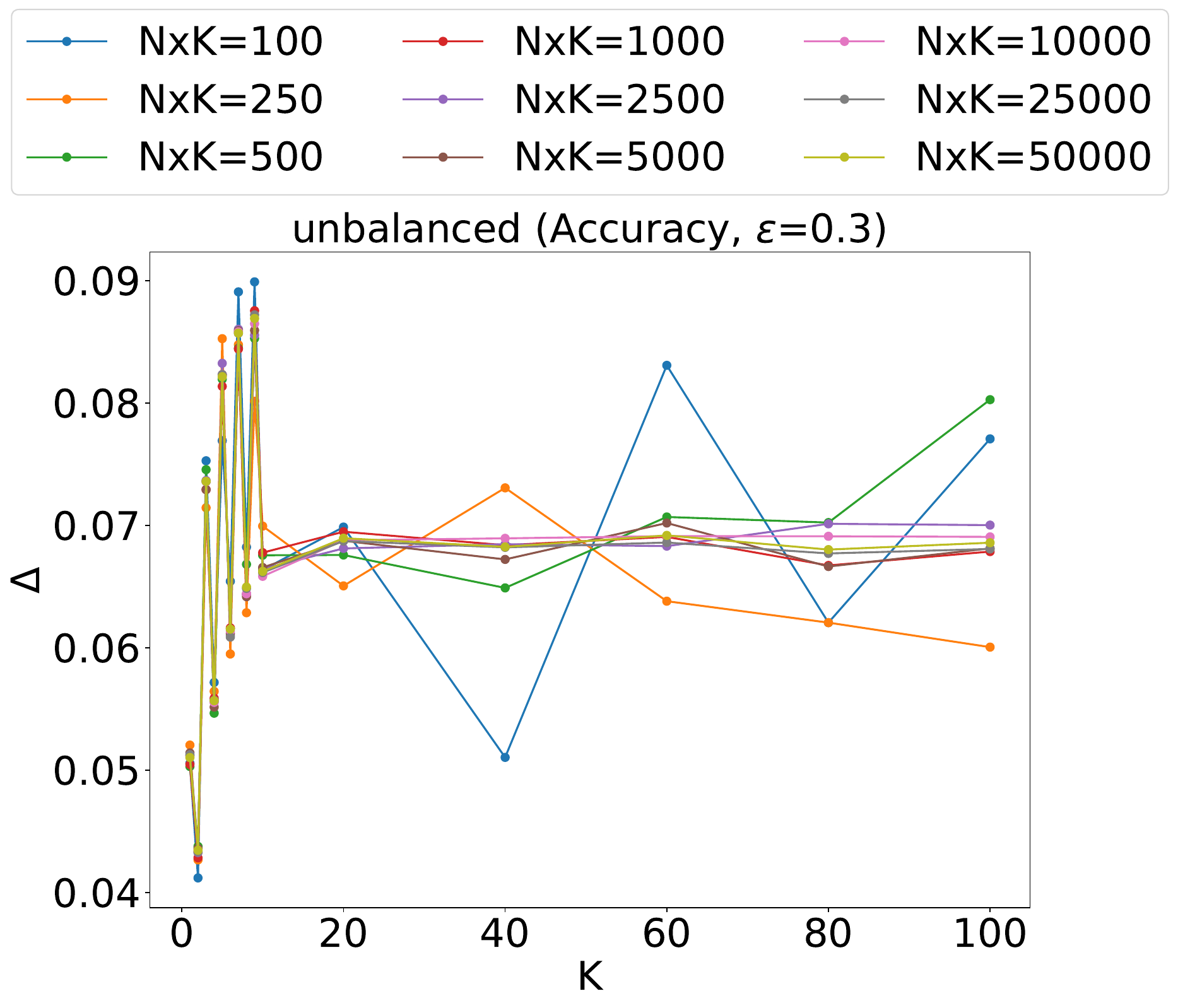}
    \caption{$\epsilon = 0.3$}
    \label{fig:gamma_delta_accuracy_cat2_e03}
  \end{subfigure} \hfill
  \begin{subfigure}[b]{0.24\linewidth}
    \centering
    \includegraphics[width=\linewidth]{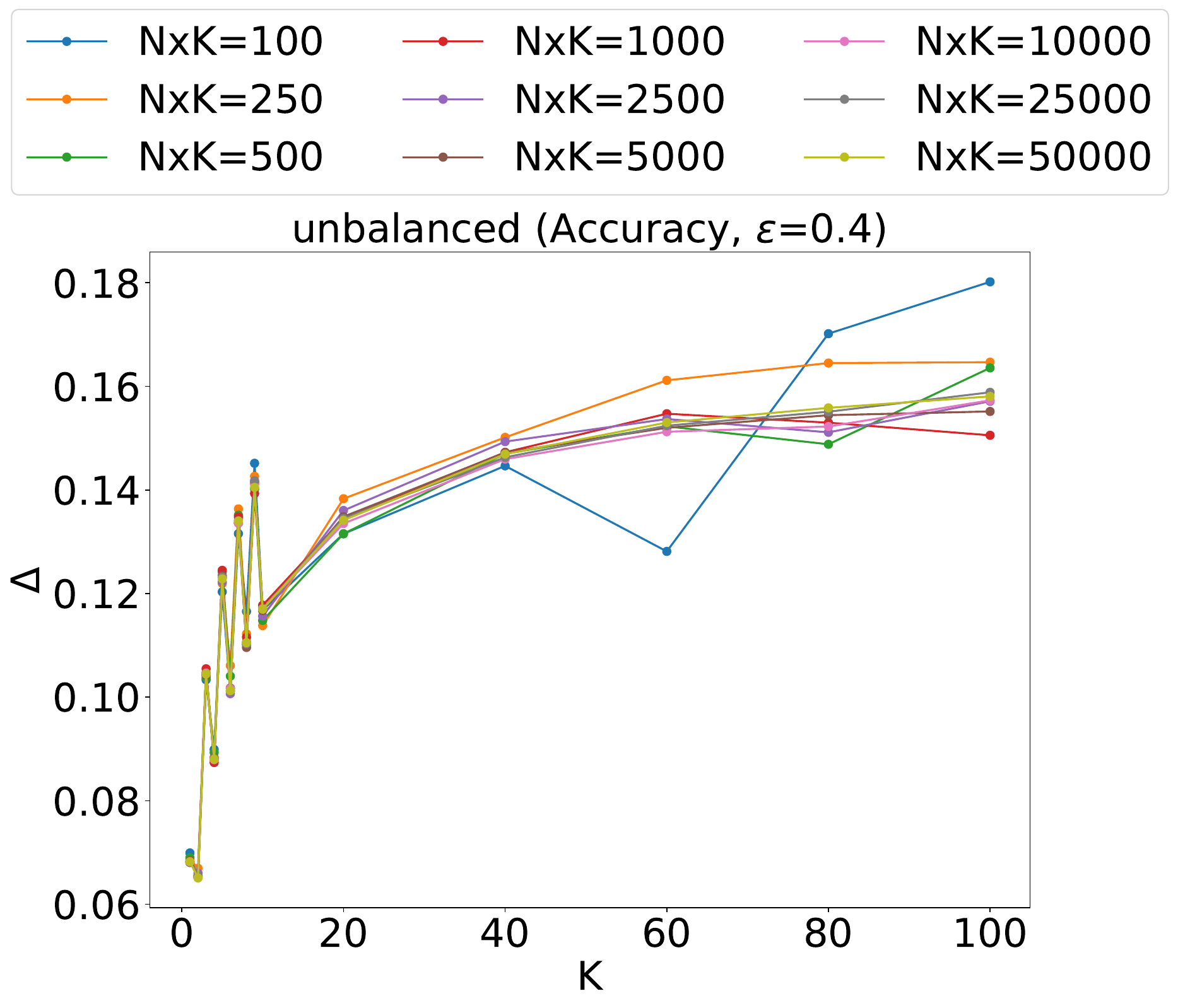}
    \caption{$\epsilon = 0.4$}
    \label{fig:gamma_delta_accuracy_cat2_e04}
  \end{subfigure}
  \caption{Effect sizes ($\Delta$) for unbalanced alphas with Accuracy as the metric ($M=2$)}
  \label{fig:gamma_delta_accuracy_cat2}
\end{figure*}

\begin{figure*}
  \centering
  \begin{subfigure}[b]{0.24\linewidth}
    \centering
    \includegraphics[width=\linewidth]{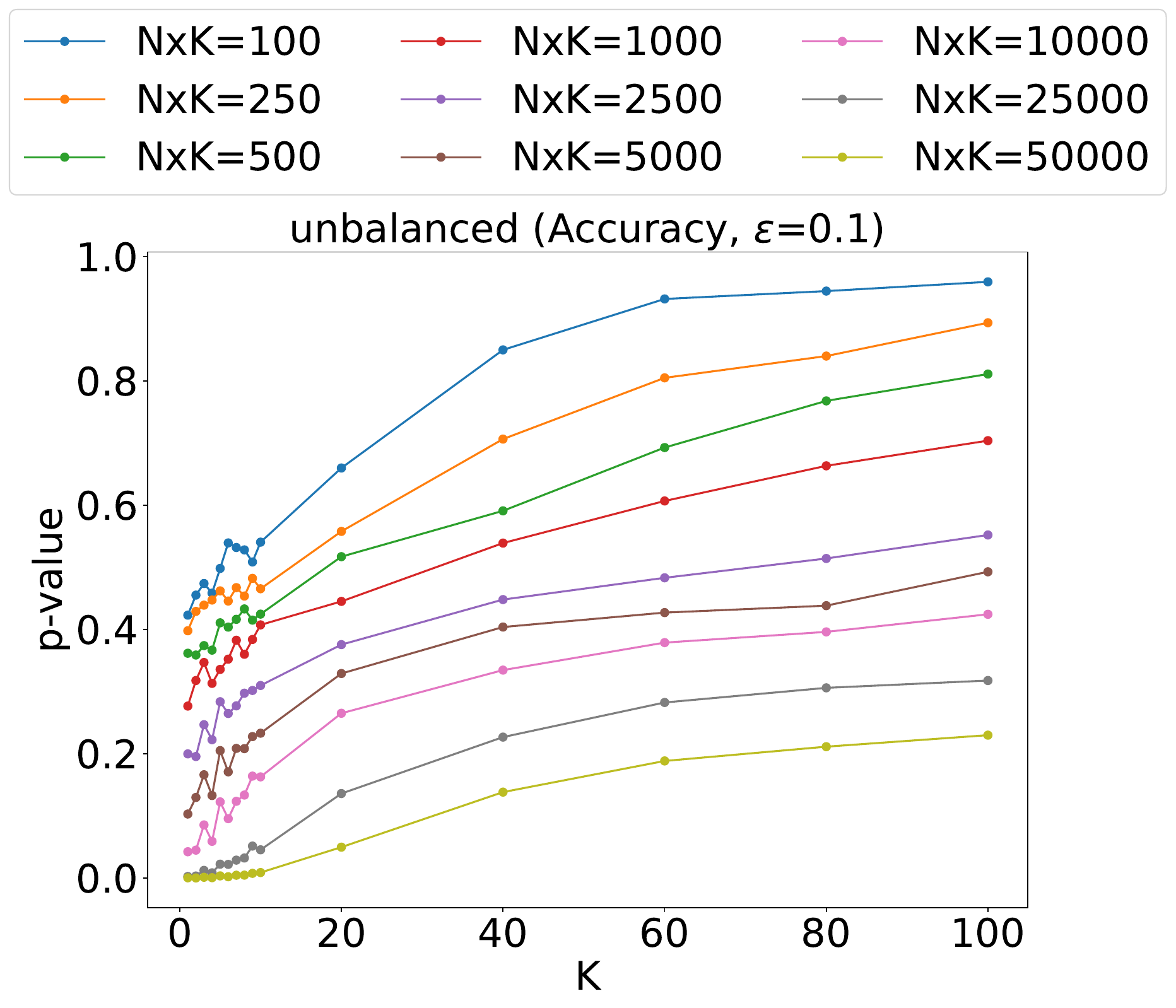}
    \caption{$\epsilon = 0.1$}
    \label{fig:gamma_accuracy_cat3_e01}
  \end{subfigure} \hfill
  \begin{subfigure}[b]{0.24\linewidth}
    \centering
    \includegraphics[width=\linewidth]{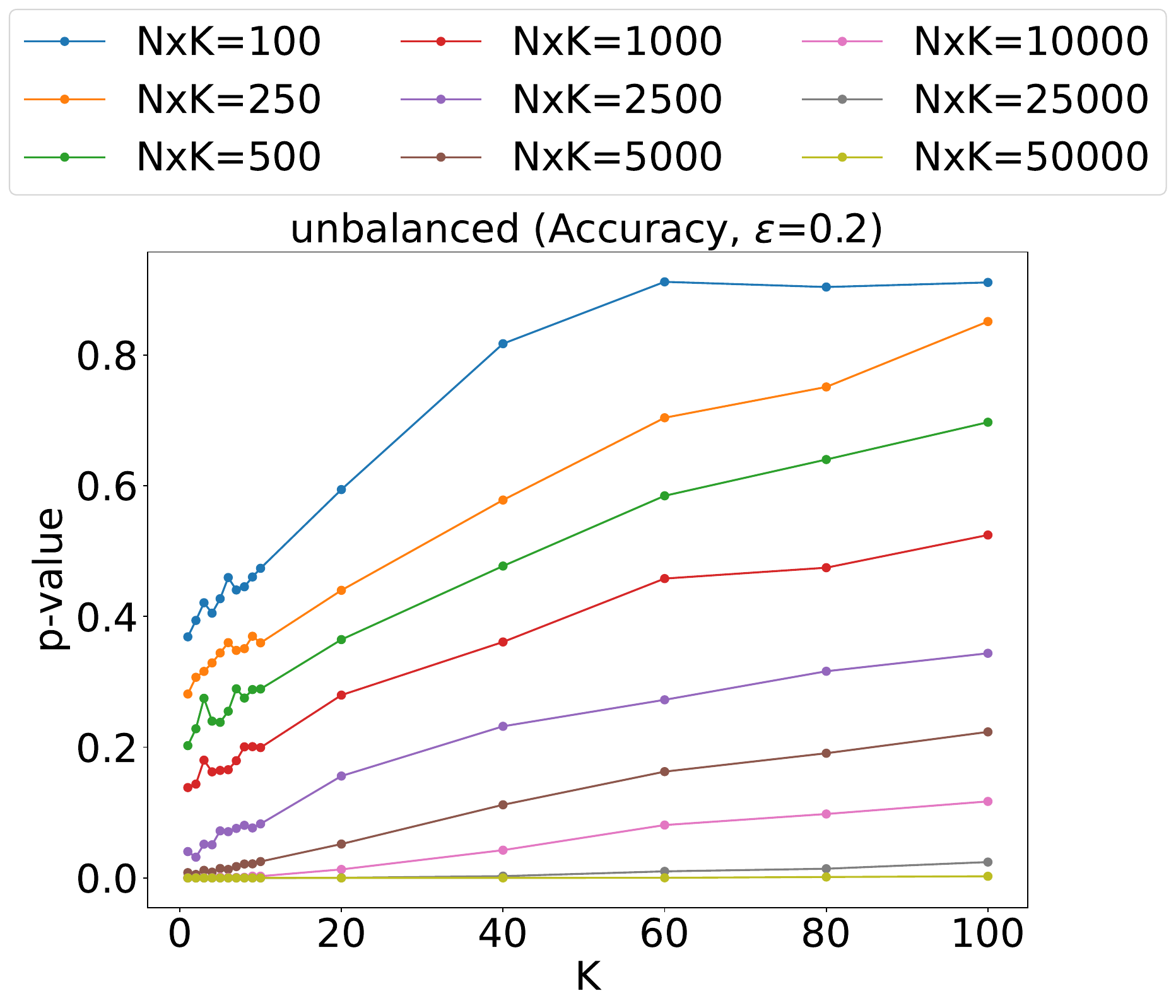}
    \caption{$\epsilon = 0.2$}
    \label{fig:gamma_accuracy_cat3_e02}
  \end{subfigure} \hfill
  \begin{subfigure}[b]{0.24\linewidth}
    \centering
    \includegraphics[width=\linewidth]{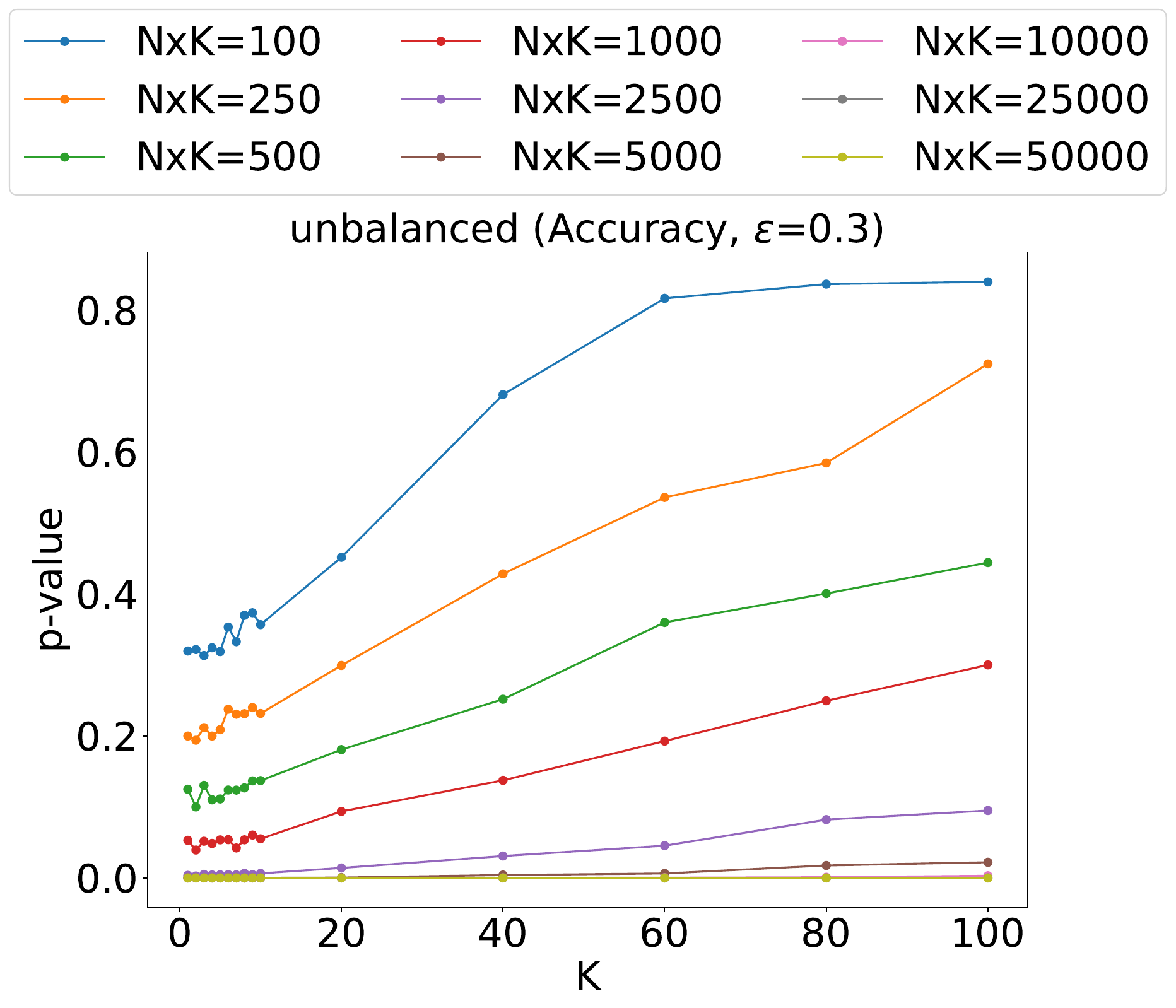}
    \caption{$\epsilon = 0.3$}
    \label{fig:gamma_accuracy_cat3_e03}
  \end{subfigure} \hfill
  \begin{subfigure}[b]{0.24\linewidth}
    \centering
    \includegraphics[width=\linewidth]{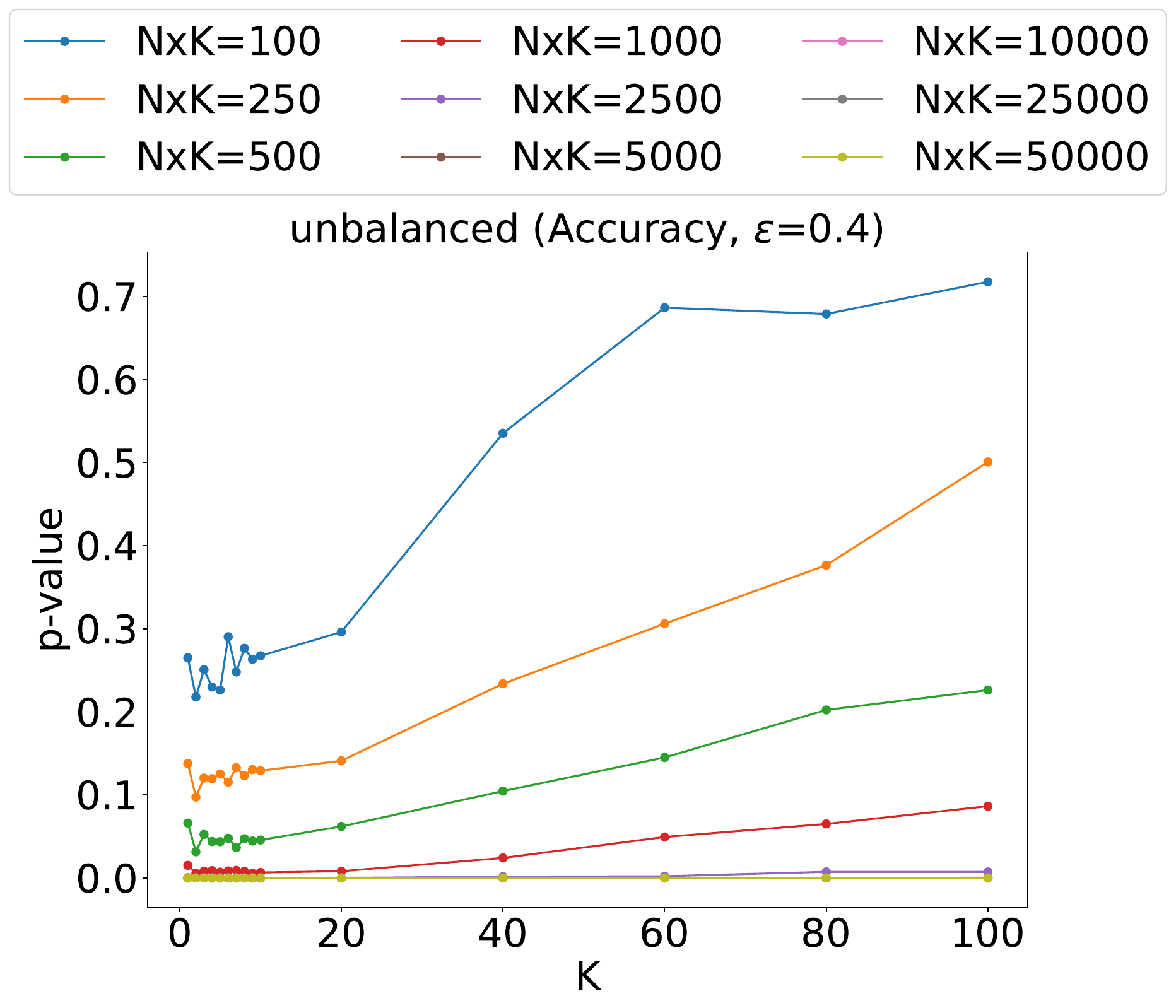}
    \caption{$\epsilon = 0.4$}
    \label{fig:gamma_accuracy_cat3_e04}
  \end{subfigure}
  \caption{P-value plots for unbalanced alphas with Accuracy as the metric ($M=3$)}
  \label{fig:gamma_accuracy_cat3}
\end{figure*}

\begin{figure*}
  \centering
  \begin{subfigure}[b]{0.24\linewidth}
    \centering
    \includegraphics[width=\linewidth]{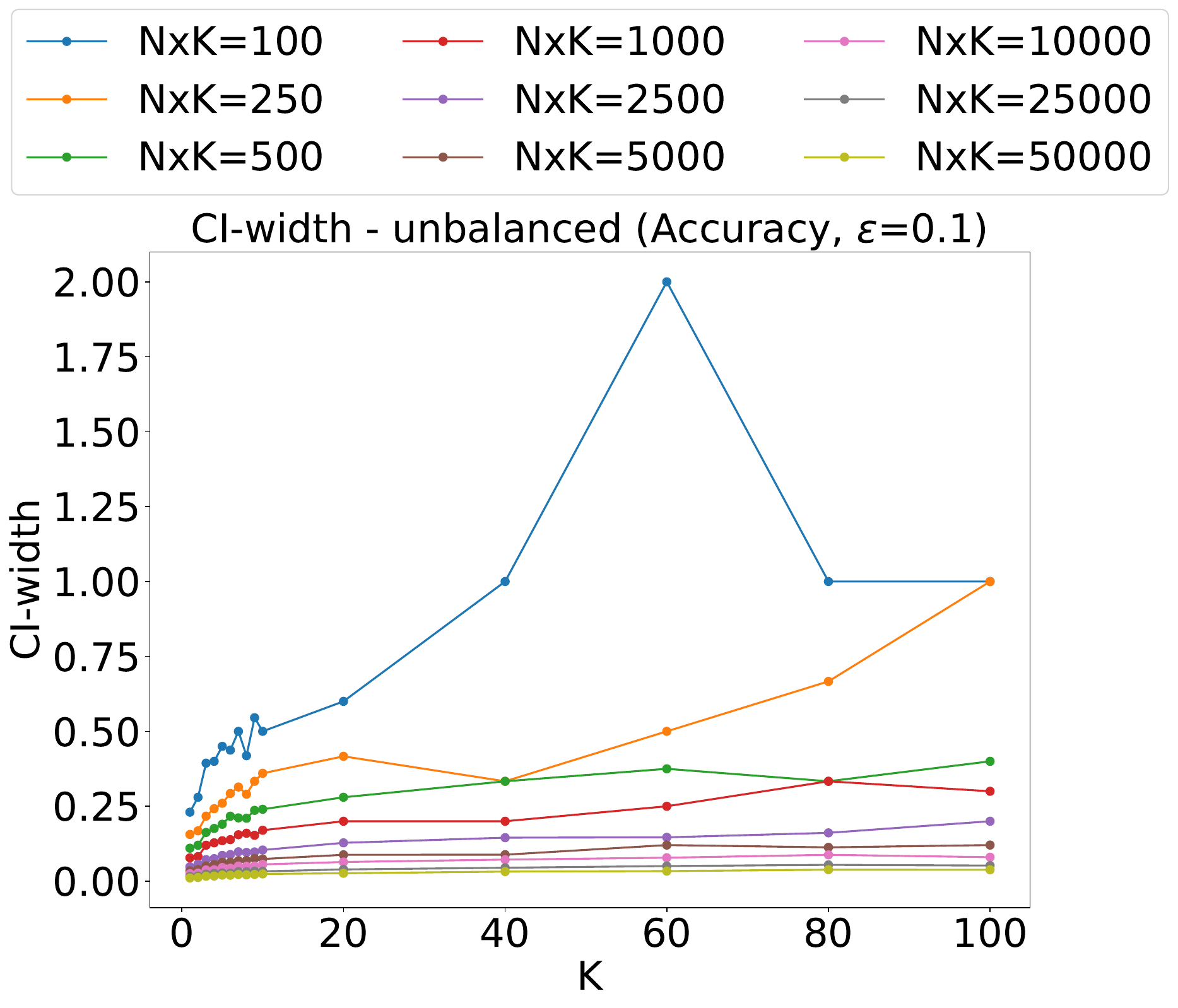}
    \caption{$\epsilon = 0.1$}
    \label{fig:gamma_ci_accuracy_cat3_e01}
  \end{subfigure} \hfill
  \begin{subfigure}[b]{0.24\linewidth}
    \centering
    \includegraphics[width=\linewidth]{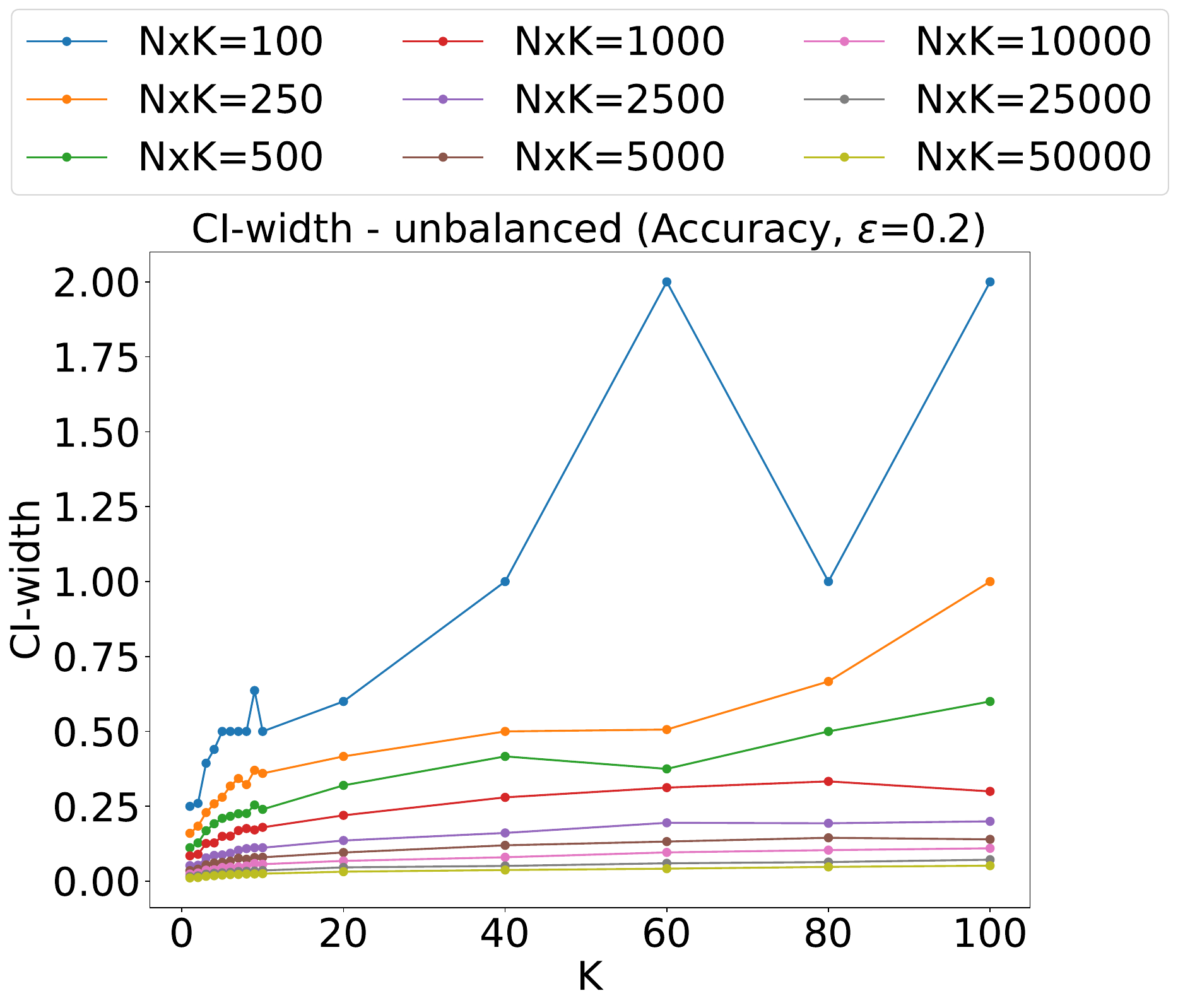}
    \caption{$\epsilon = 0.2$}
    \label{fig:gamma_ci_accuracy_cat3_e02}
  \end{subfigure} \hfill
  \begin{subfigure}[b]{0.24\linewidth}
    \centering
    \includegraphics[width=\linewidth]{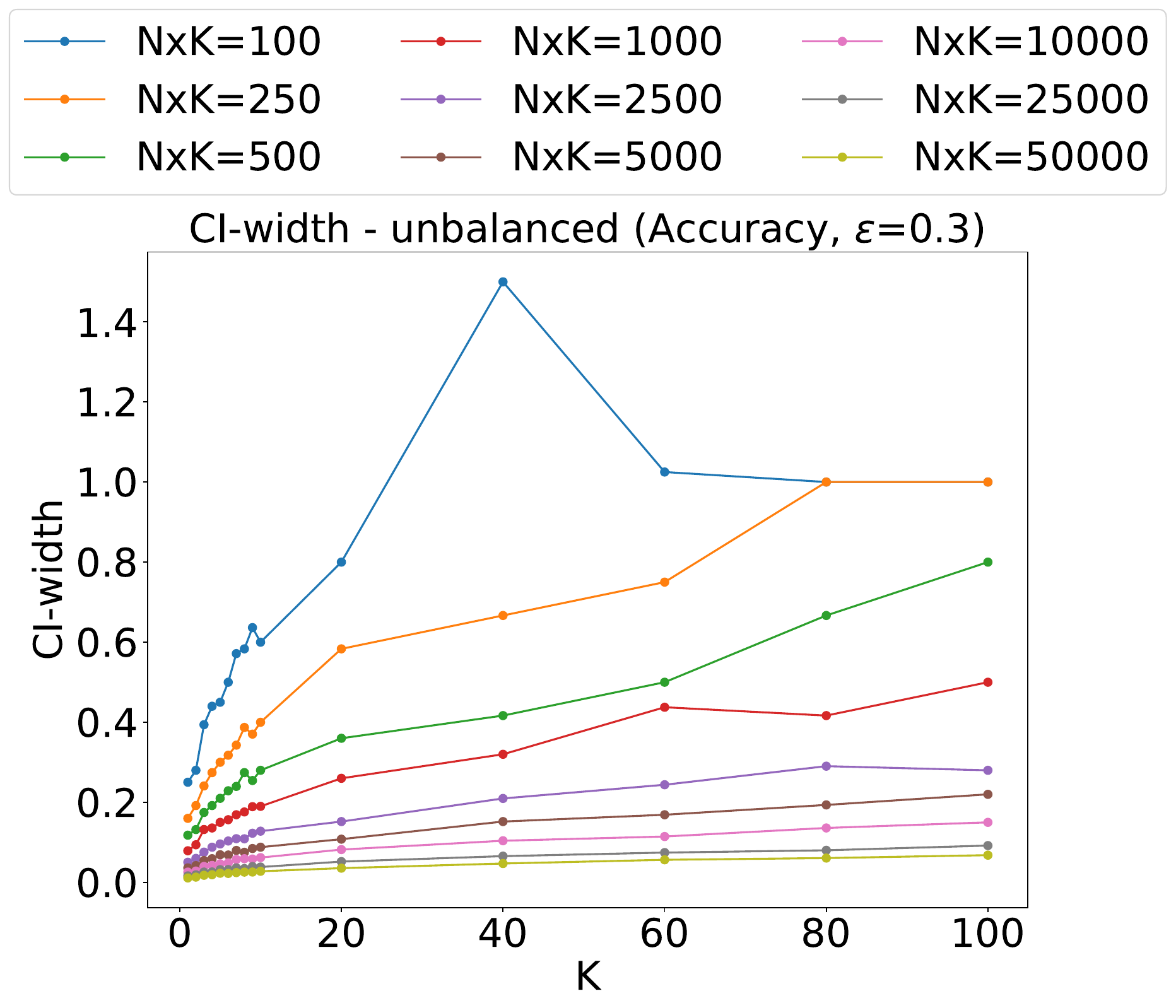}
    \caption{$\epsilon = 0.3$}
    \label{fig:gamma_ci_accuracy_cat3_e03}
  \end{subfigure} \hfill
  \begin{subfigure}[b]{0.24\linewidth}
    \centering
    \includegraphics[width=\linewidth]{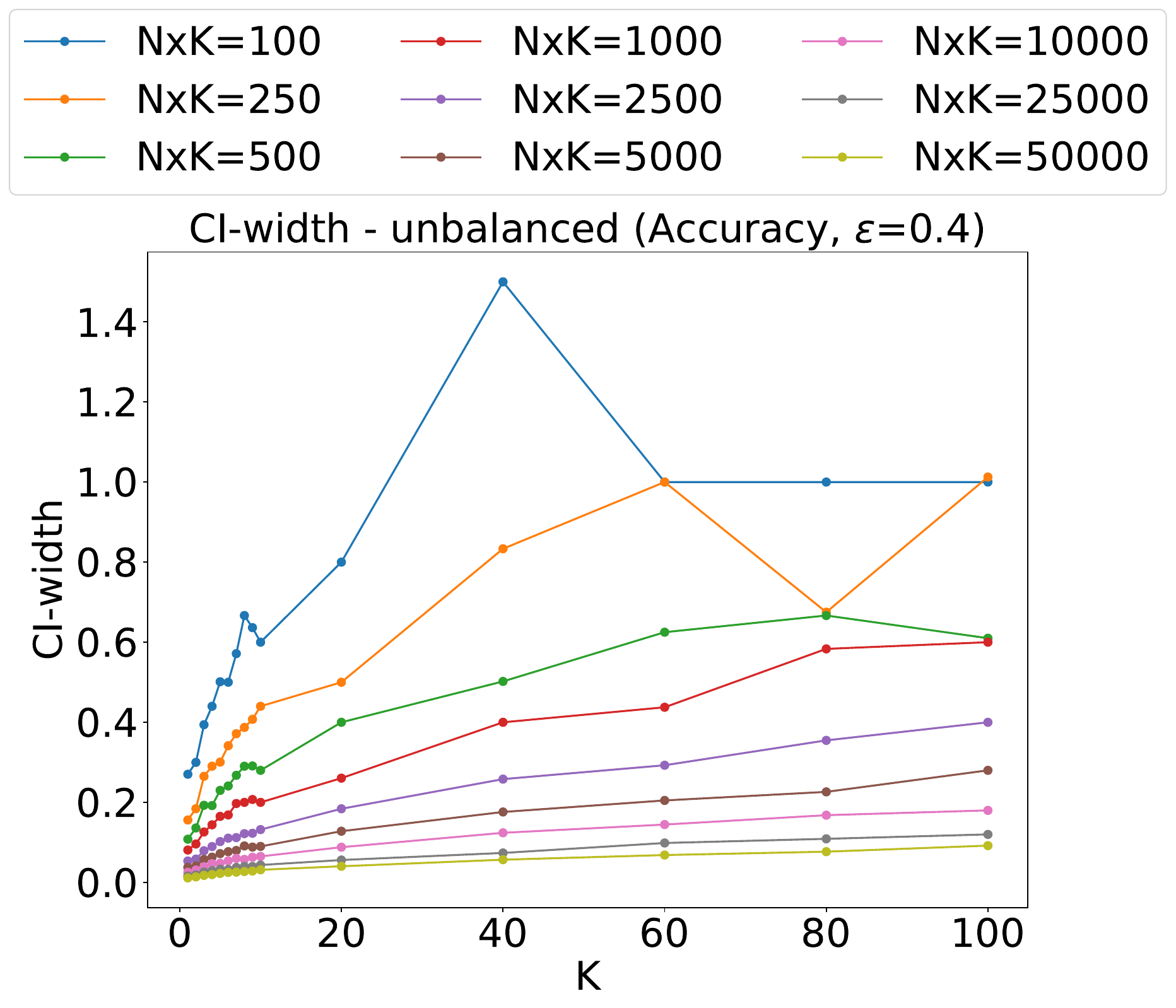}
    \caption{$\epsilon = 0.4$}
    \label{fig:gamma_ci_accuracy_cat3_e04}
  \end{subfigure}
  \caption{CI-width plots for unbalanced alphas with Accuracy as the metric ($M=3$)}
  \label{fig:gamma_ci_accuracy_cat3}
\end{figure*}

\begin{figure*}
  \centering
  \begin{subfigure}[b]{0.24\linewidth}
    \centering
    \includegraphics[width=\linewidth]{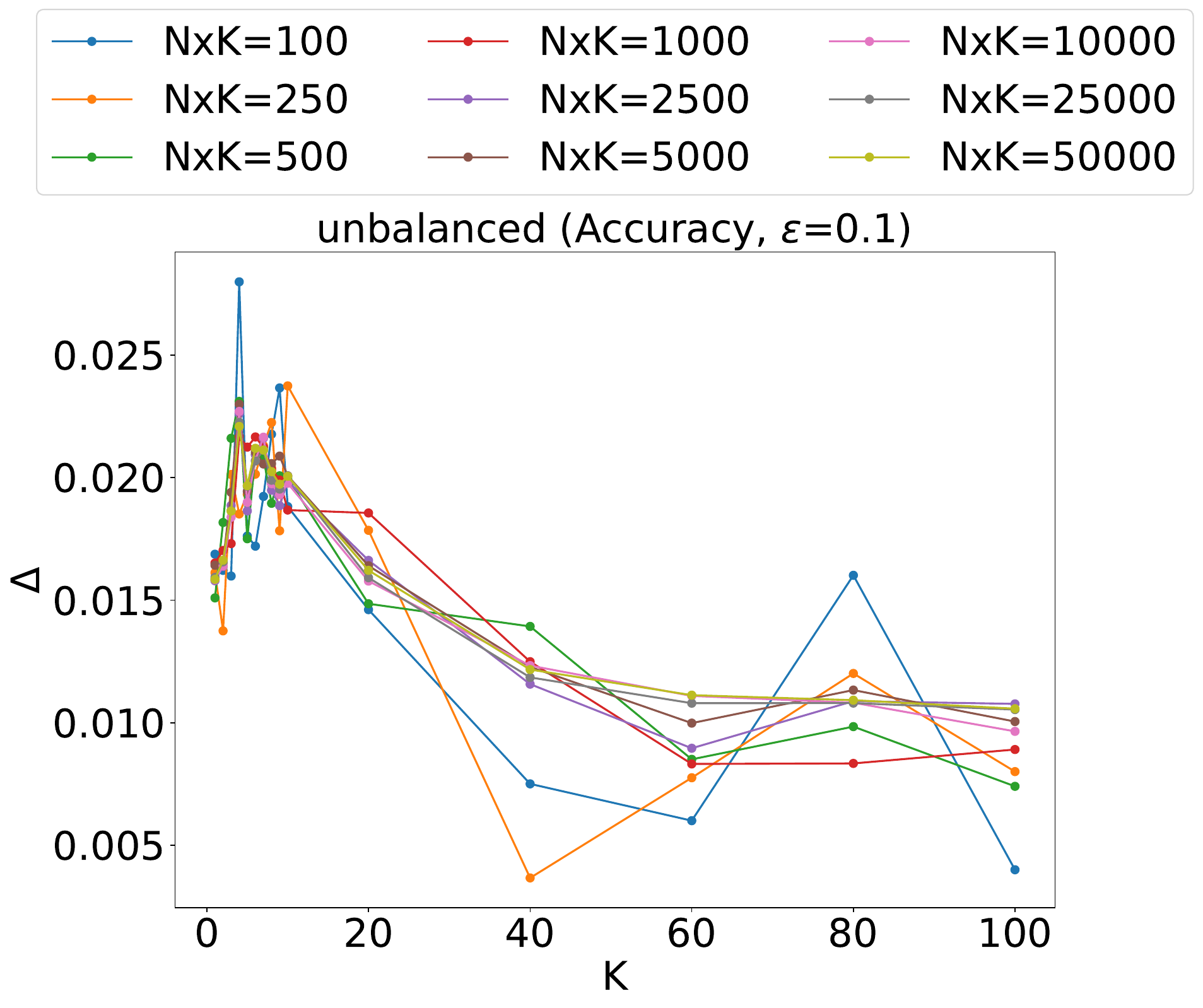}
    \caption{$\epsilon = 0.1$}
    \label{fig:gamma_delta_accuracy_cat3_e01}
  \end{subfigure} \hfill
  \begin{subfigure}[b]{0.24\linewidth}
    \centering
    \includegraphics[width=\linewidth]{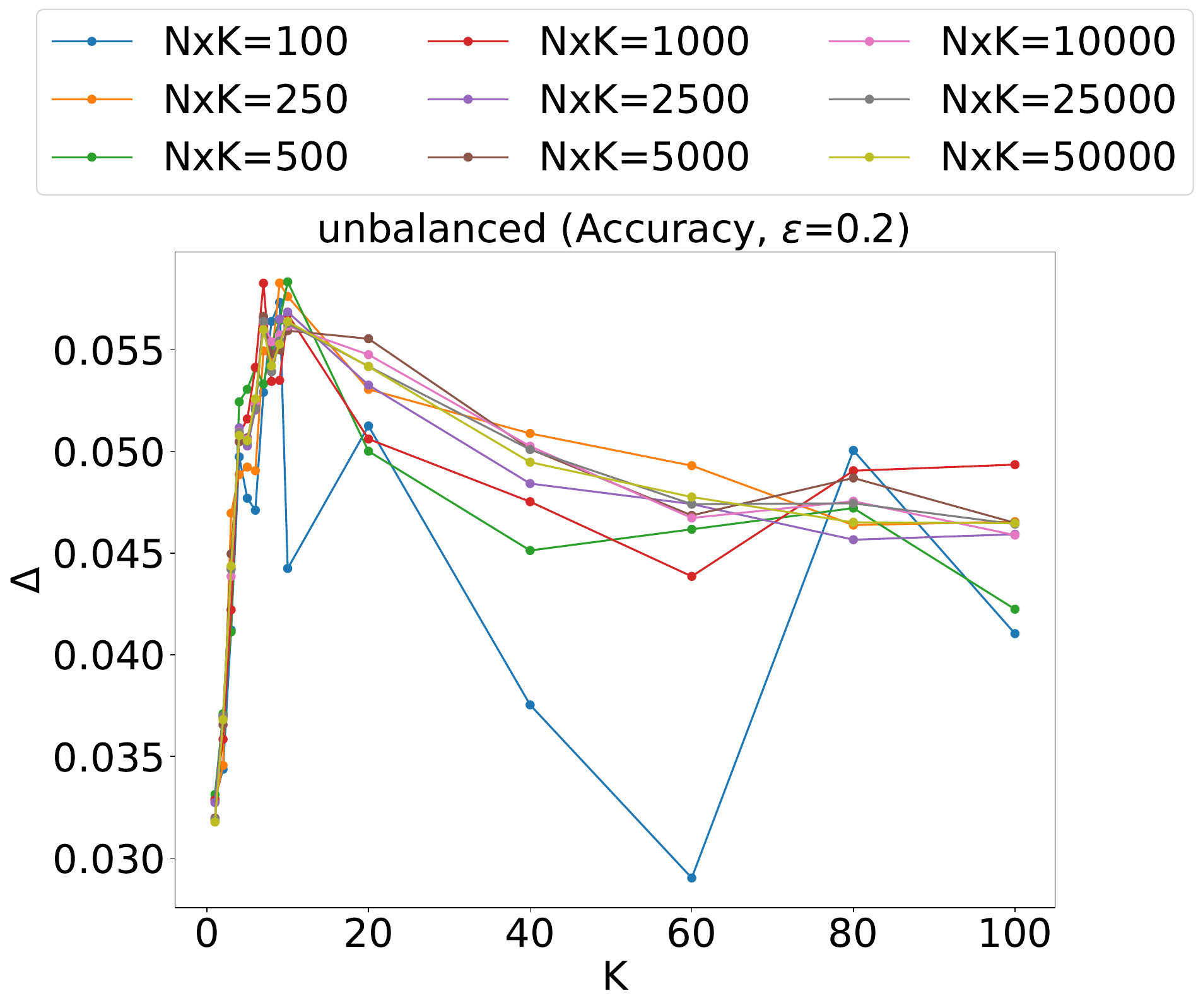}
    \caption{$\epsilon = 0.2$}
    \label{fig:gamma_delta_accuracy_cat3_e02}
  \end{subfigure} \hfill
  \begin{subfigure}[b]{0.24\linewidth}
    \centering
    \includegraphics[width=\linewidth]{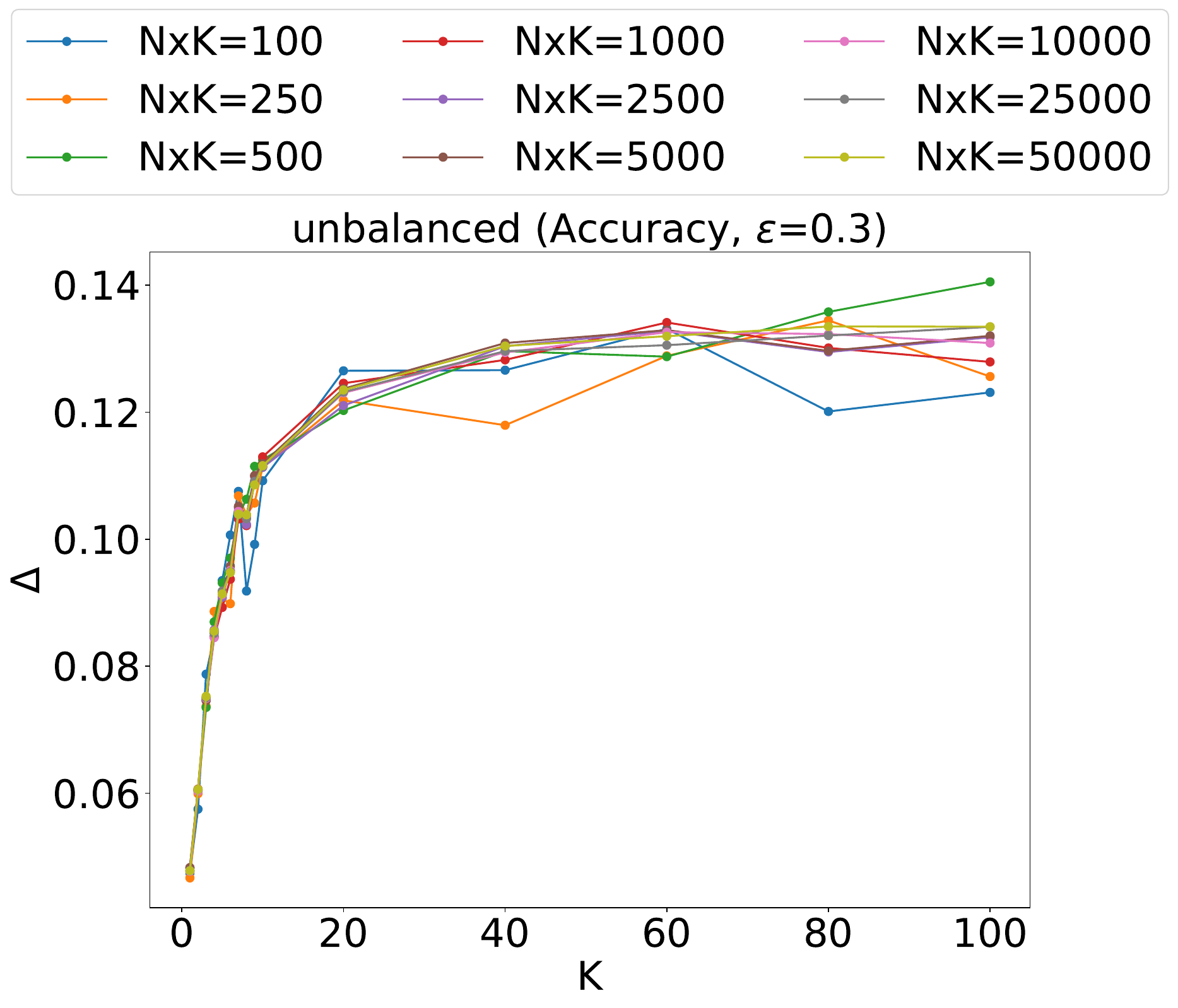}
    \caption{$\epsilon = 0.3$}
    \label{fig:gamma_delta_accuracy_cat3_e03}
  \end{subfigure} \hfill
  \begin{subfigure}[b]{0.24\linewidth}
    \centering
    \includegraphics[width=\linewidth]{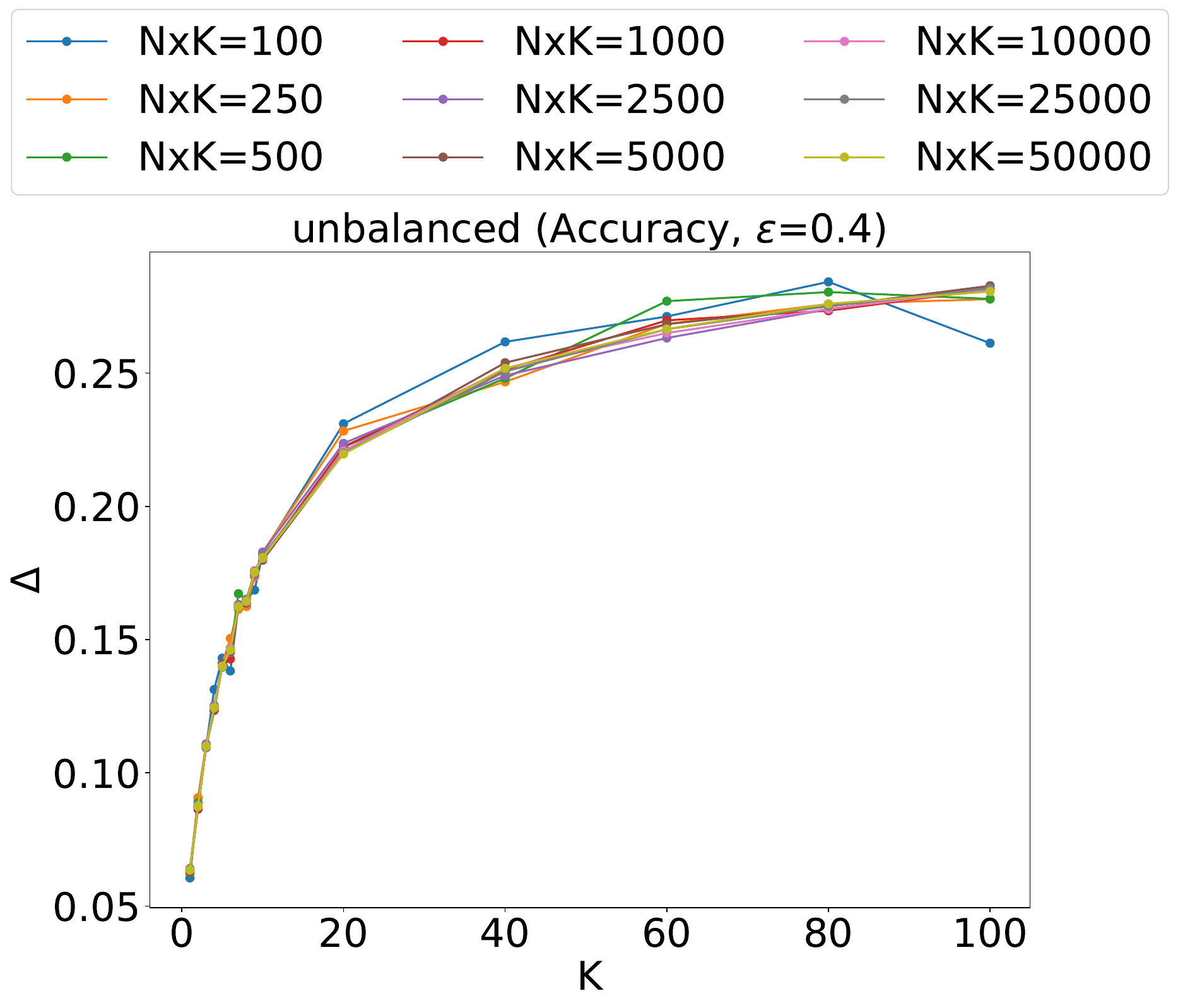}
    \caption{$\epsilon = 0.4$}
    \label{fig:gamma_delta_accuracy_cat3_e04}
  \end{subfigure}
  \caption{Effect sizes ($\Delta$) for unbalanced alphas with Accuracy as the metric ($M=3$)}
  \label{fig:gamma_delta_accuracy_cat3}
\end{figure*}

\begin{figure*}
  \centering
  \begin{subfigure}[b]{0.24\linewidth}
    \centering
    \includegraphics[width=\linewidth]{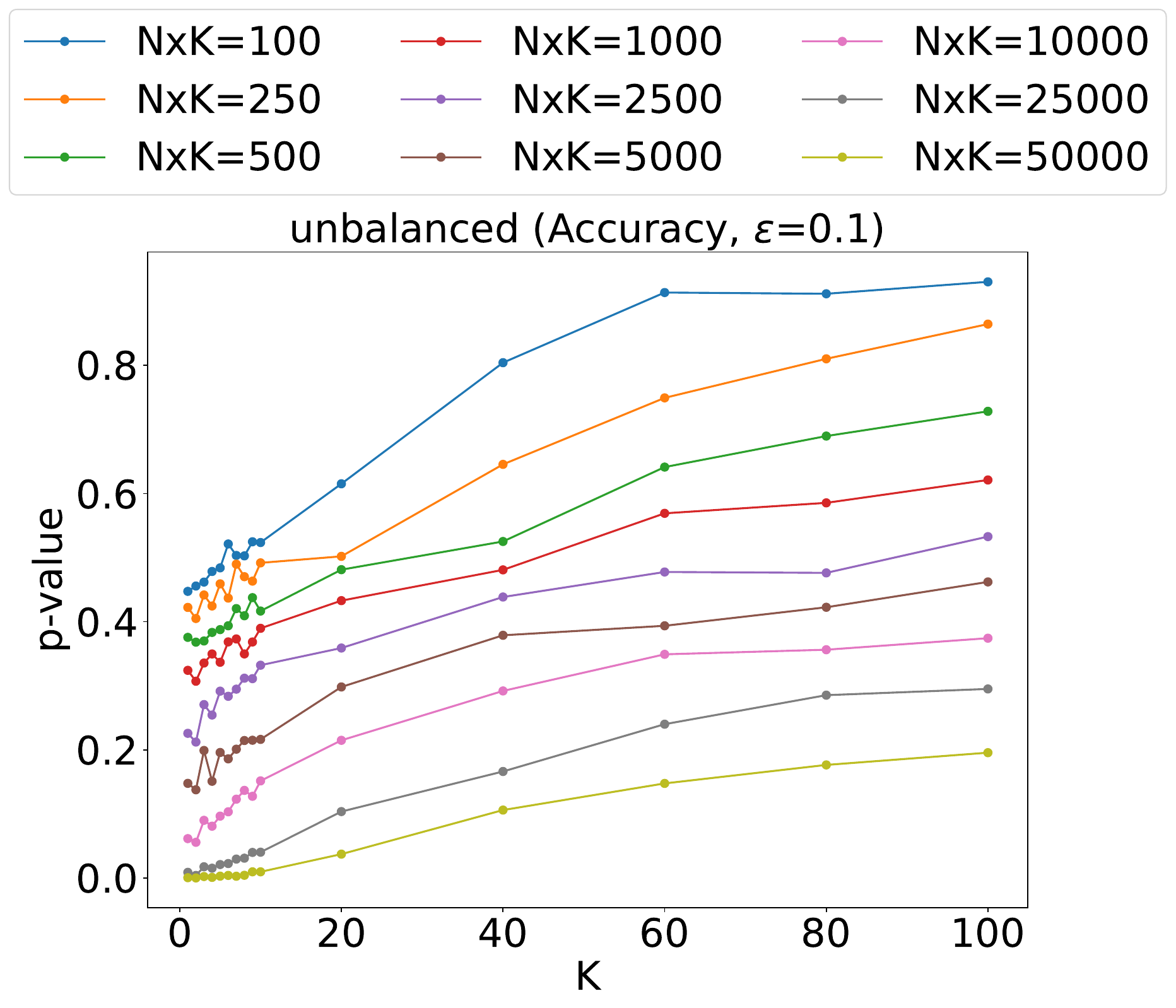}
    \caption{$\epsilon = 0.1$}
    \label{fig:gamma_accuracy_cat4_e01}
  \end{subfigure} \hfill
  \begin{subfigure}[b]{0.24\linewidth}
    \centering
    \includegraphics[width=\linewidth]{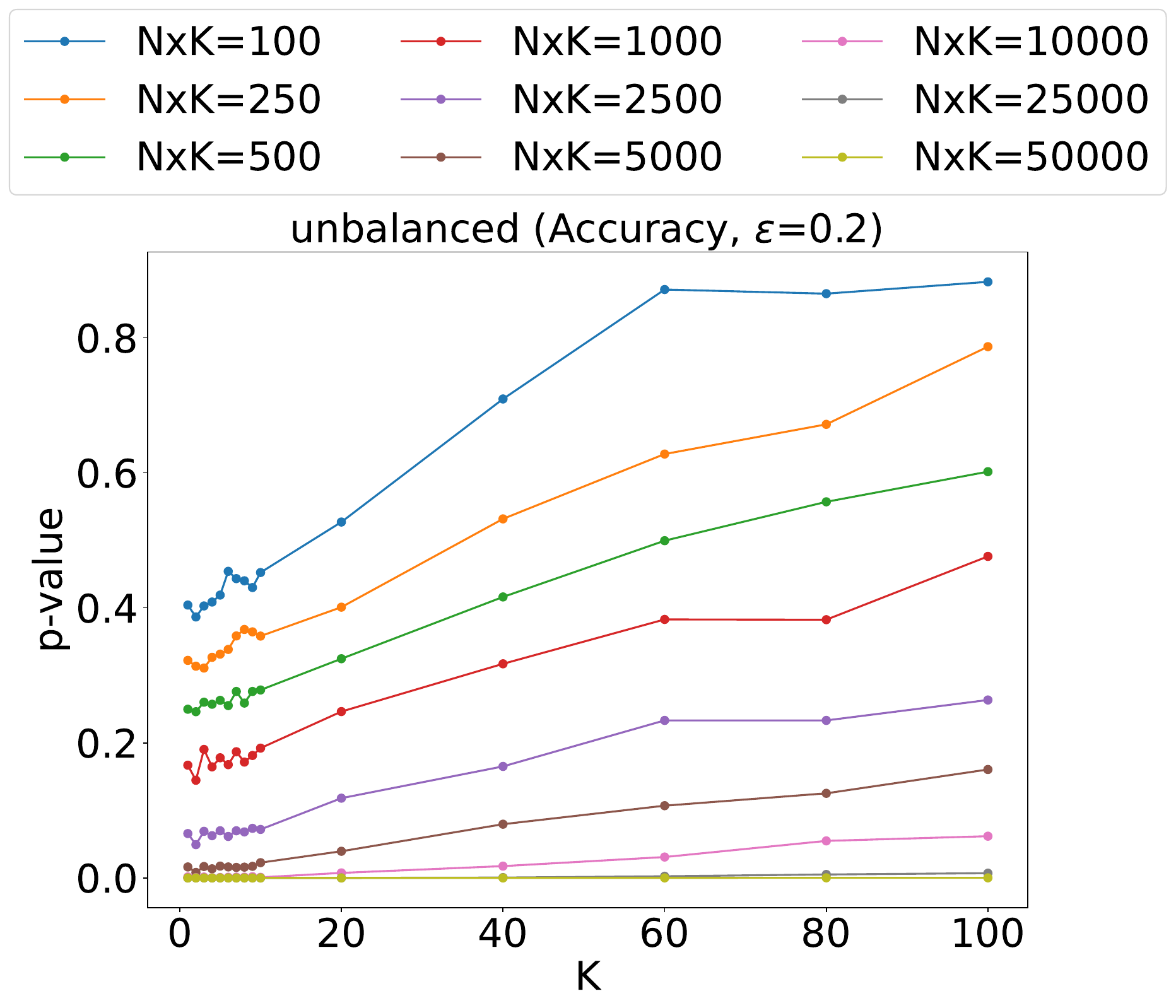}
    \caption{$\epsilon = 0.2$}
    \label{fig:gamma_accuracy_cat4_e02}
  \end{subfigure} \hfill
  \begin{subfigure}[b]{0.24\linewidth}
    \centering
    \includegraphics[width=\linewidth]{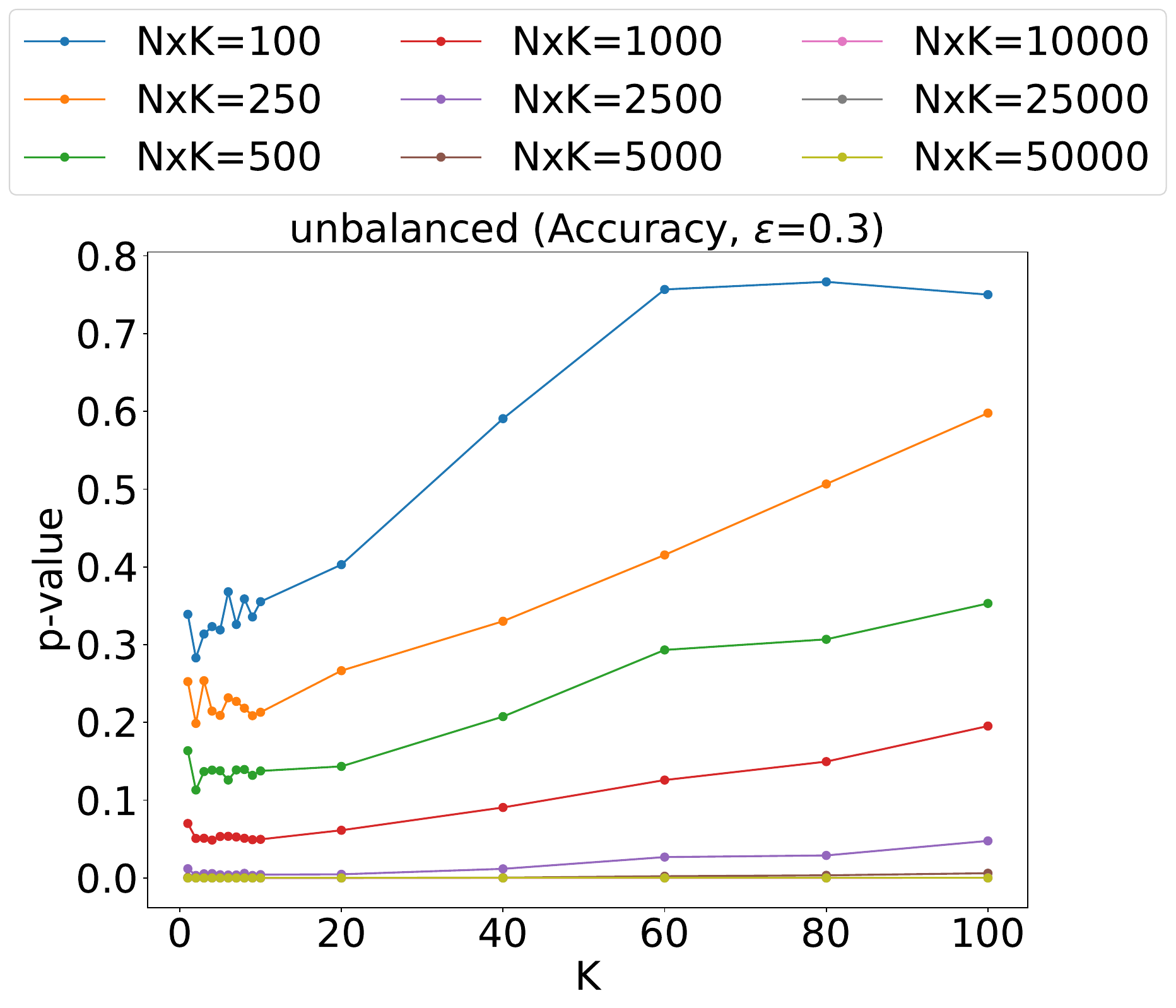}
    \caption{$\epsilon = 0.3$}
    \label{fig:gamma_accuracy_cat4_e03}
  \end{subfigure} \hfill
  \begin{subfigure}[b]{0.24\linewidth}
    \centering
    \includegraphics[width=\linewidth]{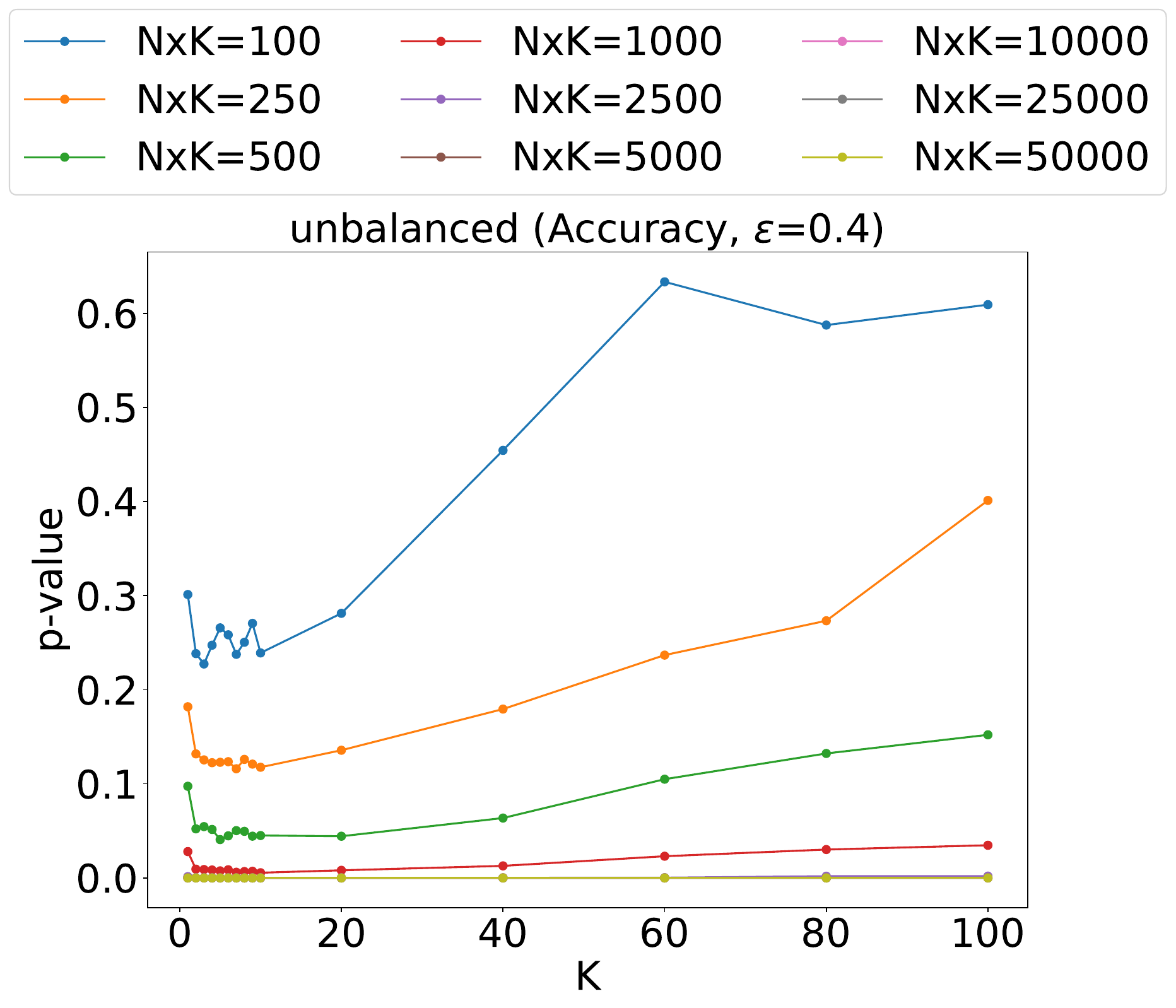}
    \caption{$\epsilon = 0.4$}
    \label{fig:gamma_accuracy_cat4_e04}
  \end{subfigure}
  \caption{P-value plots for unbalanced alphas with Accuracy as the metric ($M=4$)}
  \label{fig:gamma_accuracy_cat4}
\end{figure*}

\begin{figure*}
  \centering
  \begin{subfigure}[b]{0.24\linewidth}
    \centering
    \includegraphics[width=\linewidth]{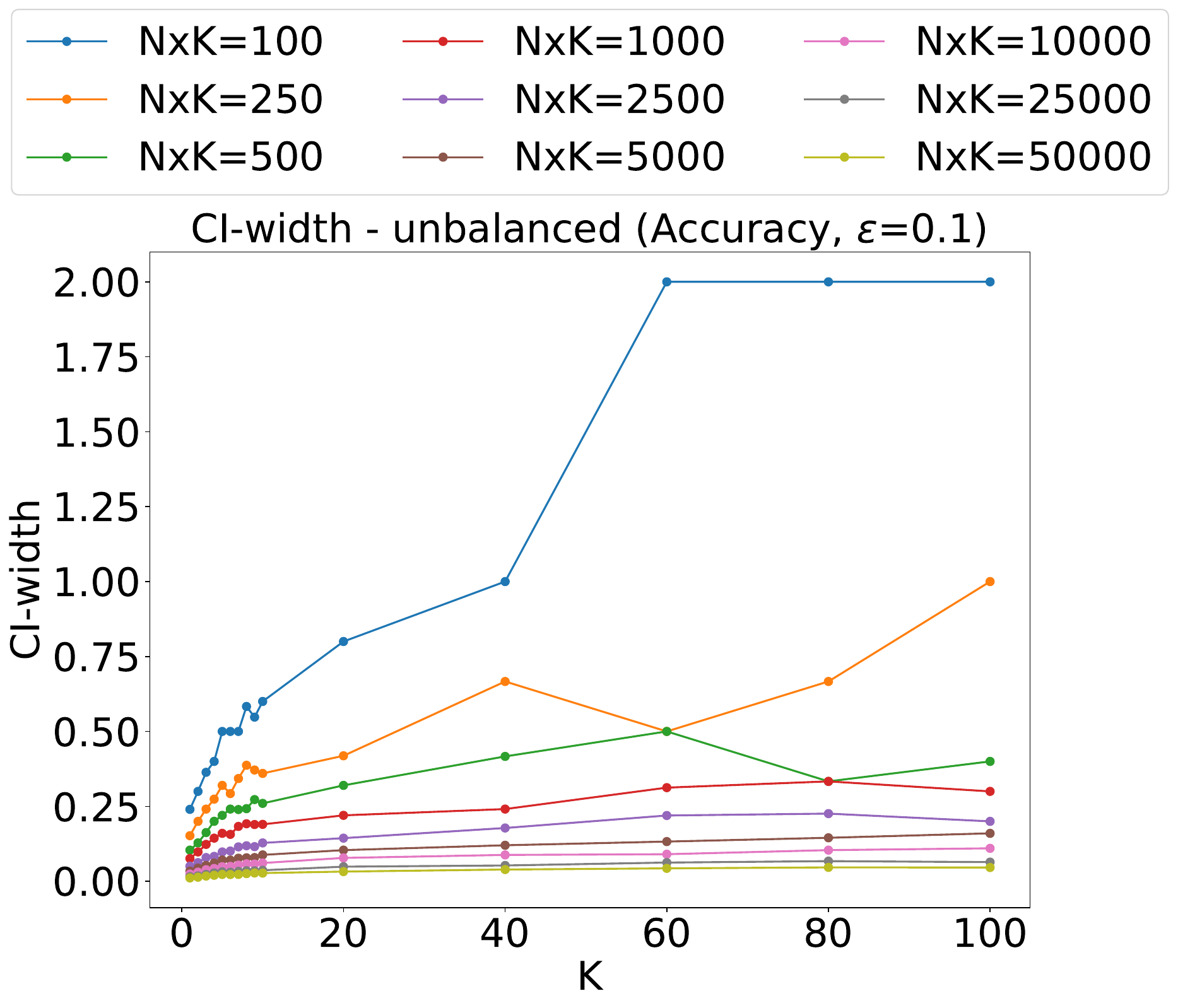}
    \caption{$\epsilon = 0.1$}
    \label{fig:gamma_ci_accuracy_cat4_e01}
  \end{subfigure} \hfill
  \begin{subfigure}[b]{0.24\linewidth}
    \centering
    \includegraphics[width=\linewidth]{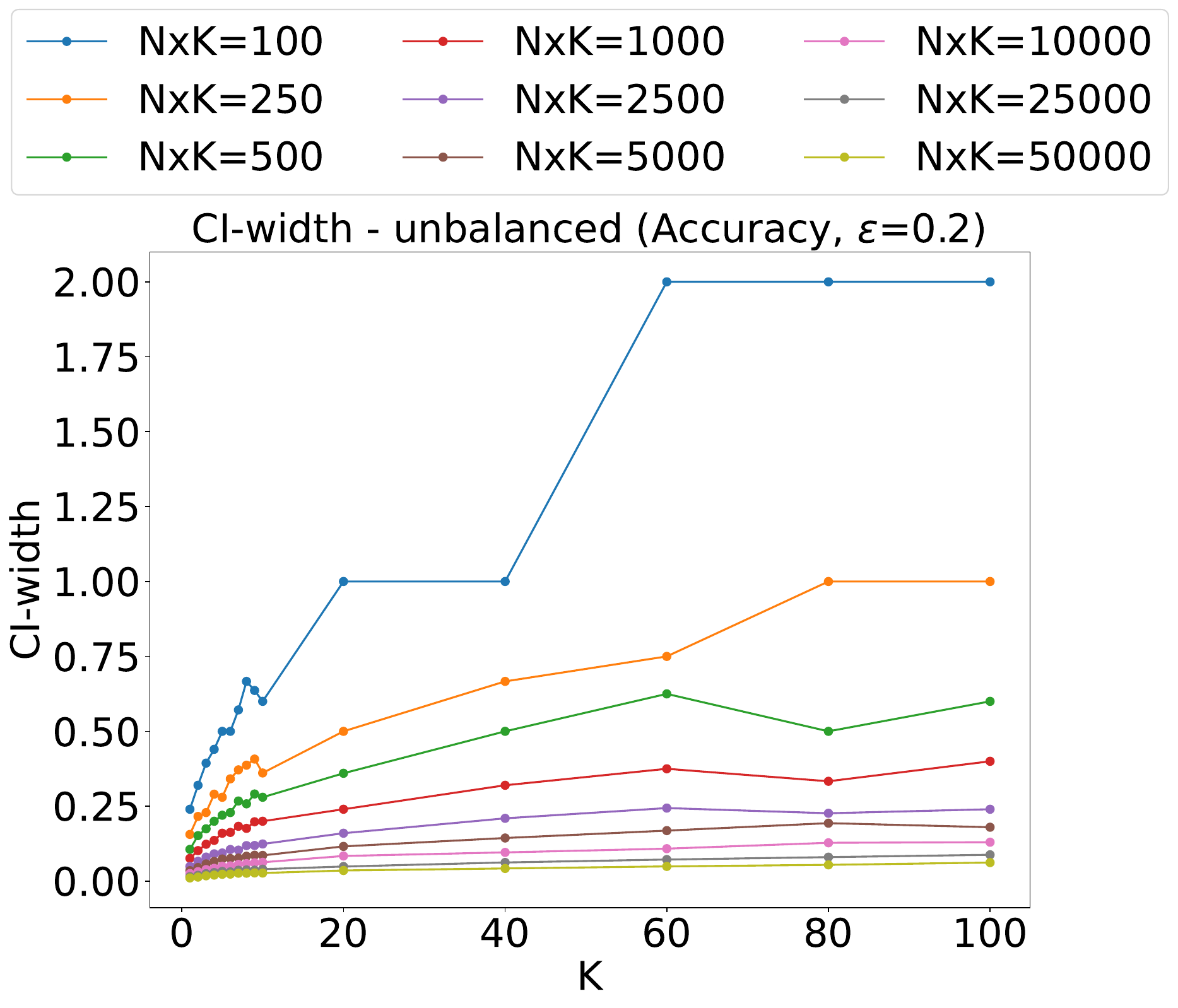}
    \caption{$\epsilon = 0.2$}
    \label{fig:gamma_ci_accuracy_cat4_e02}
  \end{subfigure} \hfill
  \begin{subfigure}[b]{0.24\linewidth}
    \centering
    \includegraphics[width=\linewidth]{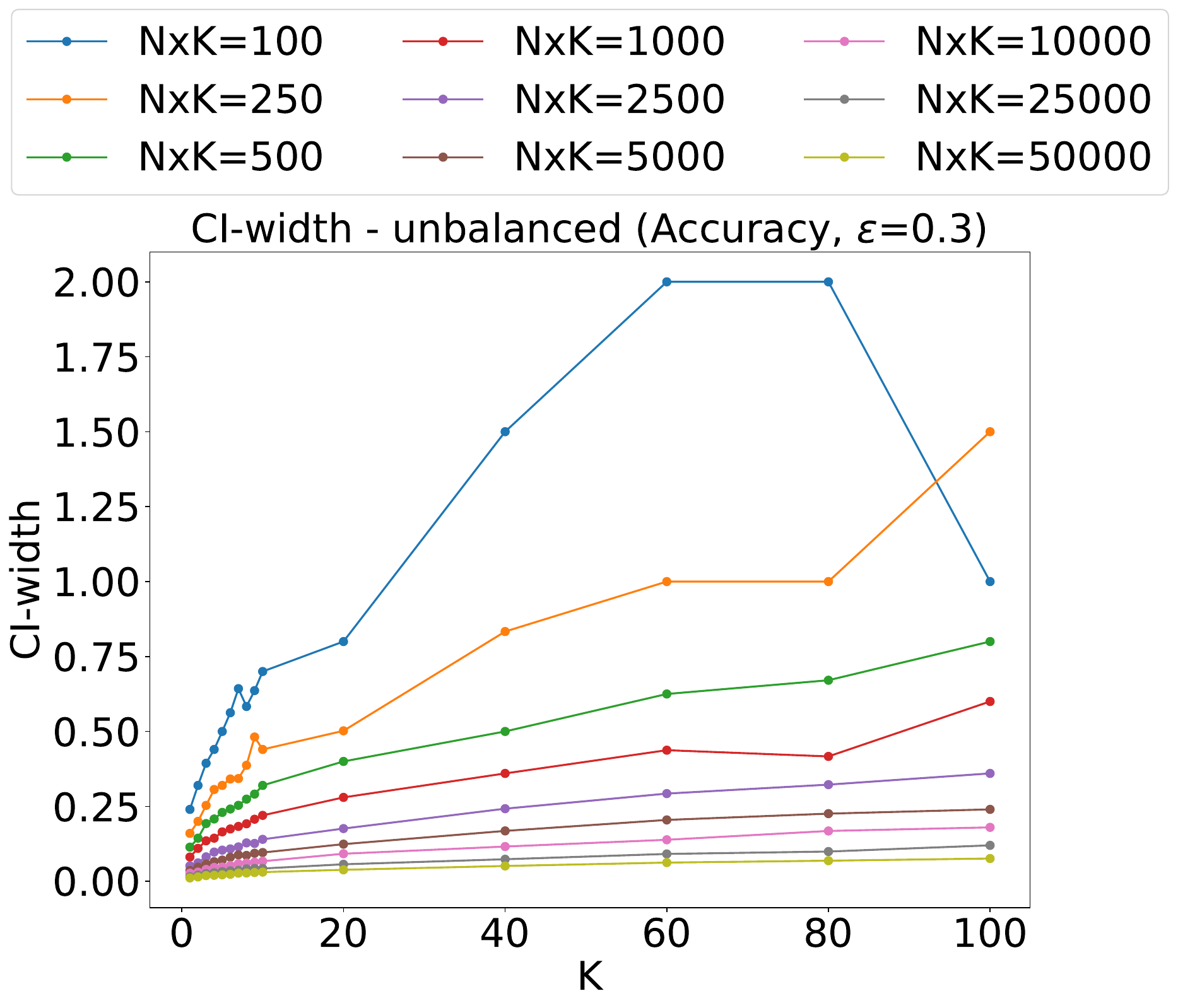}
    \caption{$\epsilon = 0.3$}
    \label{fig:gamma_ci_accuracy_cat4_e03}
  \end{subfigure} \hfill
  \begin{subfigure}[b]{0.24\linewidth}
    \centering
    \includegraphics[width=\linewidth]{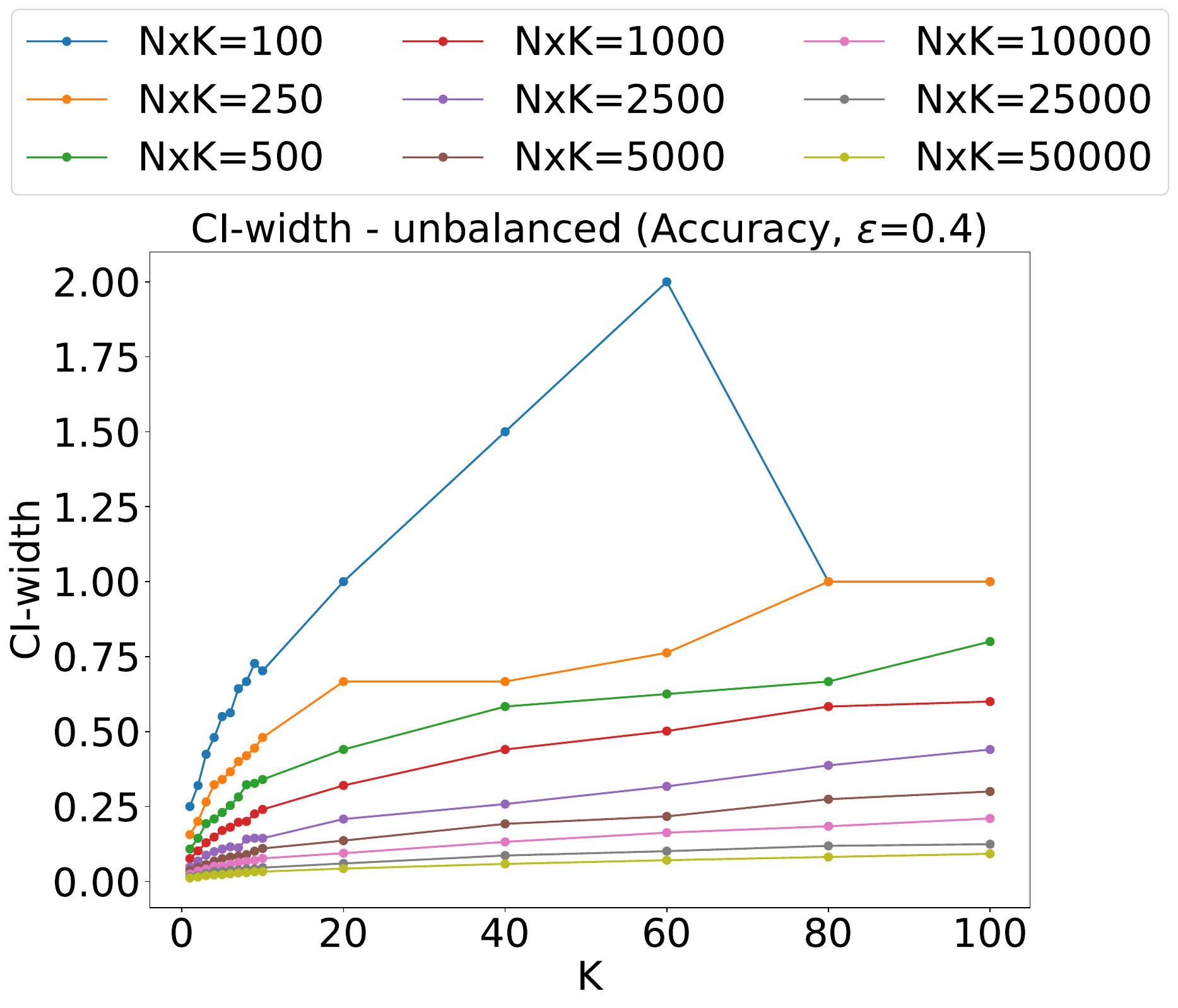}
    \caption{$\epsilon = 0.4$}
    \label{fig:gamma_ci_accuracy_cat4_e04}
  \end{subfigure}
  \caption{CI-width plots for unbalanced alphas with Accuracy as the metric ($M=4$)}
  \label{fig:gamma_ci_accuracy_cat4}
\end{figure*}

\begin{figure*}
  \centering
  \begin{subfigure}[b]{0.24\linewidth}
    \centering
    \includegraphics[width=\linewidth]{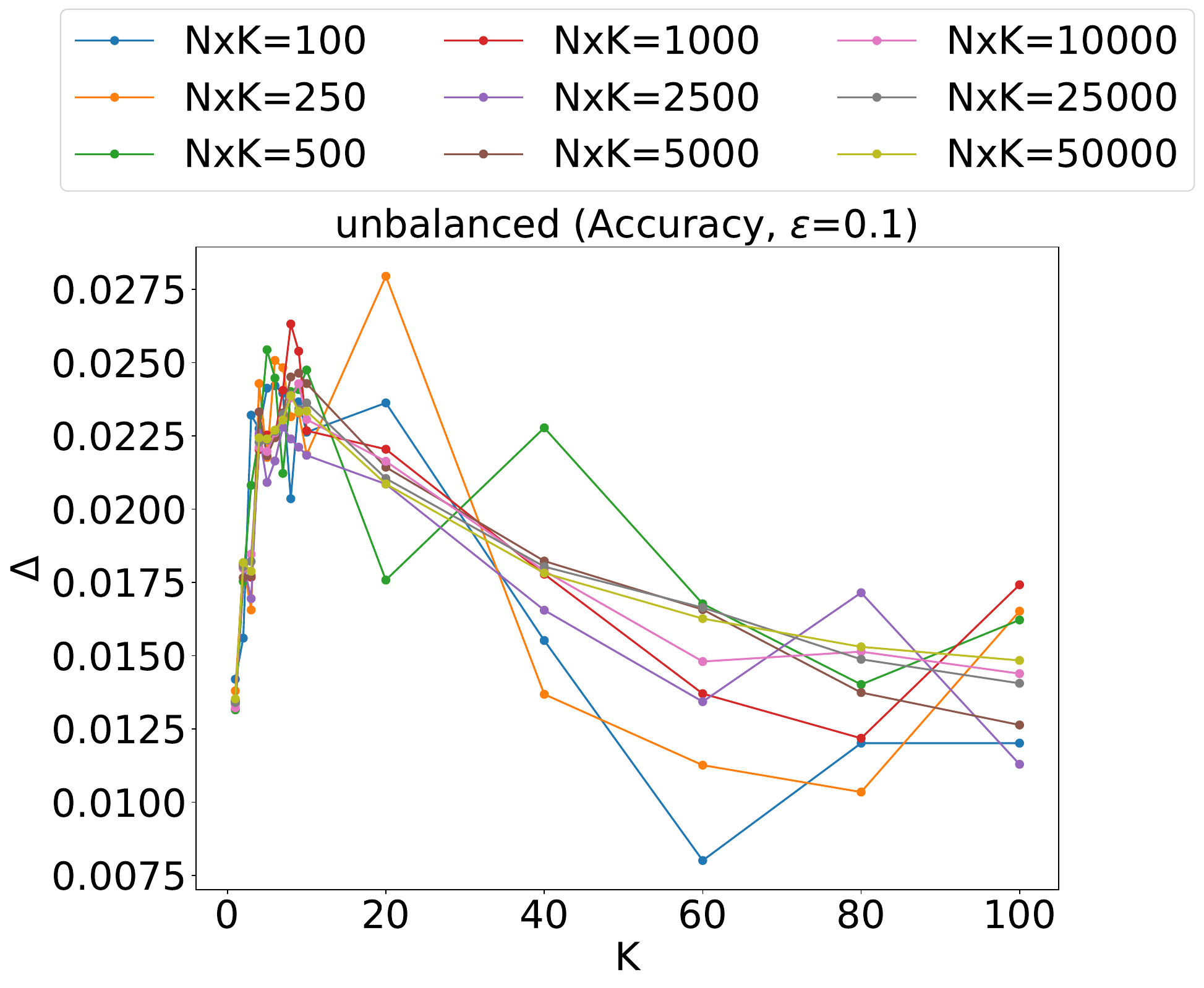}
    \caption{$\epsilon = 0.1$}
    \label{fig:gamma_delta_accuracy_cat4_e01}
  \end{subfigure} \hfill
  \begin{subfigure}[b]{0.24\linewidth}
    \centering
    \includegraphics[width=\linewidth]{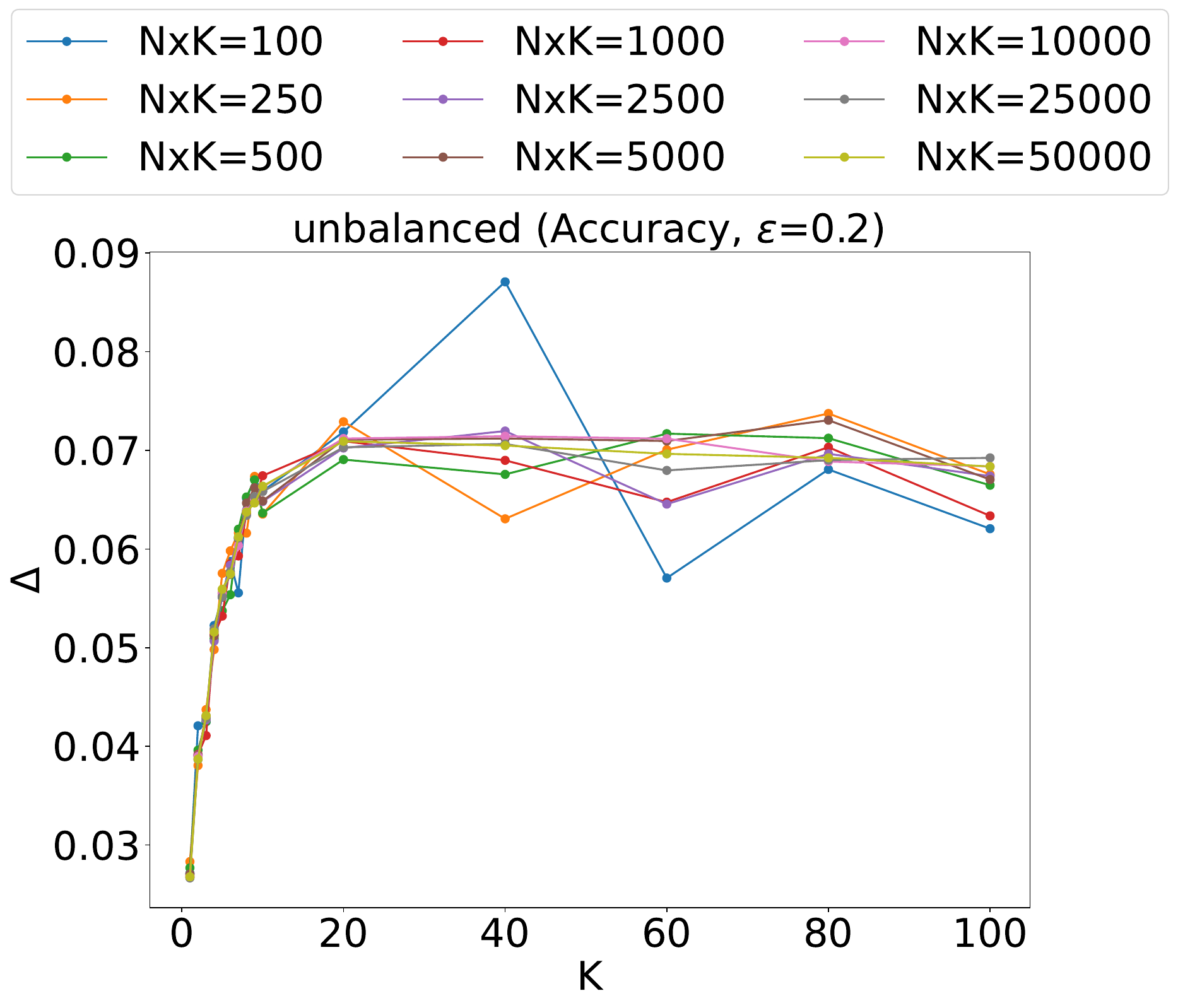}
    \caption{$\epsilon = 0.2$}
    \label{fig:gamma_delta_accuracy_cat4_e02}
  \end{subfigure} \hfill
  \begin{subfigure}[b]{0.24\linewidth}
    \centering
    \includegraphics[width=\linewidth]{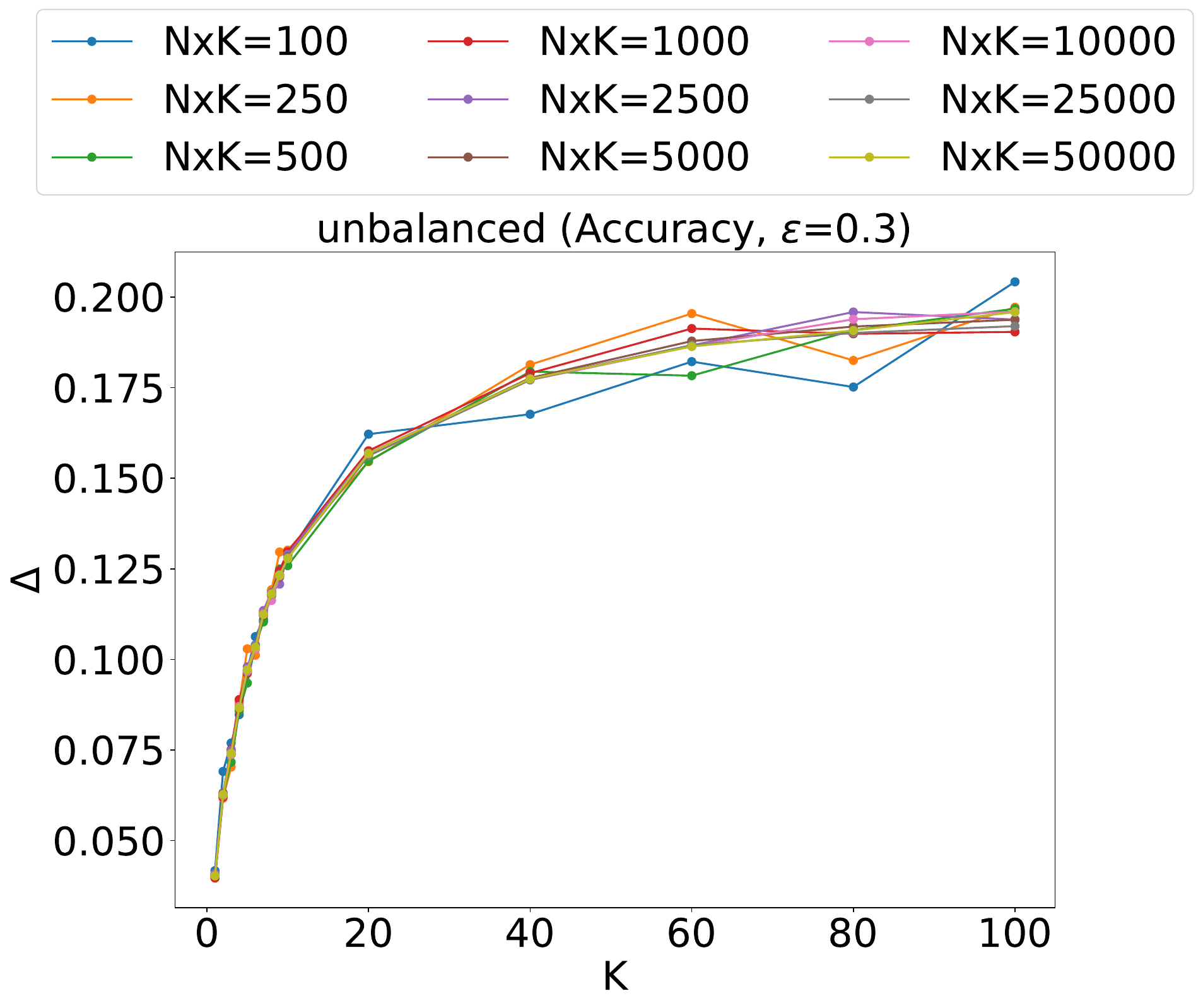}
    \caption{$\epsilon = 0.3$}
    \label{fig:gamma_delta_accuracy_cat4_e03}
  \end{subfigure} \hfill
  \begin{subfigure}[b]{0.24\linewidth}
    \centering
    \includegraphics[width=\linewidth]{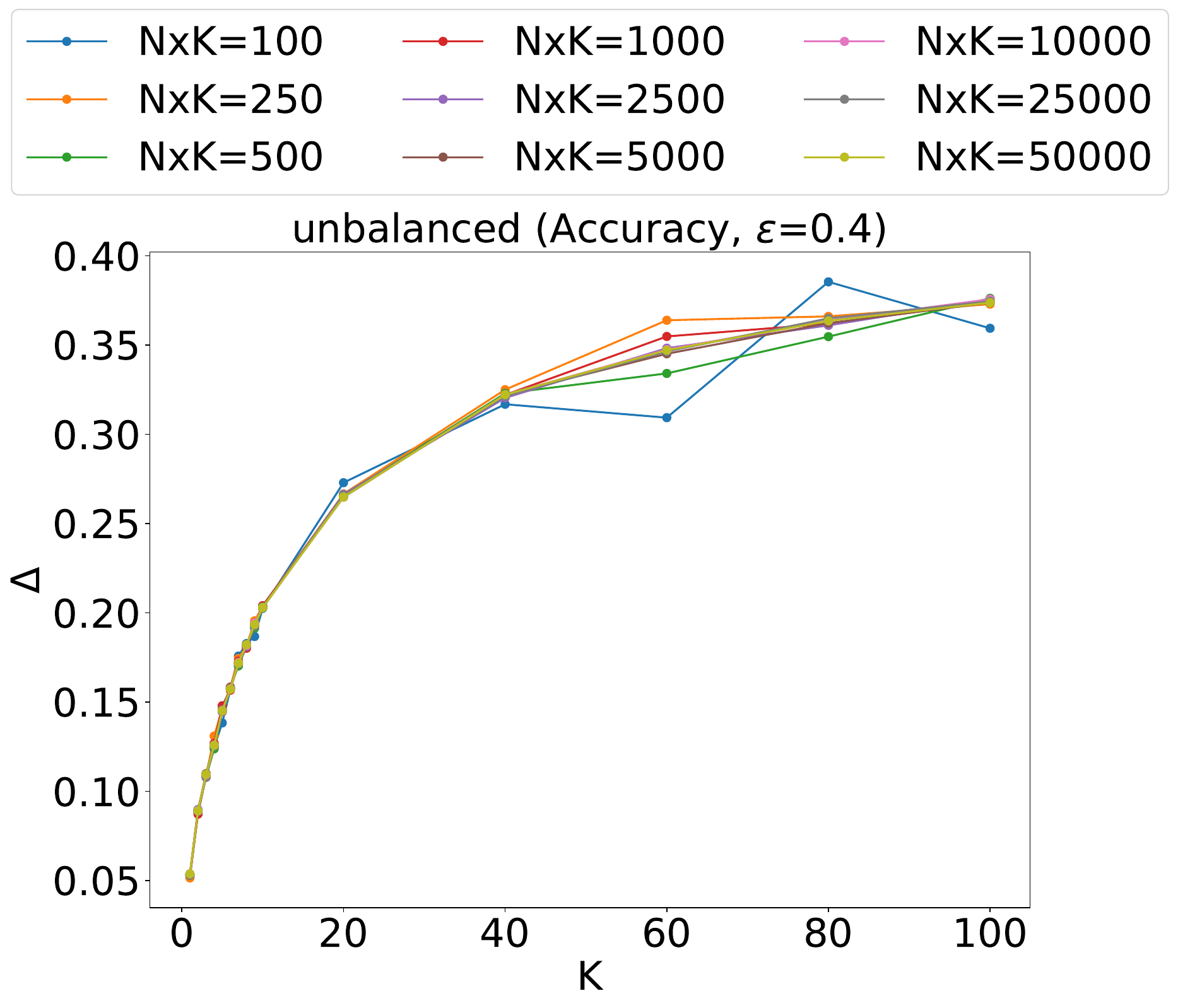}
    \caption{$\epsilon = 0.4$}
    \label{fig:gamma_delta_accuracy_cat4_e04}
  \end{subfigure}
  \caption{Effect sizes ($\Delta$) for unbalanced alphas with Accuracy as the metric ($M=4$)}
  \label{fig:gamma_delta_accuracy_cat4}
\end{figure*}

\begin{figure*}
  \centering
  \begin{subfigure}[b]{0.24\linewidth}
    \centering
    \includegraphics[width=\linewidth]{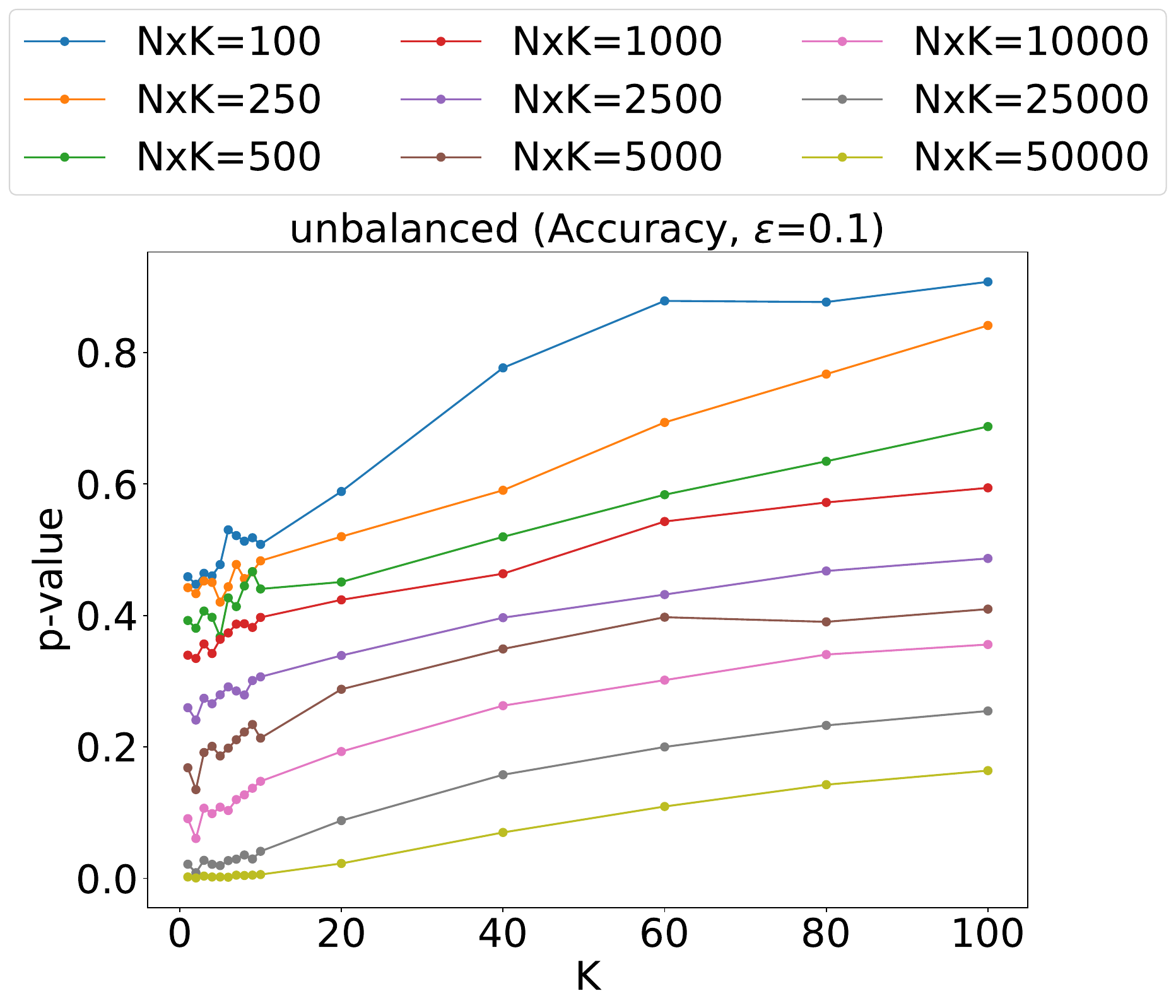}
    \caption{$\epsilon = 0.1$}
    \label{fig:gamma_accuracy_cat5_e01}
  \end{subfigure} \hfill
  \begin{subfigure}[b]{0.24\linewidth}
    \centering
    \includegraphics[width=\linewidth]{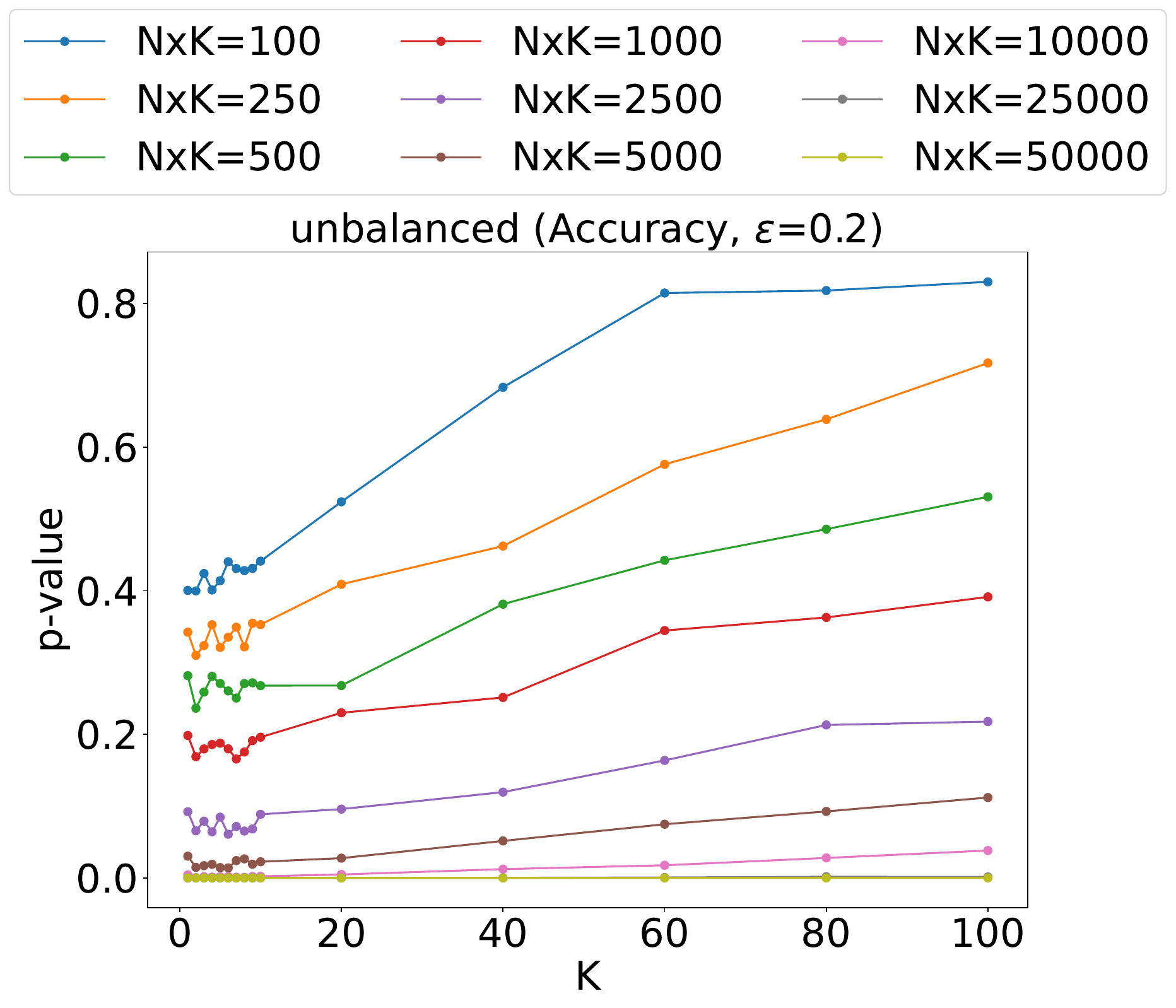}
    \caption{$\epsilon = 0.2$}
    \label{fig:gamma_accuracy_cat5_e02}
  \end{subfigure} \hfill
  \begin{subfigure}[b]{0.24\linewidth}
    \centering
    \includegraphics[width=\linewidth]{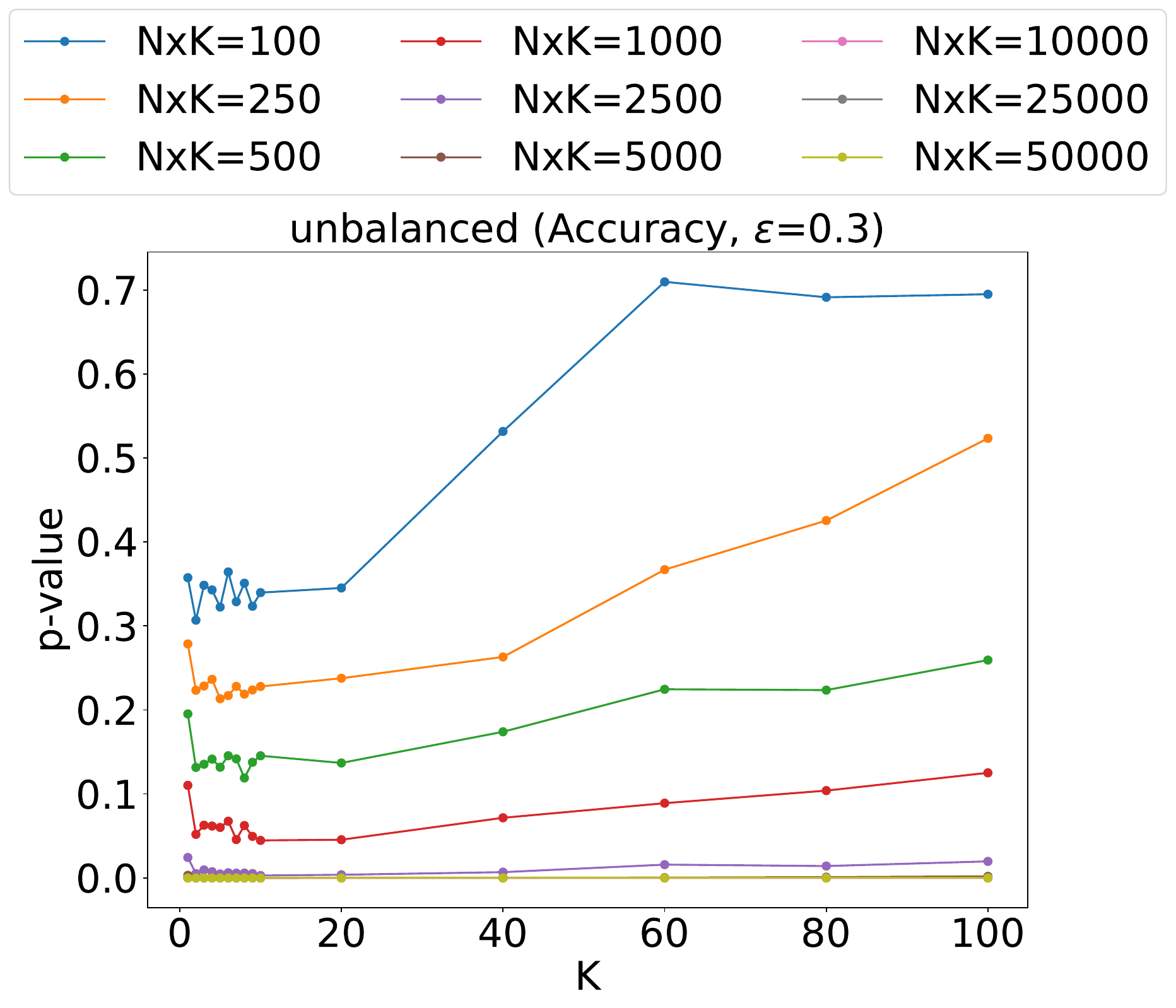}
    \caption{$\epsilon = 0.3$}
    \label{fig:gamma_accuracy_cat5_e03}
  \end{subfigure} \hfill
  \begin{subfigure}[b]{0.24\linewidth}
    \centering
    \includegraphics[width=\linewidth]{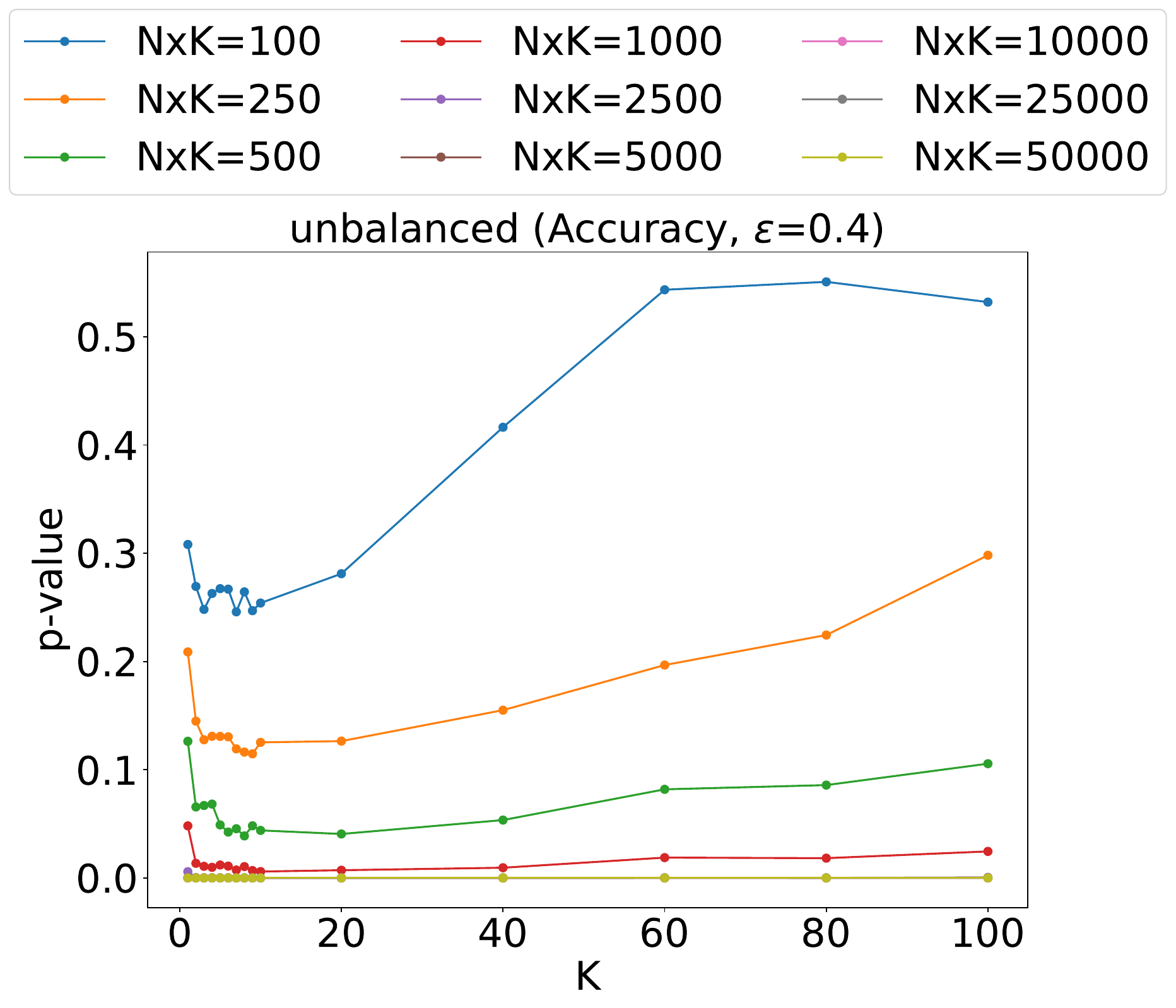}
    \caption{$\epsilon = 0.4$}
    \label{fig:gamma_accuracy_cat5_e04}
  \end{subfigure}
  \caption{P-value plots for unbalanced alphas with Accuracy as the metric ($M=5$)}
  \label{fig:gamma_accuracy_cat5}
\end{figure*}

\begin{figure*}
  \centering
  \begin{subfigure}[b]{0.24\linewidth}
    \centering
    \includegraphics[width=\linewidth]{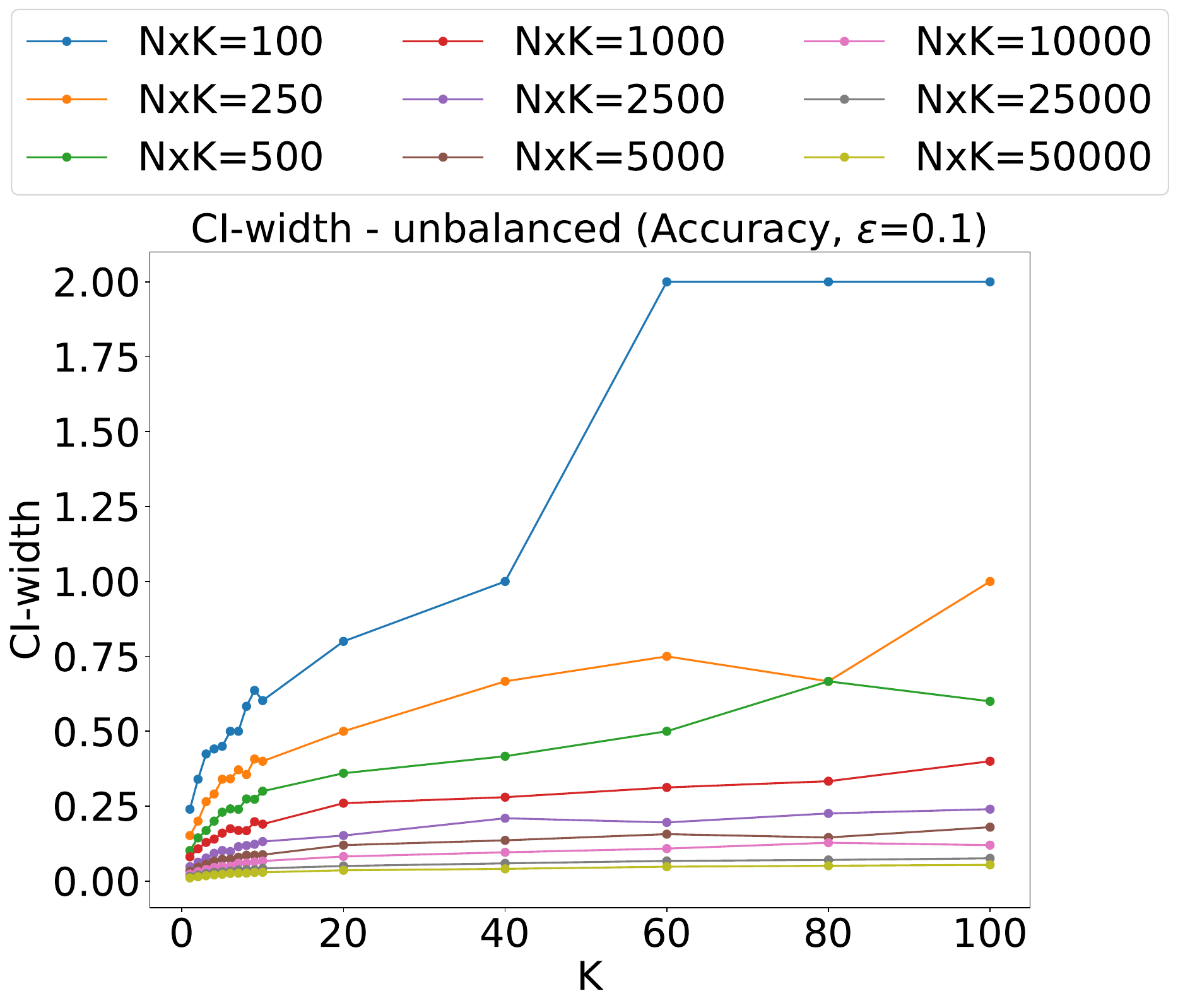}
    \caption{$\epsilon = 0.1$}
    \label{fig:gamma_ci_accuracy_cat5_e01}
  \end{subfigure} \hfill
  \begin{subfigure}[b]{0.24\linewidth}
    \centering
    \includegraphics[width=\linewidth]{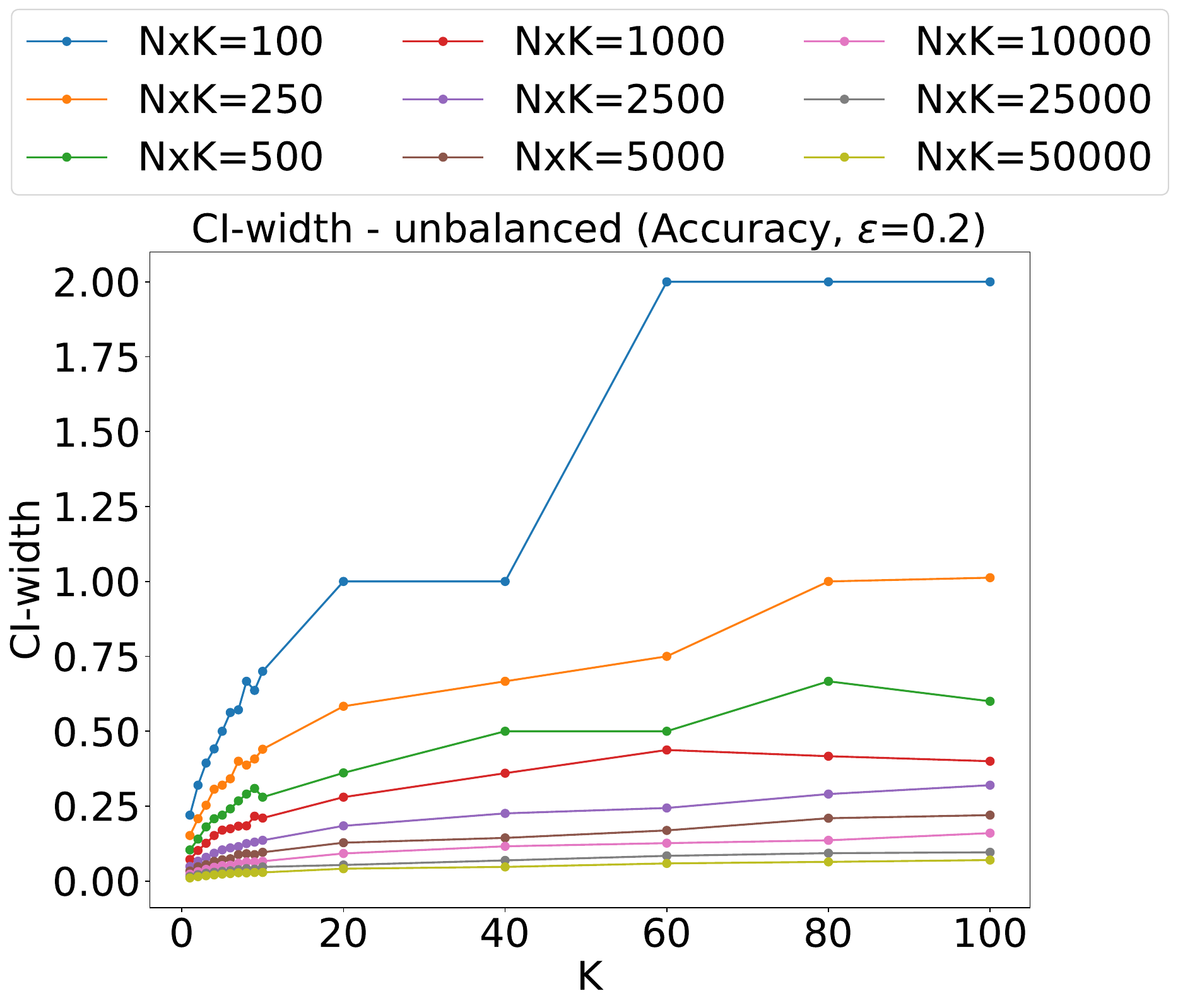}
    \caption{$\epsilon = 0.2$}
    \label{fig:gamma_ci_accuracy_cat5_e02}
  \end{subfigure} \hfill
  \begin{subfigure}[b]{0.24\linewidth}
    \centering
    \includegraphics[width=\linewidth]{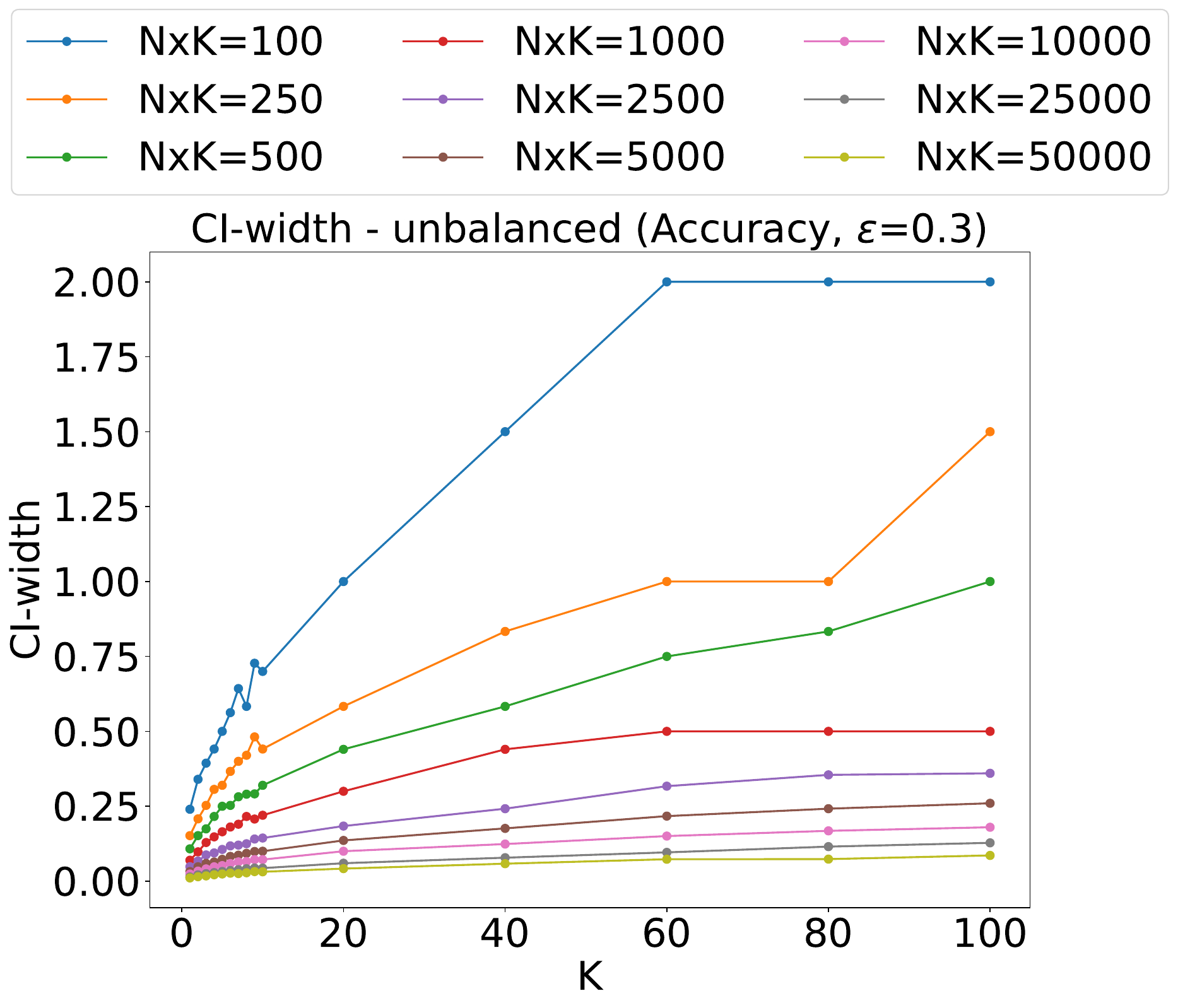}
    \caption{$\epsilon = 0.3$}
    \label{fig:gamma_ci_accuracy_cat5_e03}
  \end{subfigure} \hfill
  \begin{subfigure}[b]{0.24\linewidth}
    \centering
    \includegraphics[width=\linewidth]{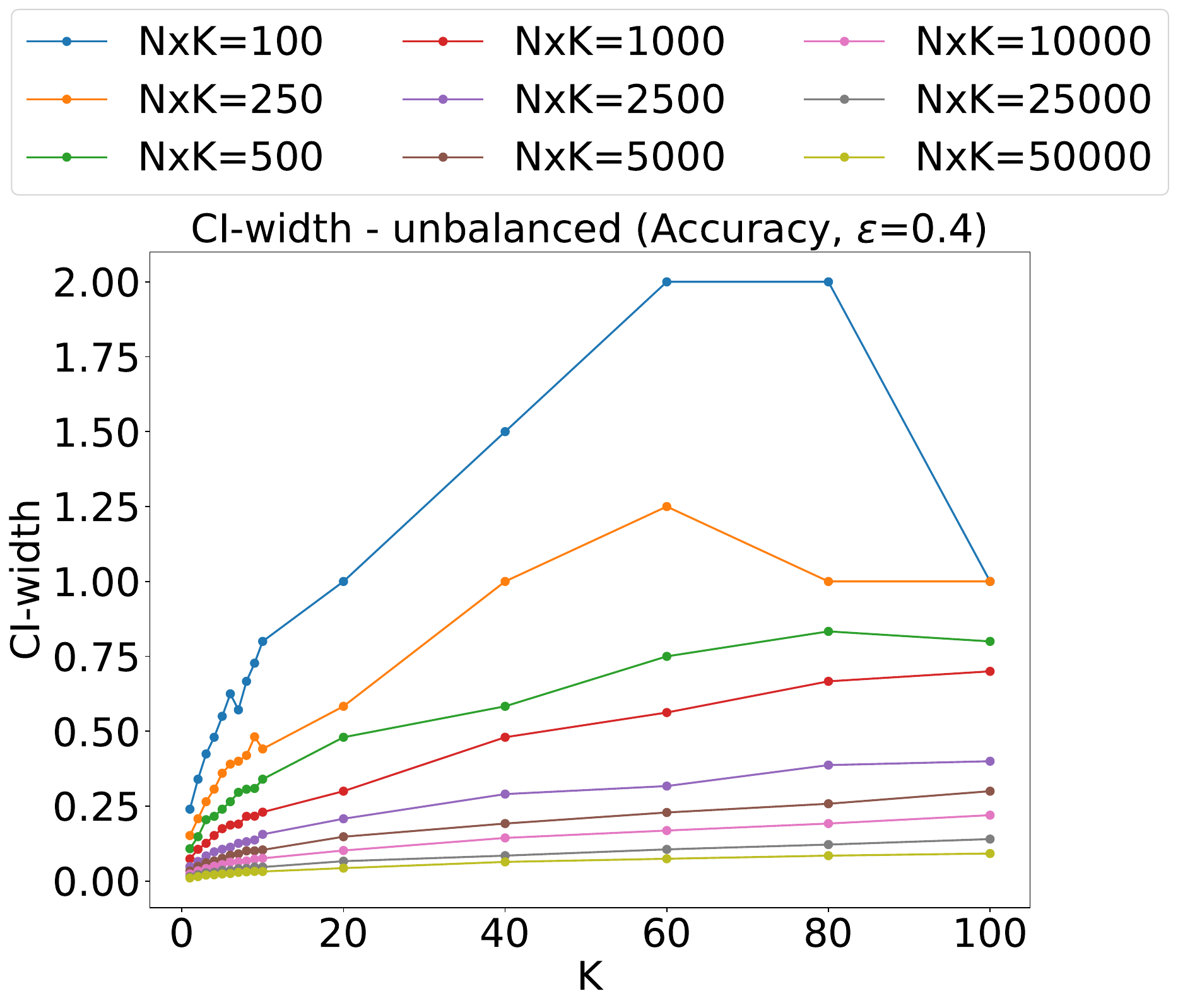}
    \caption{$\epsilon = 0.4$}
    \label{fig:gamma_ci_accuracy_cat5_e04}
  \end{subfigure}
  \caption{CI-width plots for unbalanced alphas with Accuracy as the metric ($M=5$)}
  \label{fig:gamma_ci_accuracy_cat5}
\end{figure*}

\begin{figure*}
  \centering
  \begin{subfigure}[b]{0.24\linewidth}
    \centering
    \includegraphics[width=\linewidth]{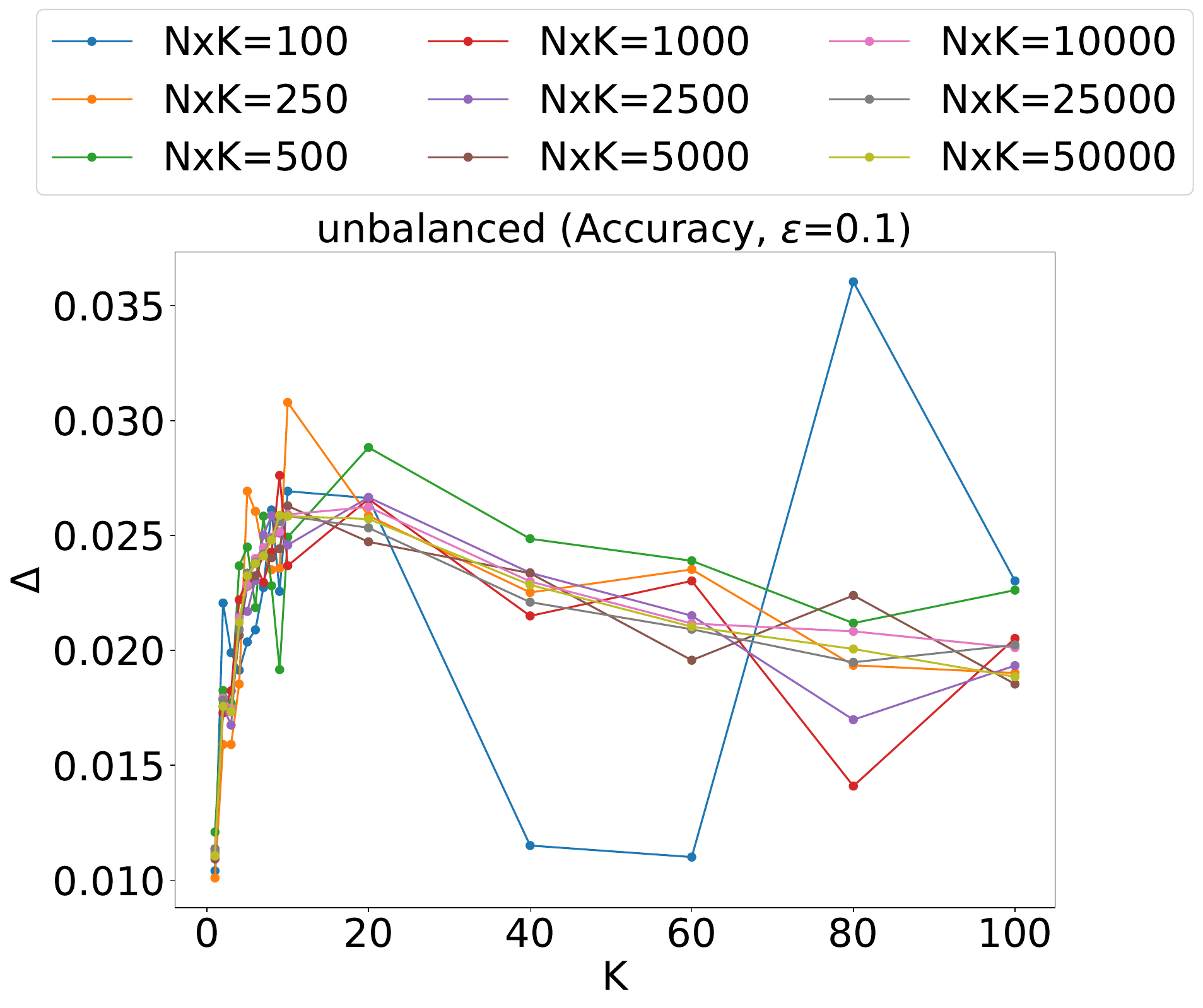}
    \caption{$\epsilon = 0.1$}
    \label{fig:gamma_delta_accuracy_cat5_e01}
  \end{subfigure} \hfill
  \begin{subfigure}[b]{0.24\linewidth}
    \centering
    \includegraphics[width=\linewidth]{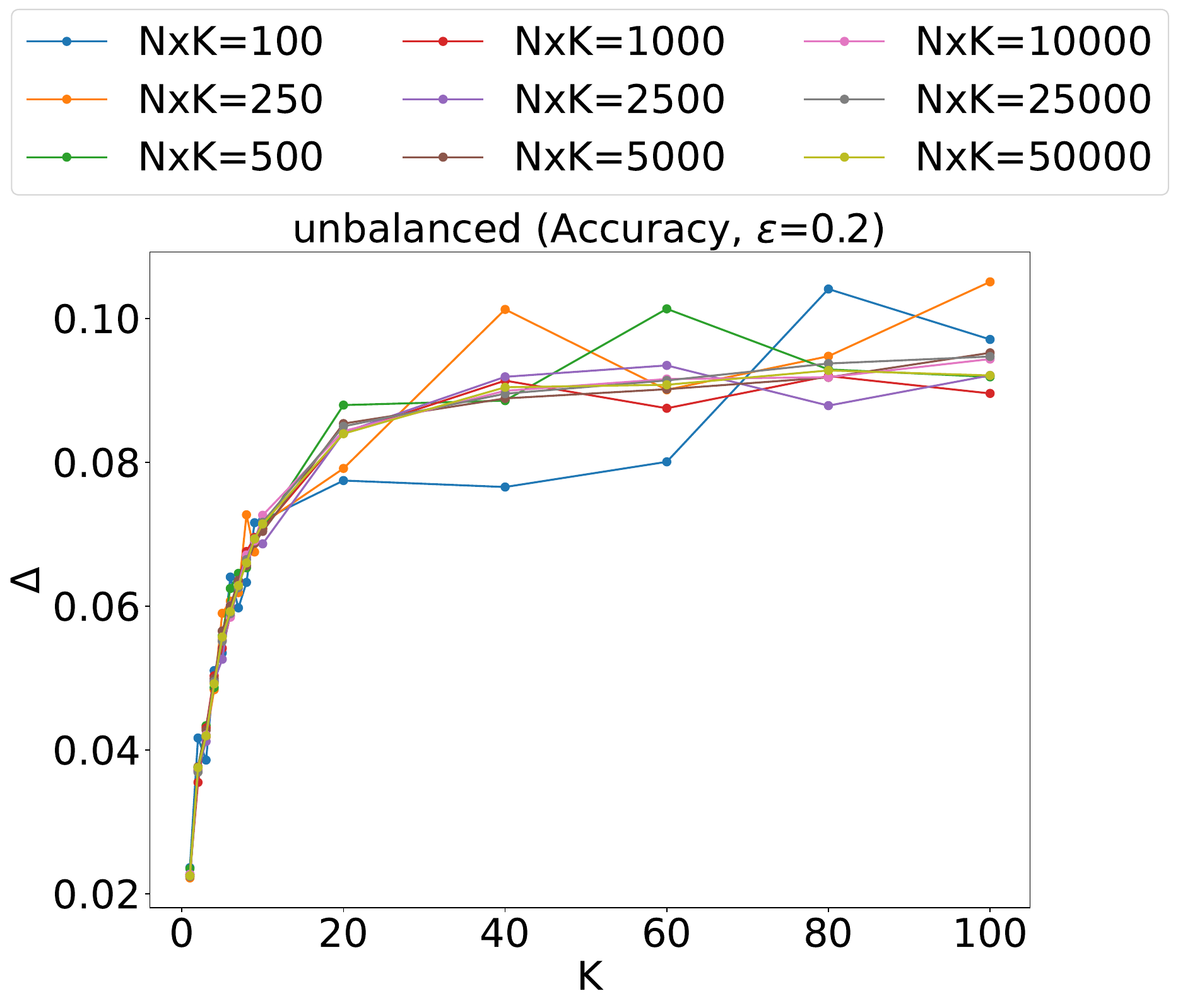}
    \caption{$\epsilon = 0.2$}
    \label{fig:gamma_delta_accuracy_cat5_e02}
  \end{subfigure} \hfill
  \begin{subfigure}[b]{0.24\linewidth}
    \centering
    \includegraphics[width=\linewidth]{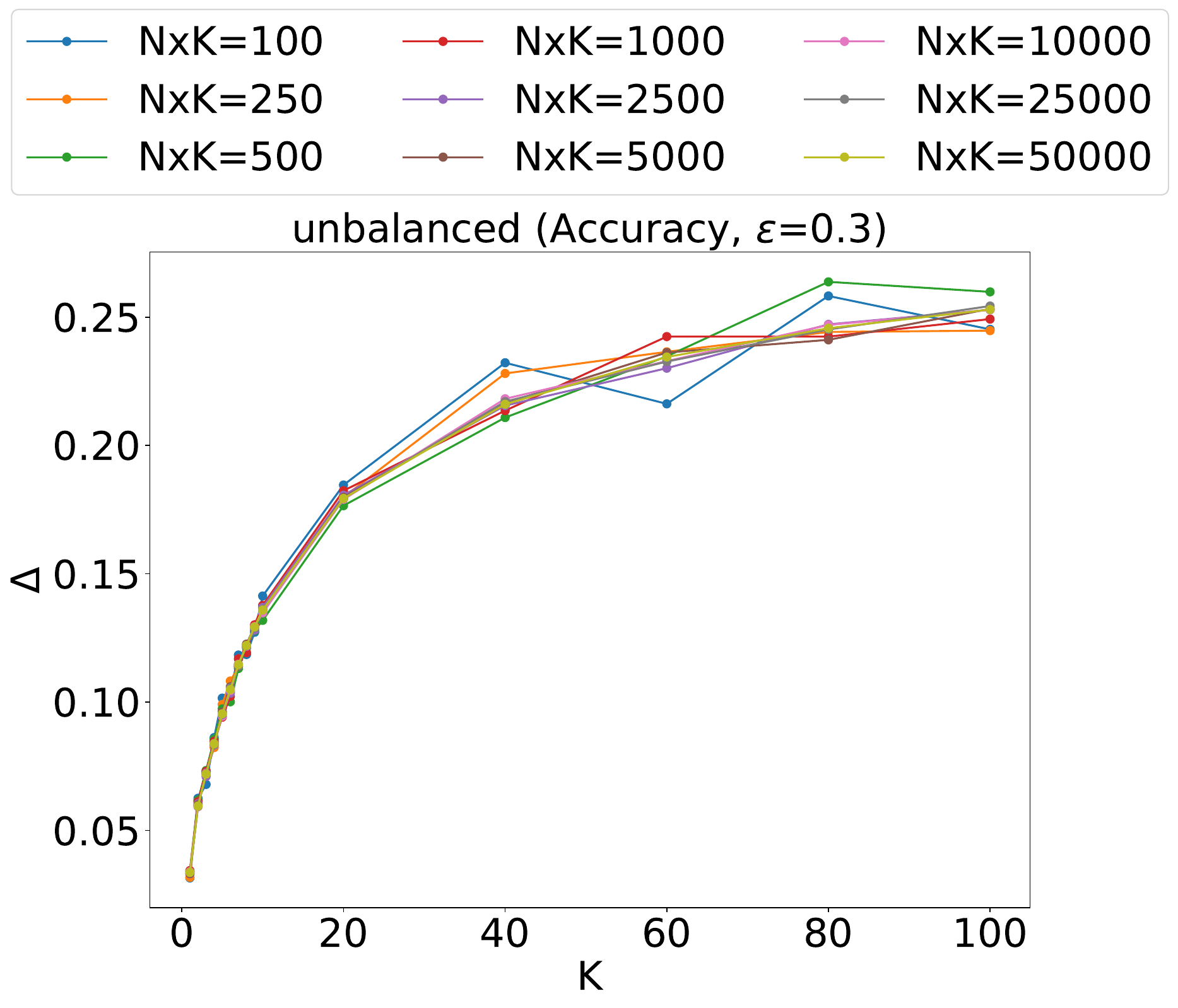}
    \caption{$\epsilon = 0.3$}
    \label{fig:gamma_delta_accuracy_cat5_e03}
  \end{subfigure} \hfill
  \begin{subfigure}[b]{0.24\linewidth}
    \centering
    \includegraphics[width=\linewidth]{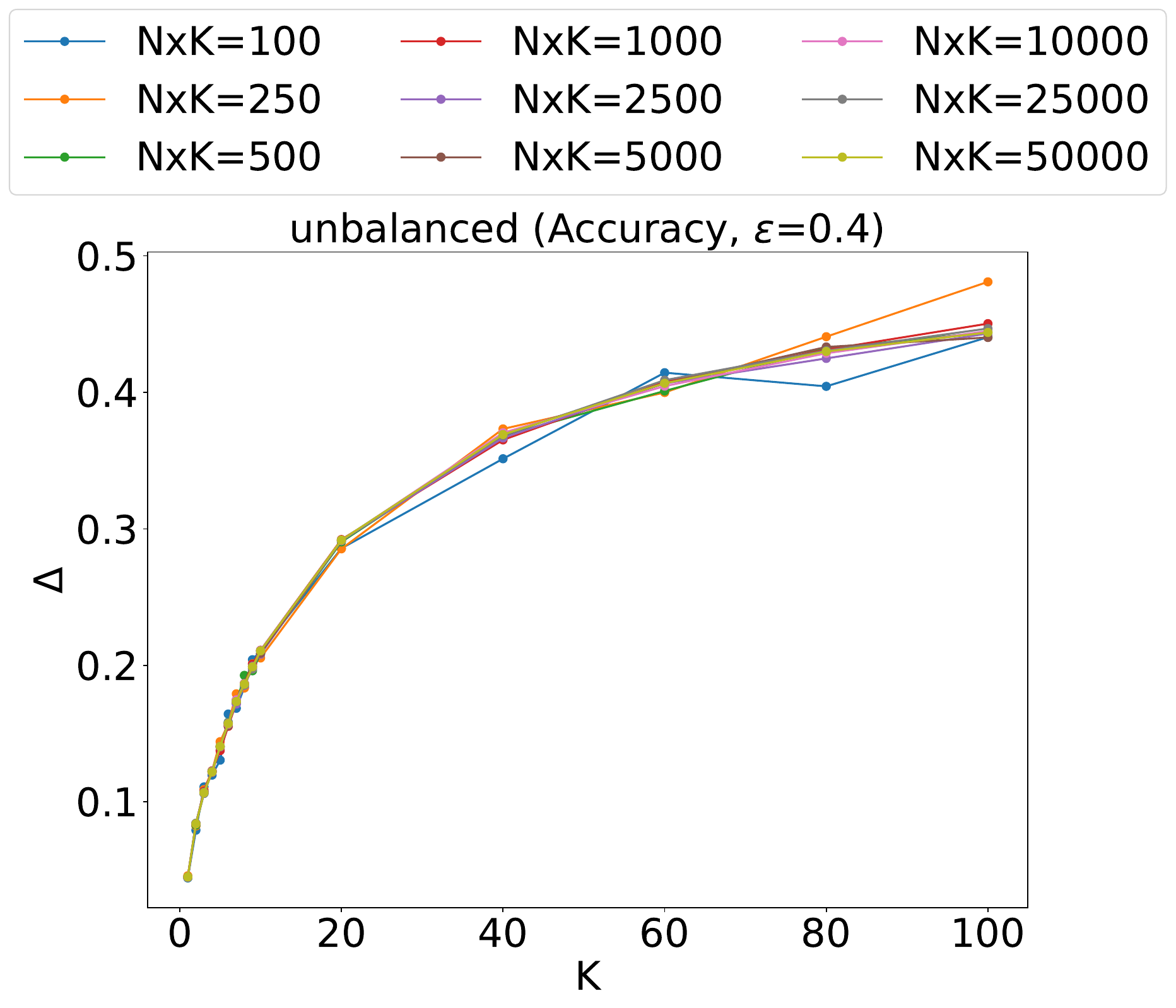}
    \caption{$\epsilon = 0.4$}
    \label{fig:gamma_delta_accuracy_cat5_e04}
  \end{subfigure}
  \caption{Effect sizes ($\Delta$) for unbalanced alphas with Accuracy as the metric ($M=5$)}
  \label{fig:gamma_delta_accuracy_cat5}
\end{figure*}

\begin{figure*}
  \centering
  \begin{subfigure}[b]{0.24\linewidth}
    \centering
    \includegraphics[width=\linewidth]{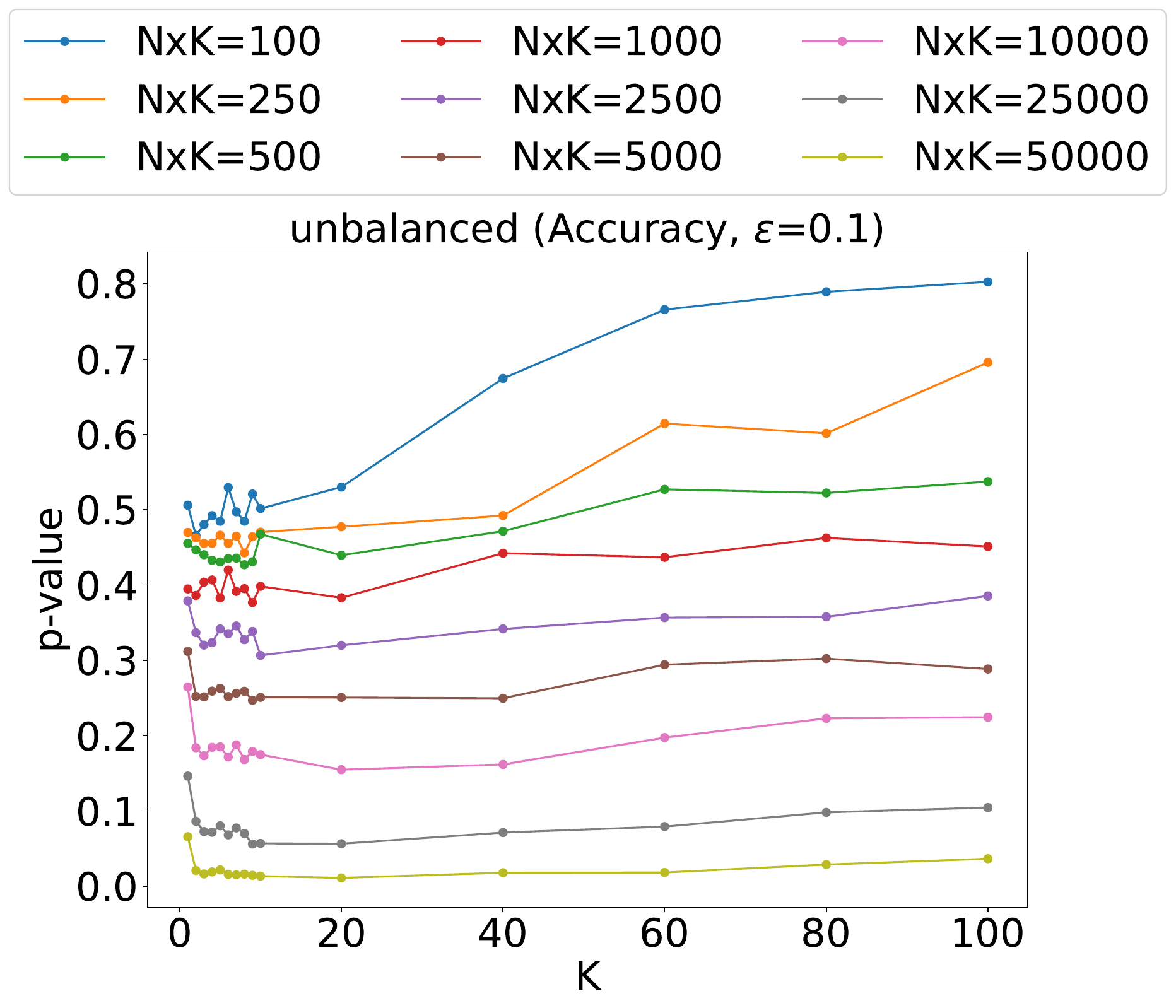}
    \caption{$\epsilon = 0.1$}
    \label{fig:gamma_accuracy_cat12_e01}
  \end{subfigure} \hfill
  \begin{subfigure}[b]{0.24\linewidth}
    \centering
    \includegraphics[width=\linewidth]{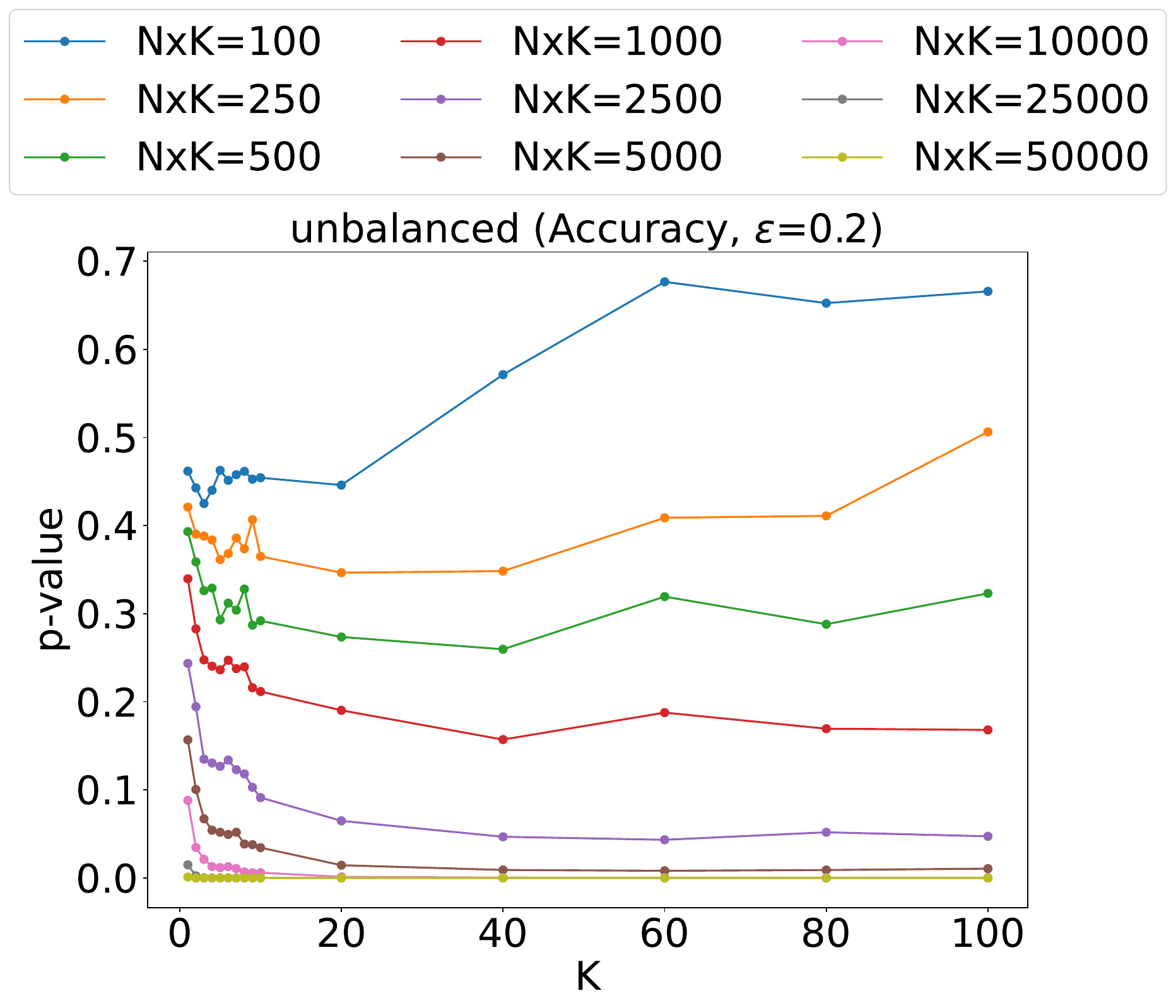}
    \caption{$\epsilon = 0.2$}
    \label{fig:gamma_accuracy_cat12_e02}
  \end{subfigure} \hfill
  \begin{subfigure}[b]{0.24\linewidth}
    \centering
    \includegraphics[width=\linewidth]{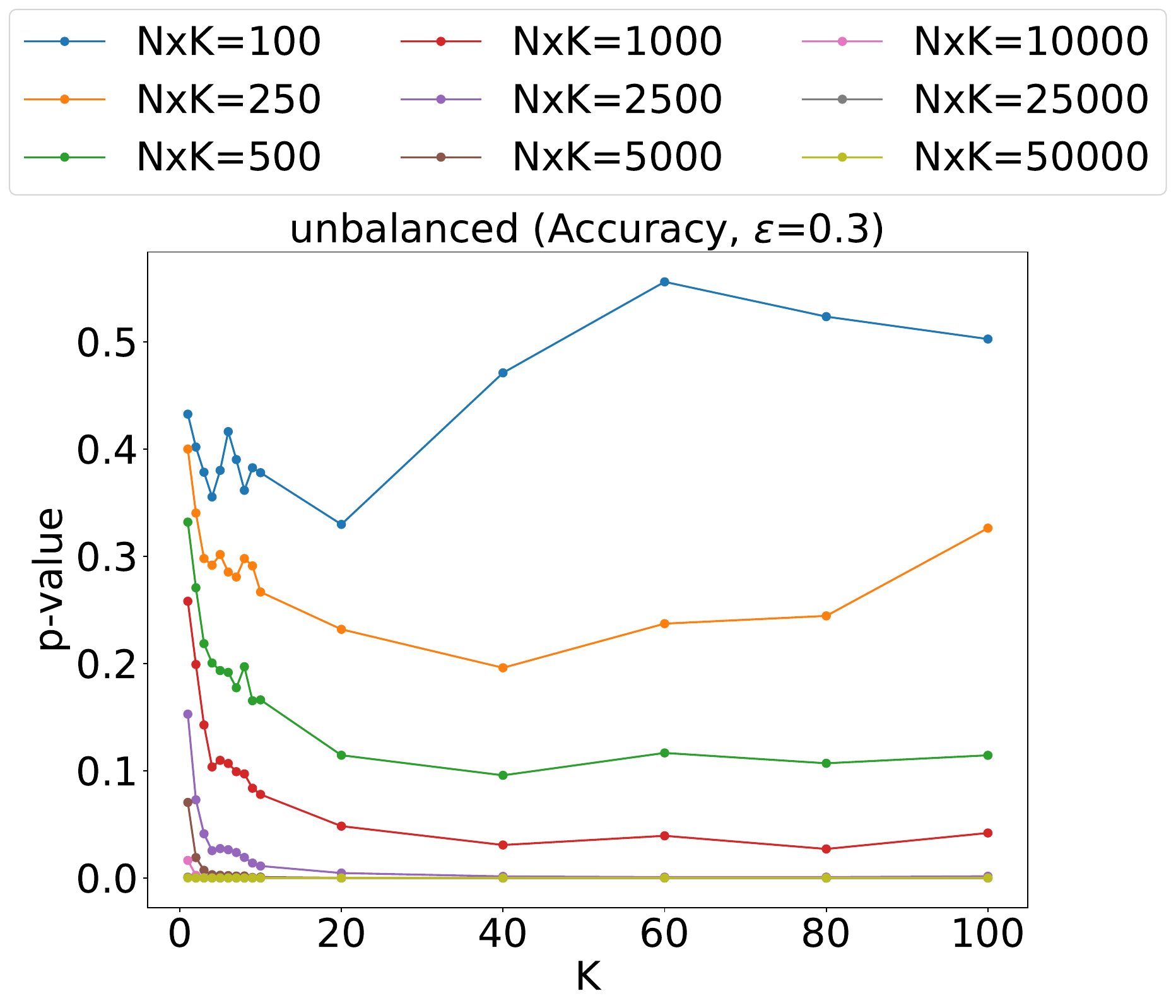}
    \caption{$\epsilon = 0.3$}
    \label{fig:gamma_accuracy_cat12_e03}
  \end{subfigure} \hfill
  \begin{subfigure}[b]{0.24\linewidth}
    \centering
    \includegraphics[width=\linewidth]{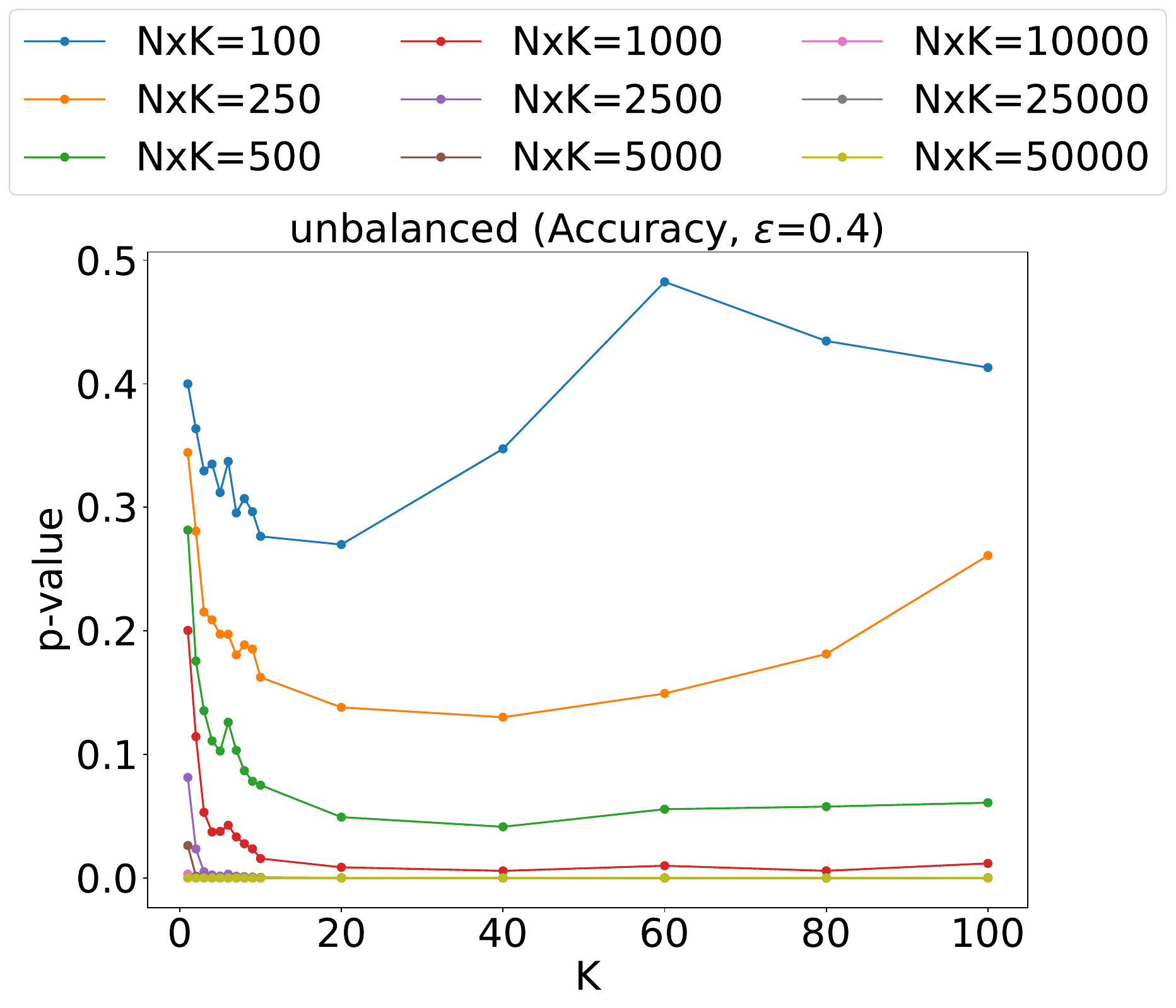}
    \caption{$\epsilon = 0.4$}
    \label{fig:gamma_accuracy_cat12_e04}
  \end{subfigure}
  \caption{P-value plots for unbalanced alphas with Accuracy as the metric ($M=12$)}
  \label{fig:gamma_accuracy_cat12}
\end{figure*}

\begin{figure*}
  \centering
  \begin{subfigure}[b]{0.24\linewidth}
    \centering
    \includegraphics[width=\linewidth]{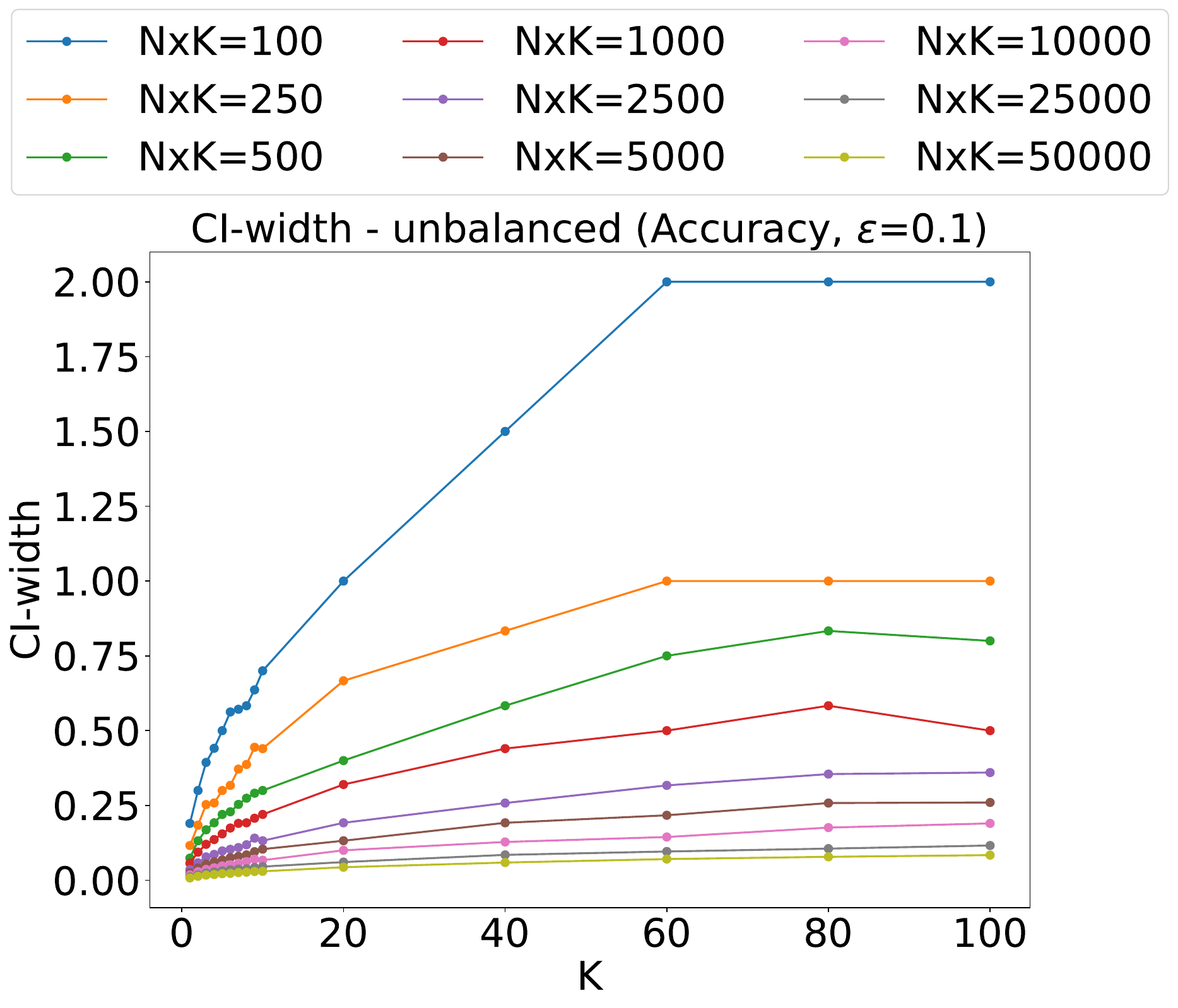}
    \caption{$\epsilon = 0.1$}
    \label{fig:gamma_ci_accuracy_cat12_e01}
  \end{subfigure} \hfill
  \begin{subfigure}[b]{0.24\linewidth}
    \centering
    \includegraphics[width=\linewidth]{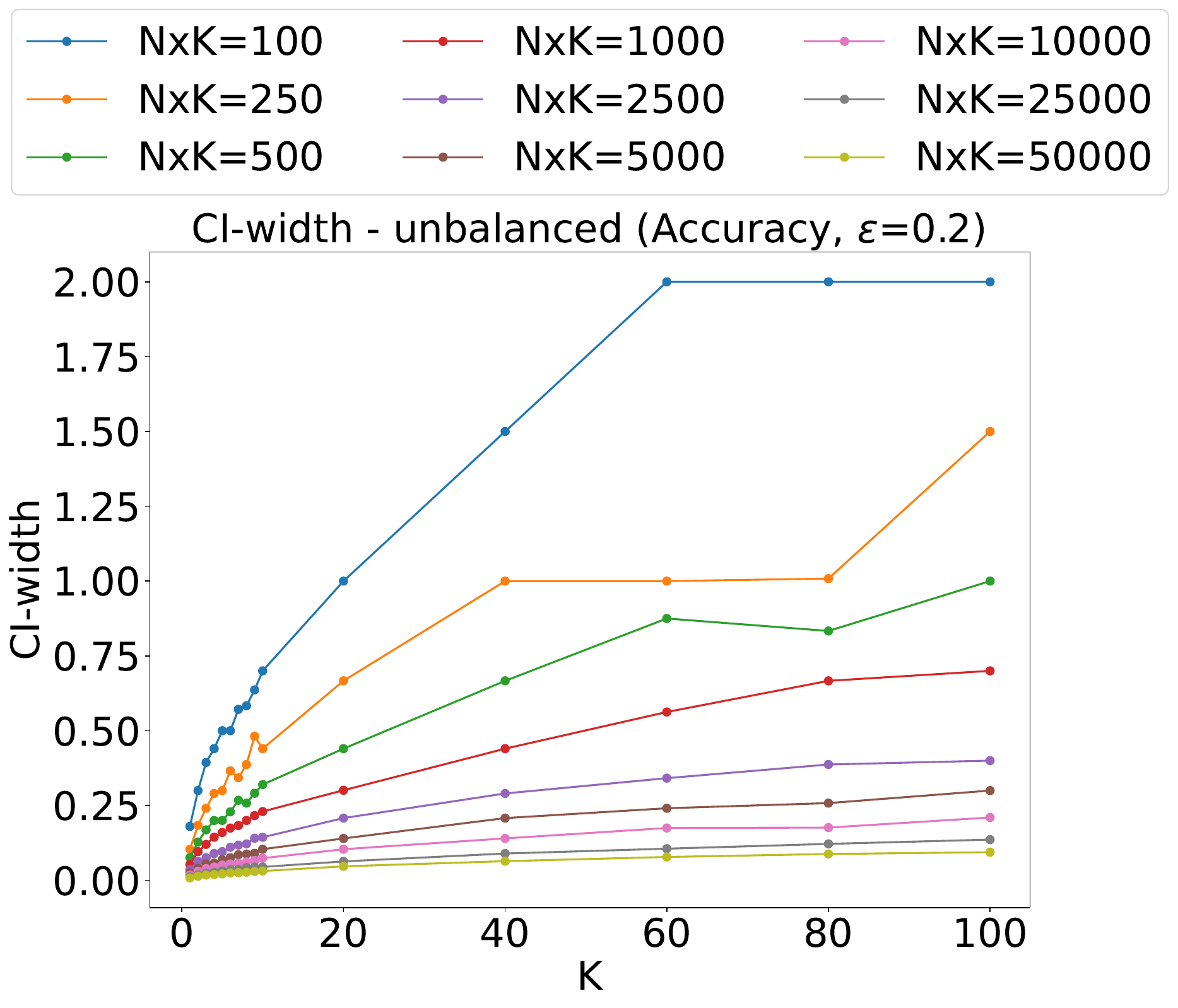}
    \caption{$\epsilon = 0.2$}
    \label{fig:gamma_ci_accuracy_cat12_e02}
  \end{subfigure} \hfill
  \begin{subfigure}[b]{0.24\linewidth}
    \centering
    \includegraphics[width=\linewidth]{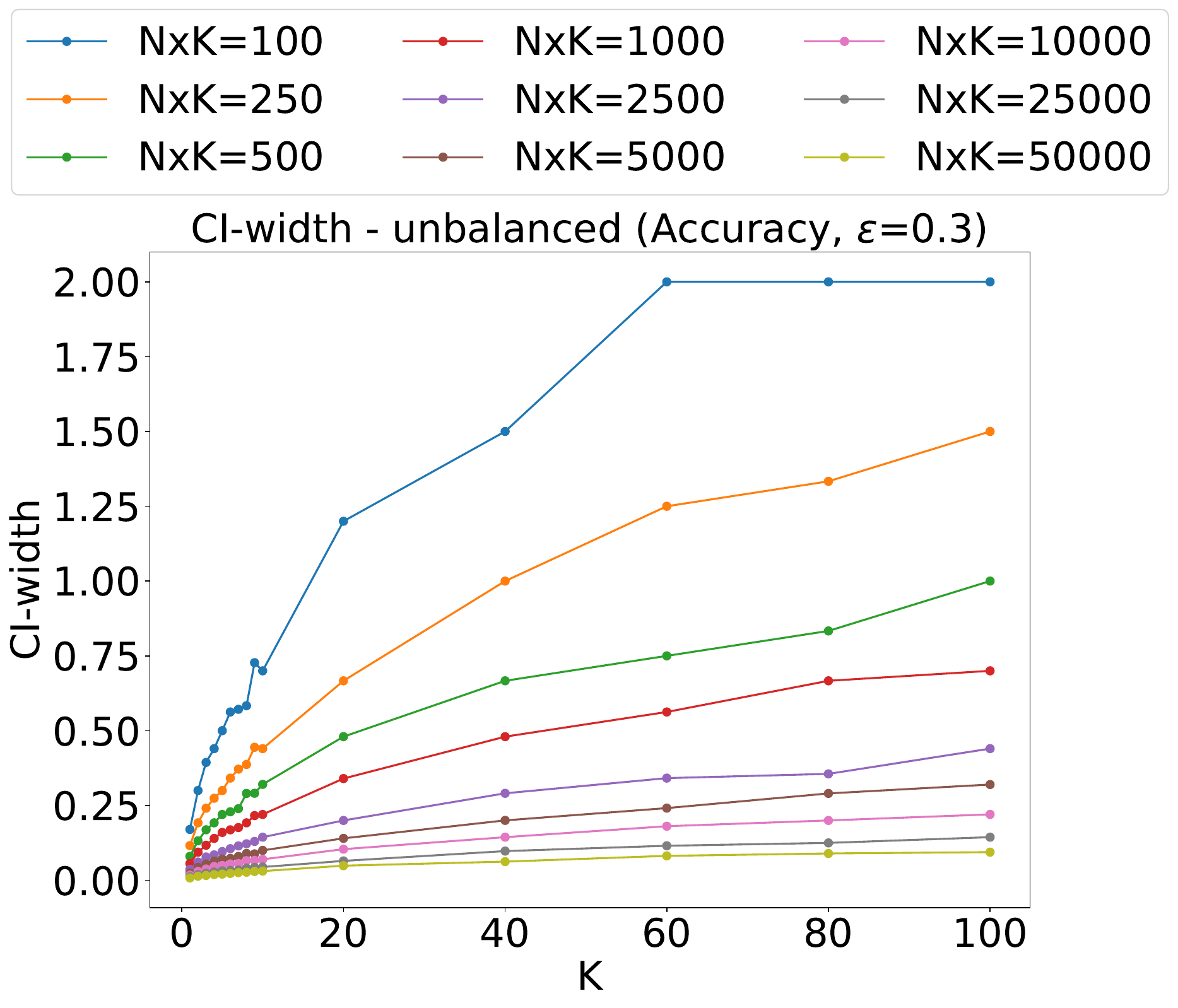}
    \caption{$\epsilon = 0.3$}
    \label{fig:gamma_ci_accuracy_cat12_e03}
  \end{subfigure} \hfill
  \begin{subfigure}[b]{0.24\linewidth}
    \centering
    \includegraphics[width=\linewidth]{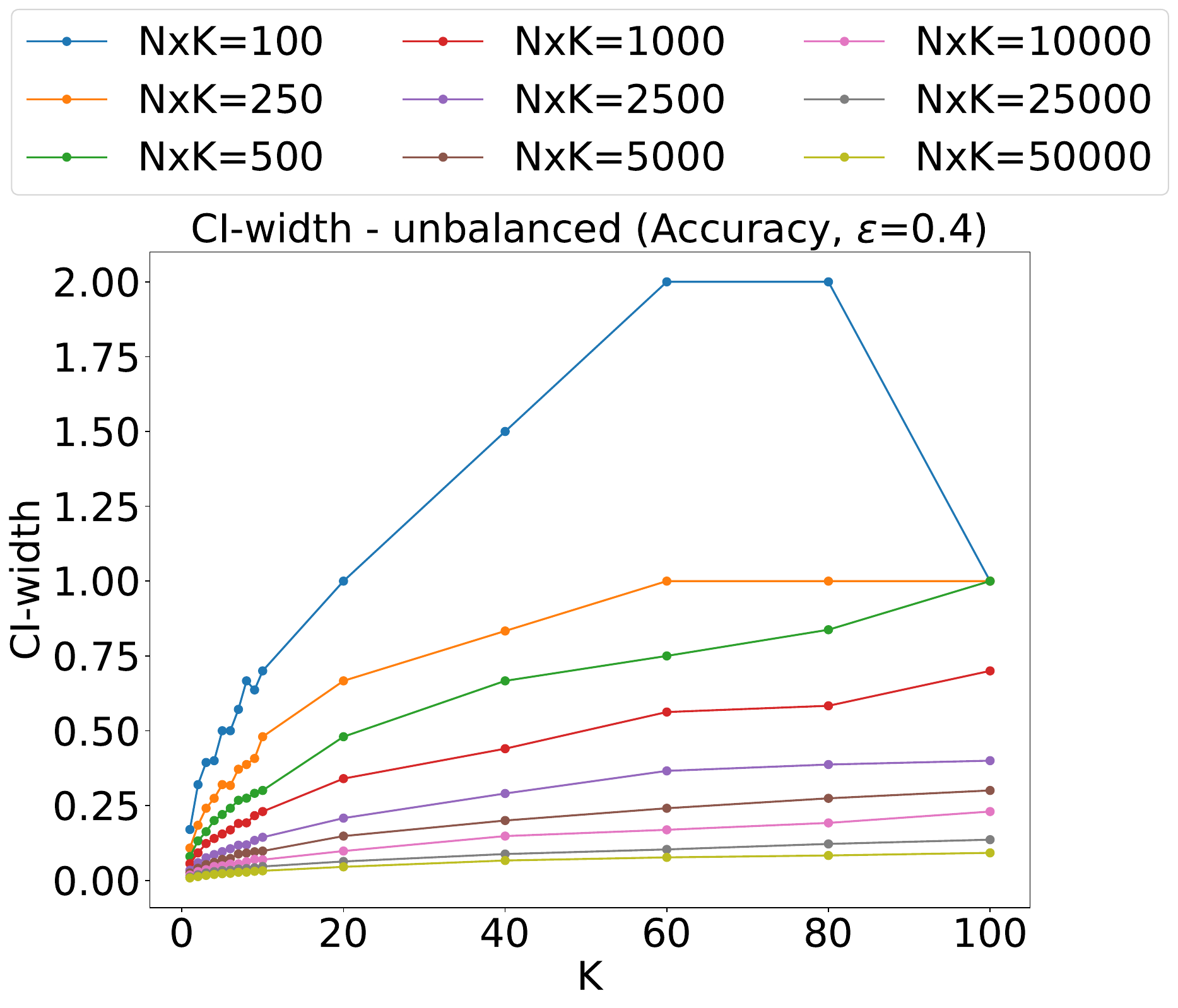}
    \caption{$\epsilon = 0.4$}
    \label{fig:gamma_ci_accuracy_cat12_e04}
  \end{subfigure}
  \caption{CI-width plots for unbalanced alphas with Accuracy as the metric ($M=12$)}
  \label{fig:gamma_ci_accuracy_cat12}
\end{figure*}

\begin{figure*}
  \centering
  \begin{subfigure}[b]{0.24\linewidth}
    \centering
    \includegraphics[width=\linewidth]{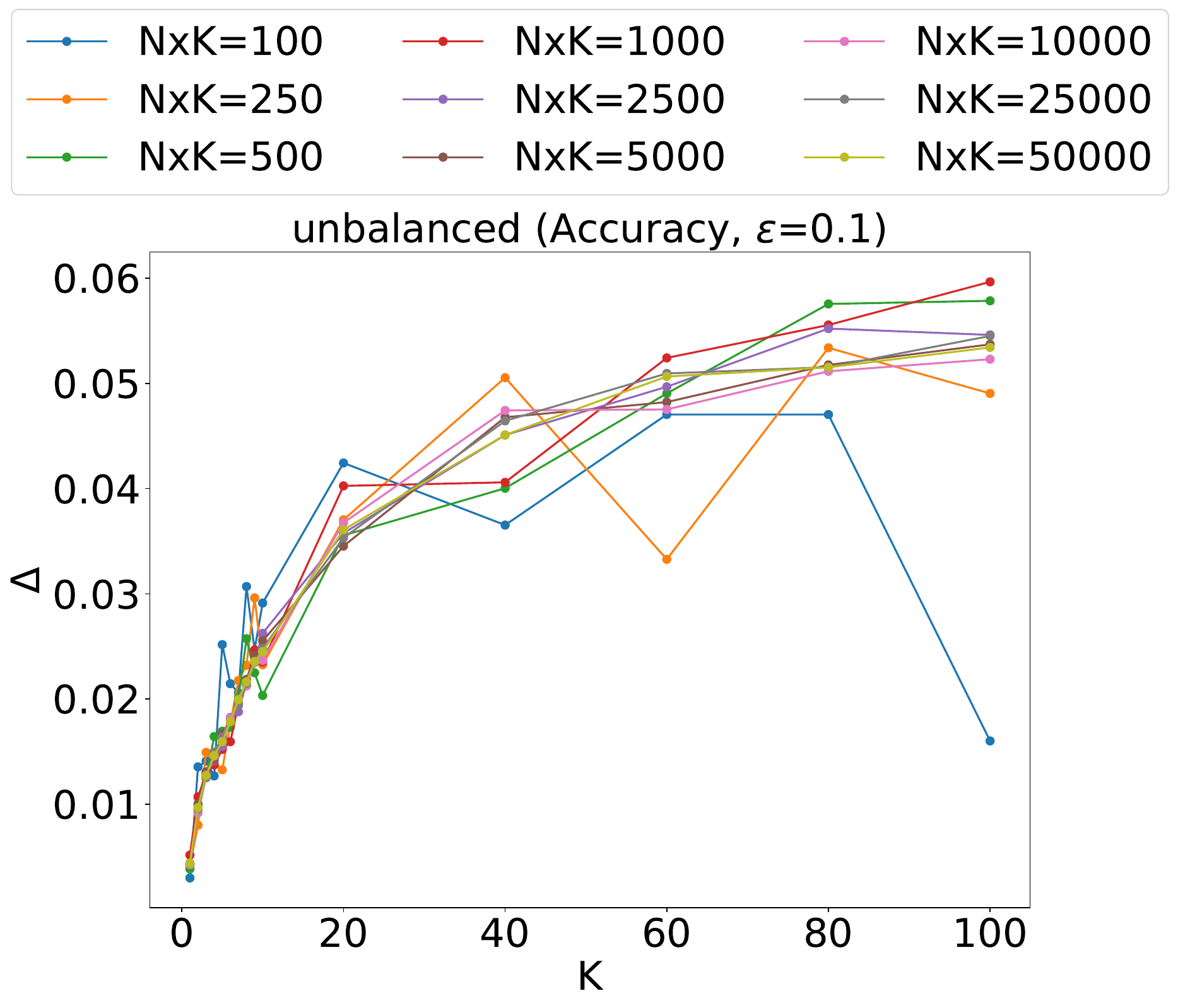}
    \caption{$\epsilon = 0.1$}
    \label{fig:gamma_delta_accuracy_cat12_e01}
  \end{subfigure} \hfill
  \begin{subfigure}[b]{0.24\linewidth}
    \centering
    \includegraphics[width=\linewidth]{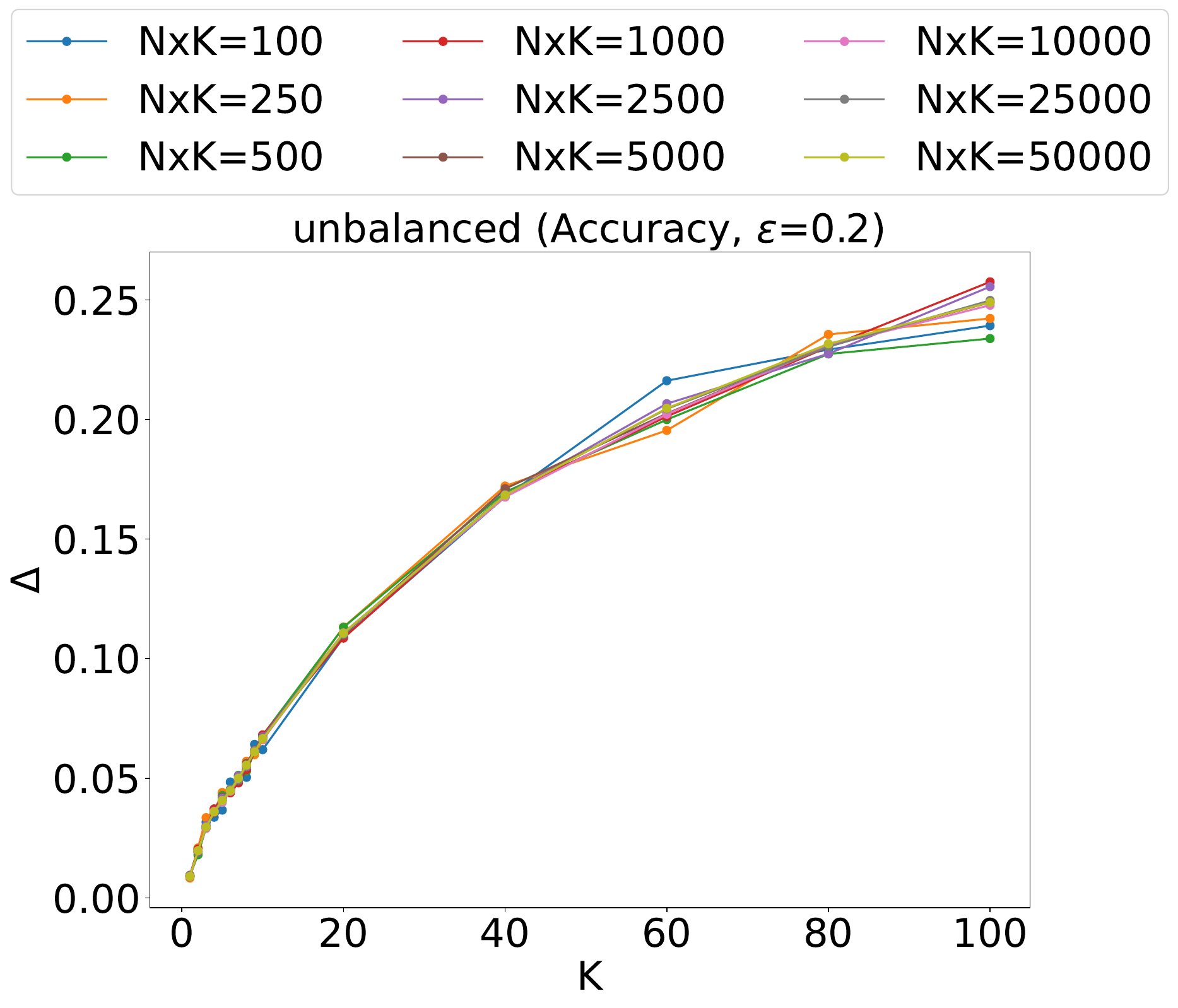}
    \caption{$\epsilon = 0.2$}
    \label{fig:gamma_delta_accuracy_cat12_e02}
  \end{subfigure} \hfill
  \begin{subfigure}[b]{0.24\linewidth}
    \centering
    \includegraphics[width=\linewidth]{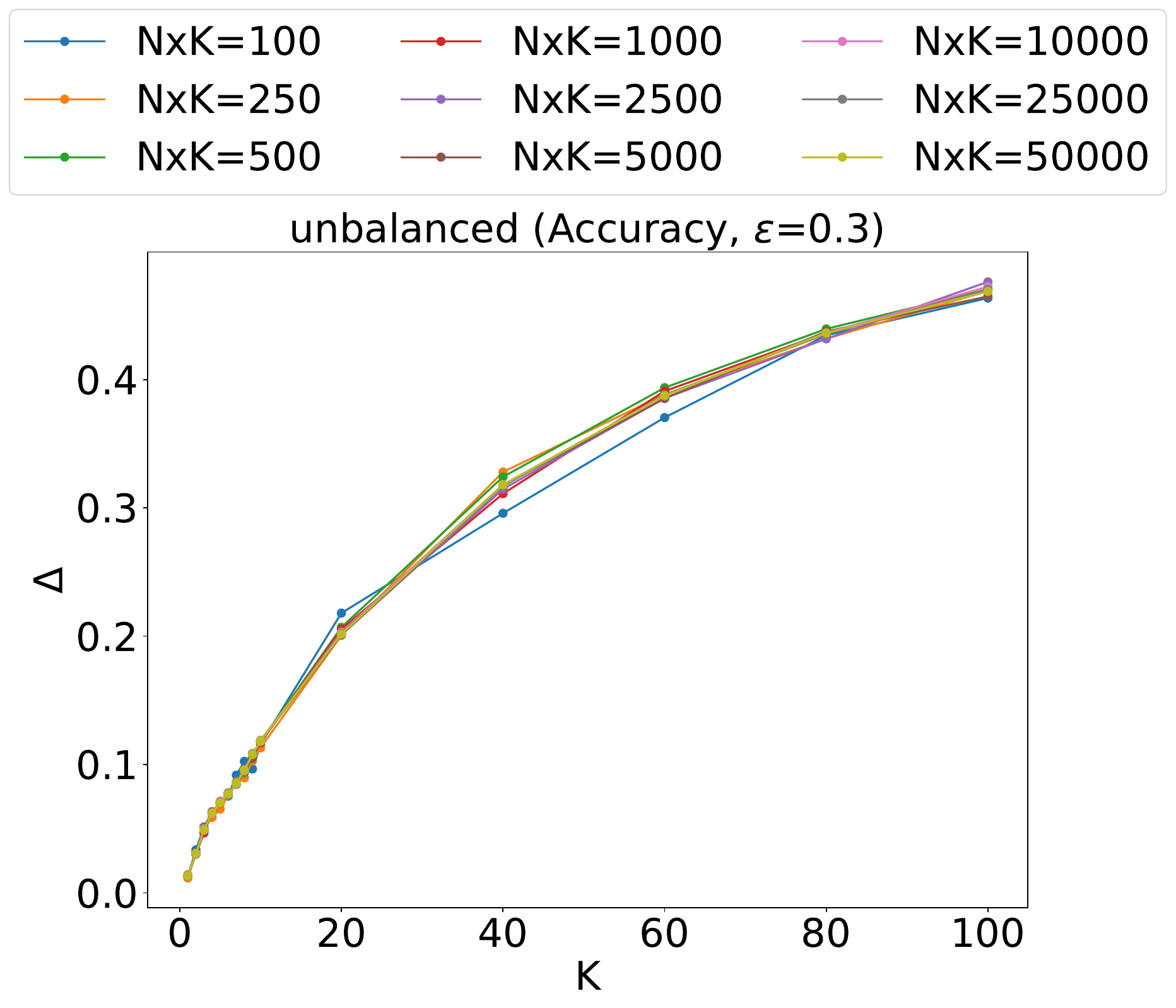}
    \caption{$\epsilon = 0.3$}
    \label{fig:gamma_delta_accuracy_cat12_e03}
  \end{subfigure} \hfill
  \begin{subfigure}[b]{0.24\linewidth}
    \centering
    \includegraphics[width=\linewidth]{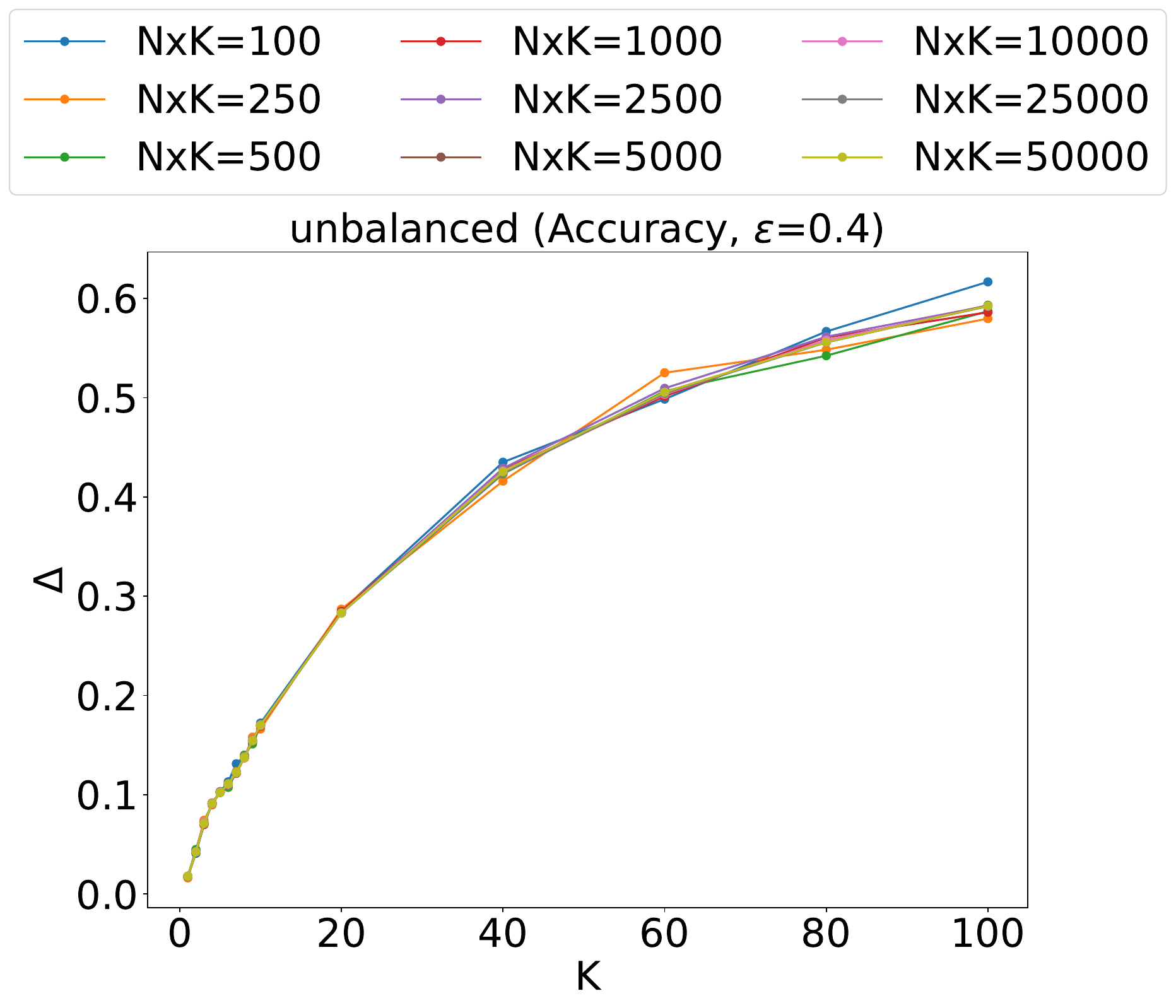}
    \caption{$\epsilon = 0.4$}
    \label{fig:gamma_delta_accuracy_cat12_e04}
  \end{subfigure}
  \caption{Effect sizes ($\Delta$) for unbalanced alphas with Accuracy as the metric ($M=12$)}
  \label{fig:gamma_delta_accuracy_cat12}
\end{figure*}

Figures \ref{fig:gamma_accuracy_cat2}--\ref{fig:gamma_delta_accuracy_cat12} show results for \pv s, confidence intervals, and effect sizes for unbalanced priors on parameter $\alpha$ with different values of $\epsilon$ and increasing $K$ for various $N \times K$. For simulations with unbalanced categories, \pv s\ and CI-width increase as $K$ increases. Effect sizes increase with higher $K$ except for $\epsilon<=0.2$; however, the behavior stabilizes as $M$ becomes higher.

\subsection{Total variation (TV)}

\subsubsection{Real-World Datasets}
\begin{figure*}
  \centering
  \begin{subfigure}[b]{0.24\linewidth}
    \centering
    \includegraphics[width=\linewidth]{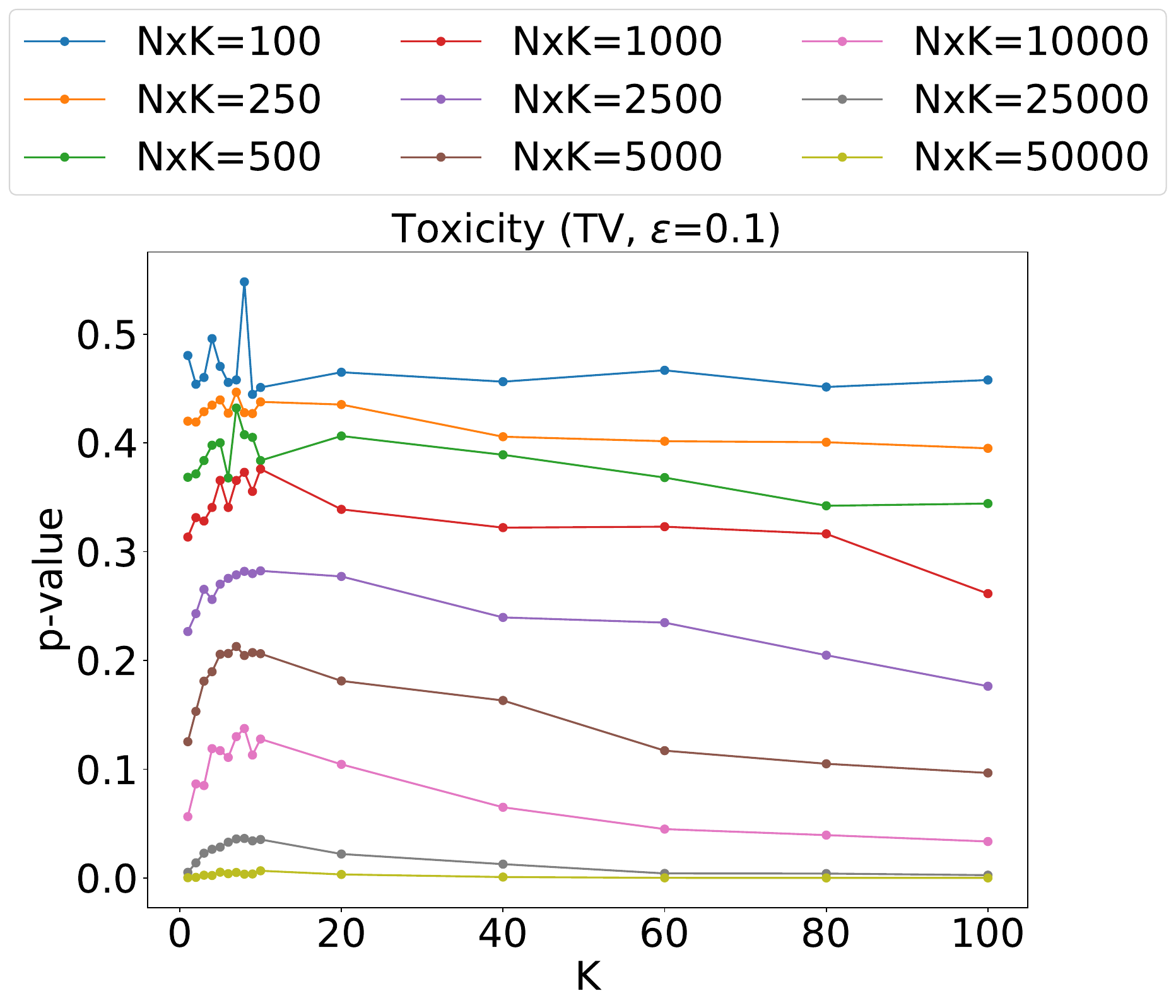}
    \caption{$\epsilon = 0.1$}
    \label{fig:toxicity_MAE_e01}
  \end{subfigure} \hfill
  \begin{subfigure}[b]{0.24\linewidth}
    \centering
    \includegraphics[width=\linewidth]{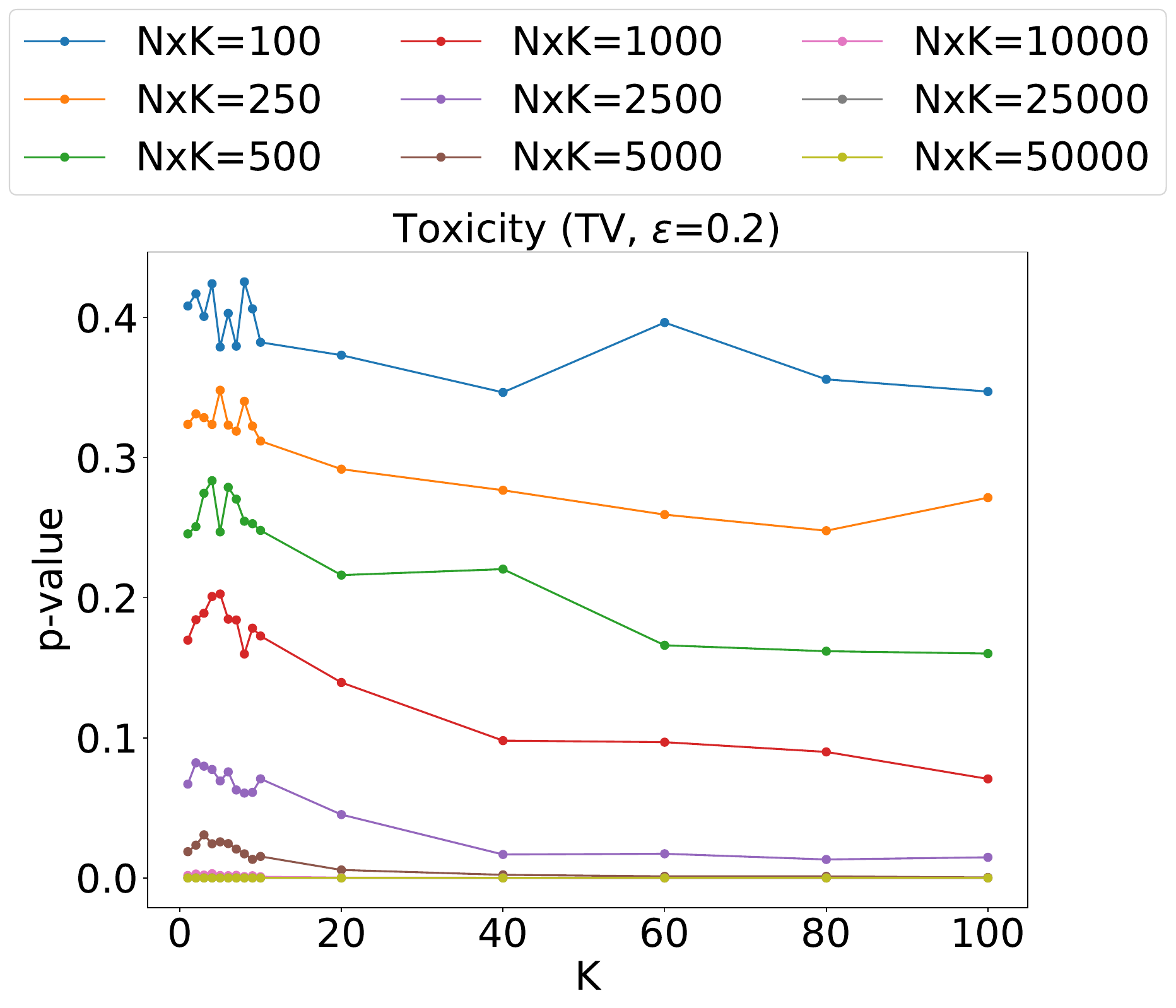}
    \caption{$\epsilon = 0.2$}
    \label{fig:toxicity_MAE_e02}
  \end{subfigure} \hfill
  \begin{subfigure}[b]{0.24\linewidth}
    \centering
    \includegraphics[width=\linewidth]{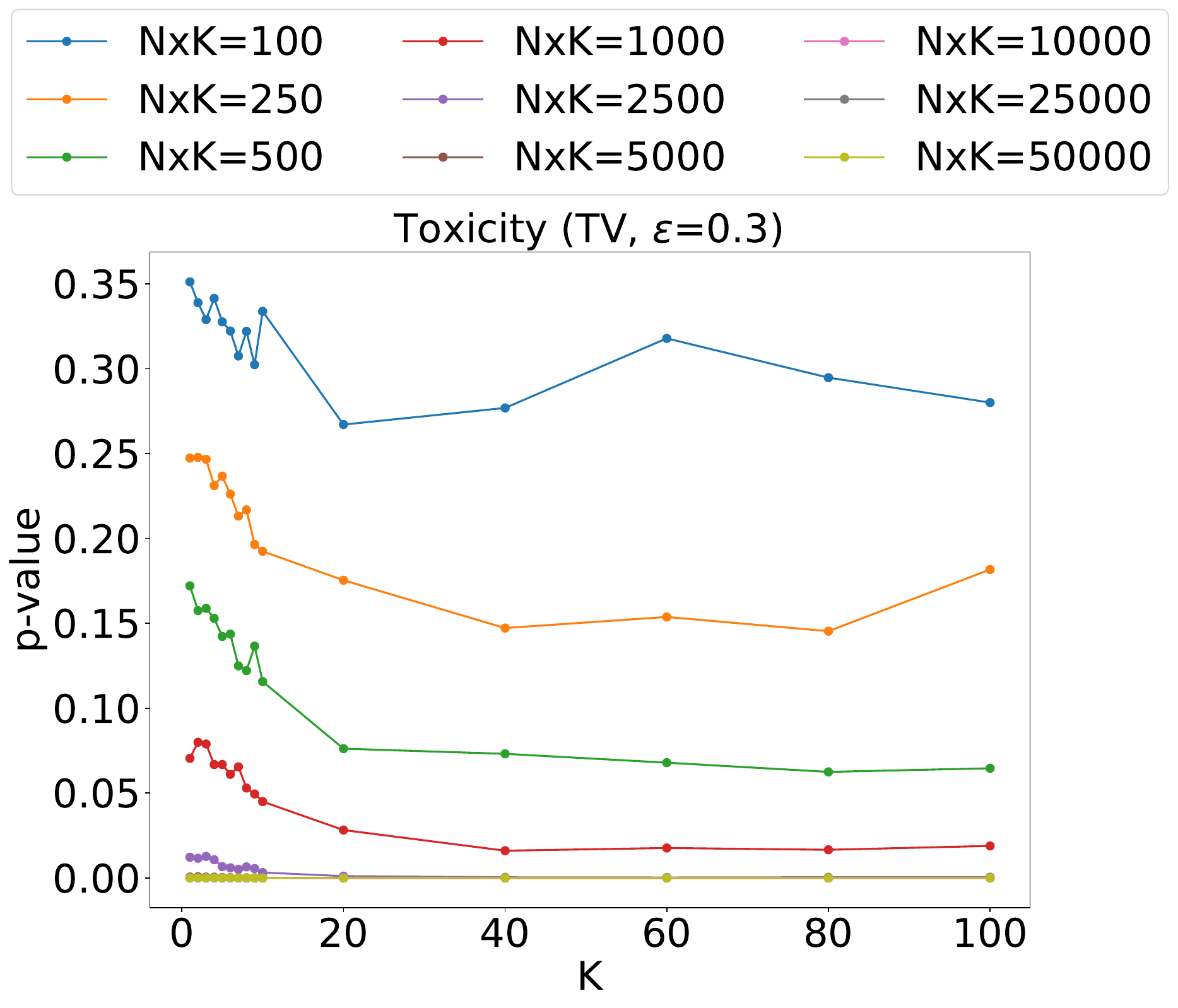}
    \caption{$\epsilon = 0.3$}
    \label{fig:toxicity_MAE_e03}
  \end{subfigure} \hfill
  \begin{subfigure}[b]{0.24\linewidth}
    \centering
    \includegraphics[width=\linewidth]{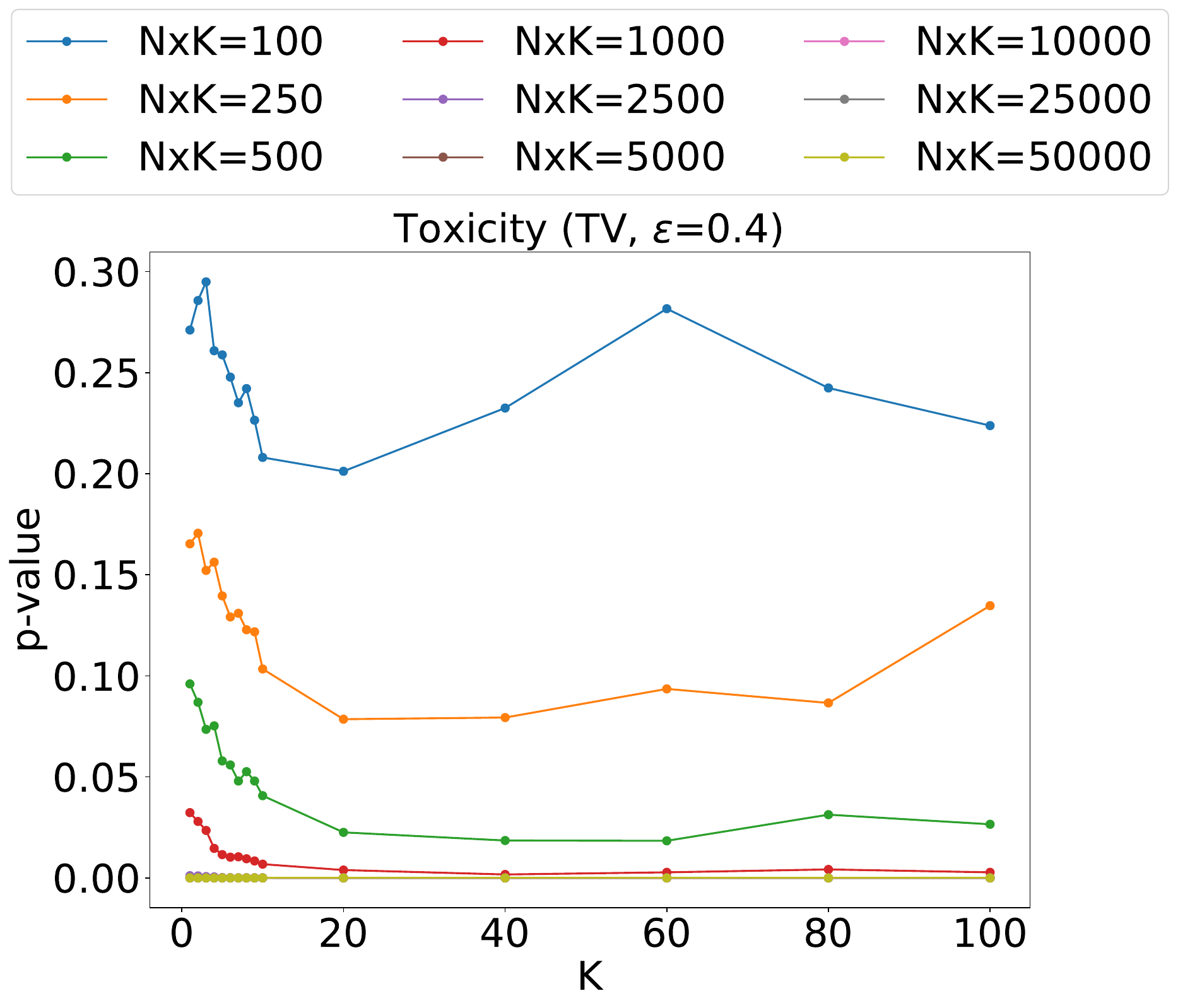}
    \caption{$\epsilon = 0.4$}
    \label{fig:toxicity_MAE_e04}
  \end{subfigure}
  \caption{P-value plots for Toxicity dataset with TV as the metric}
  \label{fig:toxicity_MAE}
\end{figure*}

\begin{figure*}
  \centering
  \begin{subfigure}[b]{0.24\linewidth}
    \centering
    \includegraphics[width=\linewidth]{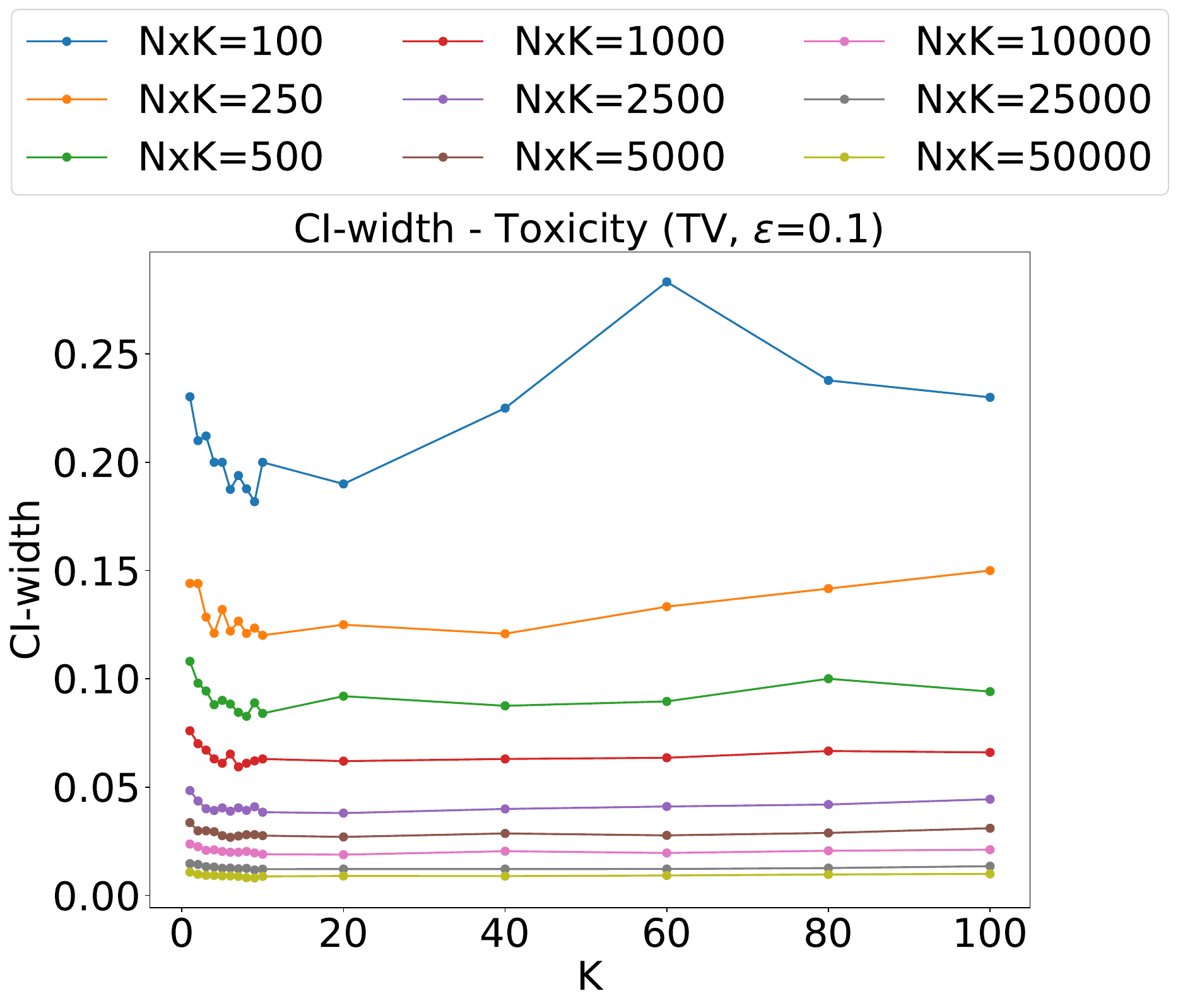}
    \caption{$\epsilon = 0.1$}
    \label{fig:toxicity_ci_MAE_e01}
  \end{subfigure} \hfill
  \begin{subfigure}[b]{0.24\linewidth}
    \centering
    \includegraphics[width=\linewidth]{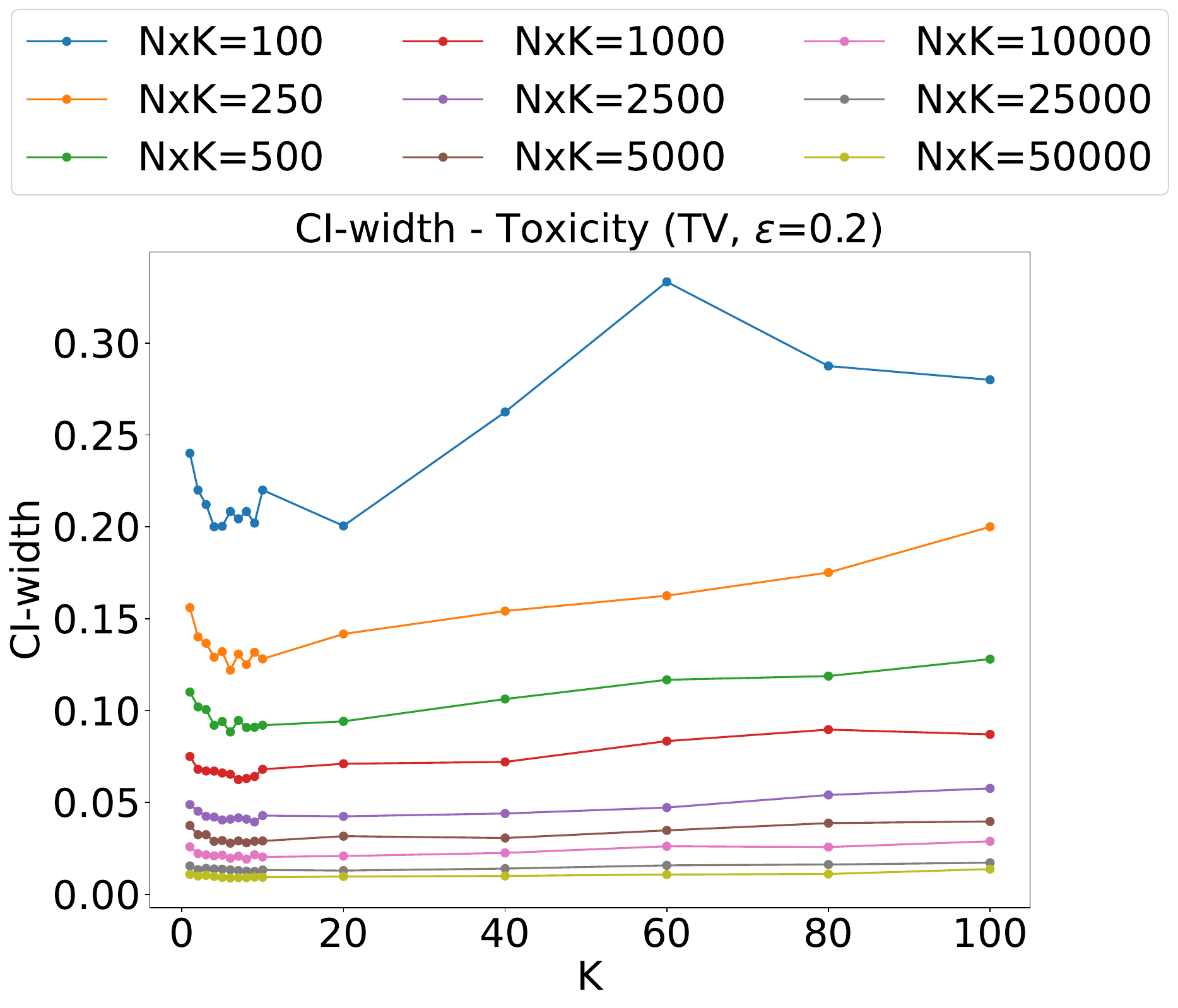}
    \caption{$\epsilon = 0.2$}
    \label{fig:toxicity_ci_MAE_e02}
  \end{subfigure} \hfill
  \begin{subfigure}[b]{0.24\linewidth}
    \centering
    \includegraphics[width=\linewidth]{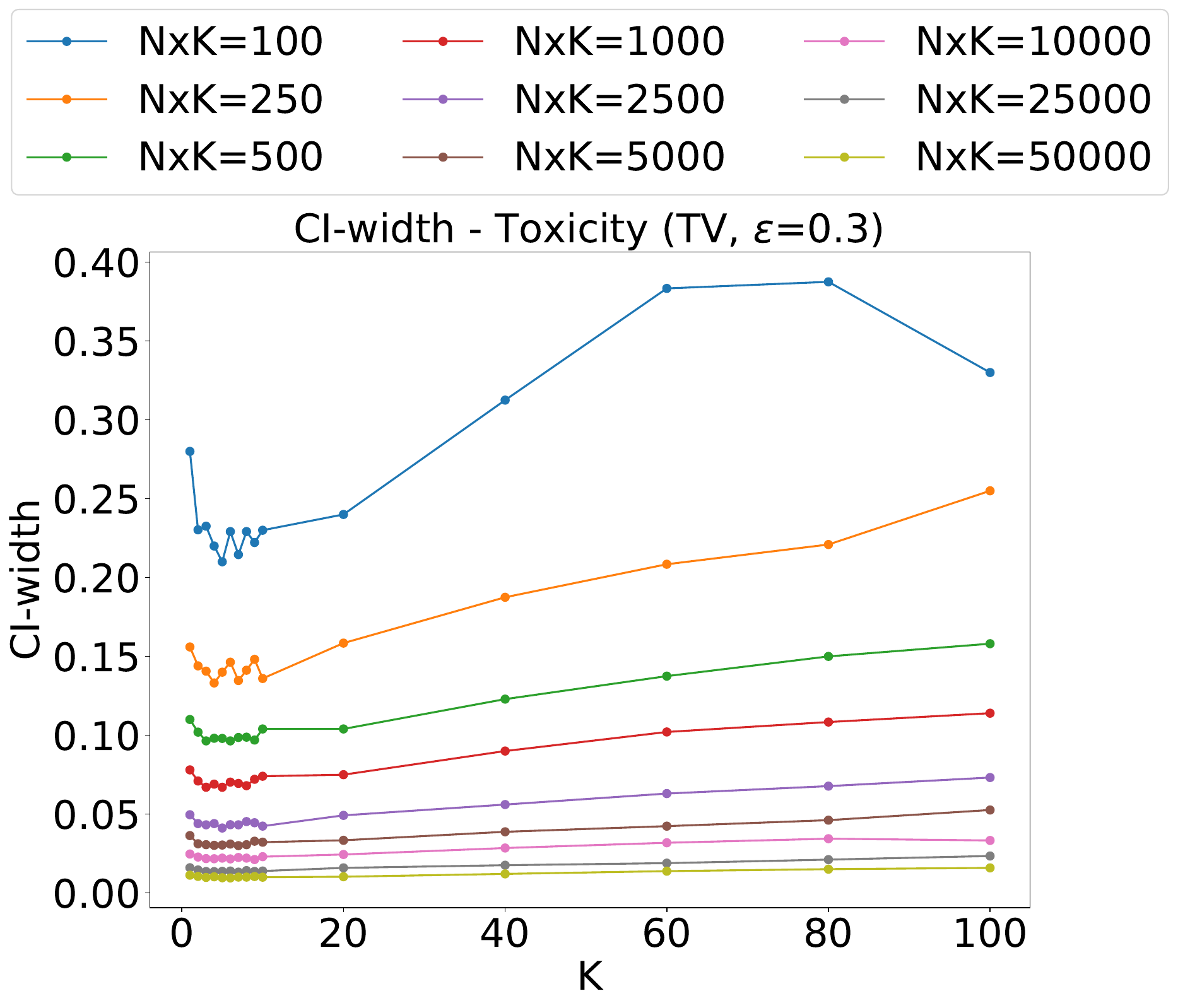}
    \caption{$\epsilon = 0.3$}
    \label{fig:toxicity_ci_MAE_e03}
  \end{subfigure} \hfill
  \begin{subfigure}[b]{0.24\linewidth}
    \centering
    \includegraphics[width=\linewidth]{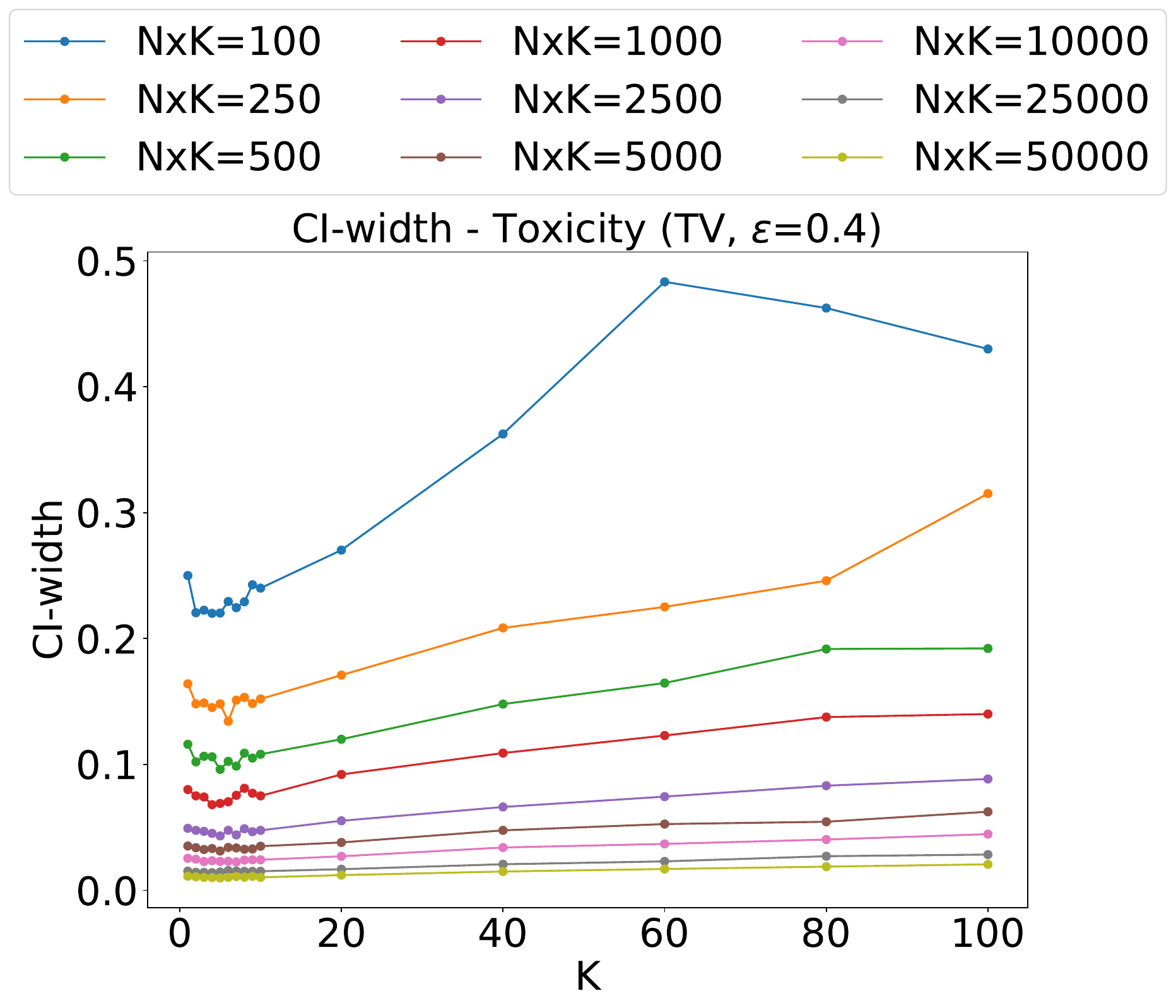}
    \caption{$\epsilon = 0.4$}
    \label{fig:toxicity_ci_MAE_e04}
  \end{subfigure}
  \caption{CI-width plots for Toxicity dataset with TV as the metric}
  \label{fig:toxicity_ci_MAE}
\end{figure*}

\begin{figure*}
  \centering
  \begin{subfigure}[b]{0.24\linewidth}
    \centering
    \includegraphics[width=\linewidth]{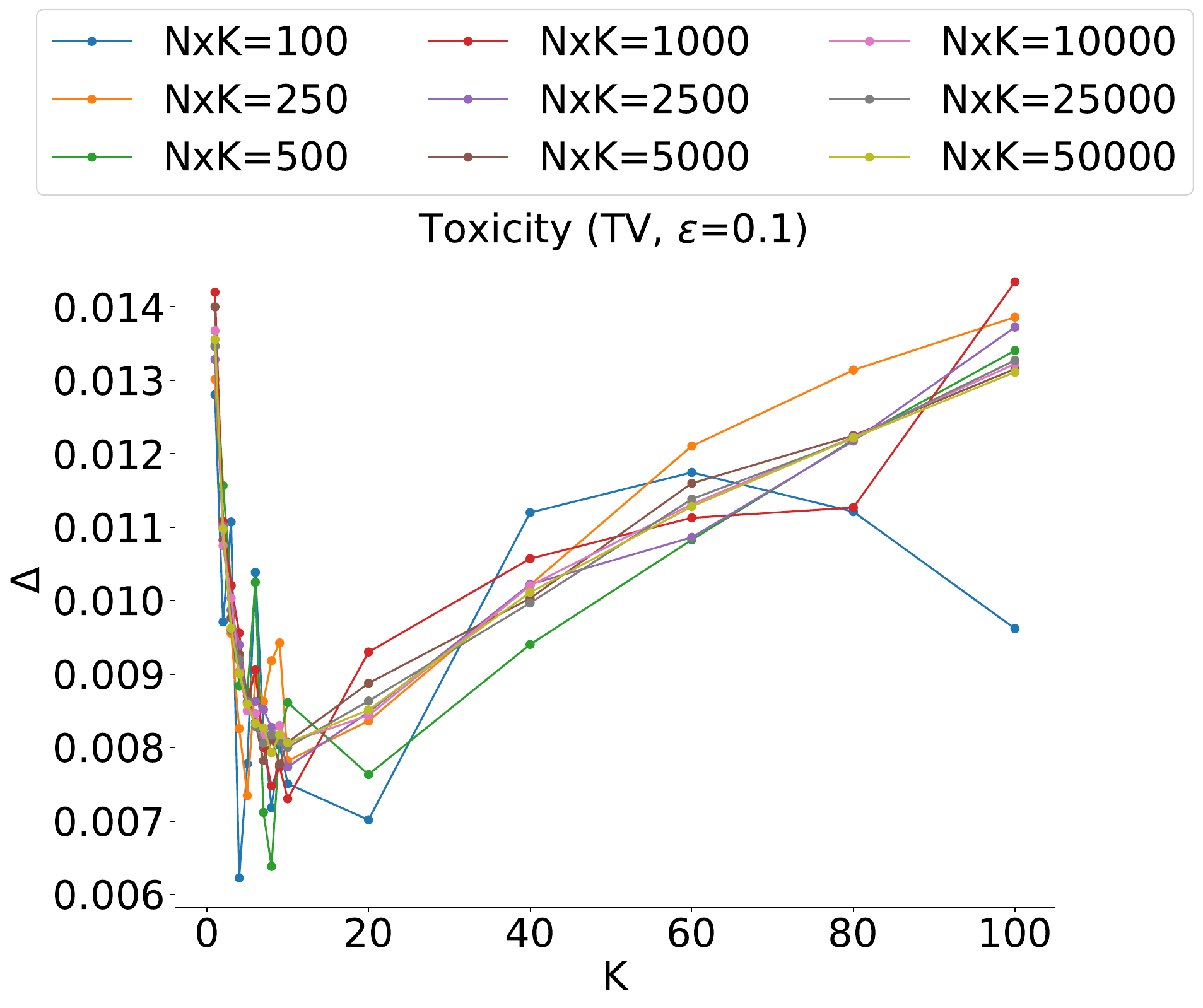}
    \caption{$\epsilon = 0.1$}
    \label{fig:toxicity_delta_MAE_e01}
  \end{subfigure} \hfill
  \begin{subfigure}[b]{0.24\linewidth}
    \centering
    \includegraphics[width=\linewidth]{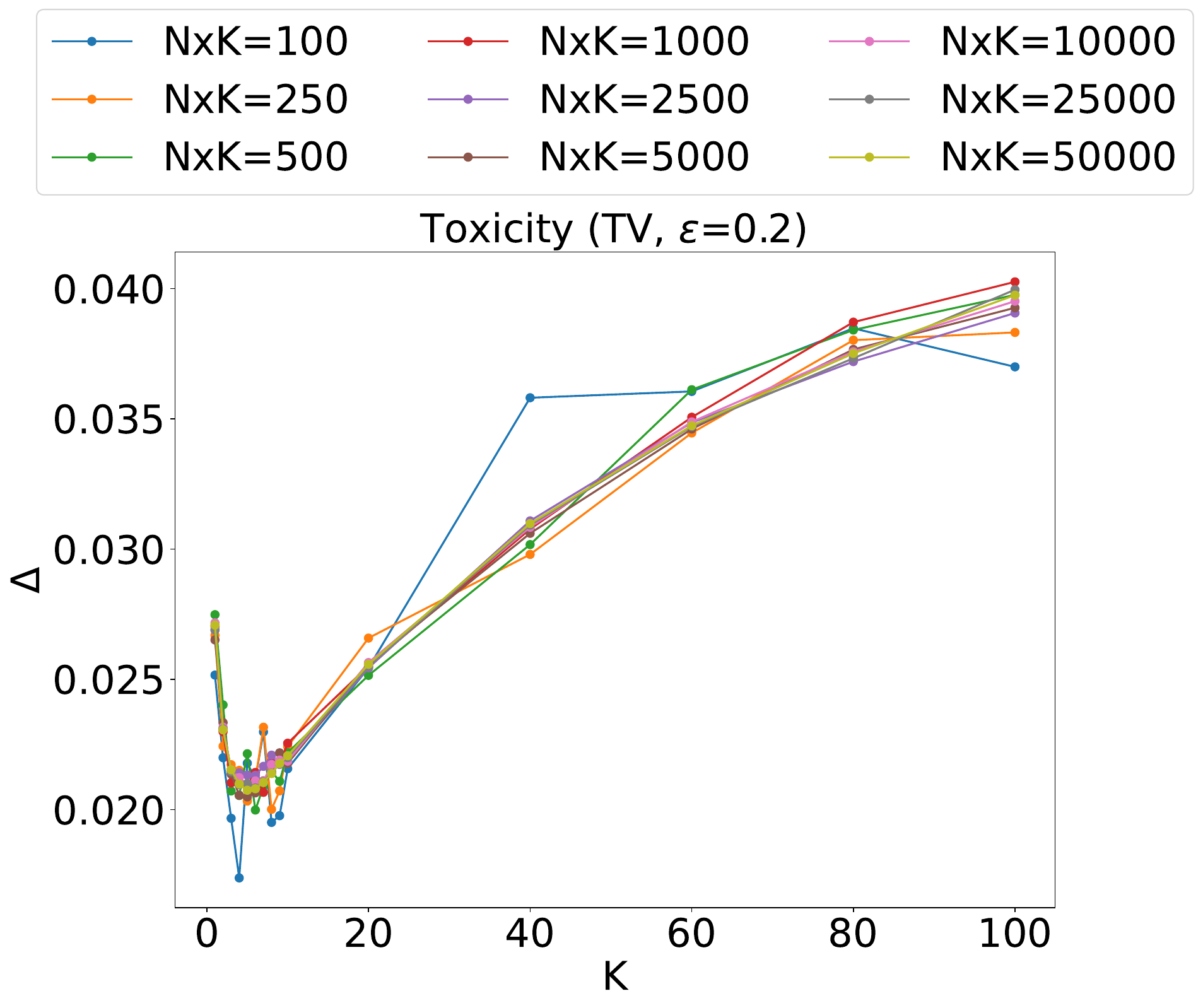}
    \caption{$\epsilon = 0.2$}
    \label{fig:toxicity_delta_MAE_e02}
  \end{subfigure} \hfill
  \begin{subfigure}[b]{0.24\linewidth}
    \centering
    \includegraphics[width=\linewidth]{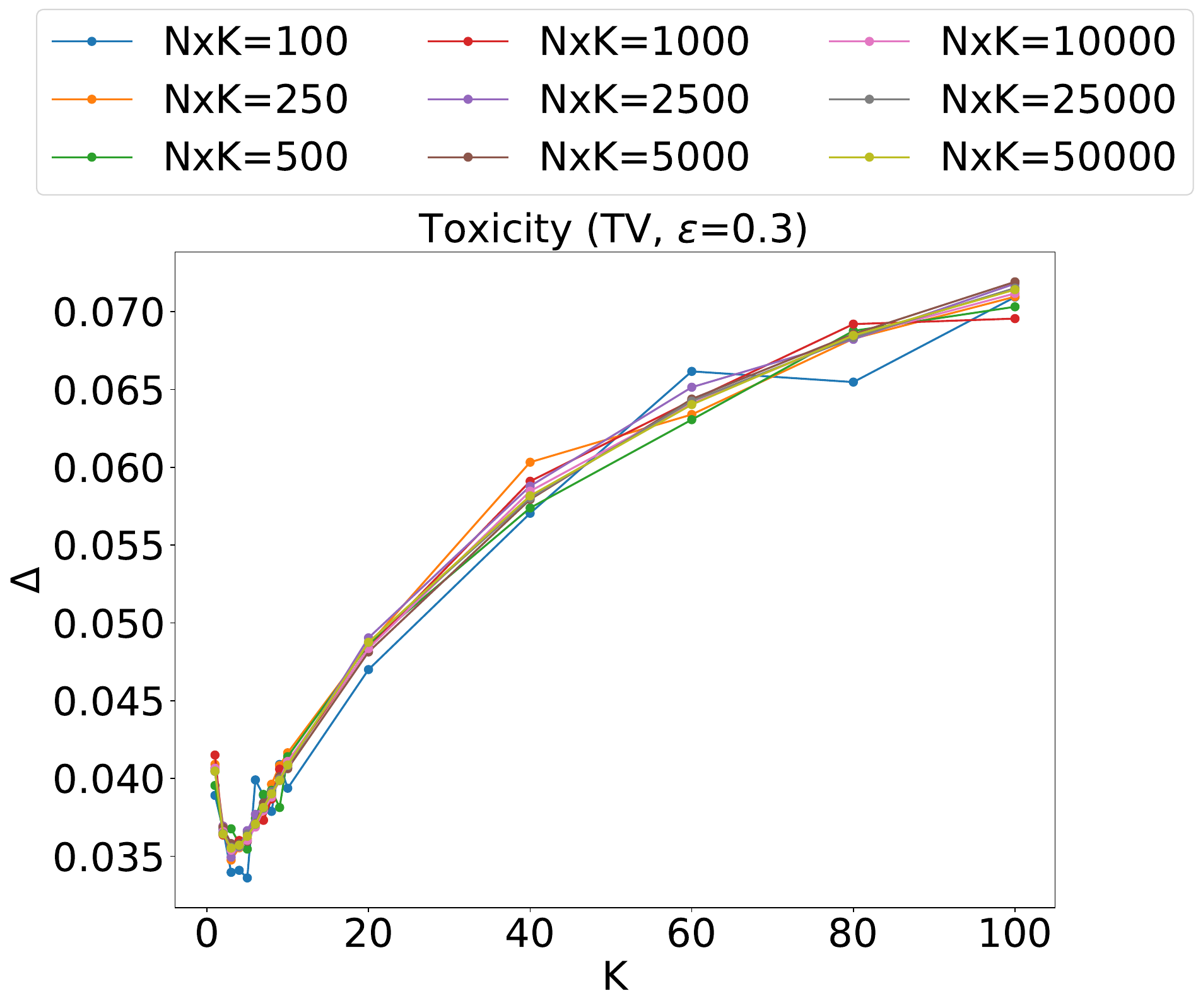}
    \caption{$\epsilon = 0.3$}
    \label{fig:toxicity_delta_MAE_e03}
  \end{subfigure} \hfill
  \begin{subfigure}[b]{0.24\linewidth}
    \centering
    \includegraphics[width=\linewidth]{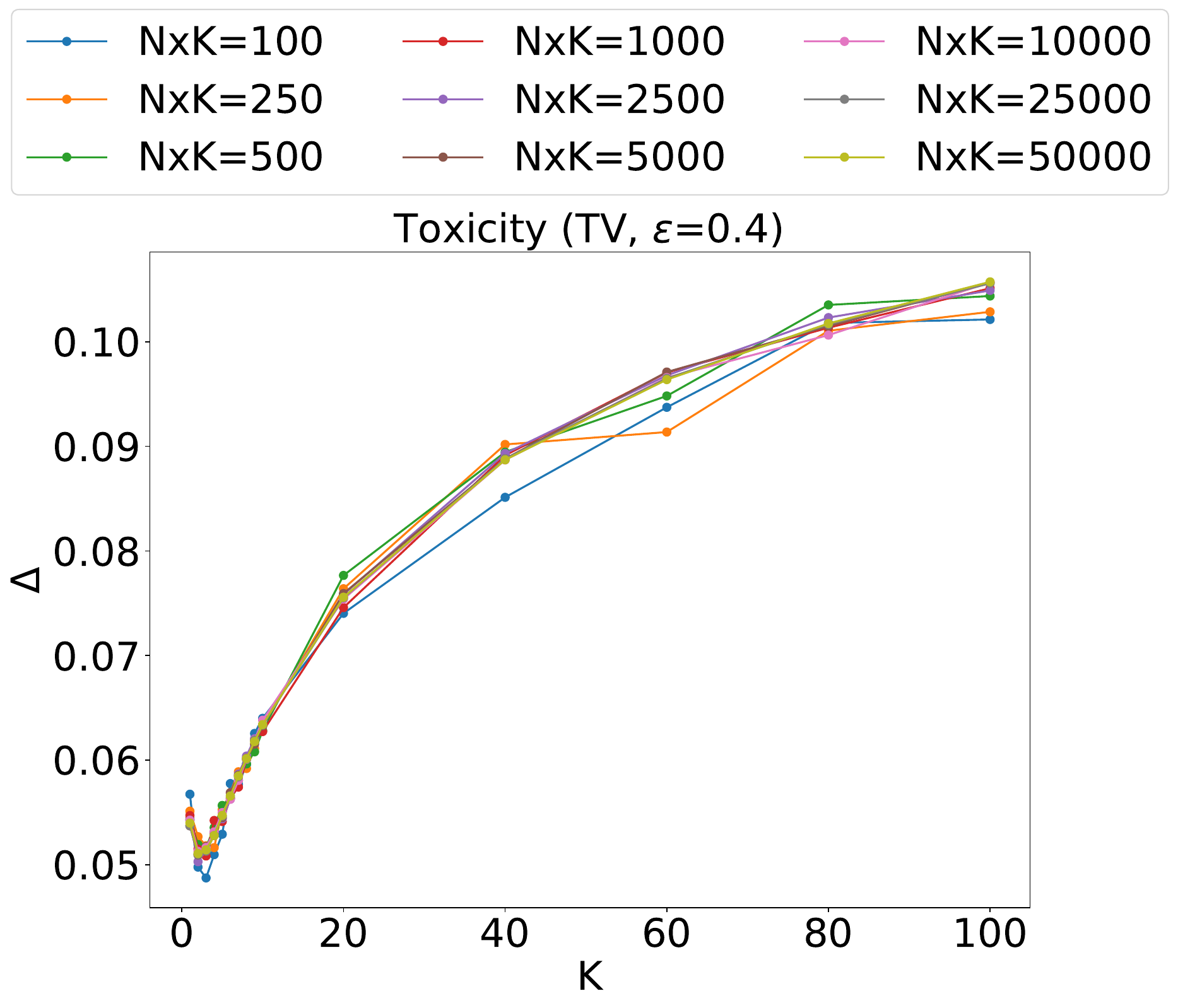}
    \caption{$\epsilon = 0.4$}
    \label{fig:toxicity_delta_MAE_e04}
  \end{subfigure}
  \caption{Effect sizes ($\Delta$) for Toxicity dataset with TV as the metric}
  \label{fig:toxicity_delta_MAE}
\end{figure*}

\begin{figure*}
  \centering
  \begin{subfigure}[b]{0.24\linewidth}
    \centering
    \includegraphics[width=\linewidth]{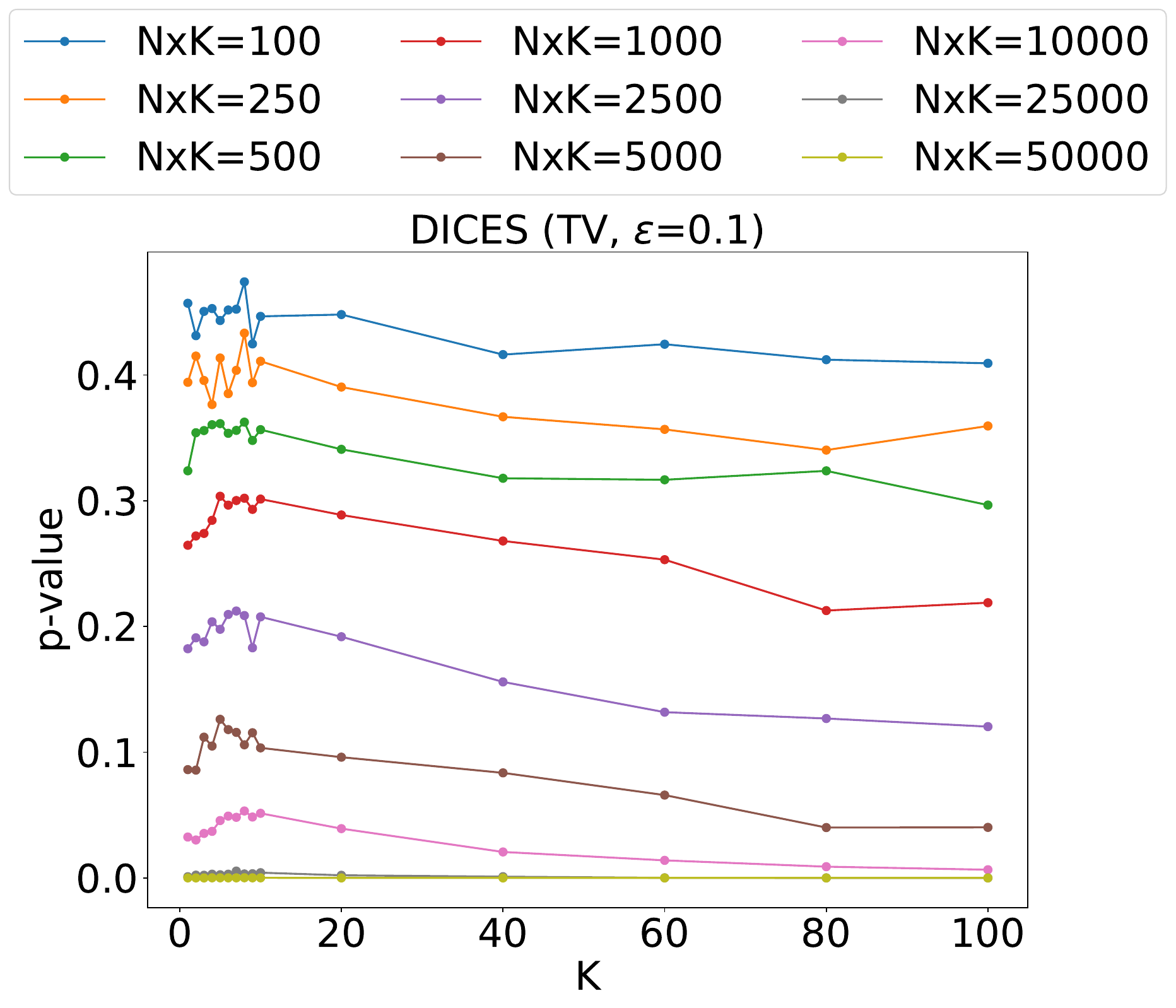}
    \caption{$\epsilon = 0.1$}
    \label{fig:dices_MAE_e01}
  \end{subfigure} \hfill
  \begin{subfigure}[b]{0.24\linewidth}
    \centering
    \includegraphics[width=\linewidth]{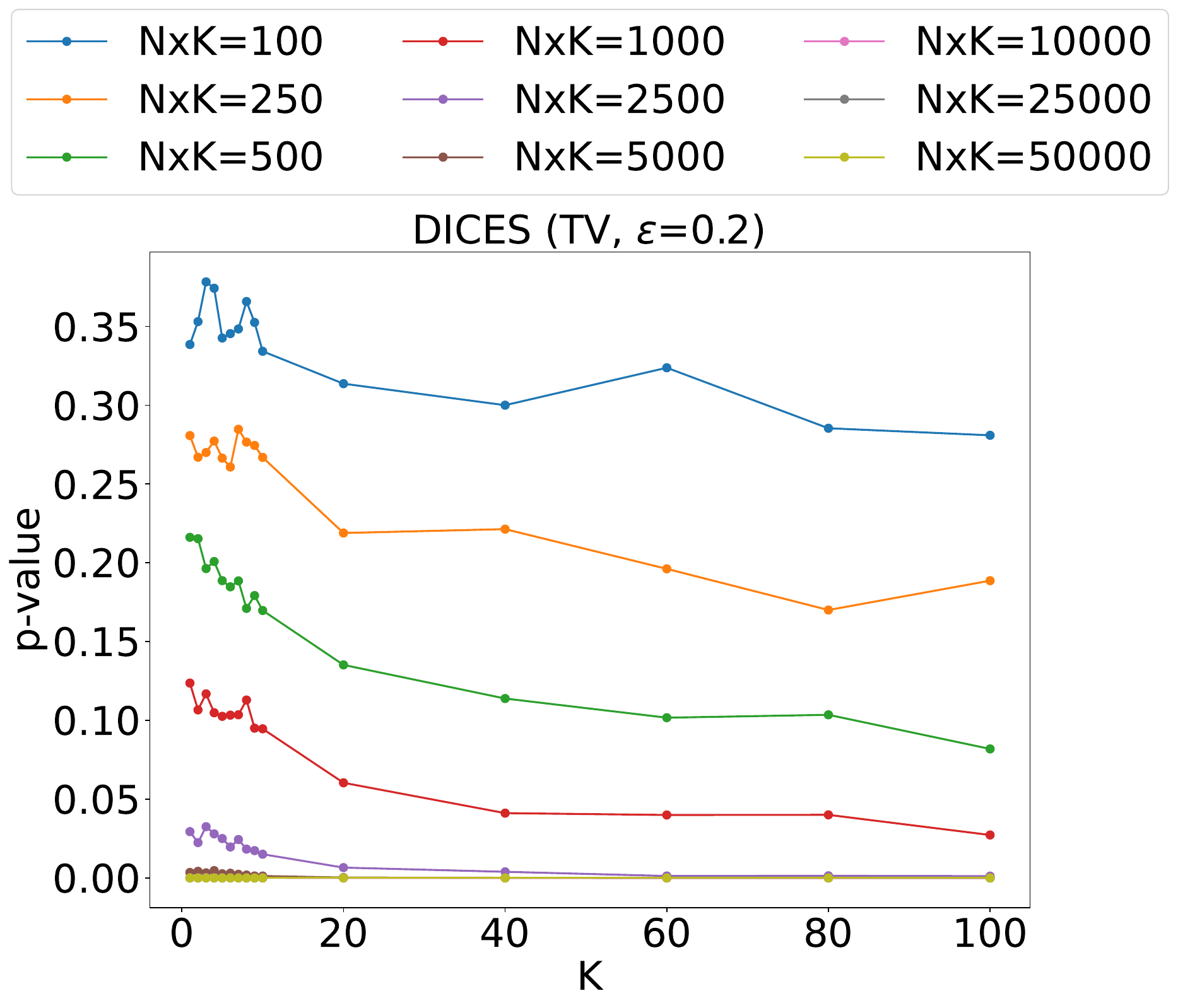}
    \caption{$\epsilon = 0.2$}
    \label{fig:dices_MAE_e02}
  \end{subfigure} \hfill
  \begin{subfigure}[b]{0.24\linewidth}
    \centering
    \includegraphics[width=\linewidth]{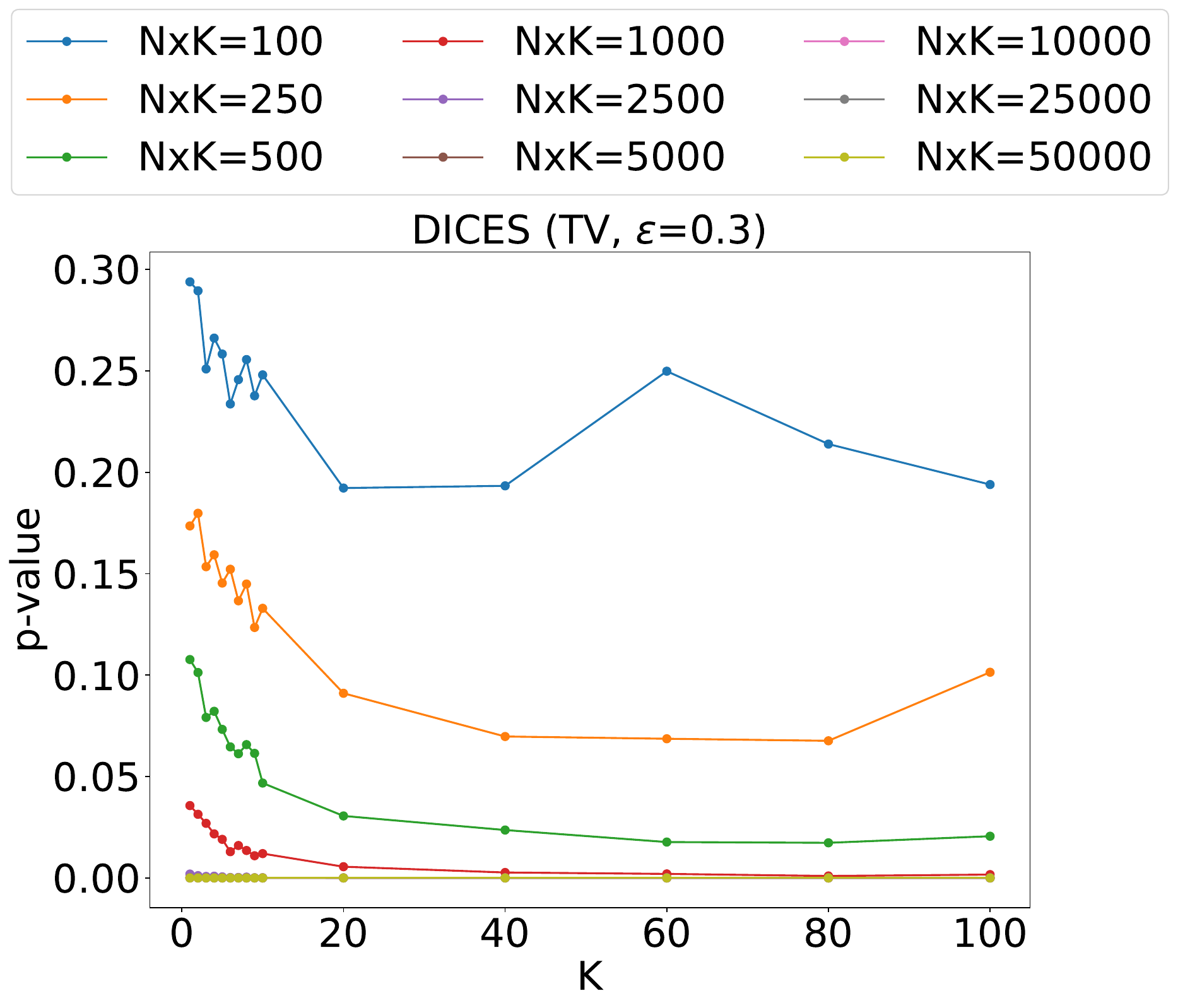}
    \caption{$\epsilon = 0.3$}
    \label{fig:dices_MAE_e03}
  \end{subfigure} \hfill
  \begin{subfigure}[b]{0.24\linewidth}
    \centering
    \includegraphics[width=\linewidth]{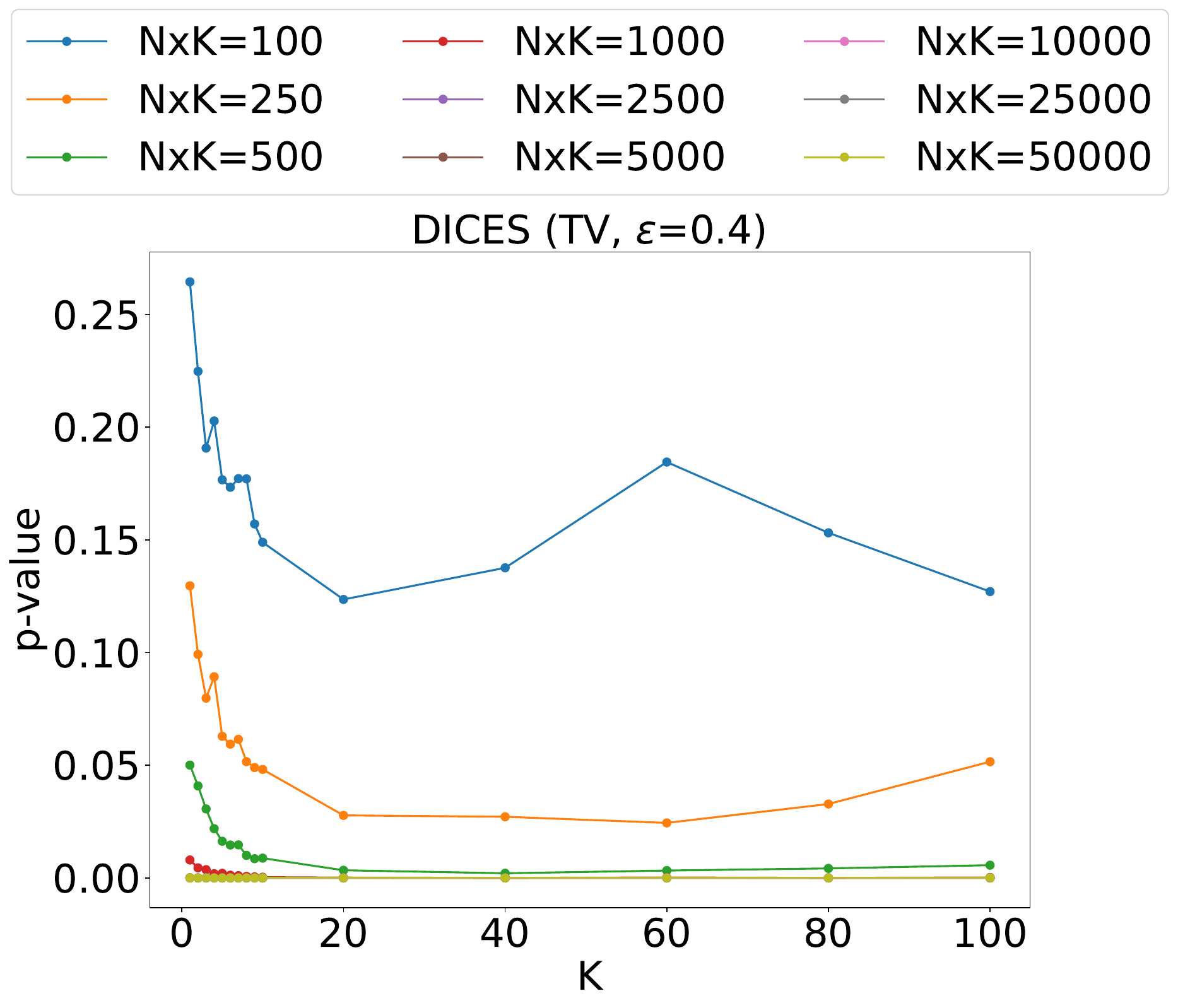}
    \caption{$\epsilon = 0.4$}
    \label{fig:dices_MAE_e04}
  \end{subfigure}
  \caption{P-value plots for DICES dataset with TV as the metric}
  \label{fig:dices_MAE}
\end{figure*}

\begin{figure*}
  \centering
  \begin{subfigure}[b]{0.24\linewidth}
    \centering
    \includegraphics[width=\linewidth]{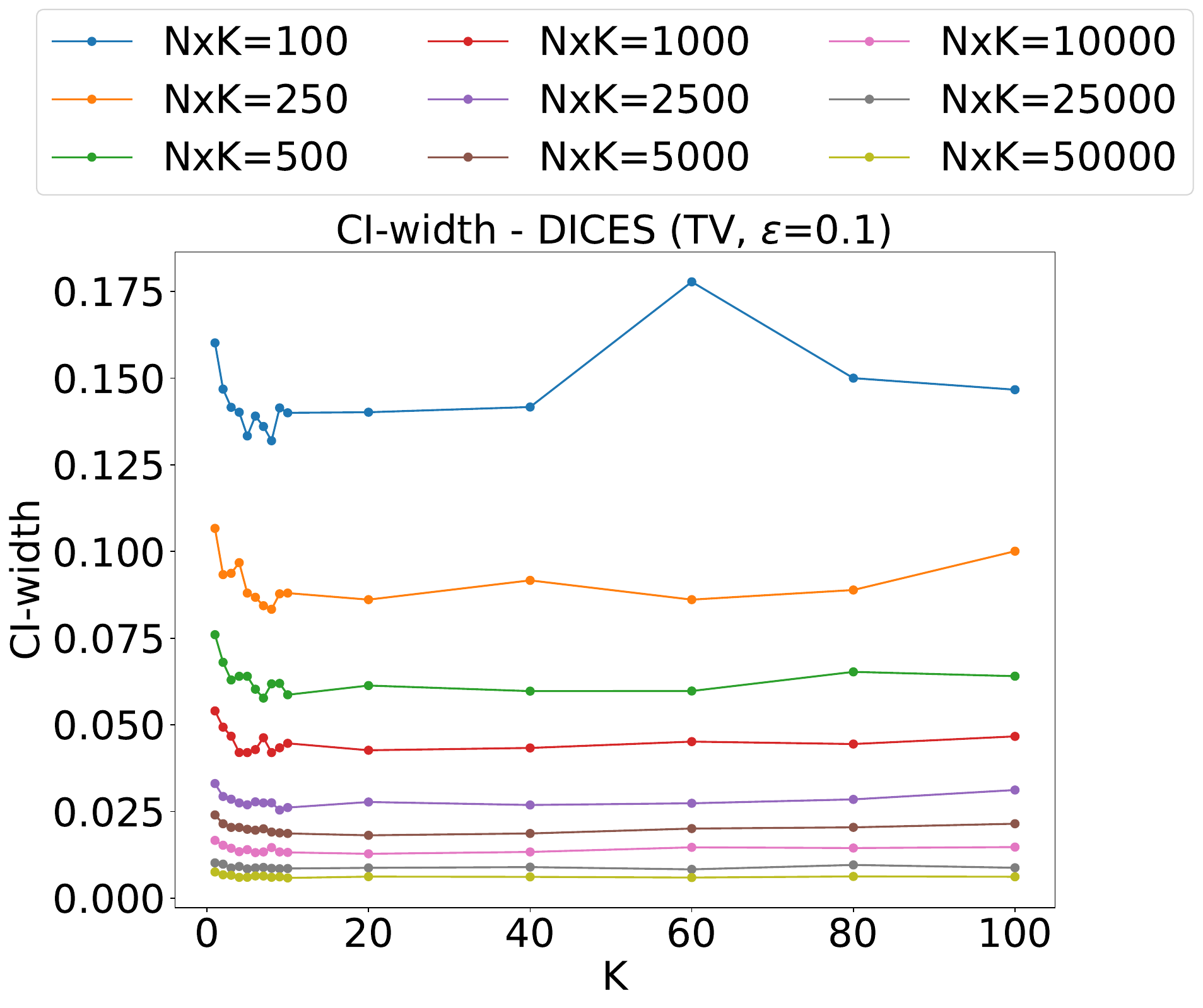}
    \caption{$\epsilon = 0.1$}
    \label{fig:dices_ci_MAE_e01}
  \end{subfigure} \hfill
  \begin{subfigure}[b]{0.24\linewidth}
    \centering
    \includegraphics[width=\linewidth]{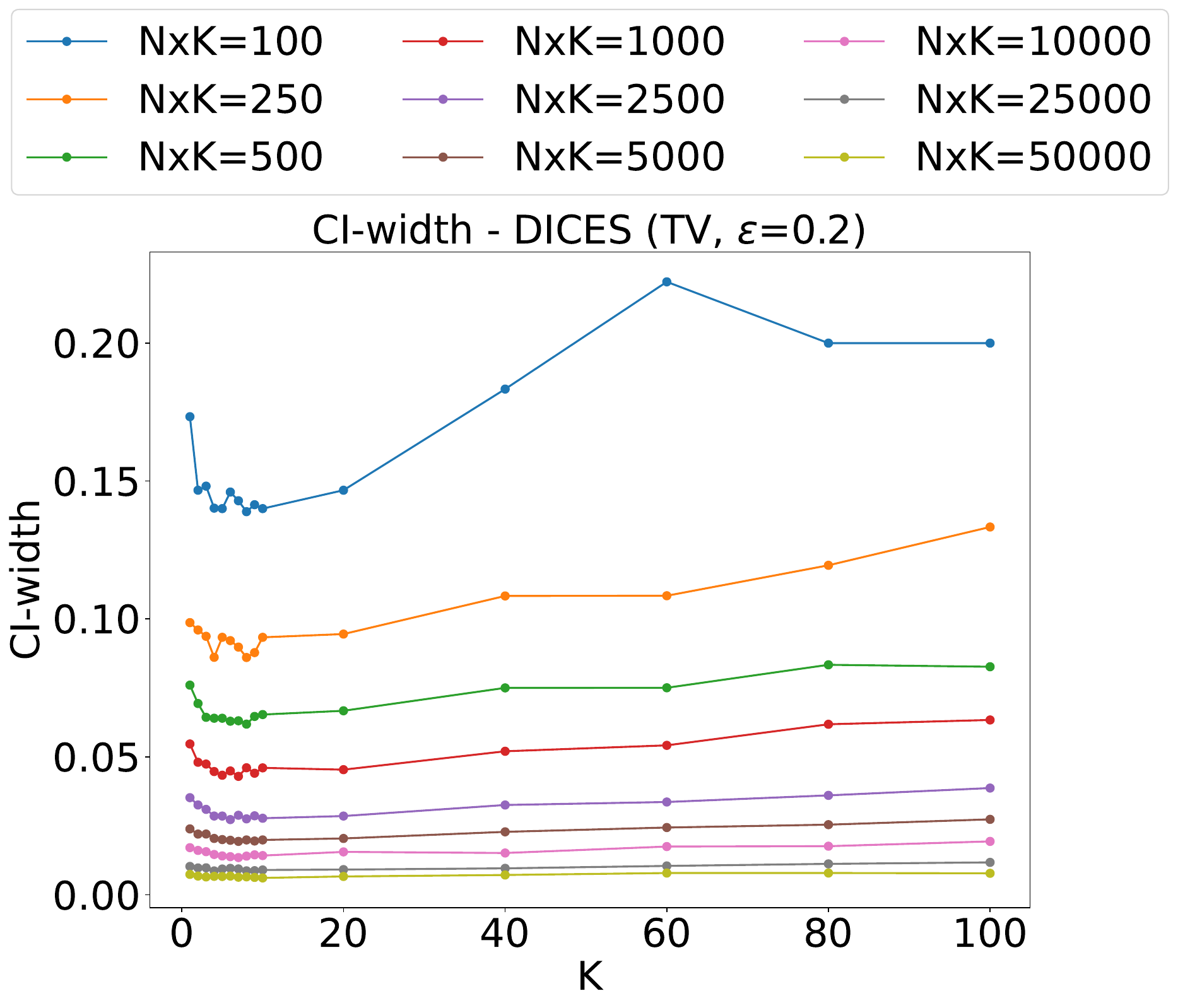}
    \caption{$\epsilon = 0.2$}
    \label{fig:dices_ci_MAE_e02}
  \end{subfigure} \hfill
  \begin{subfigure}[b]{0.24\linewidth}
    \centering
    \includegraphics[width=\linewidth]{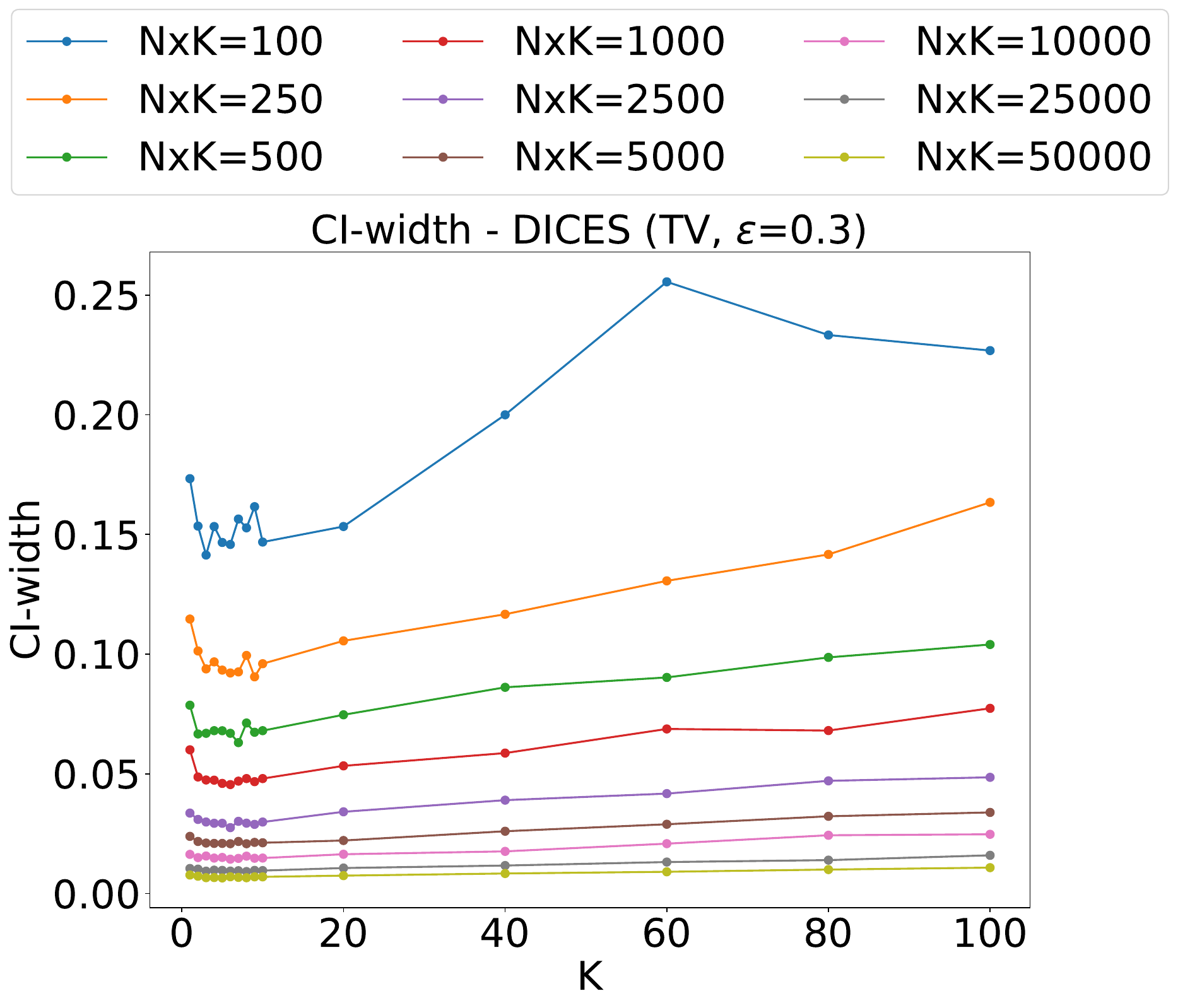}
    \caption{$\epsilon = 0.3$}
    \label{fig:dices_ci_MAE_e03}
  \end{subfigure} \hfill
  \begin{subfigure}[b]{0.24\linewidth}
    \centering
    \includegraphics[width=\linewidth]{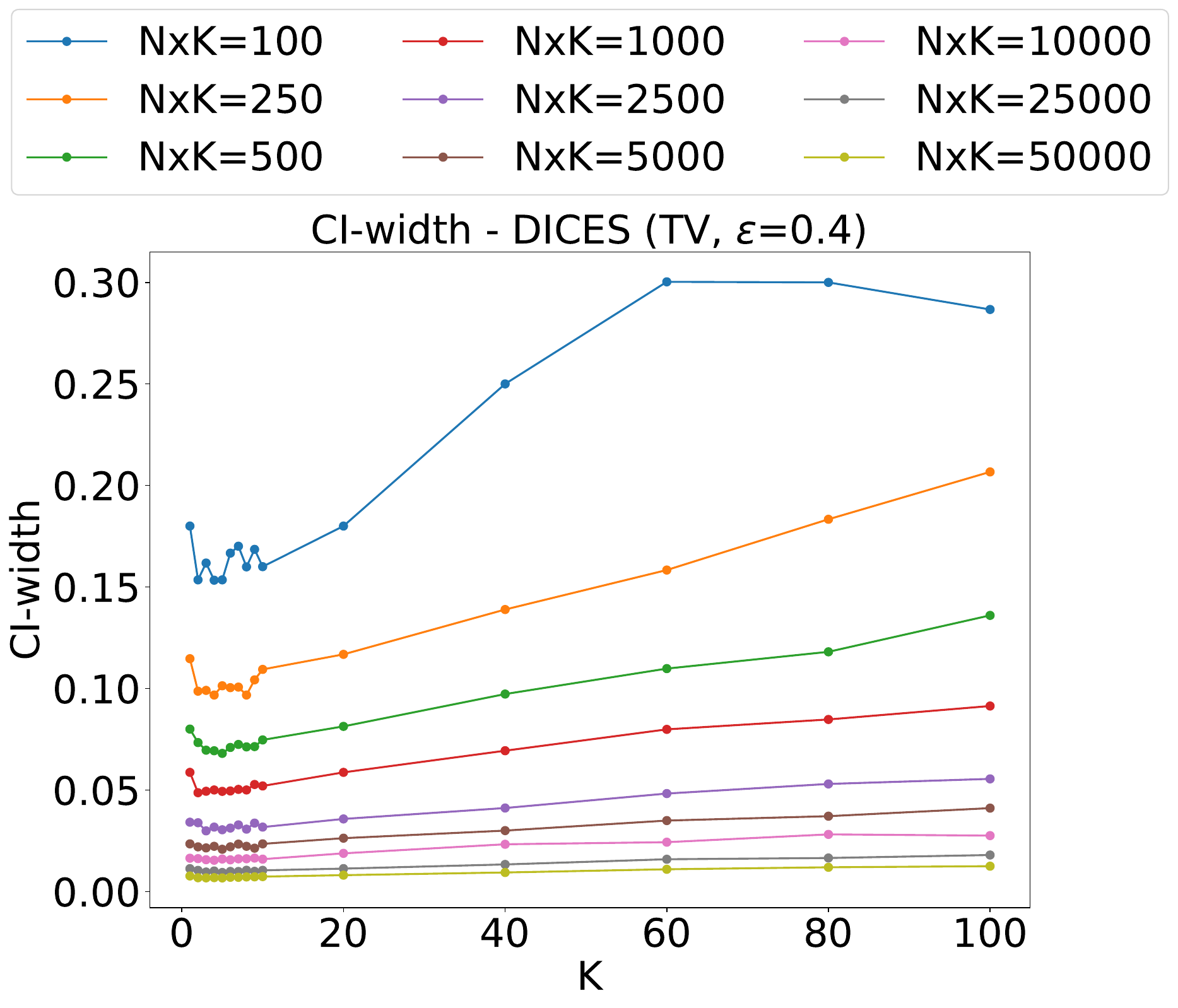}
    \caption{$\epsilon = 0.4$}
    \label{fig:dices_ci_MAE_e04}
  \end{subfigure}
  \caption{CI-width plots for DICES dataset with TV as the metric}
  \label{fig:dices_ci_MAE}
\end{figure*}

\begin{figure*}
  \centering
  \begin{subfigure}[b]{0.24\linewidth}
    \centering
    \includegraphics[width=\linewidth]{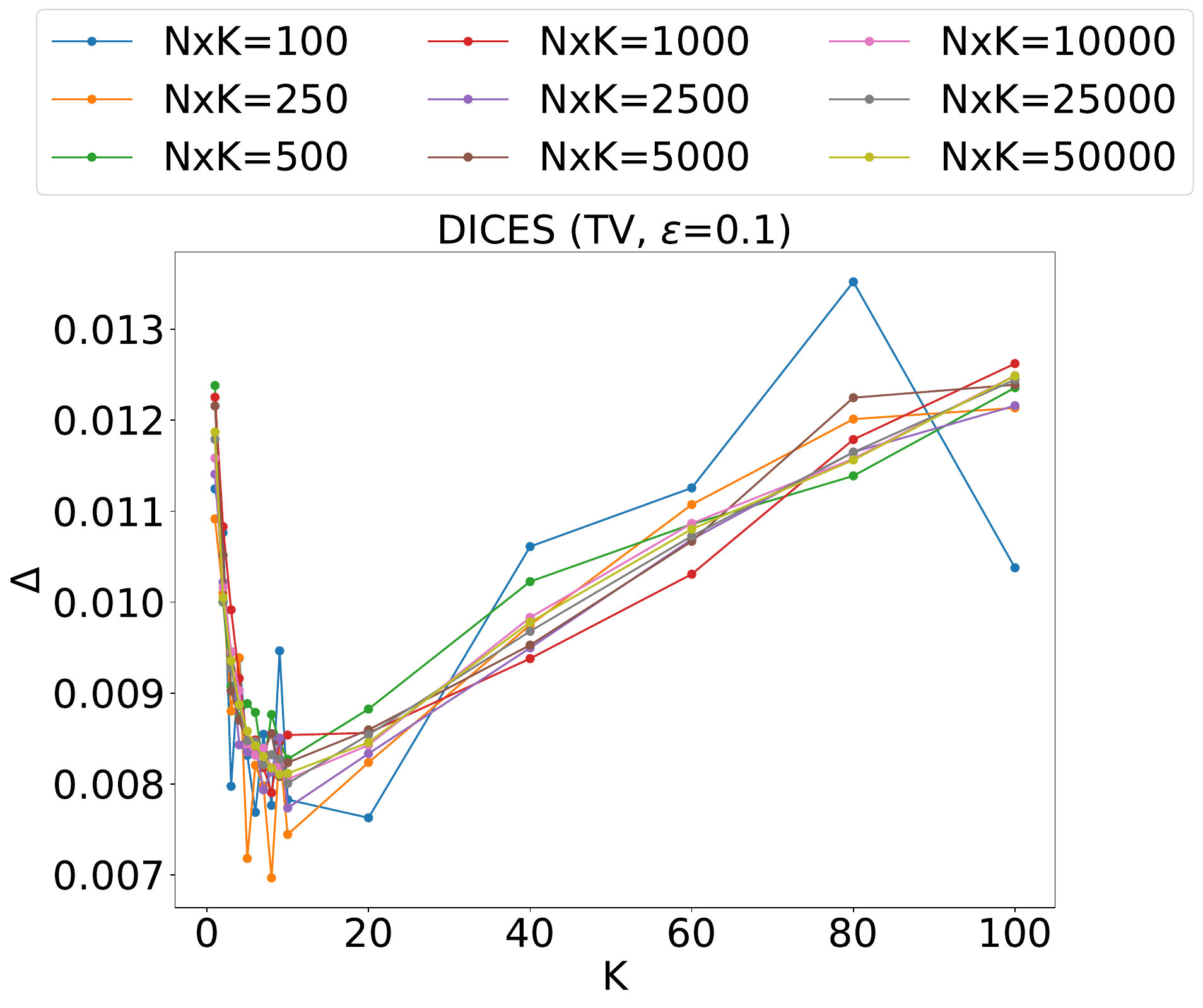}
    \caption{$\epsilon = 0.1$}
    \label{fig:dices_delta_MAE_e01}
  \end{subfigure} \hfill
  \begin{subfigure}[b]{0.24\linewidth}
    \centering
    \includegraphics[width=\linewidth]{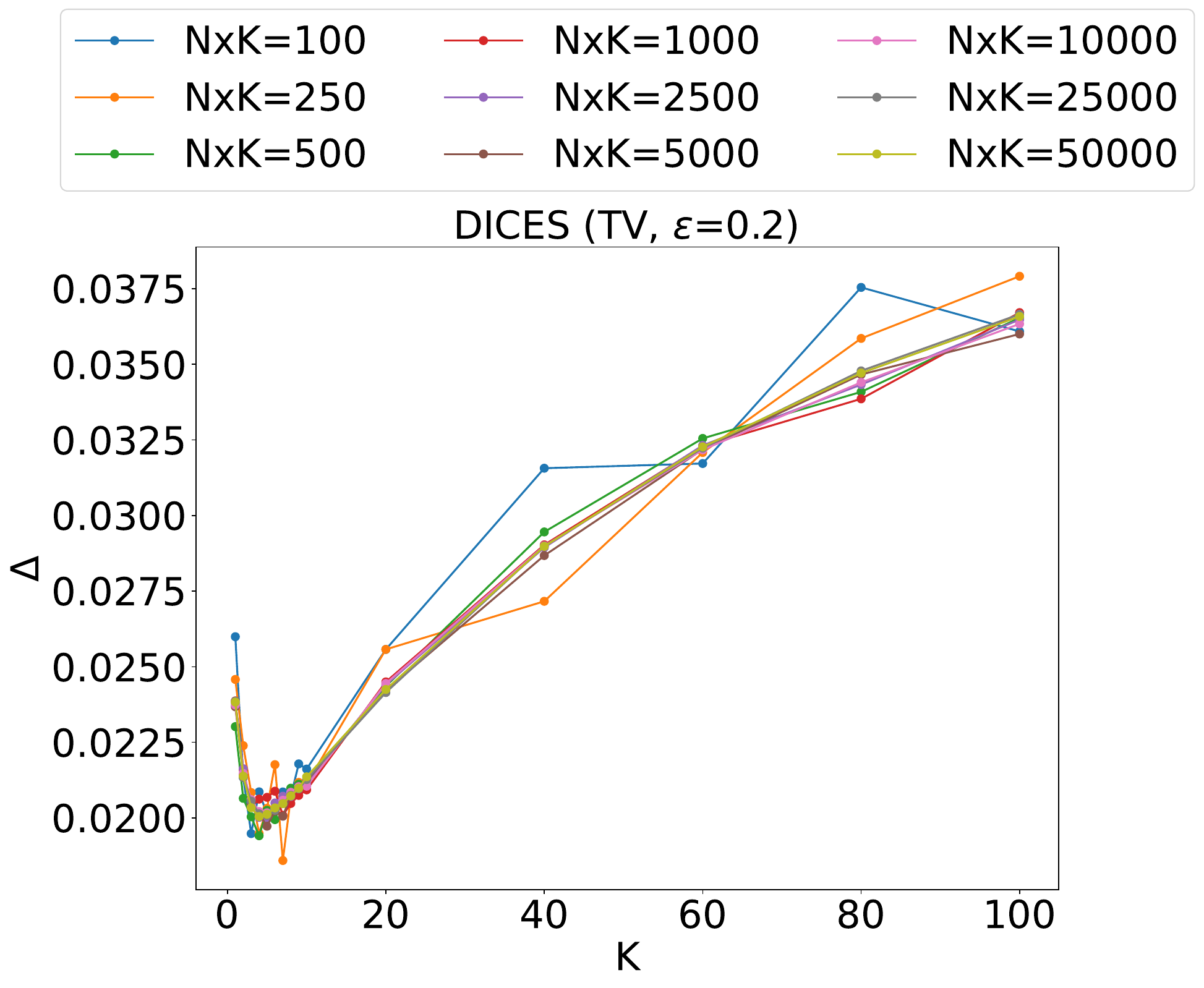}
    \caption{$\epsilon = 0.2$}
    \label{fig:dices_delta_MAE_e02}
  \end{subfigure} \hfill
  \begin{subfigure}[b]{0.24\linewidth}
    \centering
    \includegraphics[width=\linewidth]{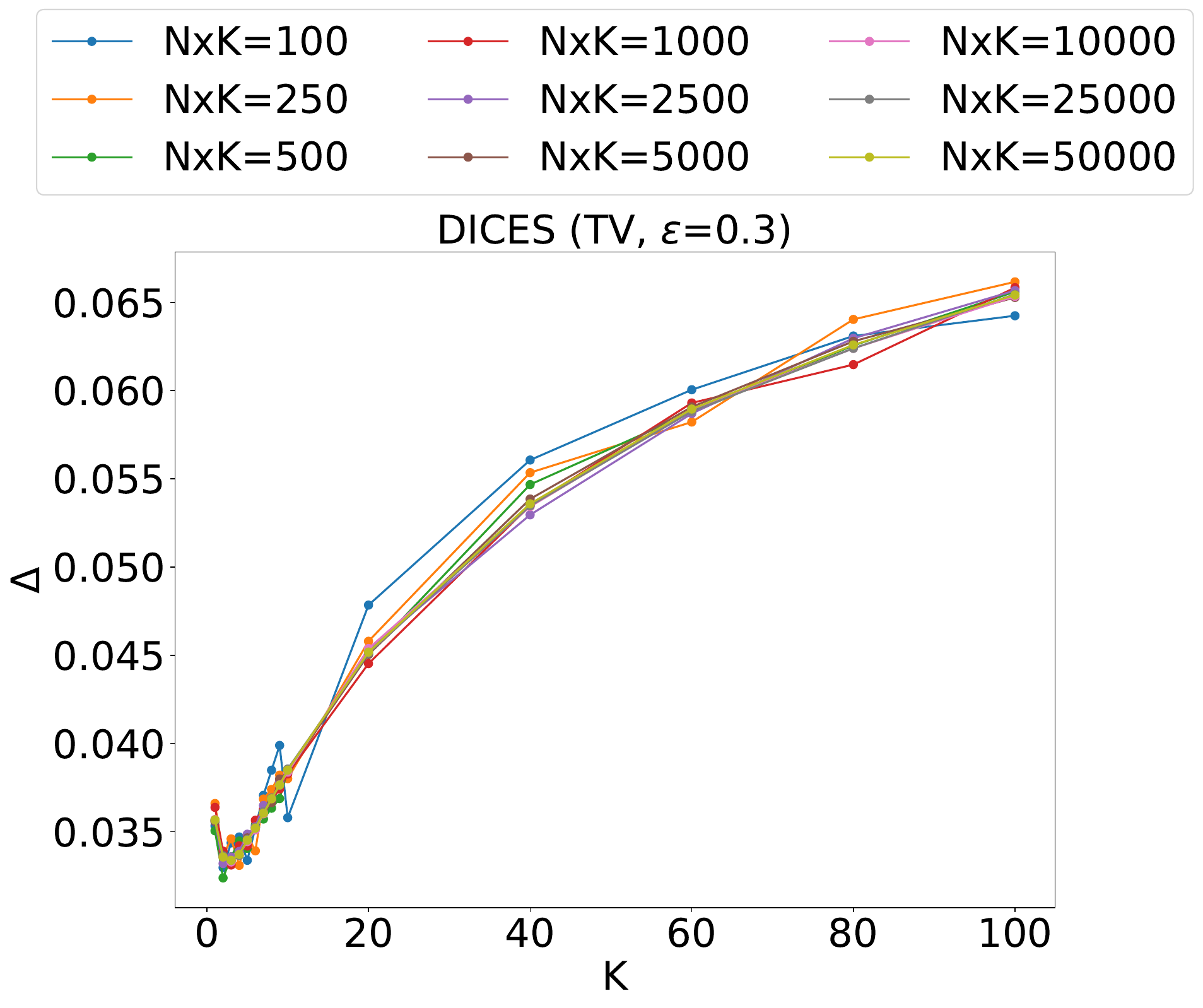}
    \caption{$\epsilon = 0.3$}
    \label{fig:dices_delta_MAE_e03}
  \end{subfigure} \hfill
  \begin{subfigure}[b]{0.24\linewidth}
    \centering
    \includegraphics[width=\linewidth]{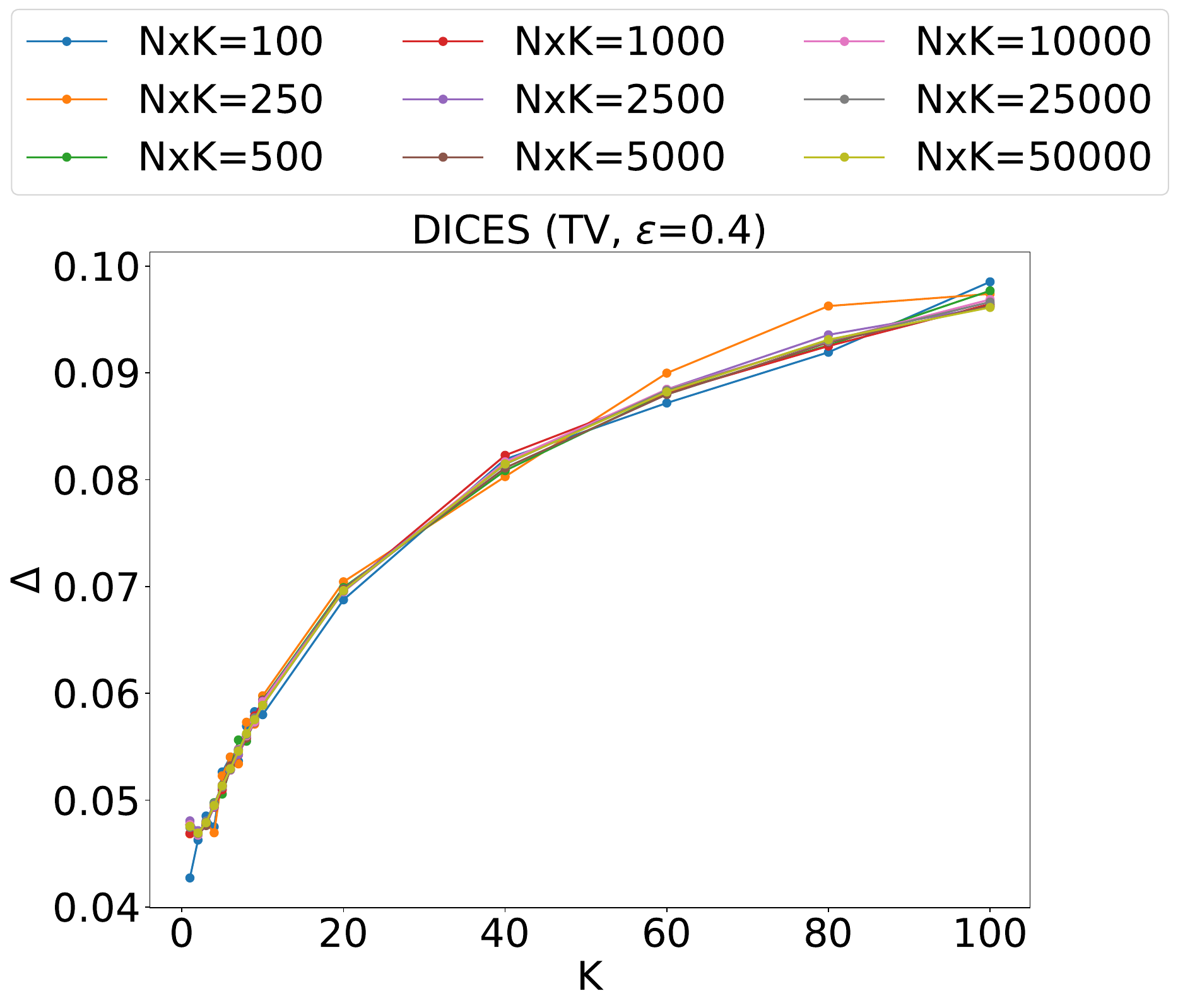}
    \caption{$\epsilon = 0.4$}
    \label{fig:dices_delta_MAE_e04}
  \end{subfigure}
  \caption{Effect sizes ($\Delta$) for DICES dataset with TV as the metric}
  \label{fig:dices_delta_MAE}
\end{figure*}

\begin{figure*}
  \centering
  \begin{subfigure}[b]{0.24\linewidth}
    \centering
    \includegraphics[width=\linewidth]{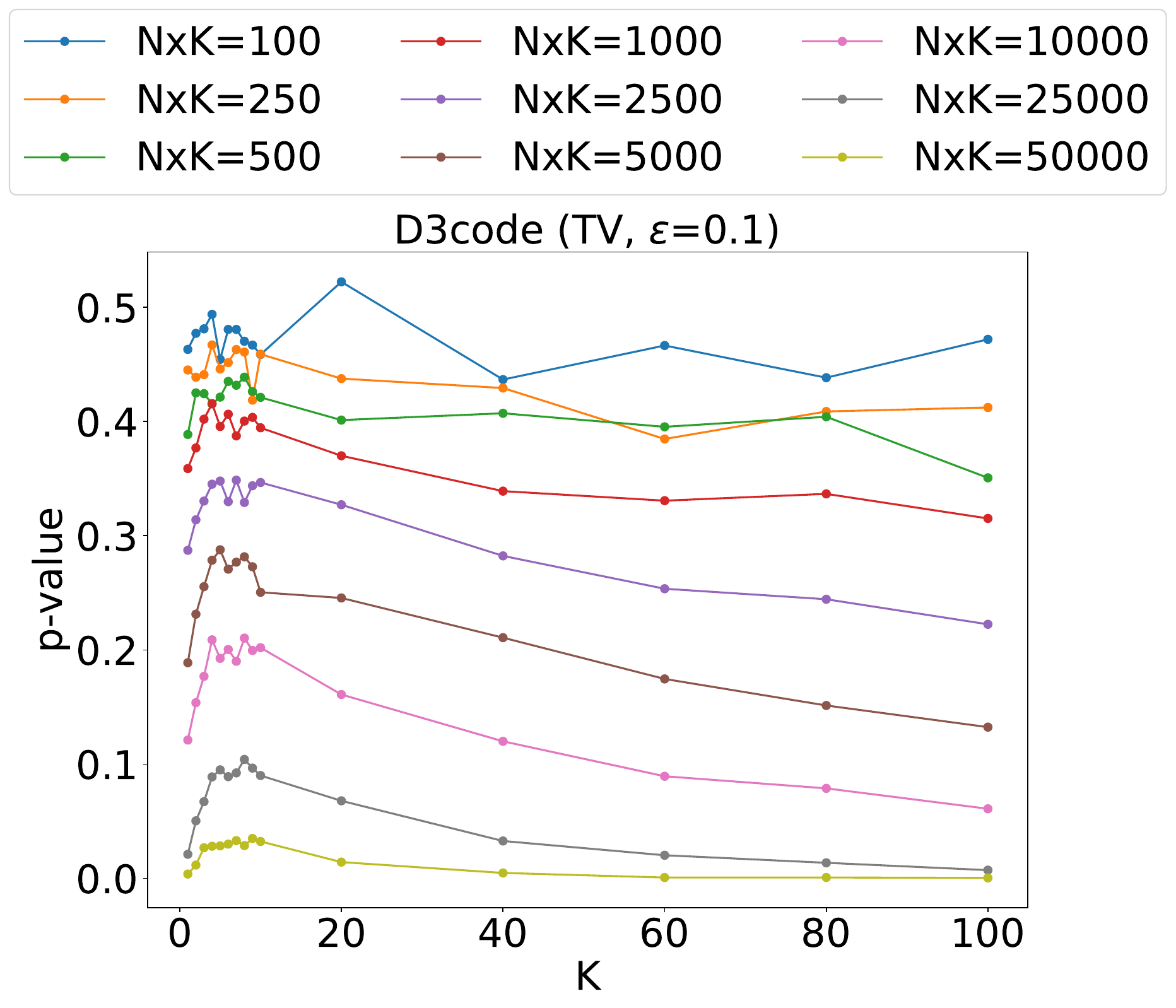}
    \caption{$\epsilon = 0.1$}
    \label{fig:d3code_MAE_e01}
  \end{subfigure} \hfill
  \begin{subfigure}[b]{0.24\linewidth}
    \centering
    \includegraphics[width=\linewidth]{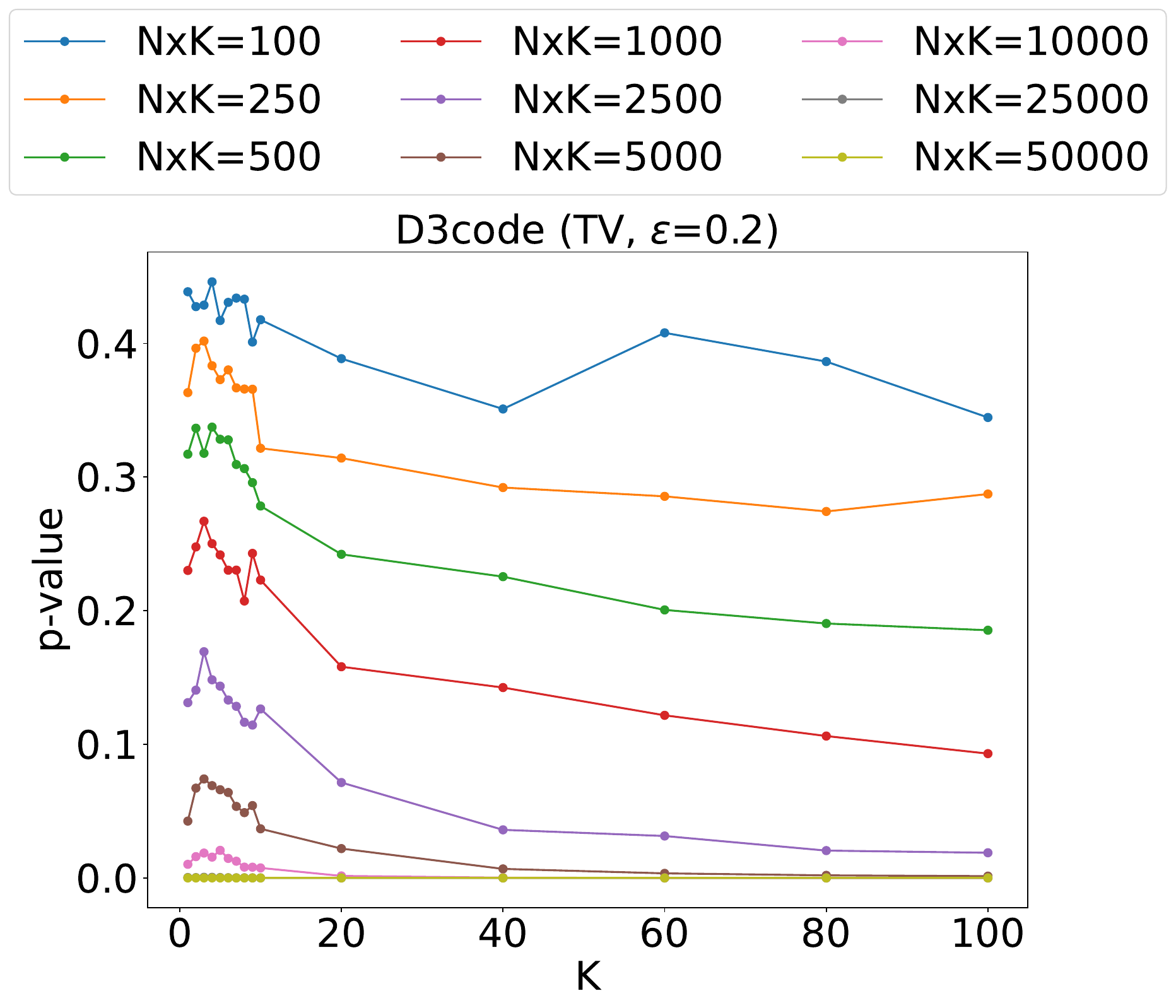}
    \caption{$\epsilon = 0.2$}
    \label{fig:d3code_MAE_e02}
  \end{subfigure} \hfill
  \begin{subfigure}[b]{0.24\linewidth}
    \centering
    \includegraphics[width=\linewidth]{figures/K100/pvals_plots/D3code/D3code_p_vals_MAE_K_100_e_0.3.pdf}
    \caption{$\epsilon = 0.3$}
    \label{fig:d3code_MAE_e03}
  \end{subfigure} \hfill
  \begin{subfigure}[b]{0.24\linewidth}
    \centering
    \includegraphics[width=\linewidth]{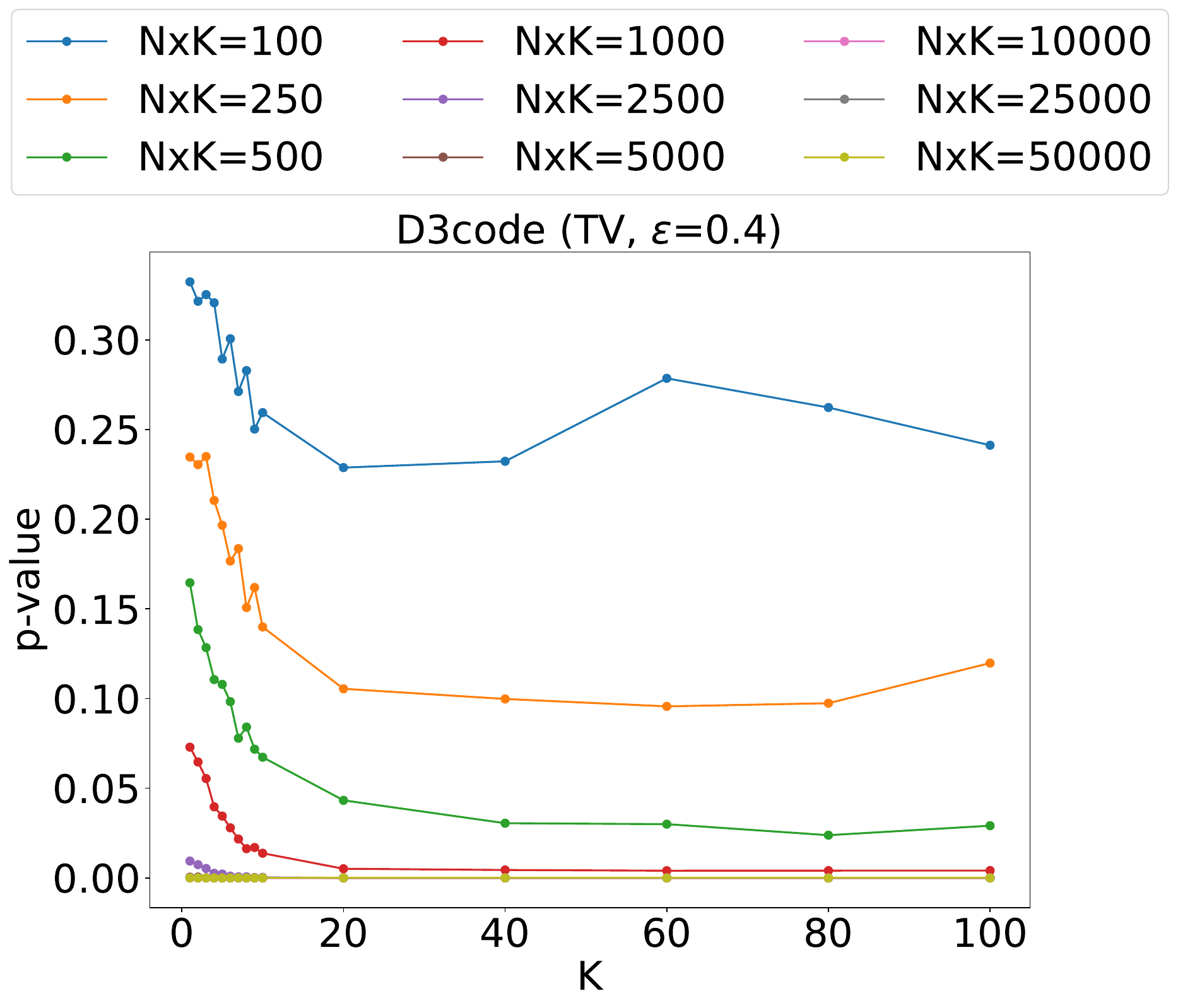}
    \caption{$\epsilon = 0.4$}
    \label{fig:d3code_MAE_e04}
  \end{subfigure}
  \caption{P-value plots for D3code dataset with TV as the metric}
  \label{fig:d3code_MAE}
\end{figure*}

\begin{figure*}
  \centering
  \begin{subfigure}[b]{0.24\linewidth}
    \centering
    \includegraphics[width=\linewidth]{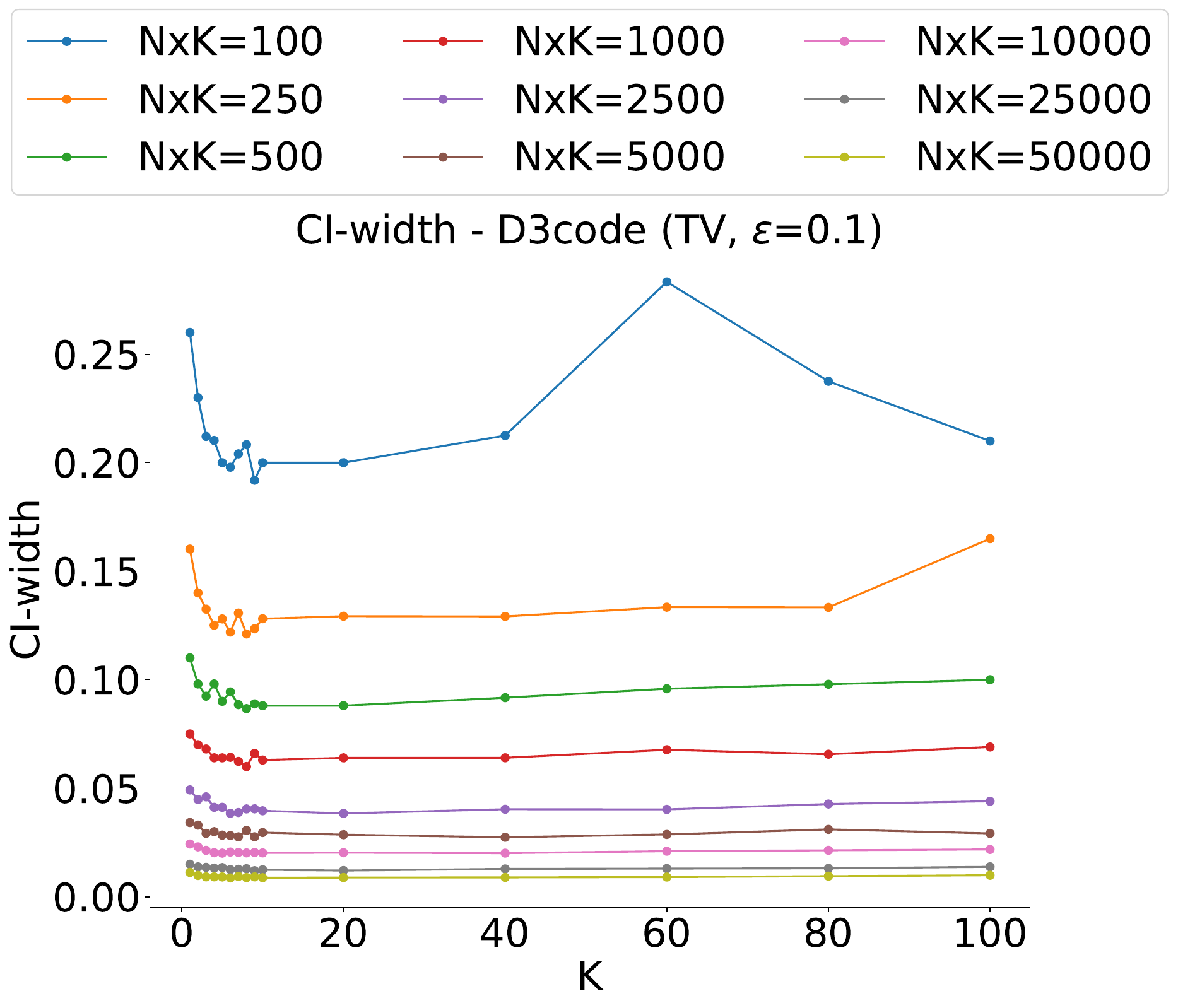}
    \caption{$\epsilon = 0.1$}
    \label{fig:d3code_ci_MAE_e01}
  \end{subfigure} \hfill
  \begin{subfigure}[b]{0.24\linewidth}
    \centering
    \includegraphics[width=\linewidth]{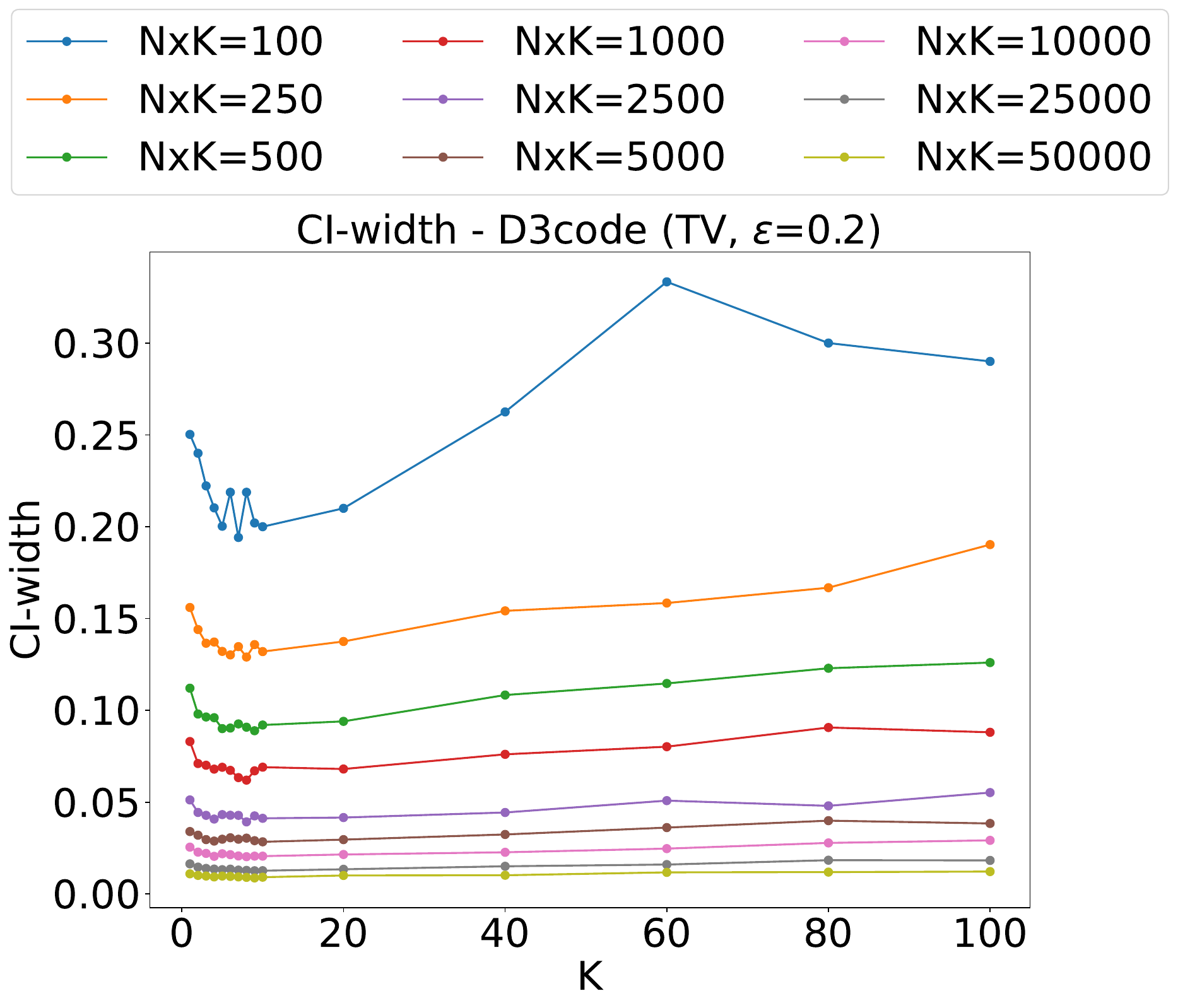}
    \caption{$\epsilon = 0.2$}
    \label{fig:d3code_ci_MAE_e02}
  \end{subfigure} \hfill
  \begin{subfigure}[b]{0.24\linewidth}
    \centering
    \includegraphics[width=\linewidth]{figures/K100/ci_plots/D3code/D3code_CI_width_MAE_K_100_e_0.3.pdf}
    \caption{$\epsilon = 0.3$}
    \label{fig:d3code_ci_MAE_e03}
  \end{subfigure} \hfill
  \begin{subfigure}[b]{0.24\linewidth}
    \centering
    \includegraphics[width=\linewidth]{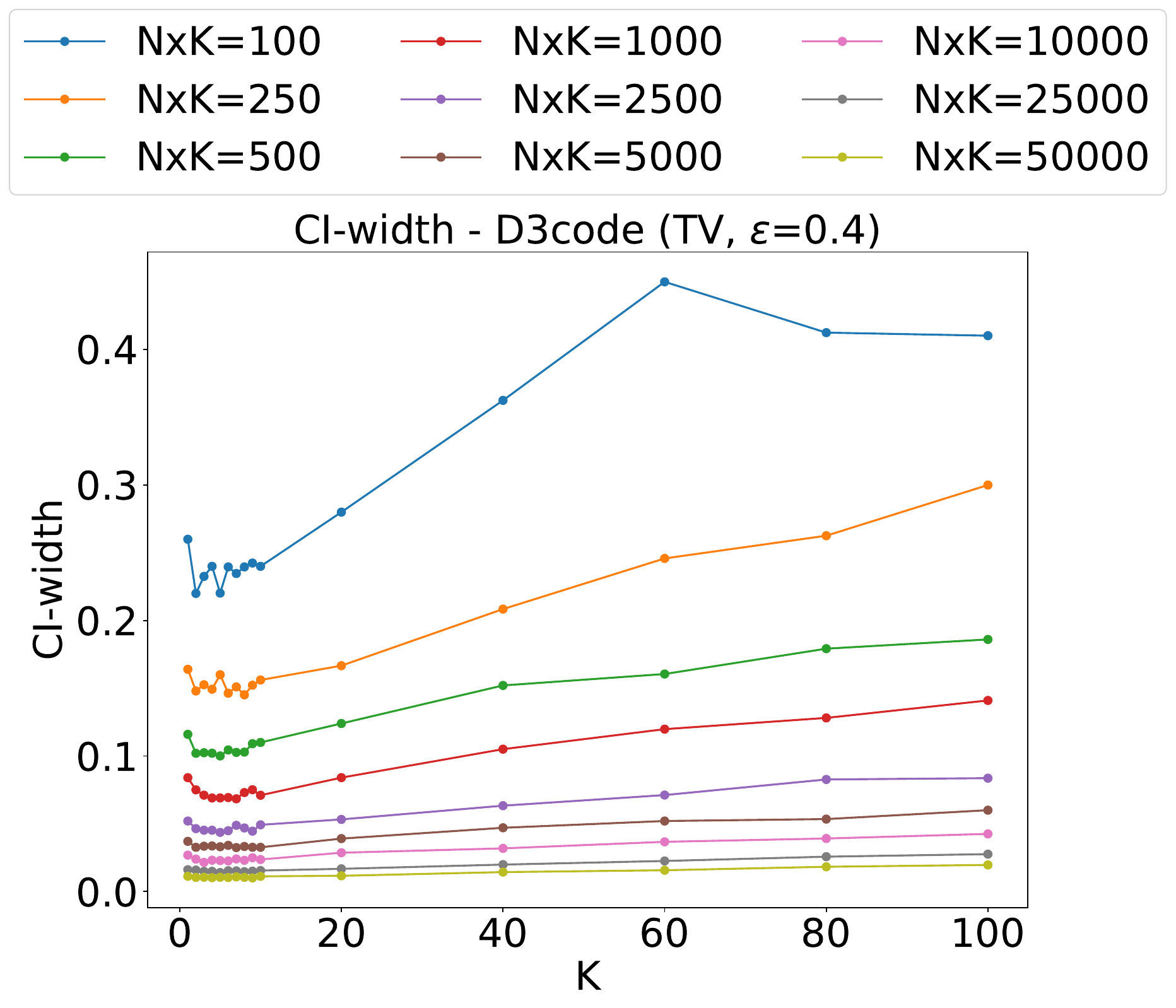}
    \caption{$\epsilon = 0.4$}
    \label{fig:d3code_ci_MAE_e04}
  \end{subfigure}
  \caption{CI-width plots for D3code dataset with TV as the metric}
  \label{fig:d3code_ci_MAE}
\end{figure*}

\begin{figure*}
  \centering
  \begin{subfigure}[b]{0.24\linewidth}
    \centering
    \includegraphics[width=\linewidth]{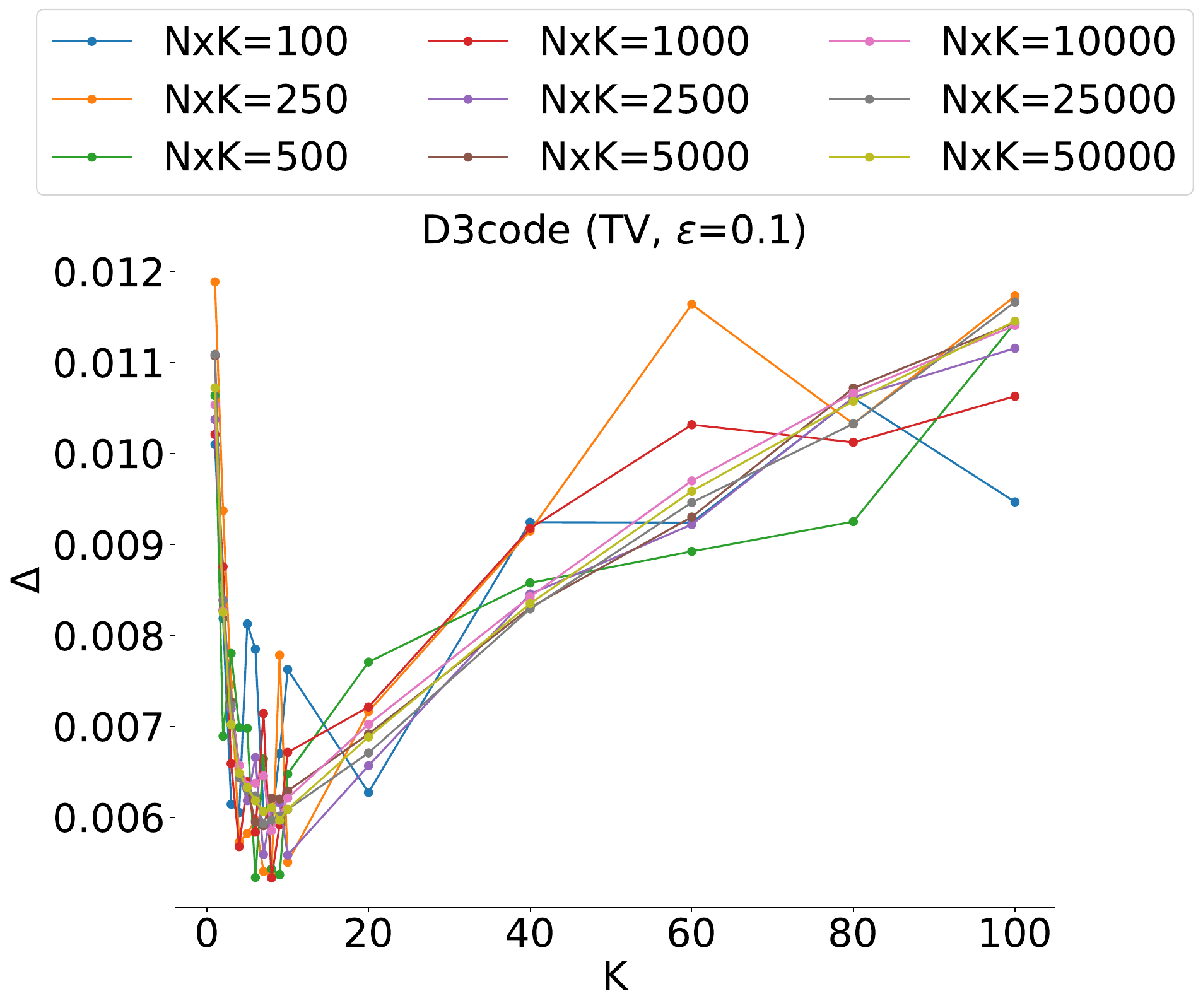}
    \caption{$\epsilon = 0.1$}
    \label{fig:d3code_delta_MAE_e01}
  \end{subfigure} \hfill
  \begin{subfigure}[b]{0.24\linewidth}
    \centering
    \includegraphics[width=\linewidth]{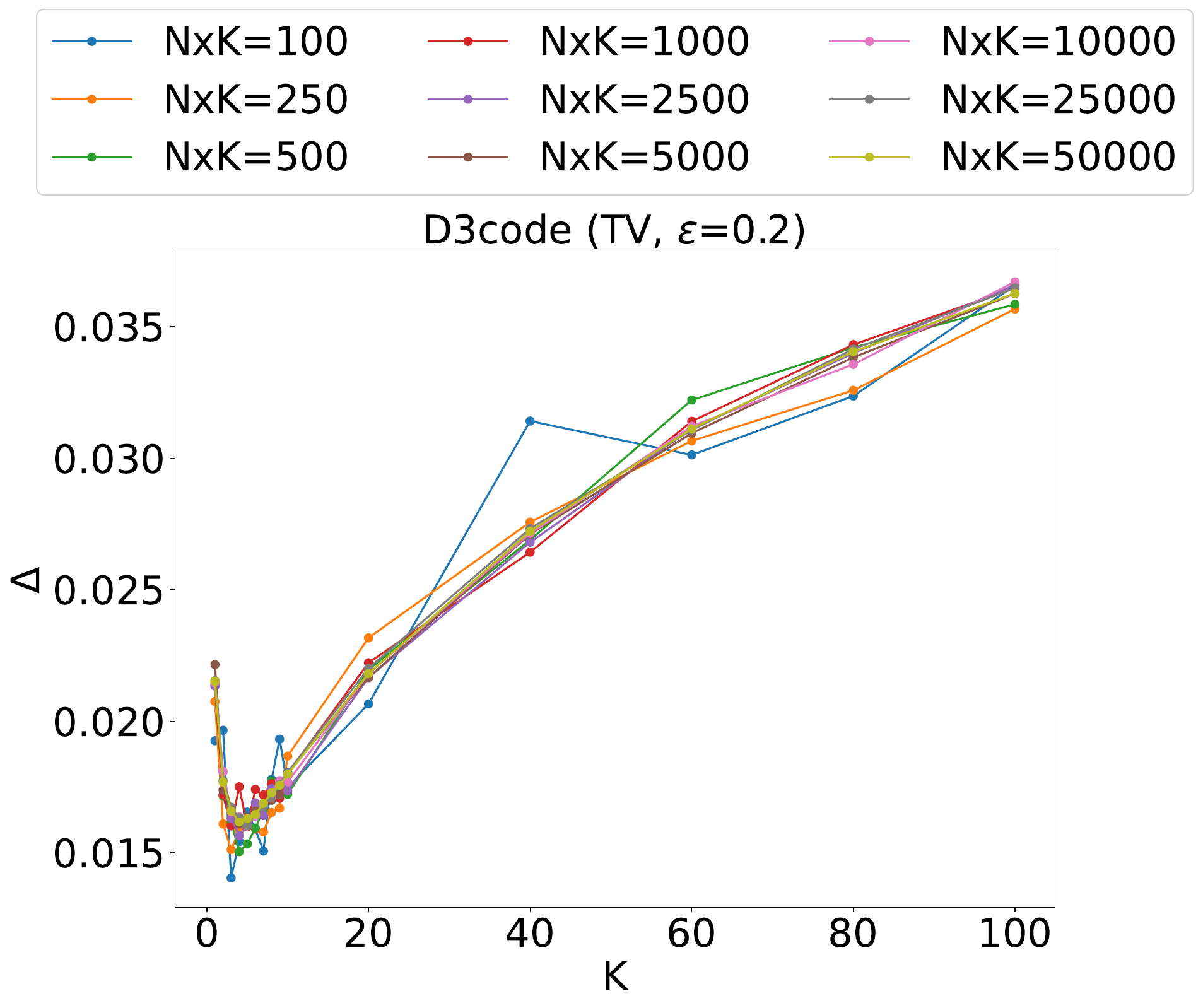}
    \caption{$\epsilon = 0.2$}
    \label{fig:d3code_delta_MAE_e02}
  \end{subfigure} \hfill
  \begin{subfigure}[b]{0.24\linewidth}
    \centering
    \includegraphics[width=\linewidth]{figures/K100/delta_plots/D3code/D3code_delta_MAE_K_100_e_0.3.pdf}
    \caption{$\epsilon = 0.3$}
    \label{fig:d3code_delta_MAE_e03}
  \end{subfigure} \hfill
  \begin{subfigure}[b]{0.24\linewidth}
    \centering
    \includegraphics[width=\linewidth]{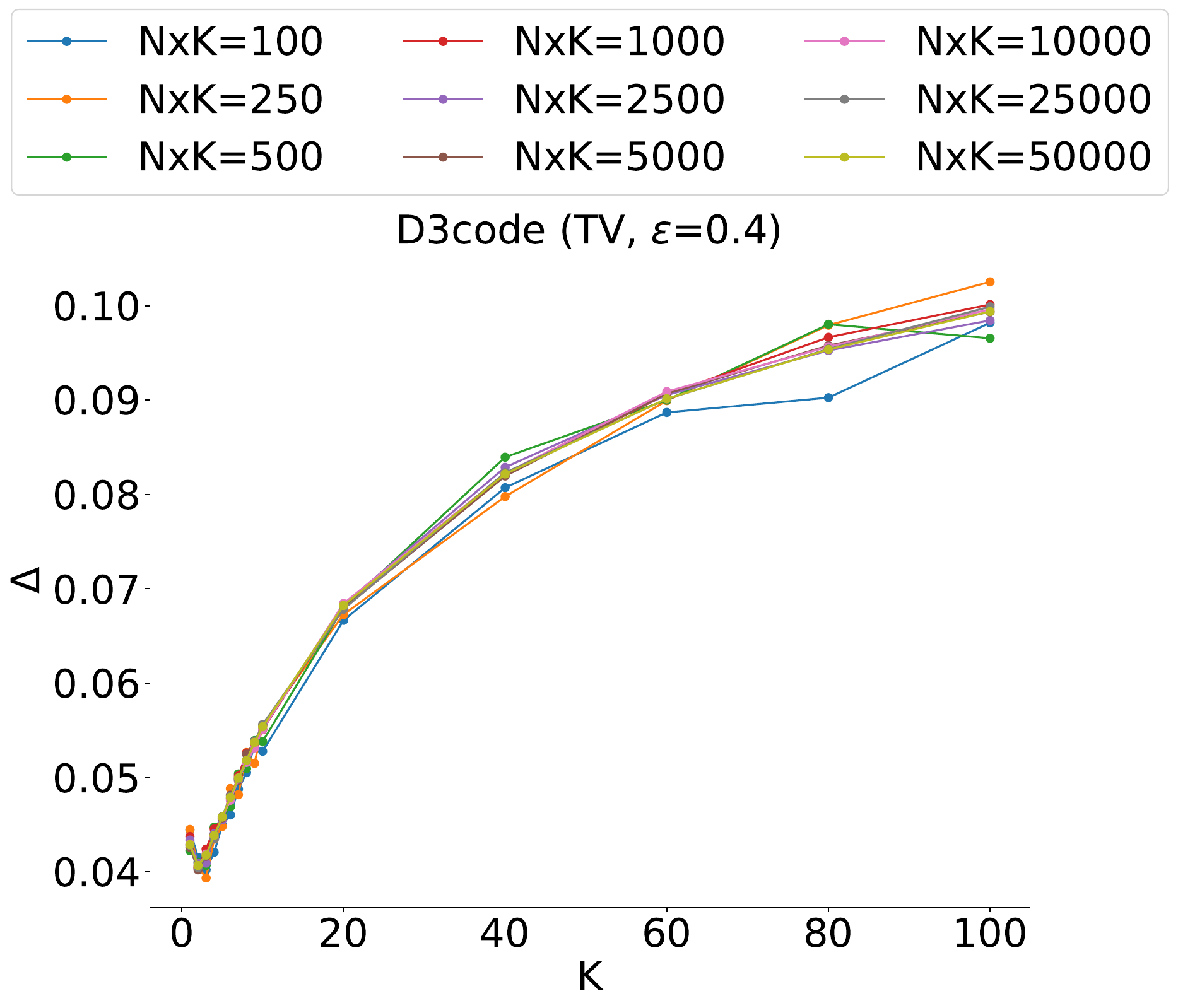}
    \caption{$\epsilon = 0.4$}
    \label{fig:d3code_delta_MAE_e04}
  \end{subfigure}
  \caption{Effect sizes ($\Delta$) for D3code dataset with TV as the metric}
  \label{fig:d3code_delta_MAE}
\end{figure*}

\begin{figure*}
  \centering
  \begin{subfigure}[b]{0.24\linewidth}
    \centering
    \includegraphics[width=\linewidth]{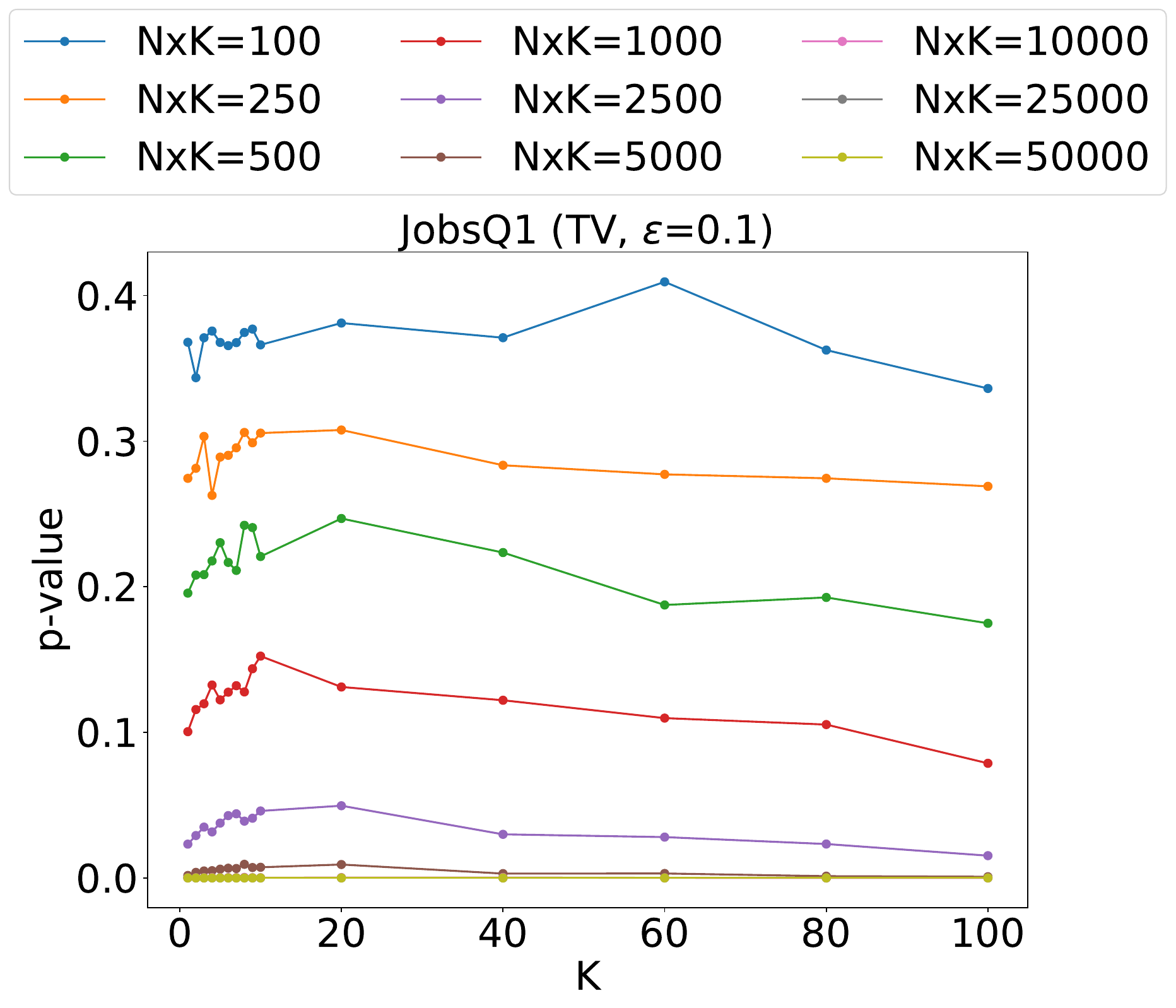}
    \caption{$\epsilon = 0.1$}
    \label{fig:jobsQ1_MAE_e01}
  \end{subfigure} \hfill
  \begin{subfigure}[b]{0.24\linewidth}
    \centering
    \includegraphics[width=\linewidth]{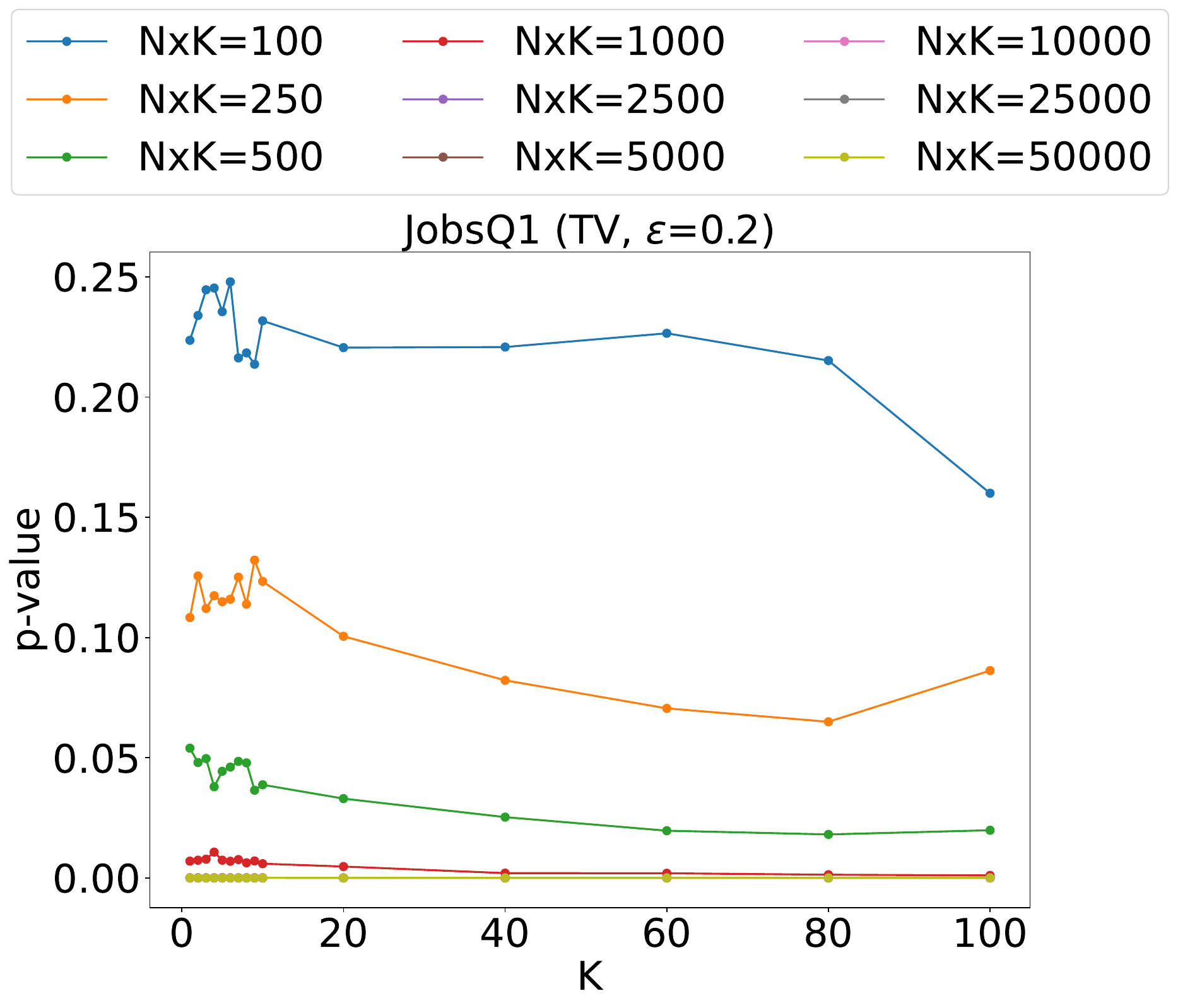}
    \caption{$\epsilon = 0.2$}
    \label{fig:jobsQ1_MAE_e02}
  \end{subfigure} \hfill
  \begin{subfigure}[b]{0.24\linewidth}
    \centering
    \includegraphics[width=\linewidth]{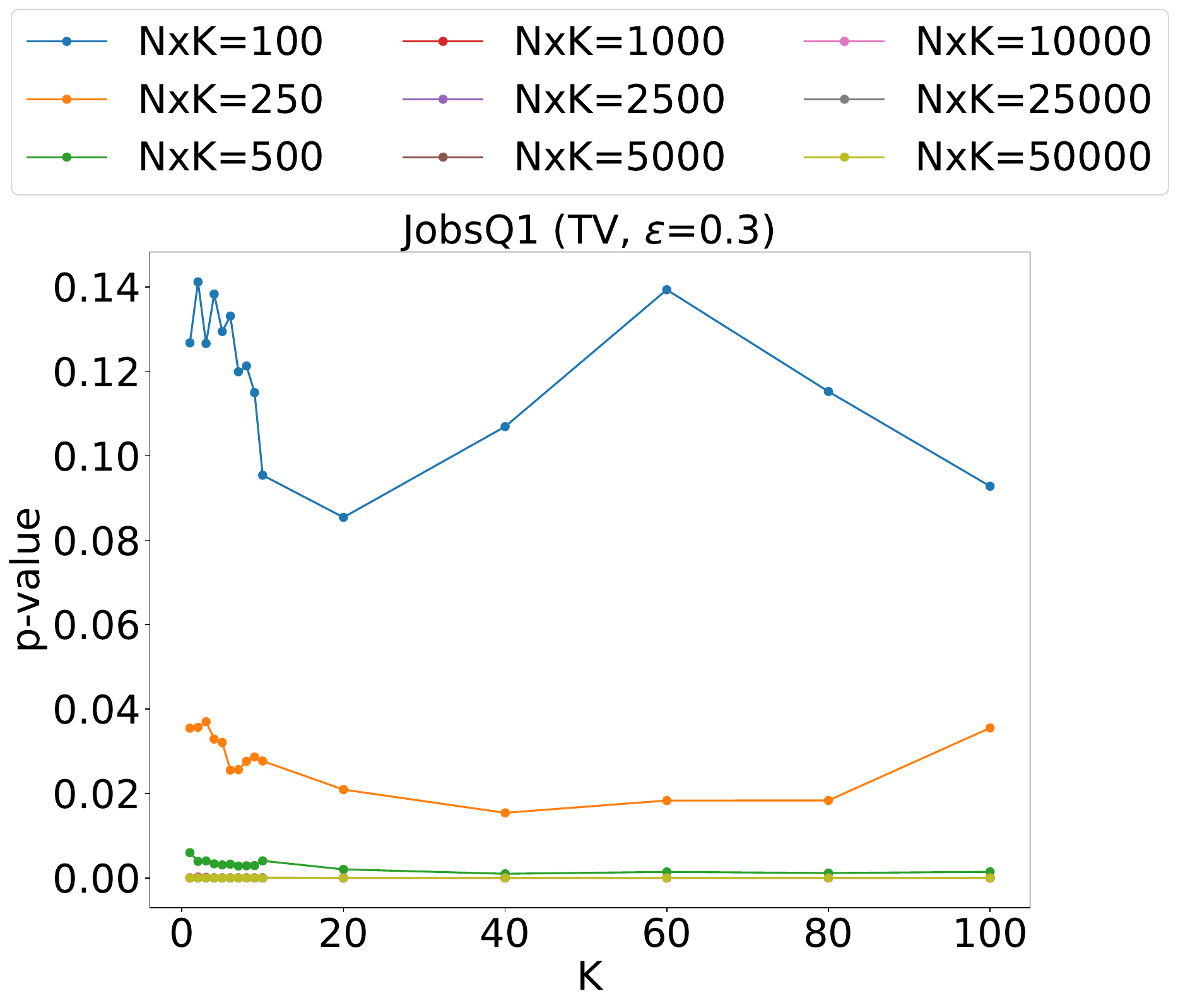}
    \caption{$\epsilon = 0.3$}
    \label{fig:jobsQ1_MAE_e03}
  \end{subfigure} \hfill
  \begin{subfigure}[b]{0.24\linewidth}
    \centering
    \includegraphics[width=\linewidth]{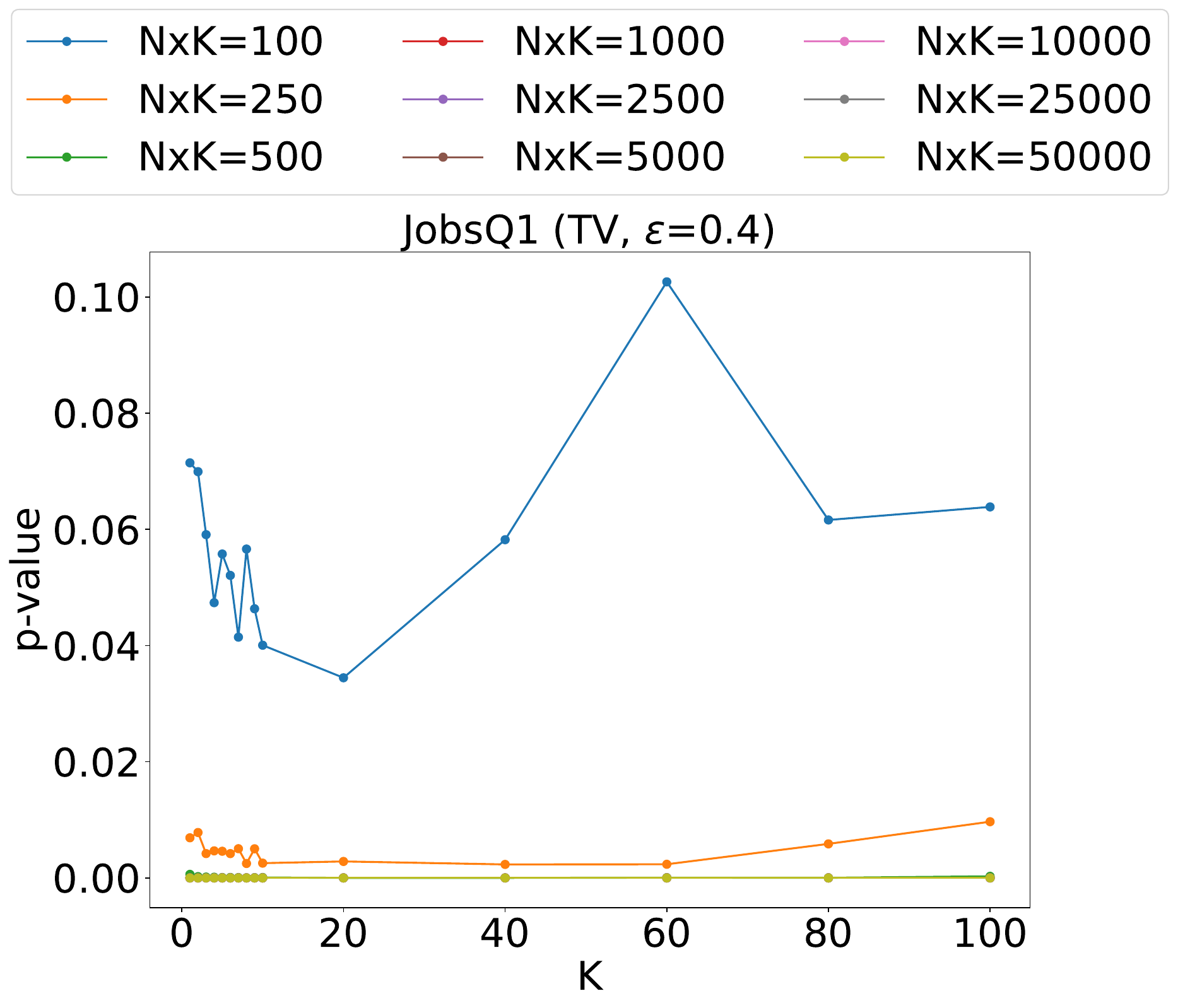}
    \caption{$\epsilon = 0.4$}
    \label{fig:jobsQ1_MAE_e04}
  \end{subfigure}
  \caption{P-value plots for JobsQ1 dataset with TV as the metric}
  \label{fig:jobsQ1_MAE}
\end{figure*}

\begin{figure*}
  \centering
  \begin{subfigure}[b]{0.24\linewidth}
    \centering
    \includegraphics[width=\linewidth]{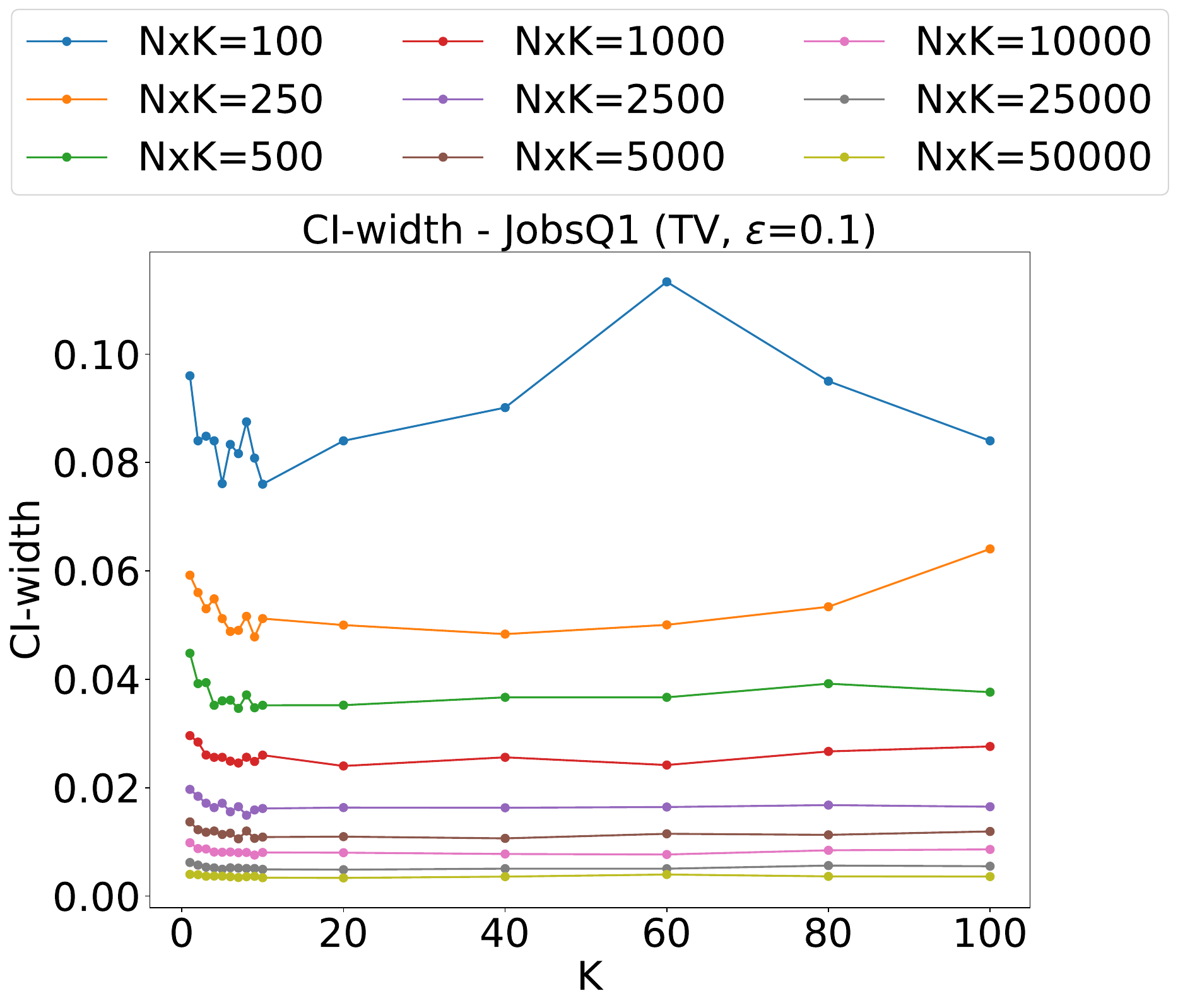}
    \caption{$\epsilon = 0.1$}
    \label{fig:jobsQ1_ci_MAE_e01}
  \end{subfigure} \hfill
  \begin{subfigure}[b]{0.24\linewidth}
    \centering
    \includegraphics[width=\linewidth]{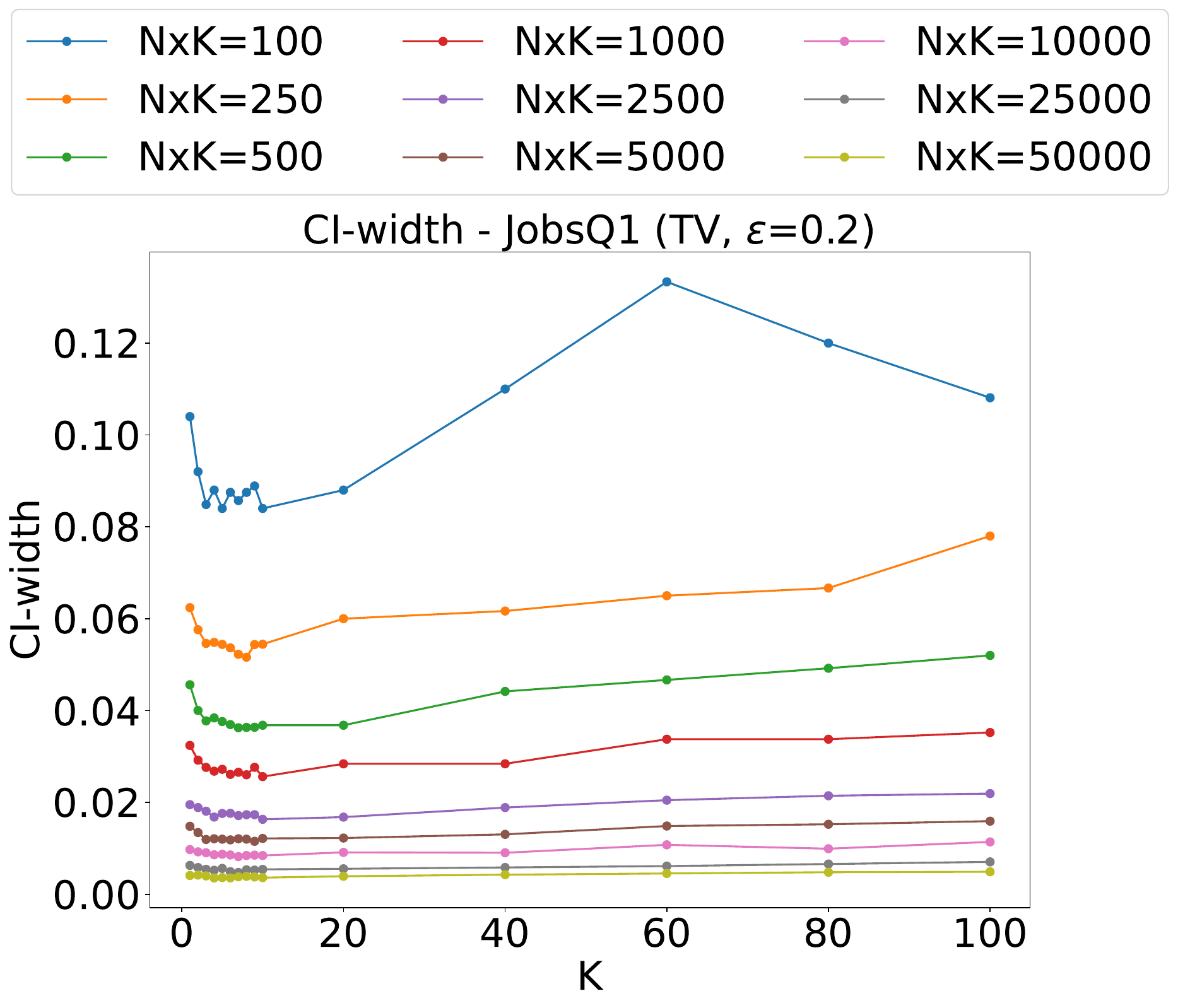}
    \caption{$\epsilon = 0.2$}
    \label{fig:jobsQ1_ci_MAE_e02}
  \end{subfigure} \hfill
  \begin{subfigure}[b]{0.24\linewidth}
    \centering
    \includegraphics[width=\linewidth]{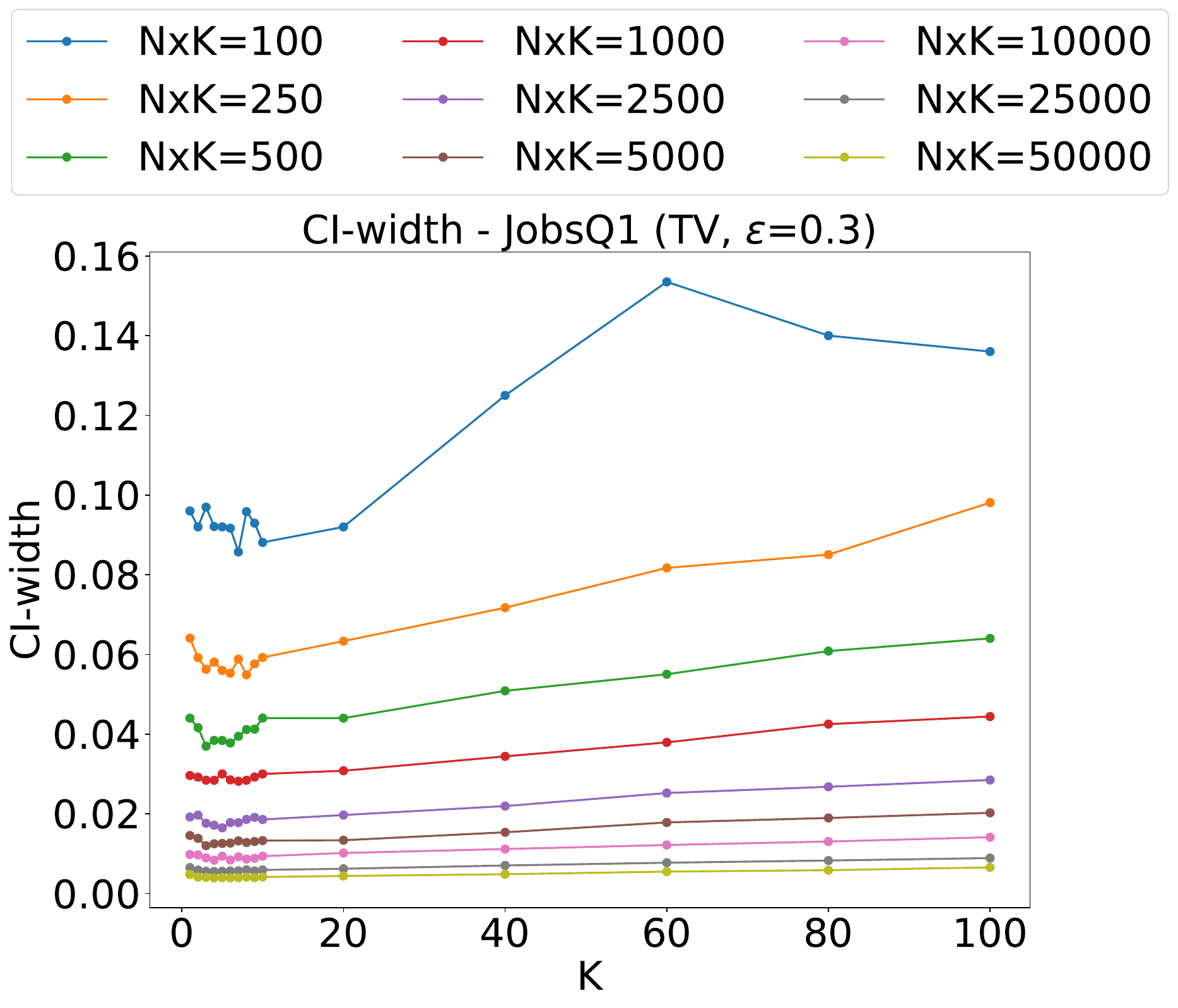}
    \caption{$\epsilon = 0.3$}
    \label{fig:jobsQ1_ci_MAE_e03}
  \end{subfigure} \hfill
  \begin{subfigure}[b]{0.24\linewidth}
    \centering
    \includegraphics[width=\linewidth]{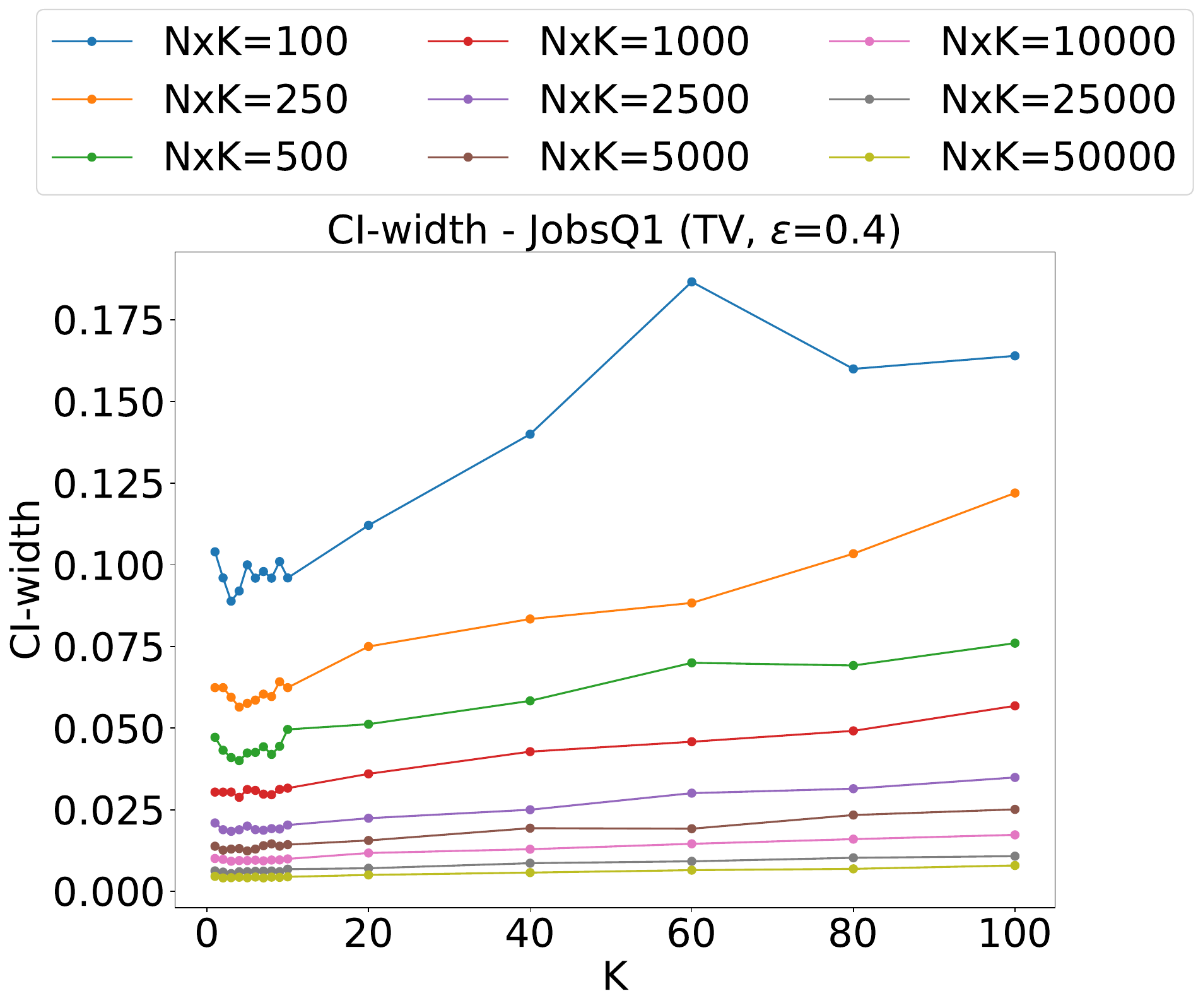}
    \caption{$\epsilon = 0.4$}
    \label{fig:jobsQ1_ci_MAE_e04}
  \end{subfigure}
  \caption{CI-width plots for JobsQ1 dataset with TV as the metric}
  \label{fig:jobsQ1_ci_MAE}
\end{figure*}

\begin{figure*}
  \centering
  \begin{subfigure}[b]{0.24\linewidth}
    \centering
    \includegraphics[width=\linewidth]{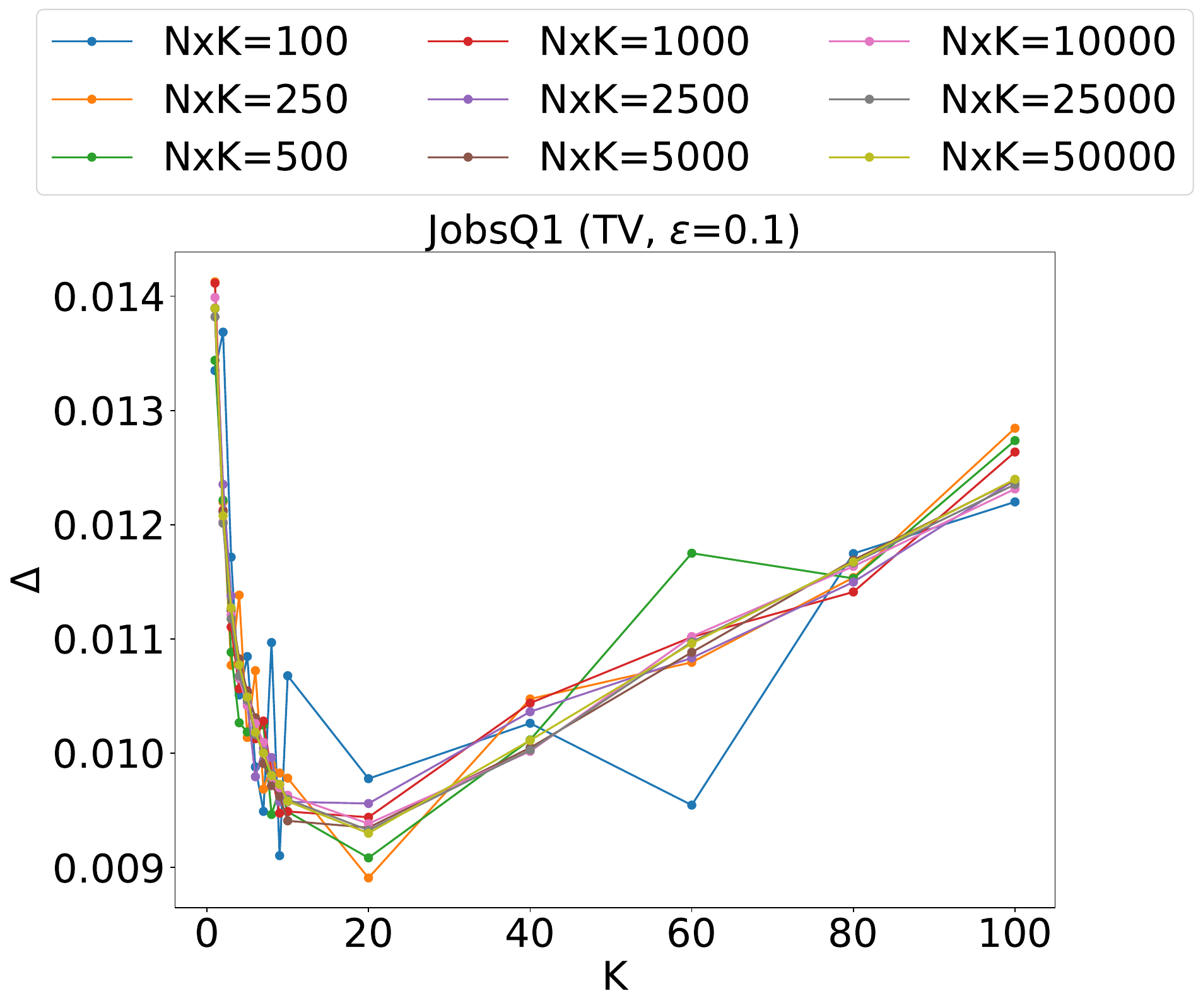}
    \caption{$\epsilon = 0.1$}
    \label{fig:jobsQ1_delta_MAE_e01}
  \end{subfigure} \hfill
  \begin{subfigure}[b]{0.24\linewidth}
    \centering
    \includegraphics[width=\linewidth]{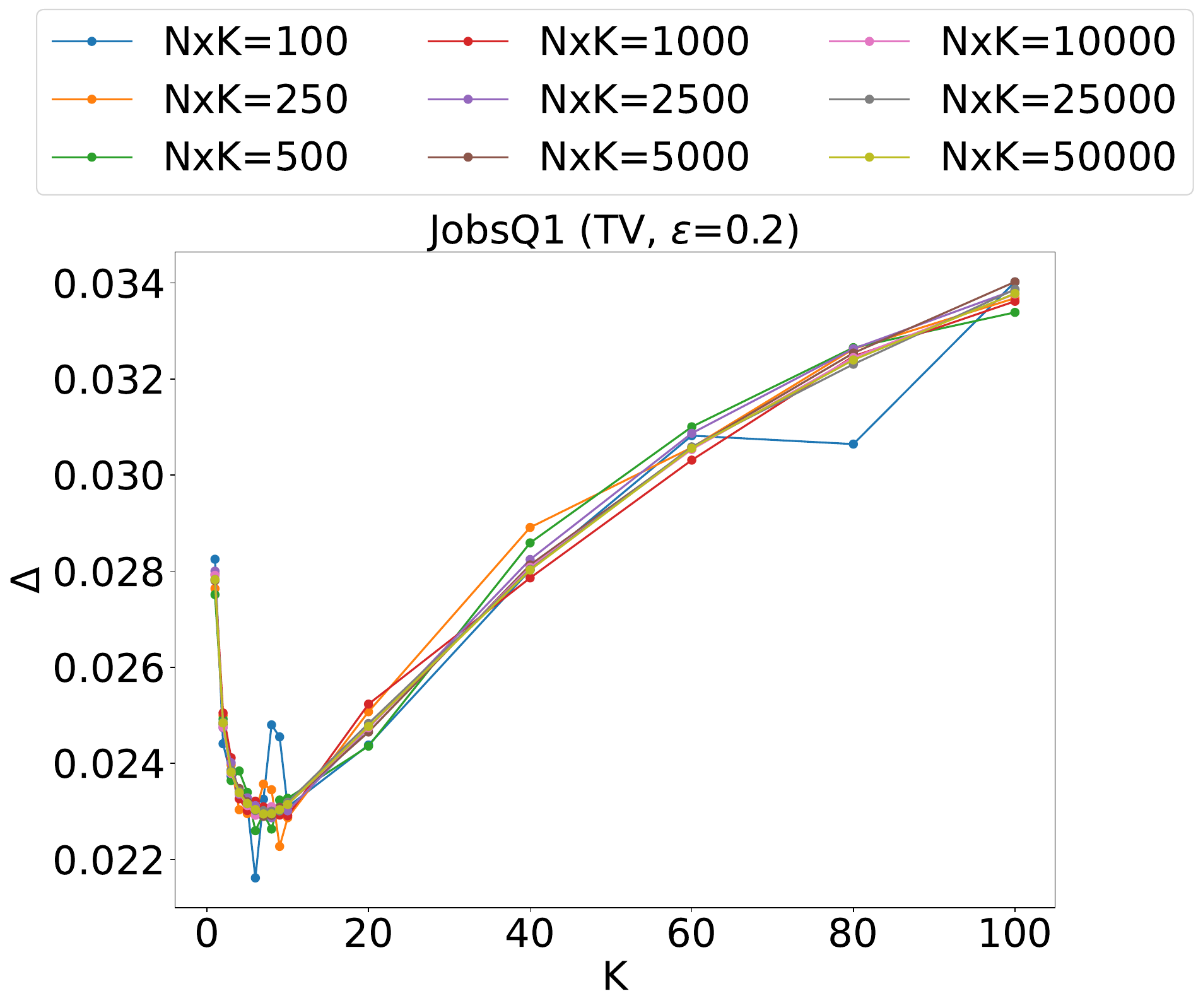}
    \caption{$\epsilon = 0.2$}
    \label{fig:jobsQ1_delta_MAE_e02}
  \end{subfigure} \hfill
  \begin{subfigure}[b]{0.24\linewidth}
    \centering
    \includegraphics[width=\linewidth]{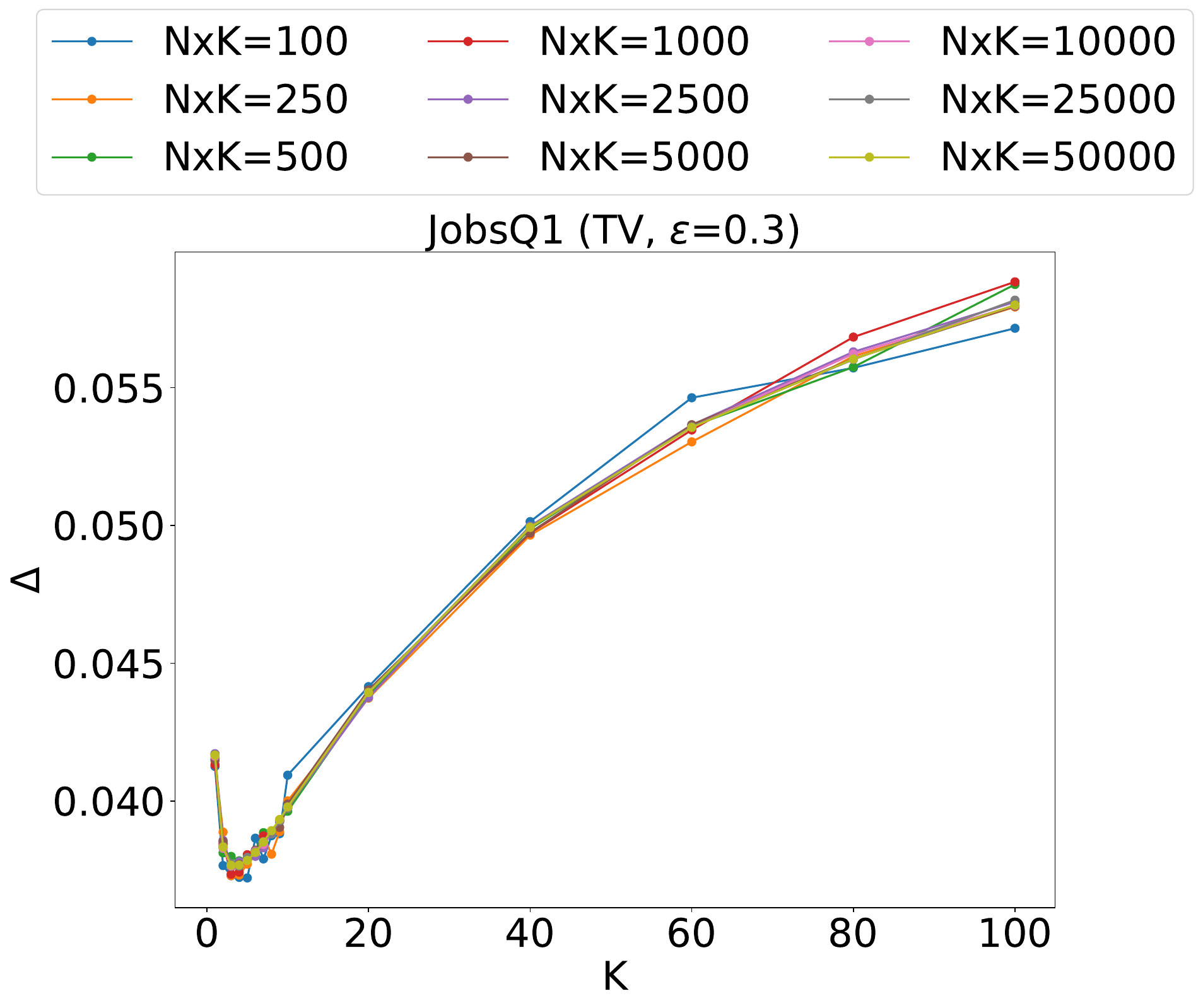}
    \caption{$\epsilon = 0.3$}
    \label{fig:jobsQ1_delta_MAE_e03}
  \end{subfigure} \hfill
  \begin{subfigure}[b]{0.24\linewidth}
    \centering
    \includegraphics[width=\linewidth]{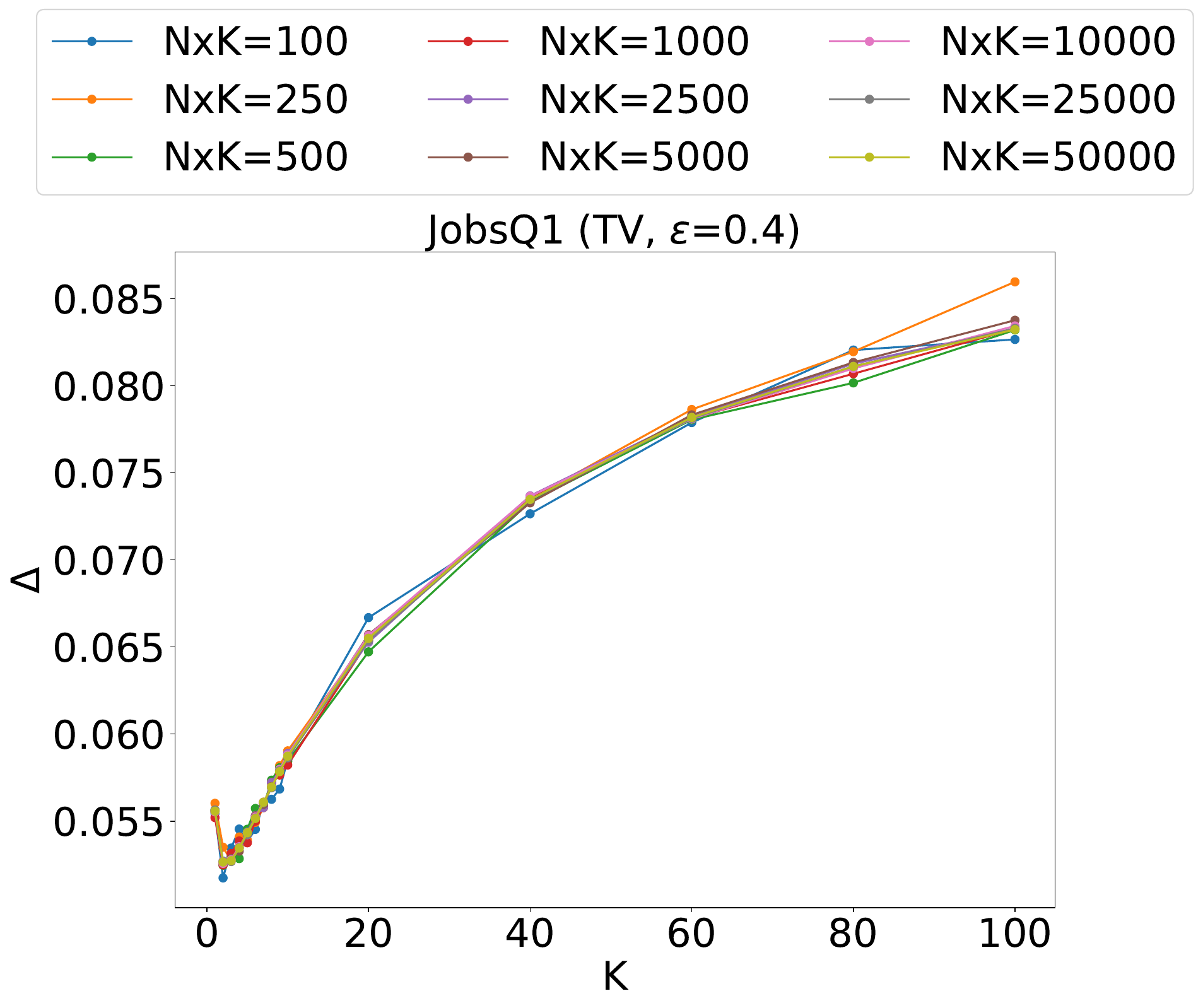}
    \caption{$\epsilon = 0.4$}
    \label{fig:jobsQ1_delta_MAE_e04}
  \end{subfigure}
  \caption{Effect sizes ($\Delta$) for JobsQ1 dataset with TV as the metric}
  \label{fig:jobsQ1_delta_MAE}
\end{figure*}

\begin{figure*}
  \centering
  \begin{subfigure}[b]{0.24\linewidth}
    \centering
    \includegraphics[width=\linewidth]{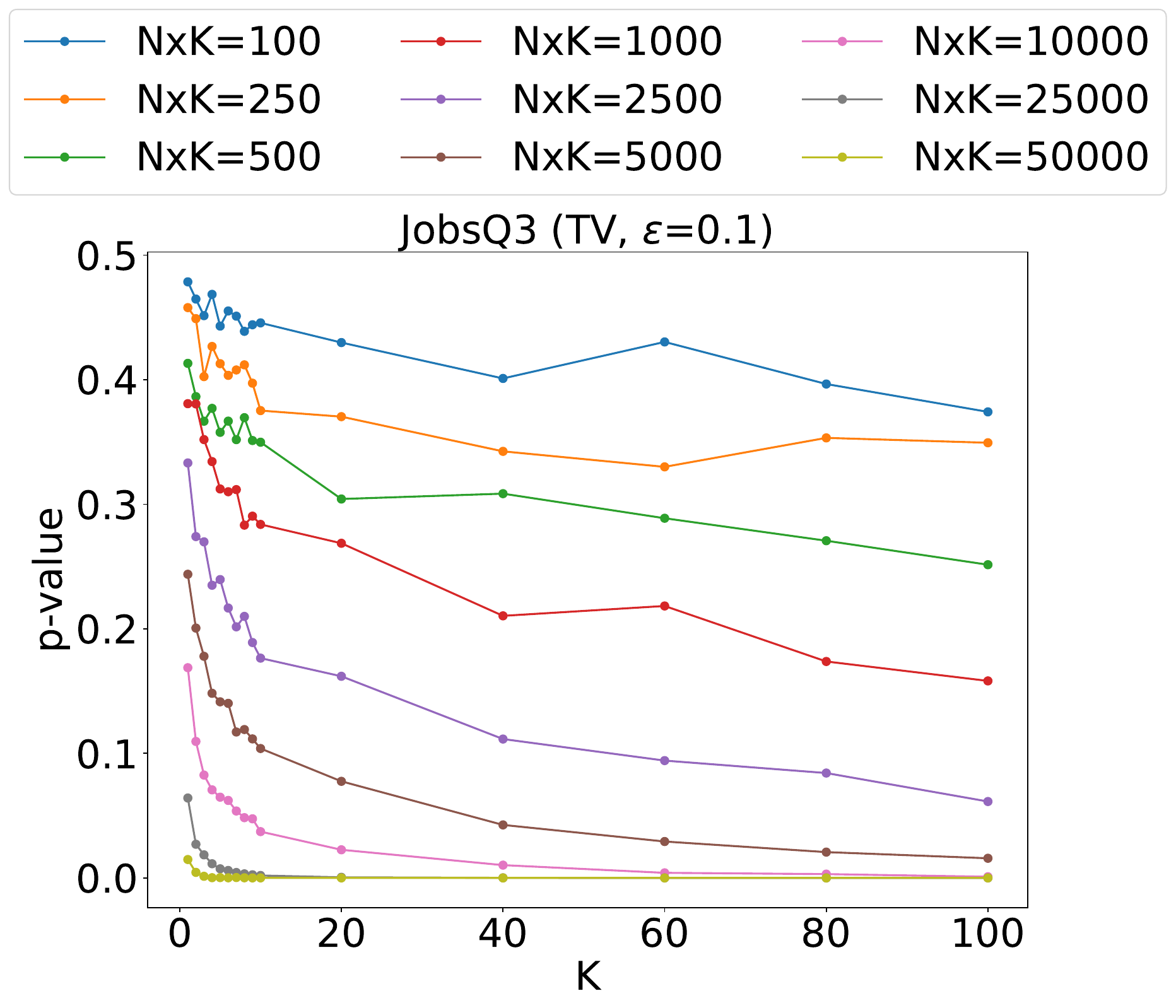}
    \caption{$\epsilon = 0.1$}
    \label{fig:jobsQ3_MAE_e01}
  \end{subfigure} \hfill
  \begin{subfigure}[b]{0.24\linewidth}
    \centering
    \includegraphics[width=\linewidth]{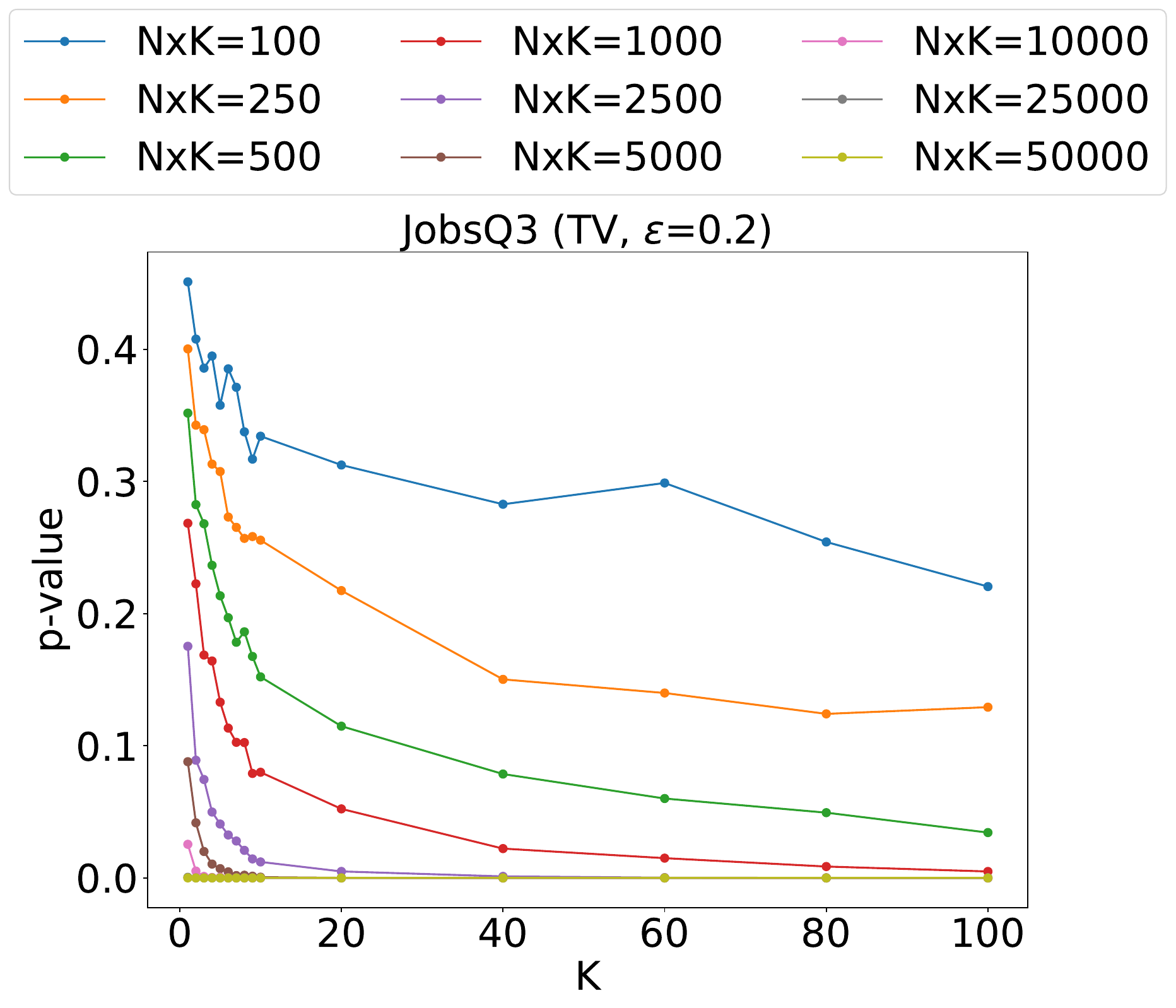}
    \caption{$\epsilon = 0.2$}
    \label{fig:jobsQ3_MAE_e02}
  \end{subfigure} \hfill
  \begin{subfigure}[b]{0.24\linewidth}
    \centering
    \includegraphics[width=\linewidth]{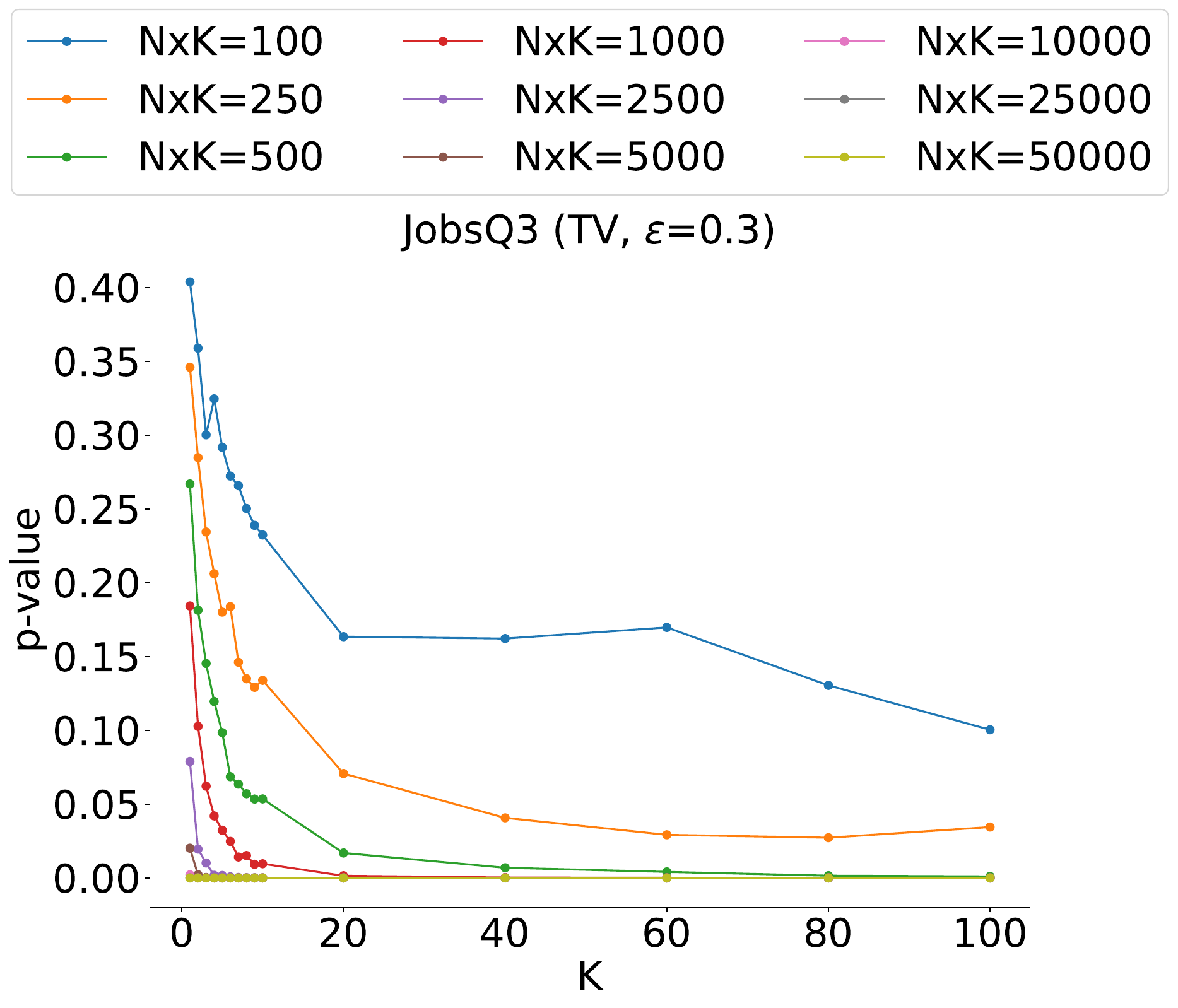}
    \caption{$\epsilon = 0.3$}
    \label{fig:jobsQ3_MAE_e03}
  \end{subfigure} \hfill
  \begin{subfigure}[b]{0.24\linewidth}
    \centering
    \includegraphics[width=\linewidth]{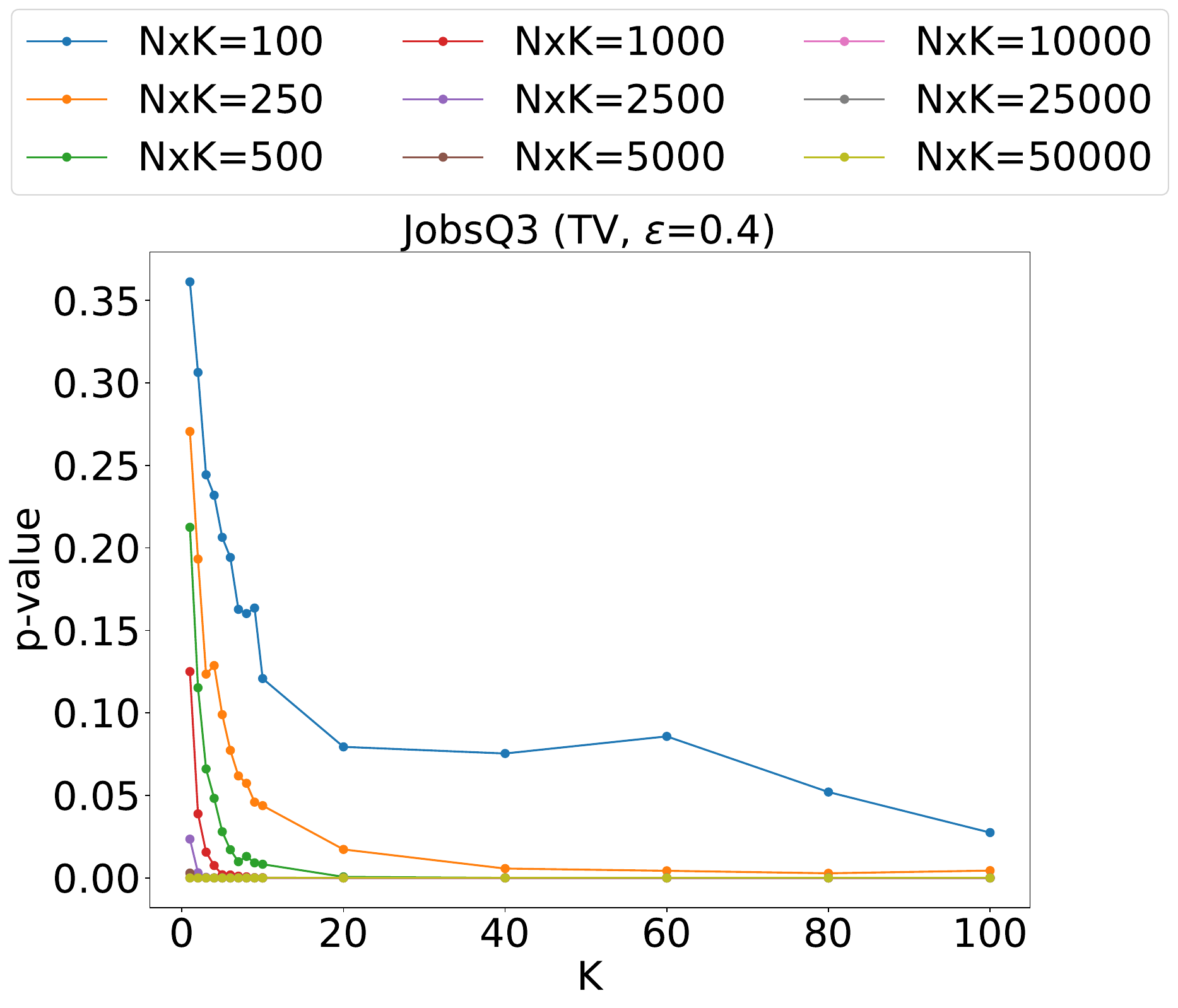}
    \caption{$\epsilon = 0.4$}
    \label{fig:jobsQ3_MAE_e04}
  \end{subfigure}
  \caption{P-value plots for JobsQ3 dataset with TV as the metric}
  \label{fig:jobsQ3_MAE}
\end{figure*}

\begin{figure*}
  \centering
  \begin{subfigure}[b]{0.24\linewidth}
    \centering
    \includegraphics[width=\linewidth]{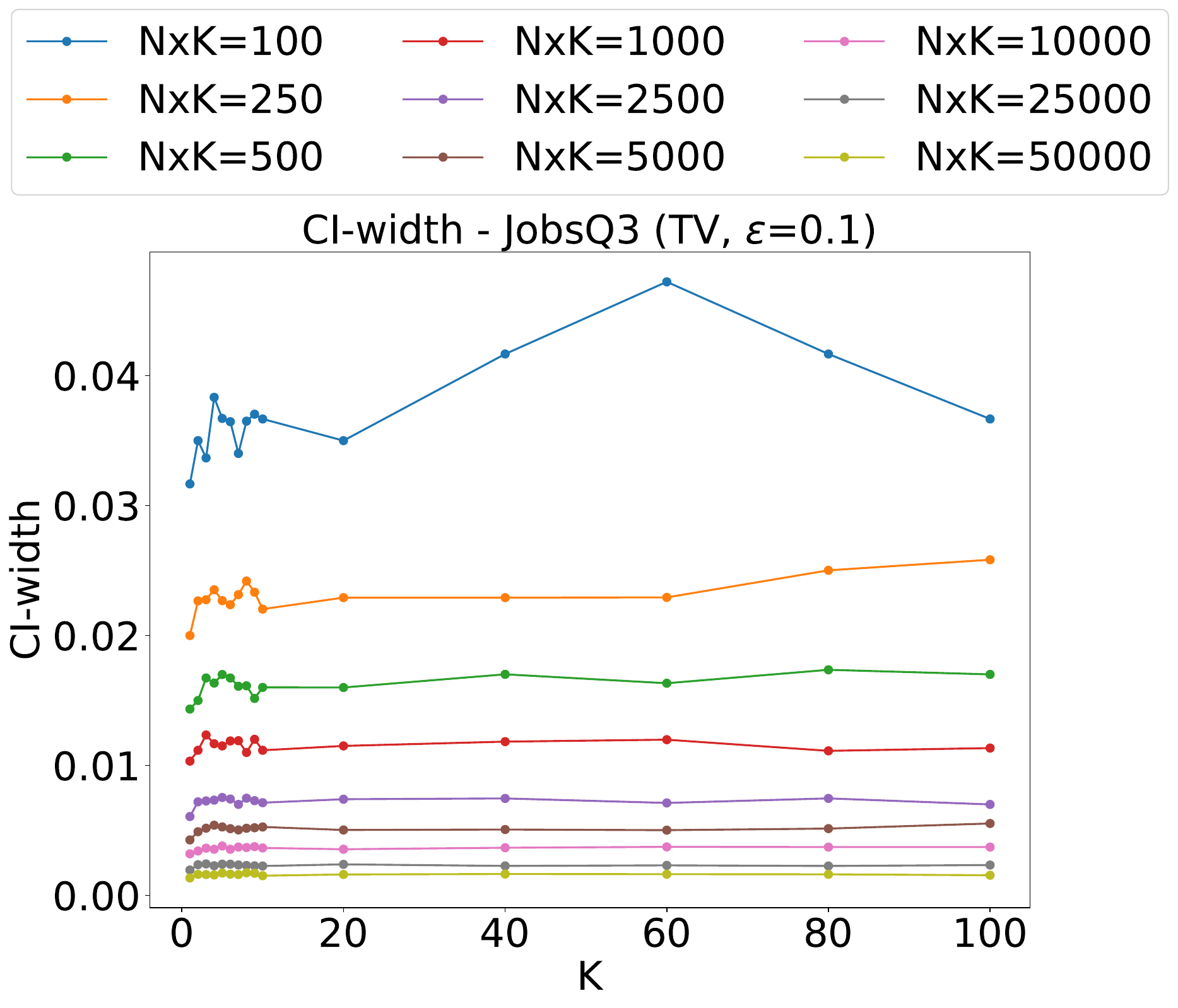}
    \caption{$\epsilon = 0.1$}
    \label{fig:jobsQ3_ci_MAE_e01}
  \end{subfigure} \hfill
  \begin{subfigure}[b]{0.24\linewidth}
    \centering
    \includegraphics[width=\linewidth]{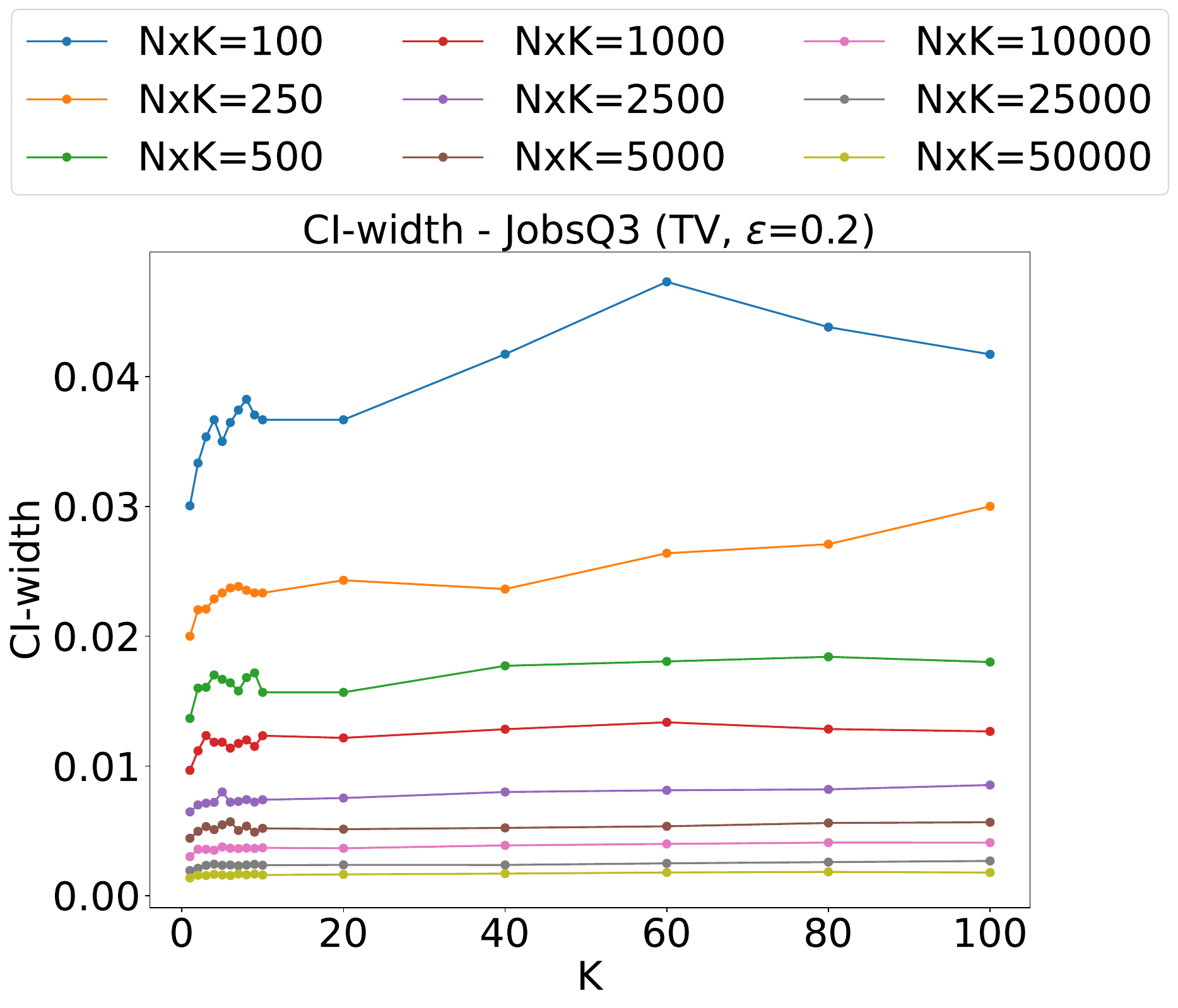}
    \caption{$\epsilon = 0.2$}
    \label{fig:jobsQ3_ci_MAE_e02}
  \end{subfigure} \hfill
  \begin{subfigure}[b]{0.24\linewidth}
    \centering
    \includegraphics[width=\linewidth]{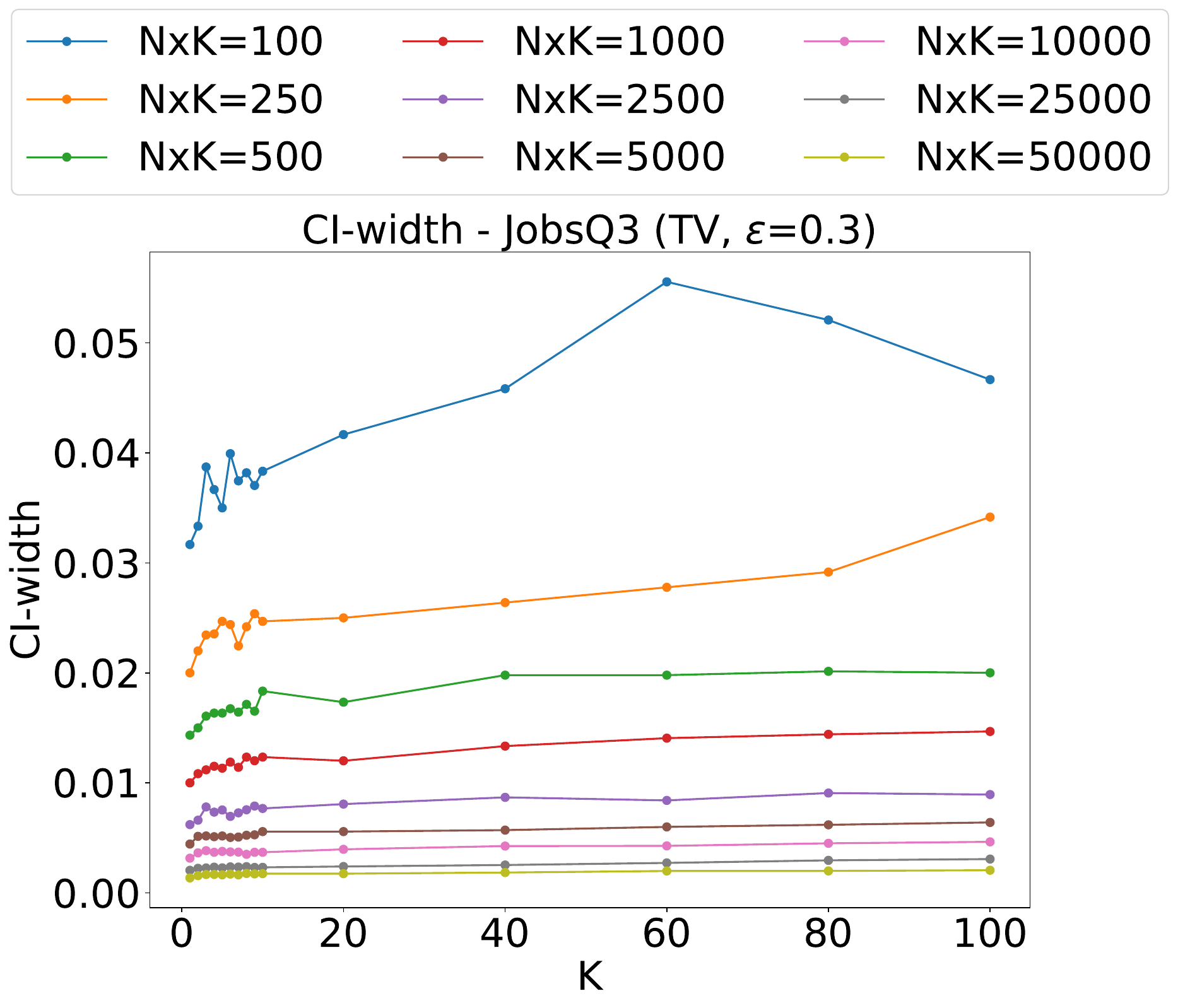}
    \caption{$\epsilon = 0.3$}
    \label{fig:jobsQ3_ci_MAE_e03}
  \end{subfigure} \hfill
  \begin{subfigure}[b]{0.24\linewidth}
    \centering
    \includegraphics[width=\linewidth]{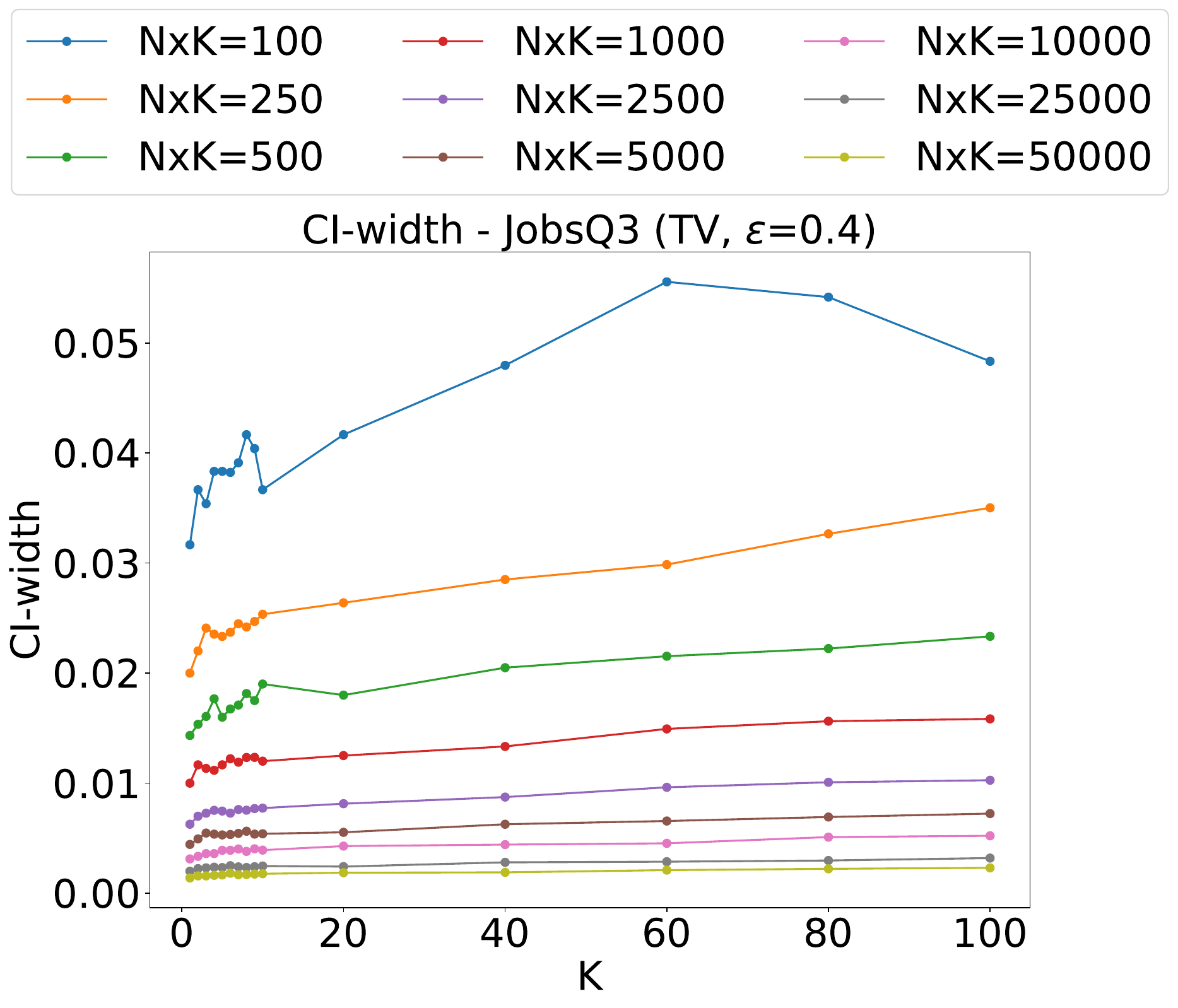}
    \caption{$\epsilon = 0.4$}
    \label{fig:jobsQ3_ci_MAE_e04}
  \end{subfigure}
  \caption{CI-width plots for JobsQ3 dataset with TV as the metric}
  \label{fig:jobsQ3_ci_MAE}
\end{figure*}

\begin{figure*}
  \centering
  \begin{subfigure}[b]{0.24\linewidth}
    \centering
    \includegraphics[width=\linewidth]{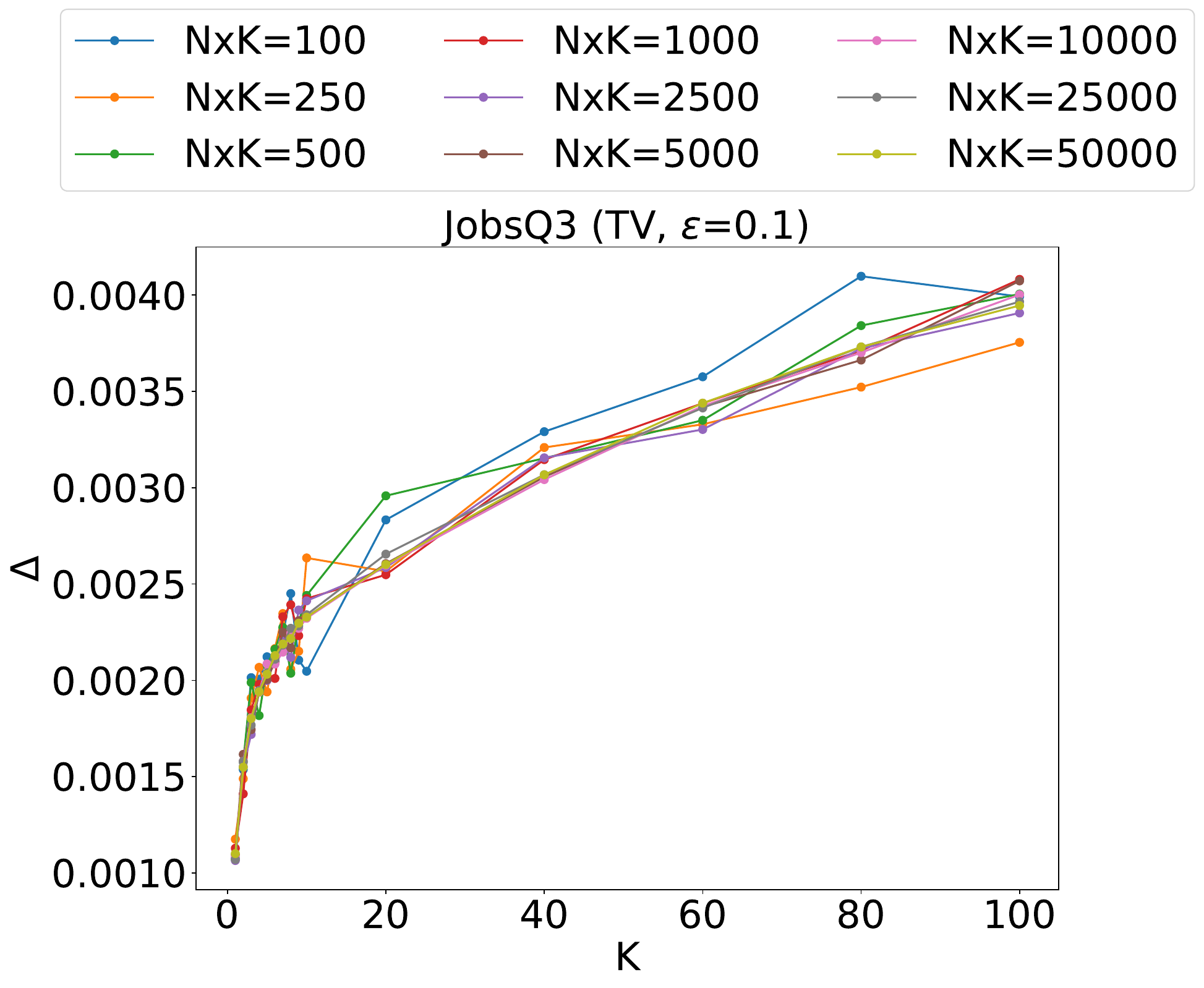}
    \caption{$\epsilon = 0.1$}
    \label{fig:jobsQ3_delta_MAE_e01}
  \end{subfigure} \hfill
  \begin{subfigure}[b]{0.24\linewidth}
    \centering
    \includegraphics[width=\linewidth]{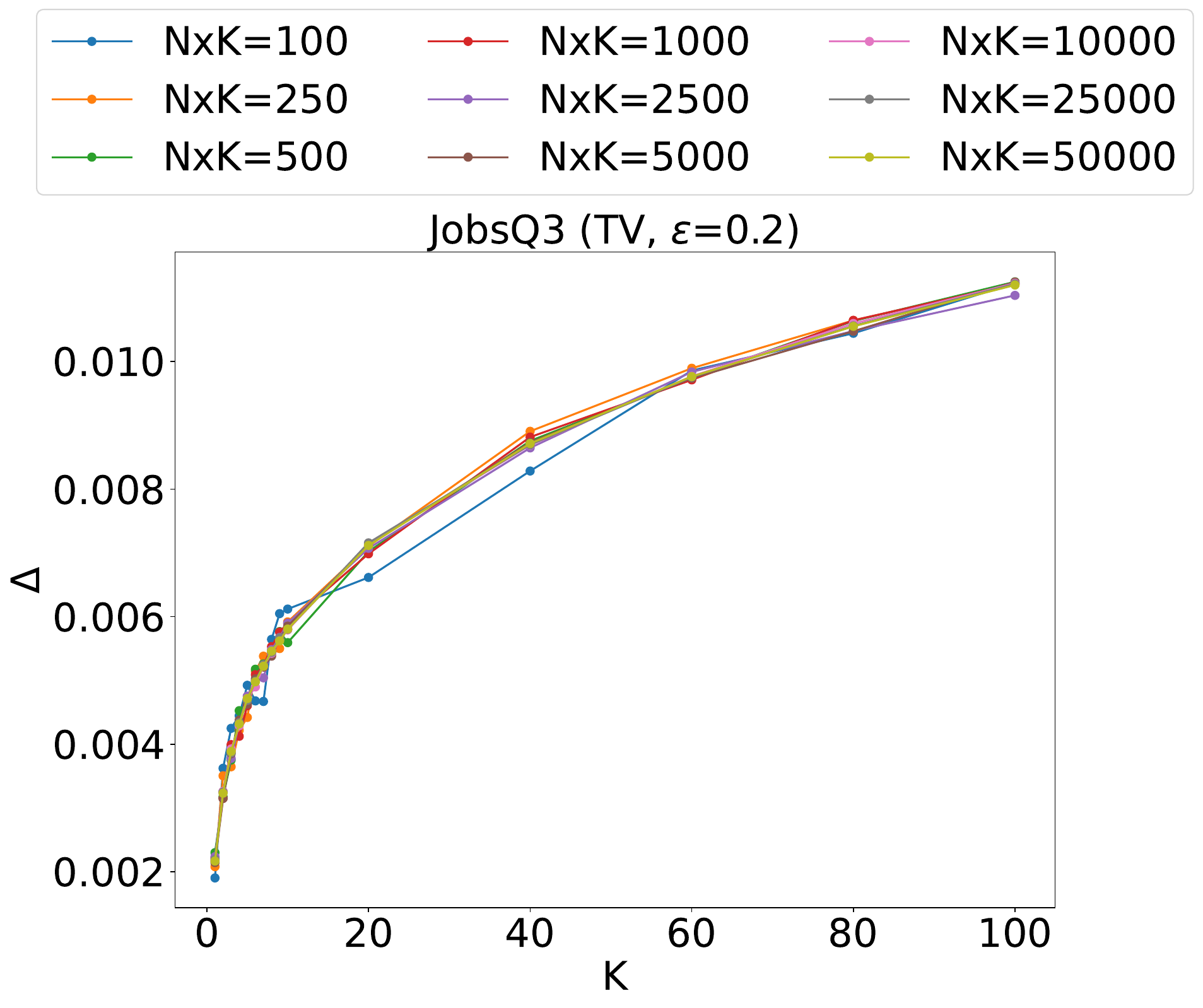}
    \caption{$\epsilon = 0.2$}
    \label{fig:jobsQ3_delta_MAE_e02}
  \end{subfigure} \hfill
  \begin{subfigure}[b]{0.24\linewidth}
    \centering
    \includegraphics[width=\linewidth]{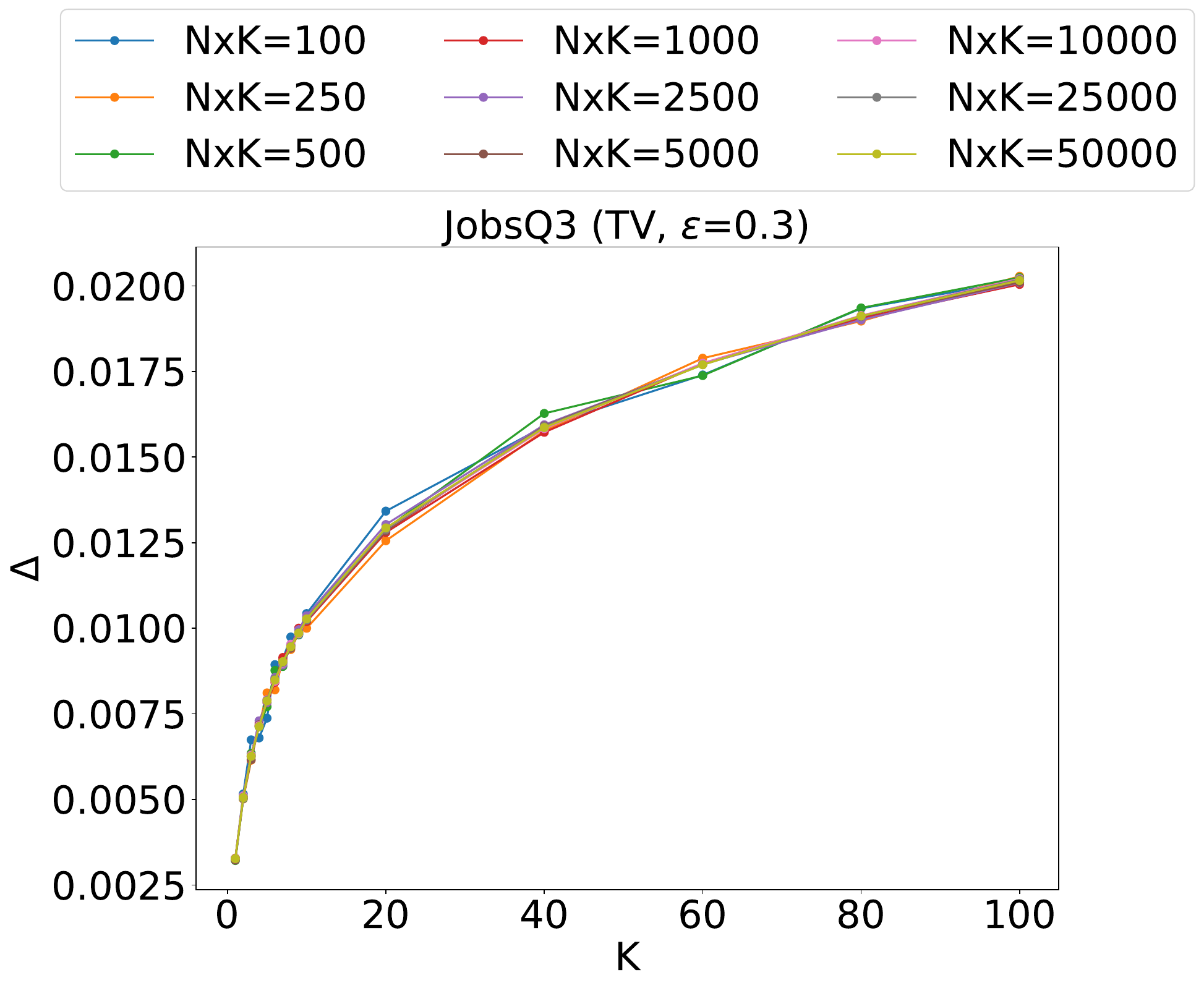}
    \caption{$\epsilon = 0.3$}
    \label{fig:jobsQ3_delta_MAE_e03}
  \end{subfigure} \hfill
  \begin{subfigure}[b]{0.24\linewidth}
    \centering
    \includegraphics[width=\linewidth]{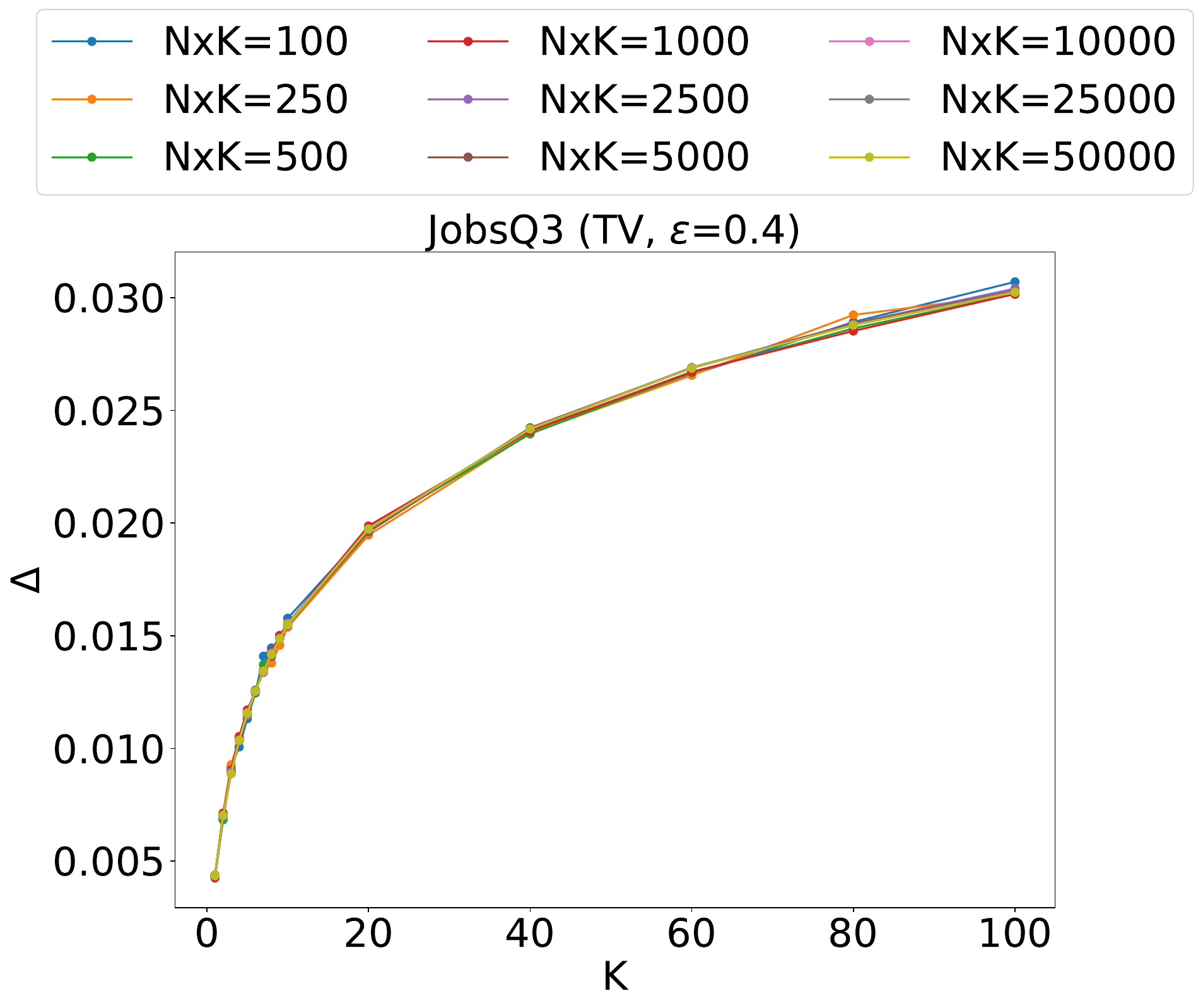}
    \caption{$\epsilon = 0.4$}
    \label{fig:jobsQ3_delta_MAE_e04}
  \end{subfigure}
  \caption{Effect sizes ($\Delta$) for JobsQ3 dataset with TV as the metric}
  \label{fig:jobsQ3_delta_MAE}
\end{figure*}

Figures \ref{fig:toxicity_MAE}--\ref{fig:jobsQ3_delta_MAE} show results for \pv s, confidence intervals, and effect sizes of TV for all datasets with different values of $\epsilon$ and increasing $K$ for various $N \times K$. We observe that \pv\ increases sharply with increasing $K$ till $K=10$, and then starts to decrease thereafter for all $N \times K$ for $\epsilon=0.1$. For the remaining values of $\epsilon$, the \pv s\ generally decrease with increasing $K$. We also notice an elbow plot emerging for total variation, suggesting an optimal value of $K$ for the datasets. As $\epsilon$ increases, \pv s\ decrease as expected. Having more responses per item ($K$) seems to be helpful for total variation and results in lower \pv s.\ CI-width and effect sizes increase as $K$ increases.

\subsubsection{Artificial Distribution with Balanced Priors}

\begin{figure*}
  \centering
  \begin{subfigure}[b]{0.24\linewidth}
    \centering
    \includegraphics[width=\linewidth]{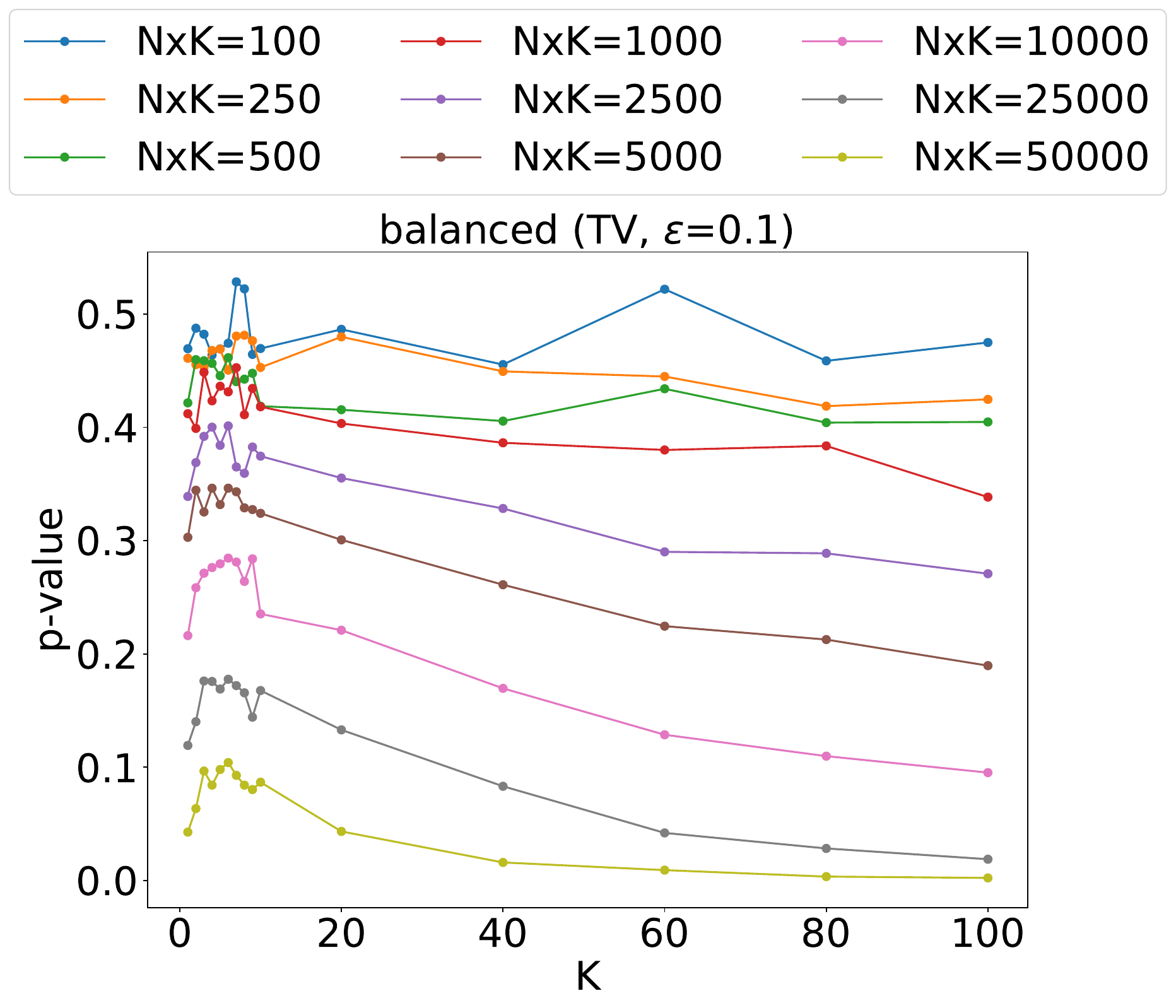}
    \caption{$\epsilon = 0.1$}
    \label{fig:uniform_MAE_cat2_e01}
  \end{subfigure} \hfill
  \begin{subfigure}[b]{0.24\linewidth}
    \centering
    \includegraphics[width=\linewidth]{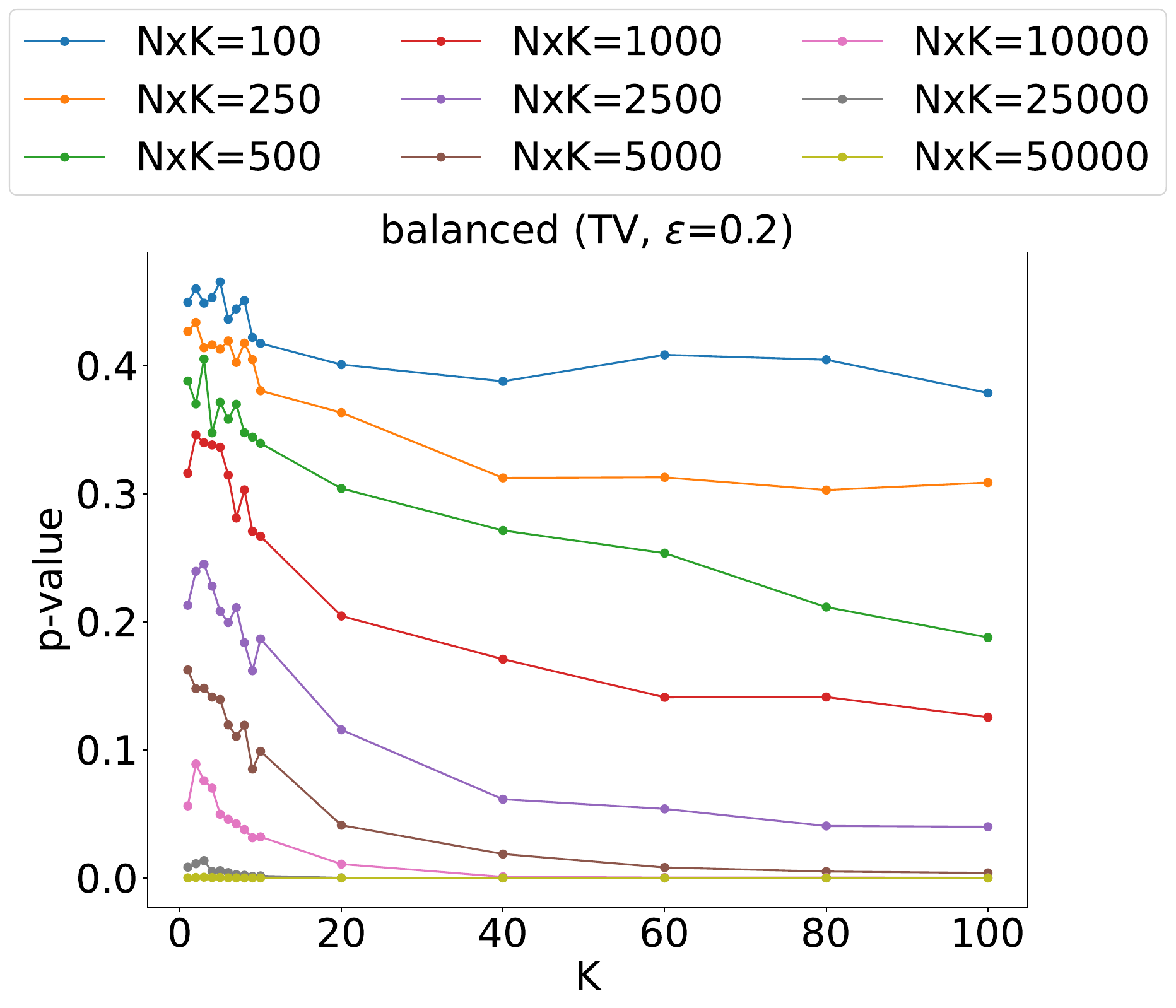}
    \caption{$\epsilon = 0.2$}
    \label{fig:uniform_MAE_cat2_e02}
  \end{subfigure} \hfill
  \begin{subfigure}[b]{0.24\linewidth}
    \centering
    \includegraphics[width=\linewidth]{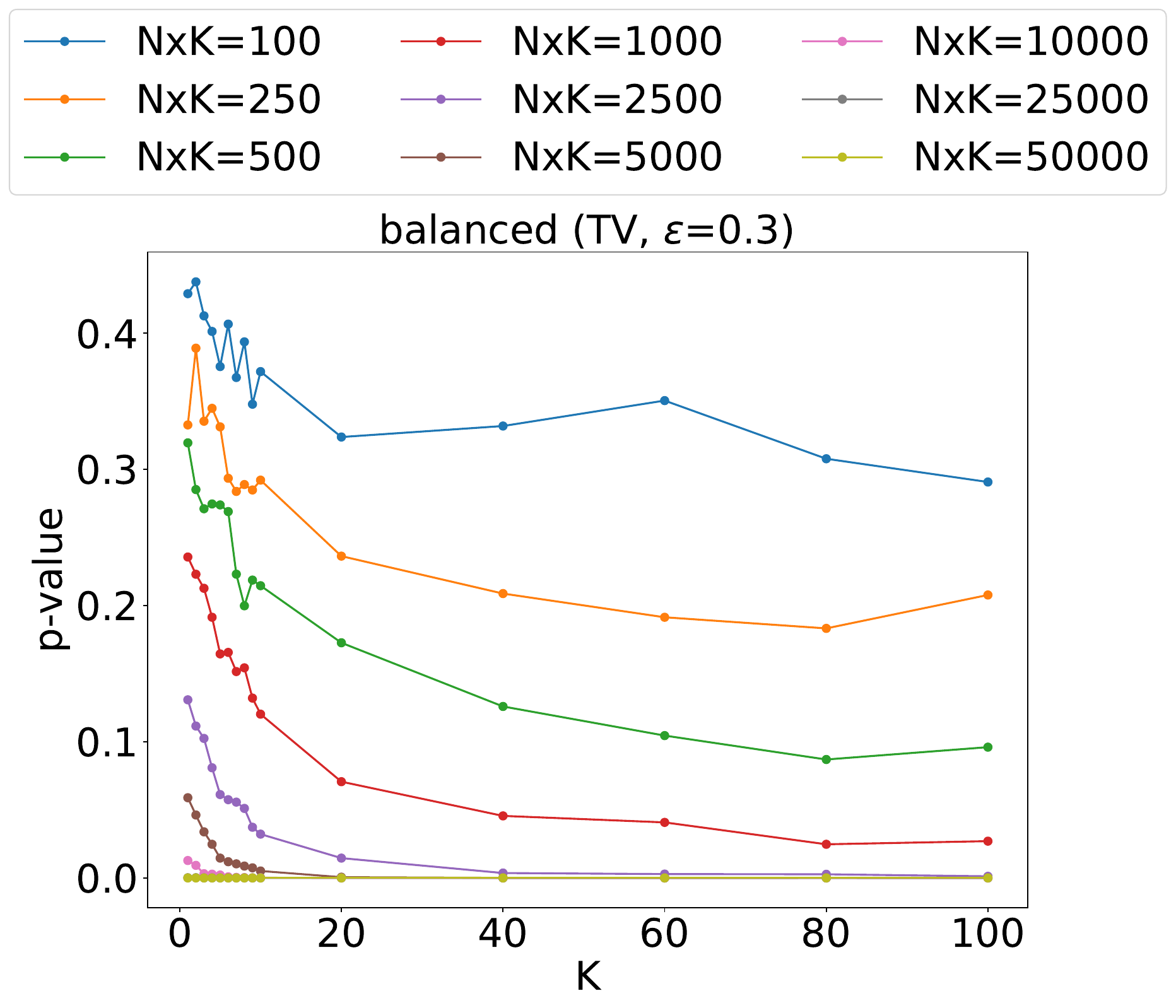}
    \caption{$\epsilon = 0.3$}
    \label{fig:uniform_MAE_cat2_e03}
  \end{subfigure} \hfill
  \begin{subfigure}[b]{0.24\linewidth}
    \centering
    \includegraphics[width=\linewidth]{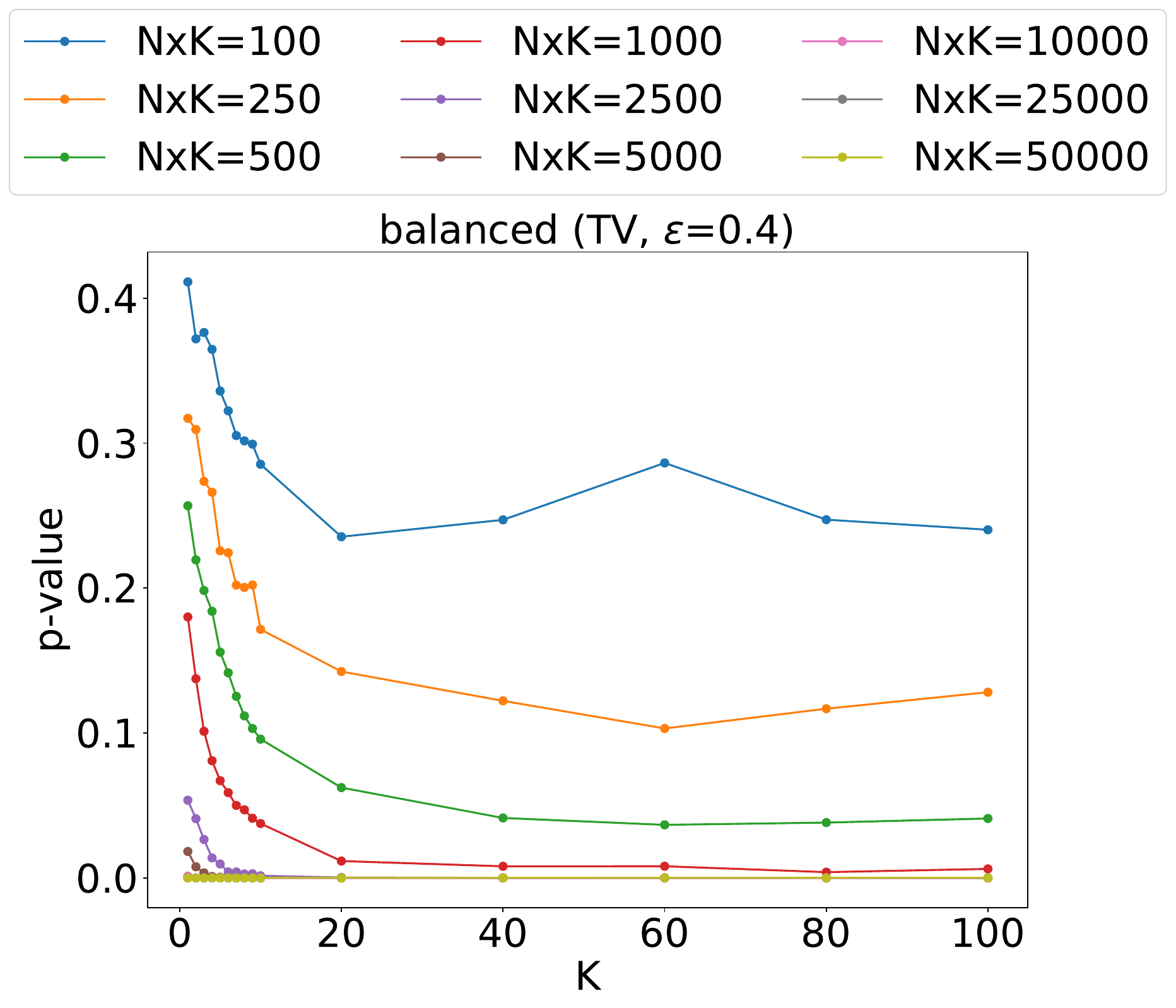}
    \caption{$\epsilon = 0.4$}
    \label{fig:uniform_MAE_cat2_e04}
  \end{subfigure}
  \caption{P-value plots for balanced alphas with TV as the metric ($M=2$)}
  \label{fig:uniform_MAE_cat2}
\end{figure*}

\begin{figure*}
  \centering
  \begin{subfigure}[b]{0.24\linewidth}
    \centering
    \includegraphics[width=\linewidth]{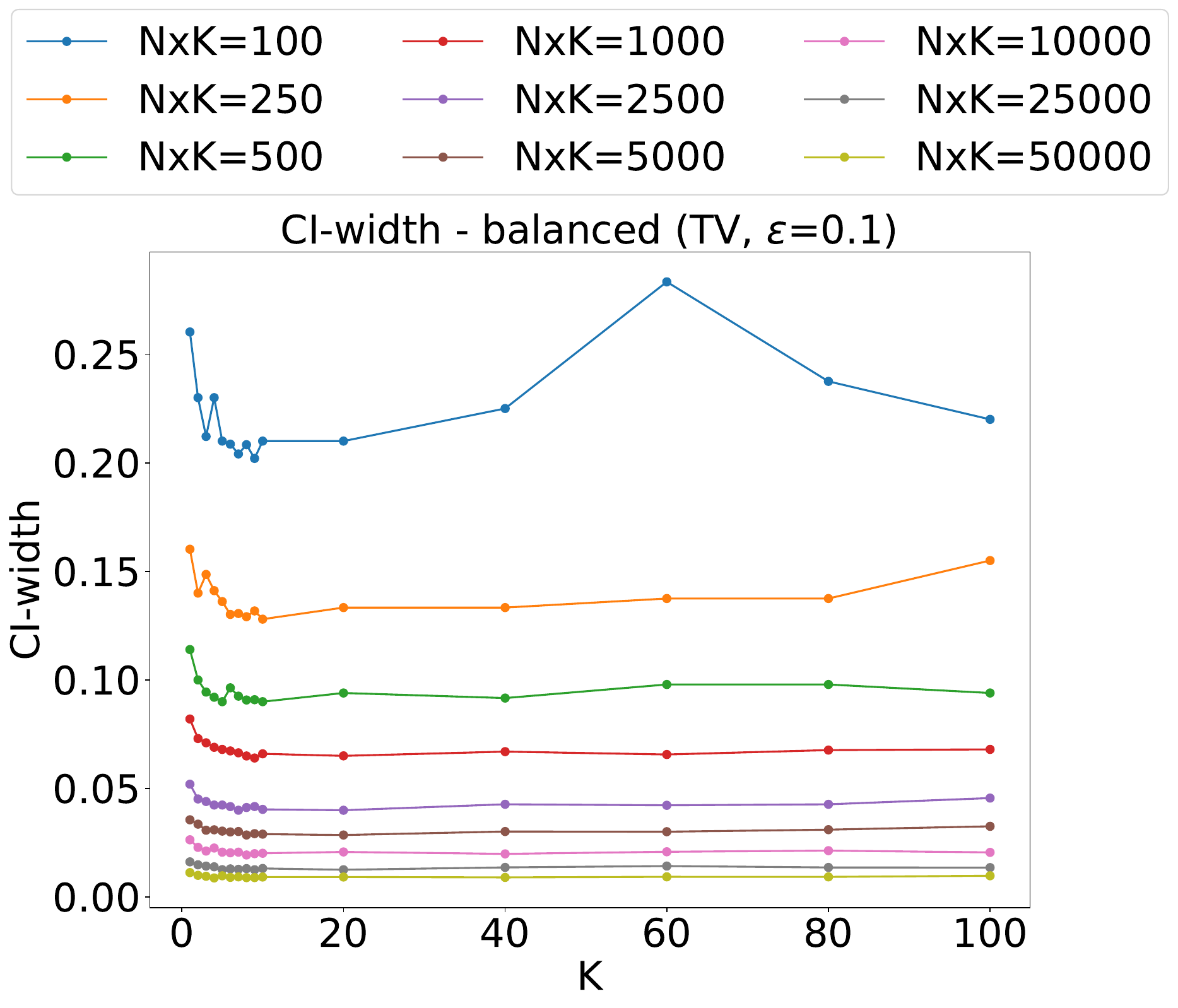}
    \caption{$\epsilon = 0.1$}
    \label{fig:uniform_ci_MAE_cat2_e01}
  \end{subfigure} \hfill
  \begin{subfigure}[b]{0.24\linewidth}
    \centering
    \includegraphics[width=\linewidth]{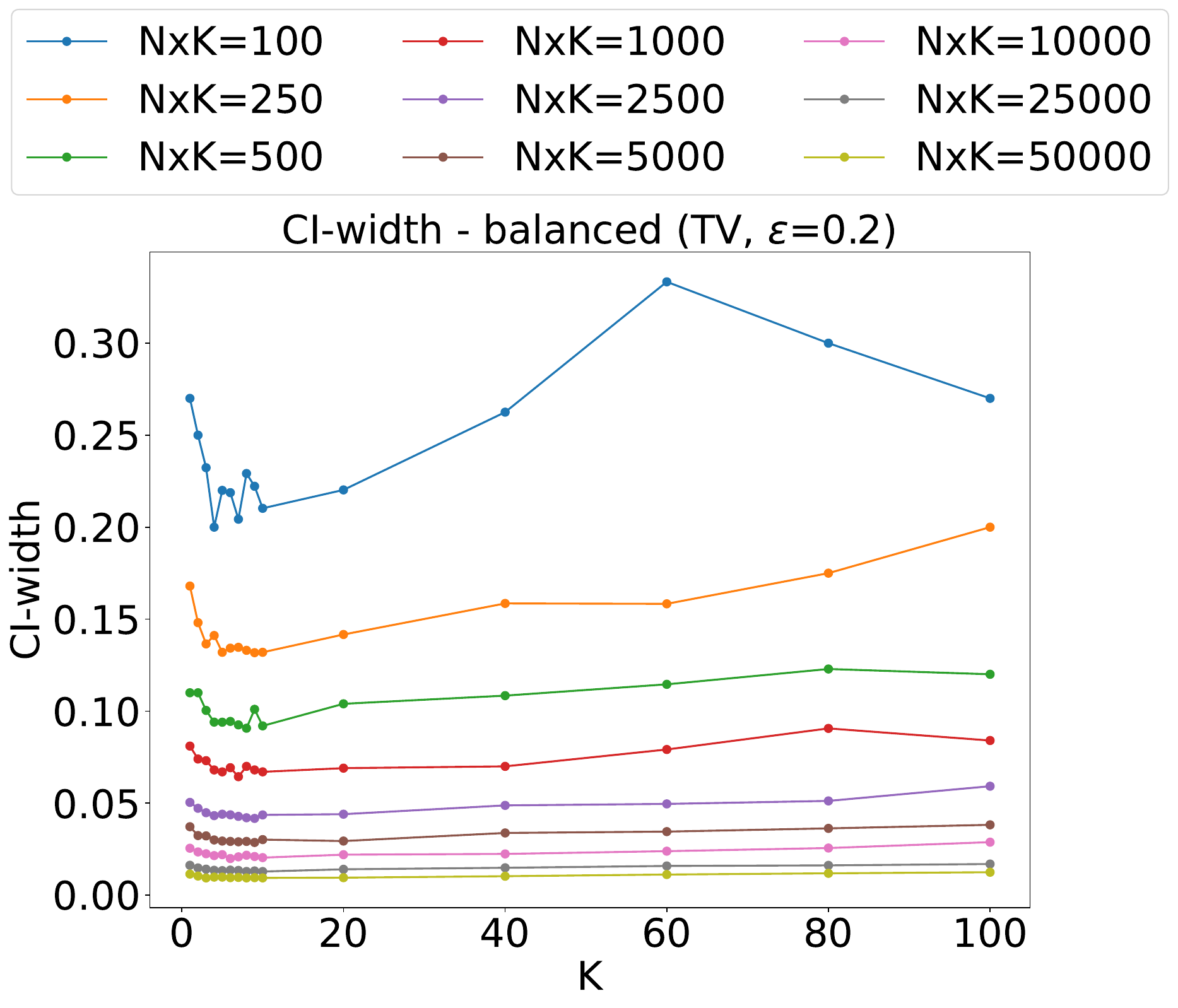}
    \caption{$\epsilon = 0.2$}
    \label{fig:uniform_ci_MAE_cat2_e02}
  \end{subfigure} \hfill
  \begin{subfigure}[b]{0.24\linewidth}
    \centering
    \includegraphics[width=\linewidth]{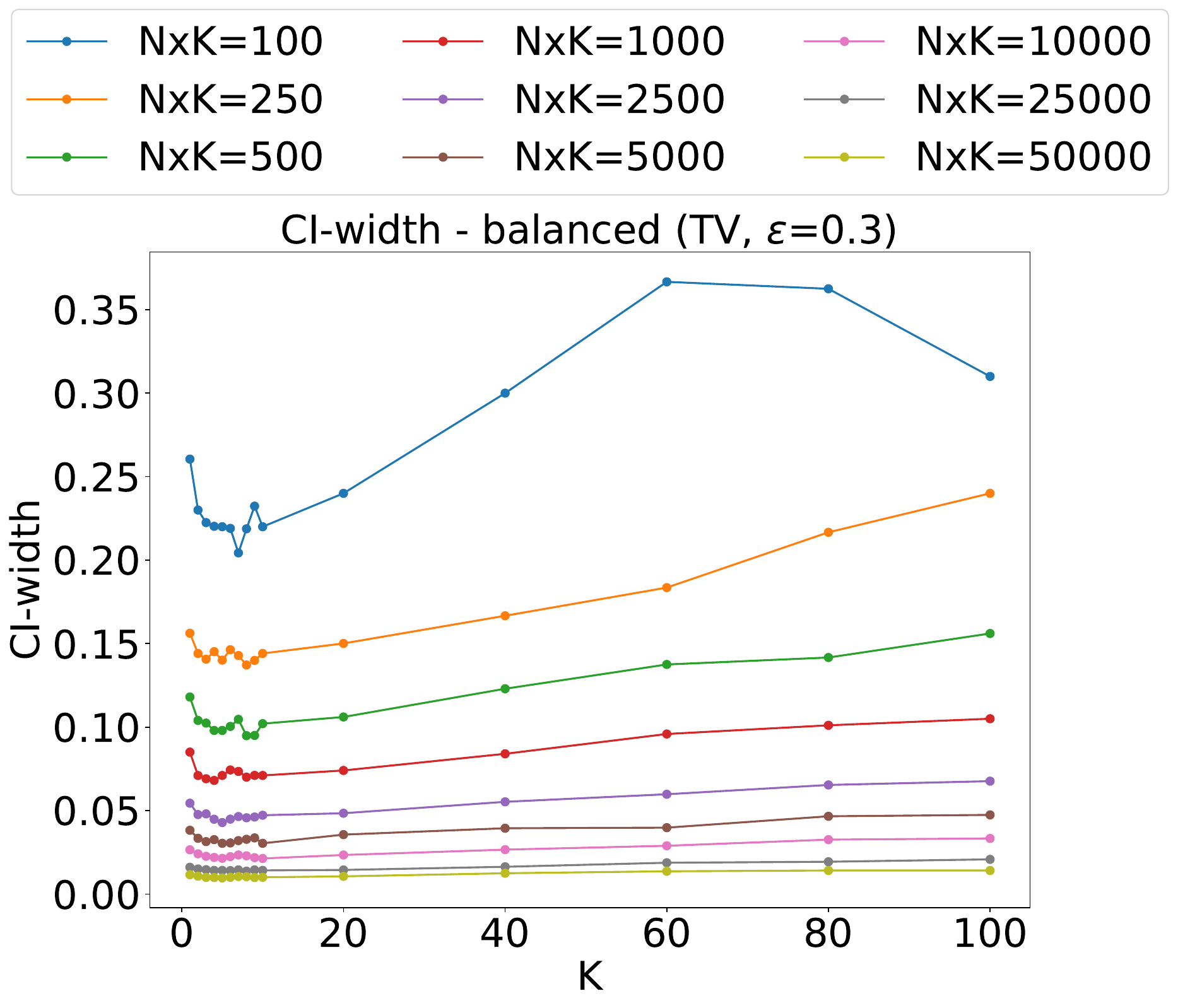}
    \caption{$\epsilon = 0.3$}
    \label{fig:uniform_ci_MAE_cat2_e03}
  \end{subfigure} \hfill
  \begin{subfigure}[b]{0.24\linewidth}
    \centering
    \includegraphics[width=\linewidth]{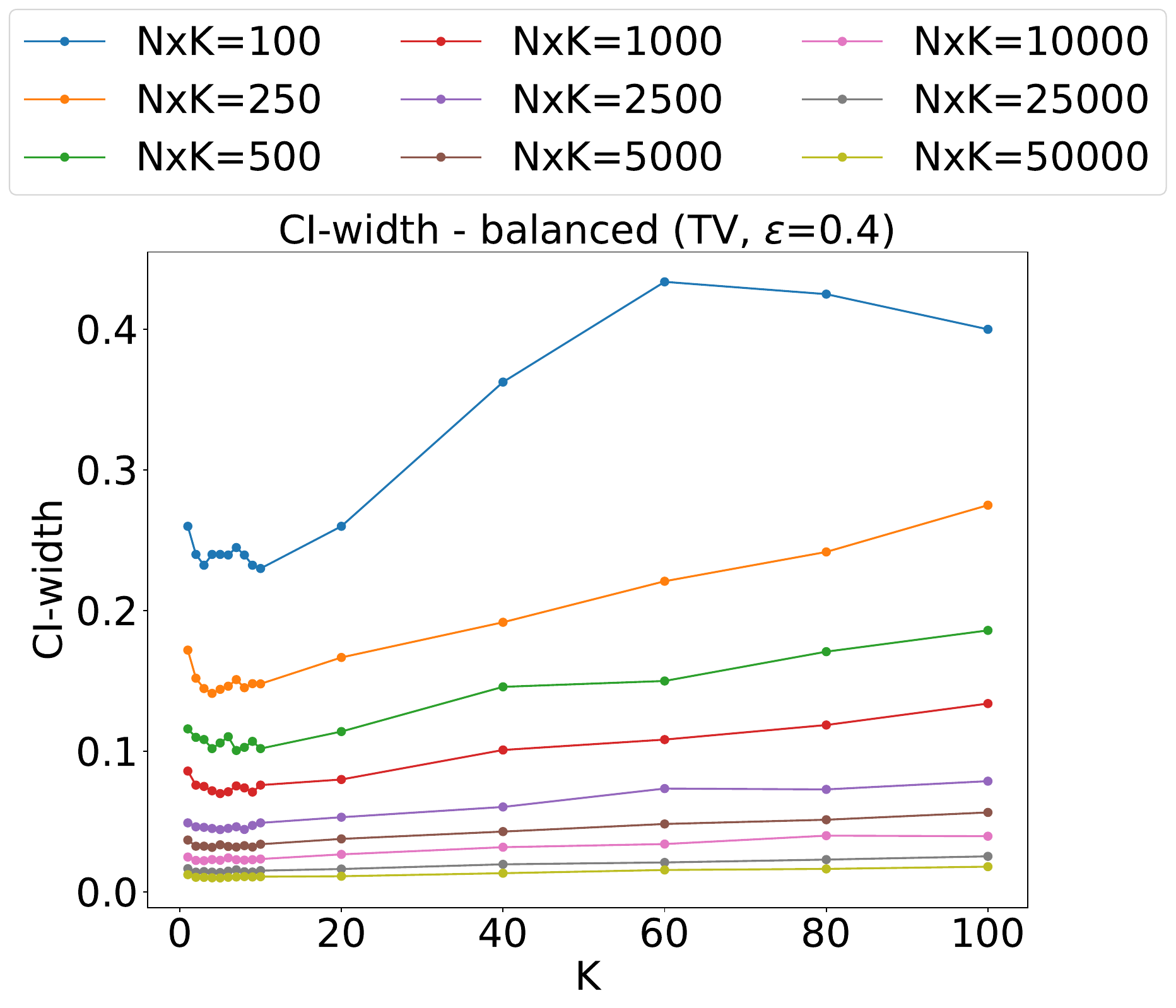}
    \caption{$\epsilon = 0.4$}
    \label{fig:uniform_ci_MAE_cat2_e04}
  \end{subfigure}
  \caption{CI-width plots for balanced alphas with TV as the metric ($M=2$)}
  \label{fig:uniform_ci_MAE_cat2}
\end{figure*}

\begin{figure*}
  \centering
  \begin{subfigure}[b]{0.24\linewidth}
    \centering
    \includegraphics[width=\linewidth]{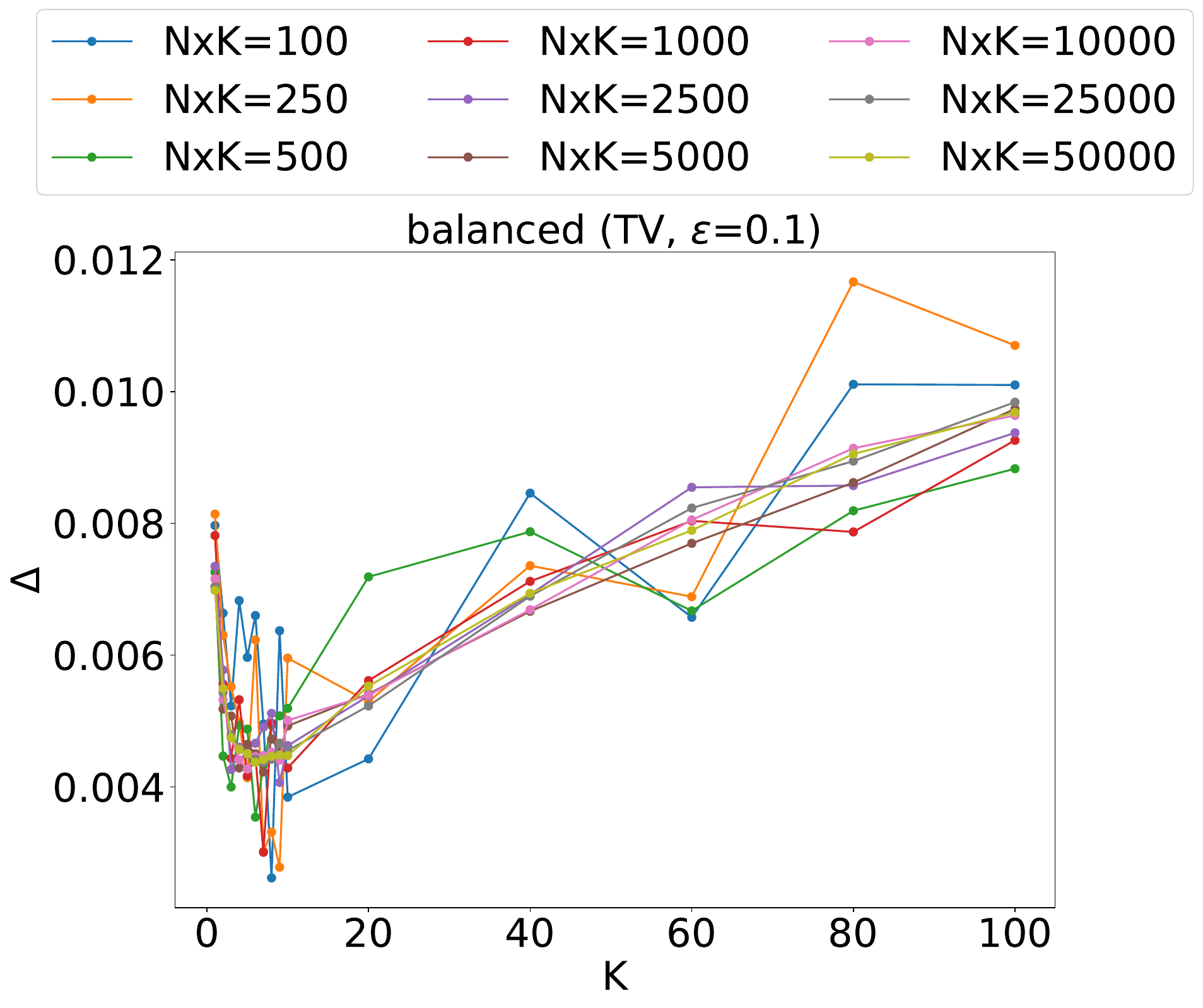}
    \caption{$\epsilon = 0.1$}
    \label{fig:uniform_delta_MAE_cat2_e01}
  \end{subfigure} \hfill
  \begin{subfigure}[b]{0.24\linewidth}
    \centering
    \includegraphics[width=\linewidth]{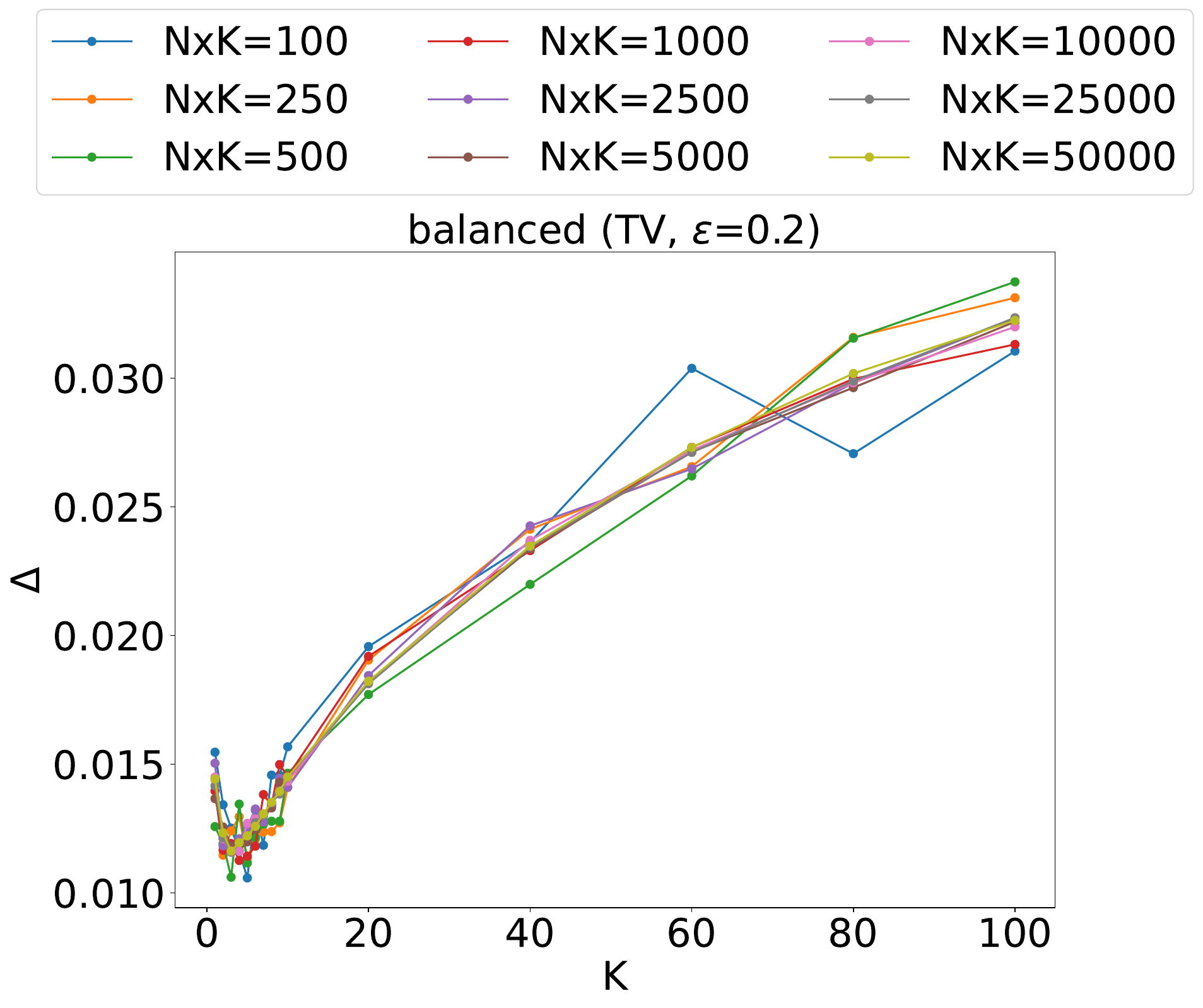}
    \caption{$\epsilon = 0.2$}
    \label{fig:uniform_delta_MAE_cat2_e02}
  \end{subfigure} \hfill
  \begin{subfigure}[b]{0.24\linewidth}
    \centering
    \includegraphics[width=\linewidth]{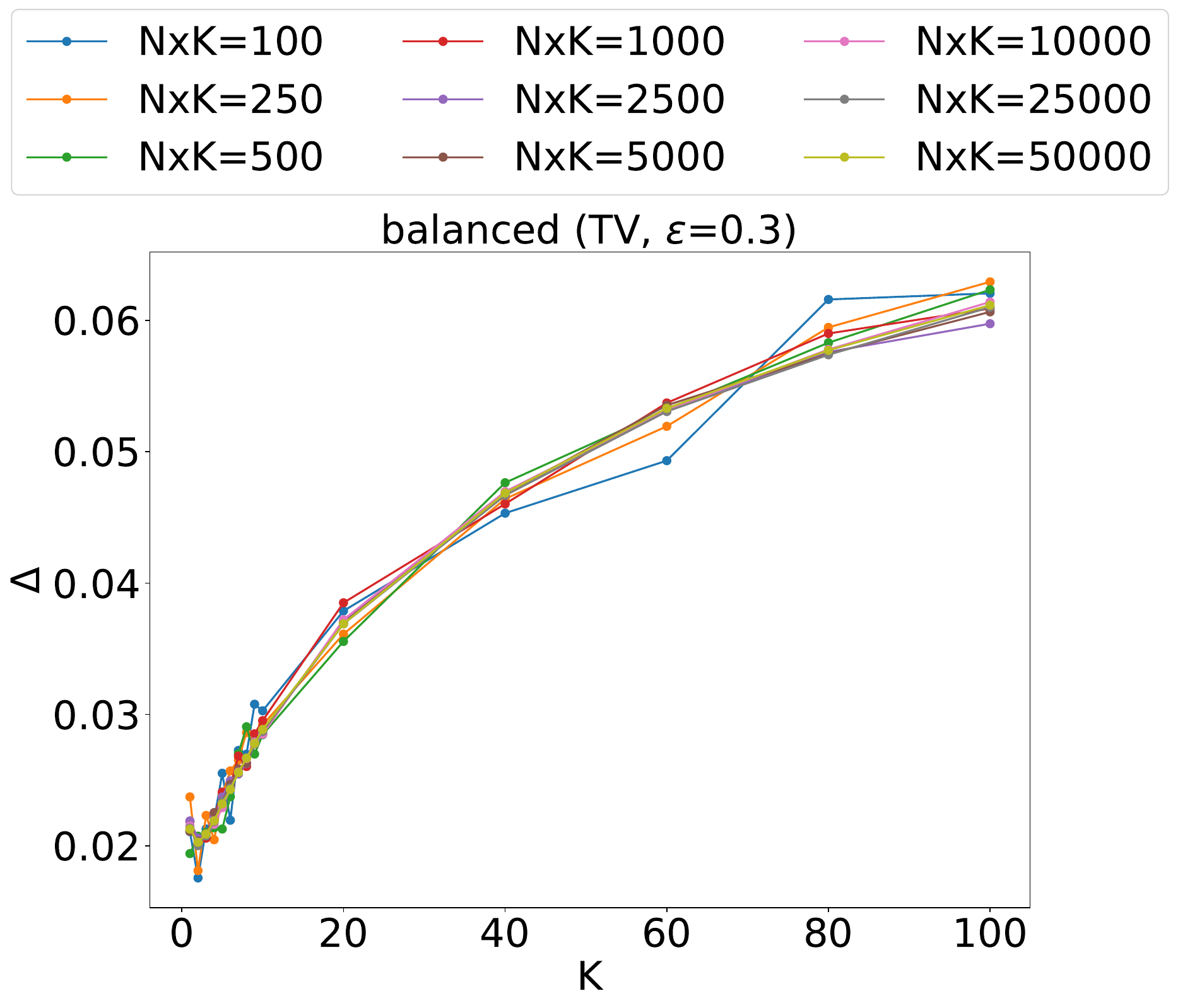}
    \caption{$\epsilon = 0.3$}
    \label{fig:uniform_delta_MAE_cat2_e03}
  \end{subfigure} \hfill
  \begin{subfigure}[b]{0.24\linewidth}
    \centering
    \includegraphics[width=\linewidth]{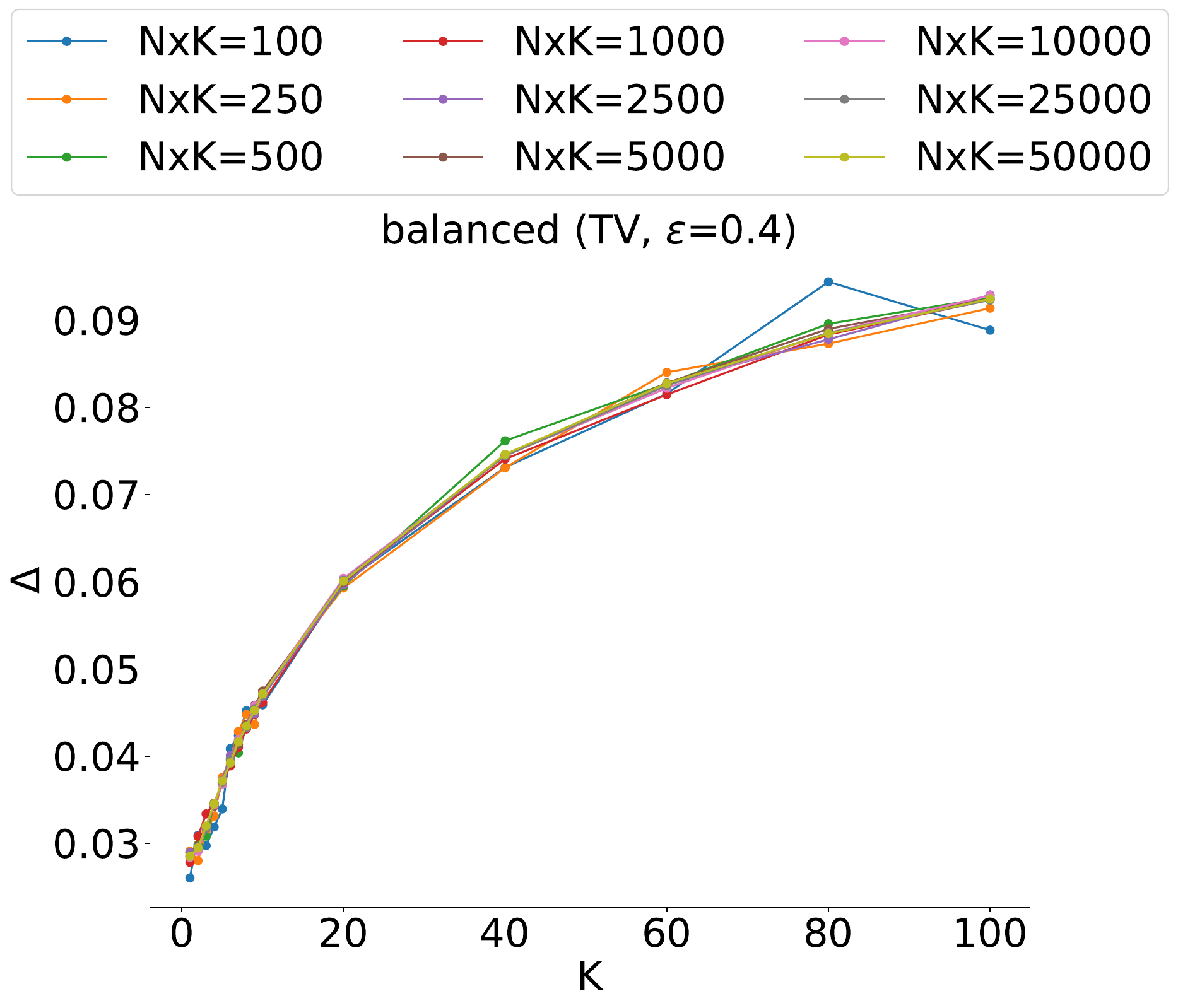}
    \caption{$\epsilon = 0.4$}
    \label{fig:uniform_delta_MAE_cat2_e04}
  \end{subfigure}
  \caption{Effect sizes ($\Delta$) for balanced alphas with TV as the metric ($M=2$)}
  \label{fig:uniform_delta_MAE_cat2}
\end{figure*}

\begin{figure*}
  \centering
  \begin{subfigure}[b]{0.24\linewidth}
    \centering
    \includegraphics[width=\linewidth]{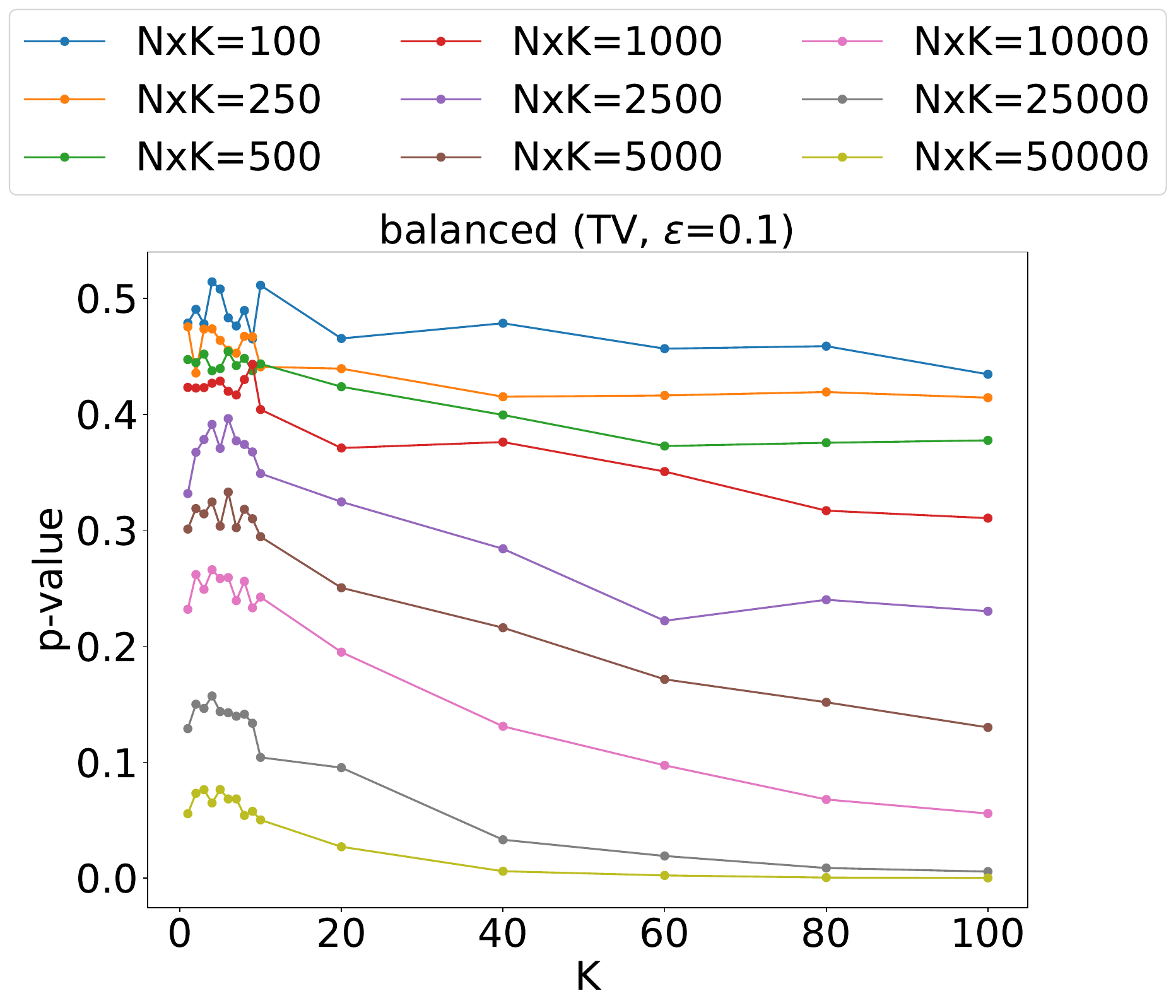}
    \caption{$\epsilon = 0.1$}
    \label{fig:uniform_MAE_cat3_e01}
  \end{subfigure} \hfill
  \begin{subfigure}[b]{0.24\linewidth}
    \centering
    \includegraphics[width=\linewidth]{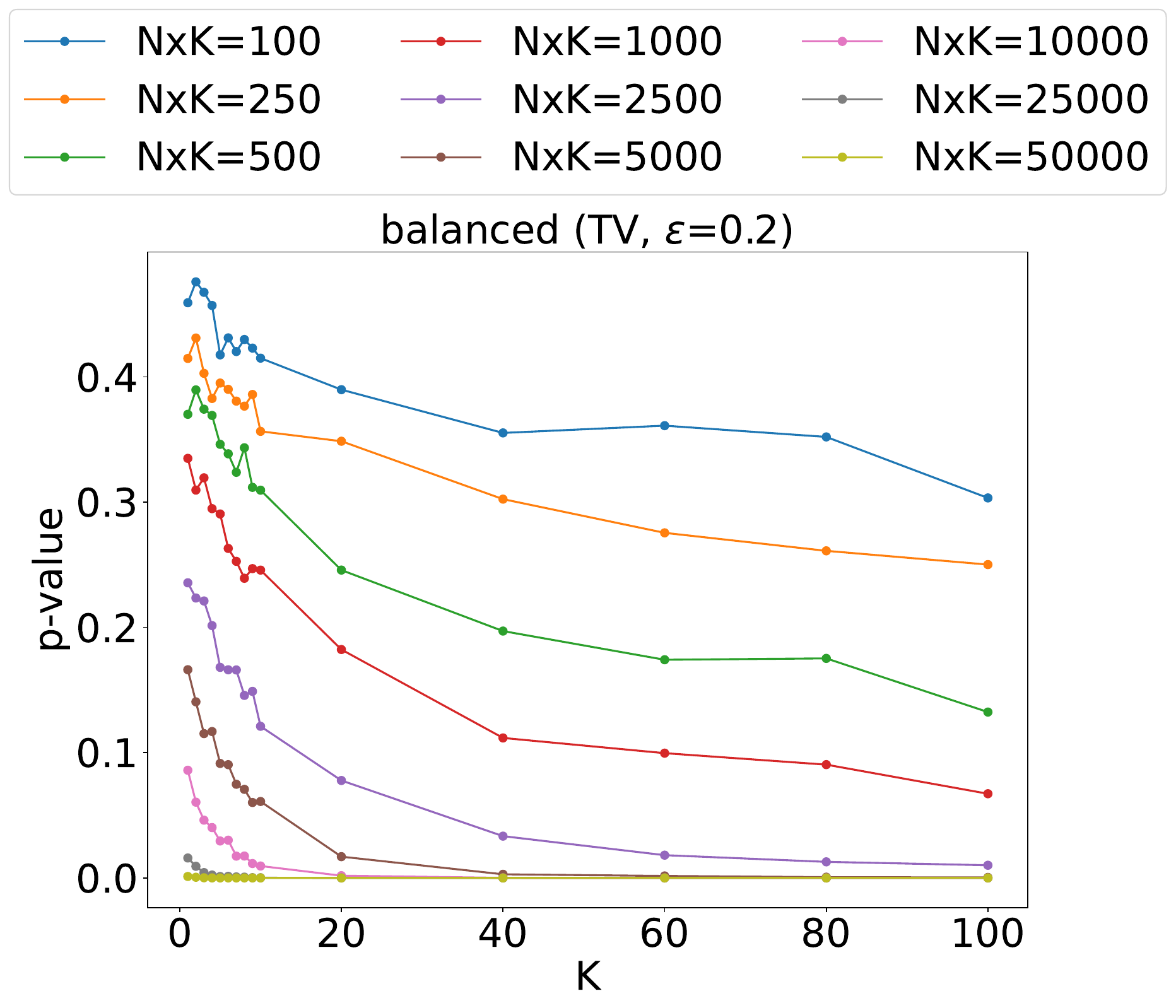}
    \caption{$\epsilon = 0.2$}
    \label{fig:uniform_MAE_cat3_e02}
  \end{subfigure} \hfill
  \begin{subfigure}[b]{0.24\linewidth}
    \centering
    \includegraphics[width=\linewidth]{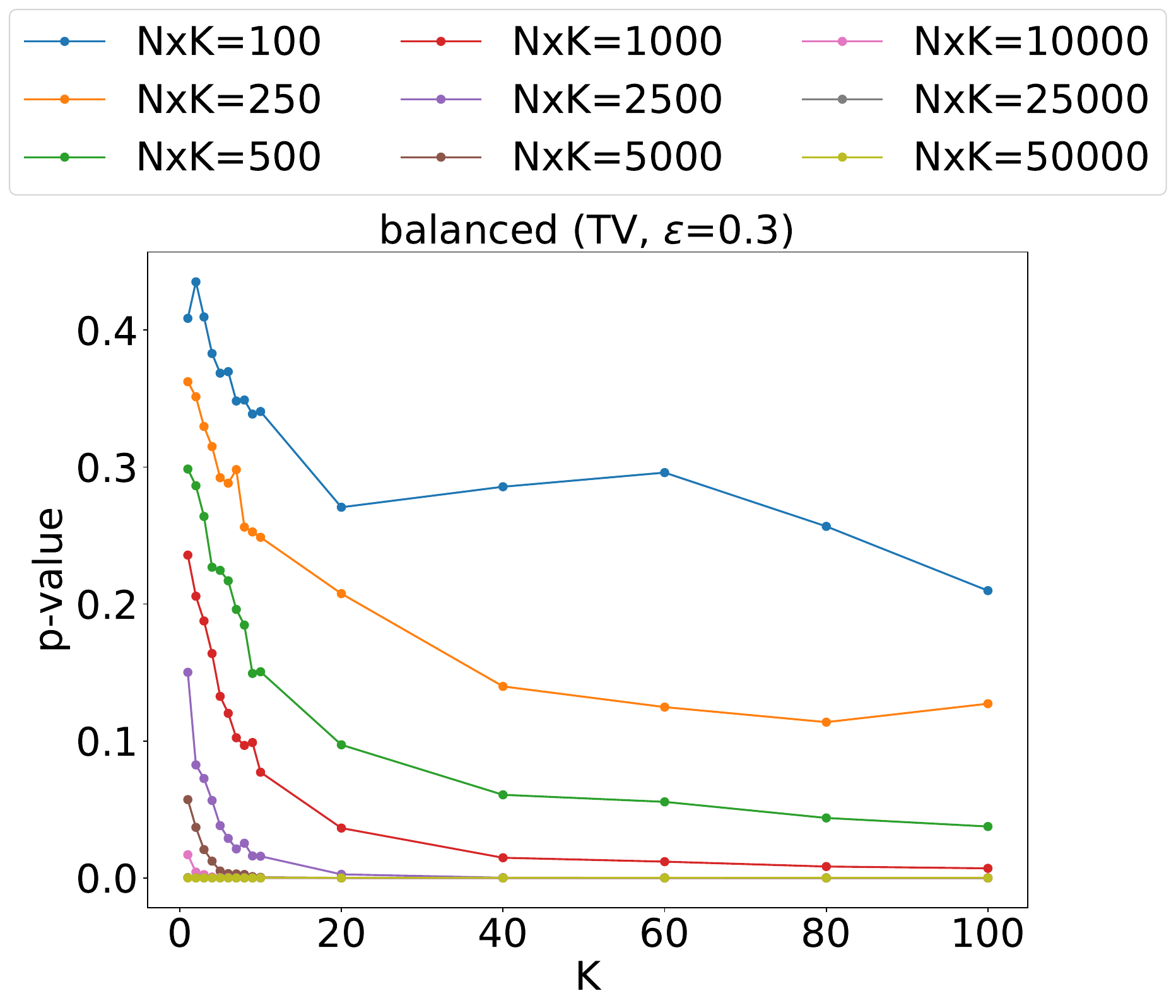}
    \caption{$\epsilon = 0.3$}
    \label{fig:uniform_MAE_cat3_e03}
  \end{subfigure} \hfill
  \begin{subfigure}[b]{0.24\linewidth}
    \centering
    \includegraphics[width=\linewidth]{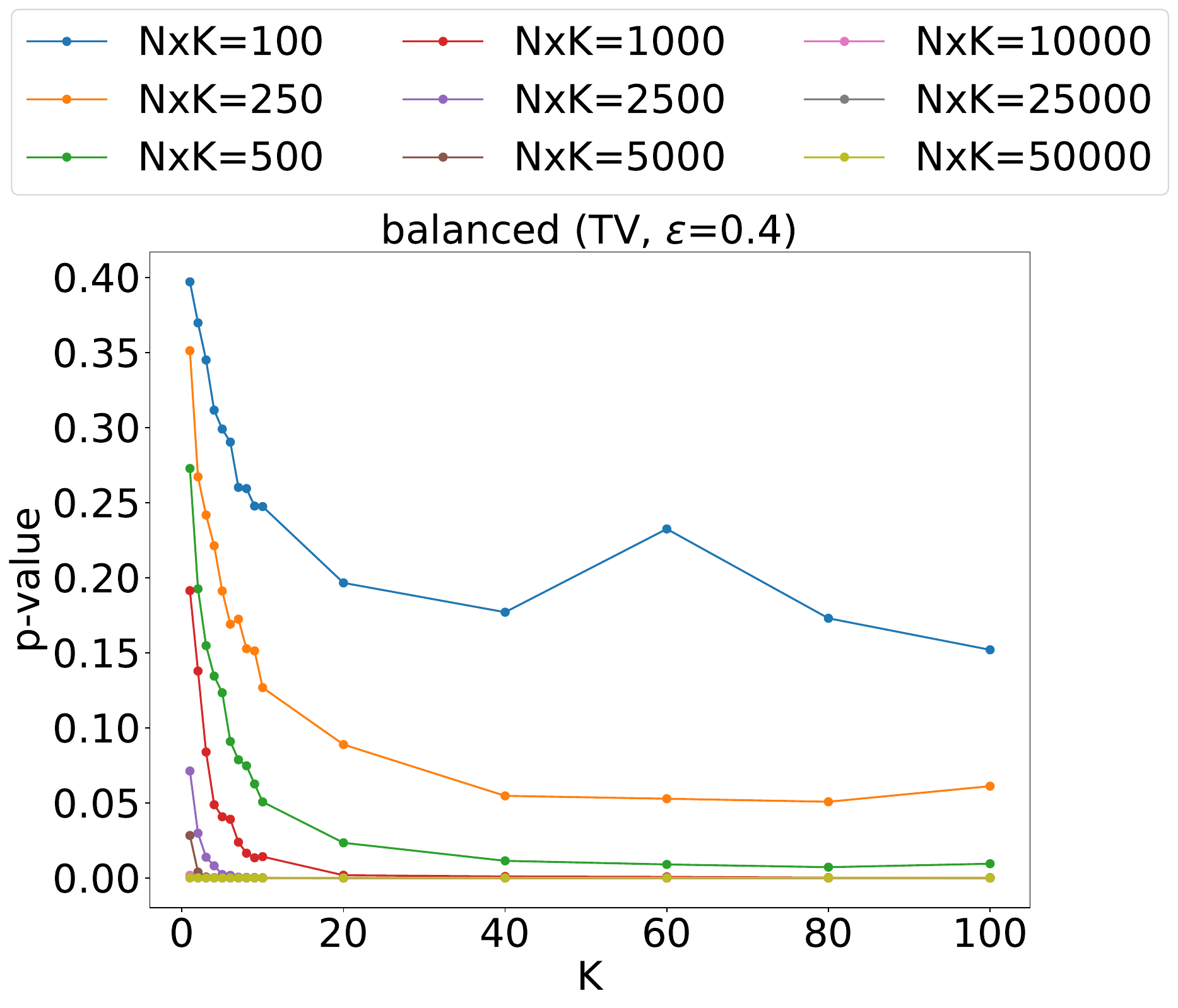}
    \caption{$\epsilon = 0.4$}
    \label{fig:uniform_MAE_cat3_e04}
  \end{subfigure}
  \caption{P-value plots for balanced alphas with TV as the metric ($M=3$)}
  \label{fig:uniform_MAE_cat3}
\end{figure*}

\begin{figure*}
  \centering
  \begin{subfigure}[b]{0.24\linewidth}
    \centering
    \includegraphics[width=\linewidth]{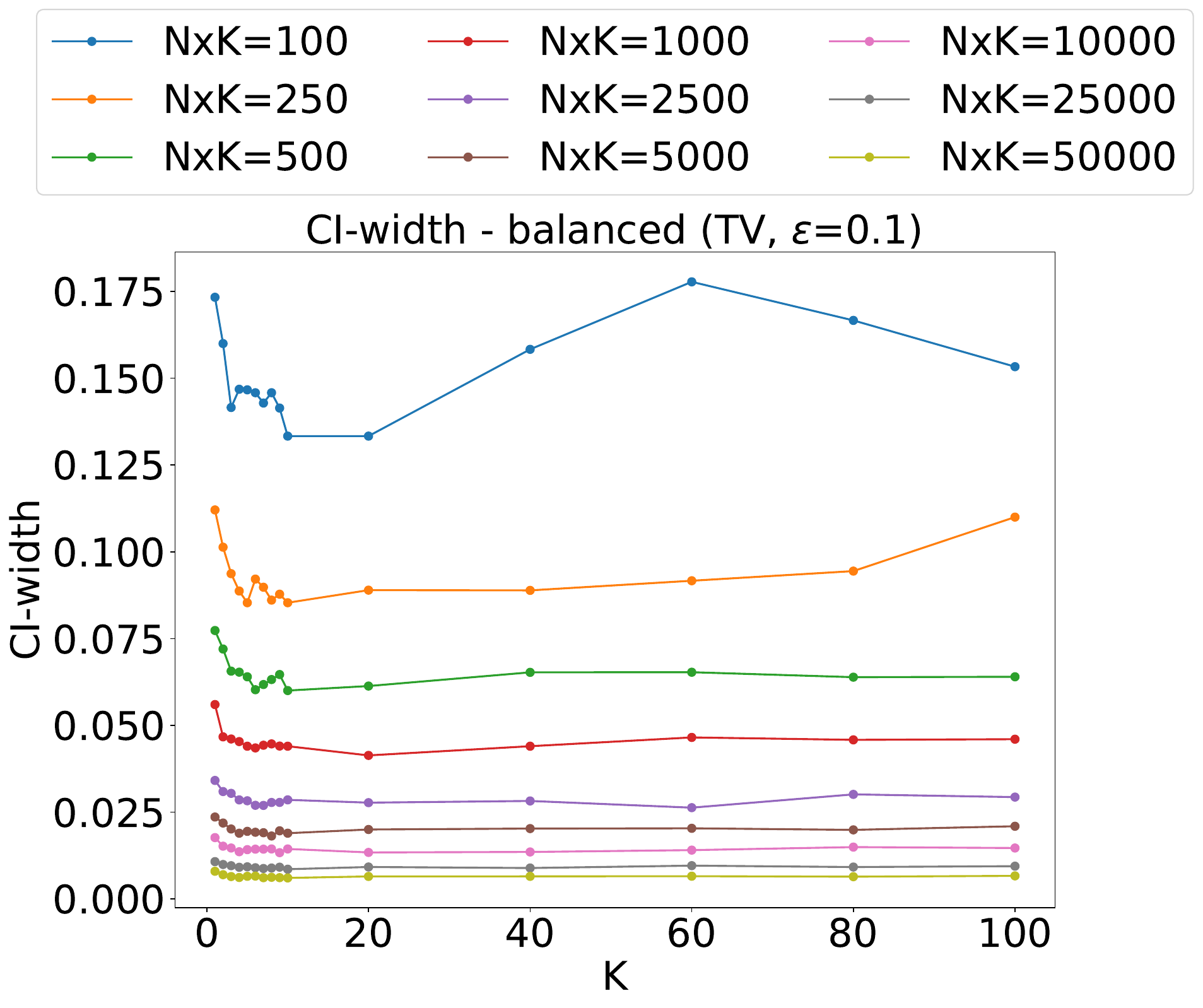}
    \caption{$\epsilon = 0.1$}
    \label{fig:uniform_ci_MAE_cat3_e01}
  \end{subfigure} \hfill
  \begin{subfigure}[b]{0.24\linewidth}
    \centering
    \includegraphics[width=\linewidth]{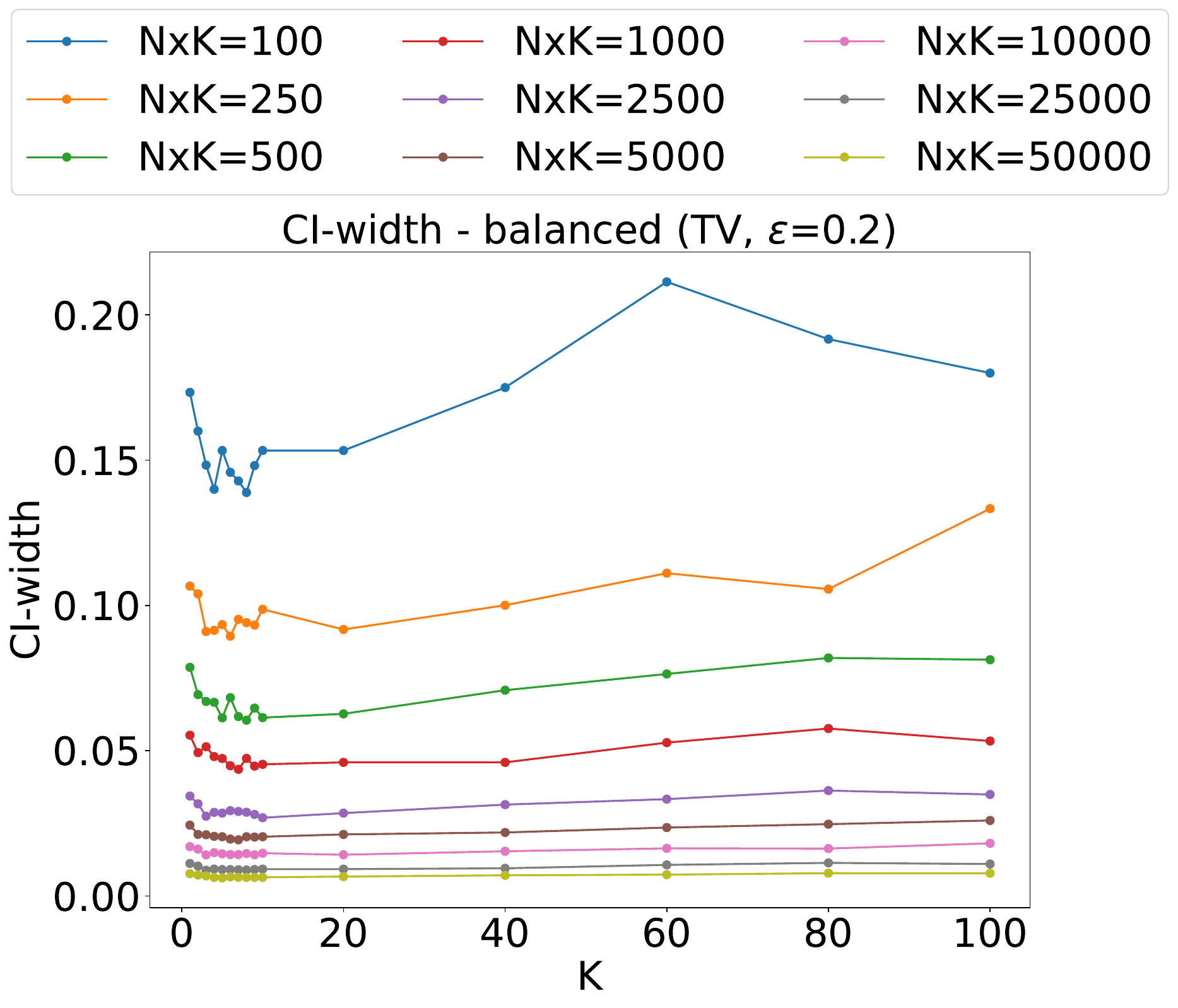}
    \caption{$\epsilon = 0.2$}
    \label{fig:uniform_ci_MAE_cat3_e02}
  \end{subfigure} \hfill
  \begin{subfigure}[b]{0.24\linewidth}
    \centering
    \includegraphics[width=\linewidth]{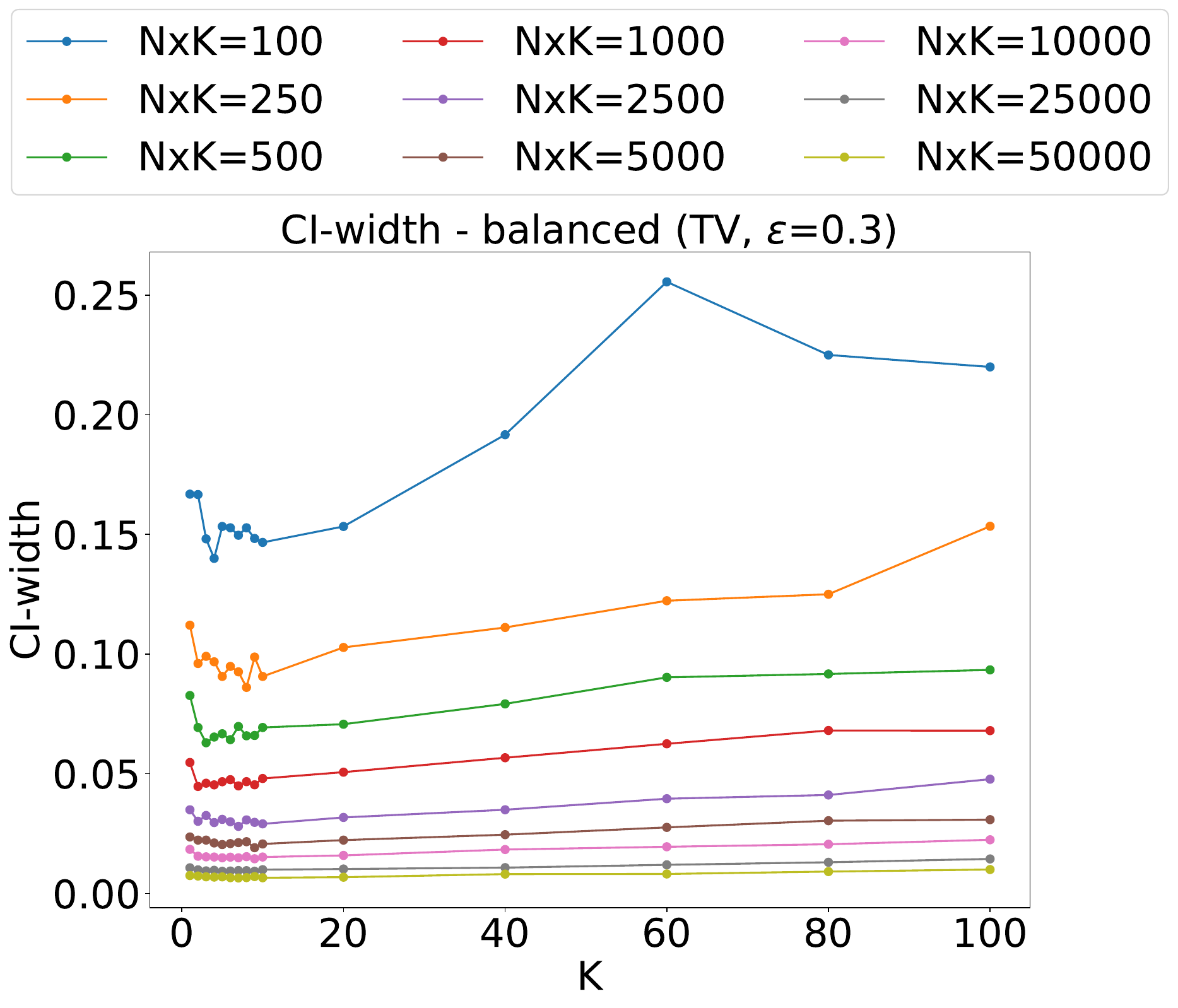}
    \caption{$\epsilon = 0.3$}
    \label{fig:uniform_ci_MAE_cat3_e03}
  \end{subfigure} \hfill
  \begin{subfigure}[b]{0.24\linewidth}
    \centering
    \includegraphics[width=\linewidth]{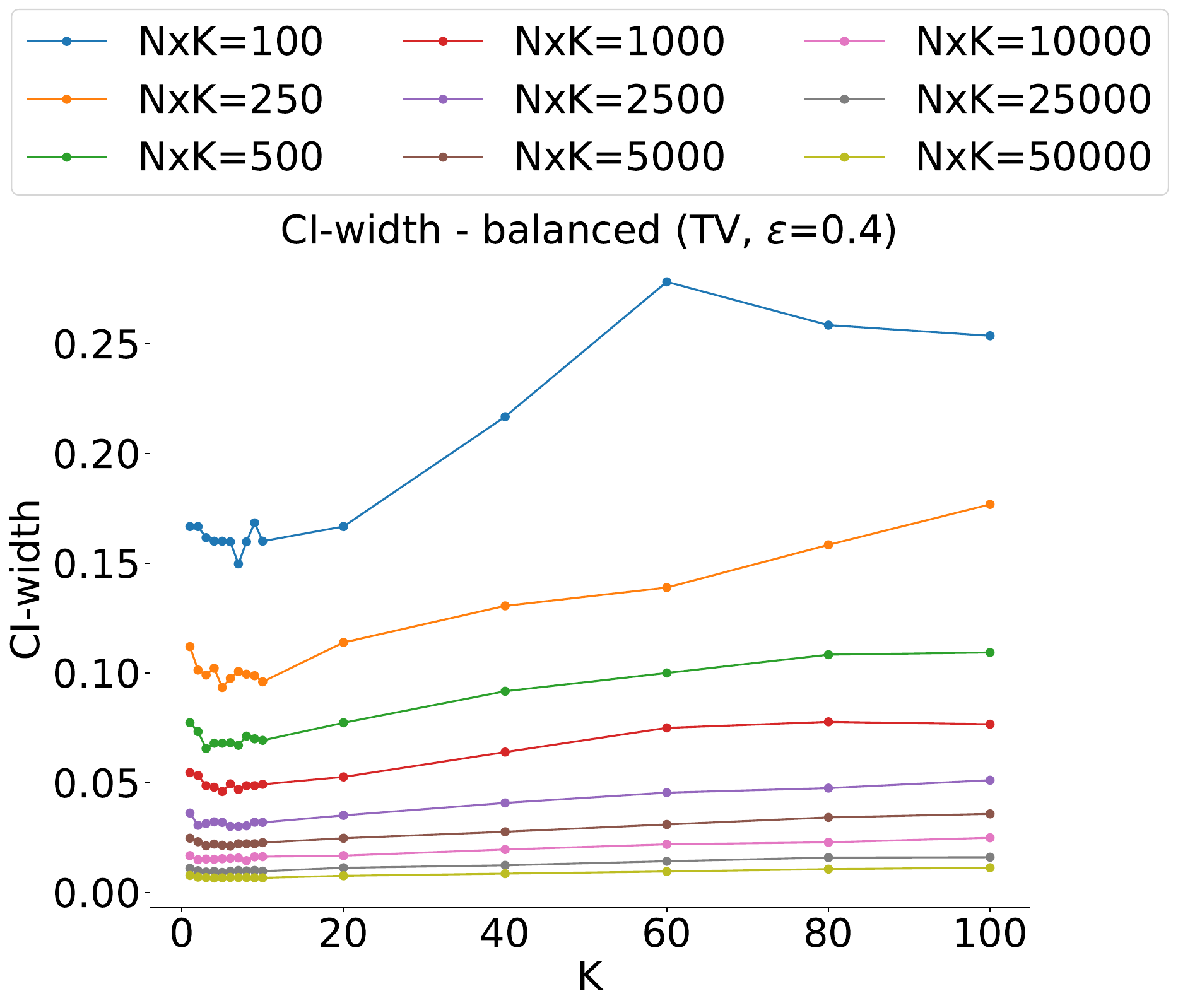}
    \caption{$\epsilon = 0.4$}
    \label{fig:uniform_ci_MAE_cat3_e04}
  \end{subfigure}
  \caption{CI-width plots for balanced alphas with TV as the metric ($M=3$)}
  \label{fig:uniform_ci_MAE_cat3}
\end{figure*}

\begin{figure*}
  \centering
  \begin{subfigure}[b]{0.24\linewidth}
    \centering
    \includegraphics[width=\linewidth]{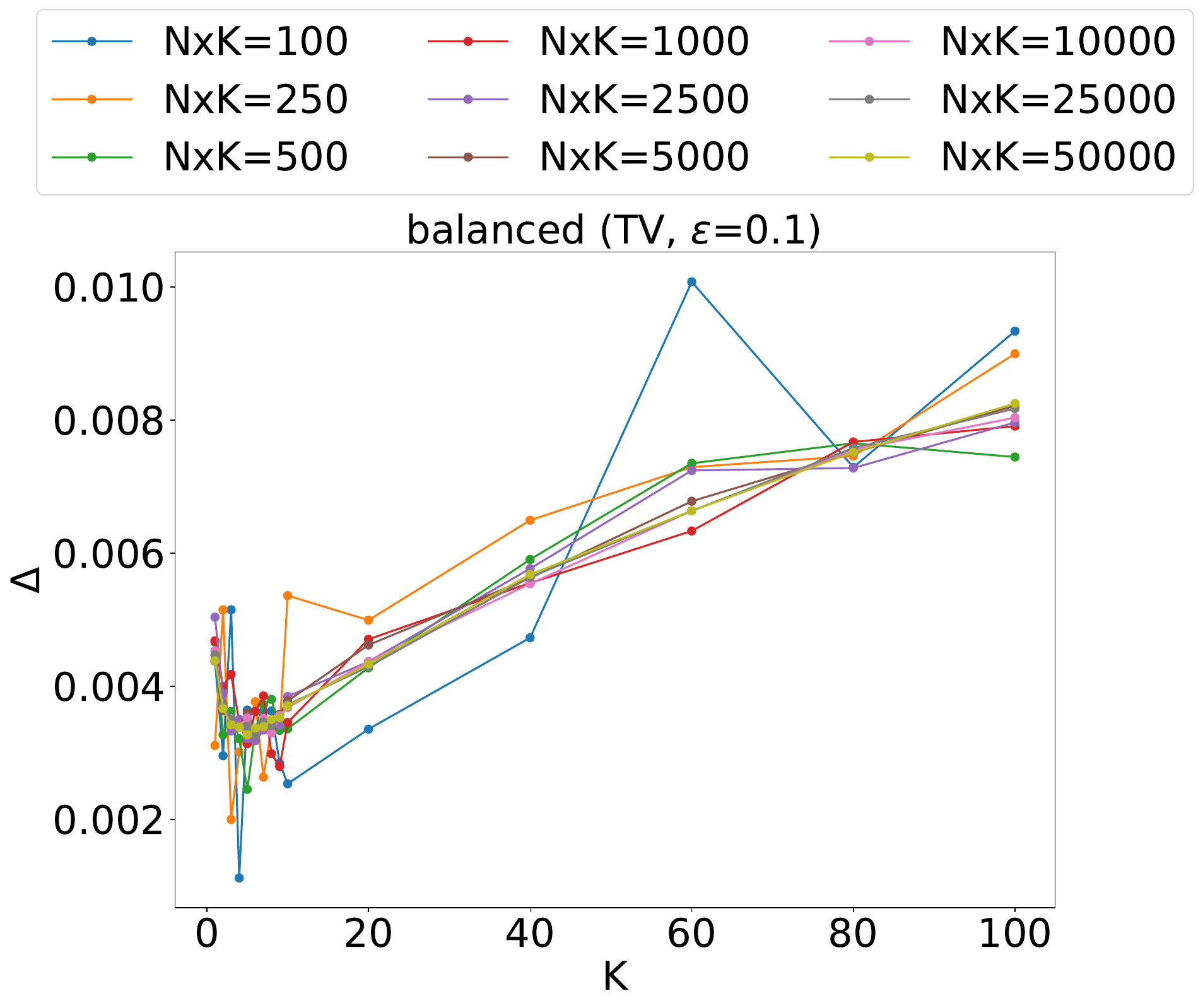}
    \caption{$\epsilon = 0.1$}
    \label{fig:uniform_delta_MAE_cat3_e01}
  \end{subfigure} \hfill
  \begin{subfigure}[b]{0.24\linewidth}
    \centering
    \includegraphics[width=\linewidth]{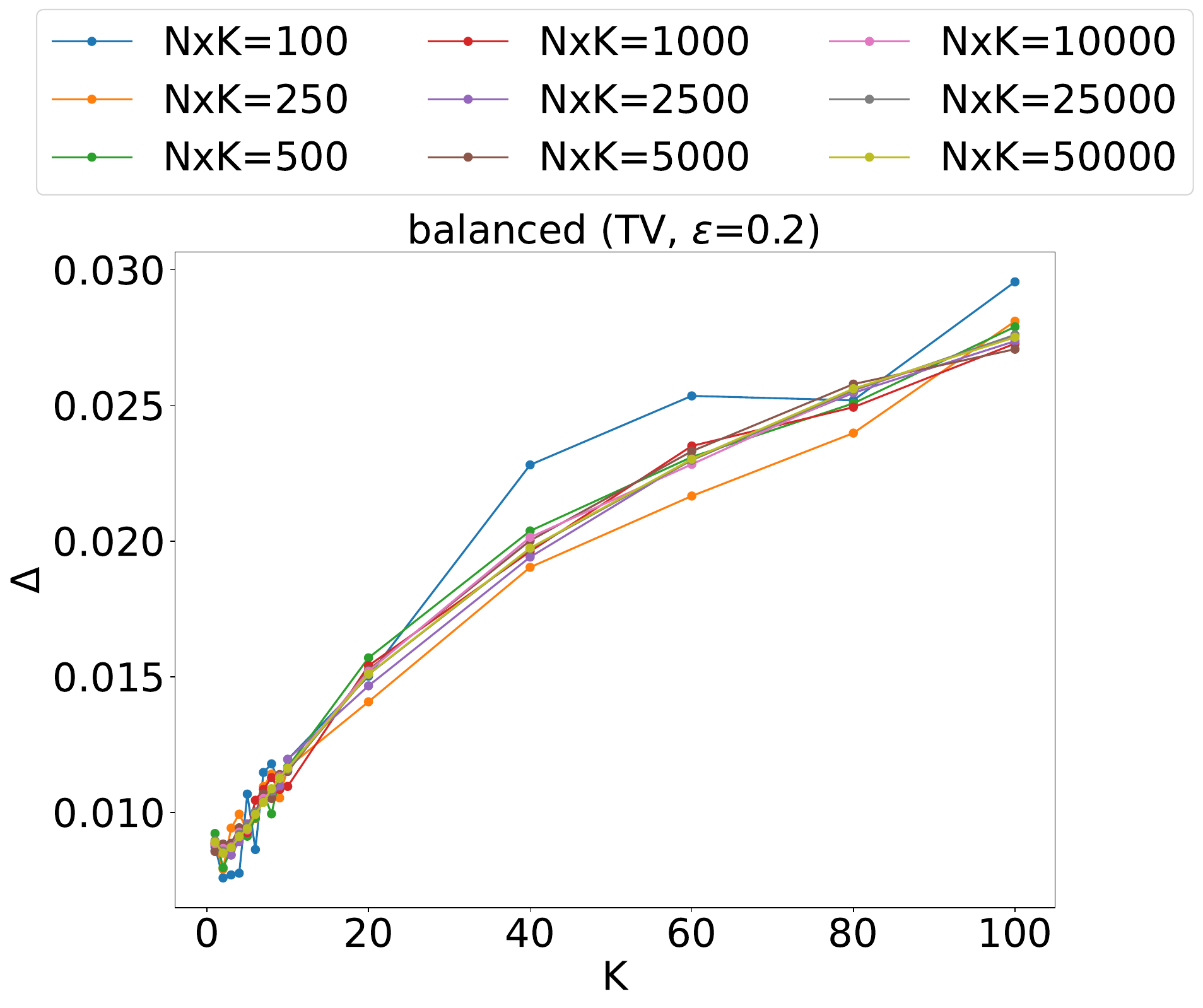}
    \caption{$\epsilon = 0.2$}
    \label{fig:uniform_delta_MAE_cat3_e02}
  \end{subfigure} \hfill
  \begin{subfigure}[b]{0.24\linewidth}
    \centering
    \includegraphics[width=\linewidth]{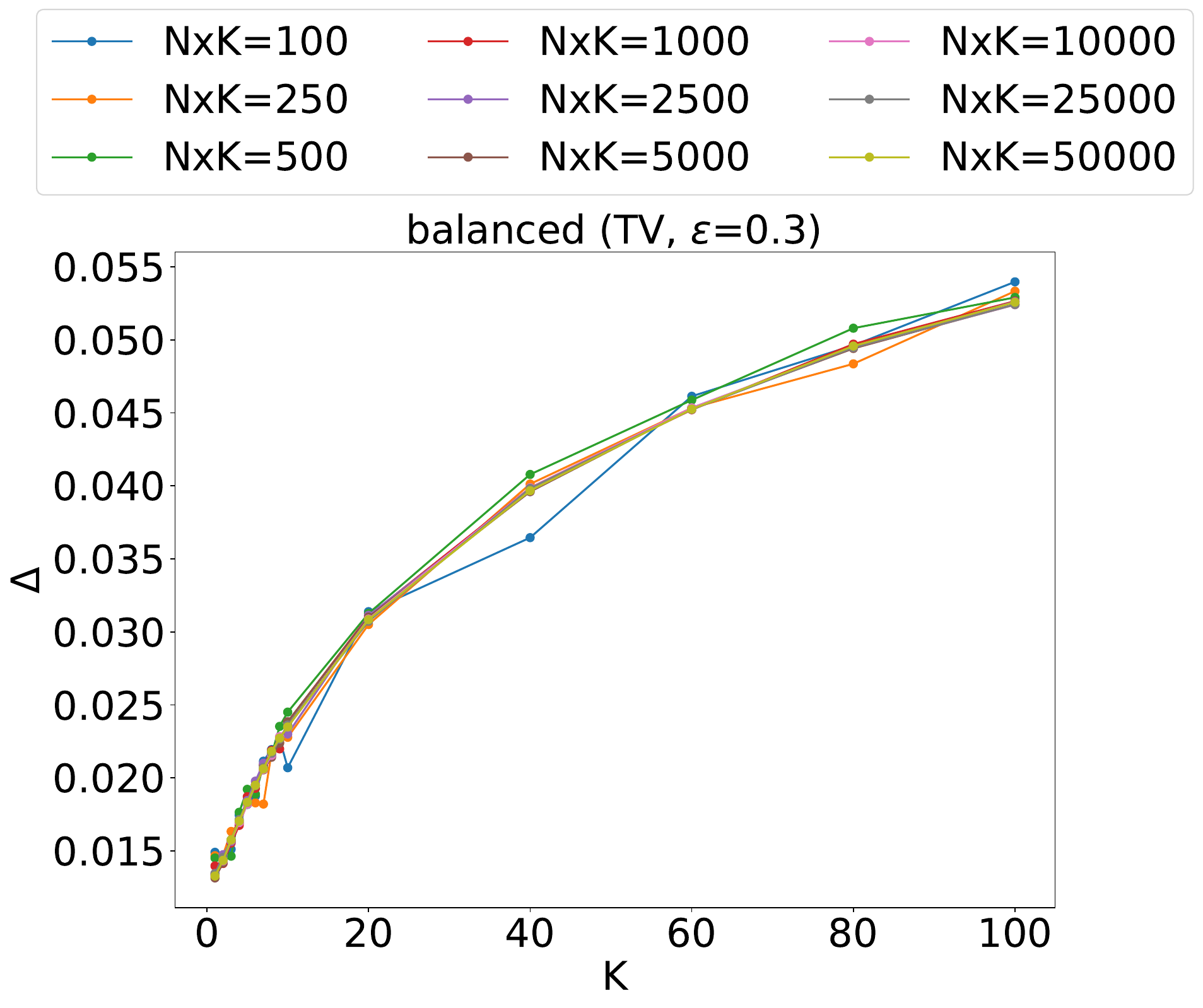}
    \caption{$\epsilon = 0.3$}
    \label{fig:uniform_delta_MAE_cat3_e03}
  \end{subfigure} \hfill
  \begin{subfigure}[b]{0.24\linewidth}
    \centering
    \includegraphics[width=\linewidth]{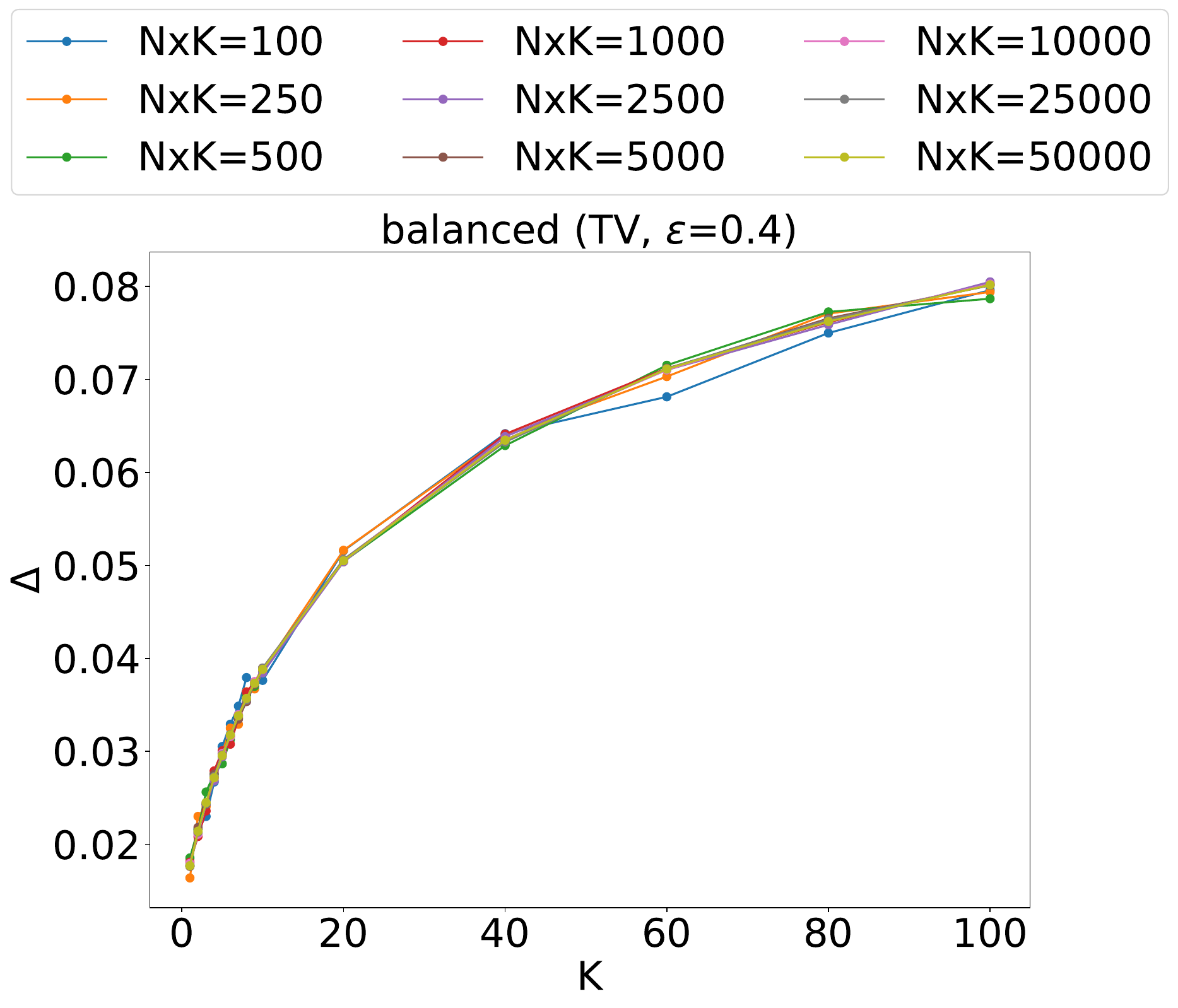}
    \caption{$\epsilon = 0.4$}
    \label{fig:uniform_delta_MAE_cat3_e04}
  \end{subfigure}
  \caption{Effect sizes ($\Delta$) for balanced alphas with TV as the metric ($M=3$)}
  \label{fig:uniform_delta_MAE_cat3}
\end{figure*}

\begin{figure*}
  \centering
  \begin{subfigure}[b]{0.24\linewidth}
    \centering
    \includegraphics[width=\linewidth]{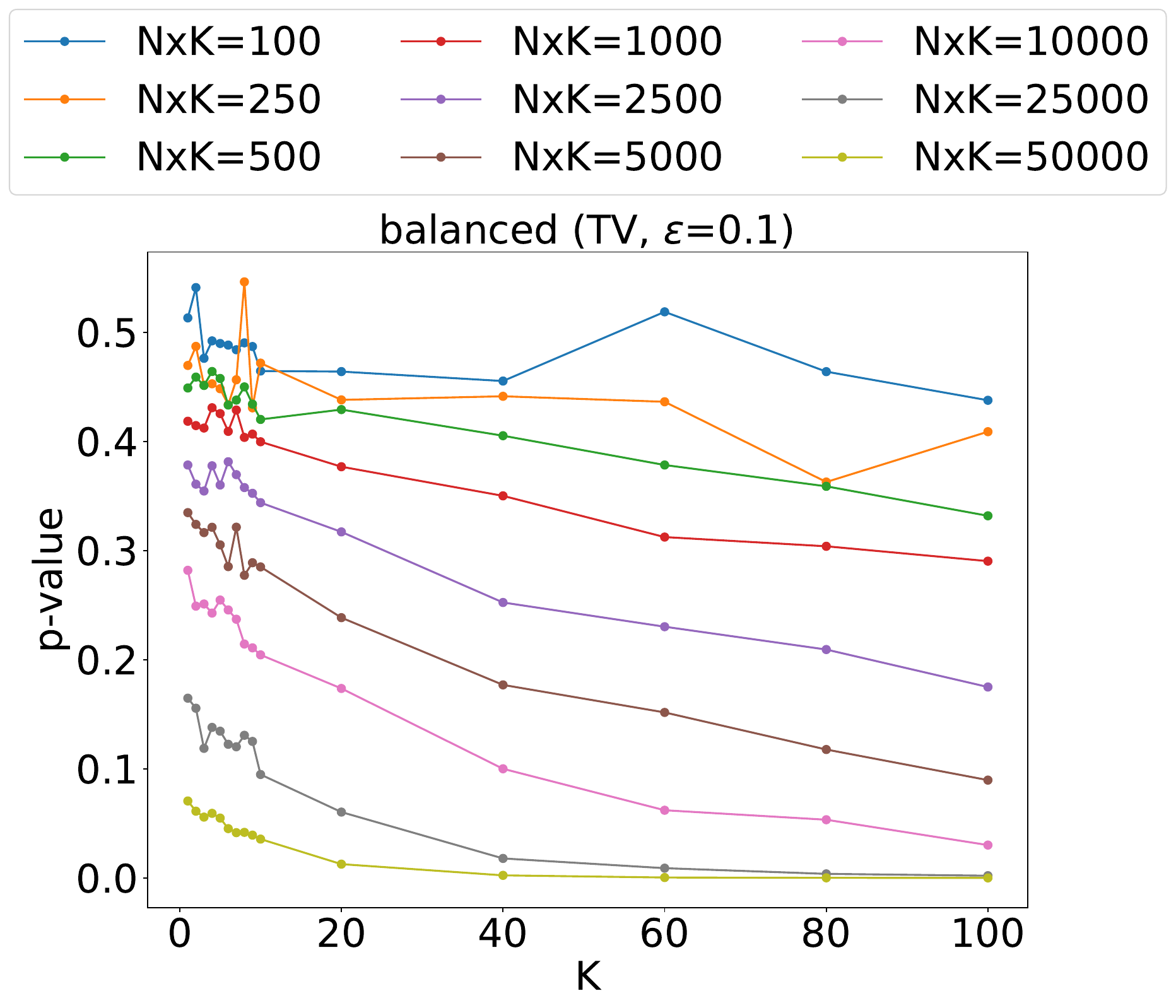}
    \caption{$\epsilon = 0.1$}
    \label{fig:uniform_MAE_cat4_e01}
  \end{subfigure} \hfill
  \begin{subfigure}[b]{0.24\linewidth}
    \centering
    \includegraphics[width=\linewidth]{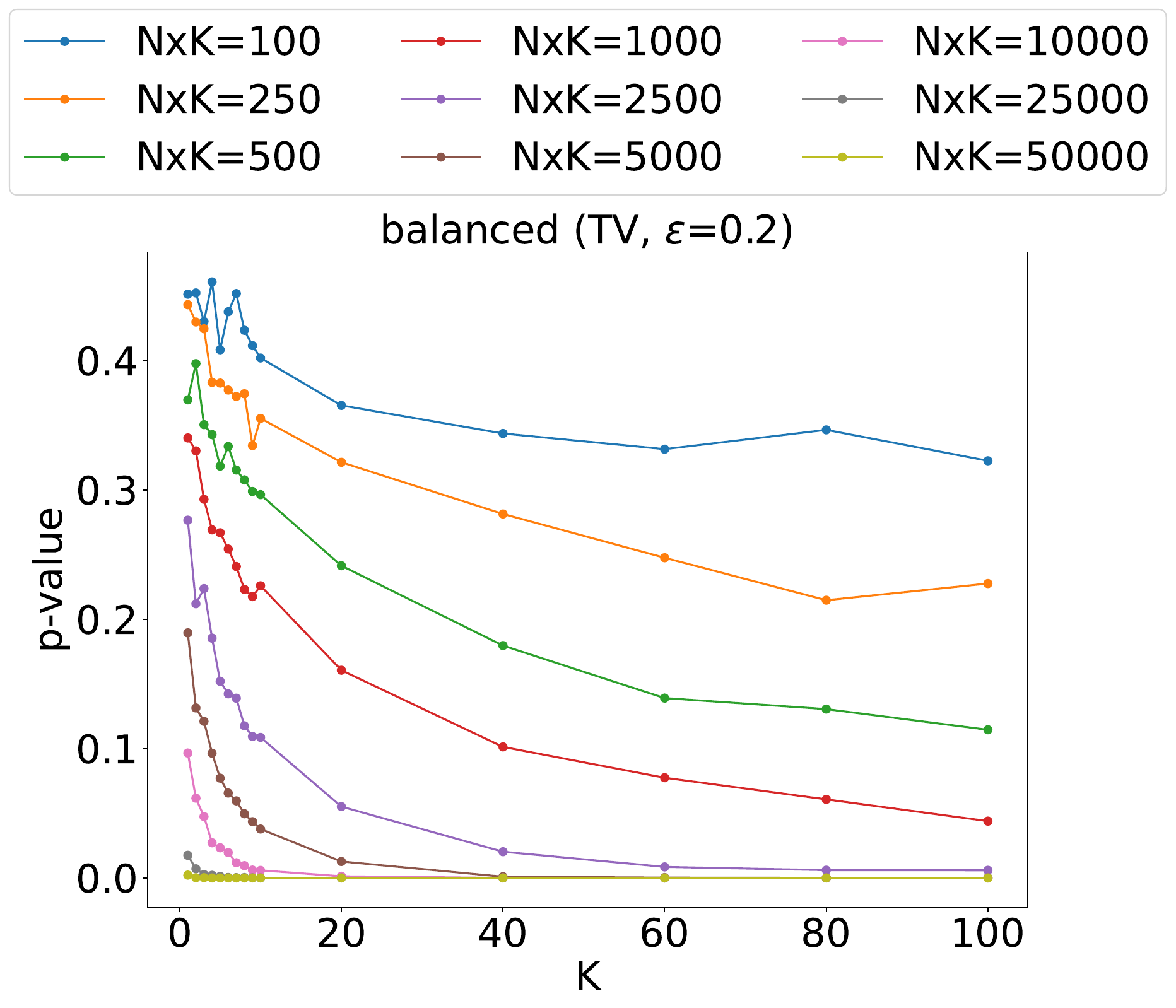}
    \caption{$\epsilon = 0.2$}
    \label{fig:uniform_MAE_cat4_e02}
  \end{subfigure} \hfill
  \begin{subfigure}[b]{0.24\linewidth}
    \centering
    \includegraphics[width=\linewidth]{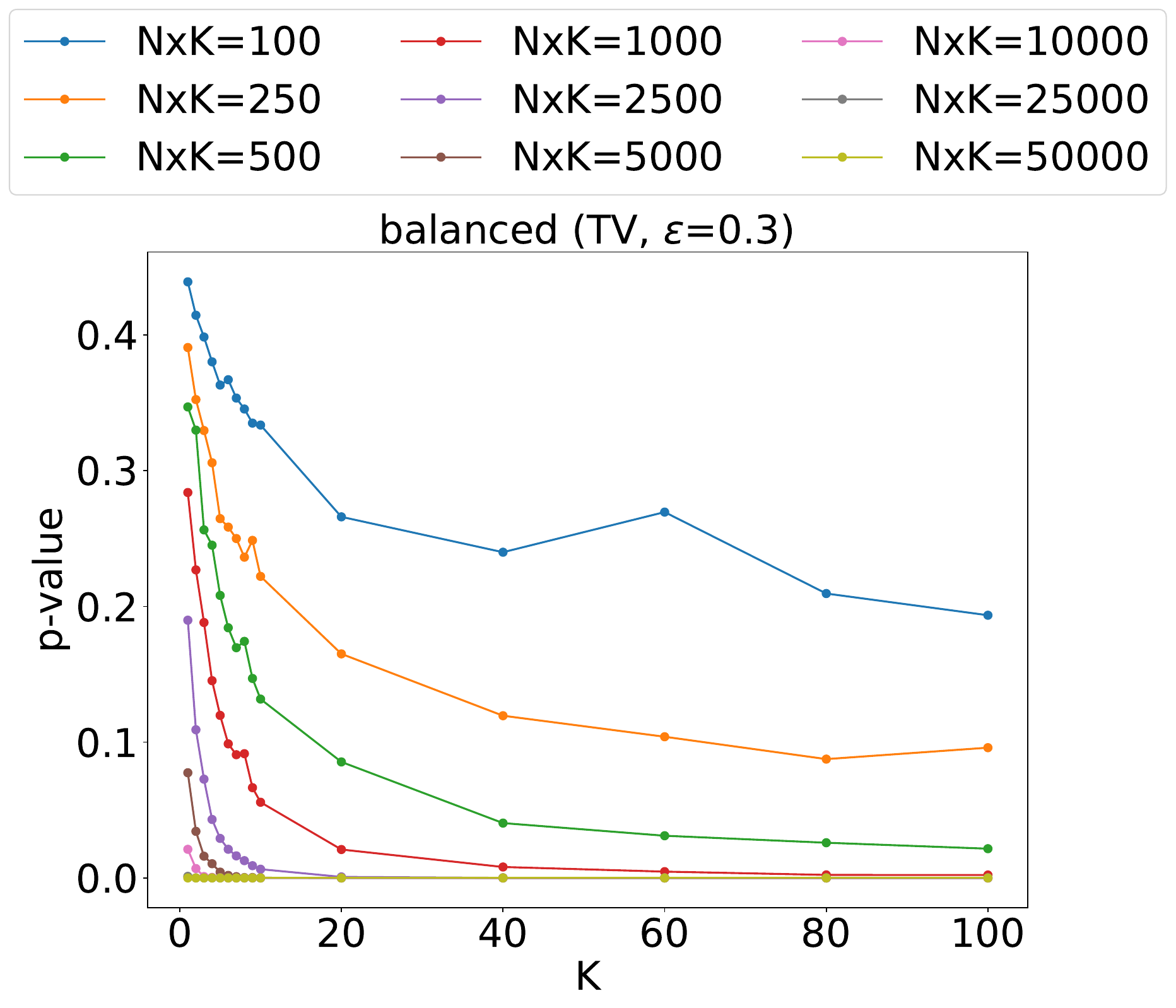}
    \caption{$\epsilon = 0.3$}
    \label{fig:uniform_MAE_cat4_e03}
  \end{subfigure} \hfill
  \begin{subfigure}[b]{0.24\linewidth}
    \centering
    \includegraphics[width=\linewidth]{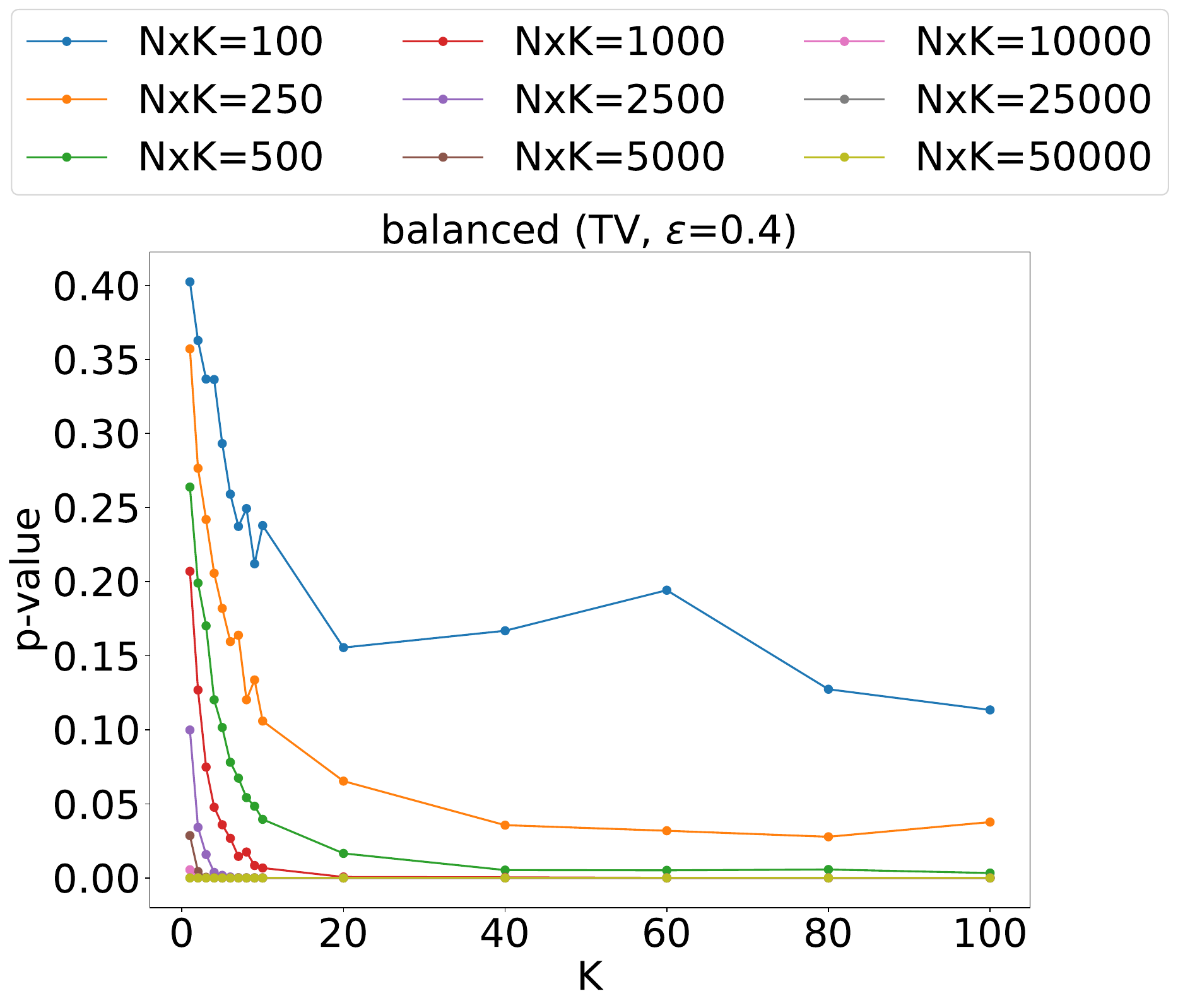}
    \caption{$\epsilon = 0.4$}
    \label{fig:uniform_MAE_cat4_e04}
  \end{subfigure}
  \caption{P-value plots for balanced alphas with TV as the metric ($M=4$)}
  \label{fig:uniform_MAE_cat4}
\end{figure*}

\begin{figure*}
  \centering
  \begin{subfigure}[b]{0.24\linewidth}
    \centering
    \includegraphics[width=\linewidth]{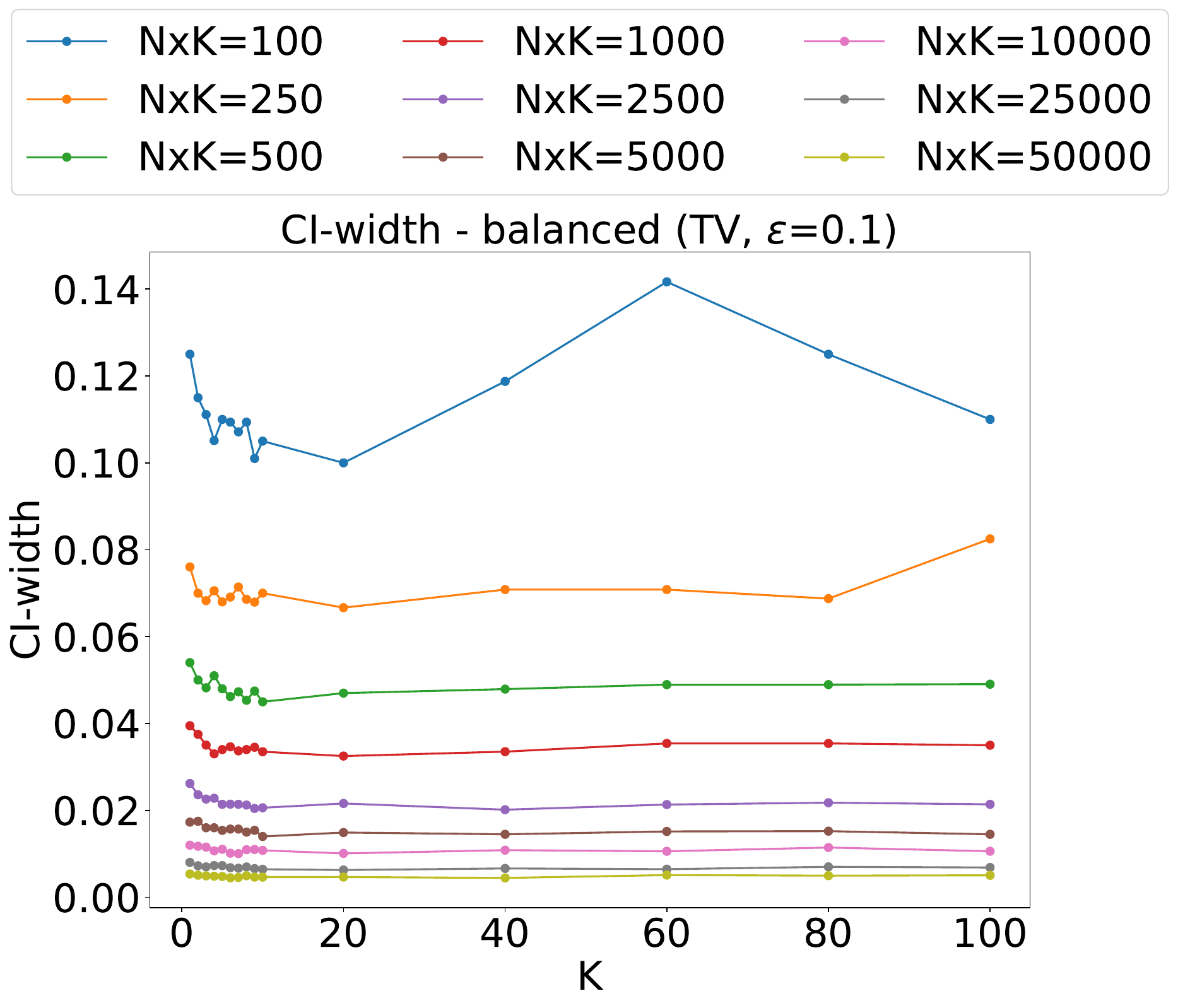}
    \caption{$\epsilon = 0.1$}
    \label{fig:uniform_ci_MAE_cat4_e01}
  \end{subfigure} \hfill
  \begin{subfigure}[b]{0.24\linewidth}
    \centering
    \includegraphics[width=\linewidth]{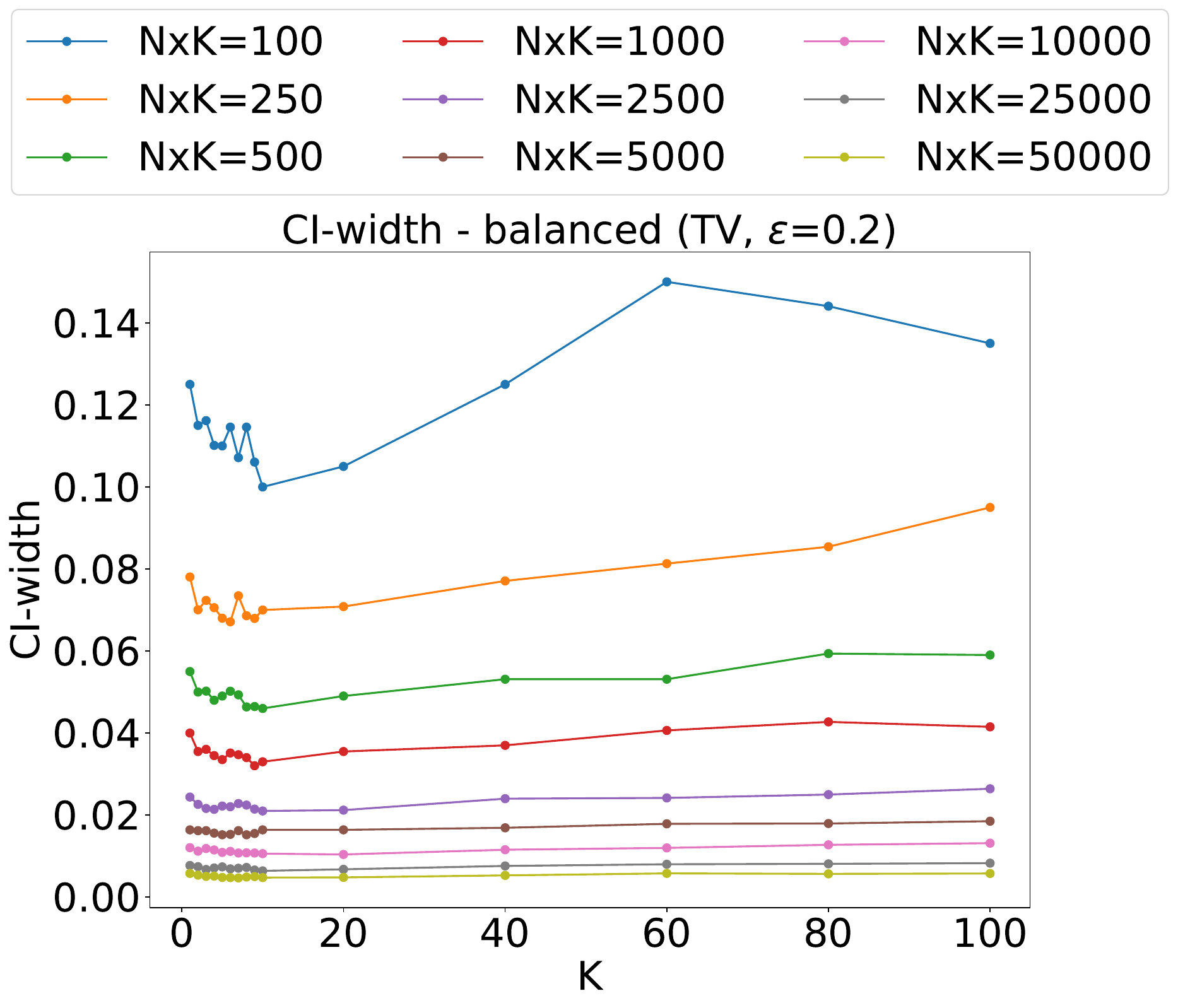}
    \caption{$\epsilon = 0.2$}
    \label{fig:uniform_ci_MAE_cat4_e02}
  \end{subfigure} \hfill
  \begin{subfigure}[b]{0.24\linewidth}
    \centering
    \includegraphics[width=\linewidth]{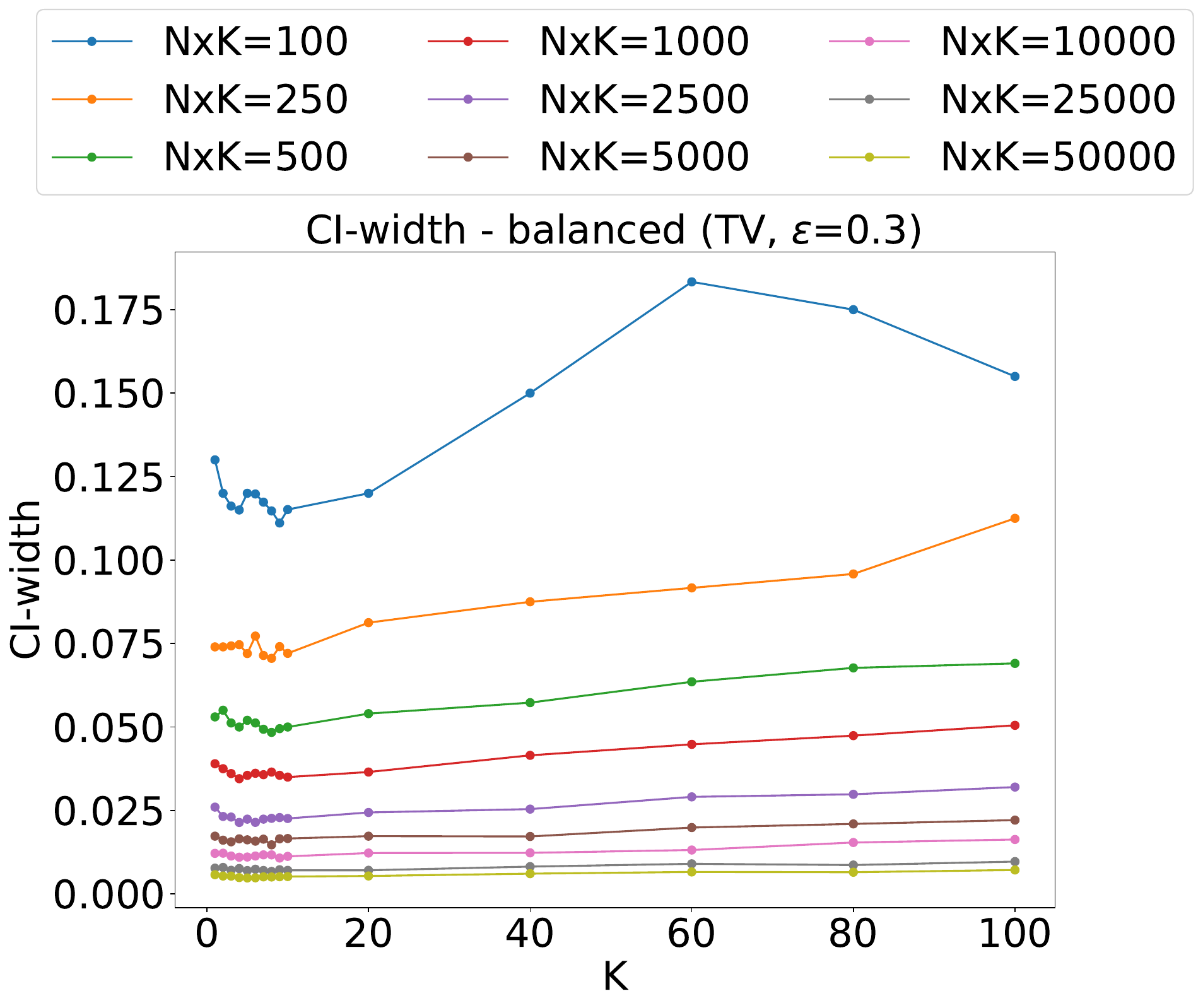}
    \caption{$\epsilon = 0.3$}
    \label{fig:uniform_ci_MAE_cat4_e03}
  \end{subfigure} \hfill
  \begin{subfigure}[b]{0.24\linewidth}
    \centering
    \includegraphics[width=\linewidth]{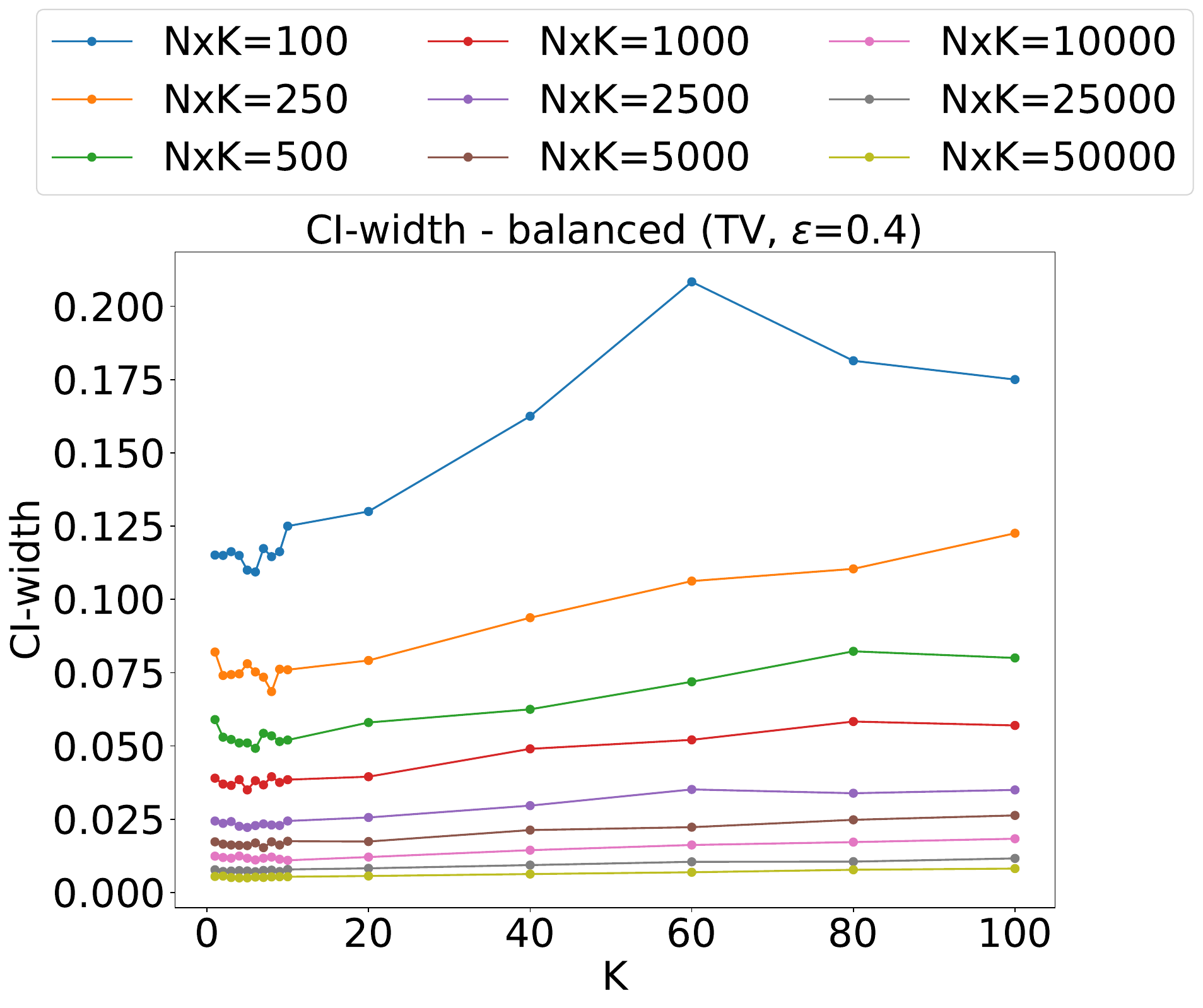}
    \caption{$\epsilon = 0.4$}
    \label{fig:uniform_ci_MAE_cat4_e04}
  \end{subfigure}
  \caption{CI-width plots for balanced alphas with TV as the metric ($M=4$)}
  \label{fig:uniform_ci_MAE_cat4}
\end{figure*}

\begin{figure*}
  \centering
  \begin{subfigure}[b]{0.24\linewidth}
    \centering
    \includegraphics[width=\linewidth]{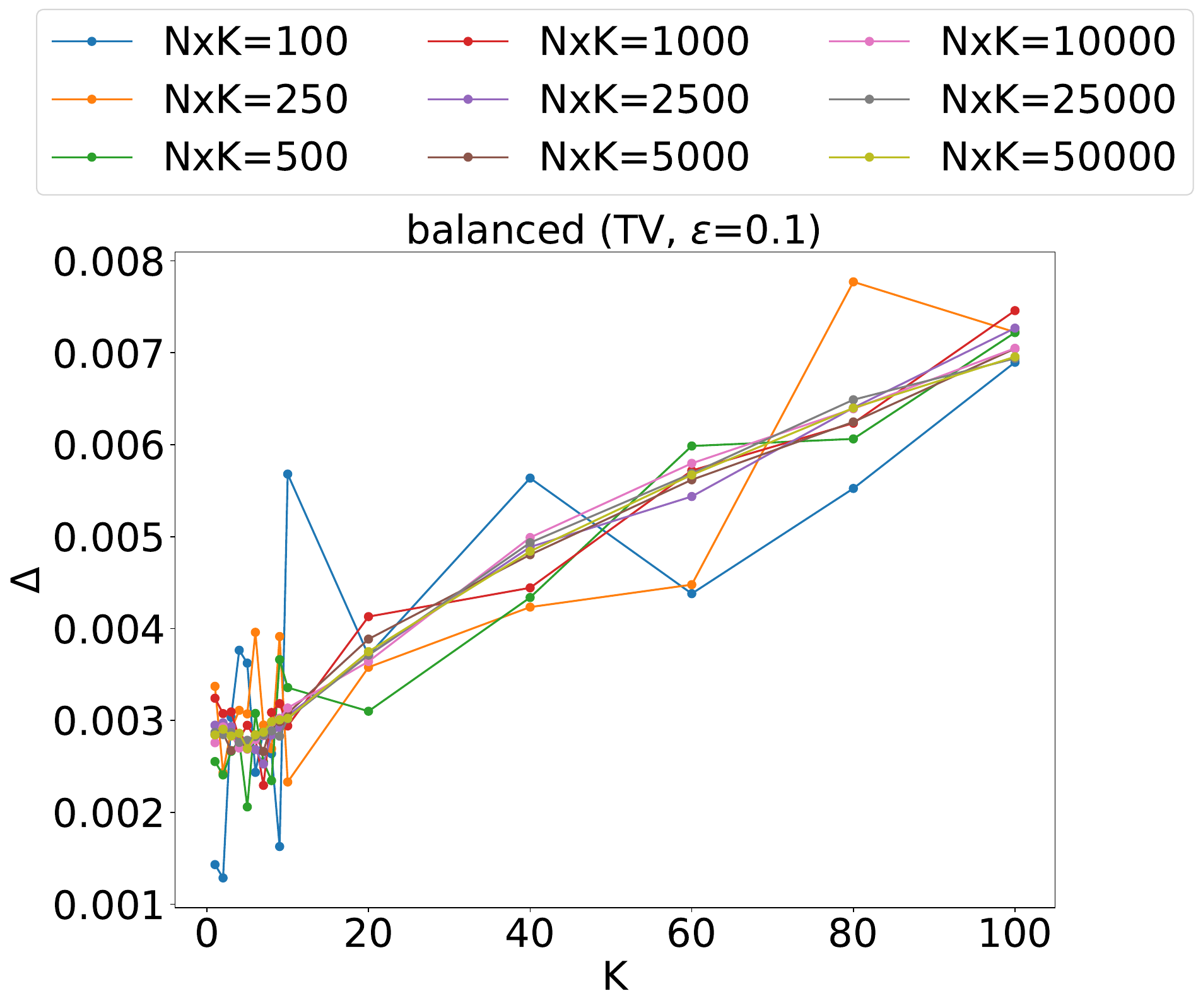}
    \caption{$\epsilon = 0.1$}
    \label{fig:uniform_delta_MAE_cat4_e01}
  \end{subfigure} \hfill
  \begin{subfigure}[b]{0.24\linewidth}
    \centering
    \includegraphics[width=\linewidth]{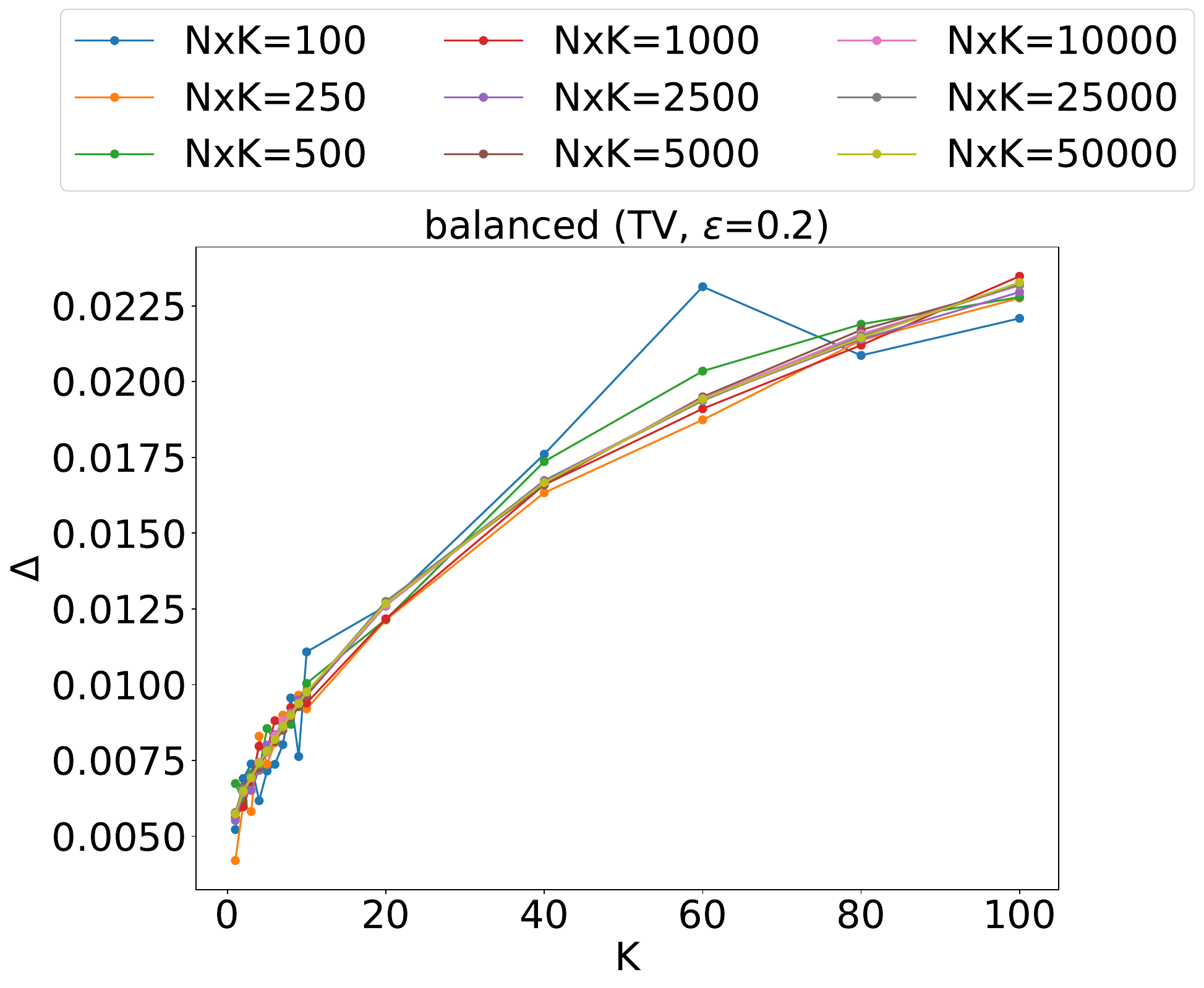}
    \caption{$\epsilon = 0.2$}
    \label{fig:uniform_delta_MAE_cat4_e02}
  \end{subfigure} \hfill
  \begin{subfigure}[b]{0.24\linewidth}
    \centering
    \includegraphics[width=\linewidth]{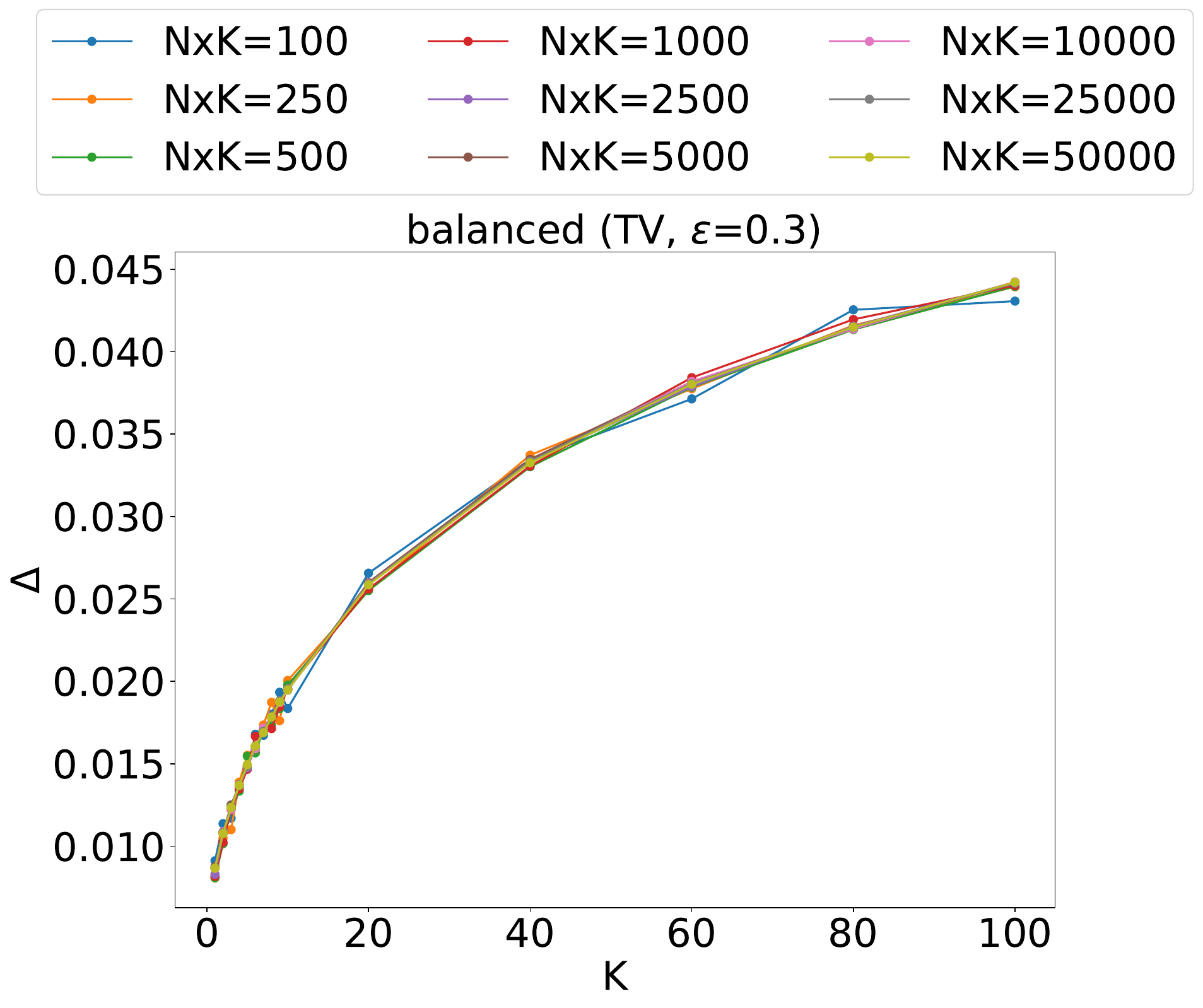}
    \caption{$\epsilon = 0.3$}
    \label{fig:uniform_delta_MAE_cat4_e03}
  \end{subfigure} \hfill
  \begin{subfigure}[b]{0.24\linewidth}
    \centering
    \includegraphics[width=\linewidth]{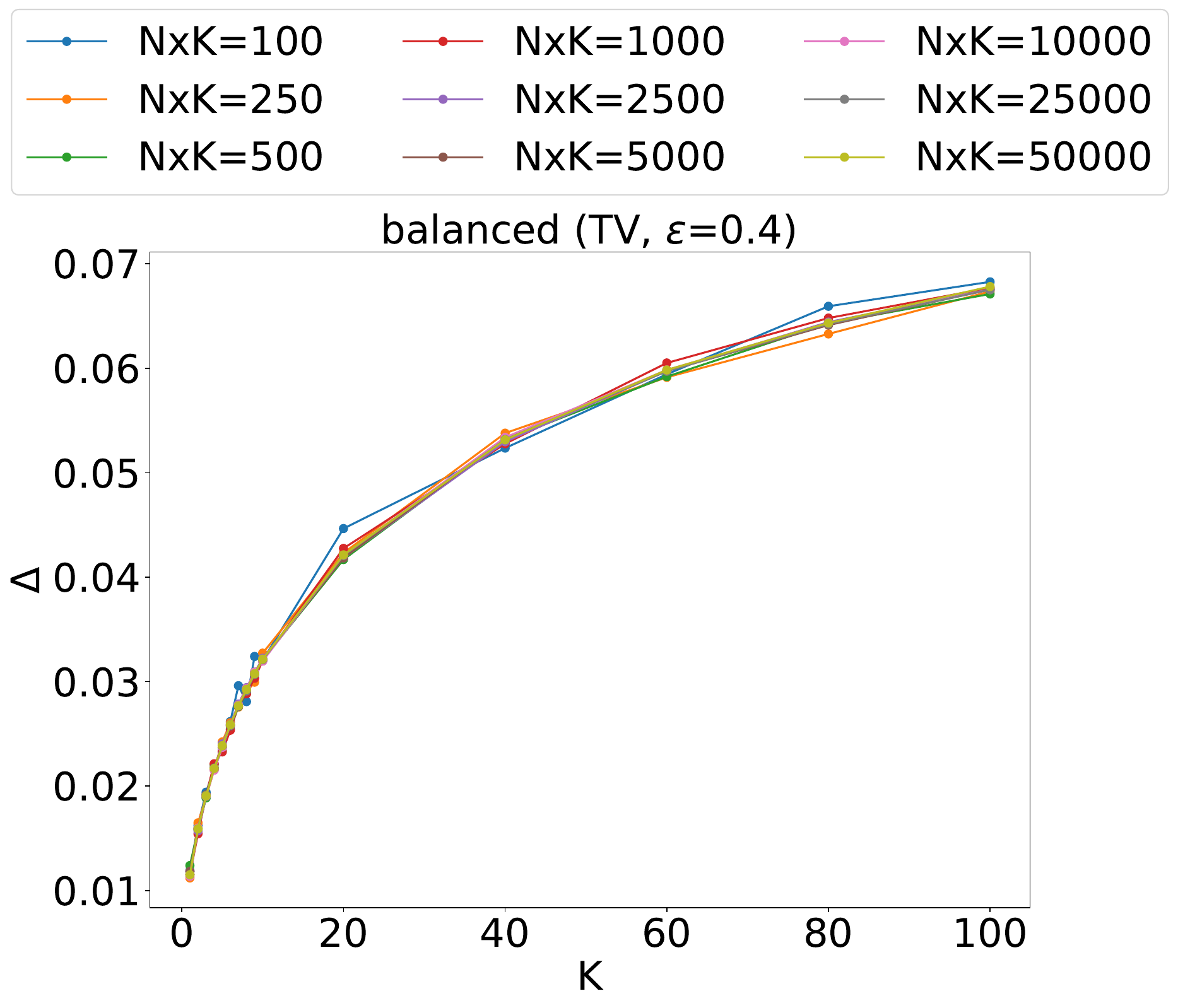}
    \caption{$\epsilon = 0.4$}
    \label{fig:uniform_delta_MAE_cat4_e04}
  \end{subfigure}
  \caption{Effect sizes ($\Delta$) for balanced alphas with TV as the metric ($M=4$)}
  \label{fig:uniform_delta_MAE_cat4}
\end{figure*}

\begin{figure*}
  \centering
  \begin{subfigure}[b]{0.24\linewidth}
    \centering
    \includegraphics[width=\linewidth]{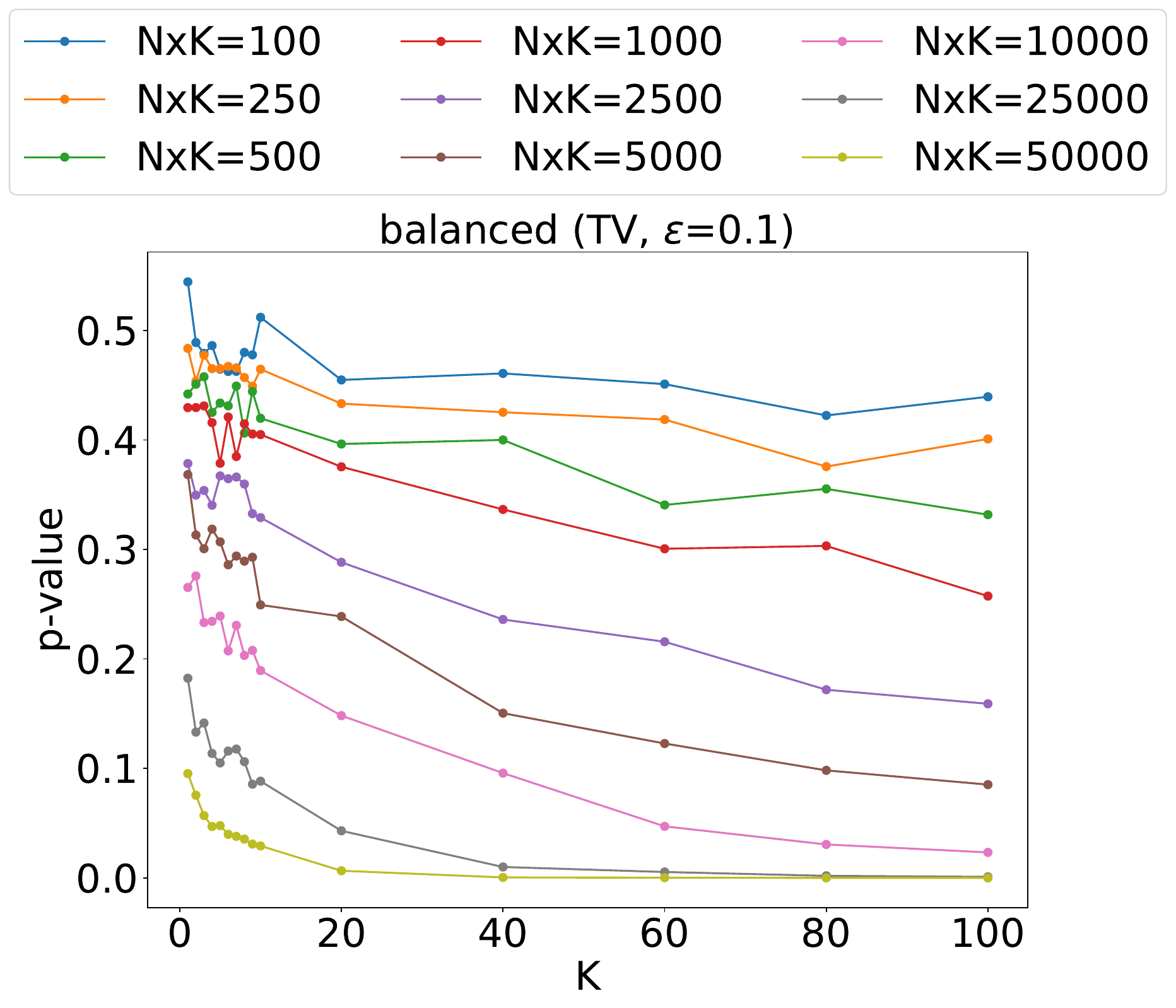}
    \caption{$\epsilon = 0.1$}
    \label{fig:uniform_MAE_cat5_e01}
  \end{subfigure} \hfill
  \begin{subfigure}[b]{0.24\linewidth}
    \centering
    \includegraphics[width=\linewidth]{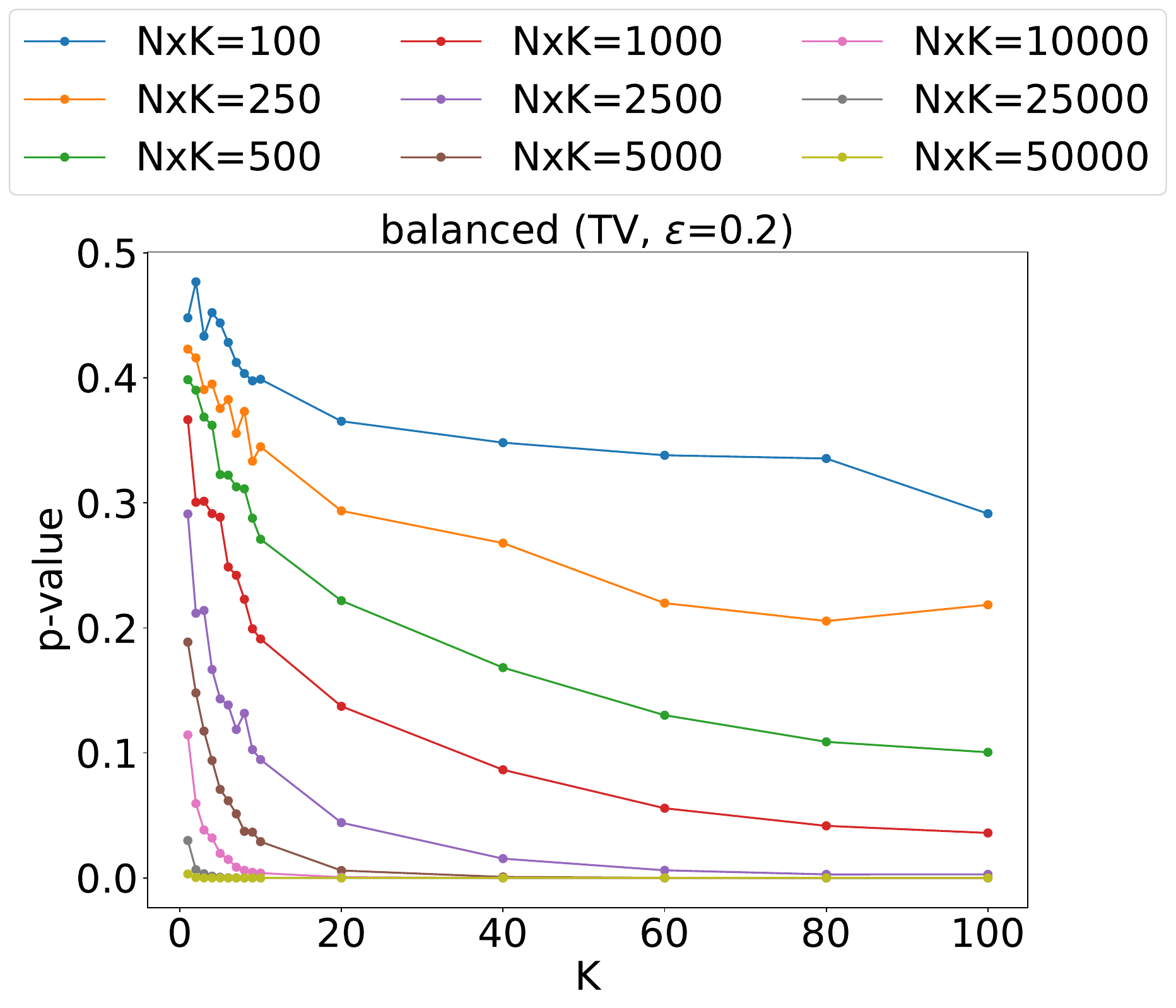}
    \caption{$\epsilon = 0.2$}
    \label{fig:uniform_MAE_cat5_e02}
  \end{subfigure} \hfill
  \begin{subfigure}[b]{0.24\linewidth}
    \centering
    \includegraphics[width=\linewidth]{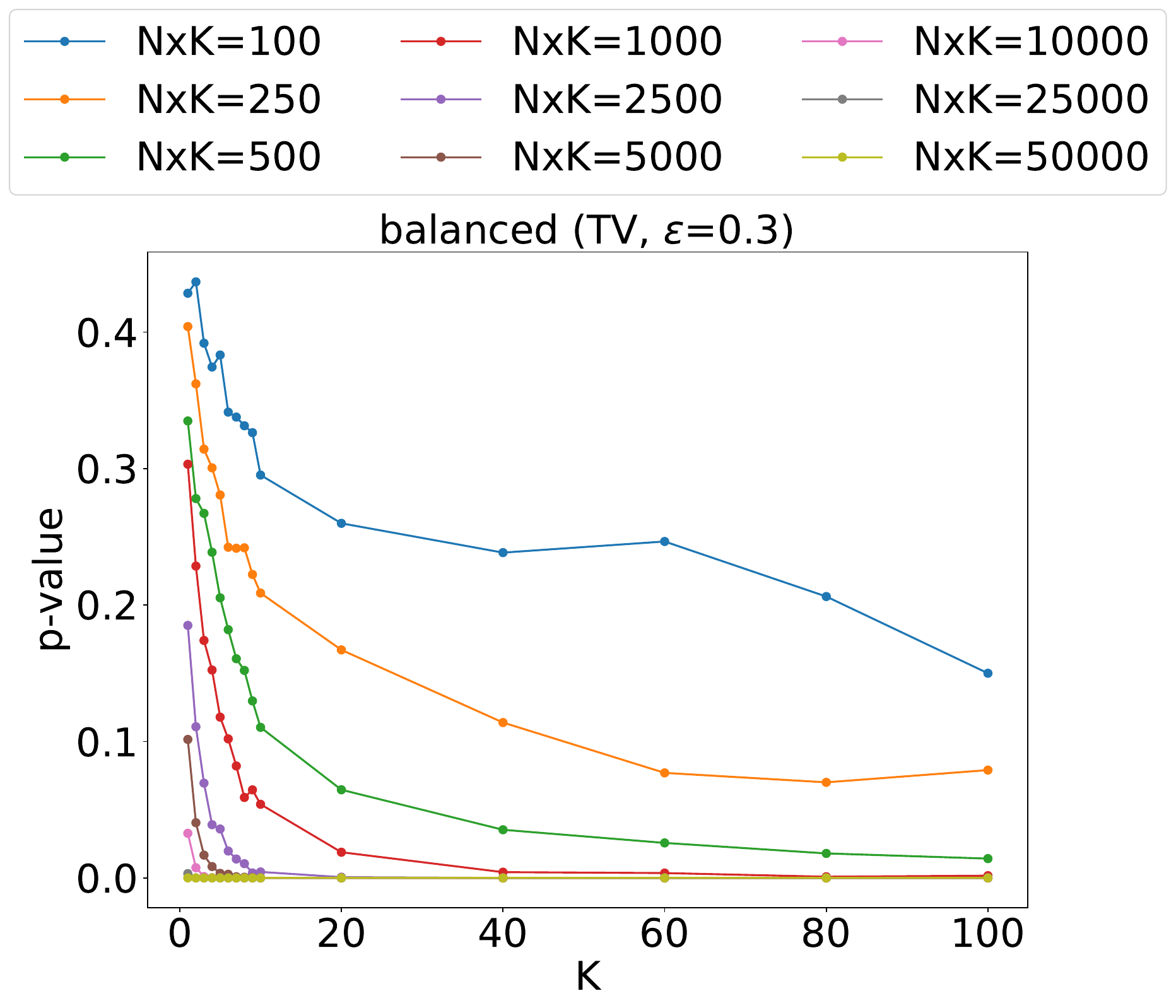}
    \caption{$\epsilon = 0.3$}
    \label{fig:uniform_MAE_cat5_e03}
  \end{subfigure} \hfill
  \begin{subfigure}[b]{0.24\linewidth}
    \centering
    \includegraphics[width=\linewidth]{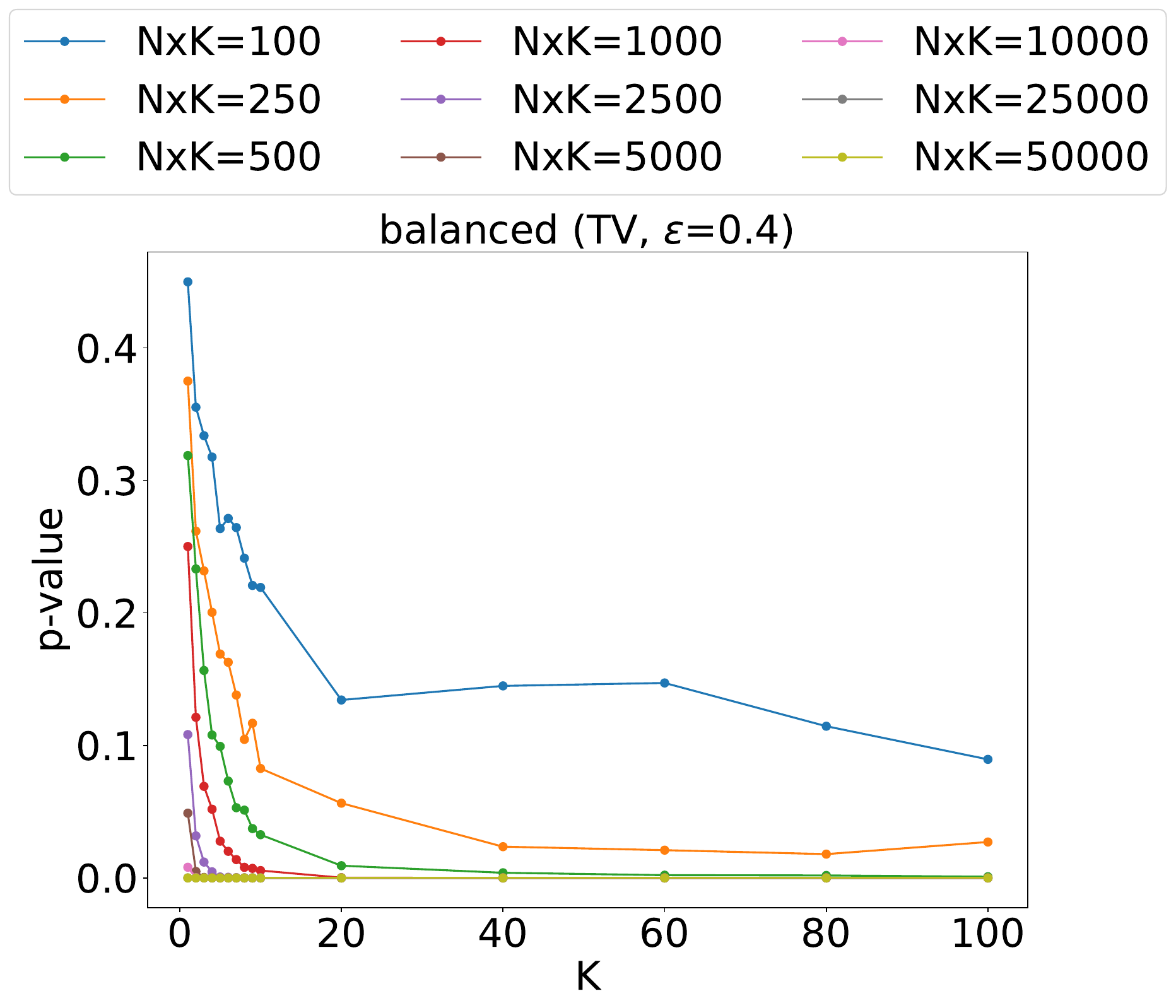}
    \caption{$\epsilon = 0.4$}
    \label{fig:uniform_MAE_cat5_e04}
  \end{subfigure}
  \caption{P-value plots for balanced alphas with TV as the metric ($M=5$)}
  \label{fig:uniform_MAE_cat5}
\end{figure*}

\begin{figure*}
  \centering
  \begin{subfigure}[b]{0.24\linewidth}
    \centering
    \includegraphics[width=\linewidth]{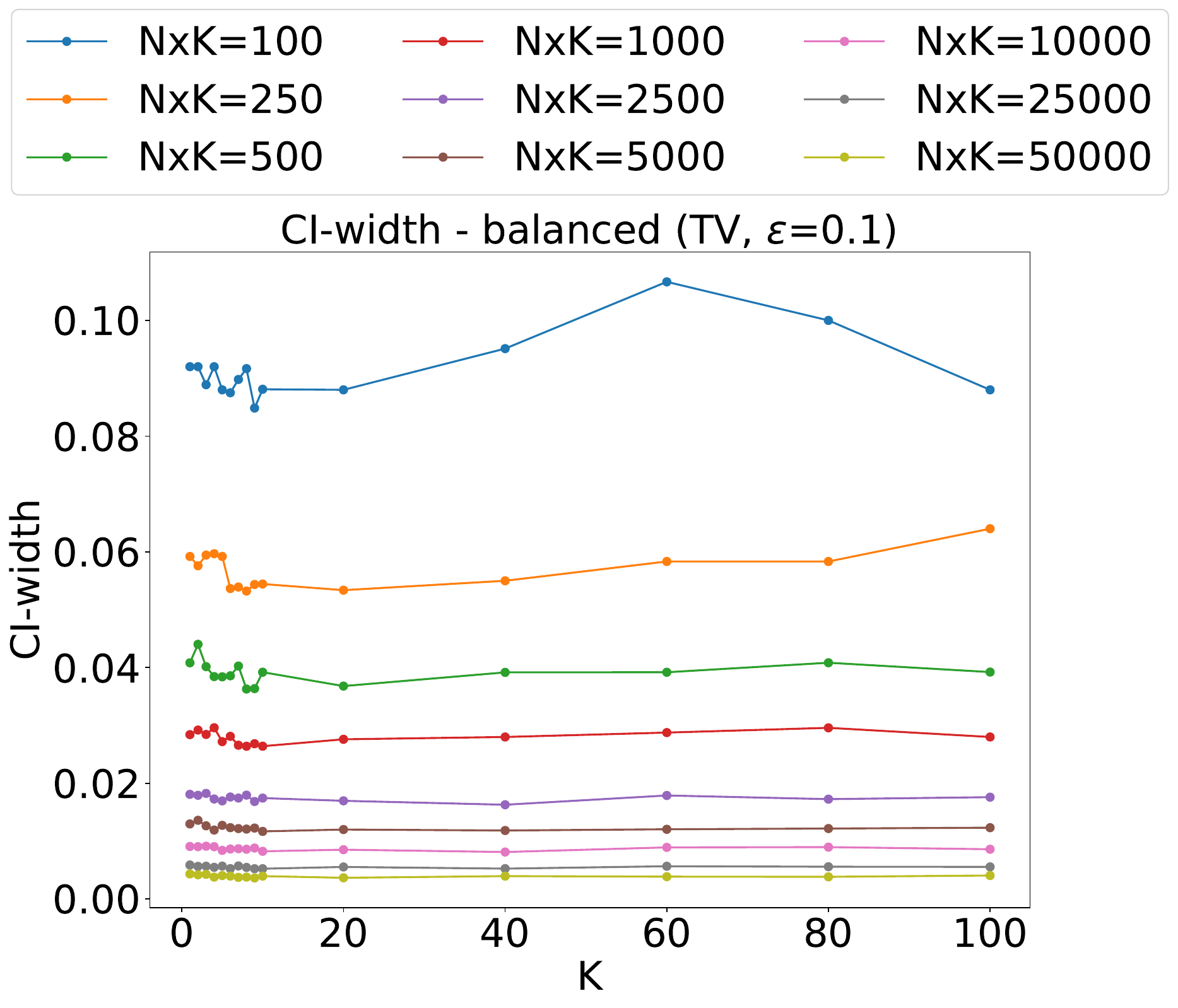}
    \caption{$\epsilon = 0.1$}
    \label{fig:uniform_ci_MAE_cat5_e01}
  \end{subfigure} \hfill
  \begin{subfigure}[b]{0.24\linewidth}
    \centering
    \includegraphics[width=\linewidth]{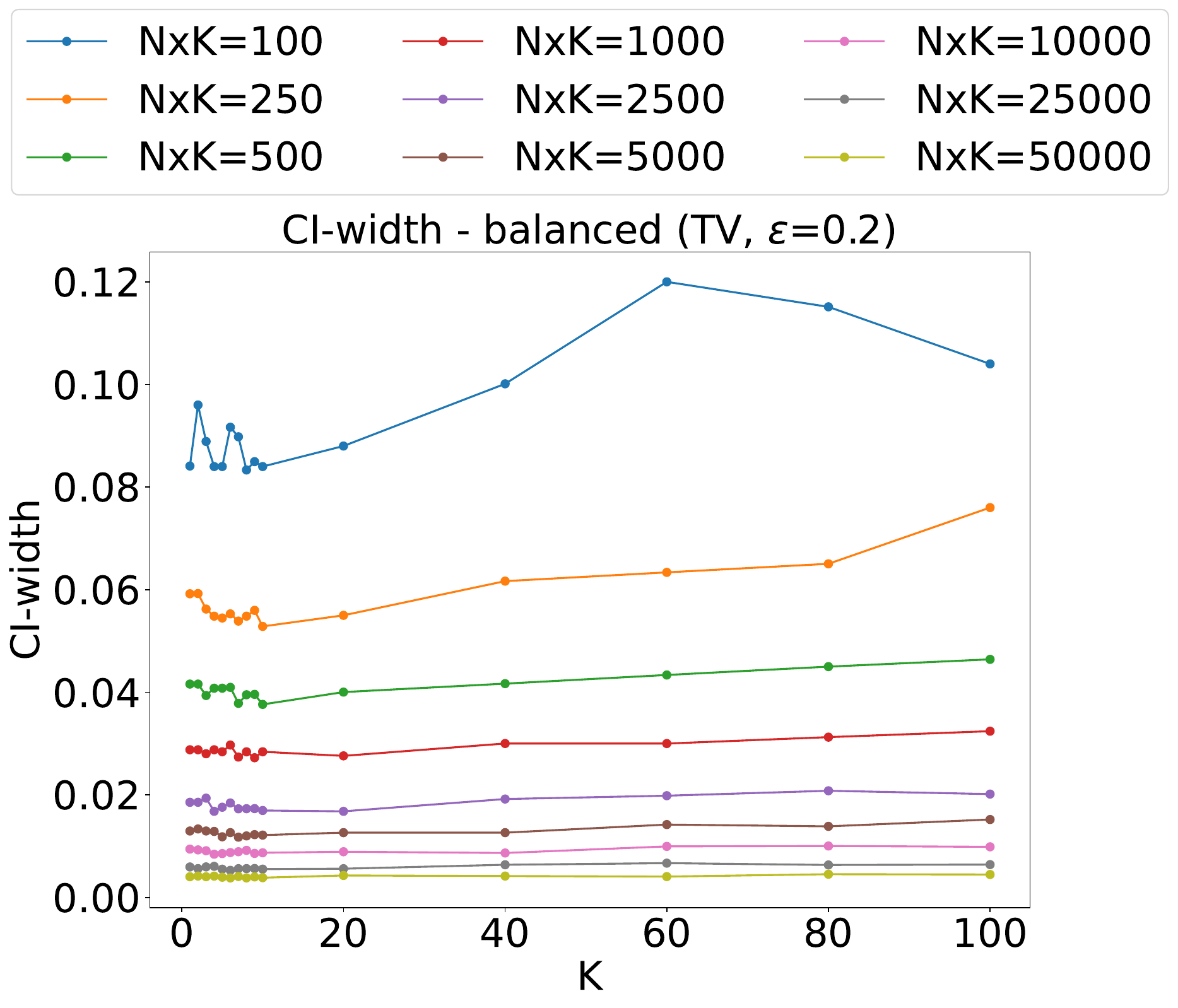}
    \caption{$\epsilon = 0.2$}
    \label{fig:uniform_ci_MAE_cat5_e02}
  \end{subfigure} \hfill
  \begin{subfigure}[b]{0.24\linewidth}
    \centering
    \includegraphics[width=\linewidth]{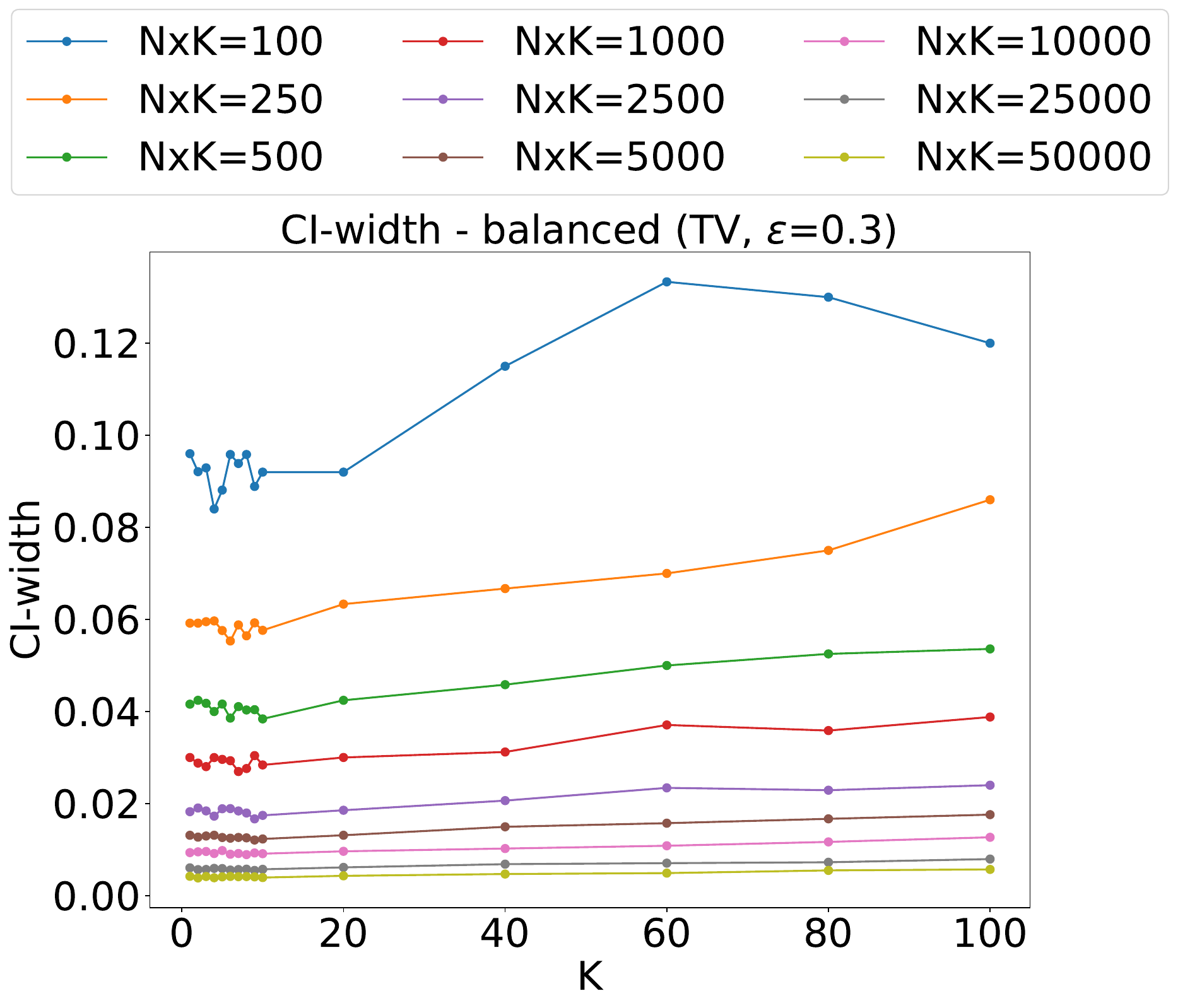}
    \caption{$\epsilon = 0.3$}
    \label{fig:uniform_ci_MAE_cat5_e03}
  \end{subfigure} \hfill
  \begin{subfigure}[b]{0.24\linewidth}
    \centering
    \includegraphics[width=\linewidth]{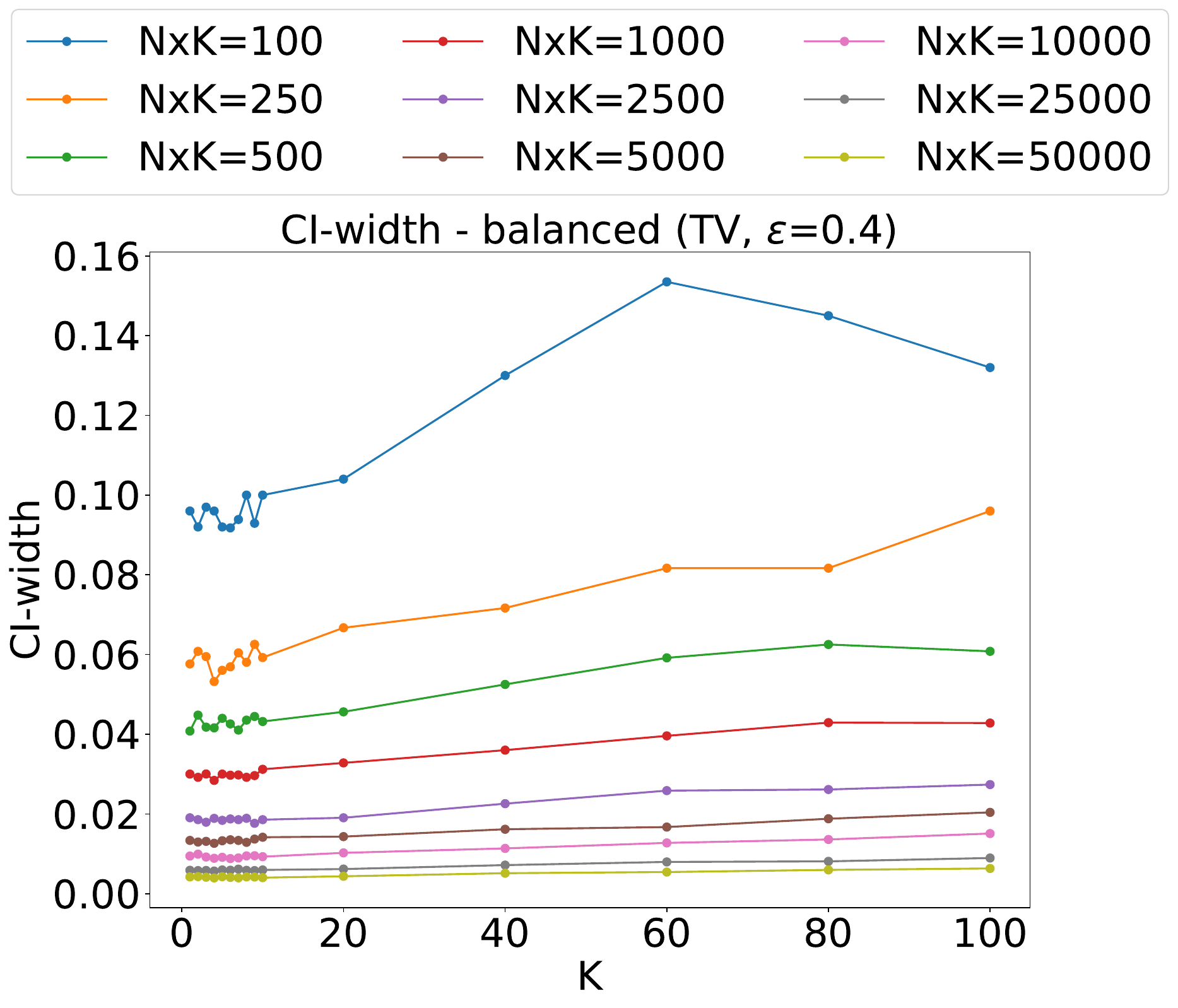}
    \caption{$\epsilon = 0.4$}
    \label{fig:uniform_ci_MAE_cat5_e04}
  \end{subfigure}
  \caption{CI-width plots for balanced alphas with TV as the metric ($M=5$)}
  \label{fig:uniform_ci_MAE_cat5}
\end{figure*}

\begin{figure*}
  \centering
  \begin{subfigure}[b]{0.24\linewidth}
    \centering
    \includegraphics[width=\linewidth]{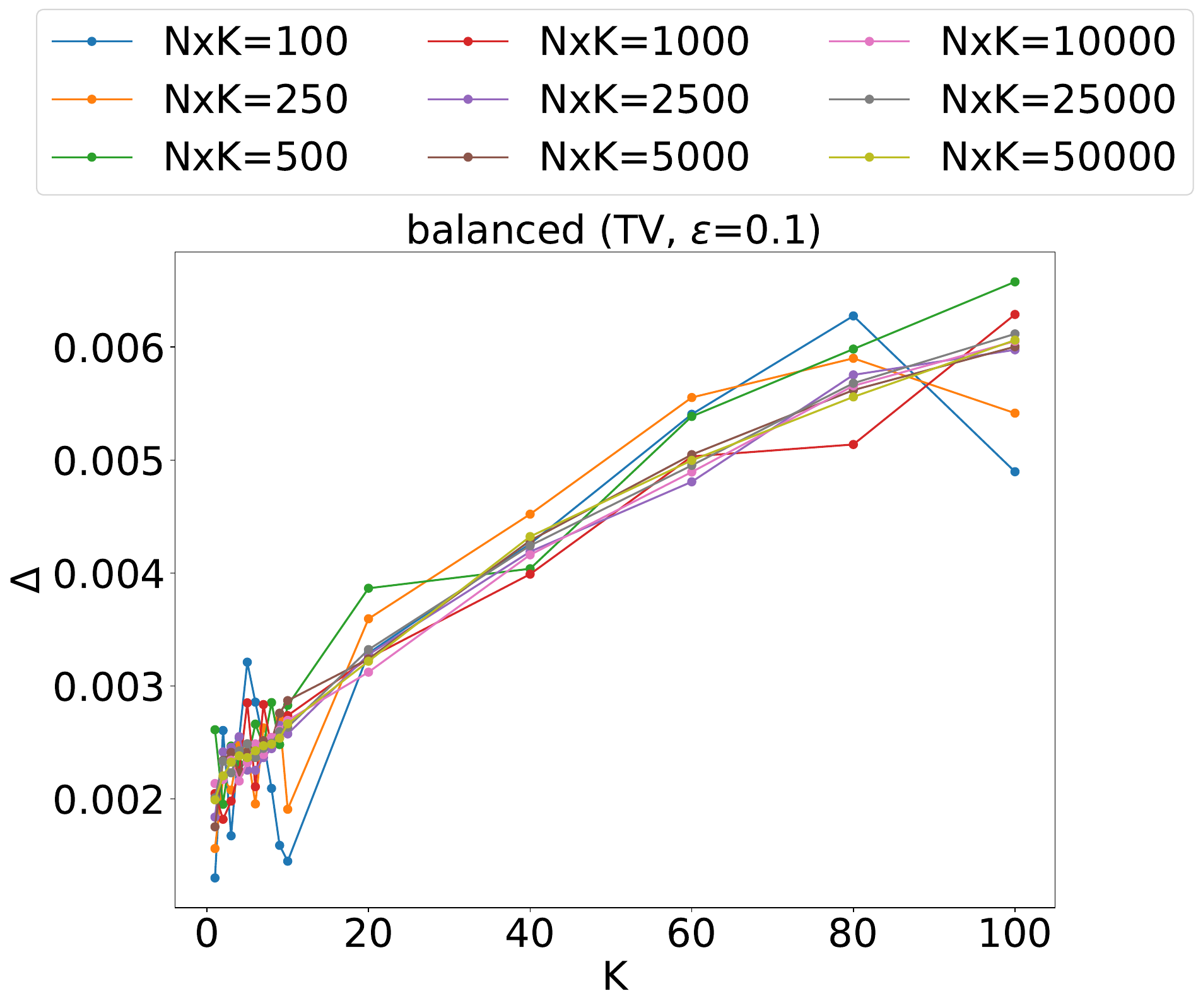}
    \caption{$\epsilon = 0.1$}
    \label{fig:uniform_delta_MAE_cat5_e01}
  \end{subfigure} \hfill
  \begin{subfigure}[b]{0.24\linewidth}
    \centering
    \includegraphics[width=\linewidth]{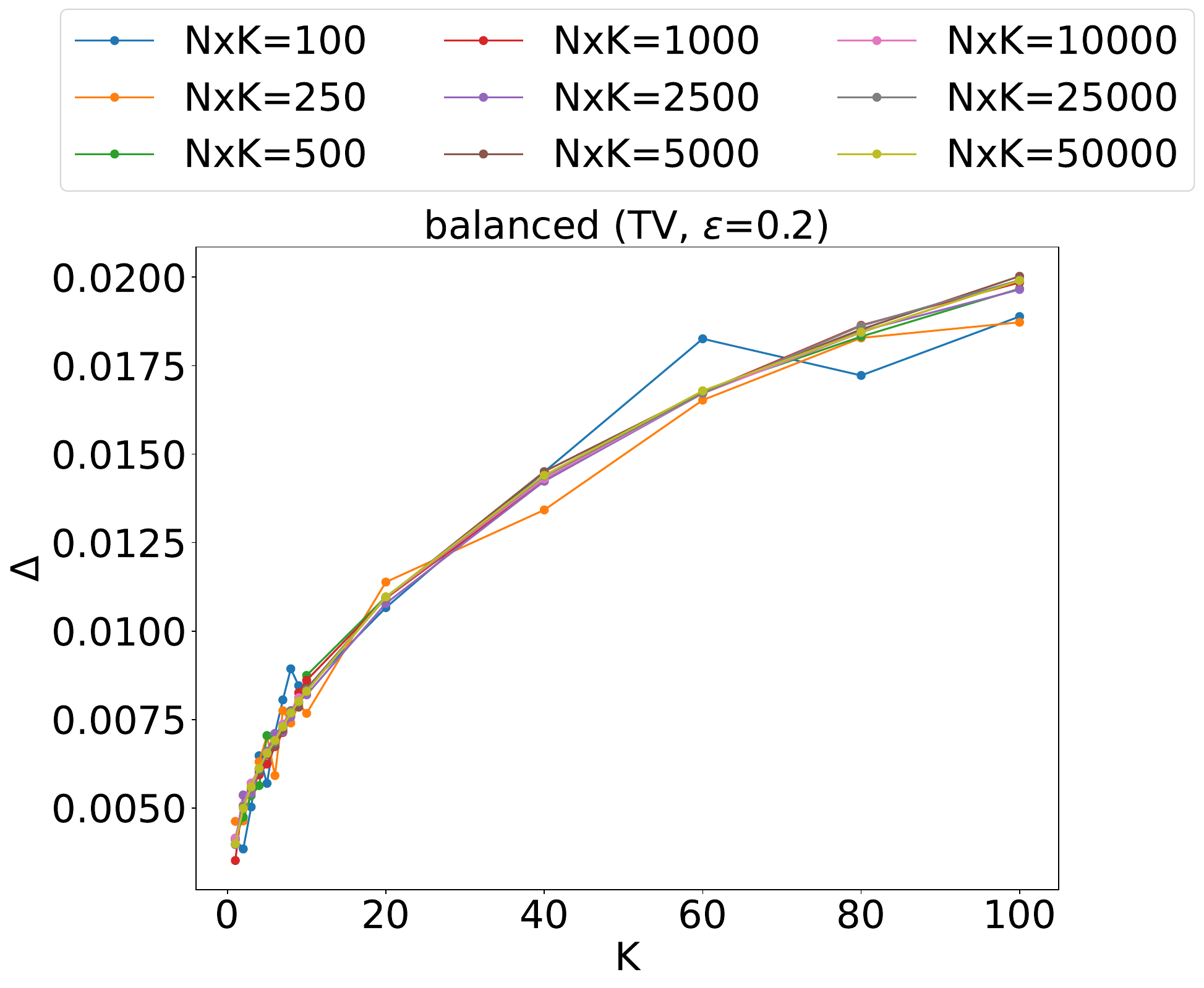}
    \caption{$\epsilon = 0.2$}
    \label{fig:uniform_delta_MAE_cat5_e02}
  \end{subfigure} \hfill
  \begin{subfigure}[b]{0.24\linewidth}
    \centering
    \includegraphics[width=\linewidth]{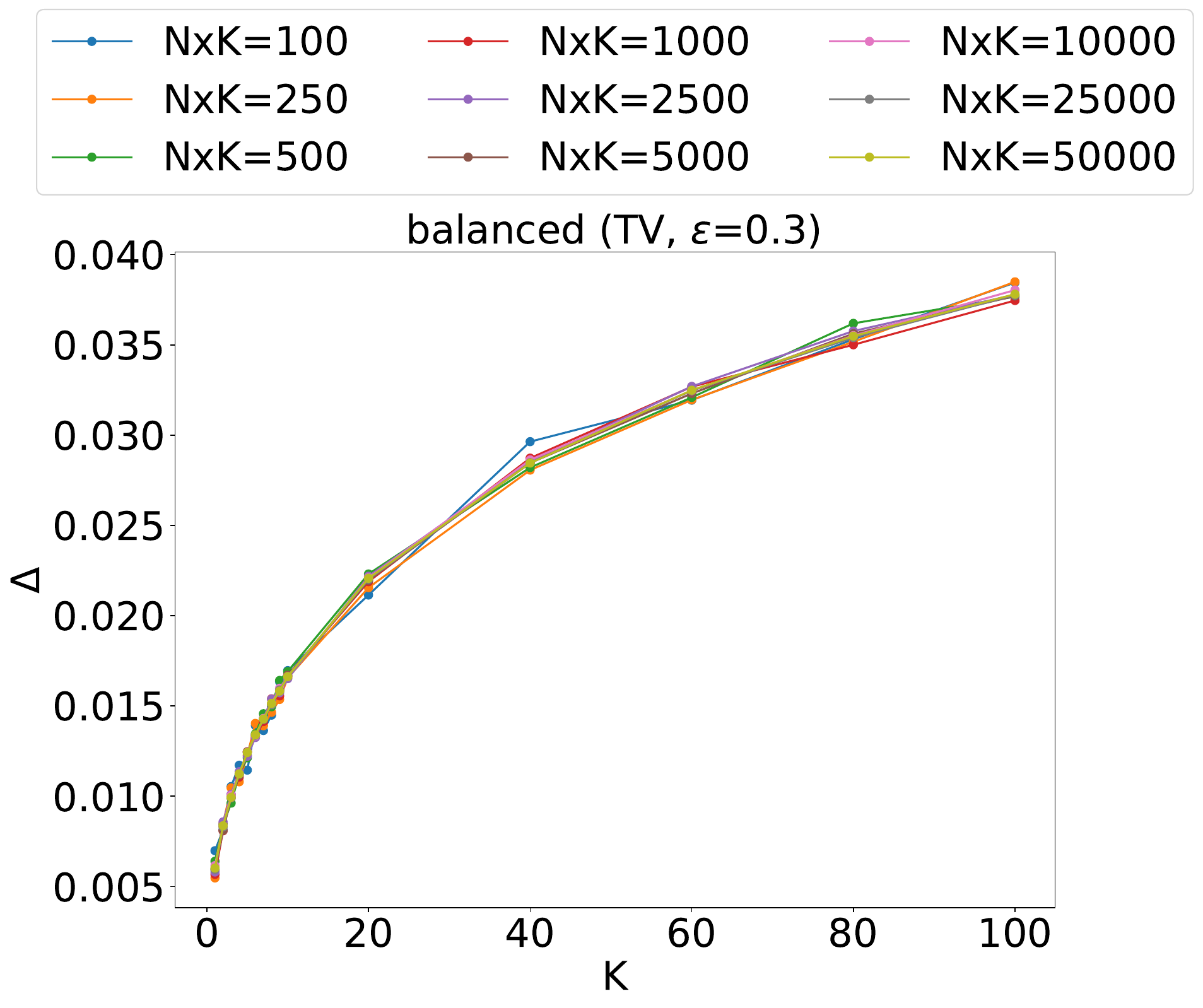}
    \caption{$\epsilon = 0.3$}
    \label{fig:uniform_delta_MAE_cat5_e03}
  \end{subfigure} \hfill
  \begin{subfigure}[b]{0.24\linewidth}
    \centering
    \includegraphics[width=\linewidth]{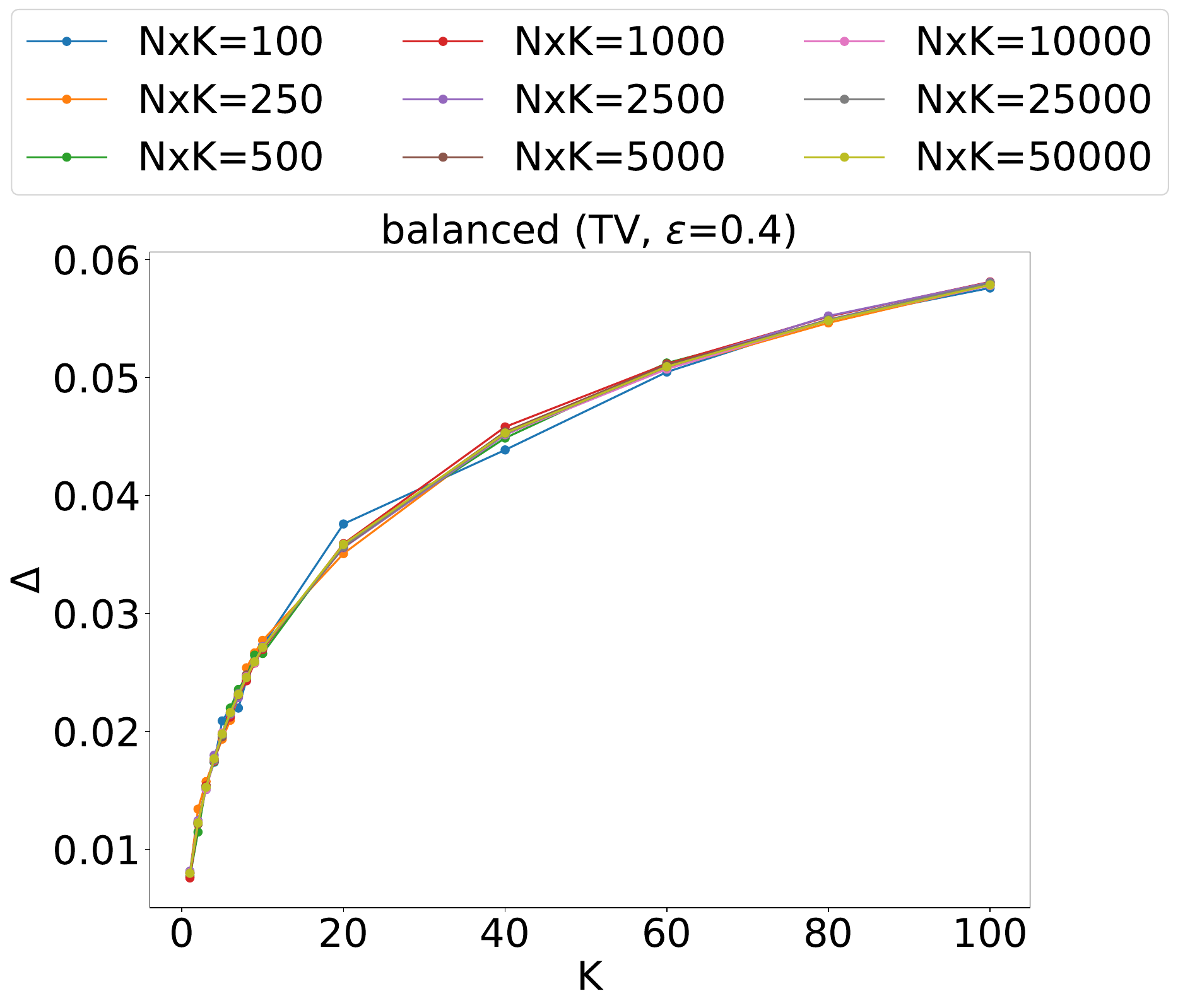}
    \caption{$\epsilon = 0.4$}
    \label{fig:uniform_delta_MAE_cat5_e04}
  \end{subfigure}
  \caption{Effect sizes ($\Delta$) for balanced alphas with TV as the metric ($M=5$)}
  \label{fig:uniform_delta_MAE_cat5}
\end{figure*}

\begin{figure*}
  \centering
  \begin{subfigure}[b]{0.24\linewidth}
    \centering
    \includegraphics[width=\linewidth]{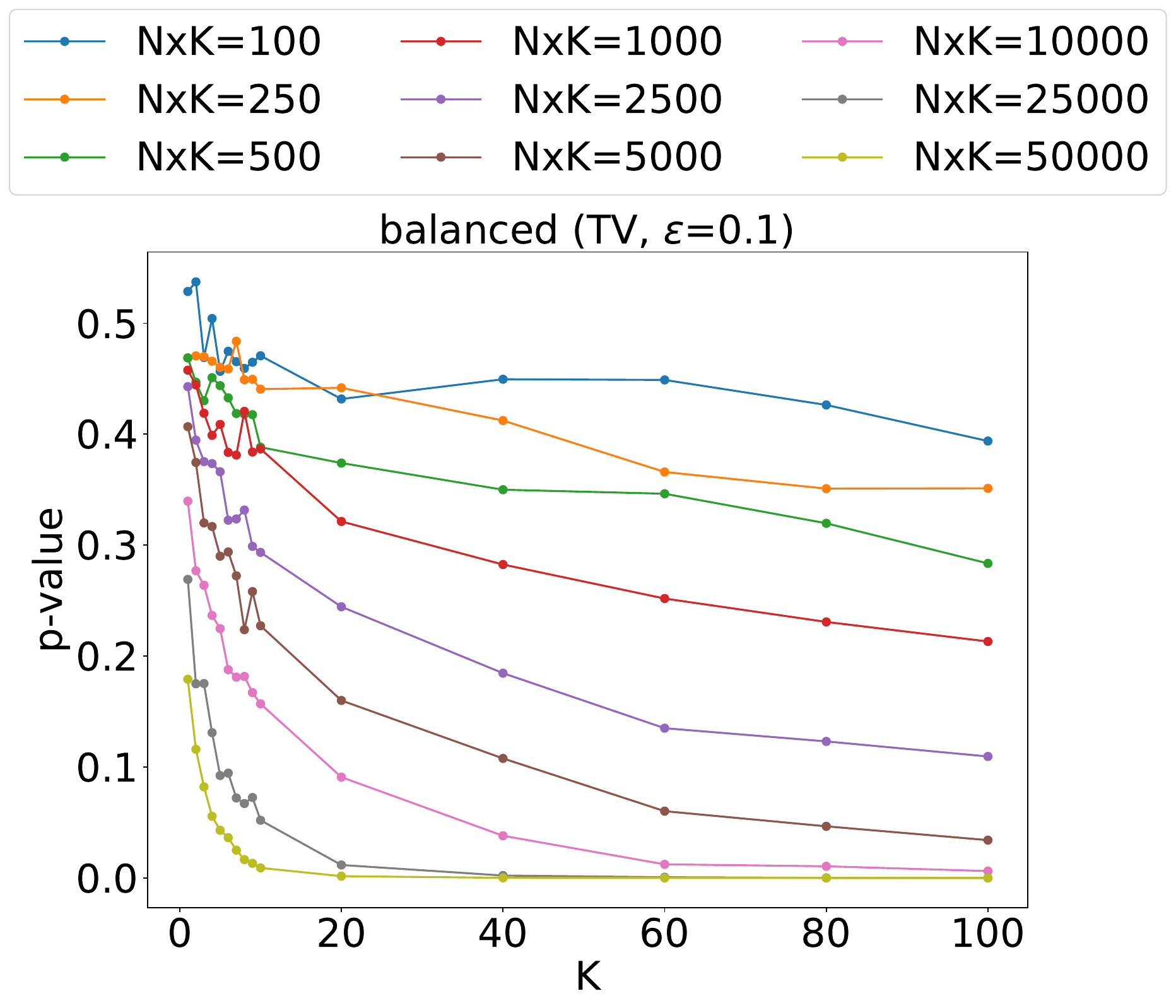}
    \caption{$\epsilon = 0.1$}
    \label{fig:uniform_MAE_cat12_e01}
  \end{subfigure} \hfill
  \begin{subfigure}[b]{0.24\linewidth}
    \centering
    \includegraphics[width=\linewidth]{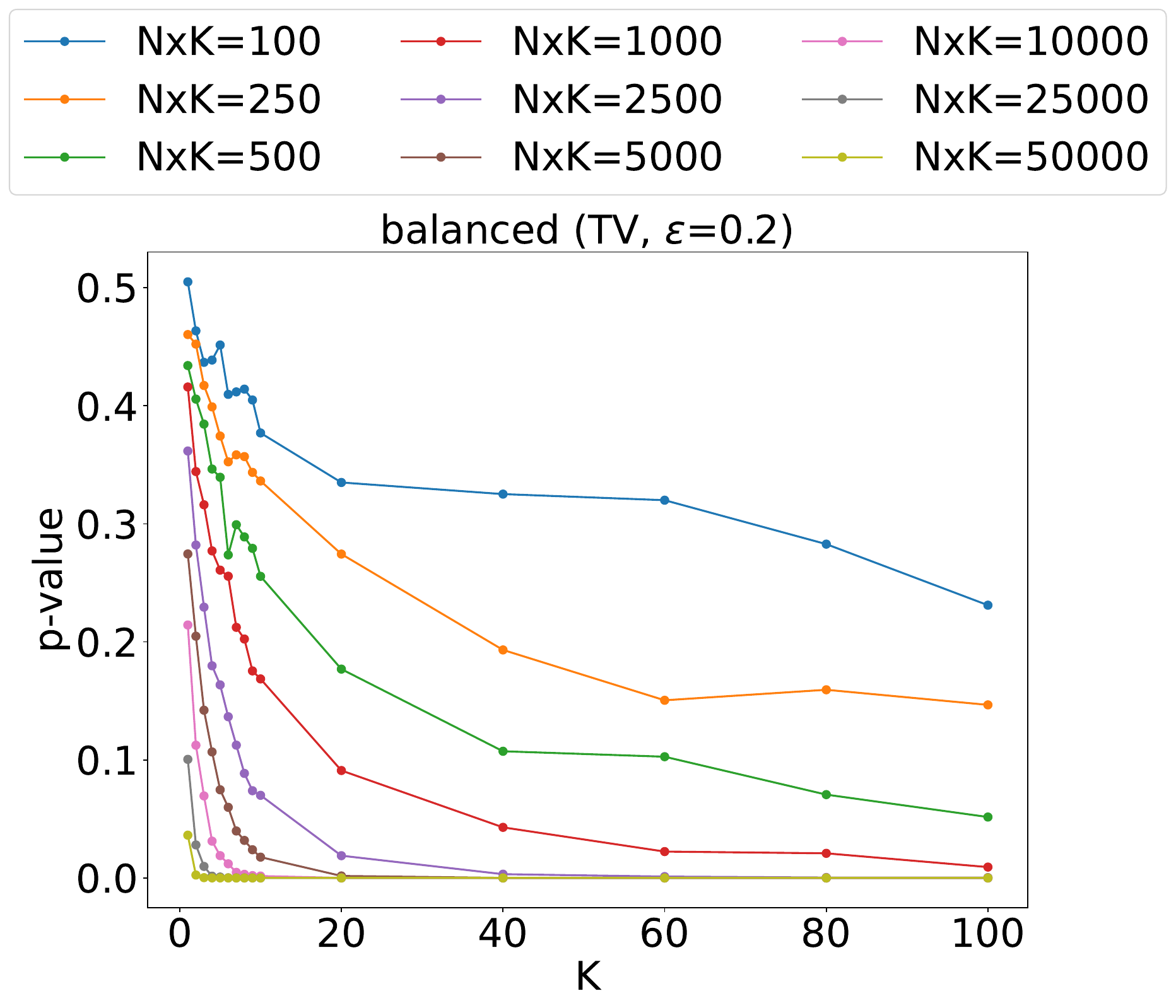}
    \caption{$\epsilon = 0.2$}
    \label{fig:uniform_MAE_cat12_e02}
  \end{subfigure} \hfill
  \begin{subfigure}[b]{0.24\linewidth}
    \centering
    \includegraphics[width=\linewidth]{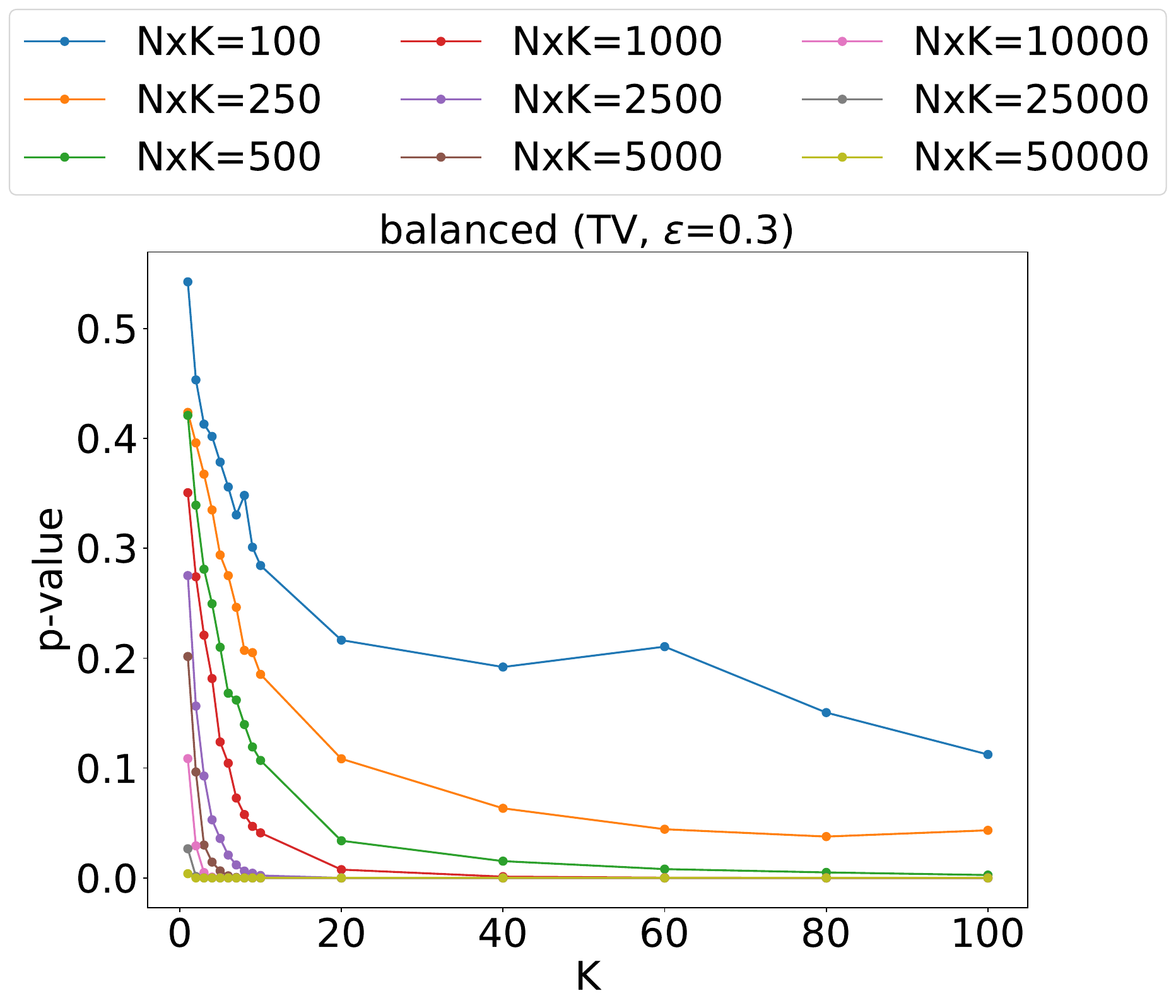}
    \caption{$\epsilon = 0.3$}
    \label{fig:uniform_MAE_cat12_e03}
  \end{subfigure} \hfill
  \begin{subfigure}[b]{0.24\linewidth}
    \centering
    \includegraphics[width=\linewidth]{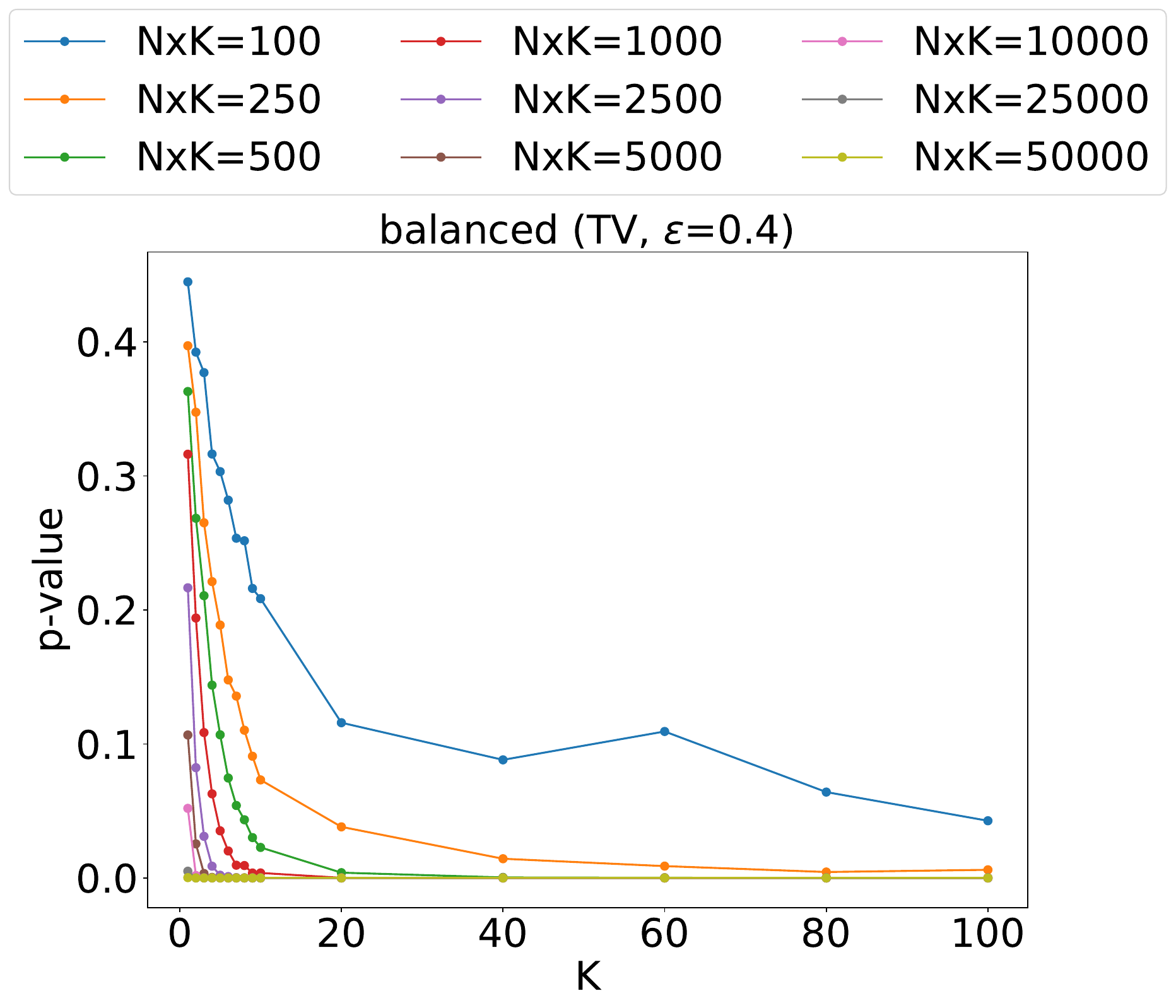}
    \caption{$\epsilon = 0.4$}
    \label{fig:uniform_MAE_cat12_e04}
  \end{subfigure}
  \caption{P-value plots for balanced alphas with TV as the metric ($M=12$)}
  \label{fig:uniform_MAE_cat12}
\end{figure*}

\begin{figure*}
  \centering
  \begin{subfigure}[b]{0.24\linewidth}
    \centering
    \includegraphics[width=\linewidth]{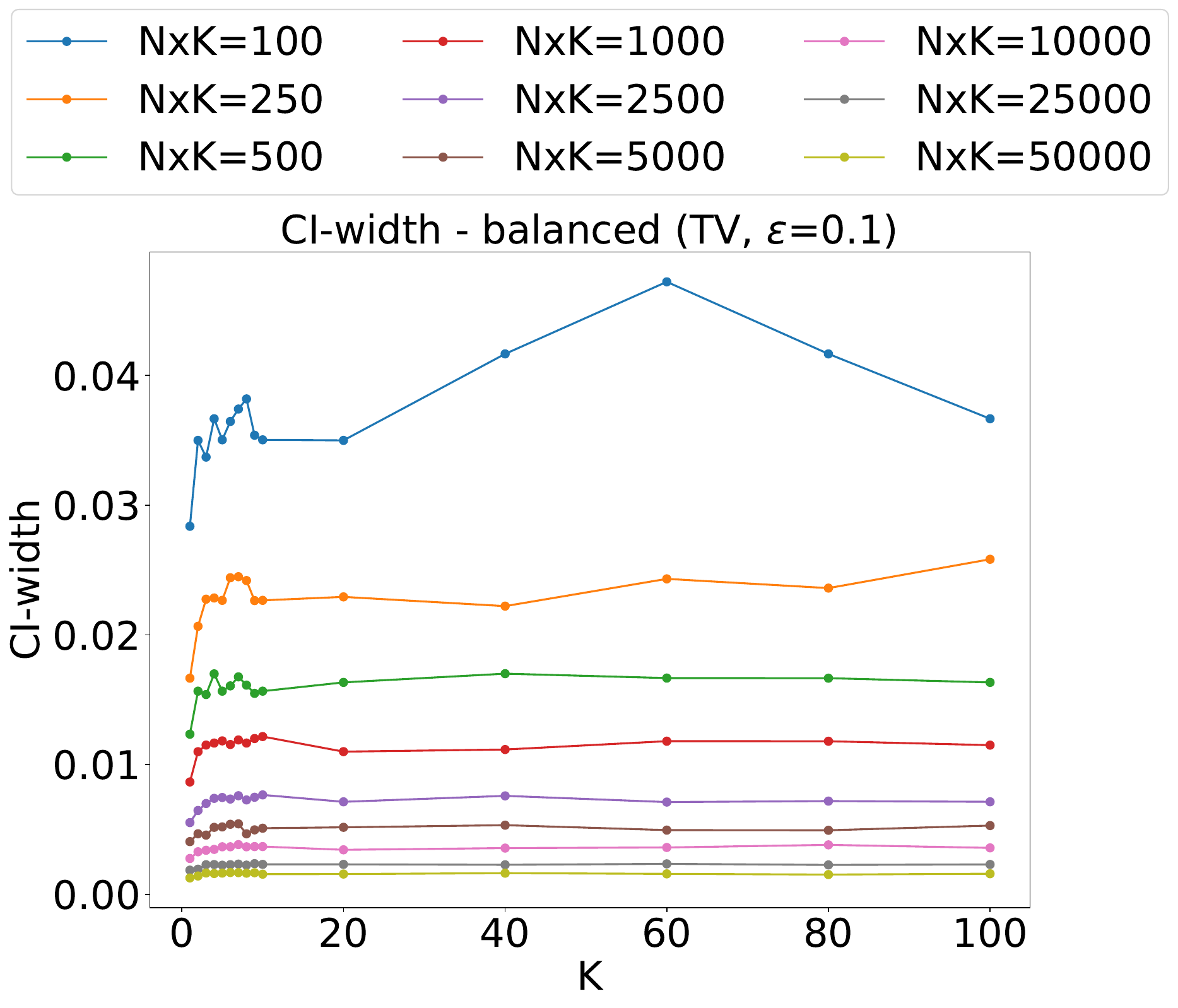}
    \caption{$\epsilon = 0.1$}
    \label{fig:uniform_ci_MAE_cat12_e01}
  \end{subfigure} \hfill
  \begin{subfigure}[b]{0.24\linewidth}
    \centering
    \includegraphics[width=\linewidth]{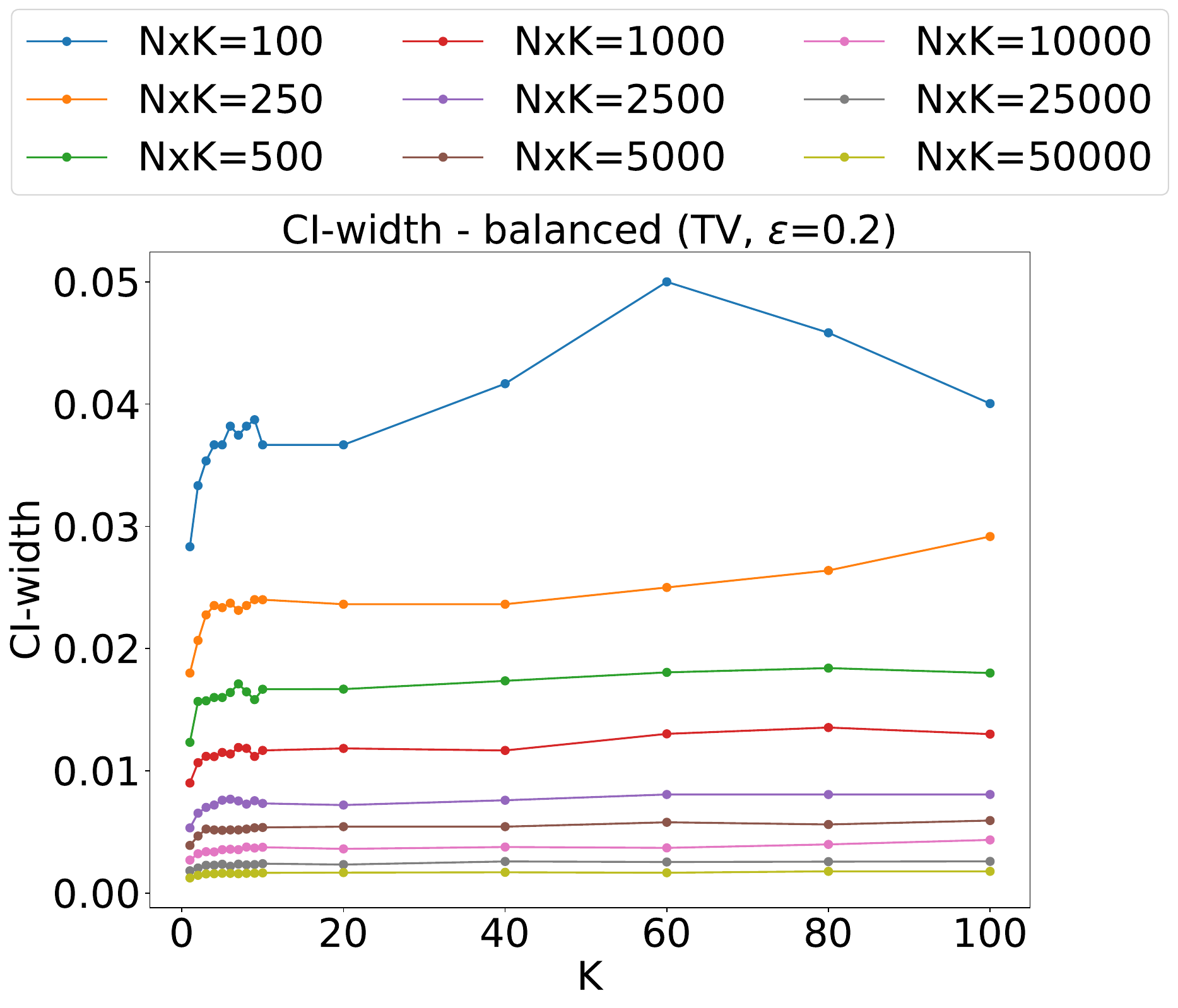}
    \caption{$\epsilon = 0.2$}
    \label{fig:uniform_ci_MAE_cat12_e02}
  \end{subfigure} \hfill
  \begin{subfigure}[b]{0.24\linewidth}
    \centering
    \includegraphics[width=\linewidth]{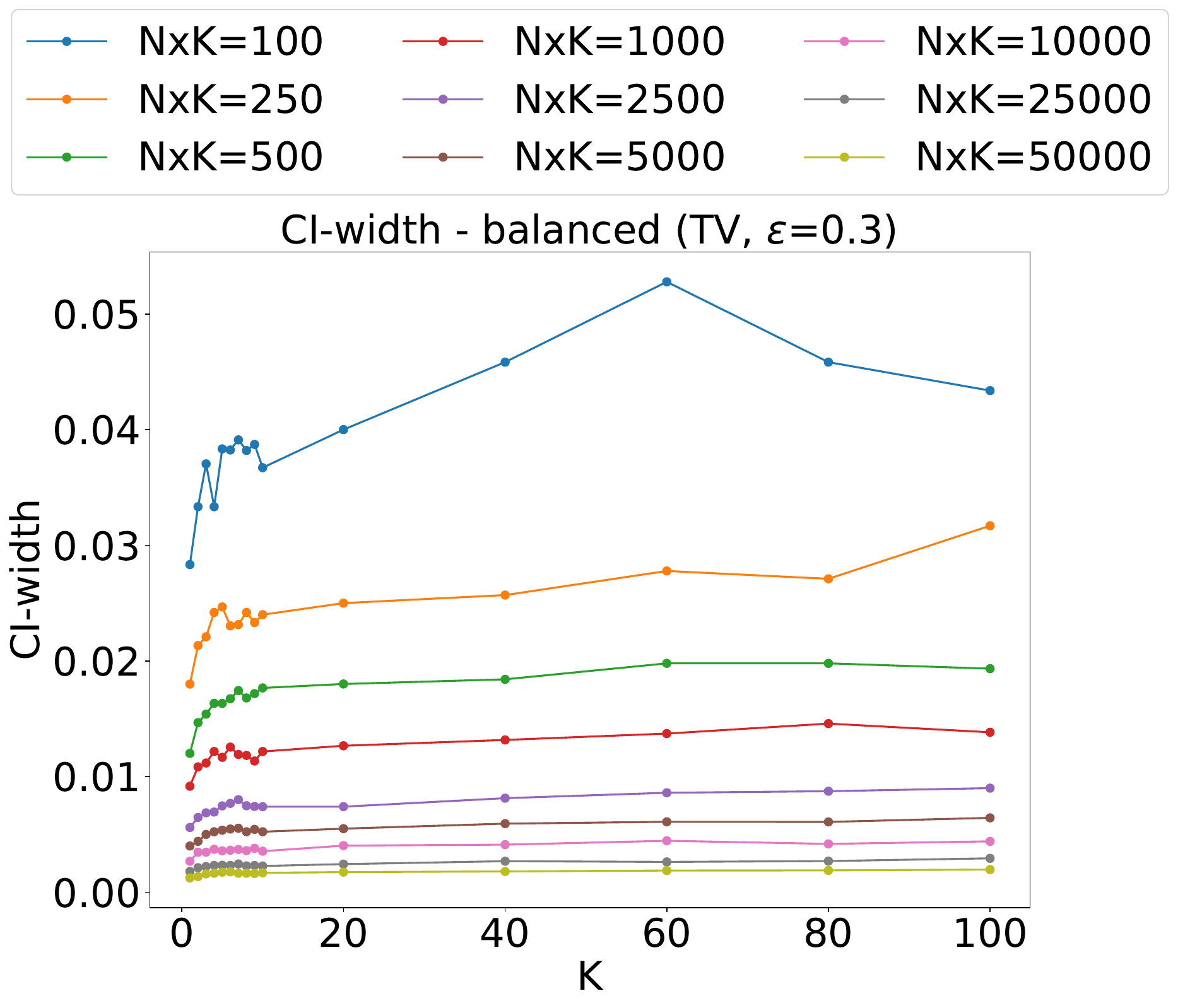}
    \caption{$\epsilon = 0.3$}
    \label{fig:uniform_ci_MAE_cat12_e03}
  \end{subfigure} \hfill
  \begin{subfigure}[b]{0.24\linewidth}
    \centering
    \includegraphics[width=\linewidth]{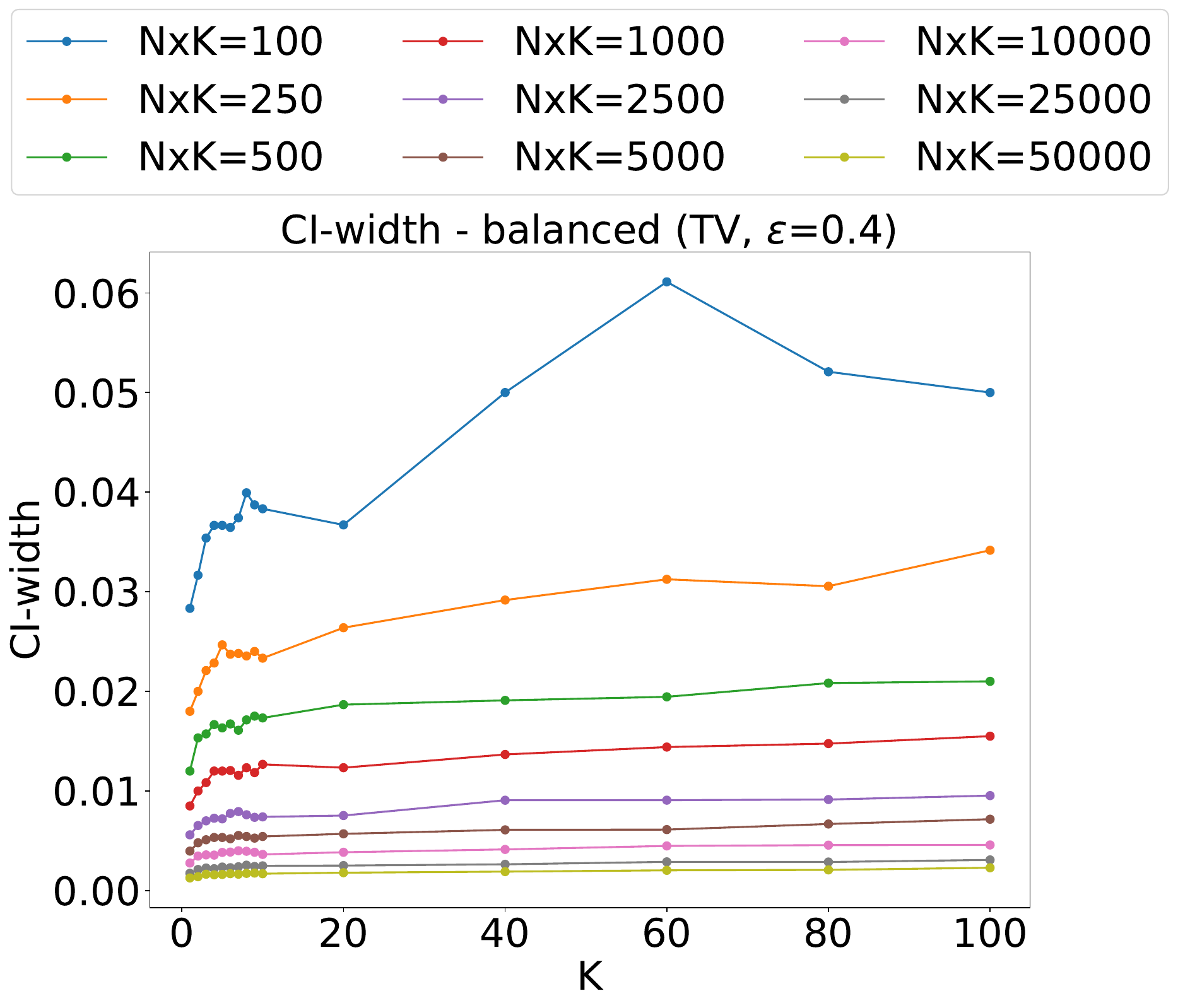}
    \caption{$\epsilon = 0.4$}
    \label{fig:uniform_ci_MAE_cat12_e04}
  \end{subfigure}
  \caption{CI-width plots for balanced alphas with TV as the metric ($M=12$)}
  \label{fig:uniform_ci_MAE_cat12}
\end{figure*}

\begin{figure*}
  \centering
  \begin{subfigure}[b]{0.24\linewidth}
    \centering
    \includegraphics[width=\linewidth]{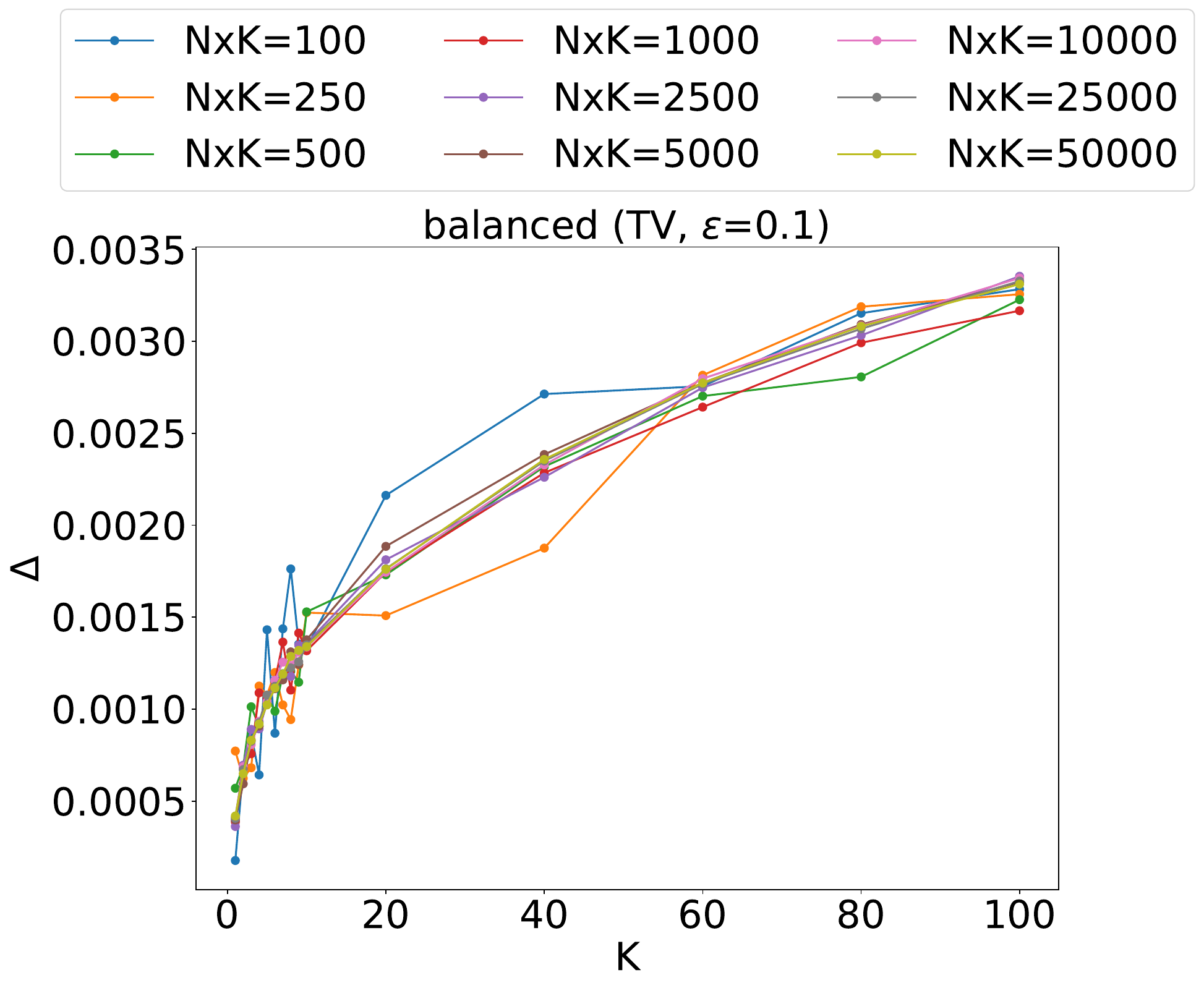}
    \caption{$\epsilon = 0.1$}
    \label{fig:uniform_delta_MAE_cat12_e01}
  \end{subfigure} \hfill
  \begin{subfigure}[b]{0.24\linewidth}
    \centering
    \includegraphics[width=\linewidth]{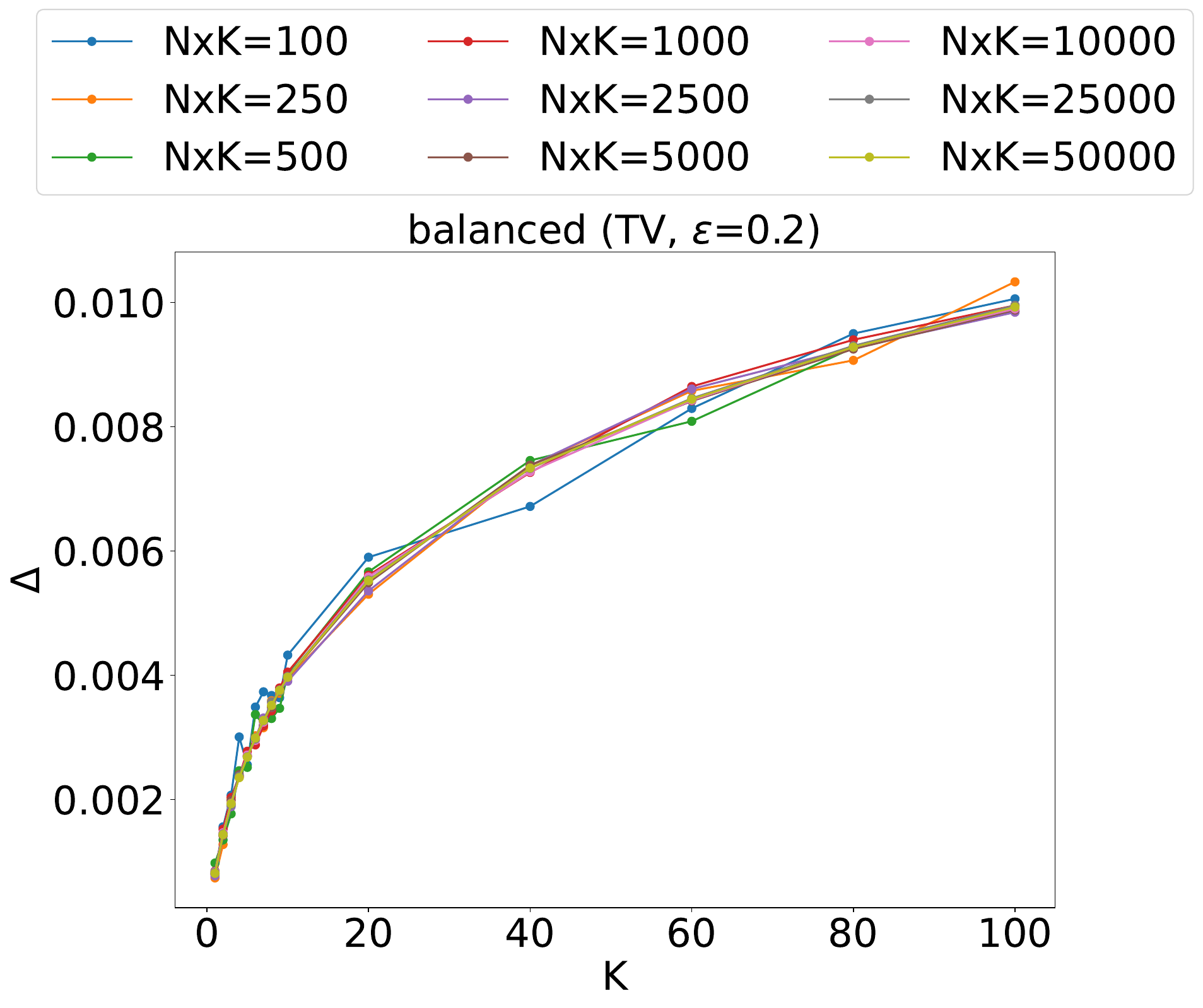}
    \caption{$\epsilon = 0.2$}
    \label{fig:uniform_delta_MAE_cat12_e02}
  \end{subfigure} \hfill
  \begin{subfigure}[b]{0.24\linewidth}
    \centering
    \includegraphics[width=\linewidth]{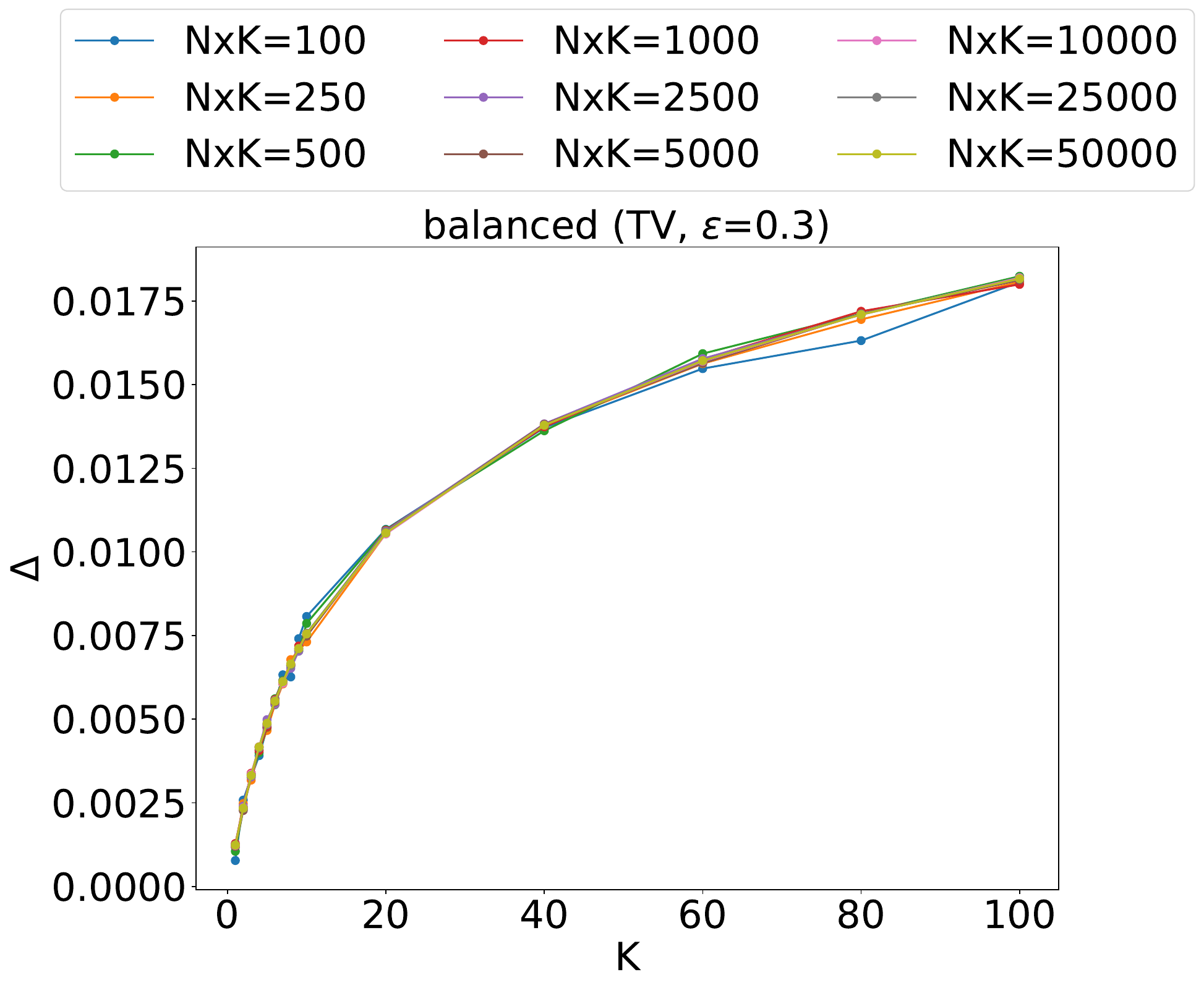}
    \caption{$\epsilon = 0.3$}
    \label{fig:uniform_delta_MAE_cat12_e03}
  \end{subfigure} \hfill
  \begin{subfigure}[b]{0.24\linewidth}
    \centering
    \includegraphics[width=\linewidth]{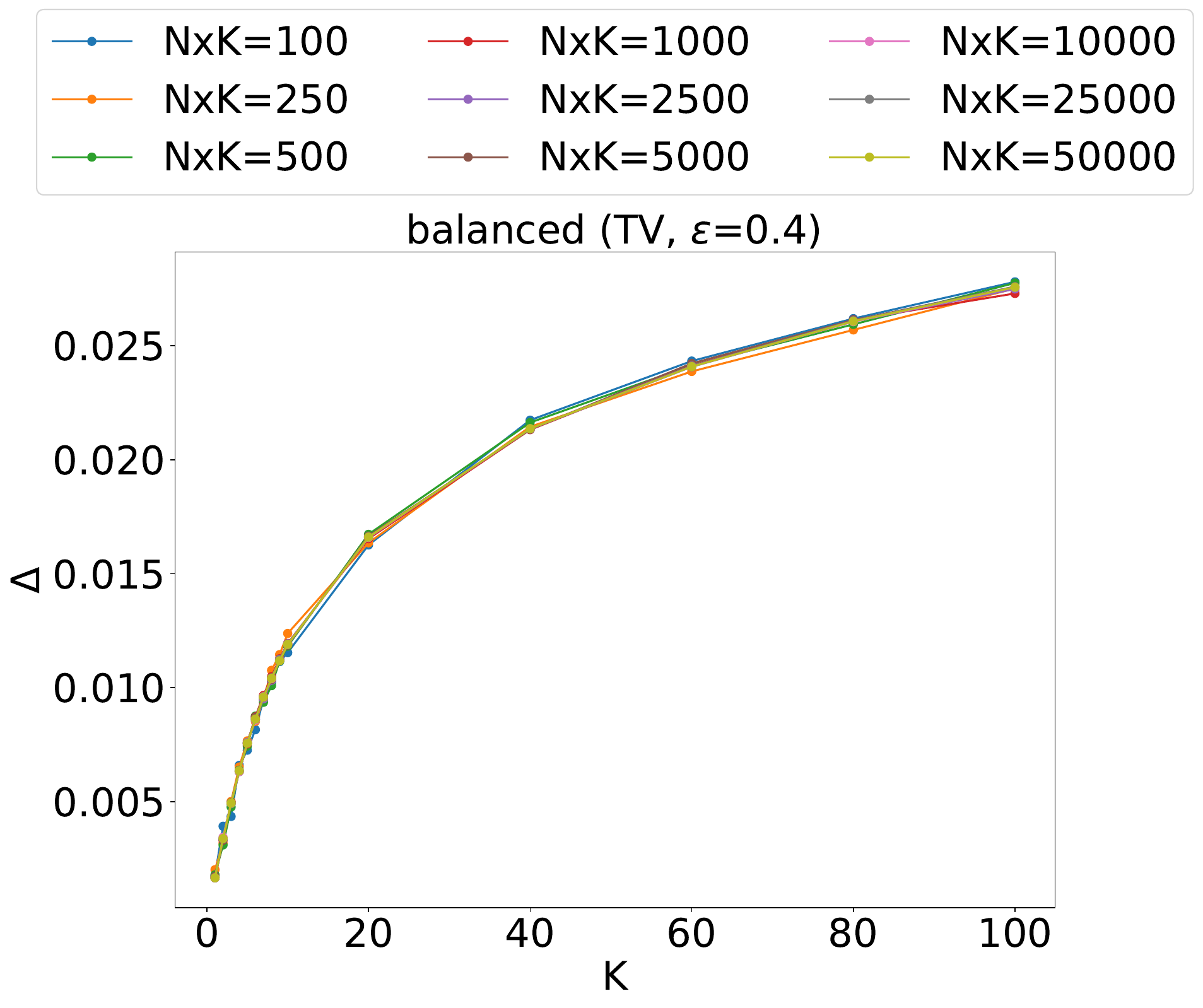}
    \caption{$\epsilon = 0.4$}
    \label{fig:uniform_delta_MAE_cat12_e04}
  \end{subfigure}
  \caption{Effect sizes ($\Delta$) for balanced alphas with TV as the metric ($M=12$)}
  \label{fig:uniform_delta_MAE_cat12}
\end{figure*}

Figures \ref{fig:uniform_MAE_cat2}--\ref{fig:uniform_delta_MAE_cat12} show results for \pv s, confidence intervals, and effect sizes of TV for balanced priors on parameter $\alpha$ with different values of $\epsilon$ and increasing $K$ for various $N \times K$. For simulations with balanced categories, \pv s\ monotonically decrease with increasing $K$ as $M$ gets larger. CI-width and effect sizes increase as $K$ increases.

\subsubsection{Artificial Distribution with Unbalanced Priors}

\begin{figure*}
  \centering
  \begin{subfigure}[b]{0.24\linewidth}
    \centering
    \includegraphics[width=\linewidth]{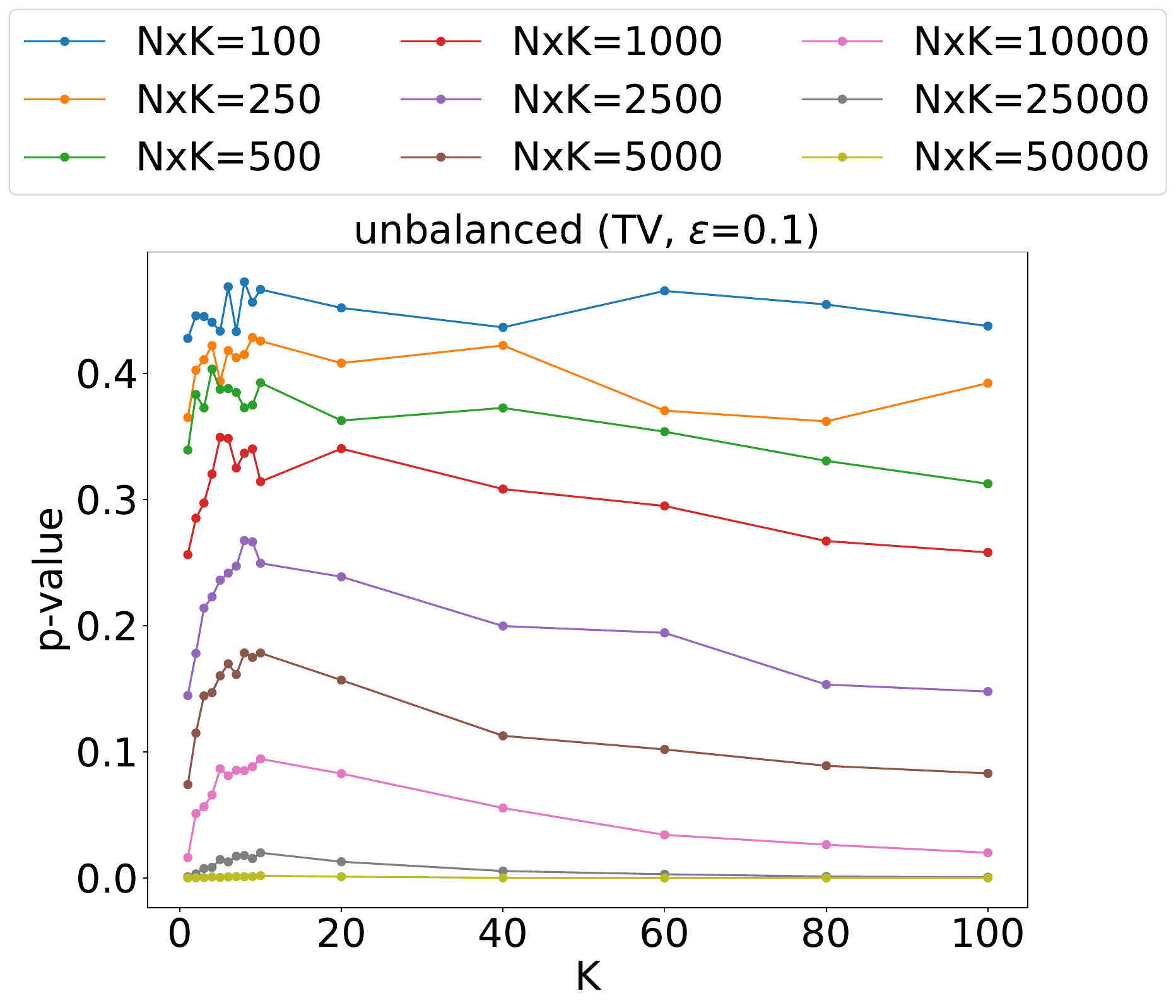}
    \caption{$\epsilon = 0.1$}
    \label{fig:gamma_MAE_cat2_e01}
  \end{subfigure} \hfill
  \begin{subfigure}[b]{0.24\linewidth}
    \centering
    \includegraphics[width=\linewidth]{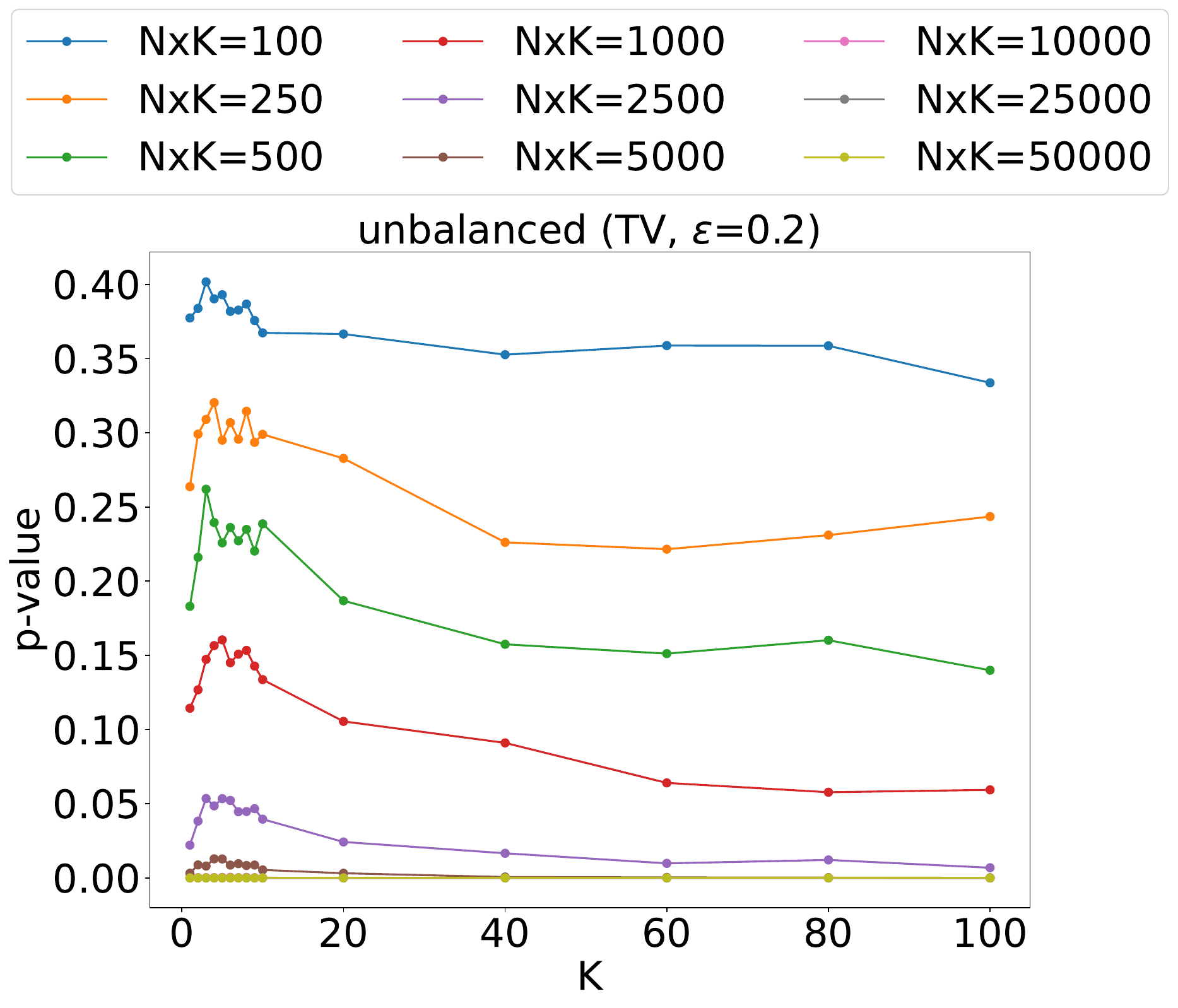}
    \caption{$\epsilon = 0.2$}
    \label{fig:gamma_MAE_cat2_e02}
  \end{subfigure} \hfill
  \begin{subfigure}[b]{0.24\linewidth}
    \centering
    \includegraphics[width=\linewidth]{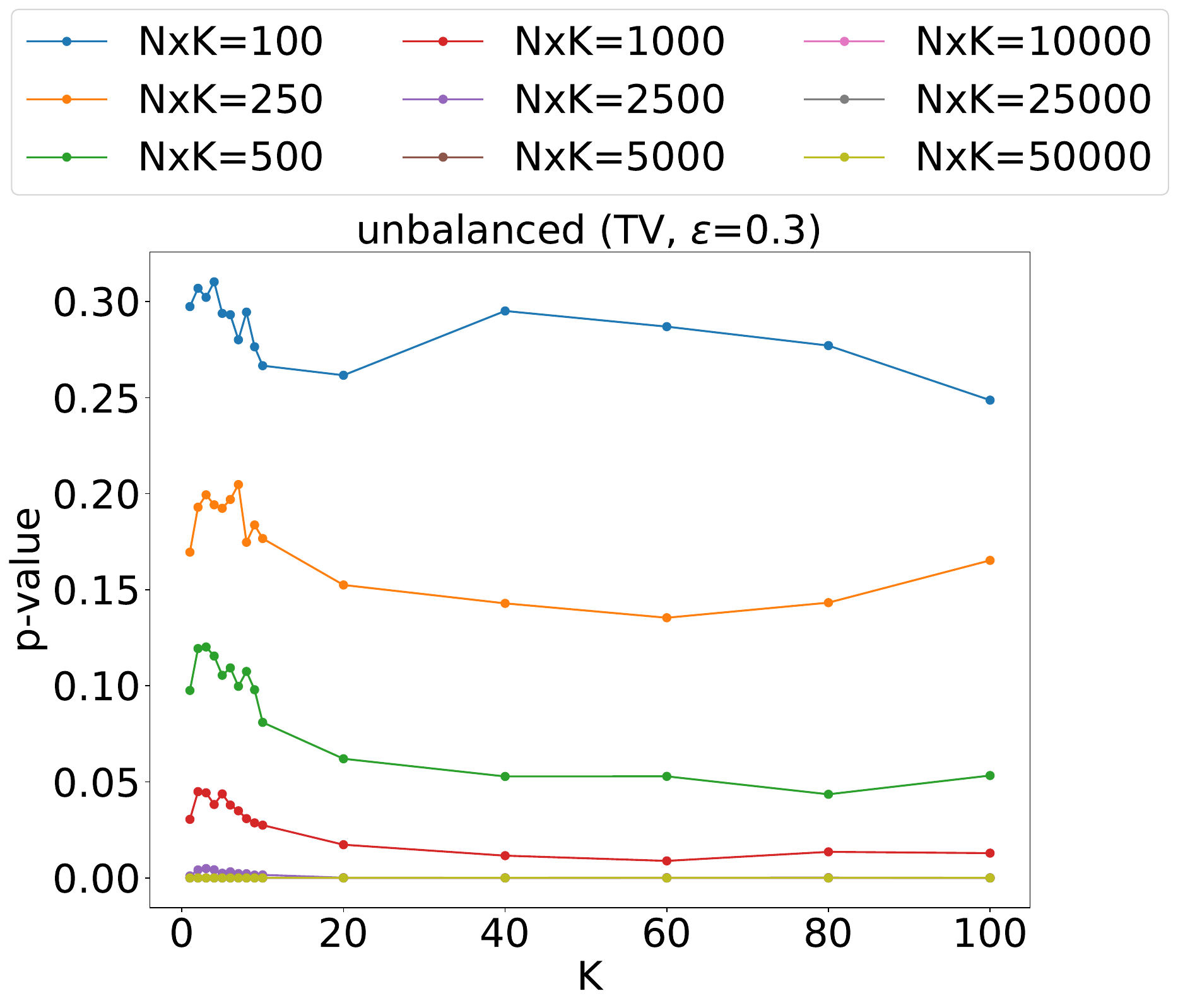}
    \caption{$\epsilon = 0.3$}
    \label{fig:gamma_MAE_cat2_e03}
  \end{subfigure} \hfill
  \begin{subfigure}[b]{0.24\linewidth}
    \centering
    \includegraphics[width=\linewidth]{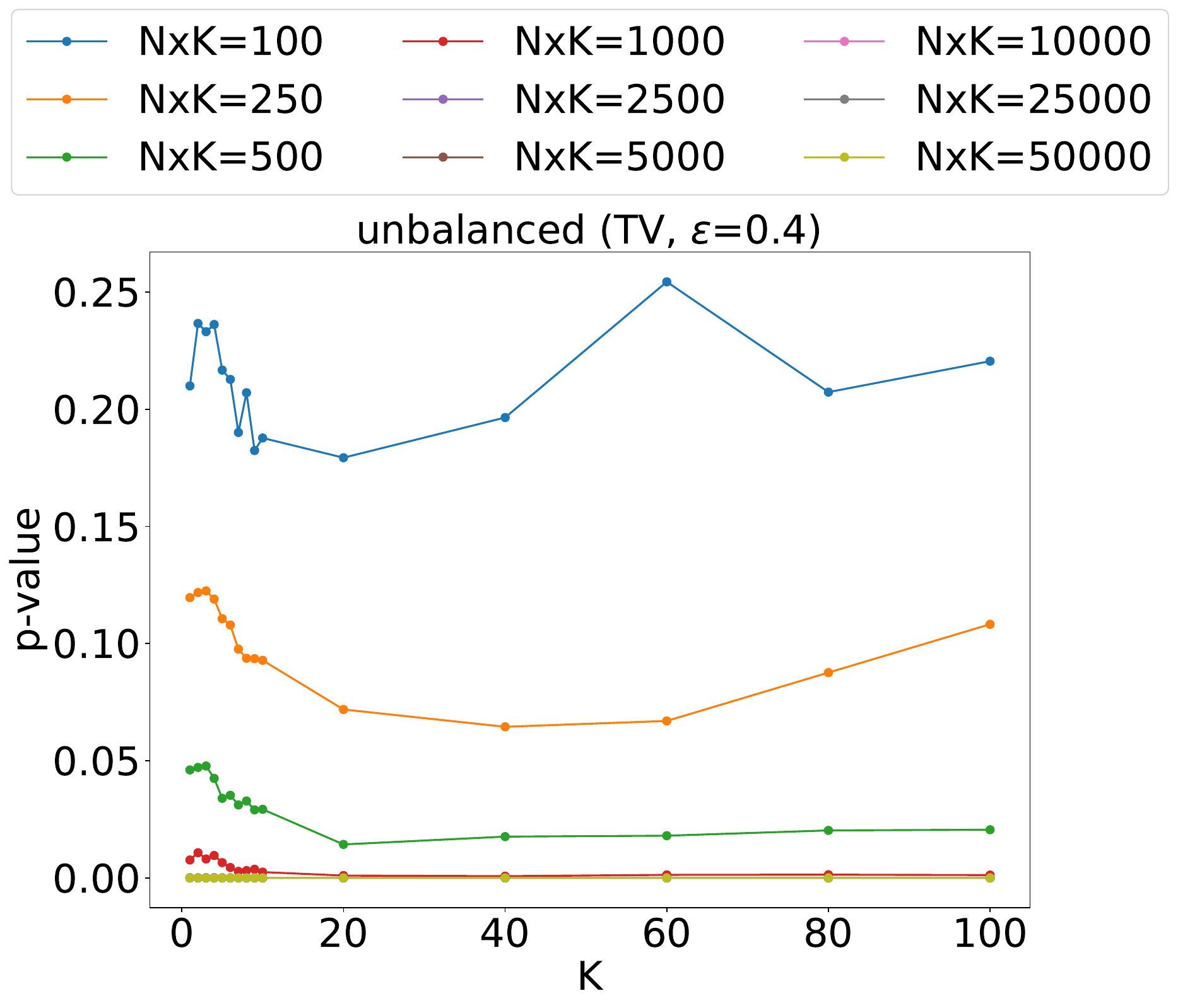}
    \caption{$\epsilon = 0.4$}
    \label{fig:gamma_MAE_cat2_e04}
  \end{subfigure}
  \caption{P-value plots for unbalanced alphas with TV as the metric ($M=2$)}
  \label{fig:gamma_MAE_cat2}
\end{figure*}

\begin{figure*}
  \centering
  \begin{subfigure}[b]{0.24\linewidth}
    \centering
    \includegraphics[width=\linewidth]{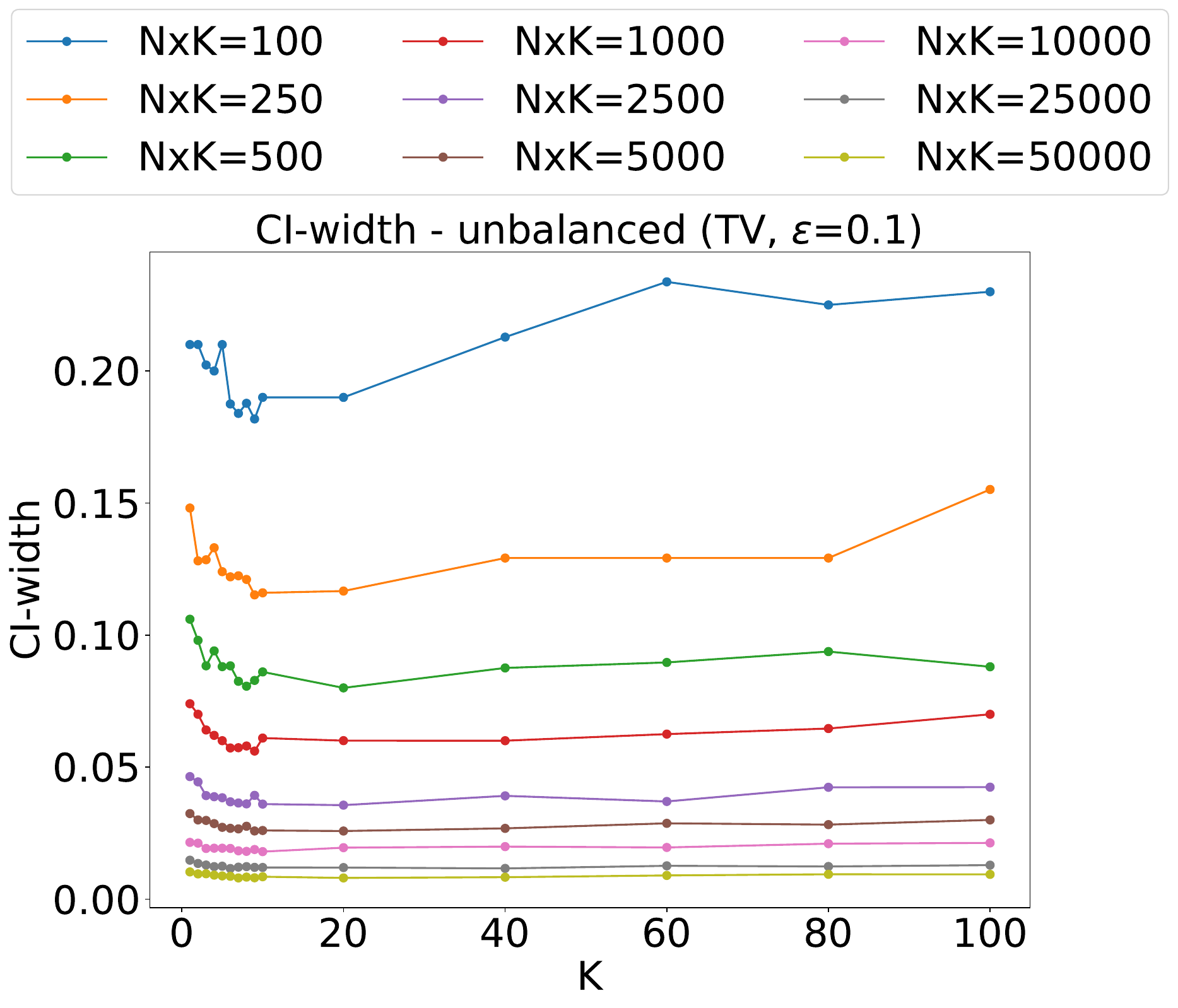}
    \caption{$\epsilon = 0.1$}
    \label{fig:gamma_ci_MAE_cat2_e01}
  \end{subfigure} \hfill
  \begin{subfigure}[b]{0.24\linewidth}
    \centering
    \includegraphics[width=\linewidth]{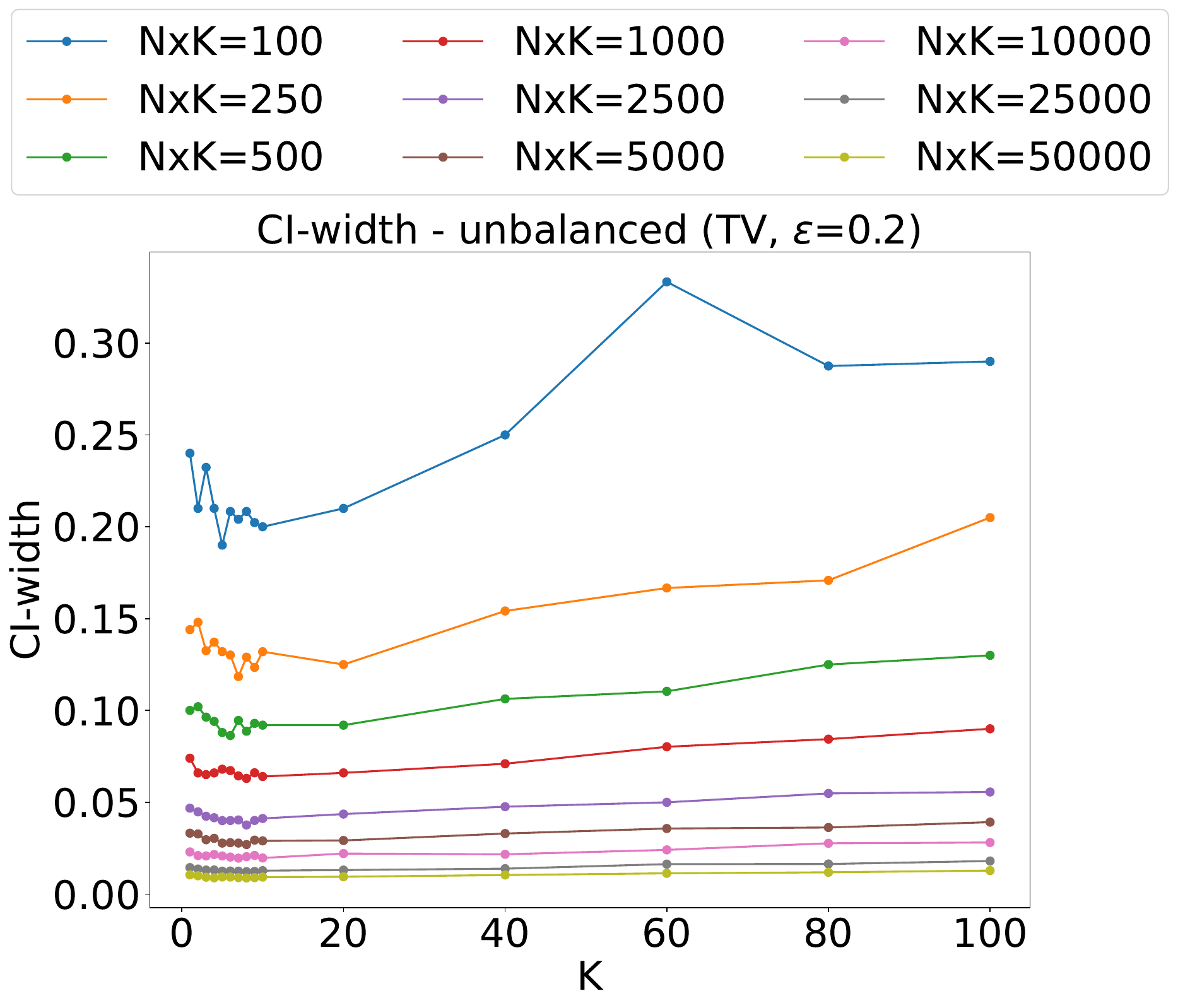}
    \caption{$\epsilon = 0.2$}
    \label{fig:gamma_ci_MAE_cat2_e02}
  \end{subfigure} \hfill
  \begin{subfigure}[b]{0.24\linewidth}
    \centering
    \includegraphics[width=\linewidth]{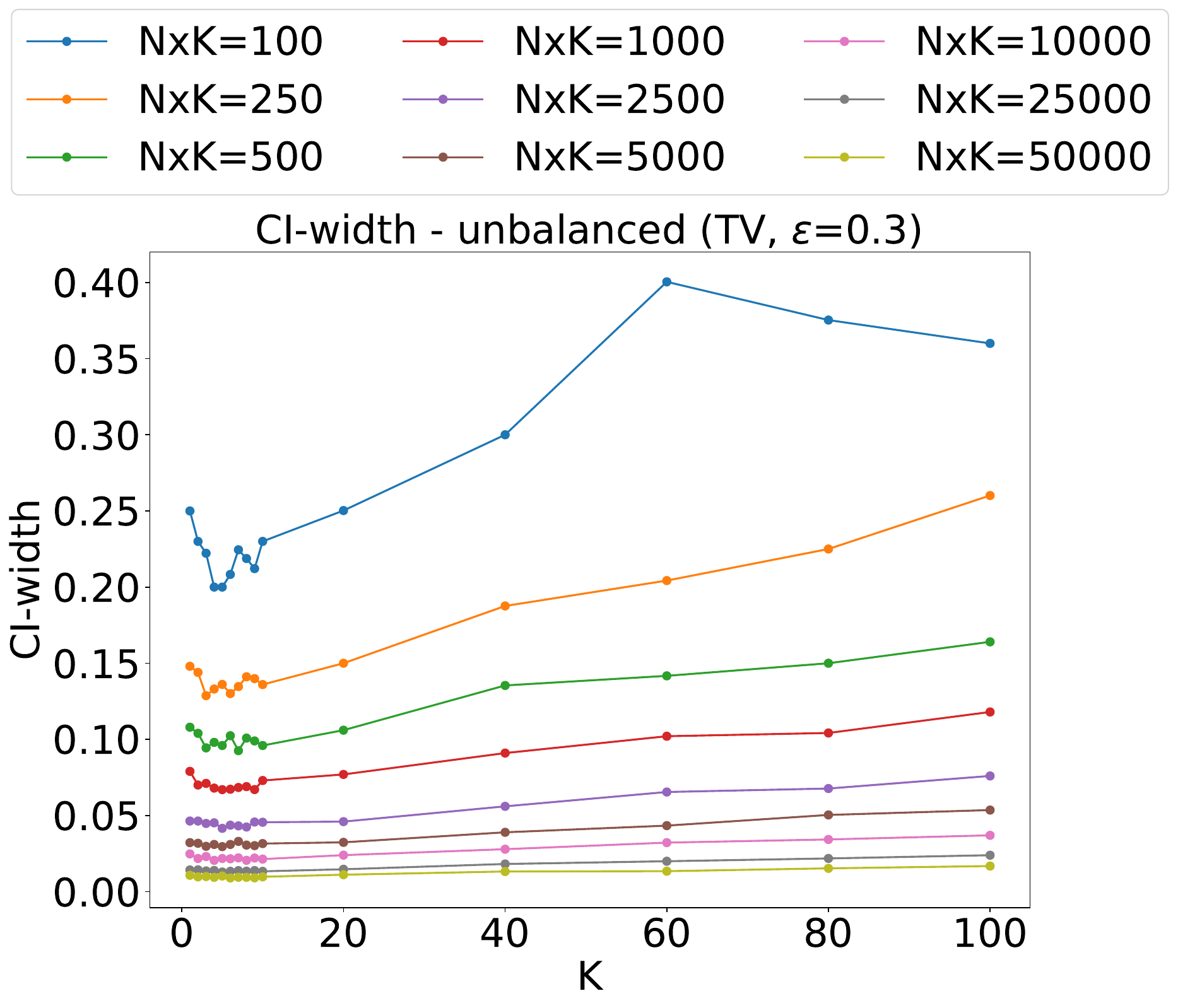}
    \caption{$\epsilon = 0.3$}
    \label{fig:gamma_ci_MAE_cat2_e03}
  \end{subfigure} \hfill
  \begin{subfigure}[b]{0.24\linewidth}
    \centering
    \includegraphics[width=\linewidth]{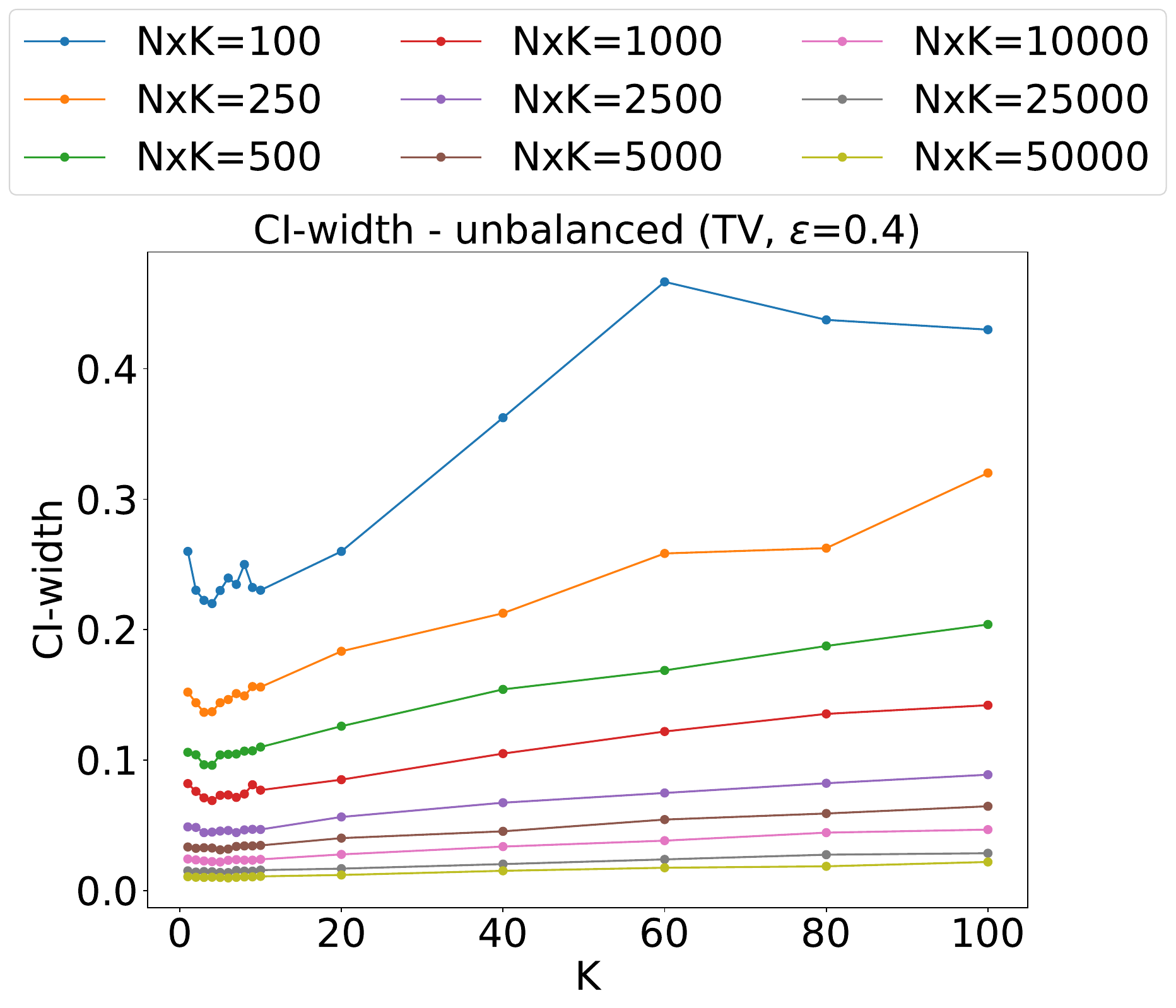}
    \caption{$\epsilon = 0.4$}
    \label{fig:gamma_ci_MAE_cat2_e04}
  \end{subfigure}
  \caption{CI-width plots for unbalanced alphas with TV as the metric ($M=2$)}
  \label{fig:gamma_ci_MAE_cat2}
\end{figure*}

\begin{figure*}
  \centering
  \begin{subfigure}[b]{0.24\linewidth}
    \centering
    \includegraphics[width=\linewidth]{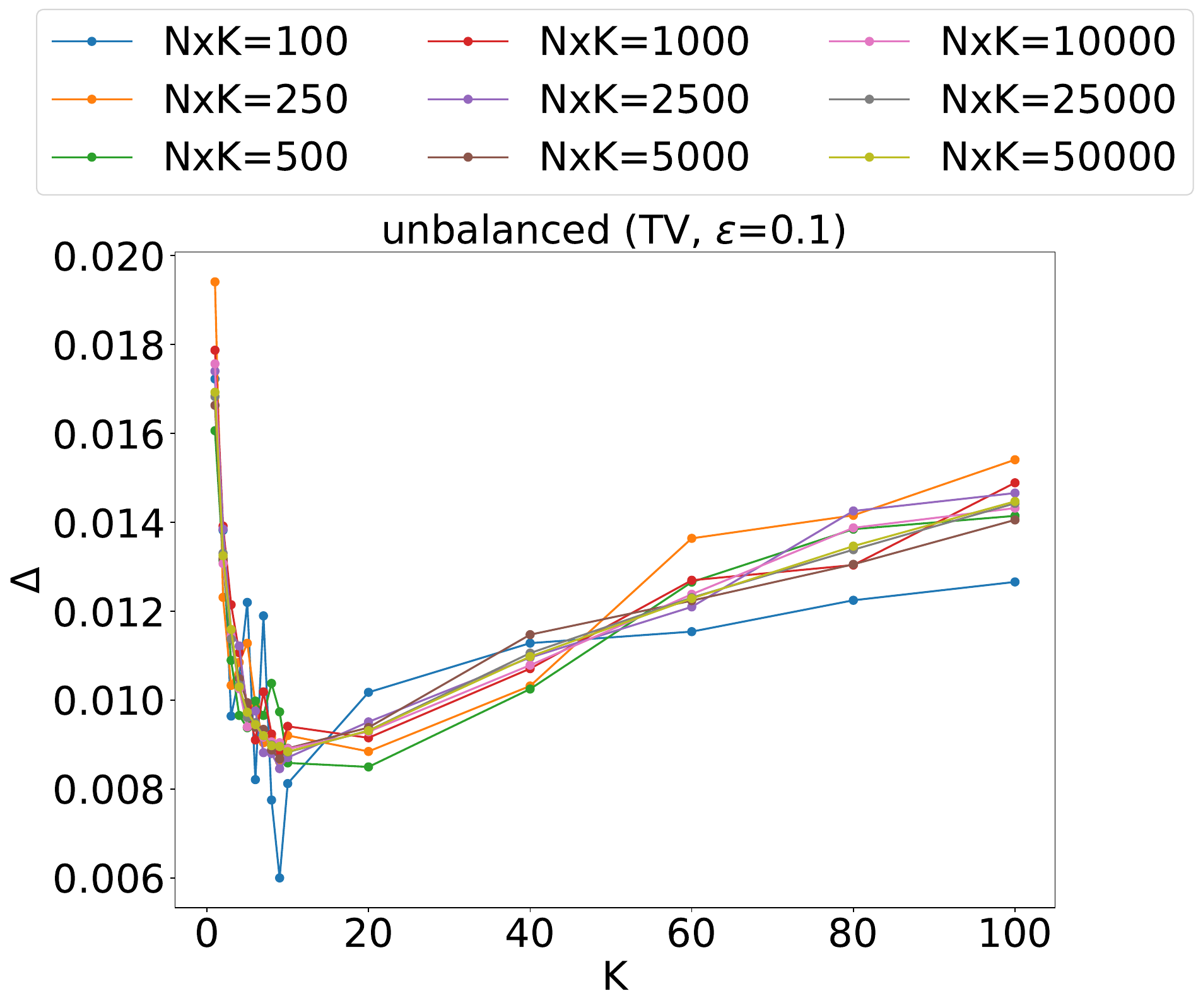}
    \caption{$\epsilon = 0.1$}
    \label{fig:gamma_delta_MAE_cat2_e01}
  \end{subfigure} \hfill
  \begin{subfigure}[b]{0.24\linewidth}
    \centering
    \includegraphics[width=\linewidth]{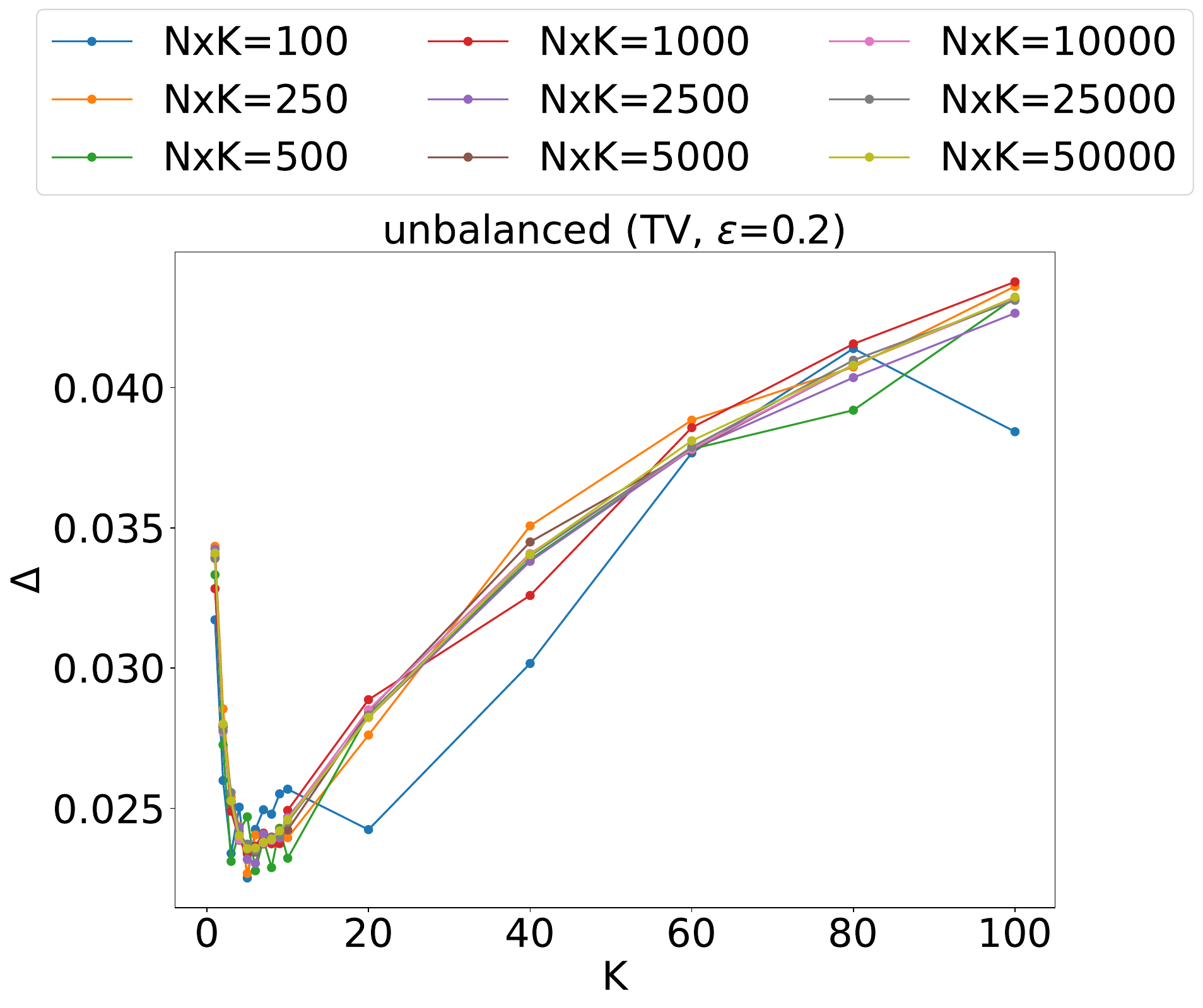}
    \caption{$\epsilon = 0.2$}
    \label{fig:gamma_delta_MAE_cat2_e02}
  \end{subfigure} \hfill
  \begin{subfigure}[b]{0.24\linewidth}
    \centering
    \includegraphics[width=\linewidth]{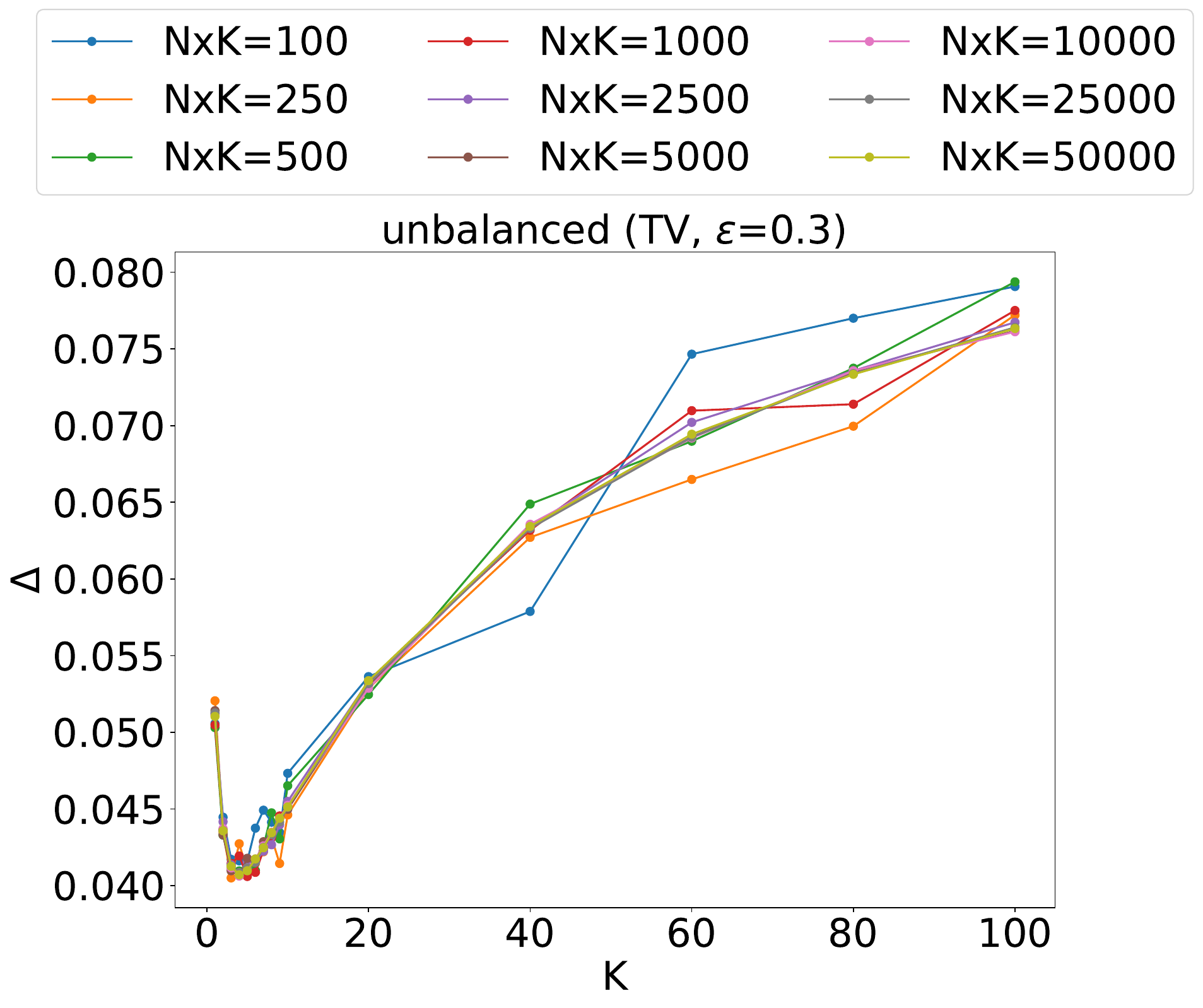}
    \caption{$\epsilon = 0.3$}
    \label{fig:gamma_delta_MAE_cat2_e03}
  \end{subfigure} \hfill
  \begin{subfigure}[b]{0.24\linewidth}
    \centering
    \includegraphics[width=\linewidth]{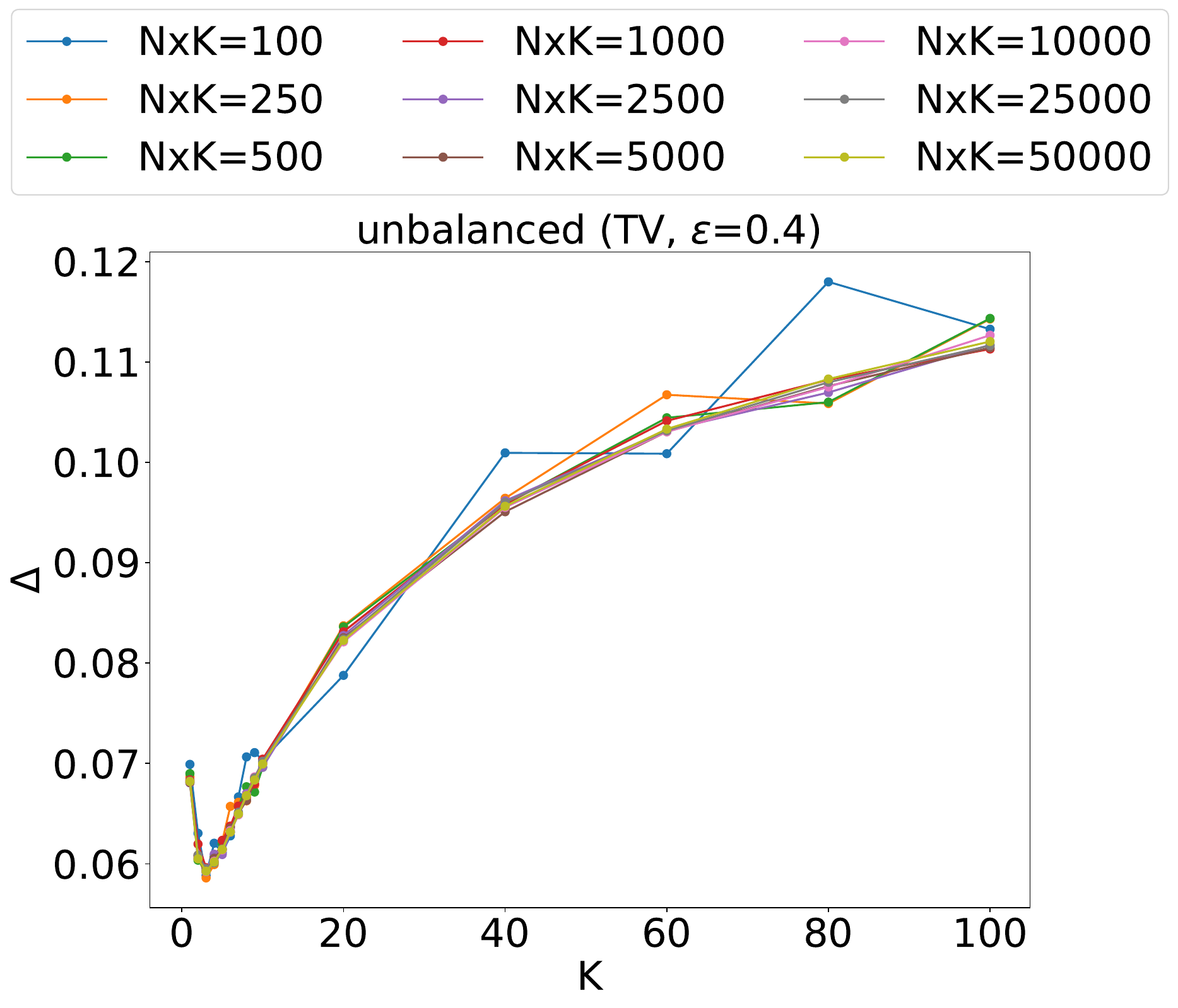}
    \caption{$\epsilon = 0.4$}
    \label{fig:gamma_delta_MAE_cat2_e04}
  \end{subfigure}
  \caption{Effect sizes ($\Delta$) for unbalanced alphas with TV as the metric ($M=2$)}
  \label{fig:gamma_delta_MAE_cat2}
\end{figure*}

\begin{figure*}
  \centering
  \begin{subfigure}[b]{0.24\linewidth}
    \centering
    \includegraphics[width=\linewidth]{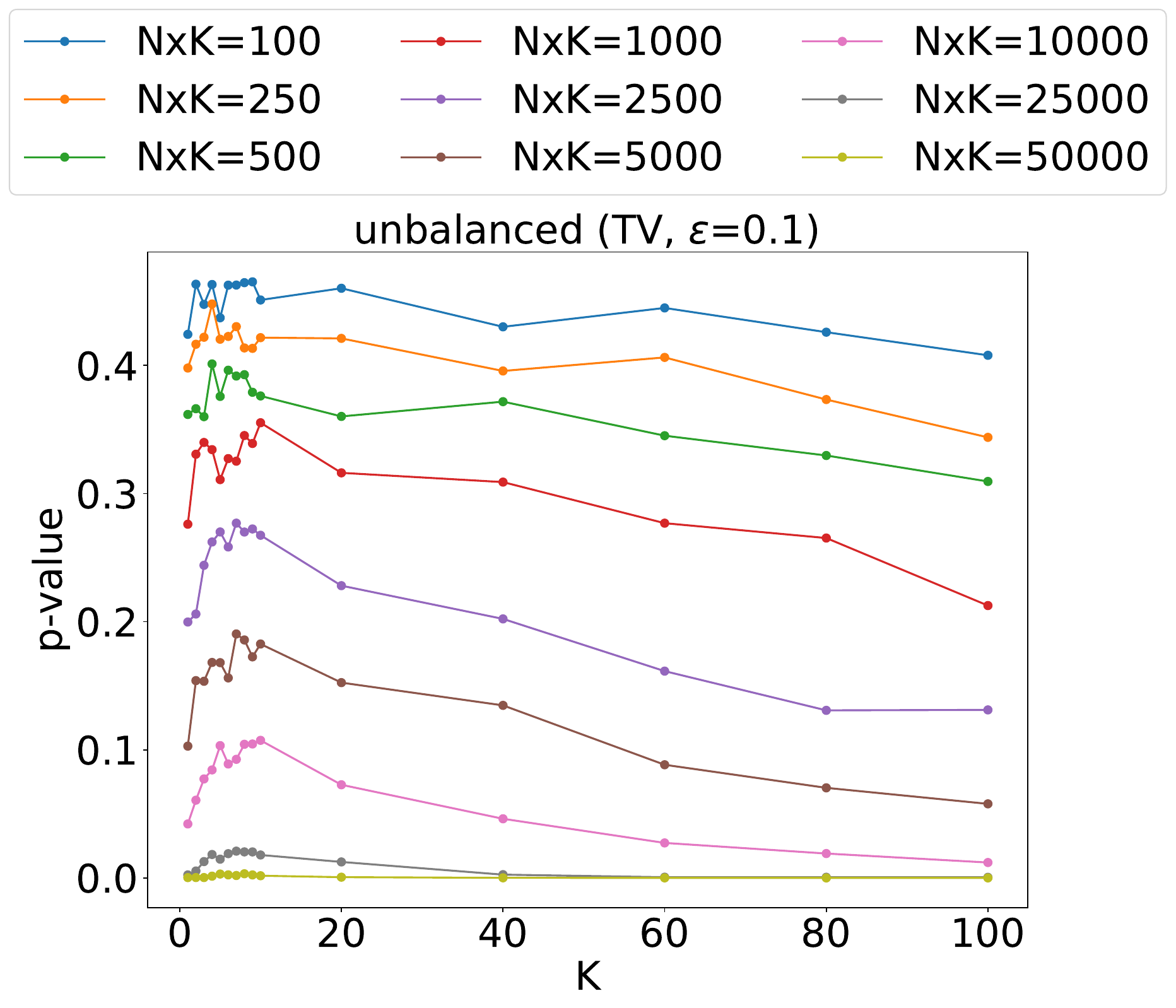}
    \caption{$\epsilon = 0.1$}
    \label{fig:gamma_MAE_cat3_e01}
  \end{subfigure} \hfill
  \begin{subfigure}[b]{0.24\linewidth}
    \centering
    \includegraphics[width=\linewidth]{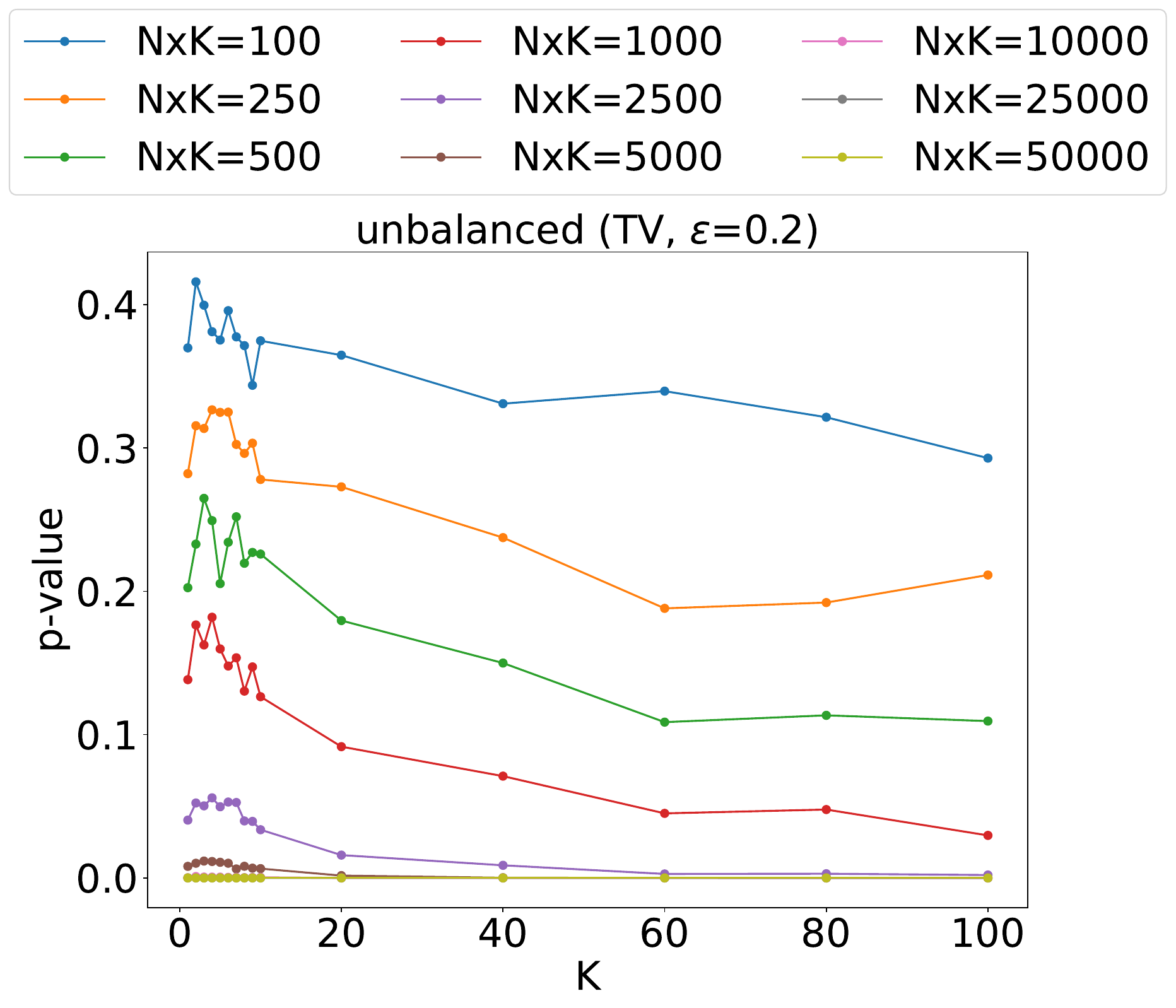}
    \caption{$\epsilon = 0.2$}
    \label{fig:gamma_MAE_cat3_e02}
  \end{subfigure} \hfill
  \begin{subfigure}[b]{0.24\linewidth}
    \centering
    \includegraphics[width=\linewidth]{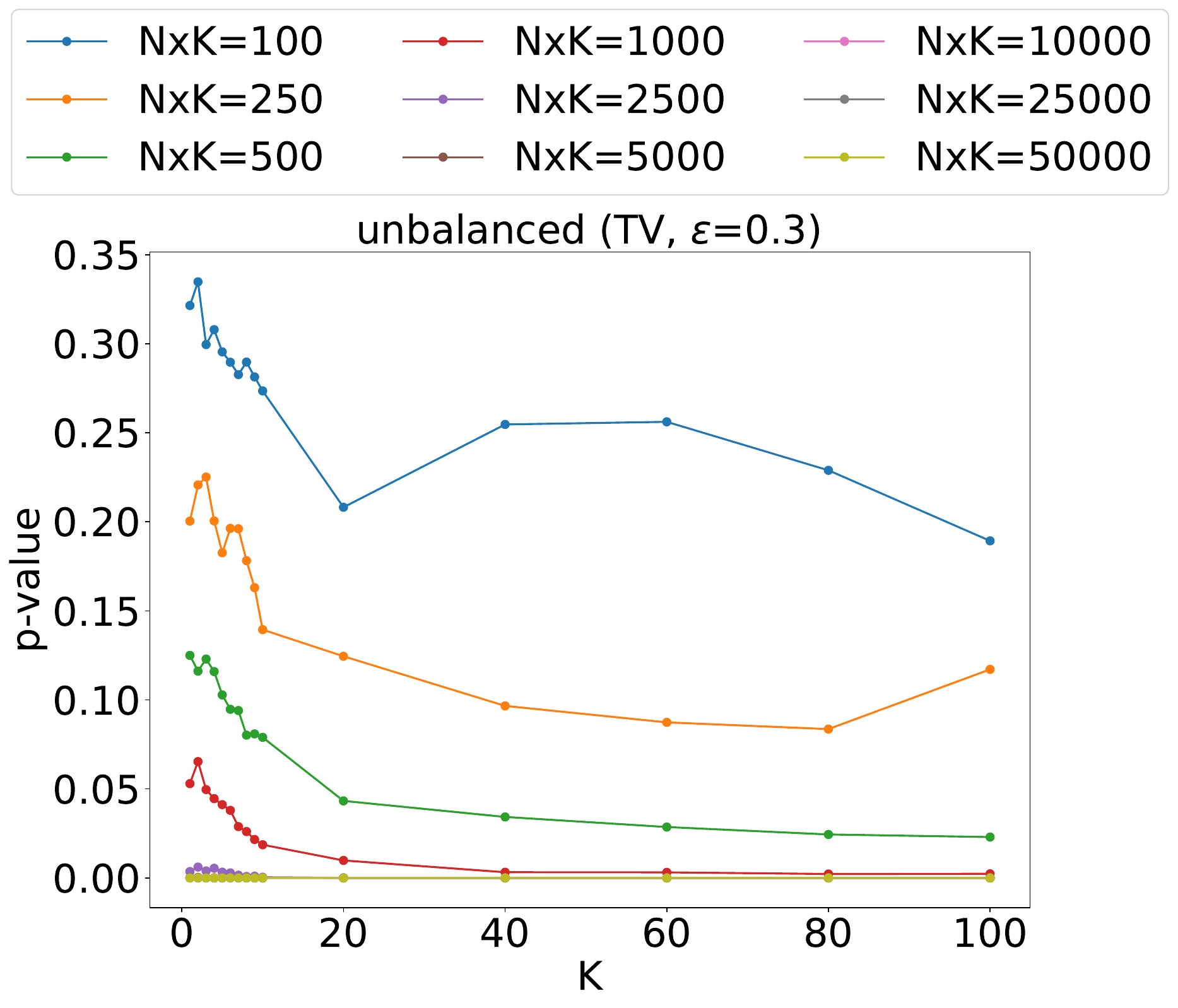}
    \caption{$\epsilon = 0.3$}
    \label{fig:gamma_MAE_cat3_e03}
  \end{subfigure} \hfill
  \begin{subfigure}[b]{0.24\linewidth}
    \centering
    \includegraphics[width=\linewidth]{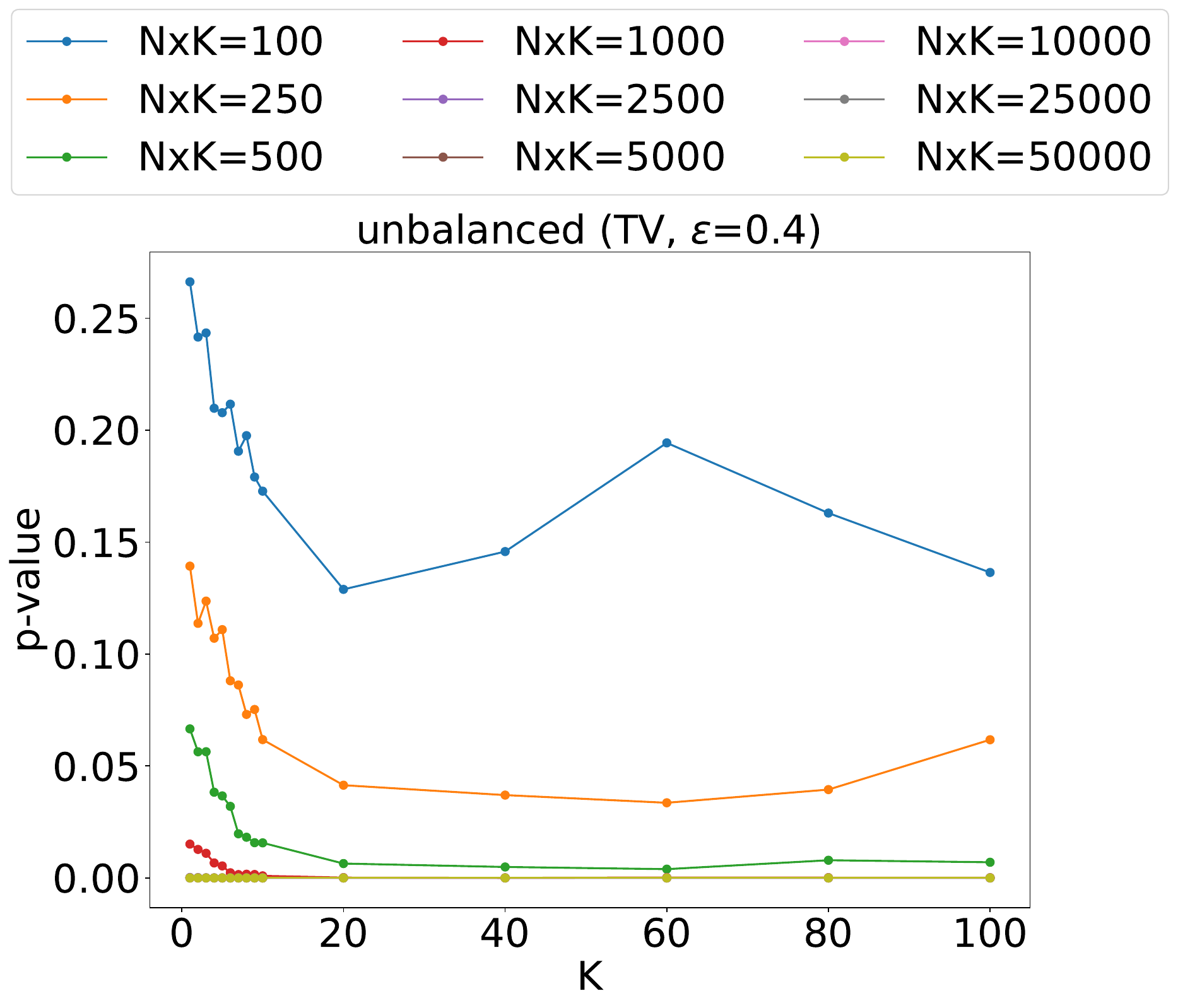}
    \caption{$\epsilon = 0.4$}
    \label{fig:gamma_MAE_cat3_e04}
  \end{subfigure}
  \caption{P-value plots for unbalanced alphas with TV as the metric ($M=3$)}
  \label{fig:gamma_MAE_cat3}
\end{figure*}

\begin{figure*}
  \centering
  \begin{subfigure}[b]{0.24\linewidth}
    \centering
    \includegraphics[width=\linewidth]{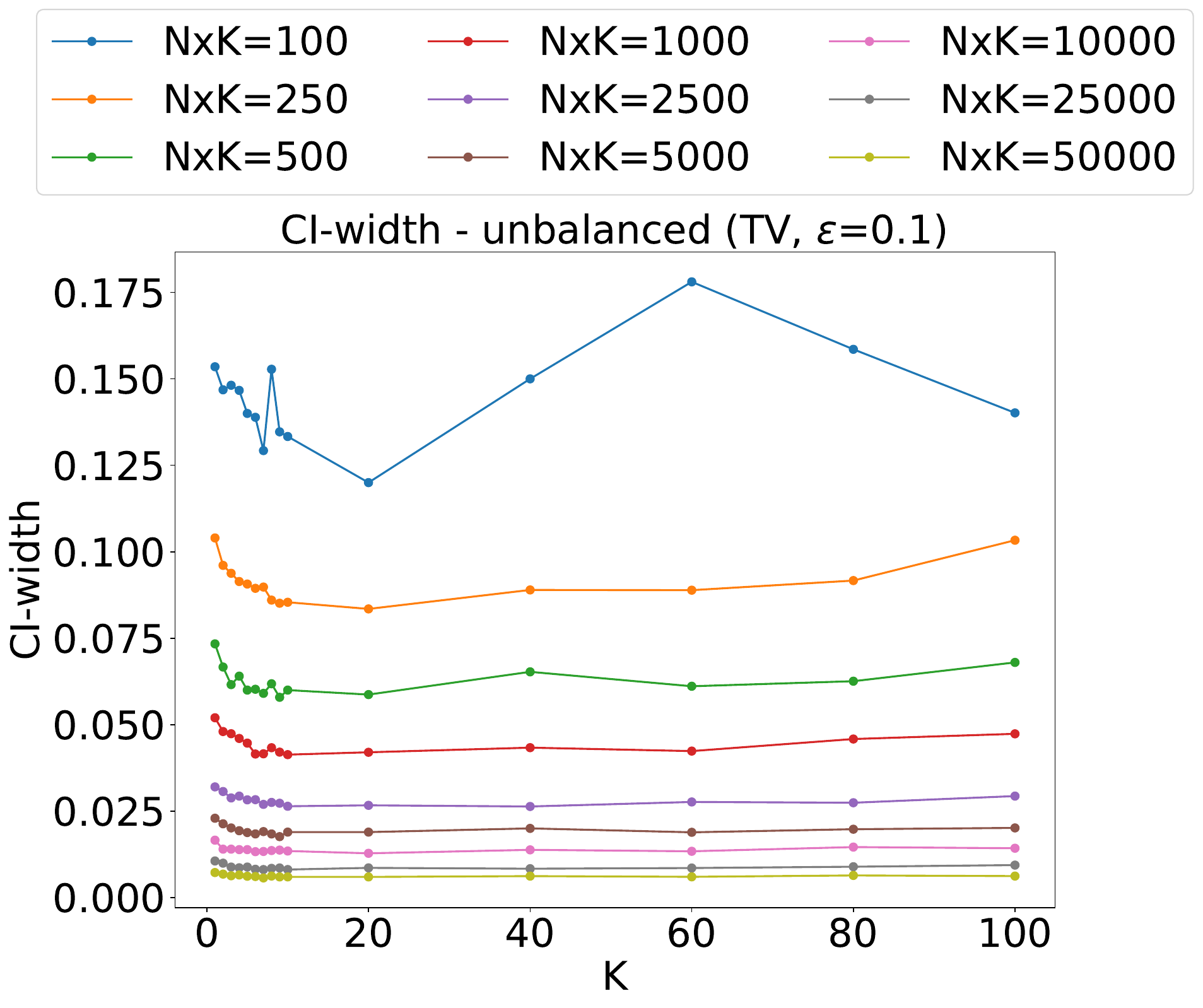}
    \caption{$\epsilon = 0.1$}
    \label{fig:gamma_ci_MAE_cat3_e01}
  \end{subfigure} \hfill
  \begin{subfigure}[b]{0.24\linewidth}
    \centering
    \includegraphics[width=\linewidth]{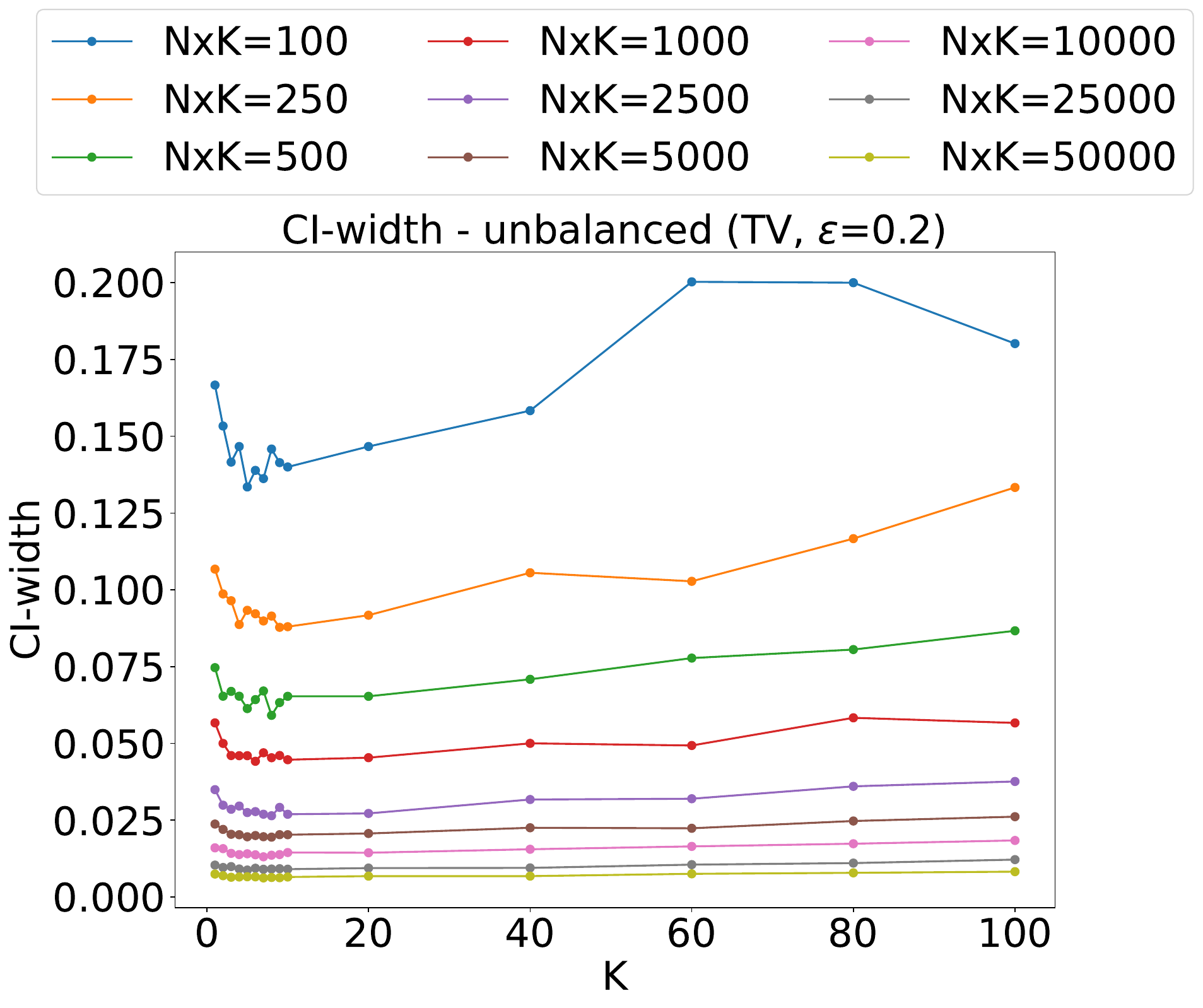}
    \caption{$\epsilon = 0.2$}
    \label{fig:gamma_ci_MAE_cat3_e02}
  \end{subfigure} \hfill
  \begin{subfigure}[b]{0.24\linewidth}
    \centering
    \includegraphics[width=\linewidth]{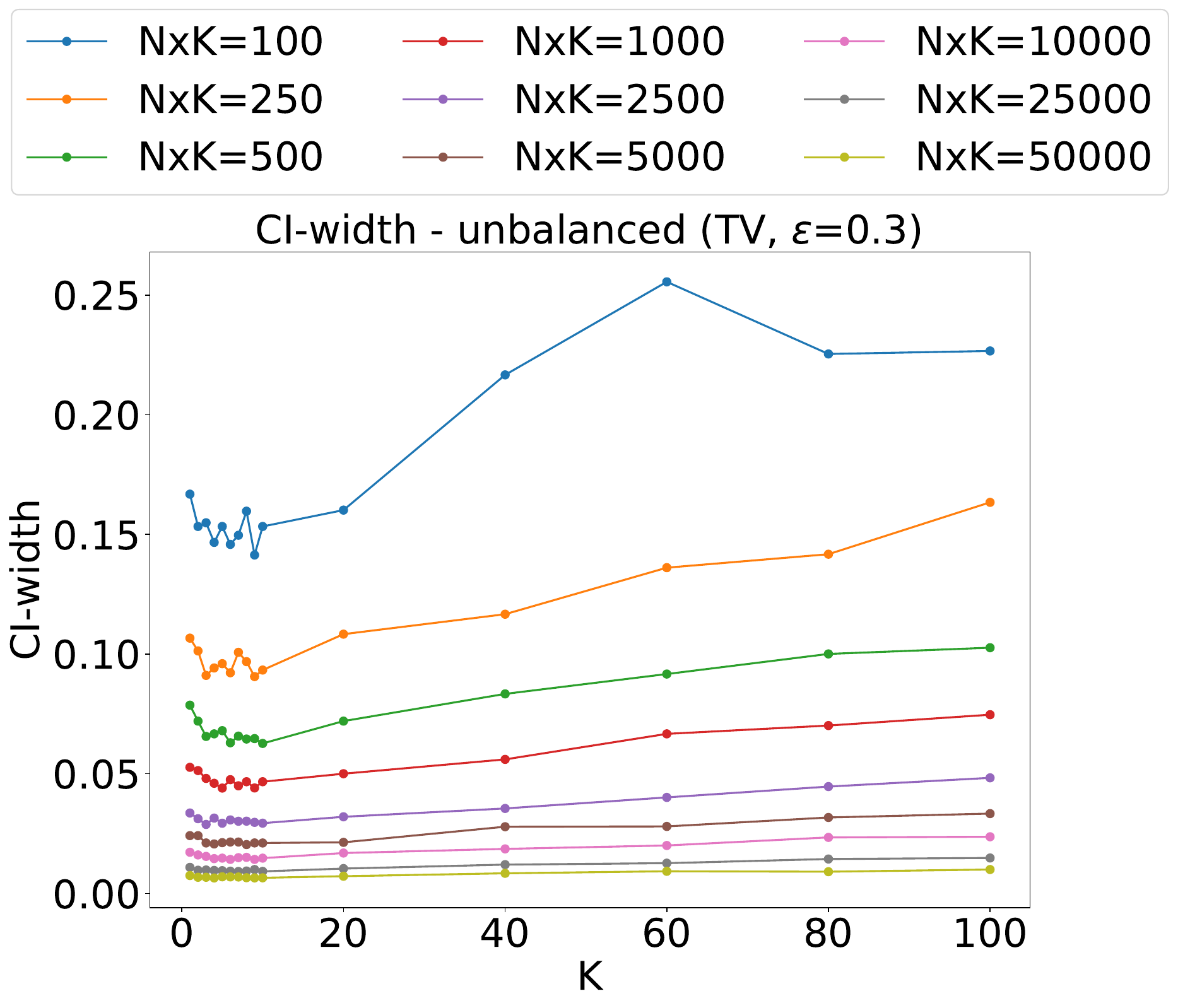}
    \caption{$\epsilon = 0.3$}
    \label{fig:gamma_ci_MAE_cat3_e03}
  \end{subfigure} \hfill
  \begin{subfigure}[b]{0.24\linewidth}
    \centering
    \includegraphics[width=\linewidth]{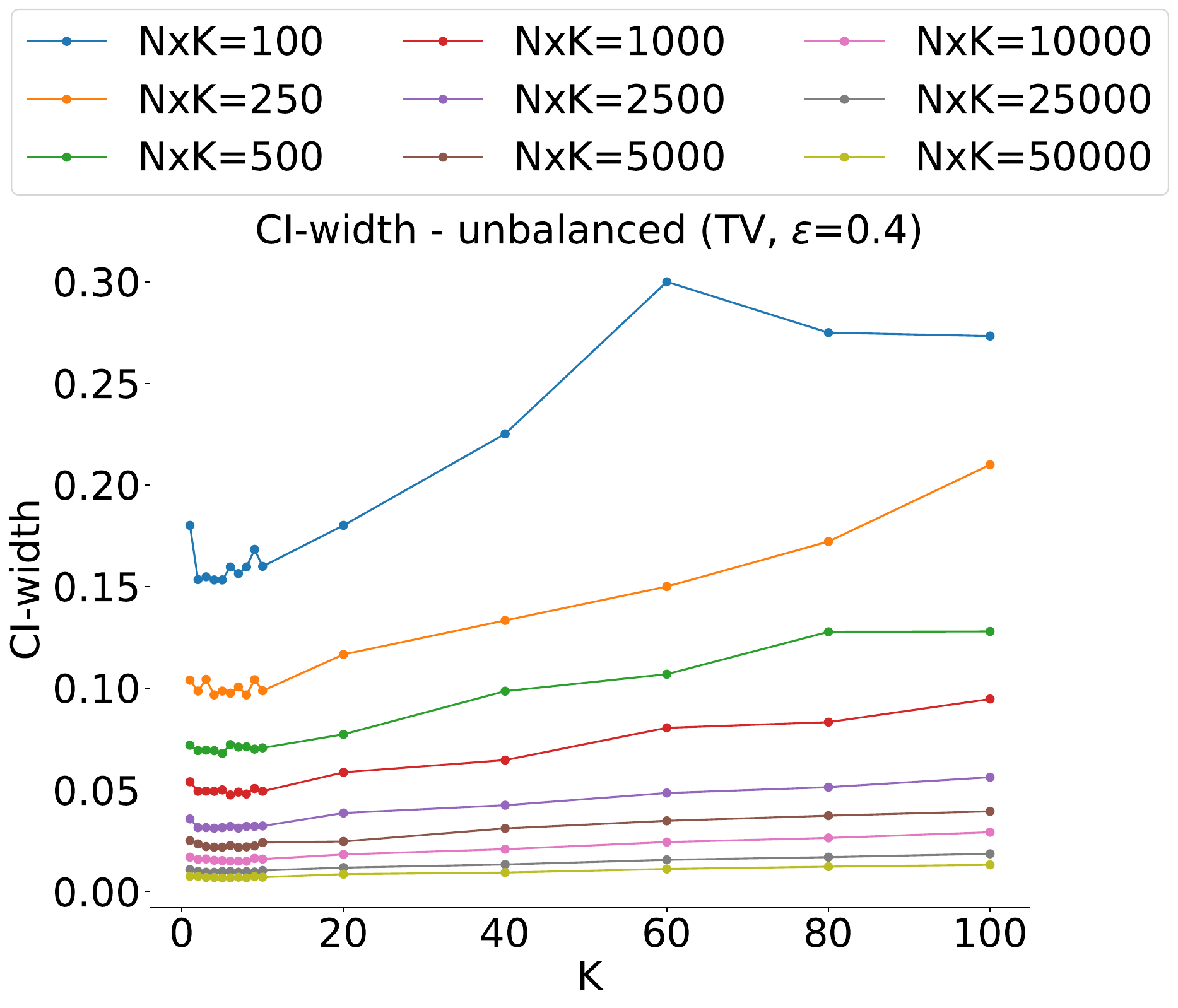}
    \caption{$\epsilon = 0.4$}
    \label{fig:gamma_ci_MAE_cat3_e04}
  \end{subfigure}
  \caption{CI-width plots for unbalanced alphas with TV as the metric ($M=3$)}
  \label{fig:gamma_ci_MAE_cat3}
\end{figure*}

\begin{figure*}
  \centering
  \begin{subfigure}[b]{0.24\linewidth}
    \centering
    \includegraphics[width=\linewidth]{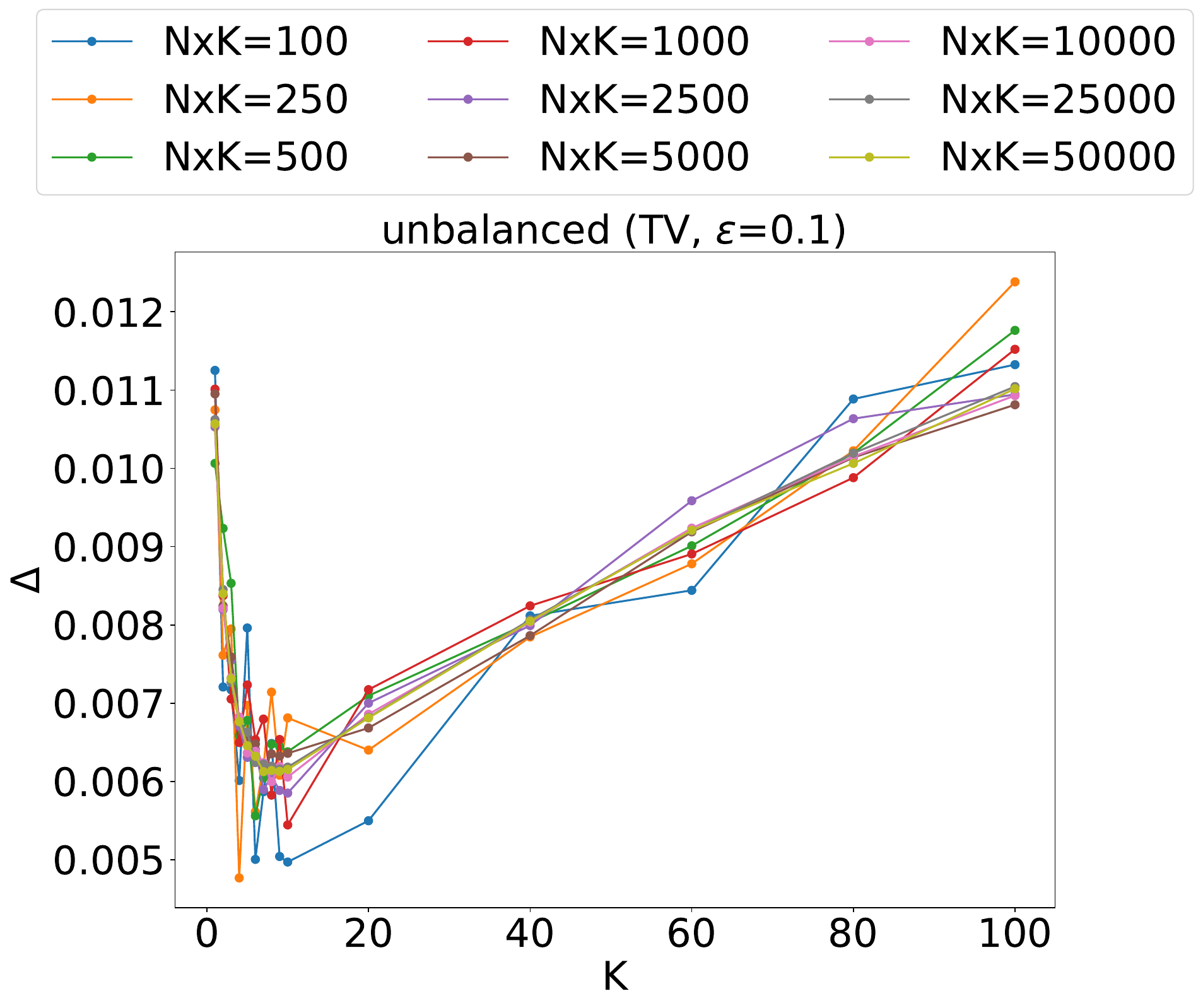}
    \caption{$\epsilon = 0.1$}
    \label{fig:gamma_delta_MAE_cat3_e01}
  \end{subfigure} \hfill
  \begin{subfigure}[b]{0.24\linewidth}
    \centering
    \includegraphics[width=\linewidth]{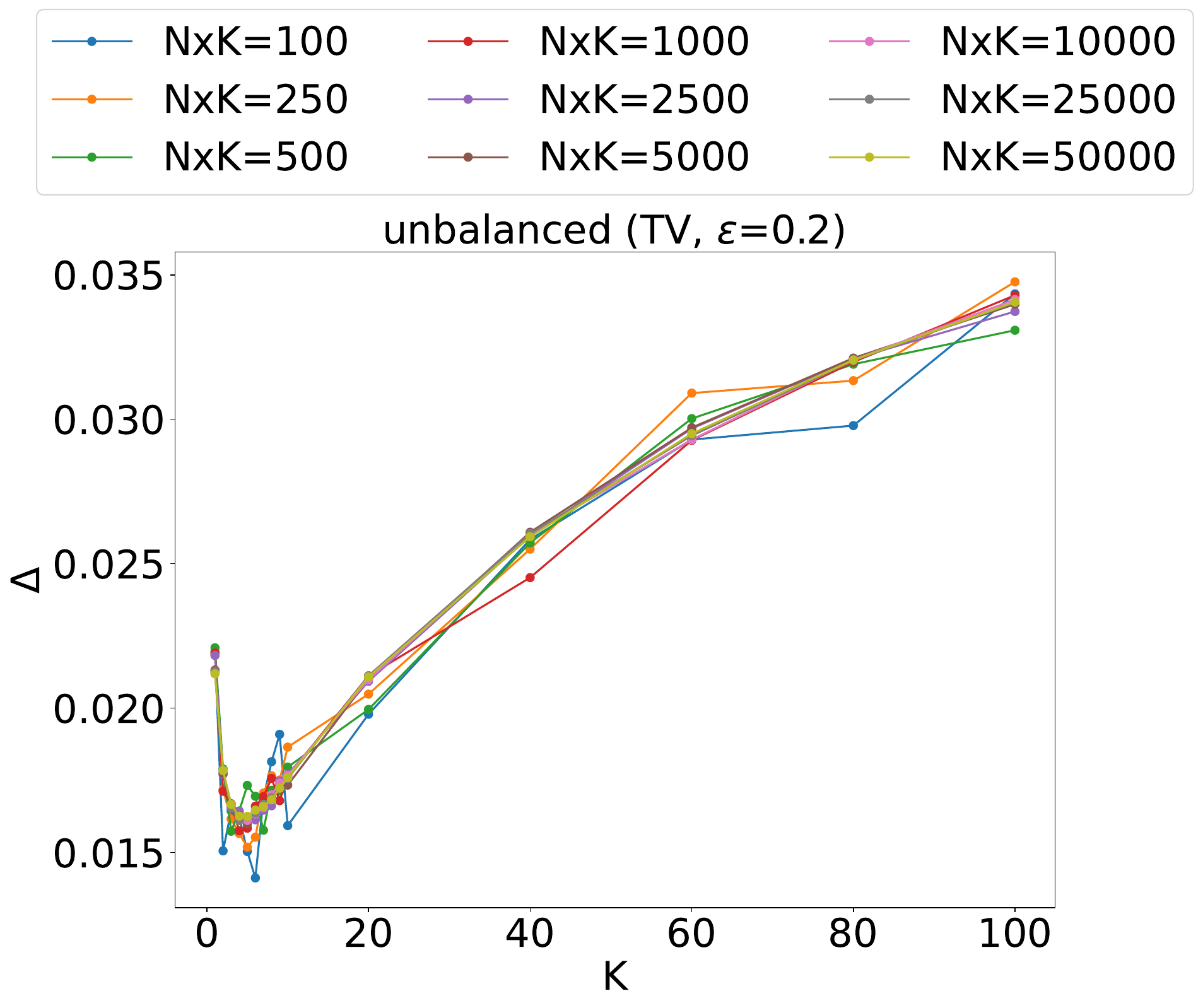}
    \caption{$\epsilon = 0.2$}
    \label{fig:gamma_delta_MAE_cat3_e02}
  \end{subfigure} \hfill
  \begin{subfigure}[b]{0.24\linewidth}
    \centering
    \includegraphics[width=\linewidth]{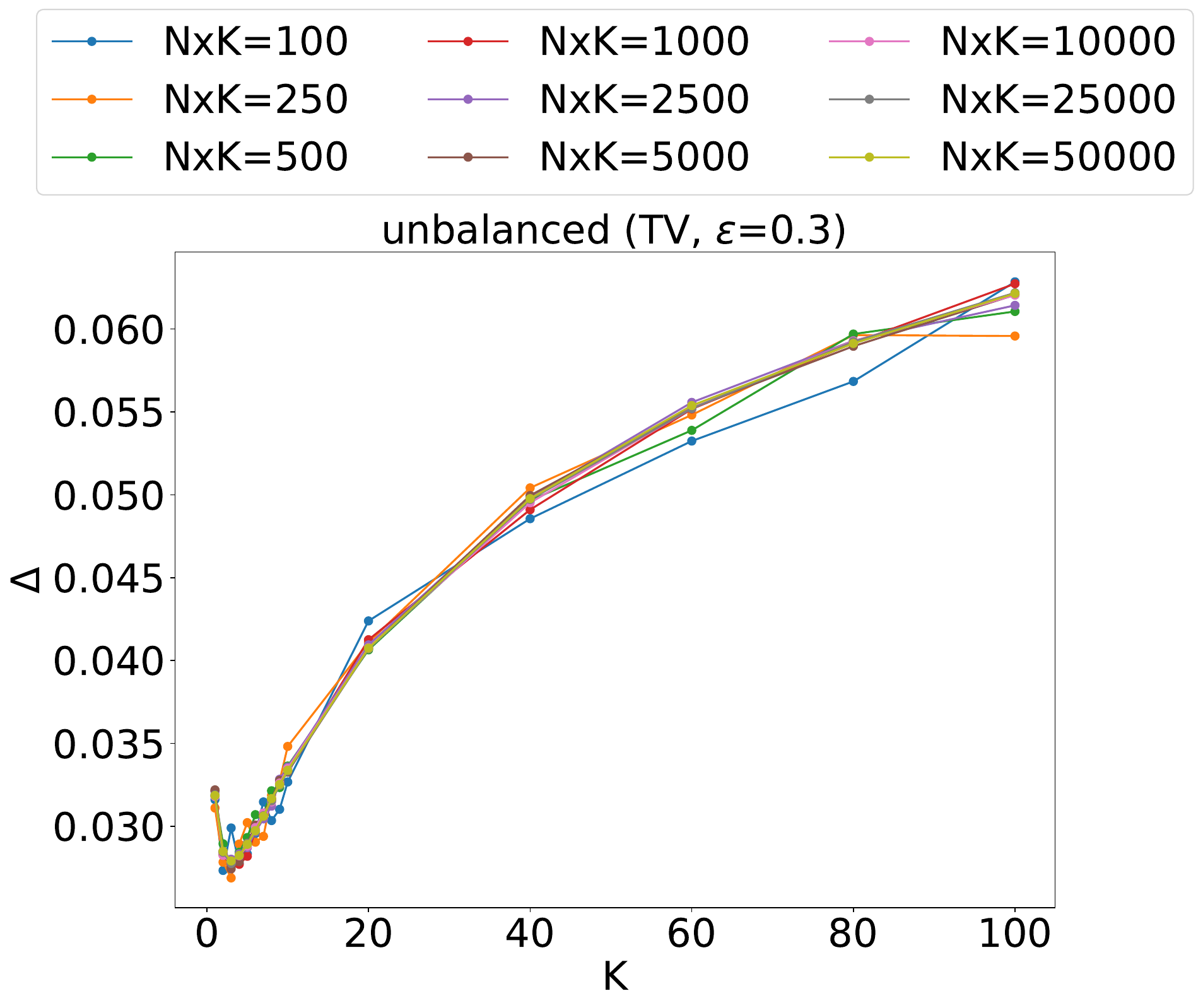}
    \caption{$\epsilon = 0.3$}
    \label{fig:gamma_delta_MAE_cat3_e03}
  \end{subfigure} \hfill
  \begin{subfigure}[b]{0.24\linewidth}
    \centering
    \includegraphics[width=\linewidth]{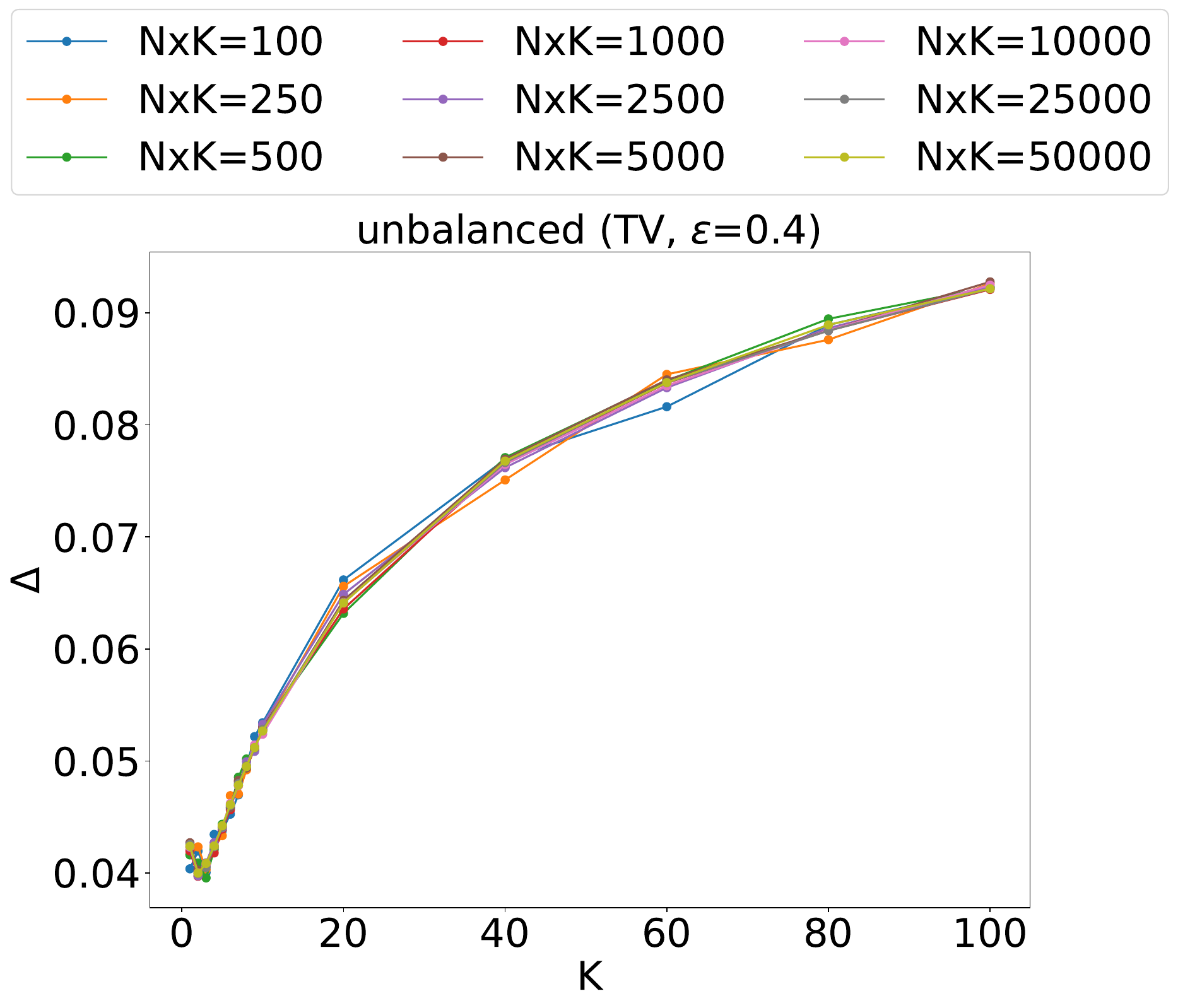}
    \caption{$\epsilon = 0.4$}
    \label{fig:gamma_delta_MAE_cat3_e04}
  \end{subfigure}
  \caption{Effect sizes ($\Delta$) for unbalanced alphas with TV as the metric ($M=3$)}
  \label{fig:gamma_delta_MAE_cat3}
\end{figure*}

\begin{figure*}
  \centering
  \begin{subfigure}[b]{0.24\linewidth}
    \centering
    \includegraphics[width=\linewidth]{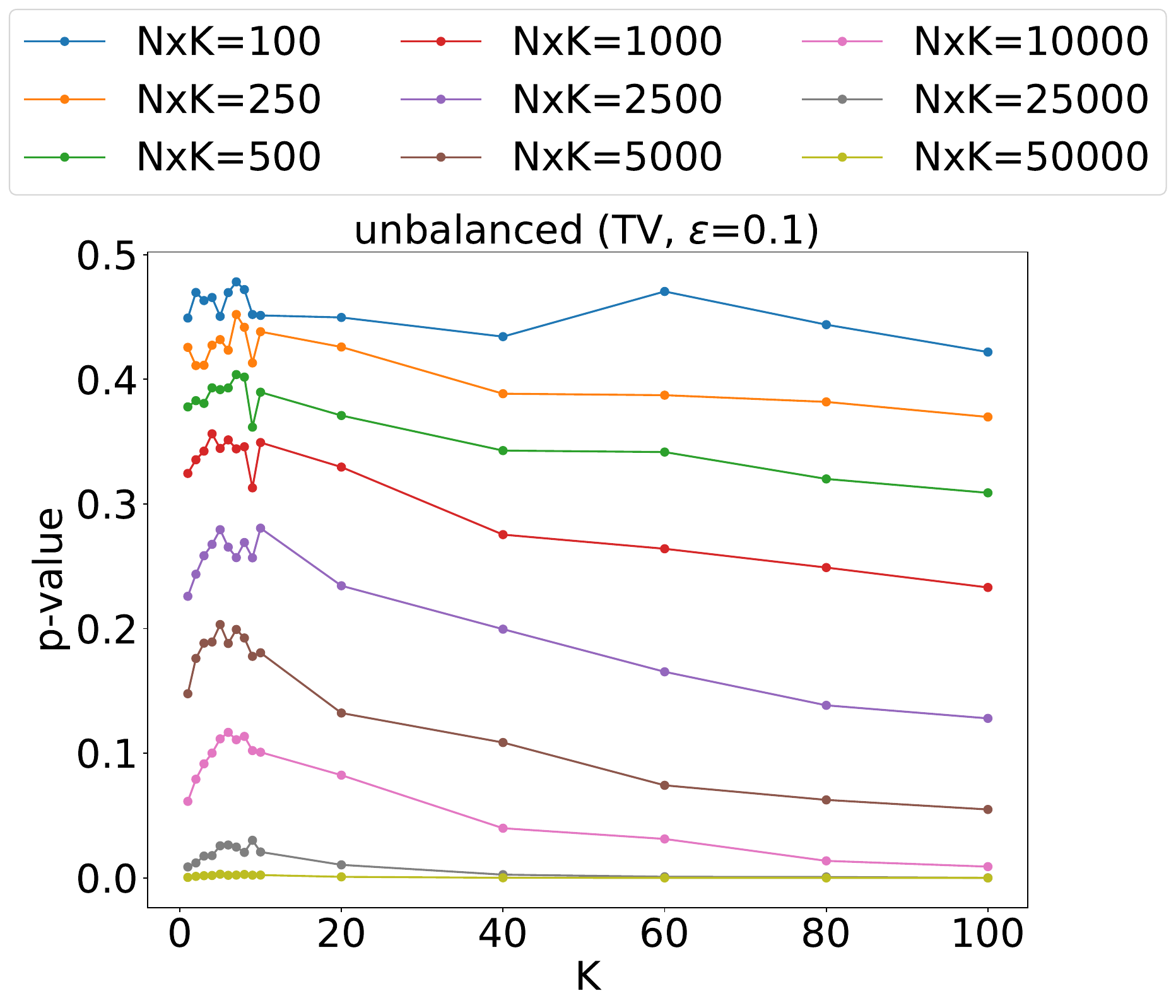}
    \caption{$\epsilon = 0.1$}
    \label{fig:gamma_MAE_cat4_e01}
  \end{subfigure} \hfill
  \begin{subfigure}[b]{0.24\linewidth}
    \centering
    \includegraphics[width=\linewidth]{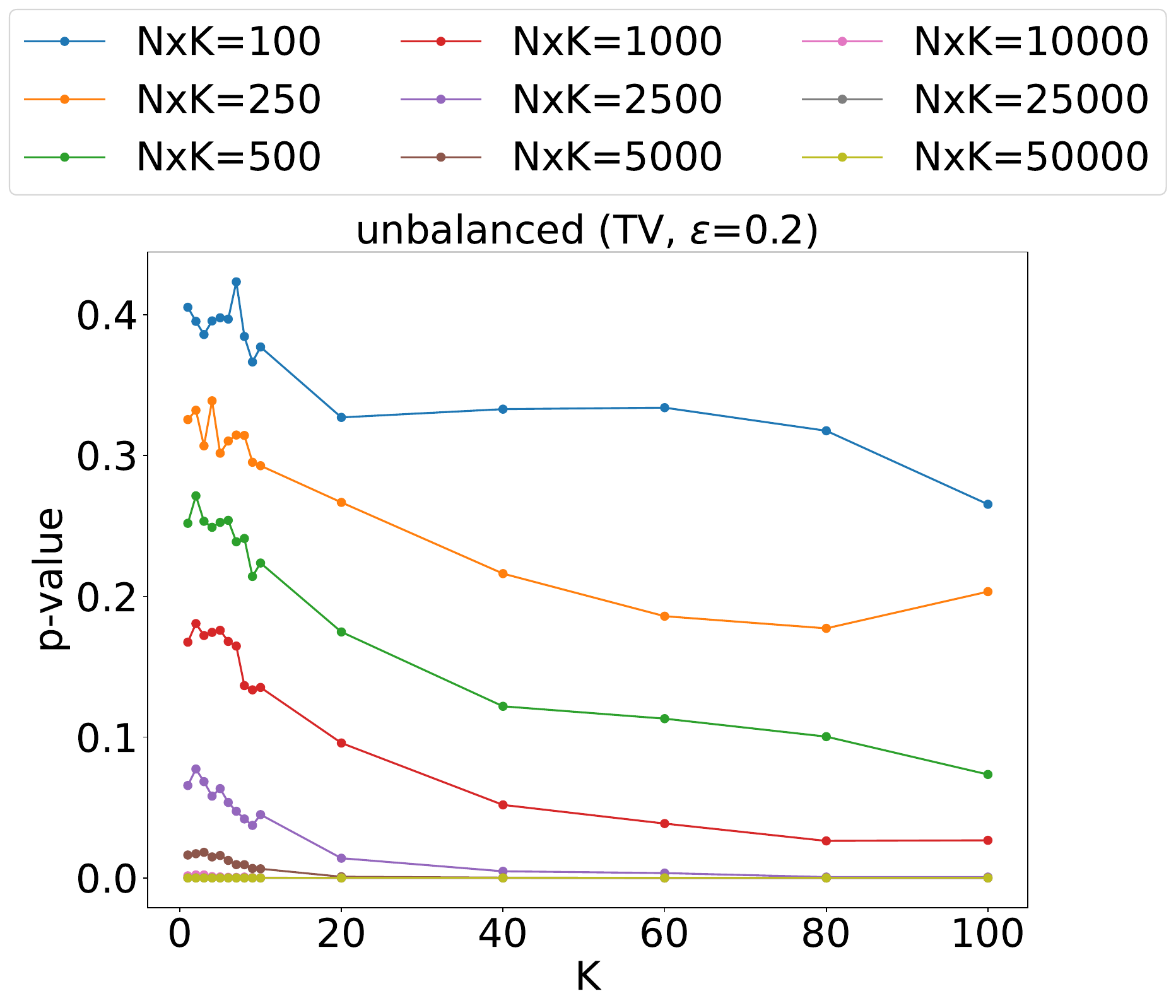}
    \caption{$\epsilon = 0.2$}
    \label{fig:gamma_MAE_cat4_e02}
  \end{subfigure} \hfill
  \begin{subfigure}[b]{0.24\linewidth}
    \centering
    \includegraphics[width=\linewidth]{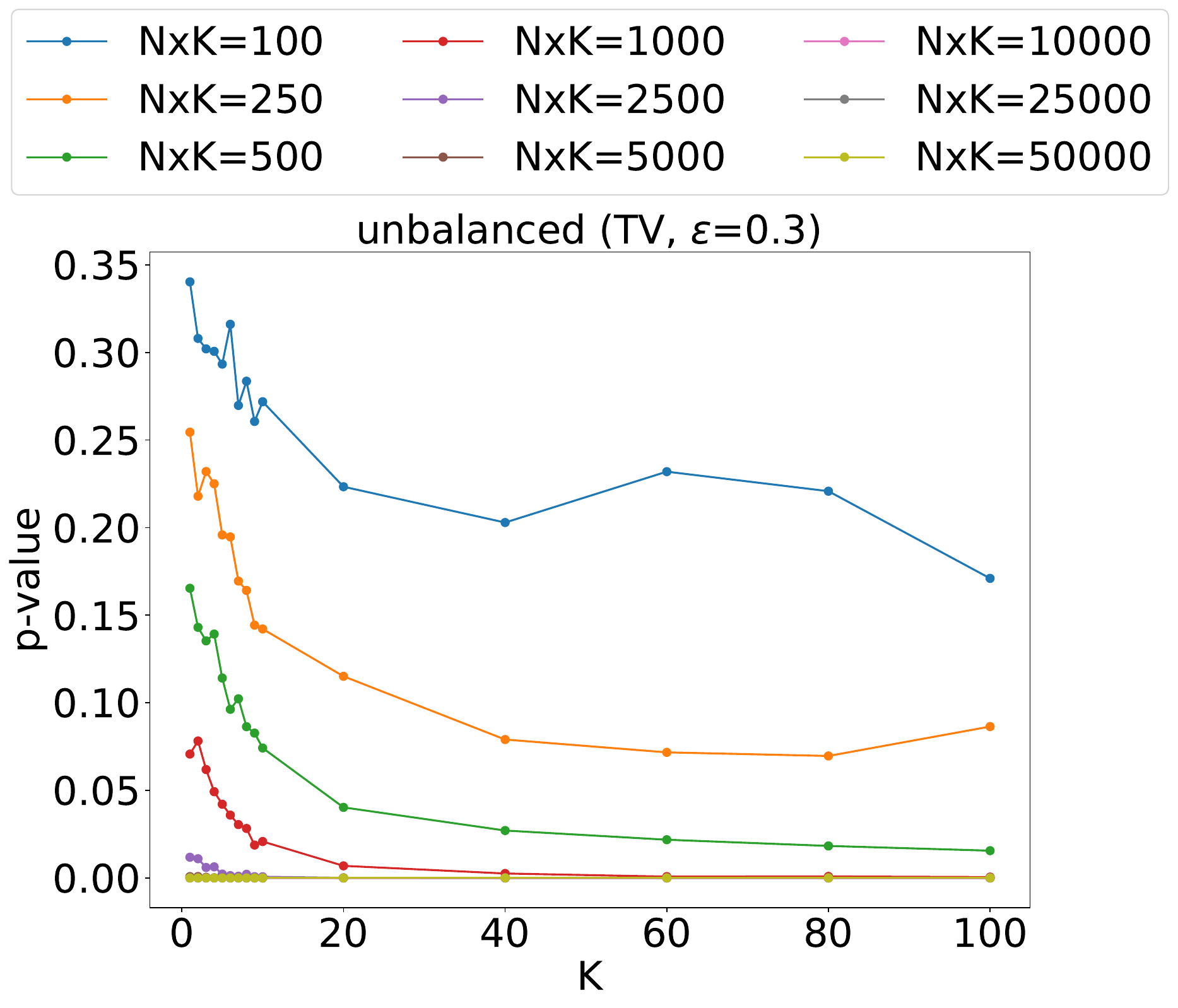}
    \caption{$\epsilon = 0.3$}
    \label{fig:gamma_MAE_cat4_e03}
  \end{subfigure} \hfill
  \begin{subfigure}[b]{0.24\linewidth}
    \centering
    \includegraphics[width=\linewidth]{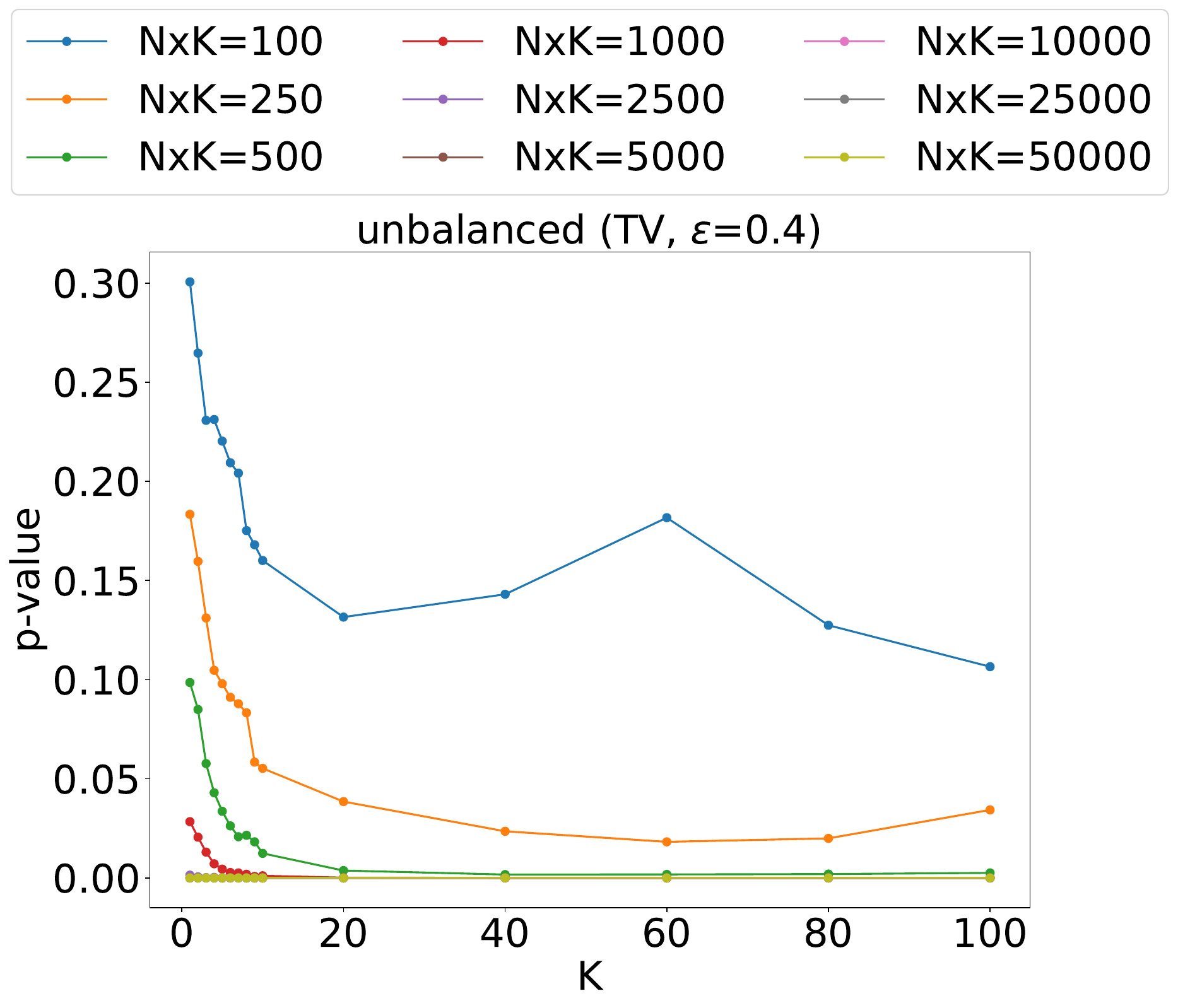}
    \caption{$\epsilon = 0.4$}
    \label{fig:gamma_MAE_cat4_e04}
  \end{subfigure}
  \caption{P-value plots for unbalanced alphas with TV as the metric ($M=4$)}
  \label{fig:gamma_MAE_cat4}
\end{figure*}

\begin{figure*}
  \centering
  \begin{subfigure}[b]{0.24\linewidth}
    \centering
    \includegraphics[width=\linewidth]{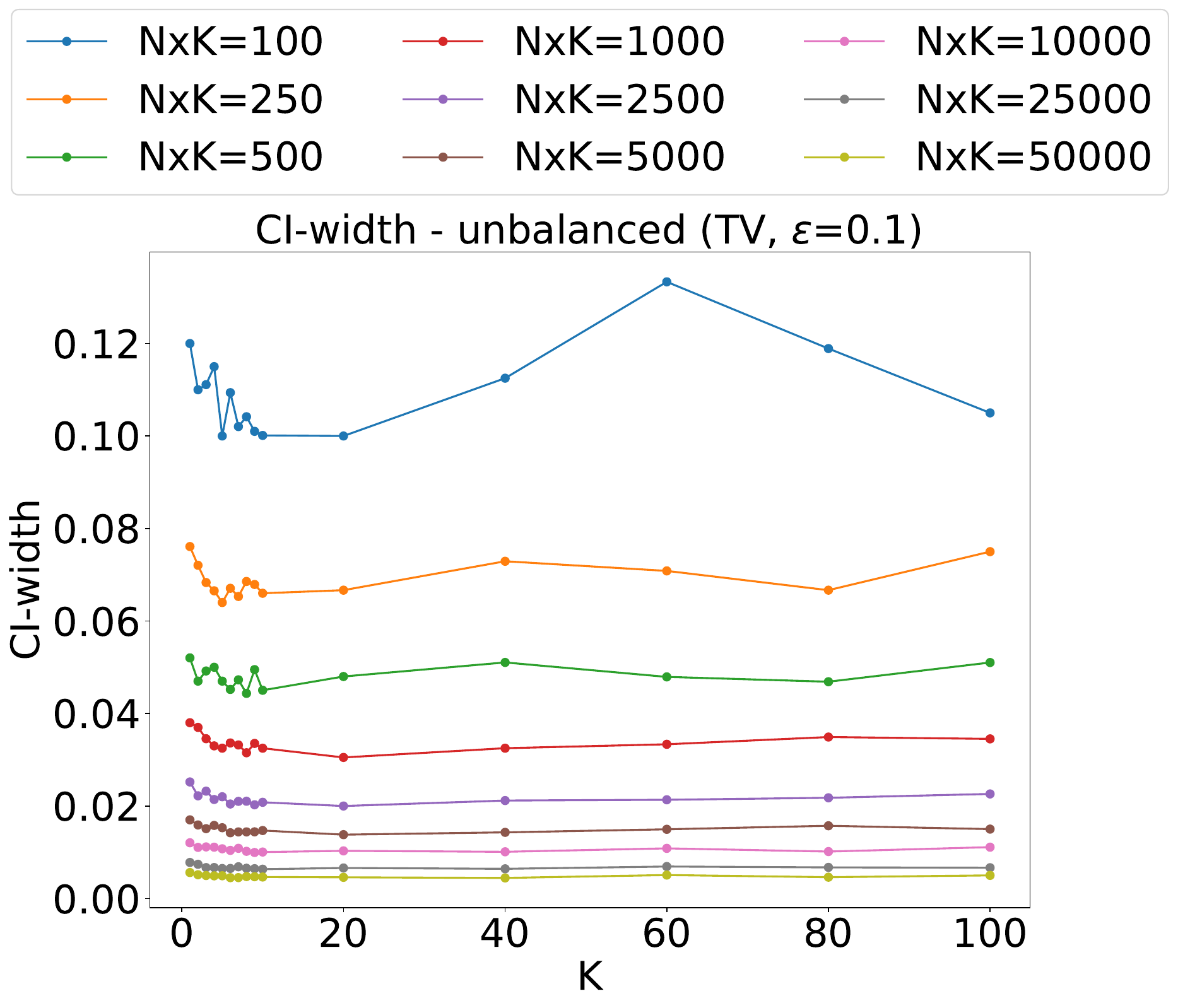}
    \caption{$\epsilon = 0.1$}
    \label{fig:gamma_ci_MAE_cat4_e01}
  \end{subfigure} \hfill
  \begin{subfigure}[b]{0.24\linewidth}
    \centering
    \includegraphics[width=\linewidth]{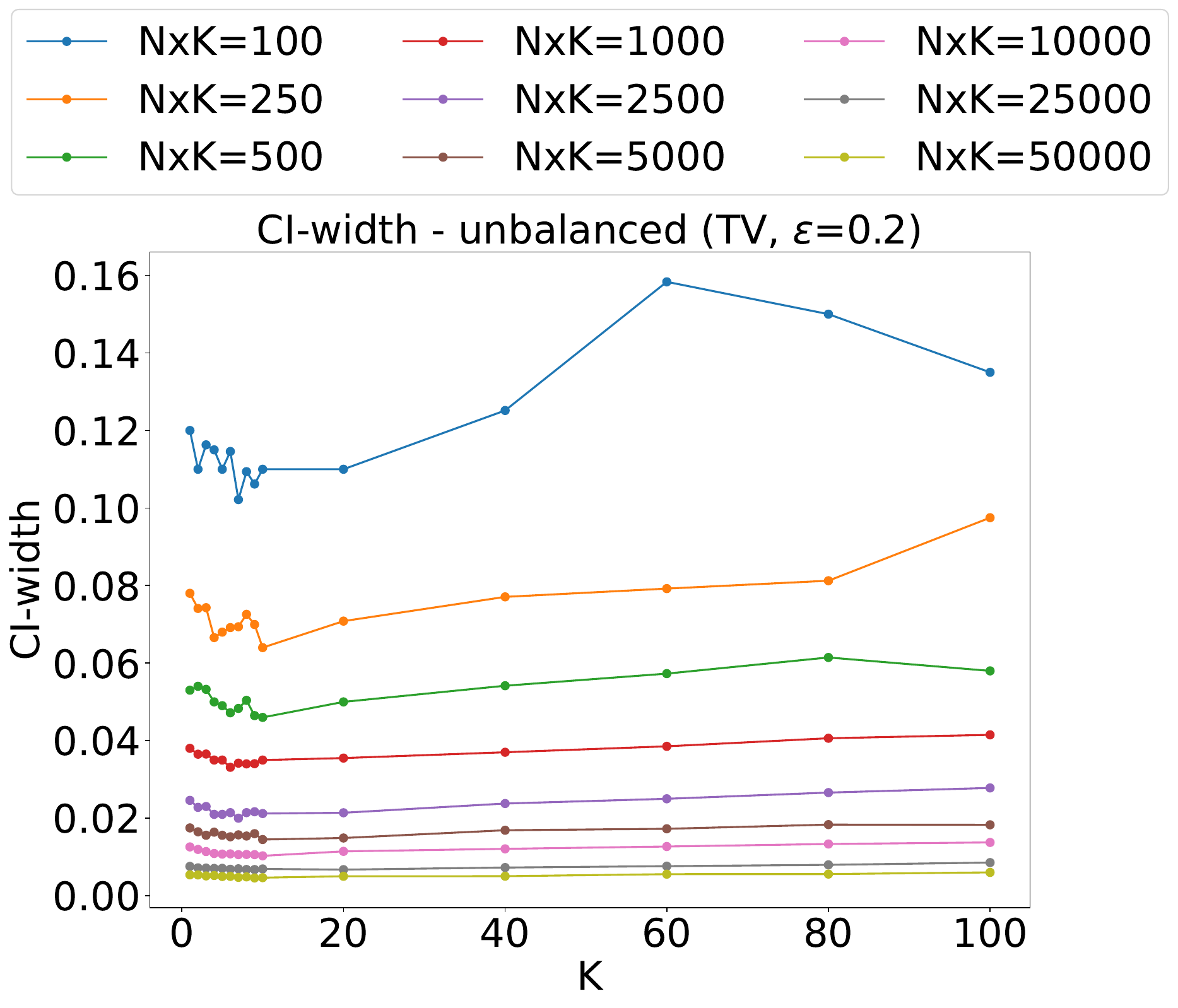}
    \caption{$\epsilon = 0.2$}
    \label{fig:gamma_ci_MAE_cat4_e02}
  \end{subfigure} \hfill
  \begin{subfigure}[b]{0.24\linewidth}
    \centering
    \includegraphics[width=\linewidth]{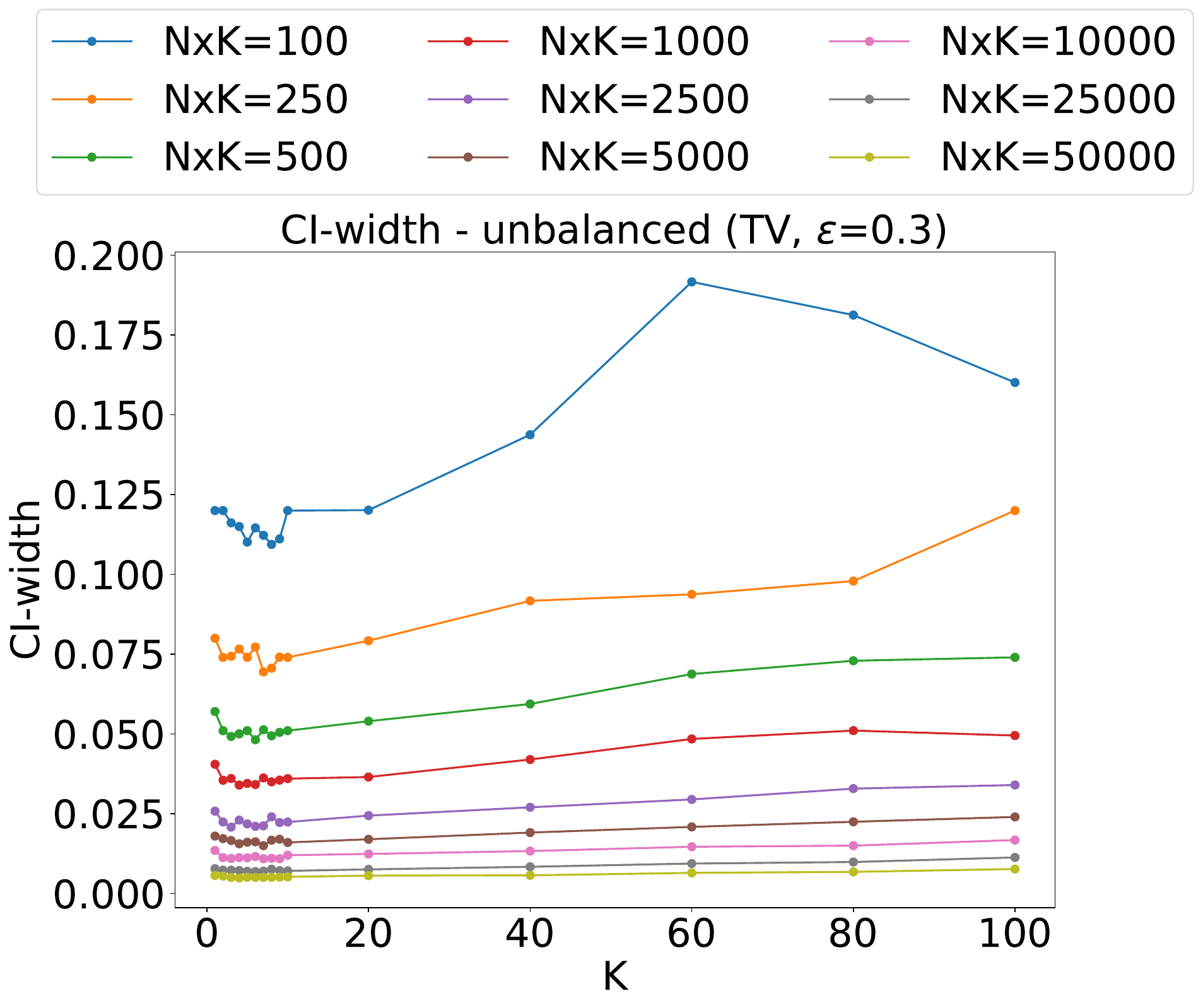}
    \caption{$\epsilon = 0.3$}
    \label{fig:gamma_ci_MAE_cat4_e03}
  \end{subfigure} \hfill
  \begin{subfigure}[b]{0.24\linewidth}
    \centering
    \includegraphics[width=\linewidth]{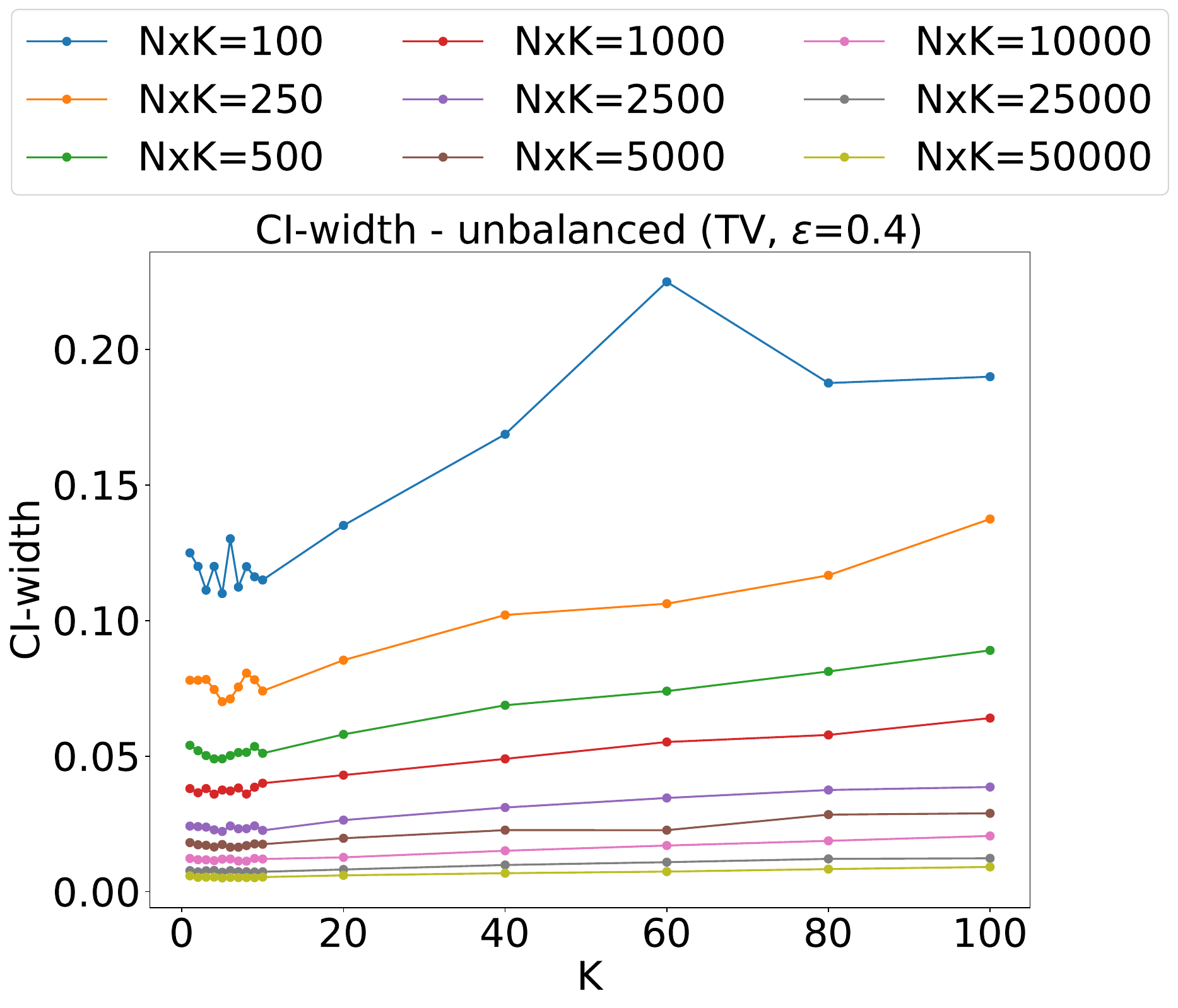}
    \caption{$\epsilon = 0.4$}
    \label{fig:gamma_ci_MAE_cat4_e04}
  \end{subfigure}
  \caption{CI-width plots for unbalanced alphas with TV as the metric ($M=4$)}
  \label{fig:gamma_ci_MAE_cat4}
\end{figure*}

\begin{figure*}
  \centering
  \begin{subfigure}[b]{0.24\linewidth}
    \centering
    \includegraphics[width=\linewidth]{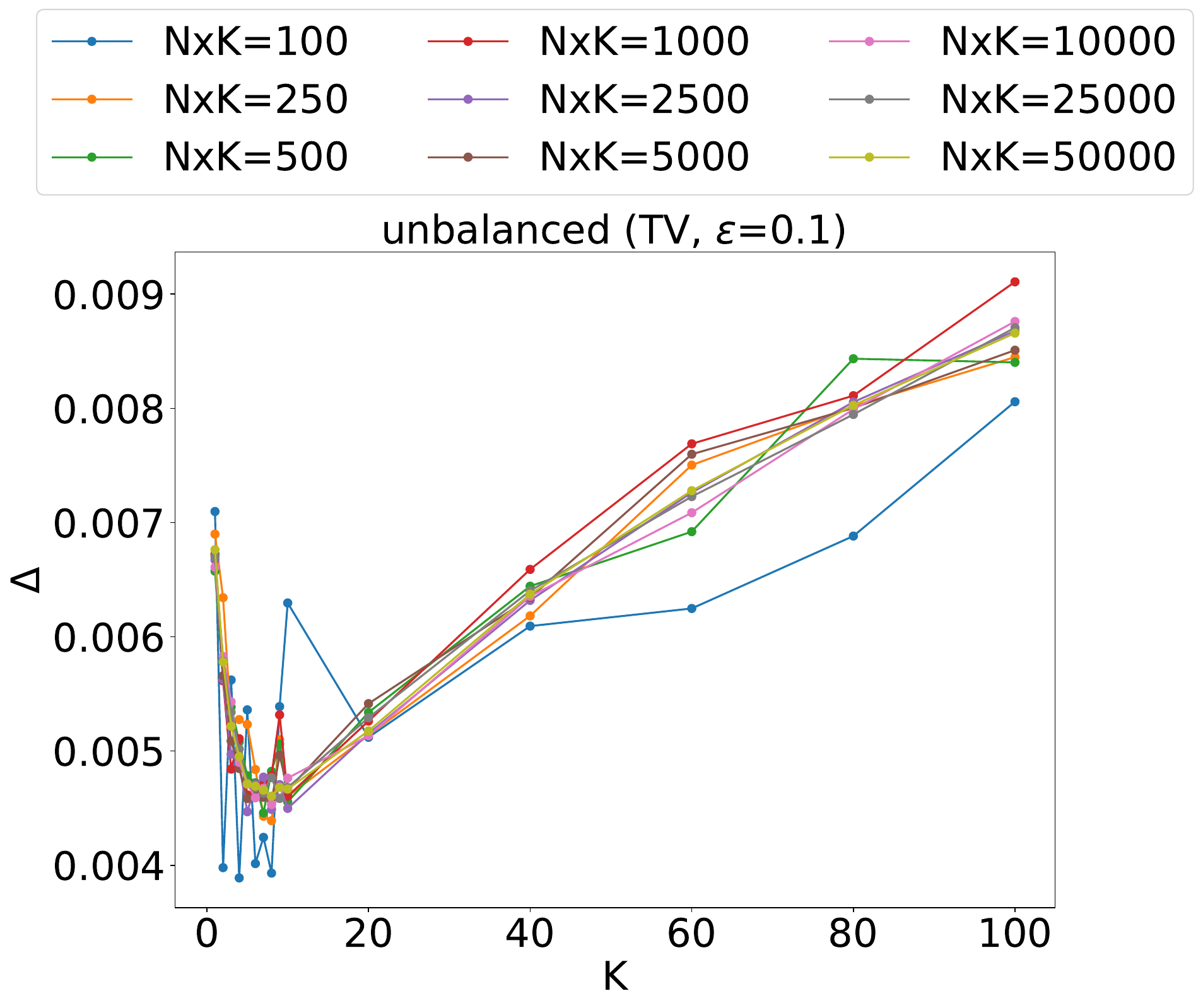}
    \caption{$\epsilon = 0.1$}
    \label{fig:gamma_delta_MAE_cat4_e01}
  \end{subfigure} \hfill
  \begin{subfigure}[b]{0.24\linewidth}
    \centering
    \includegraphics[width=\linewidth]{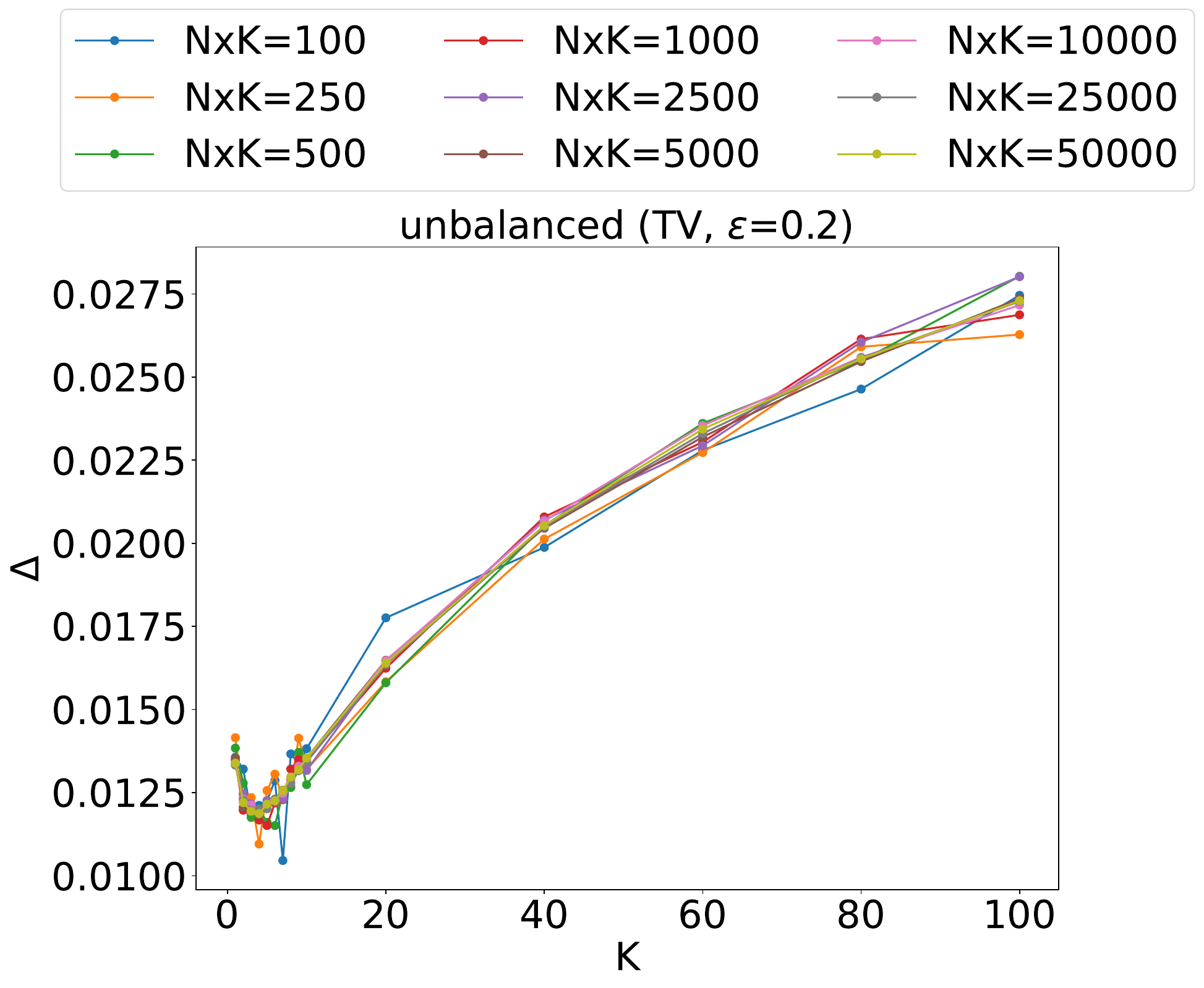}
    \caption{$\epsilon = 0.2$}
    \label{fig:gamma_delta_MAE_cat4_e02}
  \end{subfigure} \hfill
  \begin{subfigure}[b]{0.24\linewidth}
    \centering
    \includegraphics[width=\linewidth]{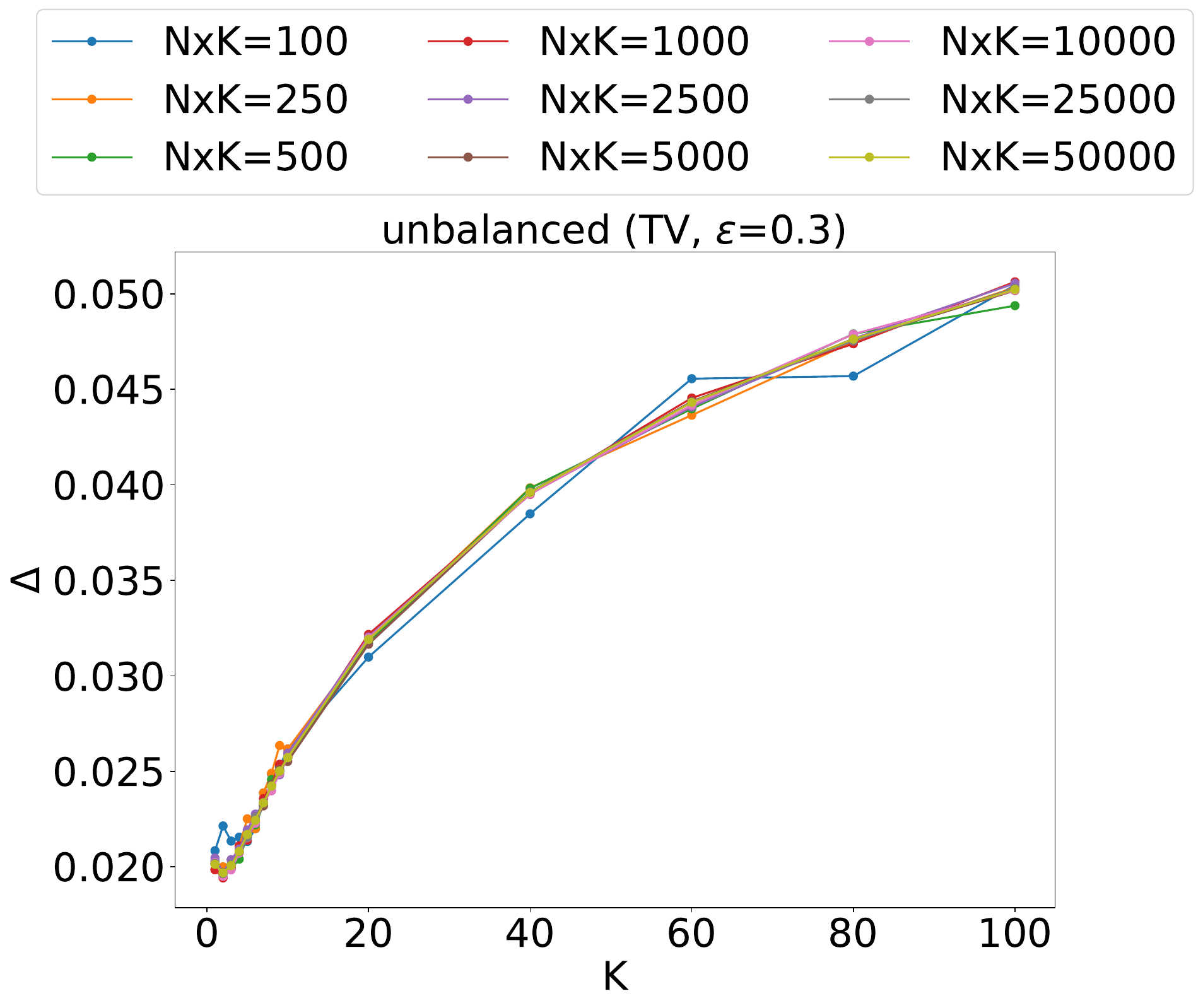}
    \caption{$\epsilon = 0.3$}
    \label{fig:gamma_delta_MAE_cat4_e03}
  \end{subfigure} \hfill
  \begin{subfigure}[b]{0.24\linewidth}
    \centering
    \includegraphics[width=\linewidth]{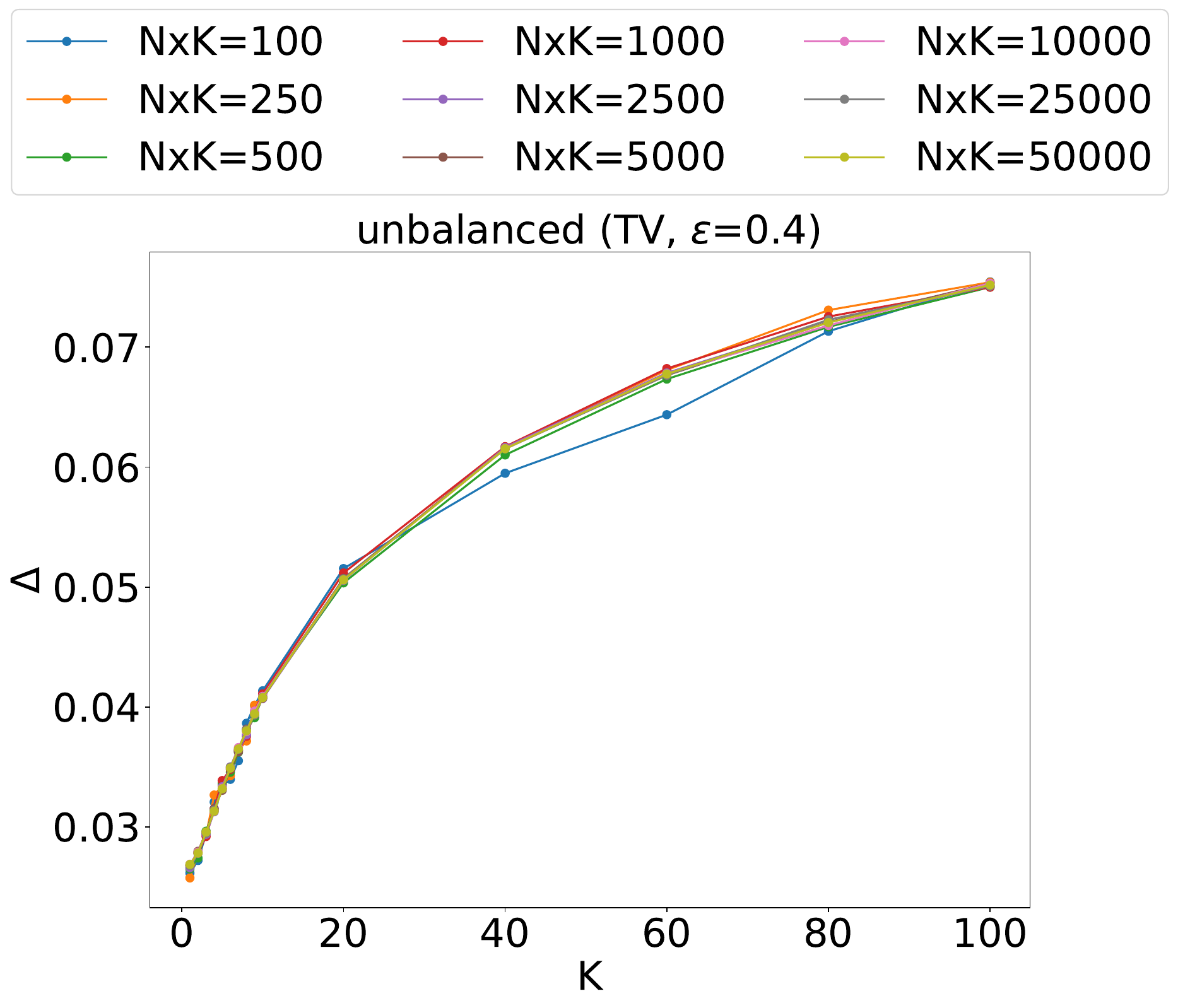}
    \caption{$\epsilon = 0.4$}
    \label{fig:gamma_delta_MAE_cat4_e04}
  \end{subfigure}
  \caption{Effect sizes ($\Delta$) for unbalanced alphas with TV as the metric ($M=4$)}
  \label{fig:gamma_delta_MAE_cat4}
\end{figure*}

\begin{figure*}
  \centering
  \begin{subfigure}[b]{0.24\linewidth}
    \centering
    \includegraphics[width=\linewidth]{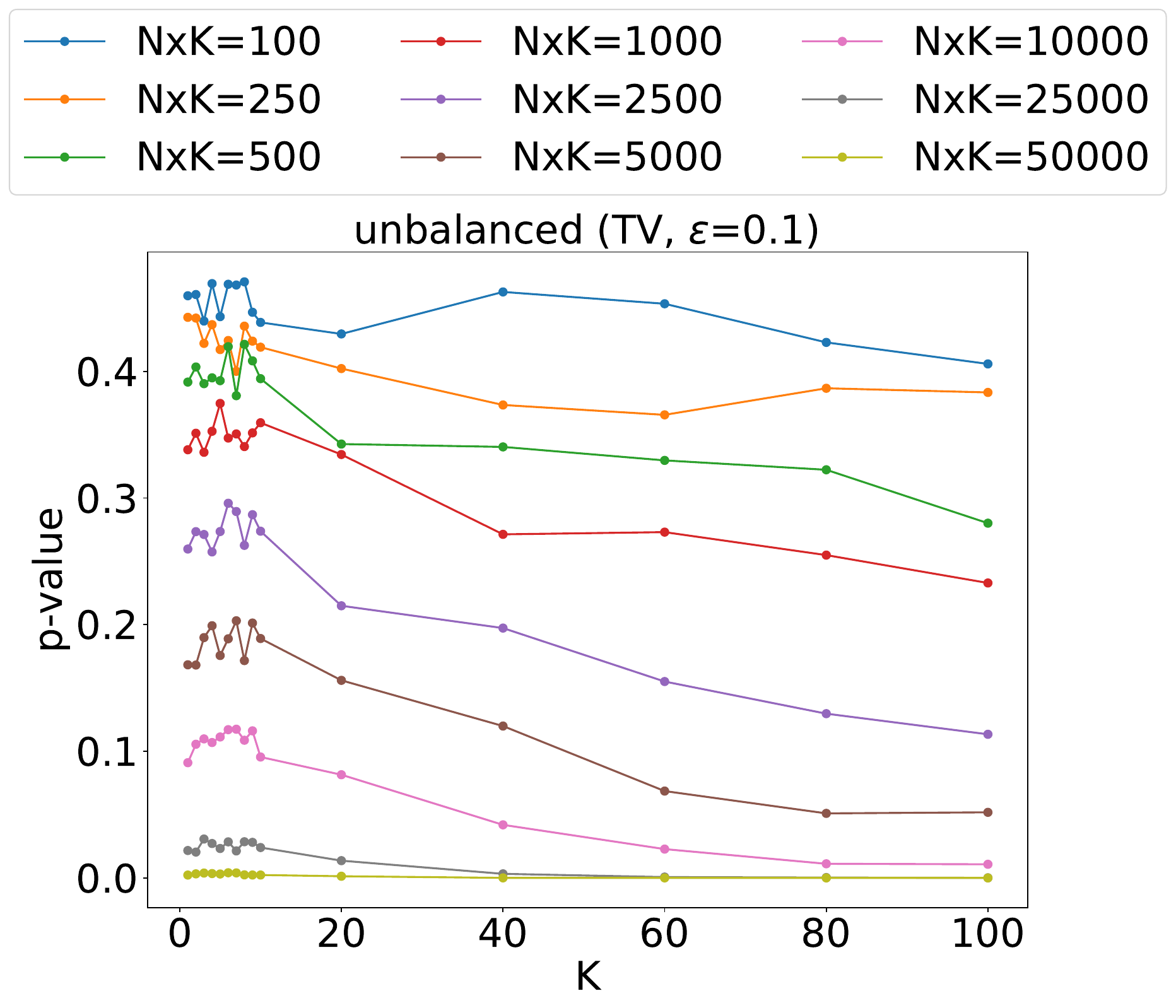}
    \caption{$\epsilon = 0.1$}
    \label{fig:gamma_MAE_cat5_e01}
  \end{subfigure} \hfill
  \begin{subfigure}[b]{0.24\linewidth}
    \centering
    \includegraphics[width=\linewidth]{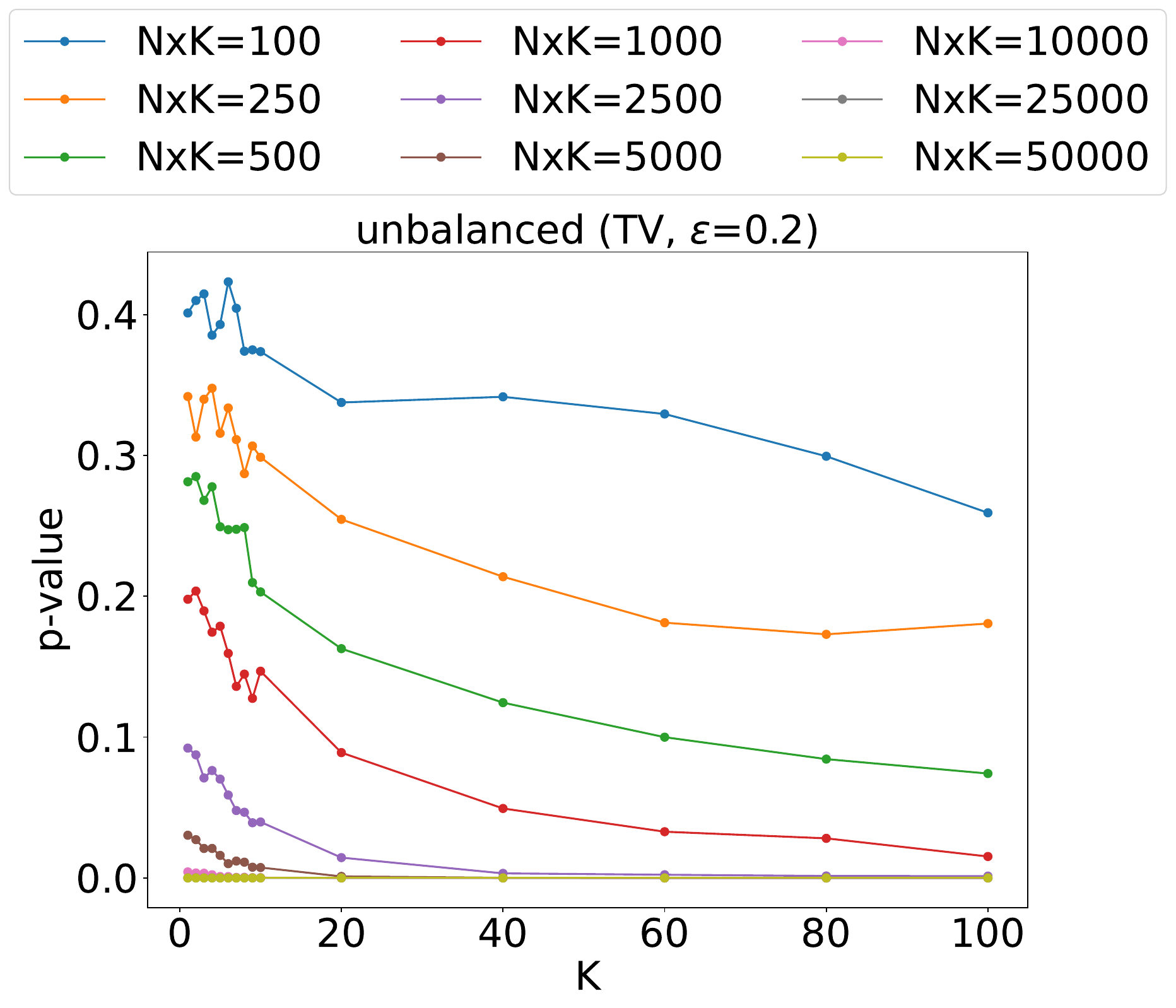}
    \caption{$\epsilon = 0.2$}
    \label{fig:gamma_MAE_cat5_e02}
  \end{subfigure} \hfill
  \begin{subfigure}[b]{0.24\linewidth}
    \centering
    \includegraphics[width=\linewidth]{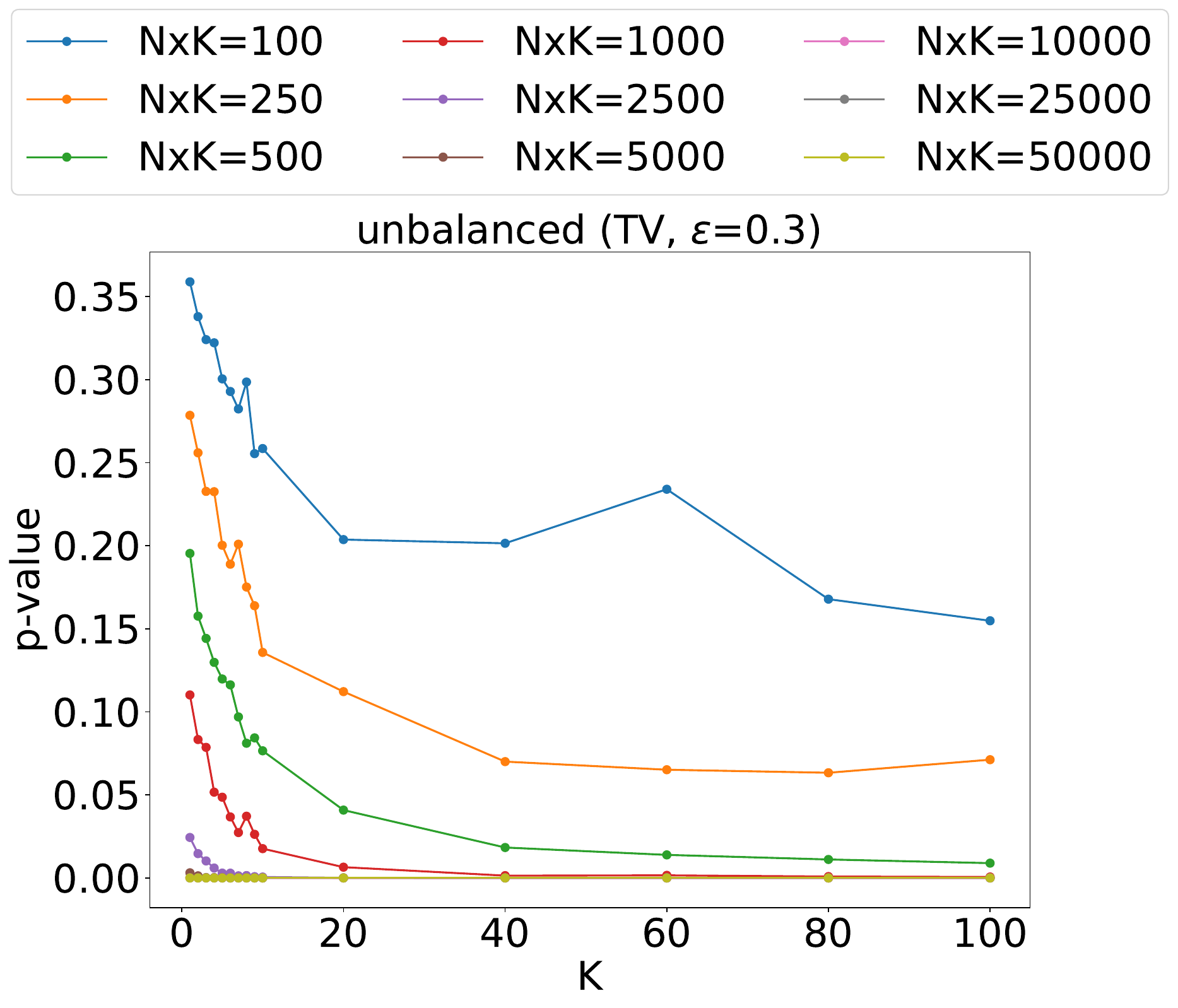}
    \caption{$\epsilon = 0.3$}
    \label{fig:gamma_MAE_cat5_e03}
  \end{subfigure} \hfill
  \begin{subfigure}[b]{0.24\linewidth}
    \centering
    \includegraphics[width=\linewidth]{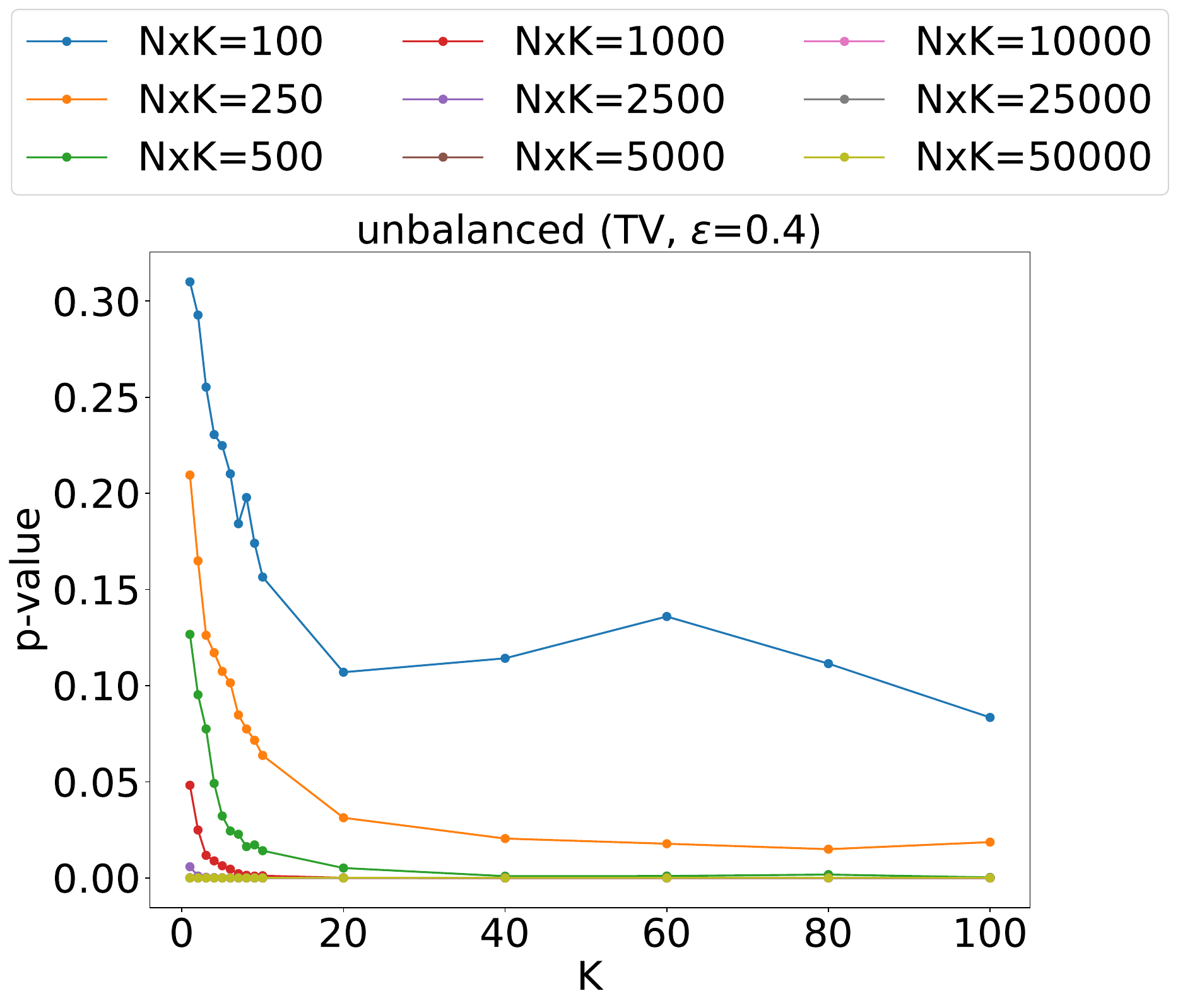}
    \caption{$\epsilon = 0.4$}
    \label{fig:gamma_MAE_cat5_e04}
  \end{subfigure}
  \caption{P-value plots for unbalanced alphas with TV as the metric ($M=5$)}
  \label{fig:gamma_MAE_cat5}
\end{figure*}

\begin{figure*}
  \centering
  \begin{subfigure}[b]{0.24\linewidth}
    \centering
    \includegraphics[width=\linewidth]{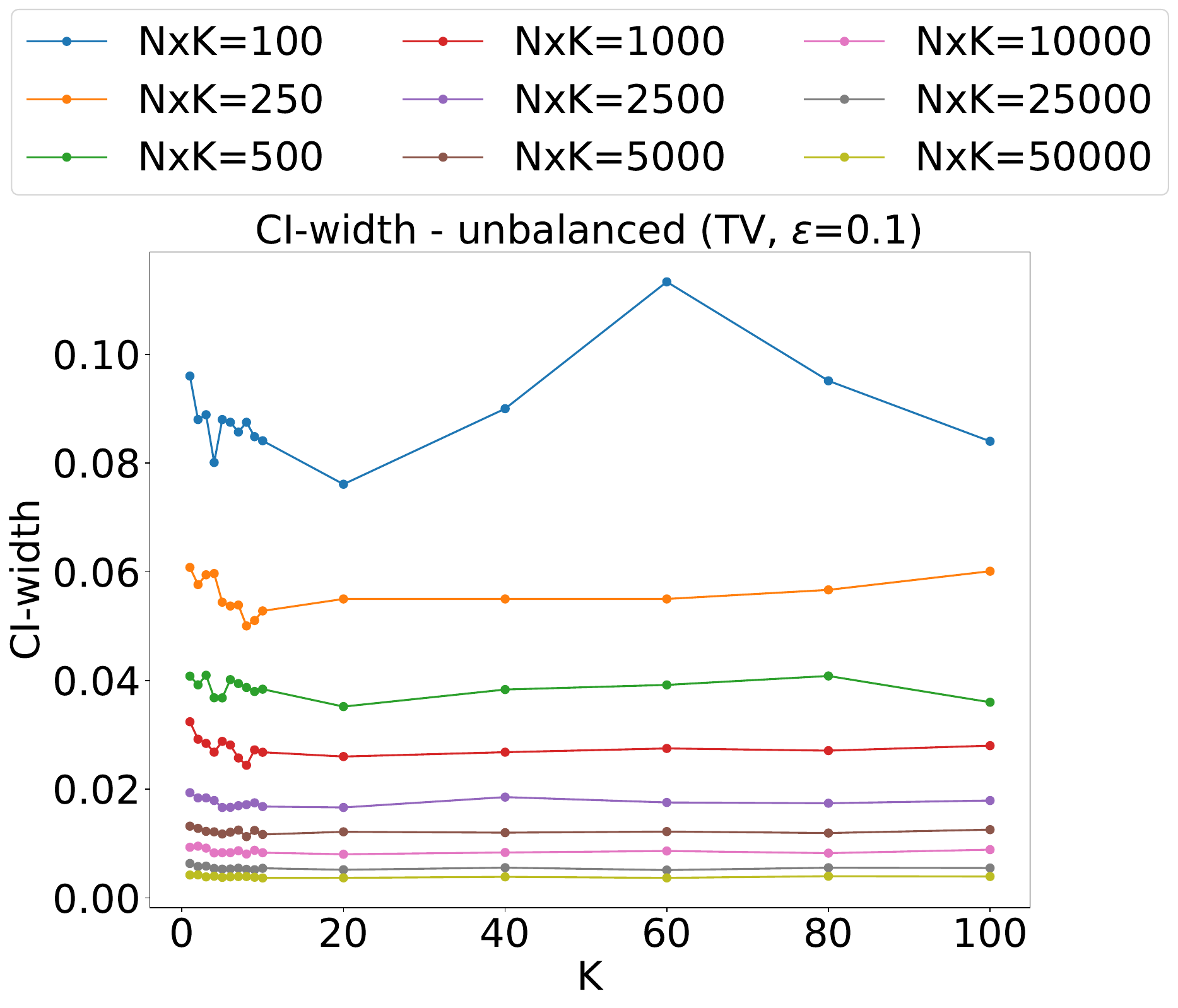}
    \caption{$\epsilon = 0.1$}
    \label{fig:gamma_ci_MAE_cat5_e01}
  \end{subfigure} \hfill
  \begin{subfigure}[b]{0.24\linewidth}
    \centering
    \includegraphics[width=\linewidth]{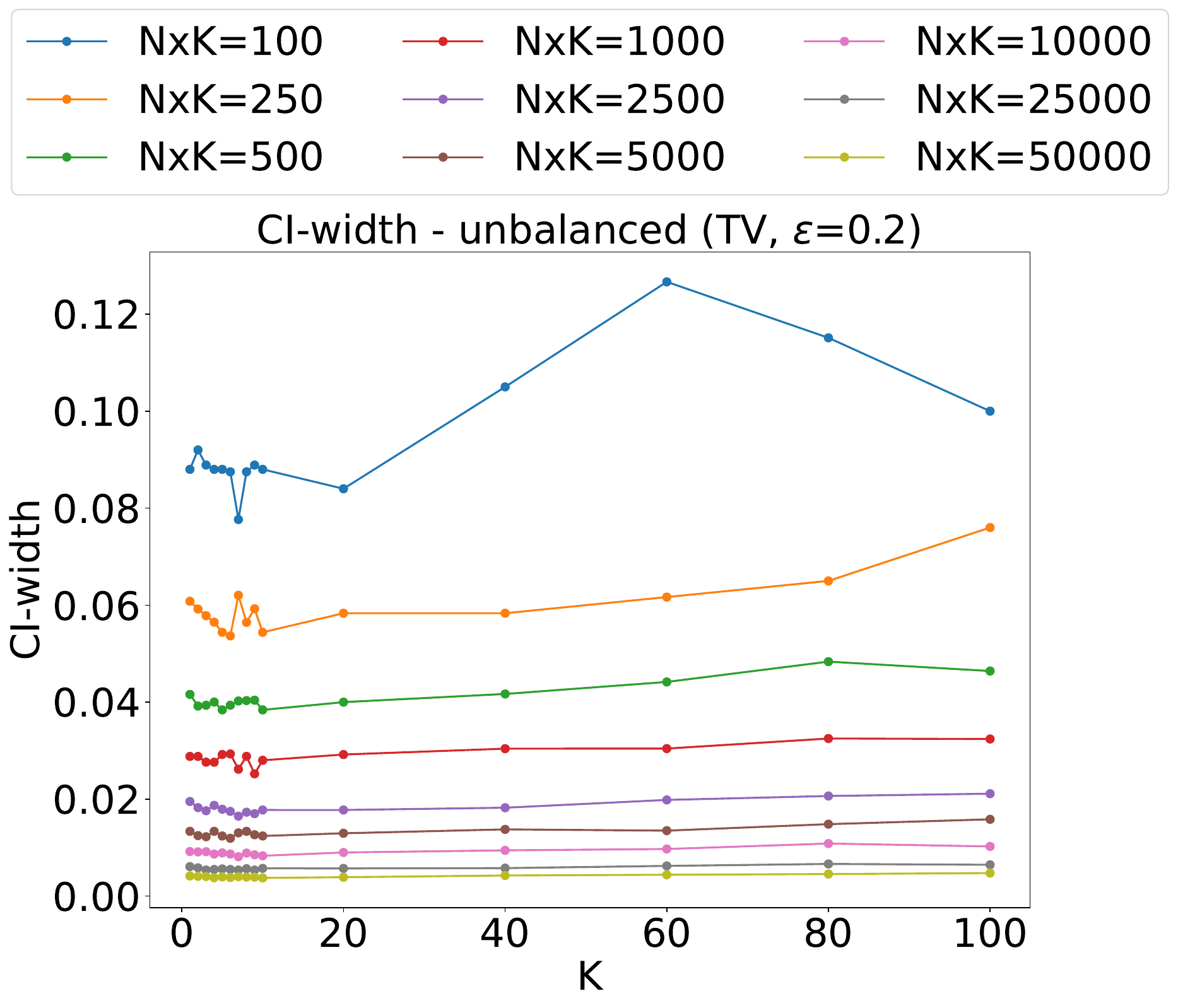}
    \caption{$\epsilon = 0.2$}
    \label{fig:gamma_ci_MAE_cat5_e02}
  \end{subfigure} \hfill
  \begin{subfigure}[b]{0.24\linewidth}
    \centering
    \includegraphics[width=\linewidth]{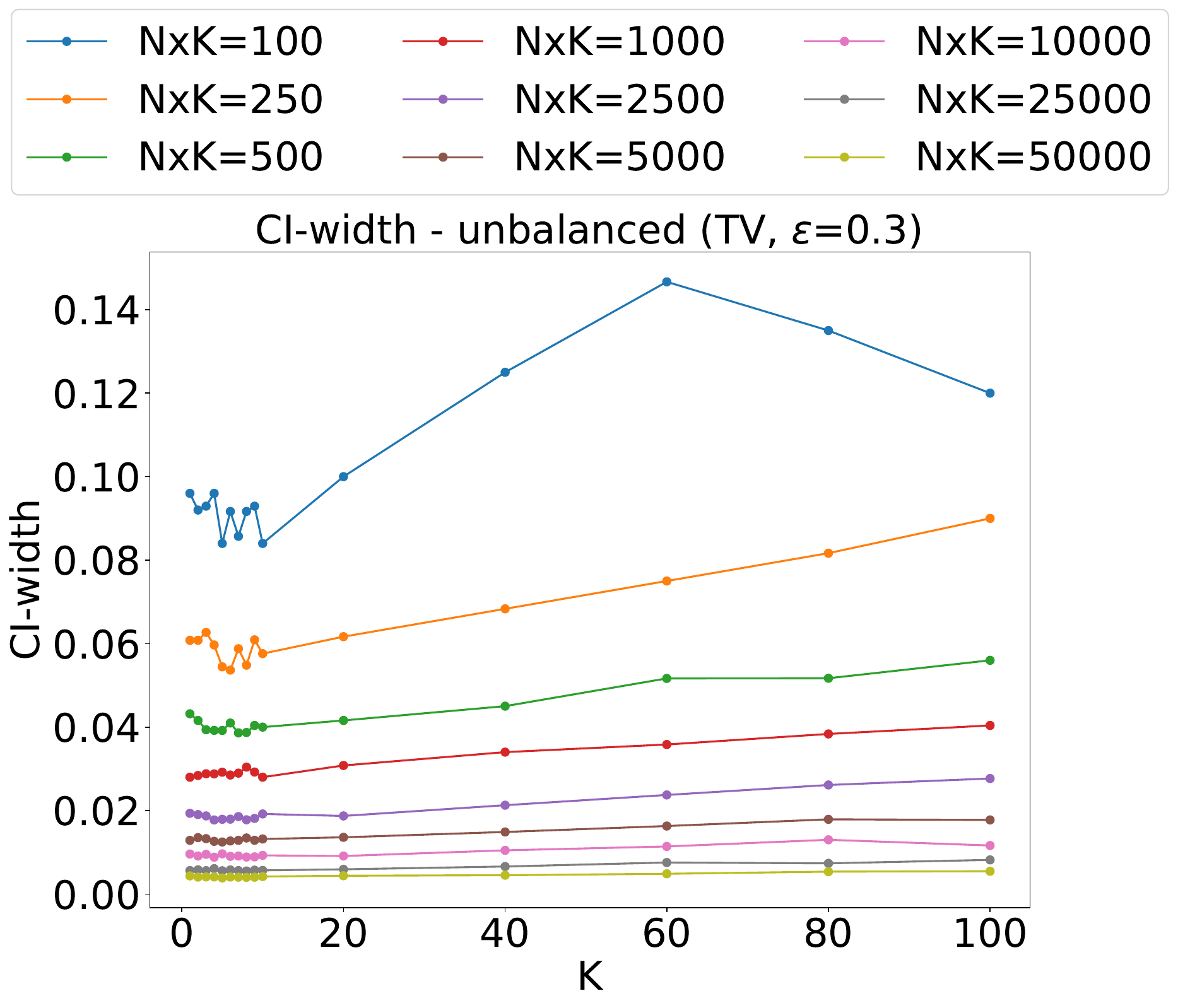}
    \caption{$\epsilon = 0.3$}
    \label{fig:gamma_ci_MAE_cat5_e03}
  \end{subfigure} \hfill
  \begin{subfigure}[b]{0.24\linewidth}
    \centering
    \includegraphics[width=\linewidth]{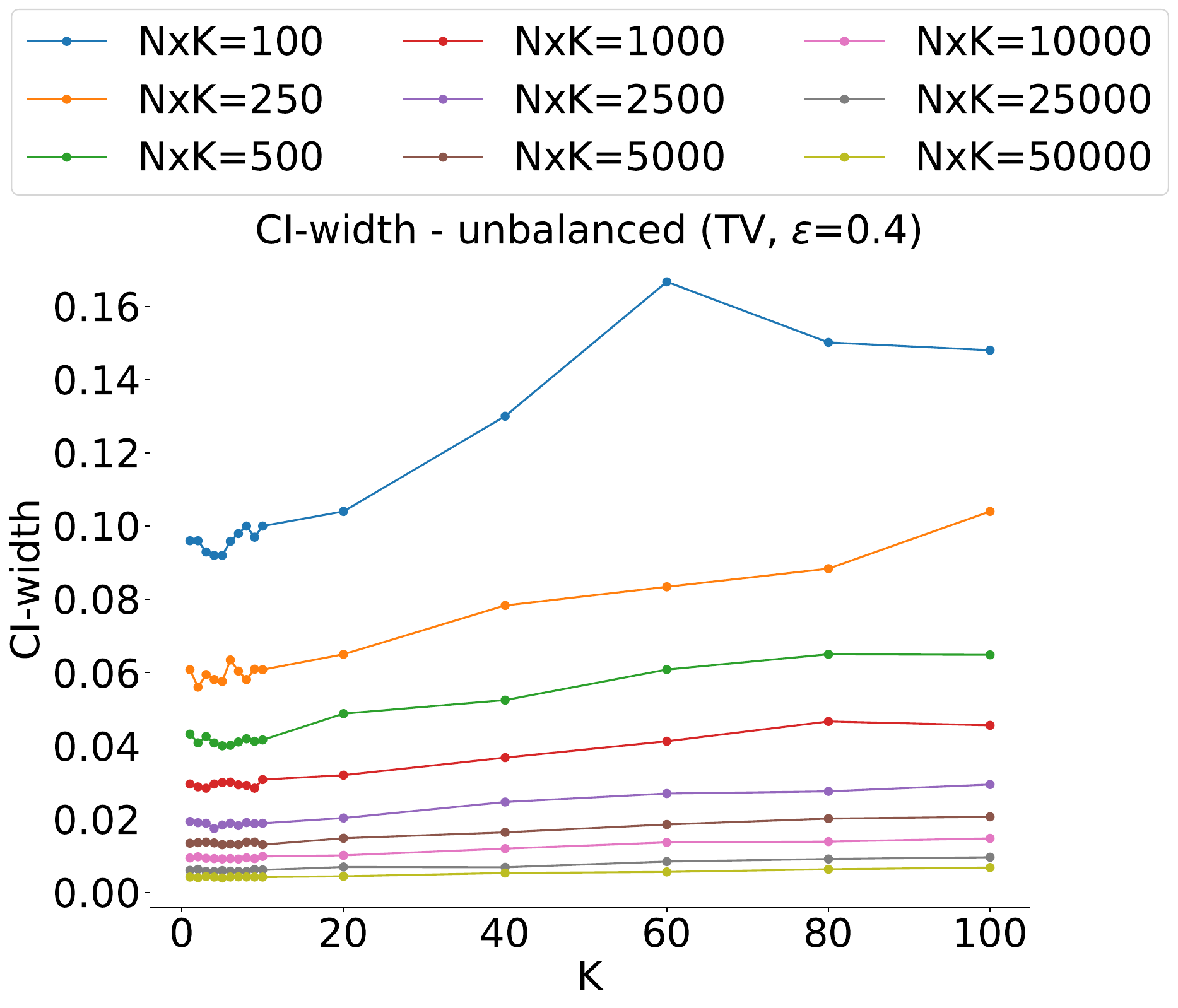}
    \caption{$\epsilon = 0.4$}
    \label{fig:gamma_ci_MAE_cat5_e04}
  \end{subfigure}
  \caption{CI-width plots for unbalanced alphas with TV as the metric ($M=5$)}
  \label{fig:gamma_ci_MAE_cat5}
\end{figure*}

\begin{figure*}
  \centering
  \begin{subfigure}[b]{0.24\linewidth}
    \centering
    \includegraphics[width=\linewidth]{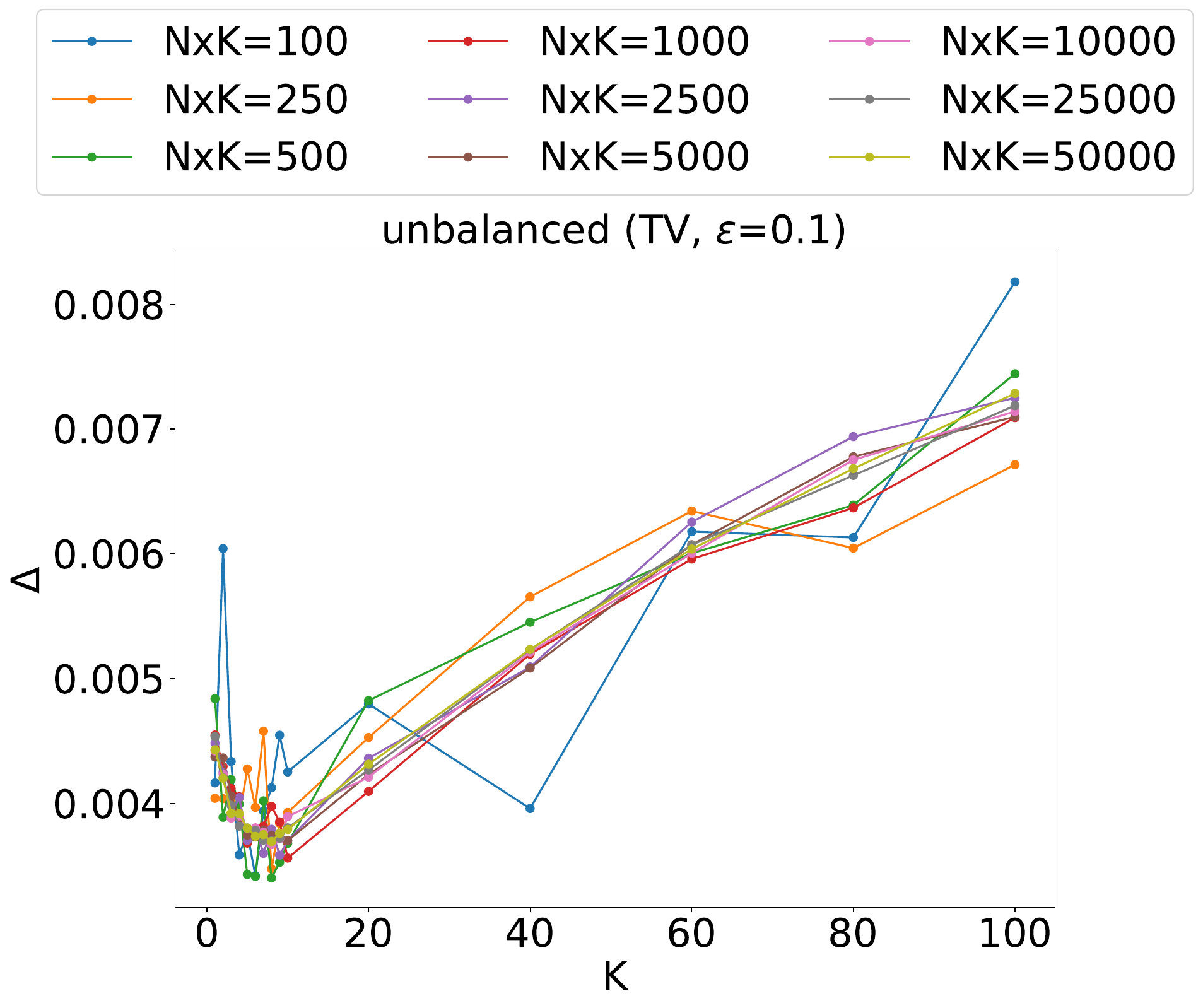}
    \caption{$\epsilon = 0.1$}
    \label{fig:gamma_delta_MAE_cat5_e01}
  \end{subfigure} \hfill
  \begin{subfigure}[b]{0.24\linewidth}
    \centering
    \includegraphics[width=\linewidth]{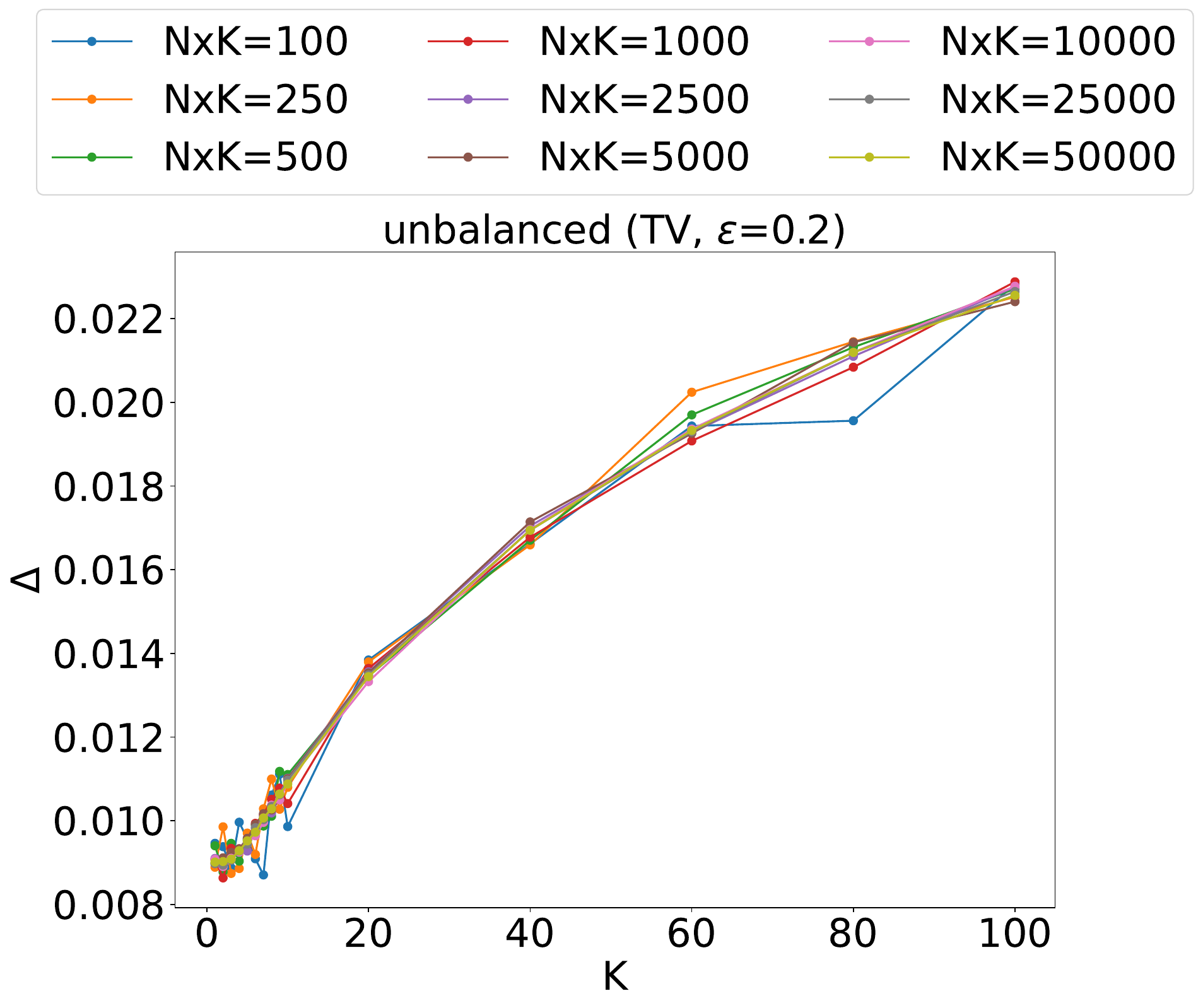}
    \caption{$\epsilon = 0.2$}
    \label{fig:gamma_delta_MAE_cat5_e02}
  \end{subfigure} \hfill
  \begin{subfigure}[b]{0.24\linewidth}
    \centering
    \includegraphics[width=\linewidth]{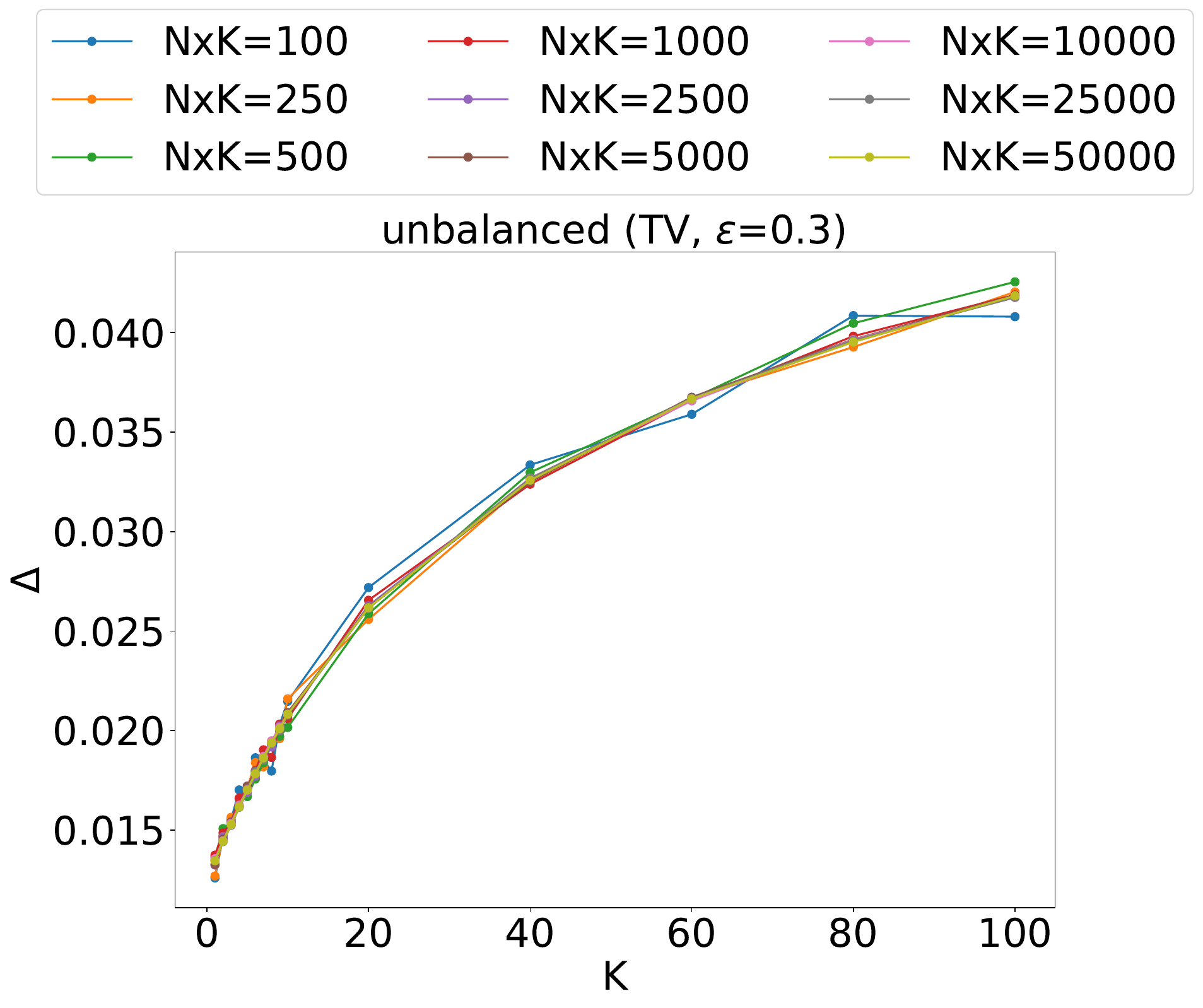}
    \caption{$\epsilon = 0.3$}
    \label{fig:gamma_delta_MAE_cat5_e03}
  \end{subfigure} \hfill
  \begin{subfigure}[b]{0.24\linewidth}
    \centering
    \includegraphics[width=\linewidth]{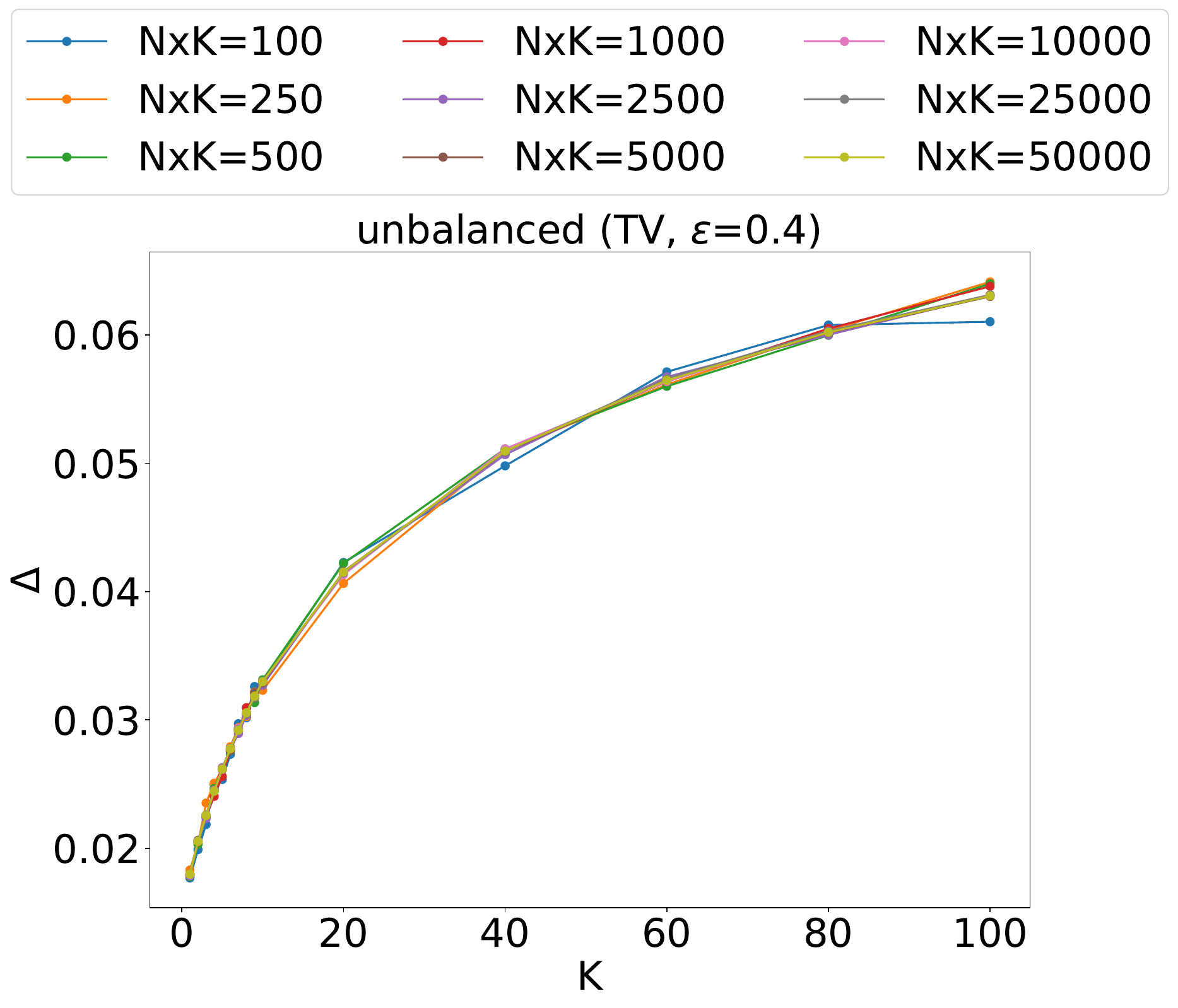}
    \caption{$\epsilon = 0.4$}
    \label{fig:gamma_delta_MAE_cat5_e04}
  \end{subfigure}
  \caption{Effect sizes ($\Delta$) for unbalanced alphas with TV as the metric ($M=5$)}
  \label{fig:gamma_delta_MAE_cat5}
\end{figure*}

\begin{figure*}
  \centering
  \begin{subfigure}[b]{0.24\linewidth}
    \centering
    \includegraphics[width=\linewidth]{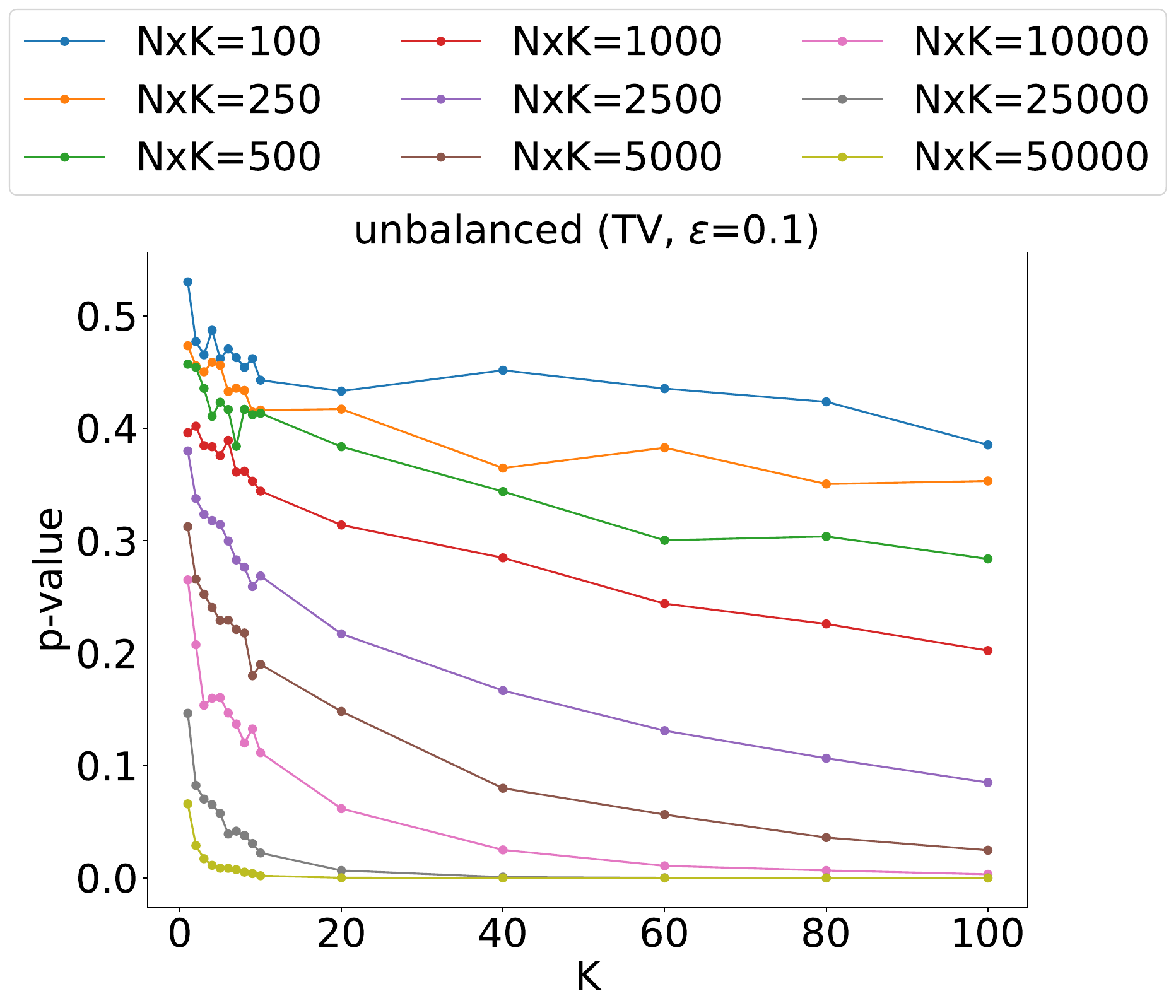}
    \caption{$\epsilon = 0.1$}
    \label{fig:gamma_MAE_cat12_e01}
  \end{subfigure} \hfill
  \begin{subfigure}[b]{0.24\linewidth}
    \centering
    \includegraphics[width=\linewidth]{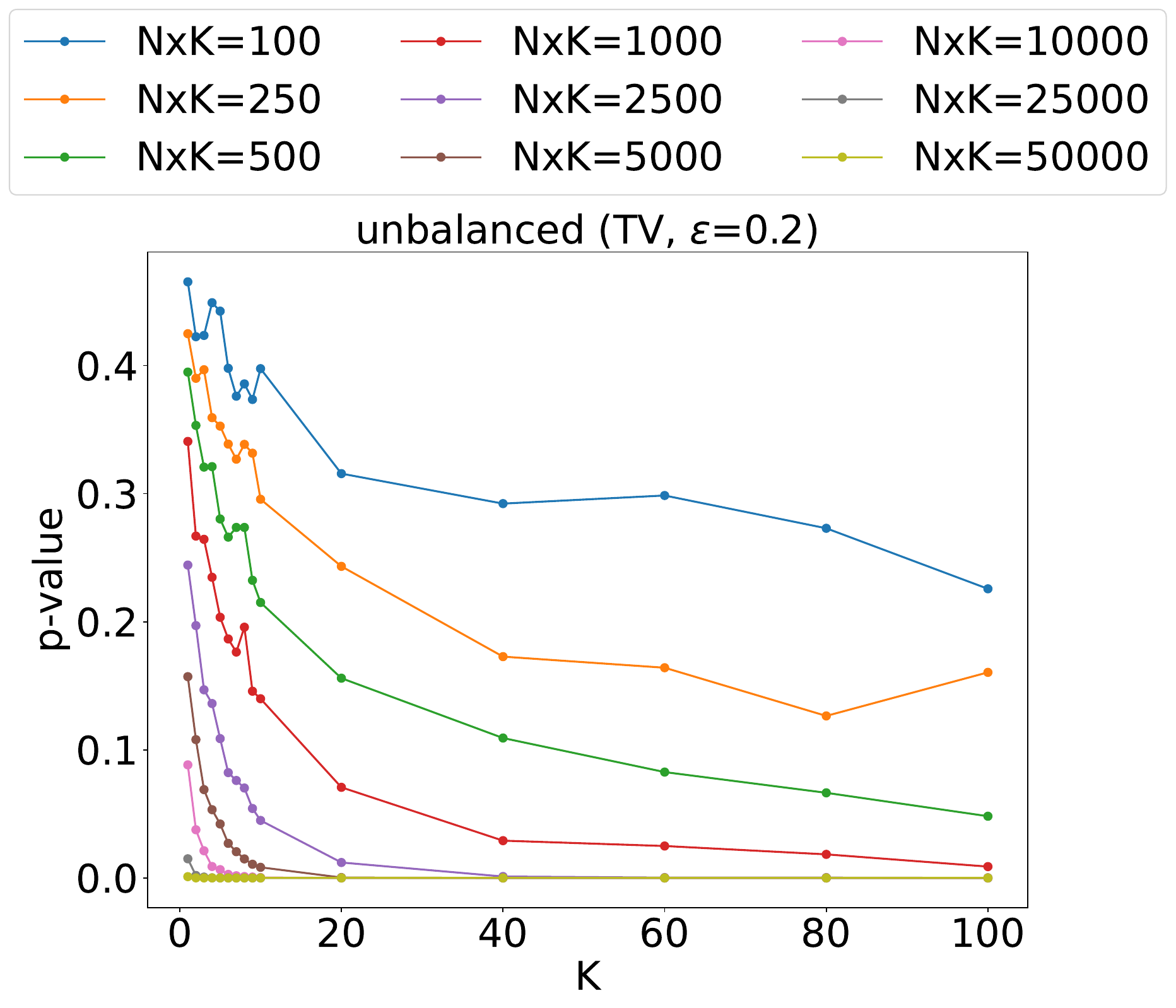}
    \caption{$\epsilon = 0.2$}
    \label{fig:gamma_MAE_cat12_e02}
  \end{subfigure} \hfill
  \begin{subfigure}[b]{0.24\linewidth}
    \centering
    \includegraphics[width=\linewidth]{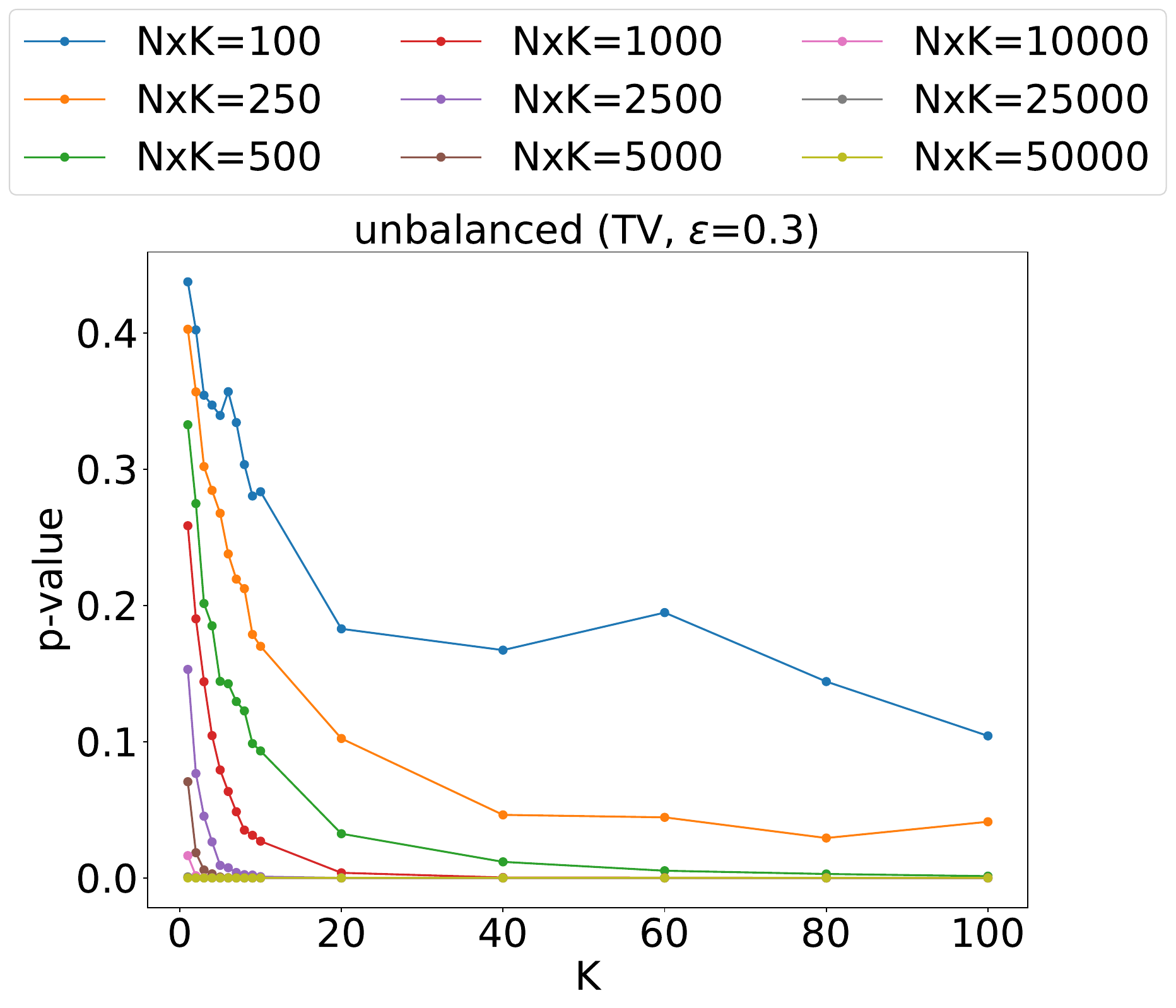}
    \caption{$\epsilon = 0.3$}
    \label{fig:gamma_MAE_cat12_e03}
  \end{subfigure} \hfill
  \begin{subfigure}[b]{0.24\linewidth}
    \centering
    \includegraphics[width=\linewidth]{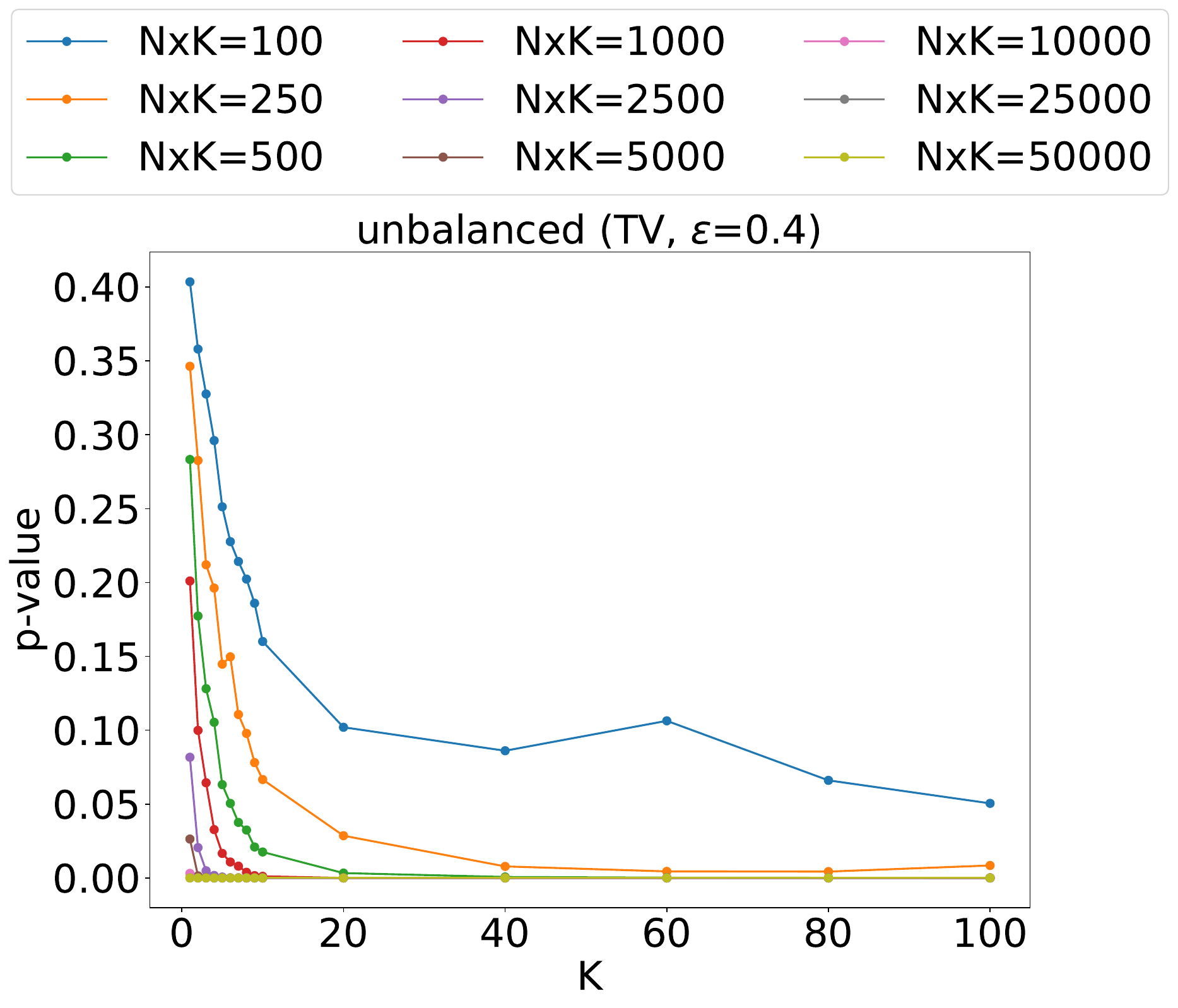}
    \caption{$\epsilon = 0.4$}
    \label{fig:gamma_MAE_cat12_e04}
  \end{subfigure}
  \caption{P-value plots for unbalanced alphas with TV as the metric ($M=12$)}
  \label{fig:gamma_MAE_cat12}
\end{figure*}

\begin{figure*}
  \centering
  \begin{subfigure}[b]{0.24\linewidth}
    \centering
    \includegraphics[width=\linewidth]{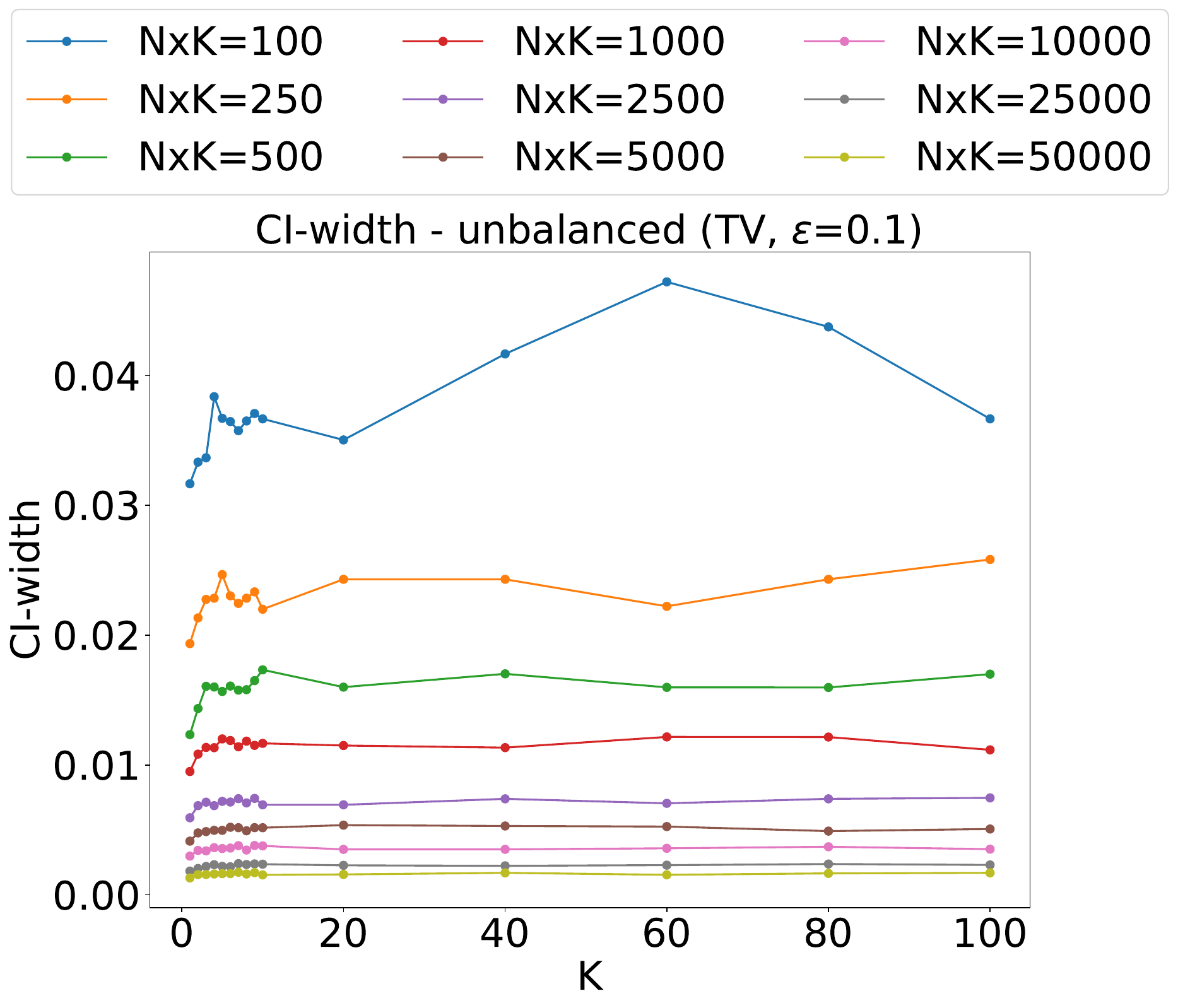}
    \caption{$\epsilon = 0.1$}
    \label{fig:gamma_ci_MAE_cat12_e01}
  \end{subfigure} \hfill
  \begin{subfigure}[b]{0.24\linewidth}
    \centering
    \includegraphics[width=\linewidth]{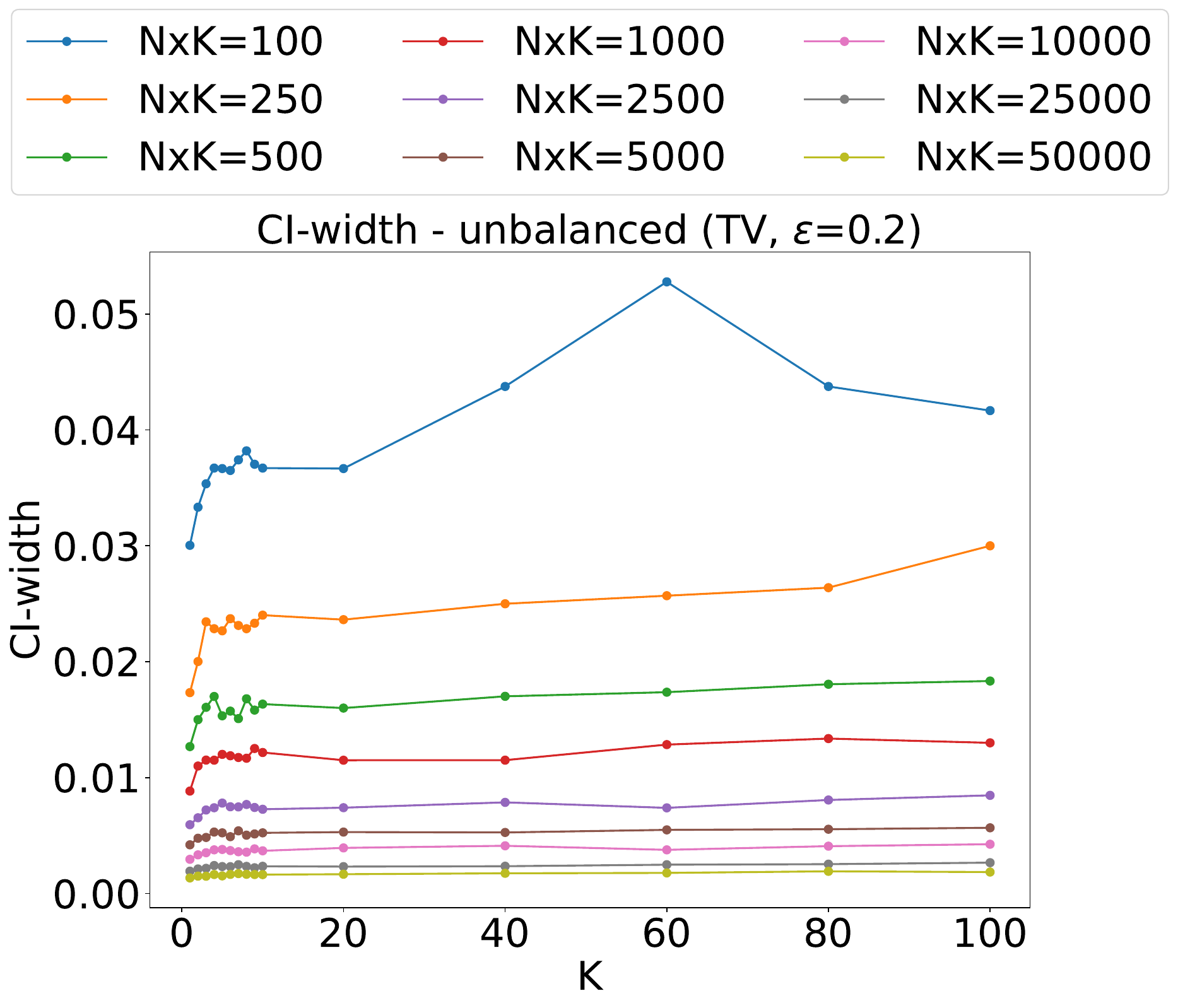}
    \caption{$\epsilon = 0.2$}
    \label{fig:gamma_ci_MAE_cat12_e02}
  \end{subfigure} \hfill
  \begin{subfigure}[b]{0.24\linewidth}
    \centering
    \includegraphics[width=\linewidth]{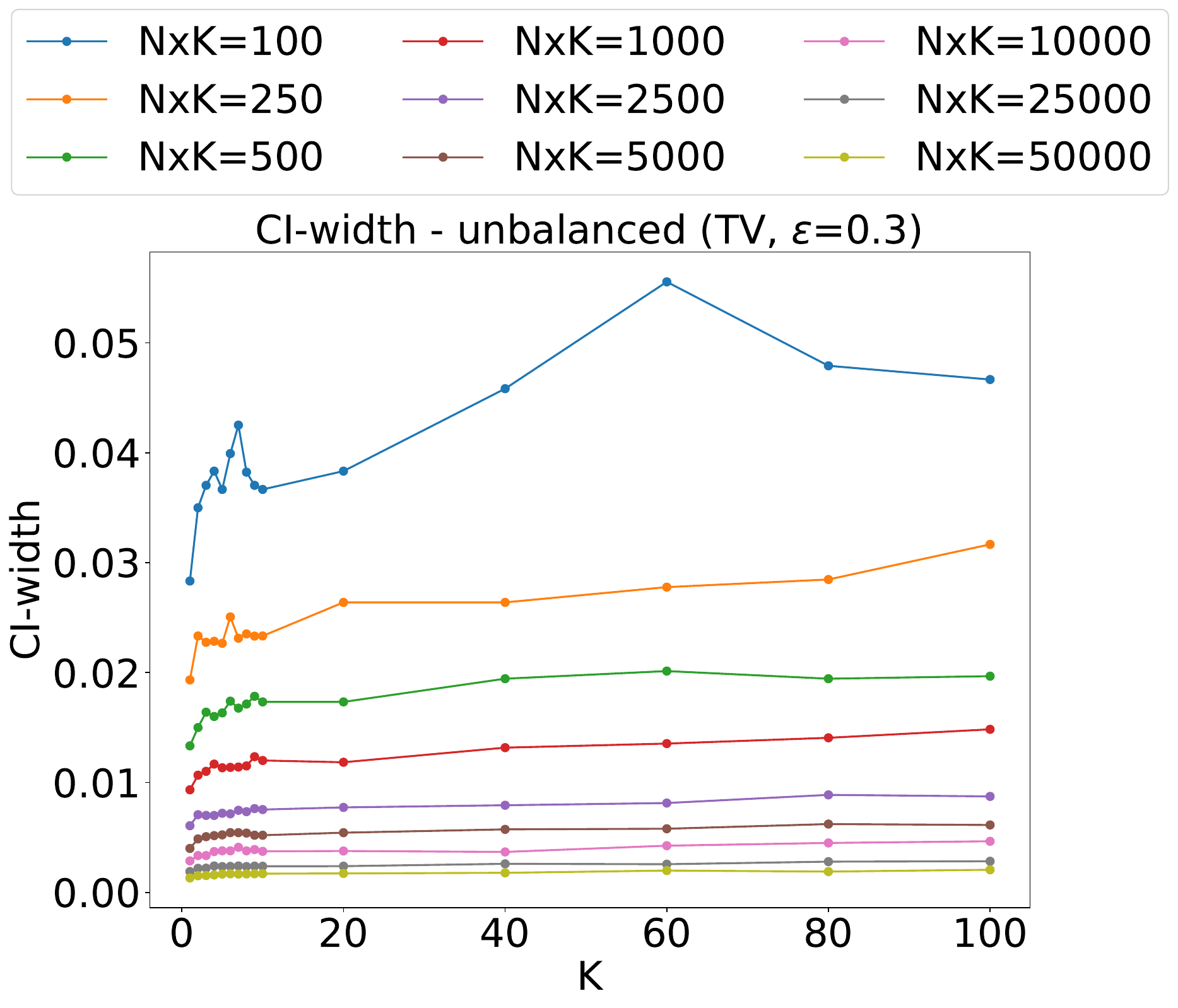}
    \caption{$\epsilon = 0.3$}
    \label{fig:gamma_ci_MAE_cat12_e03}
  \end{subfigure} \hfill
  \begin{subfigure}[b]{0.24\linewidth}
    \centering
    \includegraphics[width=\linewidth]{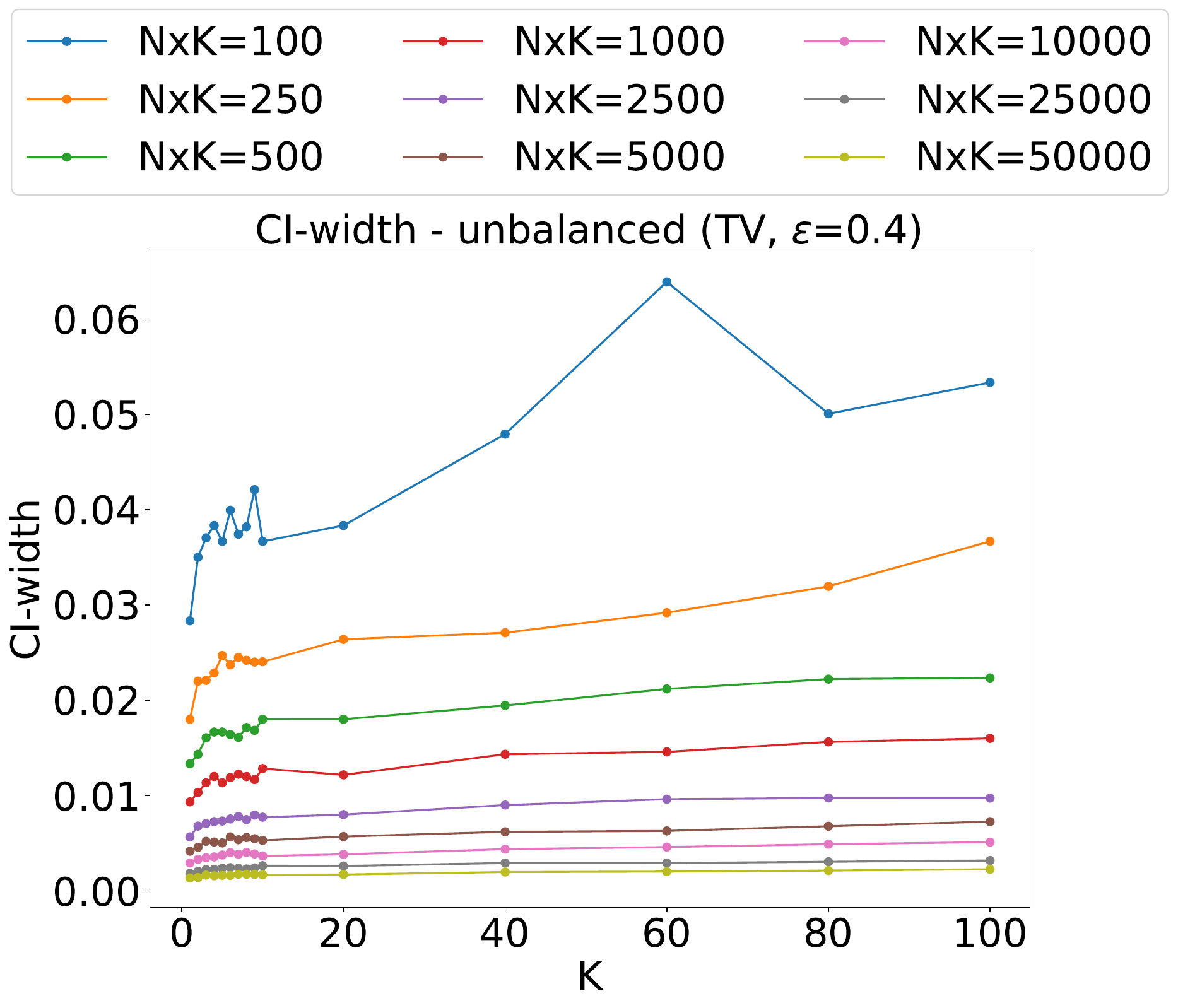}
    \caption{$\epsilon = 0.4$}
    \label{fig:gamma_ci_MAE_cat12_e04}
  \end{subfigure}
  \caption{CI-width plots for unbalanced alphas with TV as the metric ($M=12$)}
  \label{fig:gamma_ci_MAE_cat12}
\end{figure*}

\begin{figure*}
  \centering
  \begin{subfigure}[b]{0.24\linewidth}
    \centering
    \includegraphics[width=\linewidth]{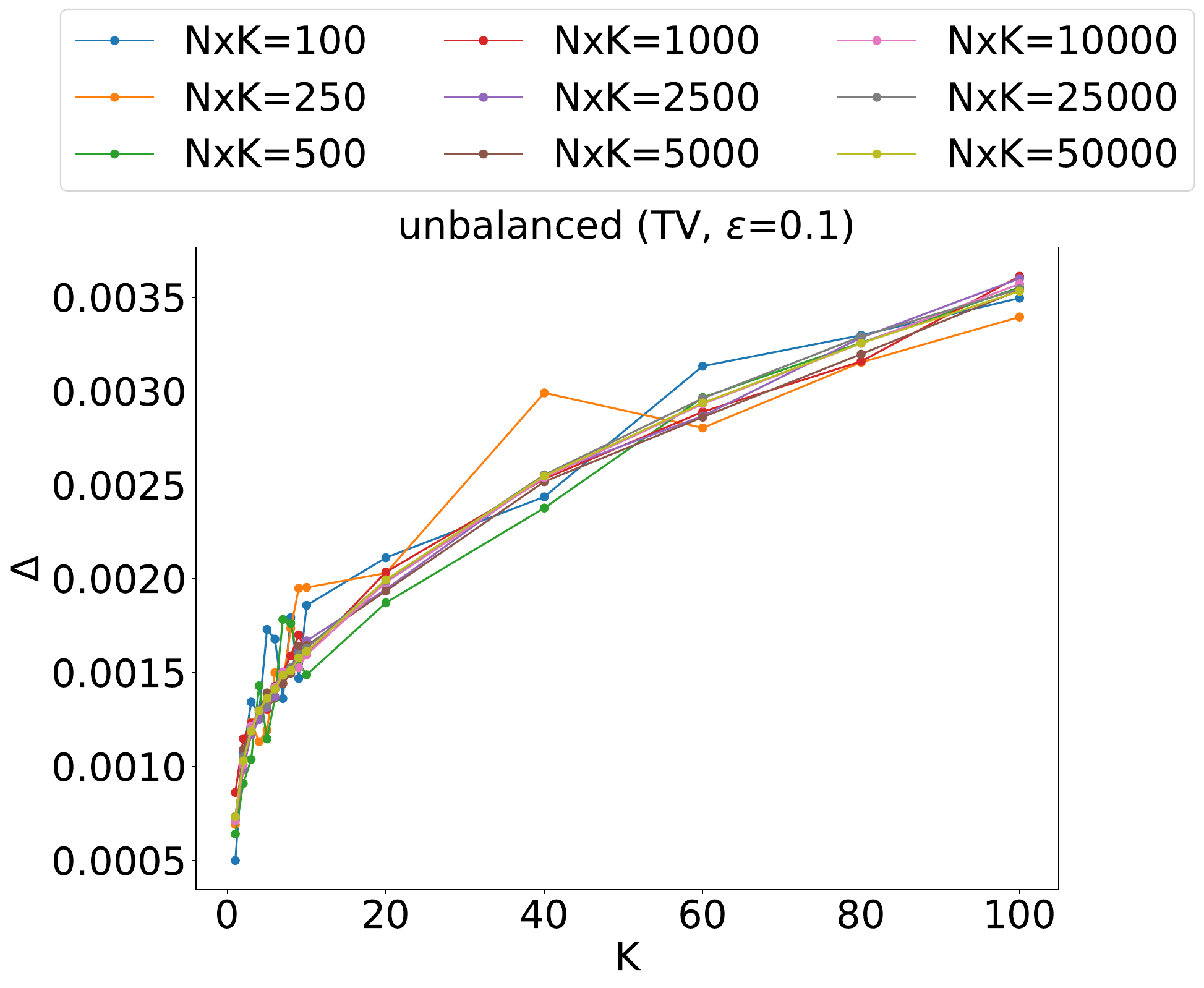}
    \caption{$\epsilon = 0.1$}
    \label{fig:gamma_delta_MAE_cat12_e01}
  \end{subfigure} \hfill
  \begin{subfigure}[b]{0.24\linewidth}
    \centering
    \includegraphics[width=\linewidth]{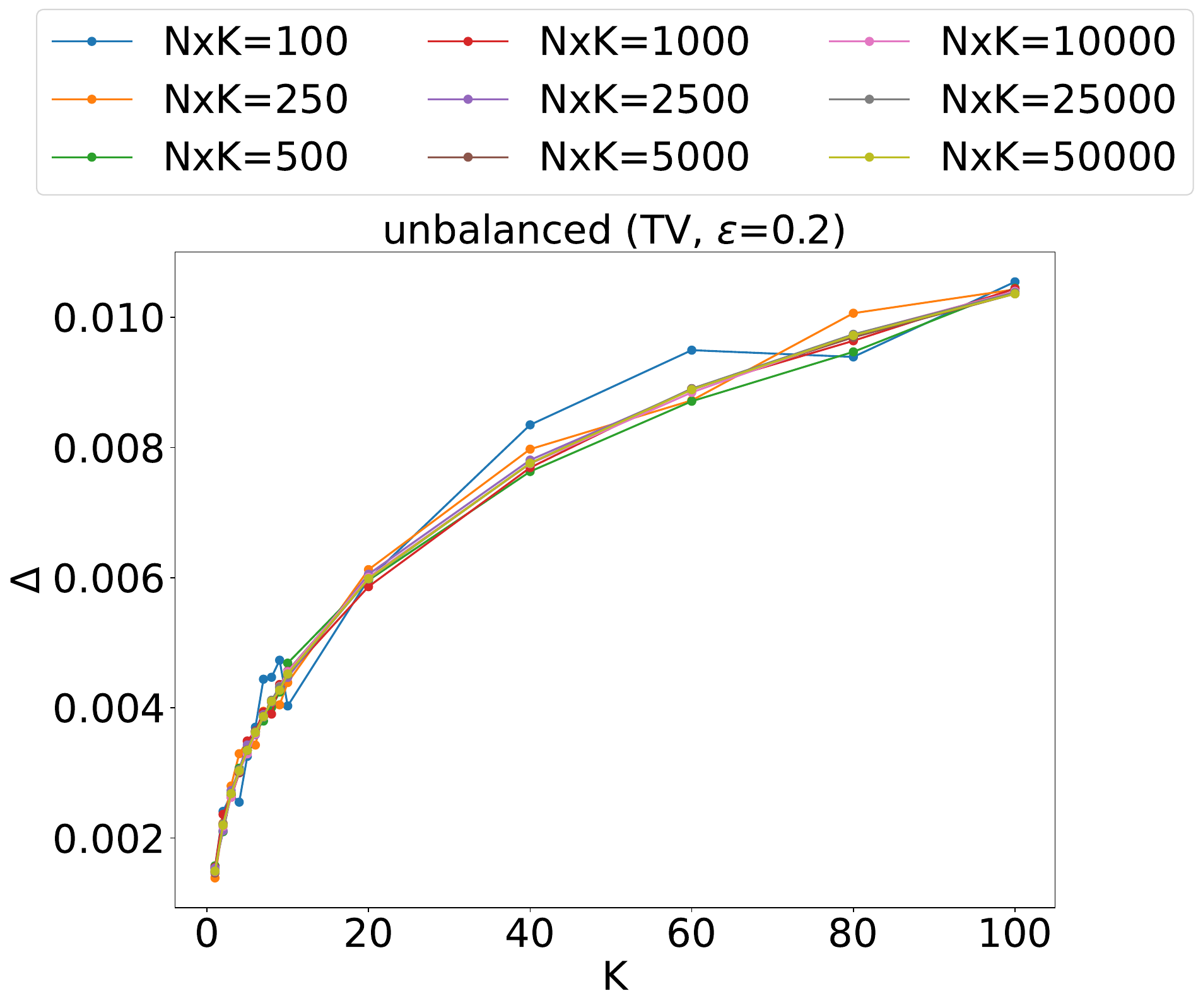}
    \caption{$\epsilon = 0.2$}
    \label{fig:gamma_delta_MAE_cat12_e02}
  \end{subfigure} \hfill
  \begin{subfigure}[b]{0.24\linewidth}
    \centering
    \includegraphics[width=\linewidth]{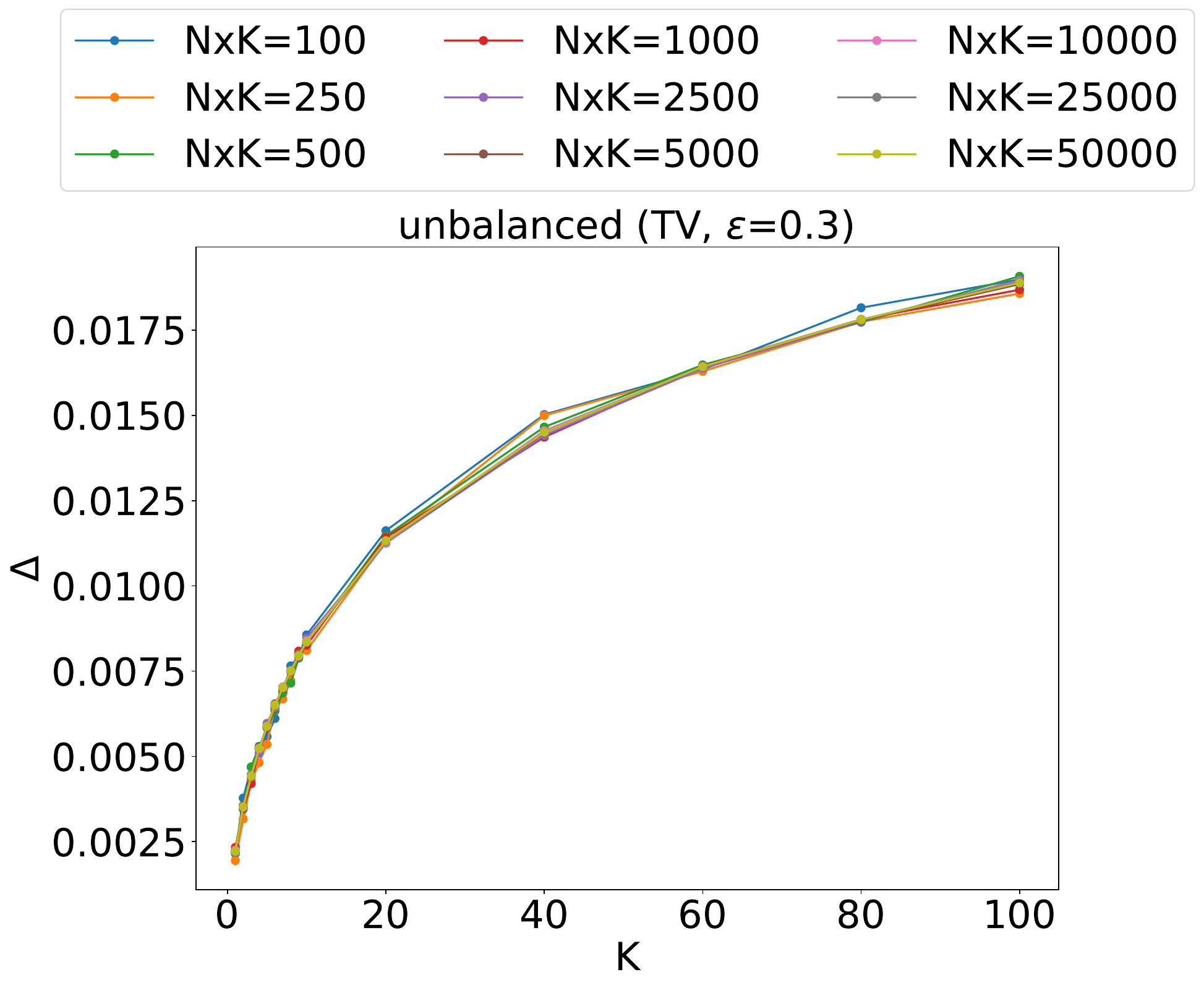}
    \caption{$\epsilon = 0.3$}
    \label{fig:gamma_delta_MAE_cat12_e03}
  \end{subfigure} \hfill
  \begin{subfigure}[b]{0.24\linewidth}
    \centering
    \includegraphics[width=\linewidth]{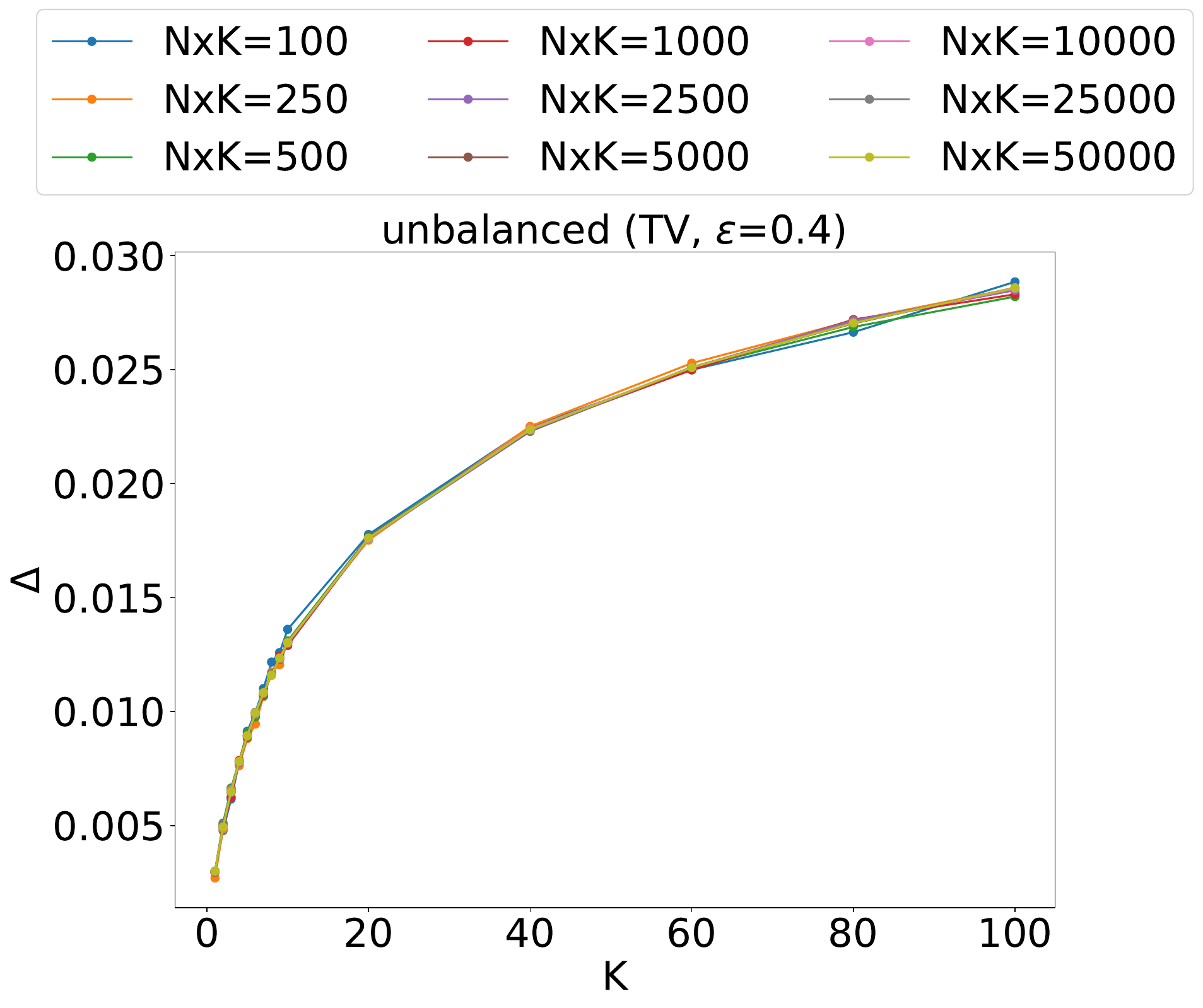}
    \caption{$\epsilon = 0.4$}
    \label{fig:gamma_delta_MAE_cat12_e04}
  \end{subfigure}
  \caption{Effect sizes ($\Delta$) for unbalanced alphas with TV as the metric ($M=12$)}
  \label{fig:gamma_delta_MAE_cat12}
\end{figure*}

Figures \ref{fig:gamma_MAE_cat2}--\ref{fig:gamma_delta_MAE_cat12} show results for \pv s, confidence intervals, and effect sizes of TV for unbalanced priors on parameter $\alpha$ with different values of $\epsilon$ and increasing $K$ for various $N \times K$. For simulations with unbalanced categories, \pv s\ decrease with increasing $K$ as $M$ gets larger, whereas CI-width and effect sizes increase as $K$ increases.

\subsection{Wins}

\subsubsection{Real-World Datasets}

\begin{figure*}
  \centering
  \begin{subfigure}[b]{0.24\linewidth}
    \centering
    \includegraphics[width=\linewidth]{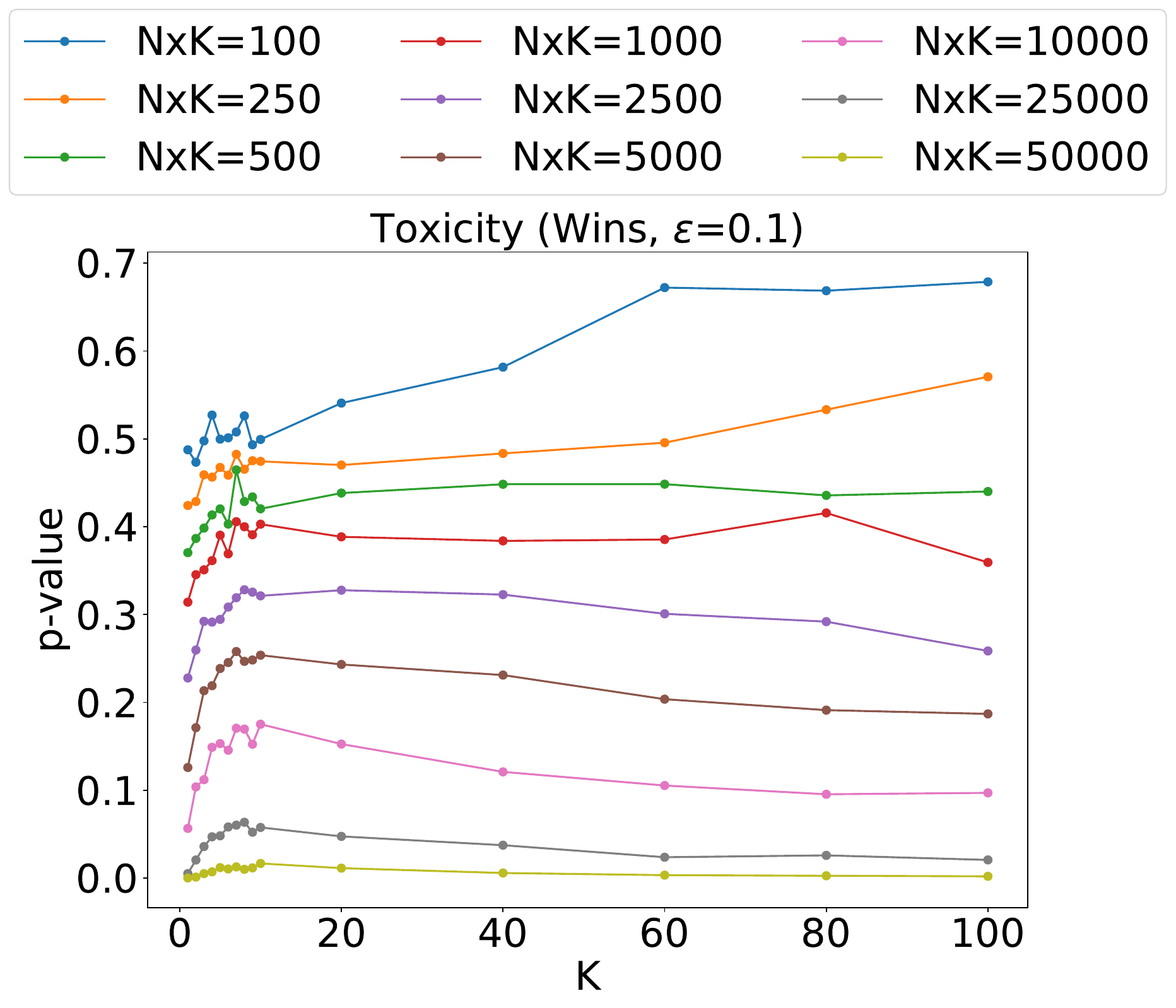}
    \caption{$\epsilon = 0.1$}
    \label{fig:toxicity_wins_e01}
  \end{subfigure} \hfill
  \begin{subfigure}[b]{0.24\linewidth}
    \centering
    \includegraphics[width=\linewidth]{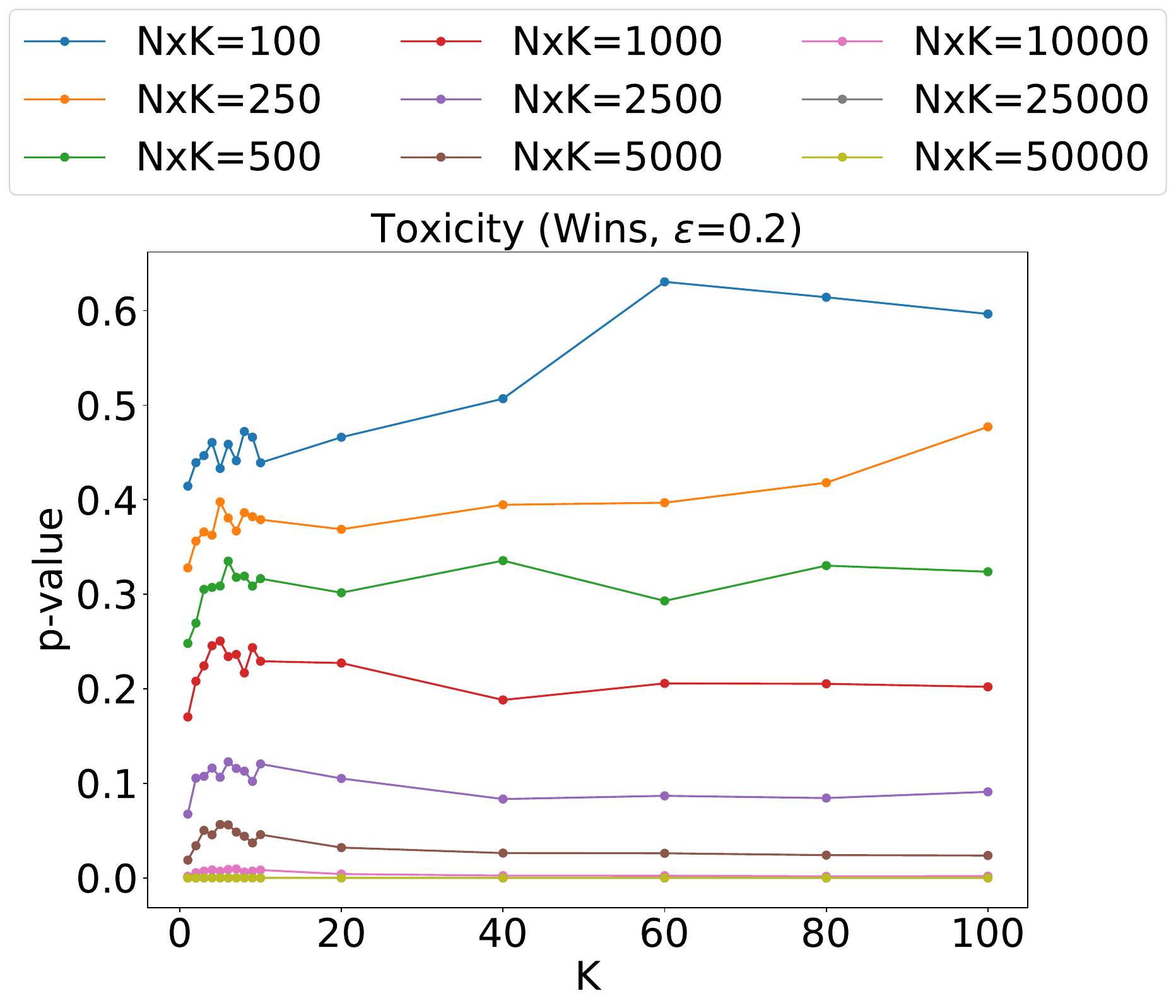}
    \caption{$\epsilon = 0.2$}
    \label{fig:toxicity_wins_e02}
  \end{subfigure} \hfill
  \begin{subfigure}[b]{0.24\linewidth}
    \centering
    \includegraphics[width=\linewidth]{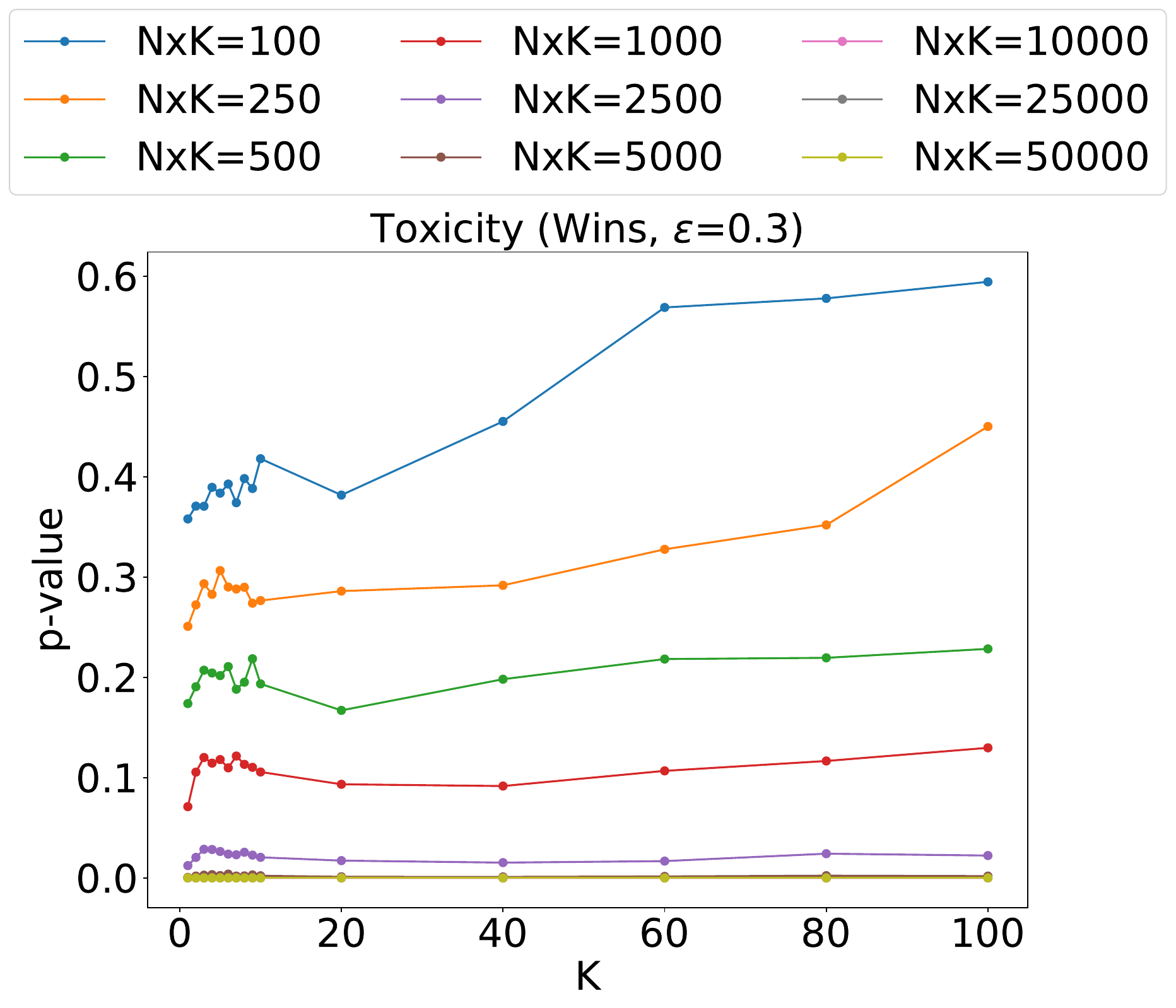}
    \caption{$\epsilon = 0.3$}
    \label{fig:toxicity_wins_e03}
  \end{subfigure} \hfill
  \begin{subfigure}[b]{0.24\linewidth}
    \centering
    \includegraphics[width=\linewidth]{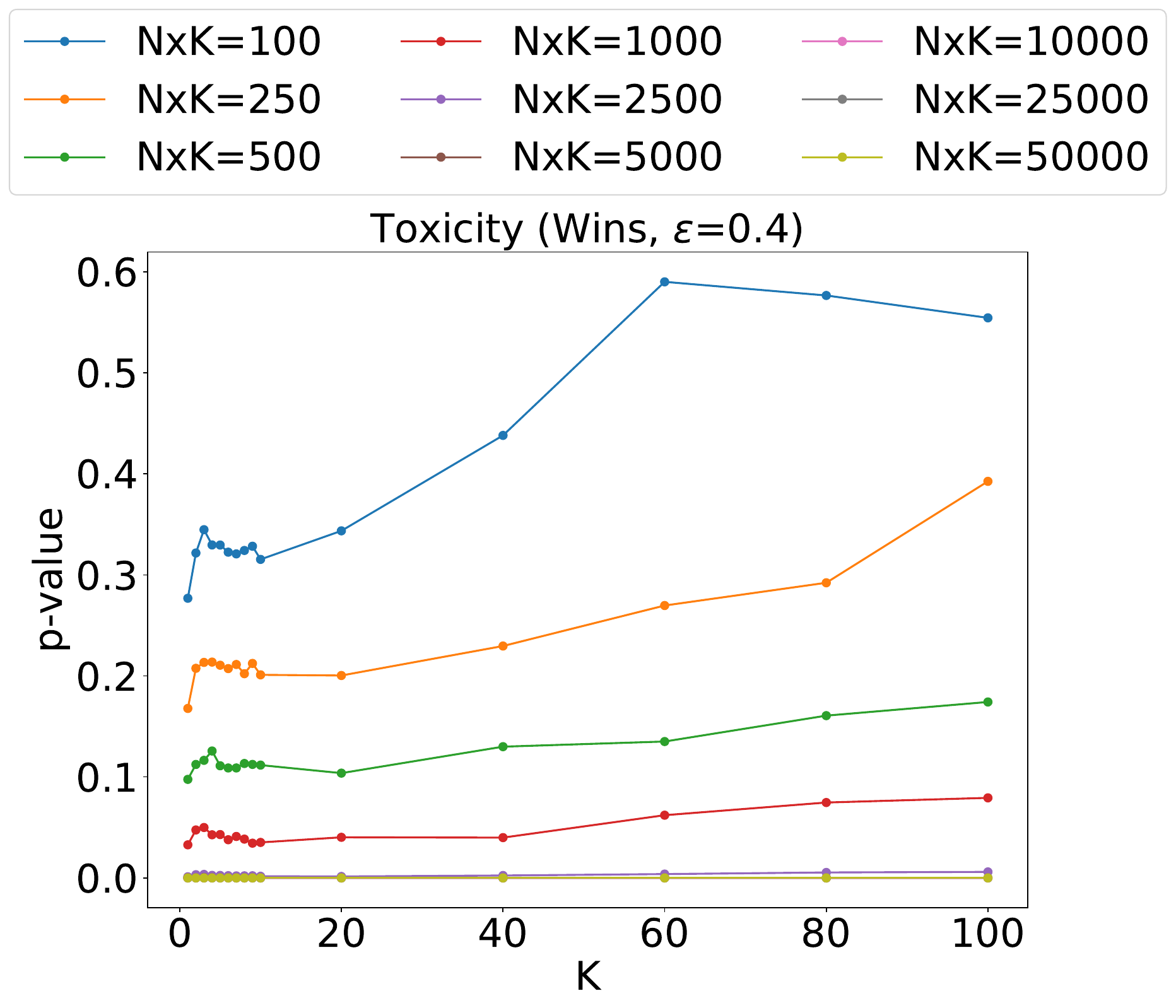}
    \caption{$\epsilon = 0.4$}
    \label{fig:toxicity_wins_e04}
  \end{subfigure}
  \caption{P-value plots for Toxicity dataset with Wins as the metric}
  \label{fig:toxicity_wins}
\end{figure*}

\begin{figure*}
  \centering
  \begin{subfigure}[b]{0.24\linewidth}
    \centering
    \includegraphics[width=\linewidth]{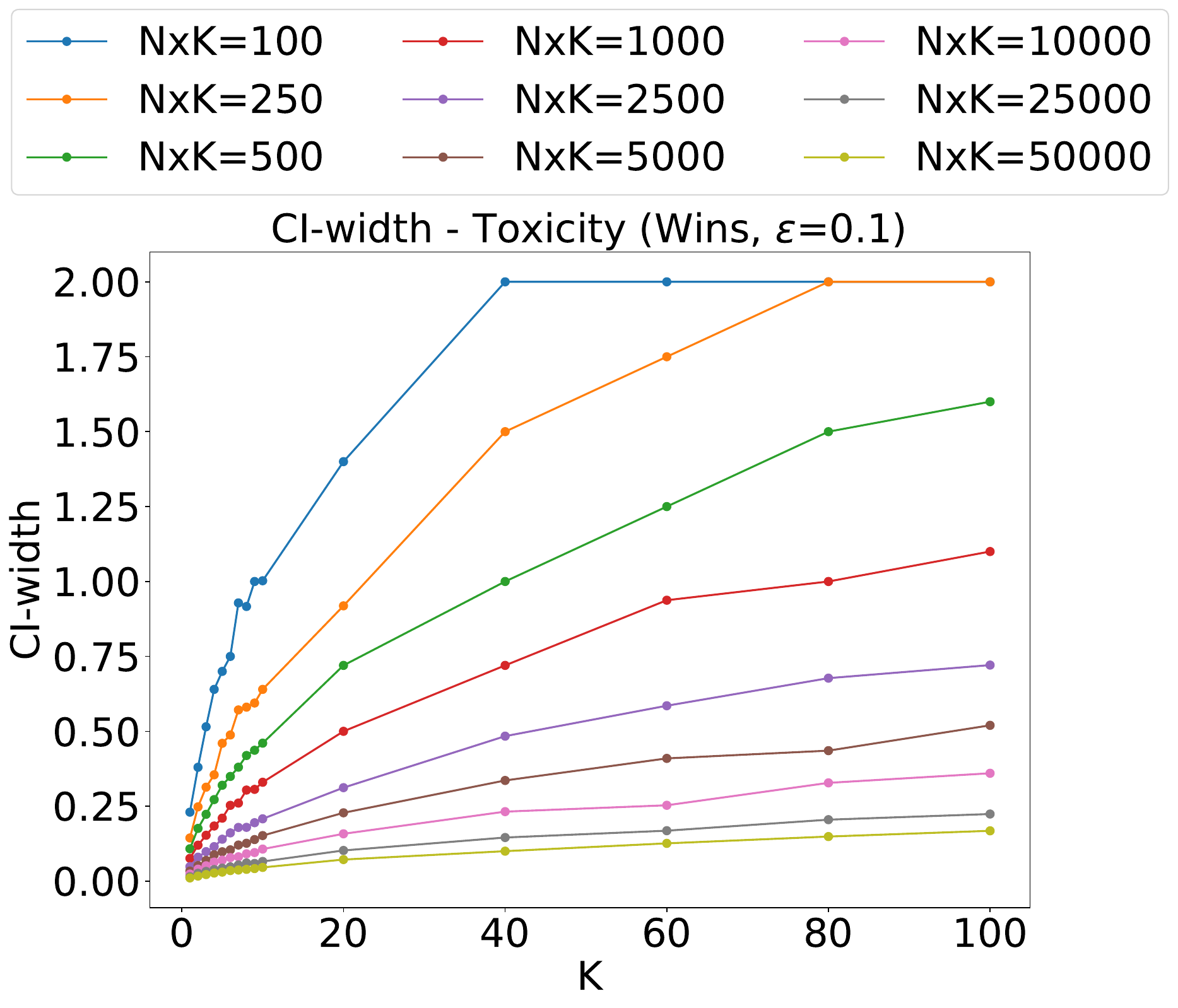}
    \caption{$\epsilon = 0.1$}
    \label{fig:toxicity_ci_wins_e01}
  \end{subfigure} \hfill
  \begin{subfigure}[b]{0.24\linewidth}
    \centering
    \includegraphics[width=\linewidth]{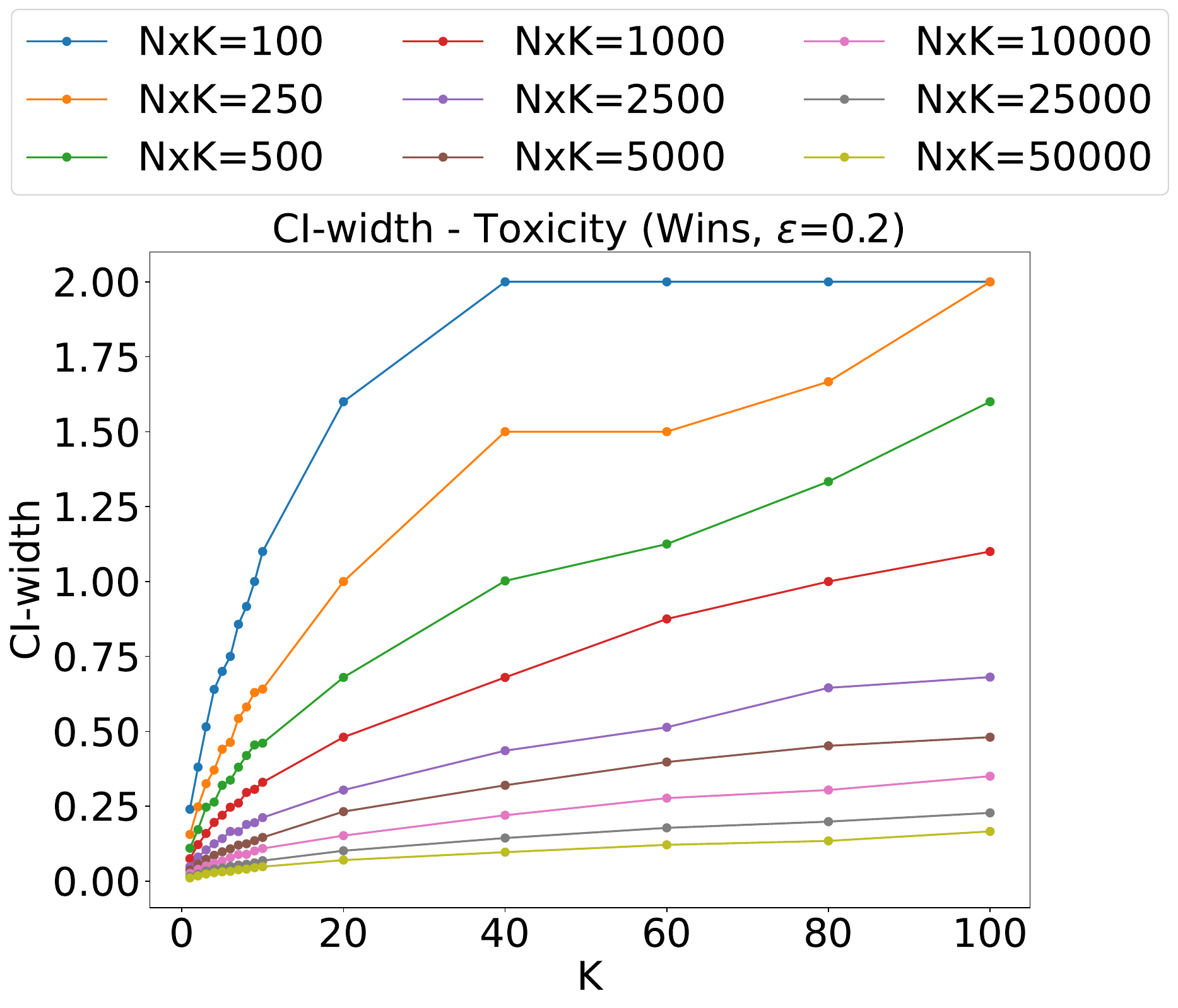}
    \caption{$\epsilon = 0.2$}
    \label{fig:toxicity_ci_wins_e02}
  \end{subfigure} \hfill
  \begin{subfigure}[b]{0.24\linewidth}
    \centering
    \includegraphics[width=\linewidth]{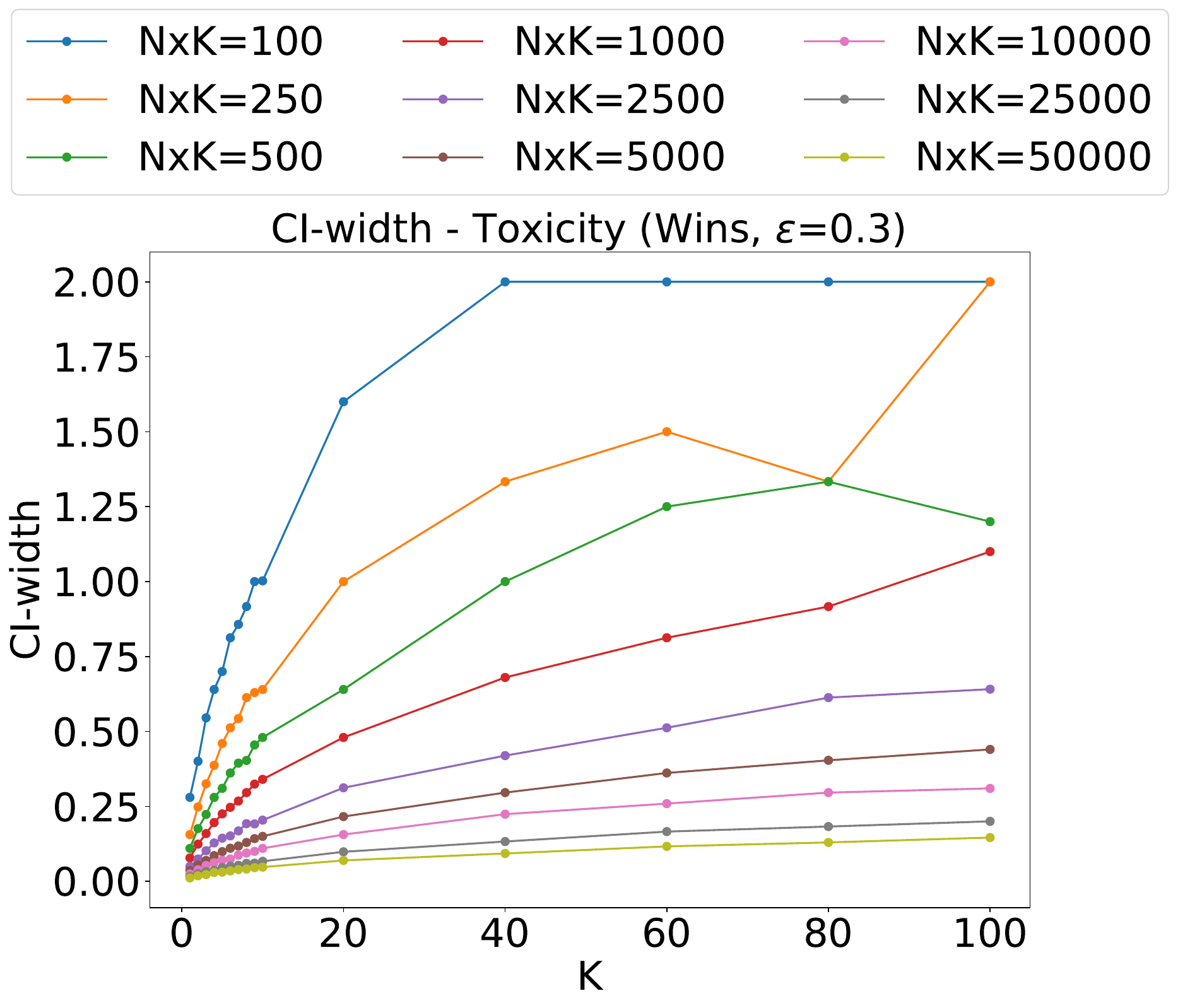}
    \caption{$\epsilon = 0.3$}
    \label{fig:toxicity_ci_wins_e03}
  \end{subfigure} \hfill
  \begin{subfigure}[b]{0.24\linewidth}
    \centering
    \includegraphics[width=\linewidth]{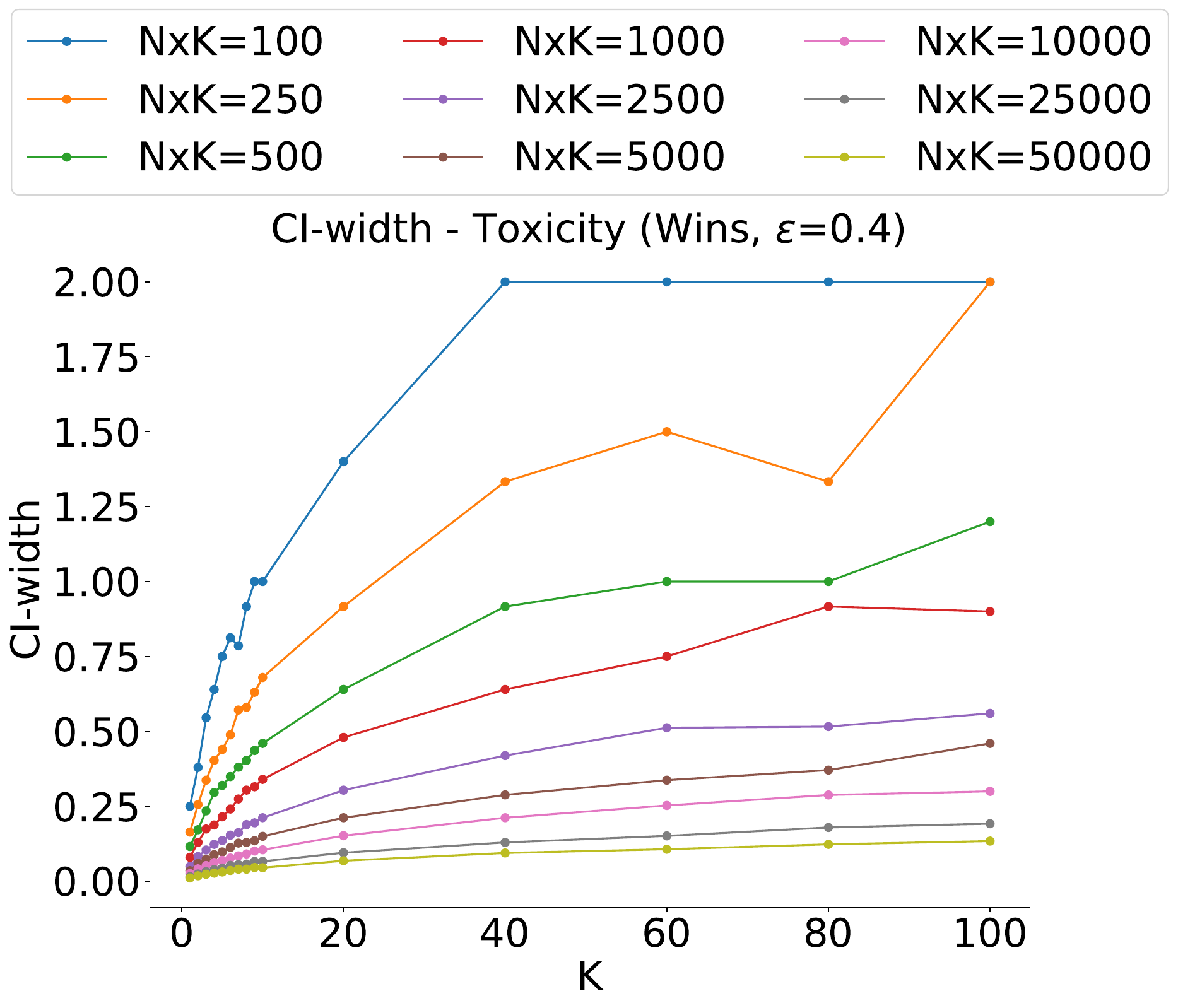}
    \caption{$\epsilon = 0.4$}
    \label{fig:toxicity_ci_wins_e04}
  \end{subfigure}
  \caption{CI-width plots for Toxicity dataset with Wins as the metric}
  \label{fig:toxicity_ci_wins}
\end{figure*}

\begin{figure*}
  \centering
  \begin{subfigure}[b]{0.24\linewidth}
    \centering
    \includegraphics[width=\linewidth]{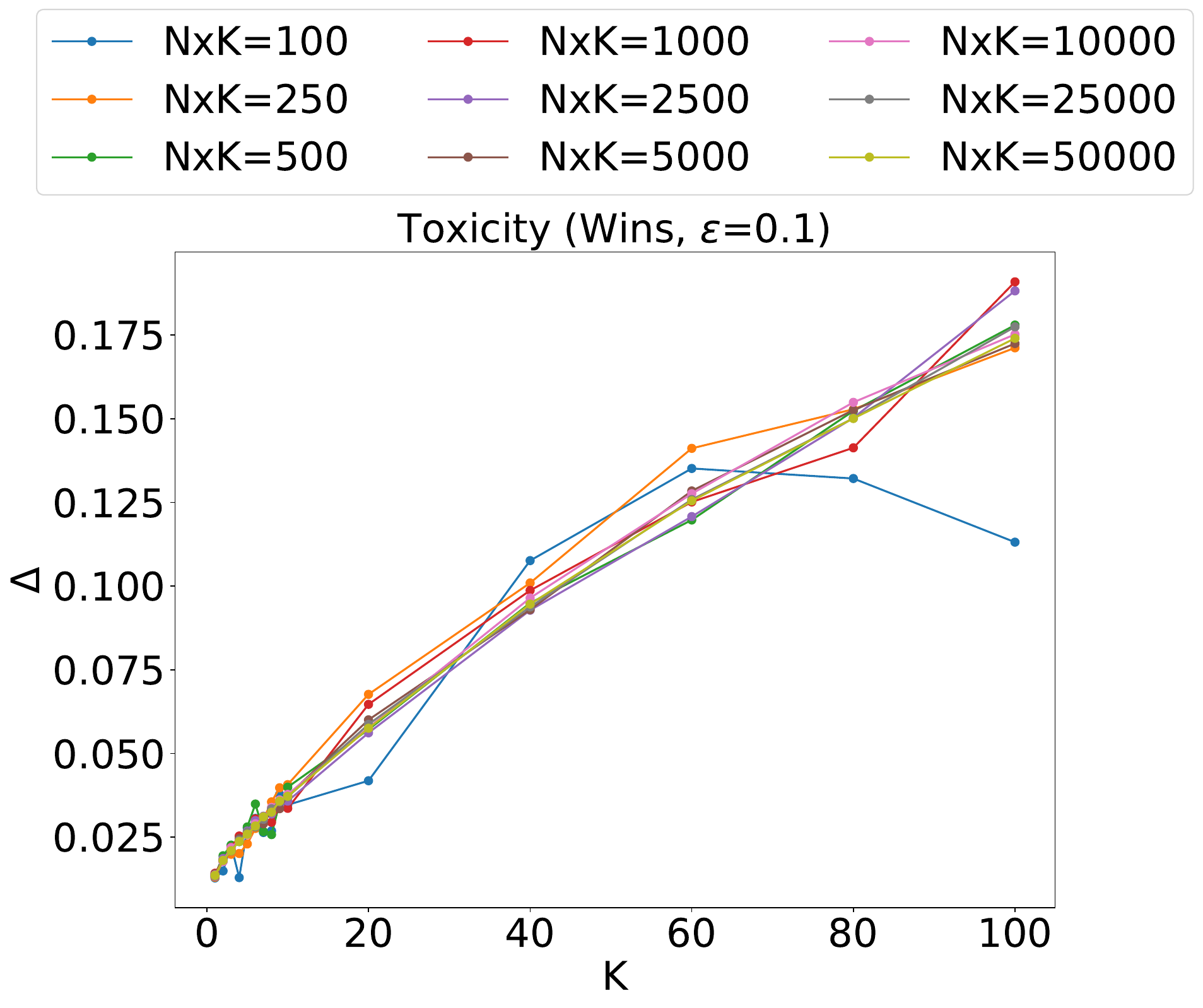}
    \caption{$\epsilon = 0.1$}
    \label{fig:toxicity_delta_wins_e01}
  \end{subfigure} \hfill
  \begin{subfigure}[b]{0.24\linewidth}
    \centering
    \includegraphics[width=\linewidth]{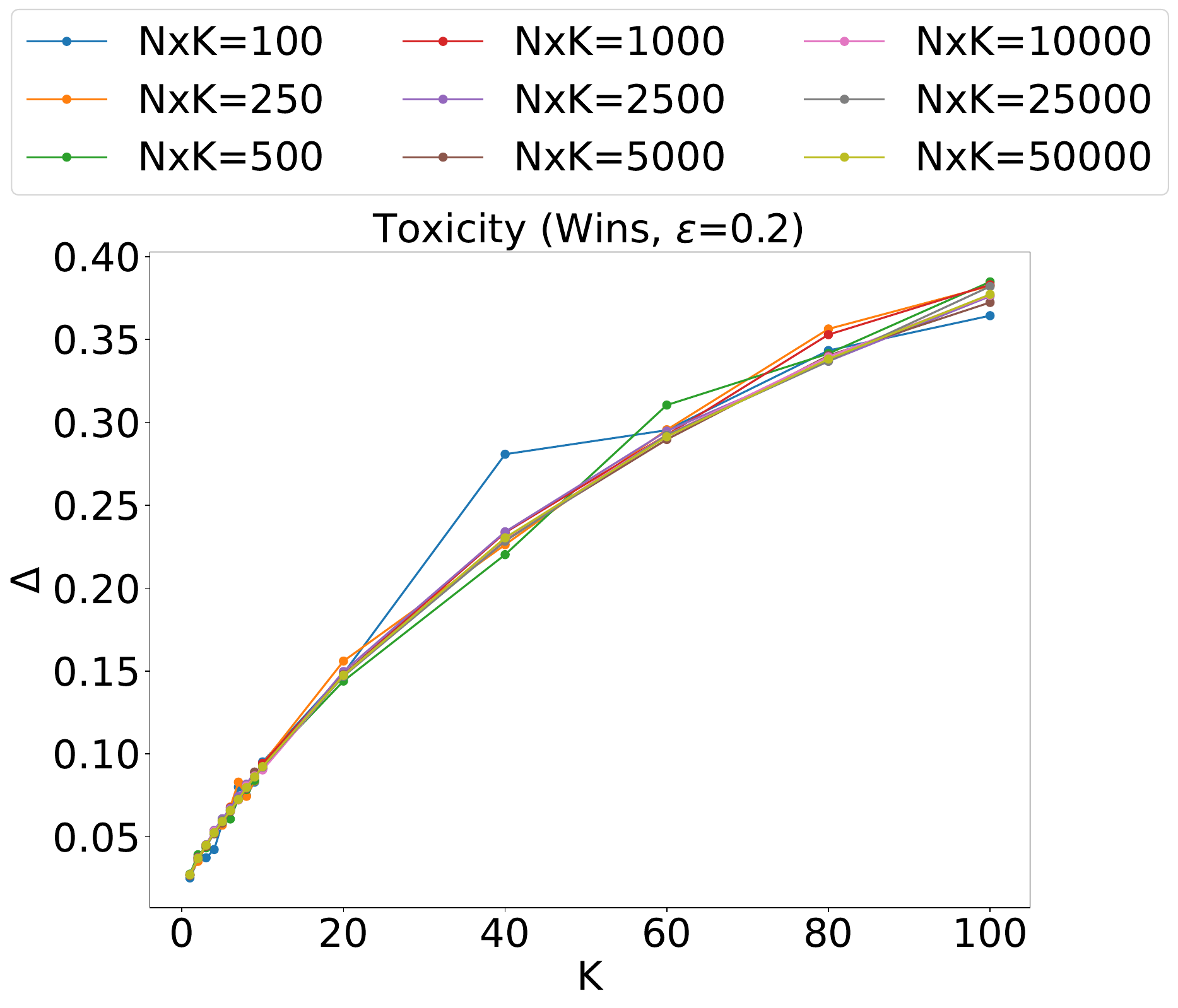}
    \caption{$\epsilon = 0.2$}
    \label{fig:toxicity_delta_wins_e02}
  \end{subfigure} \hfill
  \begin{subfigure}[b]{0.24\linewidth}
    \centering
    \includegraphics[width=\linewidth]{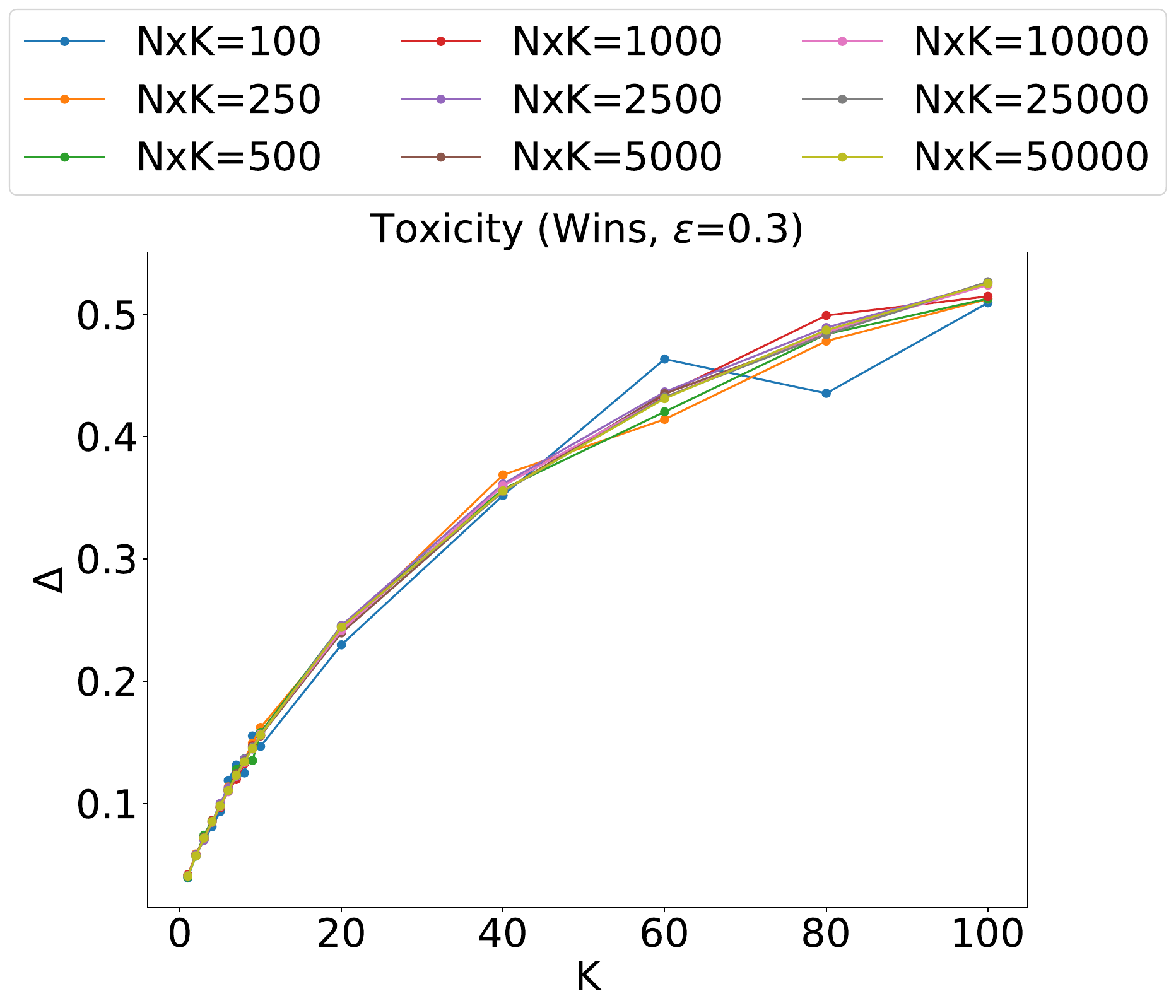}
    \caption{$\epsilon = 0.3$}
    \label{fig:toxicity_delta_wins_e03}
  \end{subfigure} \hfill
  \begin{subfigure}[b]{0.24\linewidth}
    \centering
    \includegraphics[width=\linewidth]{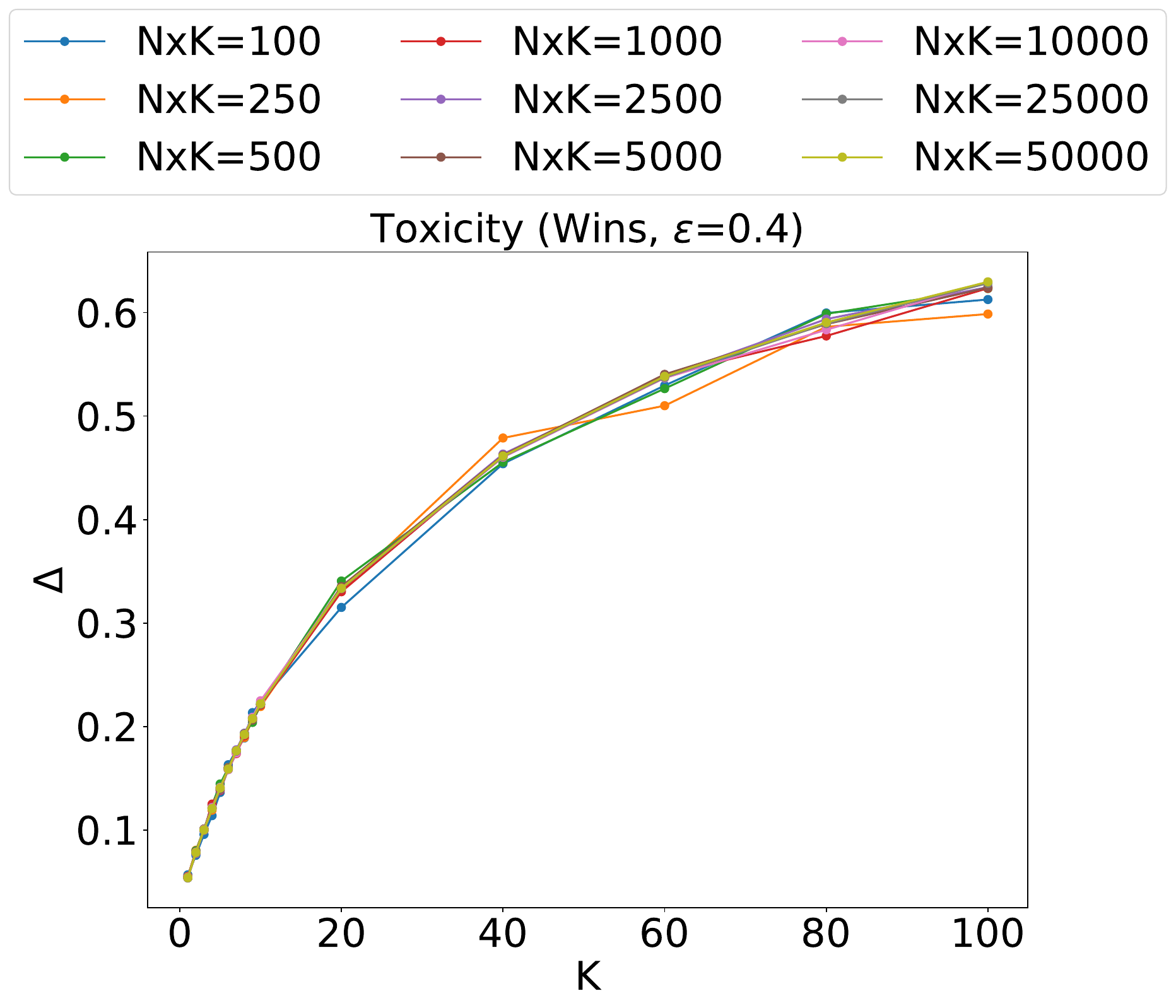}
    \caption{$\epsilon = 0.4$}
    \label{fig:toxicity_delta_wins_e04}
  \end{subfigure}
  \caption{Effect sizes ($\Delta$) for Toxicity dataset with Wins as the metric}
  \label{fig:toxicity_delta_wins}
\end{figure*}

\begin{figure*}
  \centering
  \begin{subfigure}[b]{0.24\linewidth}
    \centering
    \includegraphics[width=\linewidth]{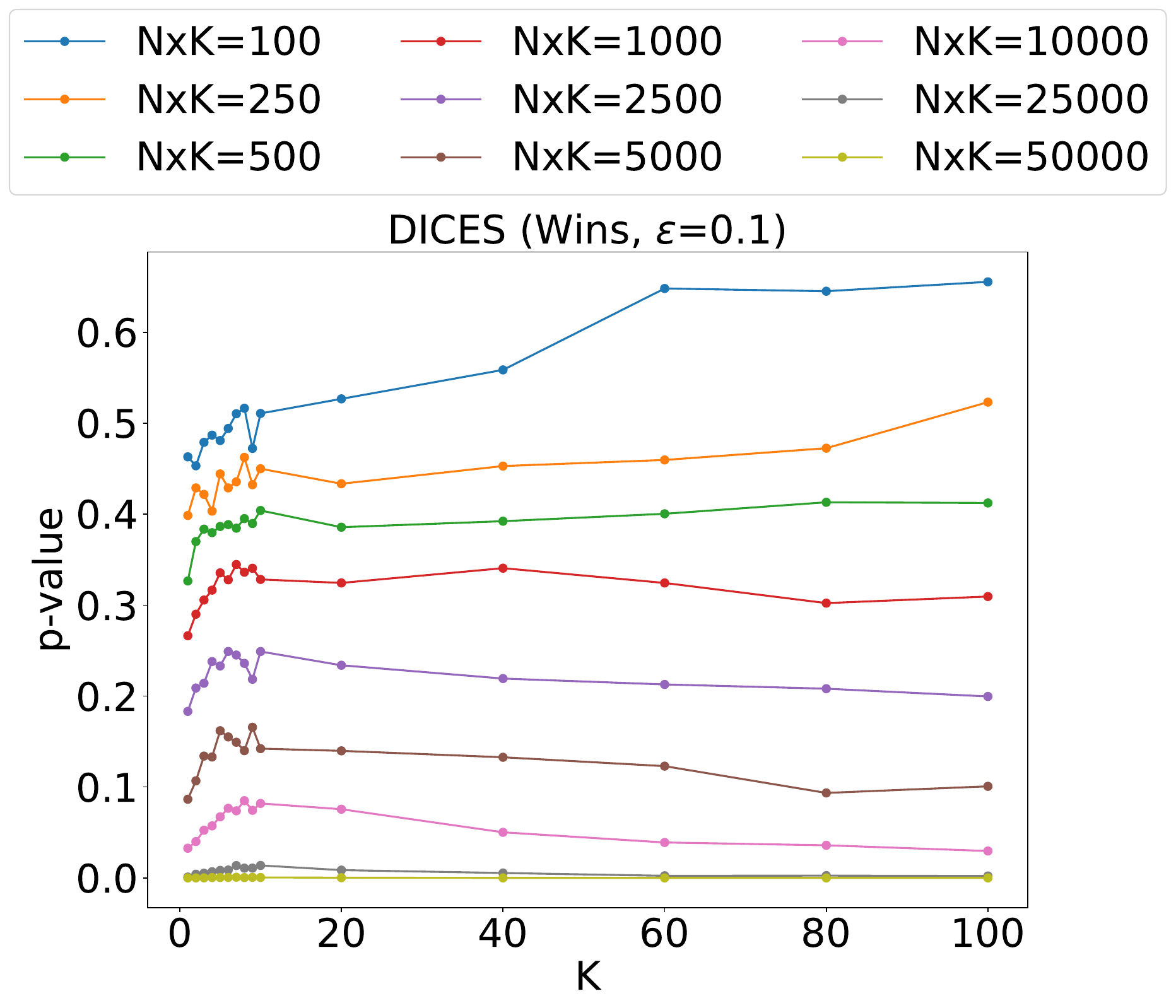}
    \caption{$\epsilon = 0.1$}
    \label{fig:dices_wins_e01}
  \end{subfigure} \hfill
  \begin{subfigure}[b]{0.24\linewidth}
    \centering
    \includegraphics[width=\linewidth]{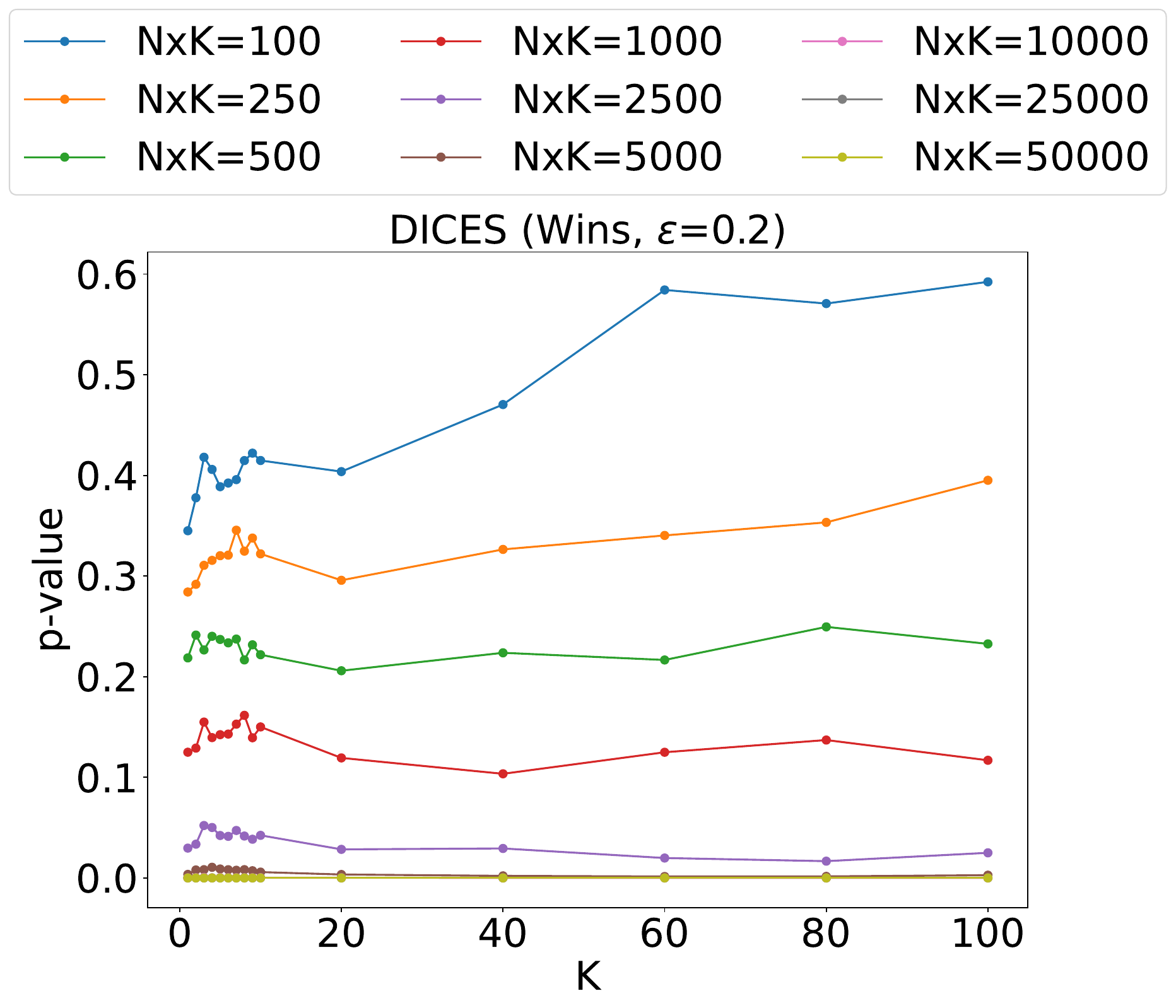}
    \caption{$\epsilon = 0.2$}
    \label{fig:dices_wins_e02}
  \end{subfigure} \hfill
  \begin{subfigure}[b]{0.24\linewidth}
    \centering
    \includegraphics[width=\linewidth]{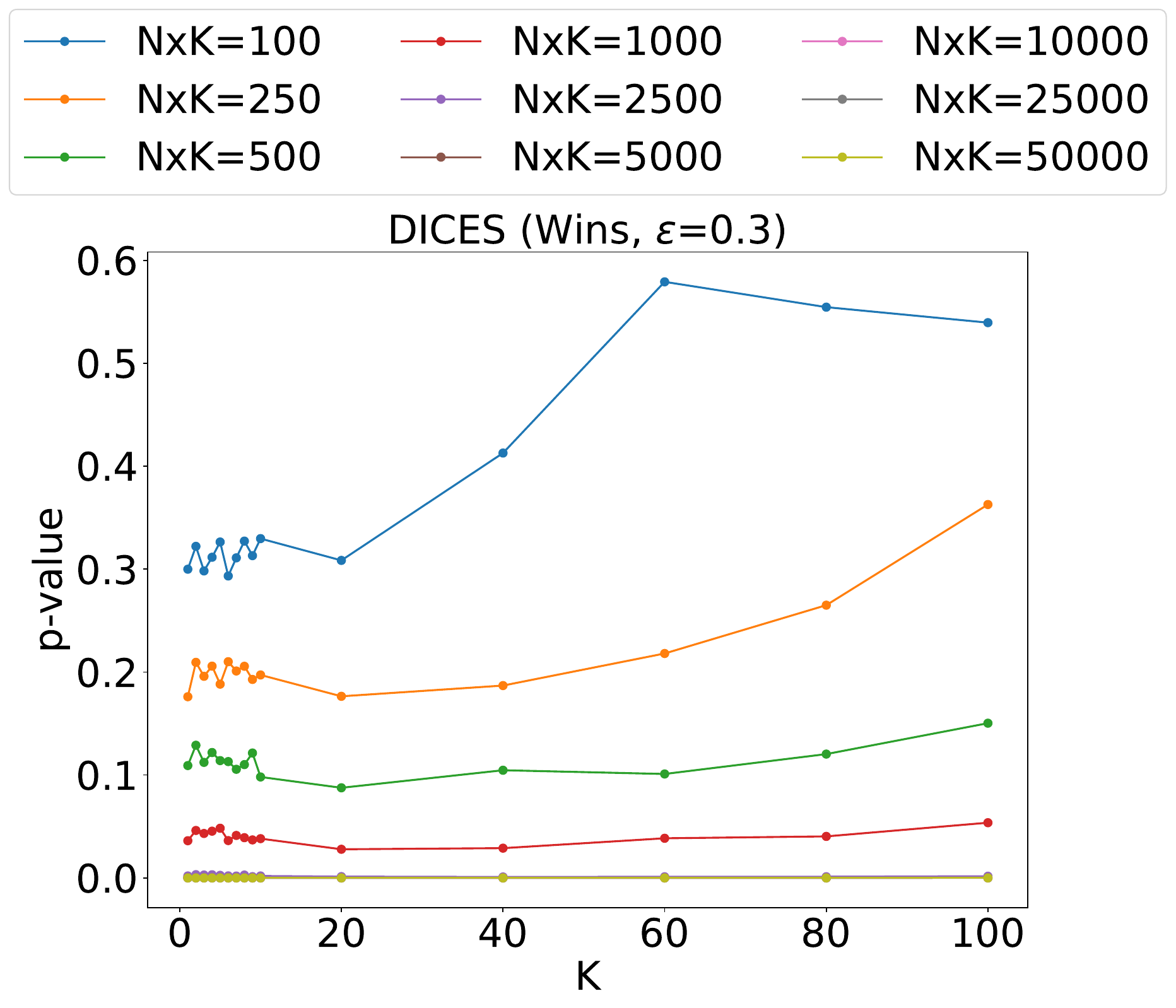}
    \caption{$\epsilon = 0.3$}
    \label{fig:dices_wins_e03}
  \end{subfigure} \hfill
  \begin{subfigure}[b]{0.24\linewidth}
    \centering
    \includegraphics[width=\linewidth]{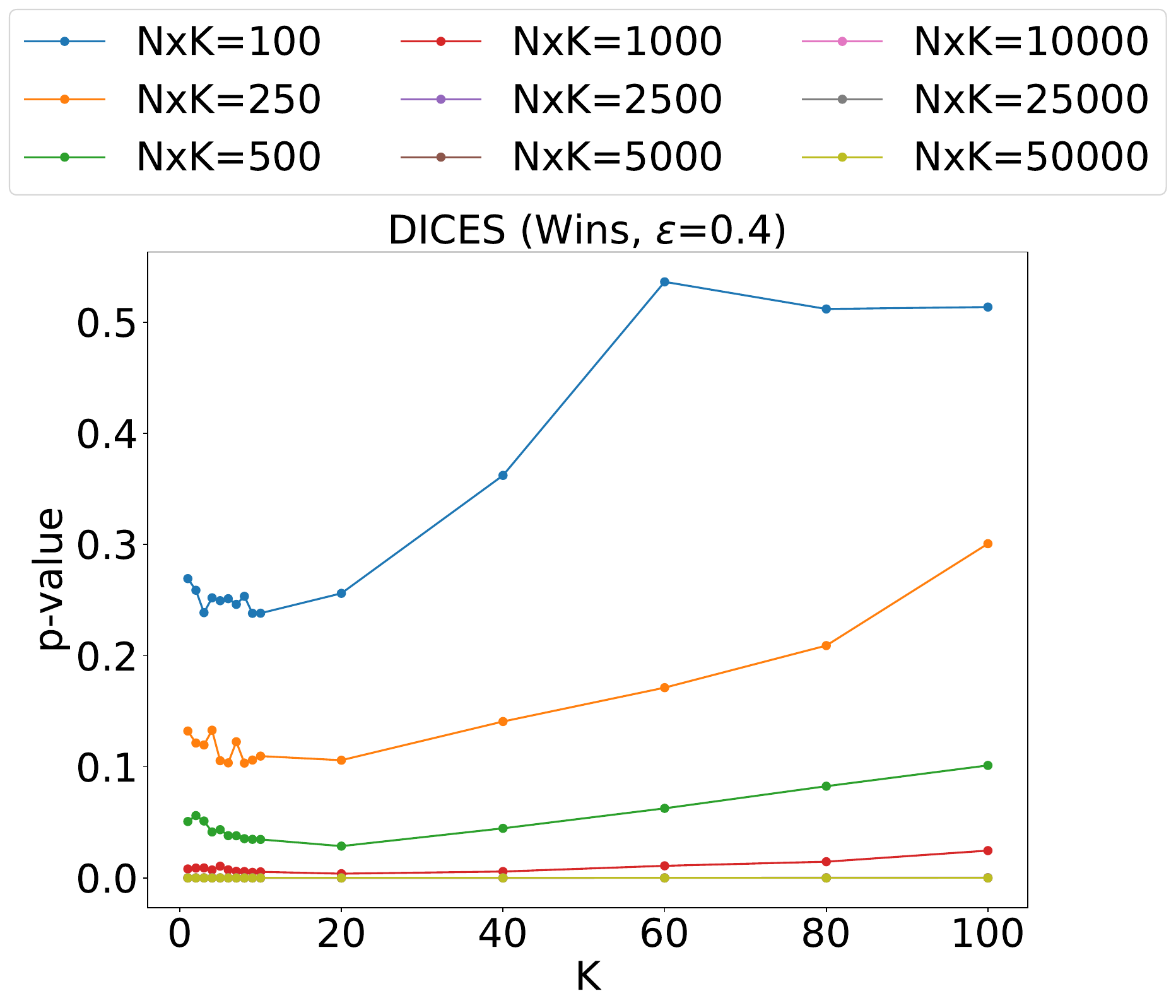}
    \caption{$\epsilon = 0.4$}
    \label{fig:dices_wins_e04}
  \end{subfigure}
  \caption{P-value plots for DICES dataset with Wins as the metric}
  \label{fig:dices_wins}
\end{figure*}

\begin{figure*}
  \centering
  \begin{subfigure}[b]{0.24\linewidth}
    \centering
    \includegraphics[width=\linewidth]{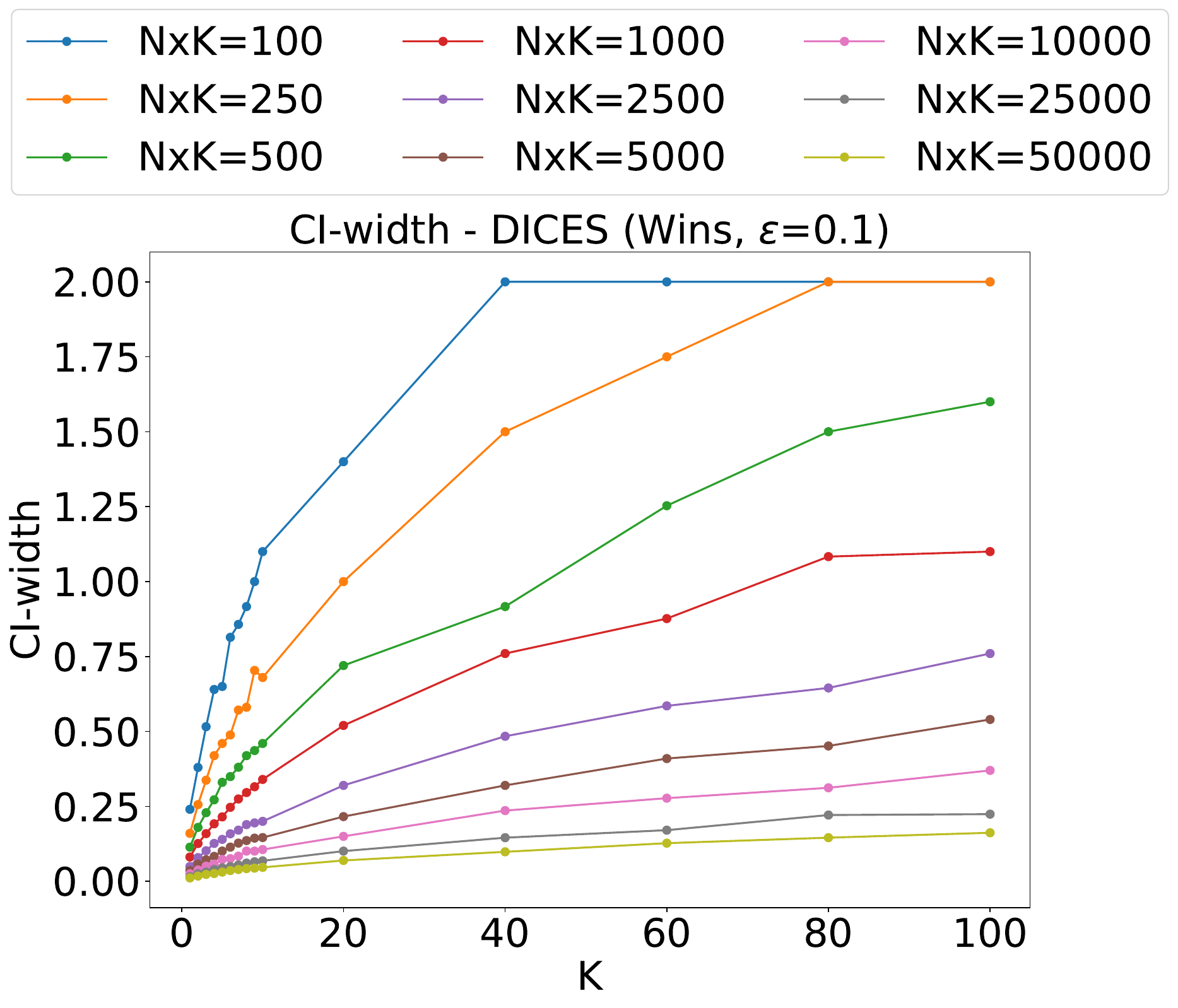}
    \caption{$\epsilon = 0.1$}
    \label{fig:dices_ci_wins_e01}
  \end{subfigure} \hfill
  \begin{subfigure}[b]{0.24\linewidth}
    \centering
    \includegraphics[width=\linewidth]{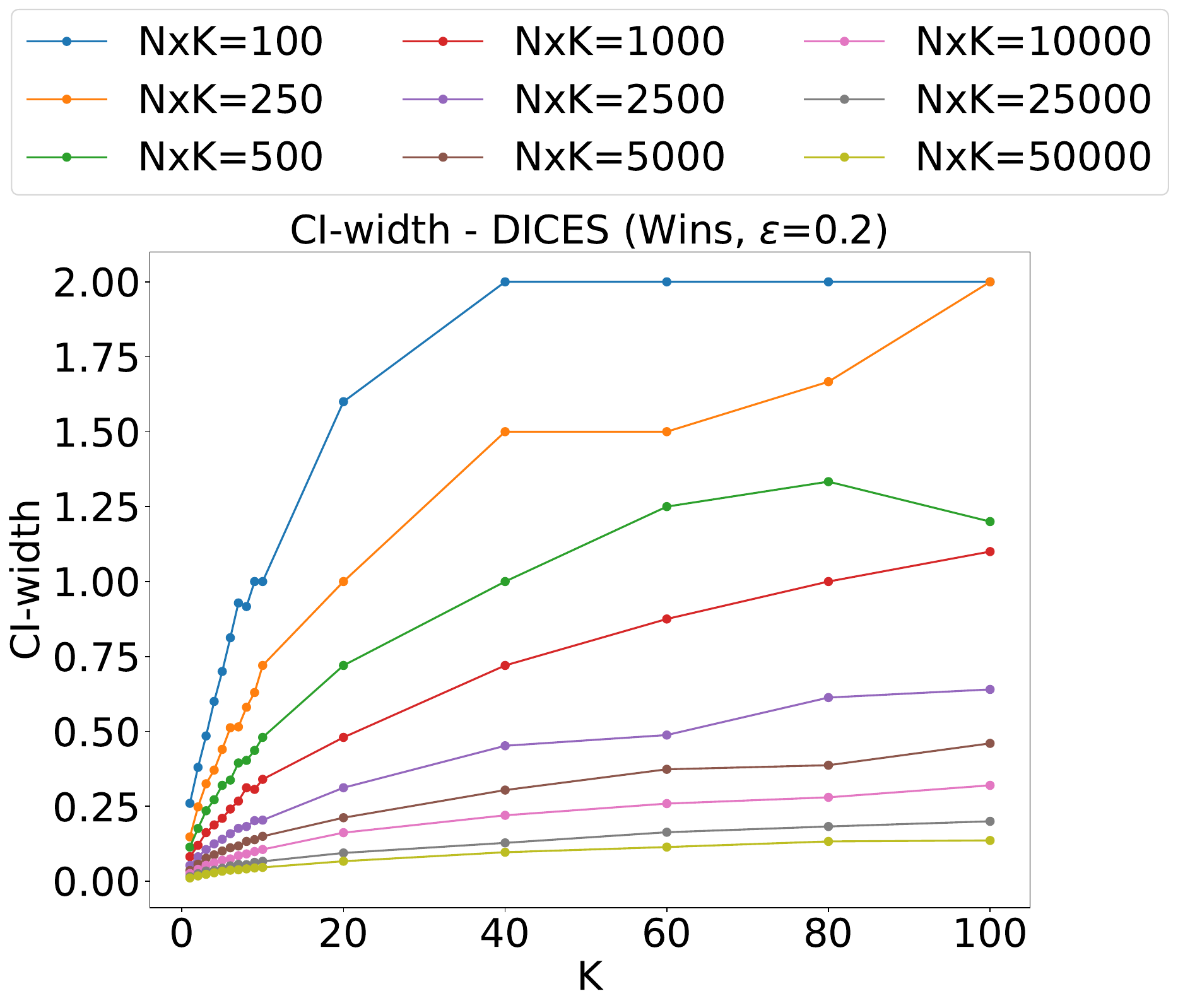}
    \caption{$\epsilon = 0.2$}
    \label{fig:dices_ci_wins_e02}
  \end{subfigure} \hfill
  \begin{subfigure}[b]{0.24\linewidth}
    \centering
    \includegraphics[width=\linewidth]{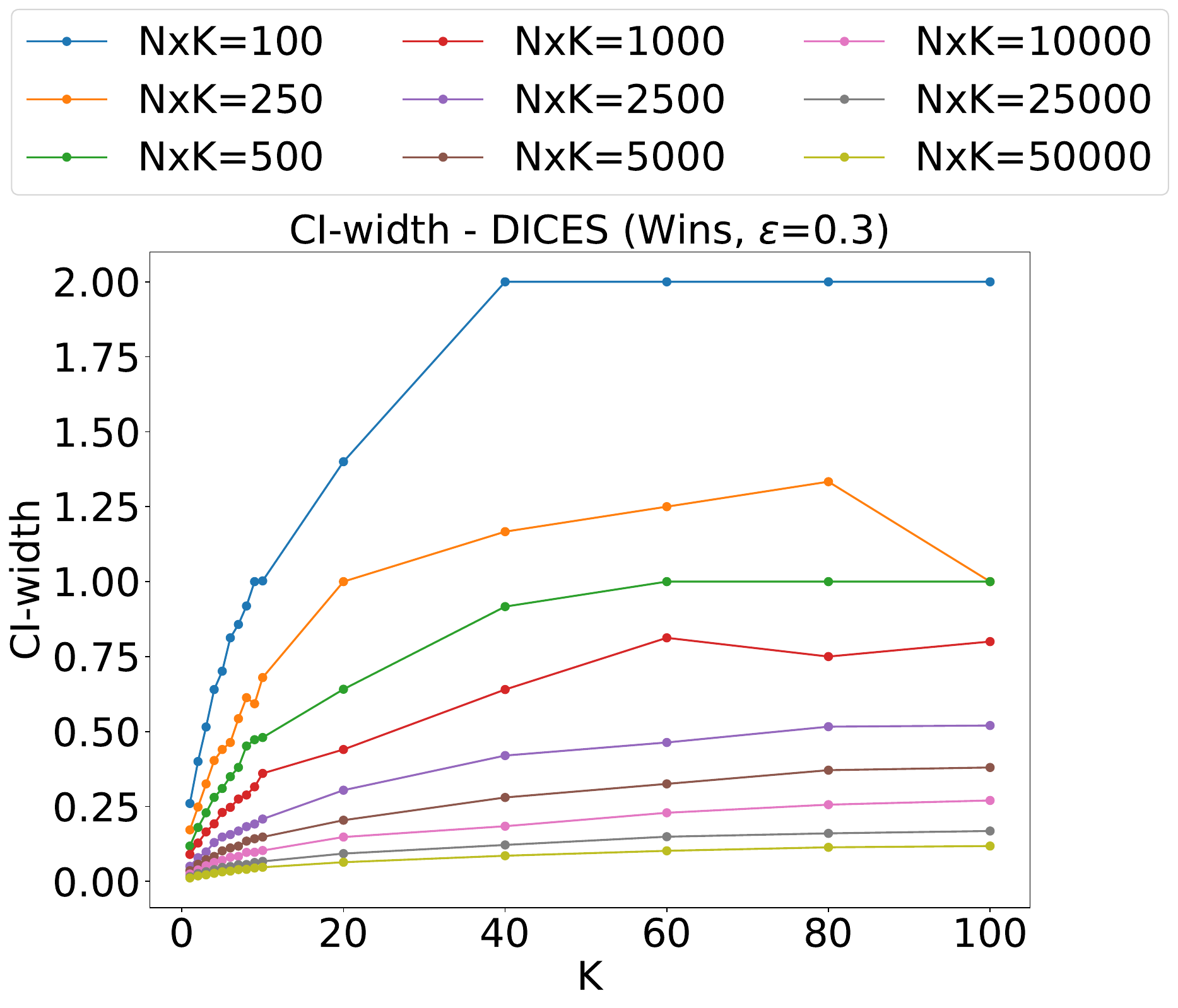}
    \caption{$\epsilon = 0.3$}
    \label{fig:dices_ci_wins_e03}
  \end{subfigure} \hfill
  \begin{subfigure}[b]{0.24\linewidth}
    \centering
    \includegraphics[width=\linewidth]{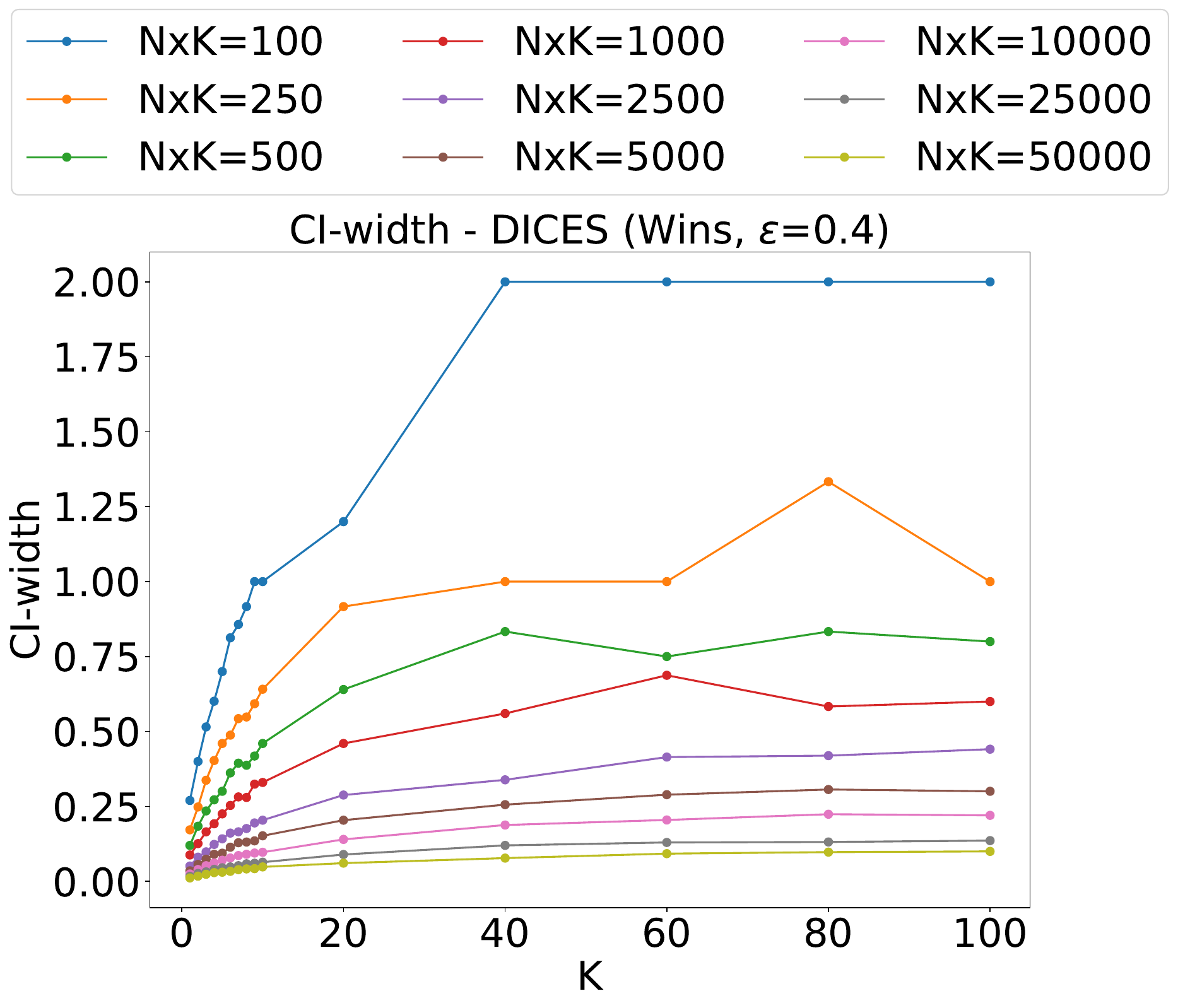}
    \caption{$\epsilon = 0.4$}
    \label{fig:dices_ci_wins_e04}
  \end{subfigure}
  \caption{CI-width plots for DICES dataset with Wins as the metric}
  \label{fig:dices_ci_wins}
\end{figure*}

\begin{figure*}
  \centering
  \begin{subfigure}[b]{0.24\linewidth}
    \centering
    \includegraphics[width=\linewidth]{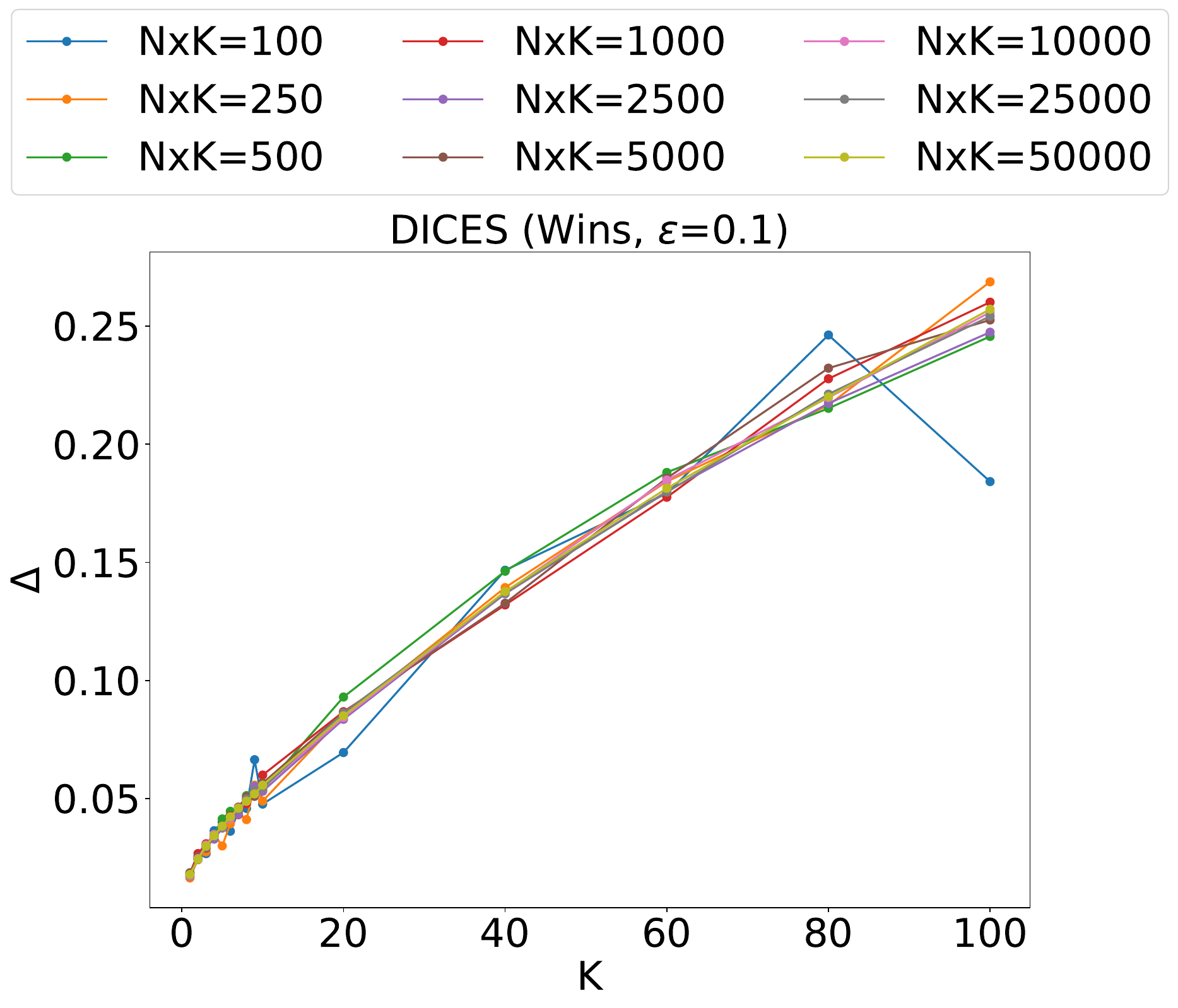}
    \caption{$\epsilon = 0.1$}
    \label{fig:dices_delta_wins_e01}
  \end{subfigure} \hfill
  \begin{subfigure}[b]{0.24\linewidth}
    \centering
    \includegraphics[width=\linewidth]{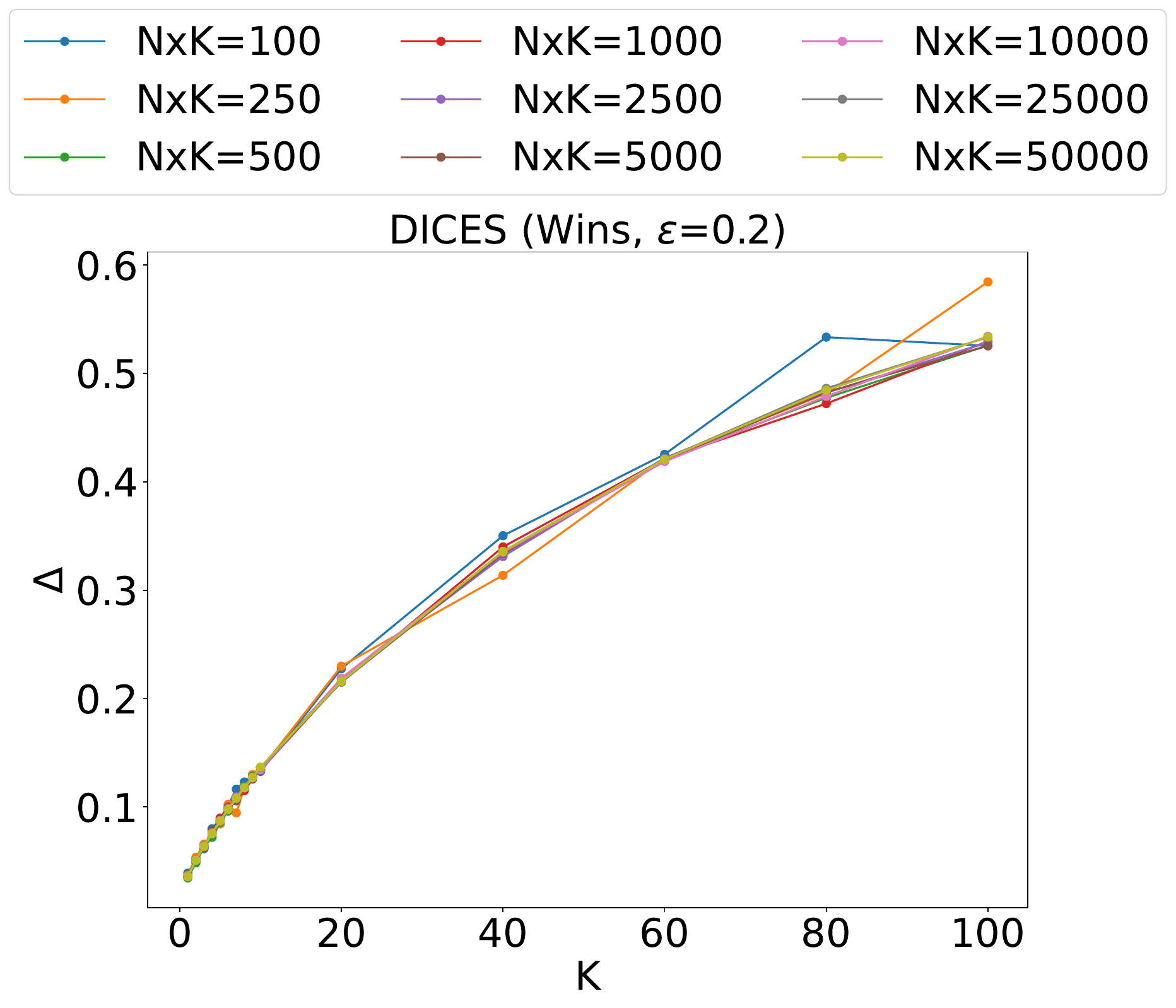}
    \caption{$\epsilon = 0.2$}
    \label{fig:dices_delta_wins_e02}
  \end{subfigure} \hfill
  \begin{subfigure}[b]{0.24\linewidth}
    \centering
    \includegraphics[width=\linewidth]{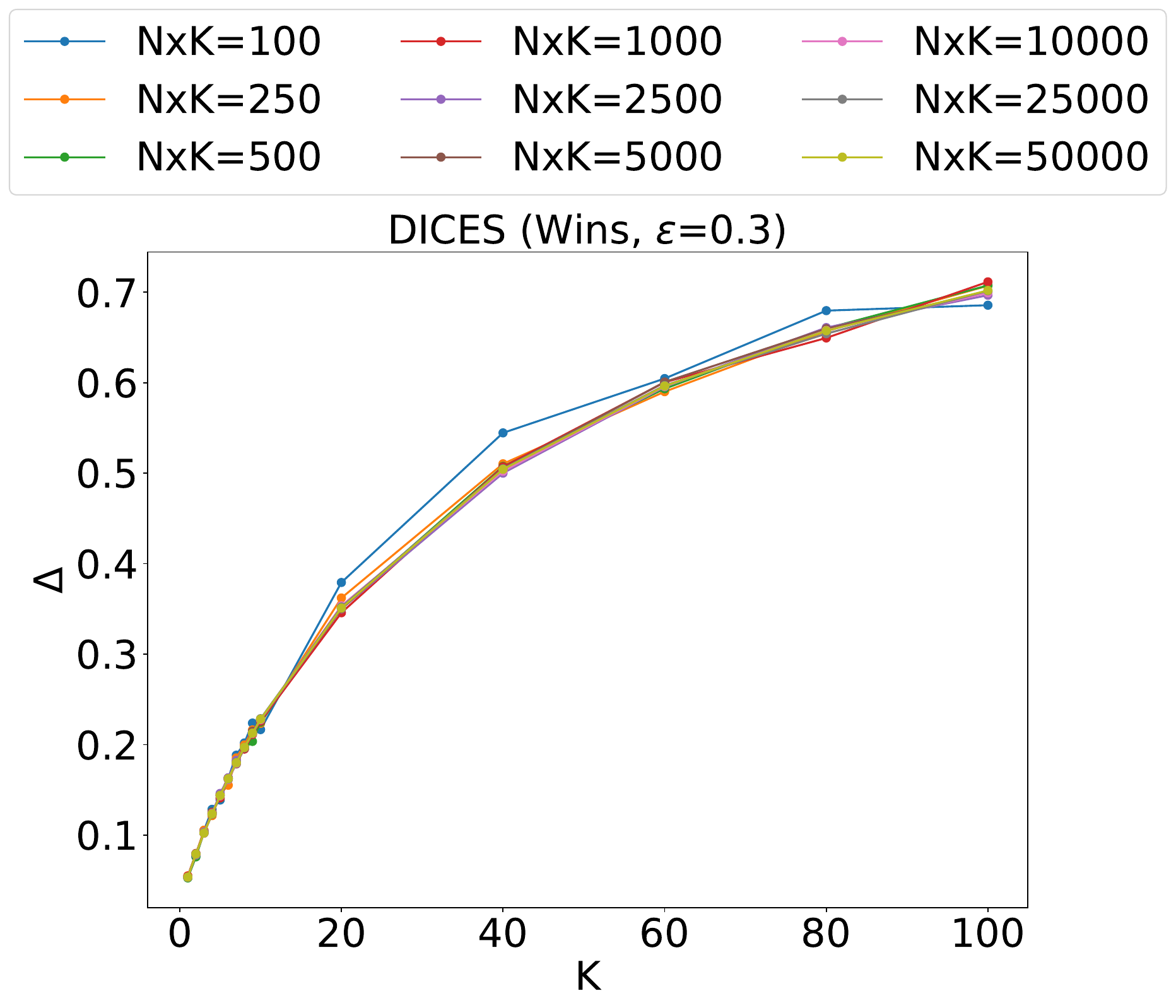}
    \caption{$\epsilon = 0.3$}
    \label{fig:dices_delta_wins_e03}
  \end{subfigure} \hfill
  \begin{subfigure}[b]{0.24\linewidth}
    \centering
    \includegraphics[width=\linewidth]{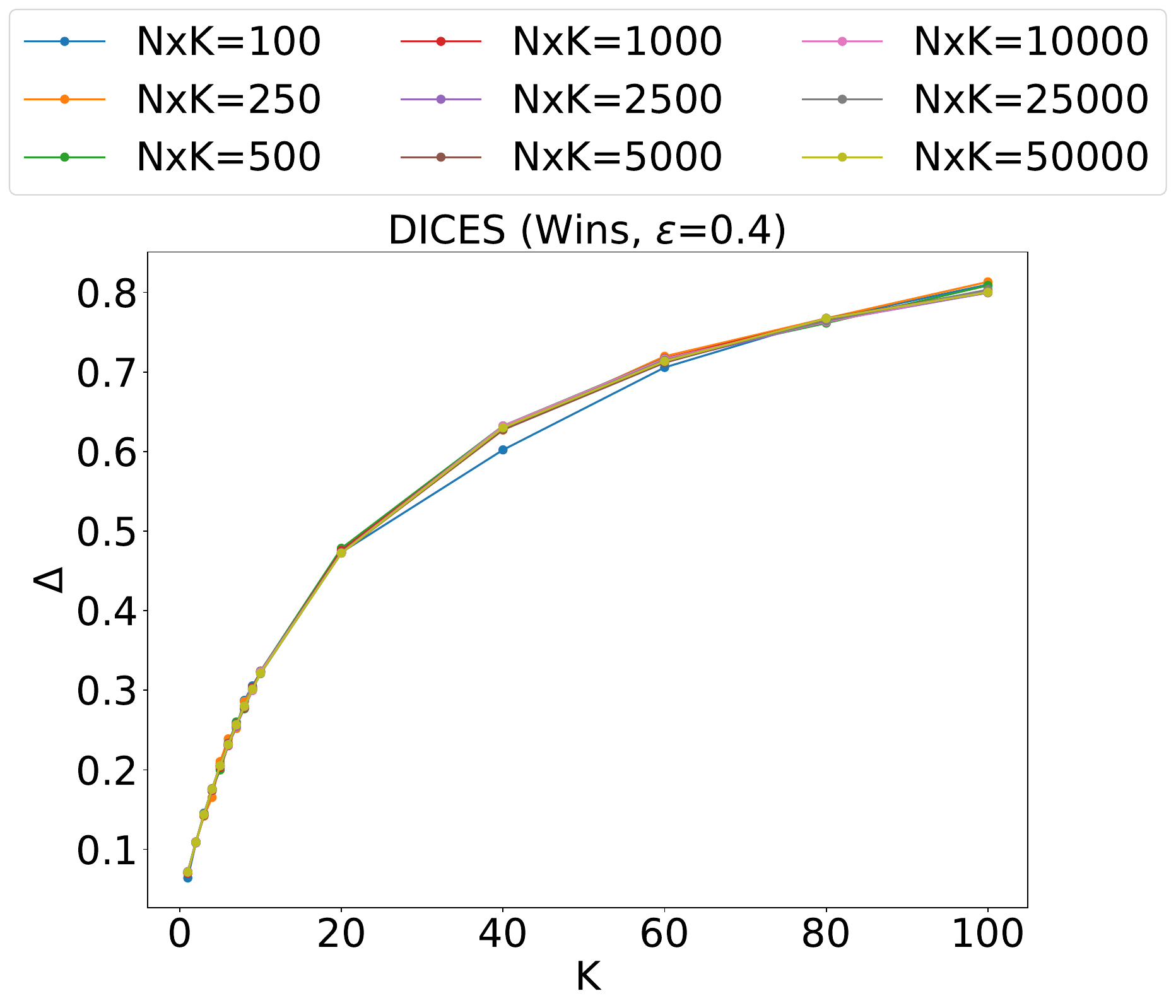}
    \caption{$\epsilon = 0.4$}
    \label{fig:dices_delta_wins_e04}
  \end{subfigure}
  \caption{Effect sizes ($\Delta$) for DICES dataset with Wins as the metric}
  \label{fig:dices_delta_wins}
\end{figure*}

\begin{figure*}
  \centering
  \begin{subfigure}[b]{0.24\linewidth}
    \centering
    \includegraphics[width=\linewidth]{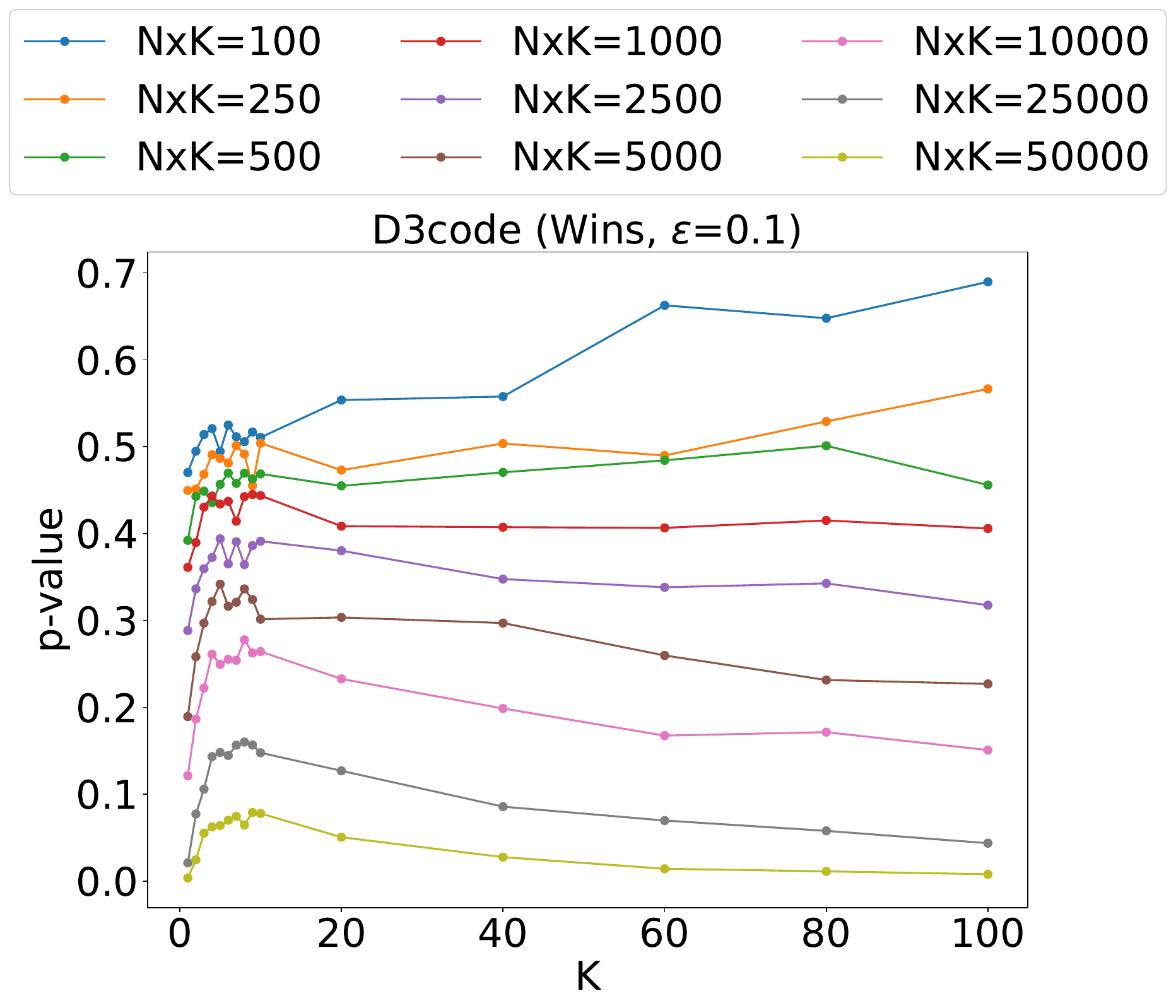}
    \caption{$\epsilon = 0.1$}
    \label{fig:d3code_wins_e01}
  \end{subfigure} \hfill
  \begin{subfigure}[b]{0.24\linewidth}
    \centering
    \includegraphics[width=\linewidth]{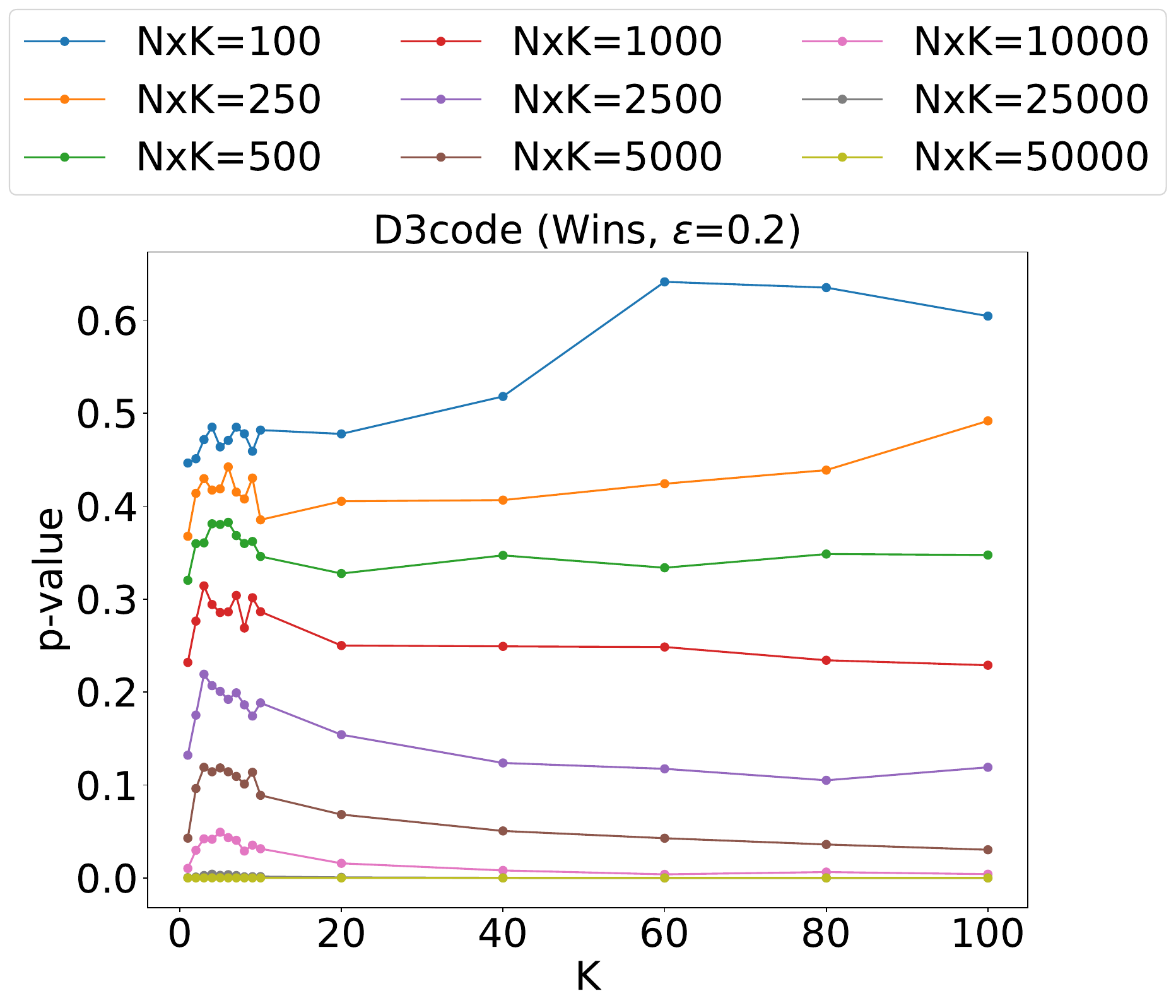}
    \caption{$\epsilon = 0.2$}
    \label{fig:d3code_wins_e02}
  \end{subfigure} \hfill
  \begin{subfigure}[b]{0.24\linewidth}
    \centering
    \includegraphics[width=\linewidth]{figures/K100/pvals_plots/D3code/D3code_p_vals_Wins_K_100_e_0.3.pdf}
    \caption{$\epsilon = 0.3$}
    \label{fig:d3code_wins_e03}
  \end{subfigure} \hfill
  \begin{subfigure}[b]{0.24\linewidth}
    \centering
    \includegraphics[width=\linewidth]{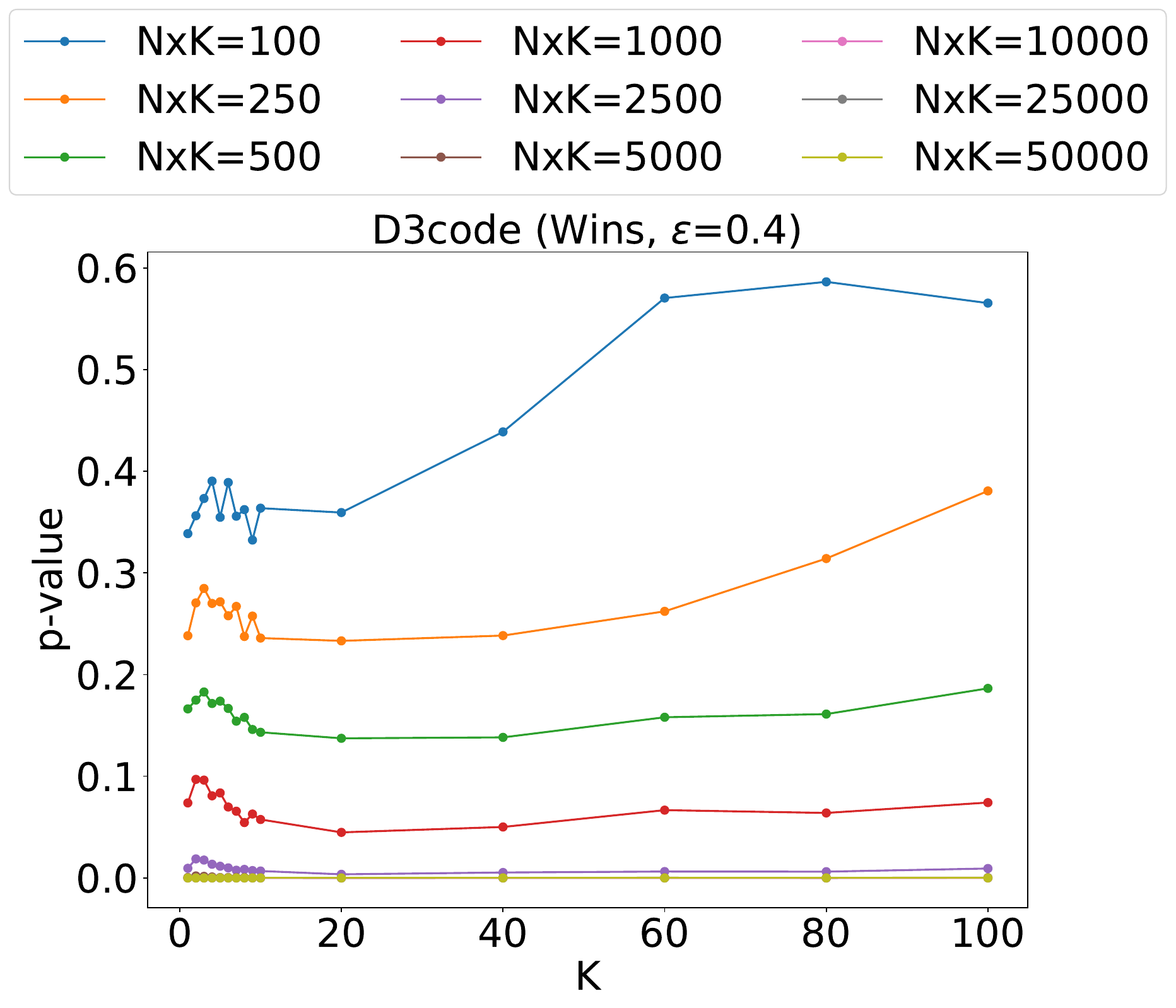}
    \caption{$\epsilon = 0.4$}
    \label{fig:d3code_wins_e04}
  \end{subfigure}
  \caption{P-value plots for D3code dataset with Wins as the metric}
  \label{fig:d3code_wins}
\end{figure*}

\begin{figure*}
  \centering
  \begin{subfigure}[b]{0.24\linewidth}
    \centering
    \includegraphics[width=\linewidth]{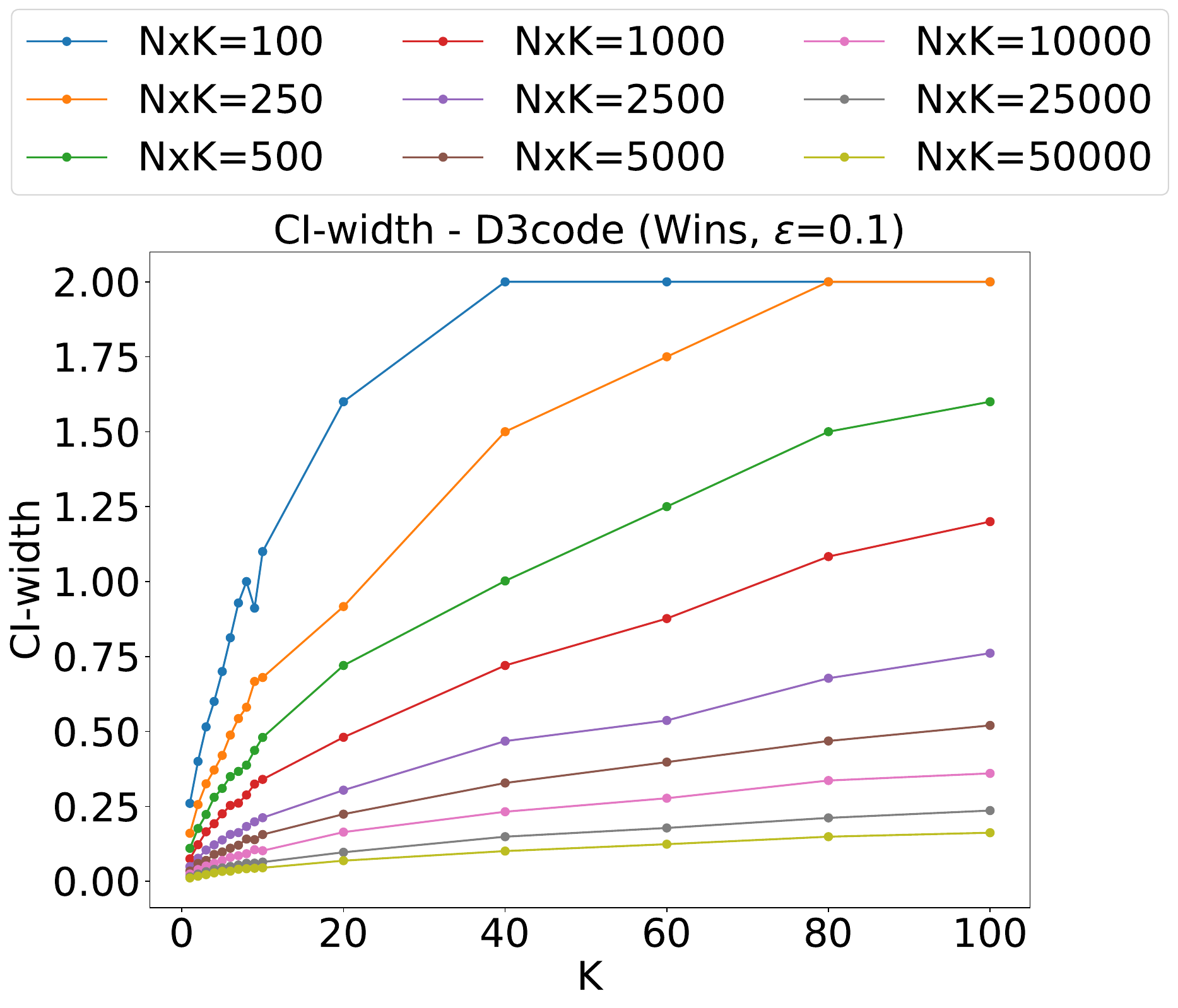}
    \caption{$\epsilon = 0.1$}
    \label{fig:d3code_ci_wins_e01}
  \end{subfigure} \hfill
  \begin{subfigure}[b]{0.24\linewidth}
    \centering
    \includegraphics[width=\linewidth]{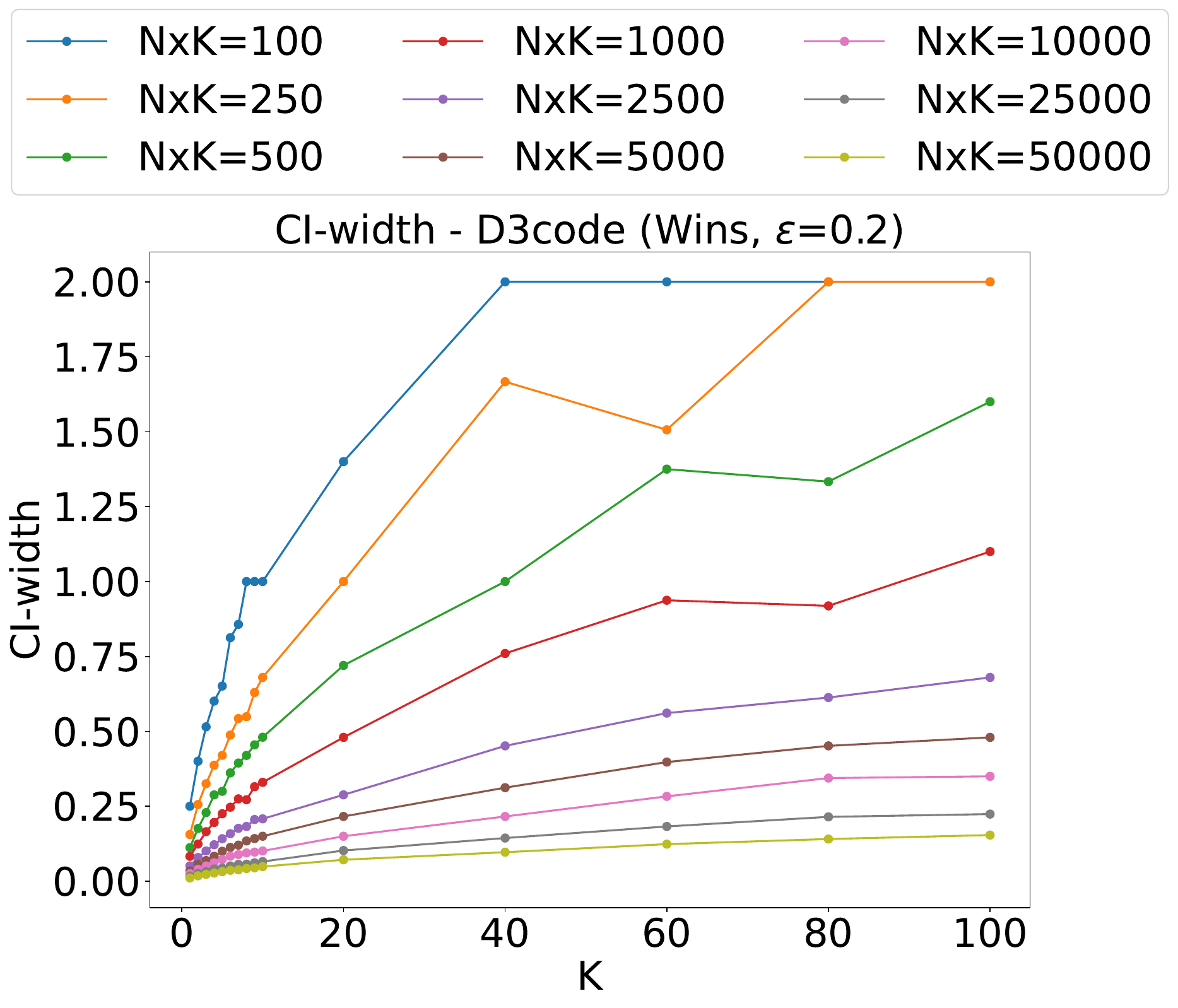}
    \caption{$\epsilon = 0.2$}
    \label{fig:d3code_ci_wins_e02}
  \end{subfigure} \hfill
  \begin{subfigure}[b]{0.24\linewidth}
    \centering
    \includegraphics[width=\linewidth]{figures/K100/ci_plots/D3code/D3code_CI_width_Wins_K_100_e_0.3.pdf}
    \caption{$\epsilon = 0.3$}
    \label{fig:d3code_ci_wins_e03}
  \end{subfigure} \hfill
  \begin{subfigure}[b]{0.24\linewidth}
    \centering
    \includegraphics[width=\linewidth]{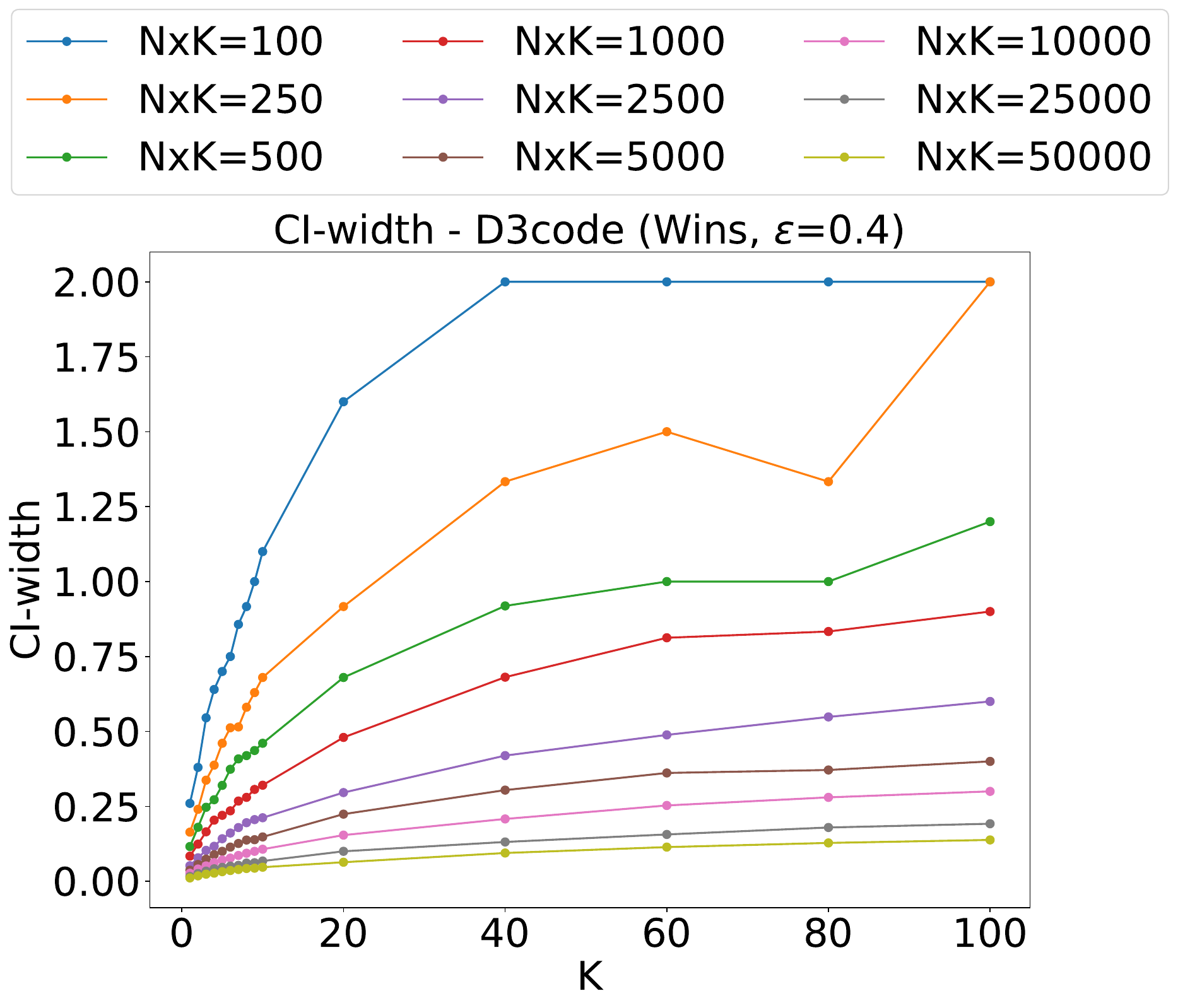}
    \caption{$\epsilon = 0.4$}
    \label{fig:d3code_ci_wins_e04}
  \end{subfigure}
  \caption{CI-width plots for D3code dataset with Wins as the metric}
  \label{fig:d3code_ci_wins}
\end{figure*}

\begin{figure*}
  \centering
  \begin{subfigure}[b]{0.24\linewidth}
    \centering
    \includegraphics[width=\linewidth]{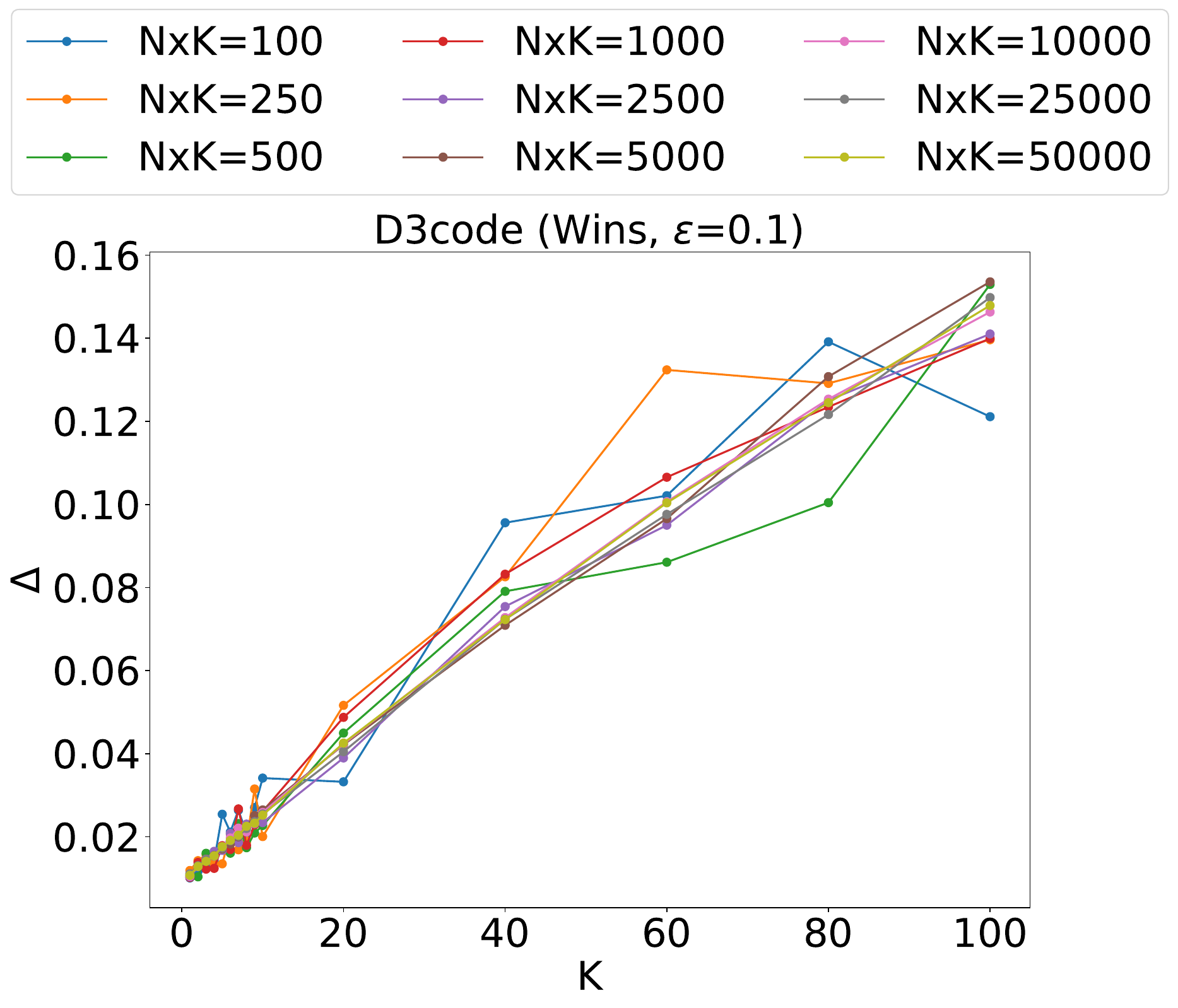}
    \caption{$\epsilon = 0.1$}
    \label{fig:d3code_delta_wins_e01}
  \end{subfigure} \hfill
  \begin{subfigure}[b]{0.24\linewidth}
    \centering
    \includegraphics[width=\linewidth]{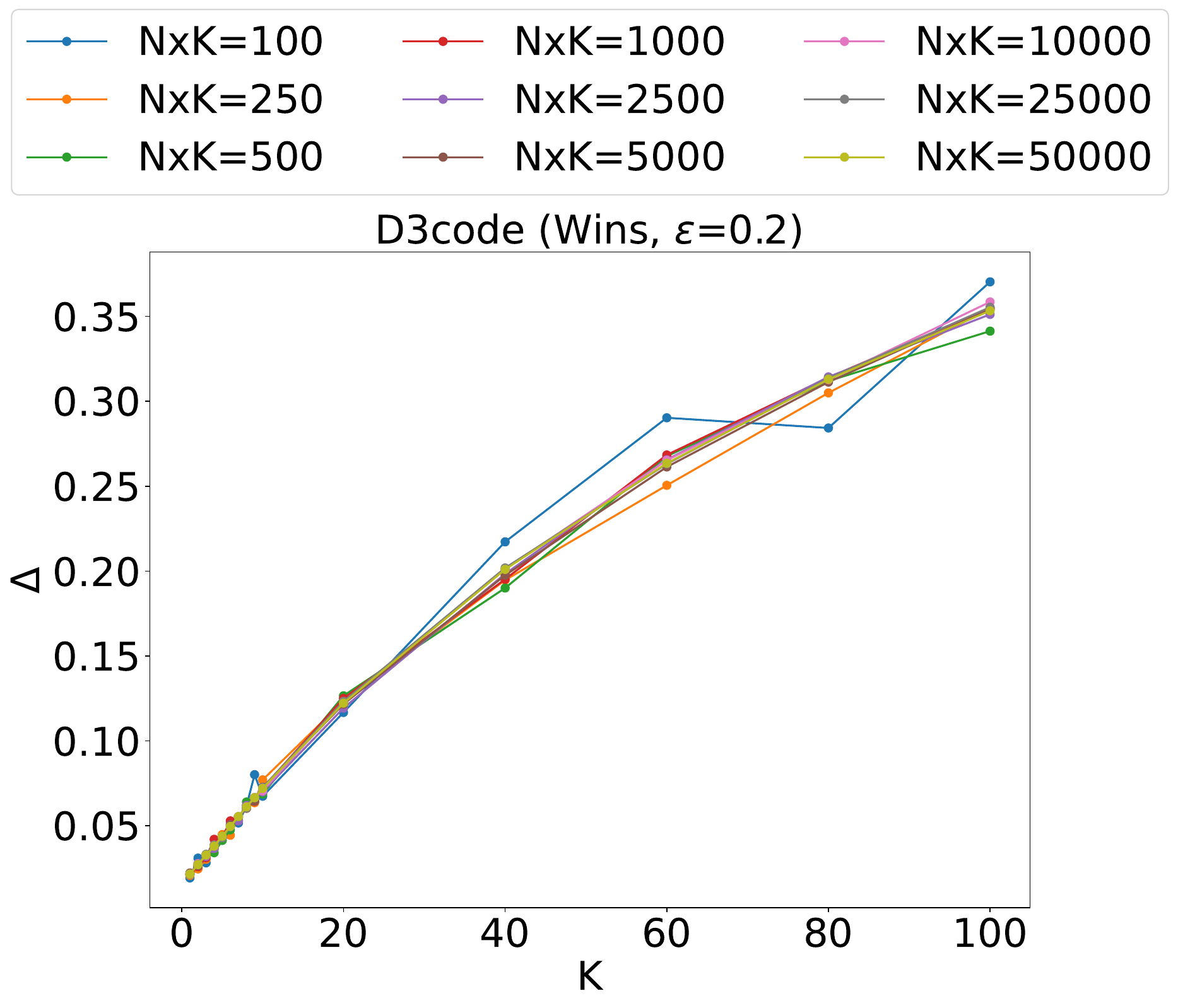}
    \caption{$\epsilon = 0.2$}
    \label{fig:d3code_delta_wins_e02}
  \end{subfigure} \hfill
  \begin{subfigure}[b]{0.24\linewidth}
    \centering
    \includegraphics[width=\linewidth]{figures/K100/delta_plots/D3code/D3code_delta_Wins_K_100_e_0.3.pdf}
    \caption{$\epsilon = 0.3$}
    \label{fig:d3code_delta_wins_e03}
  \end{subfigure} \hfill
  \begin{subfigure}[b]{0.24\linewidth}
    \centering
    \includegraphics[width=\linewidth]{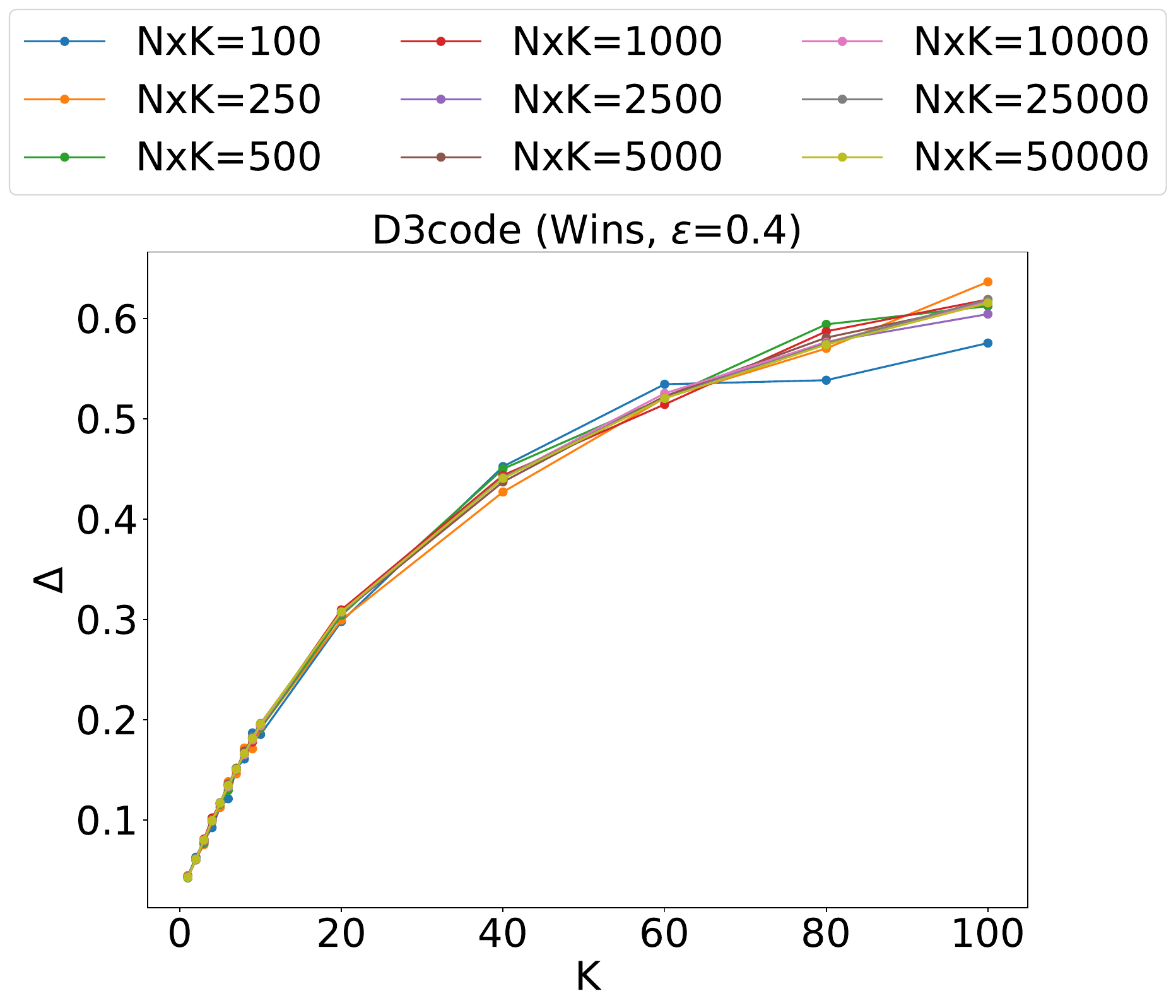}
    \caption{$\epsilon = 0.4$}
    \label{fig:d3code_delta_wins_e04}
  \end{subfigure}
  \caption{Effect sizes ($\Delta$) for D3code dataset with Wins as the metric}
  \label{fig:d3code_delta_wins}
\end{figure*}

\begin{figure*}
  \centering
  \begin{subfigure}[b]{0.24\linewidth}
    \centering
    \includegraphics[width=\linewidth]{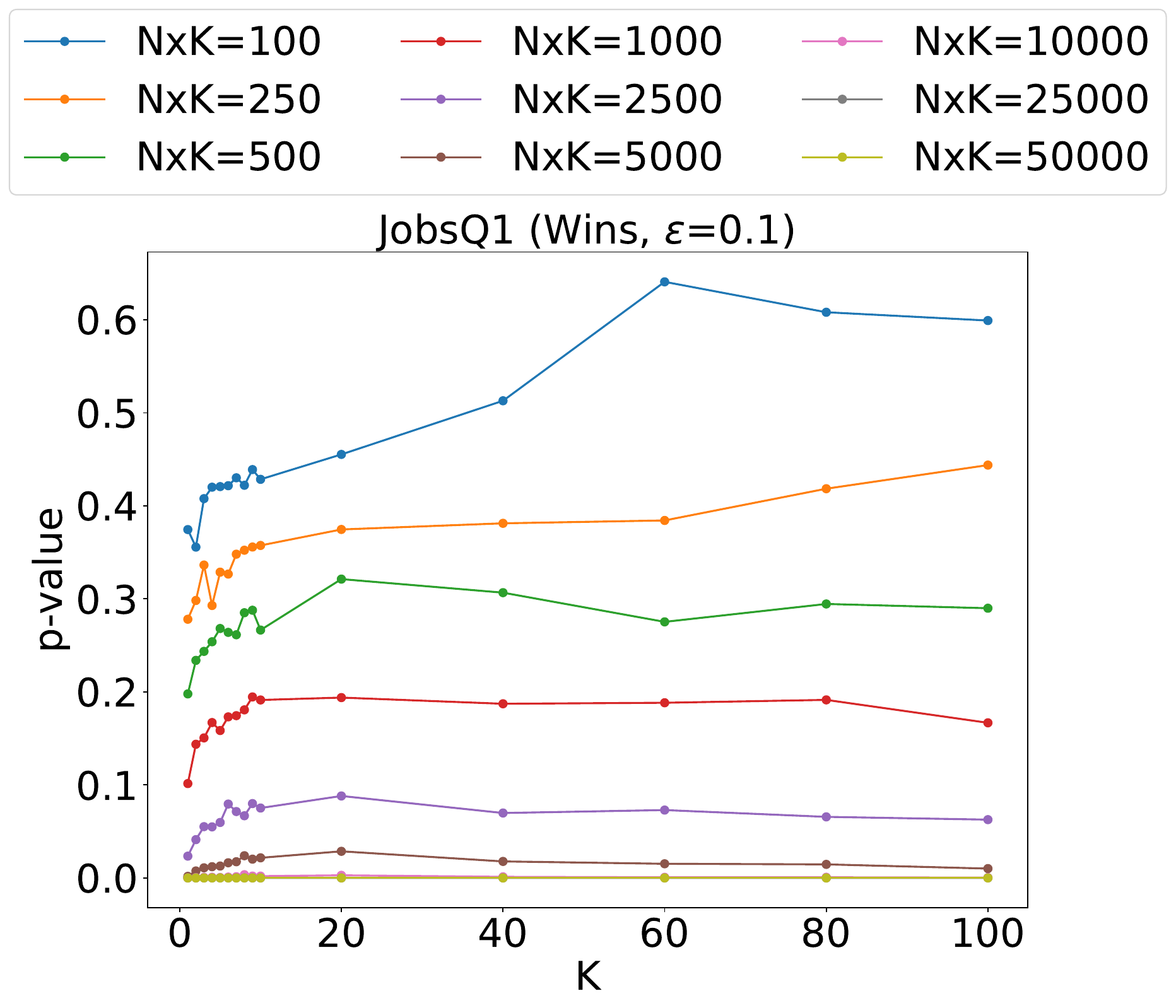}
    \caption{$\epsilon = 0.1$}
    \label{fig:jobsQ1_wins_e01}
  \end{subfigure} \hfill
  \begin{subfigure}[b]{0.24\linewidth}
    \centering
    \includegraphics[width=\linewidth]{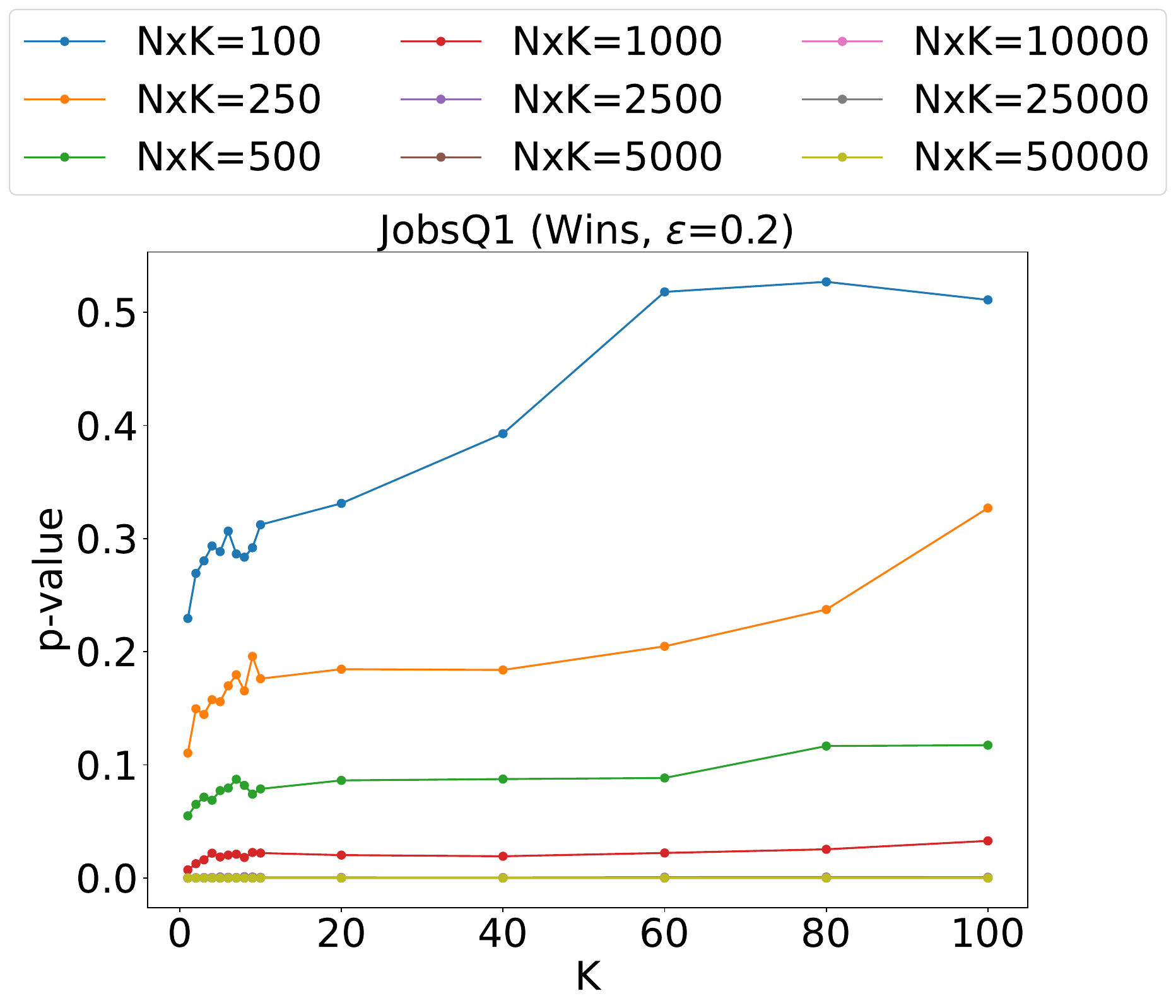}
    \caption{$\epsilon = 0.2$}
    \label{fig:jobsQ1_wins_e02}
  \end{subfigure} \hfill
  \begin{subfigure}[b]{0.24\linewidth}
    \centering
    \includegraphics[width=\linewidth]{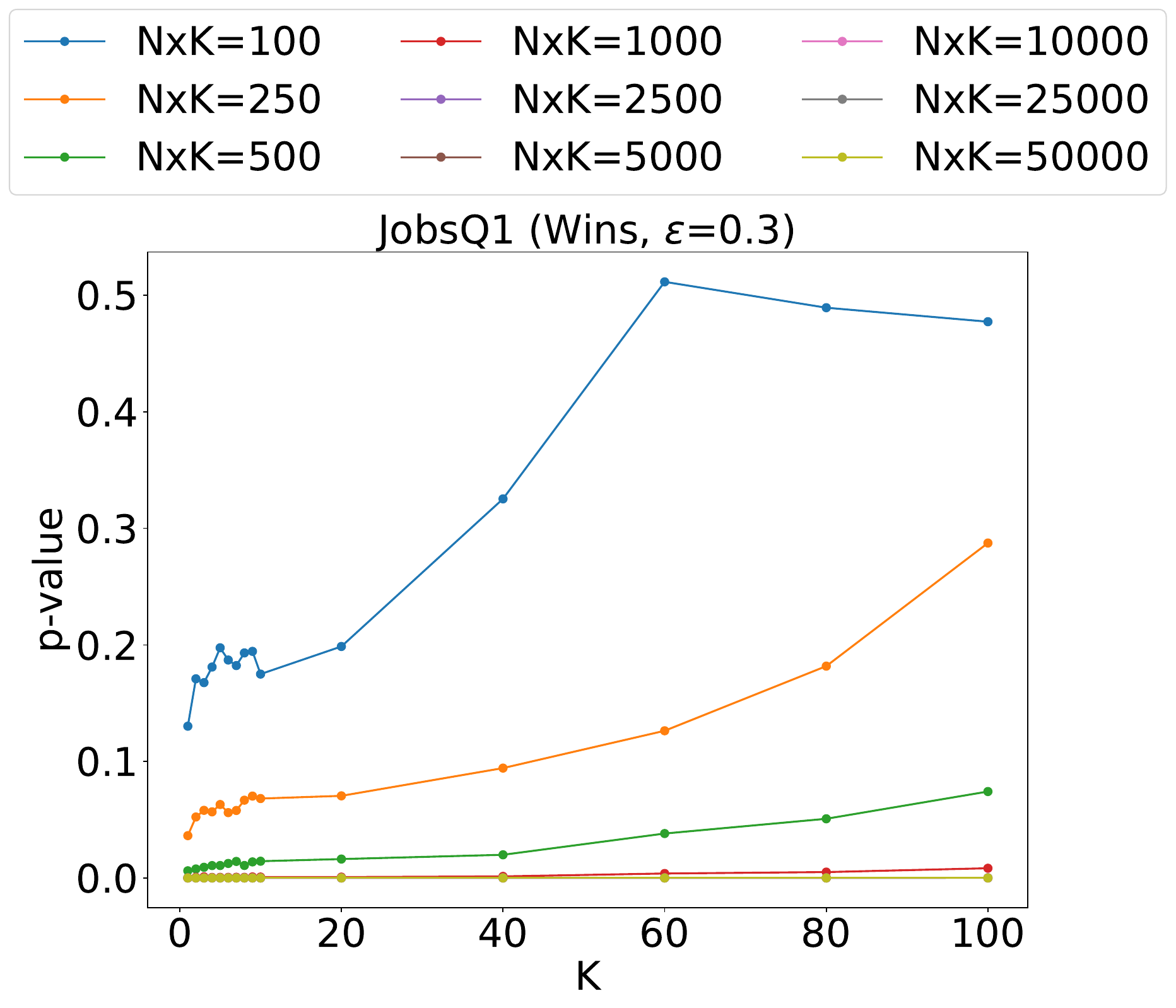}
    \caption{$\epsilon = 0.3$}
    \label{fig:jobsQ1_wins_e03}
  \end{subfigure} \hfill
  \begin{subfigure}[b]{0.24\linewidth}
    \centering
    \includegraphics[width=\linewidth]{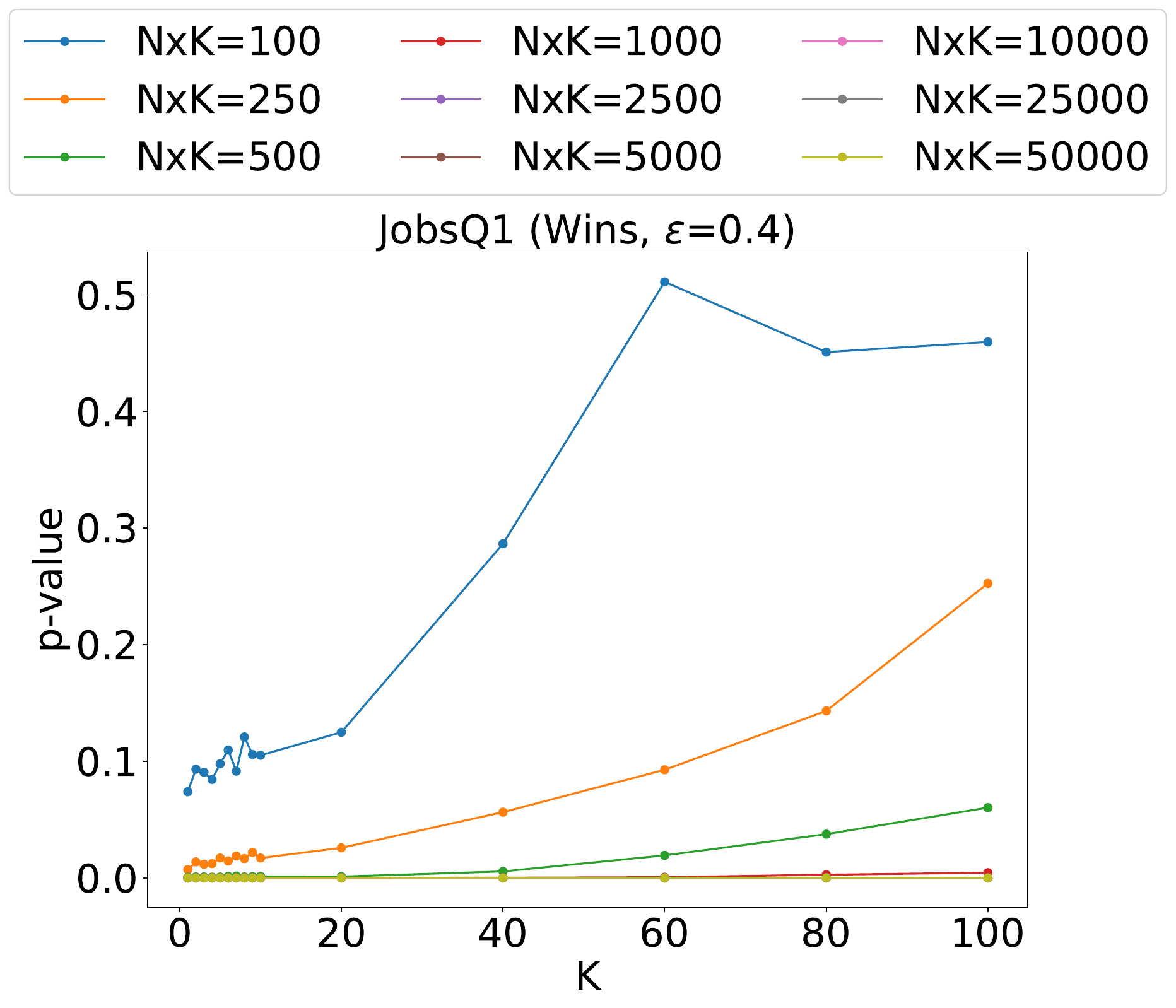}
    \caption{$\epsilon = 0.4$}
    \label{fig:jobsQ1_wins_e04}
  \end{subfigure}
  \caption{P-value plots for JobsQ1 dataset with Wins as the metric}
  \label{fig:jobsQ1_wins}
\end{figure*}

\begin{figure*}
  \centering
  \begin{subfigure}[b]{0.24\linewidth}
    \centering
    \includegraphics[width=\linewidth]{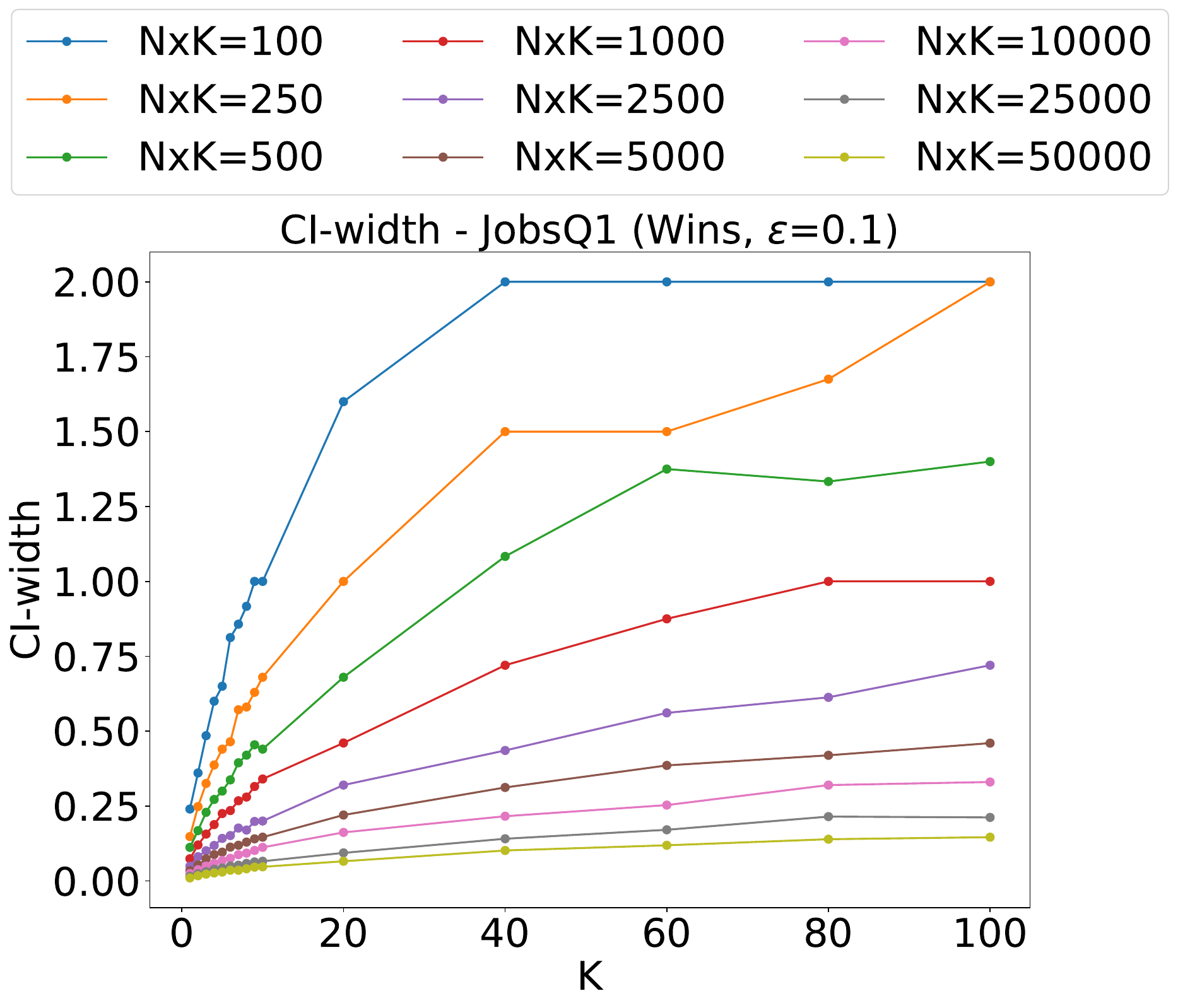}
    \caption{$\epsilon = 0.1$}
    \label{fig:jobsQ1_ci_wins_e01}
  \end{subfigure} \hfill
  \begin{subfigure}[b]{0.24\linewidth}
    \centering
    \includegraphics[width=\linewidth]{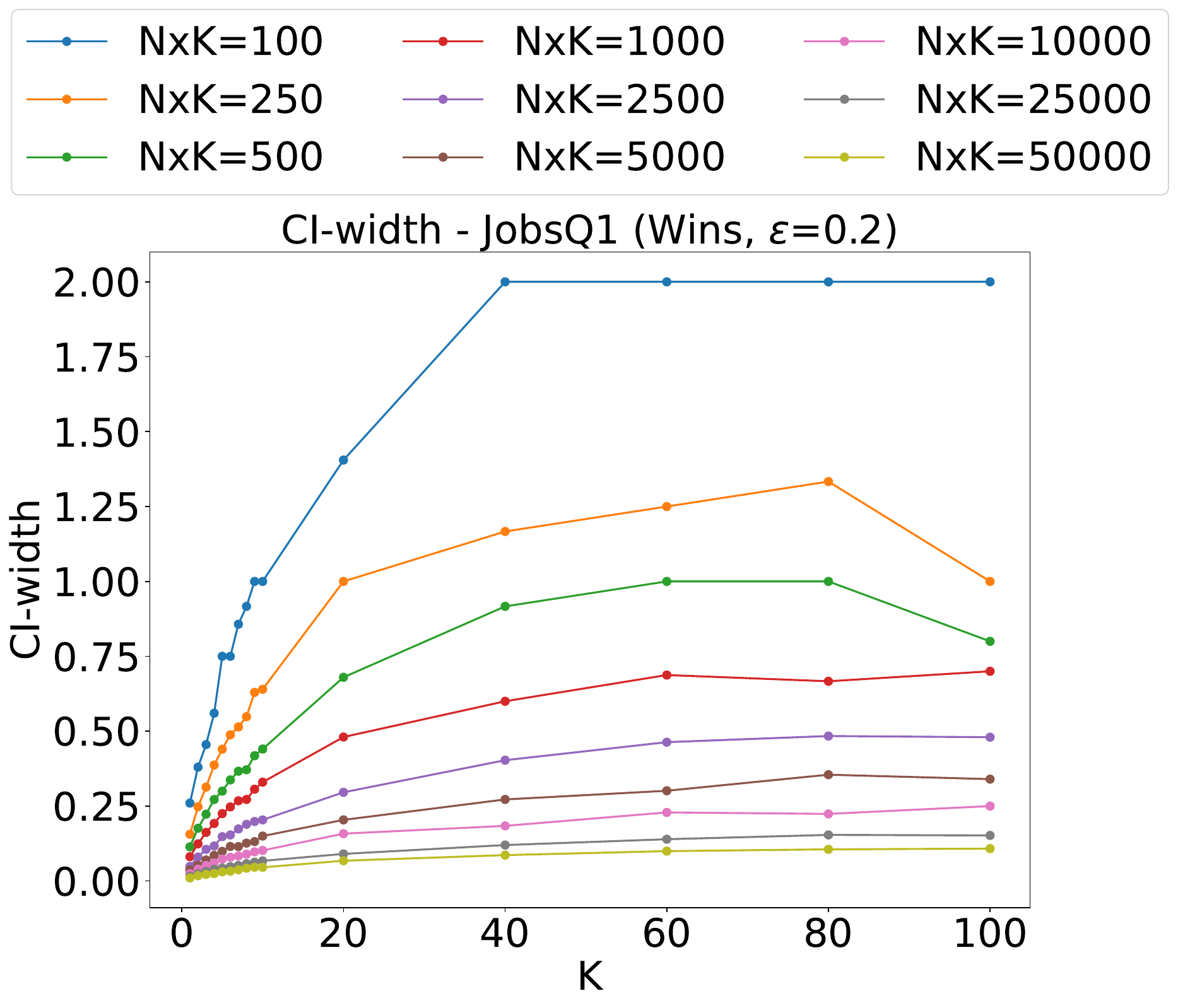}
    \caption{$\epsilon = 0.2$}
    \label{fig:jobsQ1_ci_wins_e02}
  \end{subfigure} \hfill
  \begin{subfigure}[b]{0.24\linewidth}
    \centering
    \includegraphics[width=\linewidth]{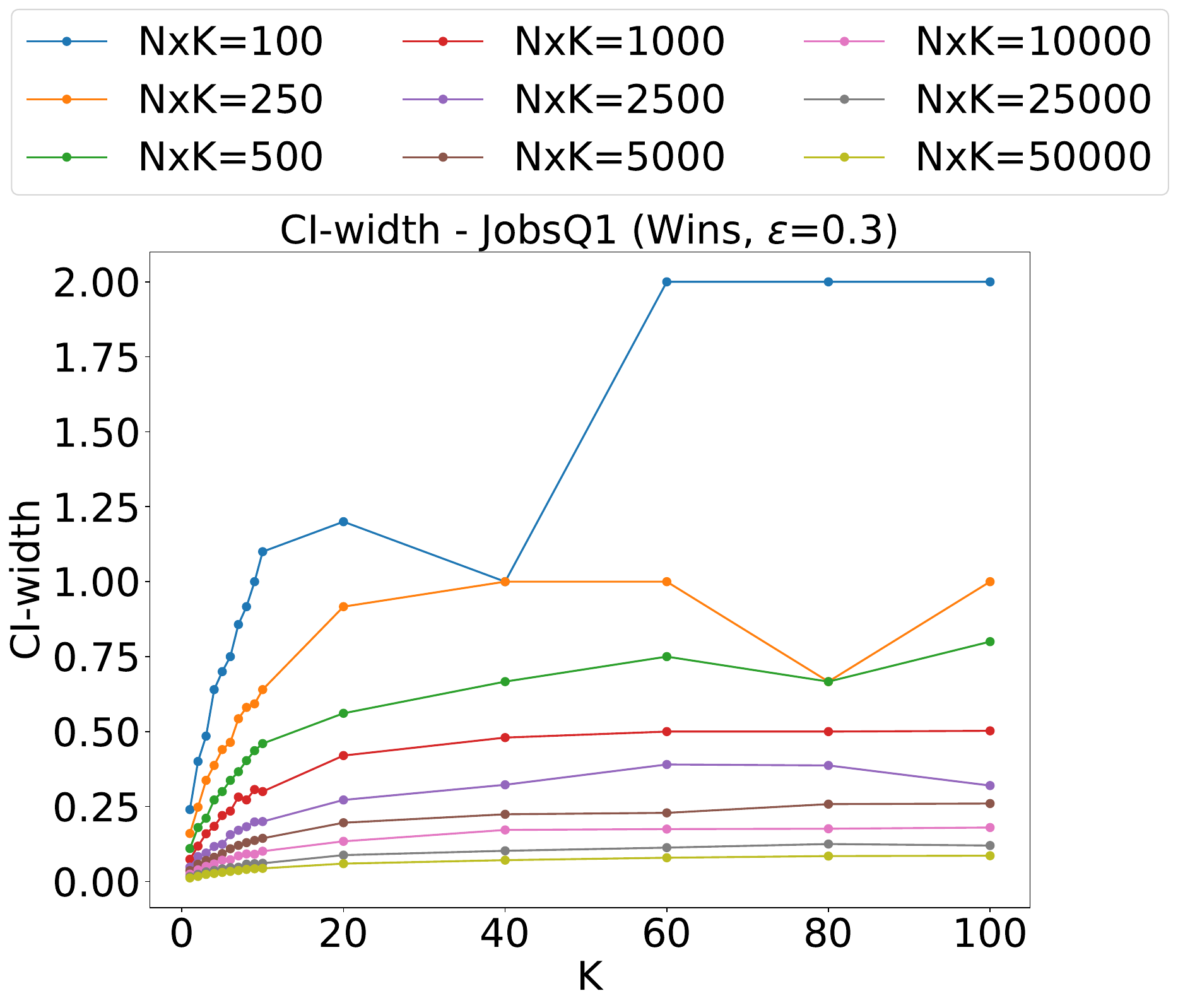}
    \caption{$\epsilon = 0.3$}
    \label{fig:jobsQ1_ci_wins_e03}
  \end{subfigure} \hfill
  \begin{subfigure}[b]{0.24\linewidth}
    \centering
    \includegraphics[width=\linewidth]{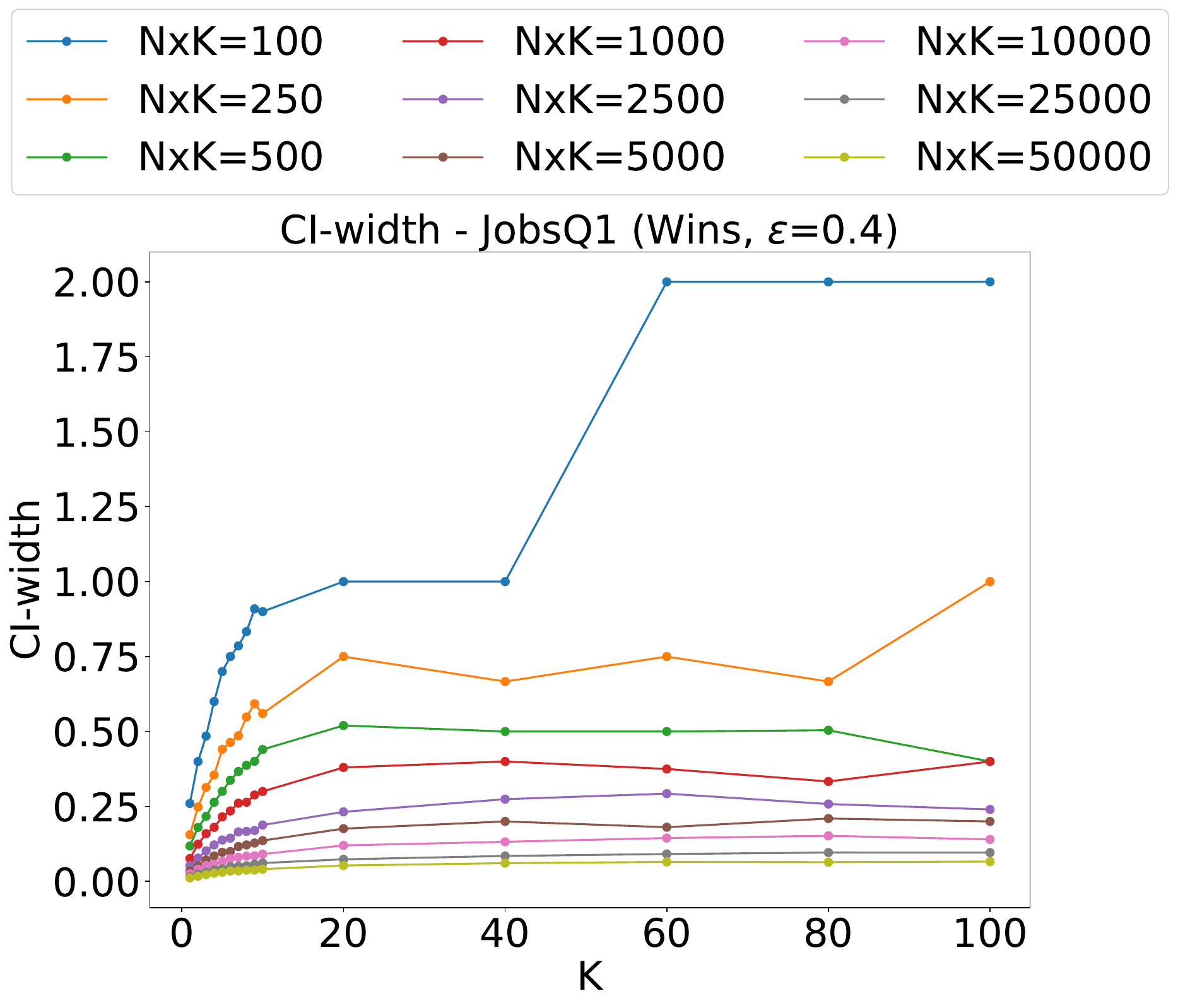}
    \caption{$\epsilon = 0.4$}
    \label{fig:jobsQ1_ci_wins_e04}
  \end{subfigure}
  \caption{CI-width plots for JobsQ1 dataset with Wins as the metric}
  \label{fig:jobsQ1_ci_wins}
\end{figure*}

\begin{figure*}
  \centering
  \begin{subfigure}[b]{0.24\linewidth}
    \centering
    \includegraphics[width=\linewidth]{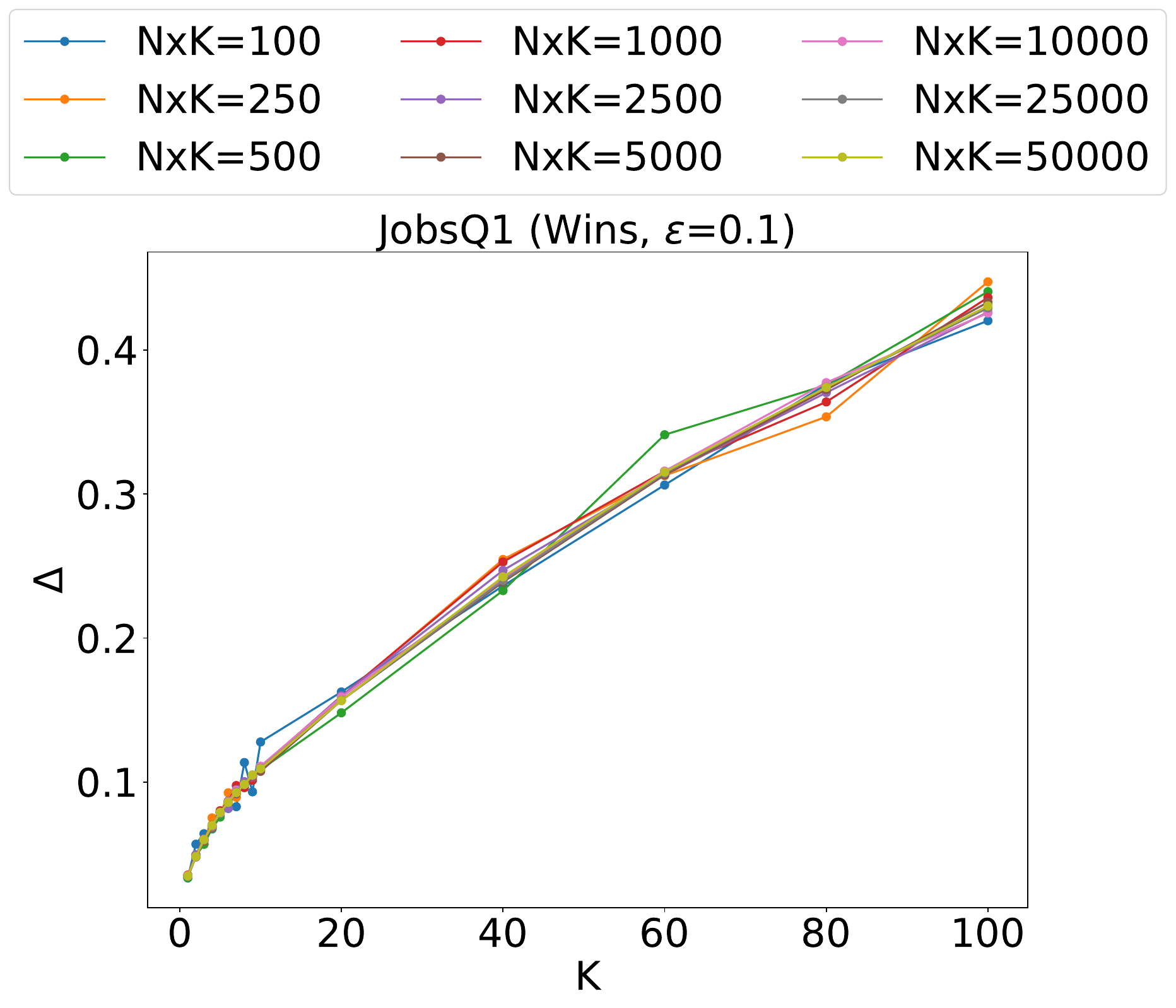}
    \caption{$\epsilon = 0.1$}
    \label{fig:jobsQ1_delta_wins_e01}
  \end{subfigure} \hfill
  \begin{subfigure}[b]{0.24\linewidth}
    \centering
    \includegraphics[width=\linewidth]{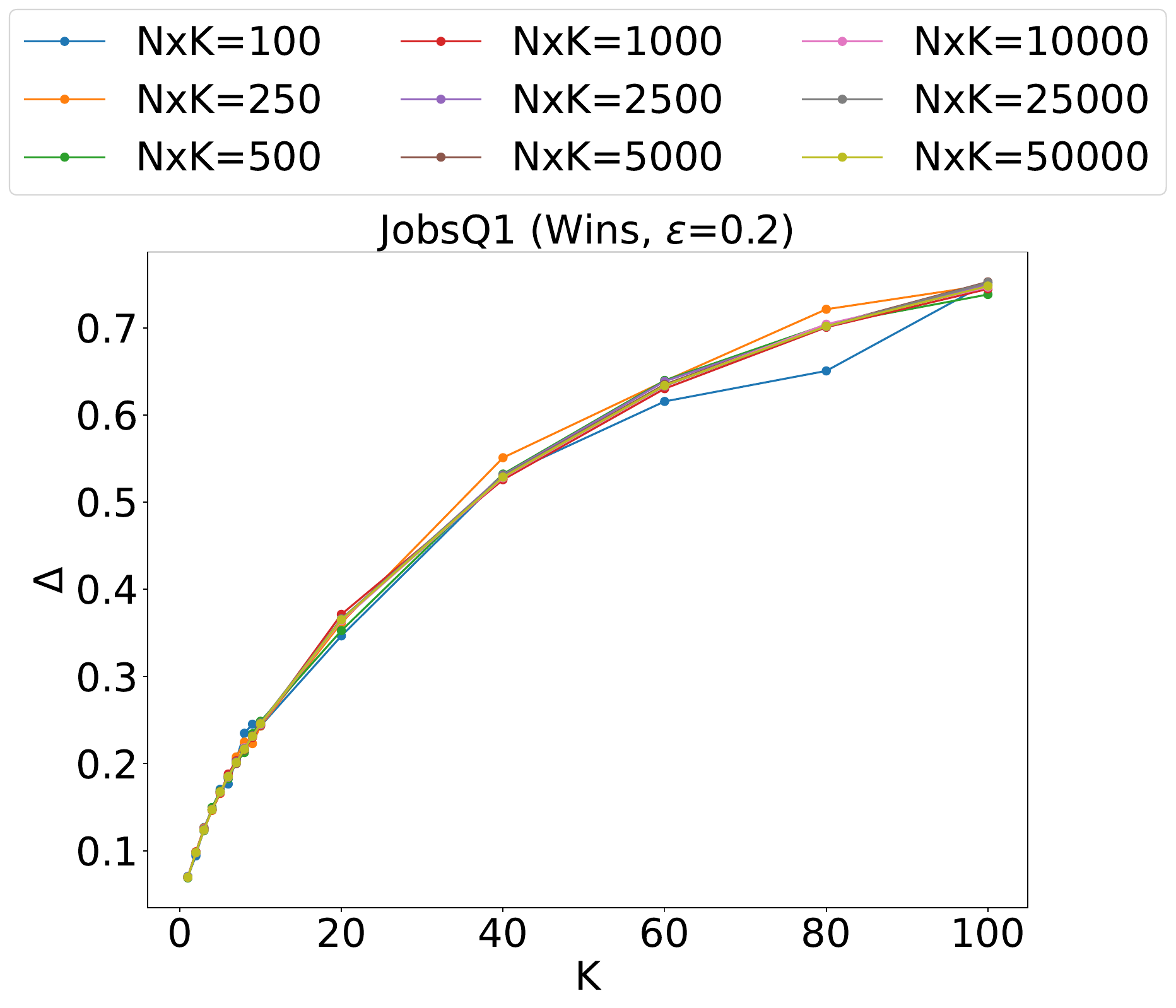}
    \caption{$\epsilon = 0.2$}
    \label{fig:jobsQ1_delta_wins_e02}
  \end{subfigure} \hfill
  \begin{subfigure}[b]{0.24\linewidth}
    \centering
    \includegraphics[width=\linewidth]{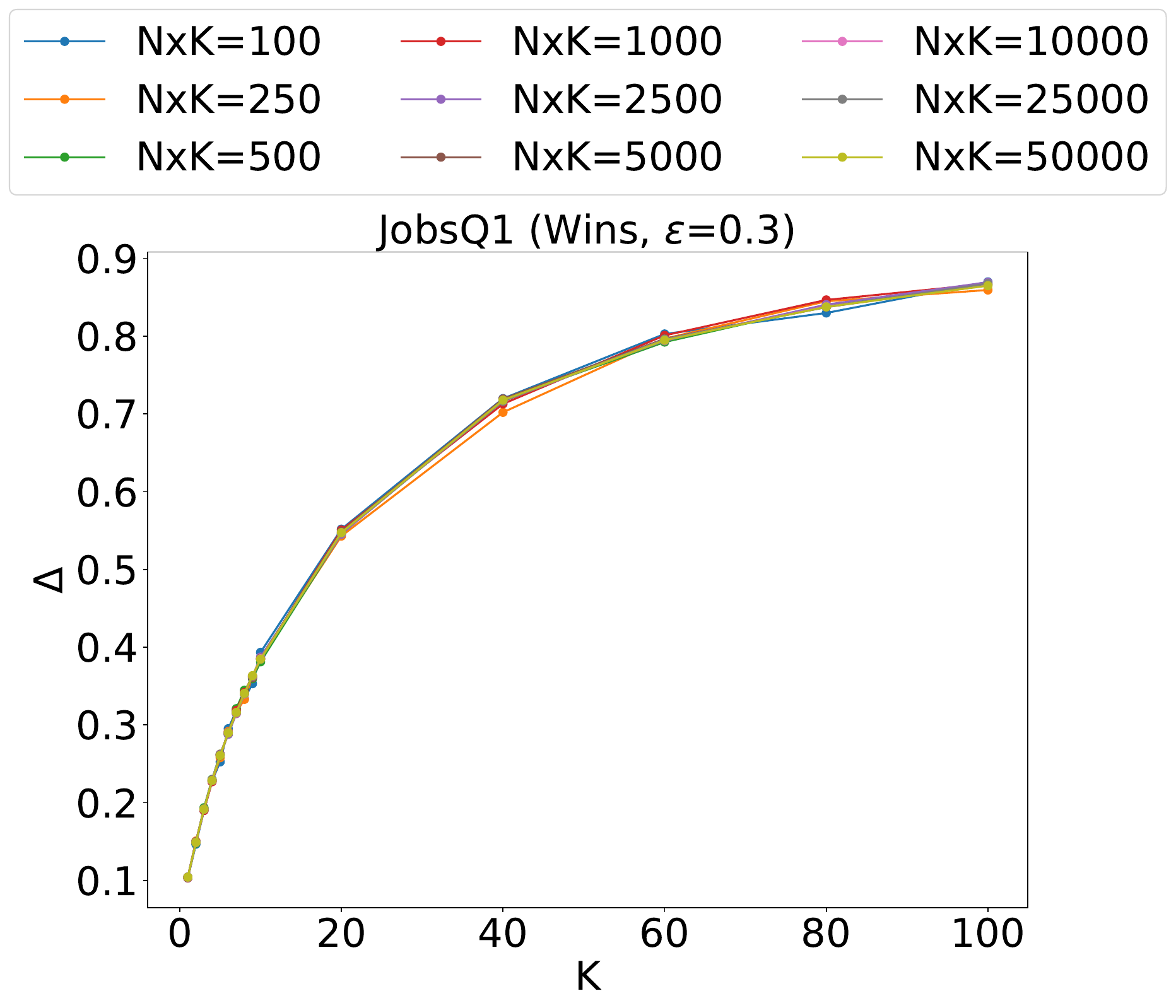}
    \caption{$\epsilon = 0.3$}
    \label{fig:jobsQ1_delta_wins_e03}
  \end{subfigure} \hfill
  \begin{subfigure}[b]{0.24\linewidth}
    \centering
    \includegraphics[width=\linewidth]{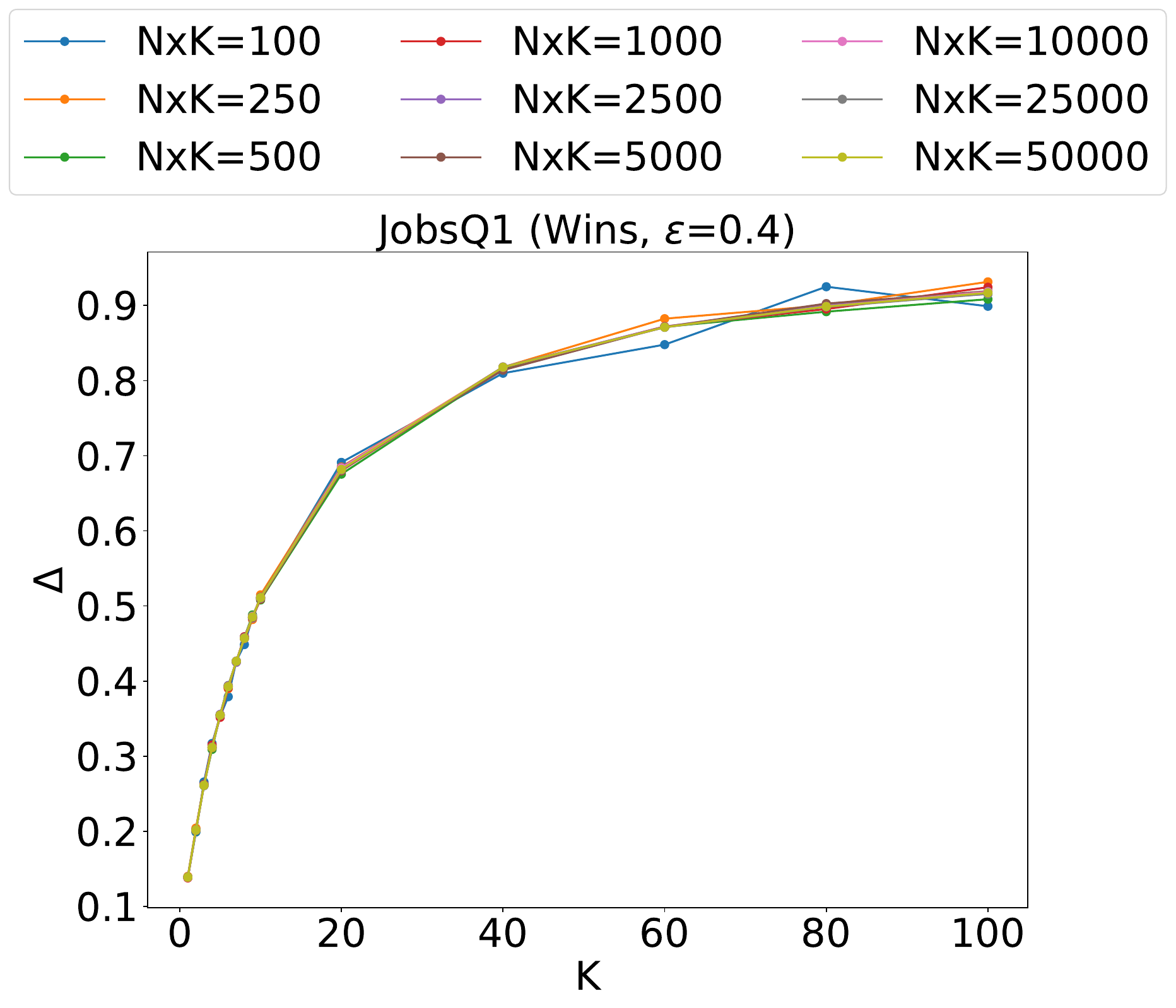}
    \caption{$\epsilon = 0.4$}
    \label{fig:jobsQ1_delta_wins_e04}
  \end{subfigure}
  \caption{Effect sizes ($\Delta$) for JobsQ1 dataset with Wins as the metric}
  \label{fig:jobsQ1_delta_wins}
\end{figure*}

\begin{figure*}
  \centering
  \begin{subfigure}[b]{0.24\linewidth}
    \centering
    \includegraphics[width=\linewidth]{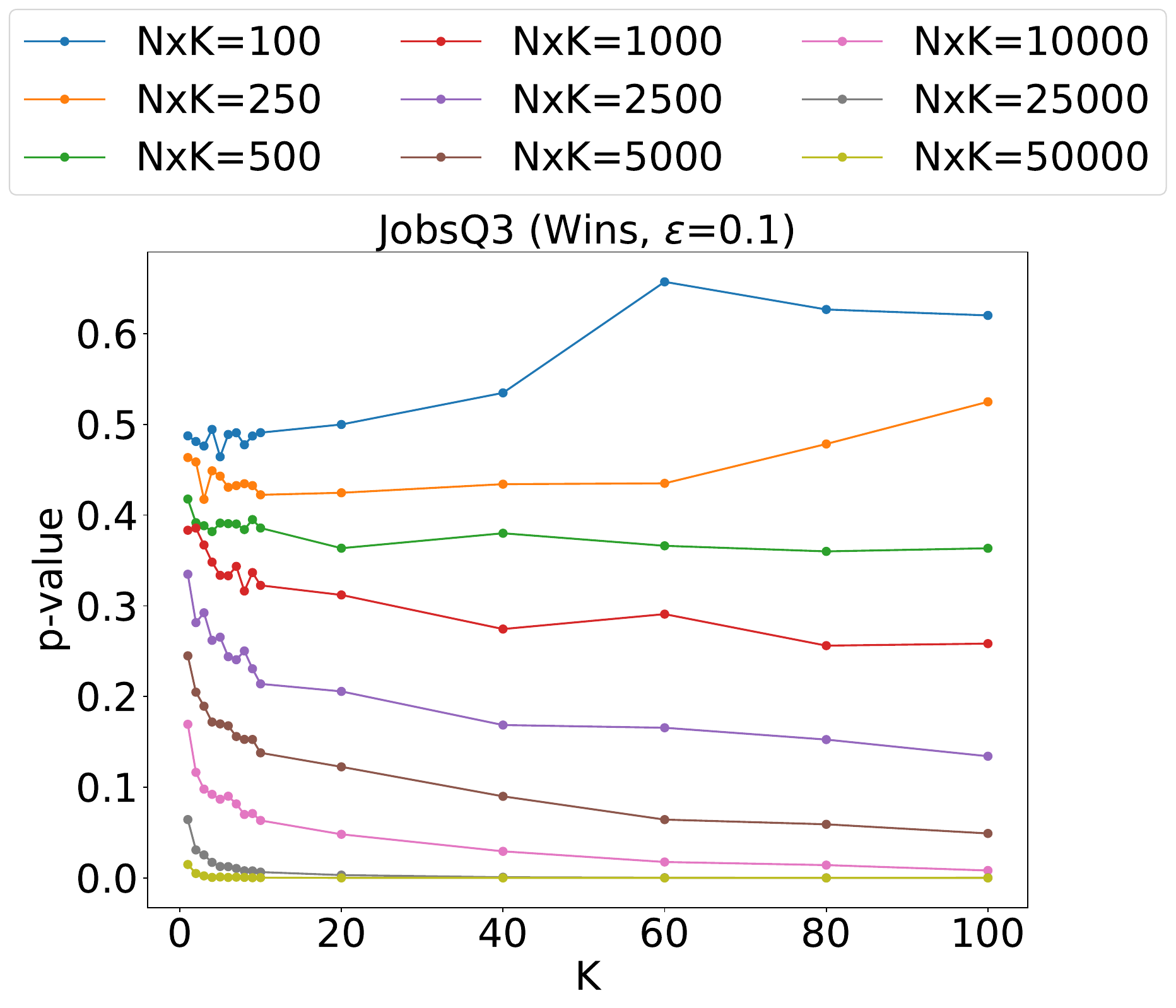}
    \caption{$\epsilon = 0.1$}
    \label{fig:jobsQ3_wins_e01}
  \end{subfigure} \hfill
  \begin{subfigure}[b]{0.24\linewidth}
    \centering
    \includegraphics[width=\linewidth]{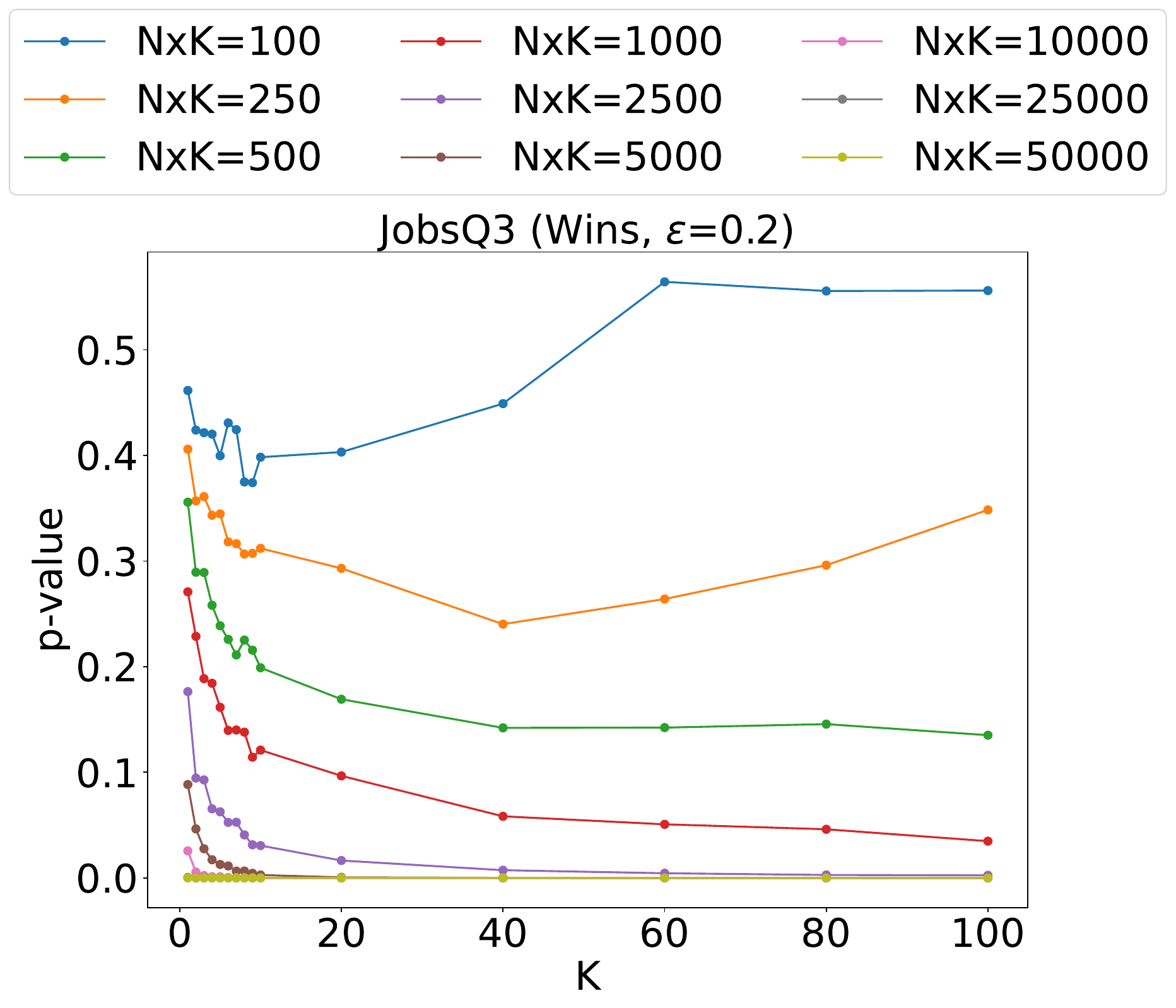}
    \caption{$\epsilon = 0.2$}
    \label{fig:jobsQ3_wins_e02}
  \end{subfigure} \hfill
  \begin{subfigure}[b]{0.24\linewidth}
    \centering
    \includegraphics[width=\linewidth]{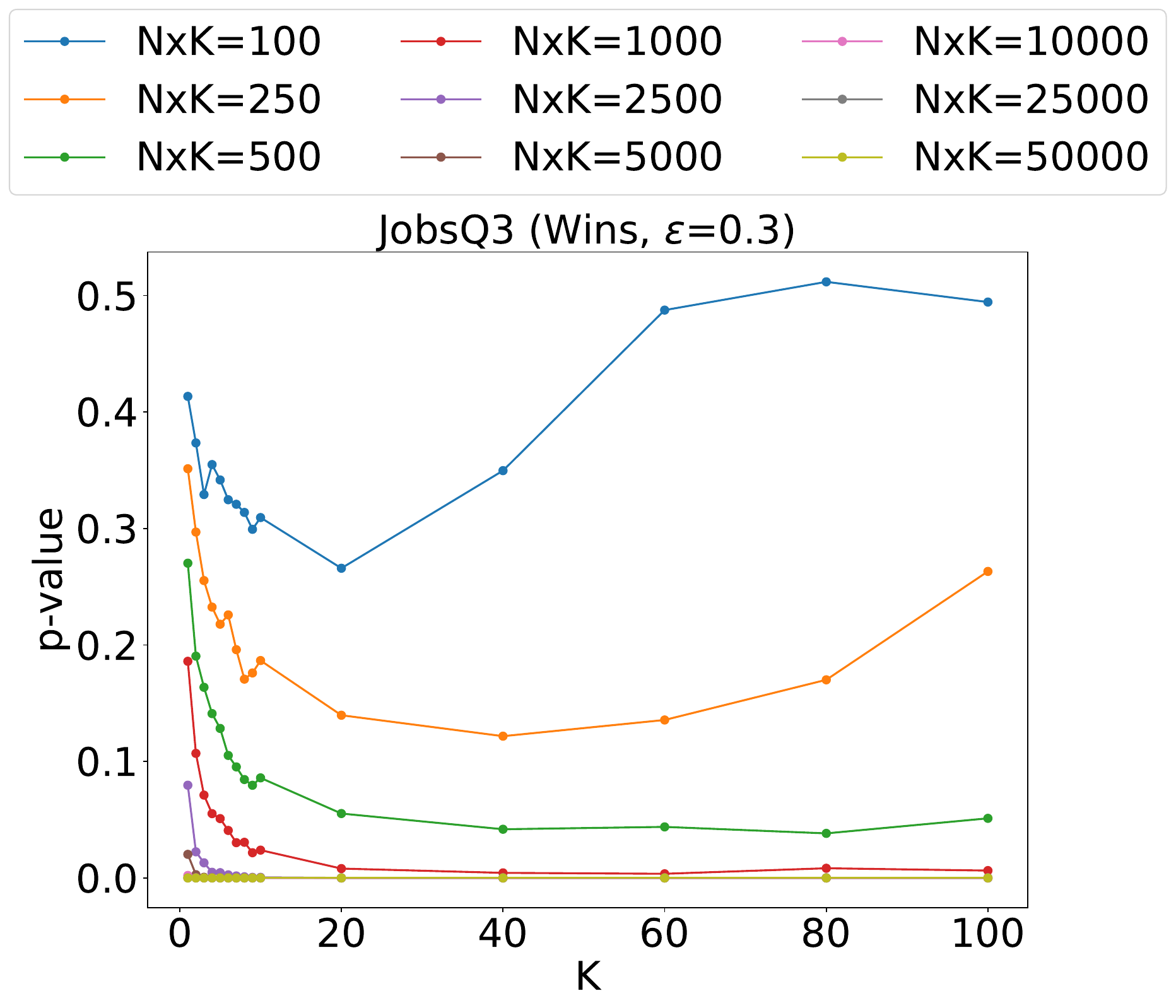}
    \caption{$\epsilon = 0.3$}
    \label{fig:jobsQ3_wins_e03}
  \end{subfigure} \hfill
  \begin{subfigure}[b]{0.24\linewidth}
    \centering
    \includegraphics[width=\linewidth]{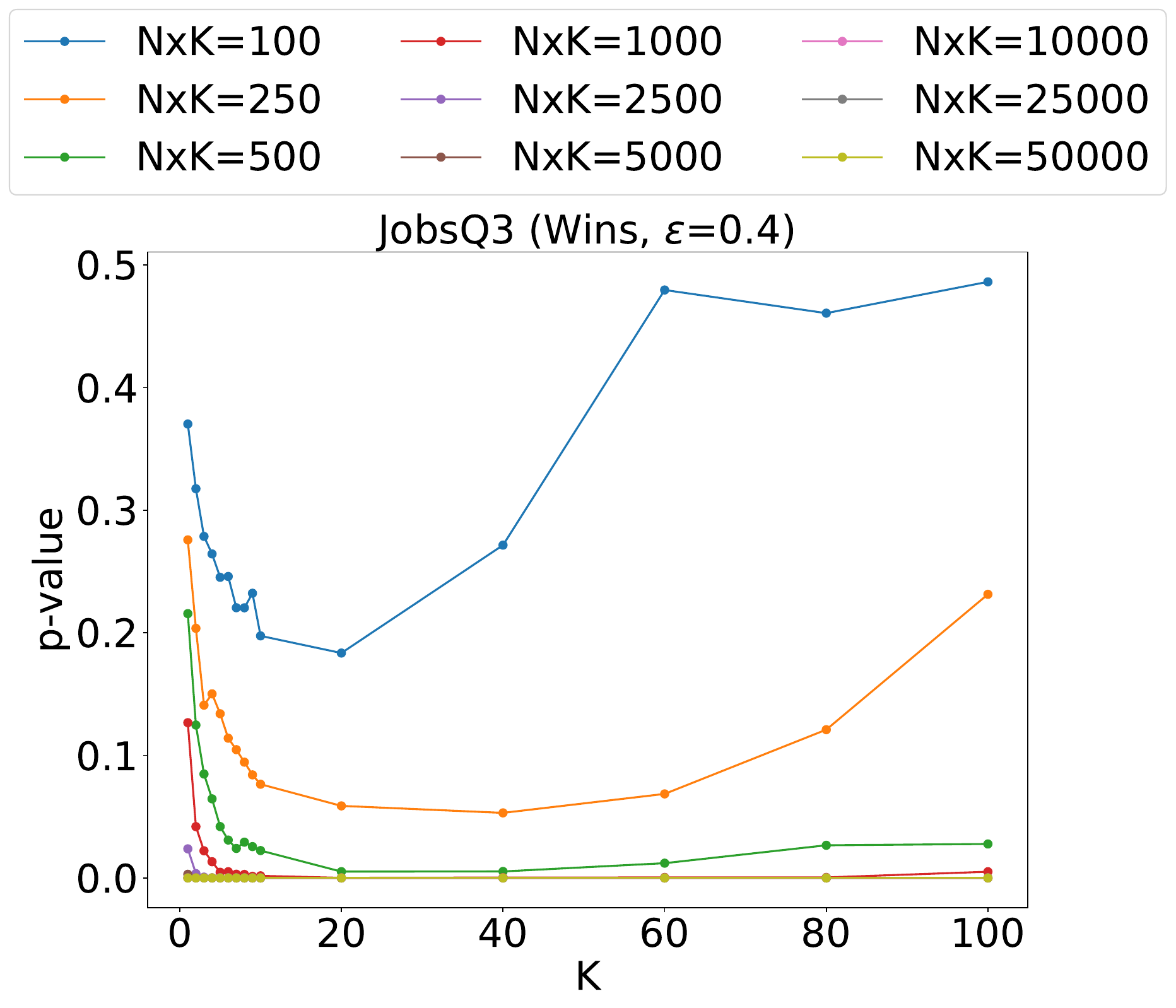}
    \caption{$\epsilon = 0.4$}
    \label{fig:jobsQ3_wins_e04}
  \end{subfigure}
  \caption{P-value plots for JobsQ3 dataset with Wins as the metric}
  \label{fig:jobsQ3_wins}
\end{figure*}

\begin{figure*}
  \centering
  \begin{subfigure}[b]{0.24\linewidth}
    \centering
    \includegraphics[width=\linewidth]{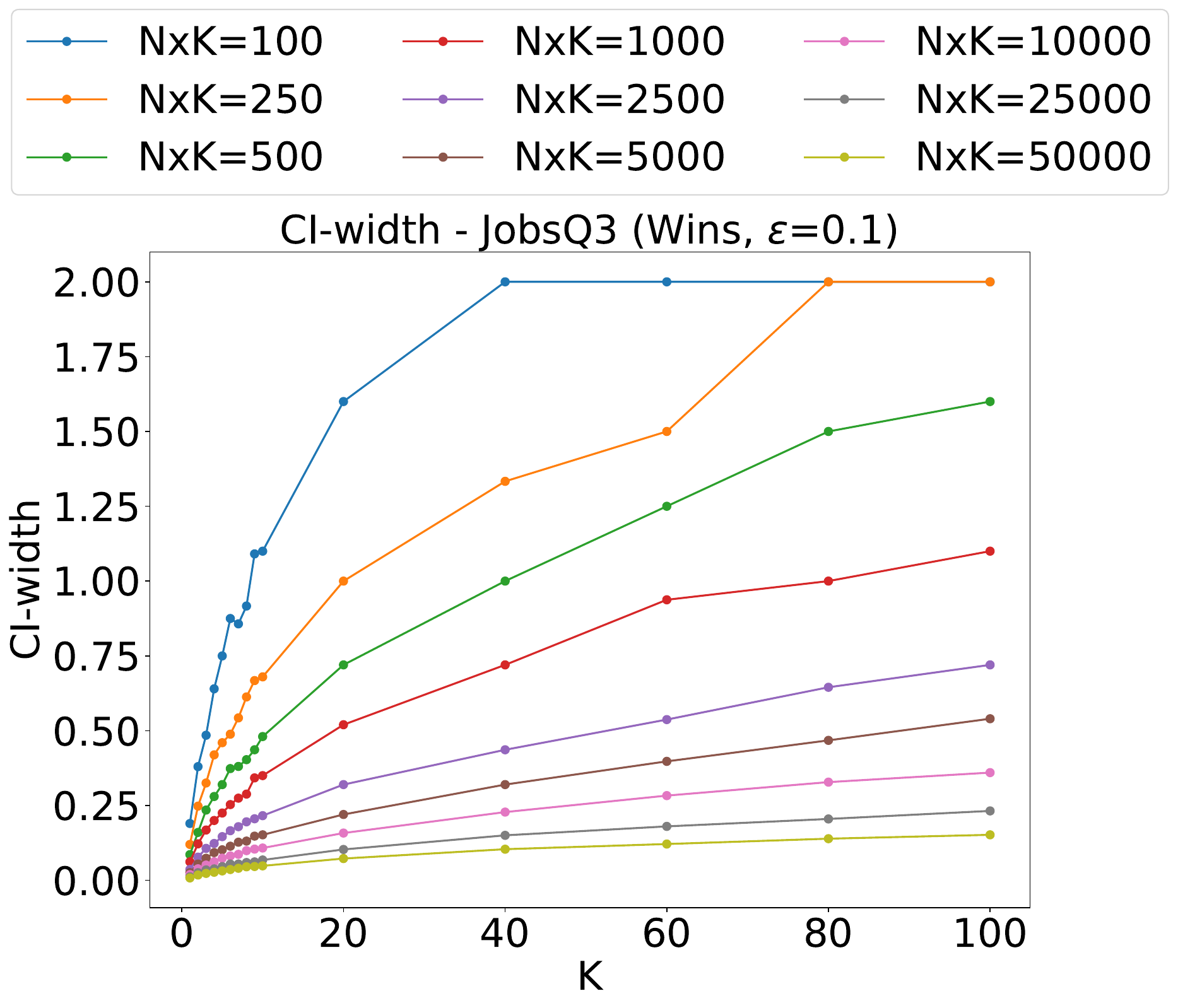}
    \caption{$\epsilon = 0.1$}
    \label{fig:jobsQ3_ci_wins_e01}
  \end{subfigure} \hfill
  \begin{subfigure}[b]{0.24\linewidth}
    \centering
    \includegraphics[width=\linewidth]{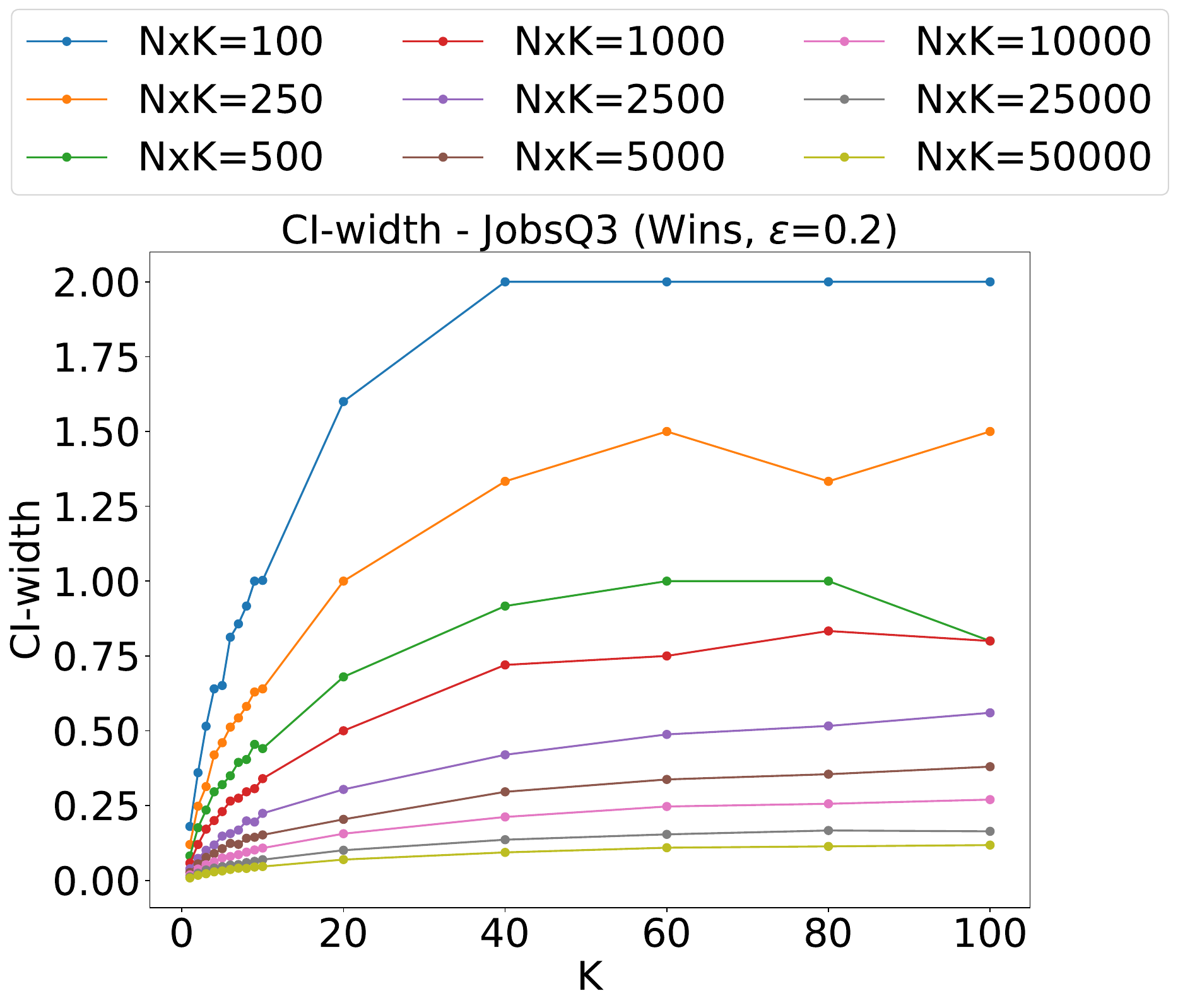}
    \caption{$\epsilon = 0.2$}
    \label{fig:jobsQ3_ci_wins_e02}
  \end{subfigure} \hfill
  \begin{subfigure}[b]{0.24\linewidth}
    \centering
    \includegraphics[width=\linewidth]{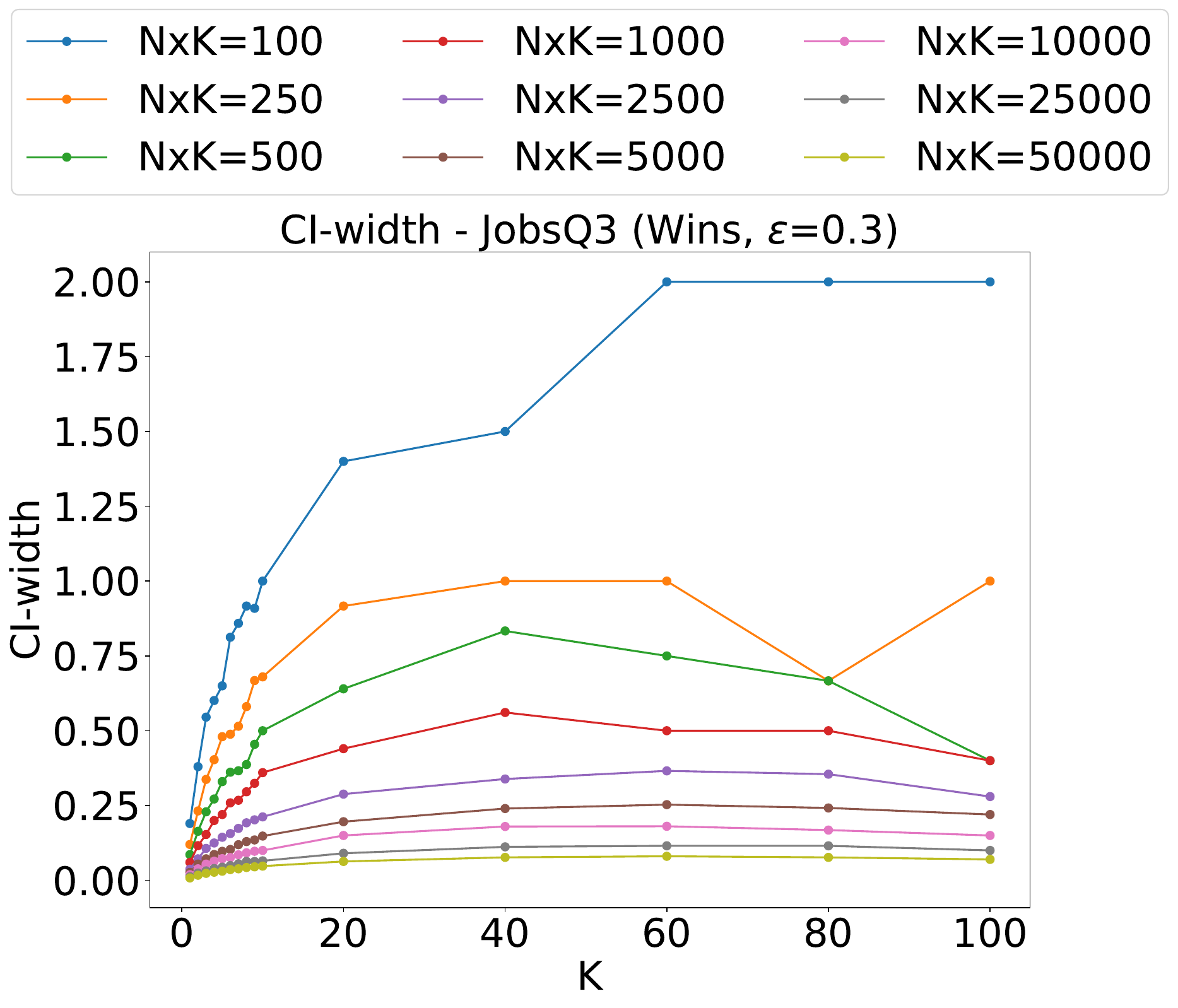}
    \caption{$\epsilon = 0.3$}
    \label{fig:jobsQ3_ci_wins_e03}
  \end{subfigure} \hfill
  \begin{subfigure}[b]{0.24\linewidth}
    \centering
    \includegraphics[width=\linewidth]{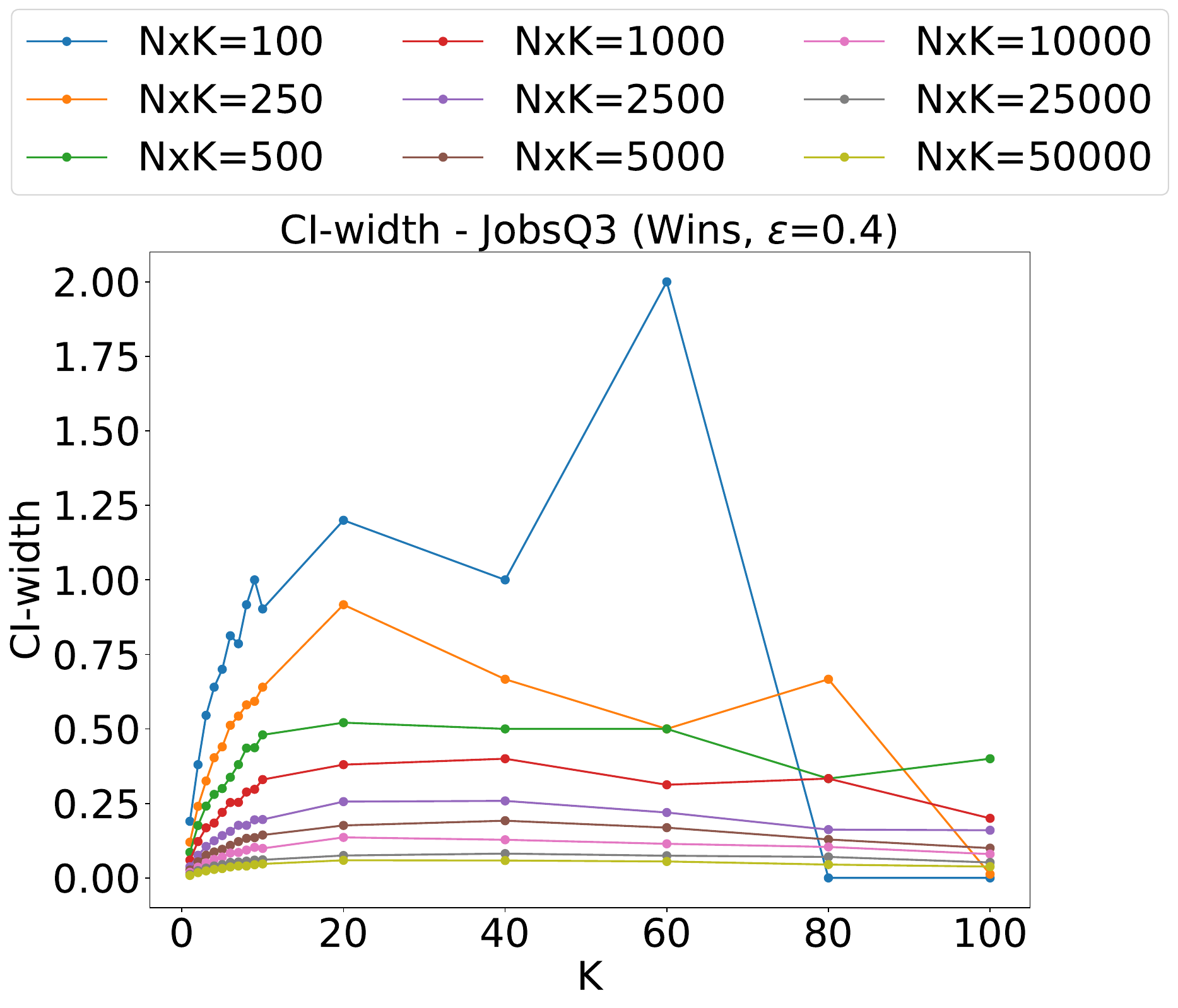}
    \caption{$\epsilon = 0.4$}
    \label{fig:jobsQ3_ci_wins_e04}
  \end{subfigure}
  \caption{CI-width plots for JobsQ3 dataset with Wins as the metric}
  \label{fig:jobsQ3_ci_wins}
\end{figure*}

\begin{figure*}
  \centering
  \begin{subfigure}[b]{0.24\linewidth}
    \centering
    \includegraphics[width=\linewidth]{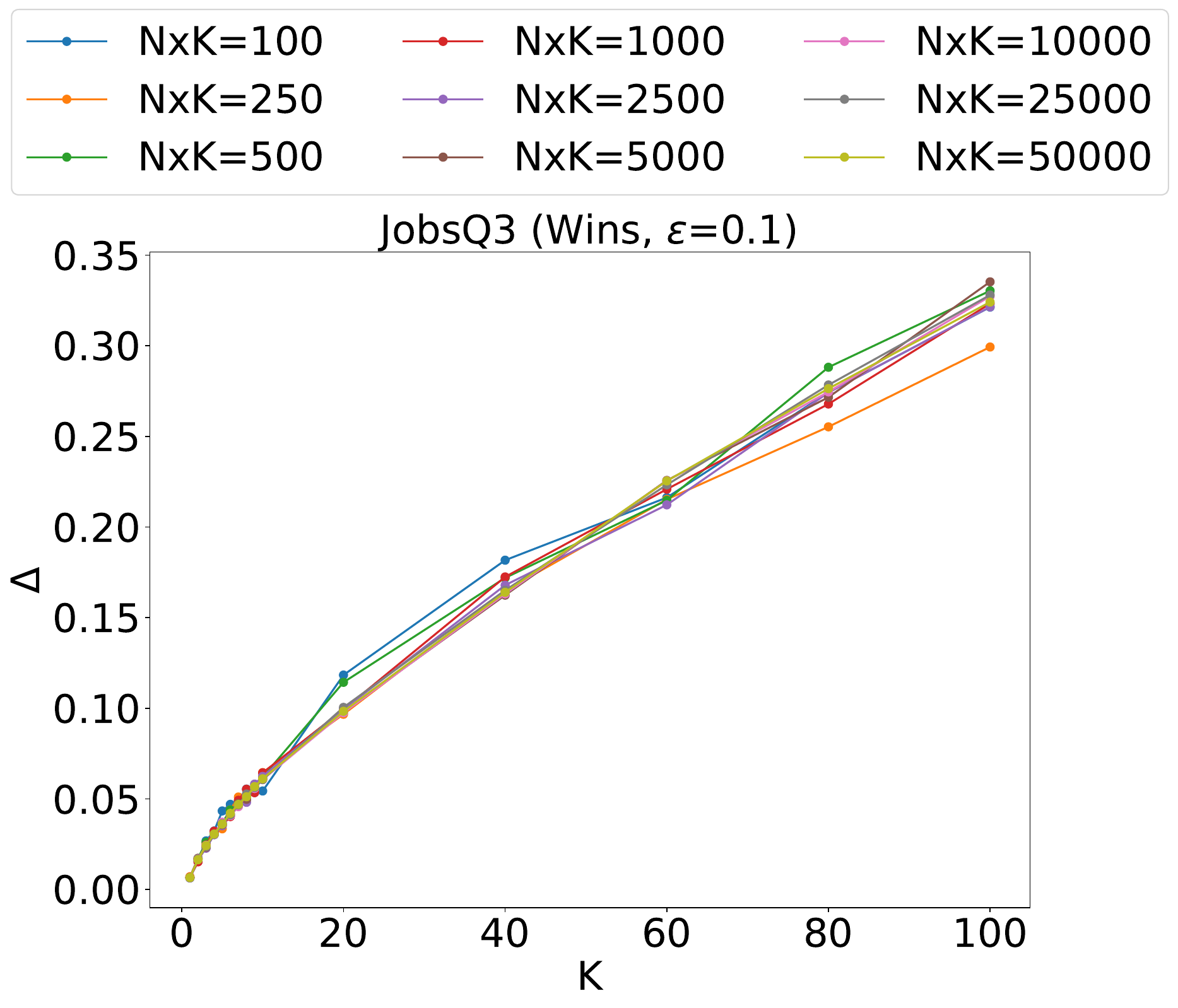}
    \caption{$\epsilon = 0.1$}
    \label{fig:jobsQ3_delta_wins_e01}
  \end{subfigure} \hfill
  \begin{subfigure}[b]{0.24\linewidth}
    \centering
    \includegraphics[width=\linewidth]{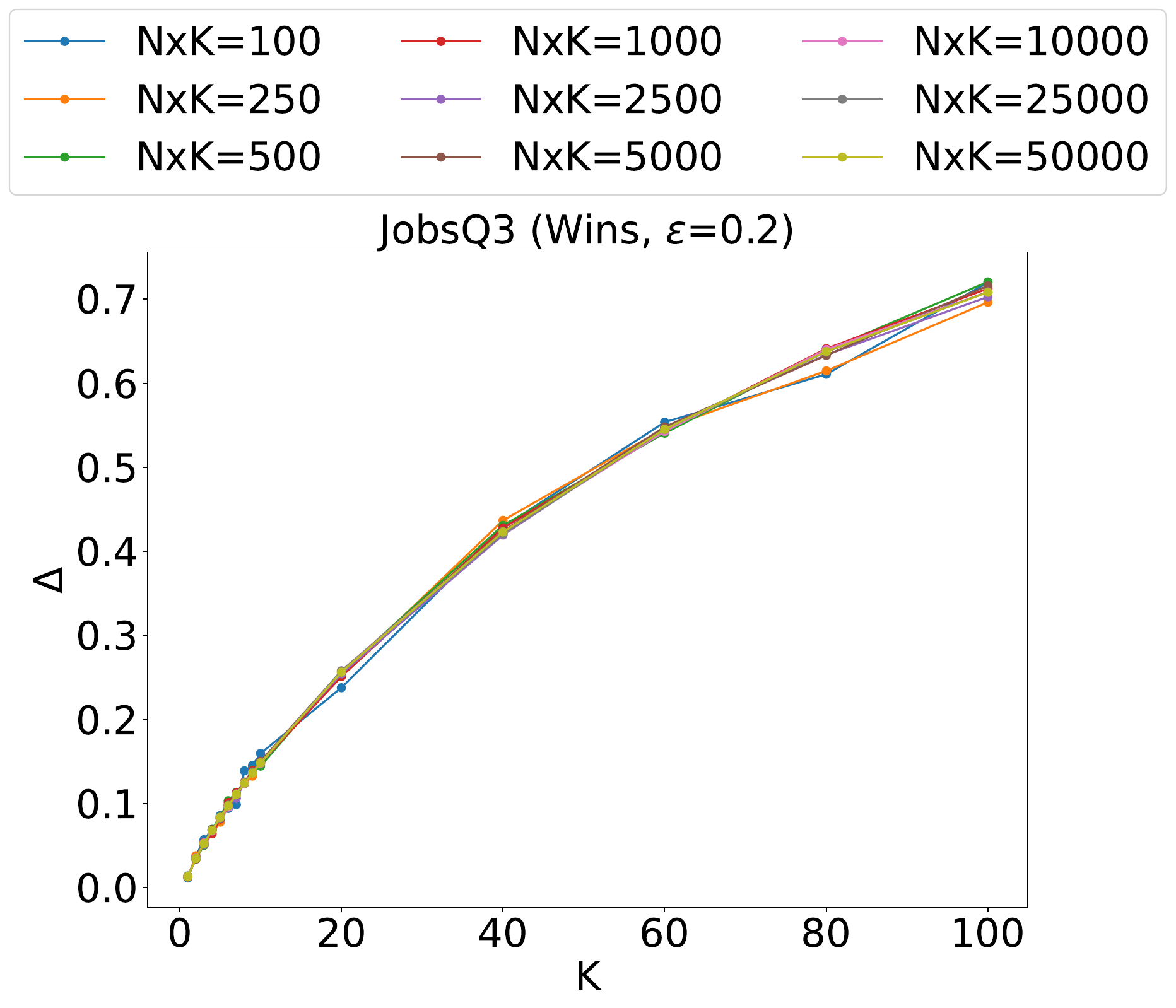}
    \caption{$\epsilon = 0.2$}
    \label{fig:jobsQ3_delta_wins_e02}
  \end{subfigure} \hfill
  \begin{subfigure}[b]{0.24\linewidth}
    \centering
    \includegraphics[width=\linewidth]{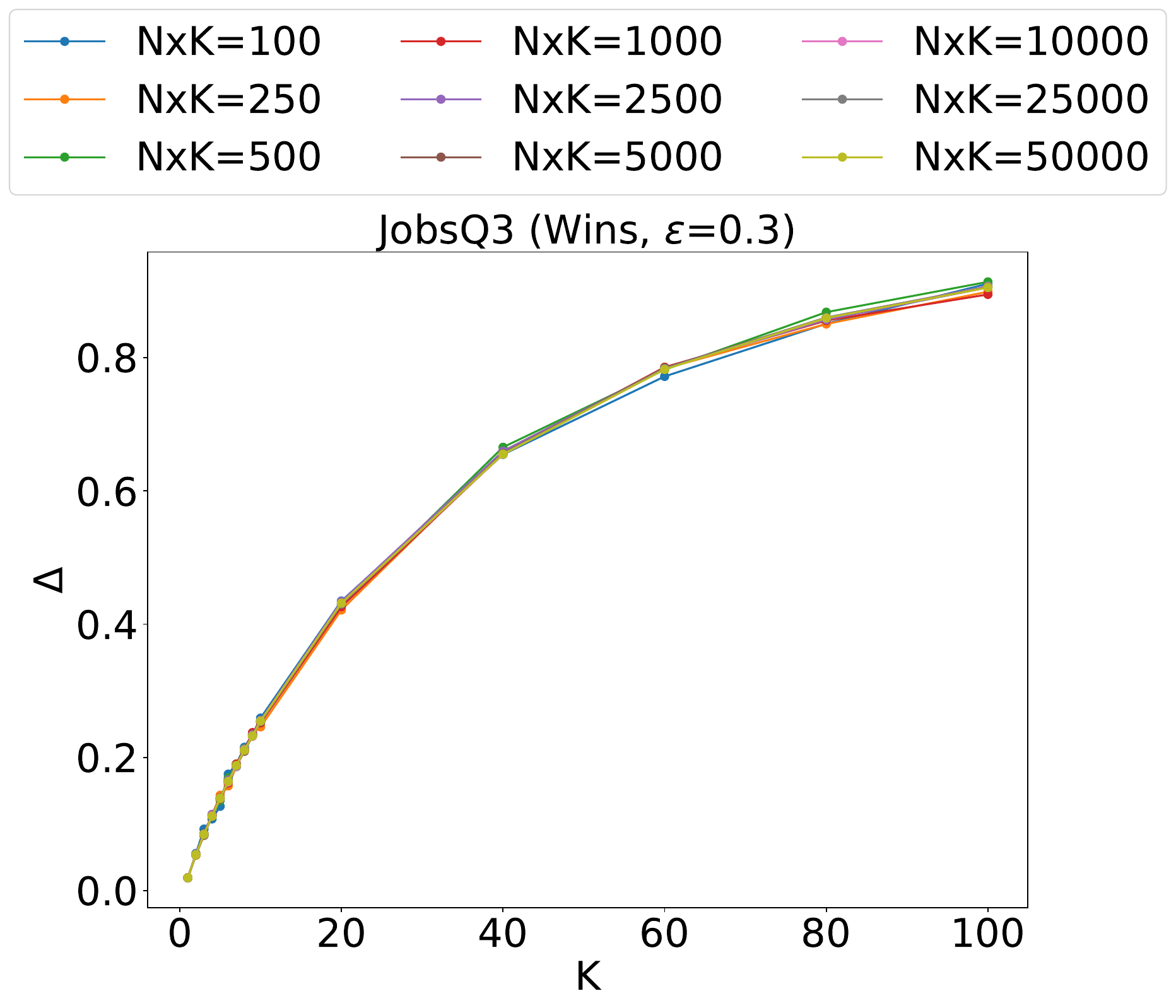}
    \caption{$\epsilon = 0.3$}
    \label{fig:jobsQ3_delta_wins_e03}
  \end{subfigure} \hfill
  \begin{subfigure}[b]{0.24\linewidth}
    \centering
    \includegraphics[width=\linewidth]{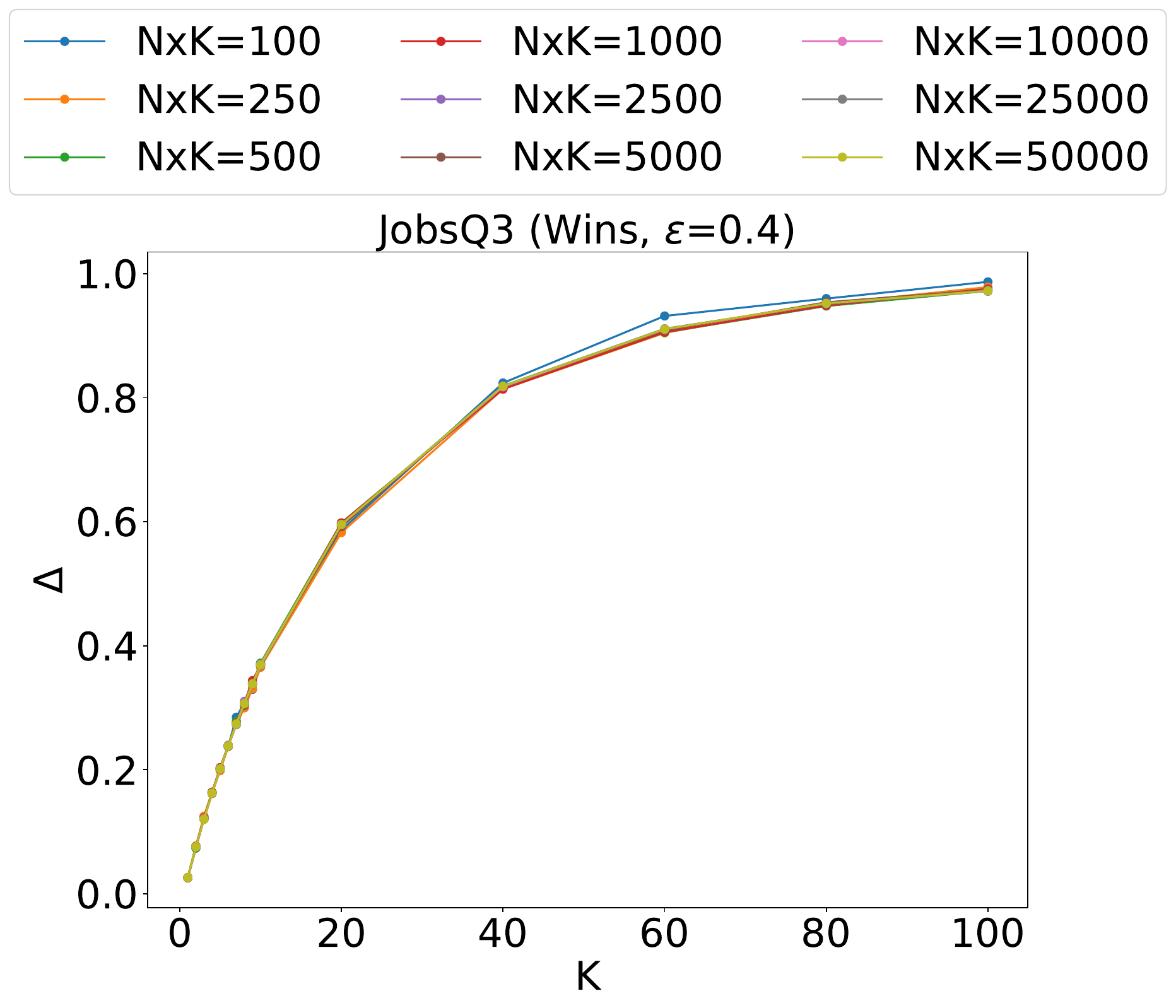}
    \caption{$\epsilon = 0.4$}
    \label{fig:jobsQ3_delta_wins_e04}
  \end{subfigure}
  \caption{Effect sizes ($\Delta$) for JobsQ3 dataset with Wins as the metric}
  \label{fig:jobsQ3_delta_wins}
\end{figure*}

Figures \ref{fig:toxicity_wins}--\ref{fig:jobsQ3_delta_wins} show results for \pv s, confidence intervals, and effect sizes for all datasets with different values of $\epsilon$ and increasing $K$ for various $N \times K$. We notice \pv\ increases sharply with increasing $K$ till $K=10$, and then starts to plateau for all $N \times K$ for $\epsilon=0.1$. For the remaining values of $\epsilon$, \pv s\ start going back up as $K$ gets higher. As $\epsilon$ increases, \pv s\ decrease as expected. There seems to be an optimal $K$ for Dices, D3code, and JobsQ3 dataset. CI-width and effect sizes increase with increasing $K$.

\subsubsection{Artificial Distribution with Balanced Priors}

\begin{figure*}
  \centering
  \begin{subfigure}[b]{0.24\linewidth}
    \centering
    \includegraphics[width=\linewidth]{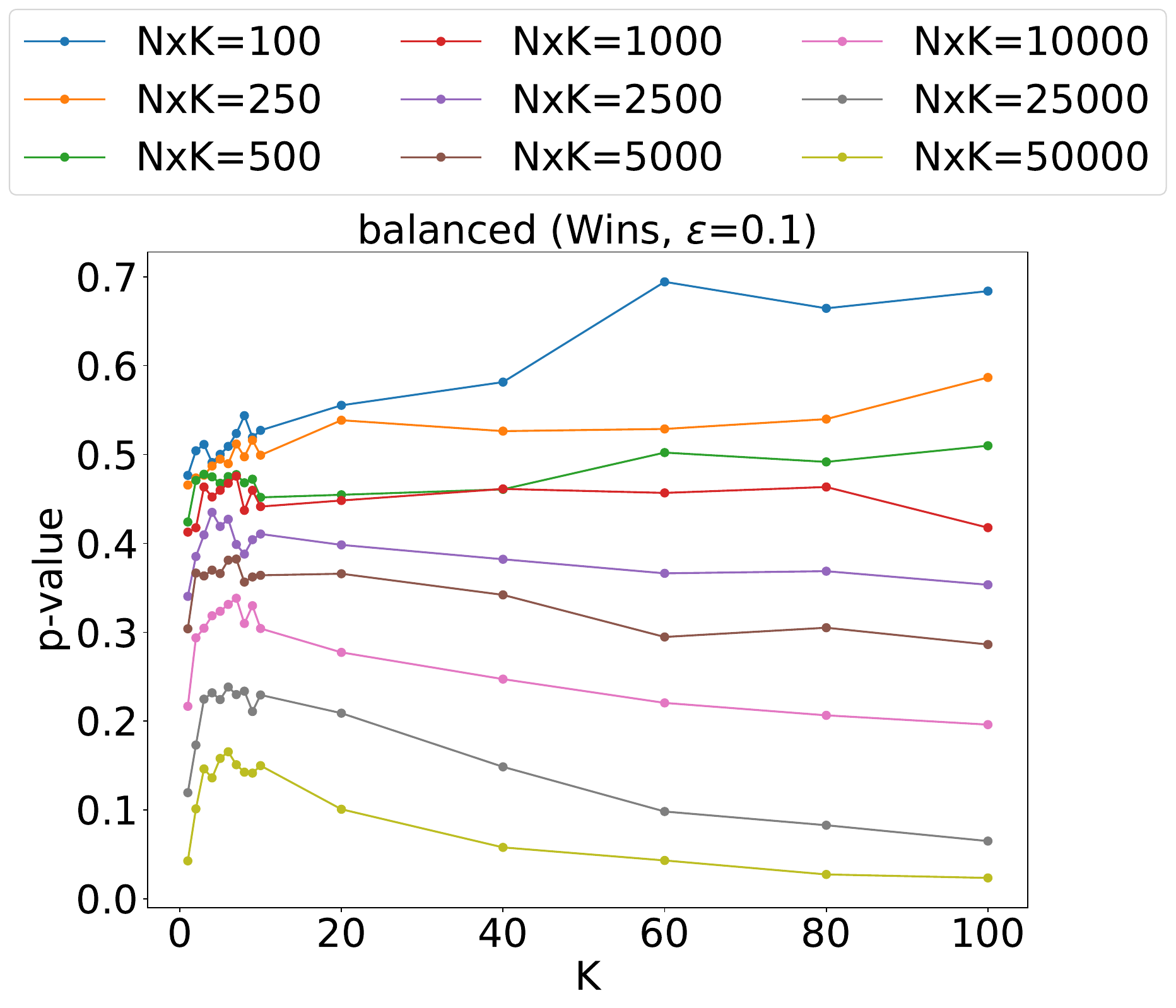}
    \caption{$\epsilon = 0.1$}
    \label{fig:uniform_wins_cat2_e01}
  \end{subfigure} \hfill
  \begin{subfigure}[b]{0.24\linewidth}
    \centering
    \includegraphics[width=\linewidth]{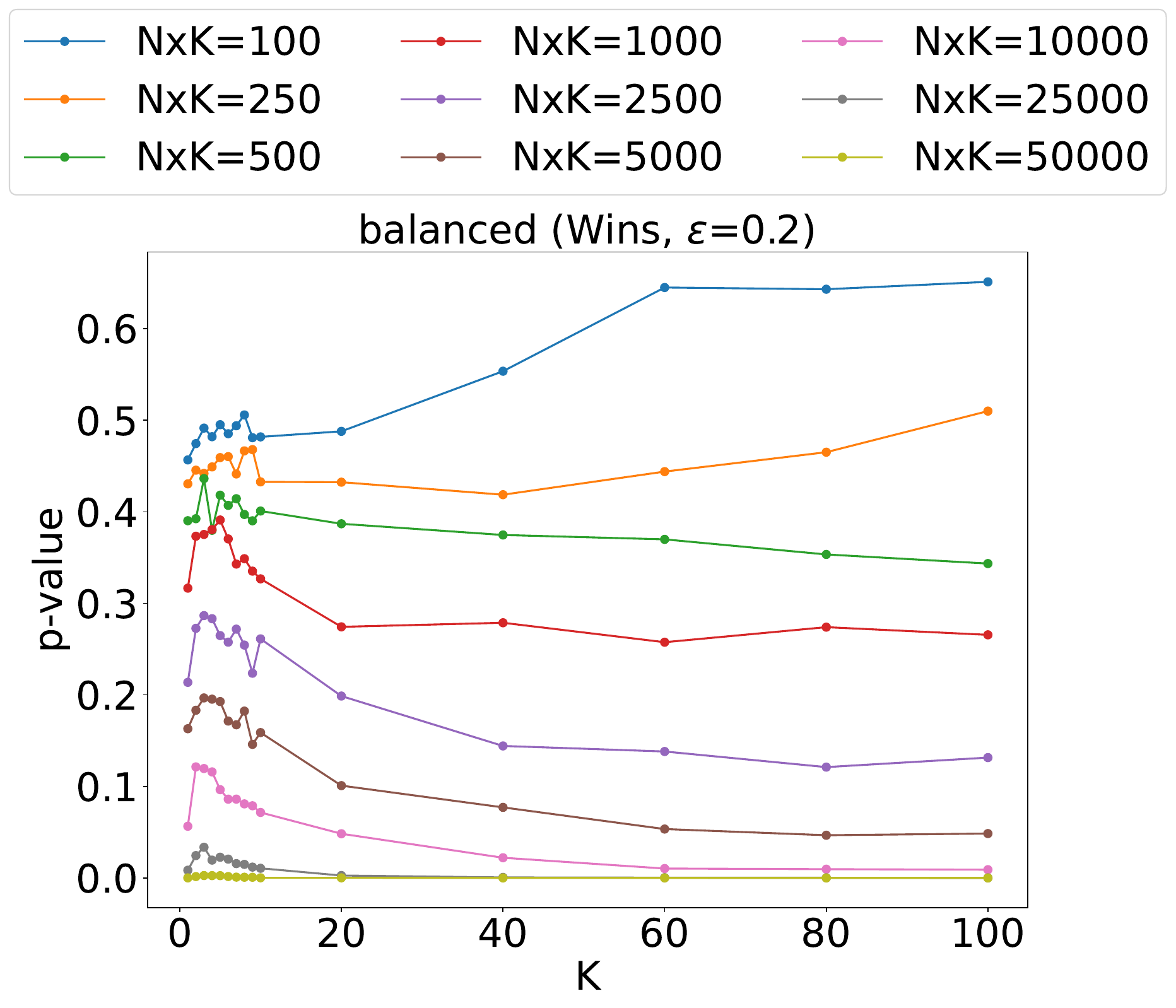}
    \caption{$\epsilon = 0.2$}
    \label{fig:uniform_wins_cat2_e02}
  \end{subfigure} \hfill
  \begin{subfigure}[b]{0.24\linewidth}
    \centering
    \includegraphics[width=\linewidth]{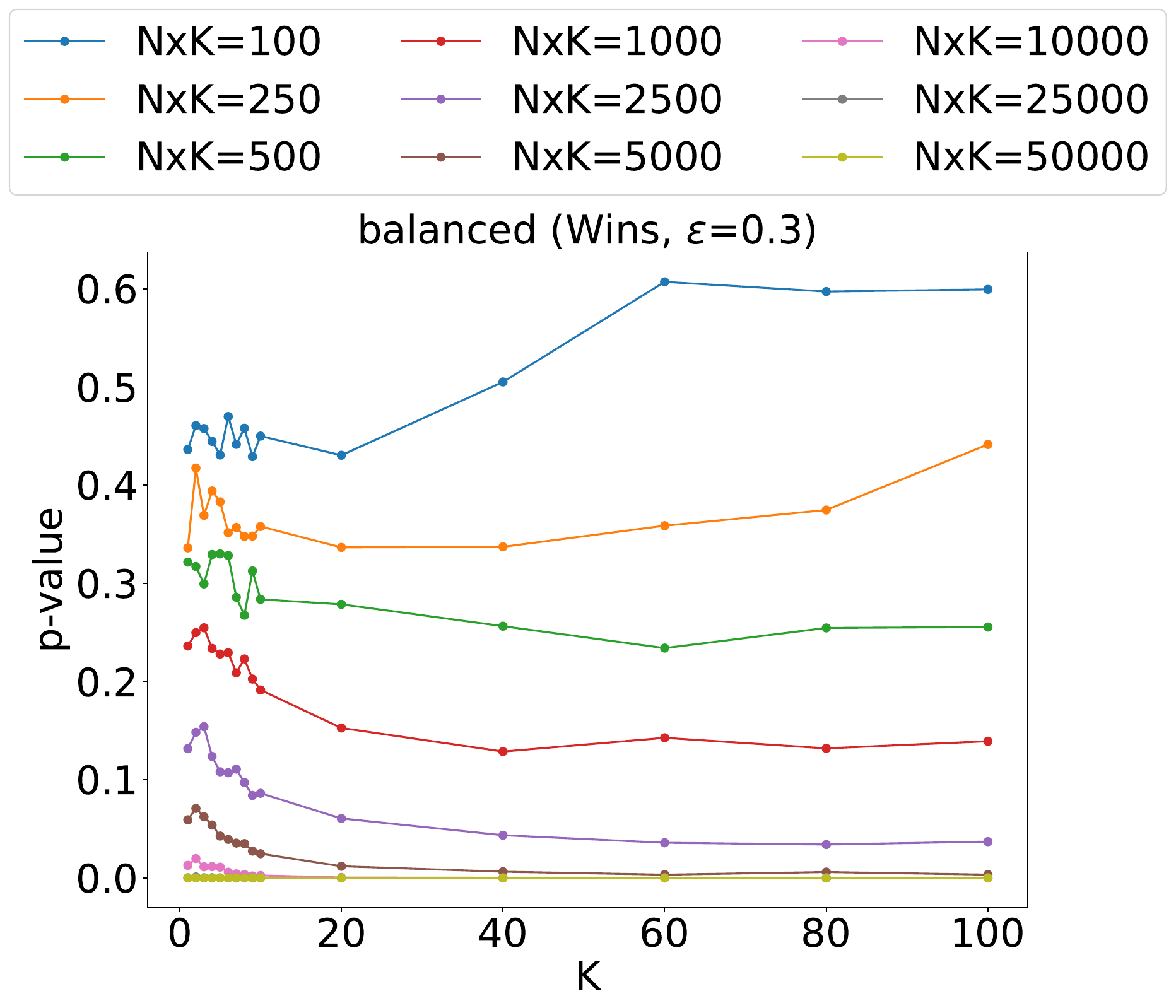}
    \caption{$\epsilon = 0.3$}
    \label{fig:uniform_wins_cat2_e03}
  \end{subfigure} \hfill
  \begin{subfigure}[b]{0.24\linewidth}
    \centering
    \includegraphics[width=\linewidth]{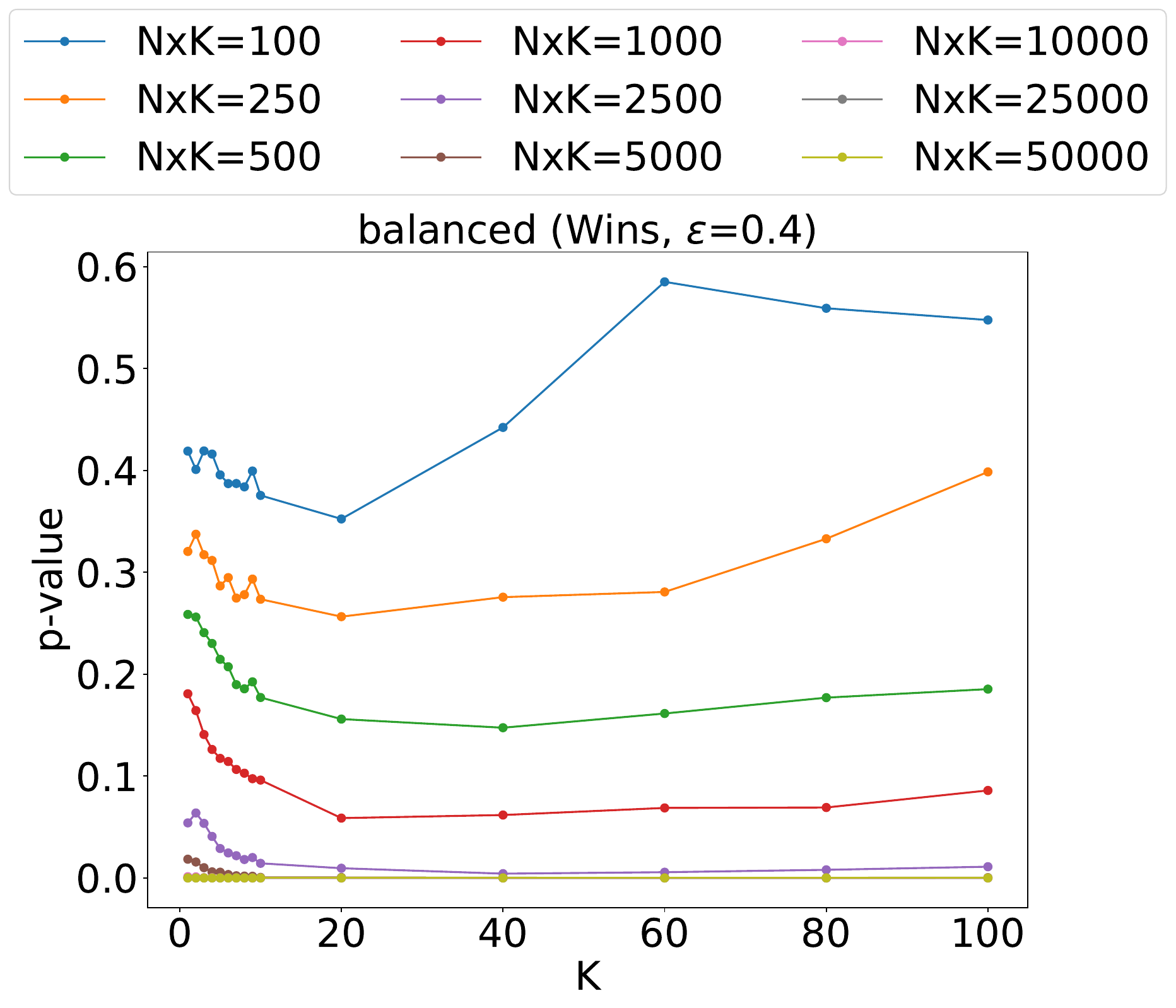}
    \caption{$\epsilon = 0.4$}
    \label{fig:uniform_wins_cat2_e04}
  \end{subfigure}
  \caption{P-value plots for balanced alphas with Wins as the metric ($M=2$)}
  \label{fig:uniform_wins_cat2}
\end{figure*}

\begin{figure*}
  \centering
  \begin{subfigure}[b]{0.24\linewidth}
    \centering
    \includegraphics[width=\linewidth]{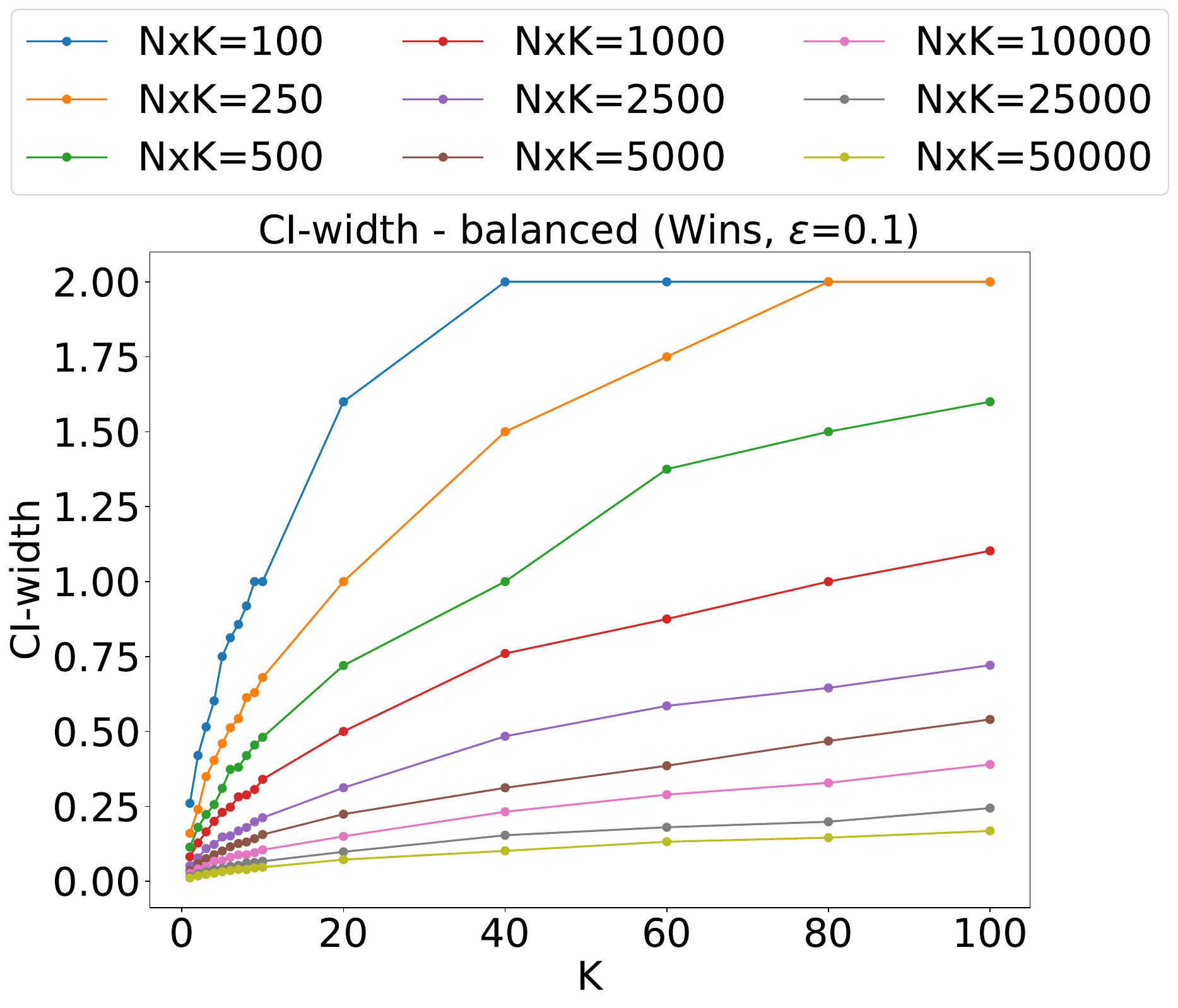}
    \caption{$\epsilon = 0.1$}
    \label{fig:uniform_ci_wins_cat2_e01}
  \end{subfigure} \hfill
  \begin{subfigure}[b]{0.24\linewidth}
    \centering
    \includegraphics[width=\linewidth]{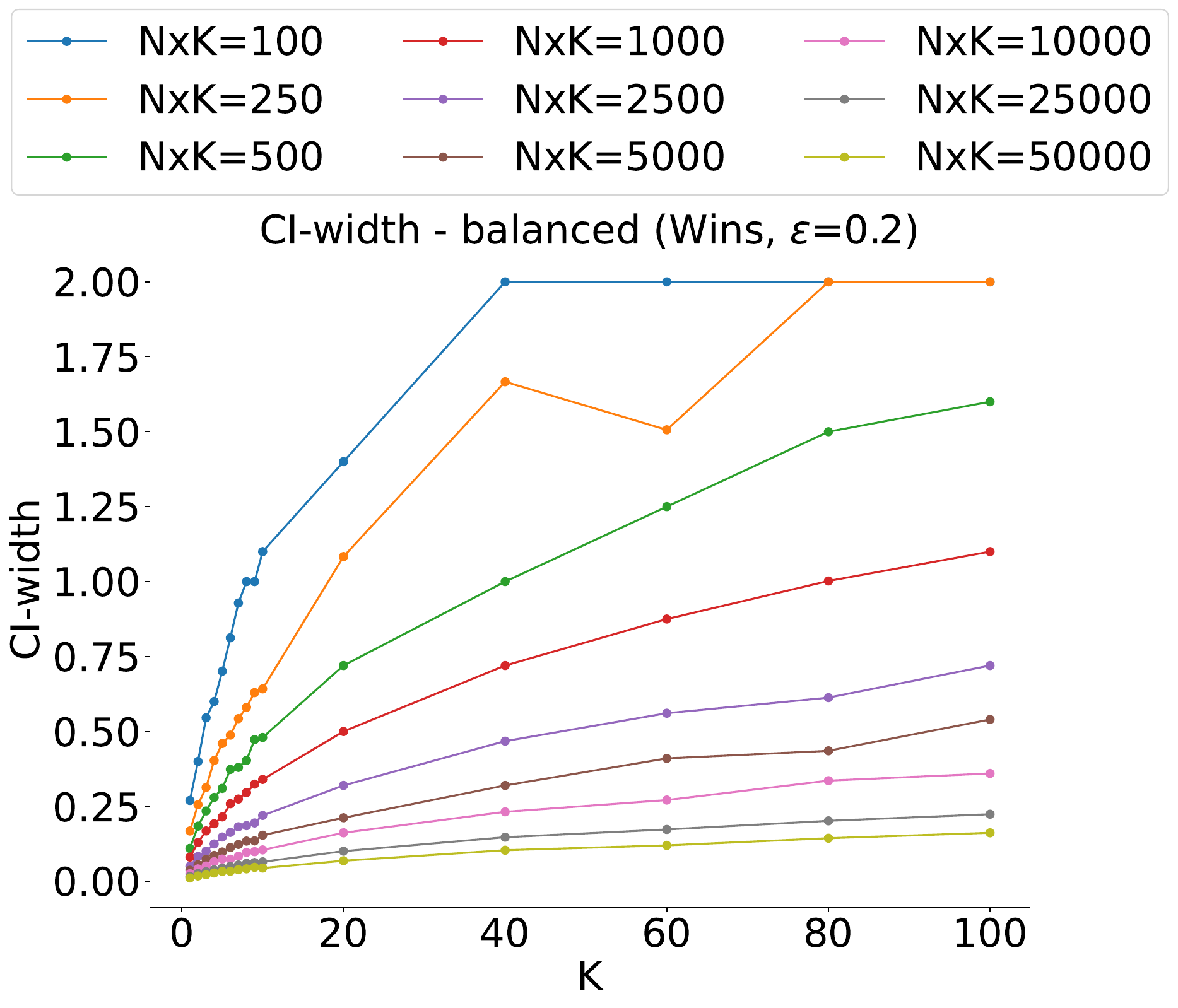}
    \caption{$\epsilon = 0.2$}
    \label{fig:uniform_ci_wins_cat2_e02}
  \end{subfigure} \hfill
  \begin{subfigure}[b]{0.24\linewidth}
    \centering
    \includegraphics[width=\linewidth]{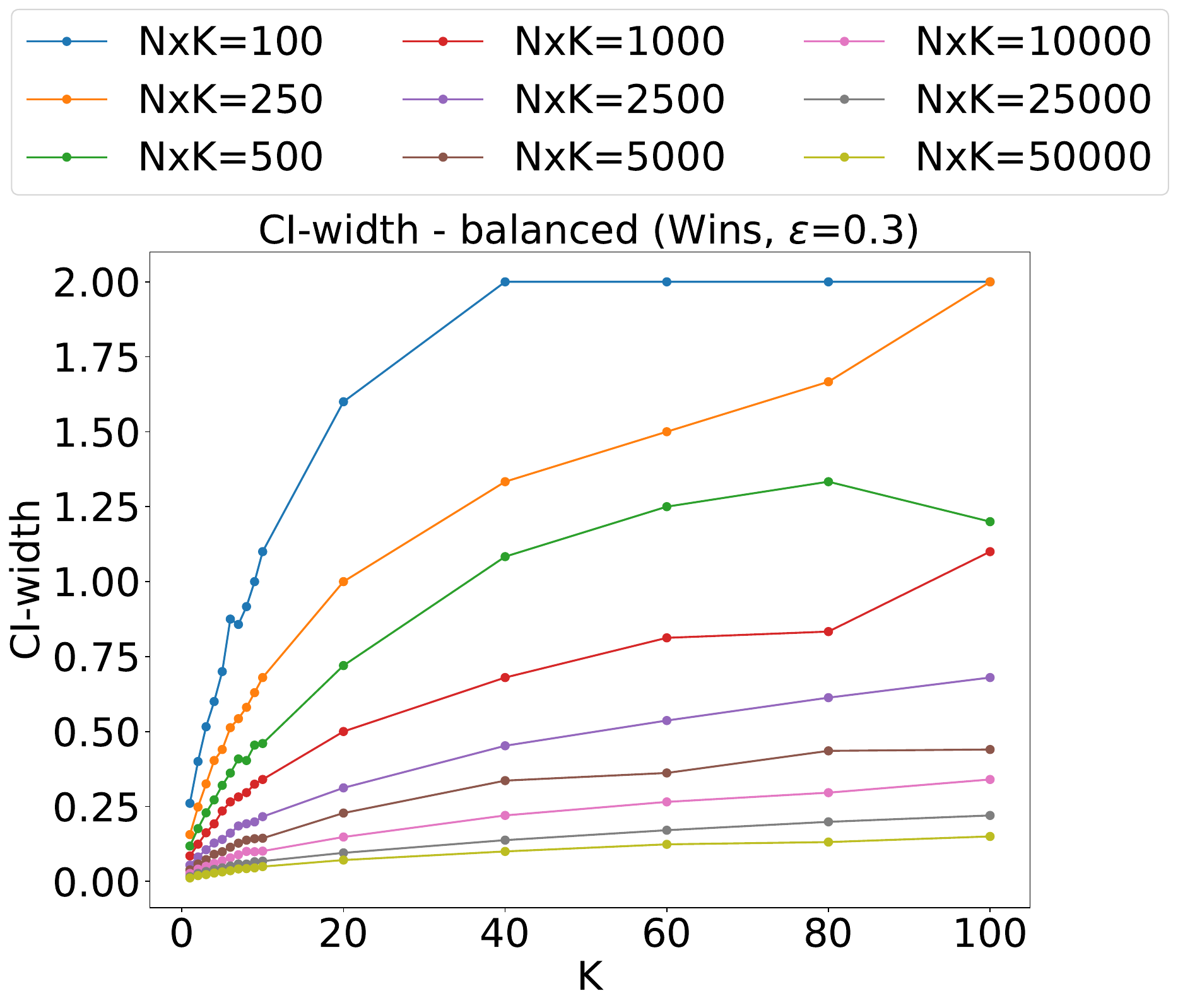}
    \caption{$\epsilon = 0.3$}
    \label{fig:uniform_ci_wins_cat2_e03}
  \end{subfigure} \hfill
  \begin{subfigure}[b]{0.24\linewidth}
    \centering
    \includegraphics[width=\linewidth]{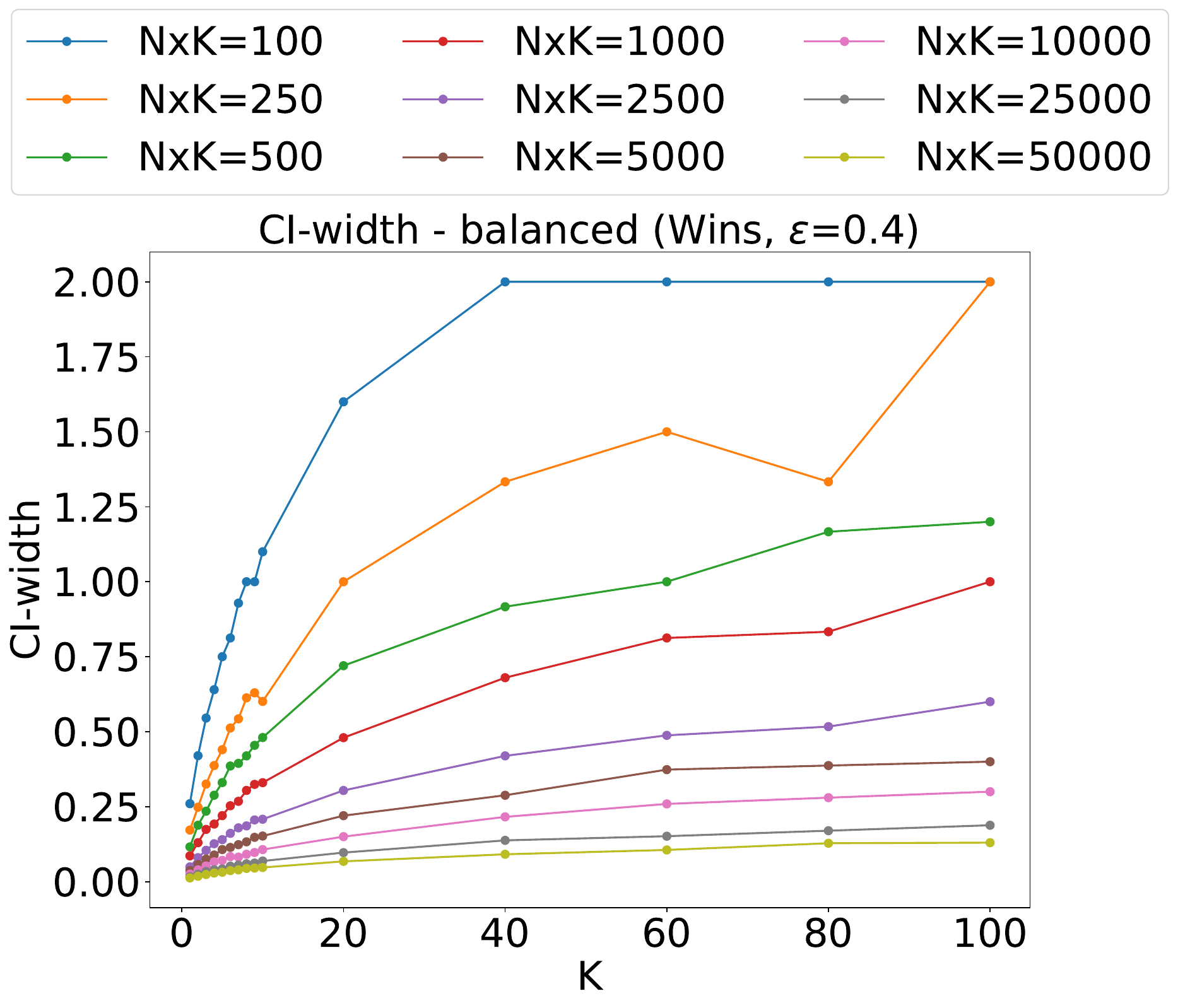}
    \caption{$\epsilon = 0.4$}
    \label{fig:uniform_ci_wins_cat2_e04}
  \end{subfigure}
  \caption{CI-width plots for balanced alphas with Wins as the metric ($M=2$)}
  \label{fig:uniform_ci_wins_cat2}
\end{figure*}

\begin{figure*}
  \centering
  \begin{subfigure}[b]{0.24\linewidth}
    \centering
    \includegraphics[width=\linewidth]{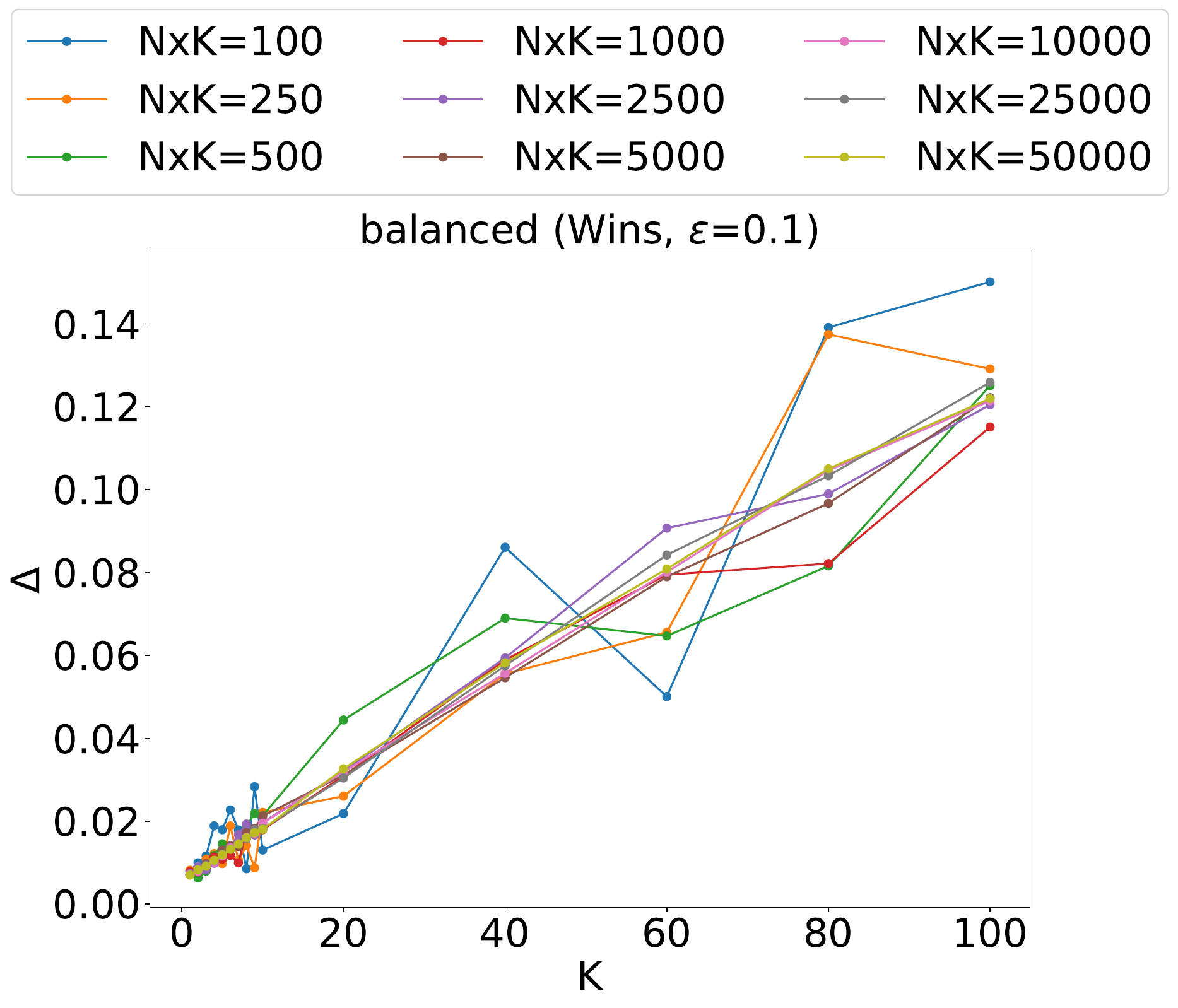}
    \caption{$\epsilon = 0.1$}
    \label{fig:uniform_delta_wins_cat2_e01}
  \end{subfigure} \hfill
  \begin{subfigure}[b]{0.24\linewidth}
    \centering
    \includegraphics[width=\linewidth]{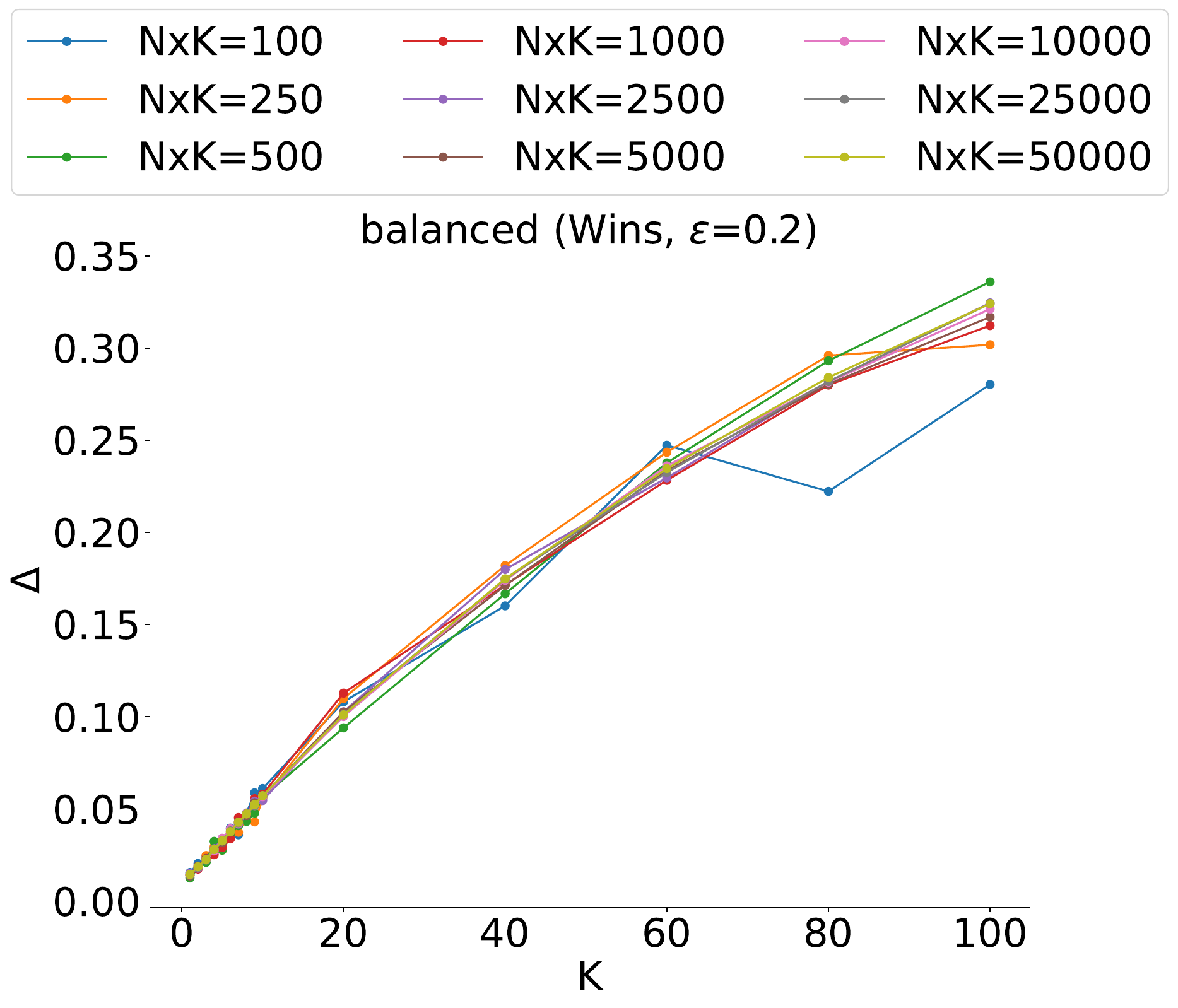}
    \caption{$\epsilon = 0.2$}
    \label{fig:uniform_delta_wins_cat2_e02}
  \end{subfigure} \hfill
  \begin{subfigure}[b]{0.24\linewidth}
    \centering
    \includegraphics[width=\linewidth]{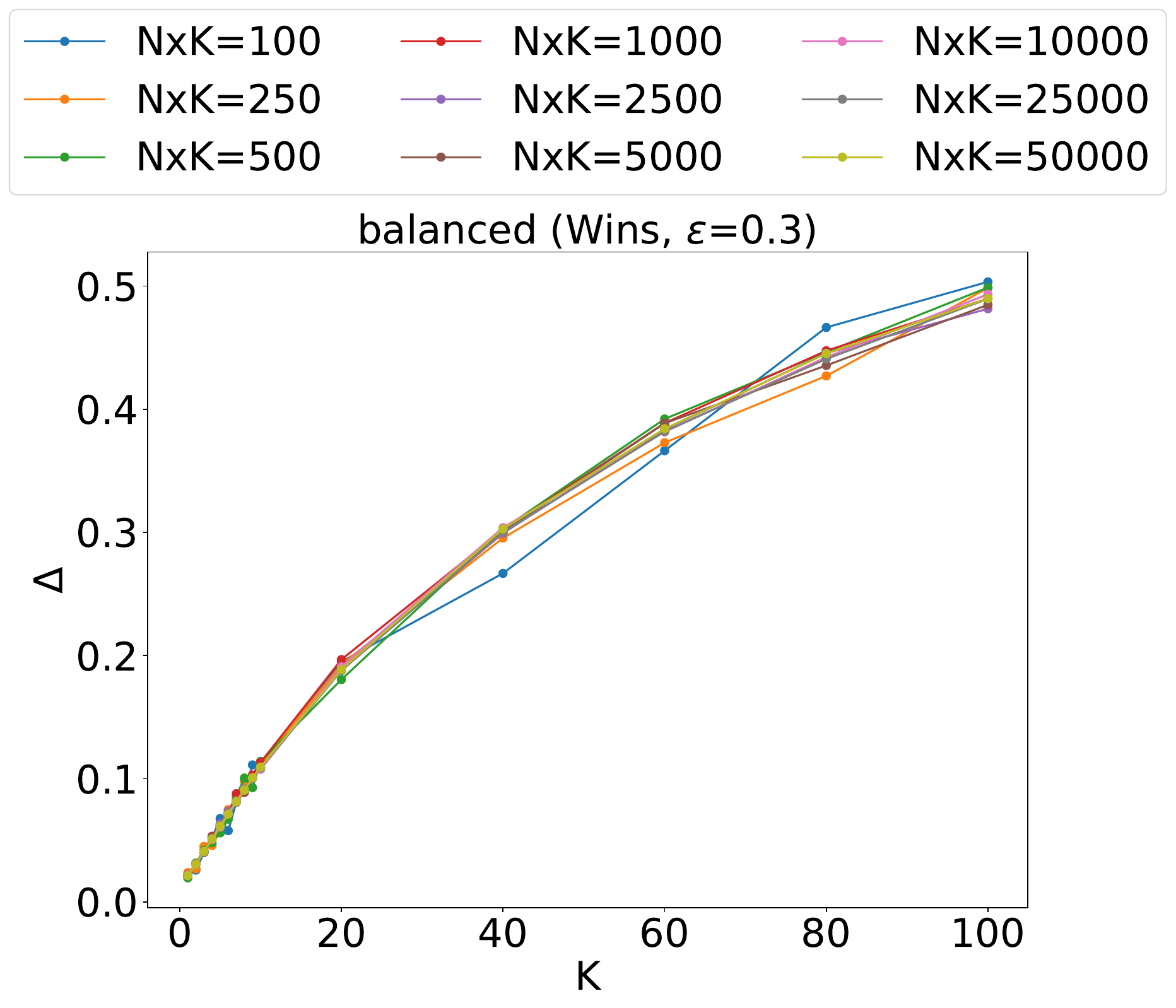}
    \caption{$\epsilon = 0.3$}
    \label{fig:uniform_delta_wins_cat2_e03}
  \end{subfigure} \hfill
  \begin{subfigure}[b]{0.24\linewidth}
    \centering
    \includegraphics[width=\linewidth]{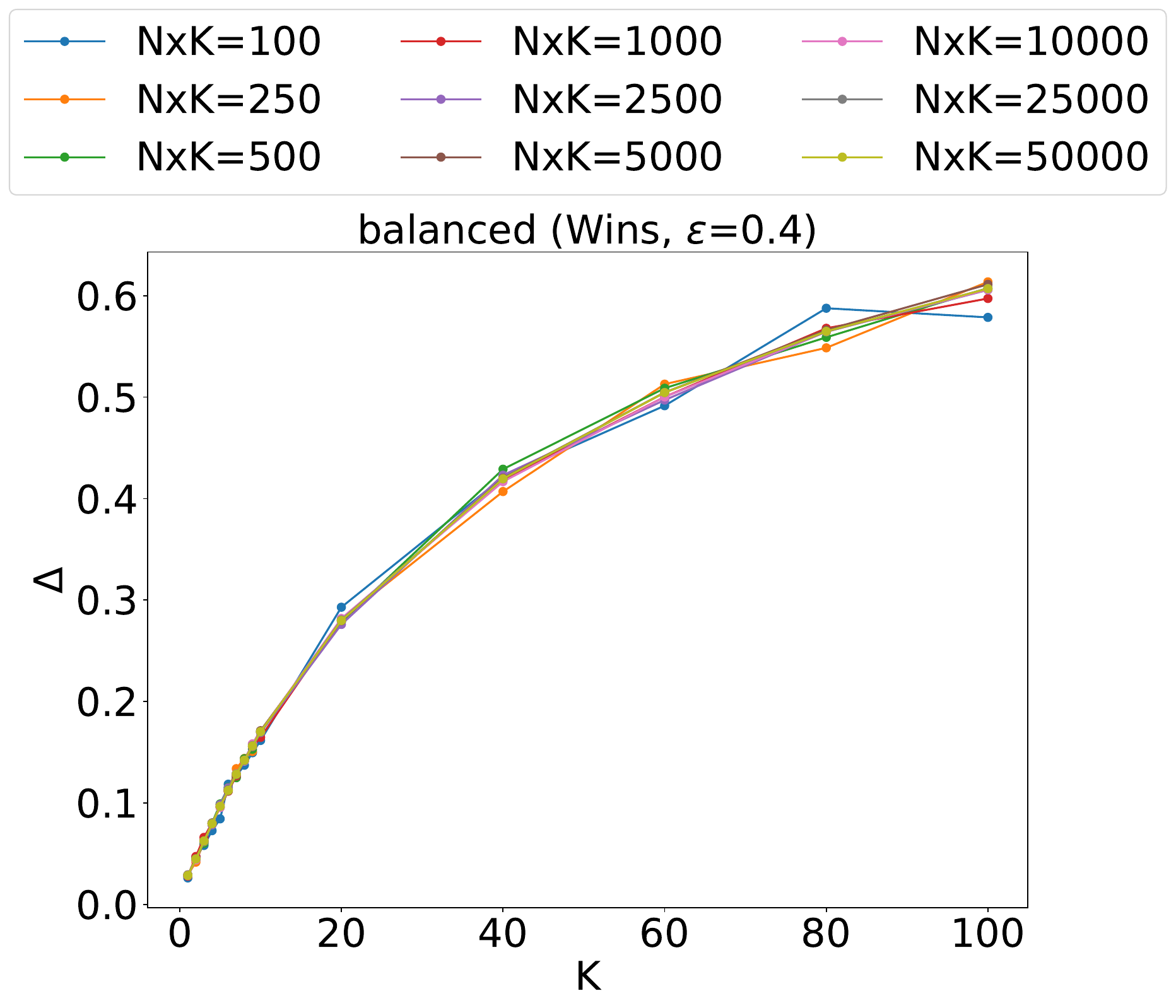}
    \caption{$\epsilon = 0.4$}
    \label{fig:uniform_delta_wins_cat2_e04}
  \end{subfigure}
  \caption{Effect sizes ($\Delta$) for balanced alphas with Wins as the metric ($M=2$)}
  \label{fig:uniform_delta_wins_cat2}
\end{figure*}

\begin{figure*}
  \centering
  \begin{subfigure}[b]{0.24\linewidth}
    \centering
    \includegraphics[width=\linewidth]{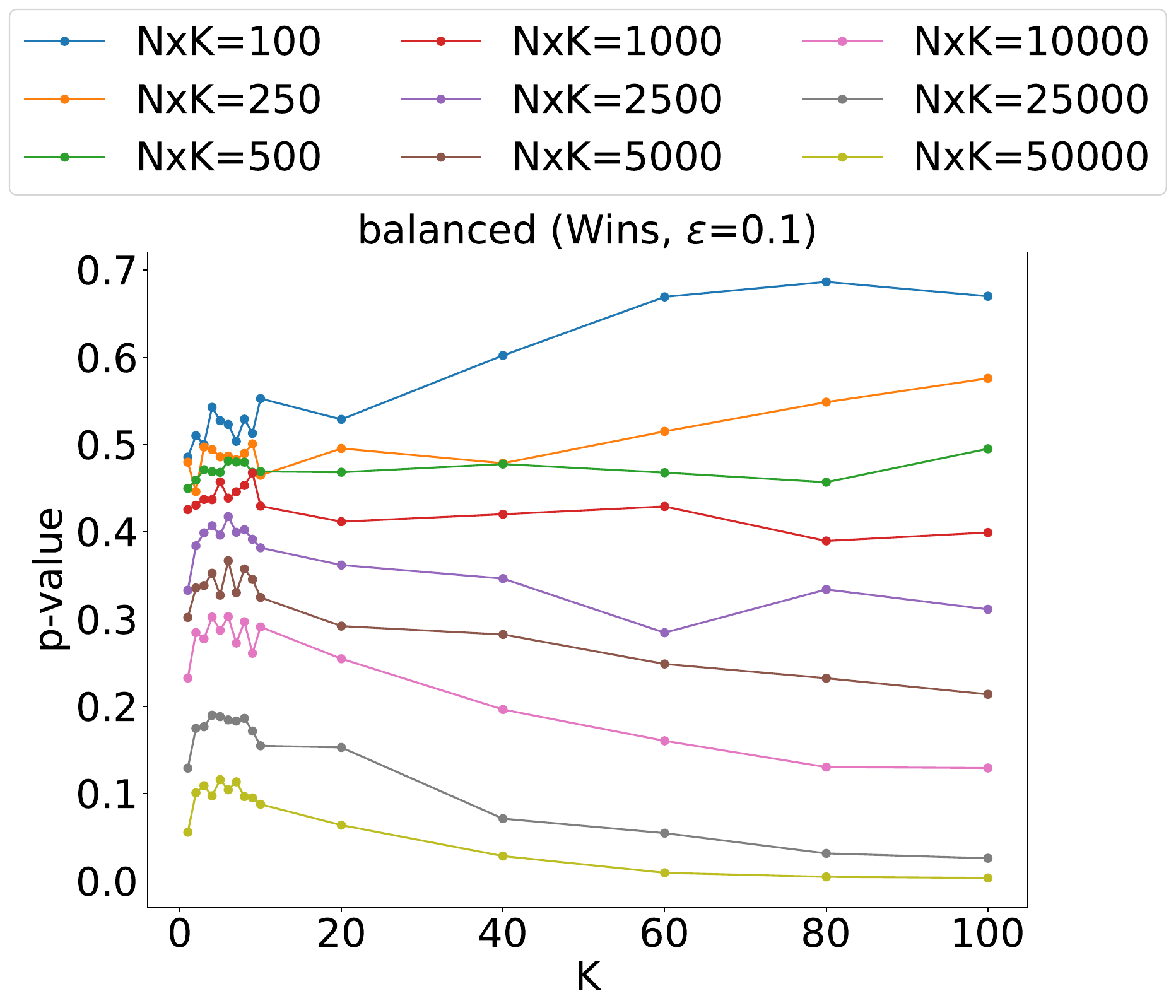}
    \caption{$\epsilon = 0.1$}
    \label{fig:uniform_wins_cat3_e01}
  \end{subfigure} \hfill
  \begin{subfigure}[b]{0.24\linewidth}
    \centering
    \includegraphics[width=\linewidth]{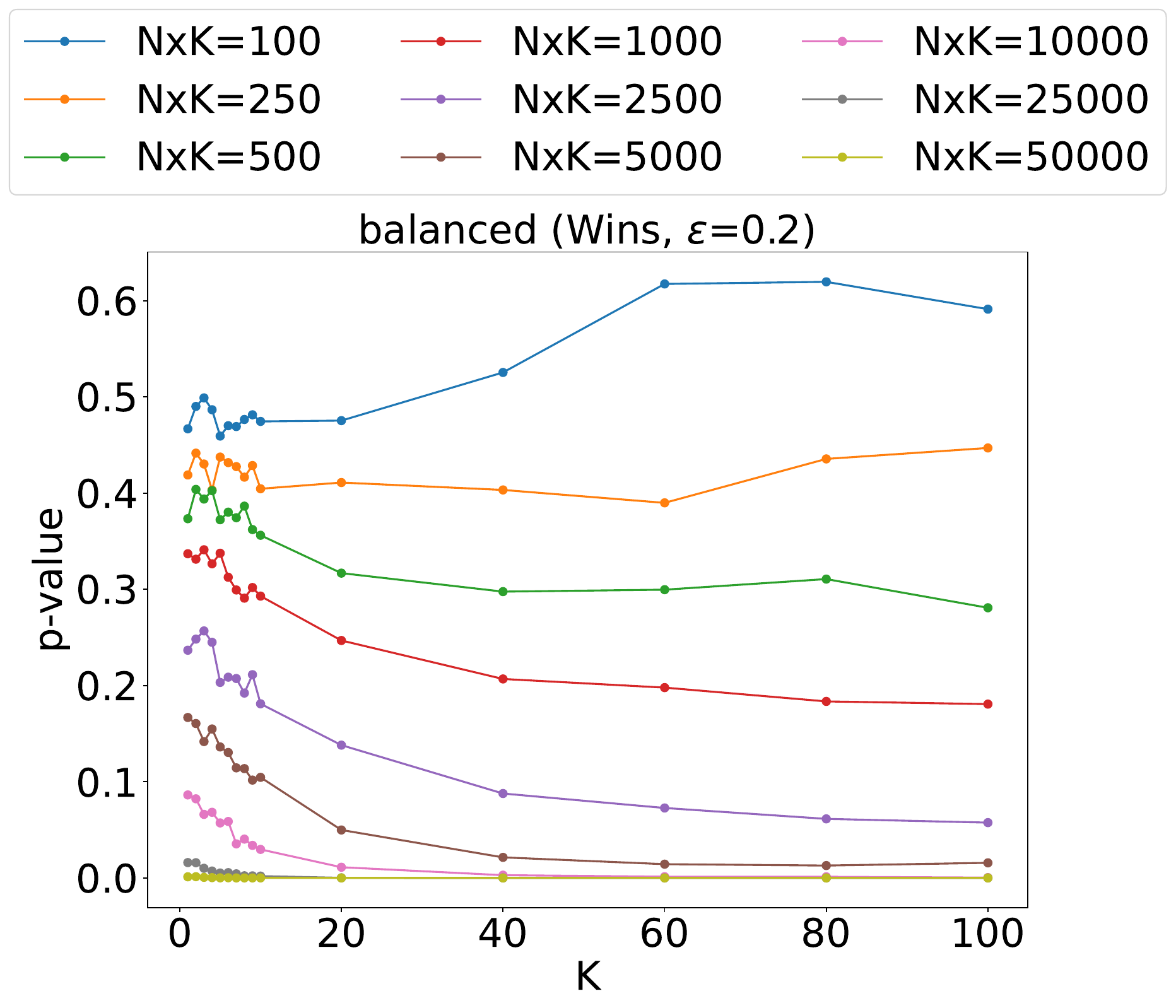}
    \caption{$\epsilon = 0.2$}
    \label{fig:uniform_wins_cat3_e02}
  \end{subfigure} \hfill
  \begin{subfigure}[b]{0.24\linewidth}
    \centering
    \includegraphics[width=\linewidth]{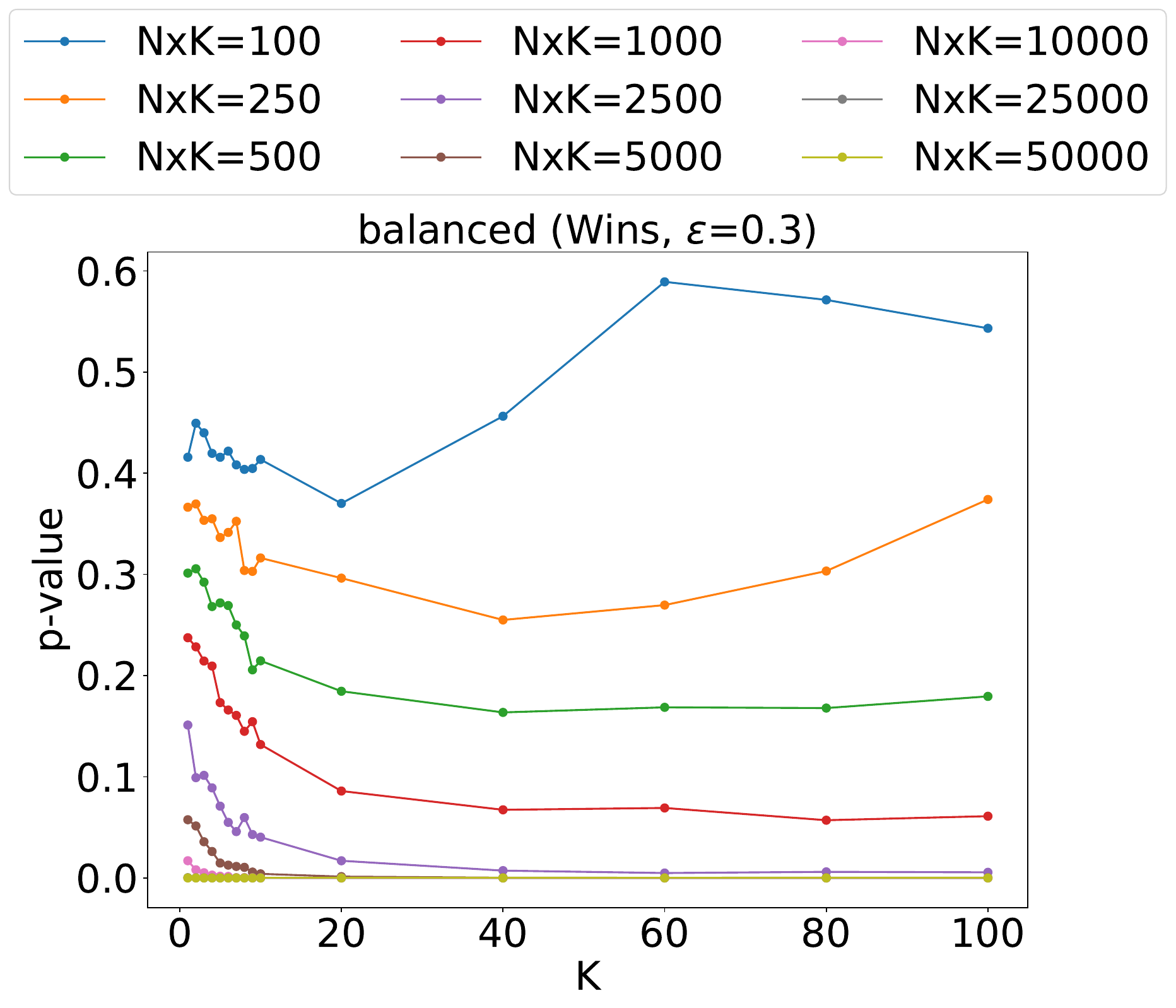}
    \caption{$\epsilon = 0.3$}
    \label{fig:uniform_wins_cat3_e03}
  \end{subfigure} \hfill
  \begin{subfigure}[b]{0.24\linewidth}
    \centering
    \includegraphics[width=\linewidth]{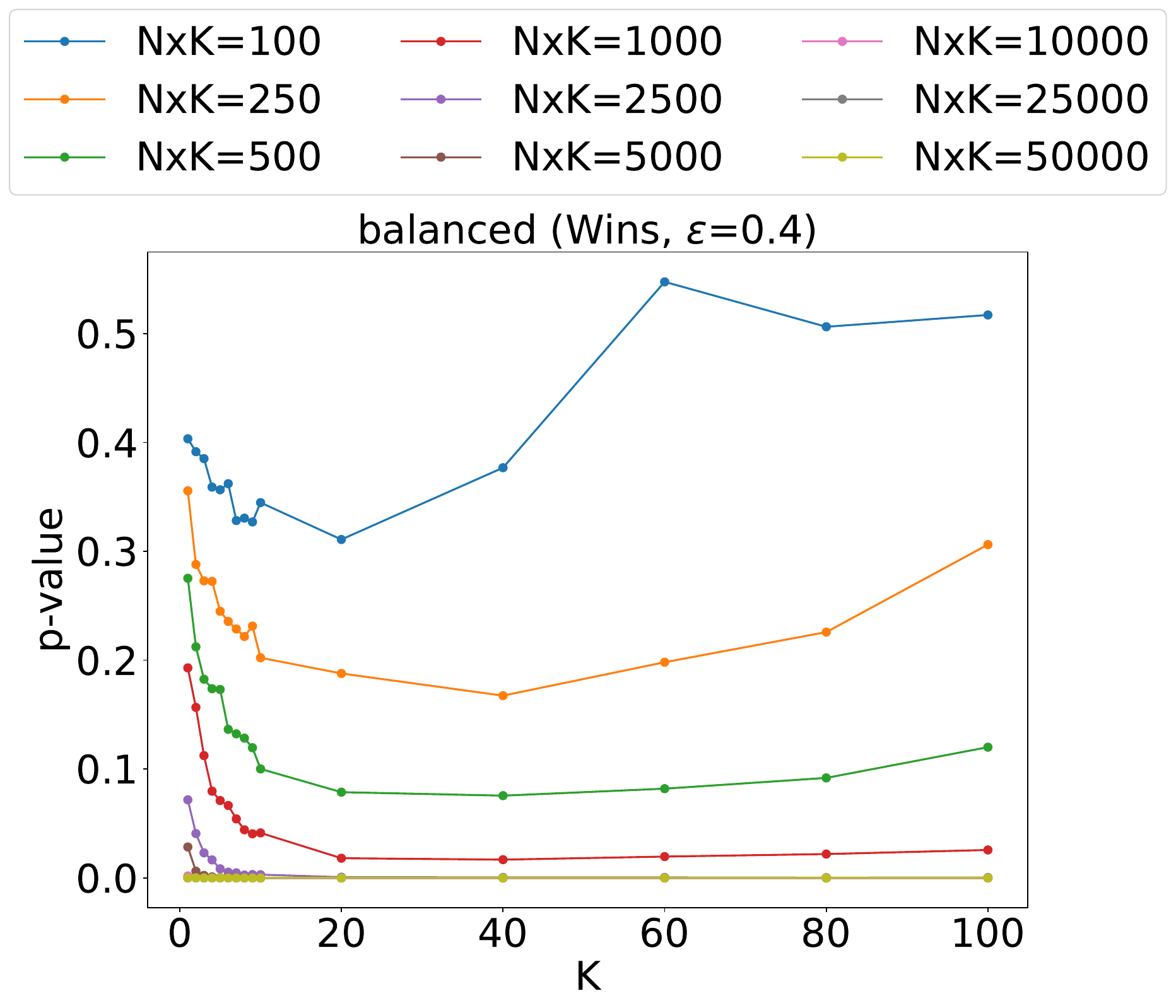}
    \caption{$\epsilon = 0.4$}
    \label{fig:uniform_wins_cat3_e04}
  \end{subfigure}
  \caption{P-value plots for balanced alphas with Wins as the metric ($M=3$)}
  \label{fig:uniform_wins_cat3}
\end{figure*}

\begin{figure*}
  \centering
  \begin{subfigure}[b]{0.24\linewidth}
    \centering
    \includegraphics[width=\linewidth]{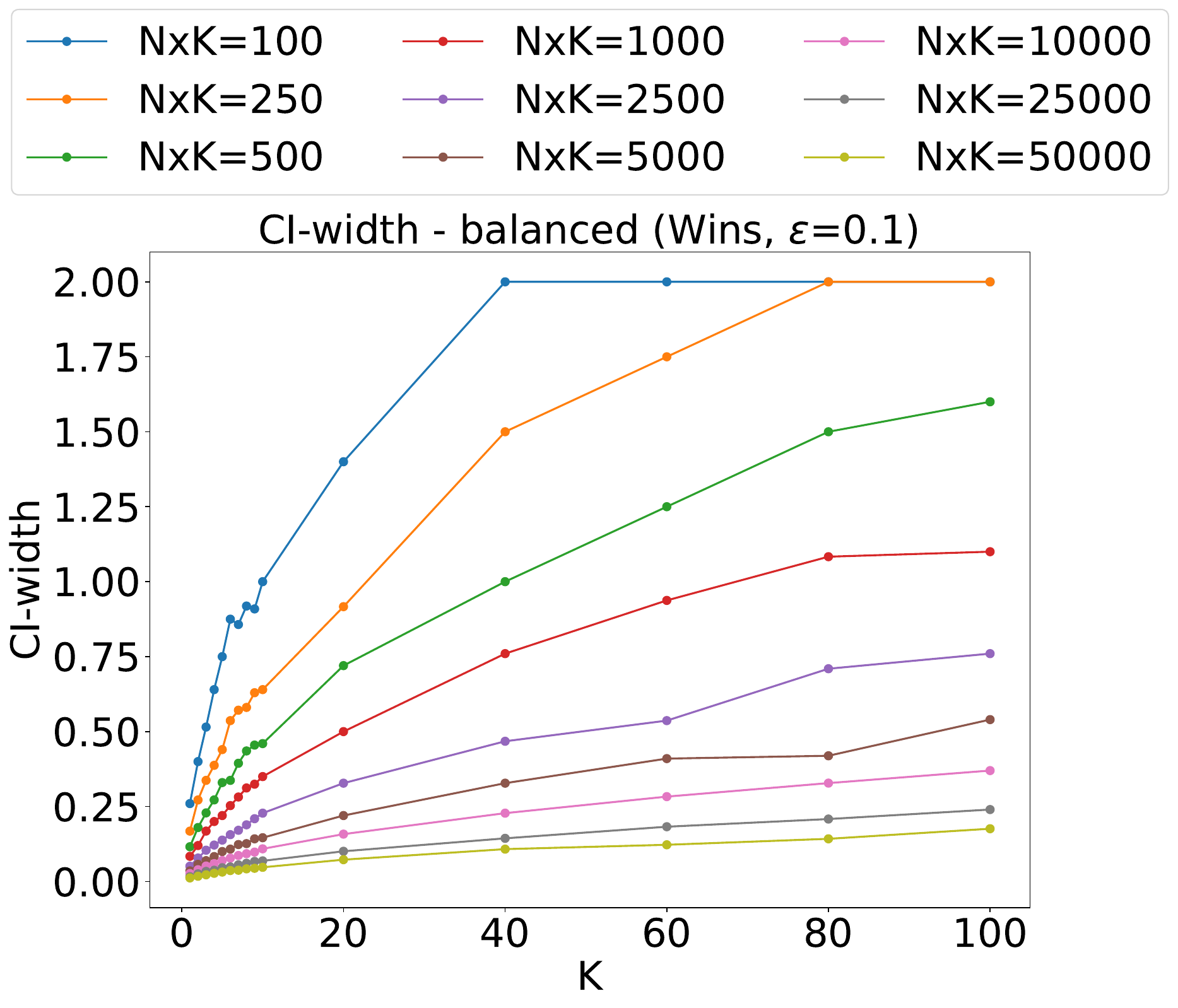}
    \caption{$\epsilon = 0.1$}
    \label{fig:uniform_ci_wins_cat3_e01}
  \end{subfigure} \hfill
  \begin{subfigure}[b]{0.24\linewidth}
    \centering
    \includegraphics[width=\linewidth]{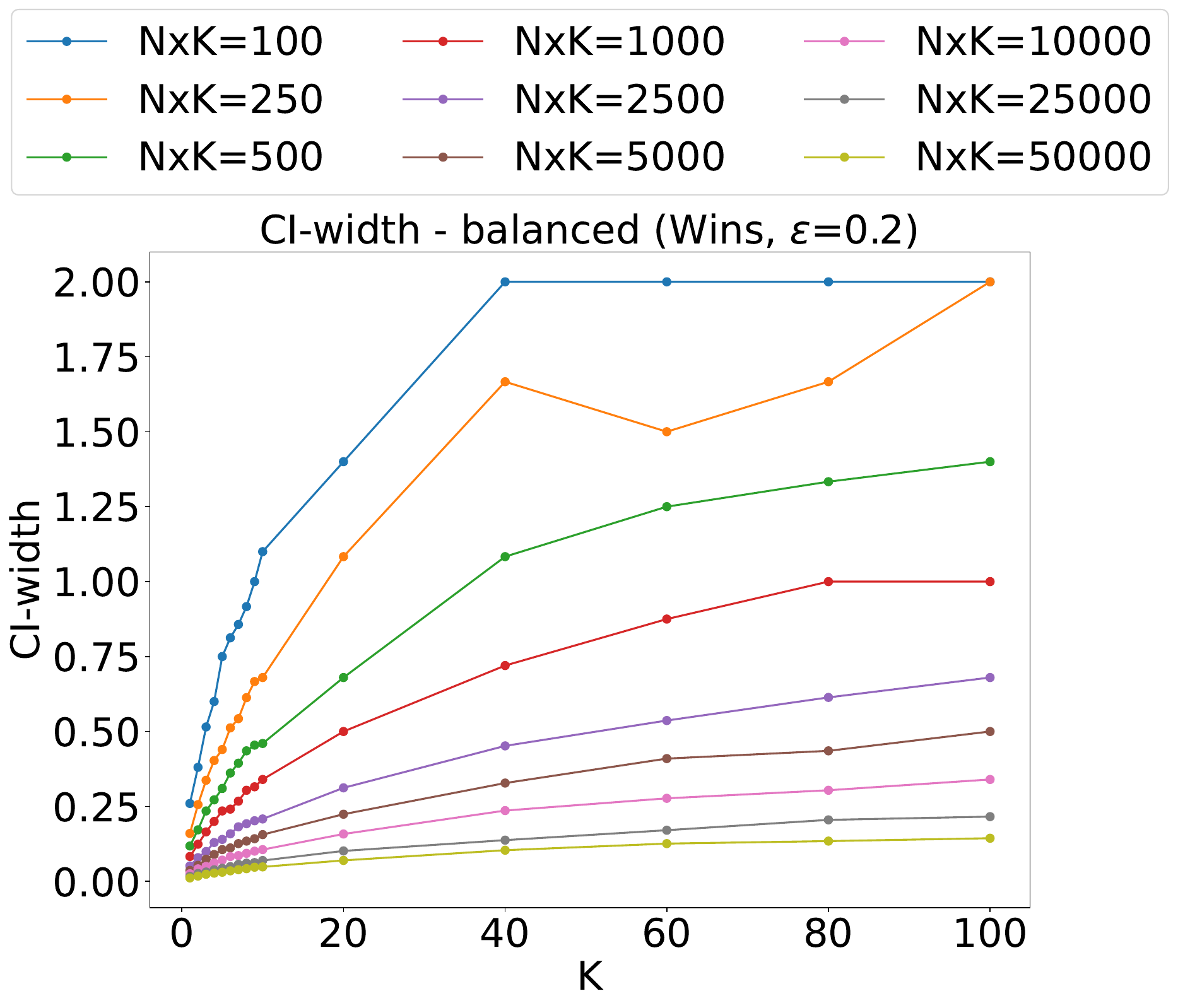}
    \caption{$\epsilon = 0.2$}
    \label{fig:uniform_ci_wins_cat3_e02}
  \end{subfigure} \hfill
  \begin{subfigure}[b]{0.24\linewidth}
    \centering
    \includegraphics[width=\linewidth]{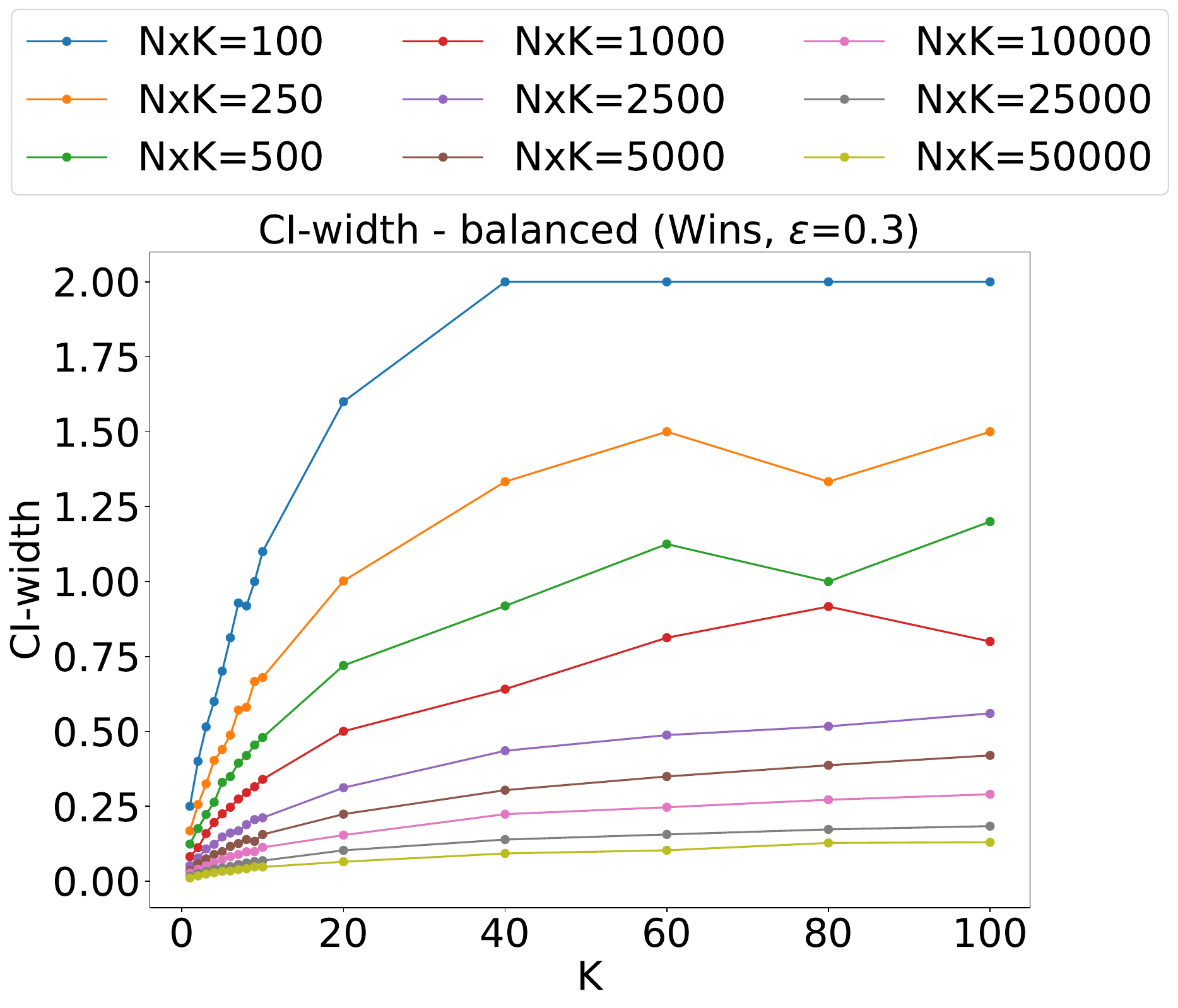}
    \caption{$\epsilon = 0.3$}
    \label{fig:uniform_ci_wins_cat3_e03}
  \end{subfigure} \hfill
  \begin{subfigure}[b]{0.24\linewidth}
    \centering
    \includegraphics[width=\linewidth]{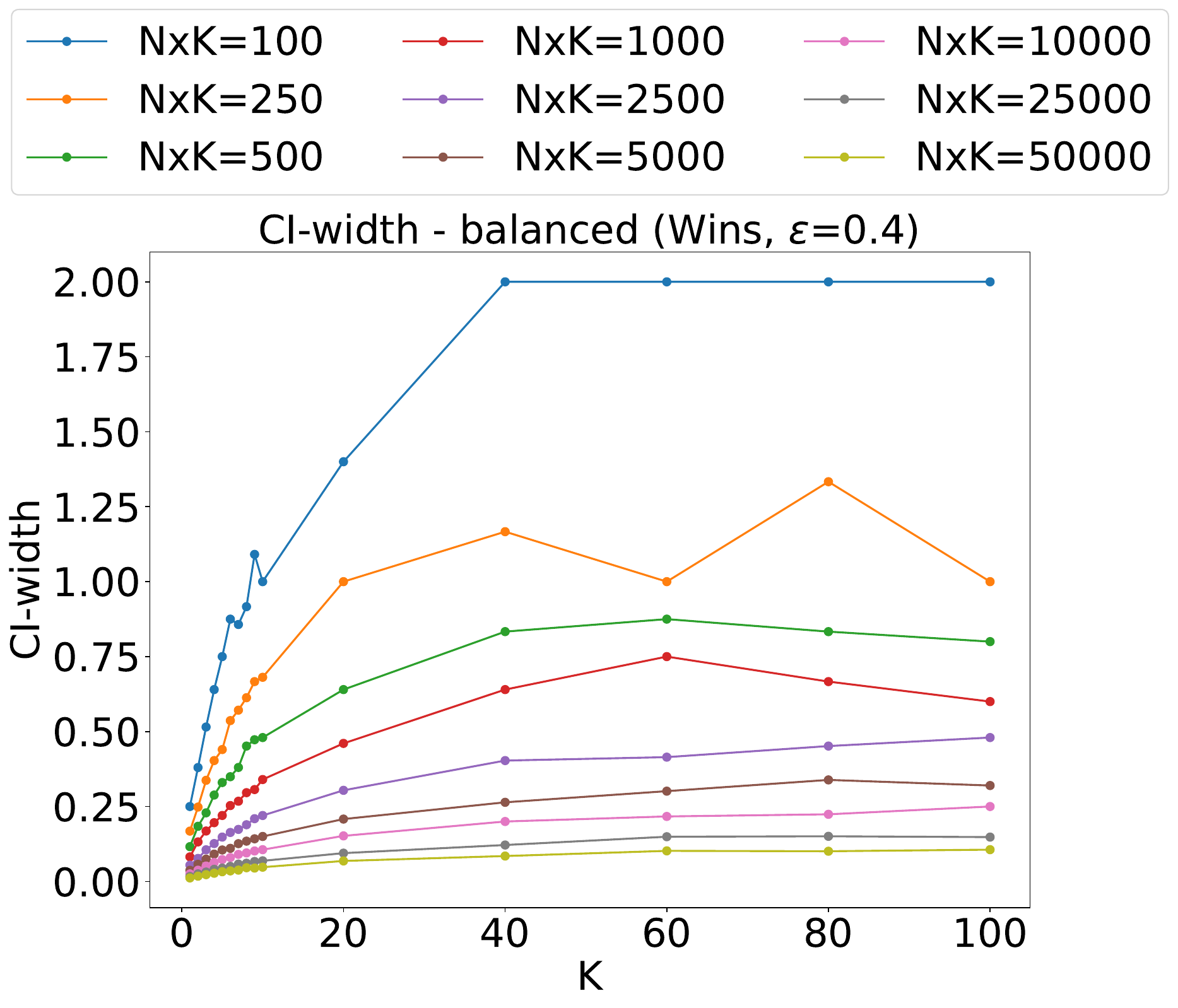}
    \caption{$\epsilon = 0.4$}
    \label{fig:uniform_ci_wins_cat3_e04}
  \end{subfigure}
  \caption{CI-width plots for balanced alphas with Wins as the metric ($M=3$)}
  \label{fig:uniform_ci_wins_cat3}
\end{figure*}

\begin{figure*}
  \centering
  \begin{subfigure}[b]{0.24\linewidth}
    \centering
    \includegraphics[width=\linewidth]{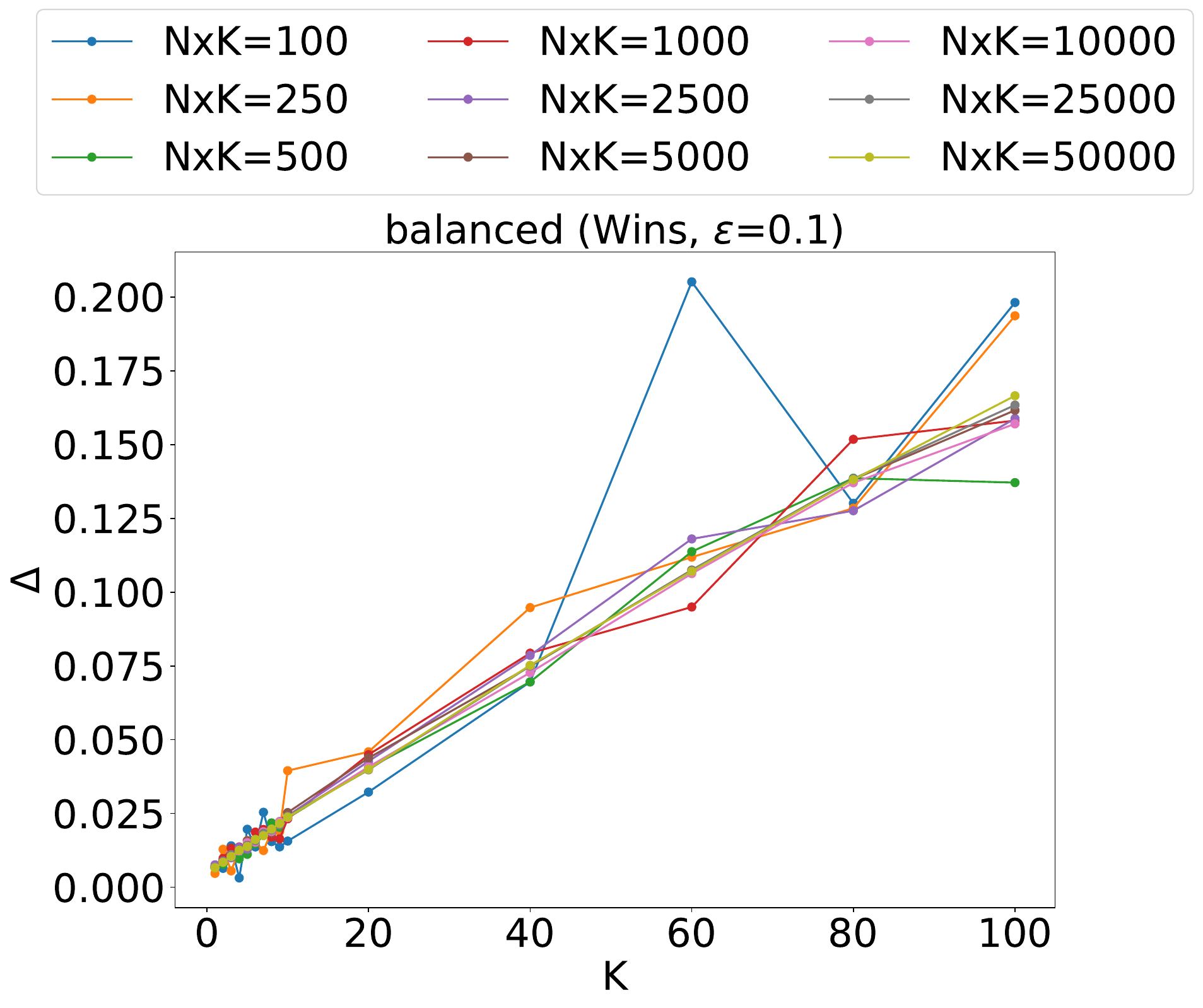}
    \caption{$\epsilon = 0.1$}
    \label{fig:uniform_delta_wins_cat3_e01}
  \end{subfigure} \hfill
  \begin{subfigure}[b]{0.24\linewidth}
    \centering
    \includegraphics[width=\linewidth]{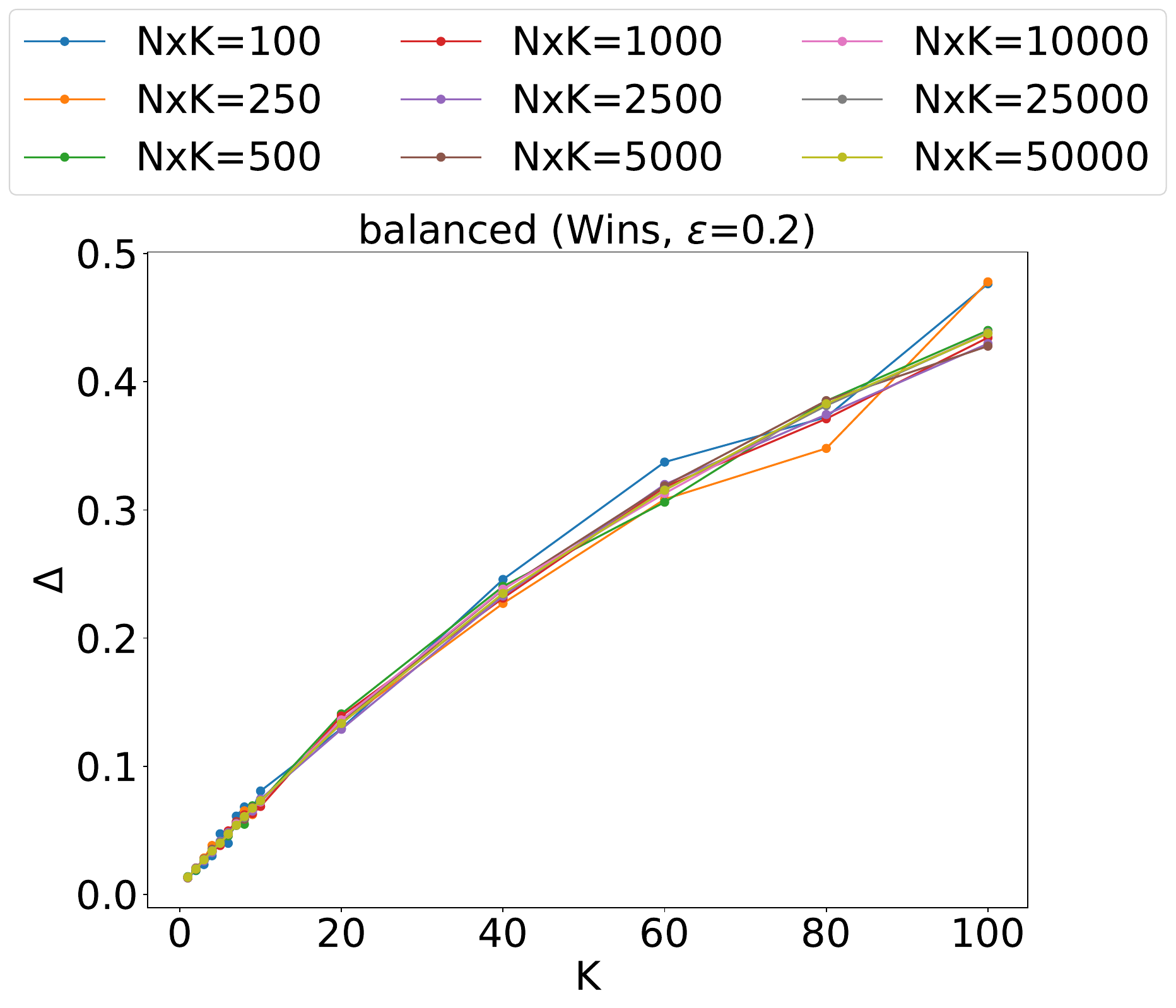}
    \caption{$\epsilon = 0.2$}
    \label{fig:uniform_delta_wins_cat3_e02}
  \end{subfigure} \hfill
  \begin{subfigure}[b]{0.24\linewidth}
    \centering
    \includegraphics[width=\linewidth]{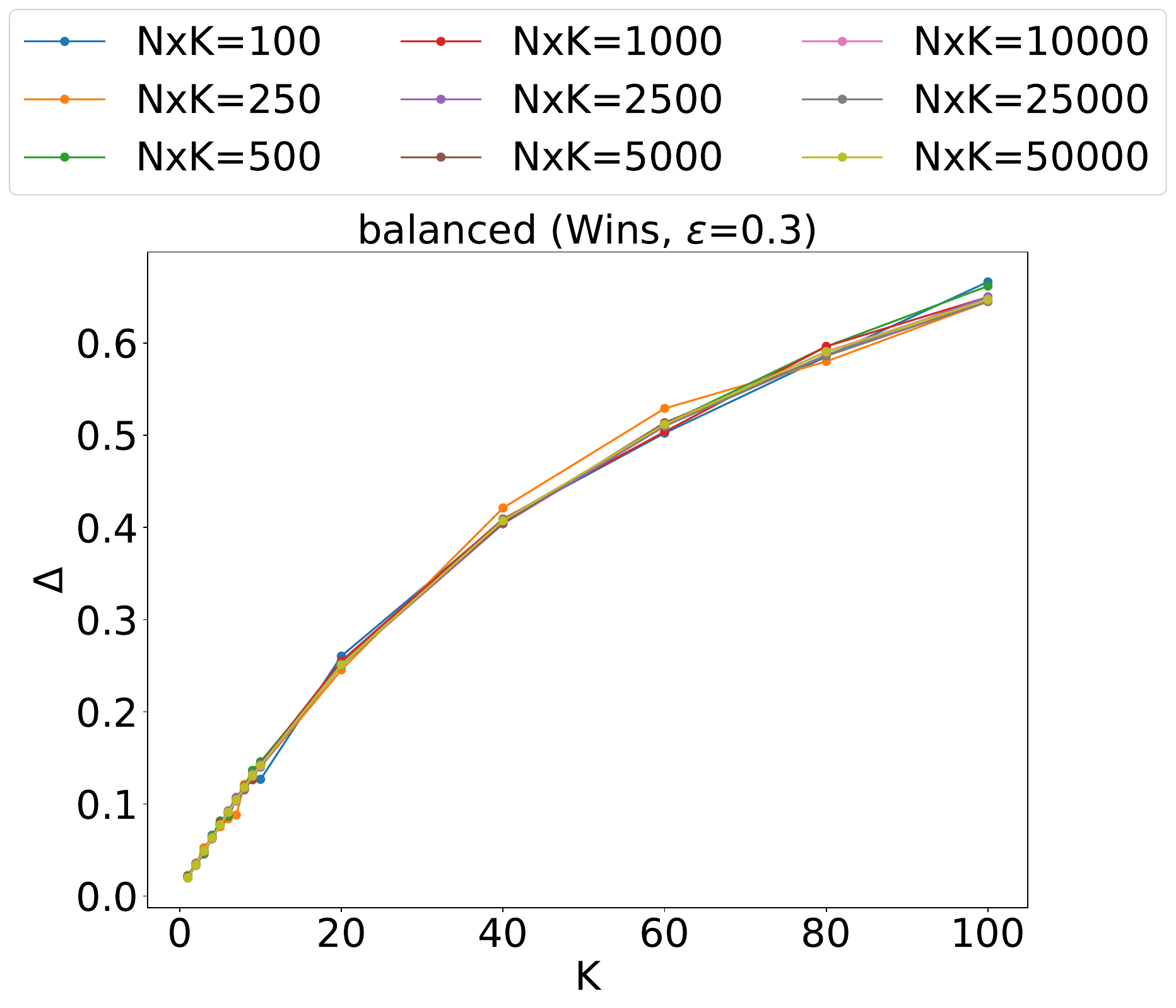}
    \caption{$\epsilon = 0.3$}
    \label{fig:uniform_delta_wins_cat3_e03}
  \end{subfigure} \hfill
  \begin{subfigure}[b]{0.24\linewidth}
    \centering
    \includegraphics[width=\linewidth]{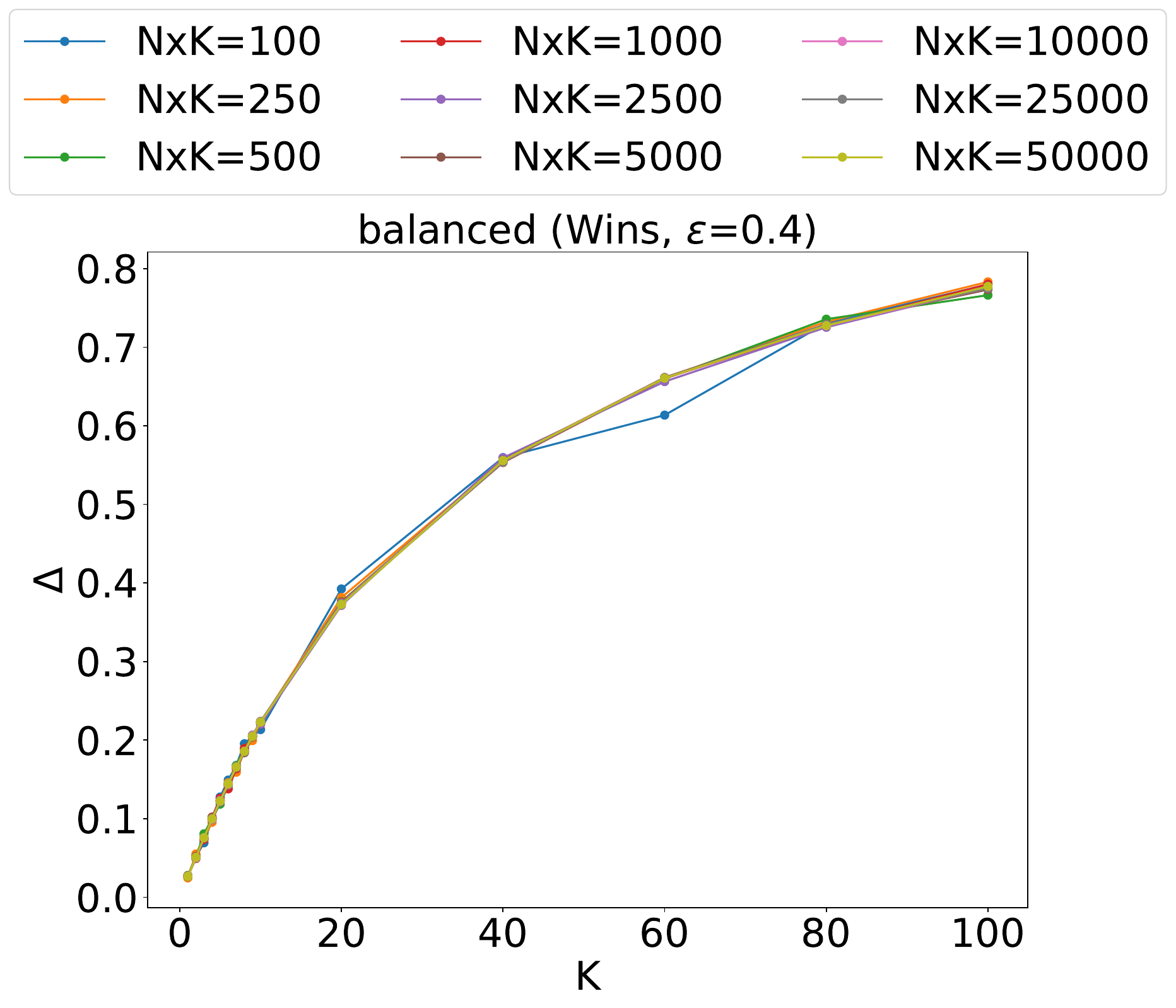}
    \caption{$\epsilon = 0.4$}
    \label{fig:uniform_delta_wins_cat3_e04}
  \end{subfigure}
  \caption{Effect sizes ($\Delta$) for balanced alphas with Wins as the metric ($M=3$)}
  \label{fig:uniform_delta_wins_cat3}
\end{figure*}

\begin{figure*}
  \centering
  \begin{subfigure}[b]{0.24\linewidth}
    \centering
    \includegraphics[width=\linewidth]{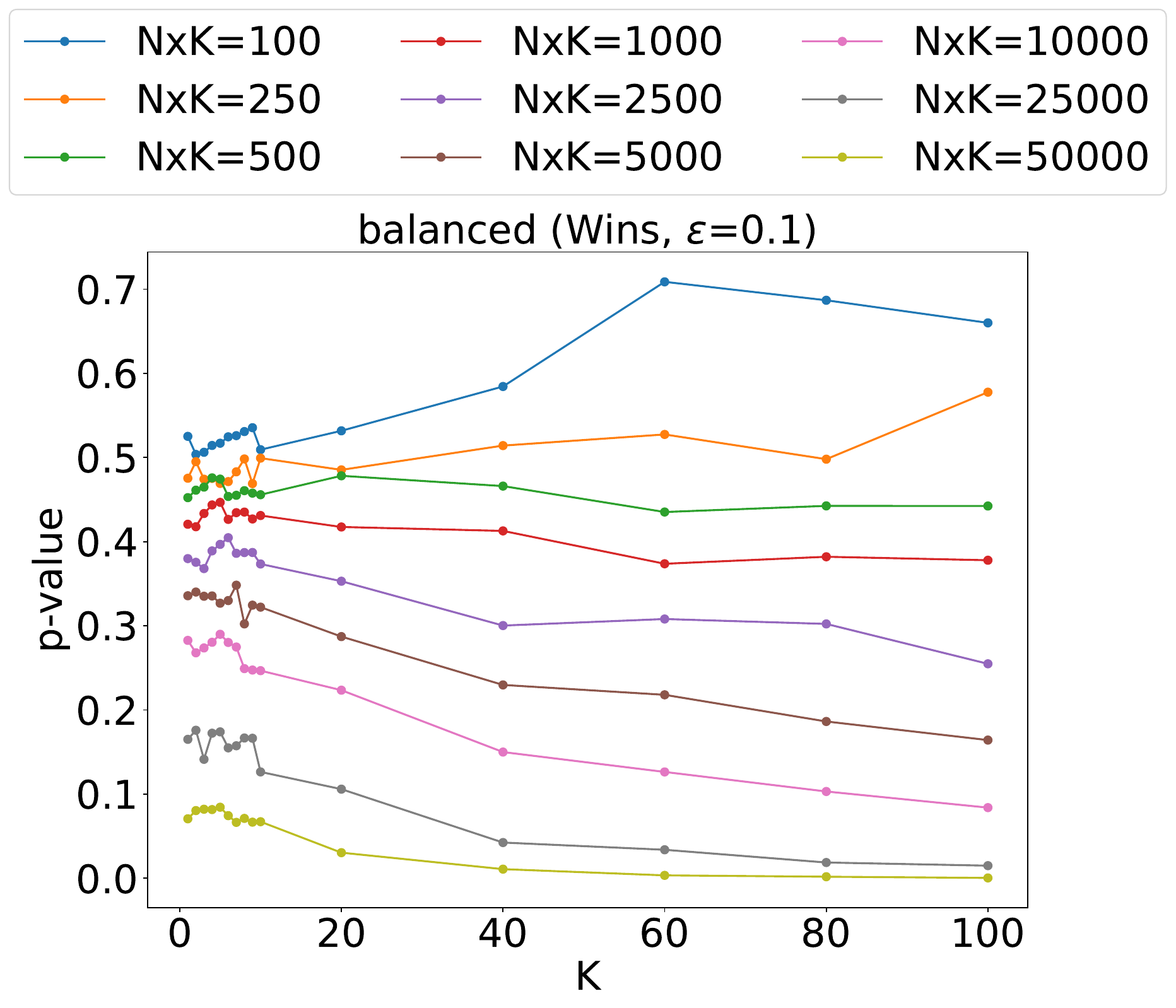}
    \caption{$\epsilon = 0.1$}
    \label{fig:uniform_wins_cat4_e01}
  \end{subfigure} \hfill
  \begin{subfigure}[b]{0.24\linewidth}
    \centering
    \includegraphics[width=\linewidth]{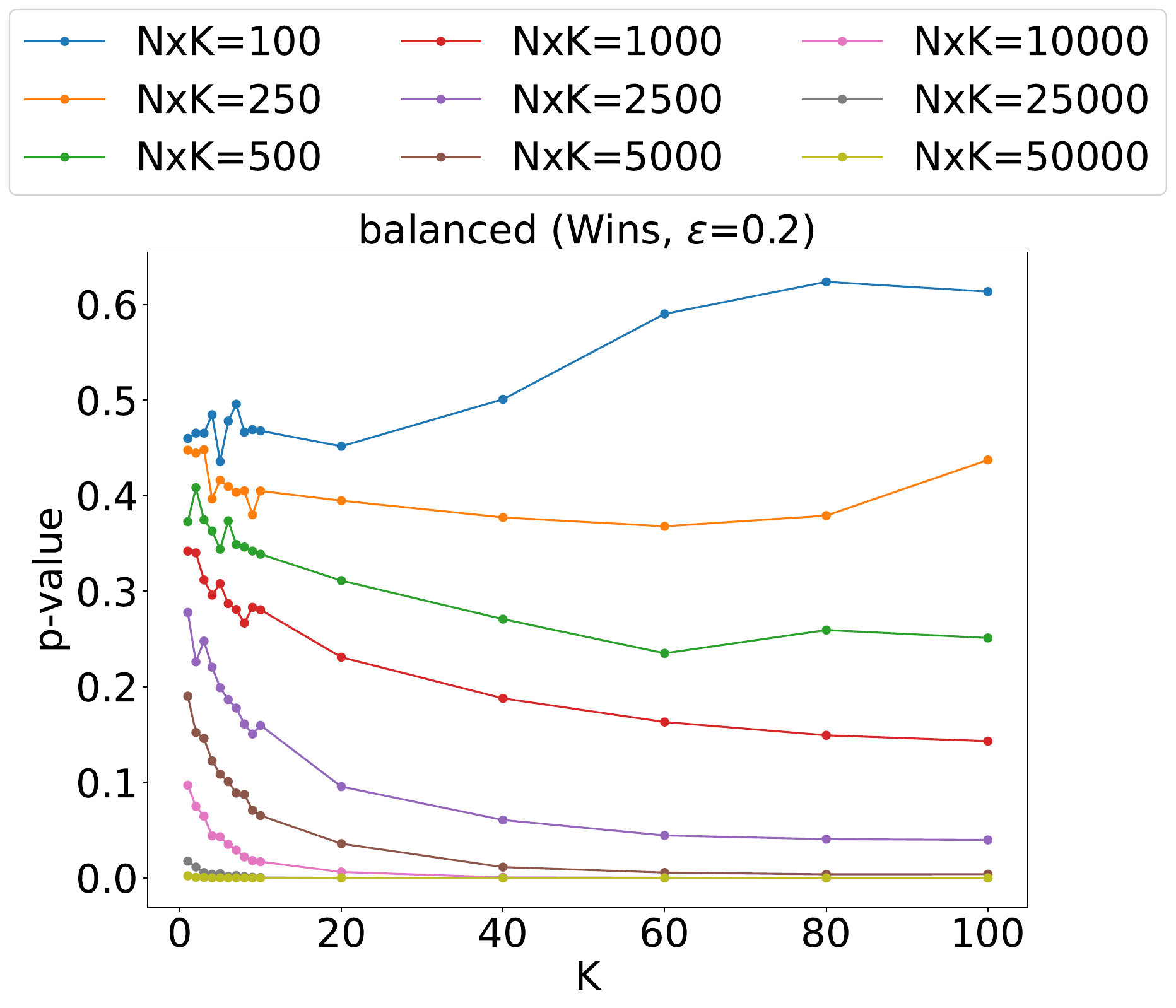}
    \caption{$\epsilon = 0.2$}
    \label{fig:uniform_wins_cat4_e02}
  \end{subfigure} \hfill
  \begin{subfigure}[b]{0.24\linewidth}
    \centering
    \includegraphics[width=\linewidth]{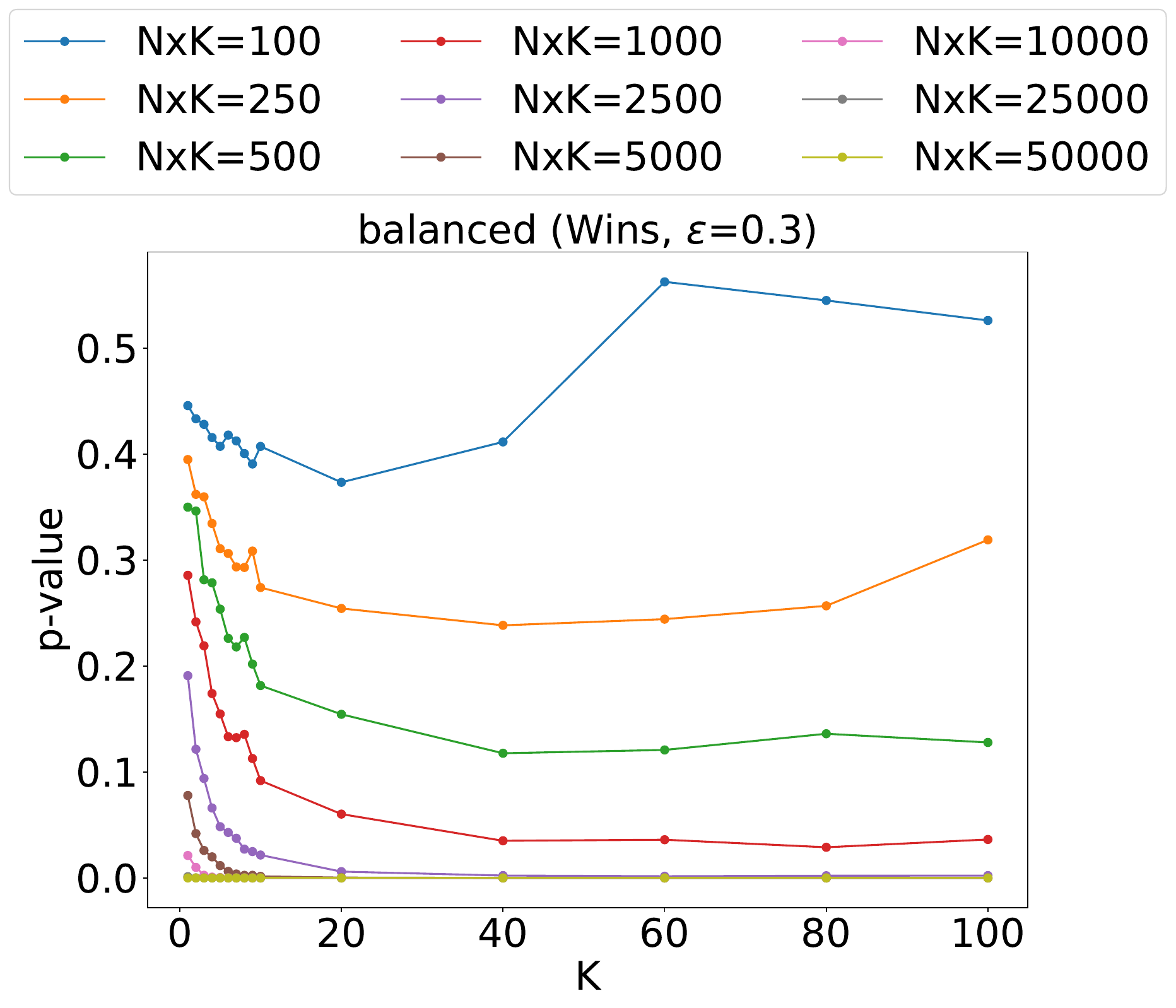}
    \caption{$\epsilon = 0.3$}
    \label{fig:uniform_wins_cat4_e03}
  \end{subfigure} \hfill
  \begin{subfigure}[b]{0.24\linewidth}
    \centering
    \includegraphics[width=\linewidth]{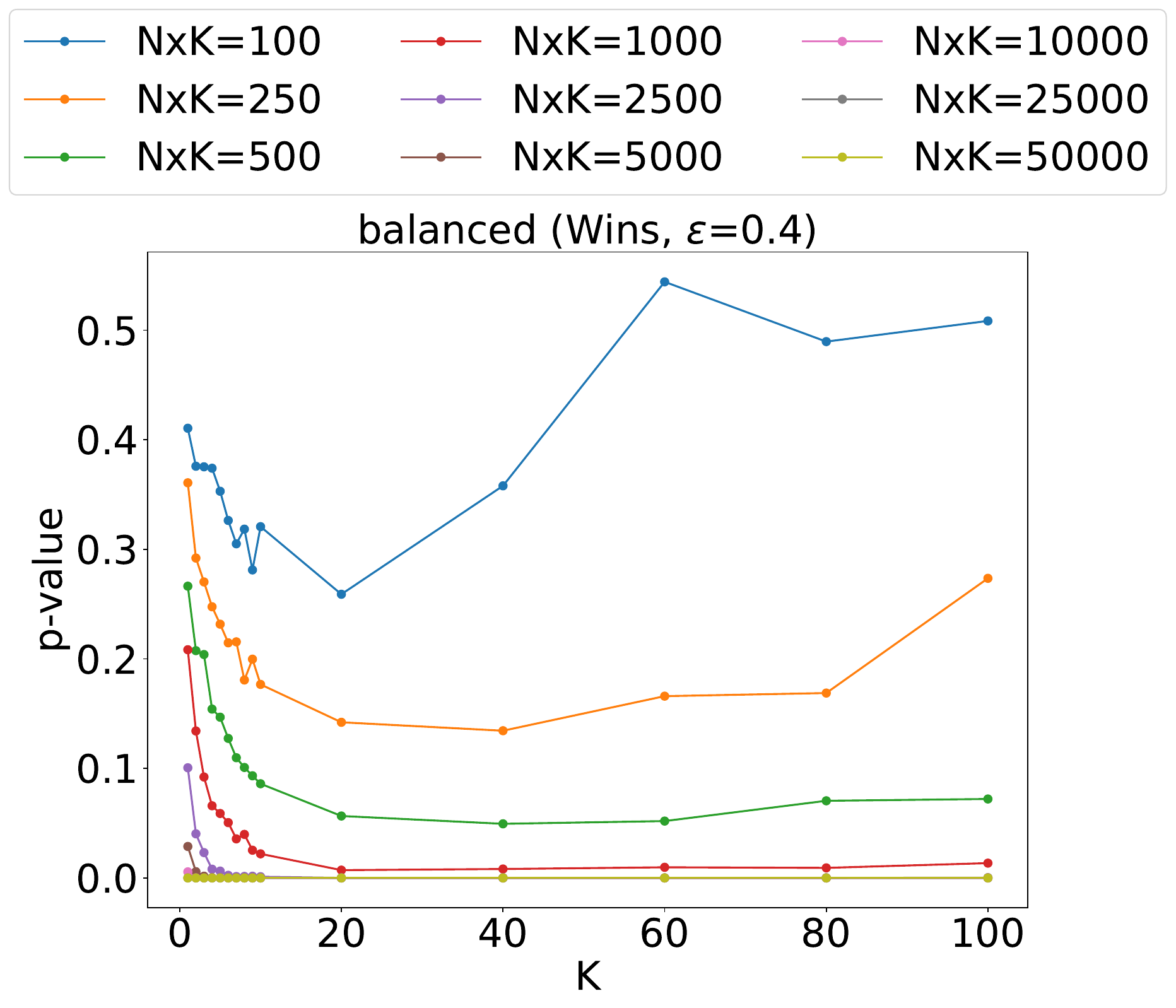}
    \caption{$\epsilon = 0.4$}
    \label{fig:uniform_wins_cat4_e04}
  \end{subfigure}
  \caption{P-value plots for balanced alphas with Wins as the metric ($M=4$)}
  \label{fig:uniform_wins_cat4}
\end{figure*}

\begin{figure*}
  \centering
  \begin{subfigure}[b]{0.24\linewidth}
    \centering
    \includegraphics[width=\linewidth]{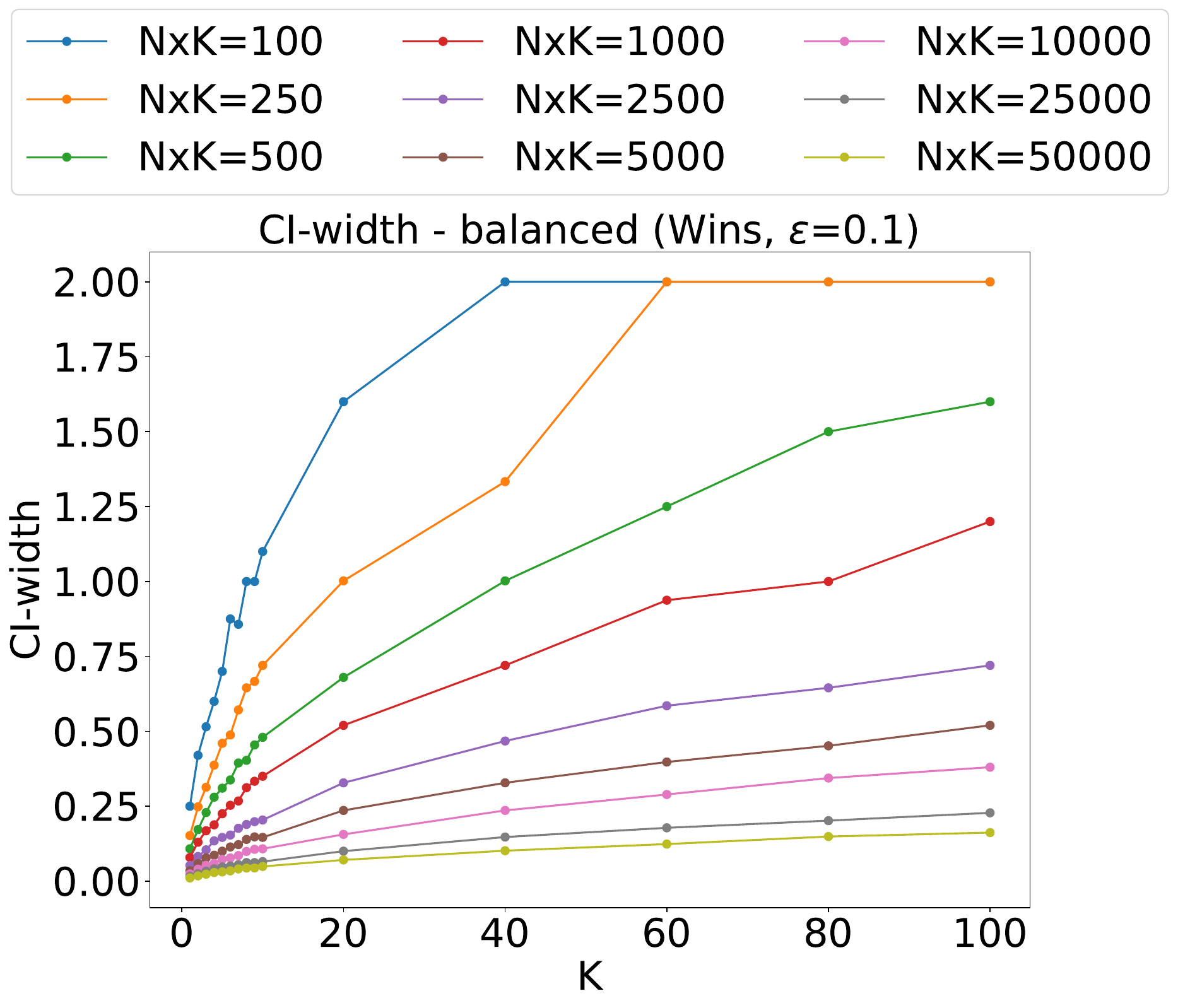}
    \caption{$\epsilon = 0.1$}
    \label{fig:uniform_ci_wins_cat4_e01}
  \end{subfigure} \hfill
  \begin{subfigure}[b]{0.24\linewidth}
    \centering
    \includegraphics[width=\linewidth]{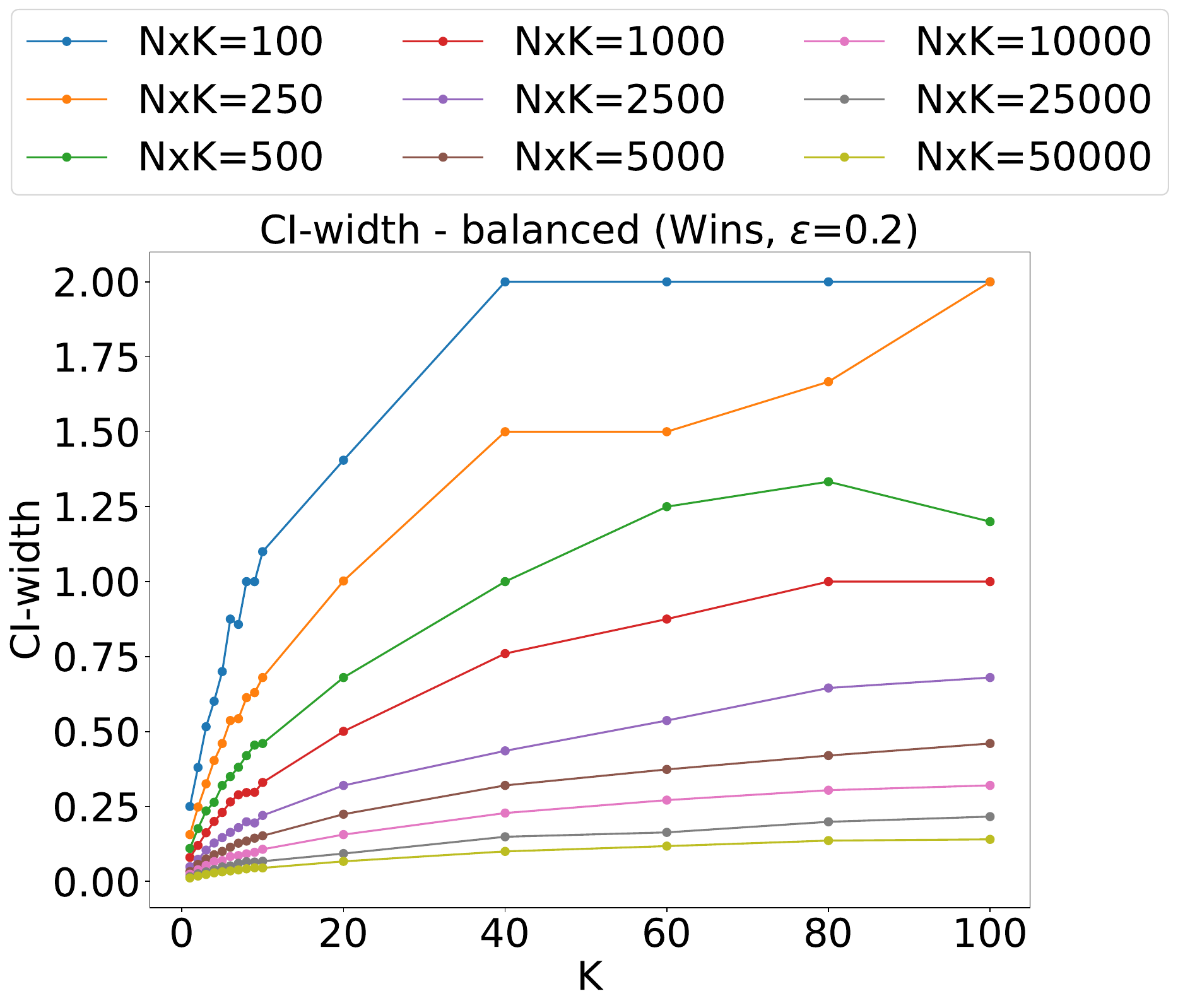}
    \caption{$\epsilon = 0.2$}
    \label{fig:uniform_ci_wins_cat4_e02}
  \end{subfigure} \hfill
  \begin{subfigure}[b]{0.24\linewidth}
    \centering
    \includegraphics[width=\linewidth]{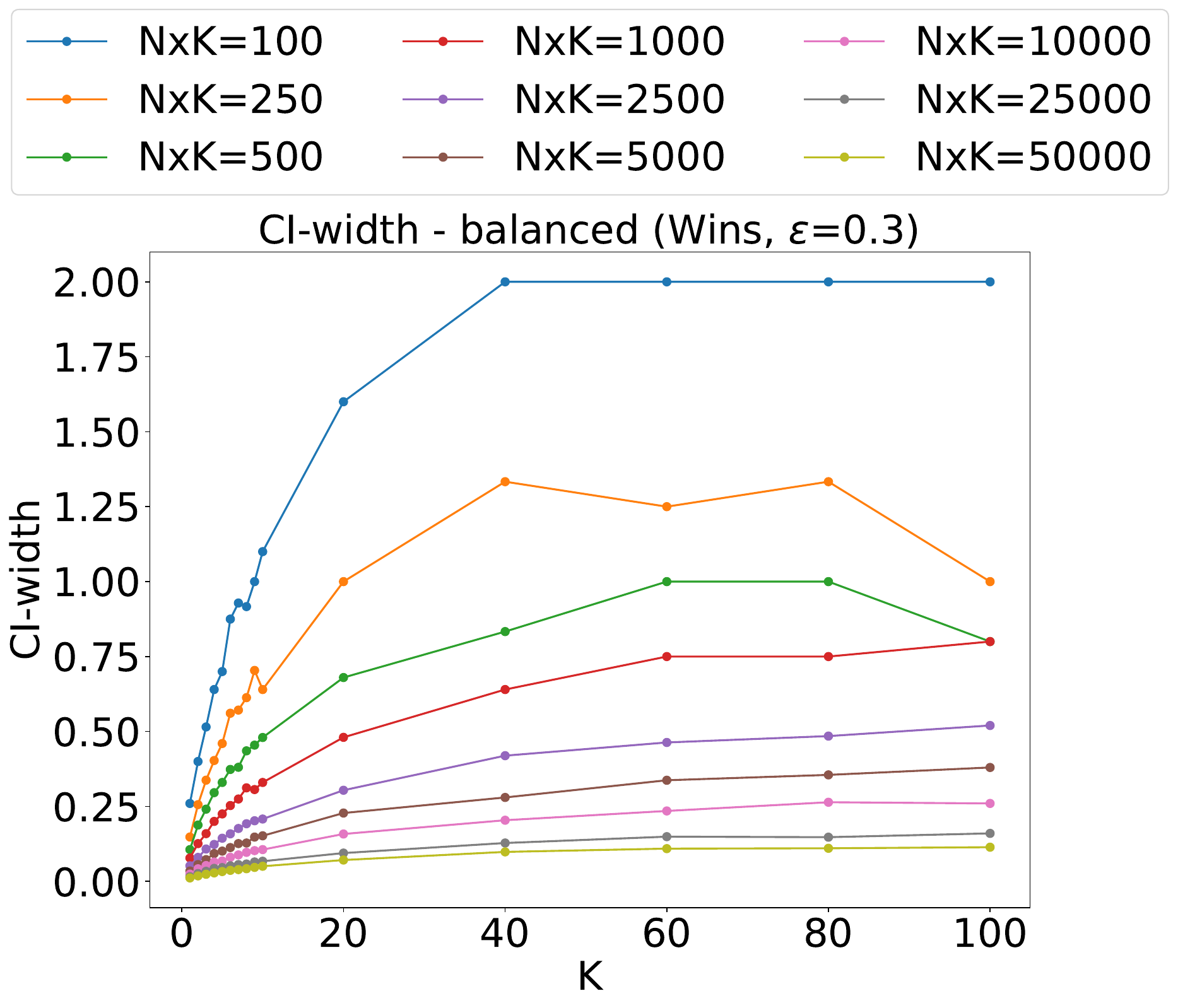}
    \caption{$\epsilon = 0.3$}
    \label{fig:uniform_ci_wins_cat4_e03}
  \end{subfigure} \hfill
  \begin{subfigure}[b]{0.24\linewidth}
    \centering
    \includegraphics[width=\linewidth]{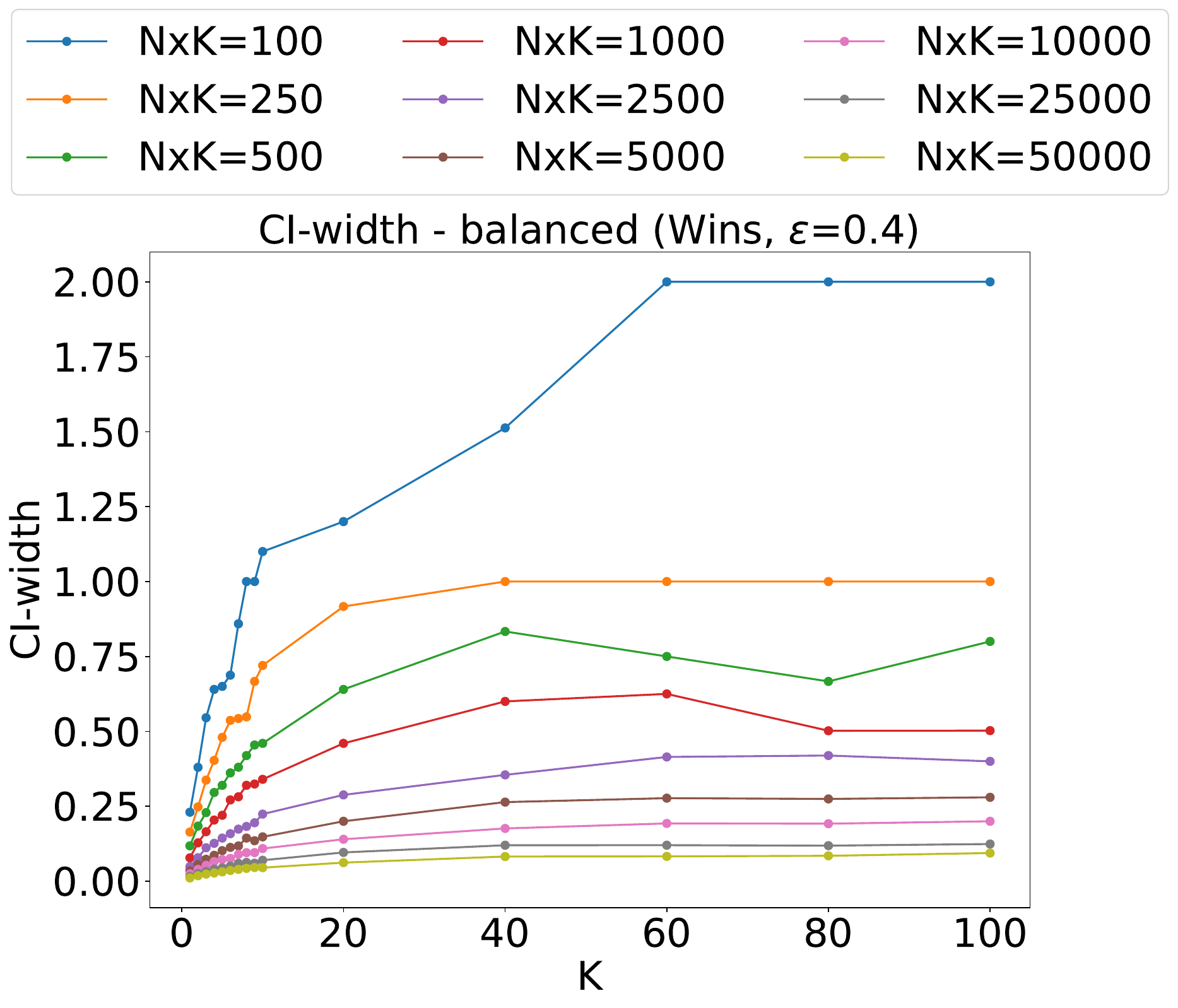}
    \caption{$\epsilon = 0.4$}
    \label{fig:uniform_ci_wins_cat4_e04}
  \end{subfigure}
  \caption{CI-width plots for balanced alphas with Wins as the metric ($M=4$)}
  \label{fig:uniform_ci_wins_cat4}
\end{figure*}

\begin{figure*}
  \centering
  \begin{subfigure}[b]{0.24\linewidth}
    \centering
    \includegraphics[width=\linewidth]{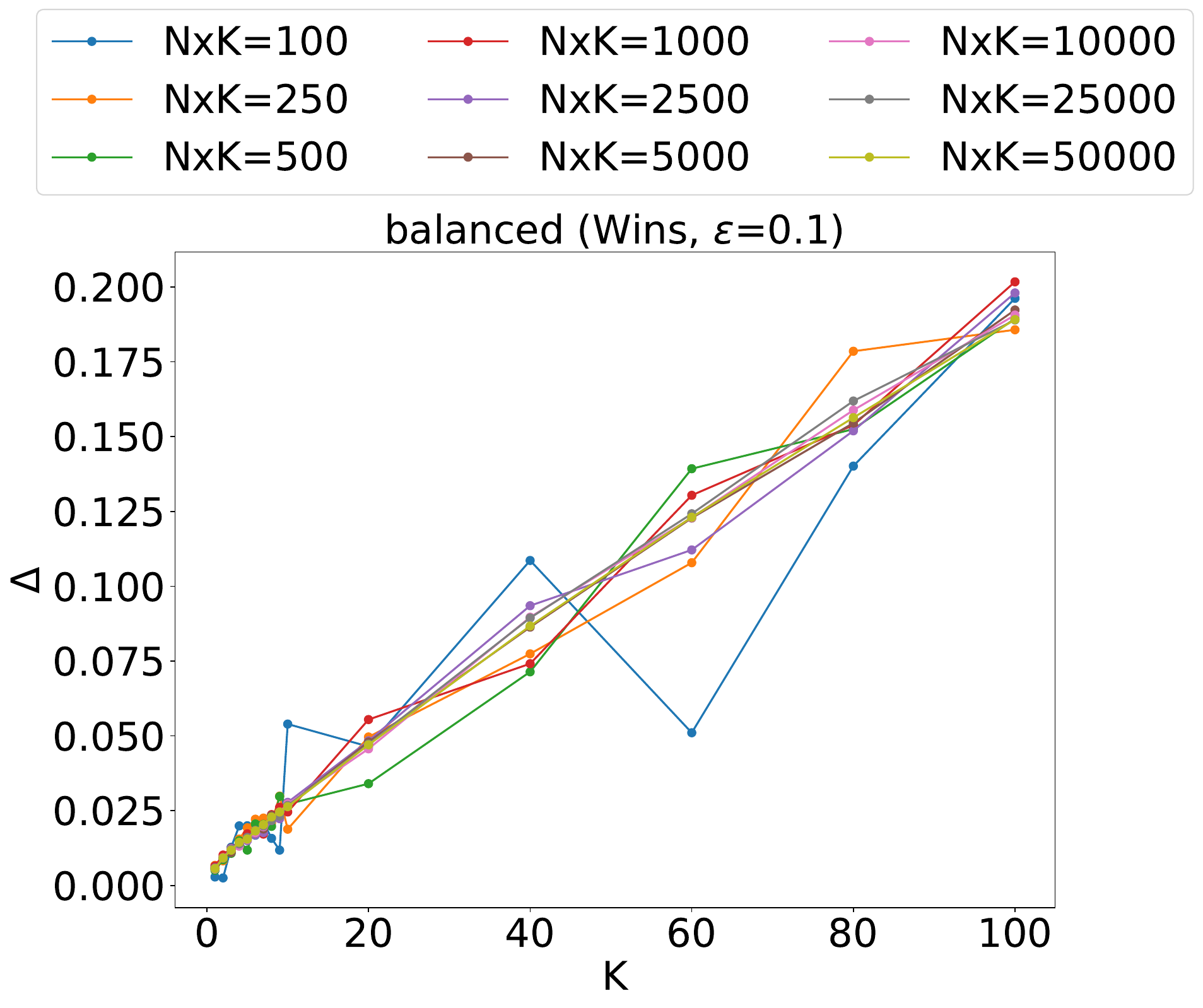}
    \caption{$\epsilon = 0.1$}
    \label{fig:uniform_delta_wins_cat4_e01}
  \end{subfigure} \hfill
  \begin{subfigure}[b]{0.24\linewidth}
    \centering
    \includegraphics[width=\linewidth]{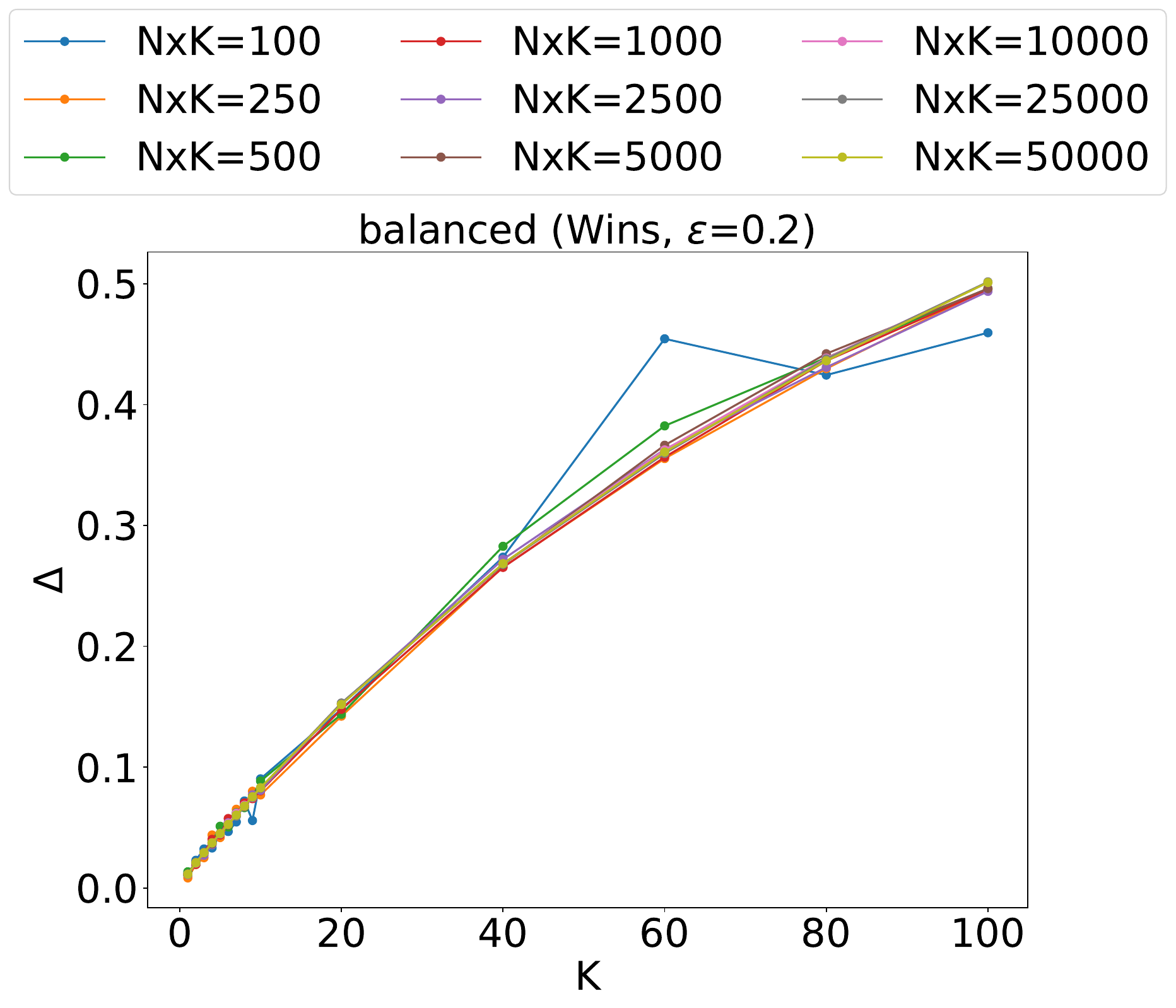}
    \caption{$\epsilon = 0.2$}
    \label{fig:uniform_delta_wins_cat4_e02}
  \end{subfigure} \hfill
  \begin{subfigure}[b]{0.24\linewidth}
    \centering
    \includegraphics[width=\linewidth]{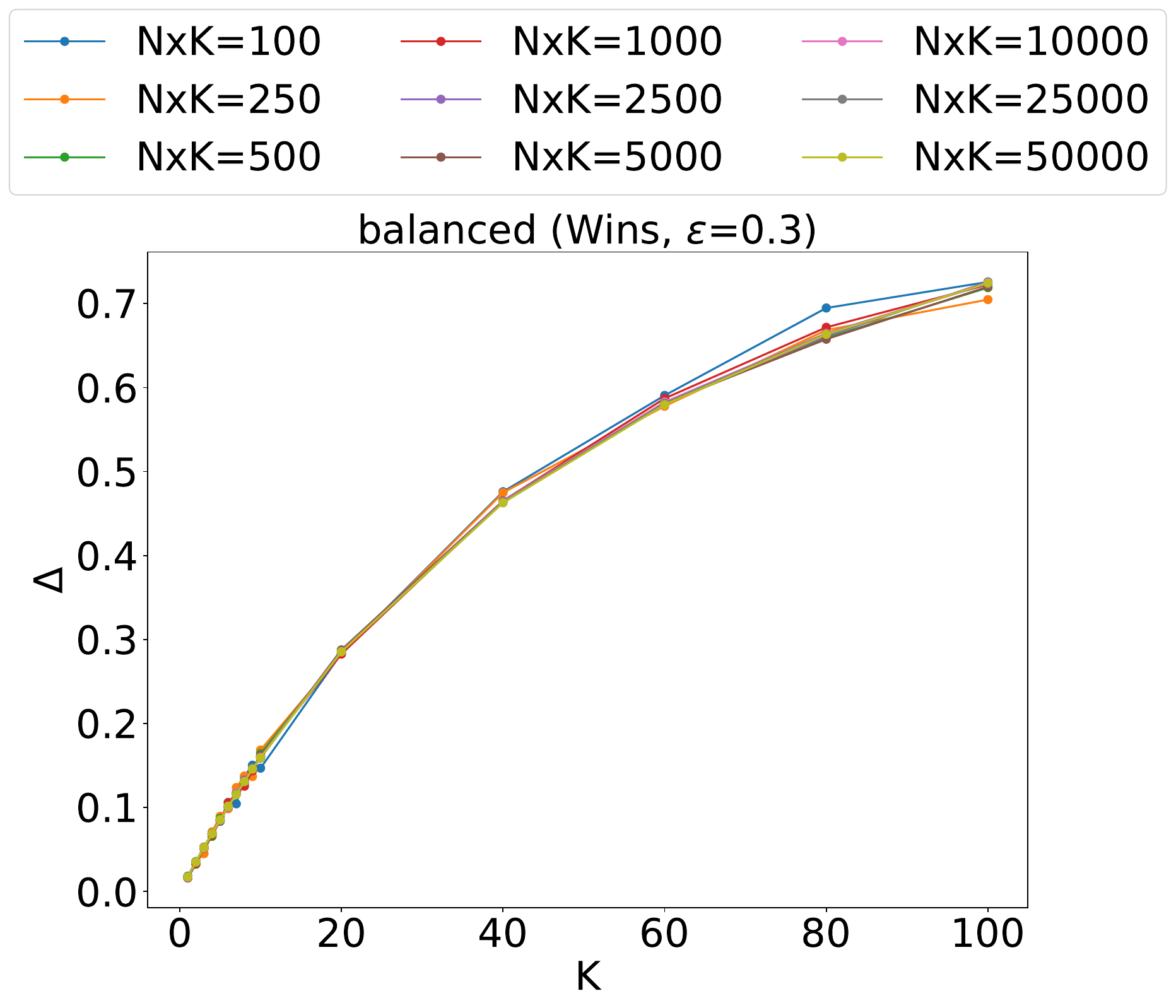}
    \caption{$\epsilon = 0.3$}
    \label{fig:uniform_delta_wins_cat4_e03}
  \end{subfigure} \hfill
  \begin{subfigure}[b]{0.24\linewidth}
    \centering
    \includegraphics[width=\linewidth]{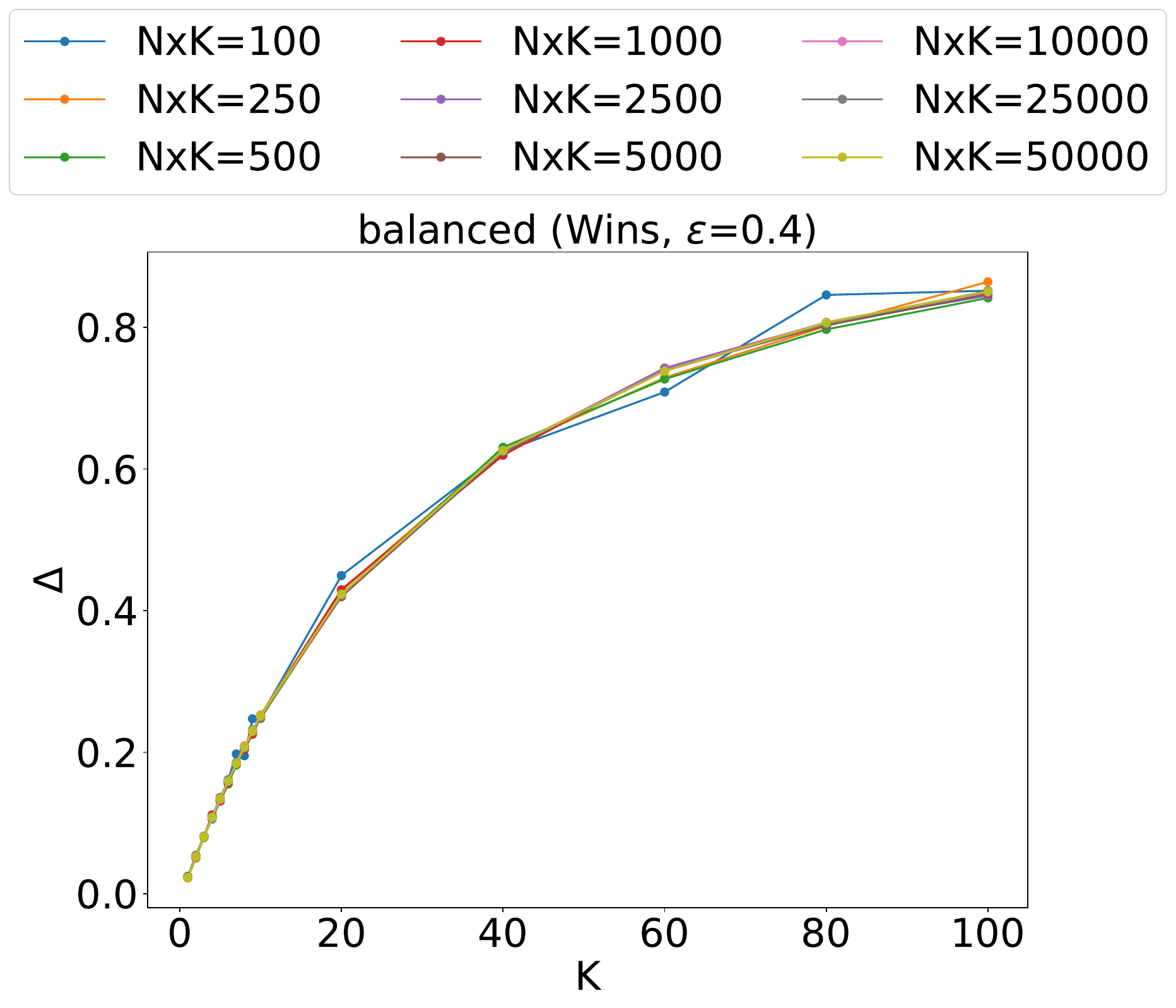}
    \caption{$\epsilon = 0.4$}
    \label{fig:uniform_delta_wins_cat4_e04}
  \end{subfigure}
  \caption{Effect sizes ($\Delta$) for balanced alphas with Wins as the metric ($M=4$)}
  \label{fig:uniform_delta_wins_cat4}
\end{figure*}

\begin{figure*}
  \centering
  \begin{subfigure}[b]{0.24\linewidth}
    \centering
    \includegraphics[width=\linewidth]{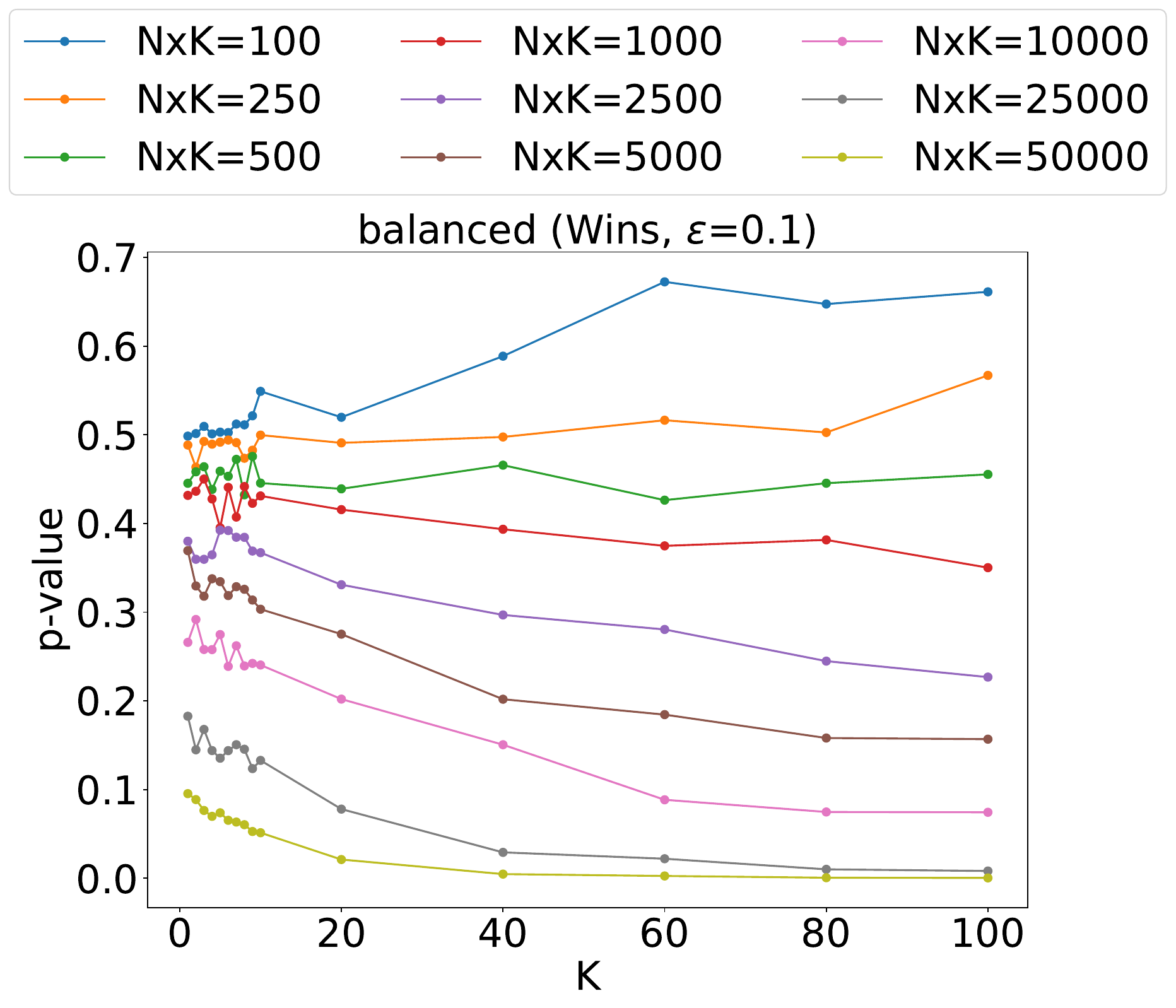}
    \caption{$\epsilon = 0.1$}
    \label{fig:uniform_wins_cat5_e01}
  \end{subfigure} \hfill
  \begin{subfigure}[b]{0.24\linewidth}
    \centering
    \includegraphics[width=\linewidth]{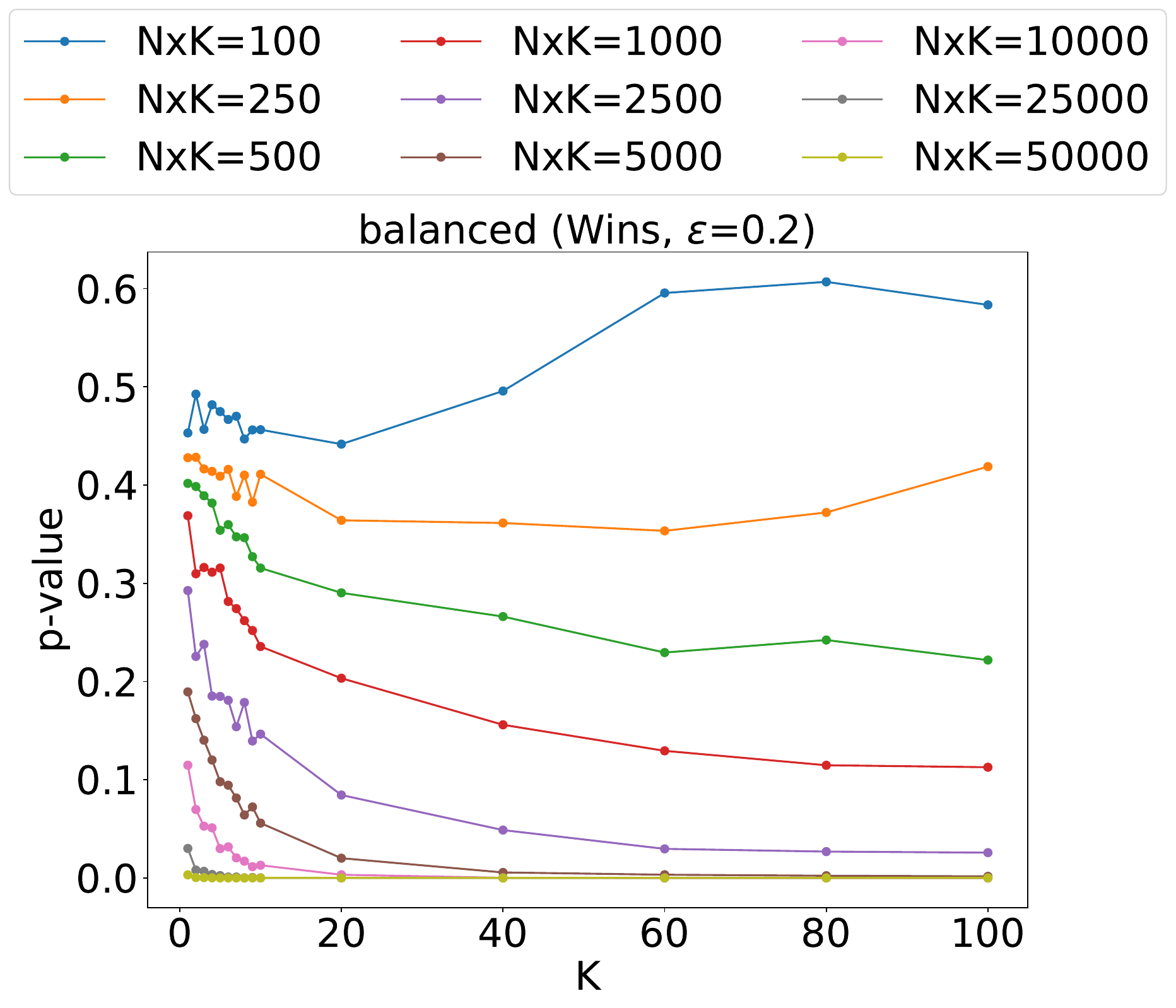}
    \caption{$\epsilon = 0.2$}
    \label{fig:uniform_wins_cat5_e02}
  \end{subfigure} \hfill
  \begin{subfigure}[b]{0.24\linewidth}
    \centering
    \includegraphics[width=\linewidth]{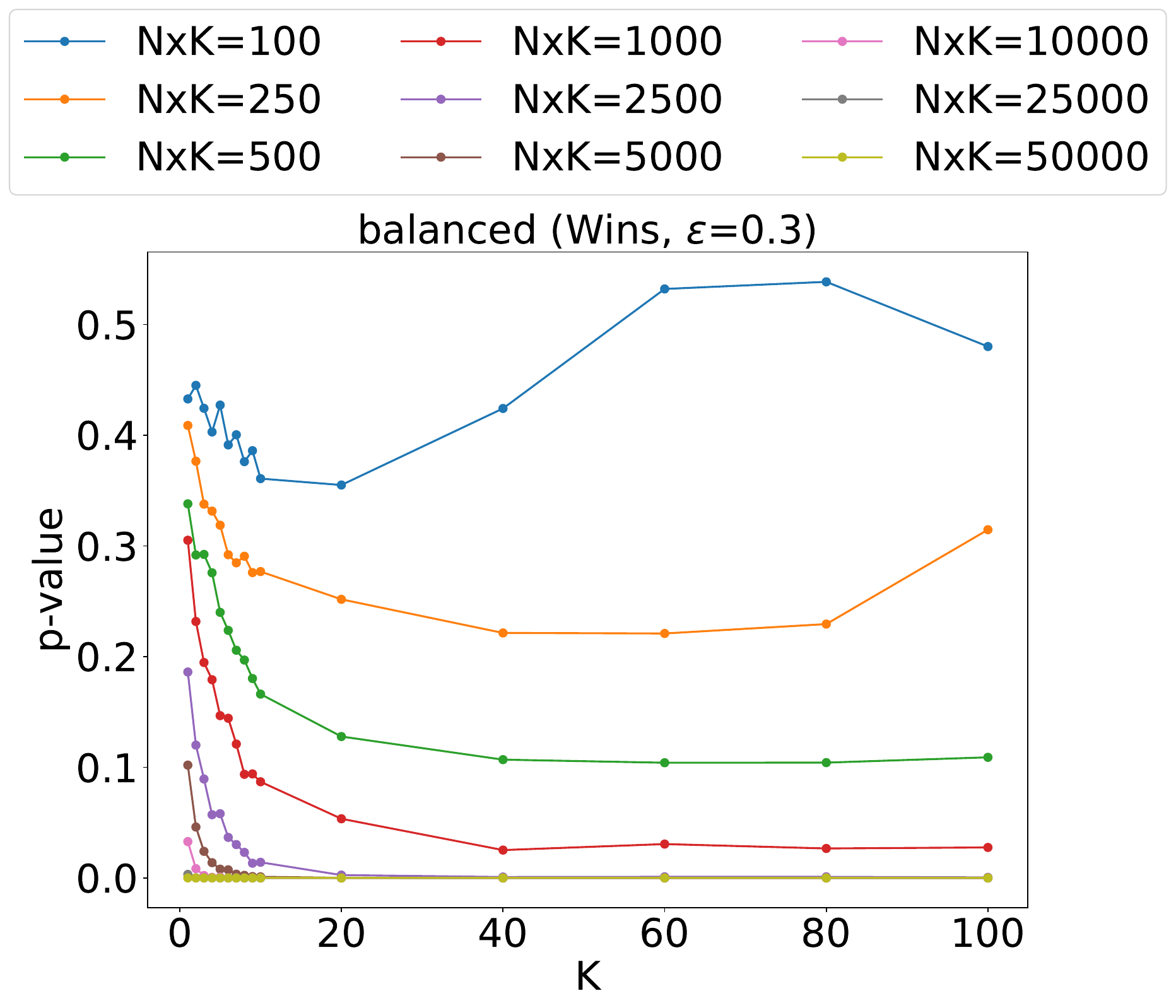}
    \caption{$\epsilon = 0.3$}
    \label{fig:uniform_wins_cat5_e03}
  \end{subfigure} \hfill
  \begin{subfigure}[b]{0.24\linewidth}
    \centering
    \includegraphics[width=\linewidth]{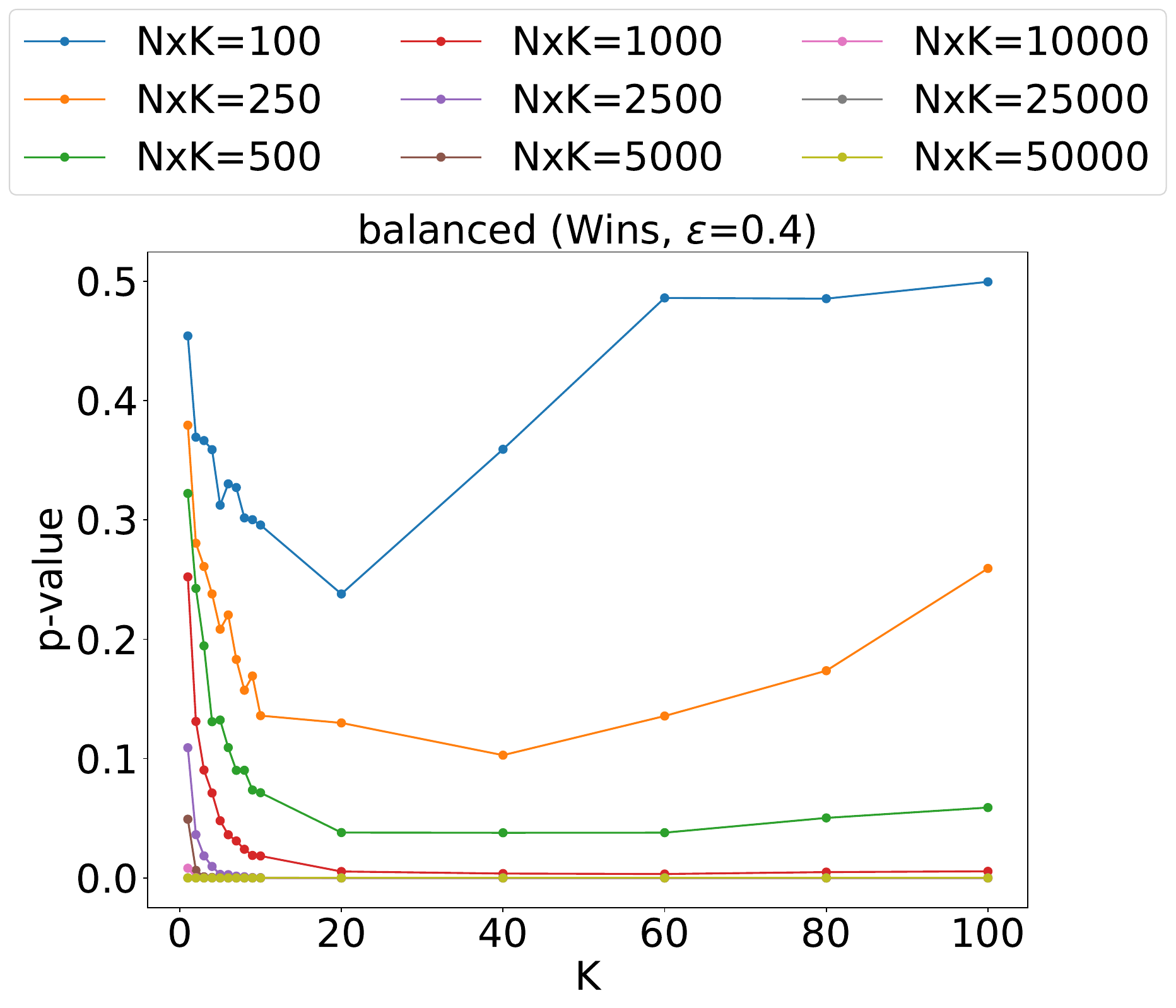}
    \caption{$\epsilon = 0.4$}
    \label{fig:uniform_wins_cat5_e04}
  \end{subfigure}
  \caption{P-value plots for balanced alphas with Wins as the metric ($M=5$)}
  \label{fig:uniform_wins_cat5}
\end{figure*}

\begin{figure*}
  \centering
  \begin{subfigure}[b]{0.24\linewidth}
    \centering
    \includegraphics[width=\linewidth]{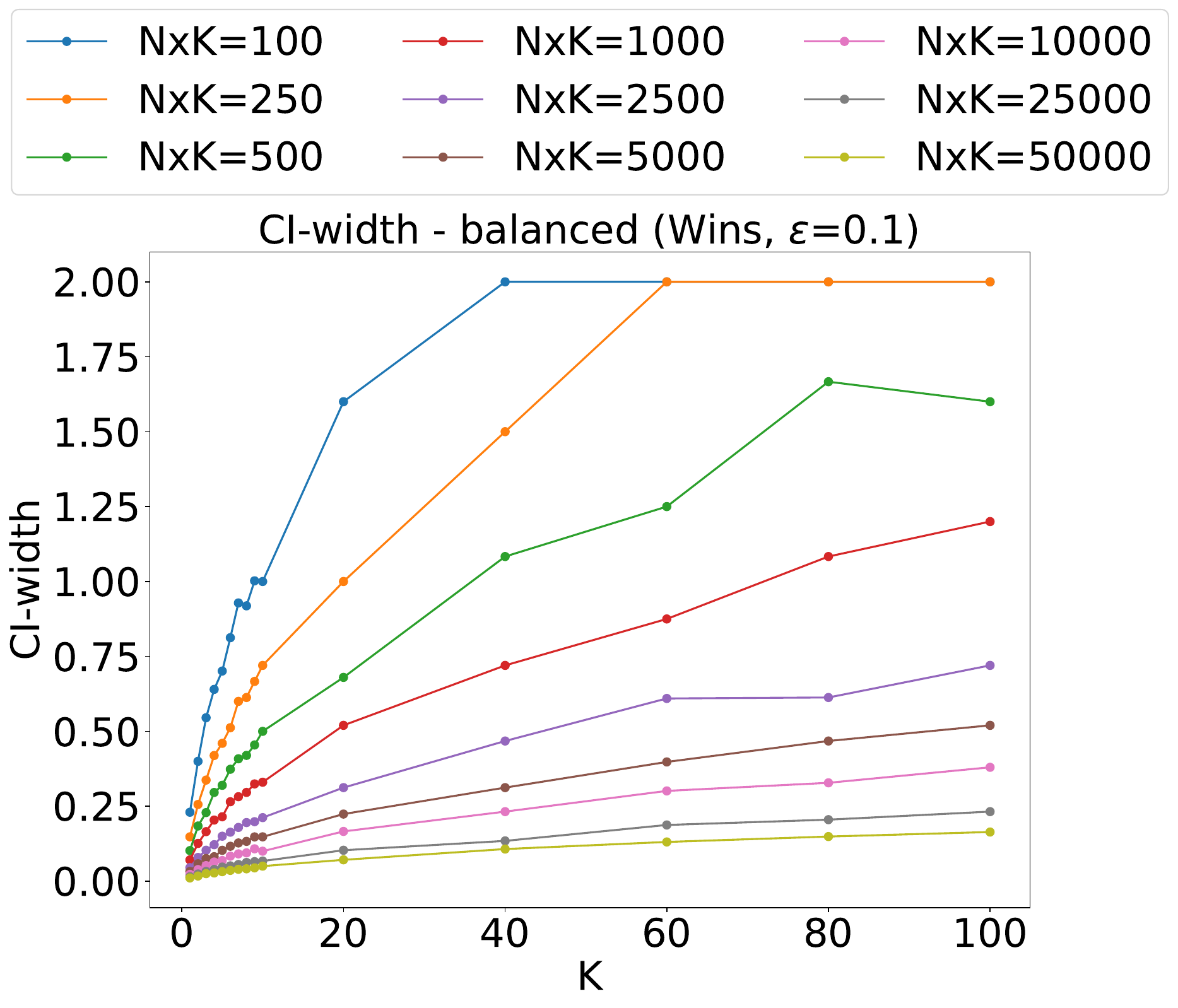}
    \caption{$\epsilon = 0.1$}
    \label{fig:uniform_ci_wins_cat5_e01}
  \end{subfigure} \hfill
  \begin{subfigure}[b]{0.24\linewidth}
    \centering
    \includegraphics[width=\linewidth]{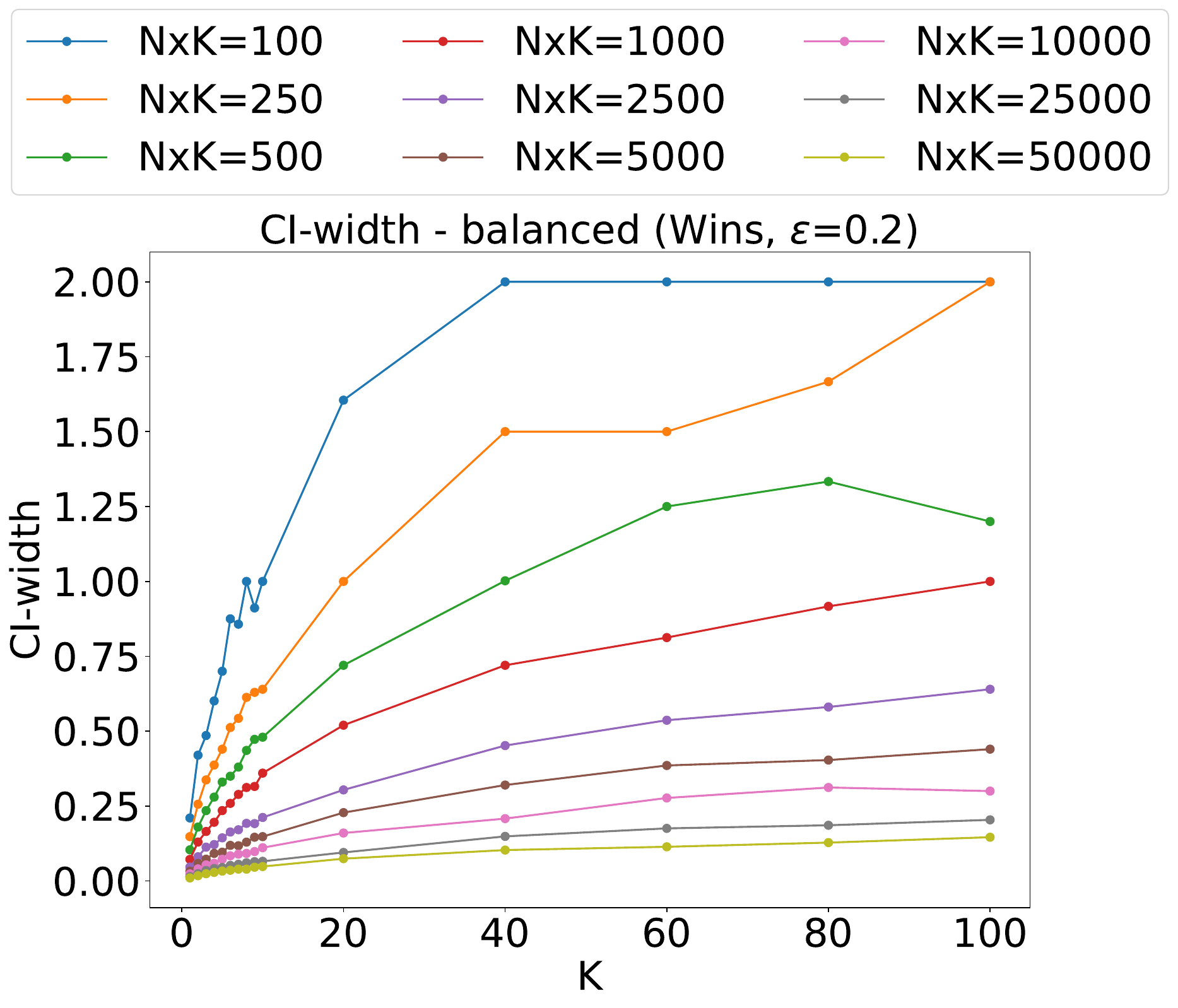}
    \caption{$\epsilon = 0.2$}
    \label{fig:uniform_ci_wins_cat5_e02}
  \end{subfigure} \hfill
  \begin{subfigure}[b]{0.24\linewidth}
    \centering
    \includegraphics[width=\linewidth]{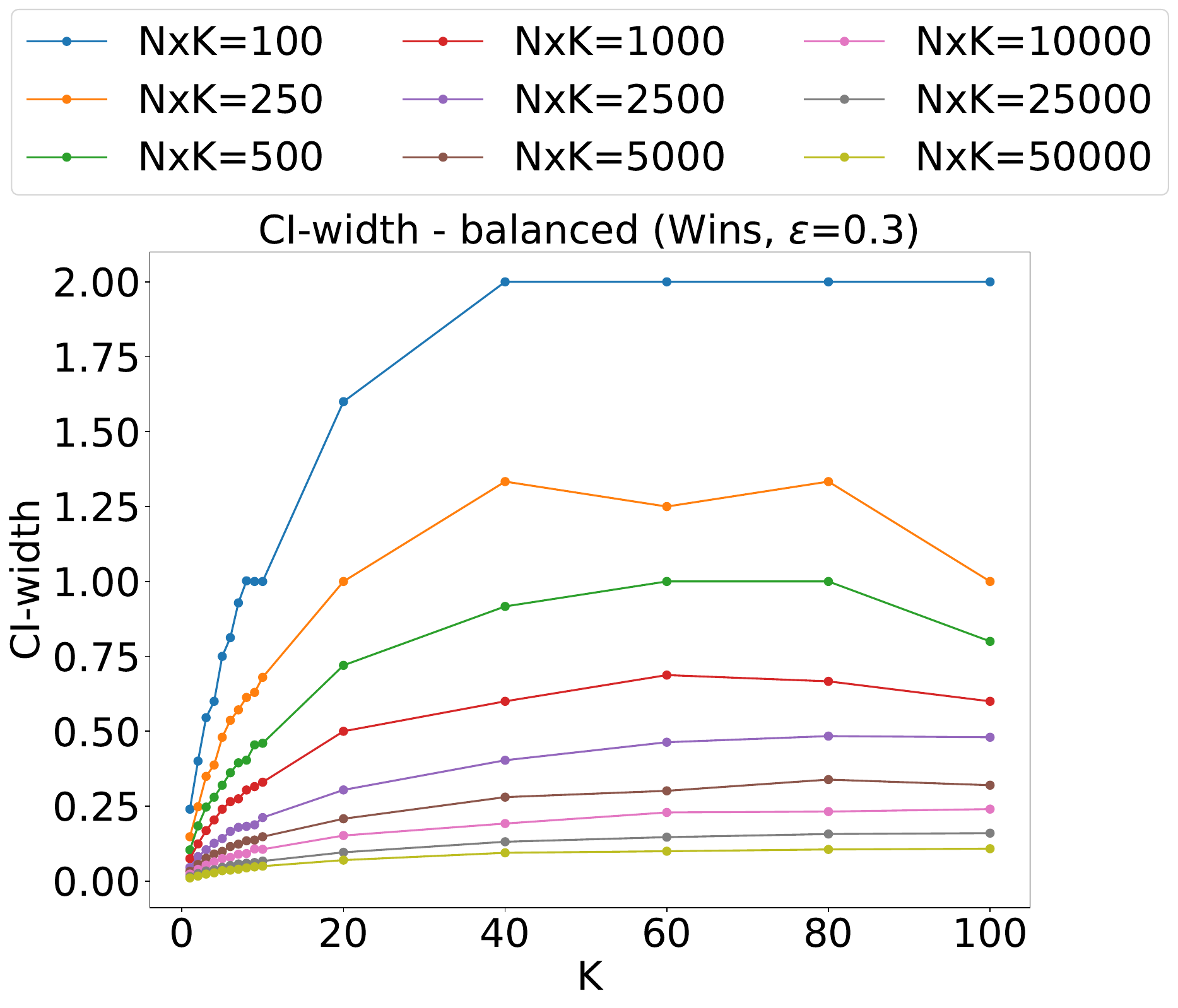}
    \caption{$\epsilon = 0.3$}
    \label{fig:uniform_ci_wins_cat5_e03}
  \end{subfigure} \hfill
  \begin{subfigure}[b]{0.24\linewidth}
    \centering
    \includegraphics[width=\linewidth]{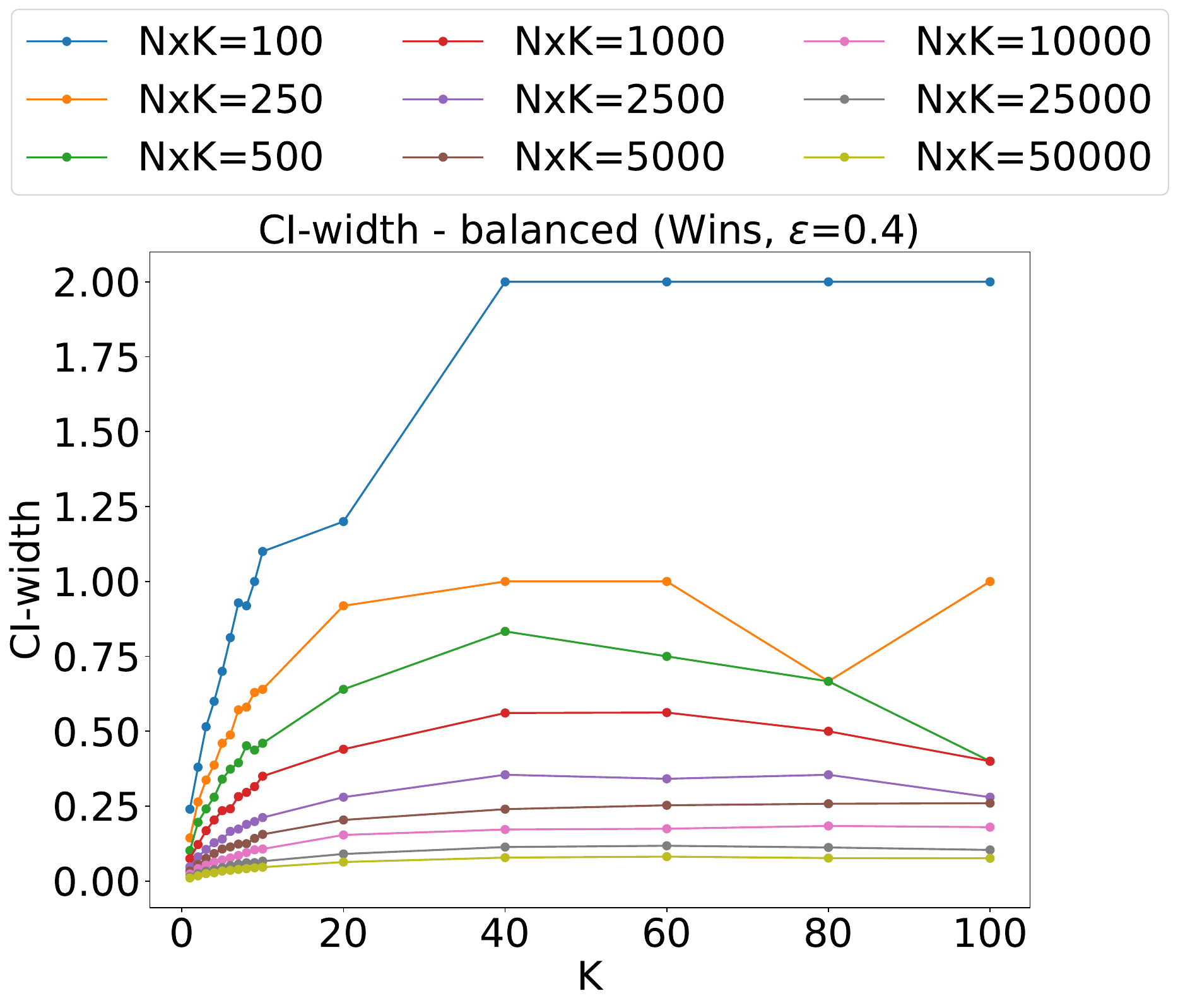}
    \caption{$\epsilon = 0.4$}
    \label{fig:uniform_ci_wins_cat5_e04}
  \end{subfigure}
  \caption{CI-width plots for balanced alphas with Wins as the metric ($M=5$)}
  \label{fig:uniform_ci_wins_cat5}
\end{figure*}

\begin{figure*}
  \centering
  \begin{subfigure}[b]{0.24\linewidth}
    \centering
    \includegraphics[width=\linewidth]{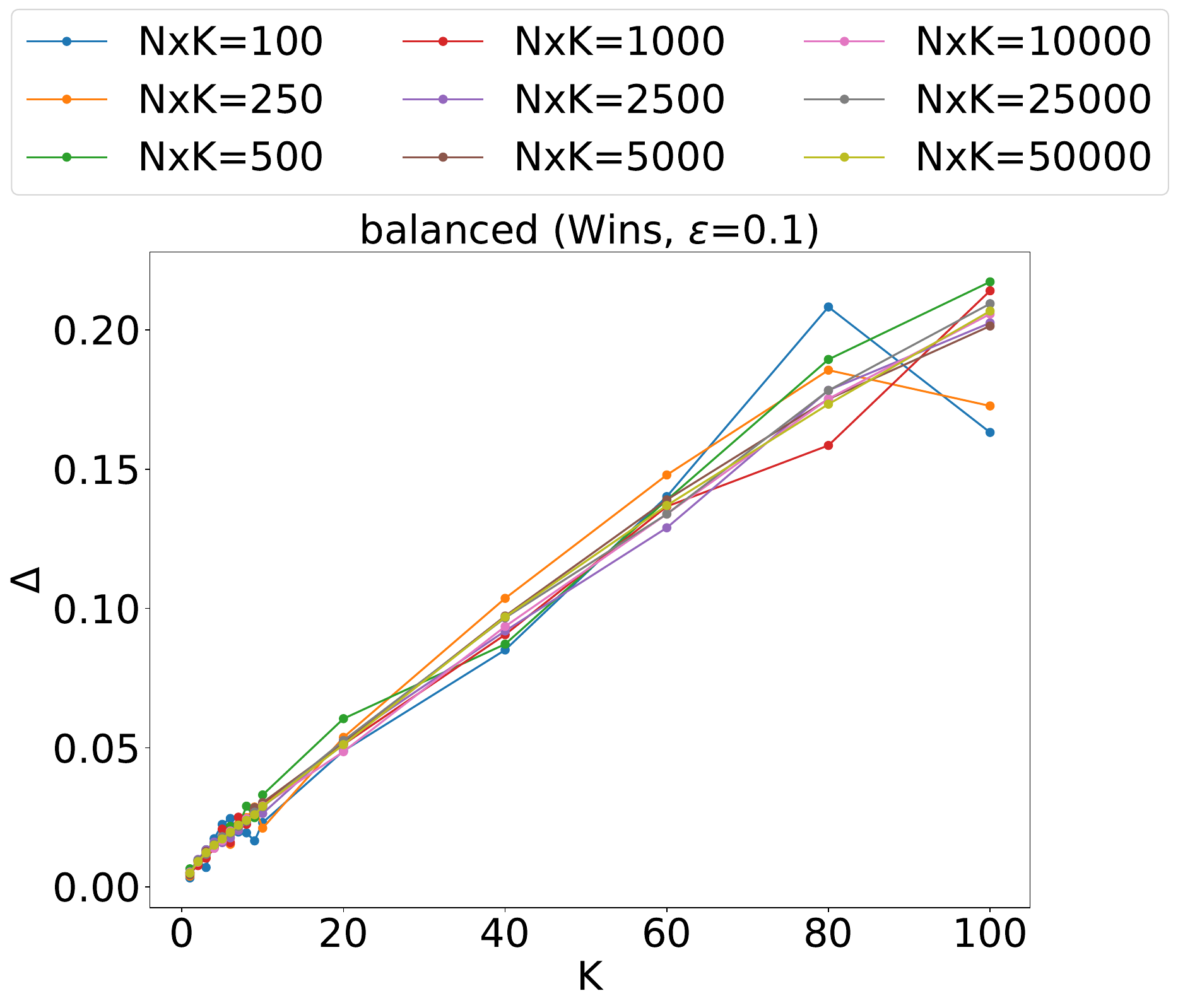}
    \caption{$\epsilon = 0.1$}
    \label{fig:uniform_delta_wins_cat5_e01}
  \end{subfigure} \hfill
  \begin{subfigure}[b]{0.24\linewidth}
    \centering
    \includegraphics[width=\linewidth]{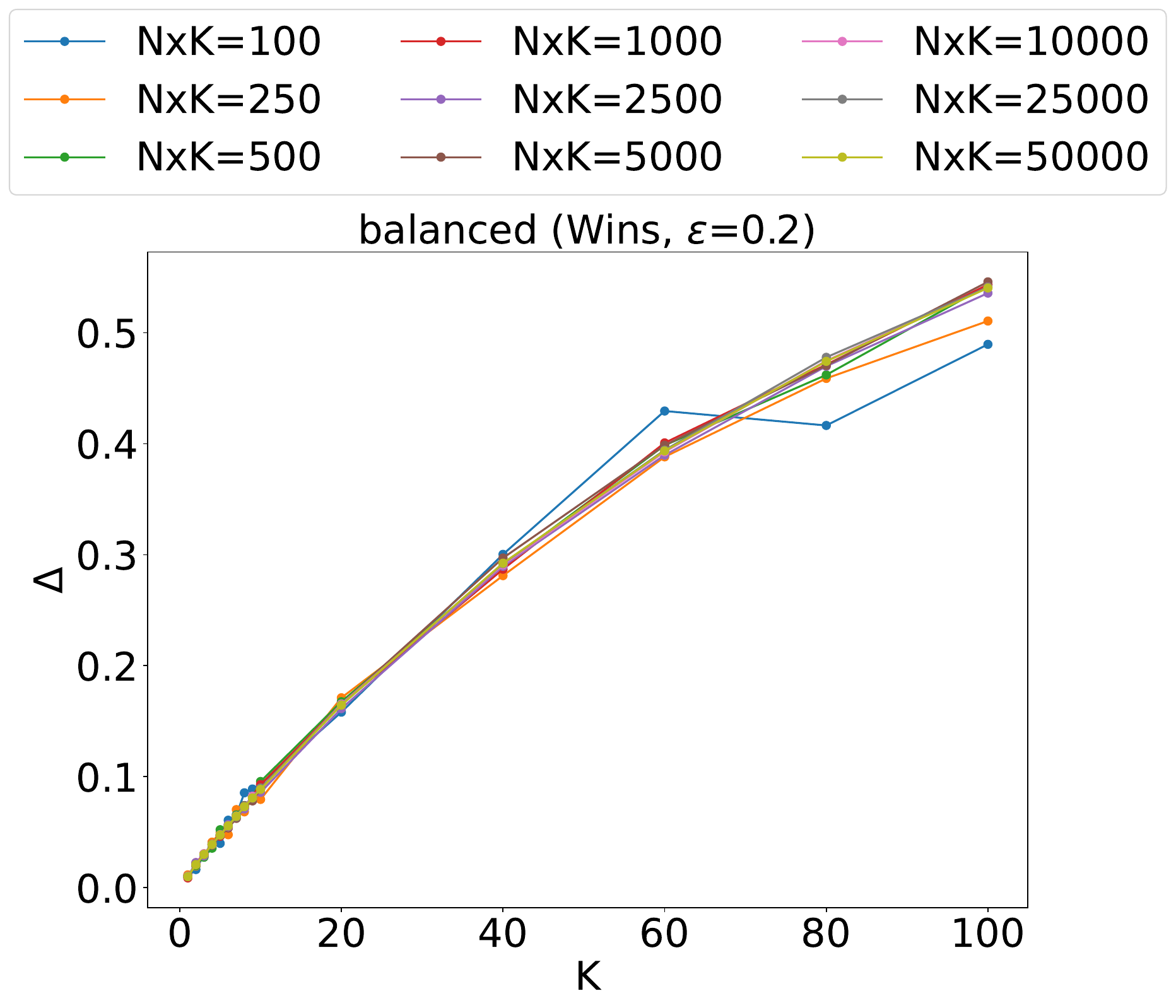}
    \caption{$\epsilon = 0.2$}
    \label{fig:uniform_delta_wins_cat5_e02}
  \end{subfigure} \hfill
  \begin{subfigure}[b]{0.24\linewidth}
    \centering
    \includegraphics[width=\linewidth]{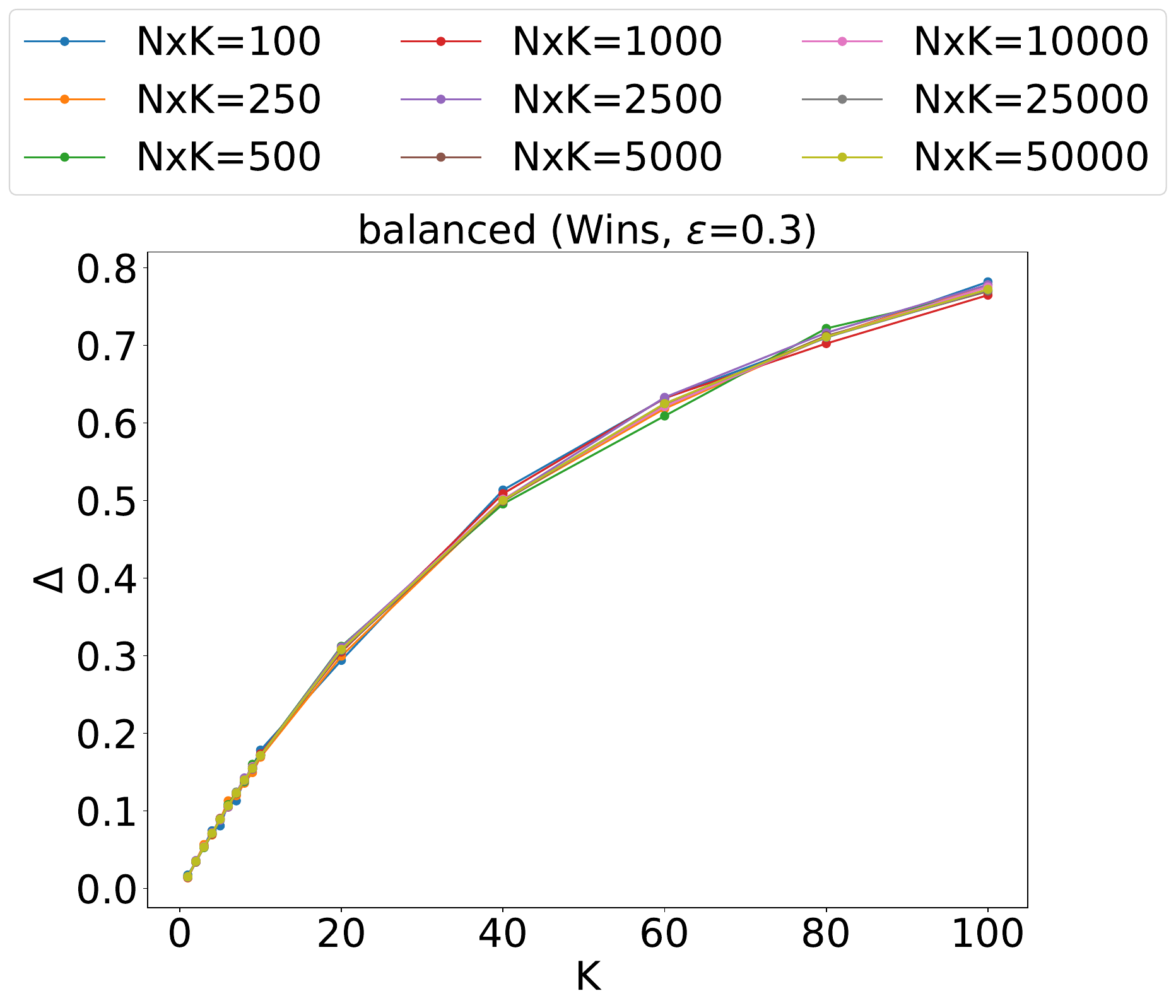}
    \caption{$\epsilon = 0.3$}
    \label{fig:uniform_delta_wins_cat5_e03}
  \end{subfigure} \hfill
  \begin{subfigure}[b]{0.24\linewidth}
    \centering
    \includegraphics[width=\linewidth]{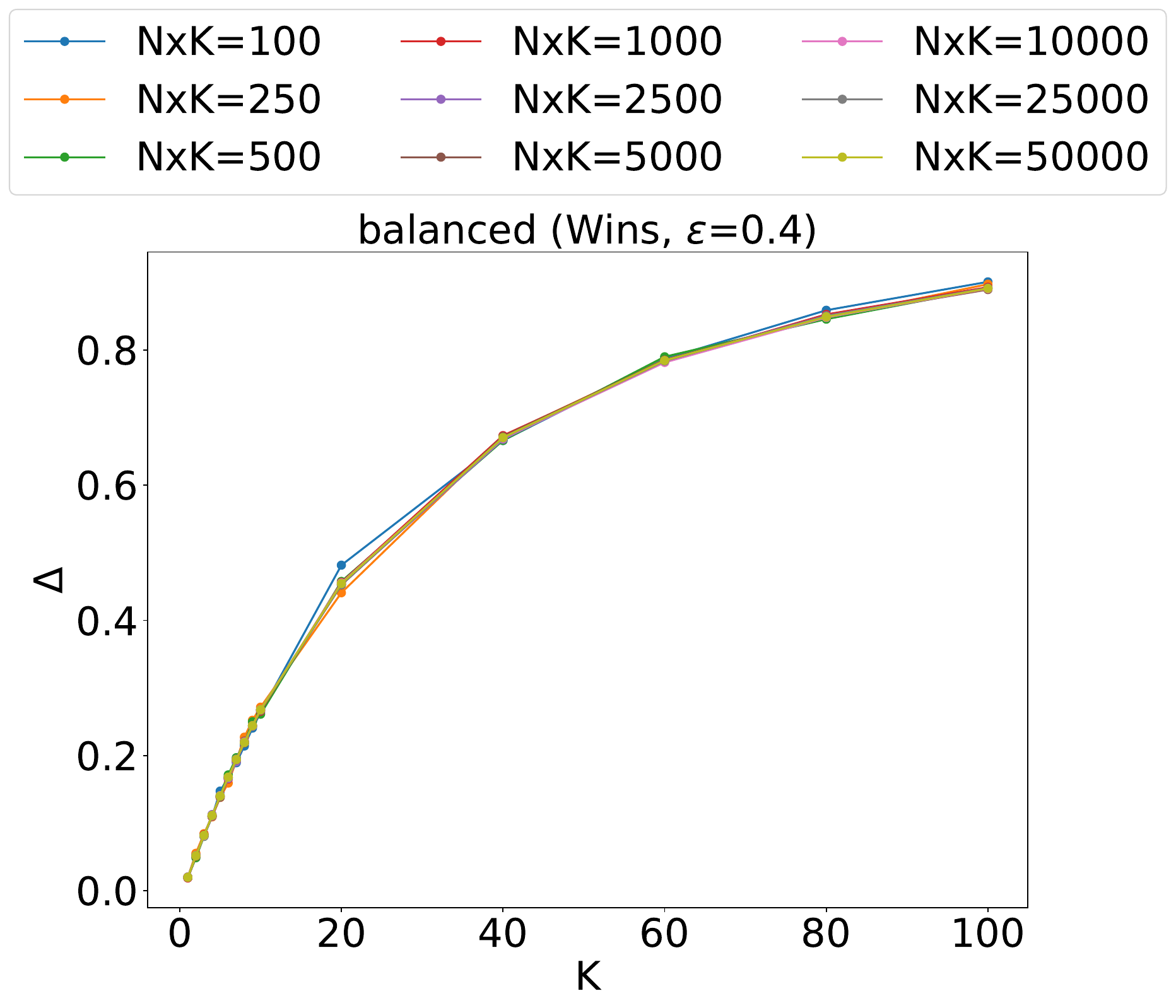}
    \caption{$\epsilon = 0.4$}
    \label{fig:uniform_delta_wins_cat5_e04}
  \end{subfigure}
  \caption{Effect sizes ($\Delta$) for balanced alphas with Wins as the metric ($M=5$)}
  \label{fig:uniform_delta_wins_cat5}
\end{figure*}

\begin{figure*}
  \centering
  \begin{subfigure}[b]{0.24\linewidth}
    \centering
    \includegraphics[width=\linewidth]{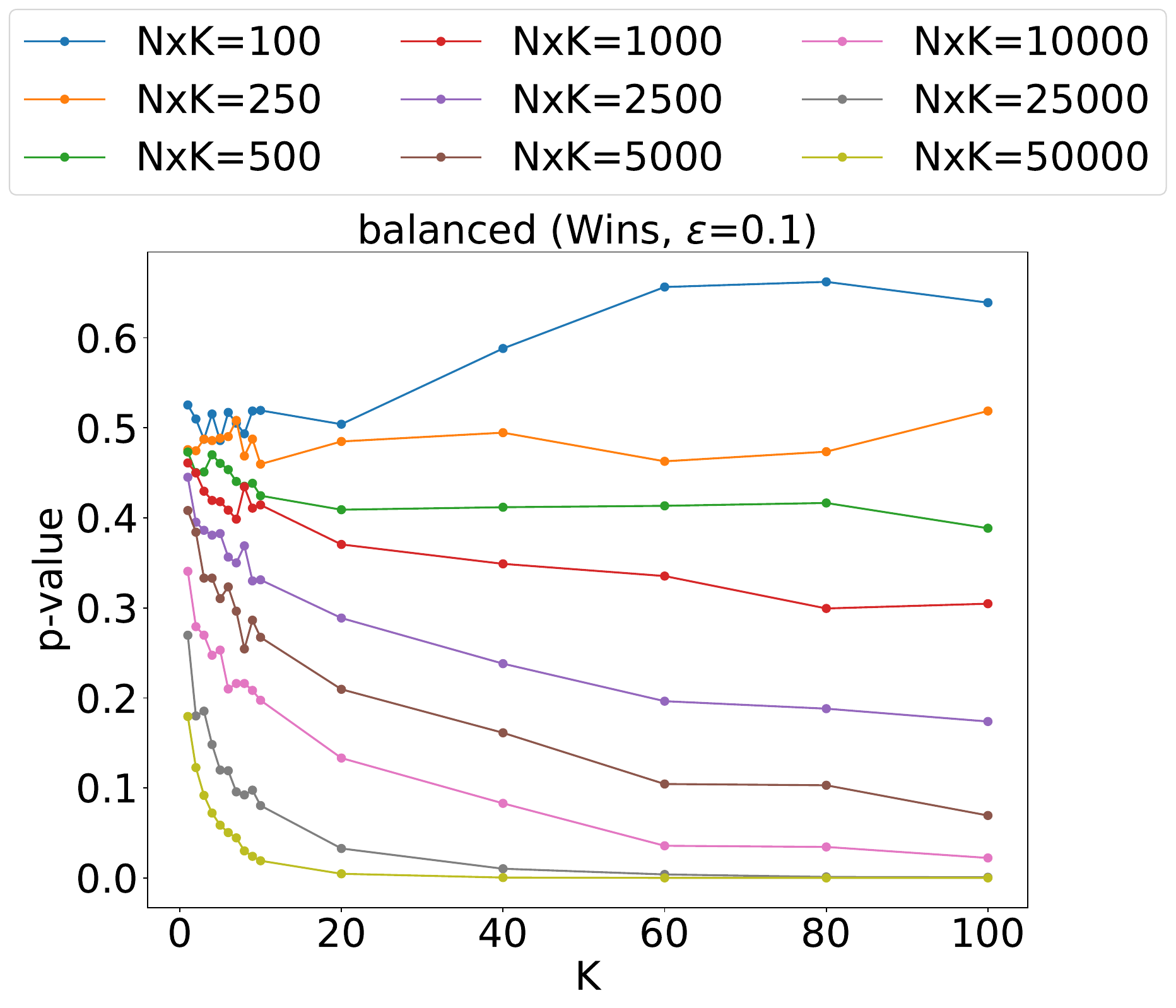}
    \caption{$\epsilon = 0.1$}
    \label{fig:uniform_wins_cat12_e01}
  \end{subfigure} \hfill
  \begin{subfigure}[b]{0.24\linewidth}
    \centering
    \includegraphics[width=\linewidth]{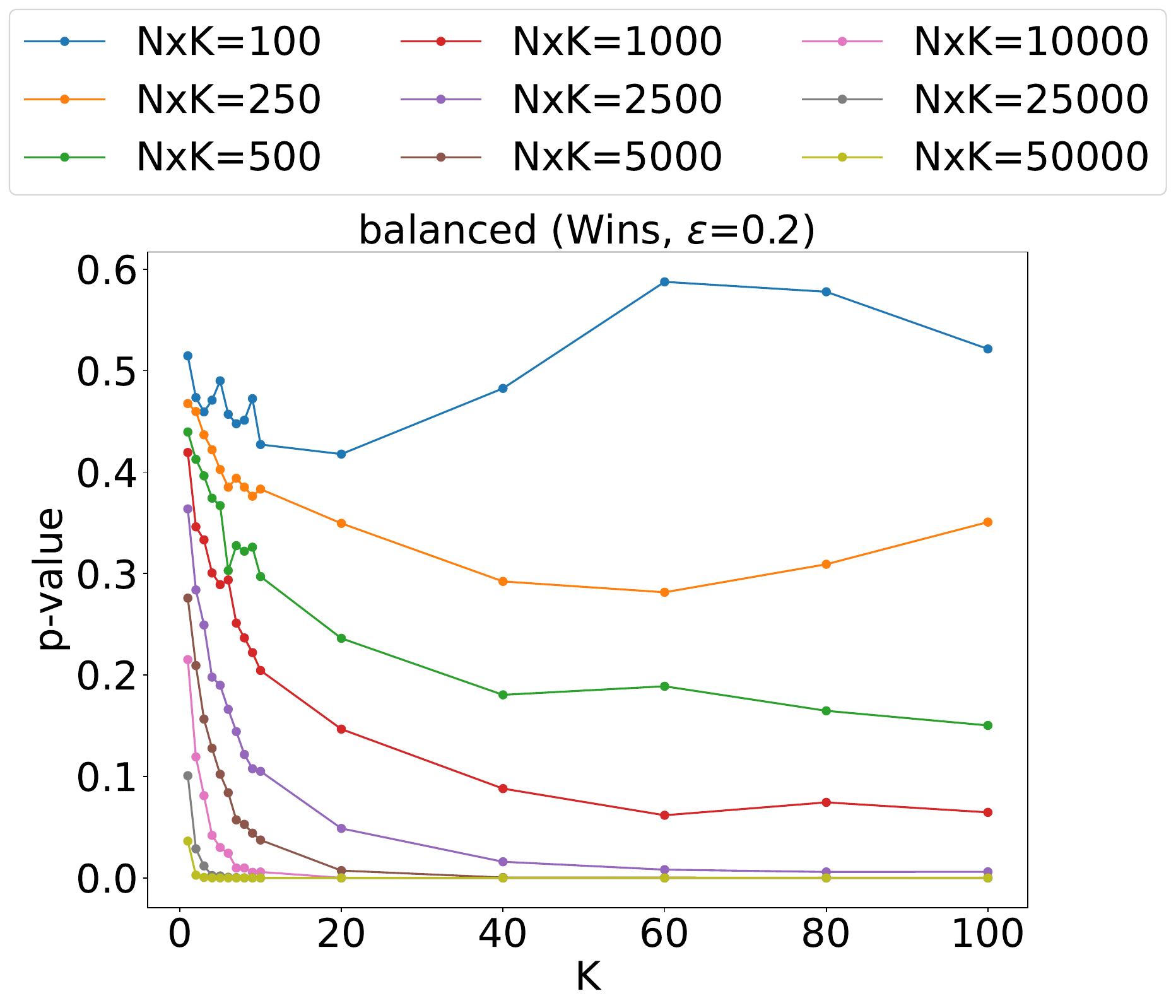}
    \caption{$\epsilon = 0.2$}
    \label{fig:uniform_wins_cat12_e02}
  \end{subfigure} \hfill
  \begin{subfigure}[b]{0.24\linewidth}
    \centering
    \includegraphics[width=\linewidth]{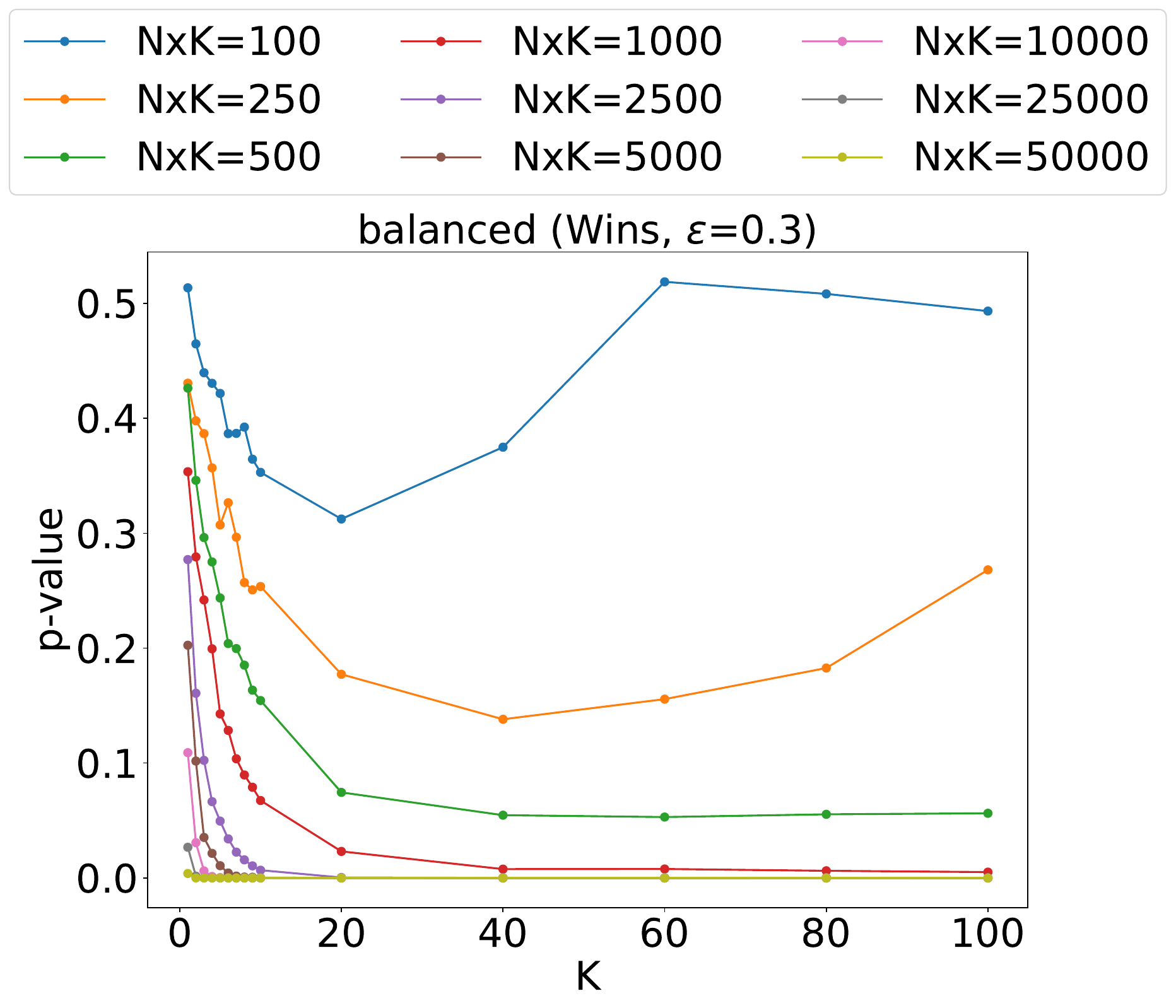}
    \caption{$\epsilon = 0.3$}
    \label{fig:uniform_wins_cat12_e03}
  \end{subfigure} \hfill
  \begin{subfigure}[b]{0.24\linewidth}
    \centering
    \includegraphics[width=\linewidth]{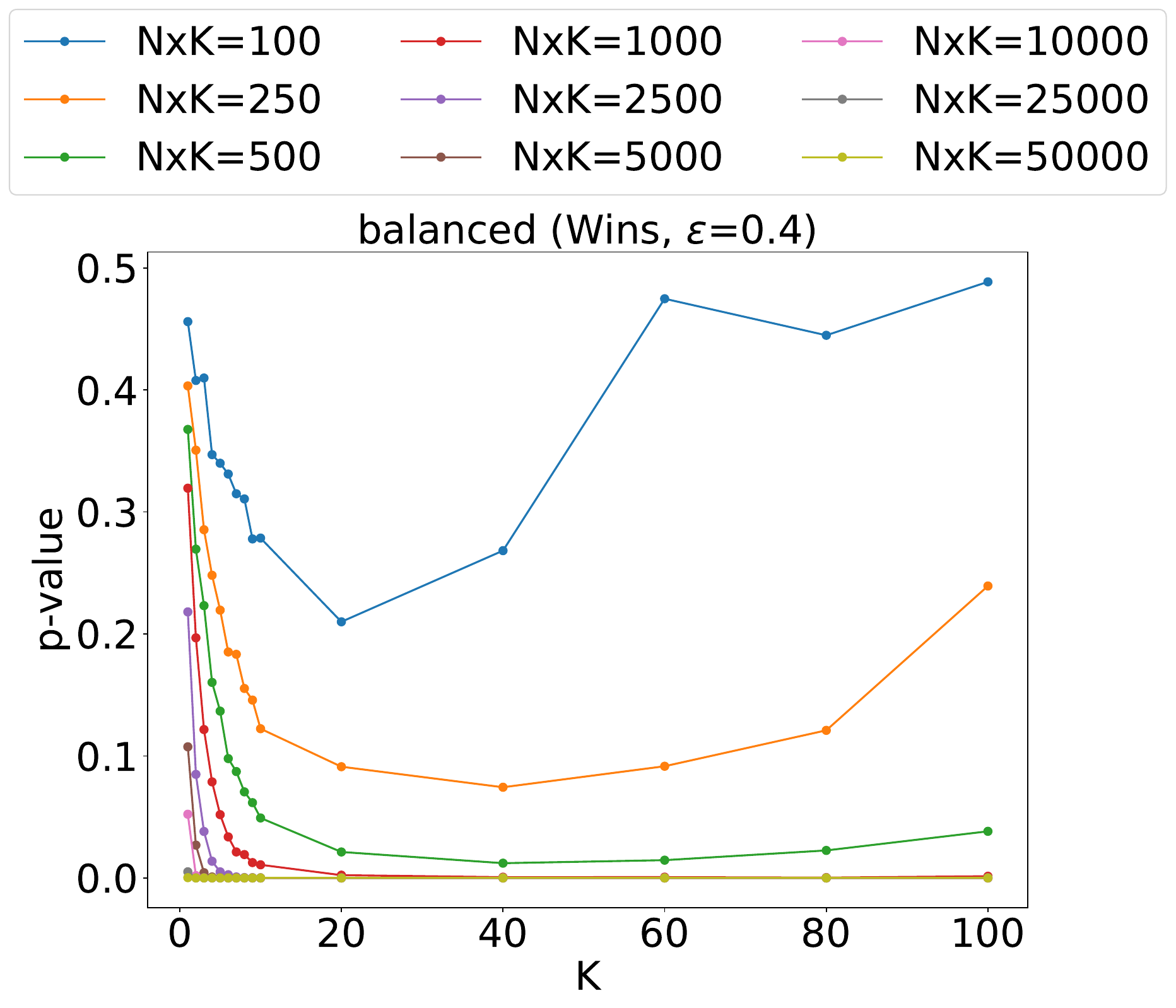}
    \caption{$\epsilon = 0.4$}
    \label{fig:uniform_wins_cat12_e04}
  \end{subfigure}
  \caption{P-value plots for balanced alphas with Wins as the metric ($M=12$)}
  \label{fig:uniform_wins_cat12}
\end{figure*}

\begin{figure*}
  \centering
  \begin{subfigure}[b]{0.24\linewidth}
    \centering
    \includegraphics[width=\linewidth]{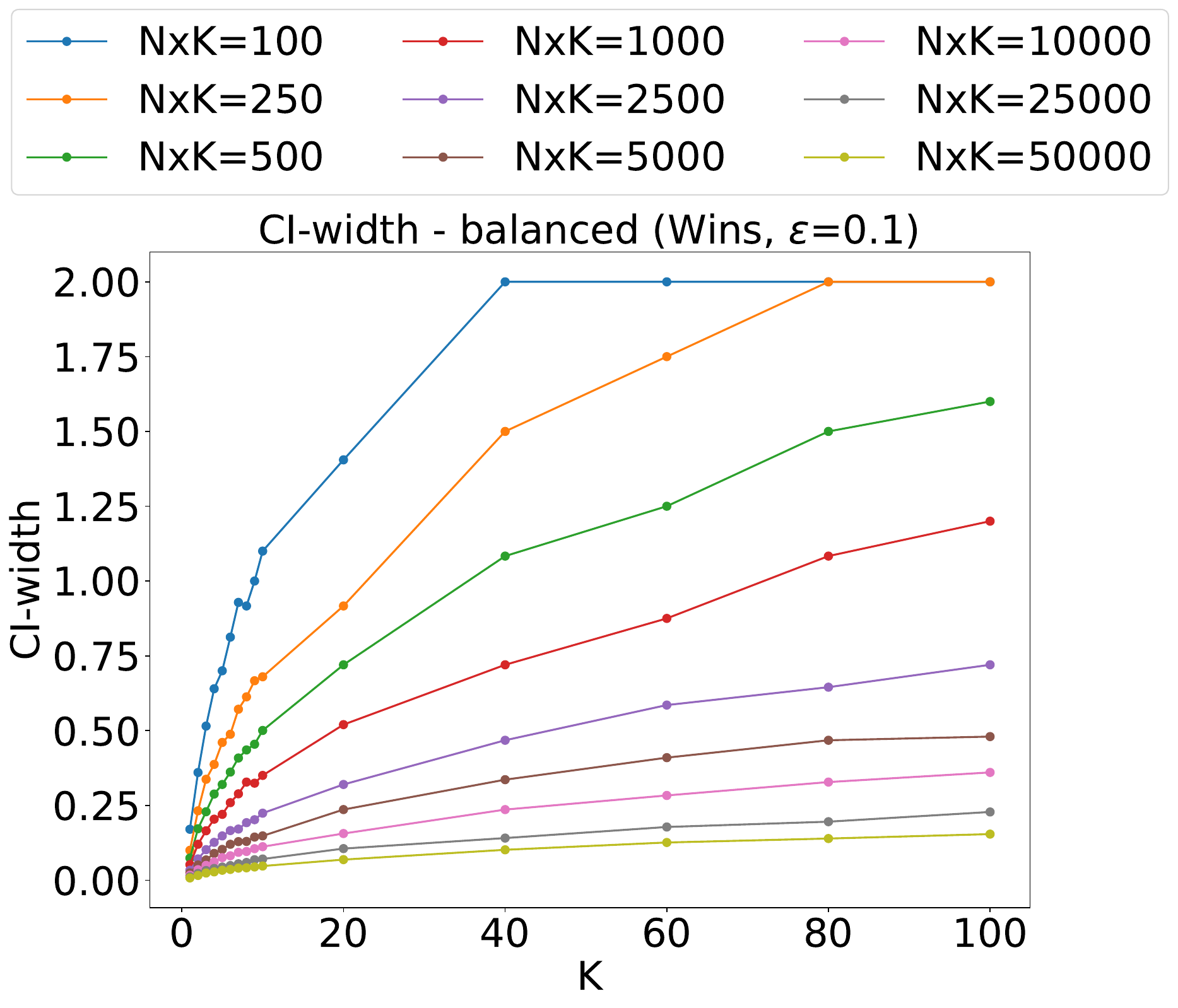}
    \caption{$\epsilon = 0.1$}
    \label{fig:uniform_ci_wins_cat12_e01}
  \end{subfigure} \hfill
  \begin{subfigure}[b]{0.24\linewidth}
    \centering
    \includegraphics[width=\linewidth]{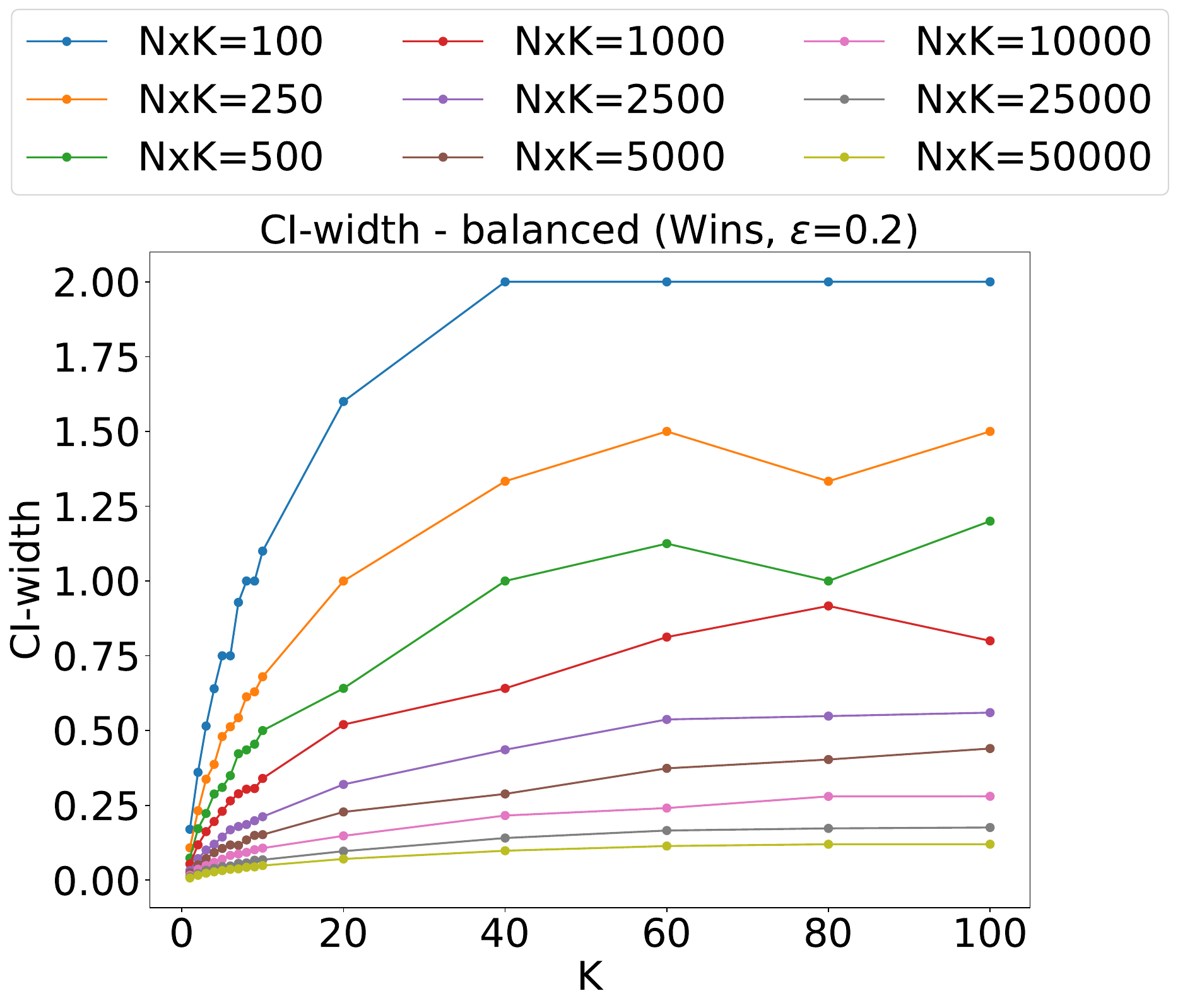}
    \caption{$\epsilon = 0.2$}
    \label{fig:uniform_ci_wins_cat12_e02}
  \end{subfigure} \hfill
  \begin{subfigure}[b]{0.24\linewidth}
    \centering
    \includegraphics[width=\linewidth]{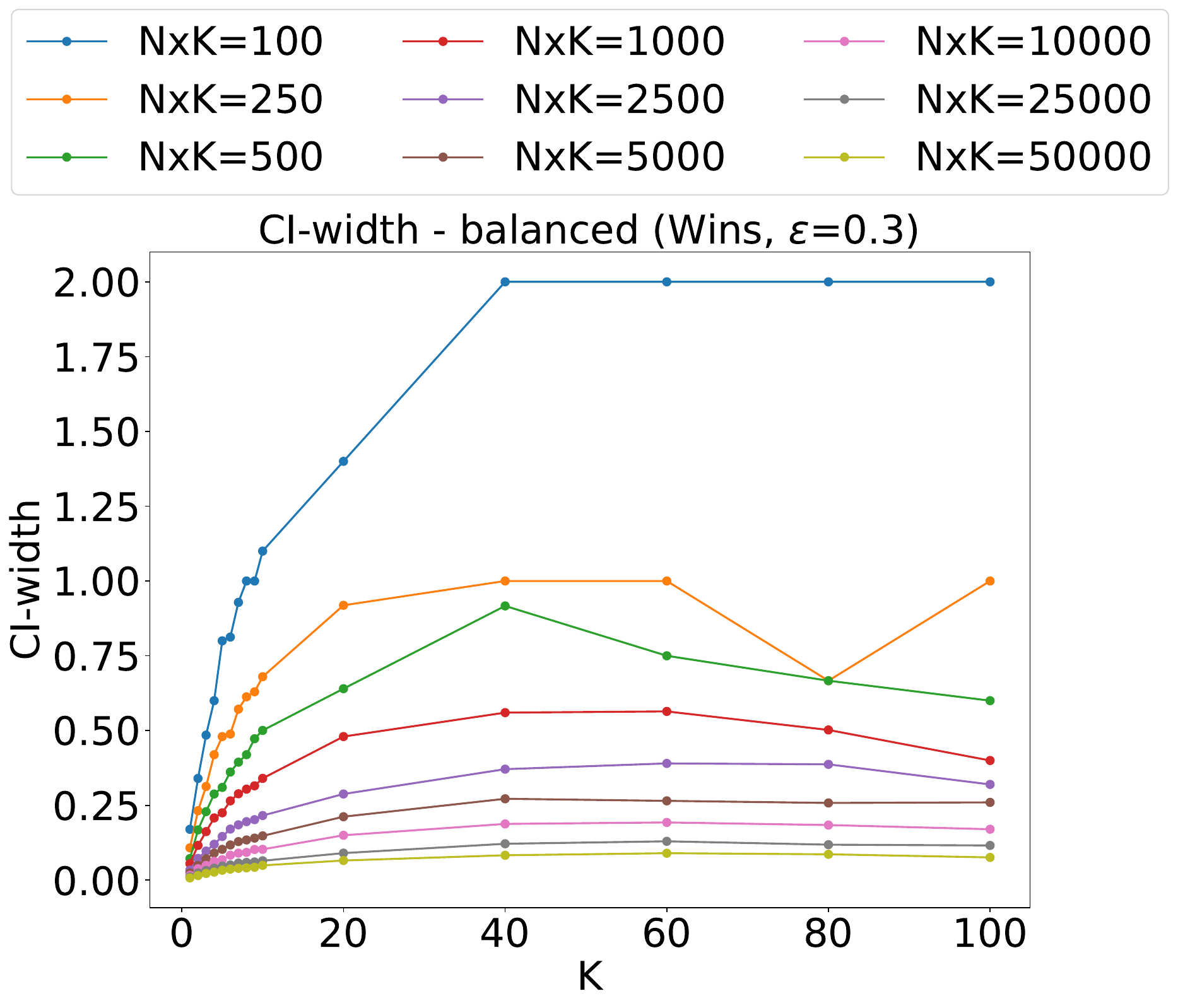}
    \caption{$\epsilon = 0.3$}
    \label{fig:uniform_ci_wins_cat12_e03}
  \end{subfigure} \hfill
  \begin{subfigure}[b]{0.24\linewidth}
    \centering
    \includegraphics[width=\linewidth]{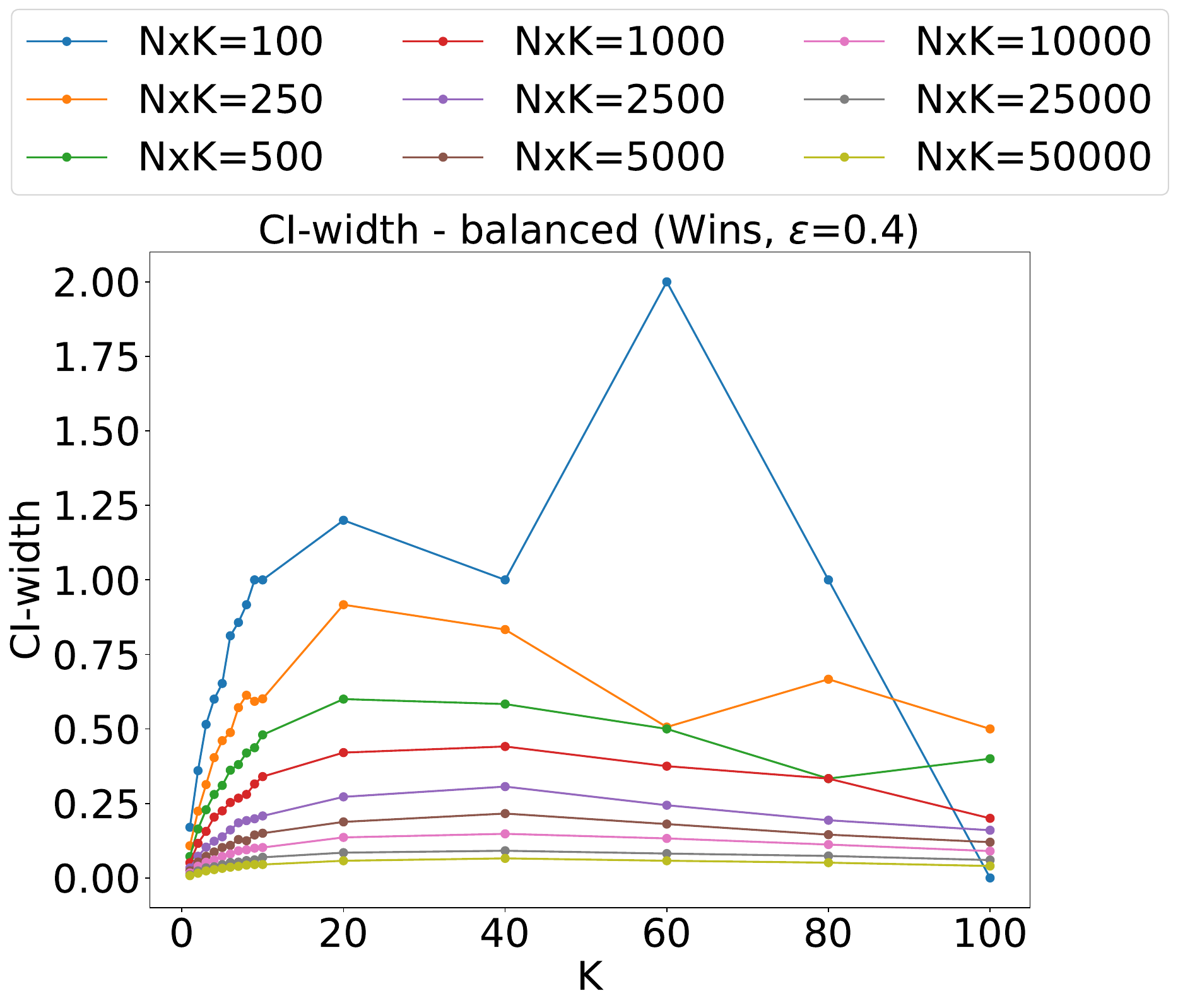}
    \caption{$\epsilon = 0.4$}
    \label{fig:uniform_ci_wins_cat12_e04}
  \end{subfigure}
  \caption{CI-width plots for balanced alphas with Wins as the metric ($M=12$)}
  \label{fig:uniform_ci_wins_cat12}
\end{figure*}

\begin{figure*}
  \centering
  \begin{subfigure}[b]{0.24\linewidth}
    \centering
    \includegraphics[width=\linewidth]{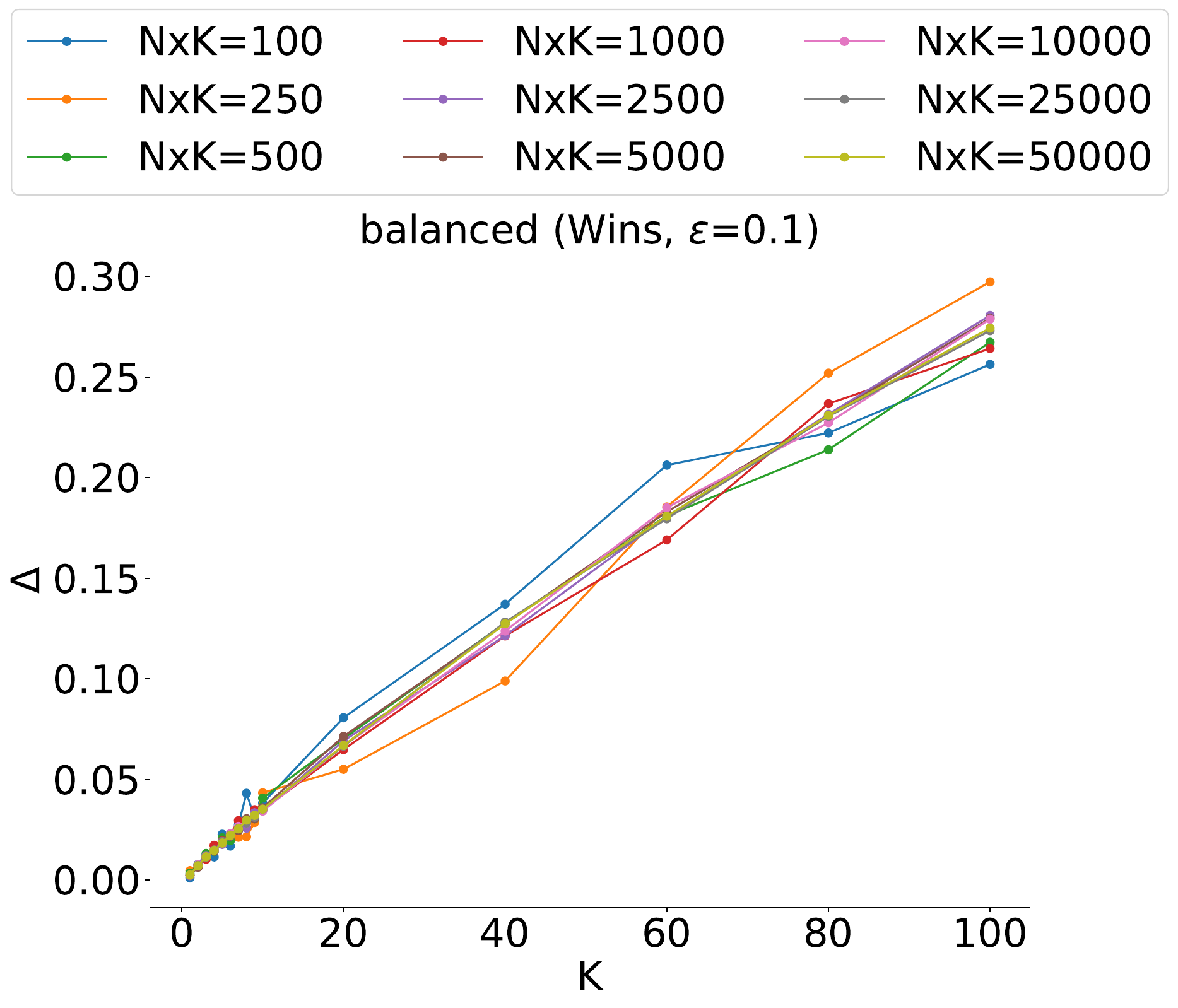}
    \caption{$\epsilon = 0.1$}
    \label{fig:uniform_delta_wins_cat12_e01}
  \end{subfigure} \hfill
  \begin{subfigure}[b]{0.24\linewidth}
    \centering
    \includegraphics[width=\linewidth]{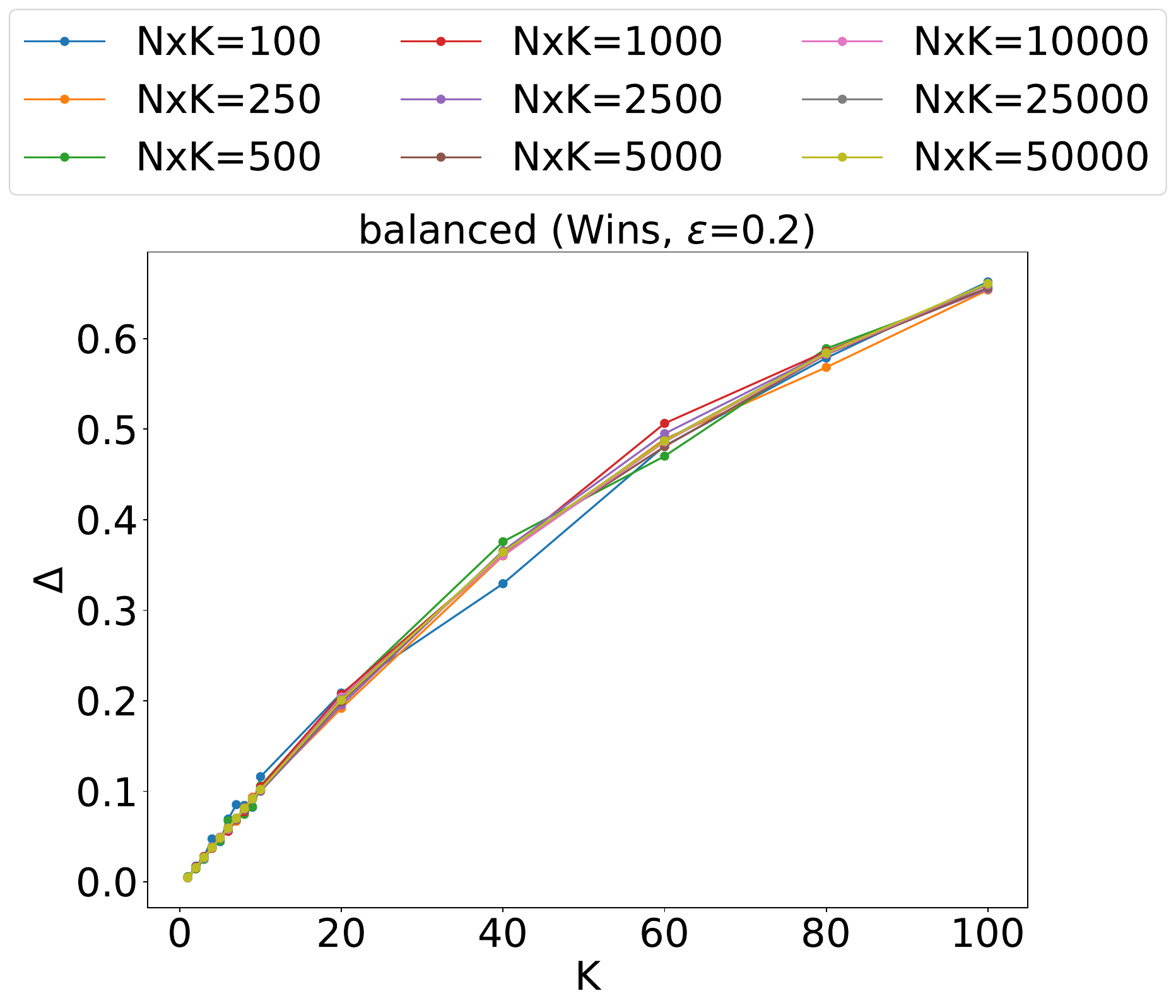}
    \caption{$\epsilon = 0.2$}
    \label{fig:uniform_delta_wins_cat12_e02}
  \end{subfigure} \hfill
  \begin{subfigure}[b]{0.24\linewidth}
    \centering
    \includegraphics[width=\linewidth]{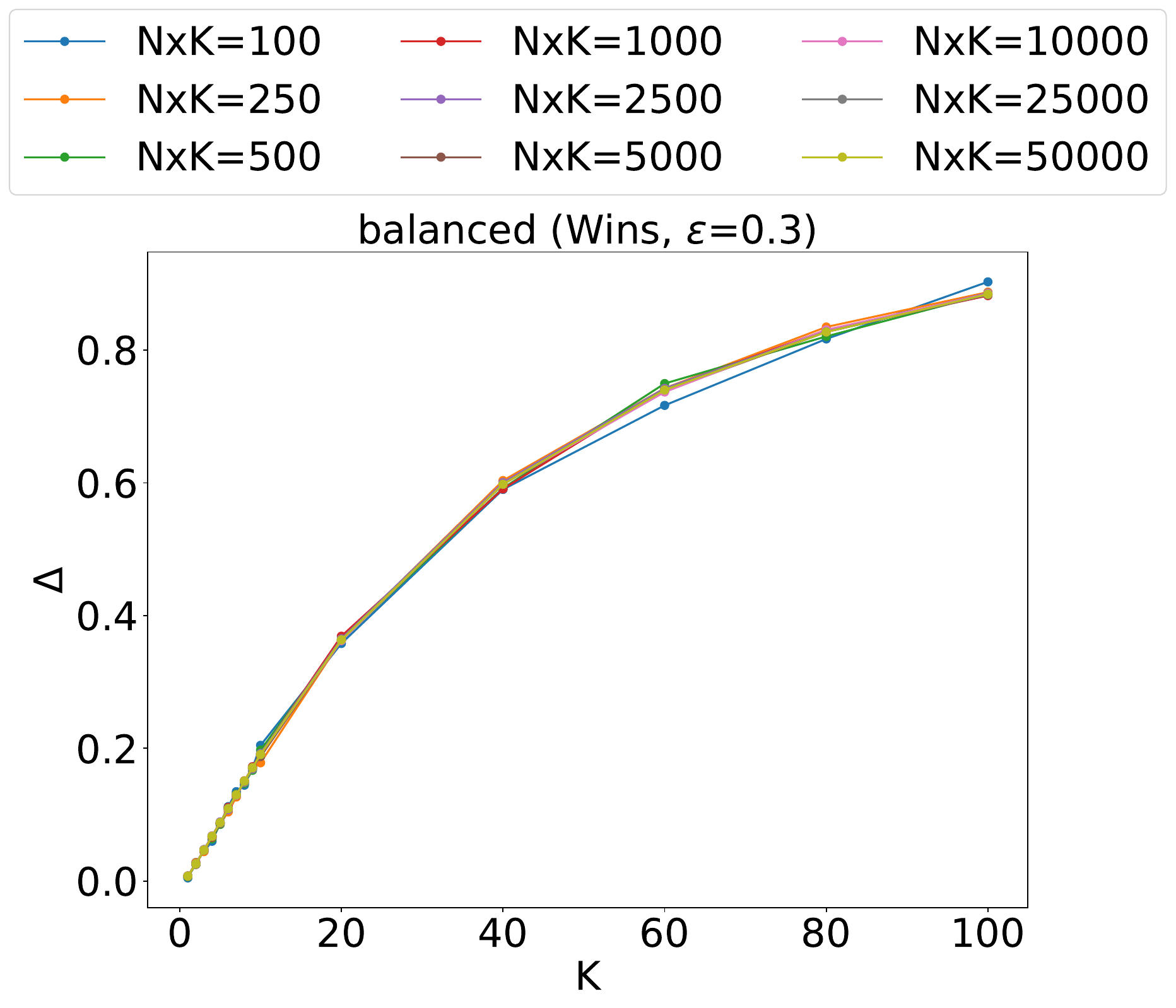}
    \caption{$\epsilon = 0.3$}
    \label{fig:uniform_delta_wins_cat12_e03}
  \end{subfigure} \hfill
  \begin{subfigure}[b]{0.24\linewidth}
    \centering
    \includegraphics[width=\linewidth]{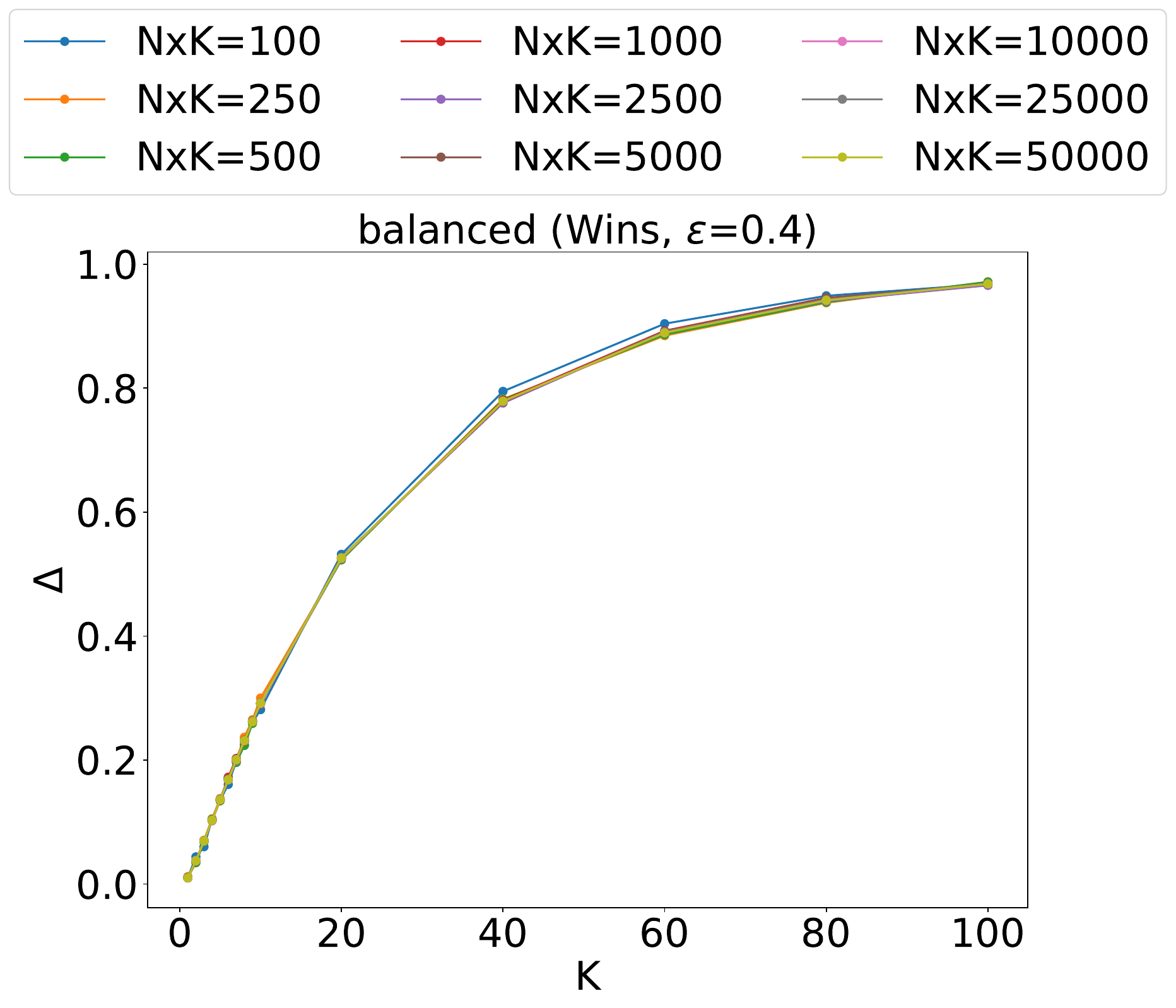}
    \caption{$\epsilon = 0.4$}
    \label{fig:uniform_delta_wins_cat12_e04}
  \end{subfigure}
  \caption{Effect sizes ($\Delta$) for balanced alphas with Wins as the metric ($M=12$)}
  \label{fig:uniform_delta_wins_cat12}
\end{figure*}

Figures \ref{fig:uniform_wins_cat2}--\ref{fig:uniform_delta_wins_cat12} show results for \pv s, confidence intervals, and effect sizes of Wins for a balanced priors on parameter $\alpha$ with different values of $\epsilon$ and increasing $K$ for various $N \times K$. For simulations with balanced categories, \pv s\ generally decrease with increasing $K$ and start to go back up for lower $N \times K$. CI-width and effect sizes increase as $K$ increases, with some exceptions.

\subsubsection{Artificial Distribution with Unbalanced Priors}

\begin{figure*}
  \centering
  \begin{subfigure}[b]{0.24\linewidth}
    \centering
    \includegraphics[width=\linewidth]{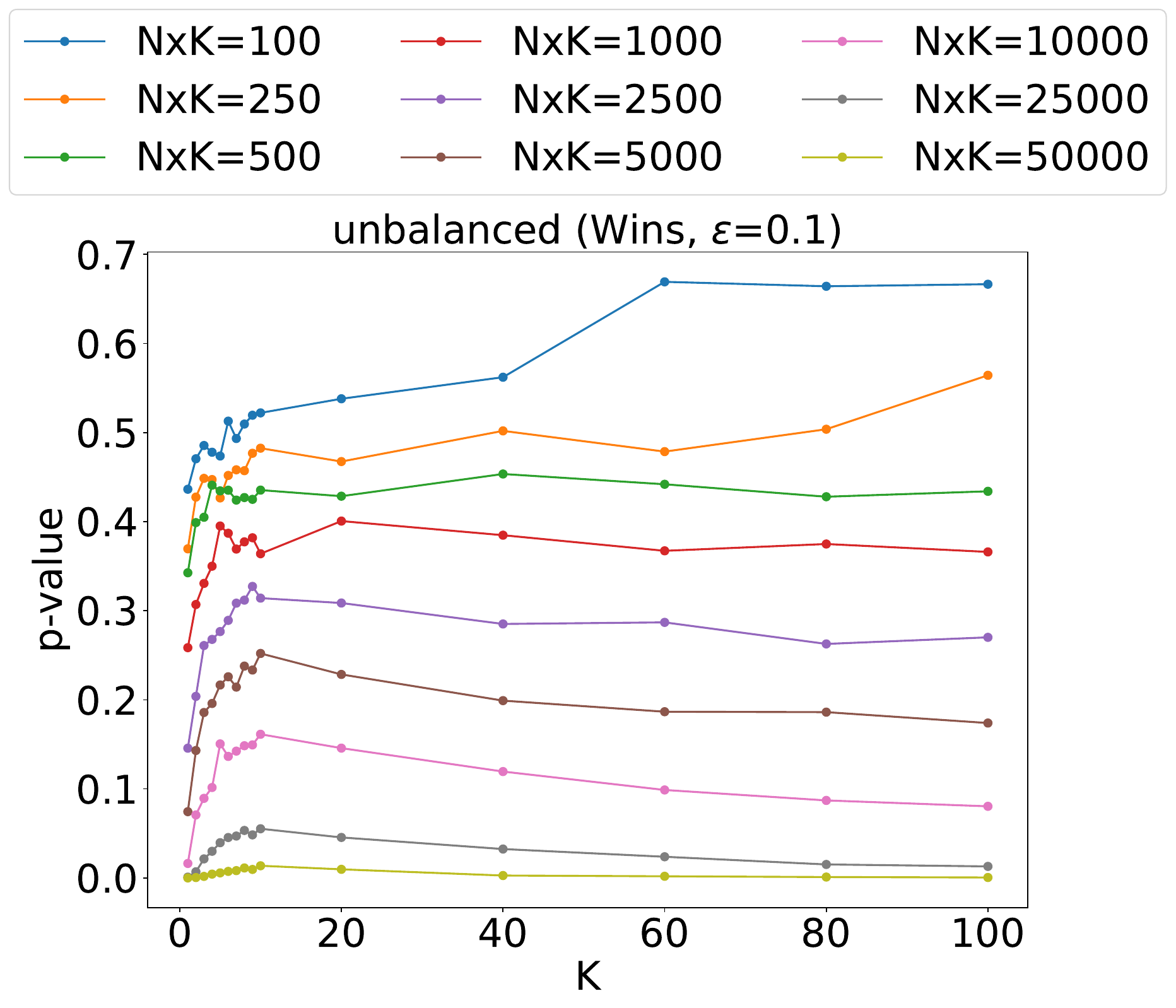}
    \caption{$\epsilon = 0.1$}
    \label{fig:gamma_wins_cat2_e01}
  \end{subfigure} \hfill
  \begin{subfigure}[b]{0.24\linewidth}
    \centering
    \includegraphics[width=\linewidth]{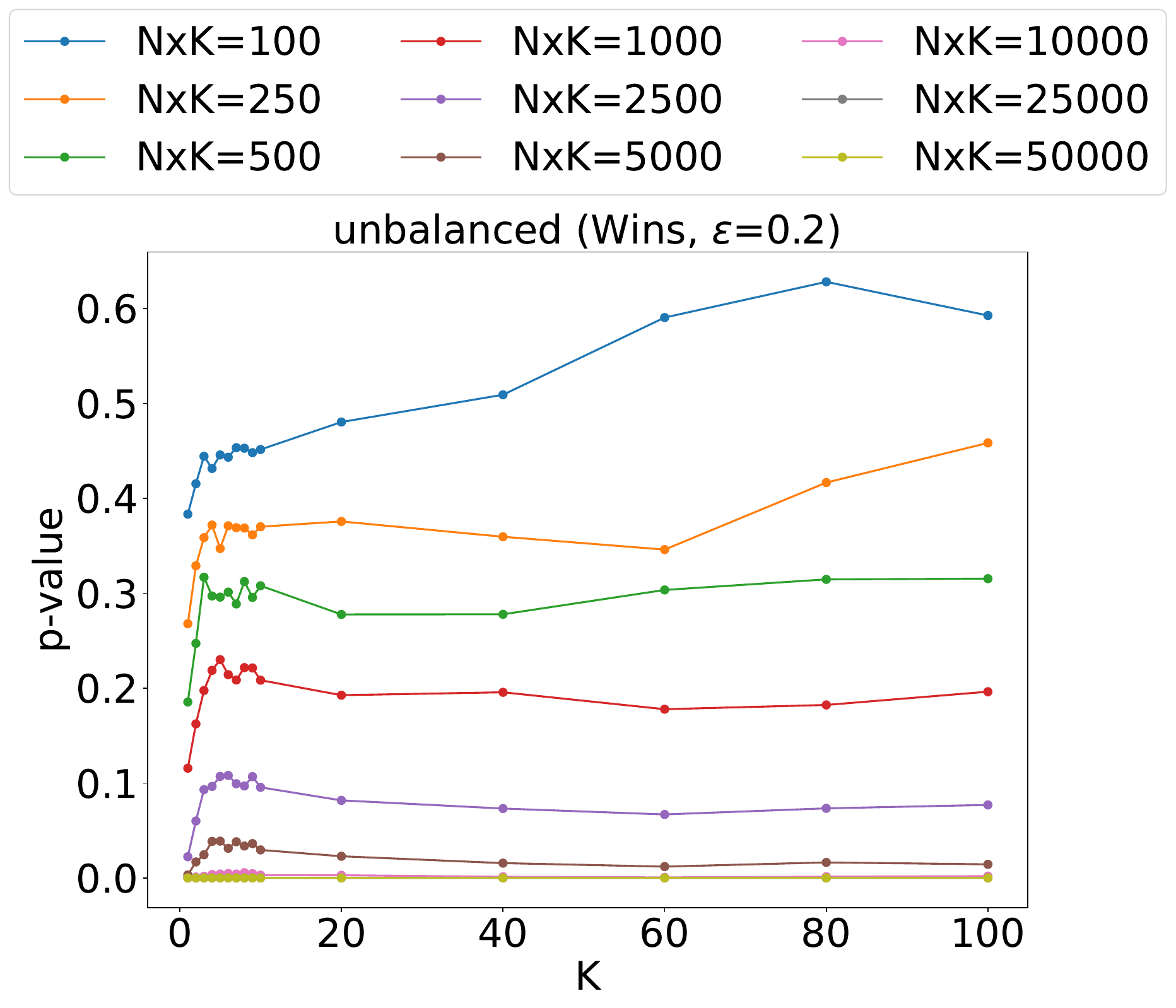}
    \caption{$\epsilon = 0.2$}
    \label{fig:gamma_wins_cat2_e02}
  \end{subfigure} \hfill
  \begin{subfigure}[b]{0.24\linewidth}
    \centering
    \includegraphics[width=\linewidth]{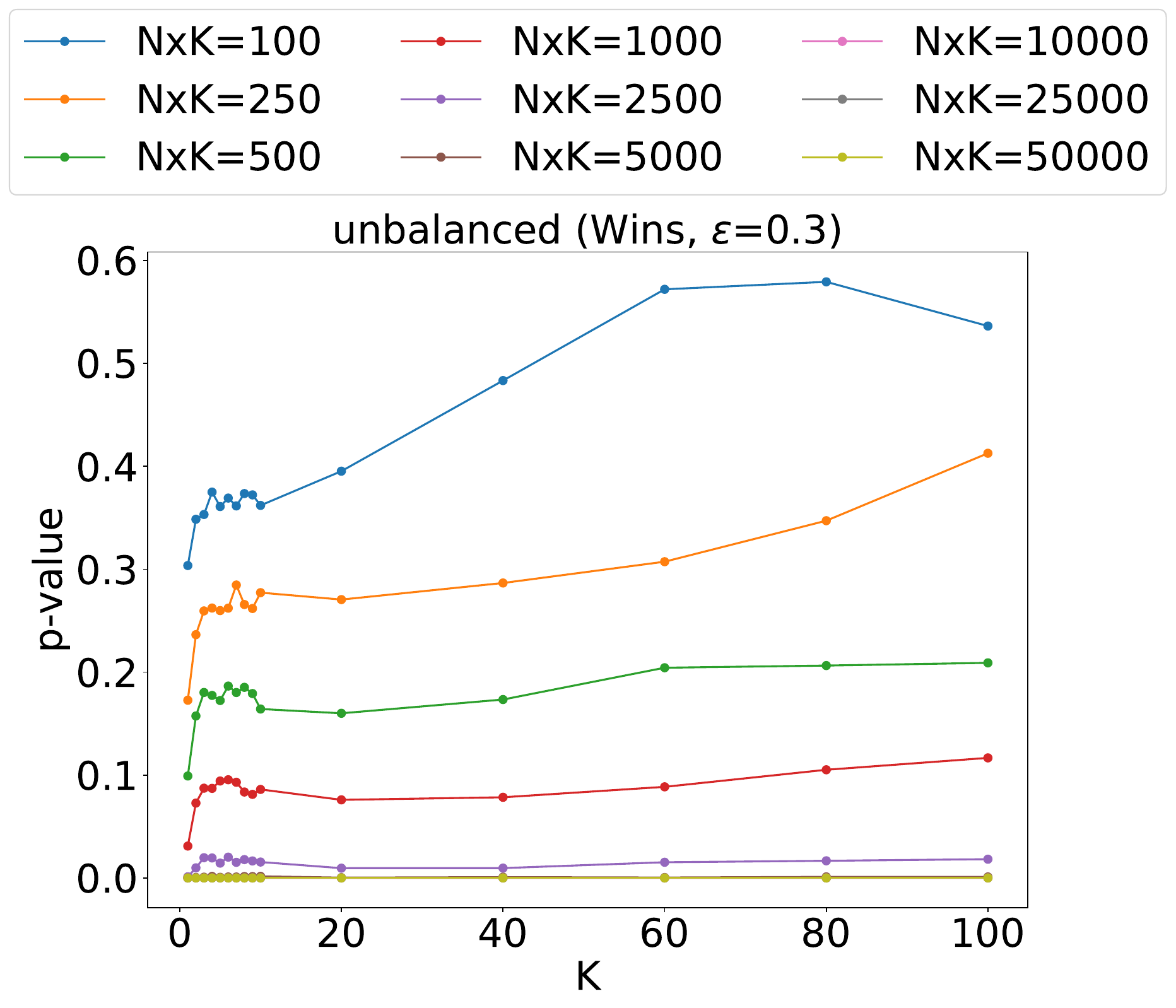}
    \caption{$\epsilon = 0.3$}
    \label{fig:gamma_wins_cat2_e03}
  \end{subfigure} \hfill
  \begin{subfigure}[b]{0.24\linewidth}
    \centering
    \includegraphics[width=\linewidth]{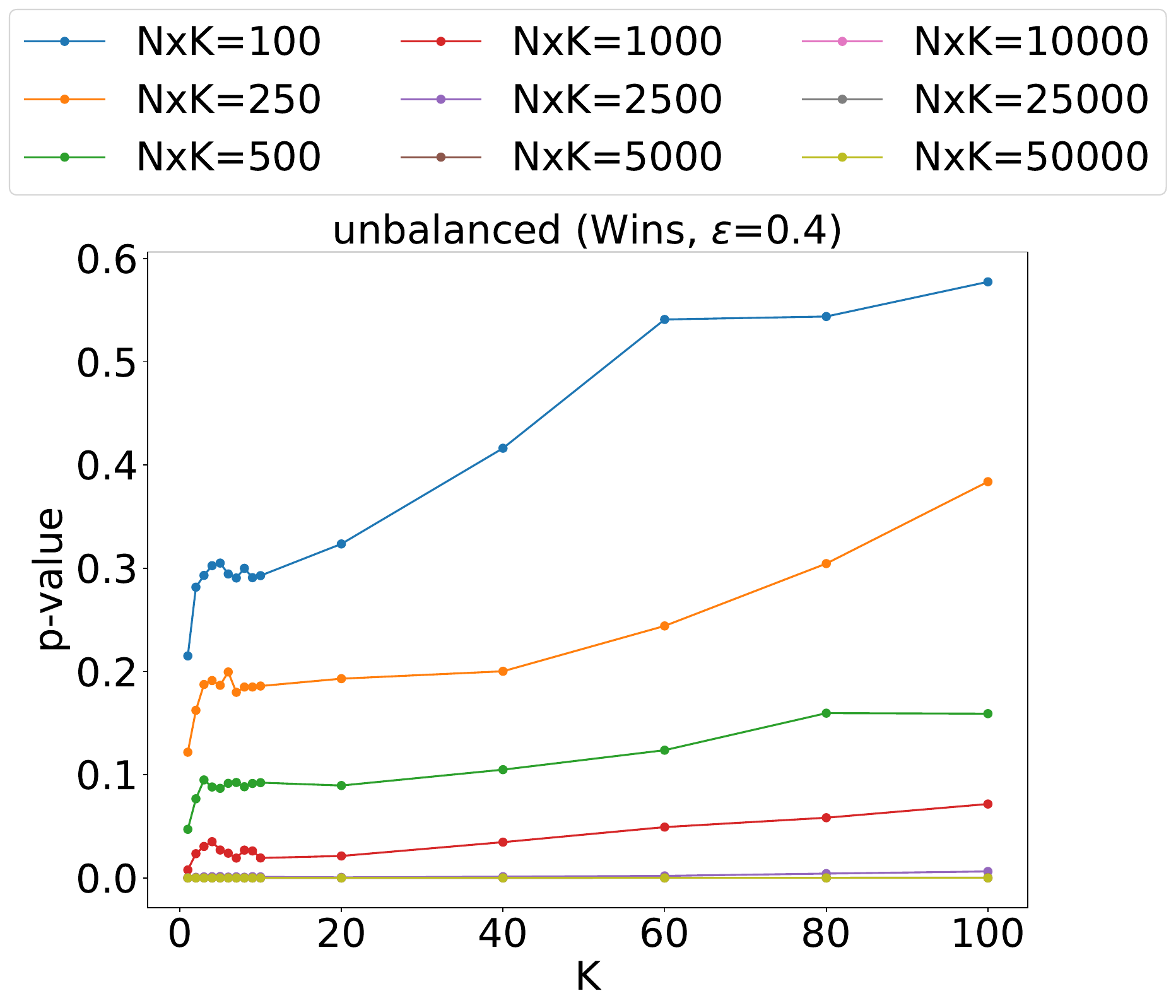}
    \caption{$\epsilon = 0.4$}
    \label{fig:gamma_wins_cat2_e04}
  \end{subfigure}
  \caption{P-value plots for unbalanced alphas with Wins as the metric ($M=2$)}
  \label{fig:gamma_wins_cat2}
\end{figure*}

\begin{figure*}
  \centering
  \begin{subfigure}[b]{0.24\linewidth}
    \centering
    \includegraphics[width=\linewidth]{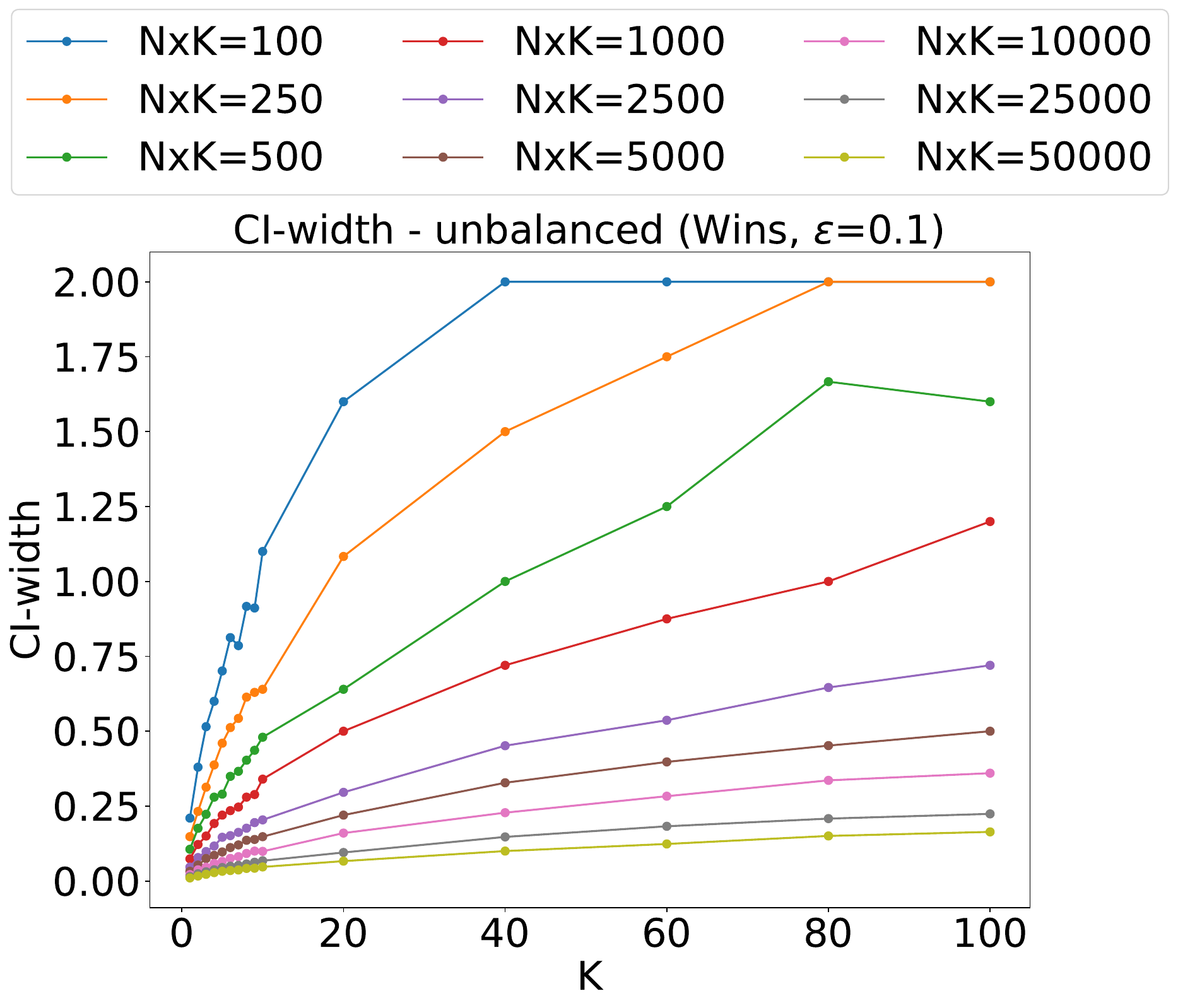}
    \caption{$\epsilon = 0.1$}
    \label{fig:gamma_ci_wins_cat2_e01}
  \end{subfigure} \hfill
  \begin{subfigure}[b]{0.24\linewidth}
    \centering
    \includegraphics[width=\linewidth]{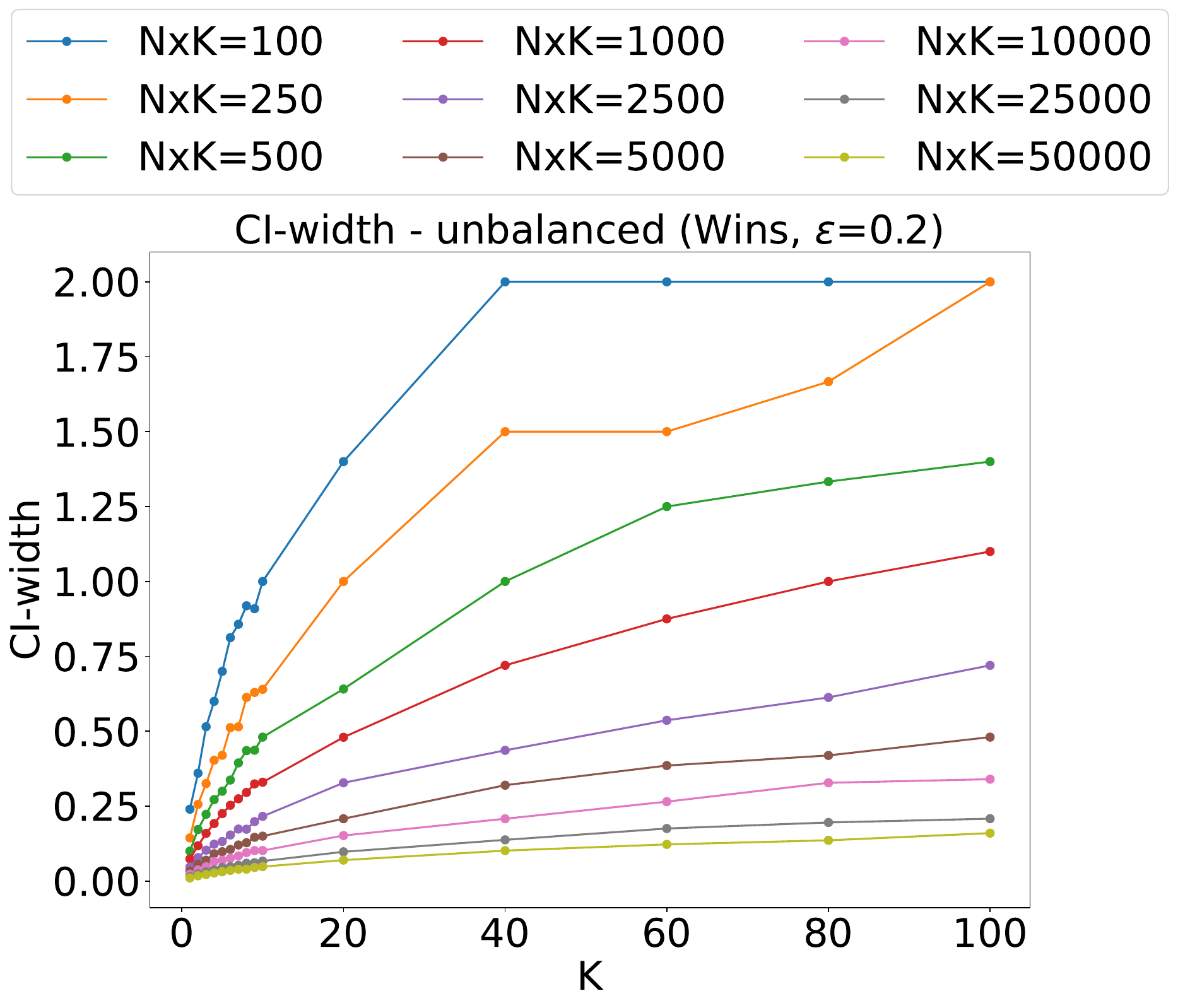}
    \caption{$\epsilon = 0.2$}
    \label{fig:gamma_ci_wins_cat2_e02}
  \end{subfigure} \hfill
  \begin{subfigure}[b]{0.24\linewidth}
    \centering
    \includegraphics[width=\linewidth]{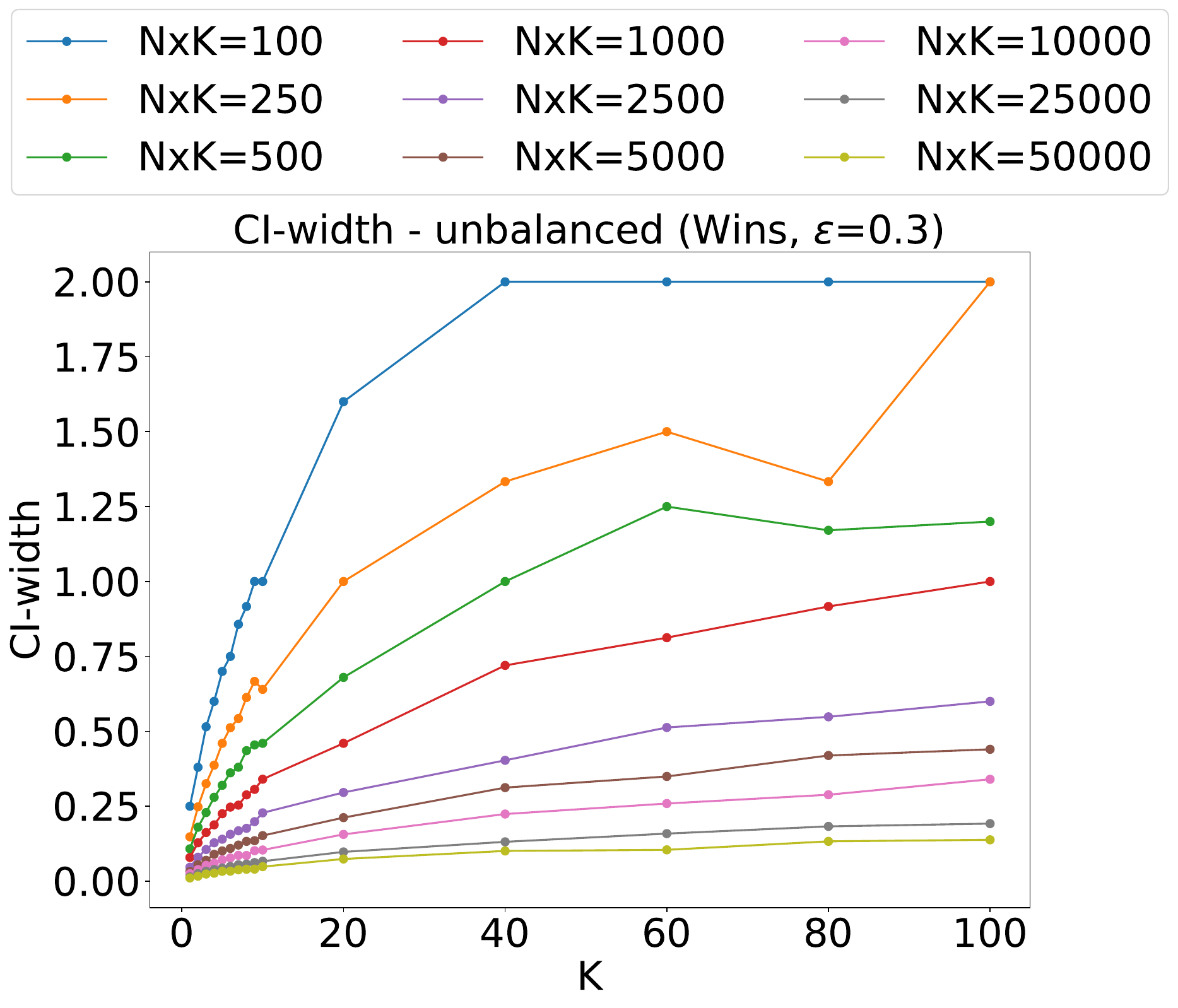}
    \caption{$\epsilon = 0.3$}
    \label{fig:gamma_ci_wins_cat2_e03}
  \end{subfigure} \hfill
  \begin{subfigure}[b]{0.24\linewidth}
    \centering
    \includegraphics[width=\linewidth]{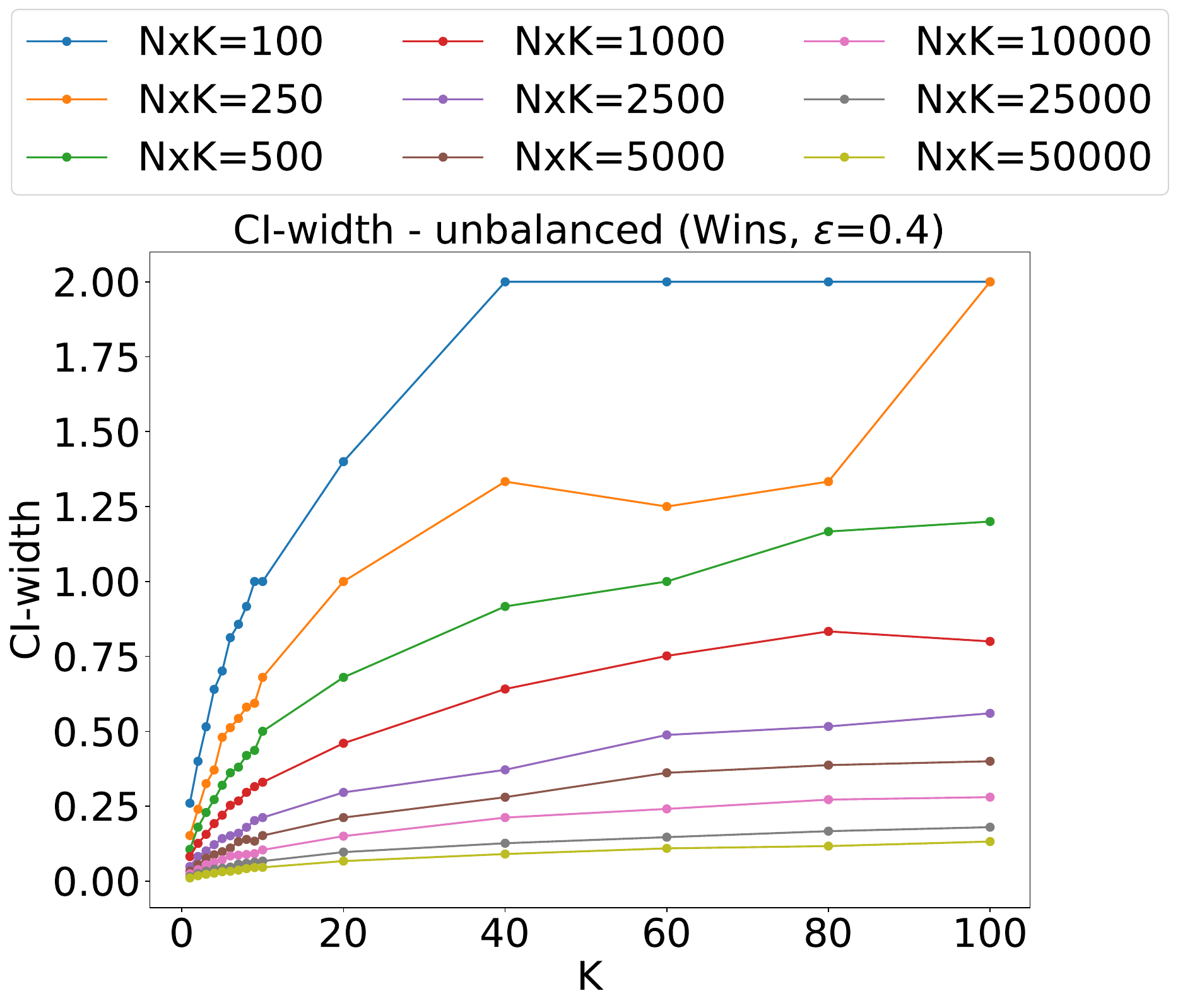}
    \caption{$\epsilon = 0.4$}
    \label{fig:gamma_ci_wins_cat2_e04}
  \end{subfigure}
  \caption{CI-width plots for unbalanced alphas with Wins as the metric ($M=2$)}
  \label{fig:gamma_ci_wins_cat2}
\end{figure*}

\begin{figure*}
  \centering
  \begin{subfigure}[b]{0.24\linewidth}
    \centering
    \includegraphics[width=\linewidth]{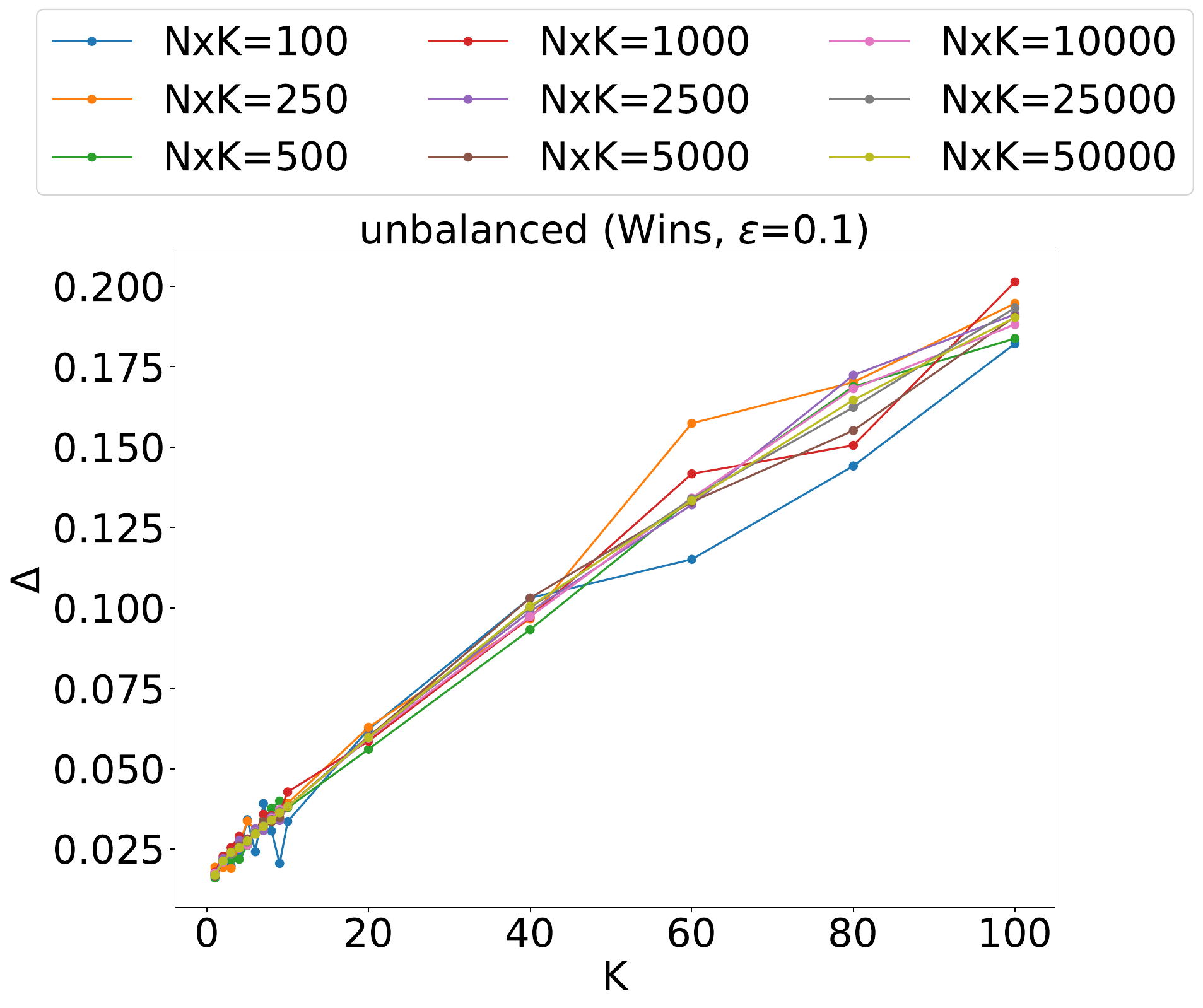}
    \caption{$\epsilon = 0.1$}
    \label{fig:gamma_delta_wins_cat2_e01}
  \end{subfigure} \hfill
  \begin{subfigure}[b]{0.24\linewidth}
    \centering
    \includegraphics[width=\linewidth]{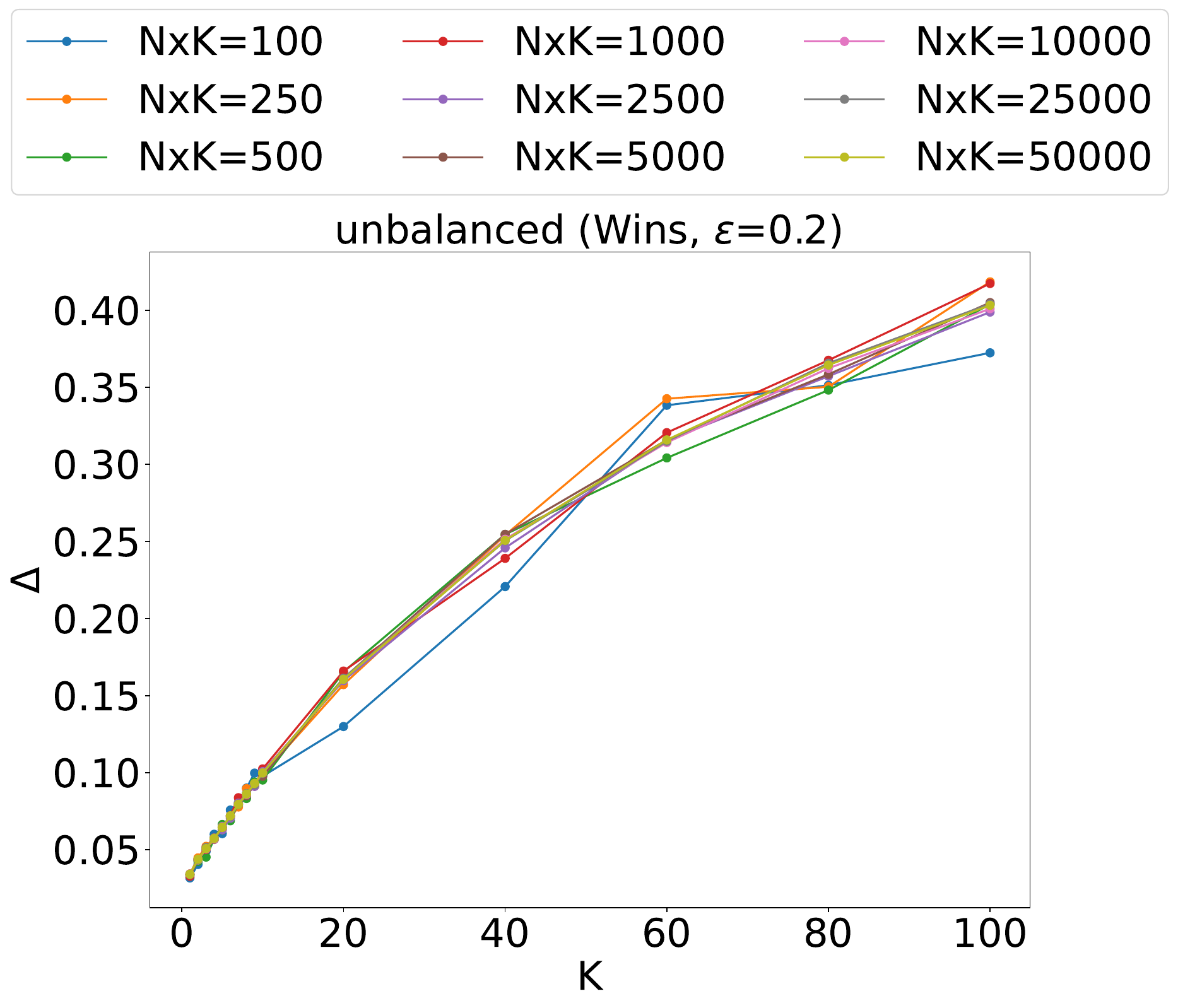}
    \caption{$\epsilon = 0.2$}
    \label{fig:gamma_delta_wins_cat2_e02}
  \end{subfigure} \hfill
  \begin{subfigure}[b]{0.24\linewidth}
    \centering
    \includegraphics[width=\linewidth]{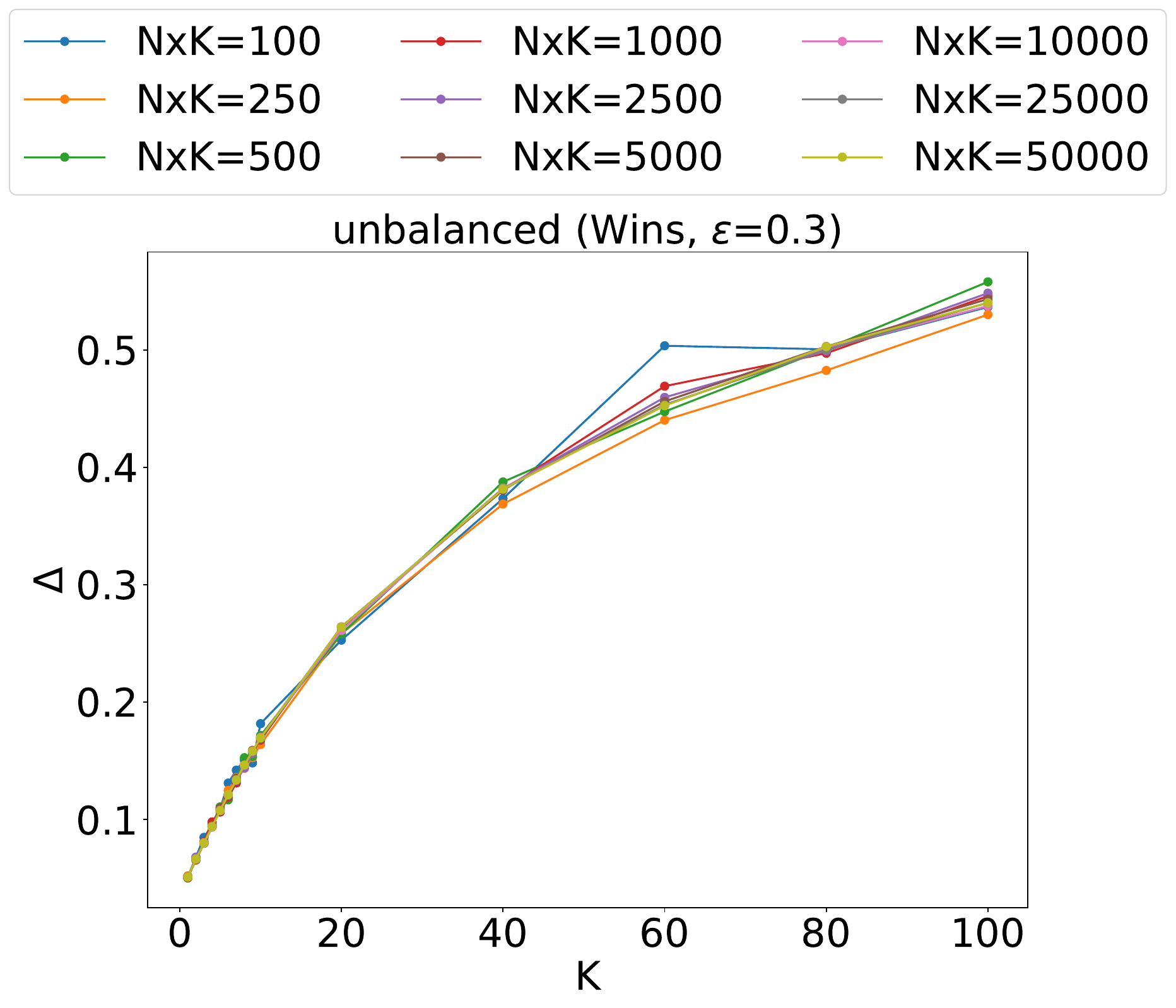}
    \caption{$\epsilon = 0.3$}
    \label{fig:gamma_delta_wins_cat2_e03}
  \end{subfigure} \hfill
  \begin{subfigure}[b]{0.24\linewidth}
    \centering
    \includegraphics[width=\linewidth]{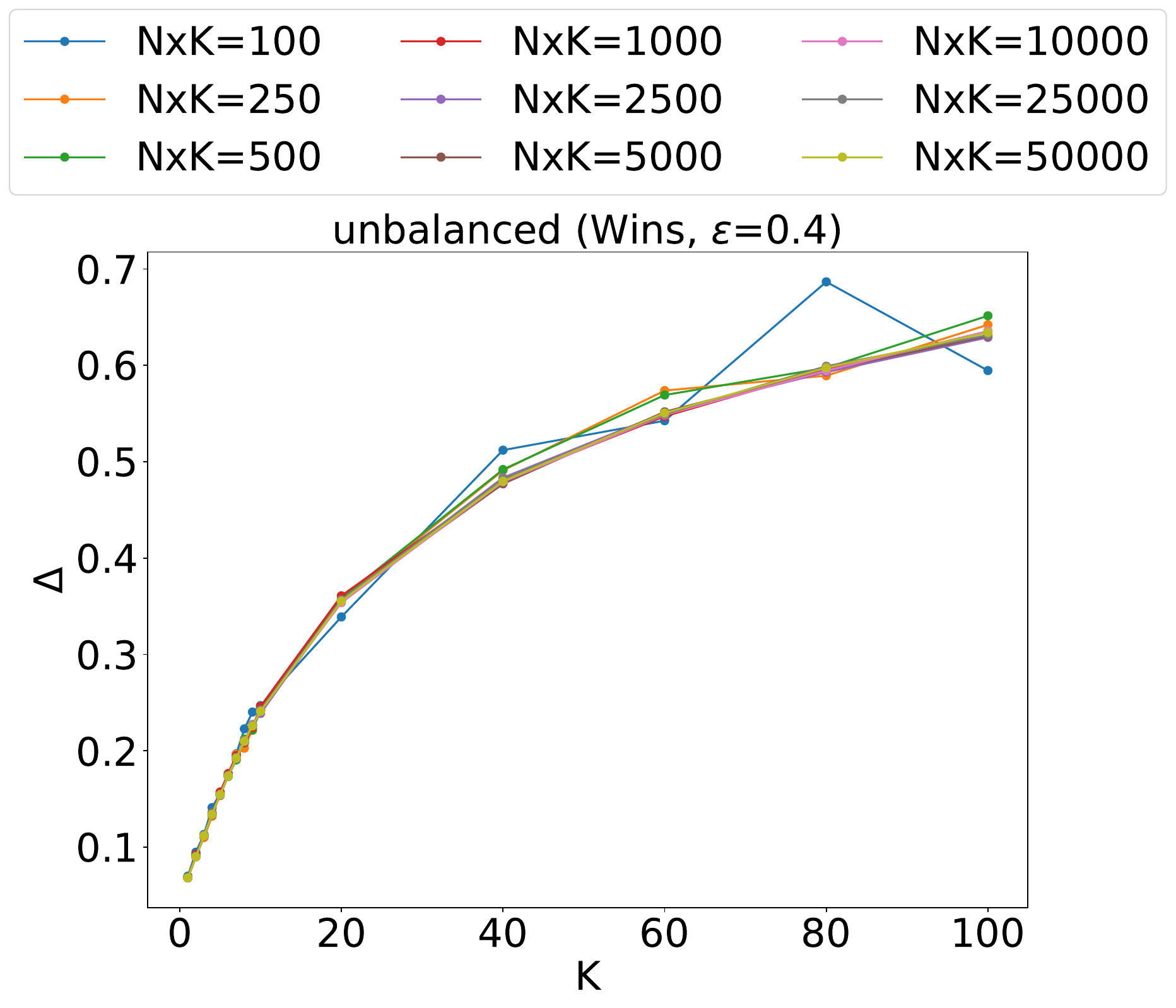}
    \caption{$\epsilon = 0.4$}
    \label{fig:gamma_delta_wins_cat2_e04}
  \end{subfigure}
  \caption{Effect sizes ($\Delta$) for unbalanced alphas with Wins as the metric ($M=2$)}
  \label{fig:gamma_delta_wins_cat2}
\end{figure*}

\begin{figure*}
  \centering
  \begin{subfigure}[b]{0.24\linewidth}
    \centering
    \includegraphics[width=\linewidth]{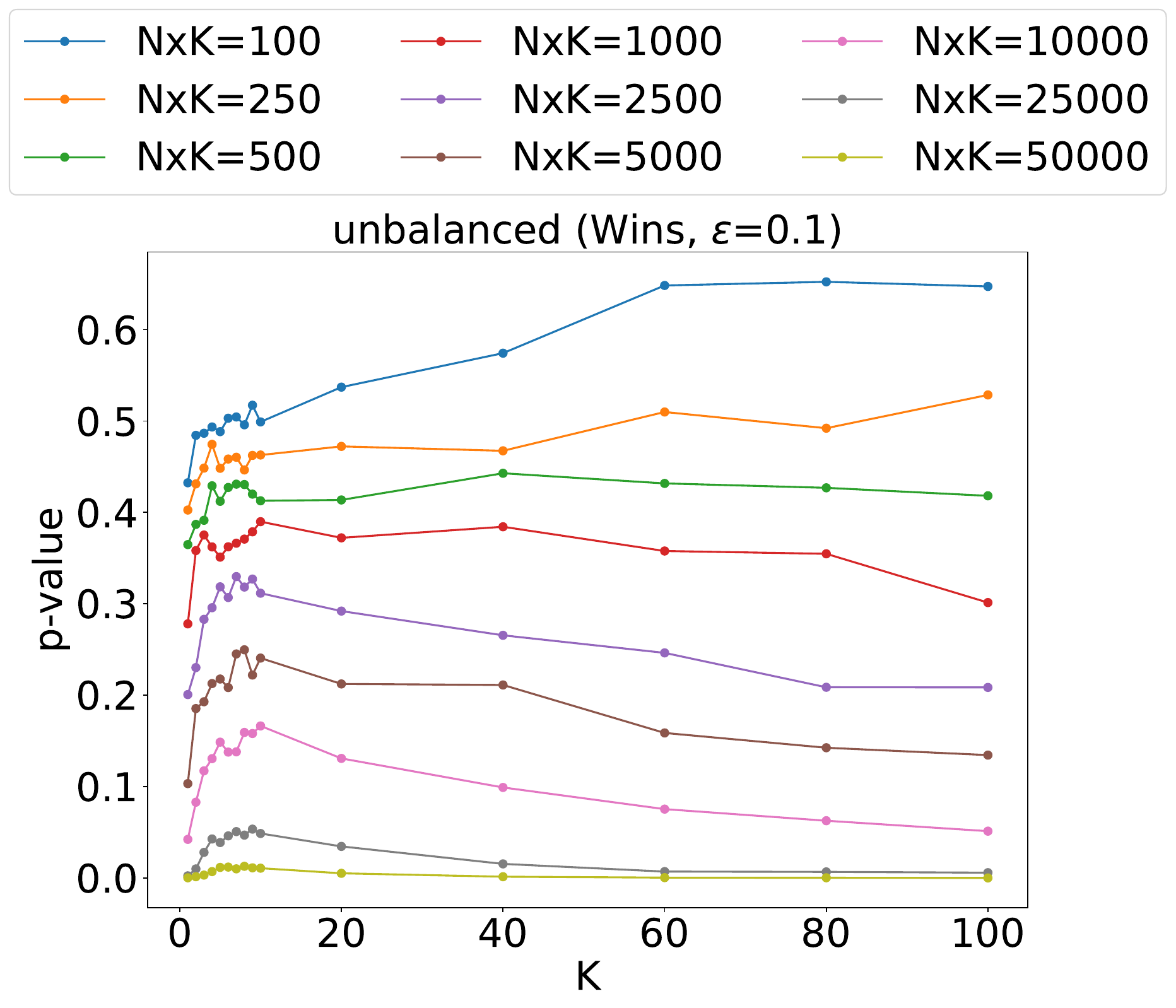}
    \caption{$\epsilon = 0.1$}
    \label{fig:gamma_wins_cat3_e01}
  \end{subfigure} \hfill
  \begin{subfigure}[b]{0.24\linewidth}
    \centering
    \includegraphics[width=\linewidth]{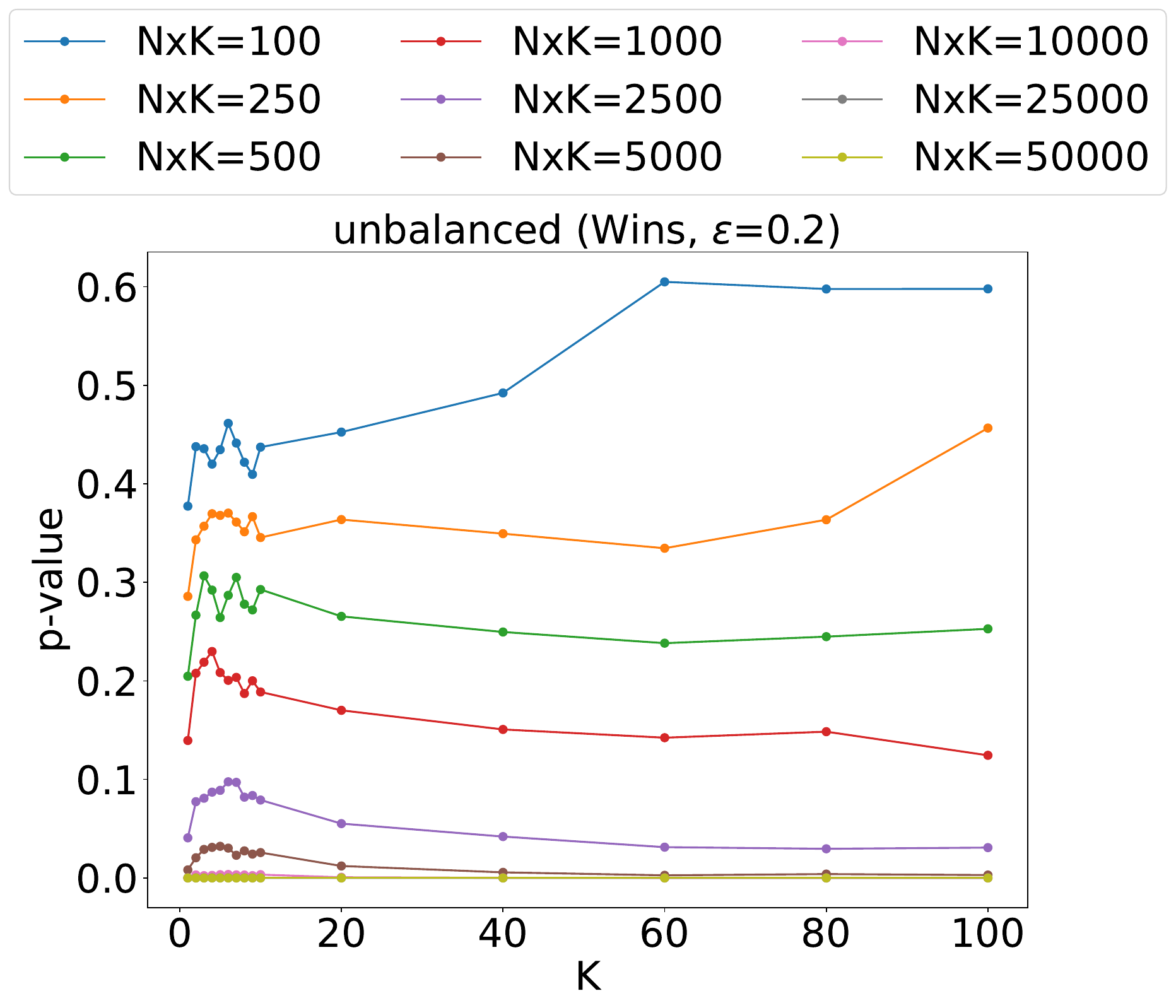}
    \caption{$\epsilon = 0.2$}
    \label{fig:gamma_wins_cat3_e02}
  \end{subfigure} \hfill
  \begin{subfigure}[b]{0.24\linewidth}
    \centering
    \includegraphics[width=\linewidth]{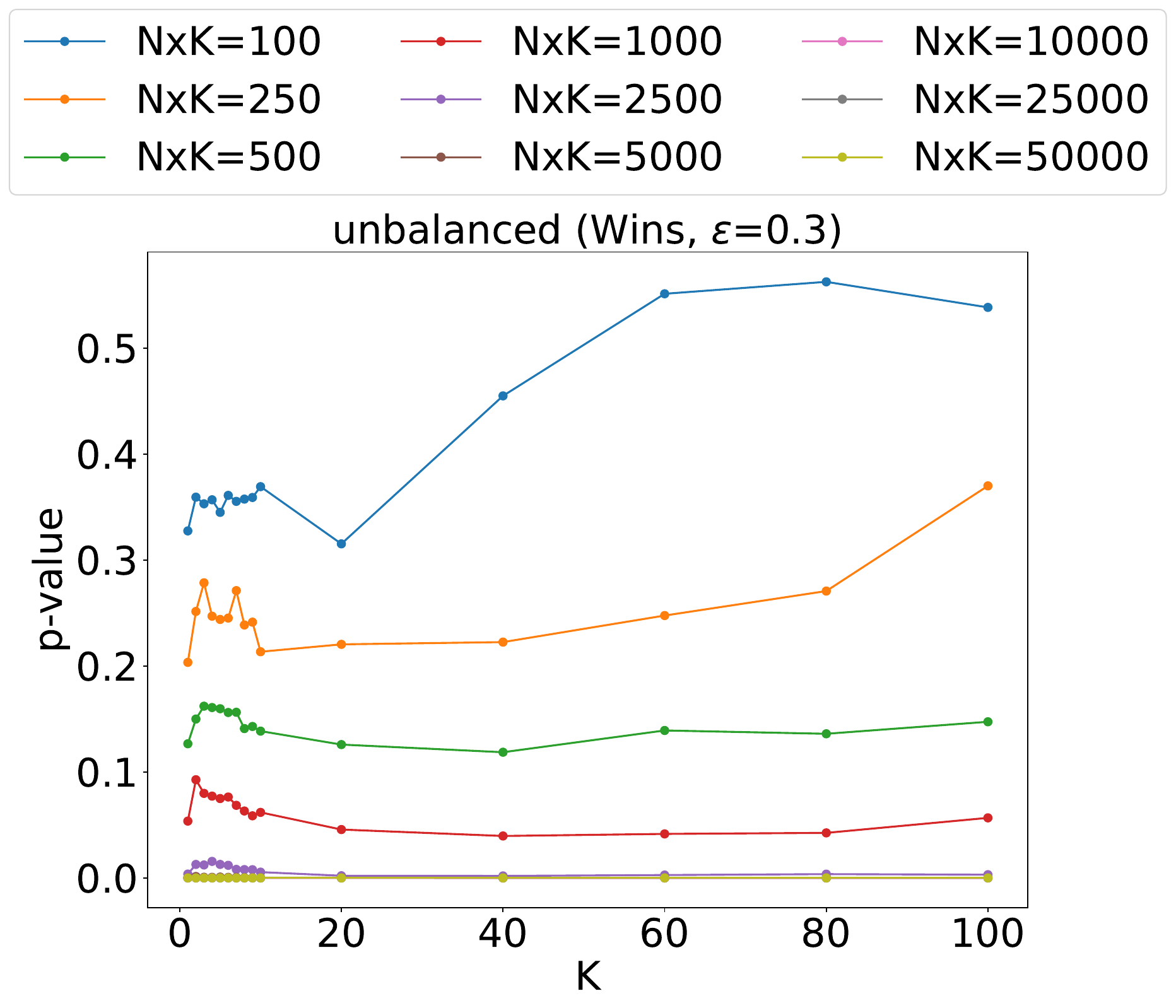}
    \caption{$\epsilon = 0.3$}
    \label{fig:gamma_wins_cat3_e03}
  \end{subfigure} \hfill
  \begin{subfigure}[b]{0.24\linewidth}
    \centering
    \includegraphics[width=\linewidth]{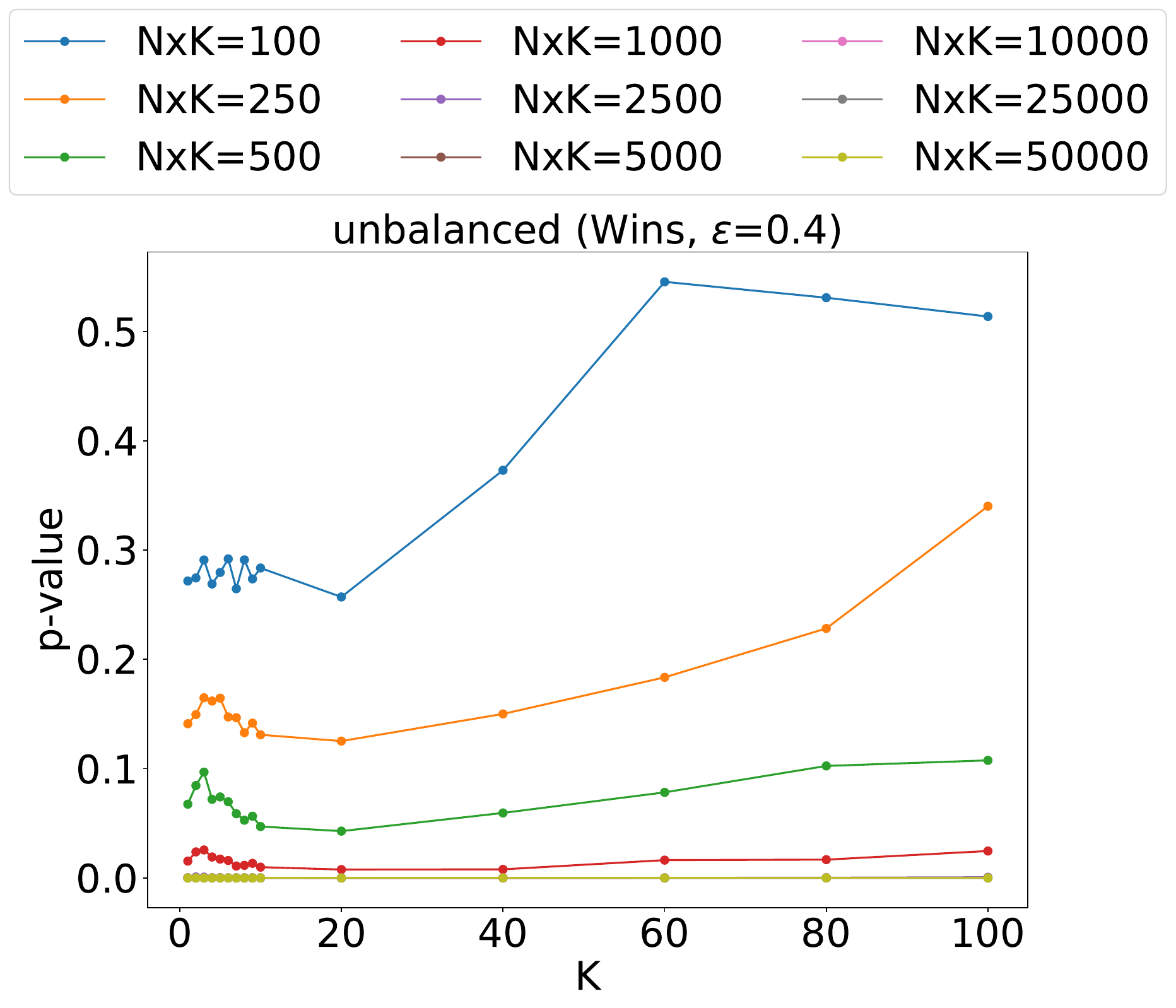}
    \caption{$\epsilon = 0.4$}
    \label{fig:gamma_wins_cat3_e04}
  \end{subfigure}
  \caption{P-value plots for unbalanced alphas with Wins as the metric ($M=3$)}
  \label{fig:gamma_wins_cat3}
\end{figure*}

\begin{figure*}
  \centering
  \begin{subfigure}[b]{0.24\linewidth}
    \centering
    \includegraphics[width=\linewidth]{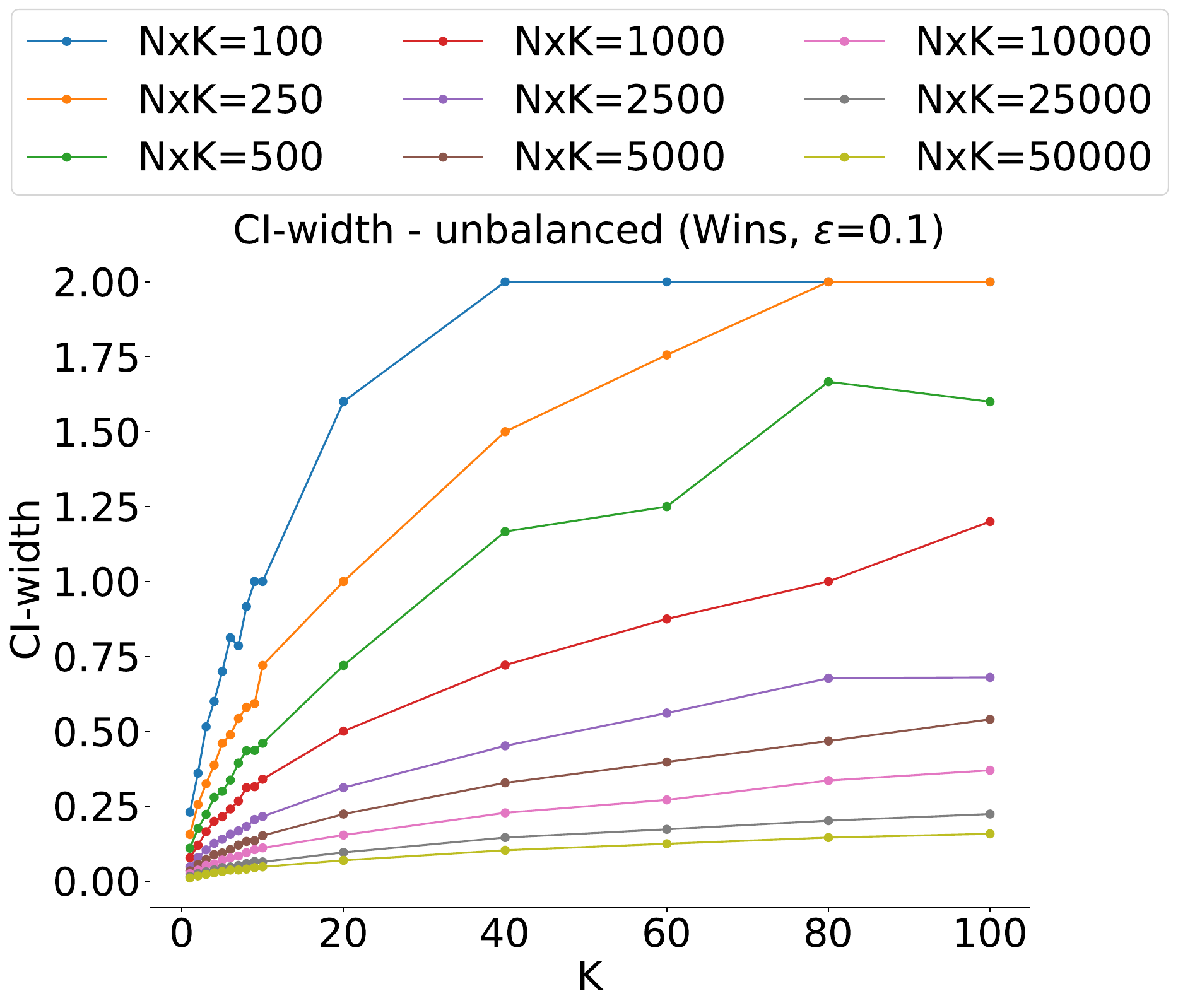}
    \caption{$\epsilon = 0.1$}
    \label{fig:gamma_ci_wins_cat3_e01}
  \end{subfigure} \hfill
  \begin{subfigure}[b]{0.24\linewidth}
    \centering
    \includegraphics[width=\linewidth]{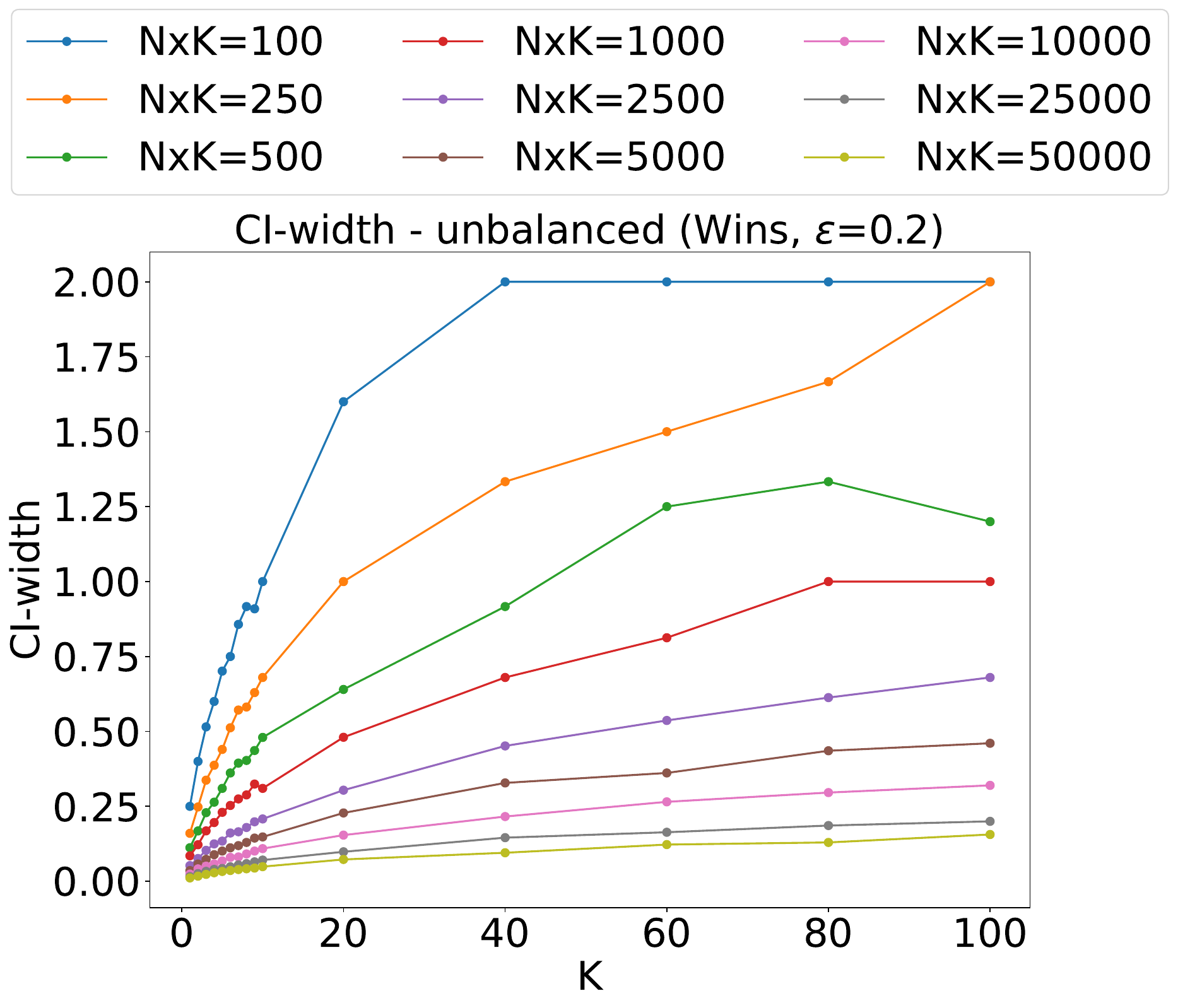}
    \caption{$\epsilon = 0.2$}
    \label{fig:gamma_ci_wins_cat3_e02}
  \end{subfigure} \hfill
  \begin{subfigure}[b]{0.24\linewidth}
    \centering
    \includegraphics[width=\linewidth]{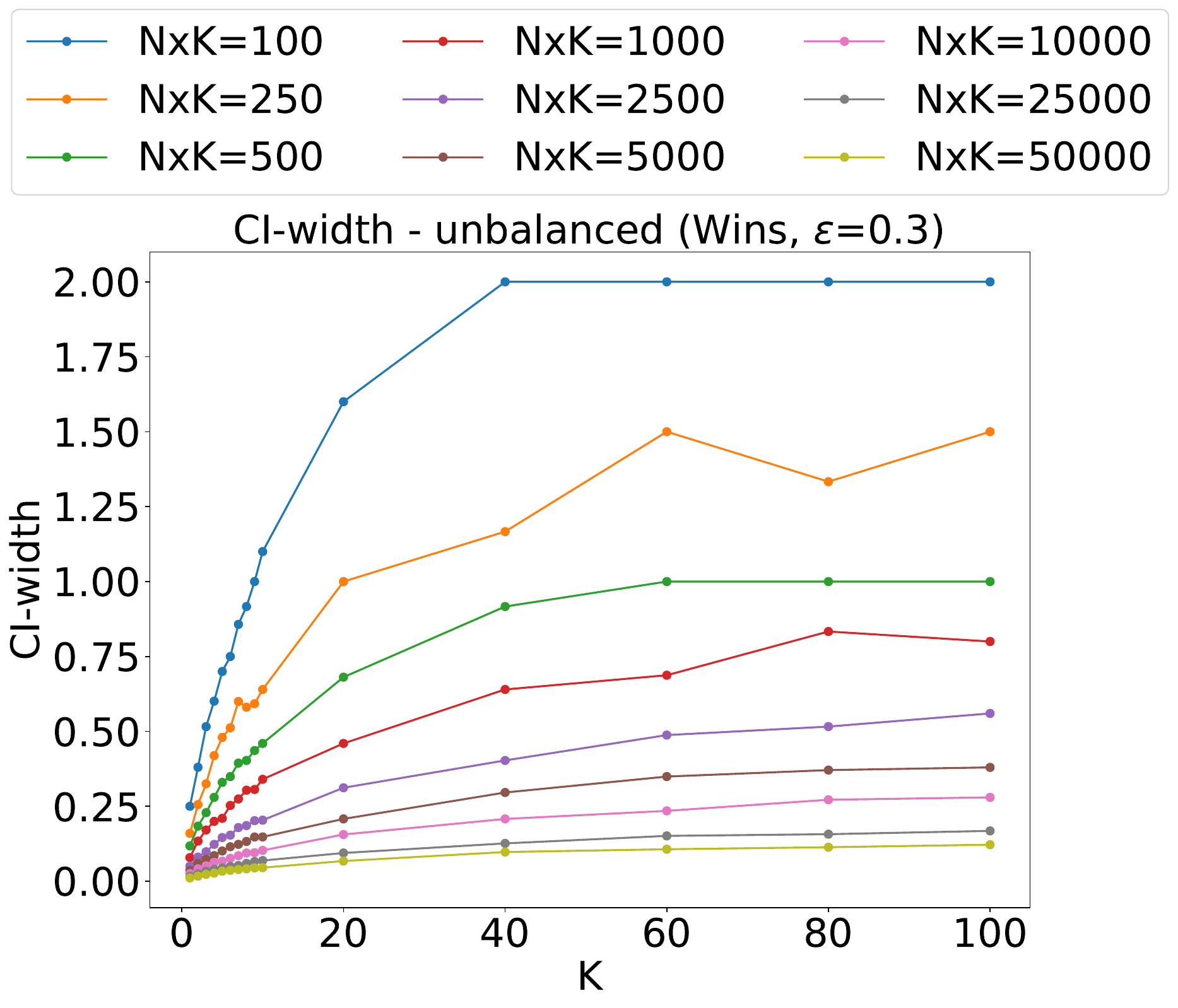}
    \caption{$\epsilon = 0.3$}
    \label{fig:gamma_ci_wins_cat3_e03}
  \end{subfigure} \hfill
  \begin{subfigure}[b]{0.24\linewidth}
    \centering
    \includegraphics[width=\linewidth]{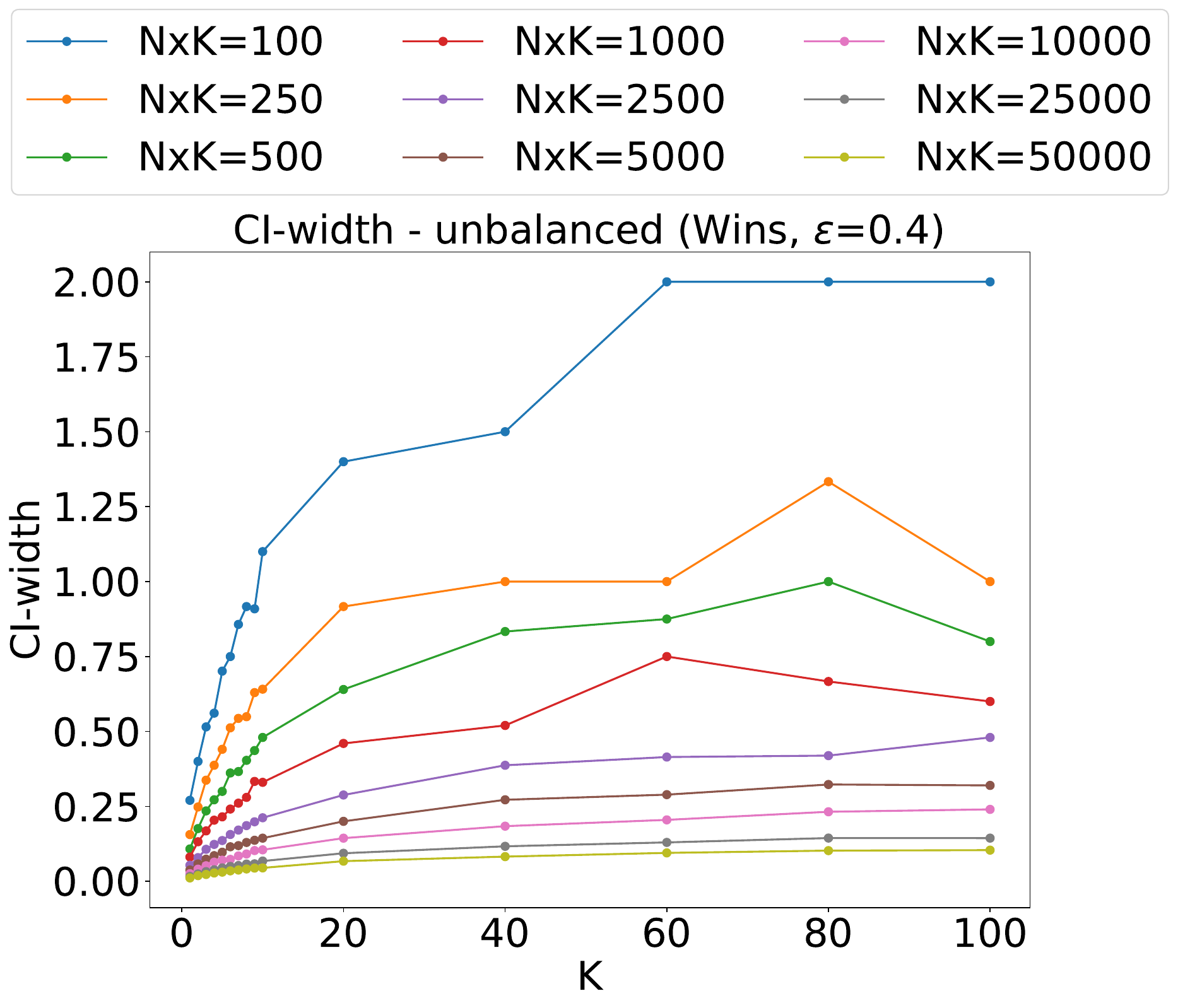}
    \caption{$\epsilon = 0.4$}
    \label{fig:gamma_ci_wins_cat3_e04}
  \end{subfigure}
  \caption{CI-width plots for unbalanced alphas with Wins as the metric ($M=3$)}
  \label{fig:gamma_ci_wins_cat3}
\end{figure*}

\begin{figure*}
  \centering
  \begin{subfigure}[b]{0.24\linewidth}
    \centering
    \includegraphics[width=\linewidth]{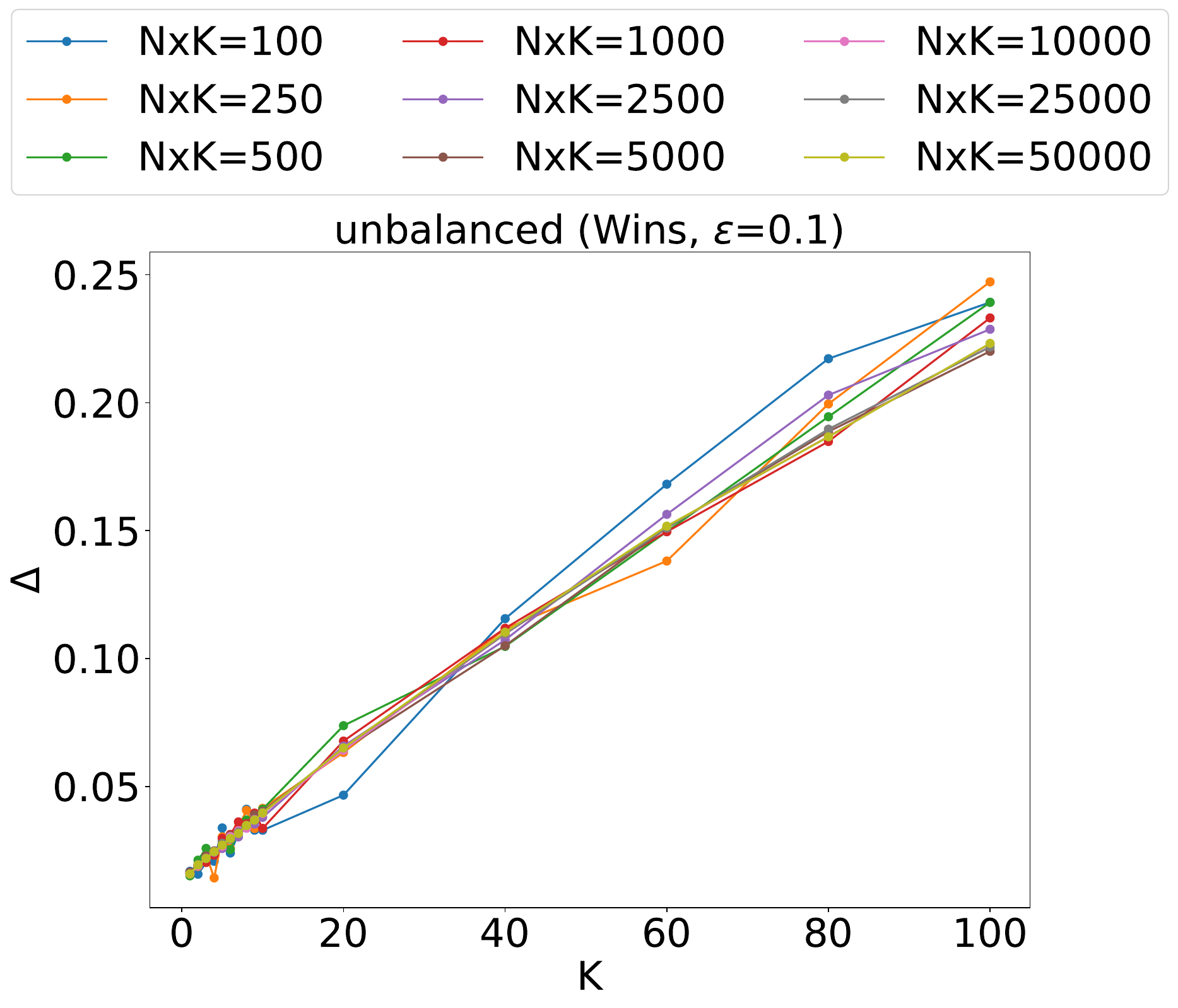}
    \caption{$\epsilon = 0.1$}
    \label{fig:gamma_delta_wins_cat3_e01}
  \end{subfigure} \hfill
  \begin{subfigure}[b]{0.24\linewidth}
    \centering
    \includegraphics[width=\linewidth]{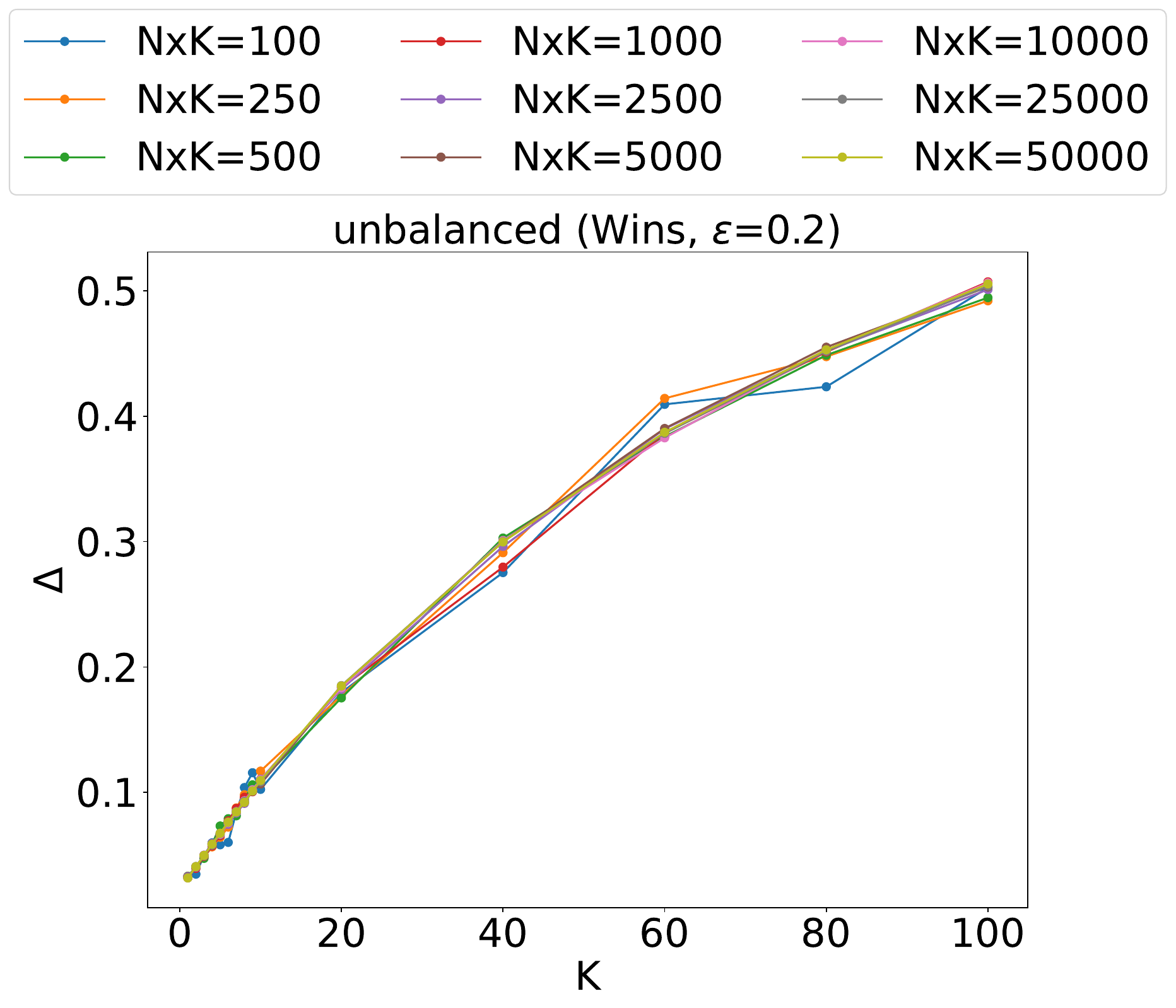}
    \caption{$\epsilon = 0.2$}
    \label{fig:gamma_delta_wins_cat3_e02}
  \end{subfigure} \hfill
  \begin{subfigure}[b]{0.24\linewidth}
    \centering
    \includegraphics[width=\linewidth]{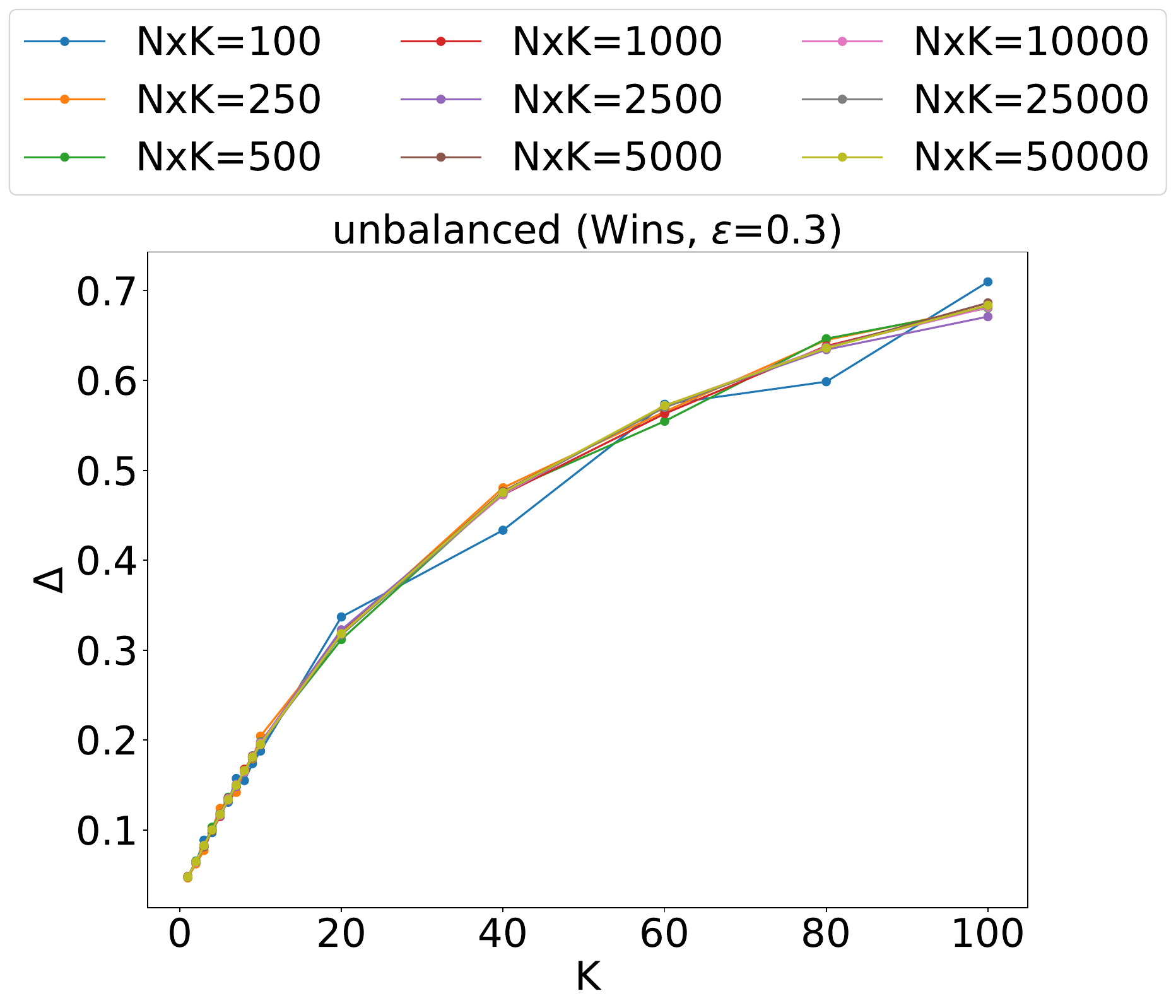}
    \caption{$\epsilon = 0.3$}
    \label{fig:gamma_delta_wins_cat3_e03}
  \end{subfigure} \hfill
  \begin{subfigure}[b]{0.24\linewidth}
    \centering
    \includegraphics[width=\linewidth]{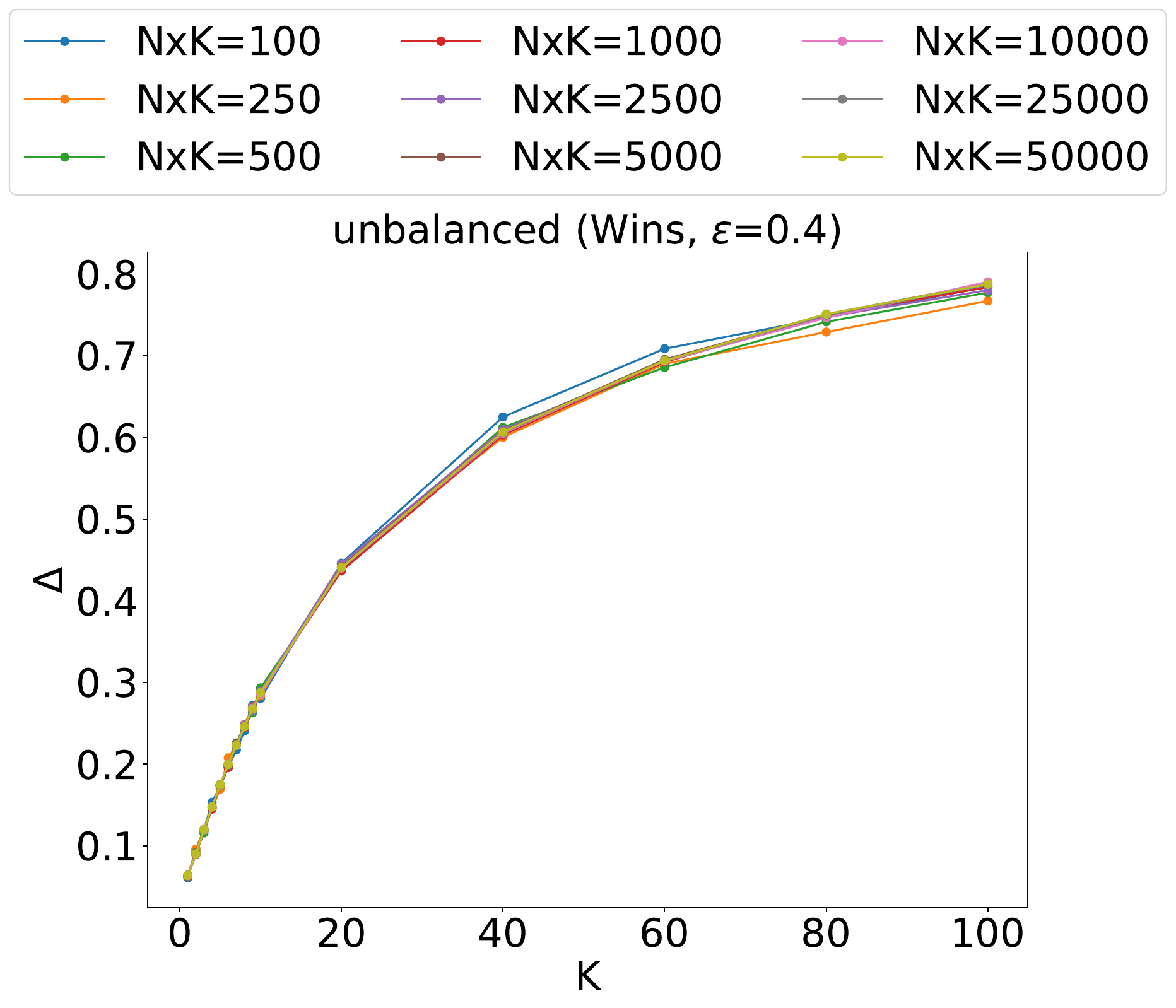}
    \caption{$\epsilon = 0.4$}
    \label{fig:gamma_delta_wins_cat3_e04}
  \end{subfigure}
  \caption{Effect sizes ($\Delta$) for unbalanced alphas with Wins as the metric ($M=3$)}
  \label{fig:gamma_delta_wins_cat3}
\end{figure*}

\begin{figure*}
  \centering
  \begin{subfigure}[b]{0.24\linewidth}
    \centering
    \includegraphics[width=\linewidth]{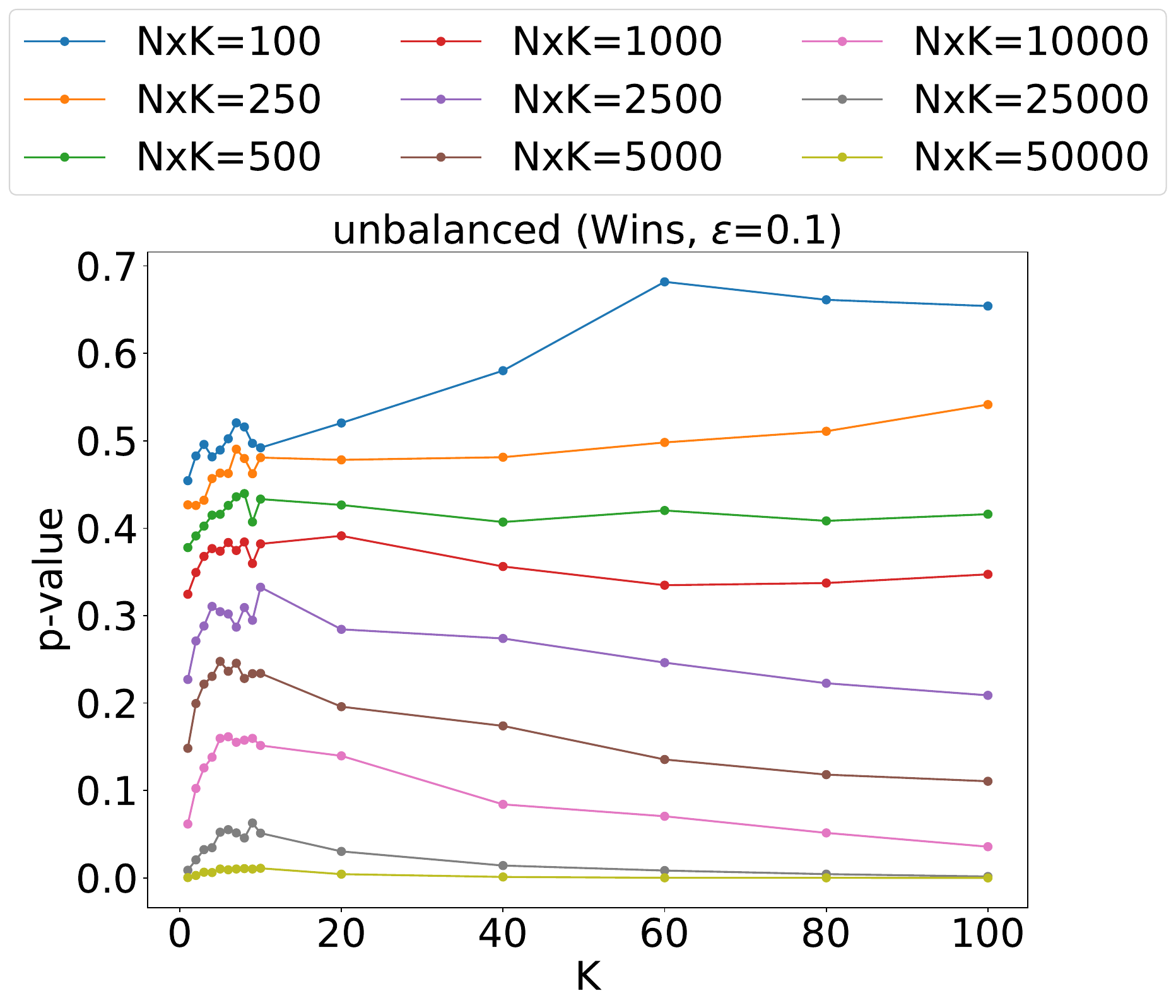}
    \caption{$\epsilon = 0.1$}
    \label{fig:gamma_wins_cat4_e01}
  \end{subfigure} \hfill
  \begin{subfigure}[b]{0.24\linewidth}
    \centering
    \includegraphics[width=\linewidth]{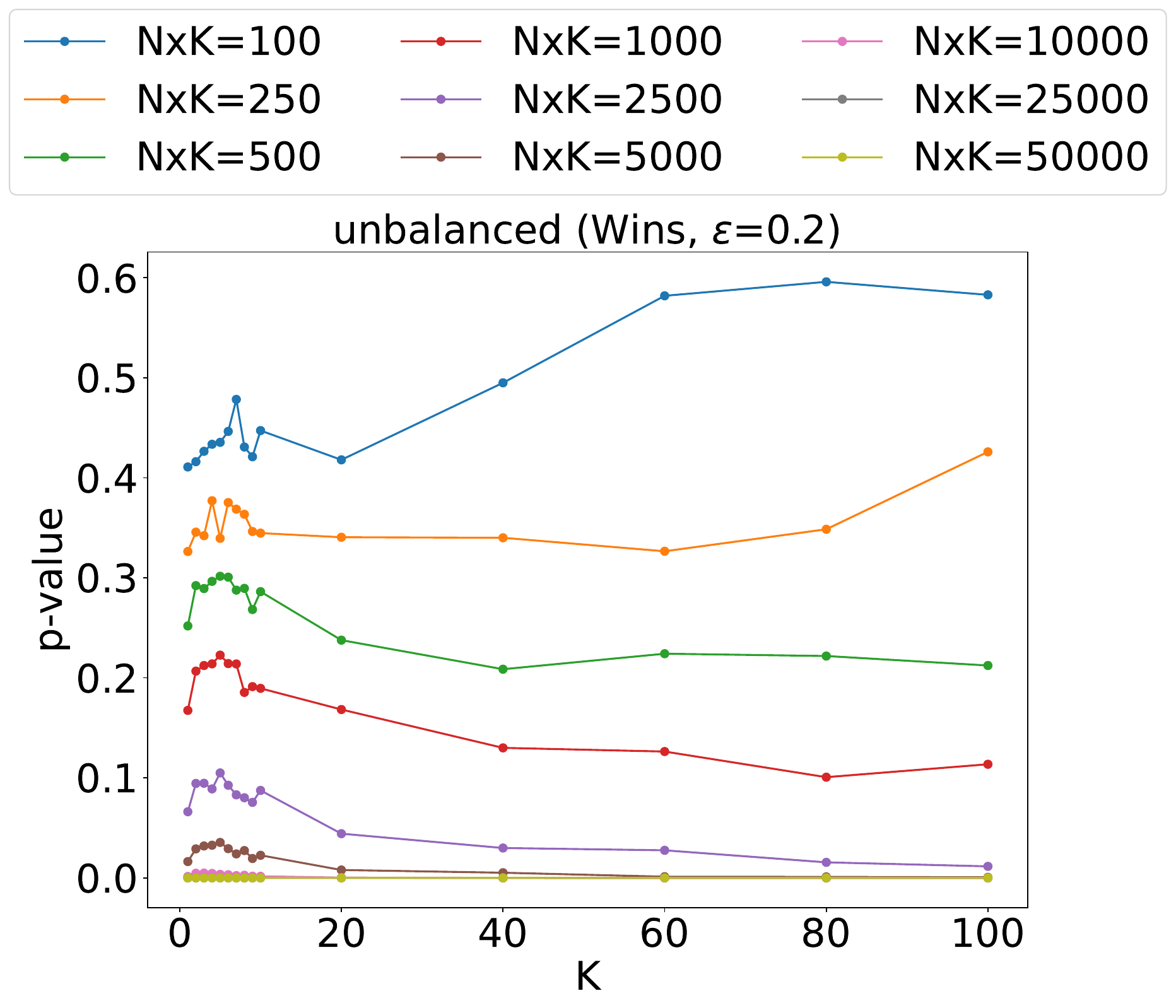}
    \caption{$\epsilon = 0.2$}
    \label{fig:gamma_wins_cat4_e02}
  \end{subfigure} \hfill
  \begin{subfigure}[b]{0.24\linewidth}
    \centering
    \includegraphics[width=\linewidth]{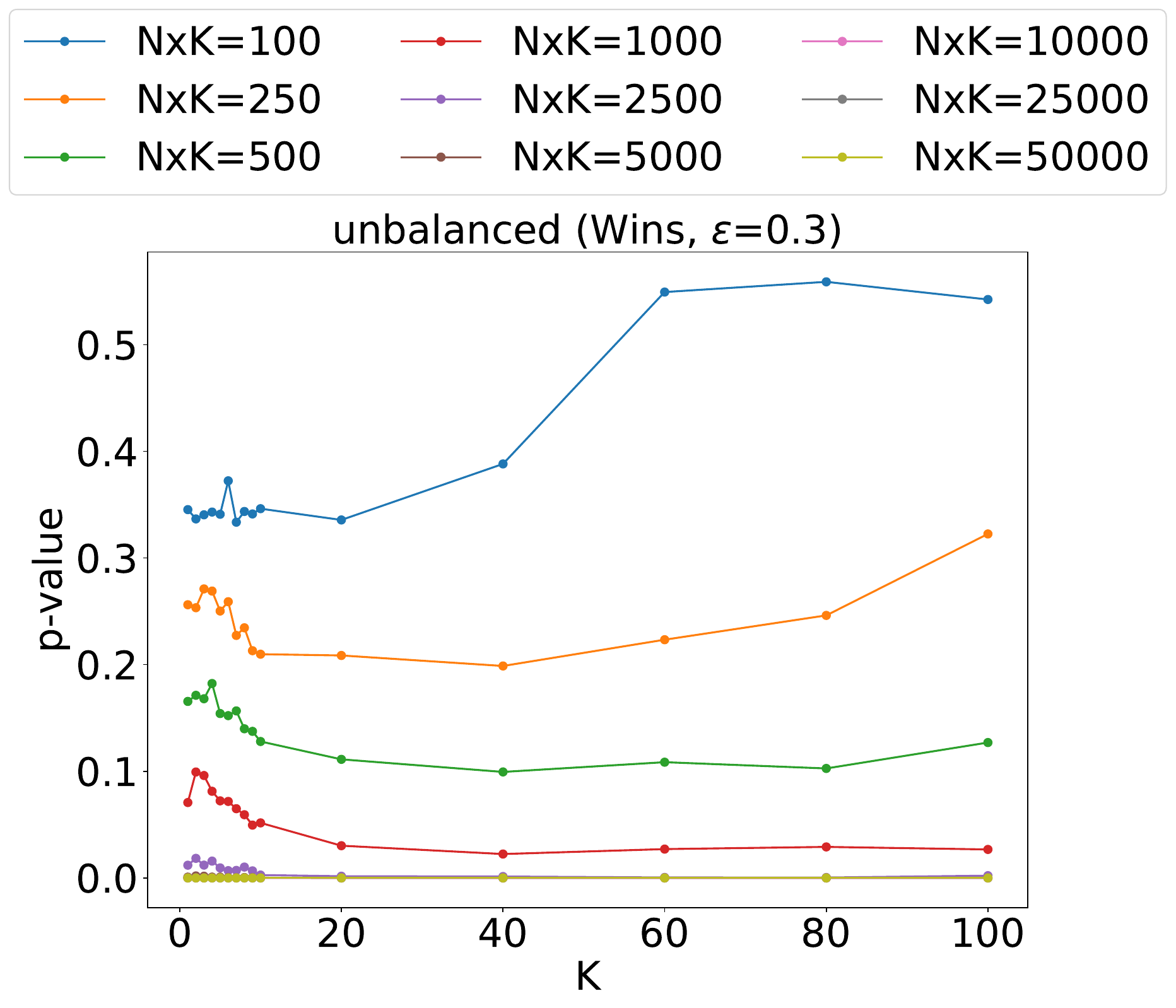}
    \caption{$\epsilon = 0.3$}
    \label{fig:gamma_wins_cat4_e03}
  \end{subfigure} \hfill
  \begin{subfigure}[b]{0.24\linewidth}
    \centering
    \includegraphics[width=\linewidth]{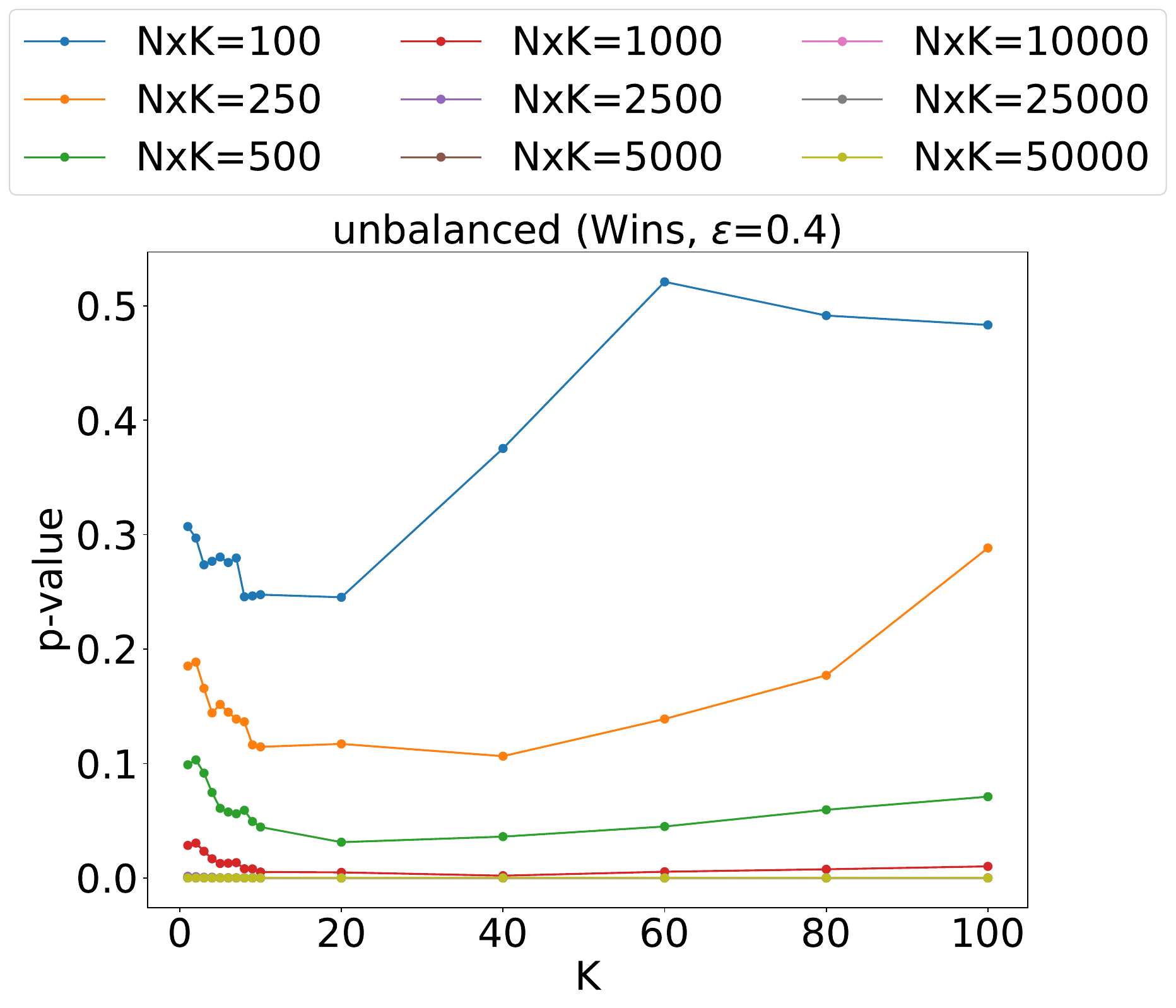}
    \caption{$\epsilon = 0.4$}
    \label{fig:gamma_wins_cat4_e04}
  \end{subfigure}
  \caption{P-value plots for unbalanced alphas with Wins as the metric ($M=4$)}
  \label{fig:gamma_wins_cat4}
\end{figure*}

\begin{figure*}
  \centering
  \begin{subfigure}[b]{0.24\linewidth}
    \centering
    \includegraphics[width=\linewidth]{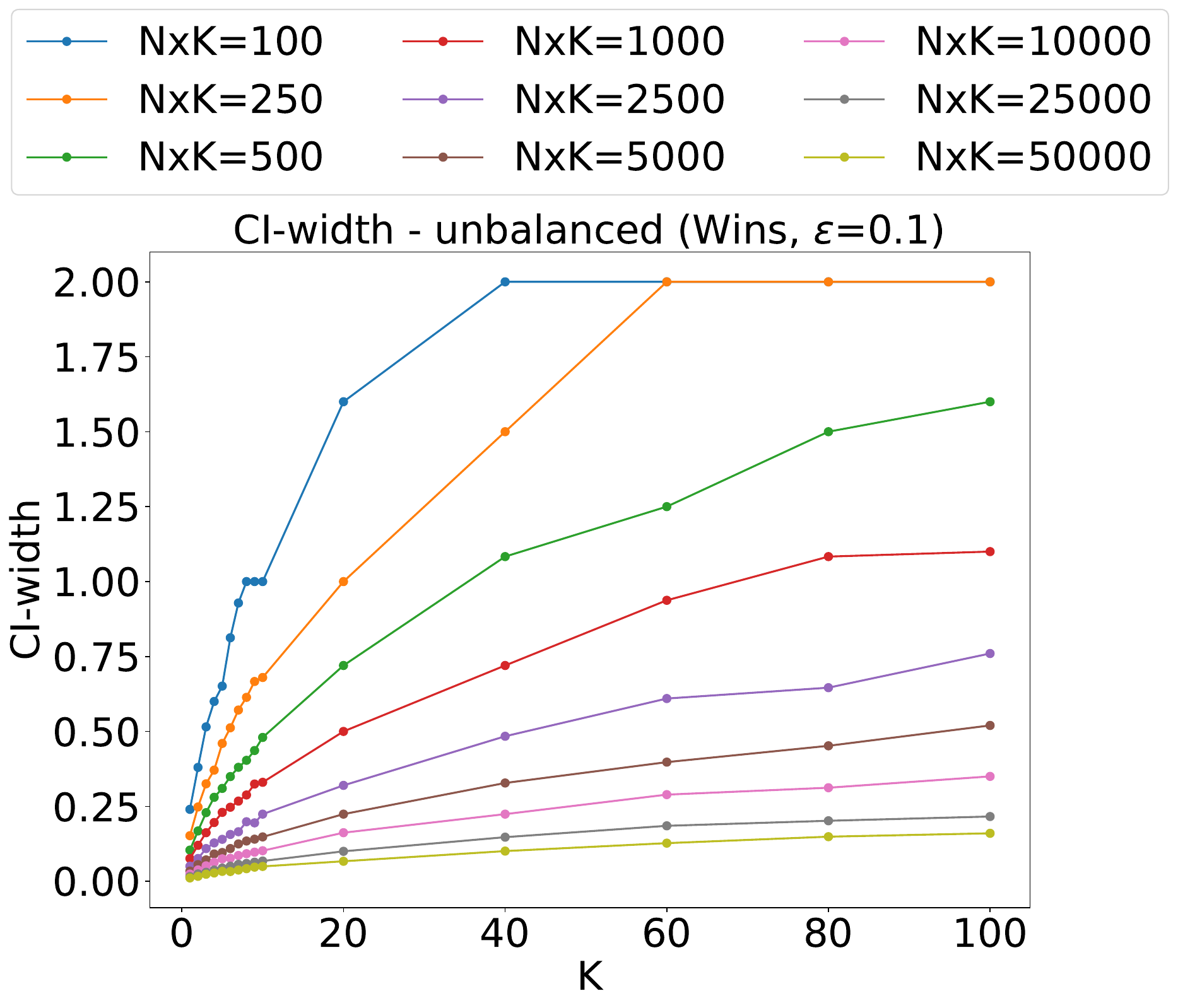}
    \caption{$\epsilon = 0.1$}
    \label{fig:gamma_ci_wins_cat4_e01}
  \end{subfigure} \hfill
  \begin{subfigure}[b]{0.24\linewidth}
    \centering
    \includegraphics[width=\linewidth]{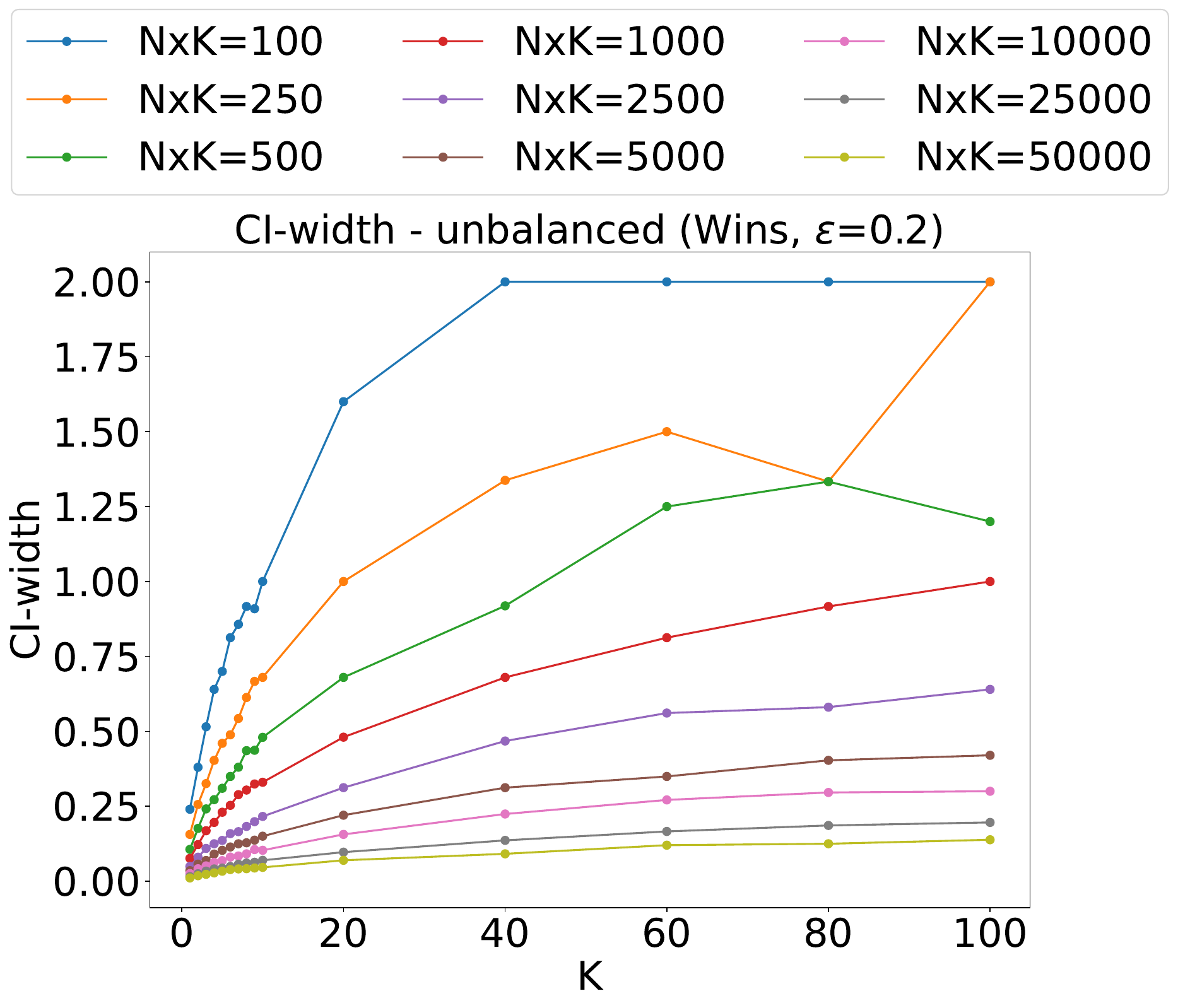}
    \caption{$\epsilon = 0.2$}
    \label{fig:gamma_ci_wins_cat4_e02}
  \end{subfigure} \hfill
  \begin{subfigure}[b]{0.24\linewidth}
    \centering
    \includegraphics[width=\linewidth]{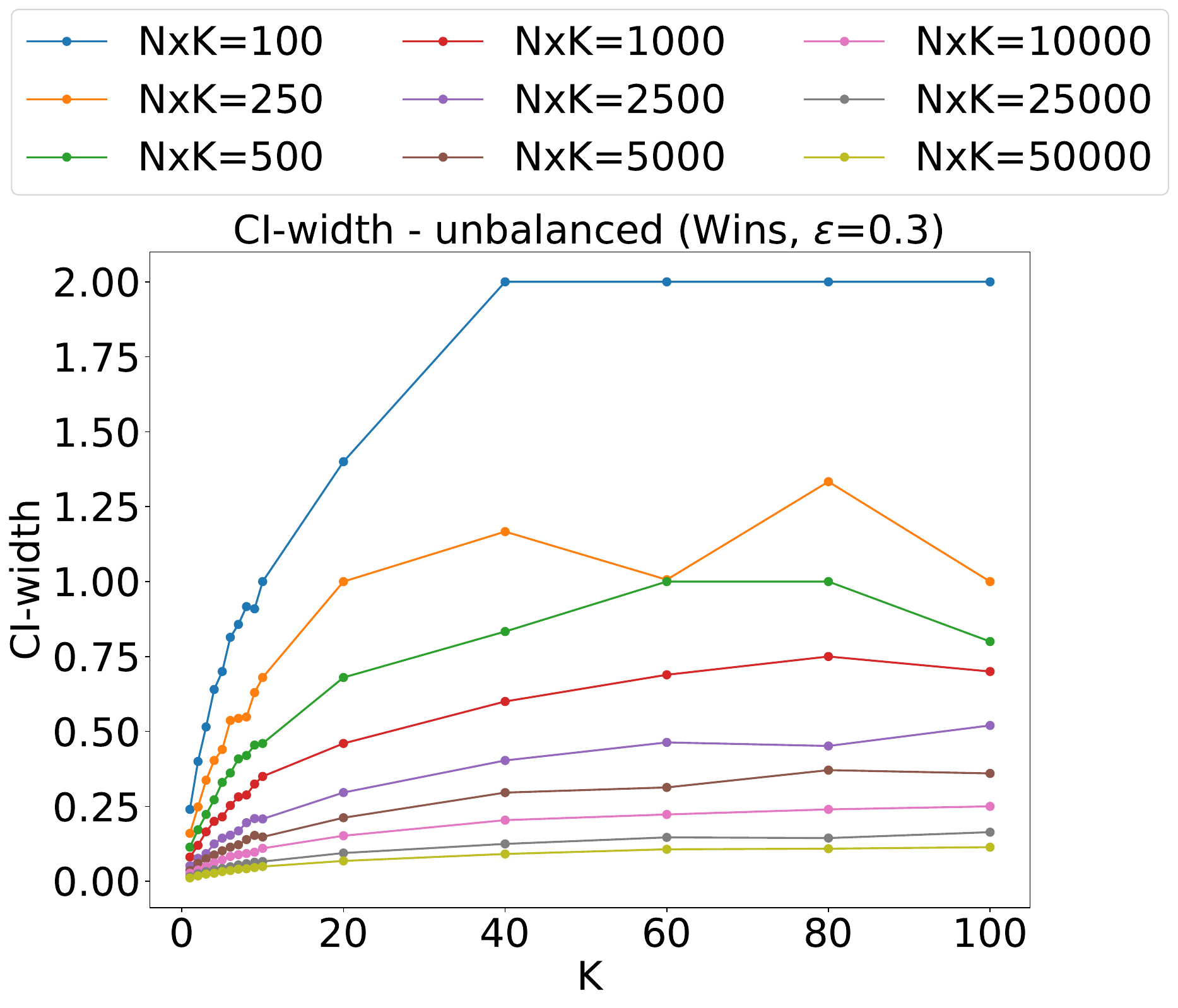}
    \caption{$\epsilon = 0.3$}
    \label{fig:gamma_ci_wins_cat4_e03}
  \end{subfigure} \hfill
  \begin{subfigure}[b]{0.24\linewidth}
    \centering
    \includegraphics[width=\linewidth]{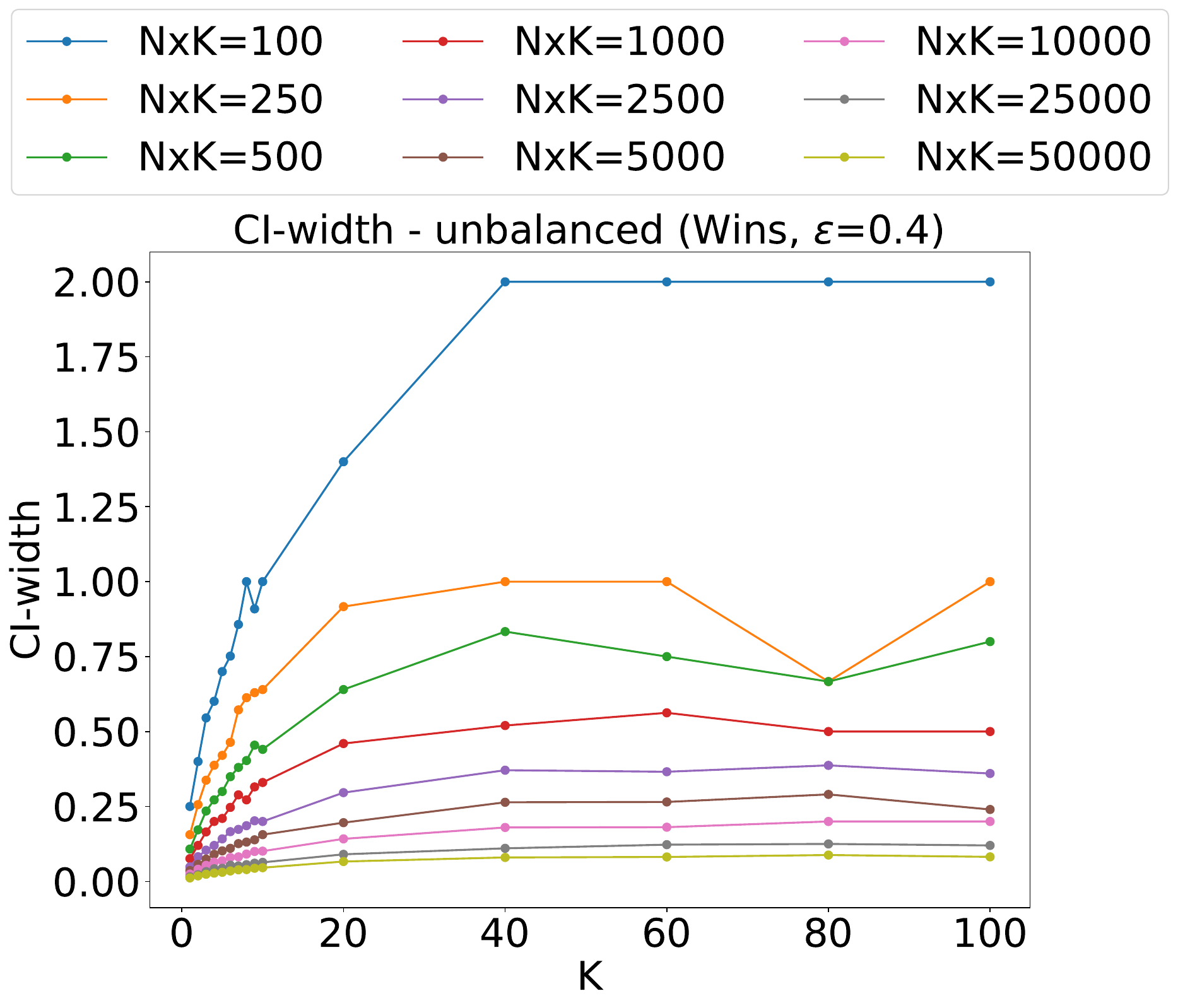}
    \caption{$\epsilon = 0.4$}
    \label{fig:gamma_ci_wins_cat4_e04}
  \end{subfigure}
  \caption{CI-width plots for unbalanced alphas with Wins as the metric ($M=4$)}
  \label{fig:gamma_ci_wins_cat4}
\end{figure*}

\begin{figure*}
  \centering
  \begin{subfigure}[b]{0.24\linewidth}
    \centering
    \includegraphics[width=\linewidth]{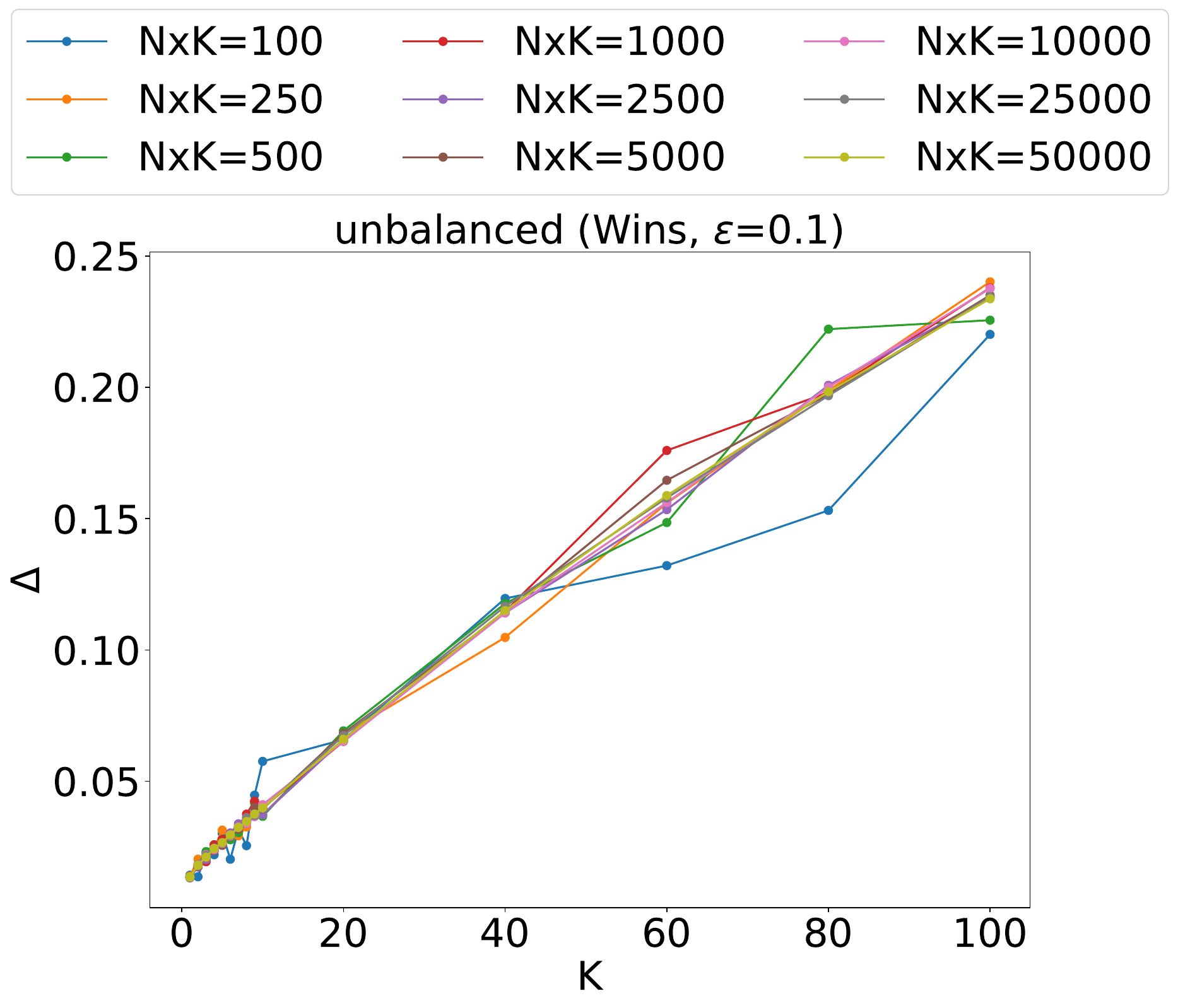}
    \caption{$\epsilon = 0.1$}
    \label{fig:gamma_delta_wins_cat4_e01}
  \end{subfigure} \hfill
  \begin{subfigure}[b]{0.24\linewidth}
    \centering
    \includegraphics[width=\linewidth]{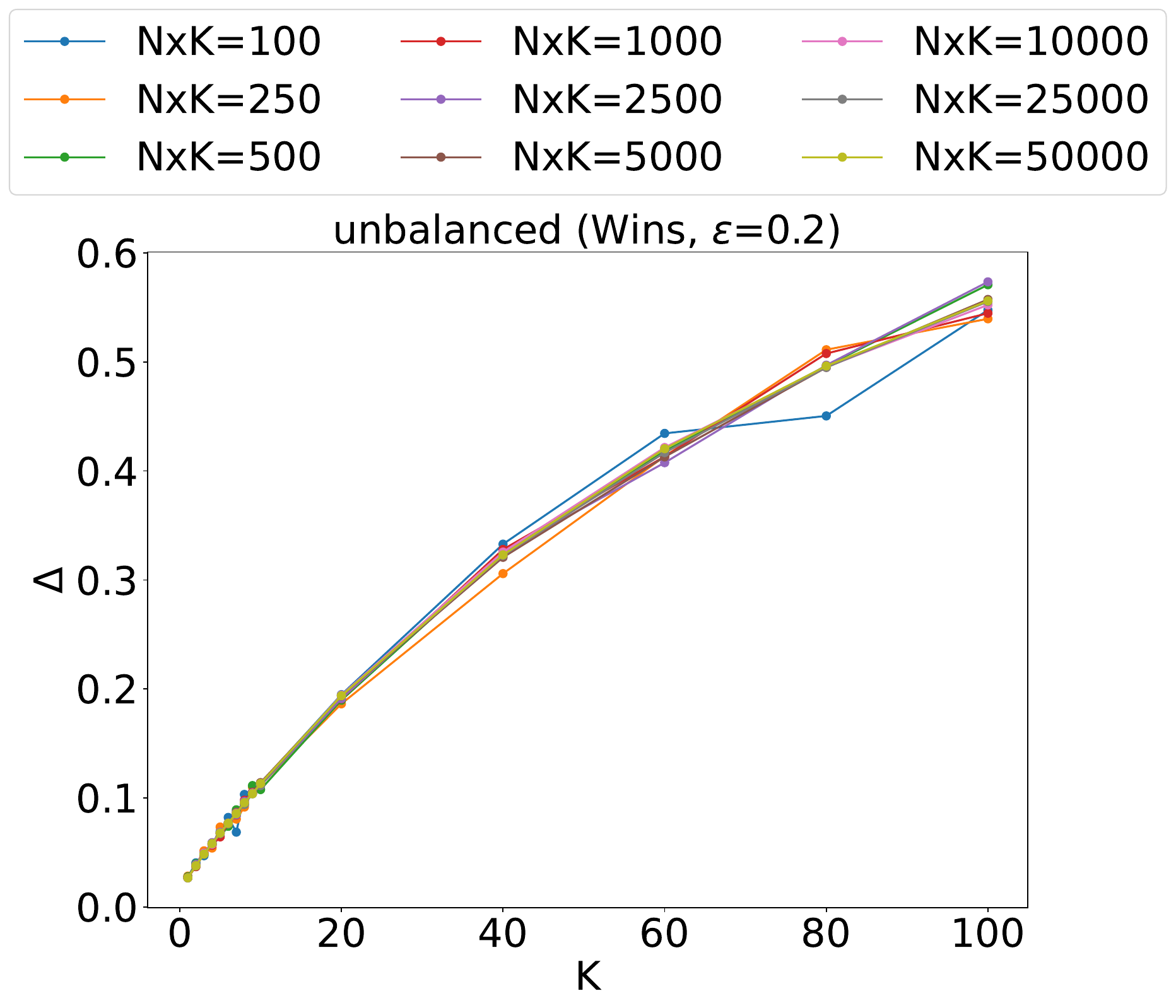}
    \caption{$\epsilon = 0.2$}
    \label{fig:gamma_delta_wins_cat4_e02}
  \end{subfigure} \hfill
  \begin{subfigure}[b]{0.24\linewidth}
    \centering
    \includegraphics[width=\linewidth]{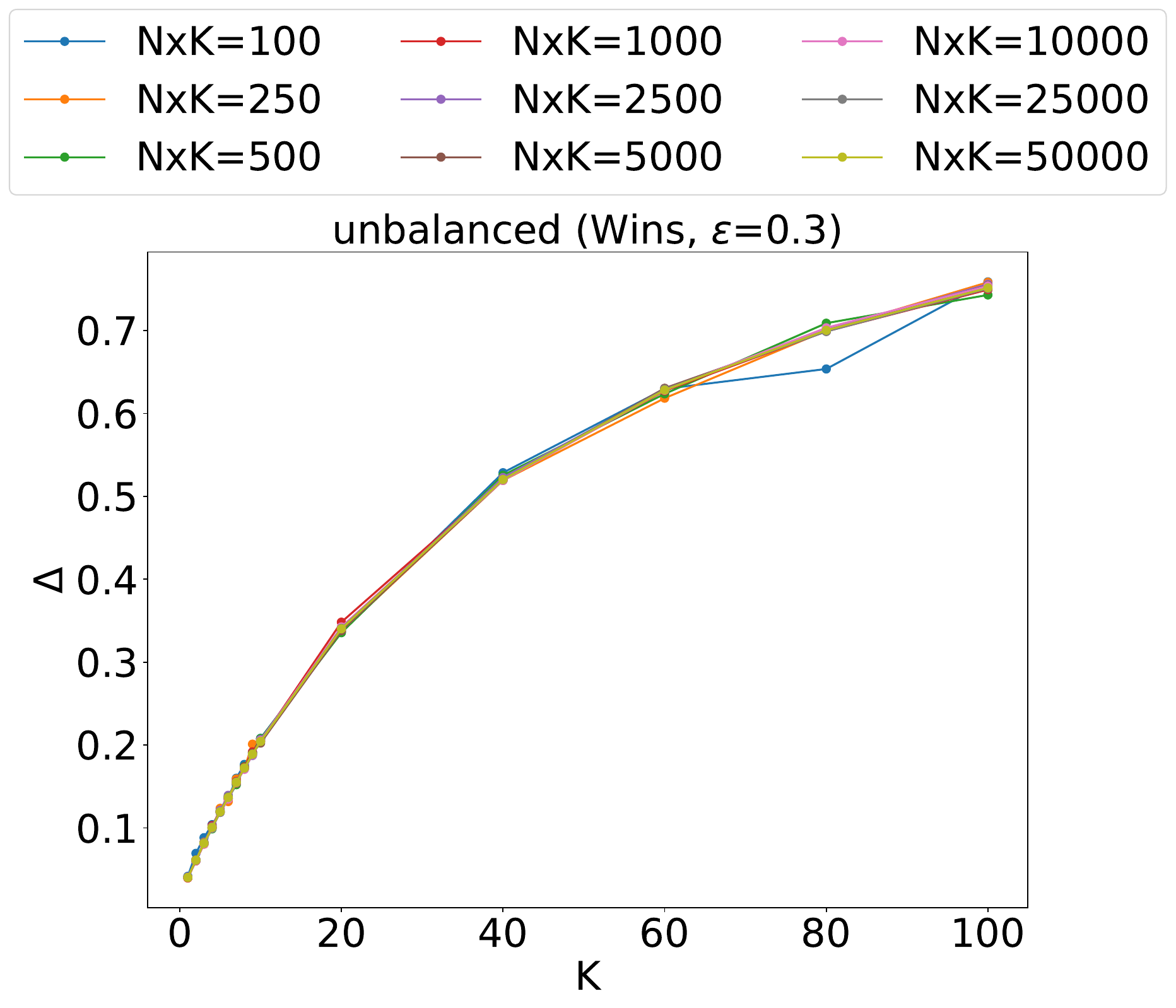}
    \caption{$\epsilon = 0.3$}
    \label{fig:gamma_delta_wins_cat4_e03}
  \end{subfigure} \hfill
  \begin{subfigure}[b]{0.24\linewidth}
    \centering
    \includegraphics[width=\linewidth]{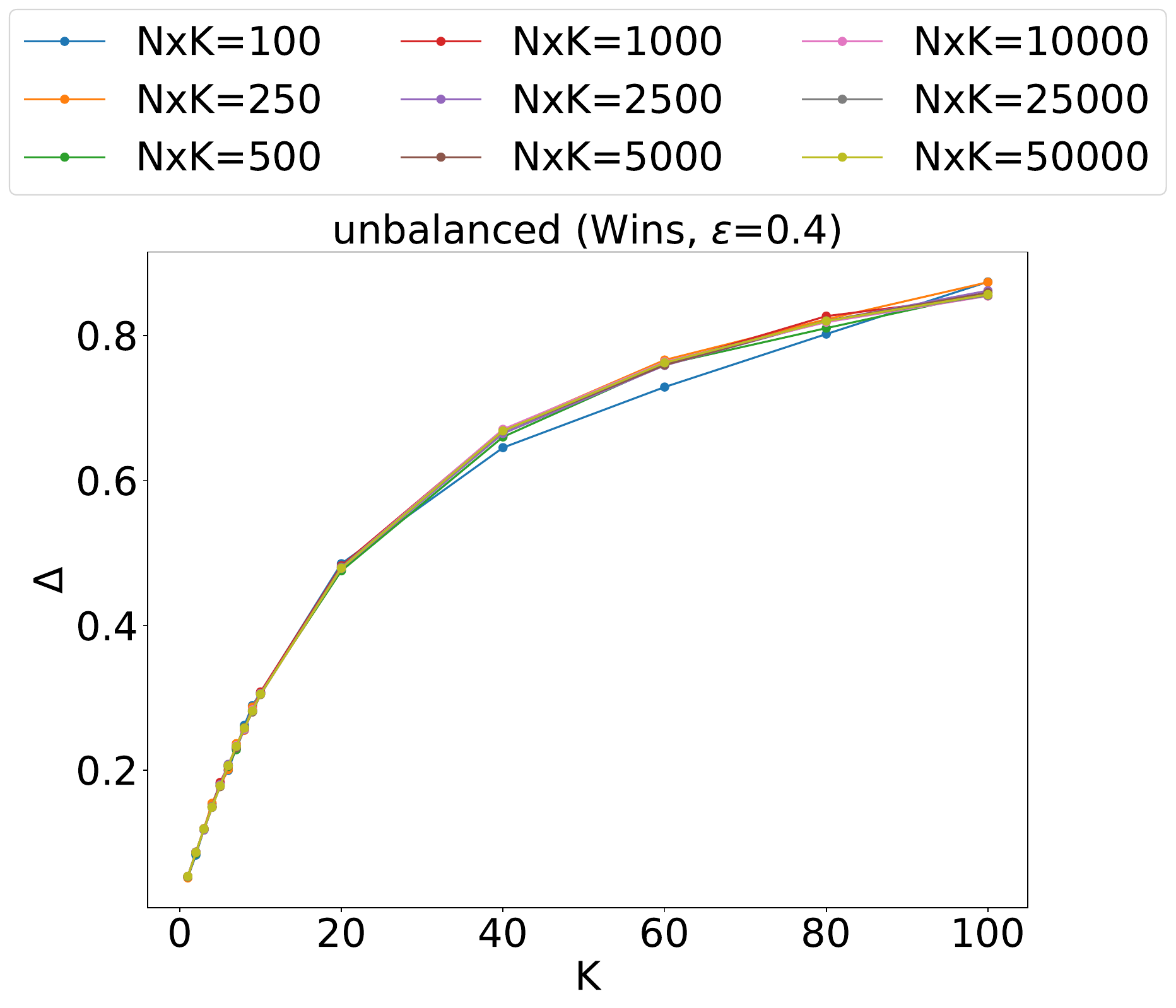}
    \caption{$\epsilon = 0.4$}
    \label{fig:gamma_delta_wins_cat4_e04}
  \end{subfigure}
  \caption{Effect sizes ($\Delta$) for unbalanced alphas with Wins as the metric ($M=4$)}
  \label{fig:gamma_delta_wins_cat4}
\end{figure*}

\begin{figure*}
  \centering
  \begin{subfigure}[b]{0.24\linewidth}
    \centering
    \includegraphics[width=\linewidth]{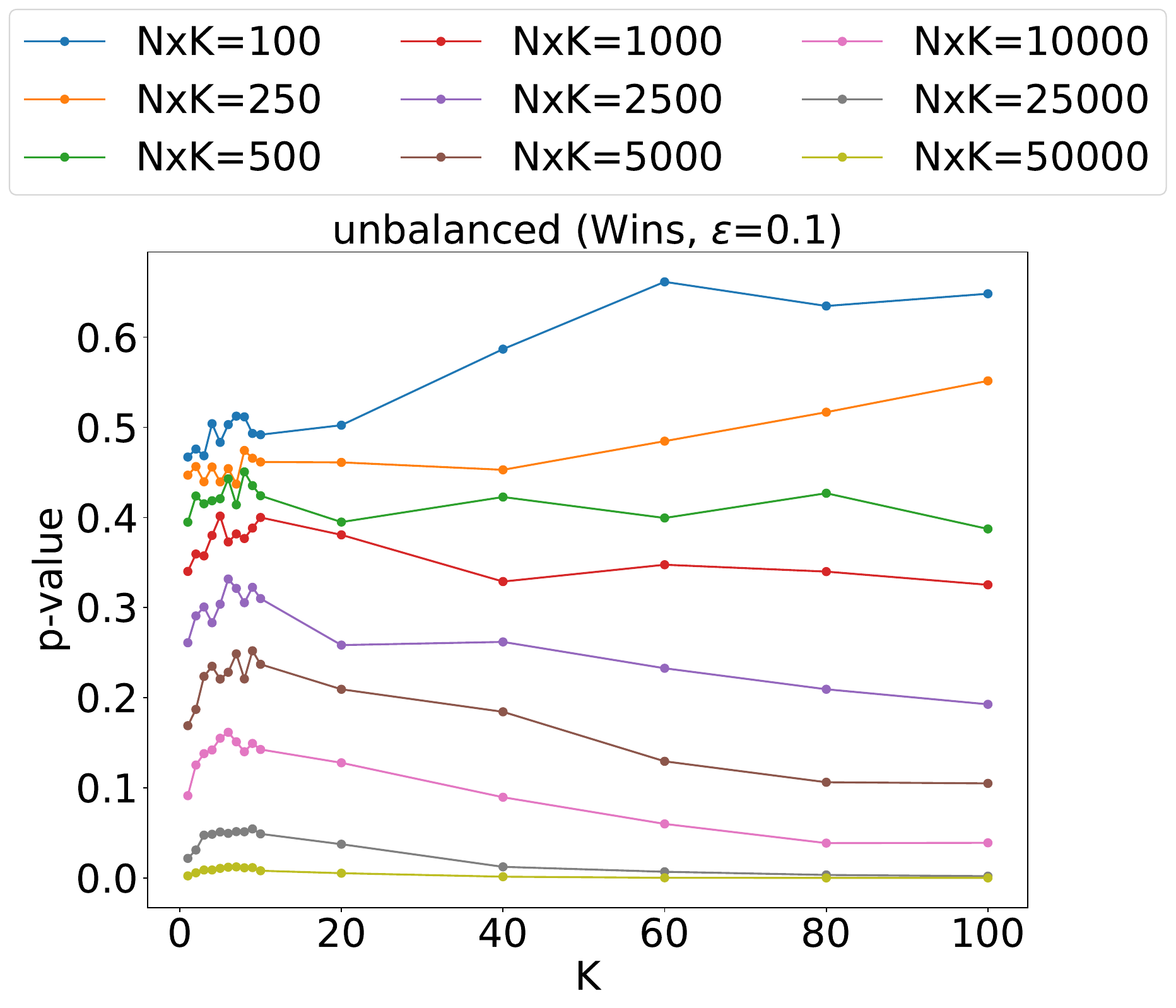}
    \caption{$\epsilon = 0.1$}
    \label{fig:gamma_wins_cat5_e01}
  \end{subfigure} \hfill
  \begin{subfigure}[b]{0.24\linewidth}
    \centering
    \includegraphics[width=\linewidth]{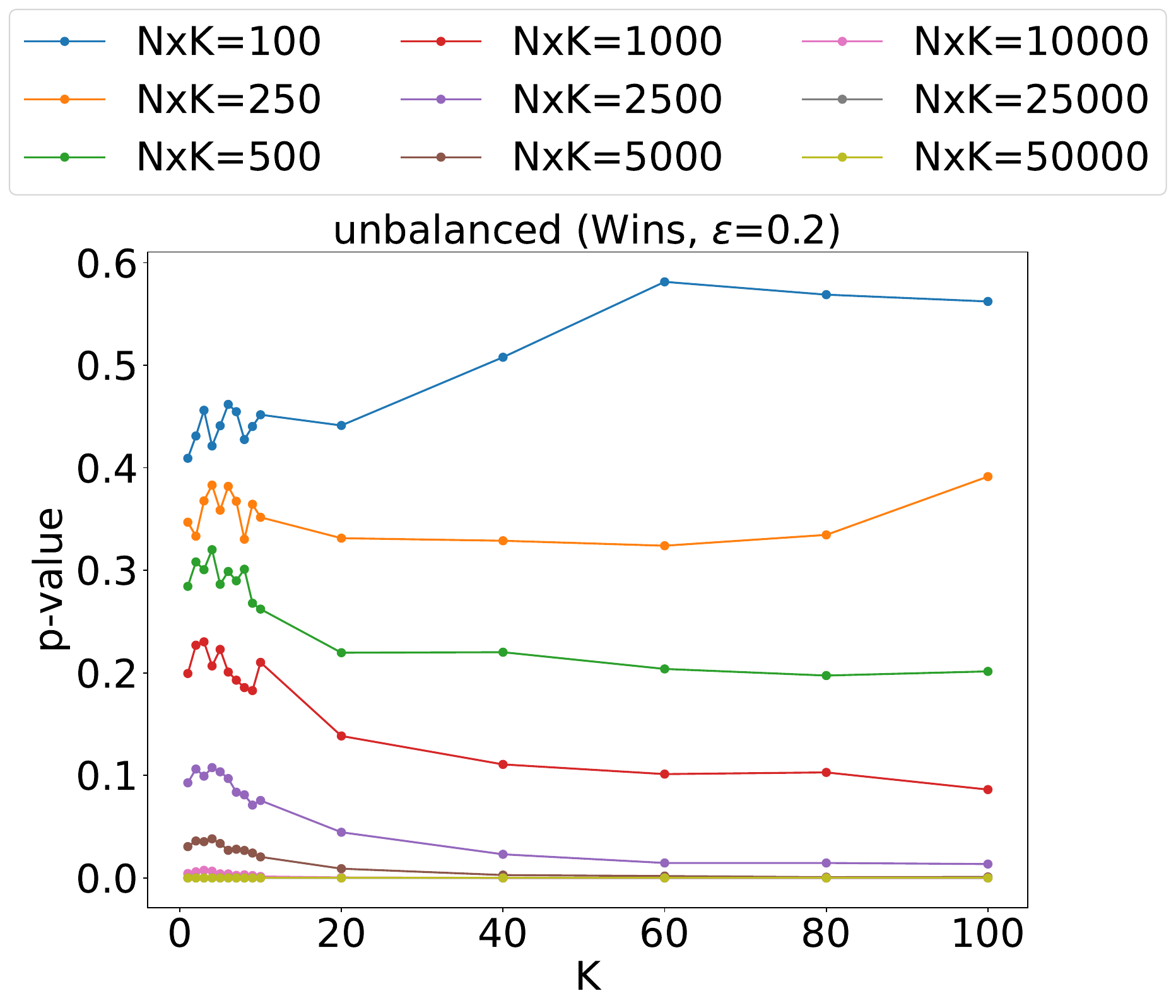}
    \caption{$\epsilon = 0.2$}
    \label{fig:gamma_wins_cat5_e02}
  \end{subfigure} \hfill
  \begin{subfigure}[b]{0.24\linewidth}
    \centering
    \includegraphics[width=\linewidth]{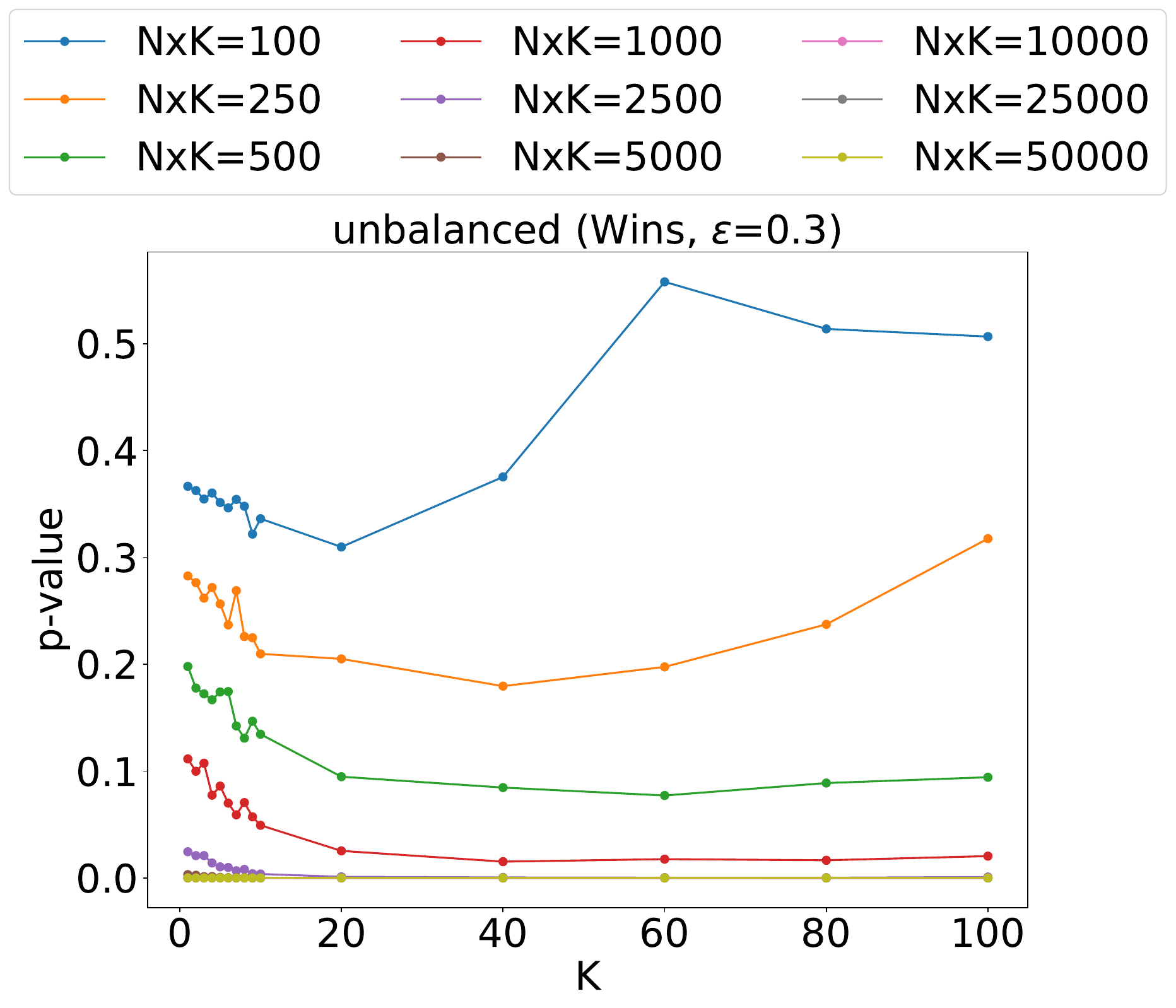}
    \caption{$\epsilon = 0.3$}
    \label{fig:gamma_wins_cat5_e03}
  \end{subfigure} \hfill
  \begin{subfigure}[b]{0.24\linewidth}
    \centering
    \includegraphics[width=\linewidth]{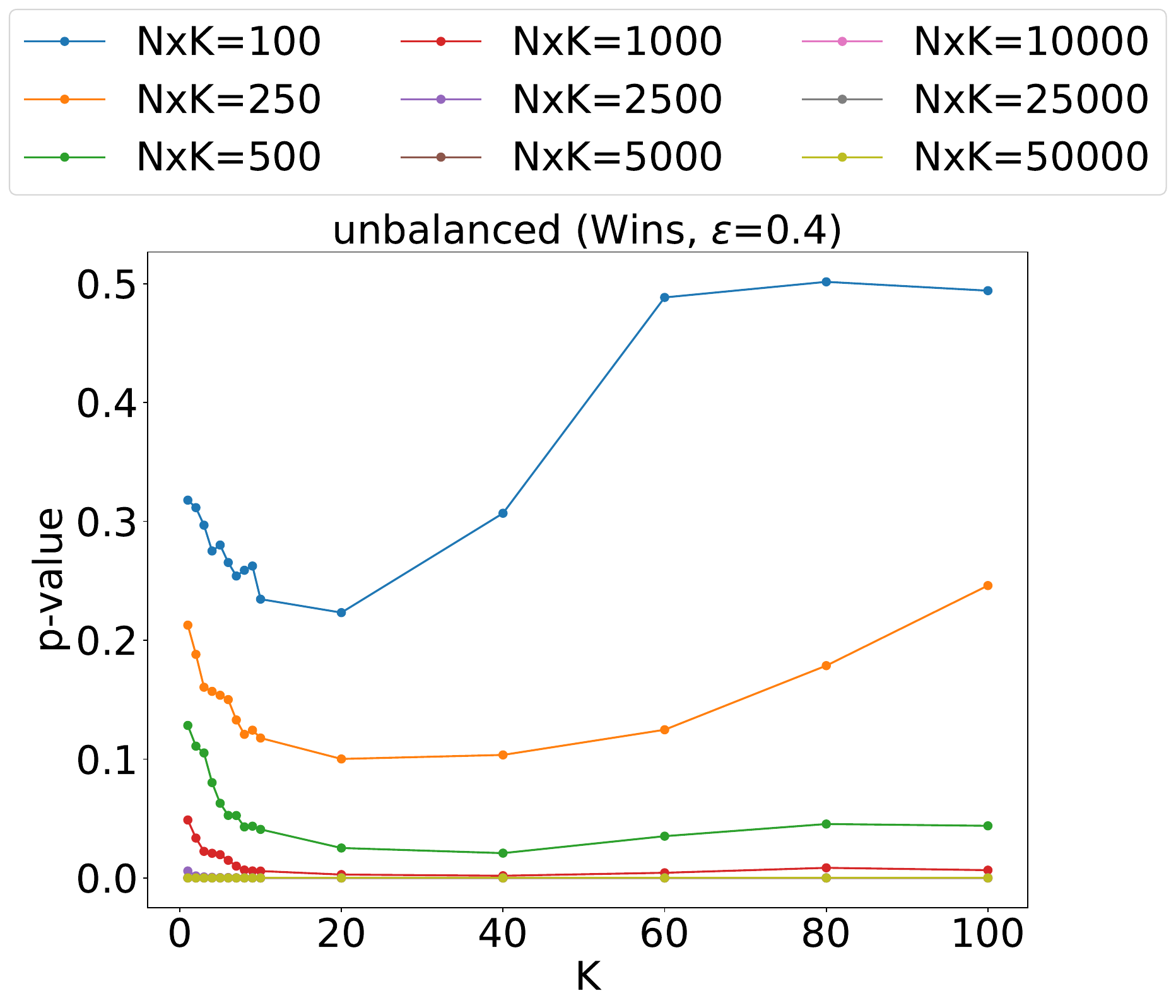}
    \caption{$\epsilon = 0.4$}
    \label{fig:gamma_wins_cat5_e04}
  \end{subfigure}
  \caption{P-value plots for unbalanced alphas with Wins as the metric ($M=5$)}
  \label{fig:gamma_wins_cat5}
\end{figure*}

\begin{figure*}
  \centering
  \begin{subfigure}[b]{0.24\linewidth}
    \centering
    \includegraphics[width=\linewidth]{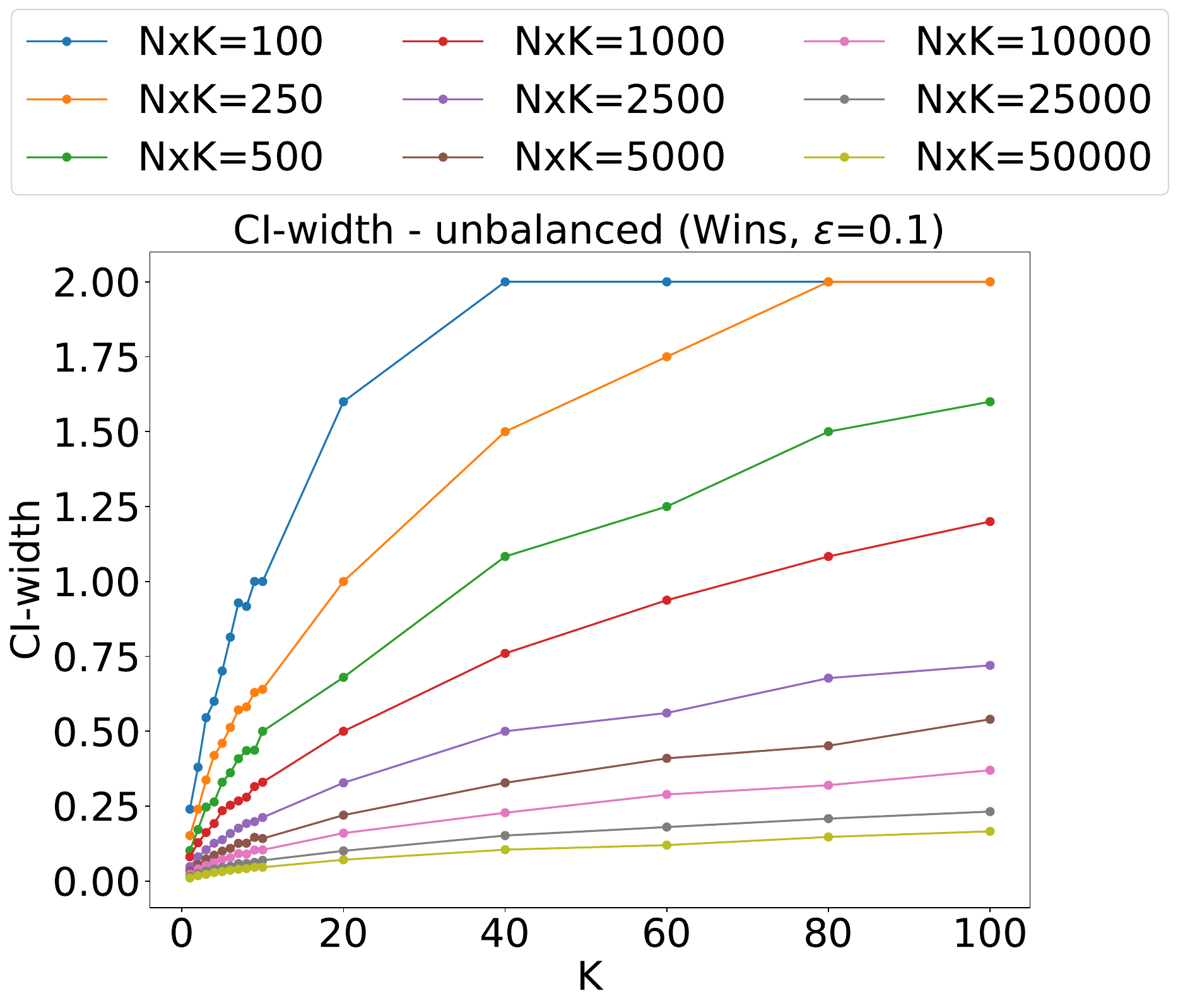}
    \caption{$\epsilon = 0.1$}
    \label{fig:gamma_ci_wins_cat5_e01}
  \end{subfigure} \hfill
  \begin{subfigure}[b]{0.24\linewidth}
    \centering
    \includegraphics[width=\linewidth]{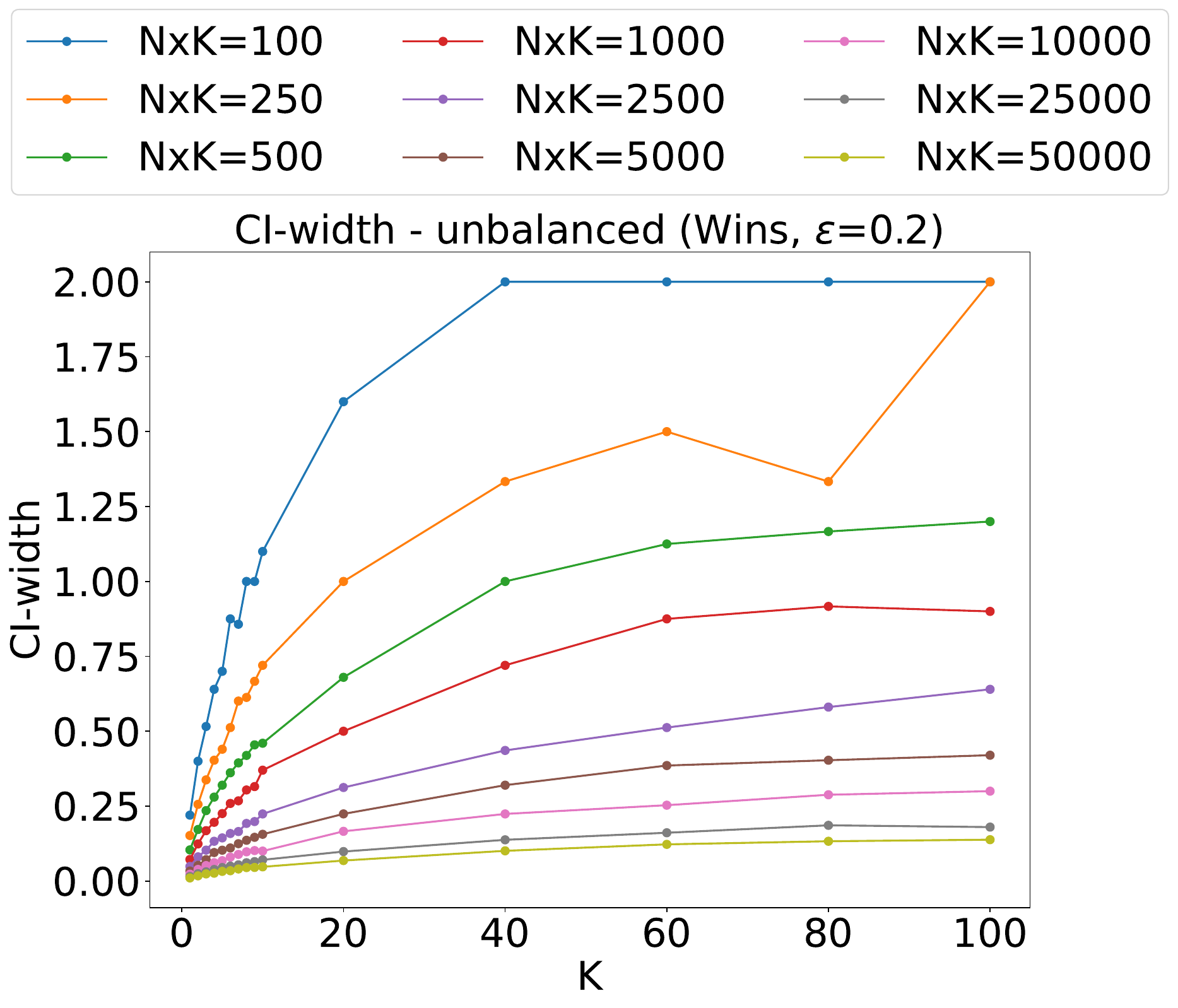}
    \caption{$\epsilon = 0.2$}
    \label{fig:gamma_ci_wins_cat5_e02}
  \end{subfigure} \hfill
  \begin{subfigure}[b]{0.24\linewidth}
    \centering
    \includegraphics[width=\linewidth]{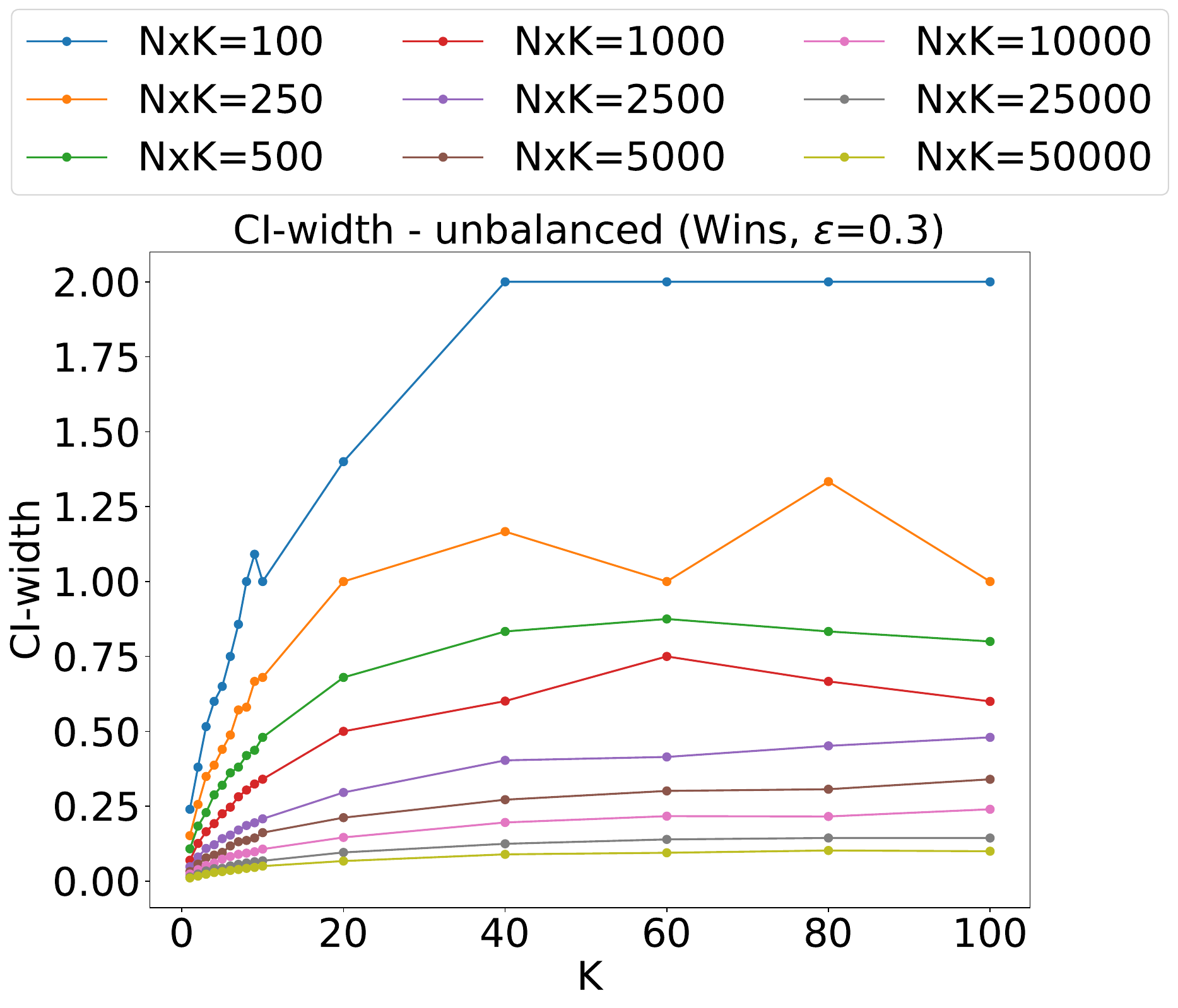}
    \caption{$\epsilon = 0.3$}
    \label{fig:gamma_ci_wins_cat5_e03}
  \end{subfigure} \hfill
  \begin{subfigure}[b]{0.24\linewidth}
    \centering
    \includegraphics[width=\linewidth]{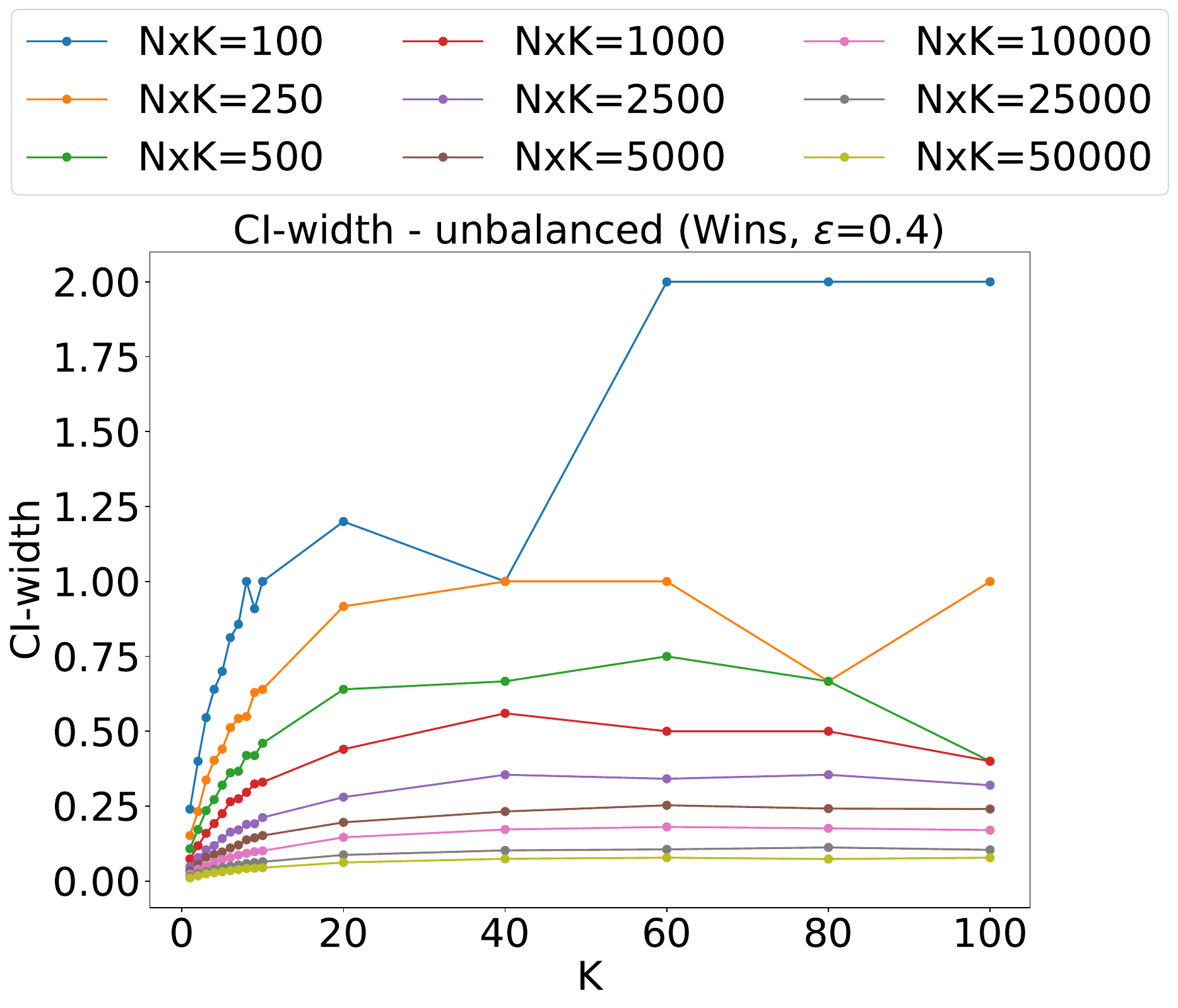}
    \caption{$\epsilon = 0.4$}
    \label{fig:gamma_ci_wins_cat5_e04}
  \end{subfigure}
  \caption{CI-width plots for unbalanced alphas with Wins as the metric ($M=5$)}
  \label{fig:gamma_ci_wins_cat5}
\end{figure*}

\begin{figure*}
  \centering
  \begin{subfigure}[b]{0.24\linewidth}
    \centering
    \includegraphics[width=\linewidth]{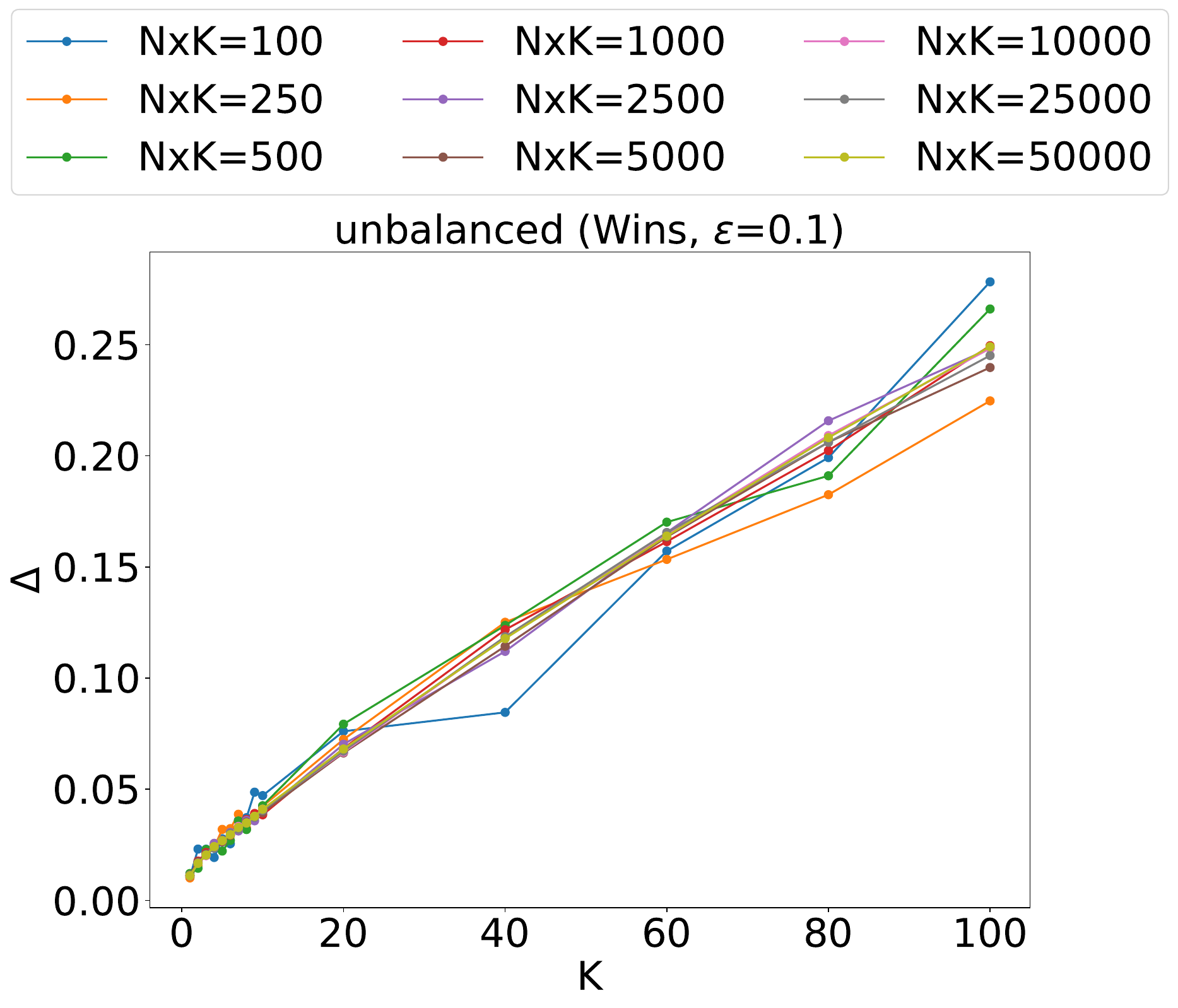}
    \caption{$\epsilon = 0.1$}
    \label{fig:gamma_delta_wins_cat5_e01}
  \end{subfigure} \hfill
  \begin{subfigure}[b]{0.24\linewidth}
    \centering
    \includegraphics[width=\linewidth]{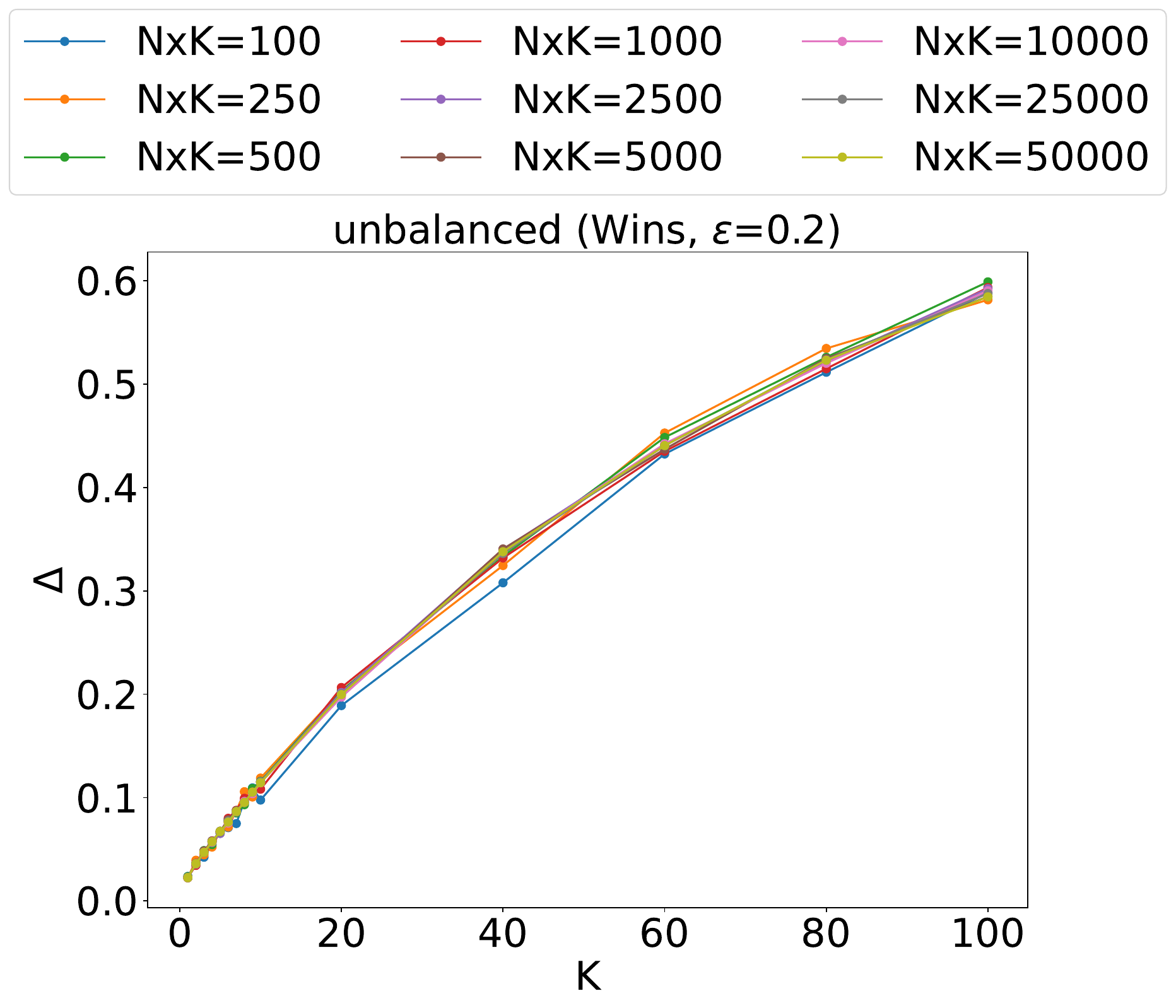}
    \caption{$\epsilon = 0.2$}
    \label{fig:gamma_delta_wins_cat5_e02}
  \end{subfigure} \hfill
  \begin{subfigure}[b]{0.24\linewidth}
    \centering
    \includegraphics[width=\linewidth]{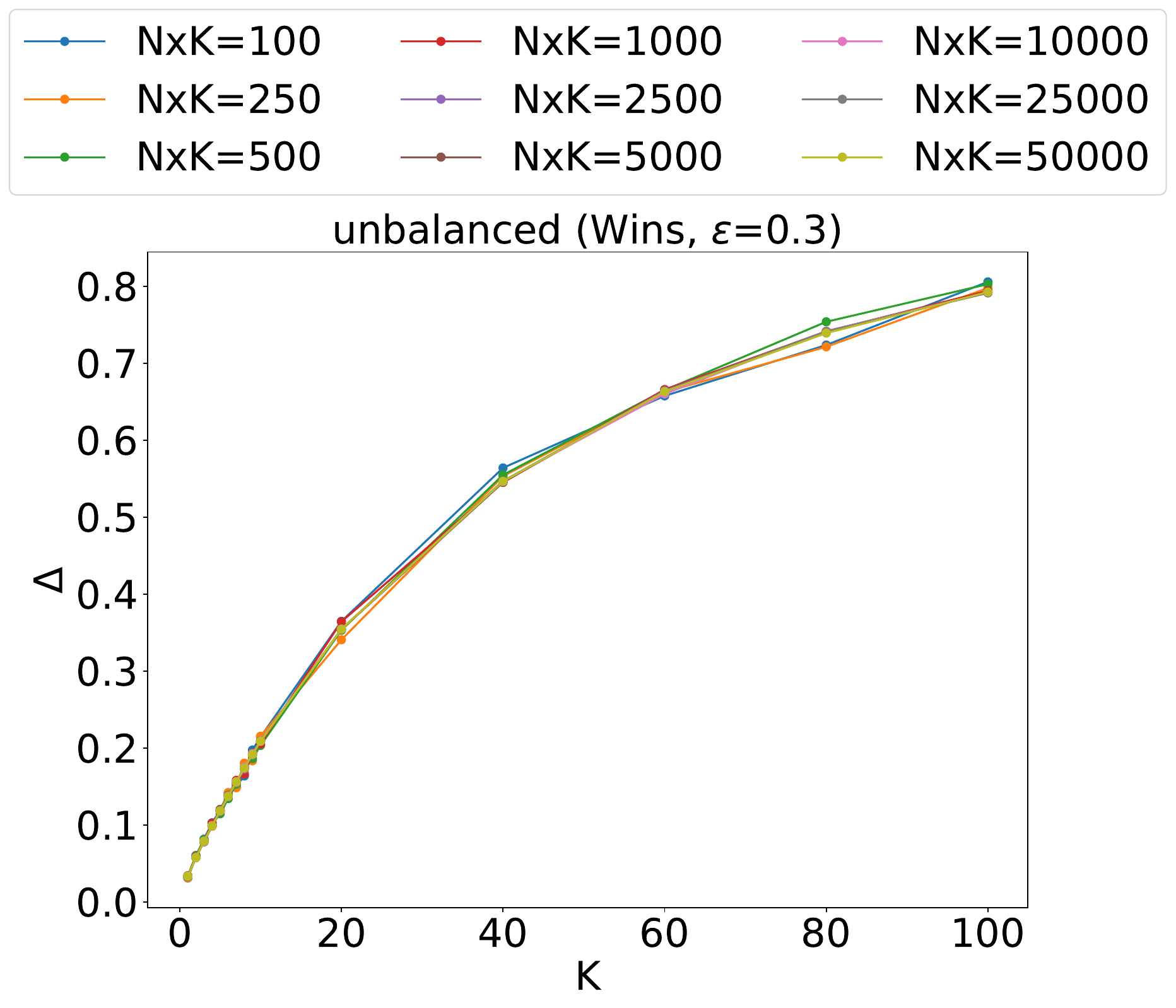}
    \caption{$\epsilon = 0.3$}
    \label{fig:gamma_delta_wins_cat5_e03}
  \end{subfigure} \hfill
  \begin{subfigure}[b]{0.24\linewidth}
    \centering
    \includegraphics[width=\linewidth]{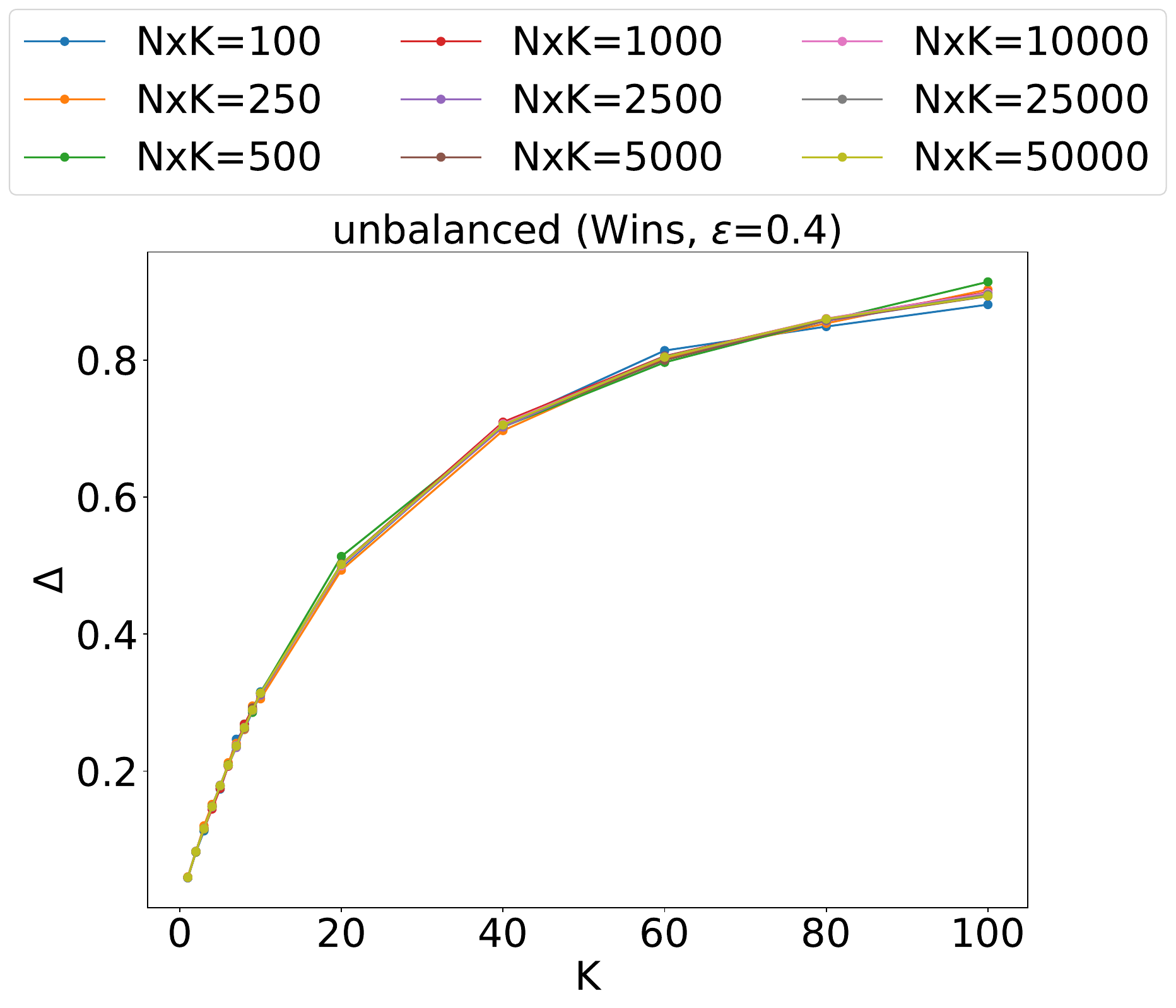}
    \caption{$\epsilon = 0.4$}
    \label{fig:gamma_delta_wins_cat5_e04}
  \end{subfigure}
  \caption{Effect sizes ($\Delta$) for unbalanced alphas with Wins as the metric ($M=5$)}
  \label{fig:gamma_delta_wins_cat5}
\end{figure*}

\begin{figure*}
  \centering
  \begin{subfigure}[b]{0.24\linewidth}
    \centering
    \includegraphics[width=\linewidth]{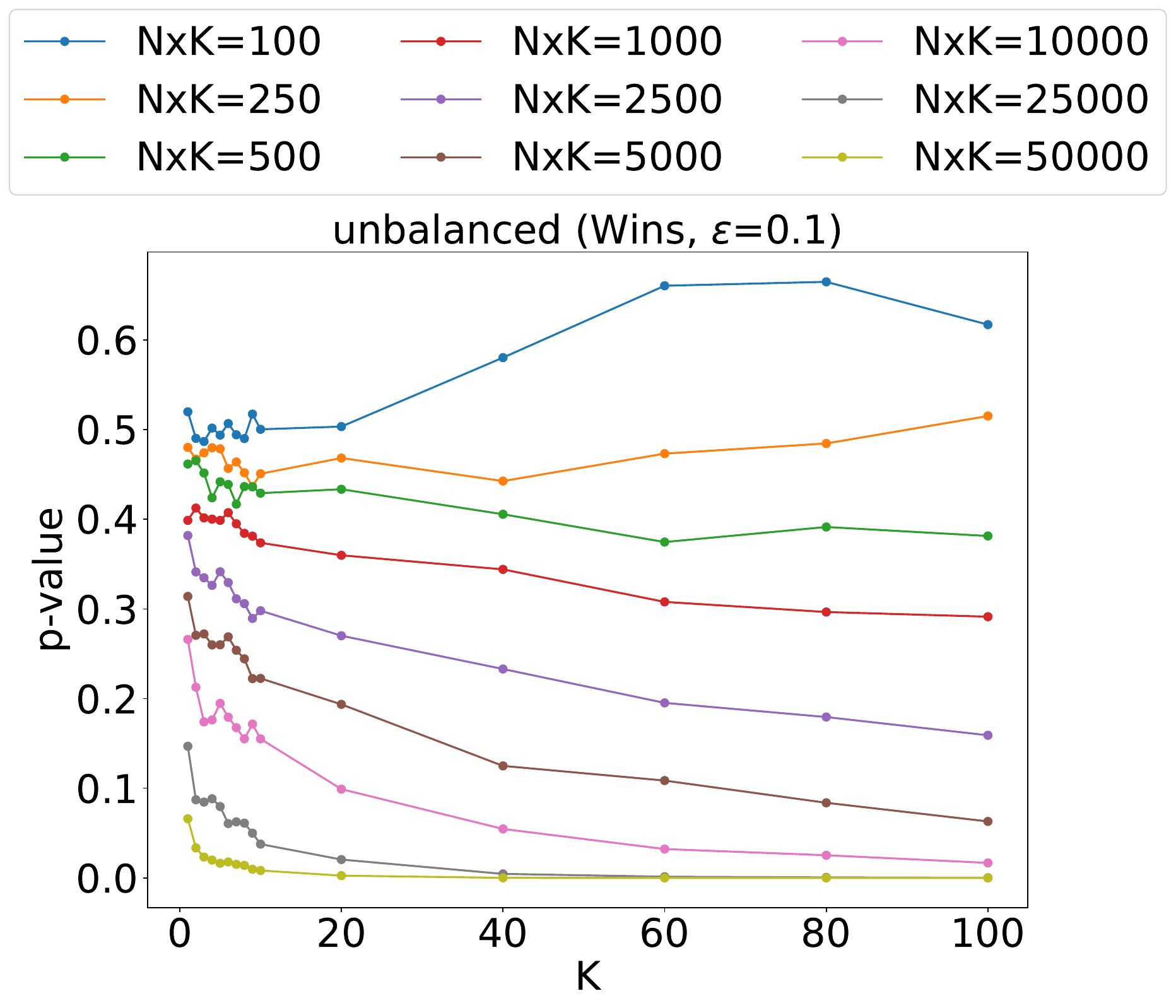}
    \caption{$\epsilon = 0.1$}
    \label{fig:gamma_wins_cat12_e01}
  \end{subfigure} \hfill
  \begin{subfigure}[b]{0.24\linewidth}
    \centering
    \includegraphics[width=\linewidth]{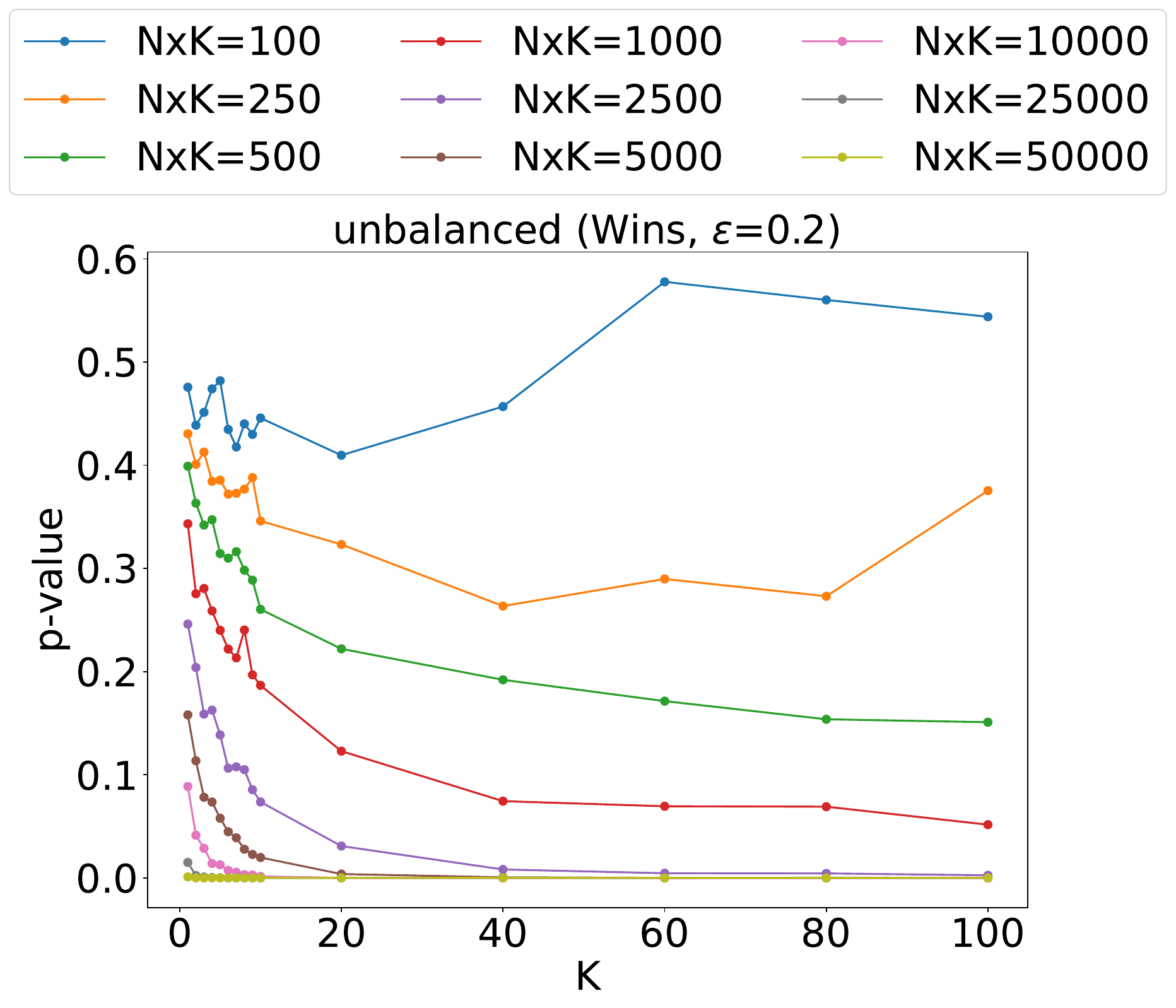}
    \caption{$\epsilon = 0.2$}
    \label{fig:gamma_wins_cat12_e02}
  \end{subfigure} \hfill
  \begin{subfigure}[b]{0.24\linewidth}
    \centering
    \includegraphics[width=\linewidth]{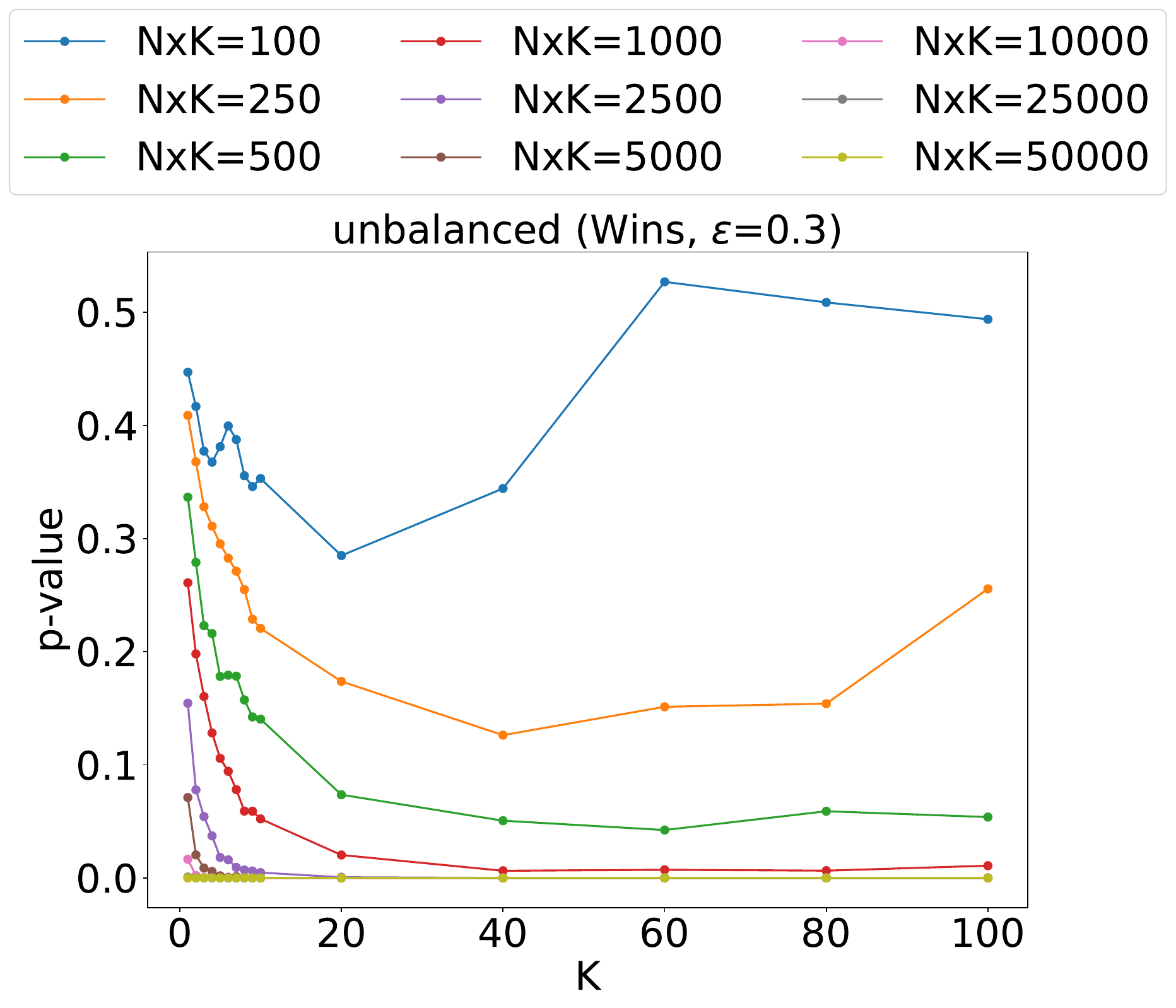}
    \caption{$\epsilon = 0.3$}
    \label{fig:gamma_wins_cat12_e03}
  \end{subfigure} \hfill
  \begin{subfigure}[b]{0.24\linewidth}
    \centering
    \includegraphics[width=\linewidth]{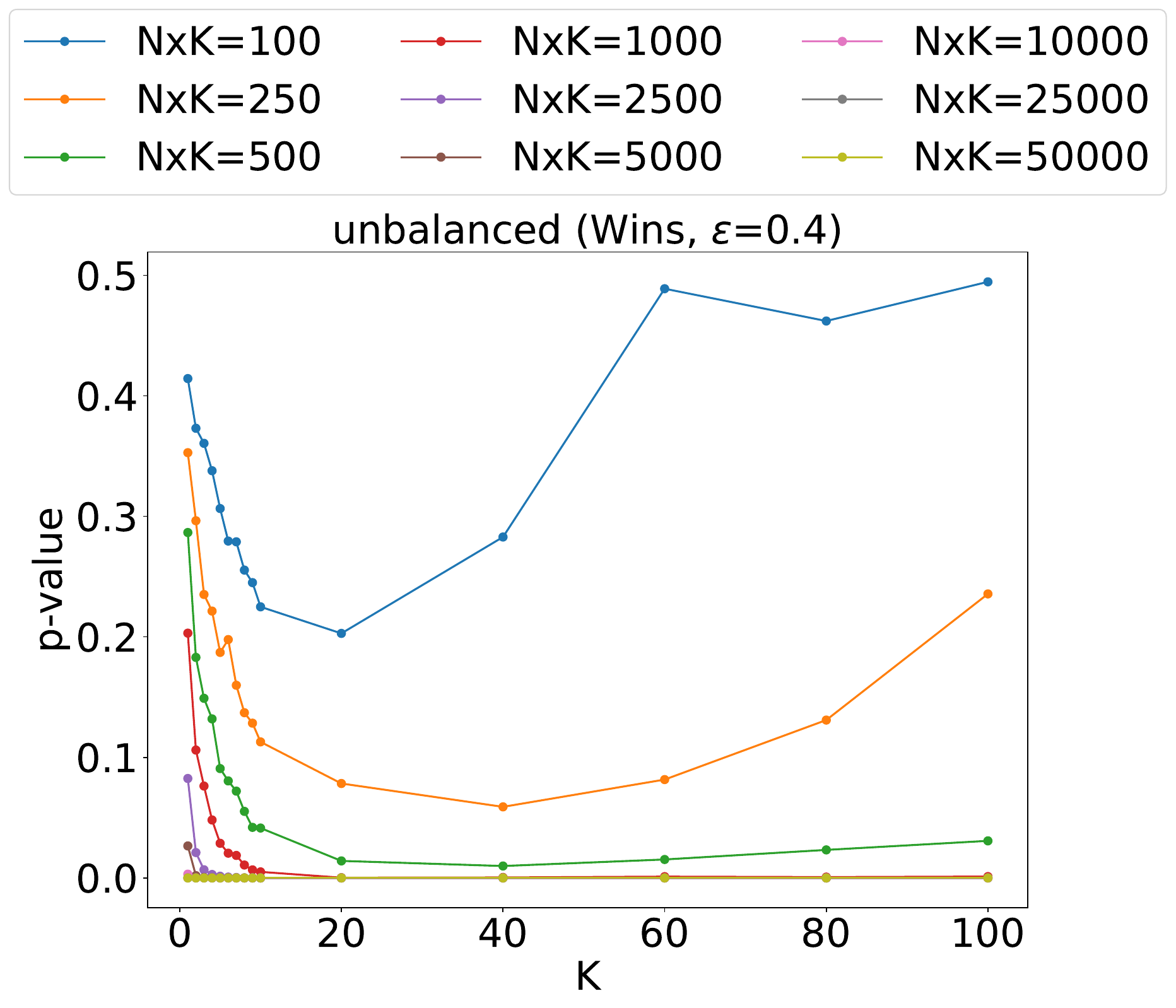}
    \caption{$\epsilon = 0.4$}
    \label{fig:gamma_wins_cat12_e04}
  \end{subfigure}
  \caption{P-value plots for unbalanced alphas with Wins as the metric ($M=12$)}
  \label{fig:gamma_wins_cat12}
\end{figure*}

\begin{figure*}
  \centering
  \begin{subfigure}[b]{0.24\linewidth}
    \centering
    \includegraphics[width=\linewidth]{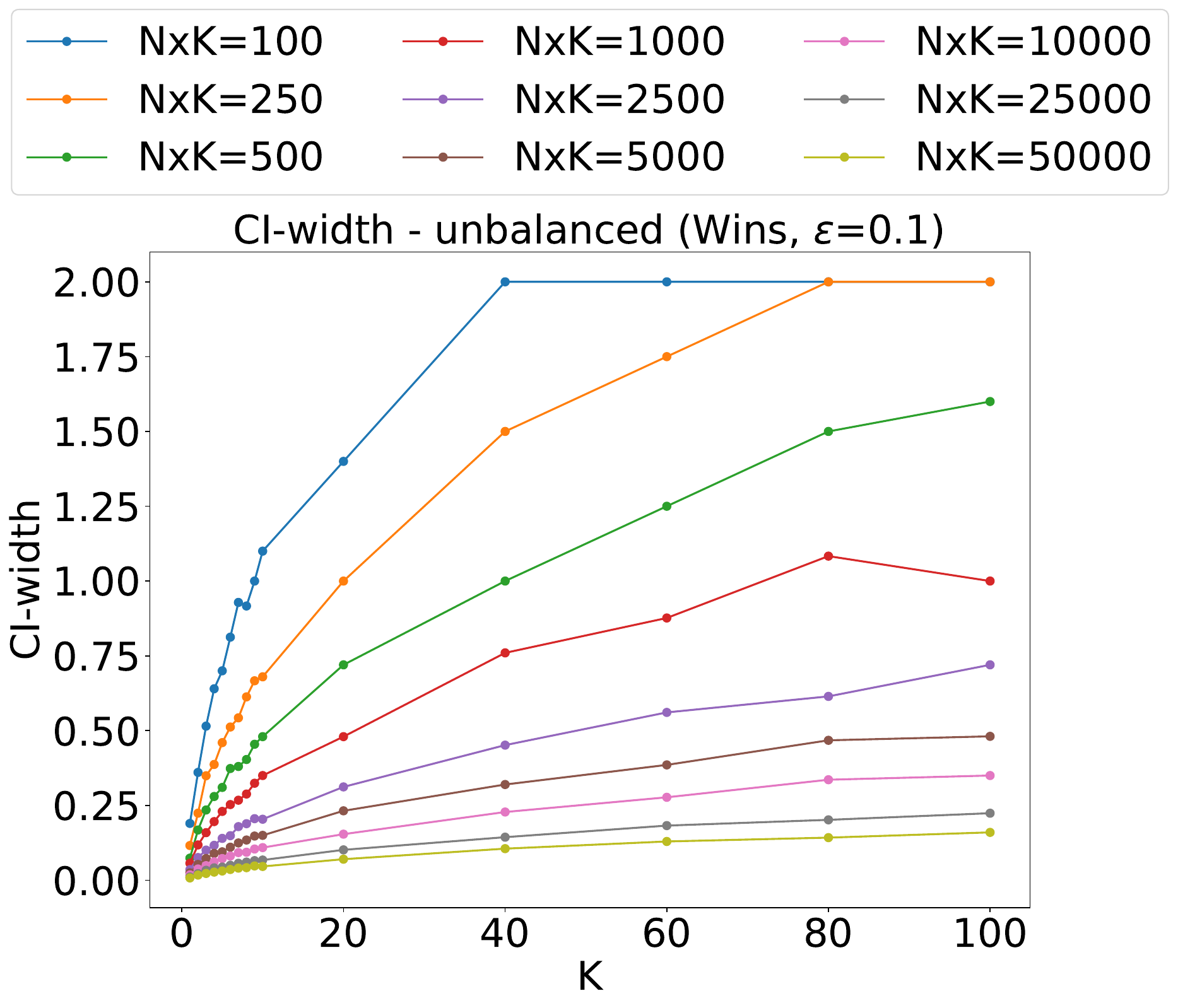}
    \caption{$\epsilon = 0.1$}
    \label{fig:gamma_ci_wins_cat12_e01}
  \end{subfigure} \hfill
  \begin{subfigure}[b]{0.24\linewidth}
    \centering
    \includegraphics[width=\linewidth]{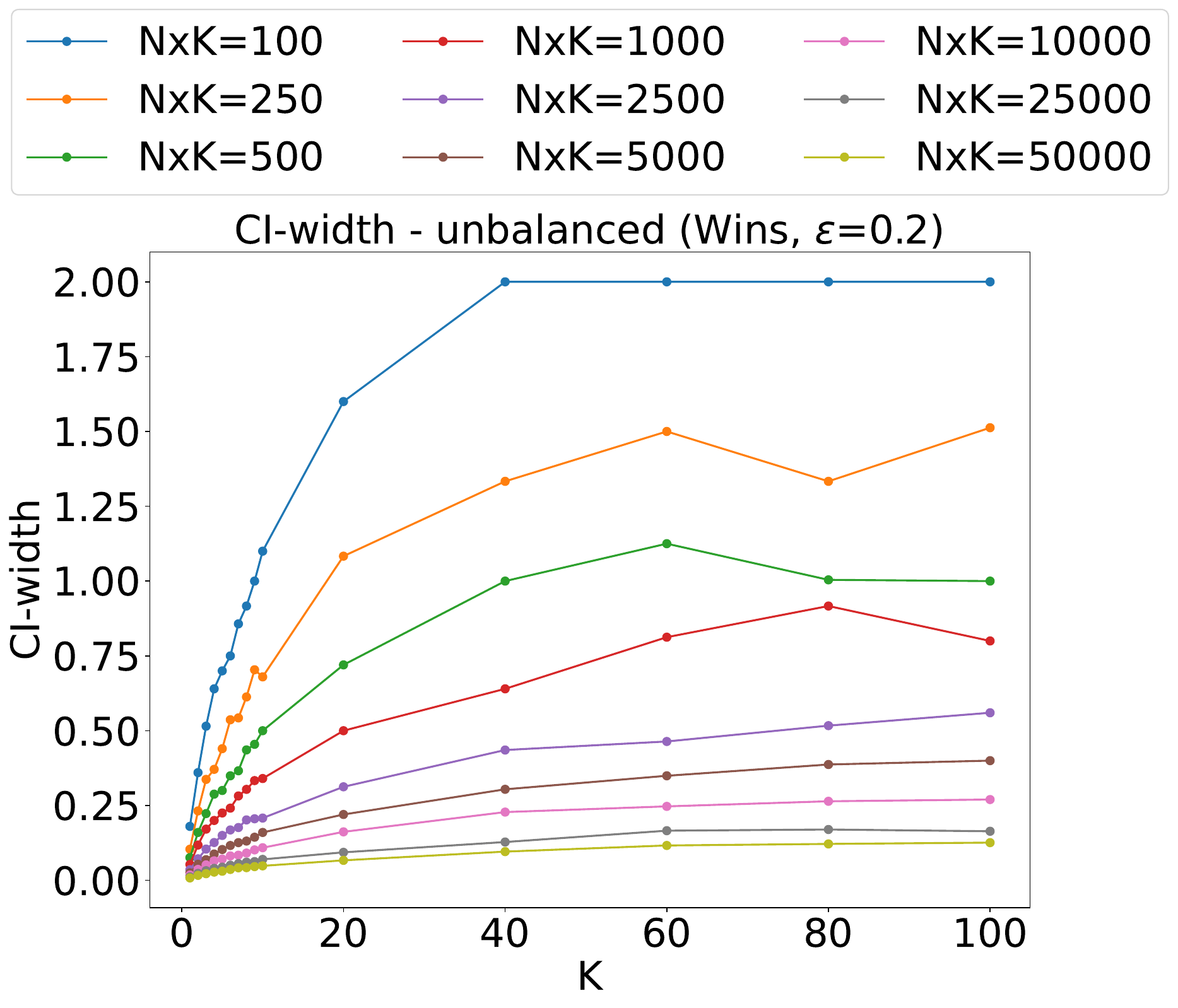}
    \caption{$\epsilon = 0.2$}
    \label{fig:gamma_ci_wins_cat12_e02}
  \end{subfigure} \hfill
  \begin{subfigure}[b]{0.24\linewidth}
    \centering
    \includegraphics[width=\linewidth]{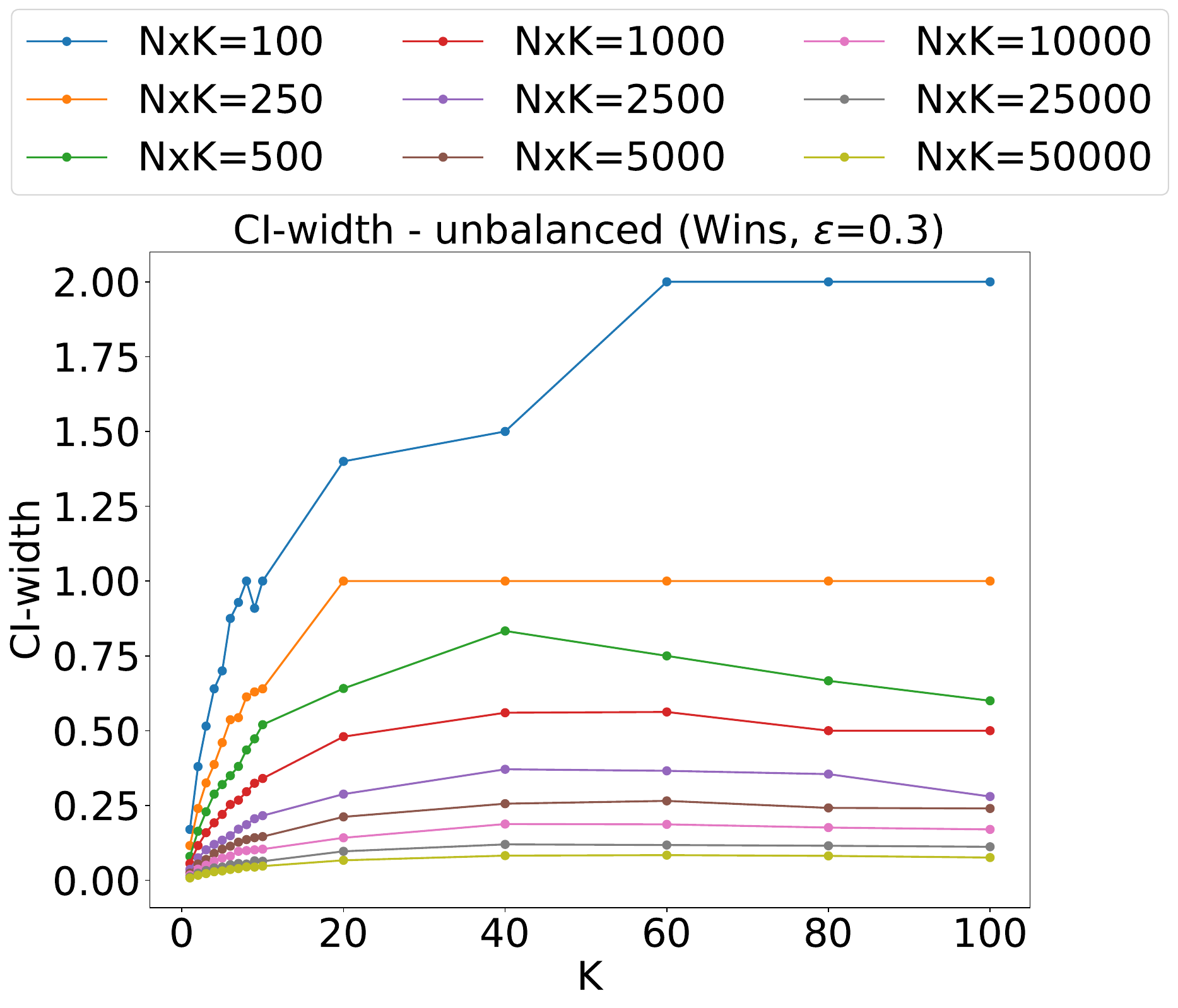}
    \caption{$\epsilon = 0.3$}
    \label{fig:gamma_ci_wins_cat12_e03}
  \end{subfigure} \hfill
  \begin{subfigure}[b]{0.24\linewidth}
    \centering
    \includegraphics[width=\linewidth]{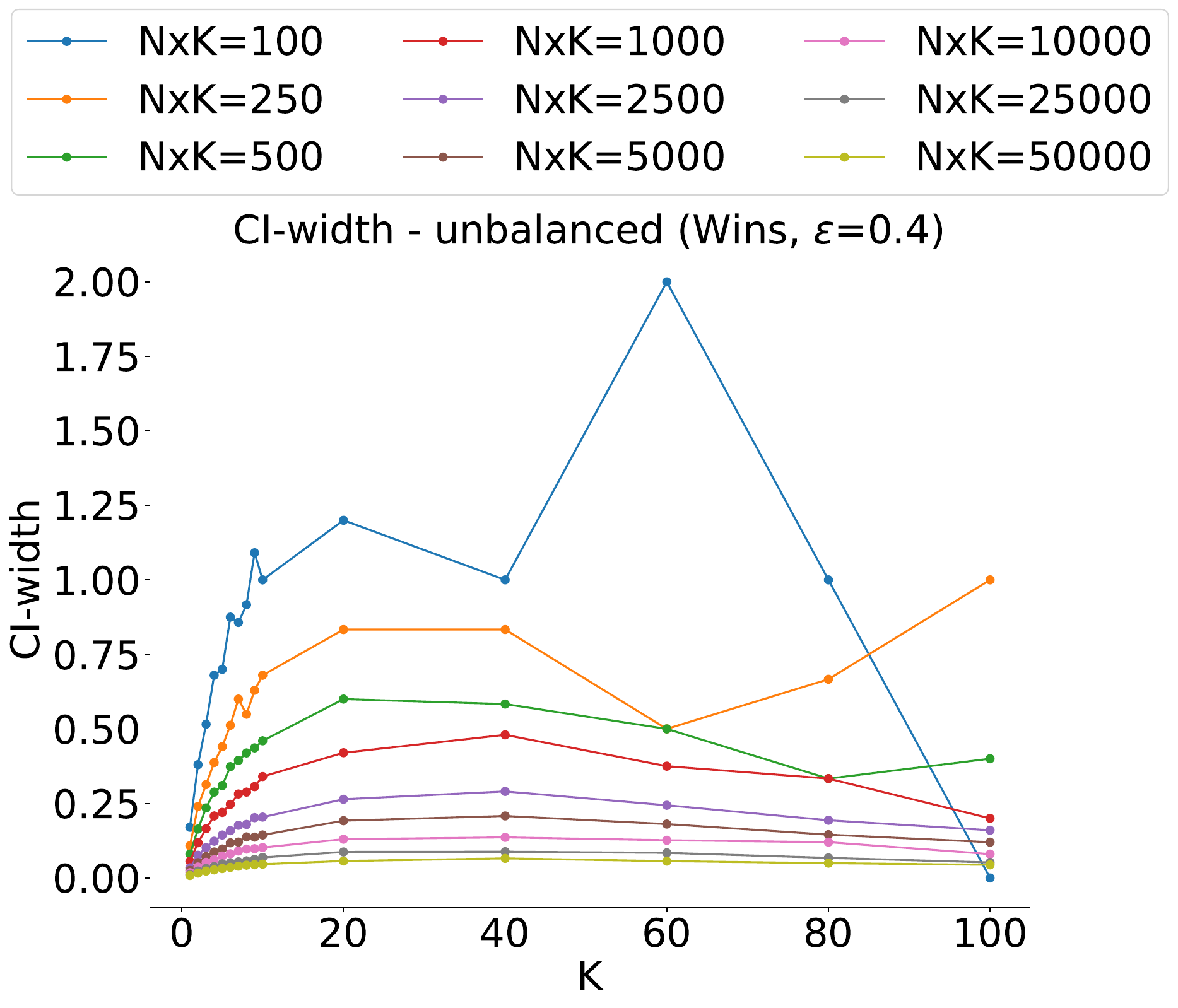}
    \caption{$\epsilon = 0.4$}
    \label{fig:gamma_ci_wins_cat12_e04}
  \end{subfigure}
  \caption{CI-width plots for unbalanced alphas with Wins as the metric ($M=12$)}
  \label{fig:gamma_ci_wins_cat12}
\end{figure*}

\begin{figure*}
  \centering
  \begin{subfigure}[b]{0.24\linewidth}
    \centering
    \includegraphics[width=\linewidth]{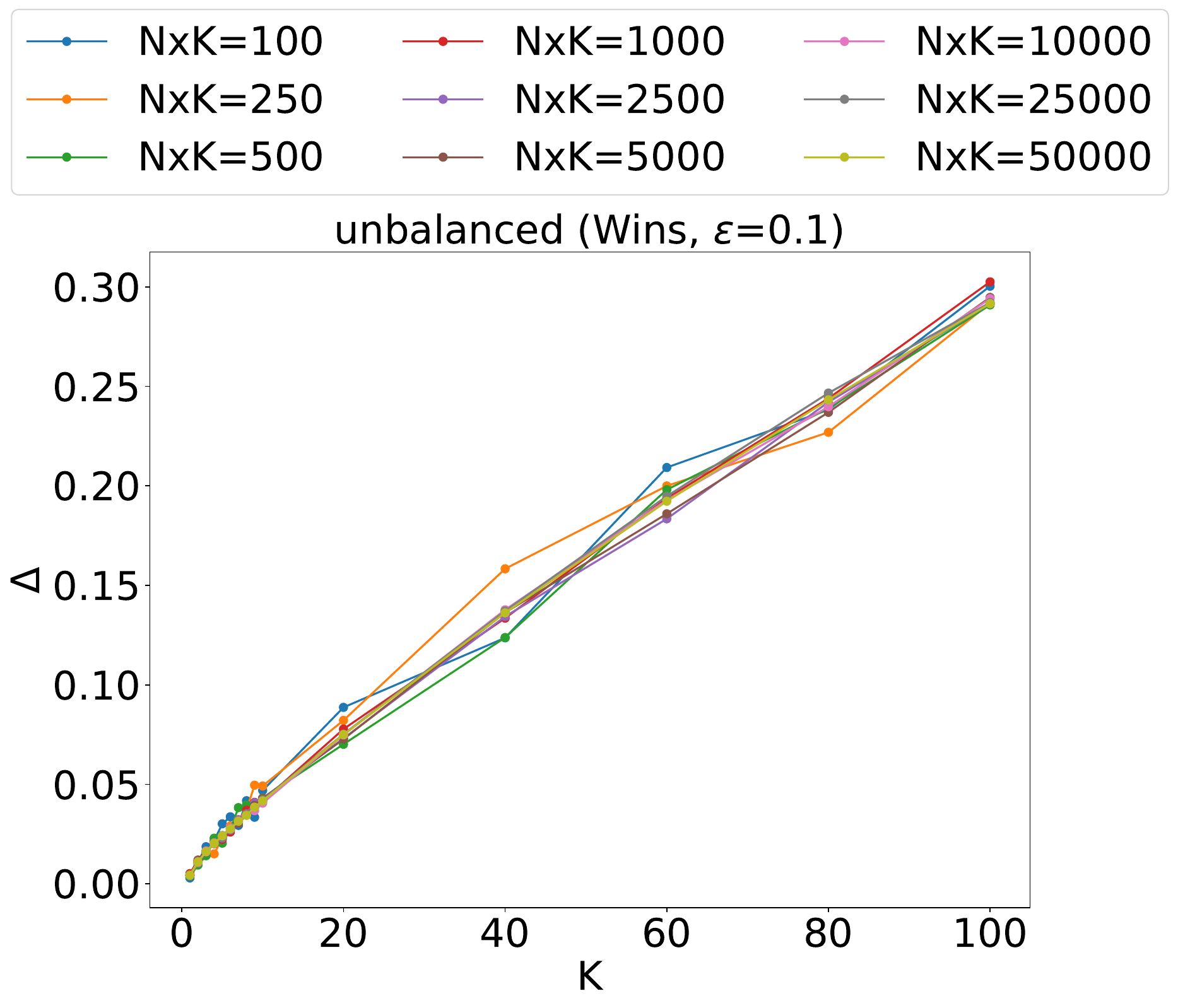}
    \caption{$\epsilon = 0.1$}
    \label{fig:gamma_delta_wins_cat12_e01}
  \end{subfigure} \hfill
  \begin{subfigure}[b]{0.24\linewidth}
    \centering
    \includegraphics[width=\linewidth]{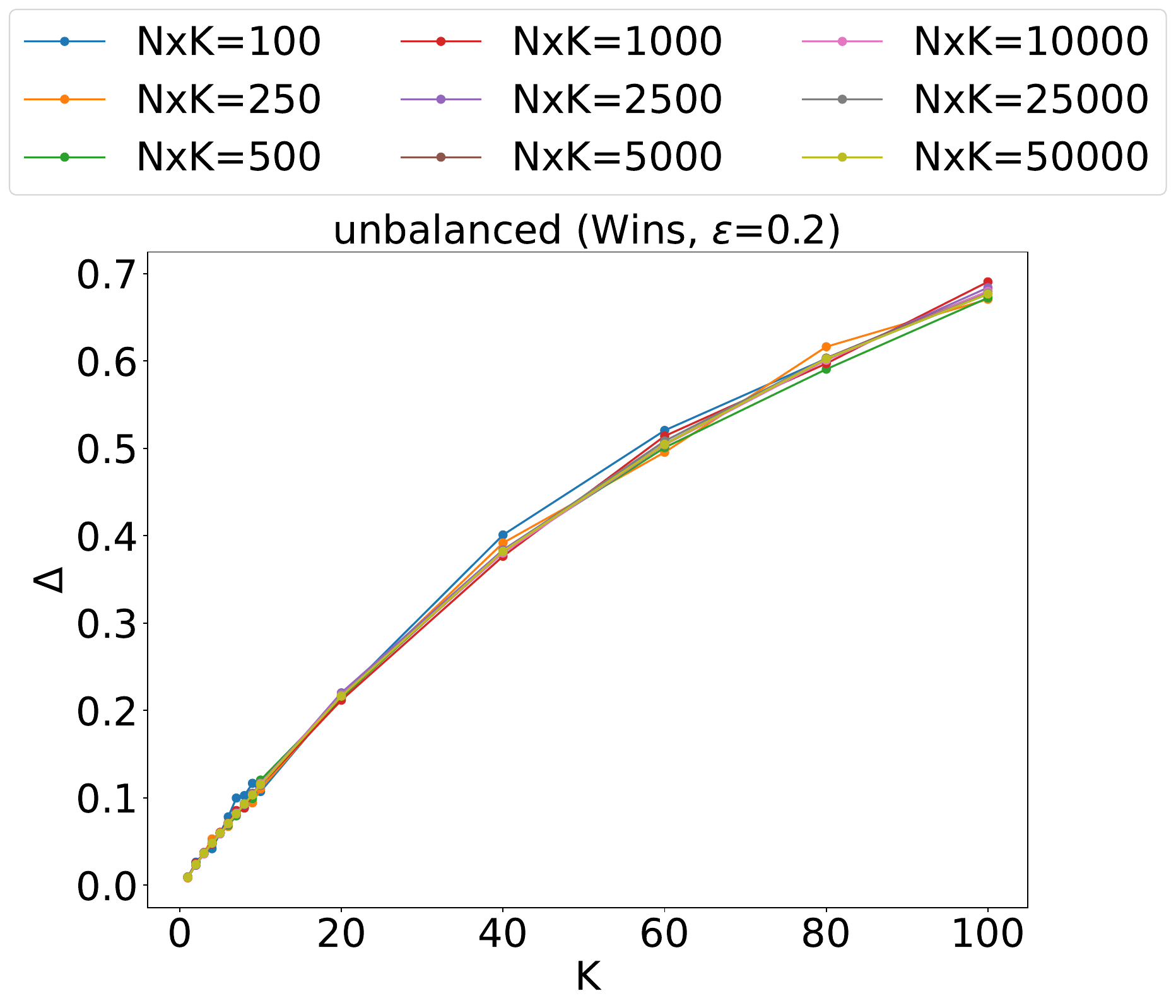}
    \caption{$\epsilon = 0.2$}
    \label{fig:gamma_delta_wins_cat12_e02}
  \end{subfigure} \hfill
  \begin{subfigure}[b]{0.24\linewidth}
    \centering
    \includegraphics[width=\linewidth]{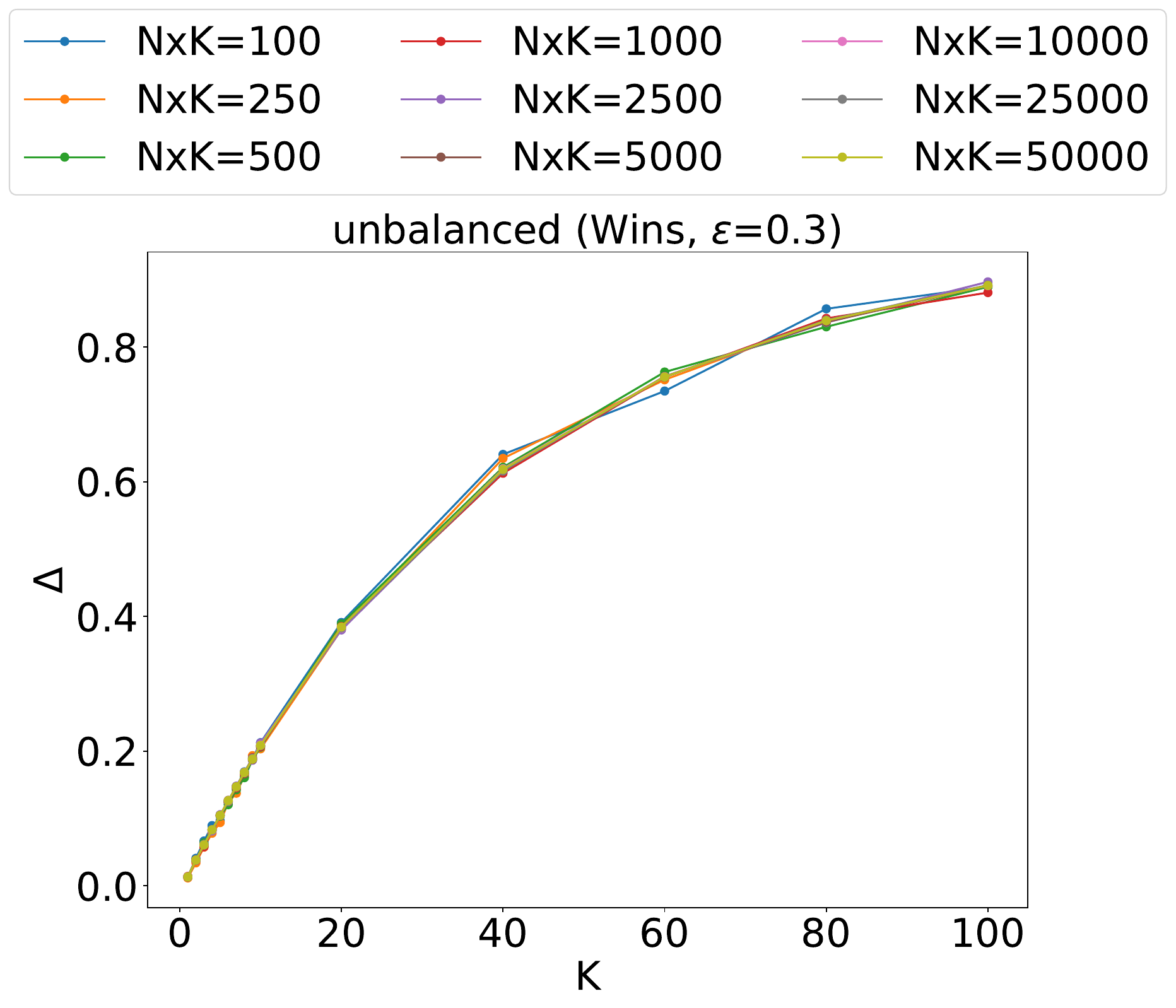}
    \caption{$\epsilon = 0.3$}
    \label{fig:gamma_delta_wins_cat12_e03}
  \end{subfigure} \hfill
  \begin{subfigure}[b]{0.24\linewidth}
    \centering
    \includegraphics[width=\linewidth]{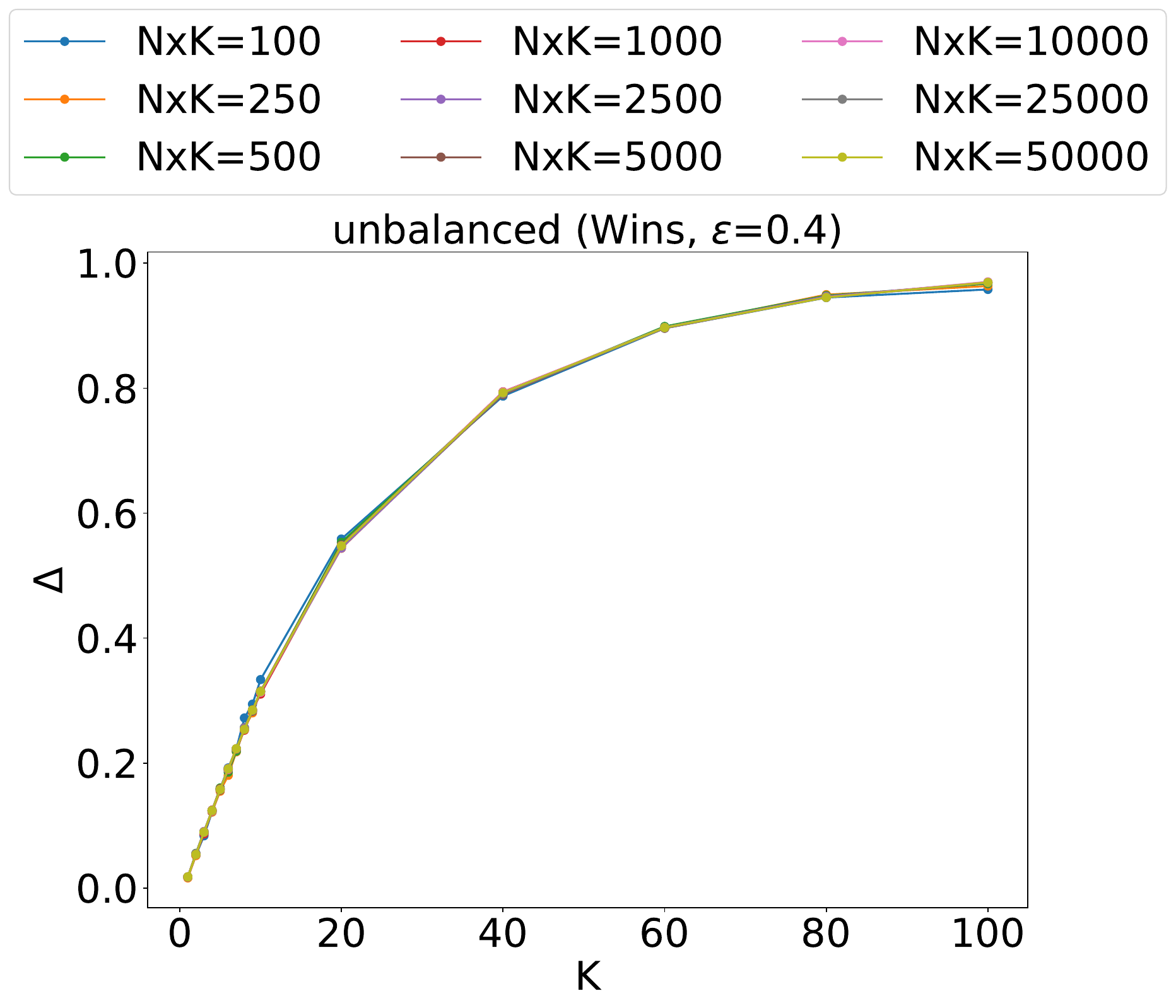}
    \caption{$\epsilon = 0.4$}
    \label{fig:gamma_delta_wins_cat12_e04}
  \end{subfigure}
  \caption{Effect sizes ($\Delta$) for unbalanced alphas with Wins as the metric ($M=12$)}
  \label{fig:gamma_delta_wins_cat12}
\end{figure*}

Figures \ref{fig:gamma_wins_cat2}--\ref{fig:gamma_delta_wins_cat12} show results for \pv s, confidence intervals, and effect sizes of Wins for unbalanced priors on parameter $\alpha$ with different values of $\epsilon$ and increasing $K$ for various $N \times K$. For simulations with unbalanced categories, \pv s\, CI-width, and effect sizes show similar trends as balanced categories.

\subsection{KL-Divergence}

\subsubsection{Real-World Datasets}

\begin{figure*}
  \centering
  \begin{subfigure}[b]{0.24\linewidth}
    \centering
    \includegraphics[width=\linewidth]{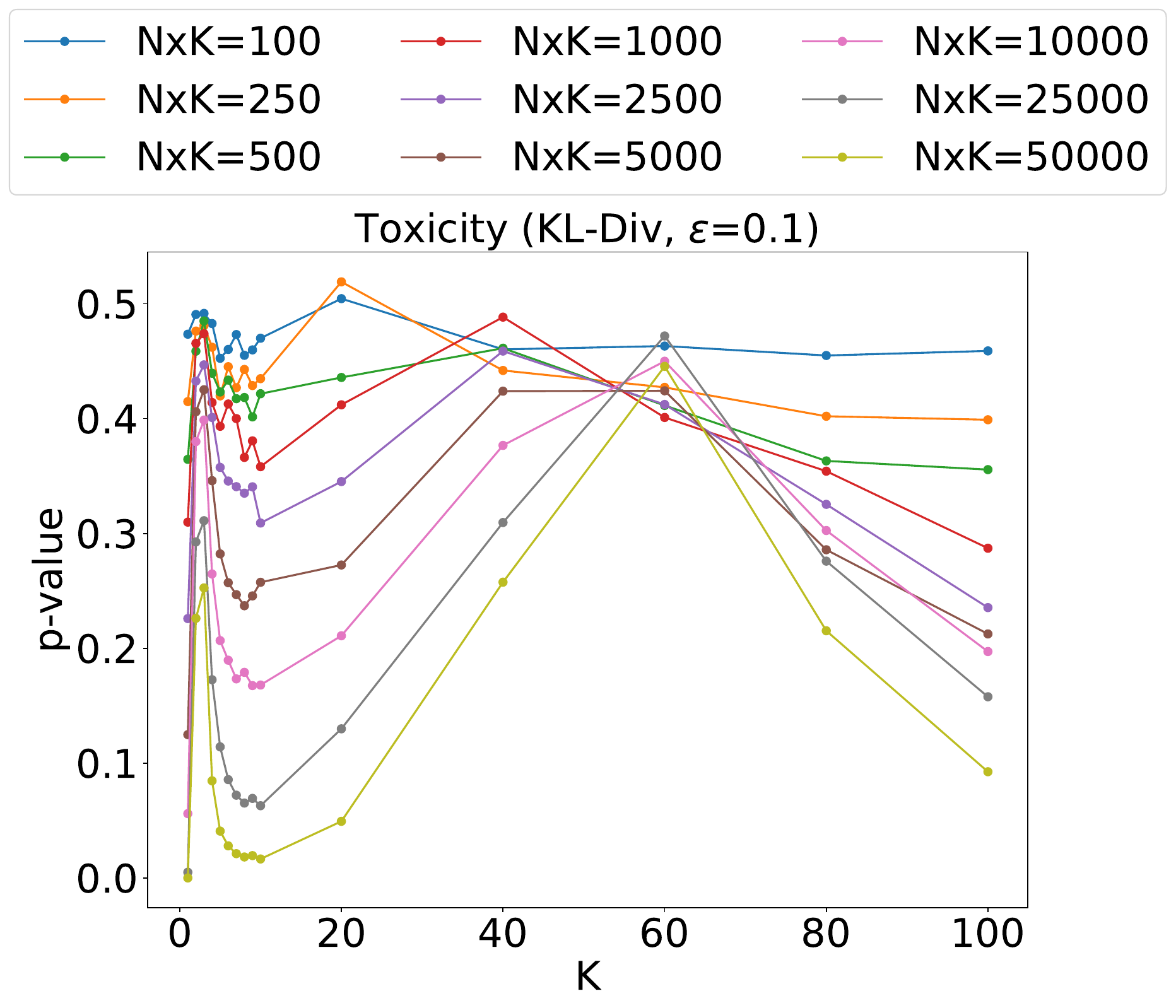}
    \caption{$\epsilon = 0.1$}
    \label{fig:toxicity_kl_e01}
  \end{subfigure} \hfill
  \begin{subfigure}[b]{0.24\linewidth}
    \centering
    \includegraphics[width=\linewidth]{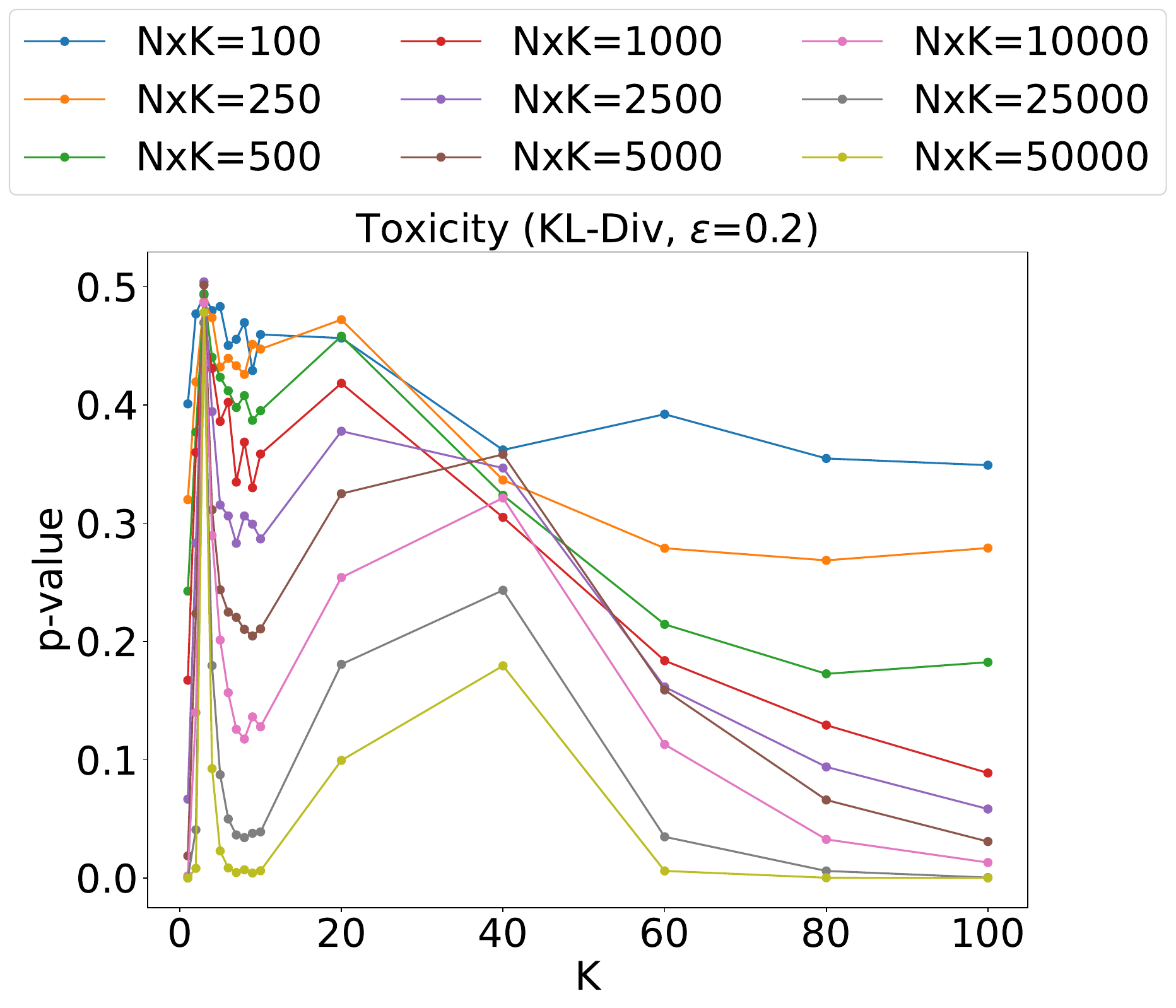}
    \caption{$\epsilon = 0.2$}
    \label{fig:toxicity_kl_e02}
  \end{subfigure} \hfill
  \begin{subfigure}[b]{0.24\linewidth}
    \centering
    \includegraphics[width=\linewidth]{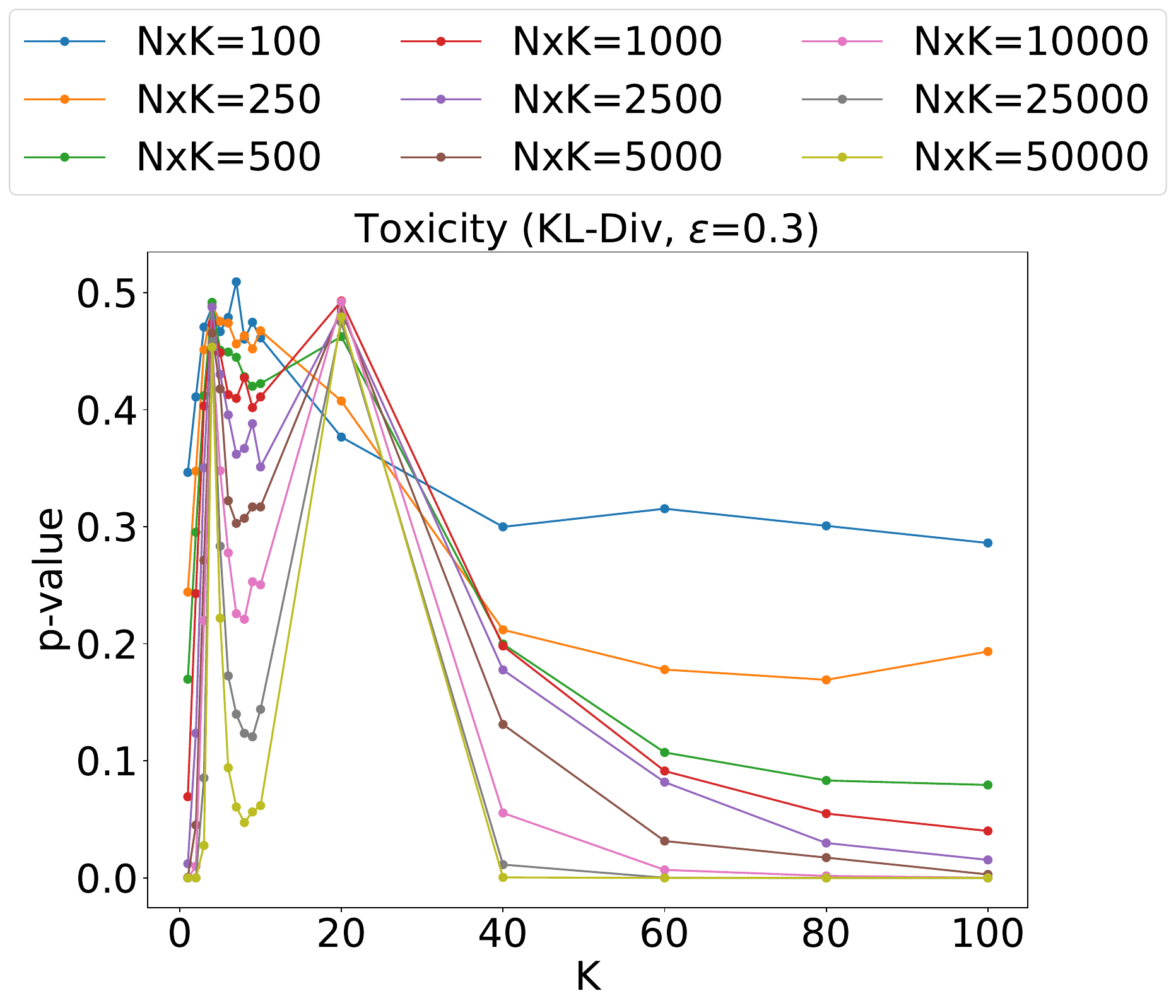}
    \caption{$\epsilon = 0.3$}
    \label{fig:toxicity_kl_e03}
  \end{subfigure} \hfill
  \begin{subfigure}[b]{0.24\linewidth}
    \centering
    \includegraphics[width=\linewidth]{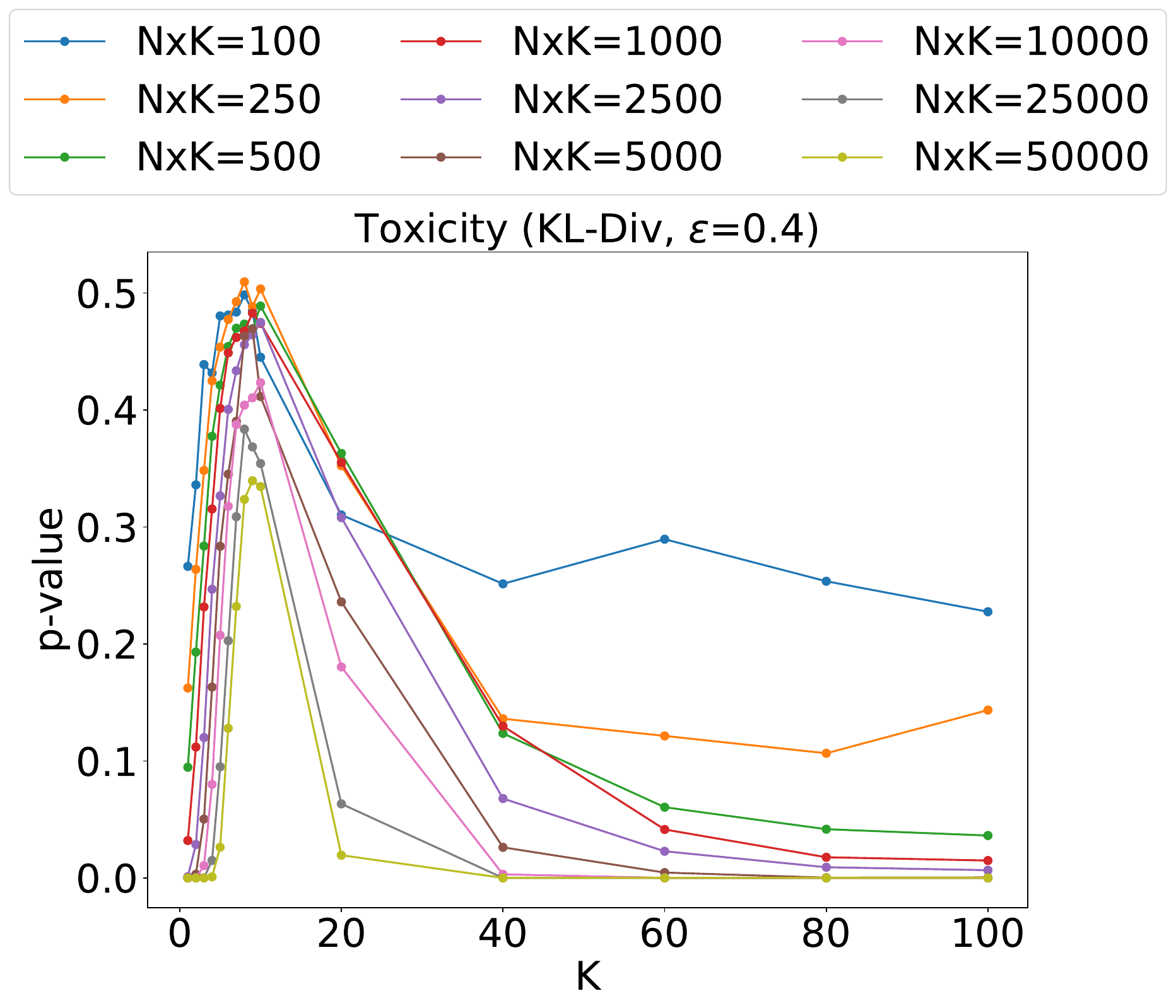}
    \caption{$\epsilon = 0.4$}
    \label{fig:toxicity_kl_e04}
  \end{subfigure}
  \caption{P-value plots for Toxicity dataset with KL-divergence as the metric}
  \label{fig:toxicity_kl}
\end{figure*}

\begin{figure*}
  \centering
  \begin{subfigure}[b]{0.24\linewidth}
    \centering
    \includegraphics[width=\linewidth]{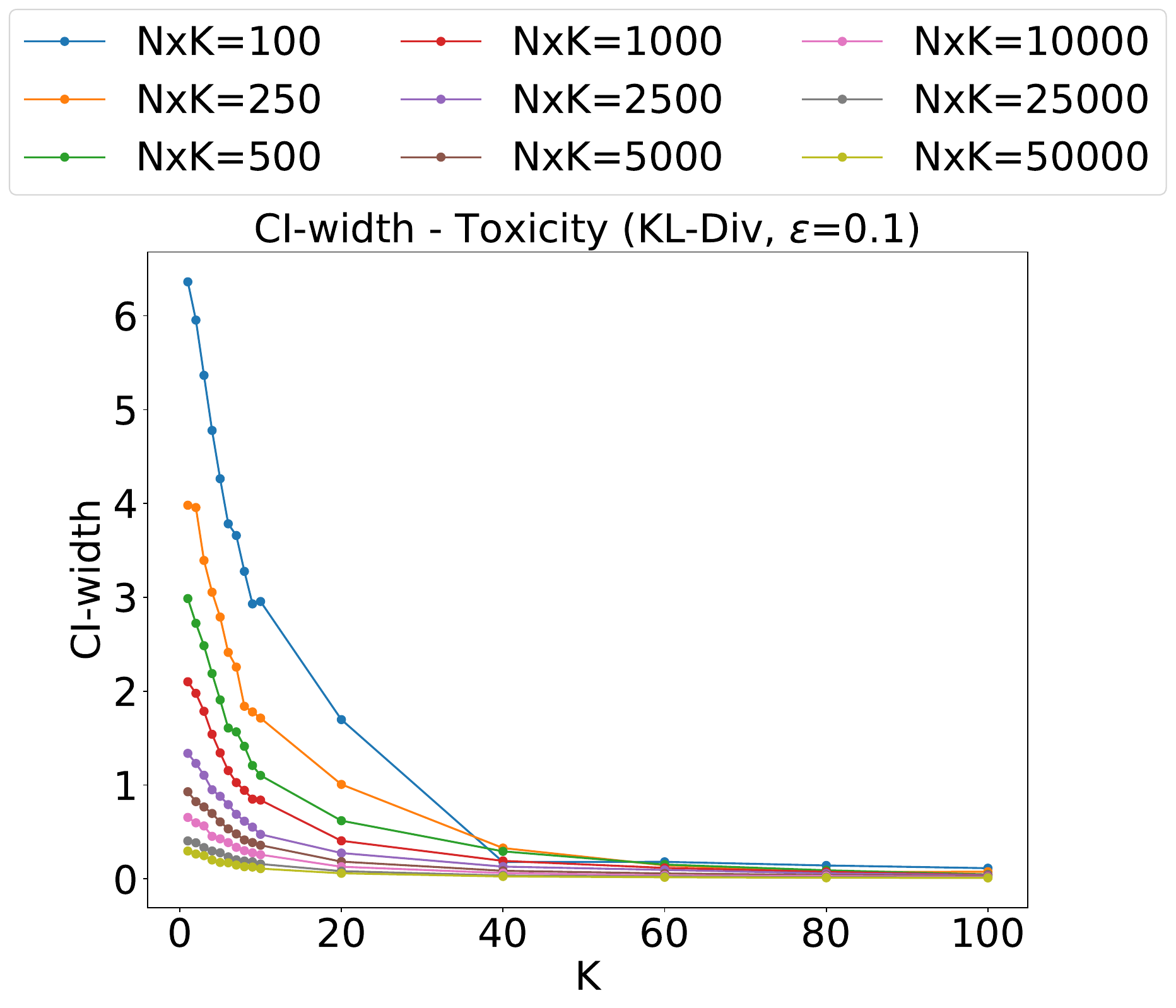}
    \caption{$\epsilon = 0.1$}
    \label{fig:toxicity_ci_kl_e01}
  \end{subfigure} \hfill
  \begin{subfigure}[b]{0.24\linewidth}
    \centering
    \includegraphics[width=\linewidth]{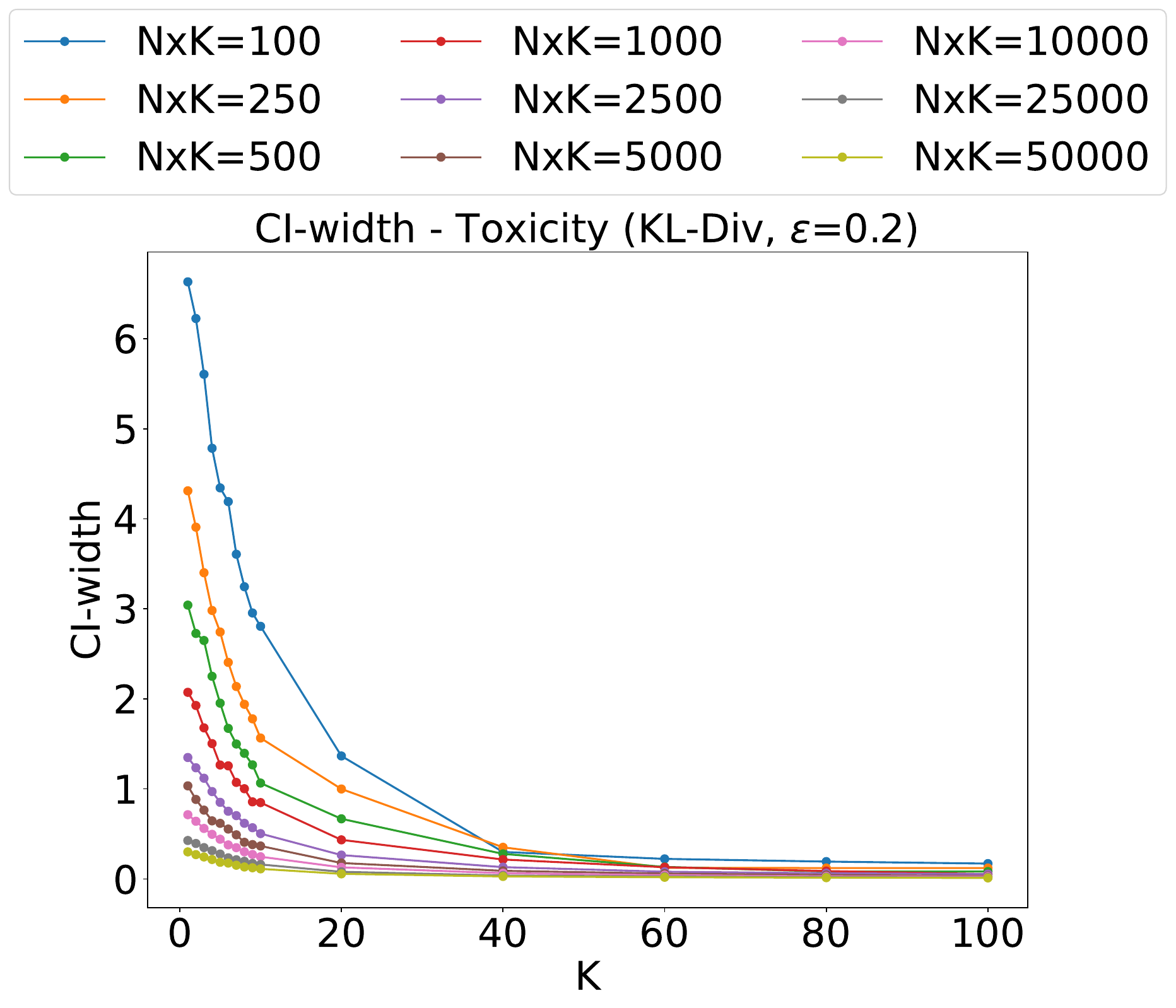}
    \caption{$\epsilon = 0.2$}
    \label{fig:toxicity_ci_kl_e02}
  \end{subfigure} \hfill
  \begin{subfigure}[b]{0.24\linewidth}
    \centering
    \includegraphics[width=\linewidth]{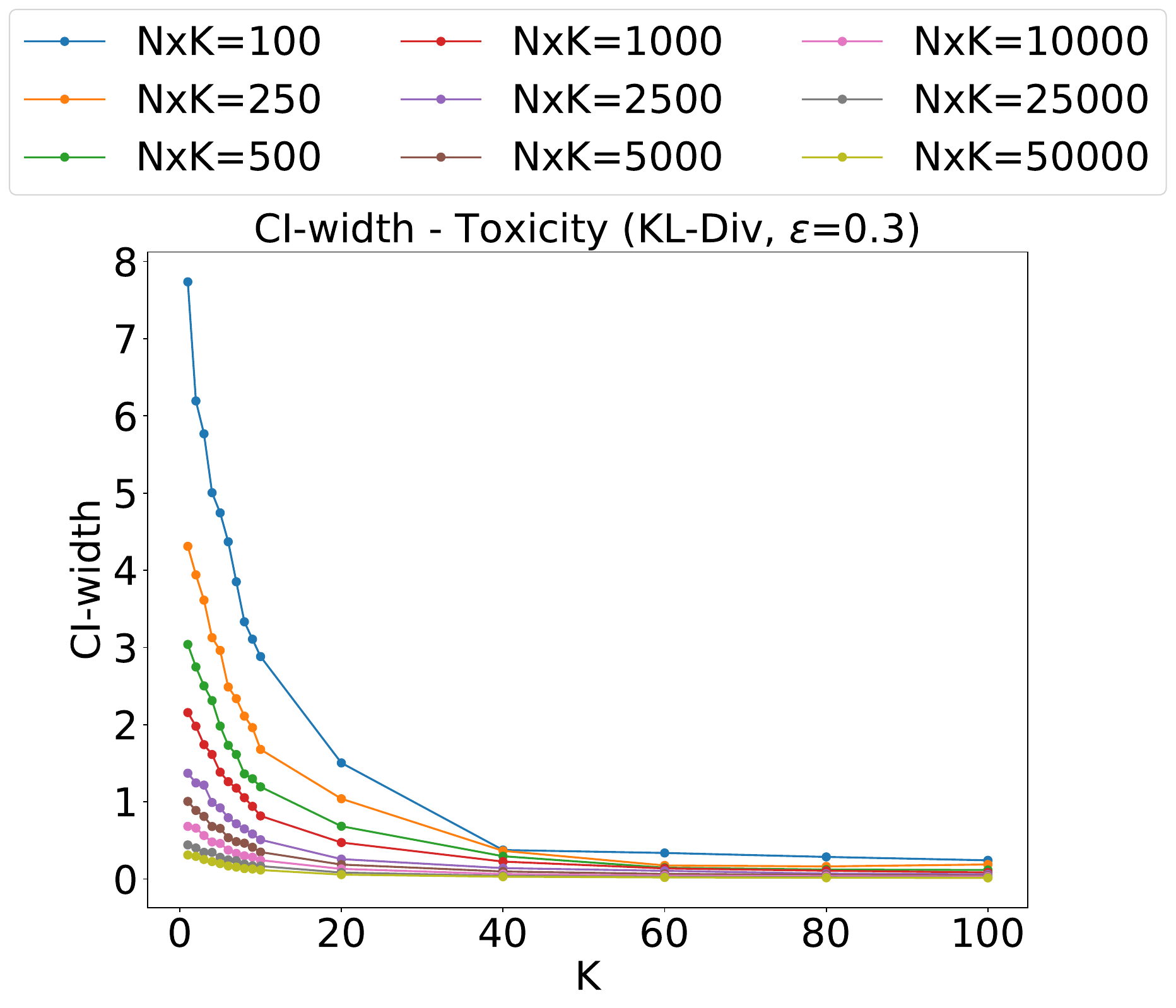}
    \caption{$\epsilon = 0.3$}
    \label{fig:toxicity_ci_kl_e03}
  \end{subfigure} \hfill
  \begin{subfigure}[b]{0.24\linewidth}
    \centering
    \includegraphics[width=\linewidth]{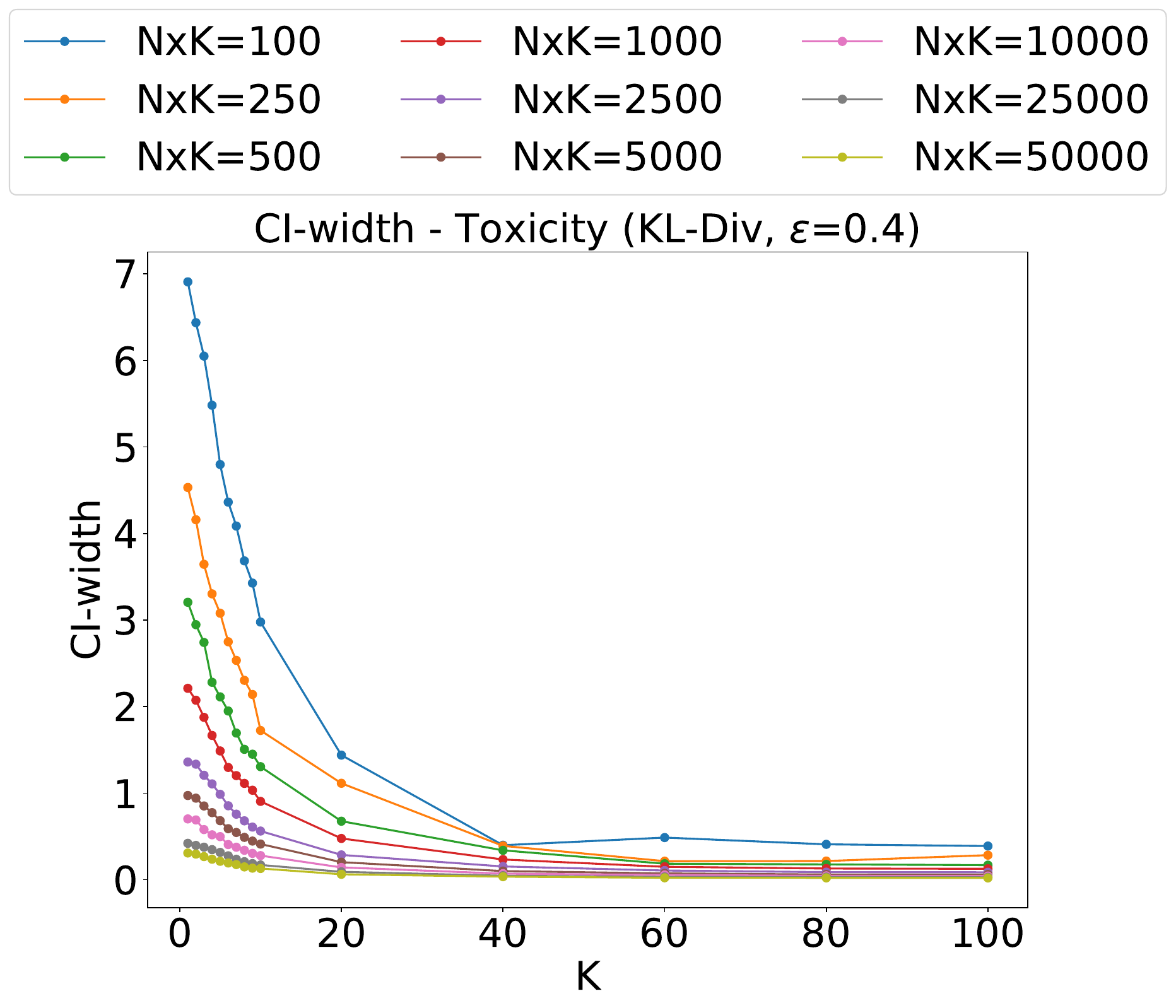}
    \caption{$\epsilon = 0.4$}
    \label{fig:toxicity_ci_kl_e04}
  \end{subfigure}
  \caption{CI-width plots for Toxicity dataset with KL-divergence as the metric}
  \label{fig:toxicity_ci_kl}
\end{figure*}

\begin{figure*}
  \centering
  \begin{subfigure}[b]{0.24\linewidth}
    \centering
    \includegraphics[width=\linewidth]{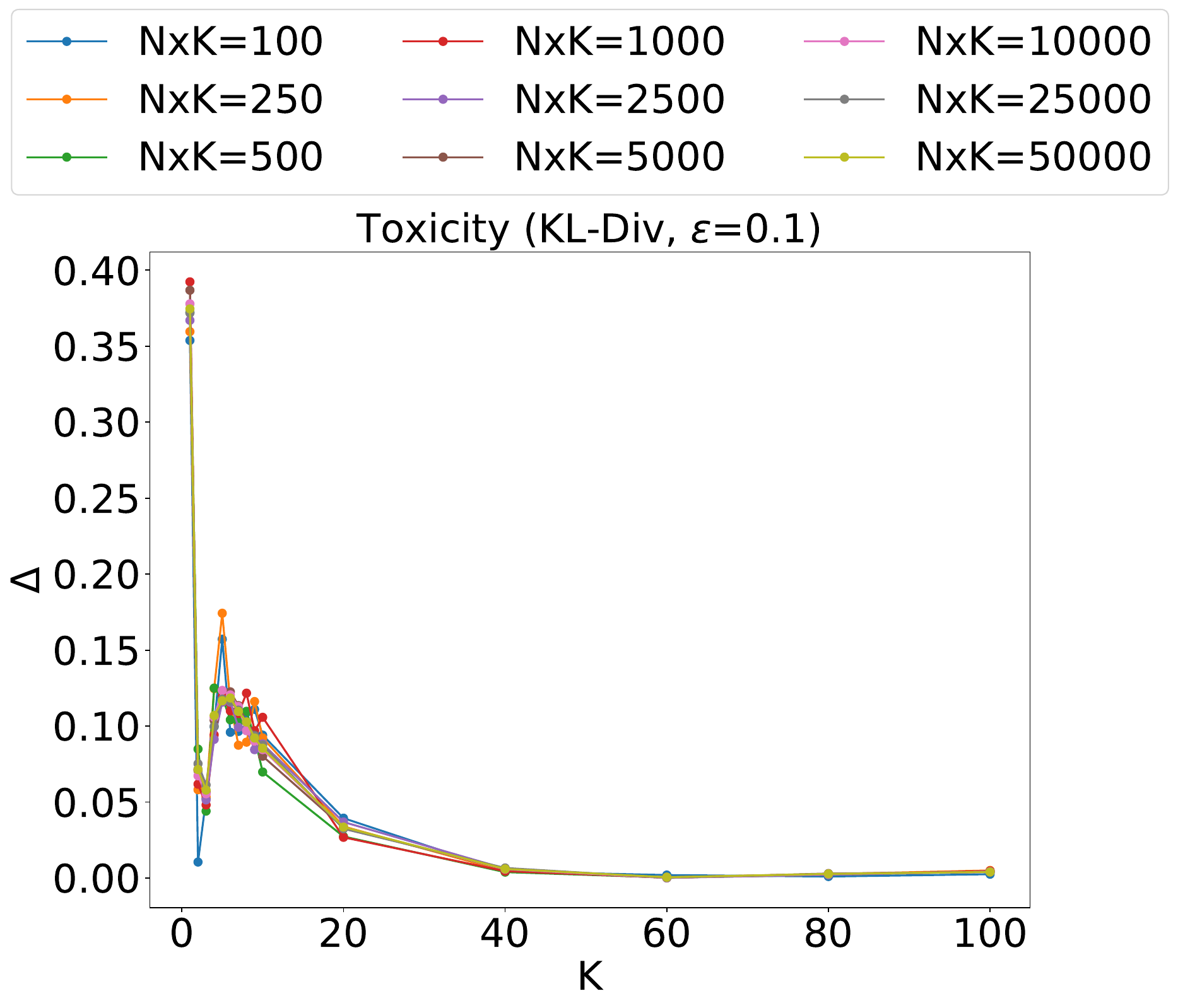}
    \caption{$\epsilon = 0.1$}
    \label{fig:toxicity_delta_kl_e01}
  \end{subfigure} \hfill
  \begin{subfigure}[b]{0.24\linewidth}
    \centering
    \includegraphics[width=\linewidth]{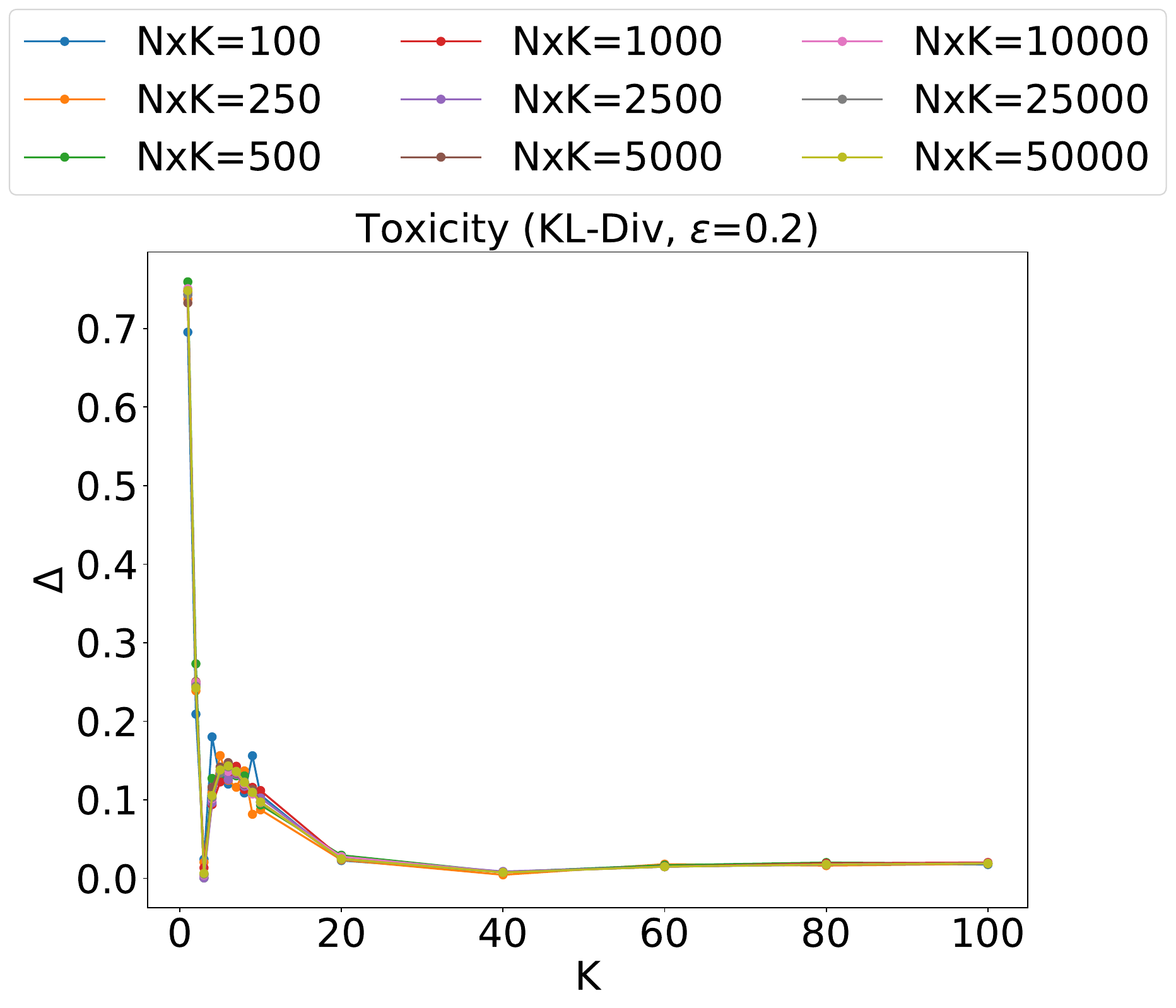}
    \caption{$\epsilon = 0.2$}
    \label{fig:toxicity_delta_kl_e02}
  \end{subfigure} \hfill
  \begin{subfigure}[b]{0.24\linewidth}
    \centering
    \includegraphics[width=\linewidth]{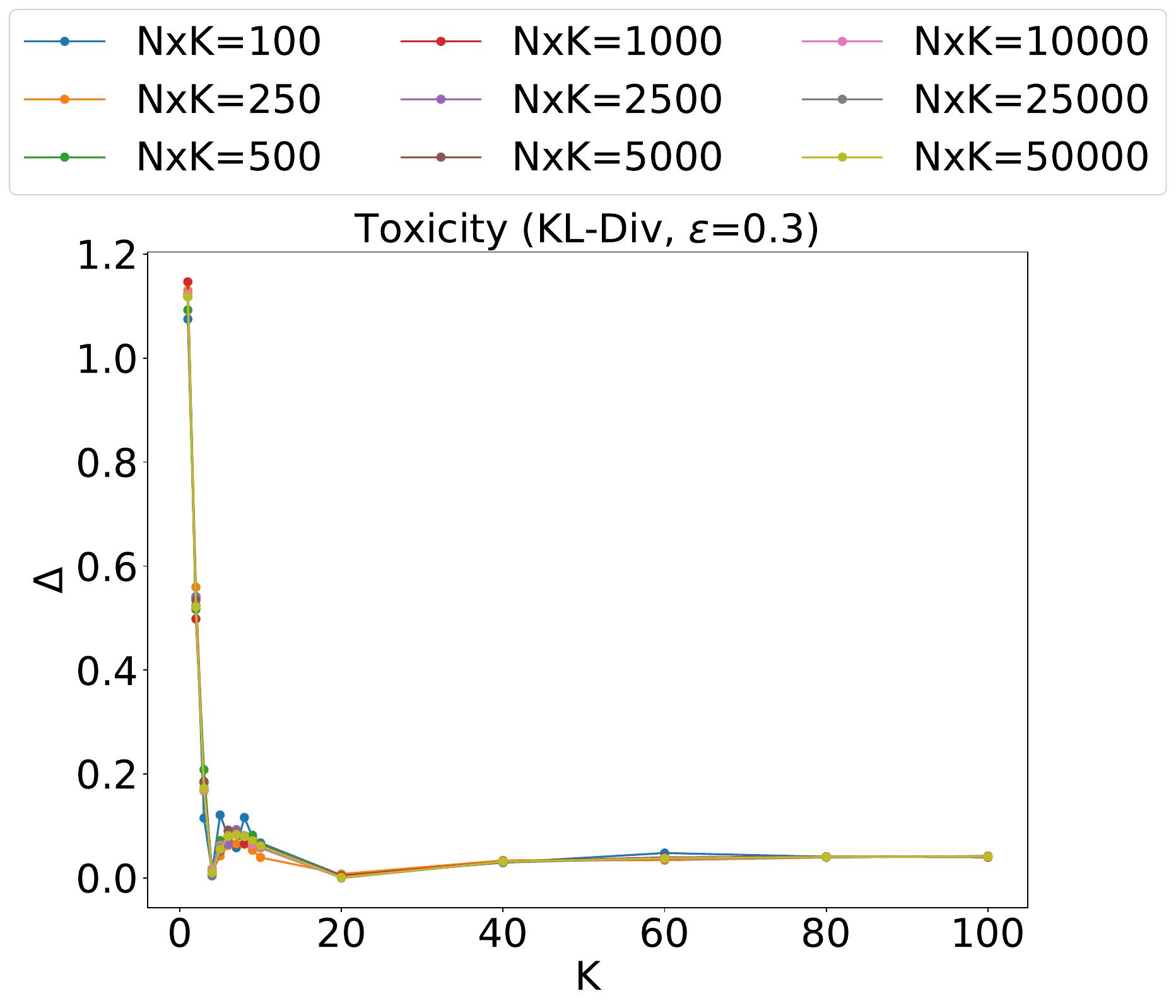}
    \caption{$\epsilon = 0.3$}
    \label{fig:toxicity_delta_kl_e03}
  \end{subfigure} \hfill
  \begin{subfigure}[b]{0.24\linewidth}
    \centering
    \includegraphics[width=\linewidth]{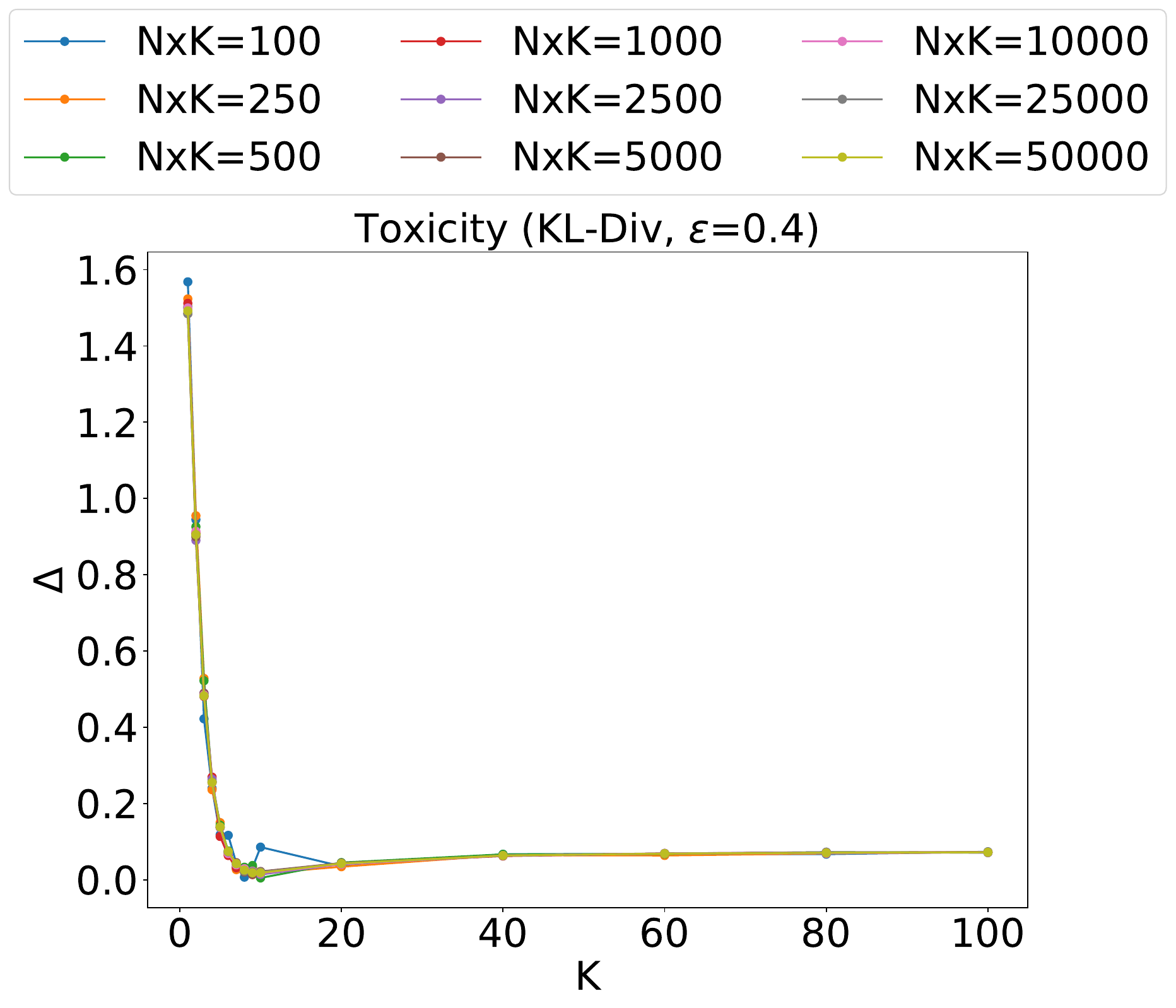}
    \caption{$\epsilon = 0.4$}
    \label{fig:toxicity_delta_kl_e04}
  \end{subfigure}
  \caption{Effect sizes ($\Delta$) for Toxicity dataset with KL-divergence as the metric}
  \label{fig:toxicity_delta_kl}
\end{figure*}

\begin{figure*}
  \centering
  \begin{subfigure}[b]{0.24\linewidth}
    \centering
    \includegraphics[width=\linewidth]{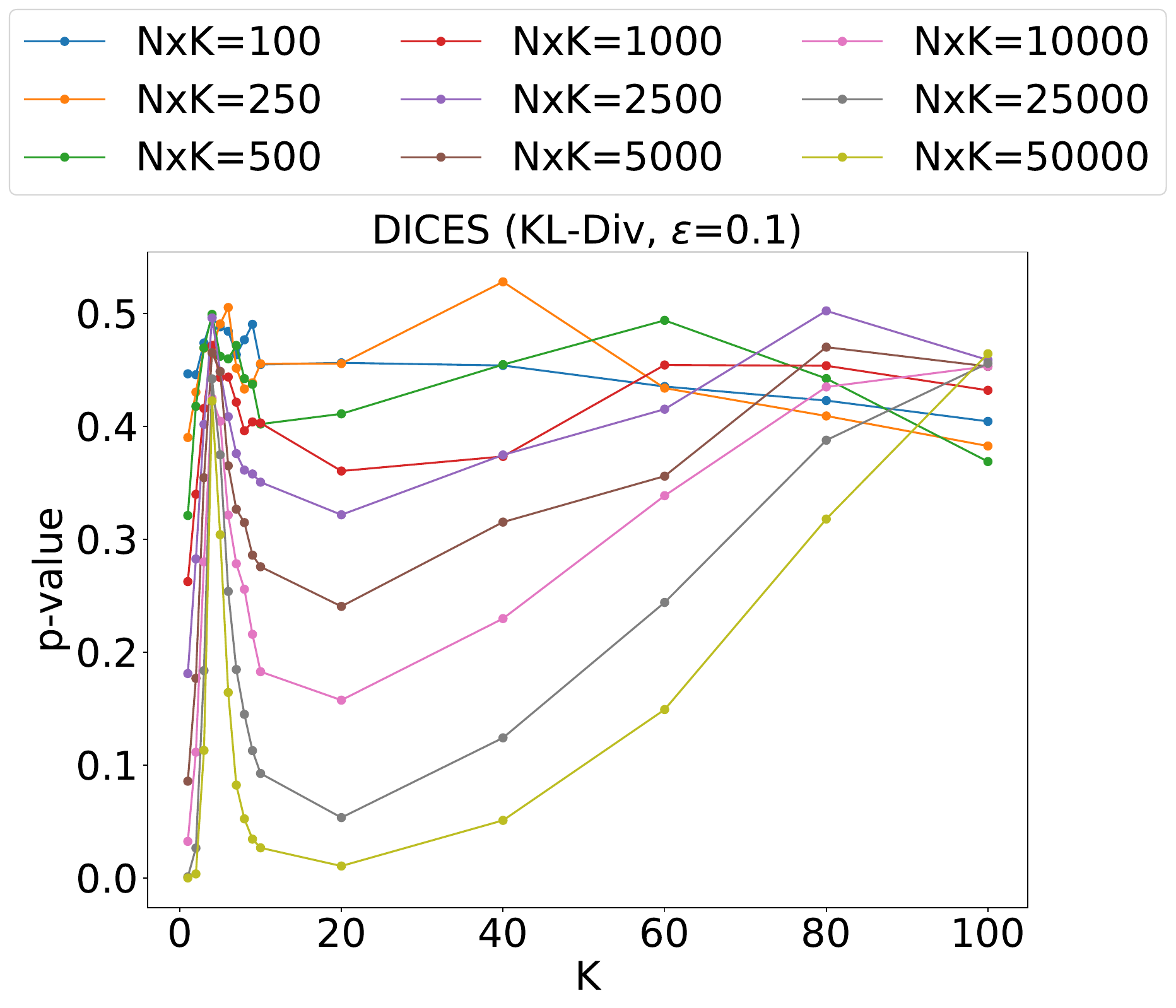}
    \caption{$\epsilon = 0.1$}
    \label{fig:dices_kl_e01}
  \end{subfigure} \hfill
  \begin{subfigure}[b]{0.24\linewidth}
    \centering
    \includegraphics[width=\linewidth]{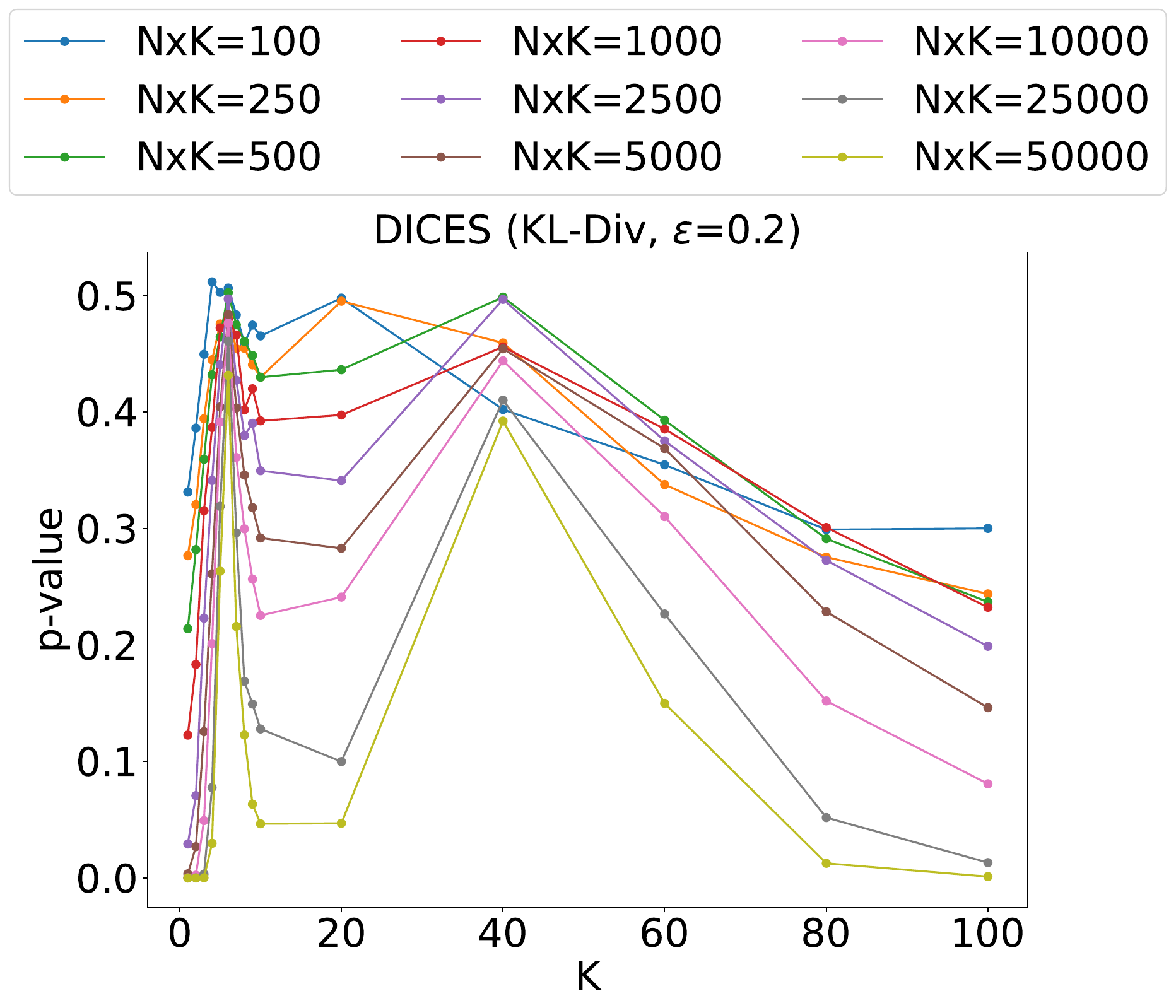}
    \caption{$\epsilon = 0.2$}
    \label{fig:dices_kl_e02}
  \end{subfigure} \hfill
  \begin{subfigure}[b]{0.24\linewidth}
    \centering
    \includegraphics[width=\linewidth]{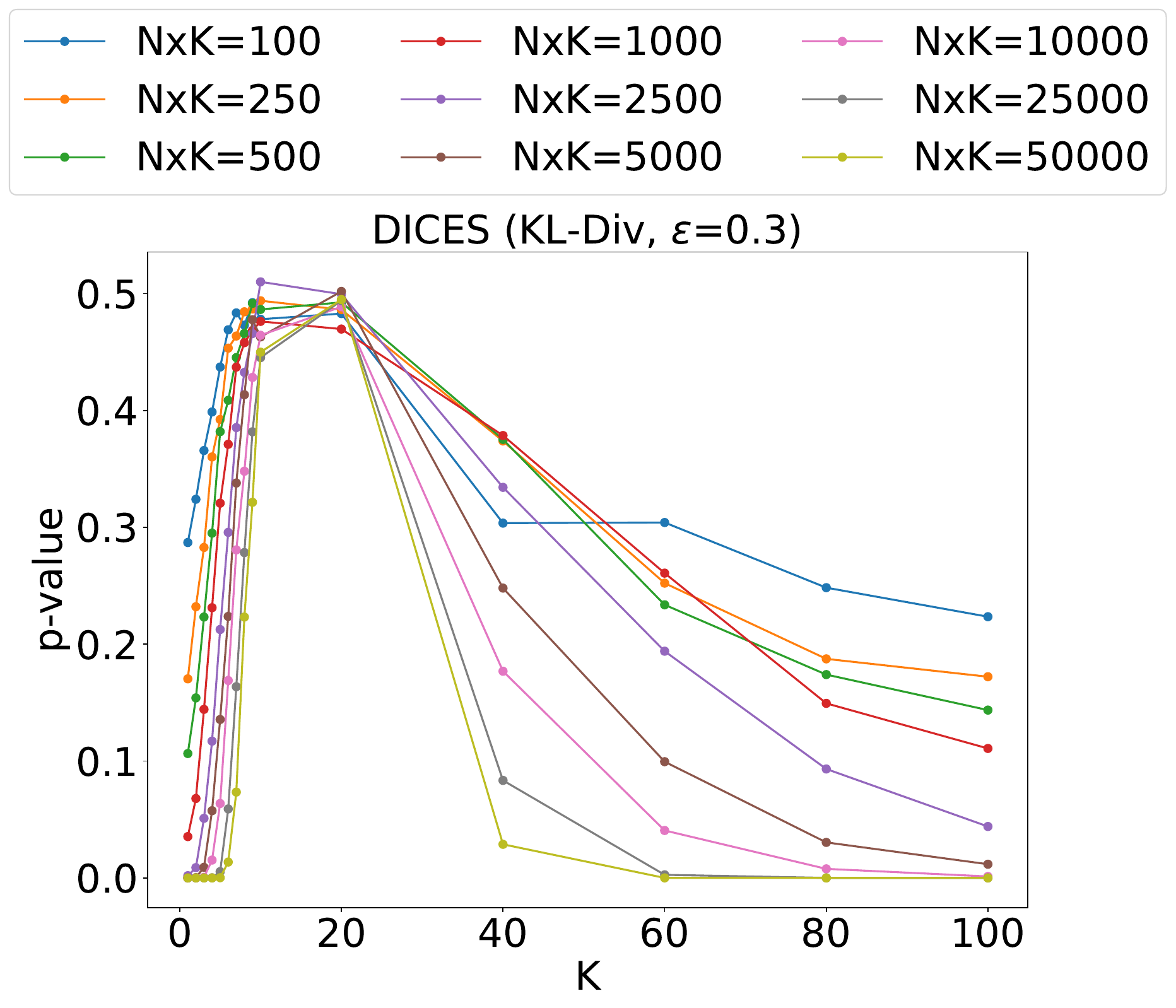}
    \caption{$\epsilon = 0.3$}
    \label{fig:dices_kl_e03}
  \end{subfigure} \hfill
  \begin{subfigure}[b]{0.24\linewidth}
    \centering
    \includegraphics[width=\linewidth]{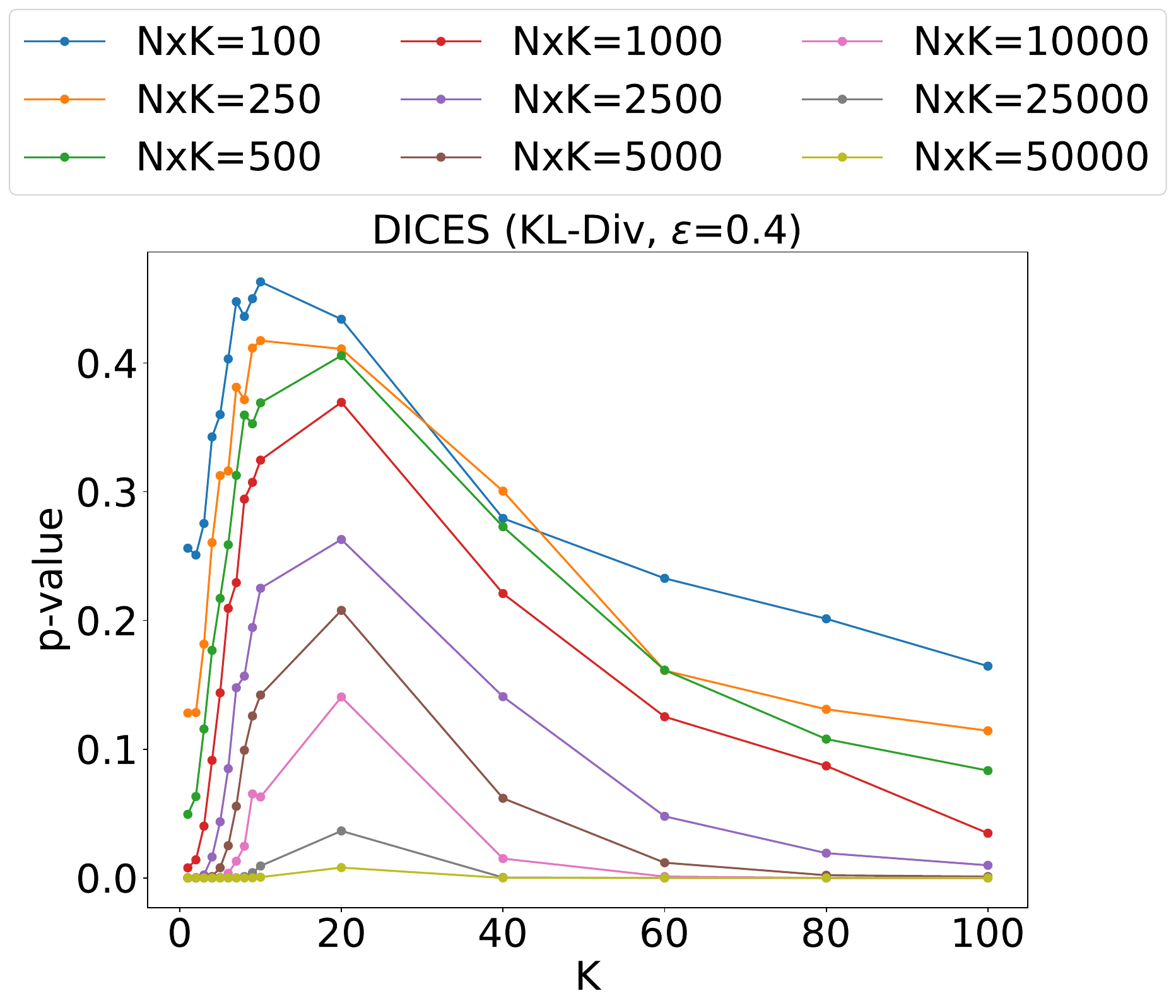}
    \caption{$\epsilon = 0.4$}
    \label{fig:dices_kl_e04}
  \end{subfigure}
  \caption{P-value plots for DICES dataset with KL-divergence as the metric}
  \label{fig:dices_kl}
\end{figure*}

\begin{figure*}
  \centering
  \begin{subfigure}[b]{0.24\linewidth}
    \centering
    \includegraphics[width=\linewidth]{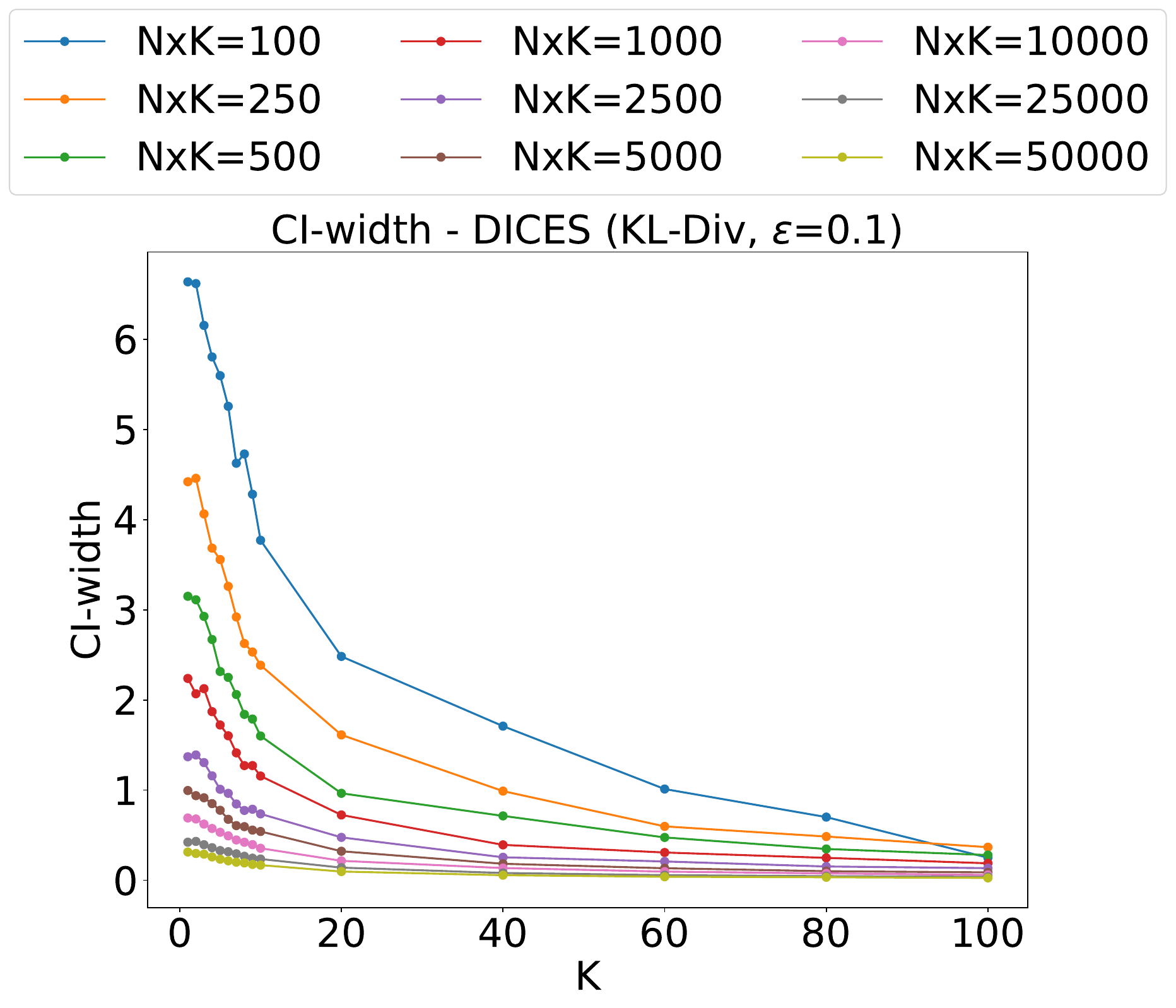}
    \caption{$\epsilon = 0.1$}
    \label{fig:dices_ci_kl_e01}
  \end{subfigure} \hfill
  \begin{subfigure}[b]{0.24\linewidth}
    \centering
    \includegraphics[width=\linewidth]{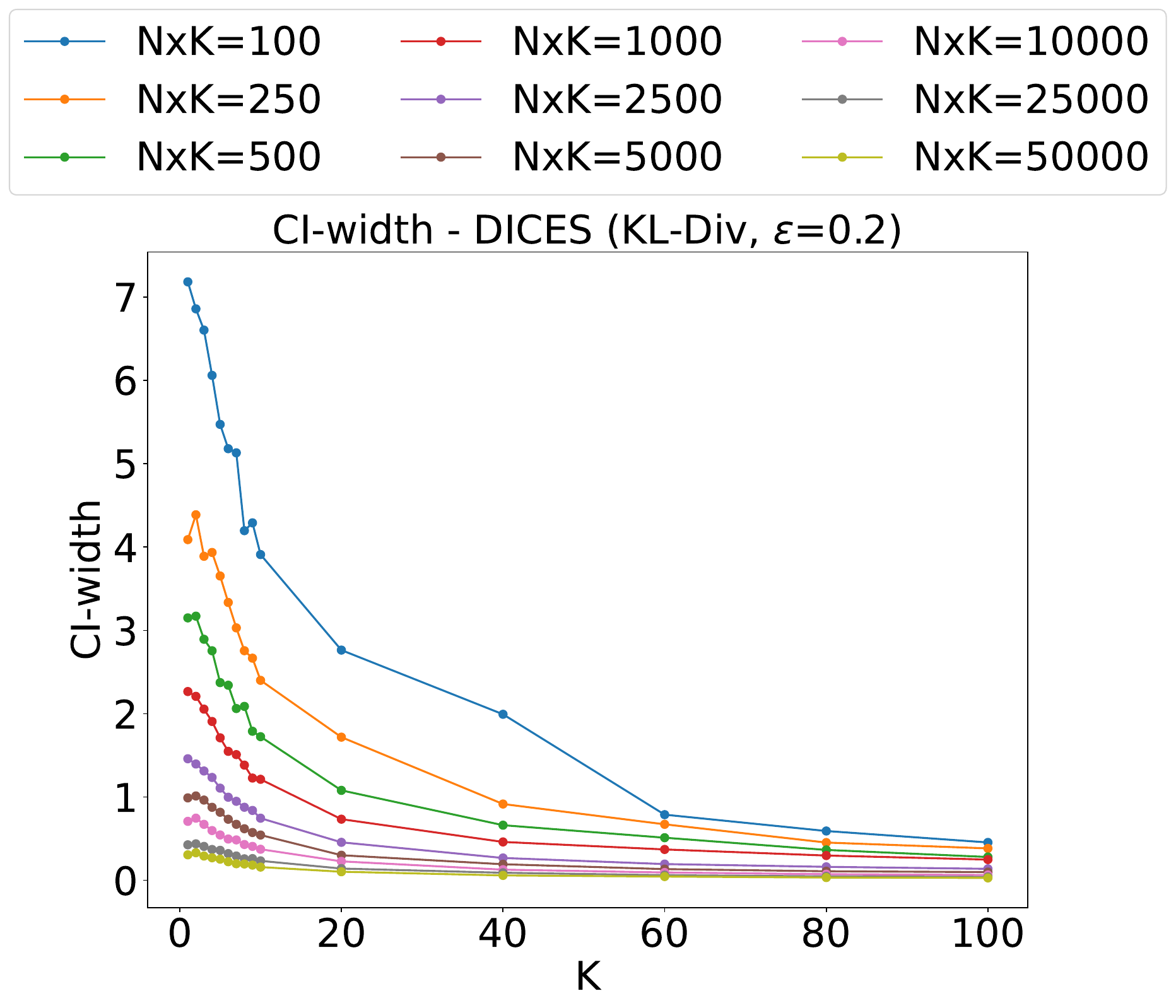}
    \caption{$\epsilon = 0.2$}
    \label{fig:dices_ci_kl_e02}
  \end{subfigure} \hfill
  \begin{subfigure}[b]{0.24\linewidth}
    \centering
    \includegraphics[width=\linewidth]{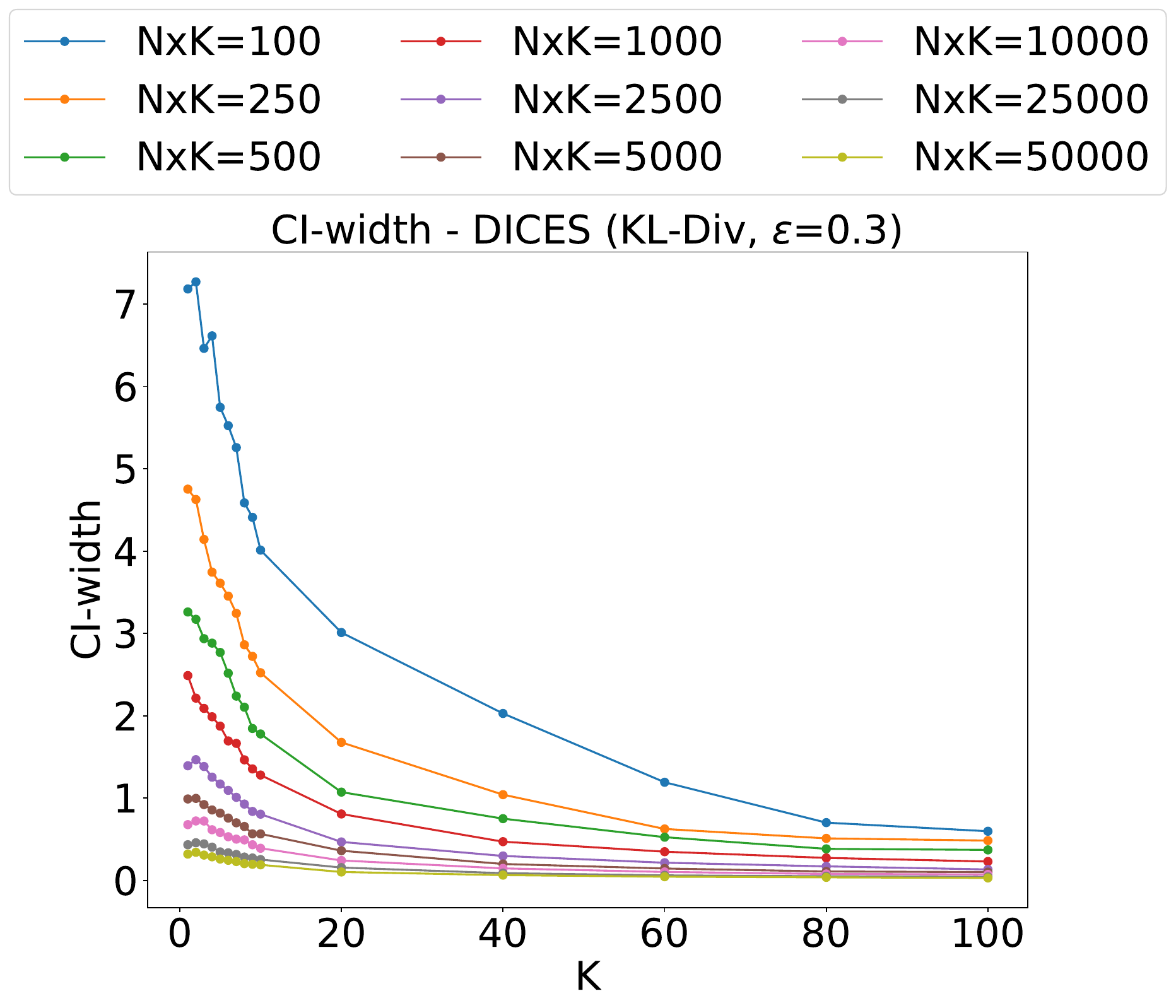}
    \caption{$\epsilon = 0.3$}
    \label{fig:dices_ci_kl_e03}
  \end{subfigure} \hfill
  \begin{subfigure}[b]{0.24\linewidth}
    \centering
    \includegraphics[width=\linewidth]{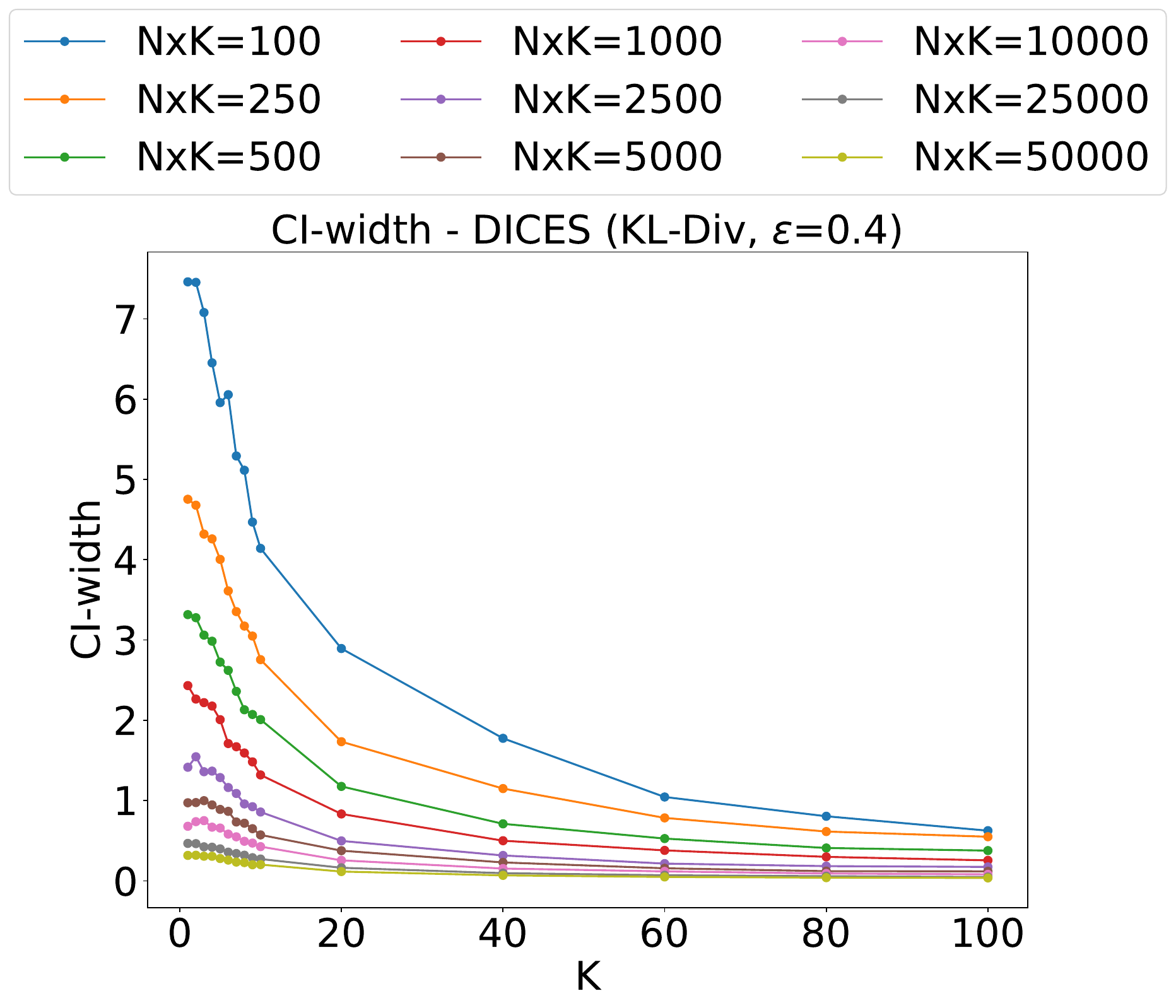}
    \caption{$\epsilon = 0.4$}
    \label{fig:dices_ci_kl_e04}
  \end{subfigure}
  \caption{CI-width plots for DICES dataset with KL-divergence as the metric}
  \label{fig:dices_ci_kl}
\end{figure*}

\begin{figure*}
  \centering
  \begin{subfigure}[b]{0.24\linewidth}
    \centering
    \includegraphics[width=\linewidth]{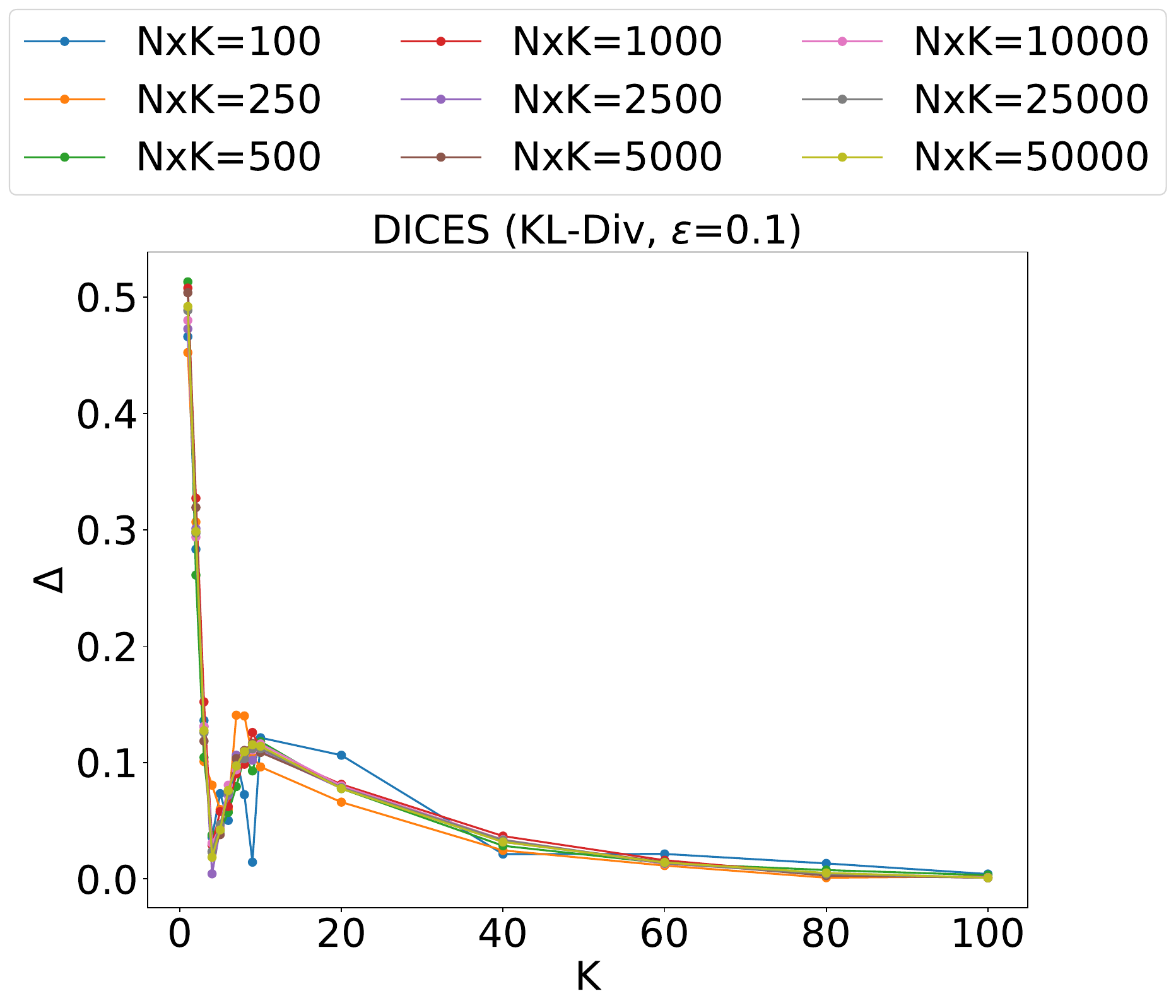}
    \caption{$\epsilon = 0.1$}
    \label{fig:dices_delta_kl_e01}
  \end{subfigure} \hfill
  \begin{subfigure}[b]{0.24\linewidth}
    \centering
    \includegraphics[width=\linewidth]{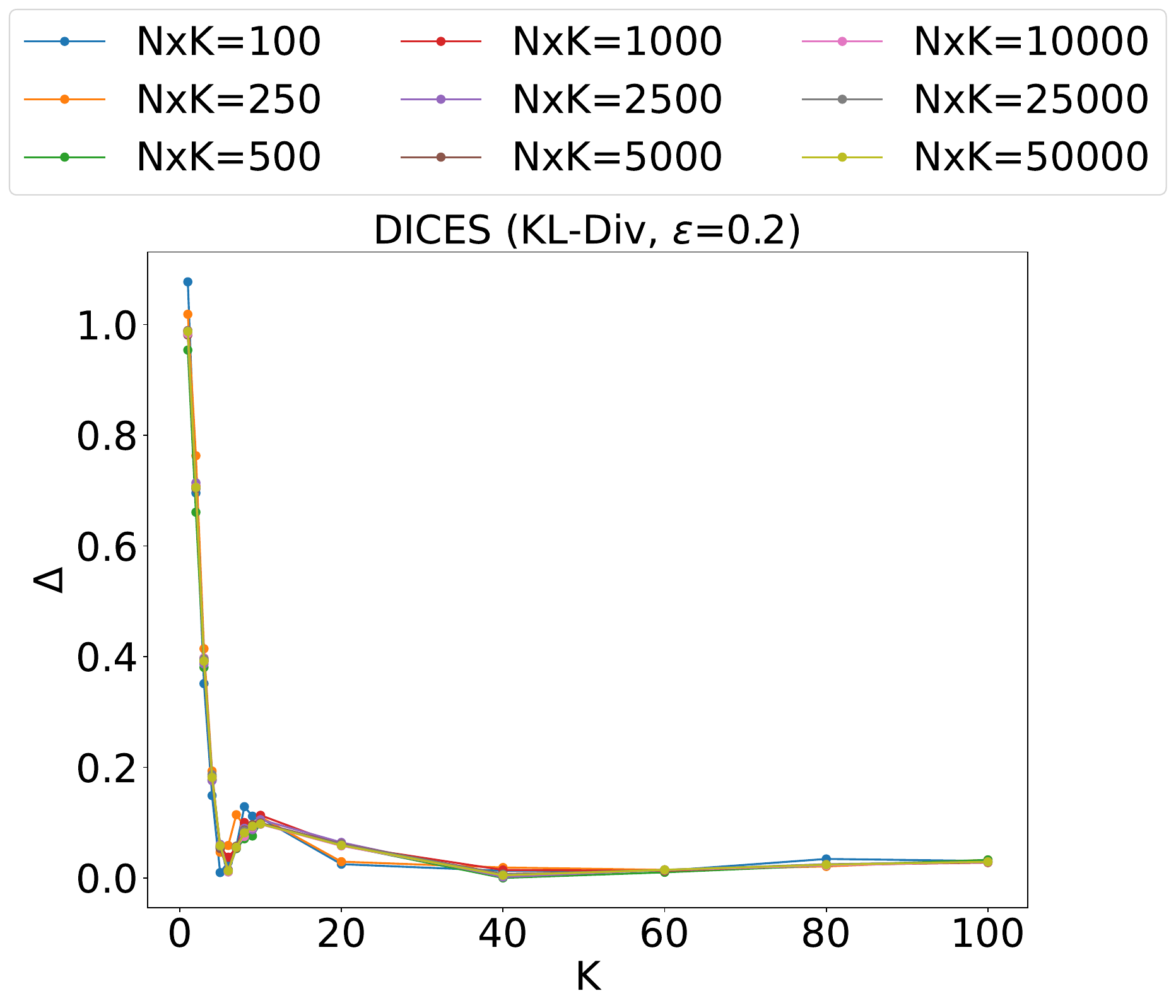}
    \caption{$\epsilon = 0.2$}
    \label{fig:dices_delta_kl_e02}
  \end{subfigure} \hfill
  \begin{subfigure}[b]{0.24\linewidth}
    \centering
    \includegraphics[width=\linewidth]{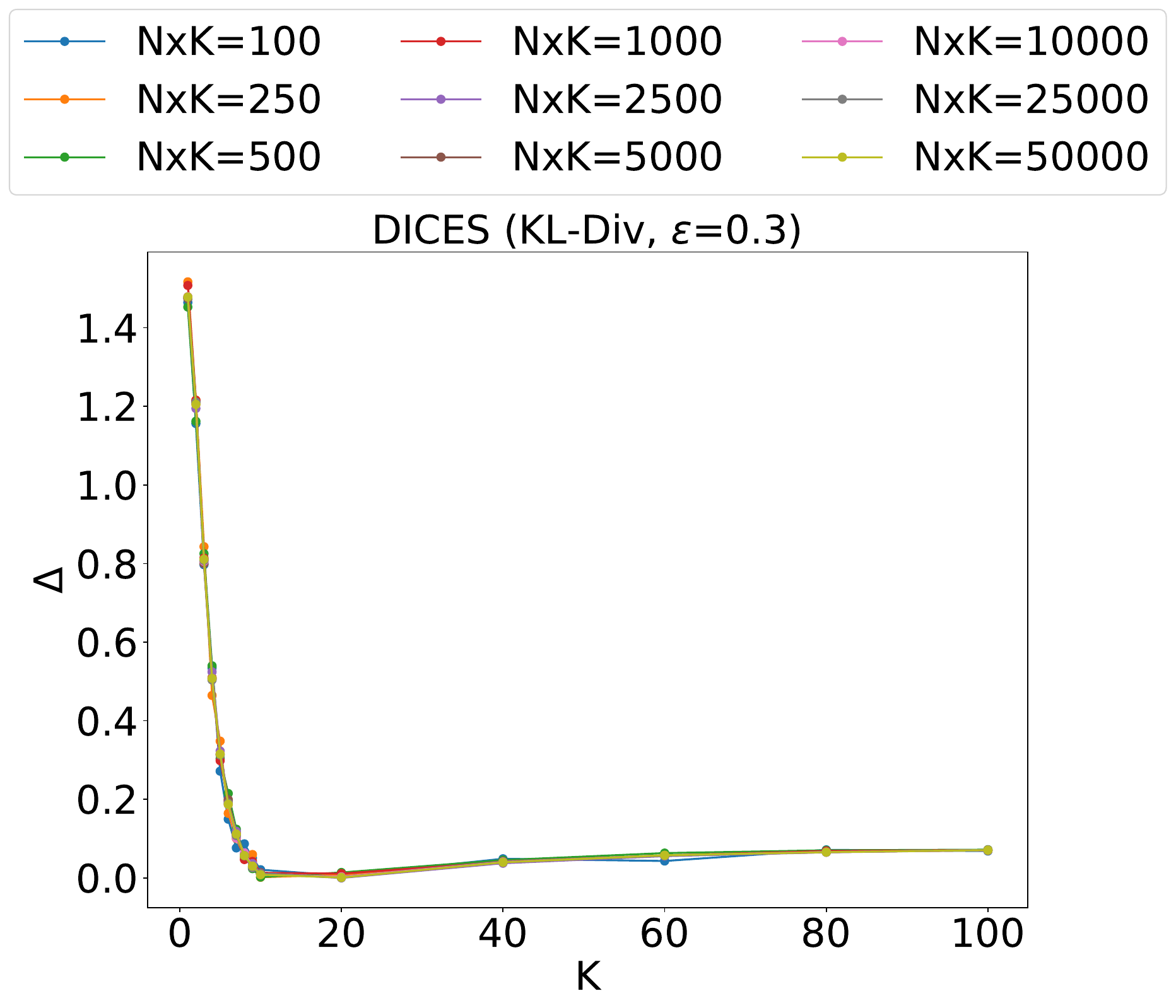}
    \caption{$\epsilon = 0.3$}
    \label{fig:dices_delta_kl_e03}
  \end{subfigure} \hfill
  \begin{subfigure}[b]{0.24\linewidth}
    \centering
    \includegraphics[width=\linewidth]{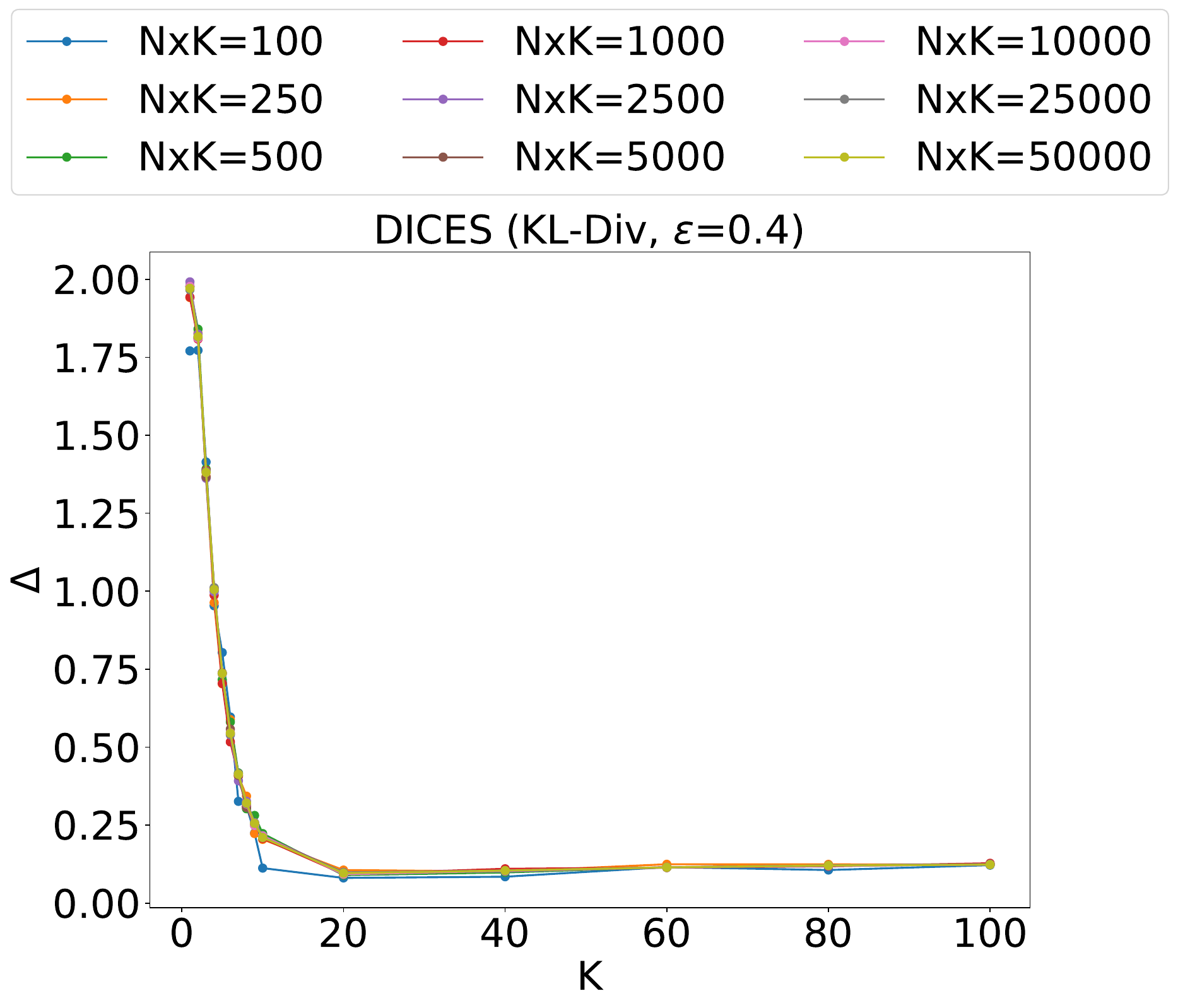}
    \caption{$\epsilon = 0.4$}
    \label{fig:dices_delta_kl_e04}
  \end{subfigure}
  \caption{Effect sizes ($\Delta$) for DICES dataset with KL-divergence as the metric}
  \label{fig:dices_delta_kl}
\end{figure*}

\begin{figure*}
  \centering
  \begin{subfigure}[b]{0.24\linewidth}
    \centering
    \includegraphics[width=\linewidth]{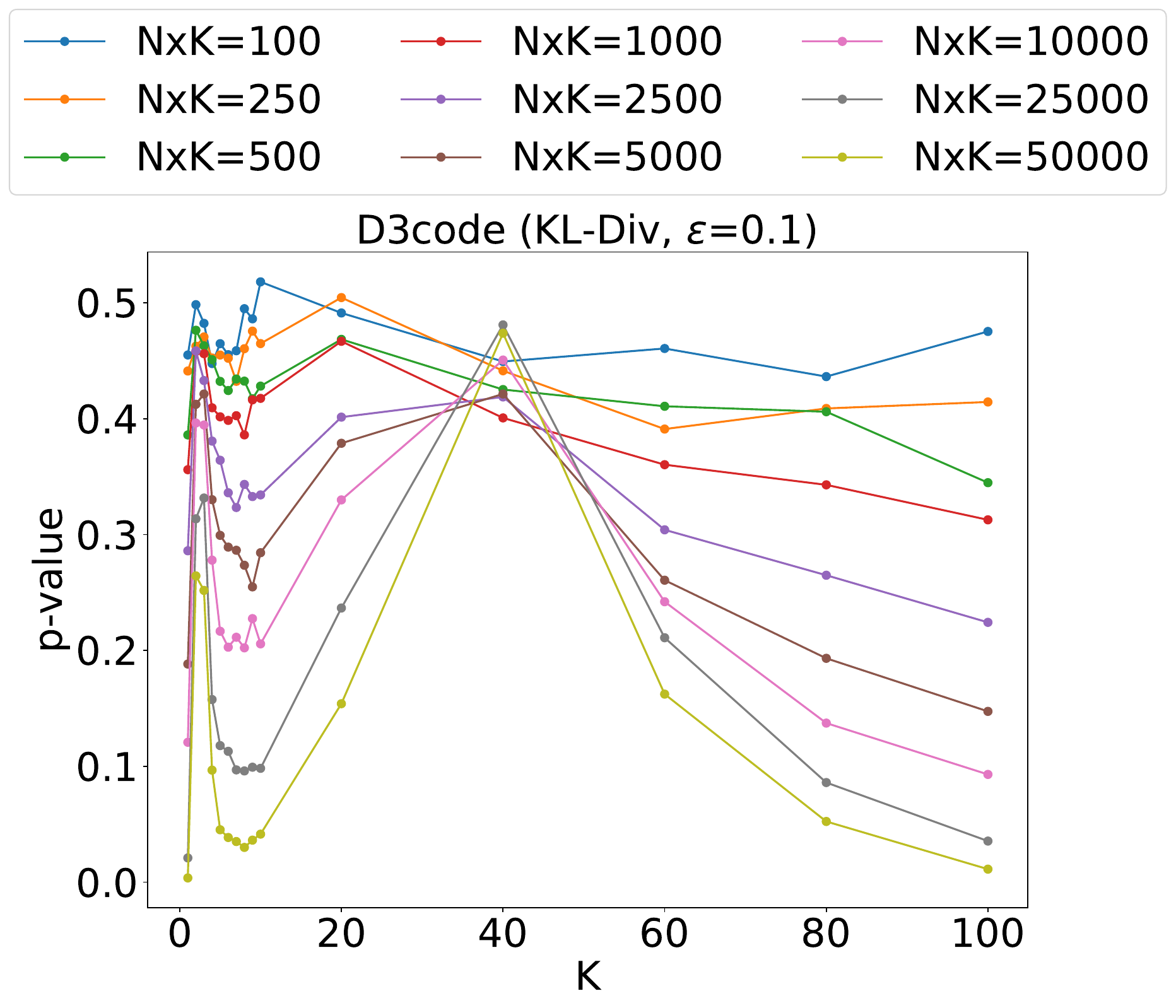}
    \caption{$\epsilon = 0.1$}
    \label{fig:d3code_kl_e01}
  \end{subfigure} \hfill
  \begin{subfigure}[b]{0.24\linewidth}
    \centering
    \includegraphics[width=\linewidth]{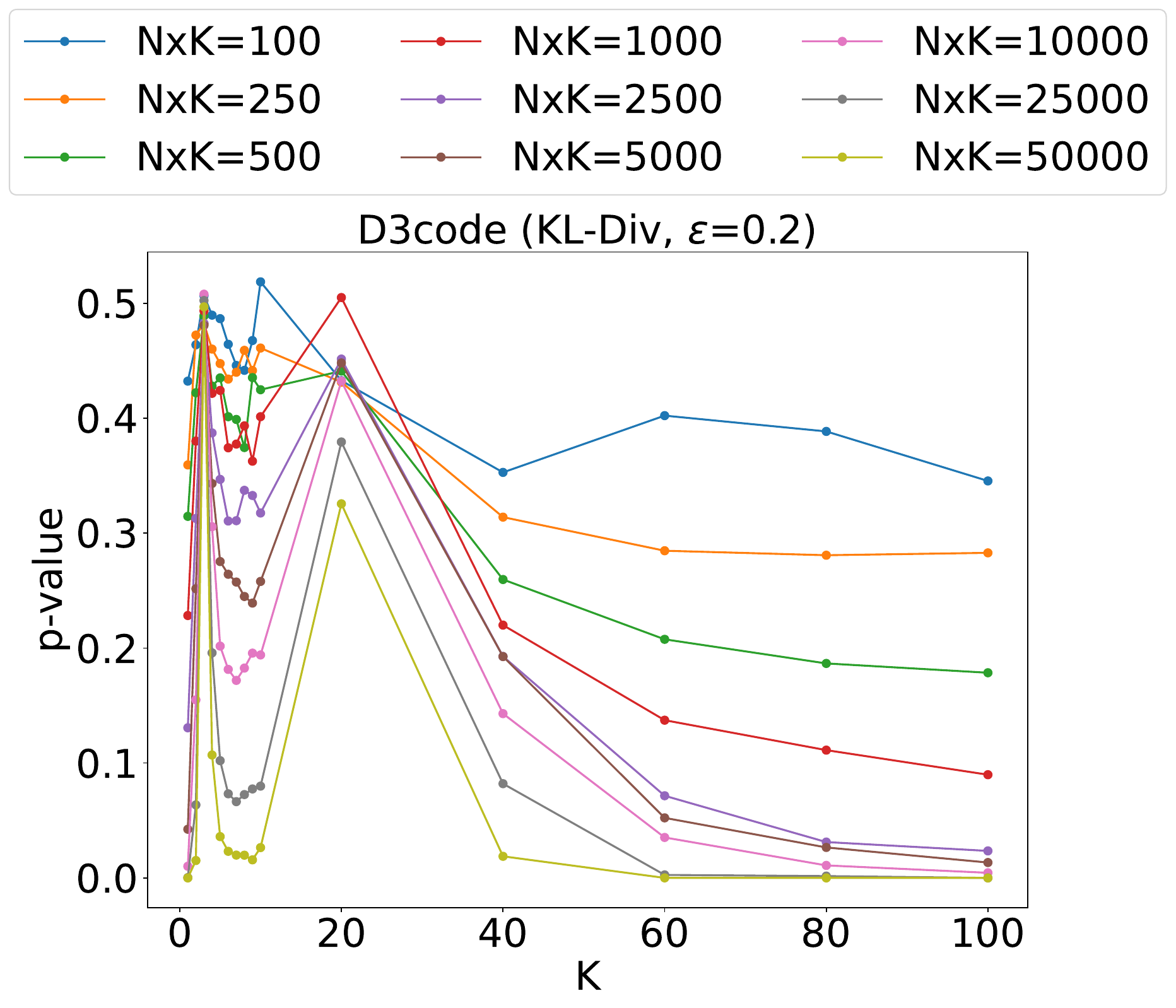}
    \caption{$\epsilon = 0.2$}
    \label{fig:d3code_kl_e02}
  \end{subfigure} \hfill
  \begin{subfigure}[b]{0.24\linewidth}
    \centering
    \includegraphics[width=\linewidth]{figures/K100/pvals_plots/D3code/D3code_p_vals_KL-Div_K_100_e_0.3.pdf}
    \caption{$\epsilon = 0.3$}
    \label{fig:d3code_kl_e03}
  \end{subfigure} \hfill
  \begin{subfigure}[b]{0.24\linewidth}
    \centering
    \includegraphics[width=\linewidth]{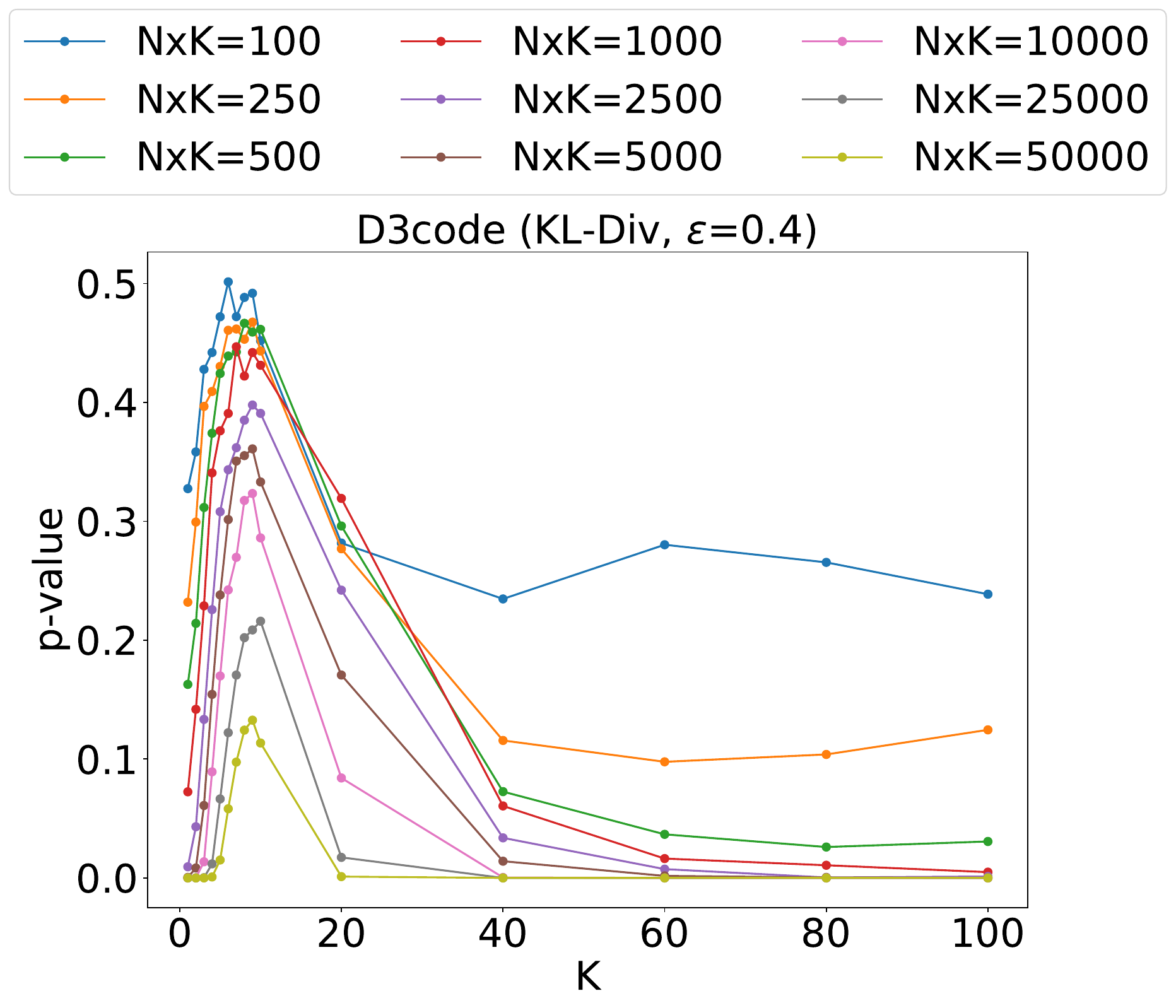}
    \caption{$\epsilon = 0.4$}
    \label{fig:d3code_kl_e04}
  \end{subfigure}
  \caption{P-value plots for D3code dataset with KL-divergence as the metric}
  \label{fig:d3code_kl}
\end{figure*}

\begin{figure*}
  \centering
  \begin{subfigure}[b]{0.24\linewidth}
    \centering
    \includegraphics[width=\linewidth]{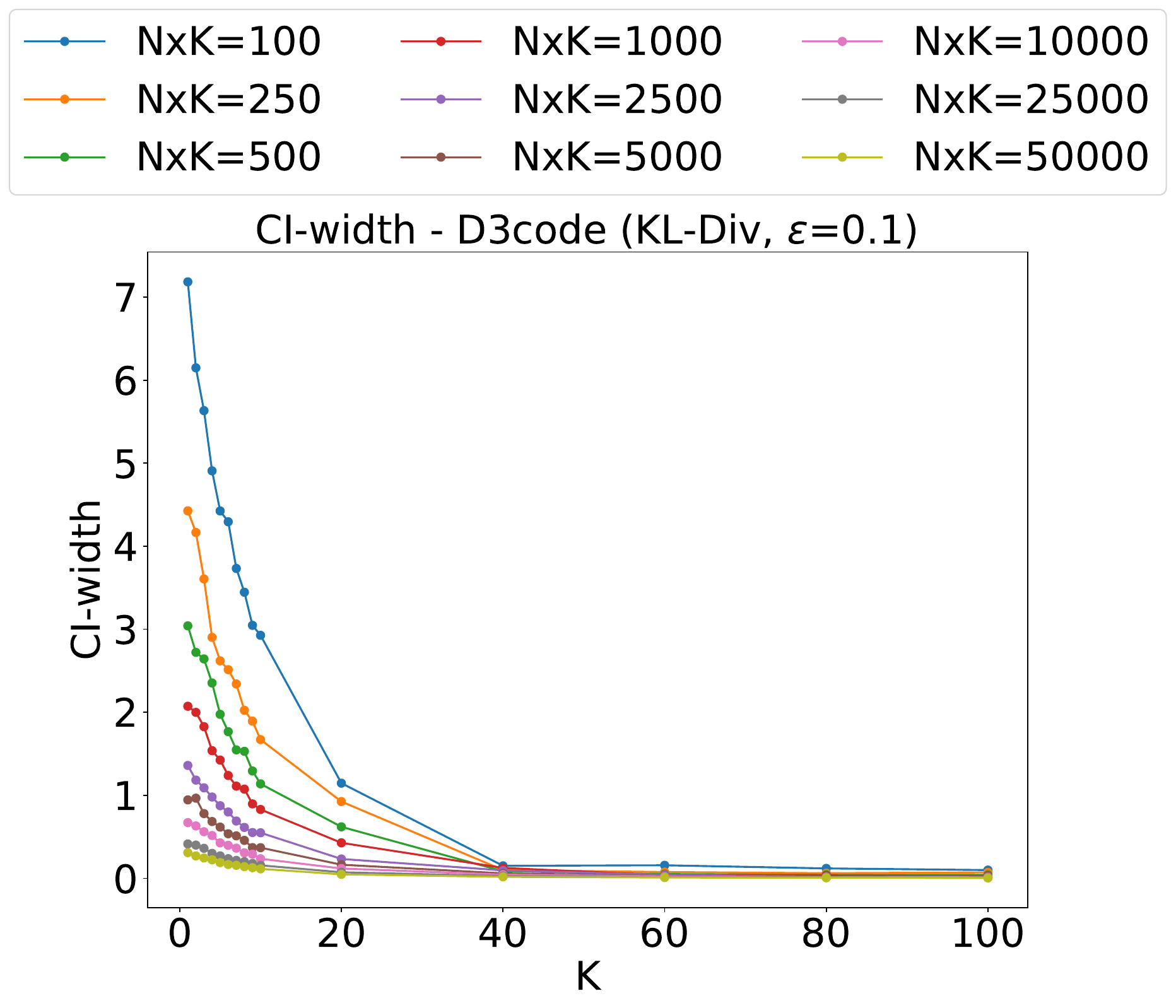}
    \caption{$\epsilon = 0.1$}
    \label{fig:d3code_ci_kl_e01}
  \end{subfigure} \hfill
  \begin{subfigure}[b]{0.24\linewidth}
    \centering
    \includegraphics[width=\linewidth]{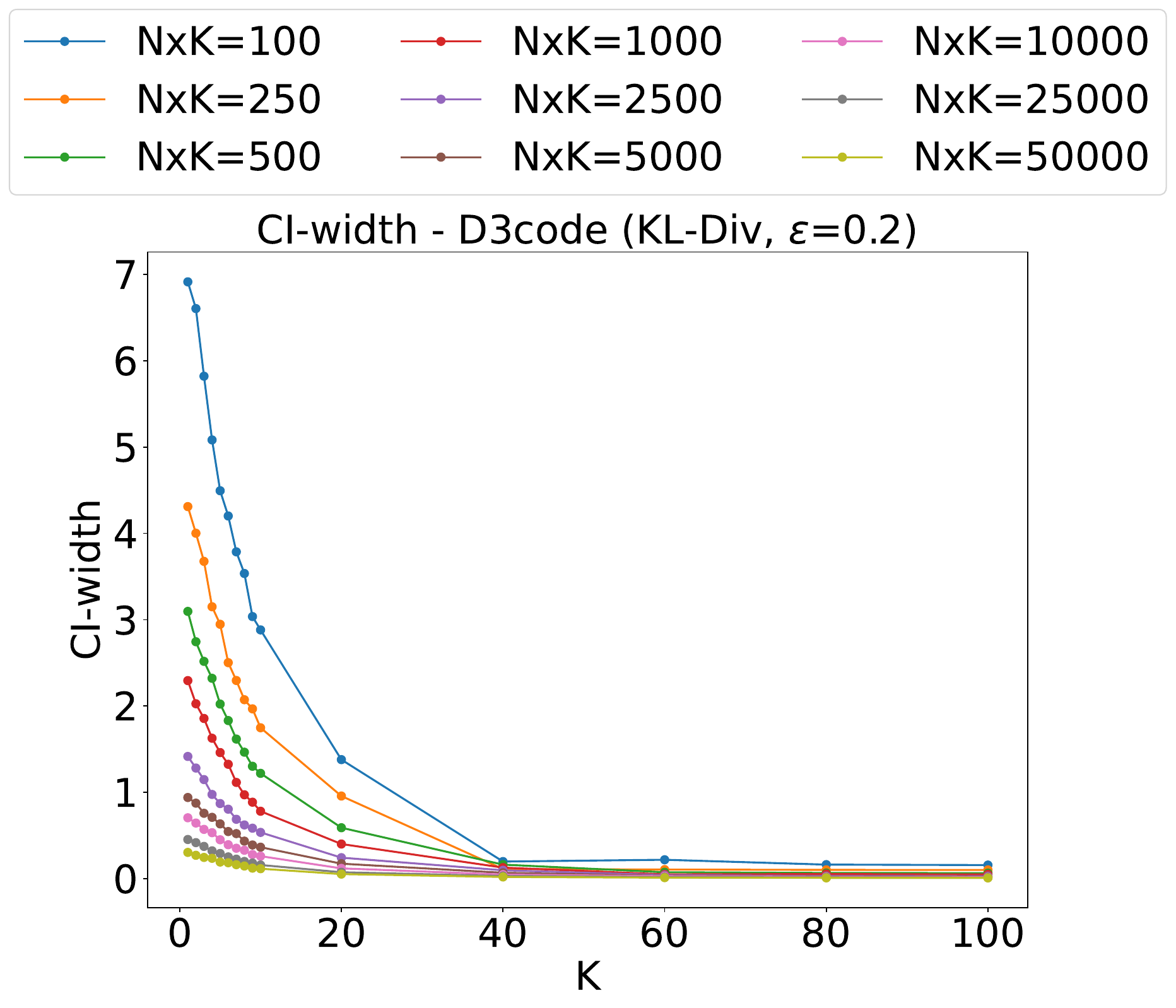}
    \caption{$\epsilon = 0.2$}
    \label{fig:d3code_ci_kl_e02}
  \end{subfigure} \hfill
  \begin{subfigure}[b]{0.24\linewidth}
    \centering
    \includegraphics[width=\linewidth]{figures/K100/ci_plots/D3code/D3code_CI_width_KL-Div_K_100_e_0.3.pdf}
    \caption{$\epsilon = 0.3$}
    \label{fig:d3code_ci_kl_e03}
  \end{subfigure} \hfill
  \begin{subfigure}[b]{0.24\linewidth}
    \centering
    \includegraphics[width=\linewidth]{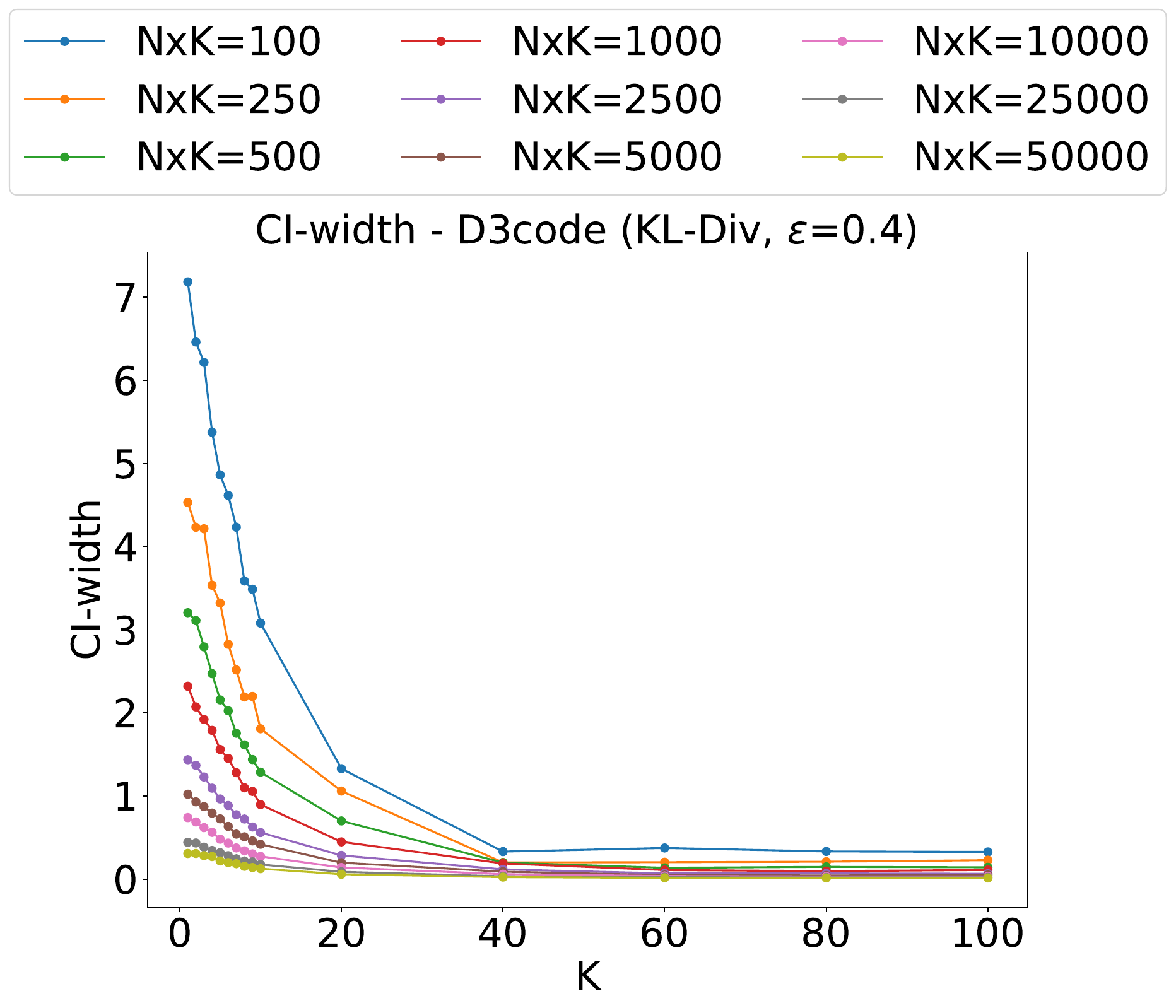}
    \caption{$\epsilon = 0.4$}
    \label{fig:d3code_ci_kl_e04}
  \end{subfigure}
  \caption{CI-width plots for D3code dataset with KL-divergence as the metric}
  \label{fig:d3code_ci_kl}
\end{figure*}

\begin{figure*}
  \centering
  \begin{subfigure}[b]{0.24\linewidth}
    \centering
    \includegraphics[width=\linewidth]{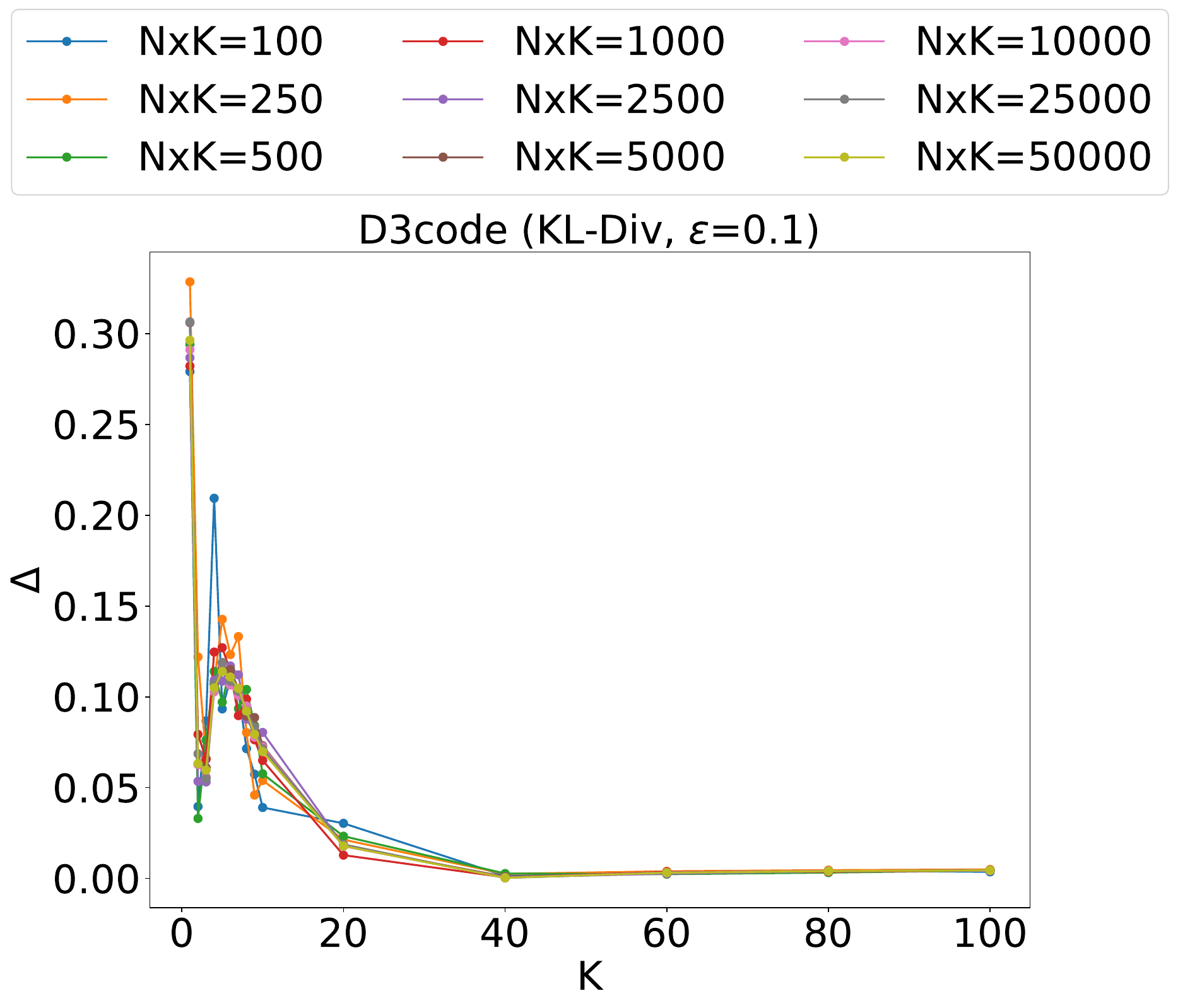}
    \caption{$\epsilon = 0.1$}
    \label{fig:d3code_delta_kl_e01}
  \end{subfigure} \hfill
  \begin{subfigure}[b]{0.24\linewidth}
    \centering
    \includegraphics[width=\linewidth]{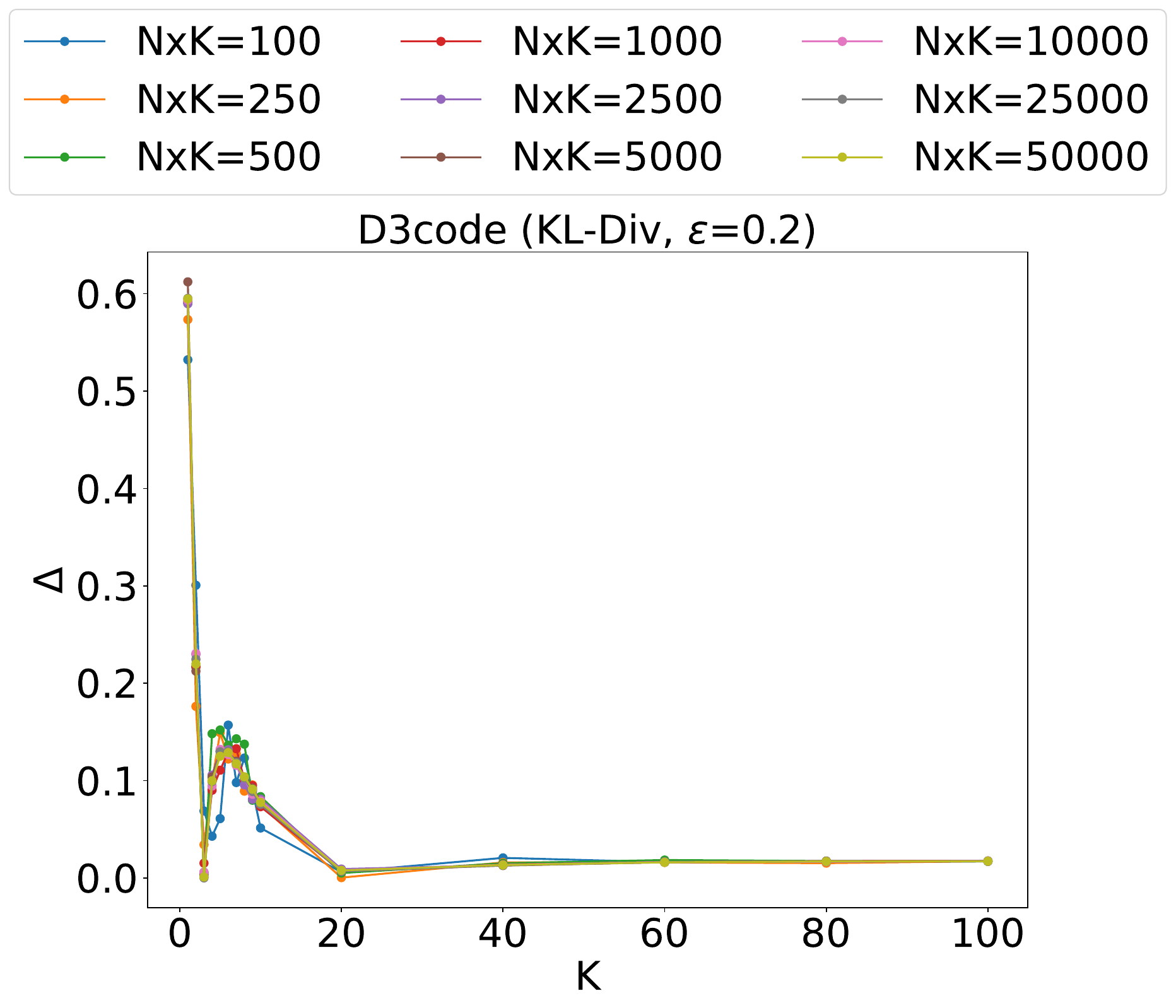}
    \caption{$\epsilon = 0.2$}
    \label{fig:d3code_delta_kl_e02}
  \end{subfigure} \hfill
  \begin{subfigure}[b]{0.24\linewidth}
    \centering
    \includegraphics[width=\linewidth]{figures/K100/delta_plots/D3code/D3code_delta_KL-Div_K_100_e_0.3.pdf}
    \caption{$\epsilon = 0.3$}
    \label{fig:d3code_delta_kl_e03}
  \end{subfigure} \hfill
  \begin{subfigure}[b]{0.24\linewidth}
    \centering
    \includegraphics[width=\linewidth]{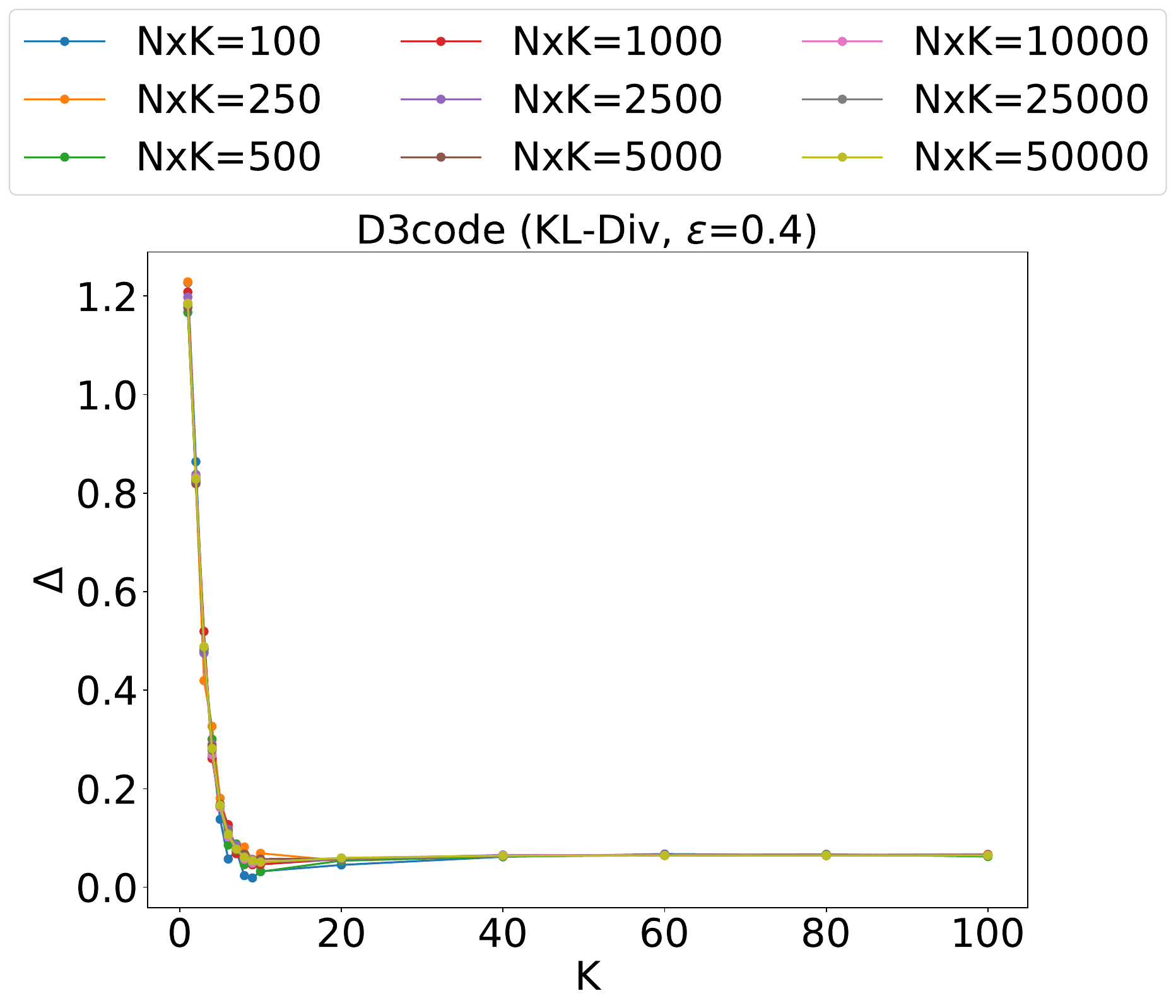}
    \caption{$\epsilon = 0.4$}
    \label{fig:d3code_delta_kl_e04}
  \end{subfigure}
  \caption{Effect sizes ($\Delta$) for D3code dataset with KL-divergence as the metric}
  \label{fig:d3code_delta_kl}
\end{figure*}

\begin{figure*}
  \centering
  \begin{subfigure}[b]{0.24\linewidth}
    \centering
    \includegraphics[width=\linewidth]{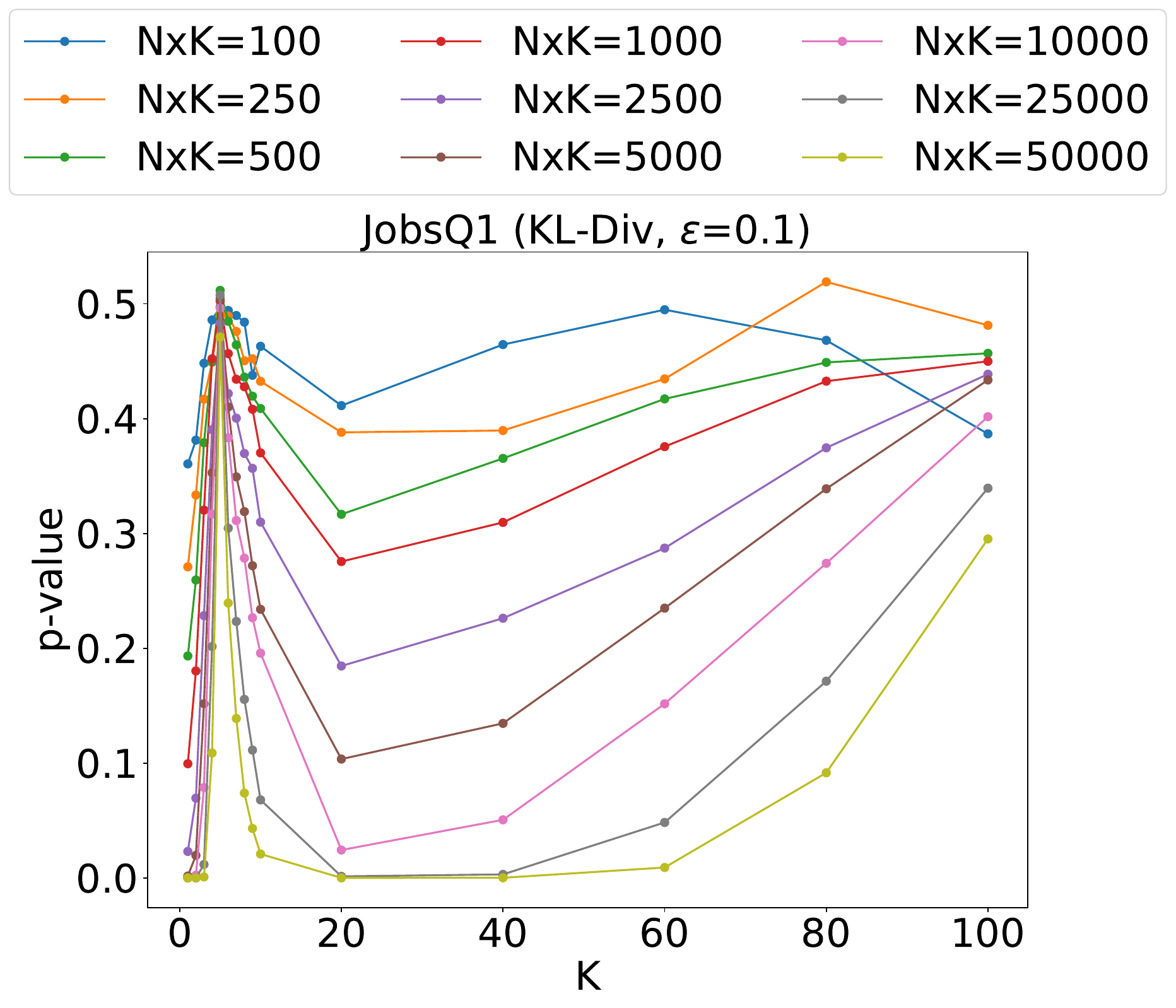}
    \caption{$\epsilon = 0.1$}
    \label{fig:jobsQ1_kl_e01}
  \end{subfigure} \hfill
  \begin{subfigure}[b]{0.24\linewidth}
    \centering
    \includegraphics[width=\linewidth]{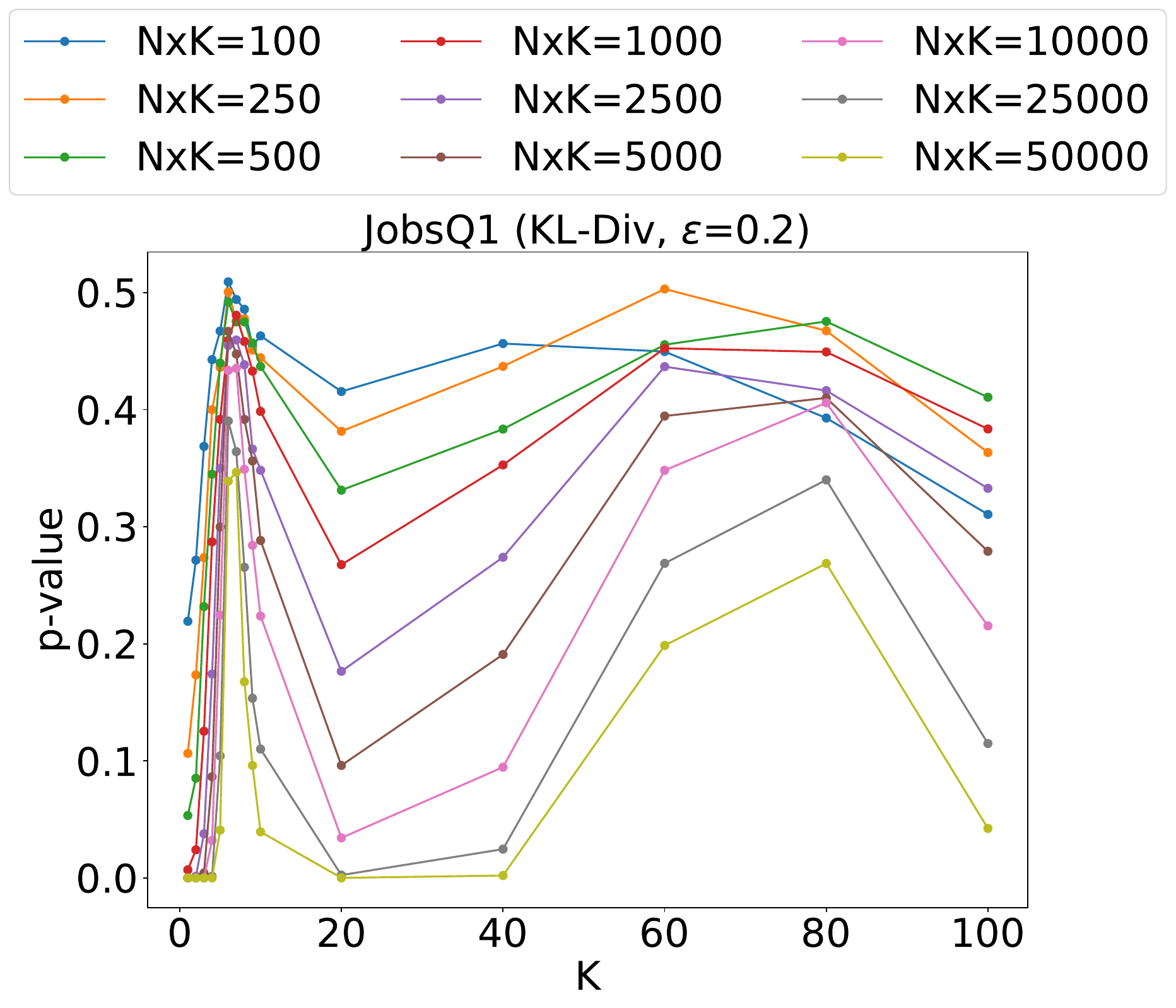}
    \caption{$\epsilon = 0.2$}
    \label{fig:jobsQ1_kl_e02}
  \end{subfigure} \hfill
  \begin{subfigure}[b]{0.24\linewidth}
    \centering
    \includegraphics[width=\linewidth]{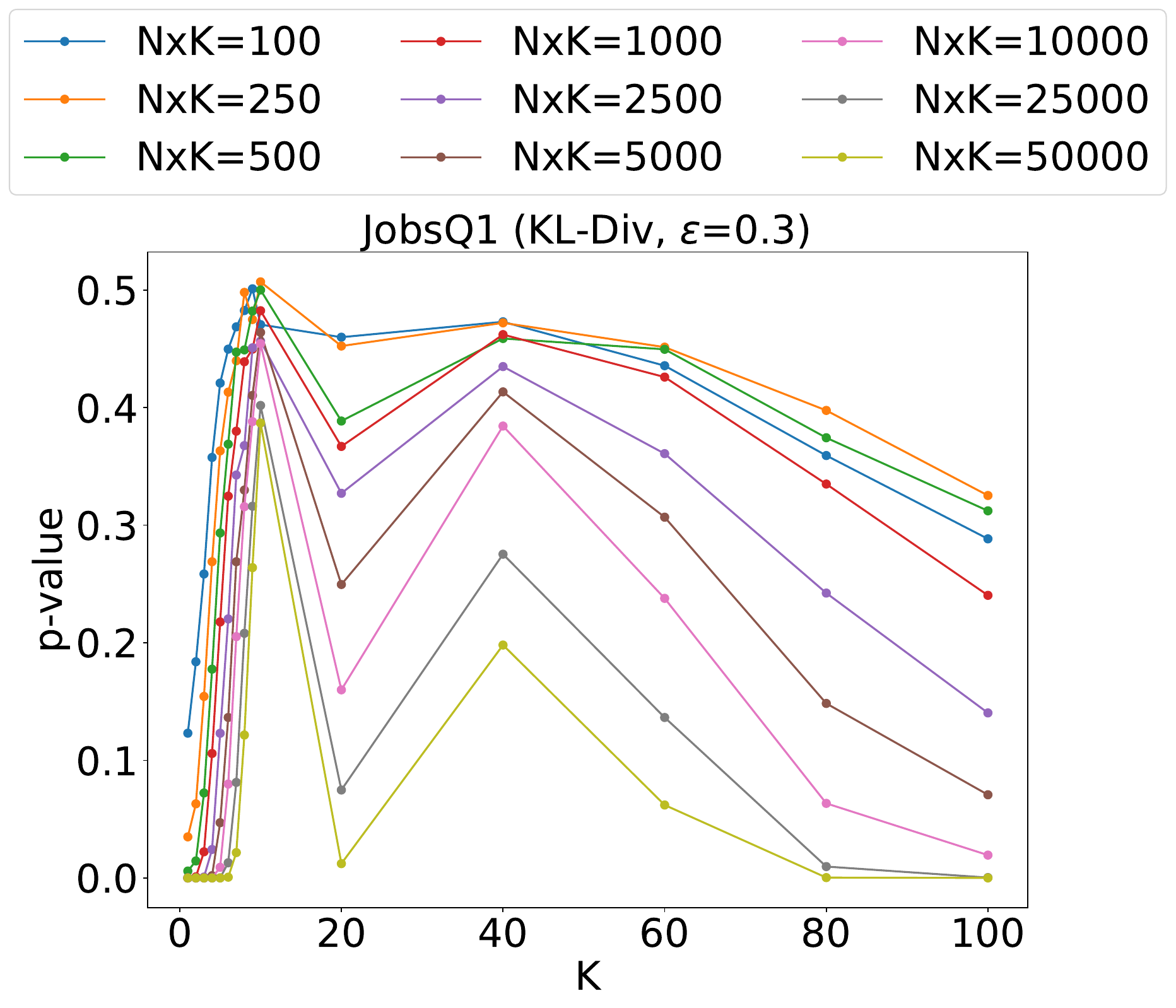}
    \caption{$\epsilon = 0.3$}
    \label{fig:jobsQ1_kl_e03}
  \end{subfigure} \hfill
  \begin{subfigure}[b]{0.24\linewidth}
    \centering
    \includegraphics[width=\linewidth]{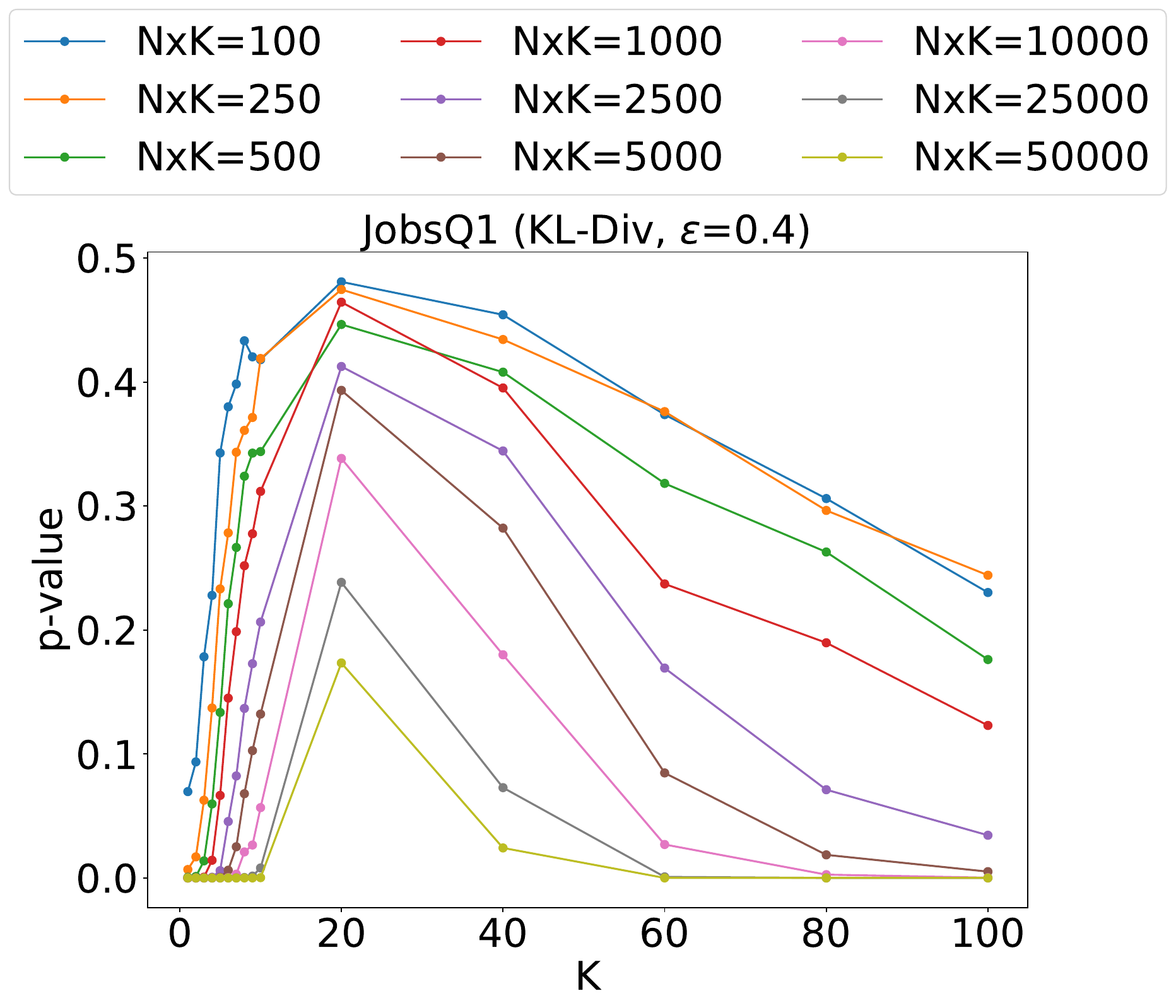}
    \caption{$\epsilon = 0.4$}
    \label{fig:jobsQ1_kl_e04}
  \end{subfigure}
  \caption{P-value plots for JobsQ1 dataset with KL-divergence as the metric}
  \label{fig:jobsQ1_kl}
\end{figure*}

\begin{figure*}
  \centering
  \begin{subfigure}[b]{0.24\linewidth}
    \centering
    \includegraphics[width=\linewidth]{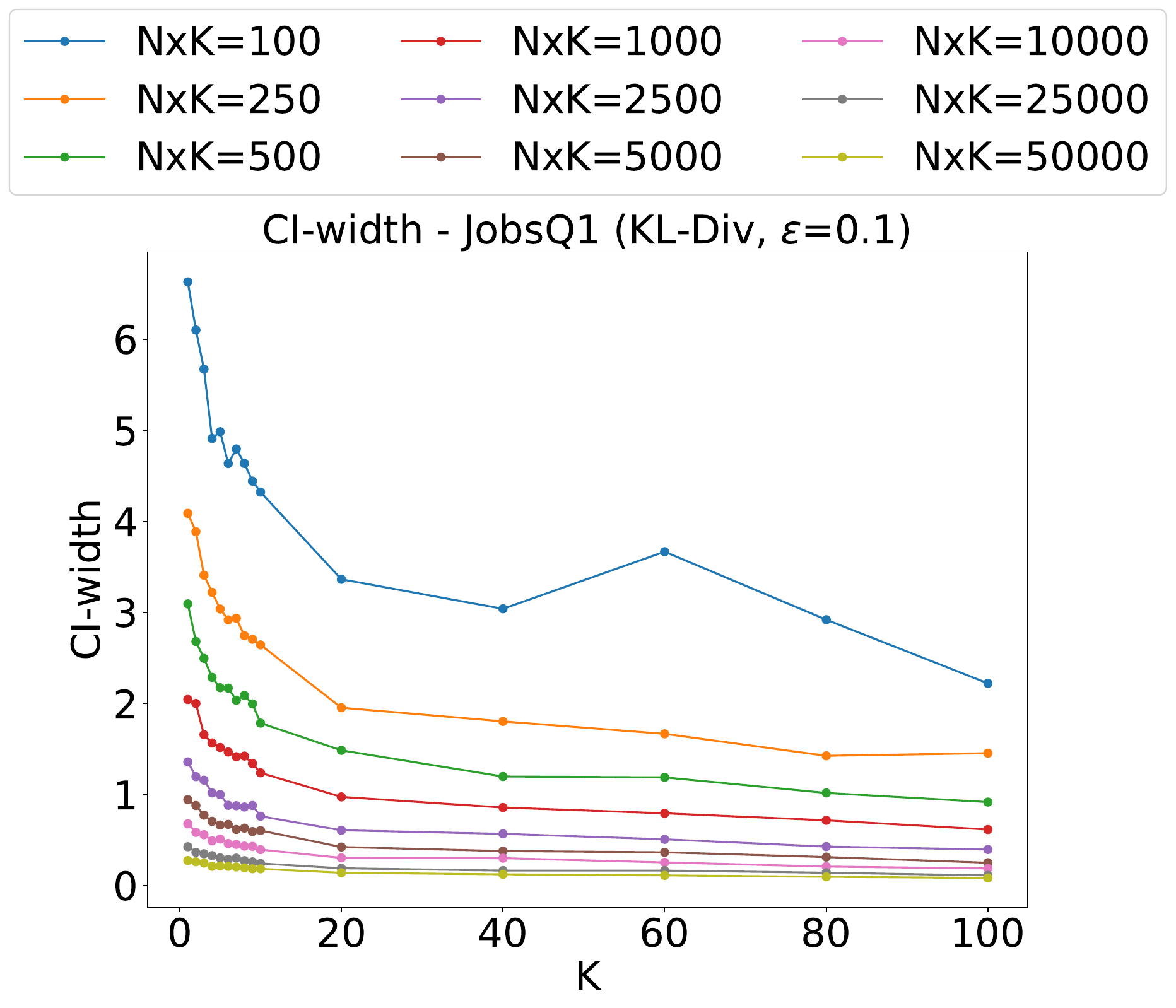}
    \caption{$\epsilon = 0.1$}
    \label{fig:jobsQ1_ci_kl_e01}
  \end{subfigure} \hfill
  \begin{subfigure}[b]{0.24\linewidth}
    \centering
    \includegraphics[width=\linewidth]{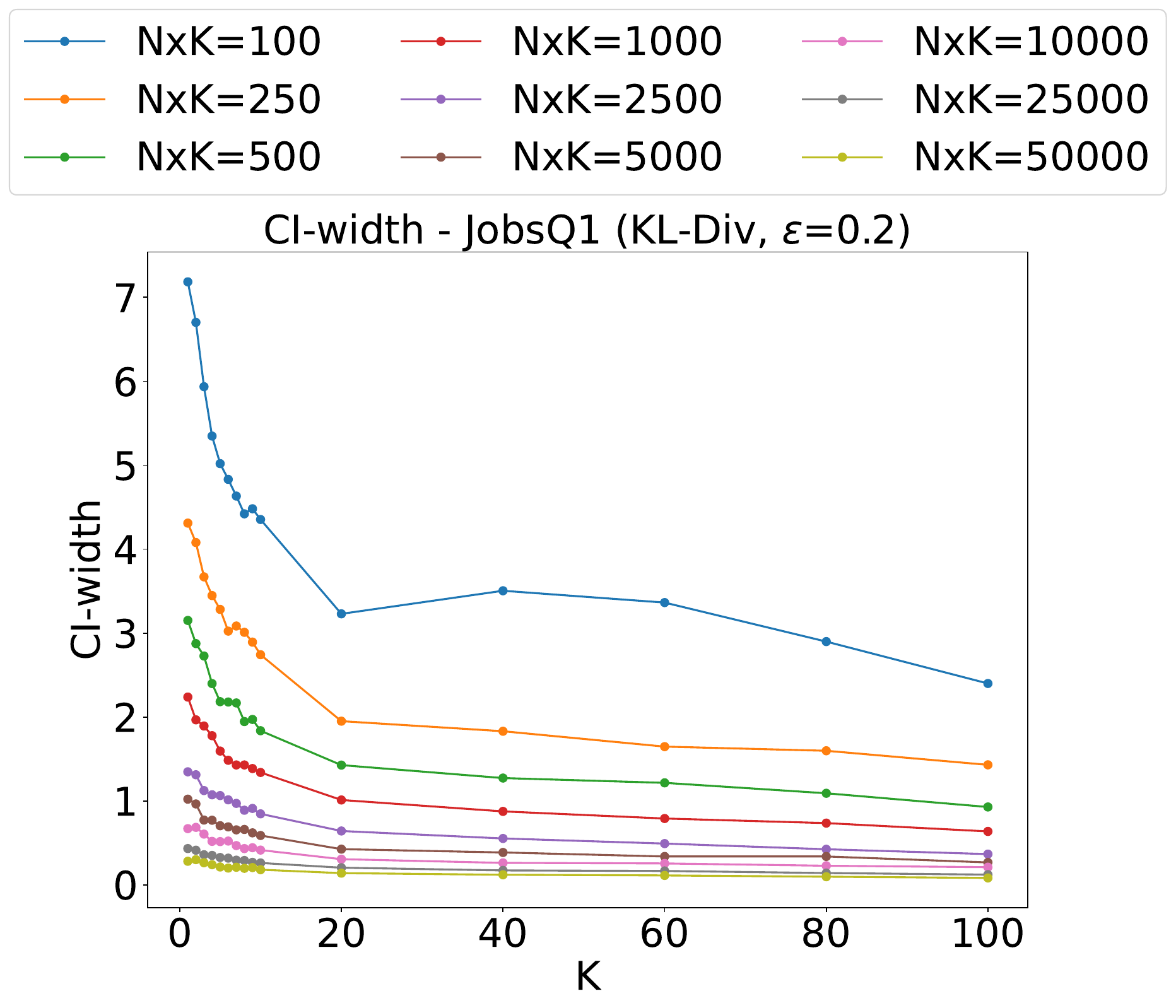}
    \caption{$\epsilon = 0.2$}
    \label{fig:jobsQ1_ci_kl_e02}
  \end{subfigure} \hfill
  \begin{subfigure}[b]{0.24\linewidth}
    \centering
    \includegraphics[width=\linewidth]{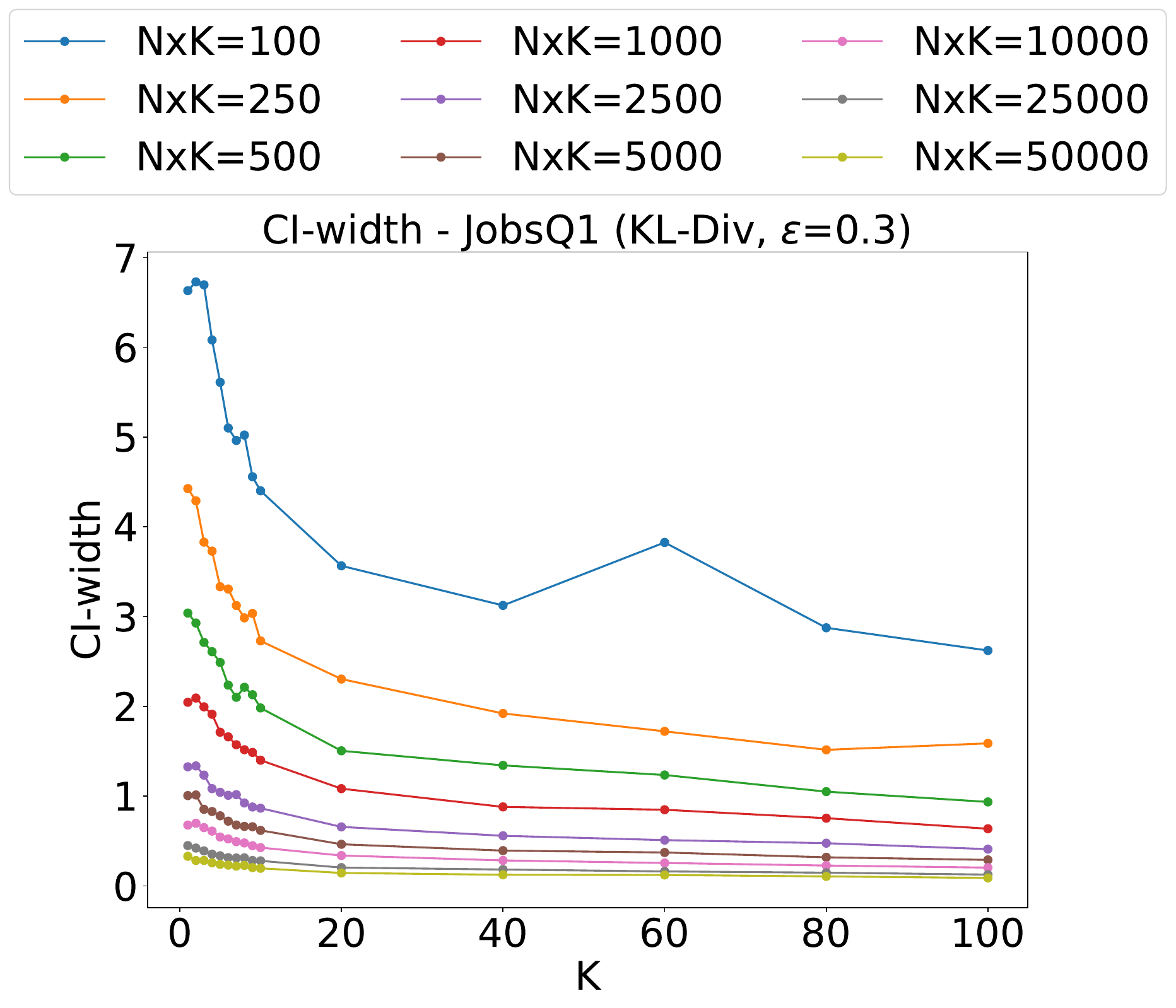}
    \caption{$\epsilon = 0.3$}
    \label{fig:jobsQ1_ci_kl_e03}
  \end{subfigure} \hfill
  \begin{subfigure}[b]{0.24\linewidth}
    \centering
    \includegraphics[width=\linewidth]{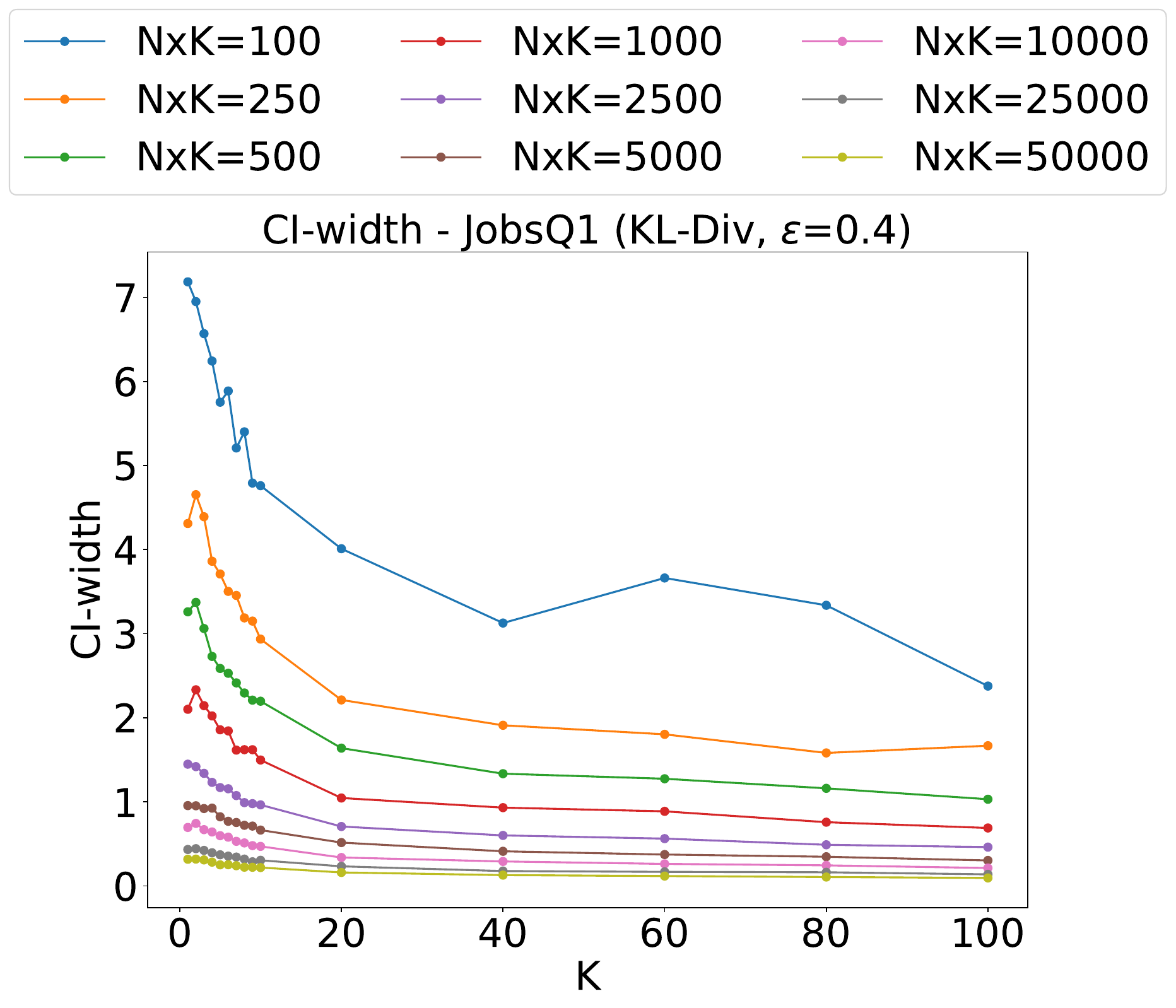}
    \caption{$\epsilon = 0.4$}
    \label{fig:jobsQ1_ci_kl_e04}
  \end{subfigure}
  \caption{CI-width plots for JobsQ1 dataset with KL-divergence as the metric}
  \label{fig:jobsQ1_ci_kl}
\end{figure*}

\begin{figure*}
  \centering
  \begin{subfigure}[b]{0.24\linewidth}
    \centering
    \includegraphics[width=\linewidth]{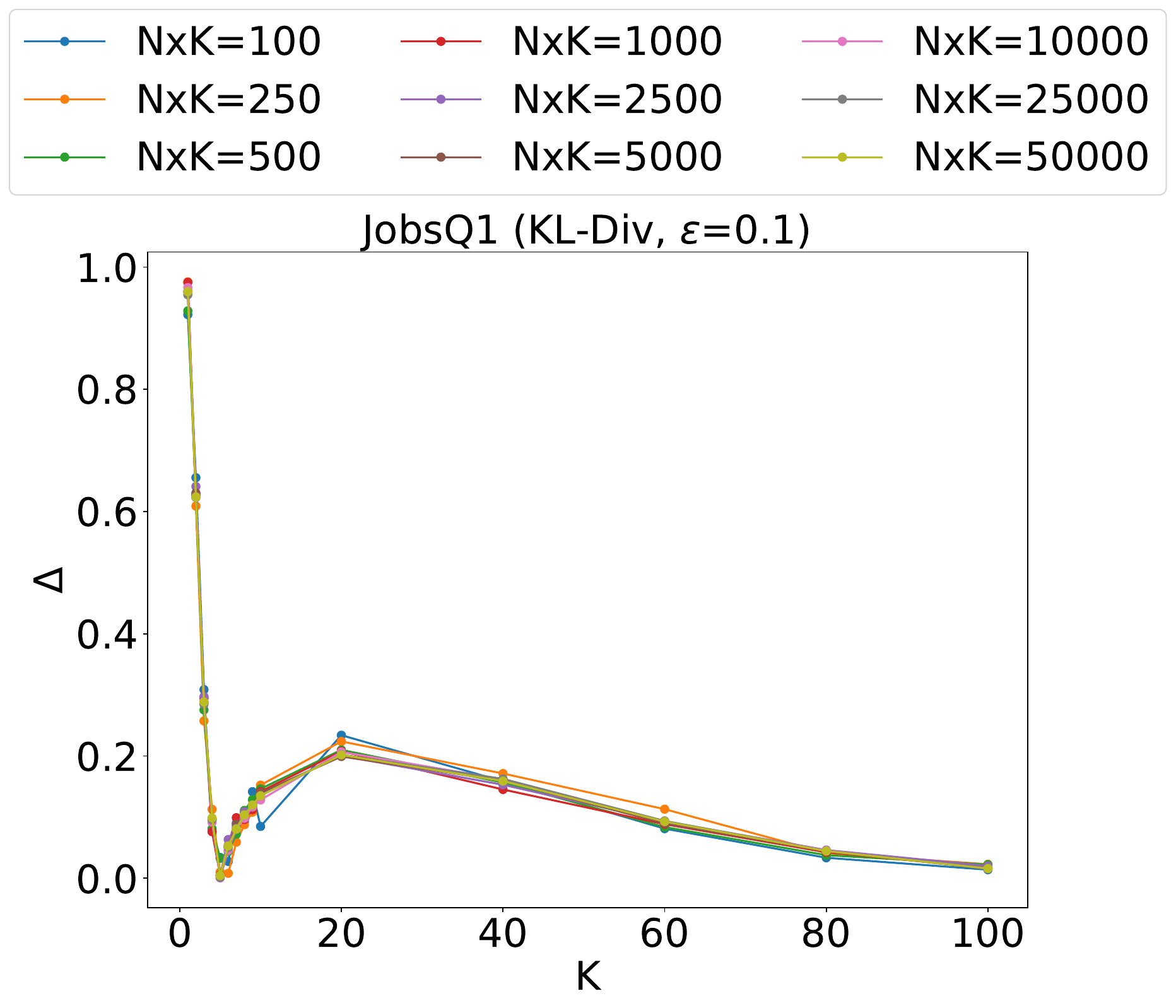}
    \caption{$\epsilon = 0.1$}
    \label{fig:jobsQ1_delta_kl_e01}
  \end{subfigure} \hfill
  \begin{subfigure}[b]{0.24\linewidth}
    \centering
    \includegraphics[width=\linewidth]{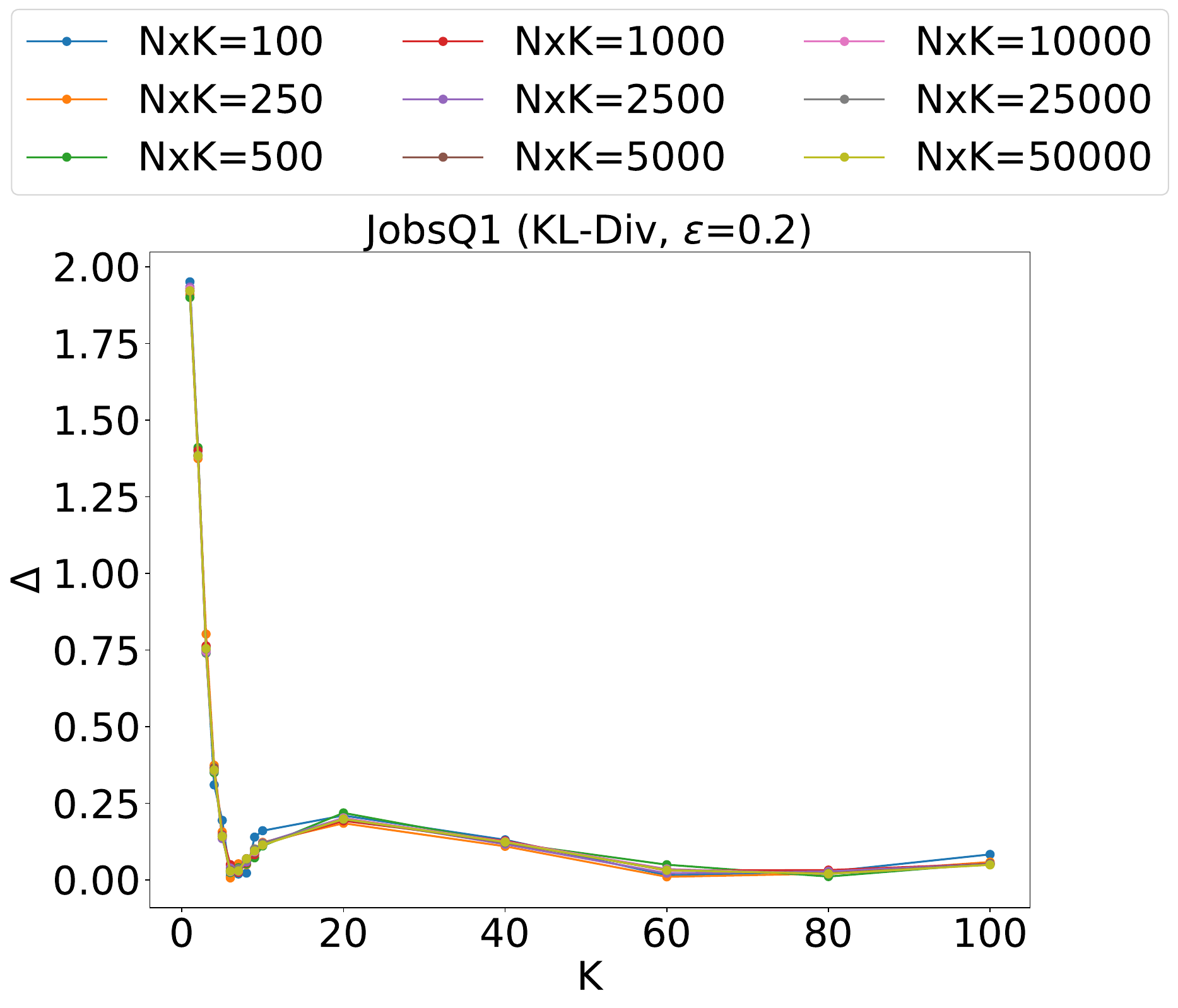}
    \caption{$\epsilon = 0.2$}
    \label{fig:jobsQ1_delta_kl_e02}
  \end{subfigure} \hfill
  \begin{subfigure}[b]{0.24\linewidth}
    \centering
    \includegraphics[width=\linewidth]{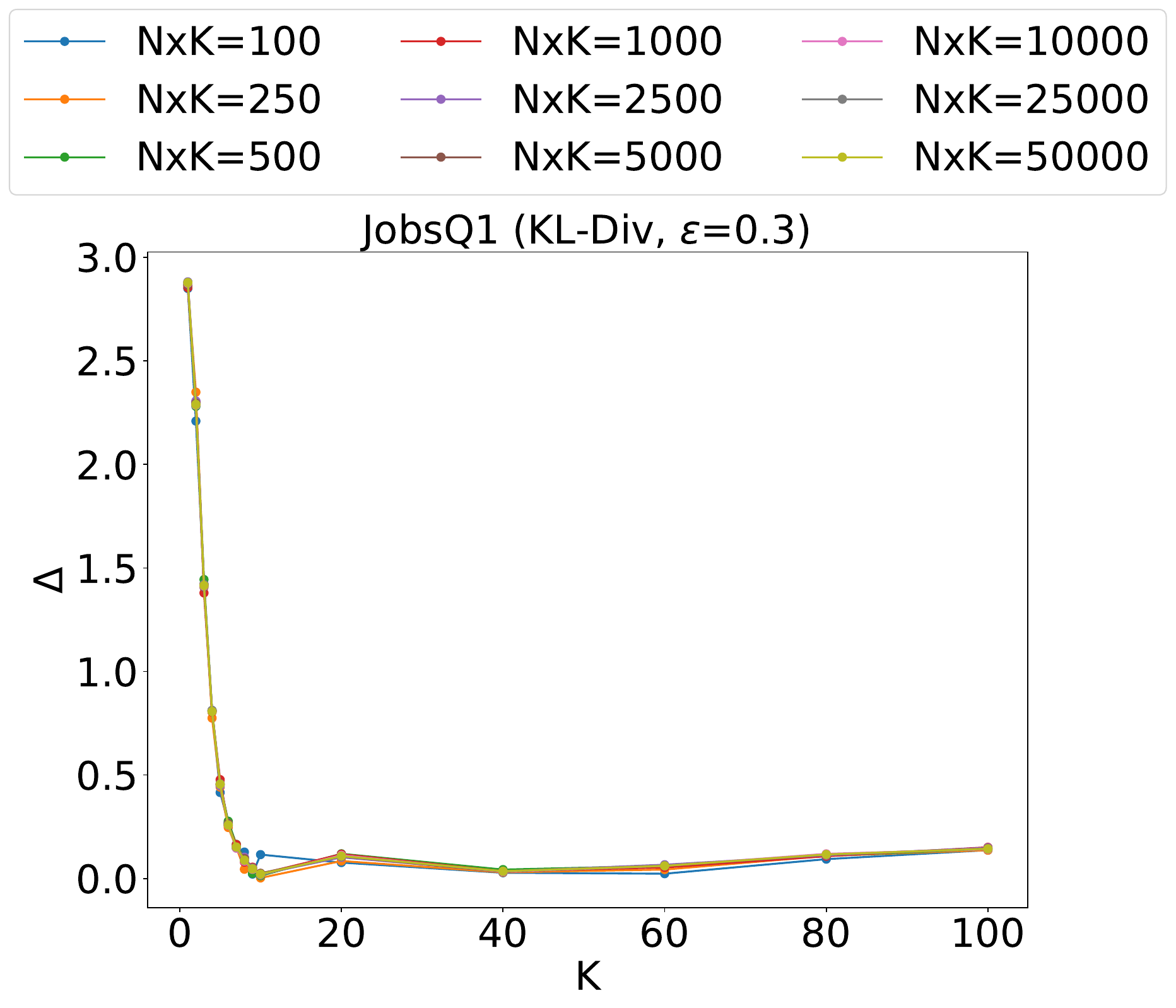}
    \caption{$\epsilon = 0.3$}
    \label{fig:jobsQ1_delta_kl_e03}
  \end{subfigure} \hfill
  \begin{subfigure}[b]{0.24\linewidth}
    \centering
    \includegraphics[width=\linewidth]{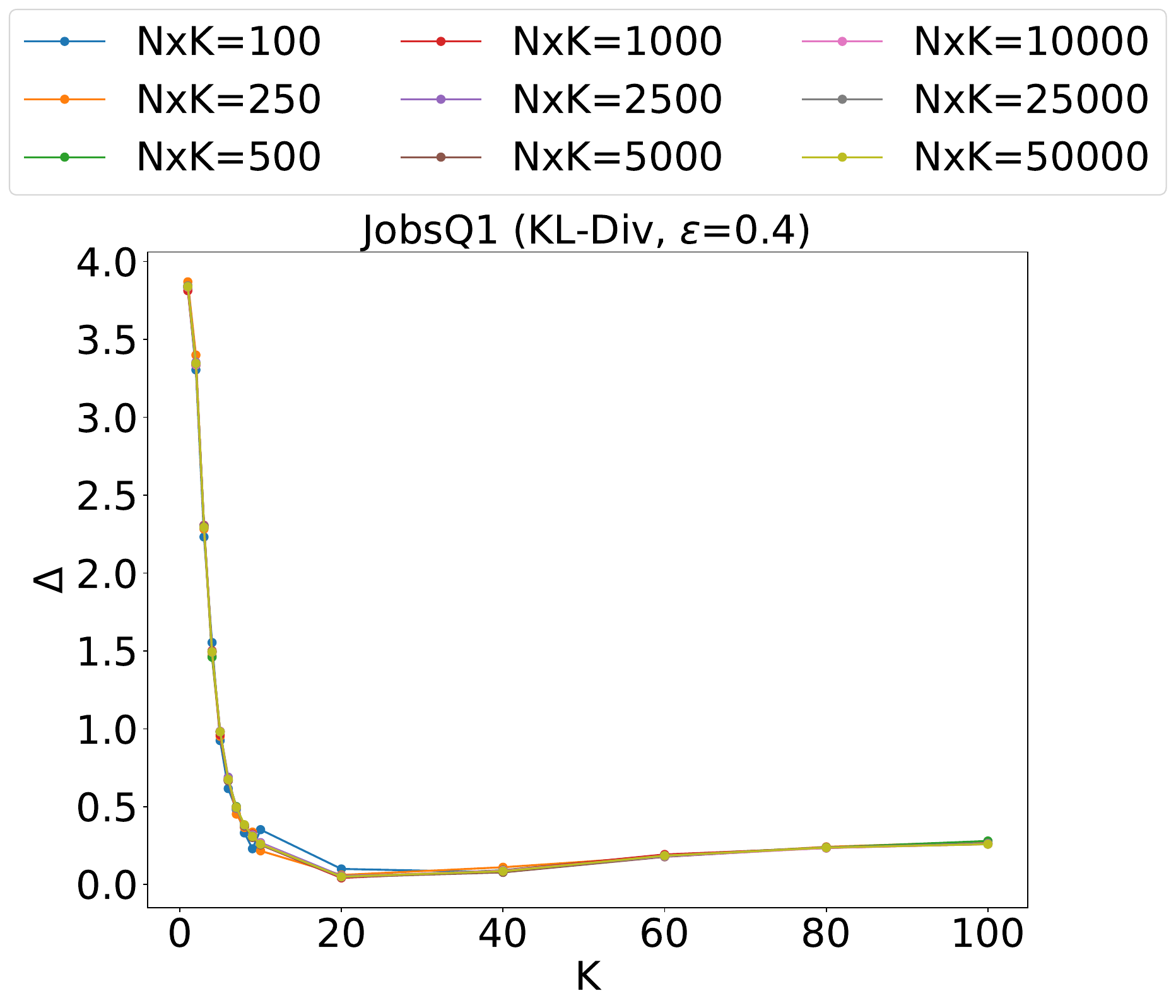}
    \caption{$\epsilon = 0.4$}
    \label{fig:jobsQ1_delta_kl_e04}
  \end{subfigure}
  \caption{Effect sizes ($\Delta$) for JobsQ1 dataset with KL-divergence as the metric}
  \label{fig:jobsQ1_delta_kl}
\end{figure*}

\begin{figure*}
  \centering
  \begin{subfigure}[b]{0.24\linewidth}
    \centering
    \includegraphics[width=\linewidth]{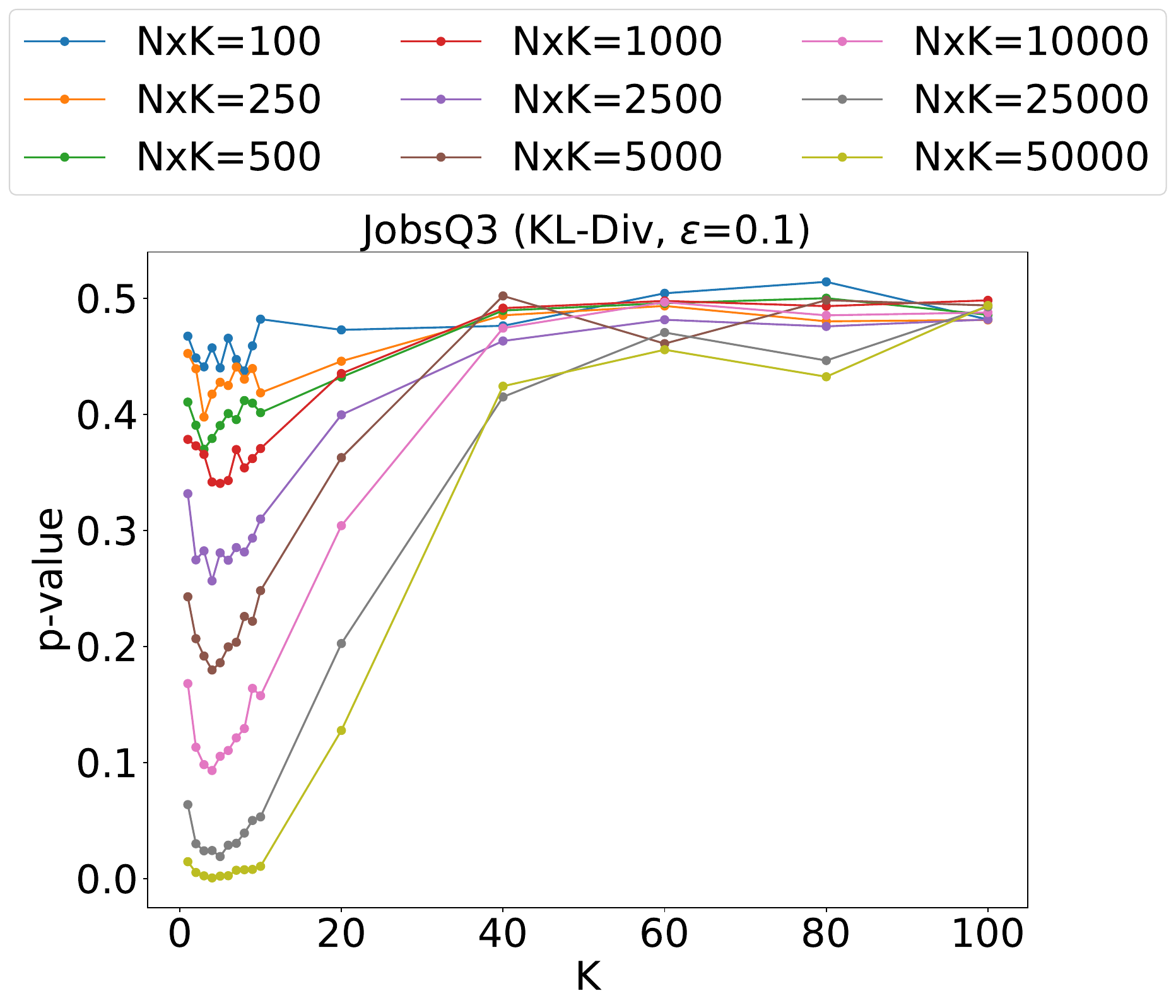}
    \caption{$\epsilon = 0.1$}
    \label{fig:jobsQ3_kl_e01}
  \end{subfigure} \hfill
  \begin{subfigure}[b]{0.24\linewidth}
    \centering
    \includegraphics[width=\linewidth]{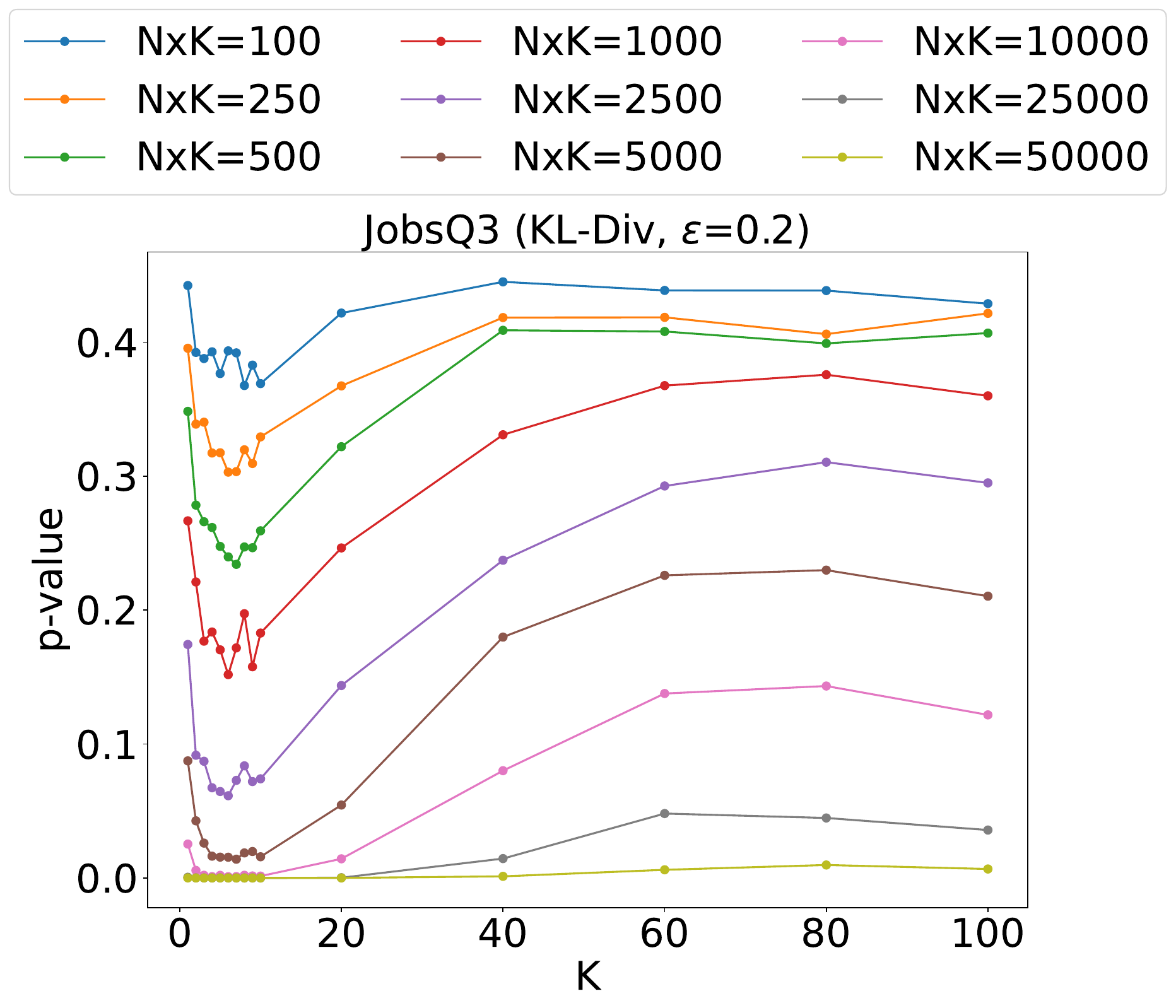}
    \caption{$\epsilon = 0.2$}
    \label{fig:jobsQ3_kl_e02}
  \end{subfigure} \hfill
  \begin{subfigure}[b]{0.24\linewidth}
    \centering
    \includegraphics[width=\linewidth]{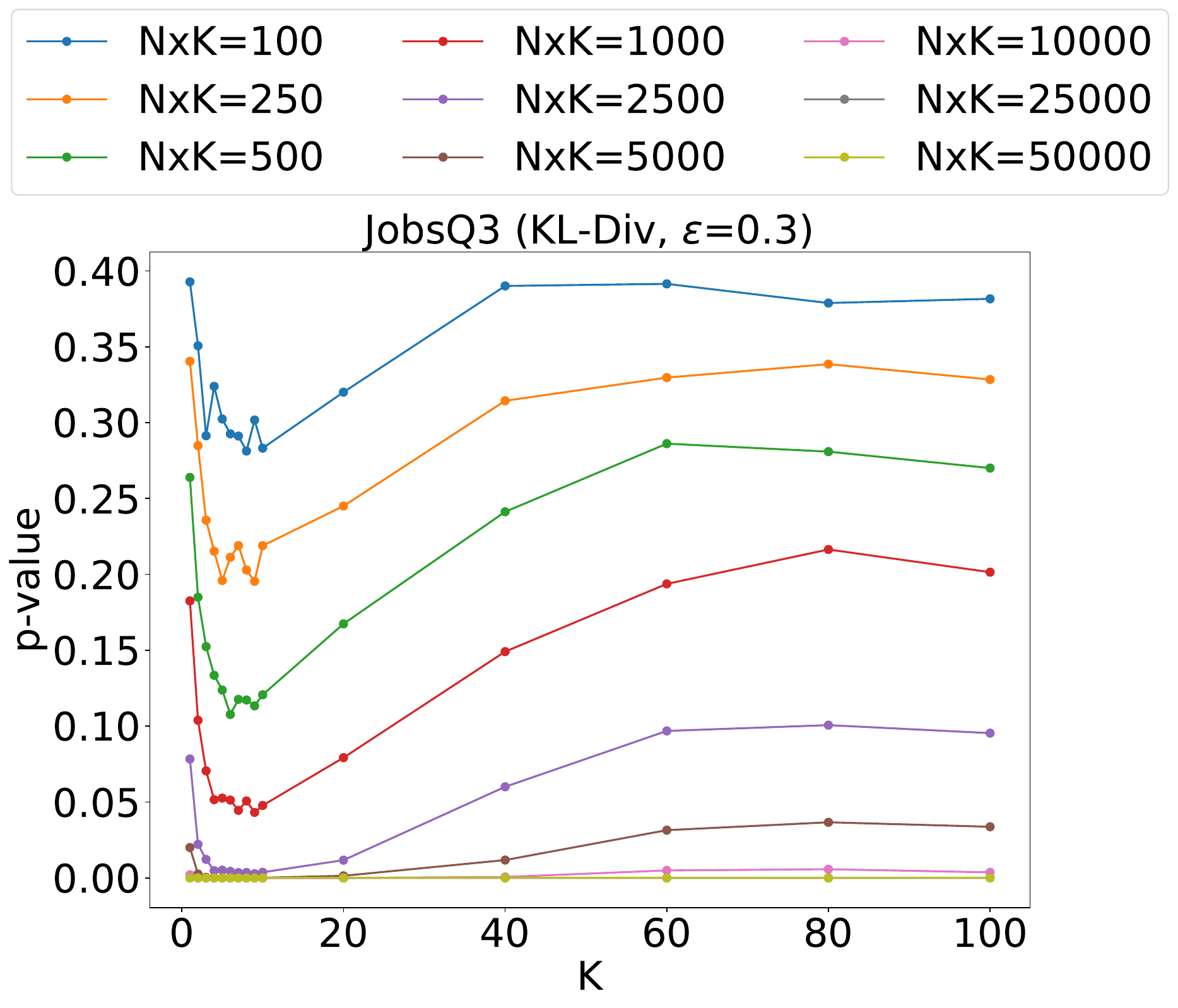}
    \caption{$\epsilon = 0.3$}
    \label{fig:jobsQ3_kl_e03}
  \end{subfigure} \hfill
  \begin{subfigure}[b]{0.24\linewidth}
    \centering
    \includegraphics[width=\linewidth]{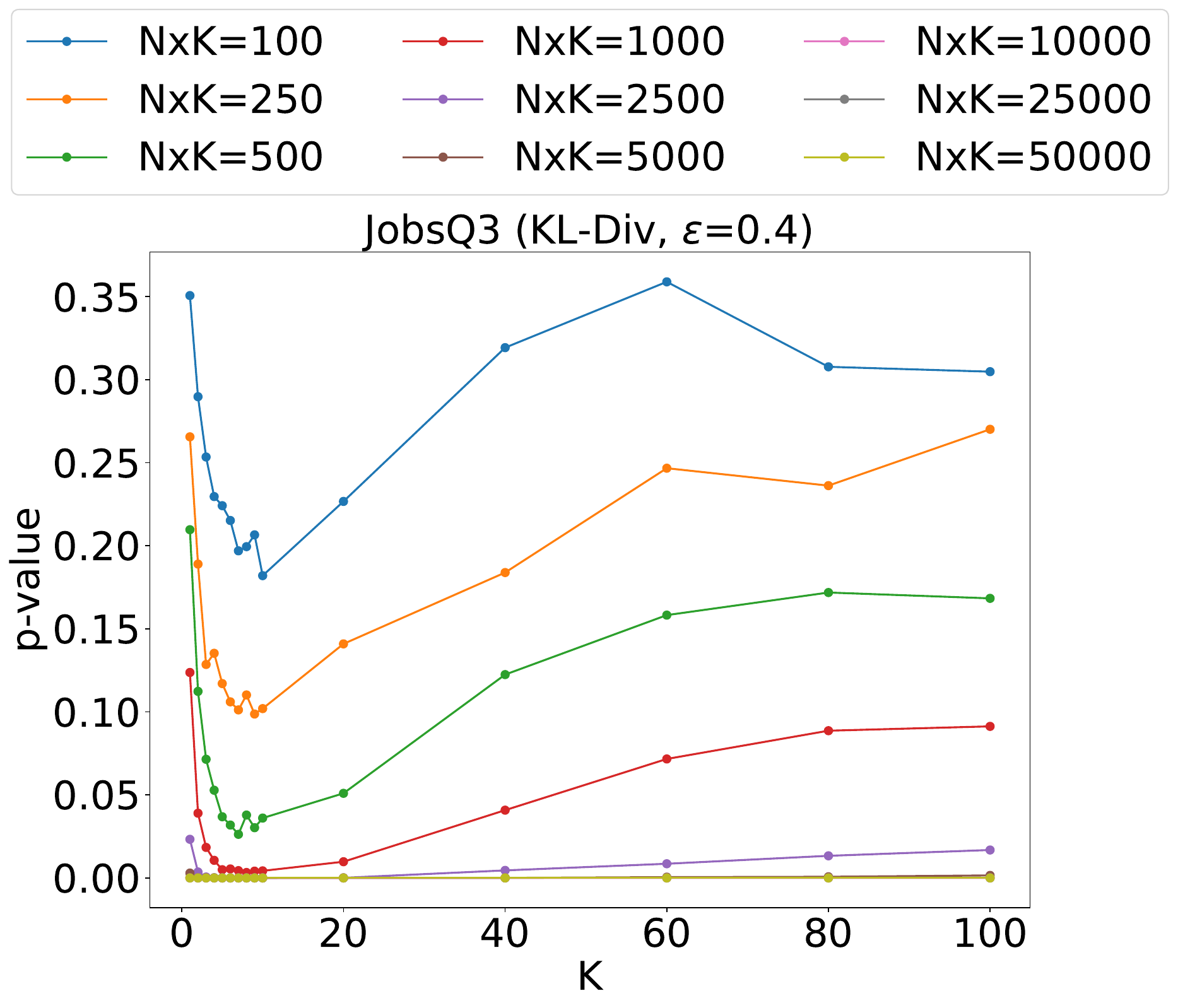}
    \caption{$\epsilon = 0.4$}
    \label{fig:jobsQ3_kl_e04}
  \end{subfigure}
  \caption{P-value plots for JobsQ3 dataset with KL-divergence as the metric}
  \label{fig:jobsQ3_kl}
\end{figure*}

\begin{figure*}
  \centering
  \begin{subfigure}[b]{0.24\linewidth}
    \centering
    \includegraphics[width=\linewidth]{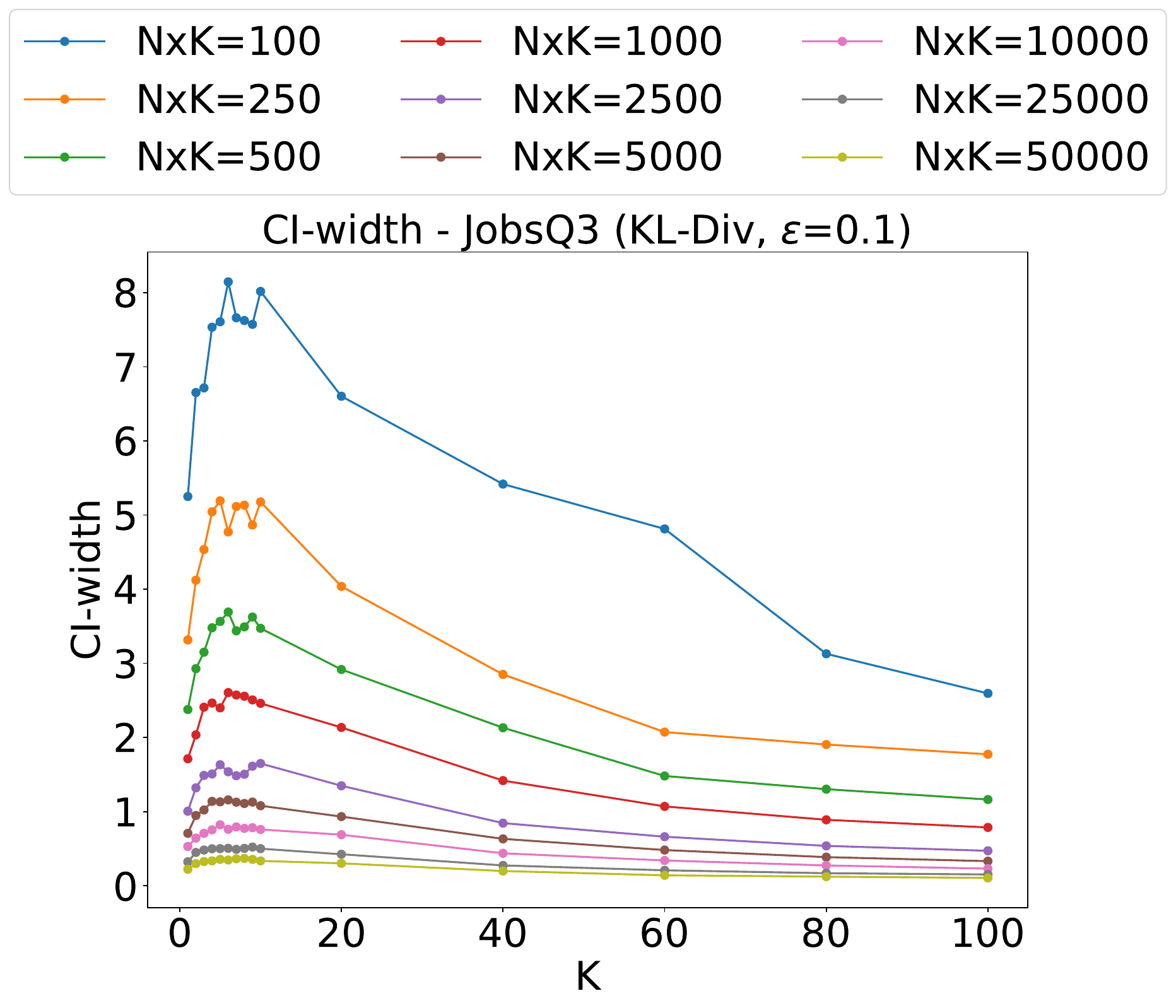}
    \caption{$\epsilon = 0.1$}
    \label{fig:jobsQ3_ci_kl_e01}
  \end{subfigure} \hfill
  \begin{subfigure}[b]{0.24\linewidth}
    \centering
    \includegraphics[width=\linewidth]{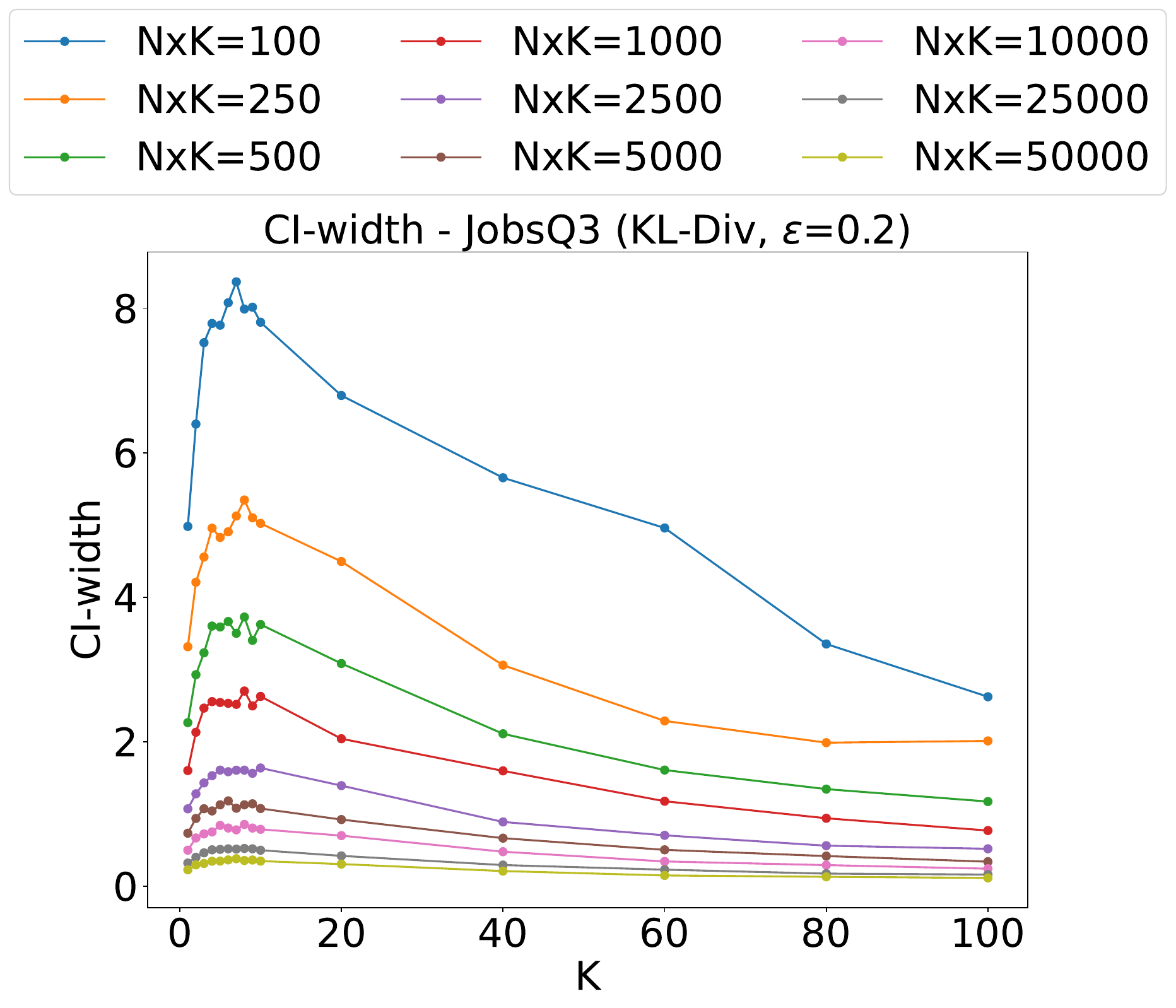}
    \caption{$\epsilon = 0.2$}
    \label{fig:jobsQ3_ci_kl_e02}
  \end{subfigure} \hfill
  \begin{subfigure}[b]{0.24\linewidth}
    \centering
    \includegraphics[width=\linewidth]{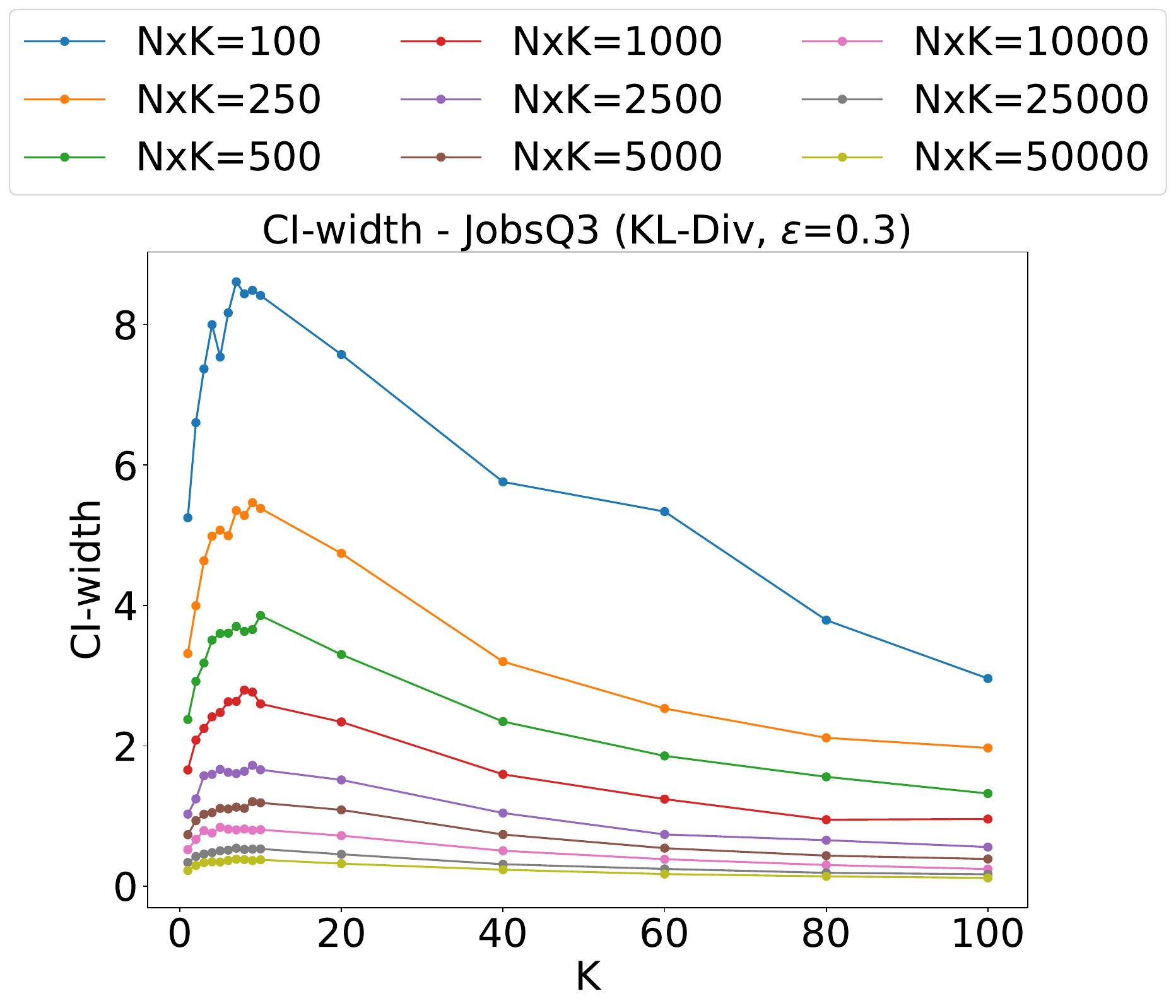}
    \caption{$\epsilon = 0.3$}
    \label{fig:jobsQ3_ci_kl_e03}
  \end{subfigure} \hfill
  \begin{subfigure}[b]{0.24\linewidth}
    \centering
    \includegraphics[width=\linewidth]{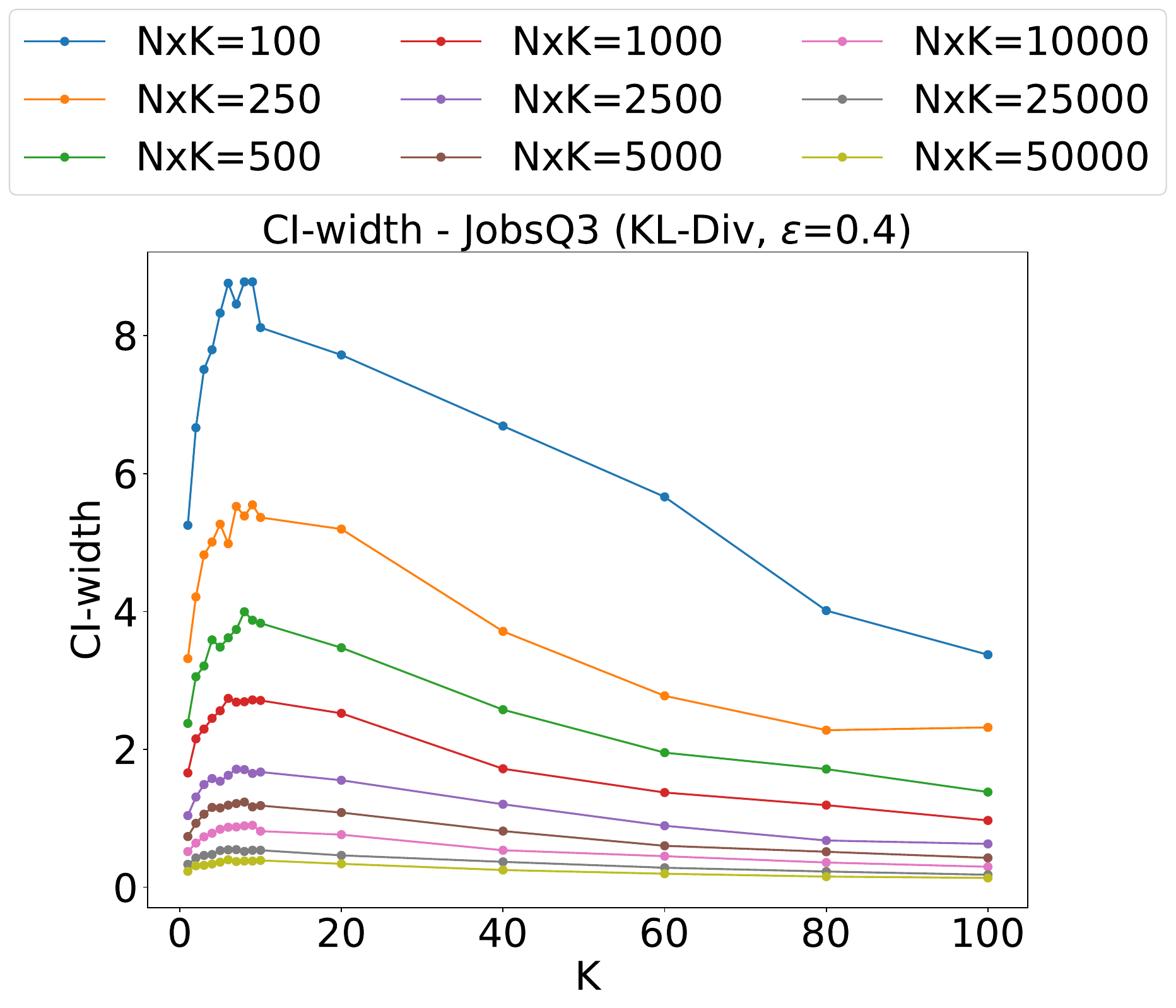}
    \caption{$\epsilon = 0.4$}
    \label{fig:jobsQ3_ci_kl_e04}
  \end{subfigure}
  \caption{CI-width plots for JobsQ3 dataset with KL-divergence as the metric}
  \label{fig:jobsQ3_ci_kl}
\end{figure*}

\begin{figure*}
  \centering
  \begin{subfigure}[b]{0.24\linewidth}
    \centering
    \includegraphics[width=\linewidth]{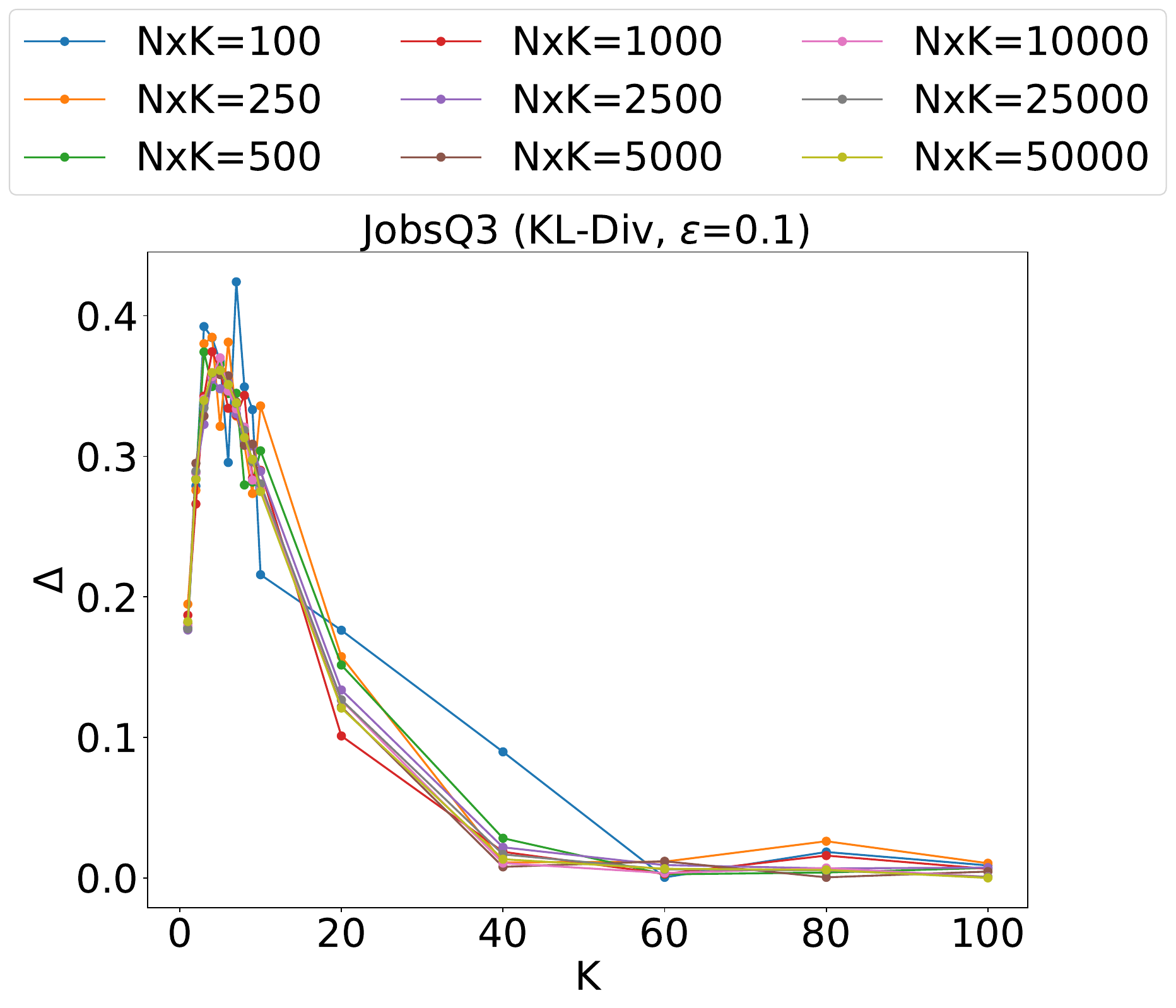}
    \caption{$\epsilon = 0.1$}
    \label{fig:jobsQ3_delta_kl_e01}
  \end{subfigure} \hfill
  \begin{subfigure}[b]{0.24\linewidth}
    \centering
    \includegraphics[width=\linewidth]{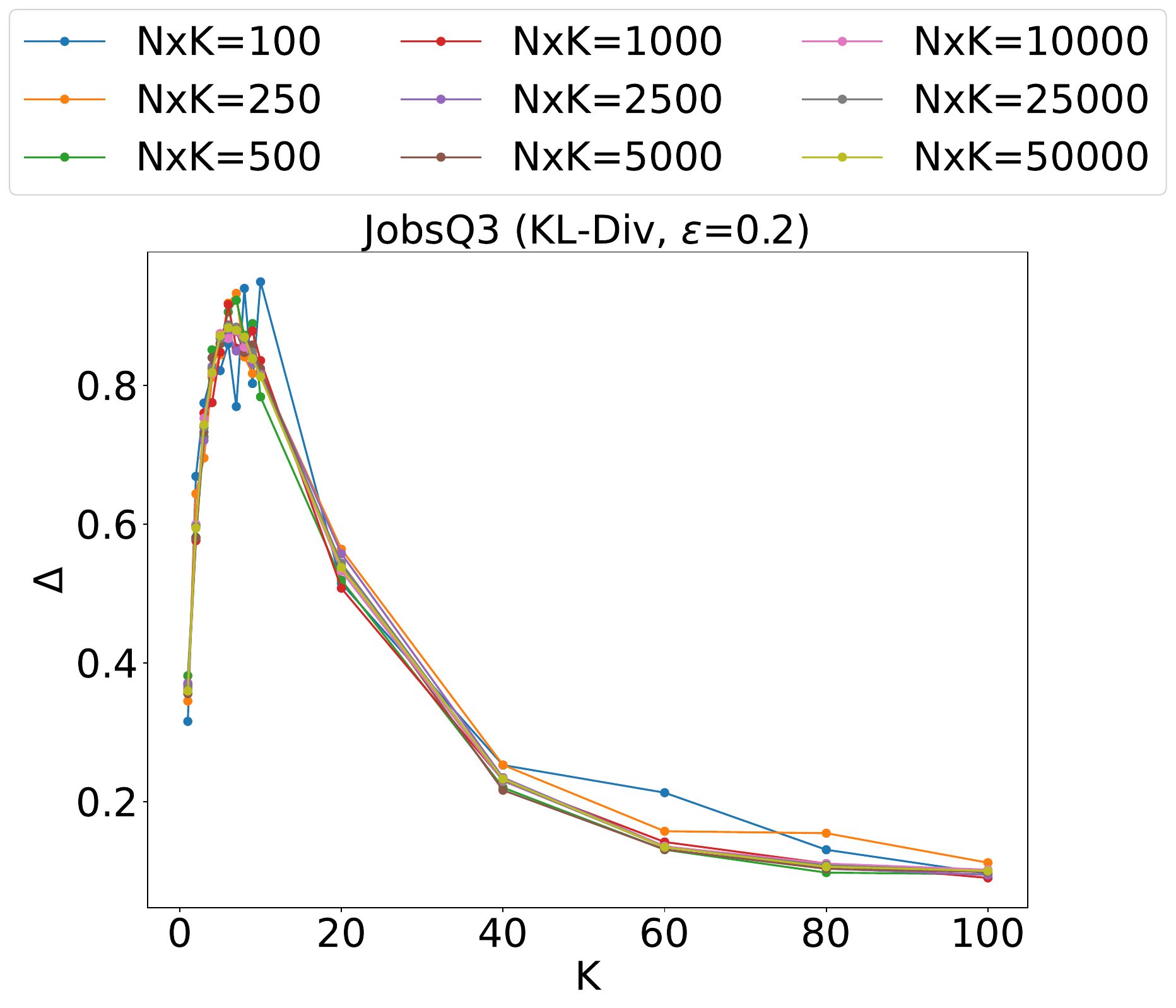}
    \caption{$\epsilon = 0.2$}
    \label{fig:jobsQ3_delta_kl_e02}
  \end{subfigure} \hfill
  \begin{subfigure}[b]{0.24\linewidth}
    \centering
    \includegraphics[width=\linewidth]{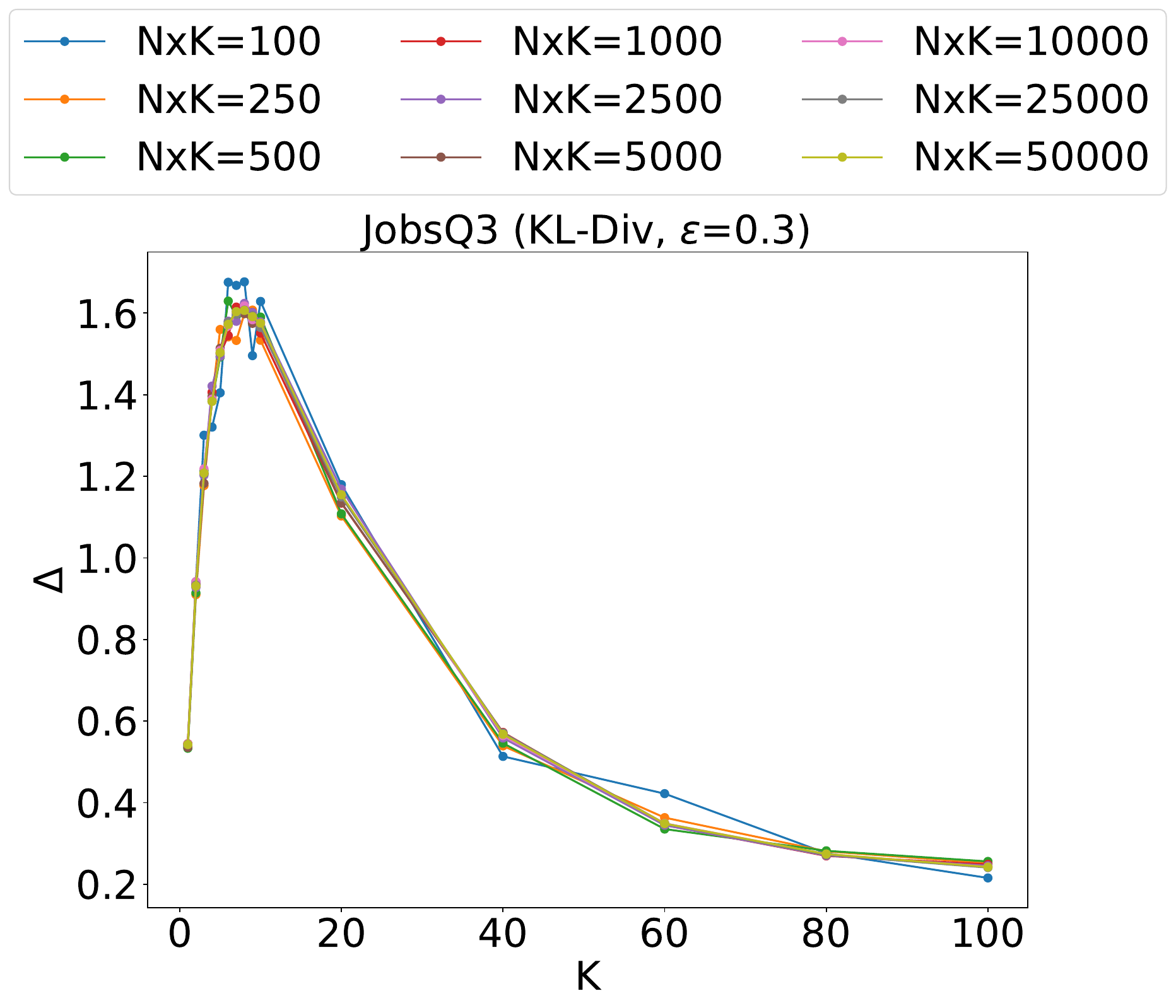}
    \caption{$\epsilon = 0.3$}
    \label{fig:jobsQ3_delta_kl_e03}
  \end{subfigure} \hfill
  \begin{subfigure}[b]{0.24\linewidth}
    \centering
    \includegraphics[width=\linewidth]{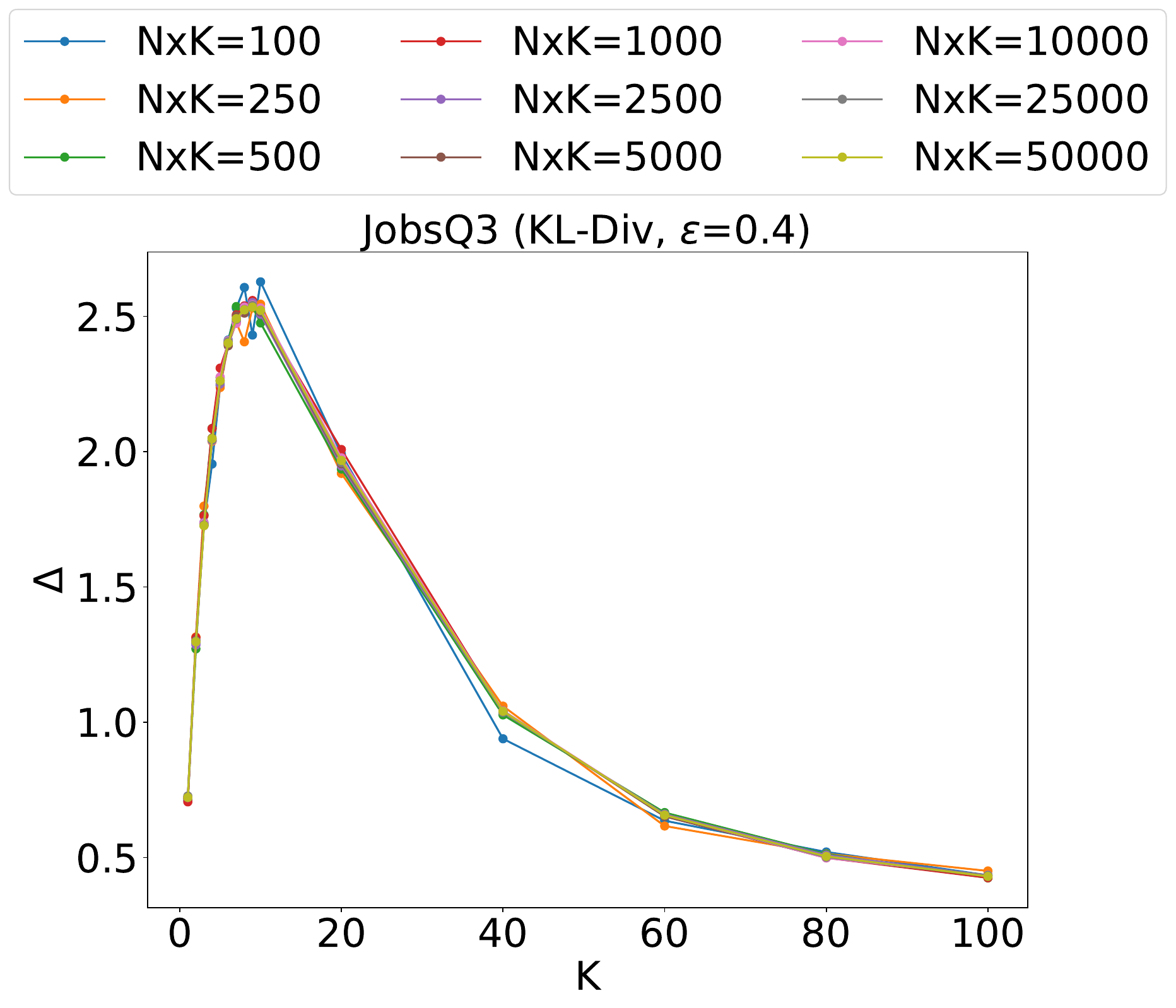}
    \caption{$\epsilon = 0.4$}
    \label{fig:jobsQ3_delta_kl_e04}
  \end{subfigure}
  \caption{Effect sizes ($\Delta$) for JobsQ3 dataset with KL-divergence as the metric}
  \label{fig:jobsQ3_delta_kl}
\end{figure*}

Figures \ref{fig:toxicity_kl}--\ref{fig:jobsQ3_delta_kl} show results for \pv s, confidence intervals, and effect sizes of KL-Divergence for all datasets with different values of $\epsilon$ and increasing $K$ for various $N \times K$. We notice that \pv s\ exhibit double peaks till about $K=40$ for all datasets and for all $N \times K$. The \pv s\ continue to decrease for higher values of $K$. As $\epsilon$ increases, \pv s\ decrease as expected. CI-width and effect sizes generally decrease as $K$ increases, except for JobsQ3 with one initial peak.

\subsubsection{Artificial Distribution with Balanced Priors}

\begin{figure*}
  \centering
  \begin{subfigure}[b]{0.24\linewidth}
    \centering
    \includegraphics[width=\linewidth]{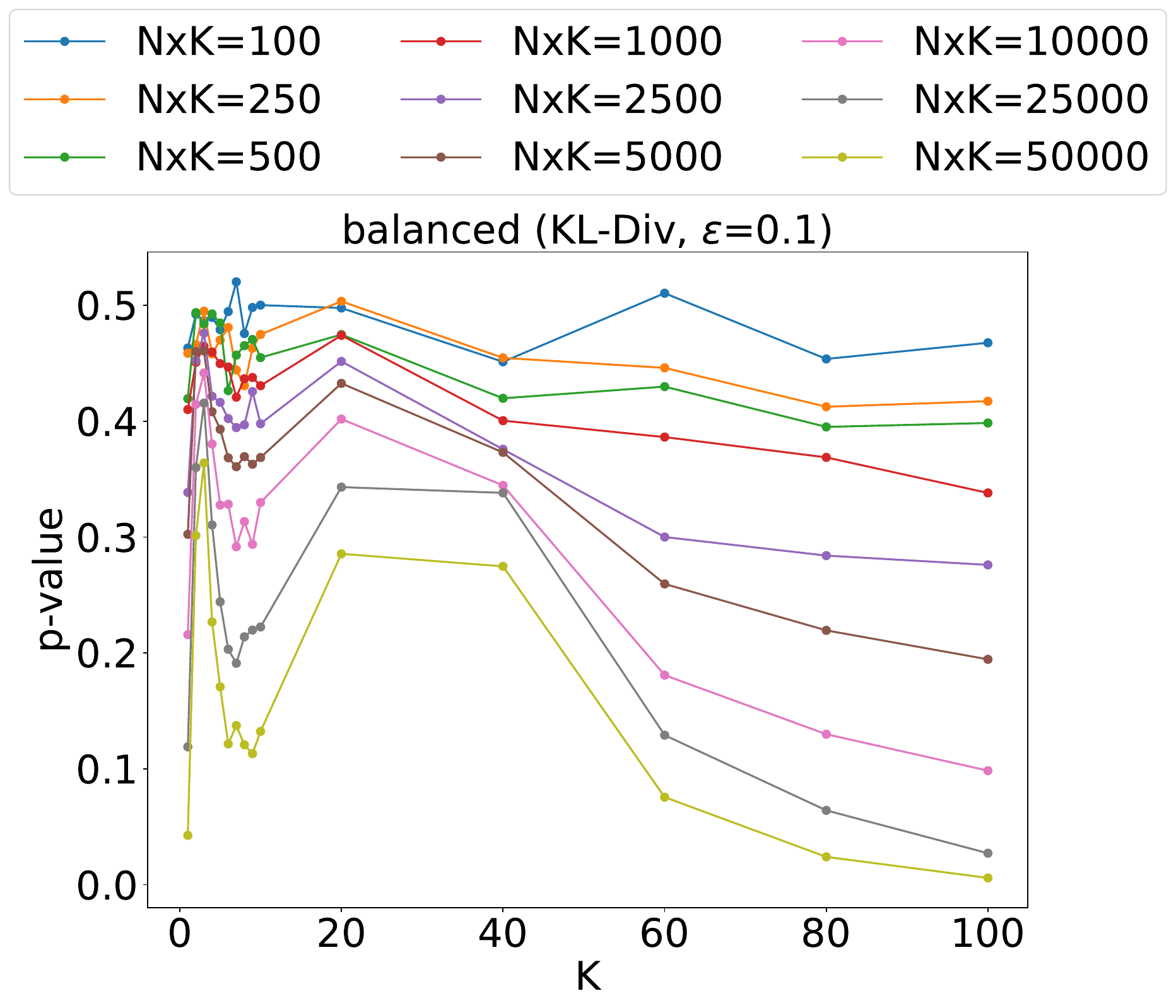}
    \caption{$\epsilon = 0.1$}
    \label{fig:uniform_kl_cat2_e01}
  \end{subfigure} \hfill
  \begin{subfigure}[b]{0.24\linewidth}
    \centering
    \includegraphics[width=\linewidth]{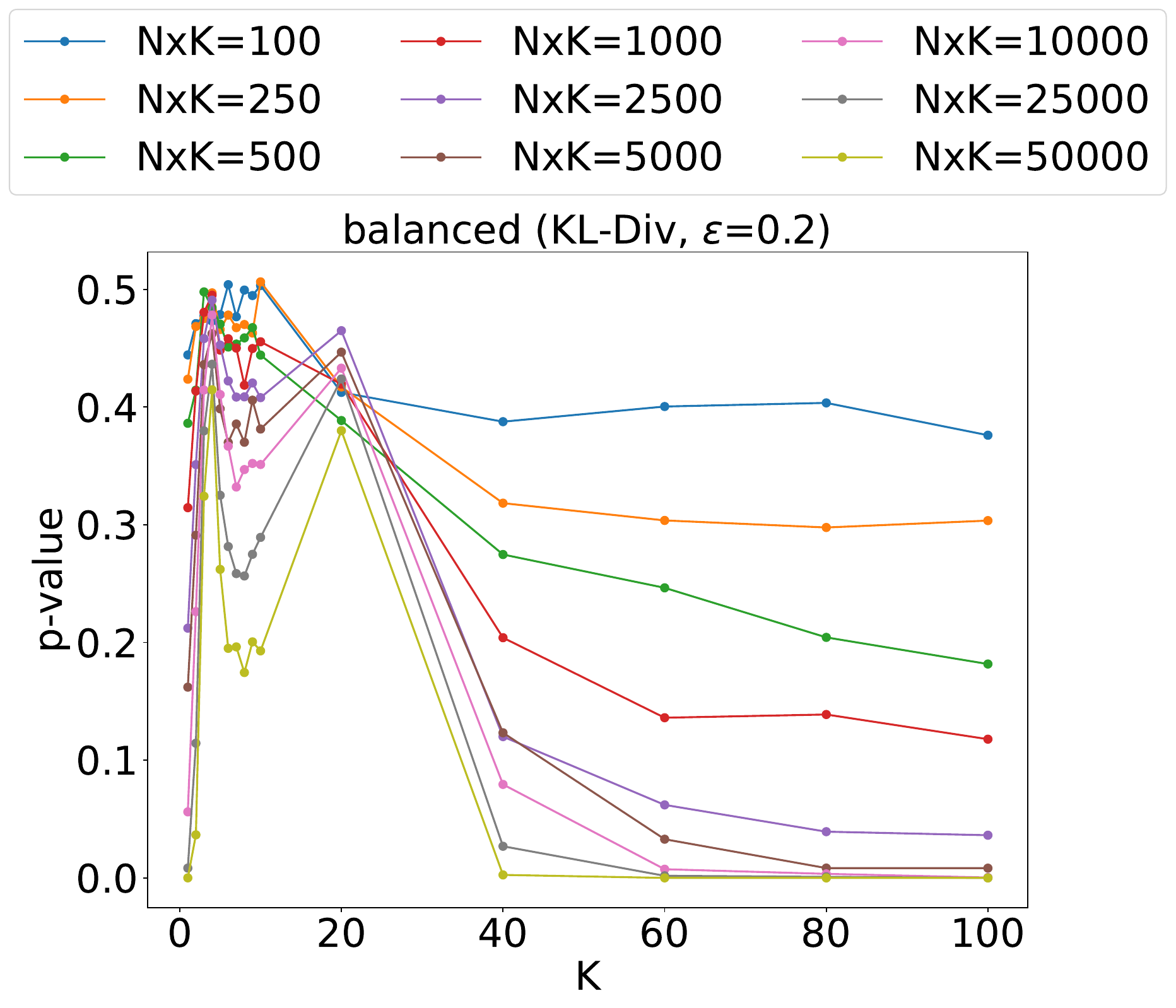}
    \caption{$\epsilon = 0.2$}
    \label{fig:uniform_kl_cat2_e02}
  \end{subfigure} \hfill
  \begin{subfigure}[b]{0.24\linewidth}
    \centering
    \includegraphics[width=\linewidth]{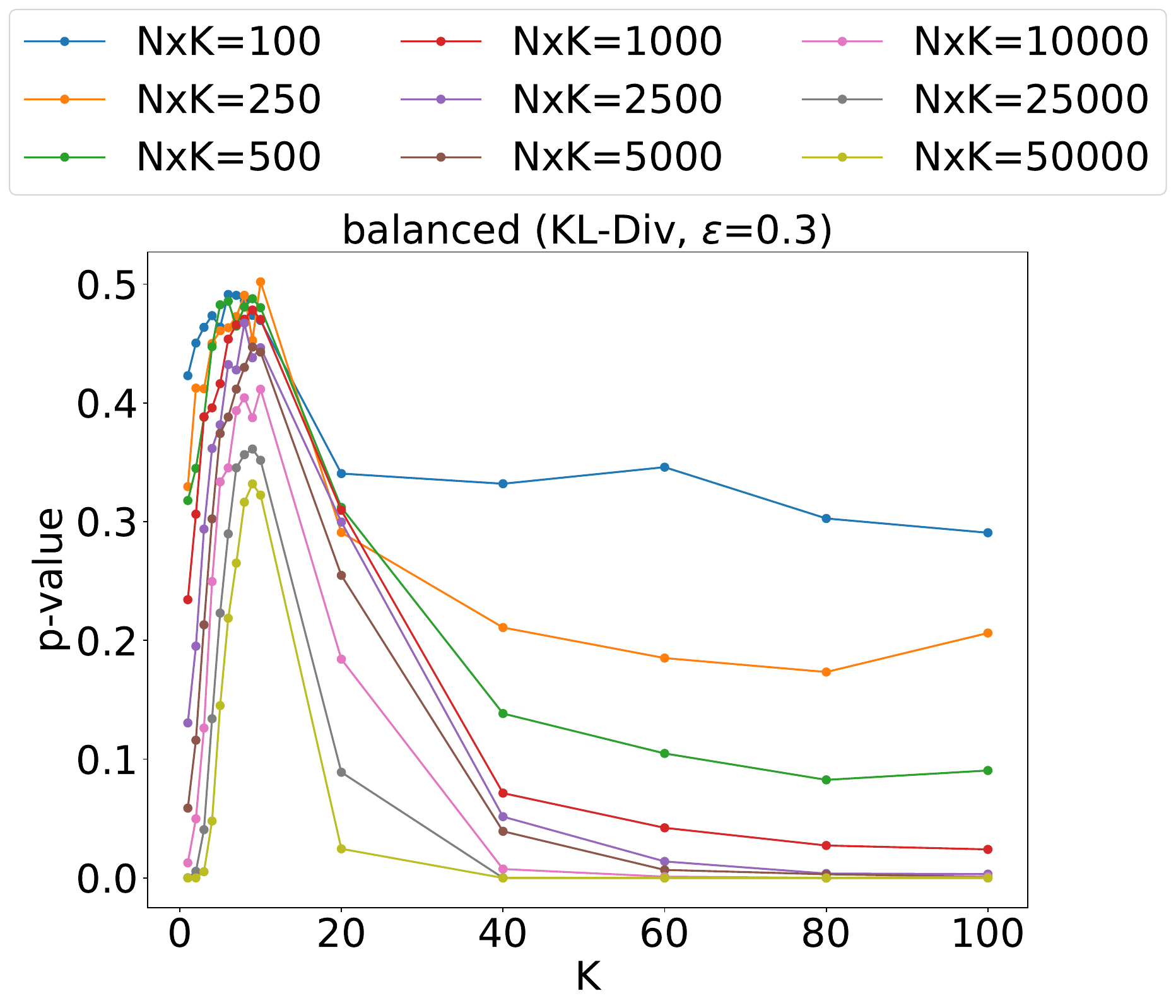}
    \caption{$\epsilon = 0.3$}
    \label{fig:uniform_kl_cat2_e03}
  \end{subfigure} \hfill
  \begin{subfigure}[b]{0.24\linewidth}
    \centering
    \includegraphics[width=\linewidth]{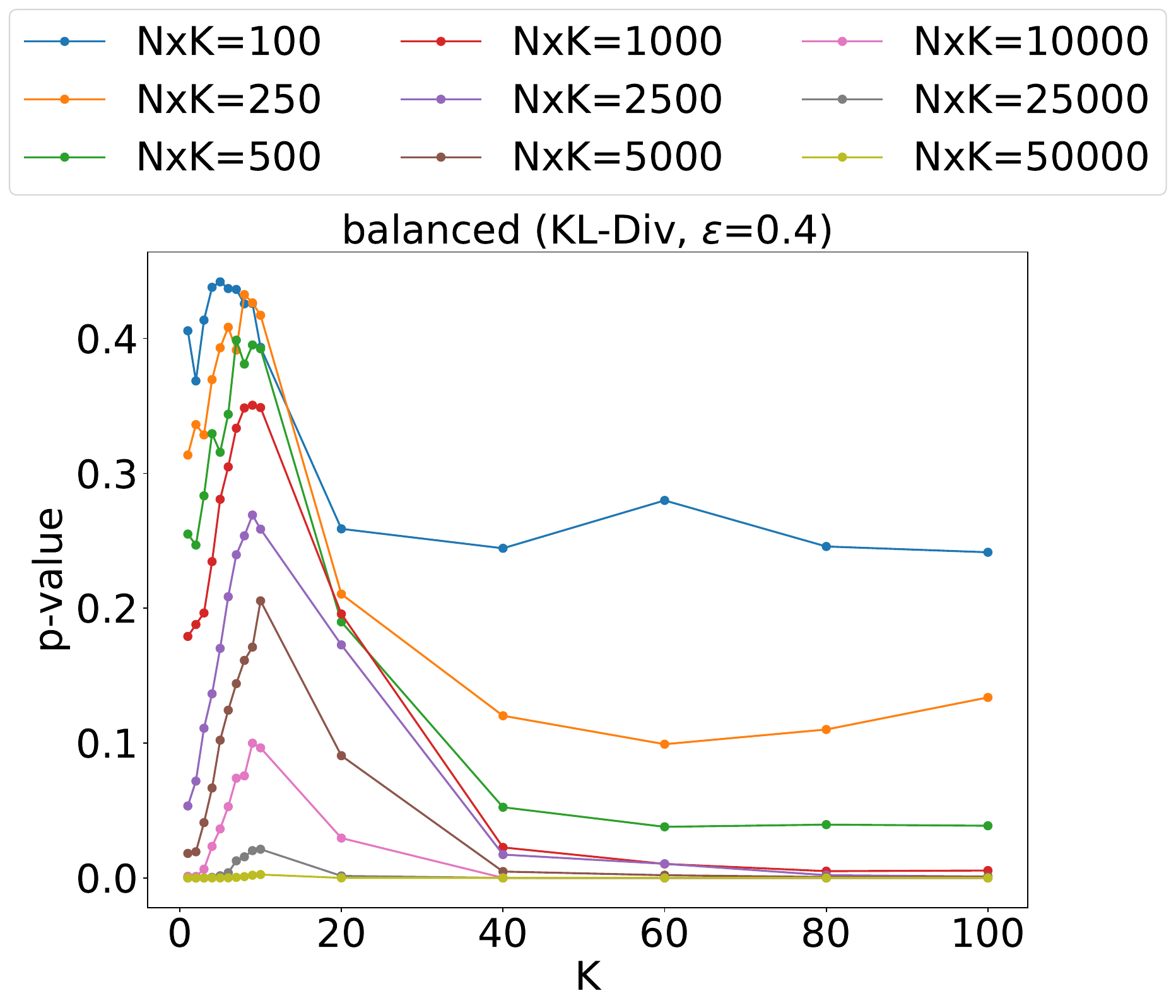}
    \caption{$\epsilon = 0.4$}
    \label{fig:uniform_kl_cat2_e04}
  \end{subfigure}
  \caption{P-value plots for balanced alphas with KL-divergence as the metric ($M=2$)}
  \label{fig:uniform_kl_cat2}
\end{figure*}

\begin{figure*}
  \centering
  \begin{subfigure}[b]{0.24\linewidth}
    \centering
    \includegraphics[width=\linewidth]{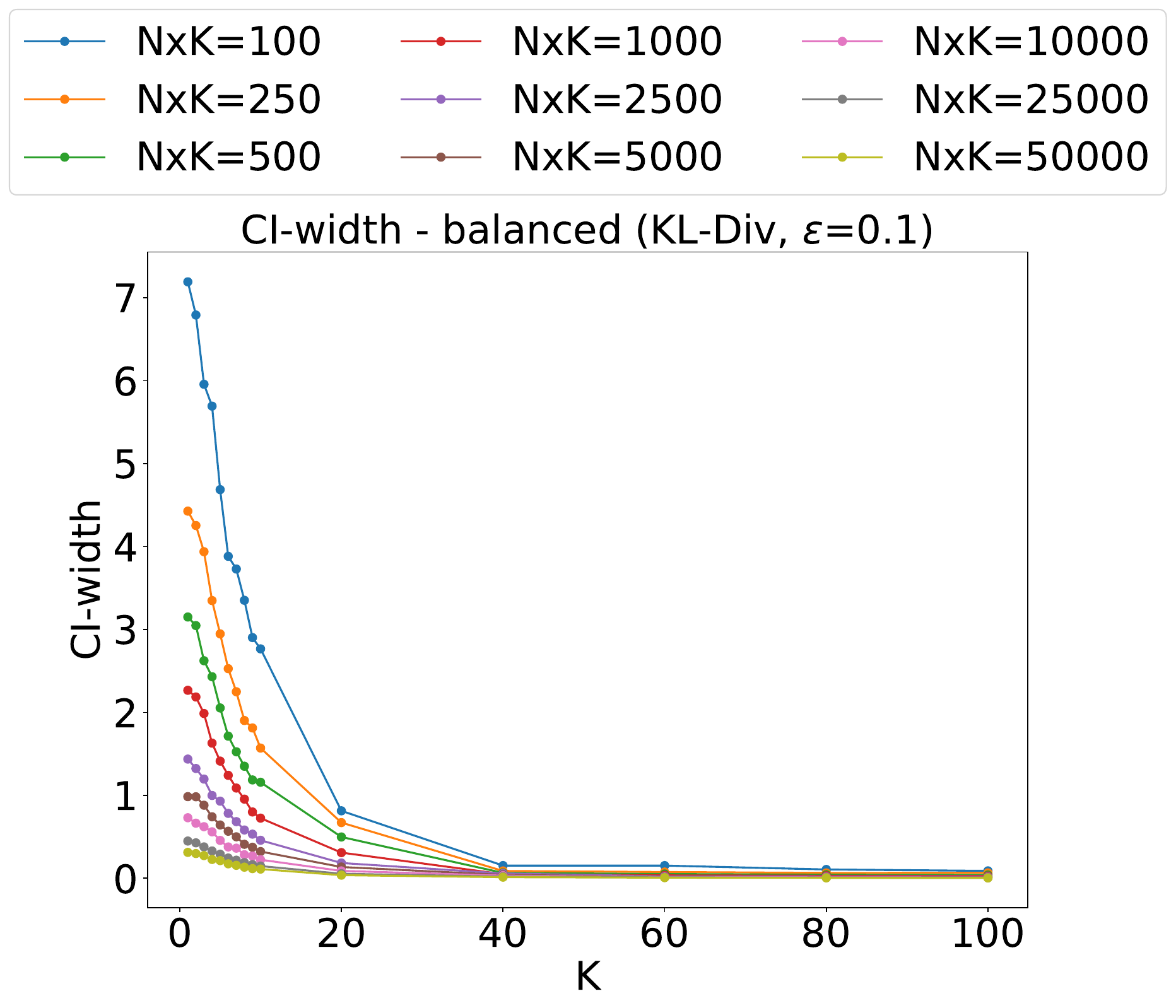}
    \caption{$\epsilon = 0.1$}
    \label{fig:uniform_ci_kl_cat2_e01}
  \end{subfigure} \hfill
  \begin{subfigure}[b]{0.24\linewidth}
    \centering
    \includegraphics[width=\linewidth]{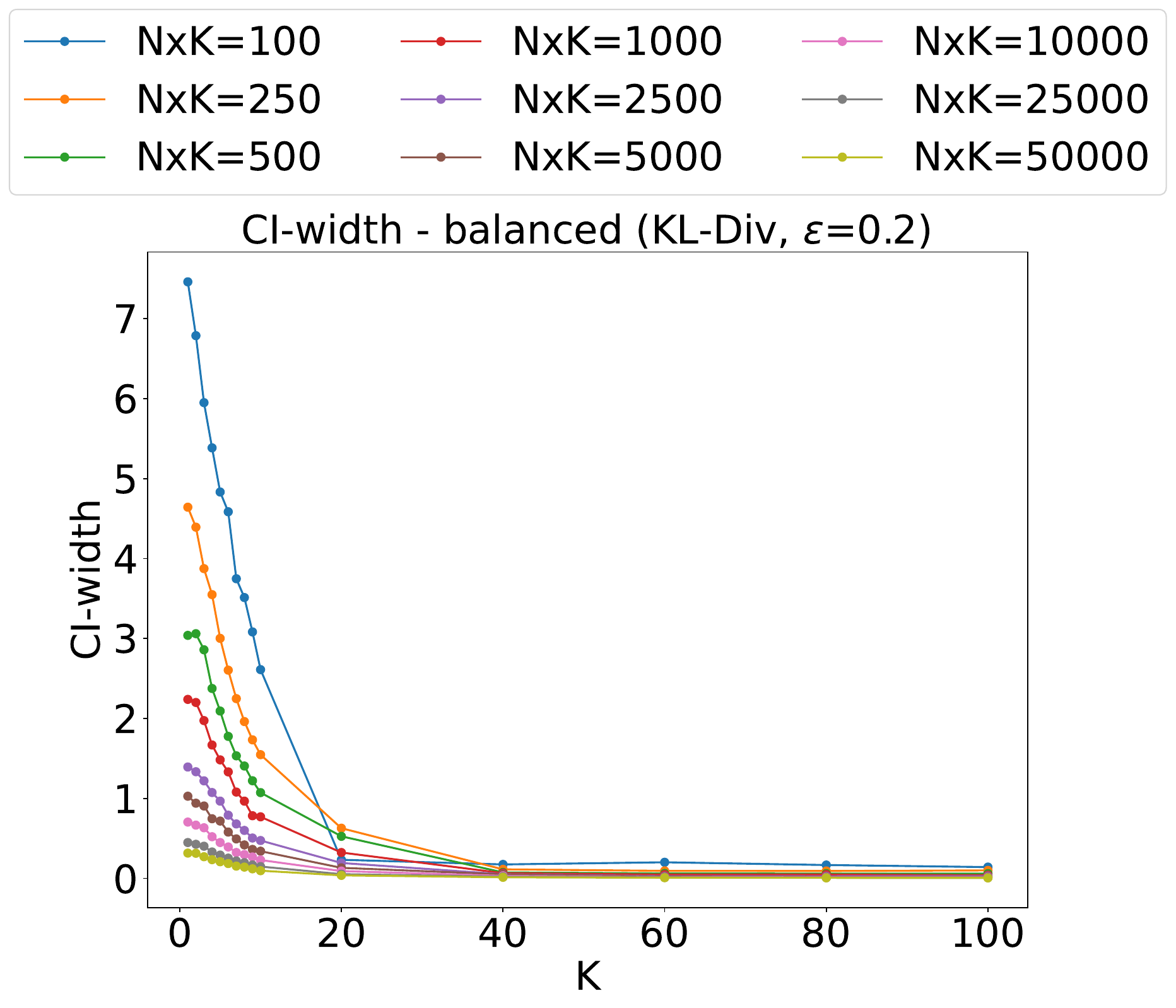}
    \caption{$\epsilon = 0.2$}
    \label{fig:uniform_ci_kl_cat2_e02}
  \end{subfigure} \hfill
  \begin{subfigure}[b]{0.24\linewidth}
    \centering
    \includegraphics[width=\linewidth]{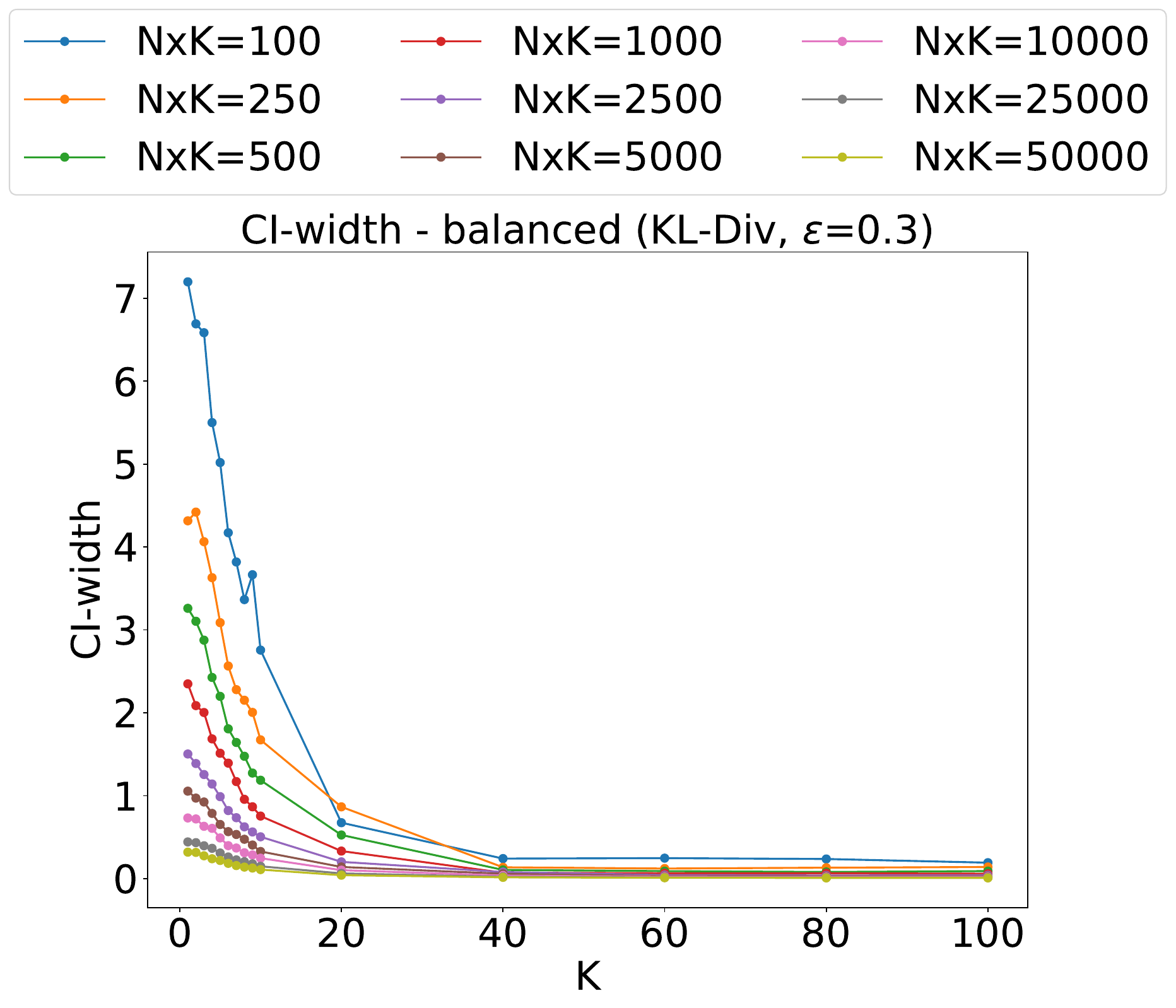}
    \caption{$\epsilon = 0.3$}
    \label{fig:uniform_ci_kl_cat2_e03}
  \end{subfigure} \hfill
  \begin{subfigure}[b]{0.24\linewidth}
    \centering
    \includegraphics[width=\linewidth]{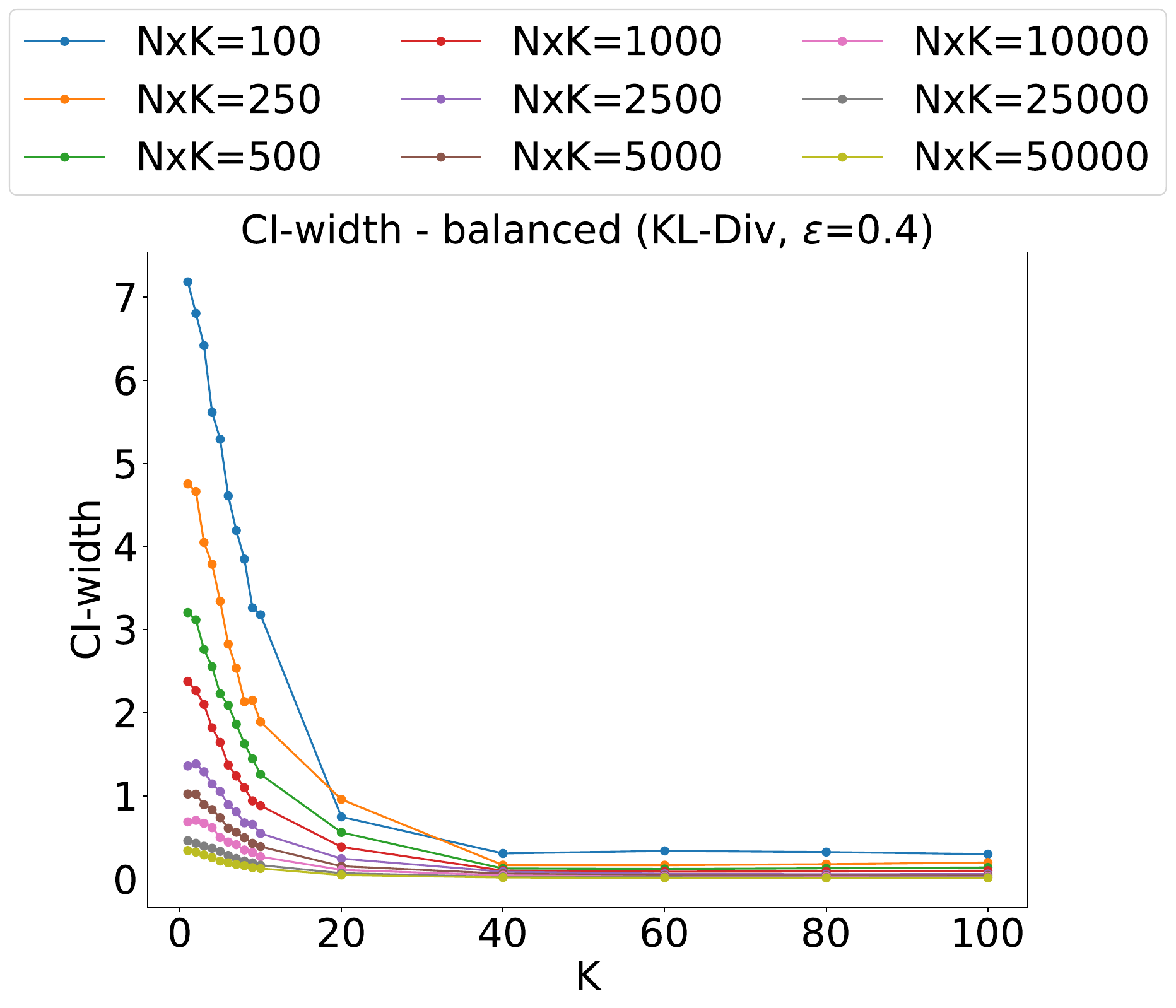}
    \caption{$\epsilon = 0.4$}
    \label{fig:uniform_ci_kl_cat2_e04}
  \end{subfigure}
  \caption{CI-width plots for balanced alphas with KL-divergence as the metric ($M=2$)}
  \label{fig:uniform_ci_kl_cat2}
\end{figure*}

\begin{figure*}
  \centering
  \begin{subfigure}[b]{0.24\linewidth}
    \centering
    \includegraphics[width=\linewidth]{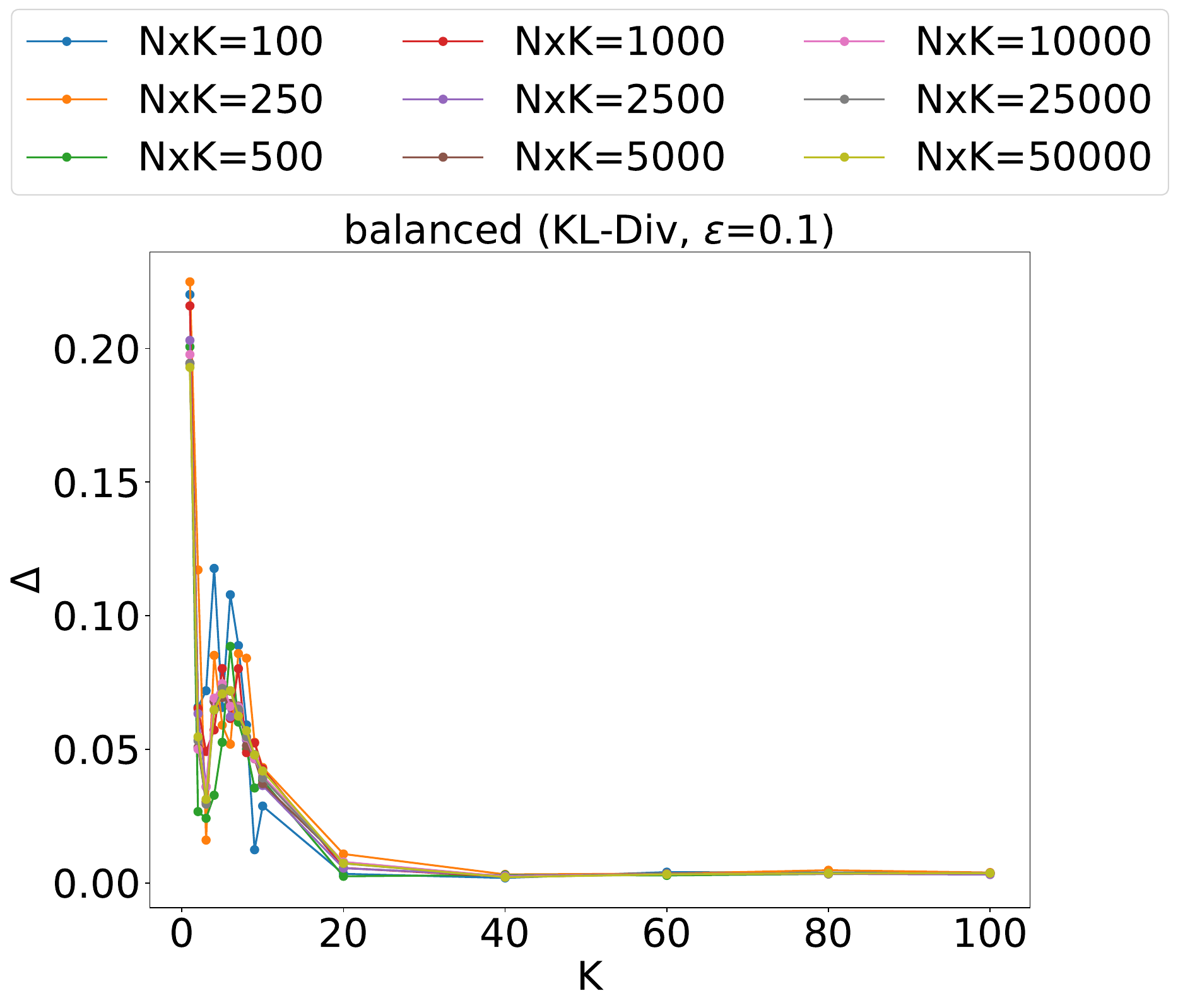}
    \caption{$\epsilon = 0.1$}
    \label{fig:uniform_delta_kl_cat2_e01}
  \end{subfigure} \hfill
  \begin{subfigure}[b]{0.24\linewidth}
    \centering
    \includegraphics[width=\linewidth]{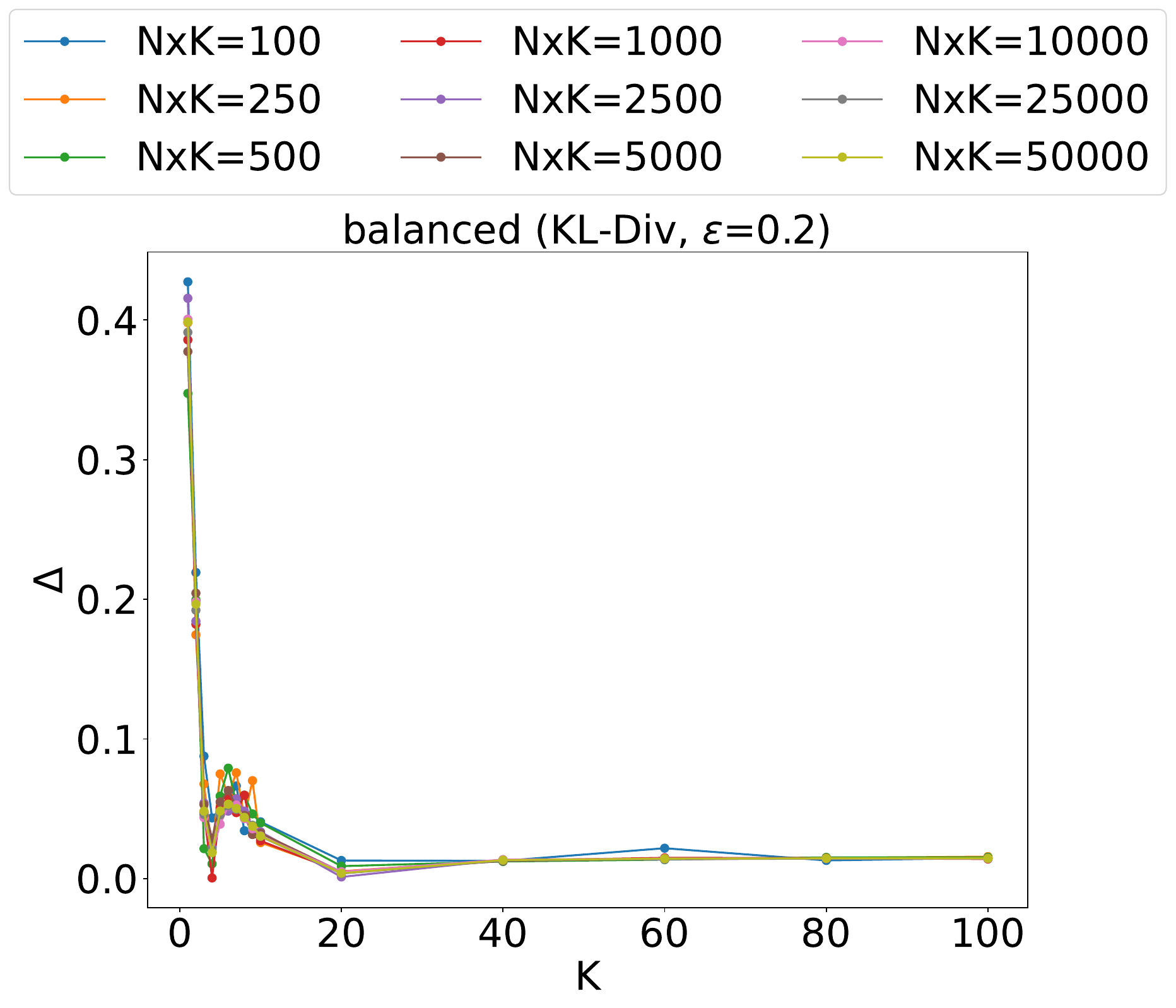}
    \caption{$\epsilon = 0.2$}
    \label{fig:uniform_delta_kl_cat2_e02}
  \end{subfigure} \hfill
  \begin{subfigure}[b]{0.24\linewidth}
    \centering
    \includegraphics[width=\linewidth]{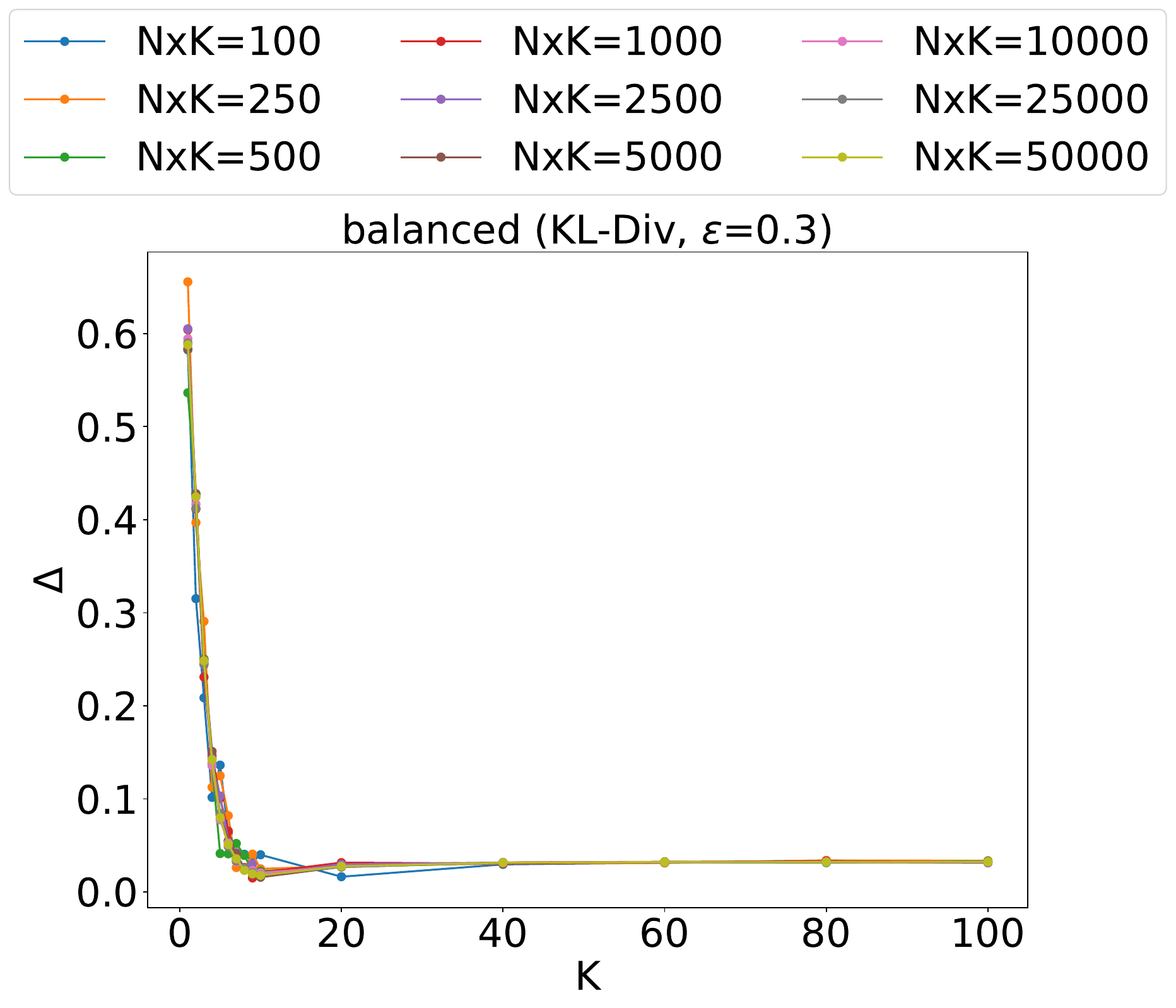}
    \caption{$\epsilon = 0.3$}
    \label{fig:uniform_delta_kl_cat2_e03}
  \end{subfigure} \hfill
  \begin{subfigure}[b]{0.24\linewidth}
    \centering
    \includegraphics[width=\linewidth]{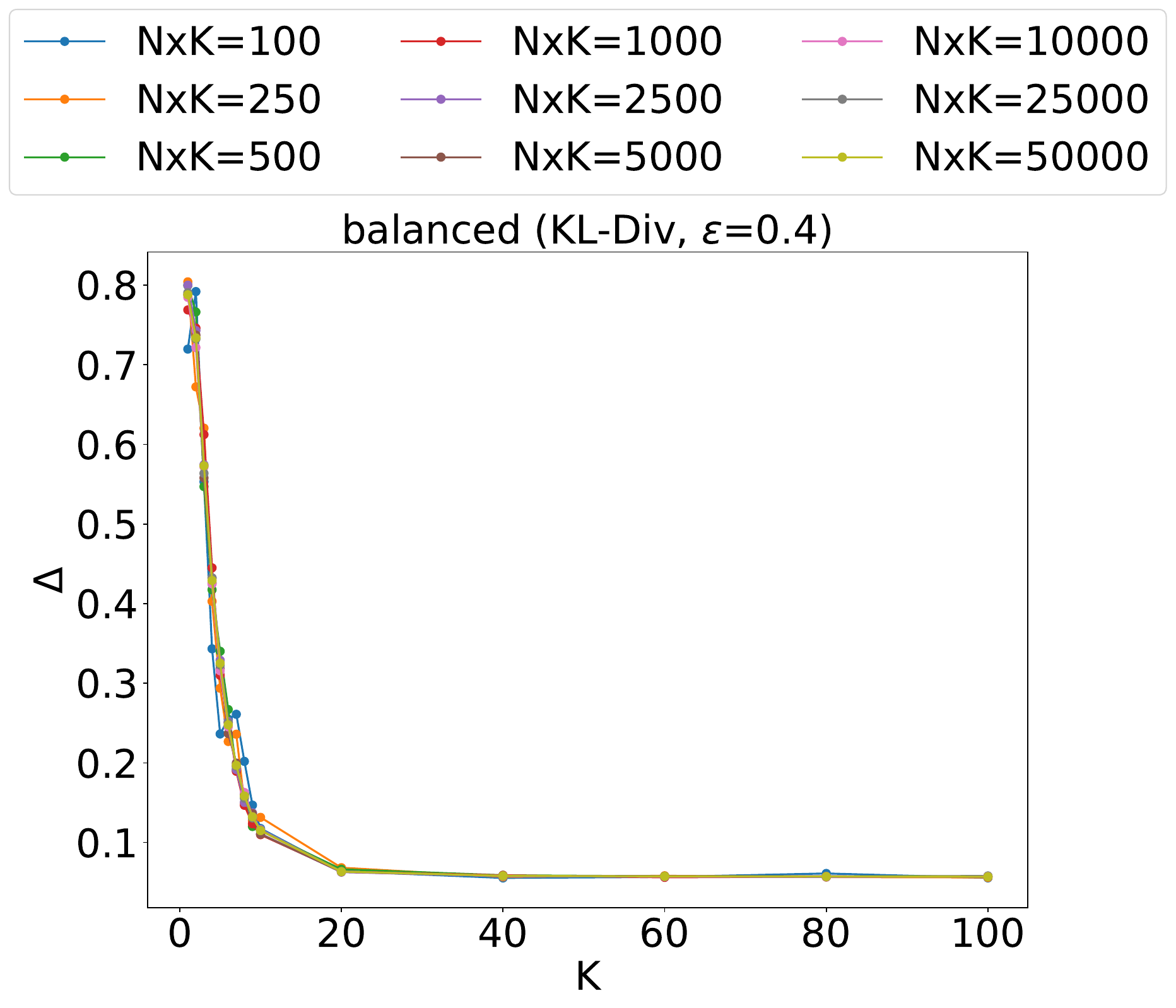}
    \caption{$\epsilon = 0.4$}
    \label{fig:uniform_delta_kl_cat2_e04}
  \end{subfigure}
  \caption{Effect sizes ($\Delta$) for balanced alphas with KL-divergence as the metric ($M=2$)}
  \label{fig:uniform_delta_kl_cat2}
\end{figure*}

\begin{figure*}
  \centering
  \begin{subfigure}[b]{0.24\linewidth}
    \centering
    \includegraphics[width=\linewidth]{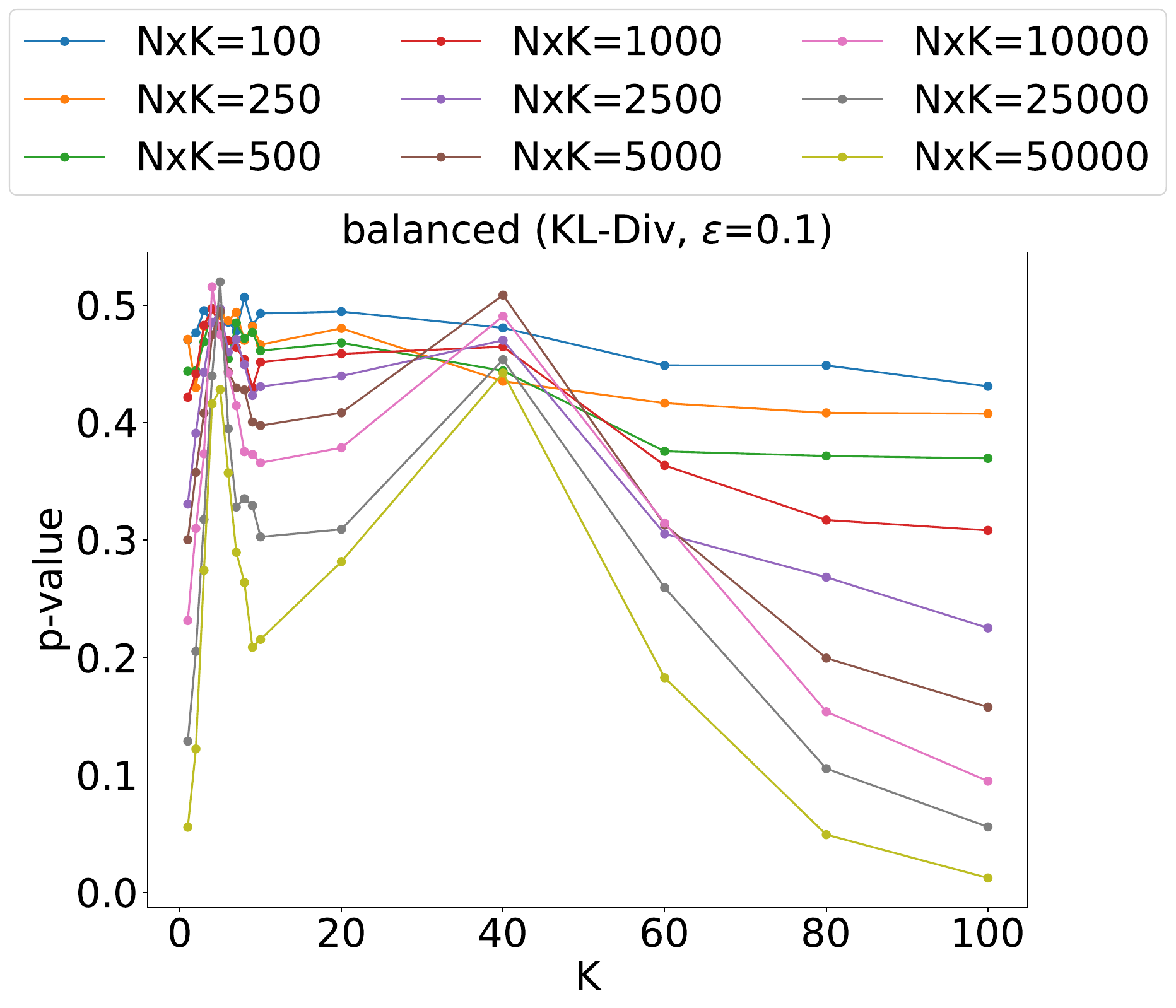}
    \caption{$\epsilon = 0.1$}
    \label{fig:uniform_kl_cat3_e01}
  \end{subfigure} \hfill
  \begin{subfigure}[b]{0.24\linewidth}
    \centering
    \includegraphics[width=\linewidth]{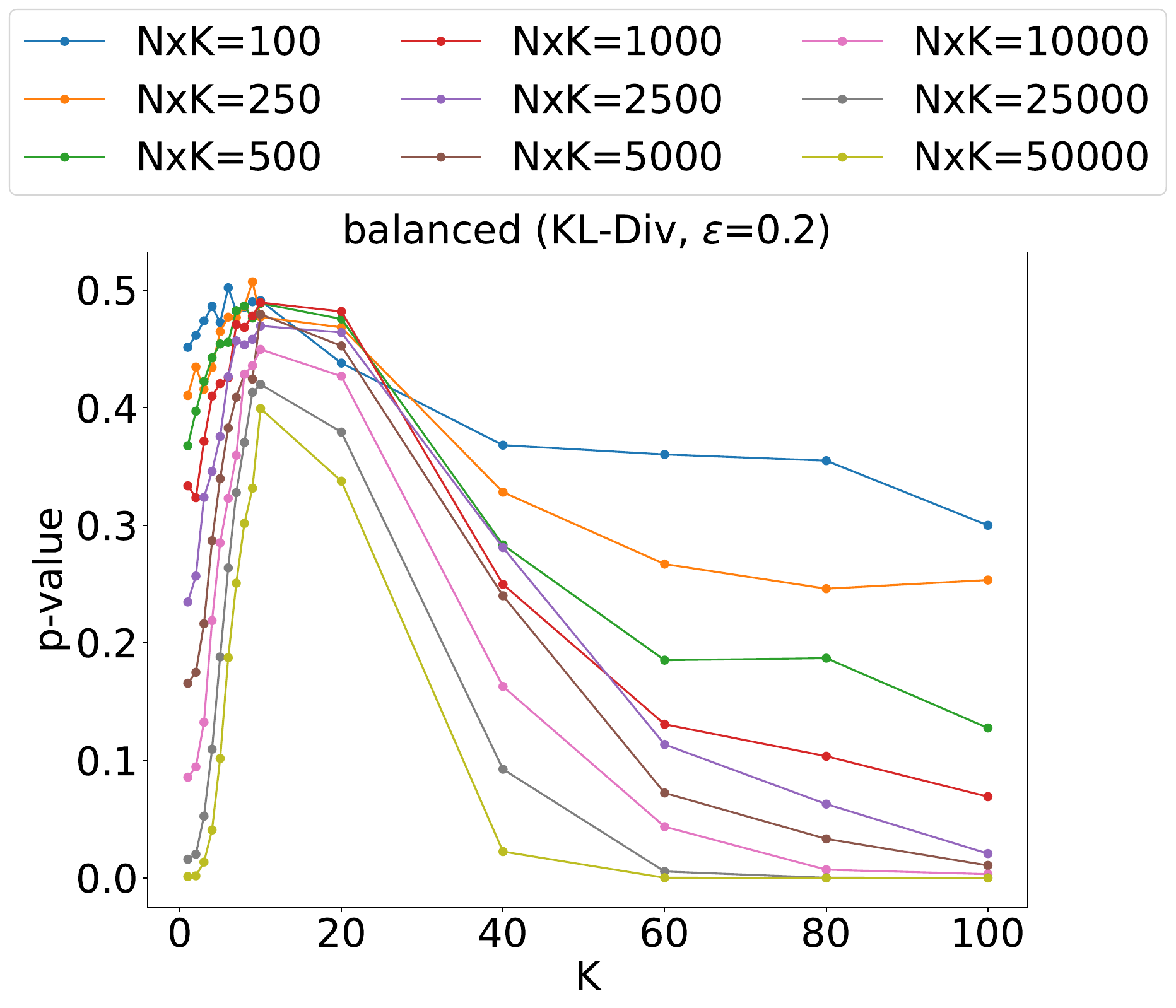}
    \caption{$\epsilon = 0.2$}
    \label{fig:uniform_kl_cat3_e02}
  \end{subfigure} \hfill
  \begin{subfigure}[b]{0.24\linewidth}
    \centering
    \includegraphics[width=\linewidth]{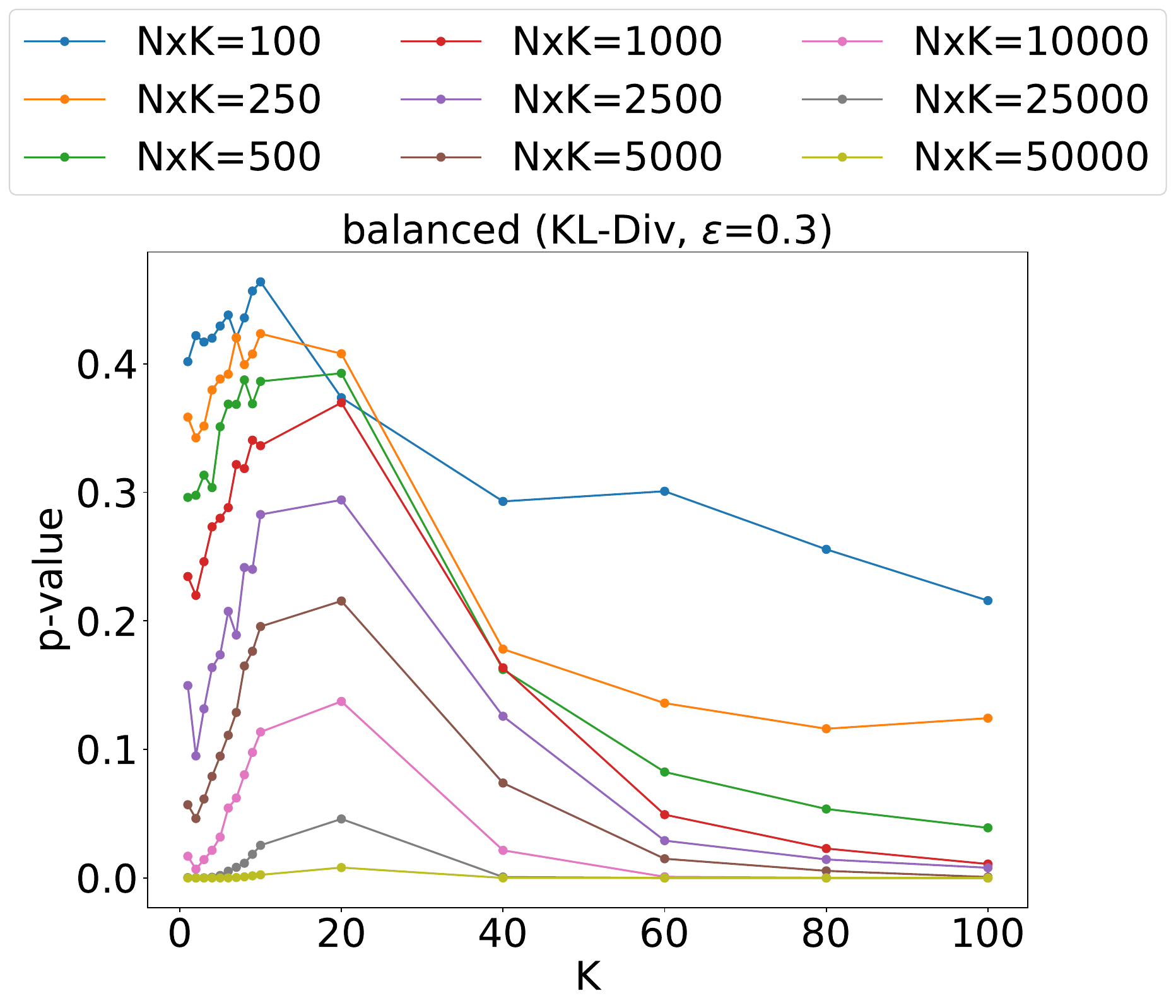}
    \caption{$\epsilon = 0.3$}
    \label{fig:uniform_kl_cat3_e03}
  \end{subfigure} \hfill
  \begin{subfigure}[b]{0.24\linewidth}
    \centering
    \includegraphics[width=\linewidth]{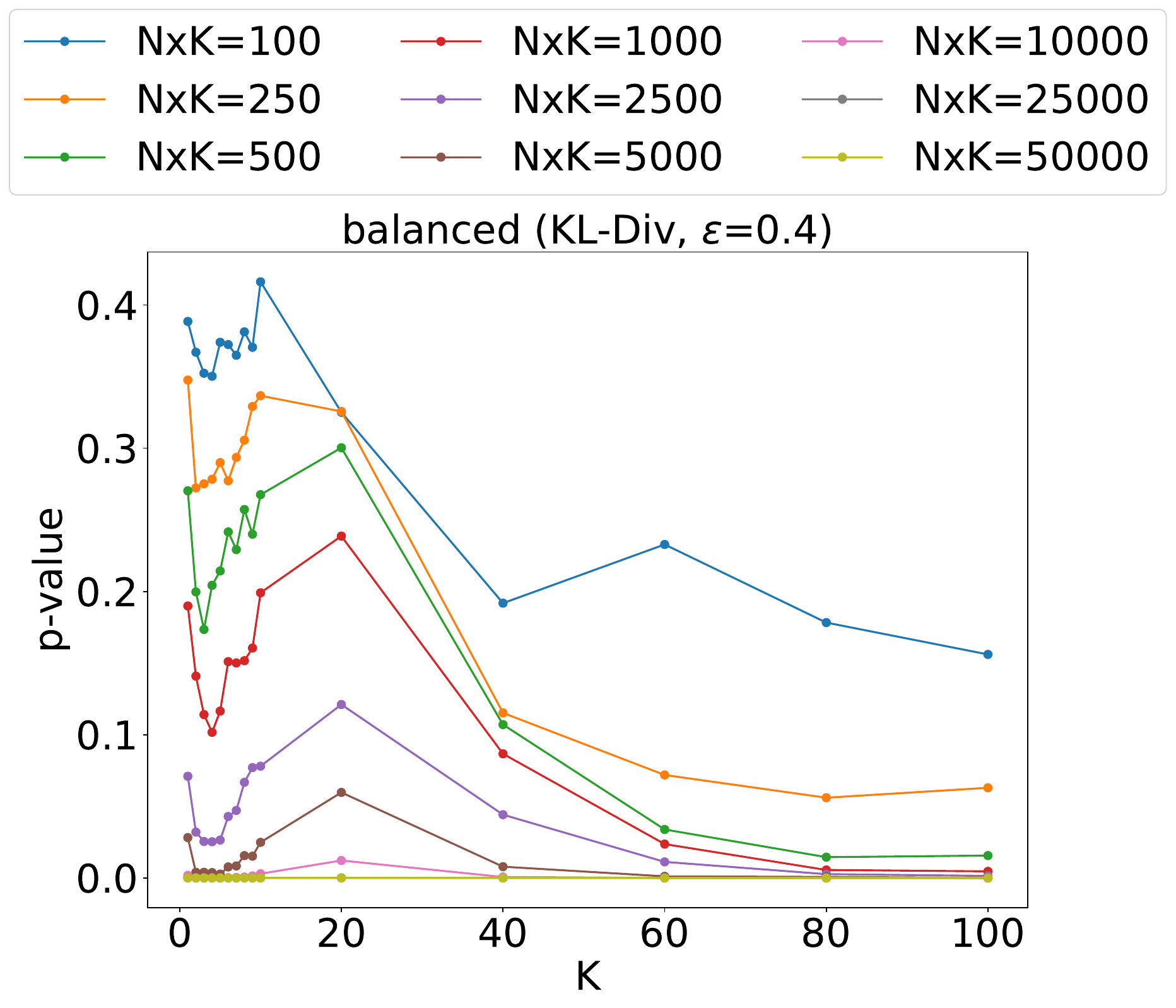}
    \caption{$\epsilon = 0.4$}
    \label{fig:uniform_kl_cat3_e04}
  \end{subfigure}
  \caption{P-value plots for balanced alphas with KL-divergence as the metric ($M=3$)}
  \label{fig:uniform_kl_cat3}
\end{figure*}

\begin{figure*}
  \centering
  \begin{subfigure}[b]{0.24\linewidth}
    \centering
    \includegraphics[width=\linewidth]{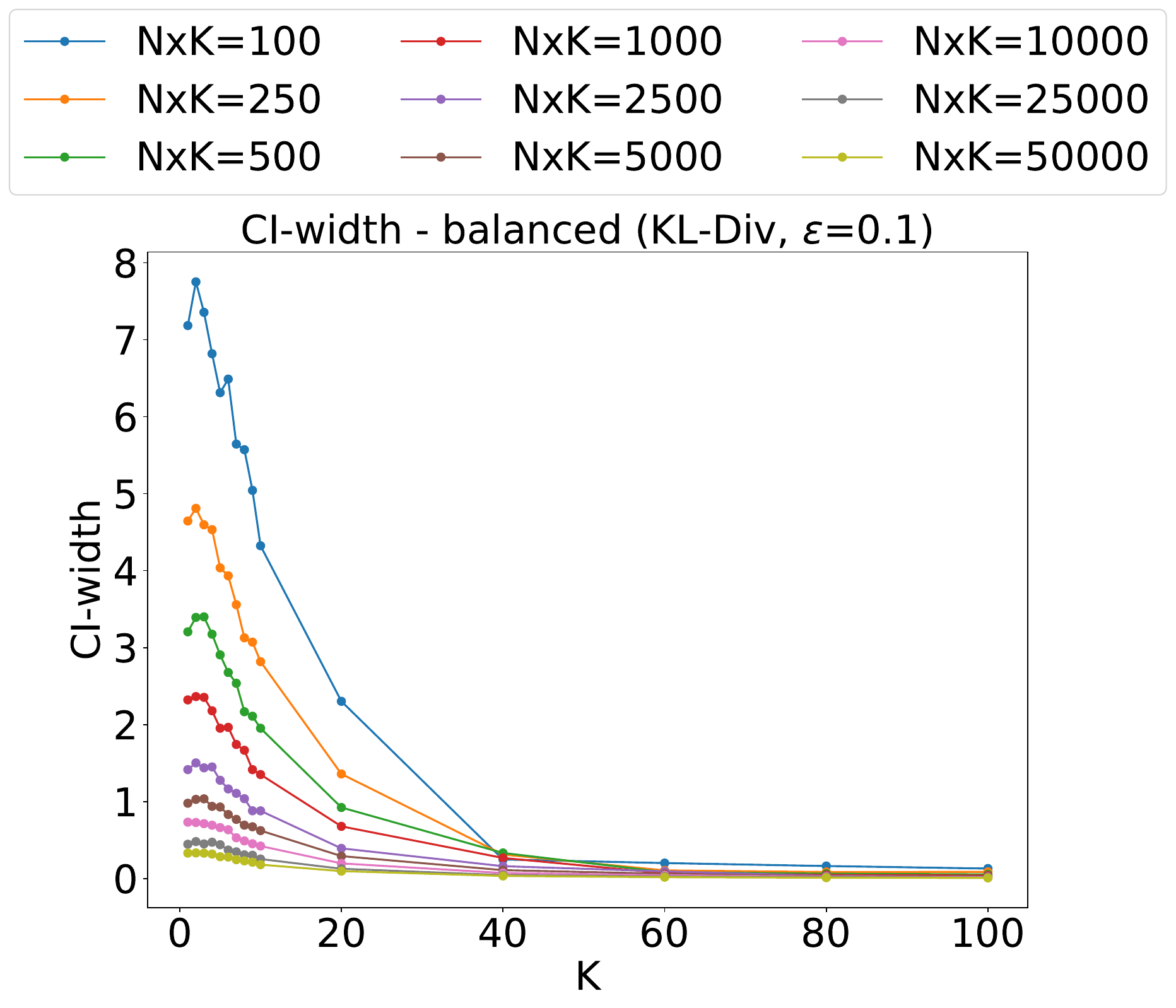}
    \caption{$\epsilon = 0.1$}
    \label{fig:uniform_ci_kl_cat3_e01}
  \end{subfigure} \hfill
  \begin{subfigure}[b]{0.24\linewidth}
    \centering
    \includegraphics[width=\linewidth]{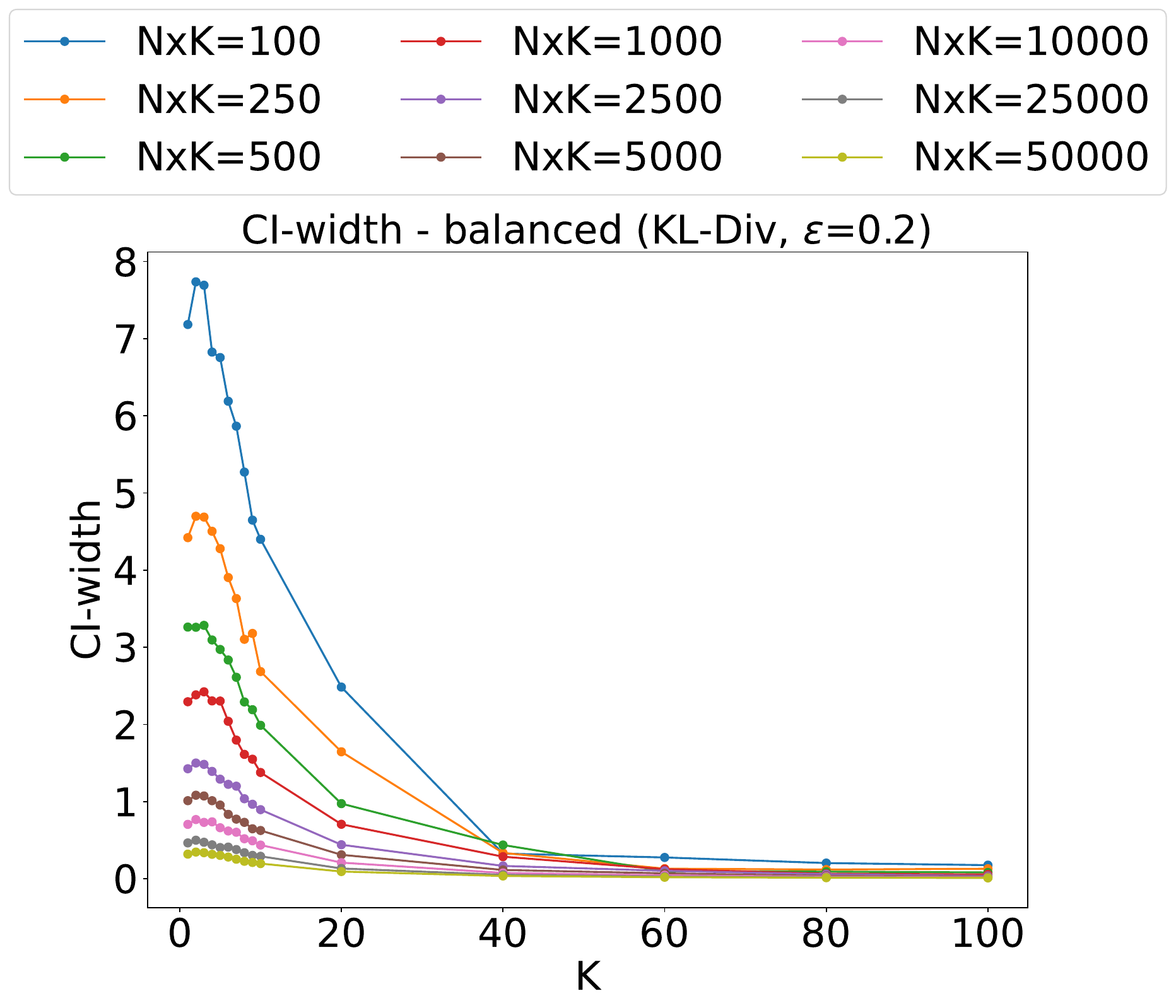}
    \caption{$\epsilon = 0.2$}
    \label{fig:uniform_ci_kl_cat3_e02}
  \end{subfigure} \hfill
  \begin{subfigure}[b]{0.24\linewidth}
    \centering
    \includegraphics[width=\linewidth]{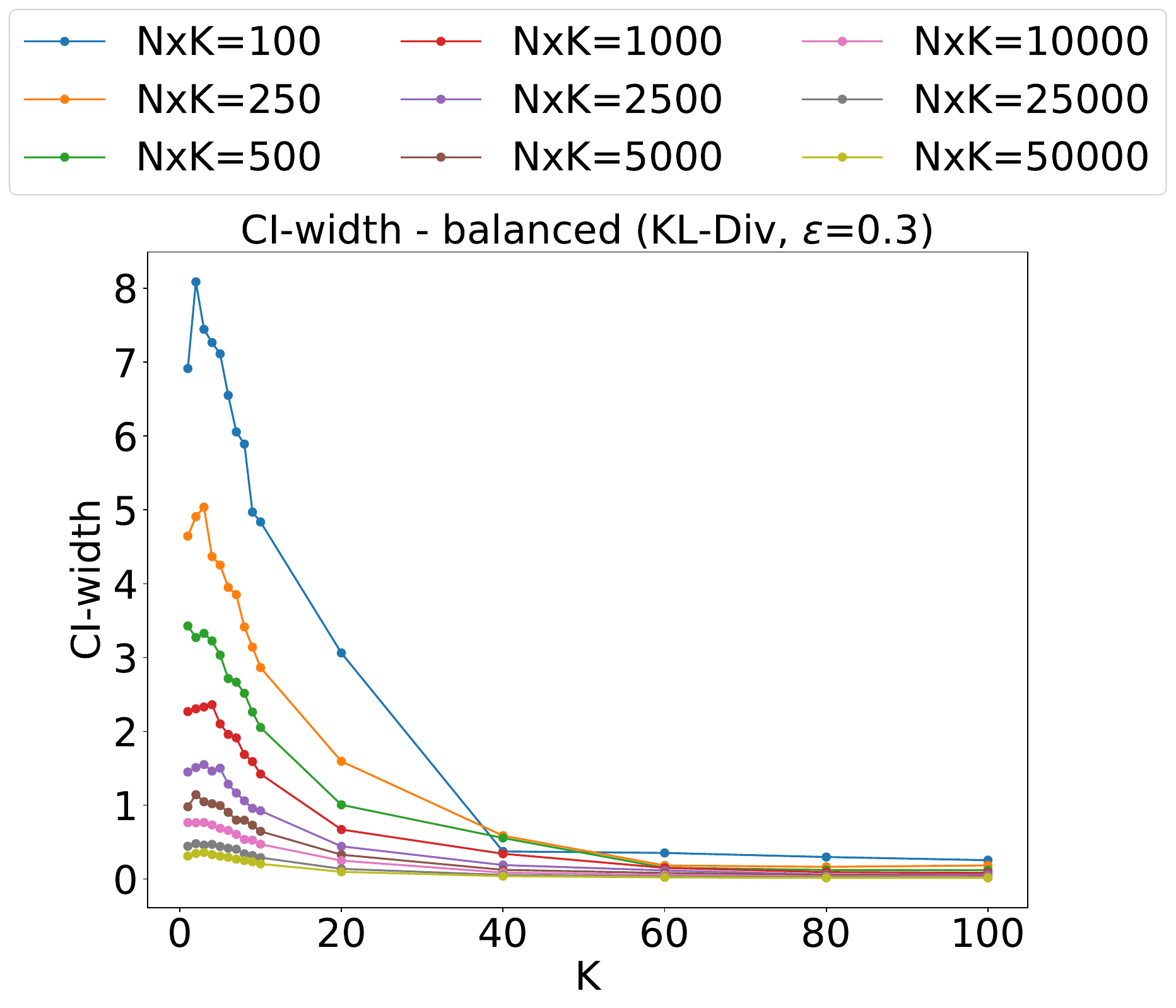}
    \caption{$\epsilon = 0.3$}
    \label{fig:uniform_ci_kl_cat3_e03}
  \end{subfigure} \hfill
  \begin{subfigure}[b]{0.24\linewidth}
    \centering
    \includegraphics[width=\linewidth]{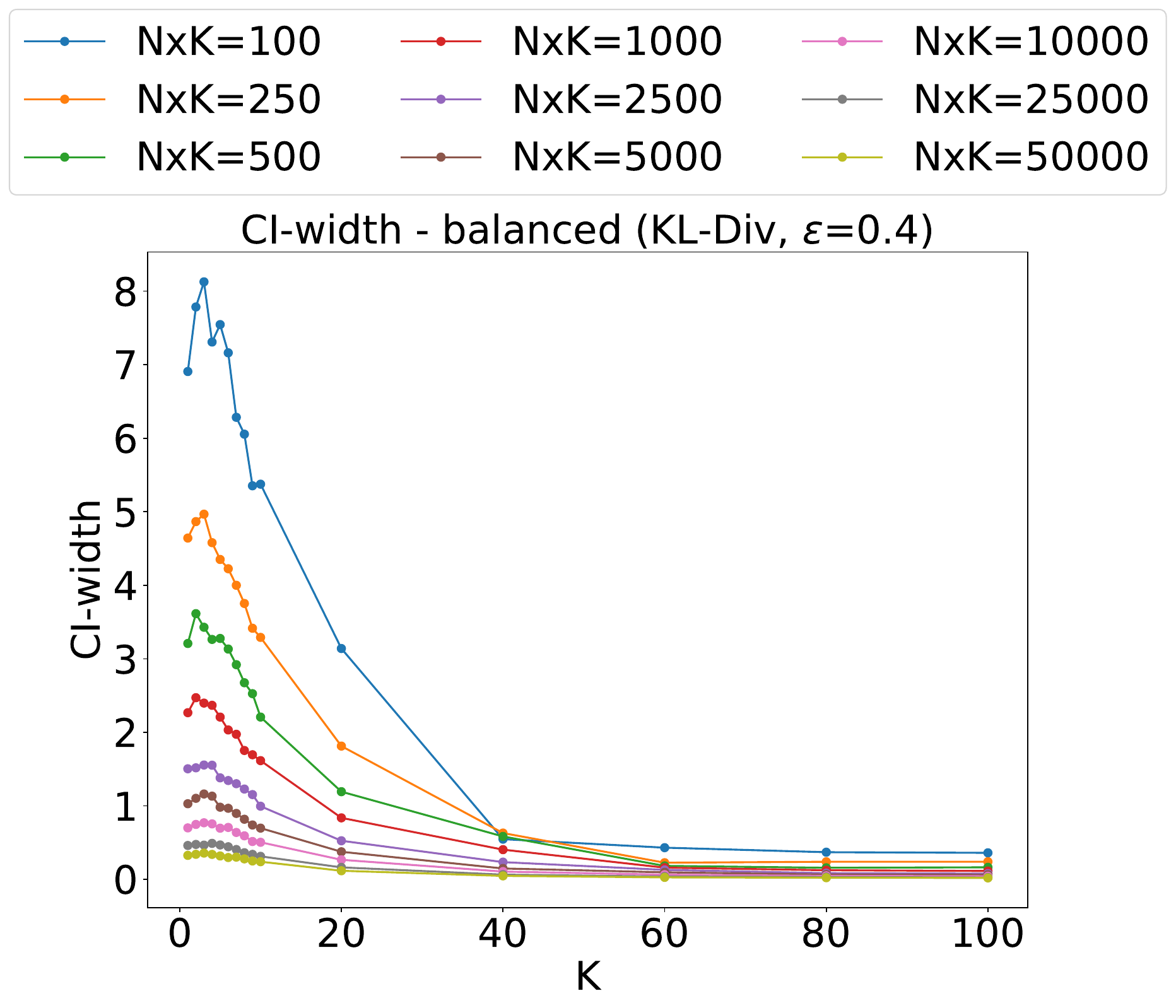}
    \caption{$\epsilon = 0.4$}
    \label{fig:uniform_ci_kl_cat3_e04}
  \end{subfigure}
  \caption{CI-width plots for balanced alphas with KL-divergence as the metric ($M=3$)}
  \label{fig:uniform_ci_kl_cat3}
\end{figure*}

\begin{figure*}
  \centering
  \begin{subfigure}[b]{0.24\linewidth}
    \centering
    \includegraphics[width=\linewidth]{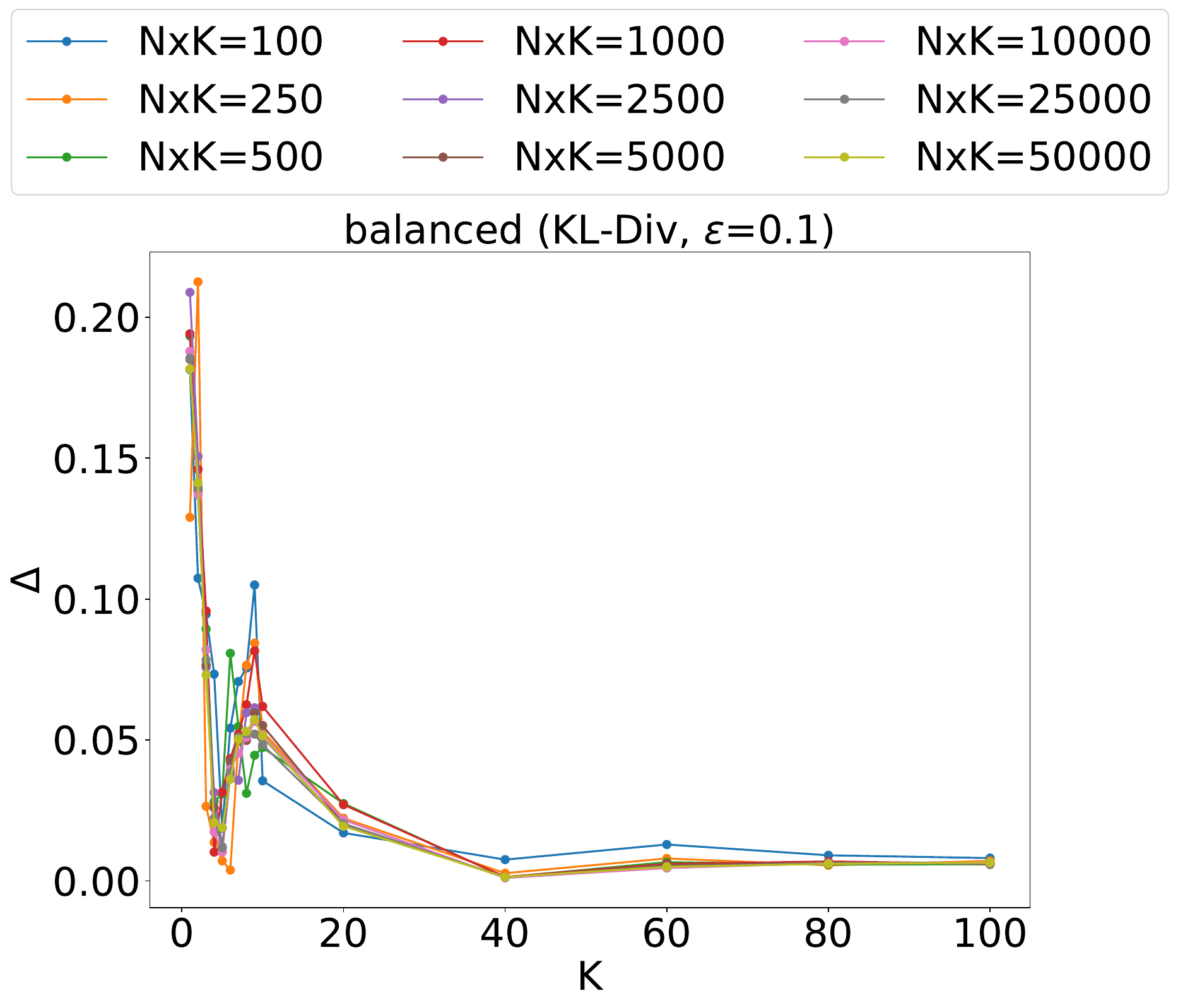}
    \caption{$\epsilon = 0.1$}
    \label{fig:uniform_delta_kl_cat3_e01}
  \end{subfigure} \hfill
  \begin{subfigure}[b]{0.24\linewidth}
    \centering
    \includegraphics[width=\linewidth]{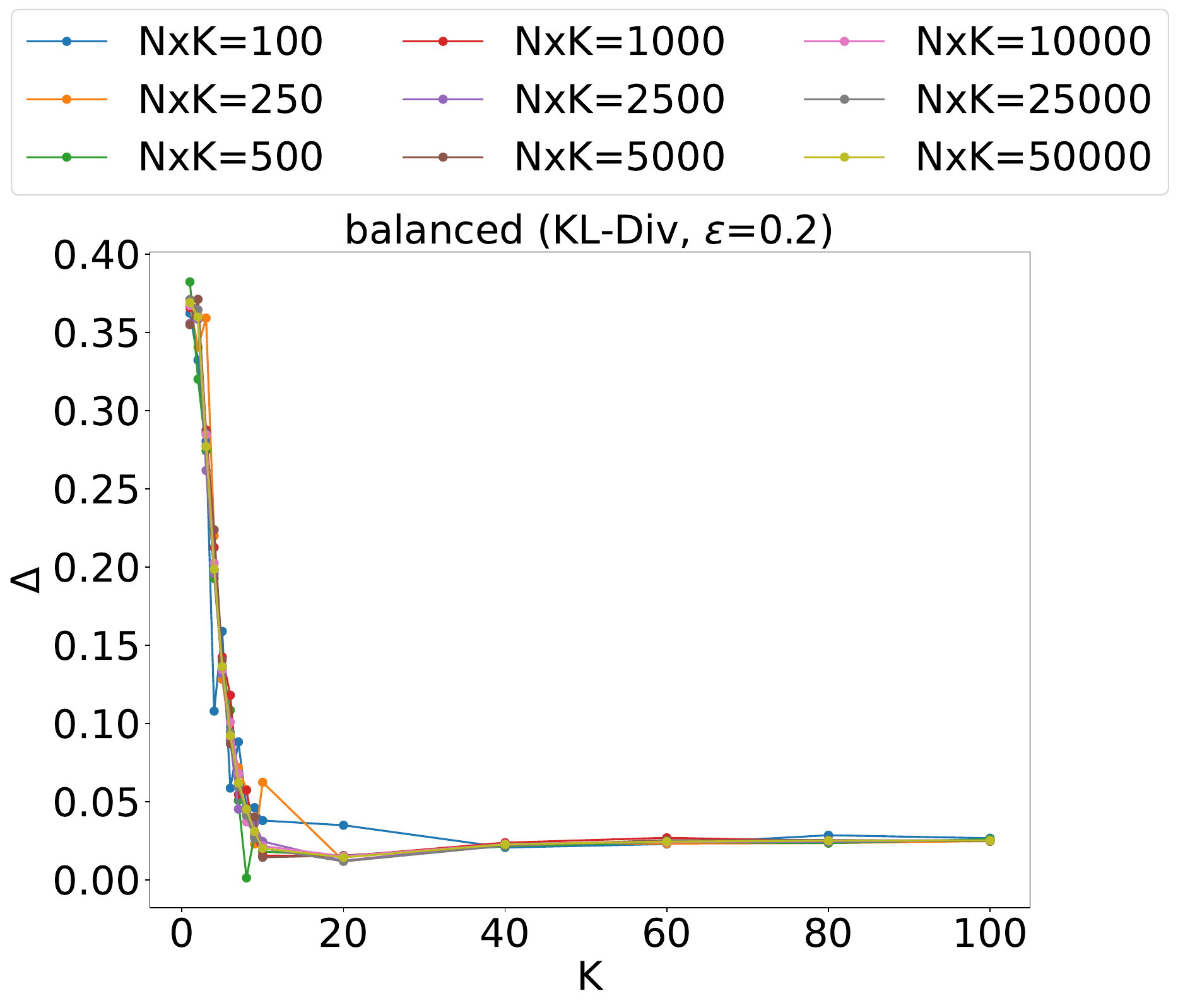}
    \caption{$\epsilon = 0.2$}
    \label{fig:uniform_delta_kl_cat3_e02}
  \end{subfigure} \hfill
  \begin{subfigure}[b]{0.24\linewidth}
    \centering
    \includegraphics[width=\linewidth]{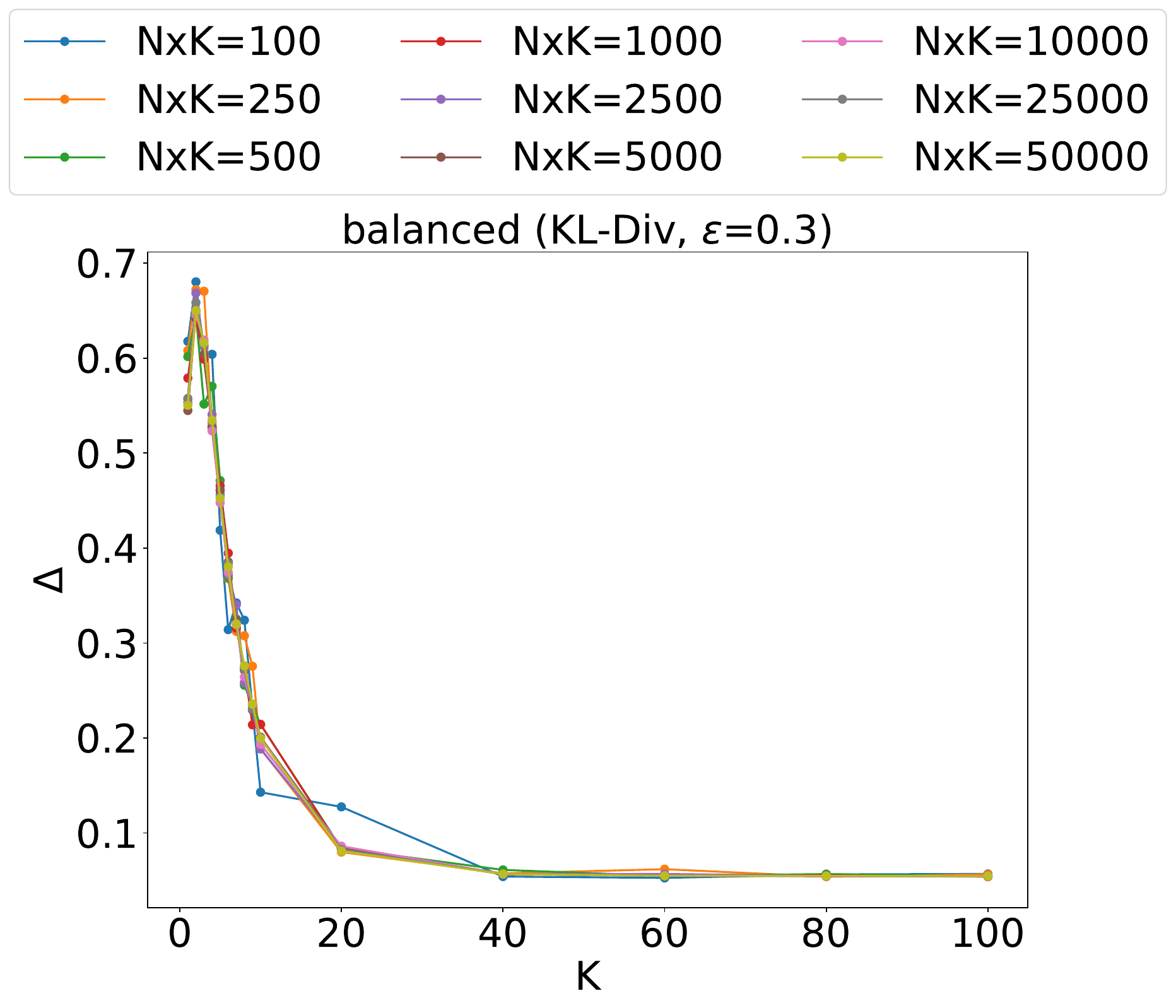}
    \caption{$\epsilon = 0.3$}
    \label{fig:uniform_delta_kl_cat3_e03}
  \end{subfigure} \hfill
  \begin{subfigure}[b]{0.24\linewidth}
    \centering
    \includegraphics[width=\linewidth]{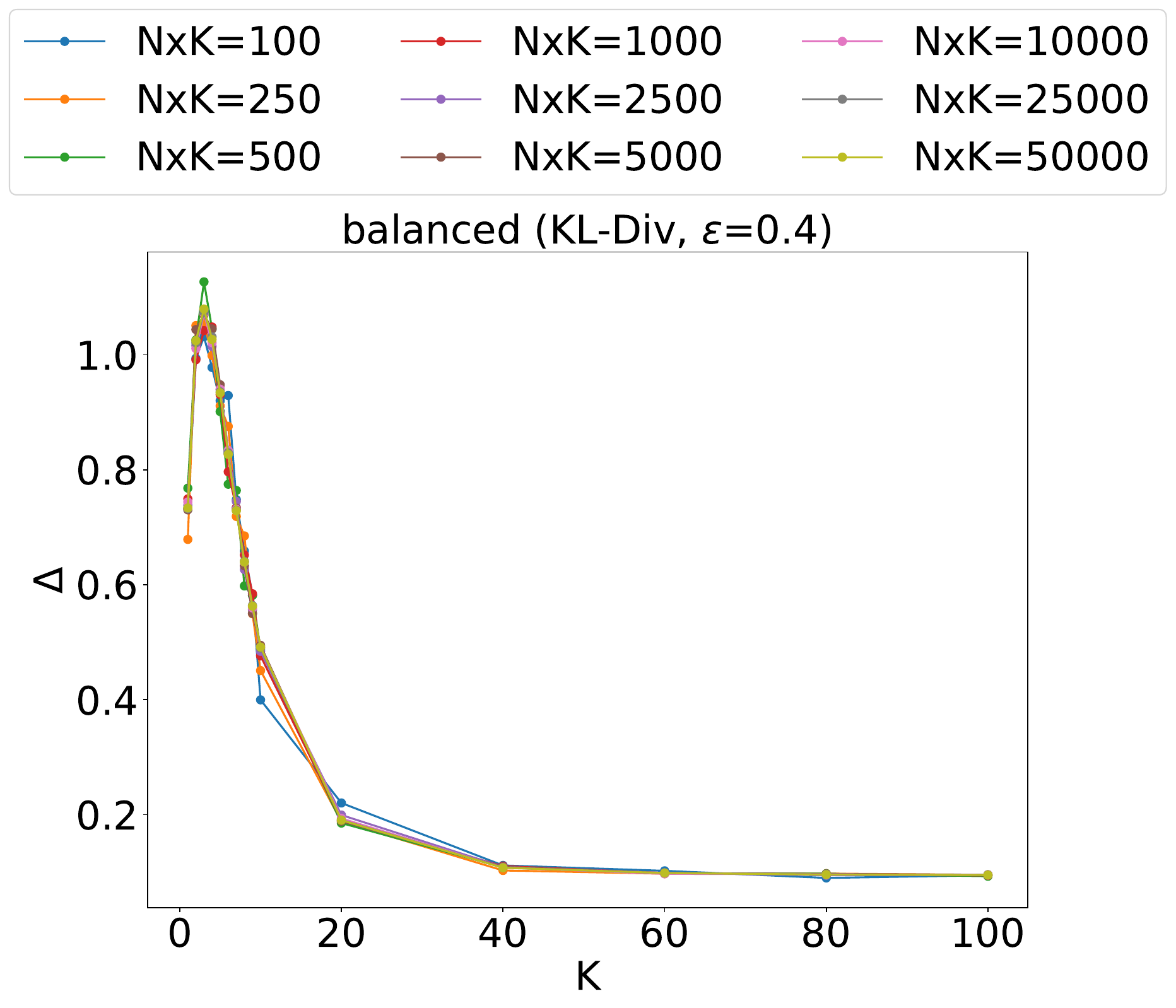}
    \caption{$\epsilon = 0.4$}
    \label{fig:uniform_delta_kl_cat3_e04}
  \end{subfigure}
  \caption{Effect sizes ($\Delta$) for balanced alphas with KL-divergence as the metric ($M=3$)}
  \label{fig:uniform_delta_kl_cat3}
\end{figure*}

\begin{figure*}
  \centering
  \begin{subfigure}[b]{0.24\linewidth}
    \centering
    \includegraphics[width=\linewidth]{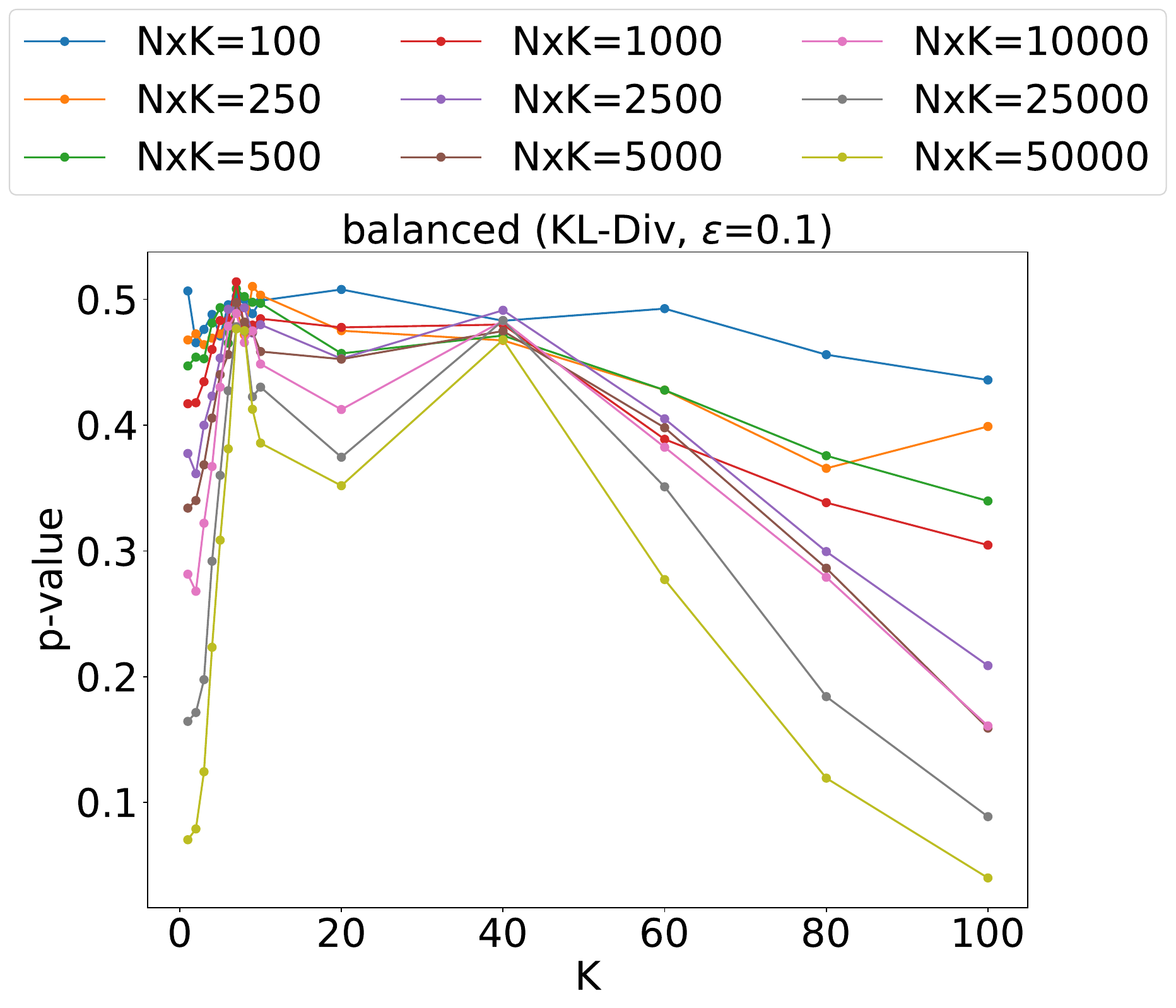}
    \caption{$\epsilon = 0.1$}
    \label{fig:uniform_kl_cat4_e01}
  \end{subfigure} \hfill
  \begin{subfigure}[b]{0.24\linewidth}
    \centering
    \includegraphics[width=\linewidth]{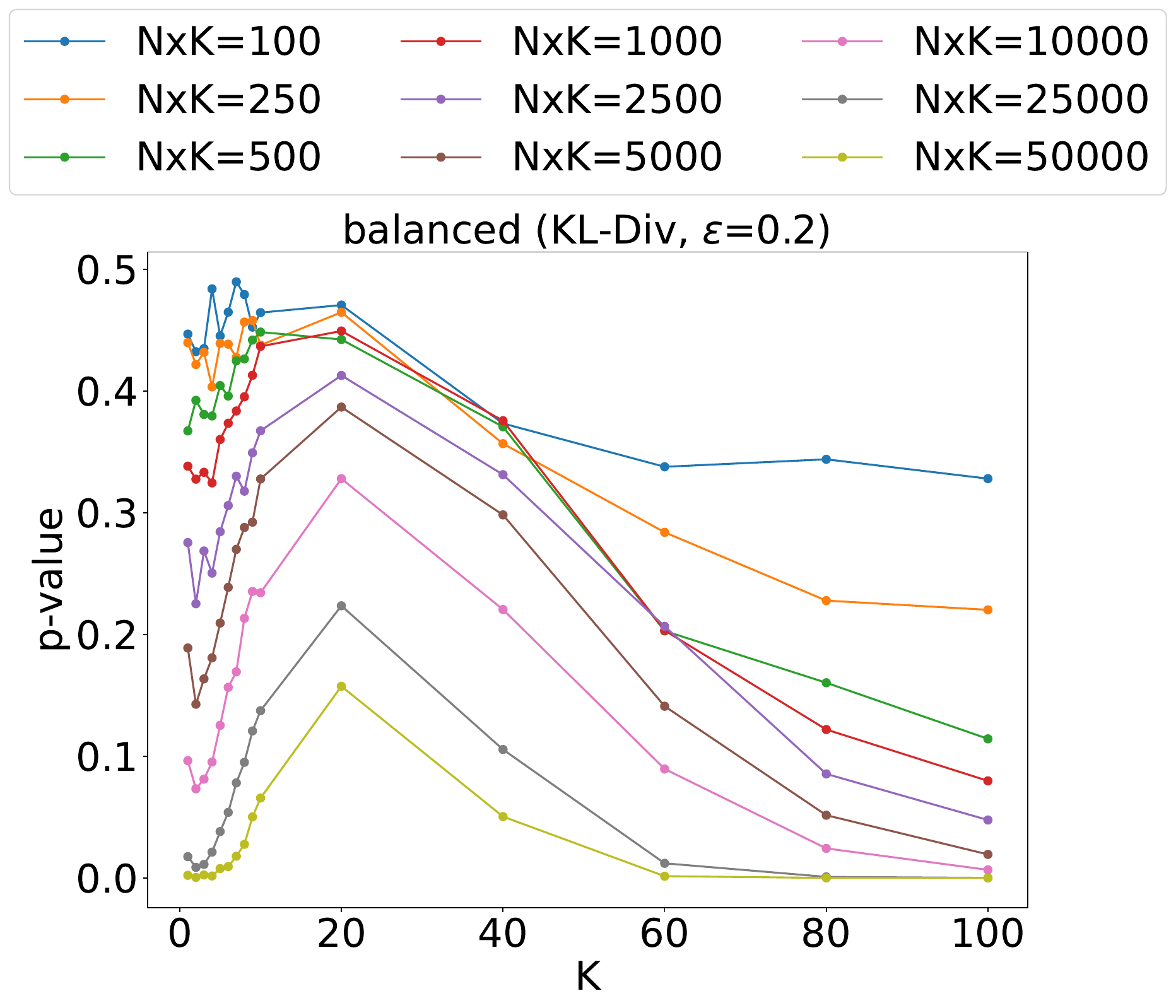}
    \caption{$\epsilon = 0.2$}
    \label{fig:uniform_kl_cat4_e02}
  \end{subfigure} \hfill
  \begin{subfigure}[b]{0.24\linewidth}
    \centering
    \includegraphics[width=\linewidth]{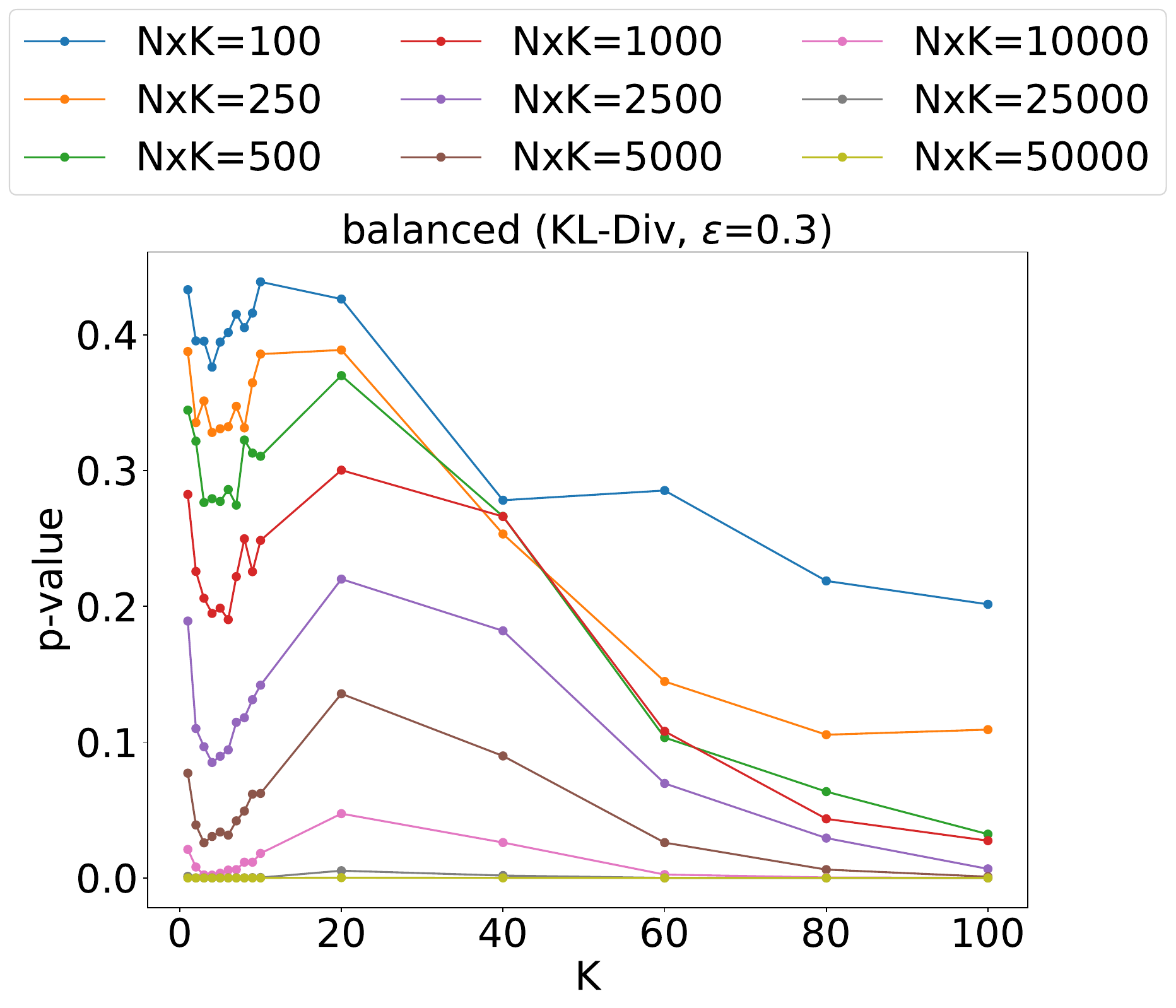}
    \caption{$\epsilon = 0.3$}
    \label{fig:uniform_kl_cat4_e03}
  \end{subfigure} \hfill
  \begin{subfigure}[b]{0.24\linewidth}
    \centering
    \includegraphics[width=\linewidth]{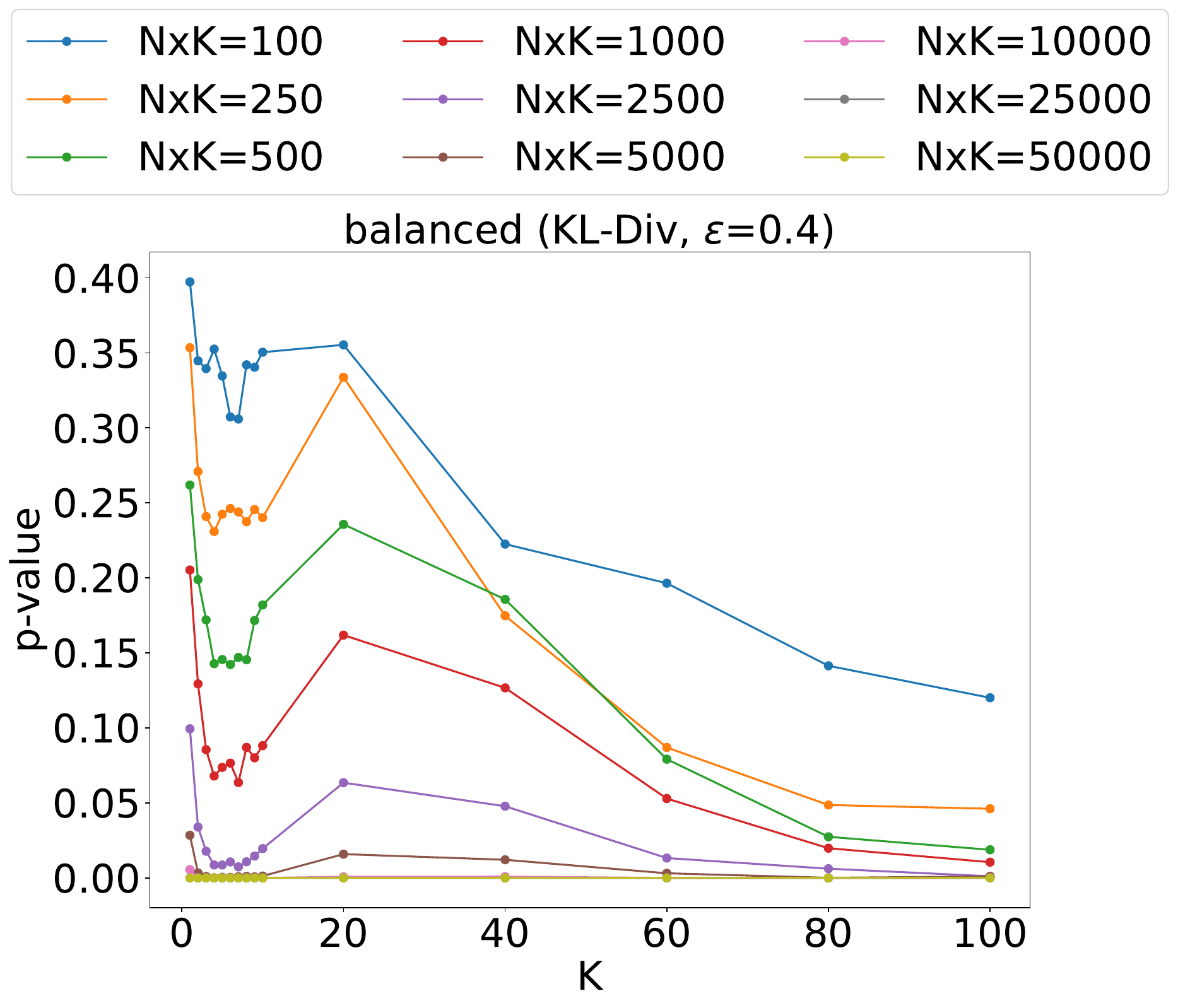}
    \caption{$\epsilon = 0.4$}
    \label{fig:uniform_kl_cat4_e04}
  \end{subfigure}
  \caption{P-value plots for balanced alphas with KL-divergence as the metric ($M=4$)}
  \label{fig:uniform_kl_cat4}
\end{figure*}

\begin{figure*}
  \centering
  \begin{subfigure}[b]{0.24\linewidth}
    \centering
    \includegraphics[width=\linewidth]{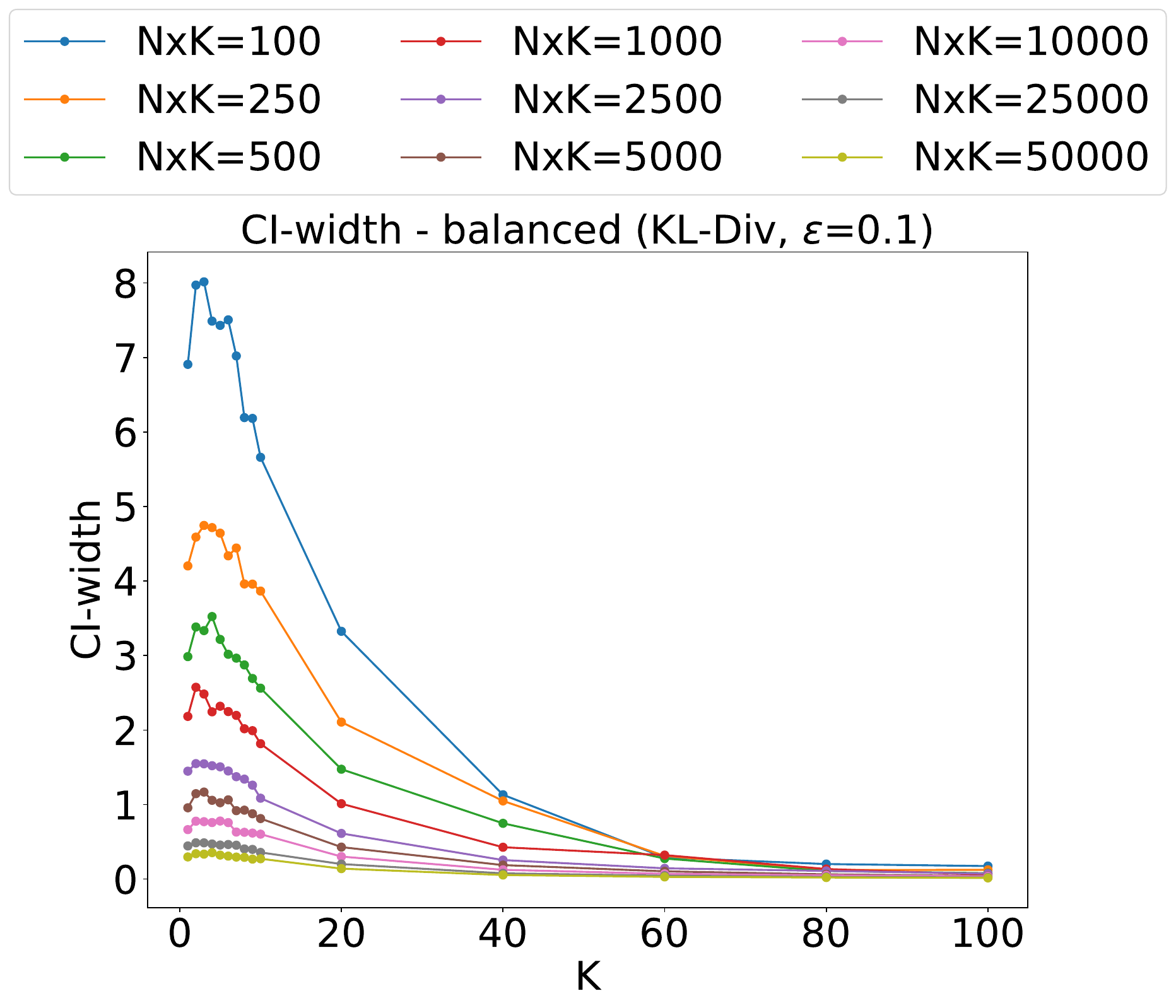}
    \caption{$\epsilon = 0.1$}
    \label{fig:uniform_ci_kl_cat4_e01}
  \end{subfigure} \hfill
  \begin{subfigure}[b]{0.24\linewidth}
    \centering
    \includegraphics[width=\linewidth]{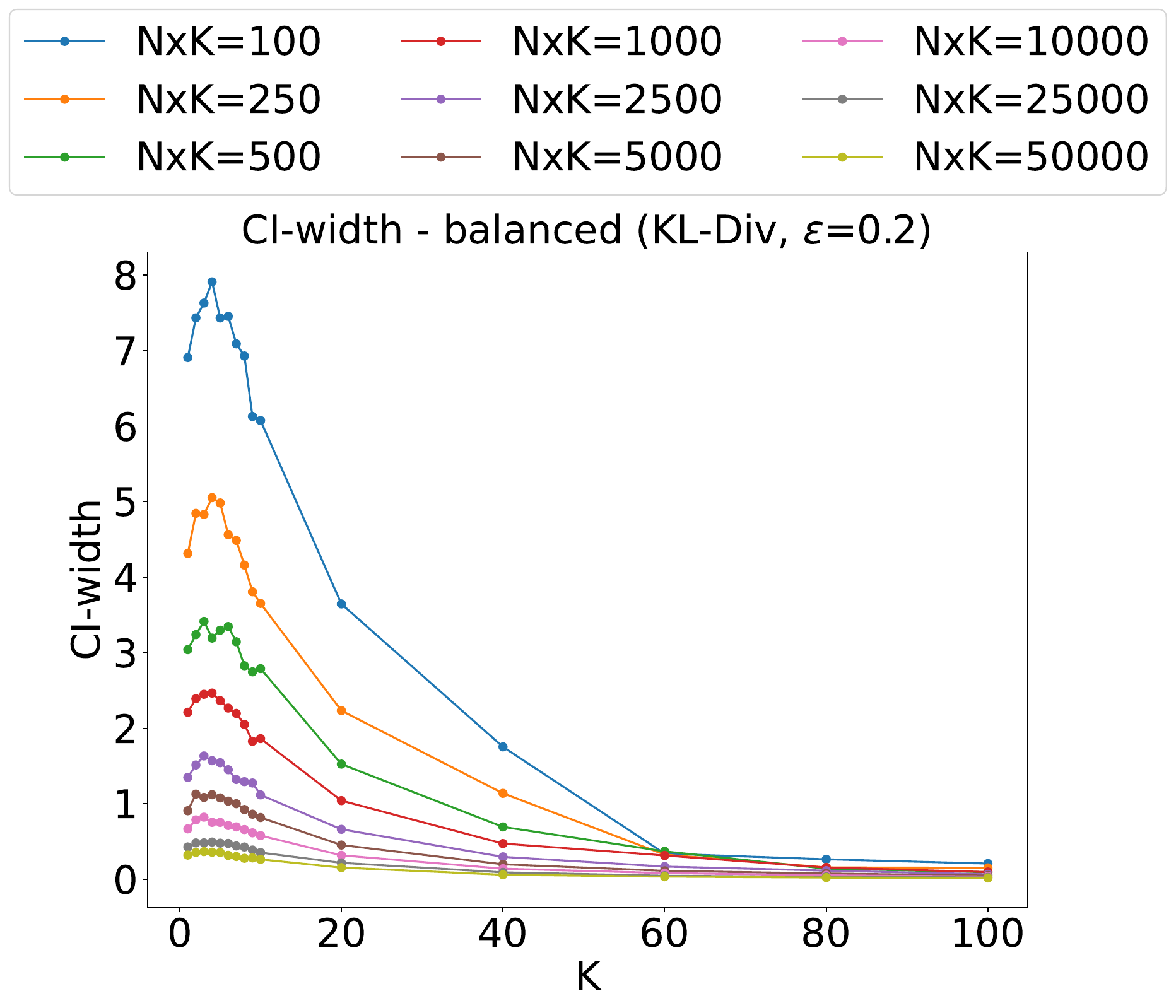}
    \caption{$\epsilon = 0.2$}
    \label{fig:uniform_ci_kl_cat4_e02}
  \end{subfigure} \hfill
  \begin{subfigure}[b]{0.24\linewidth}
    \centering
    \includegraphics[width=\linewidth]{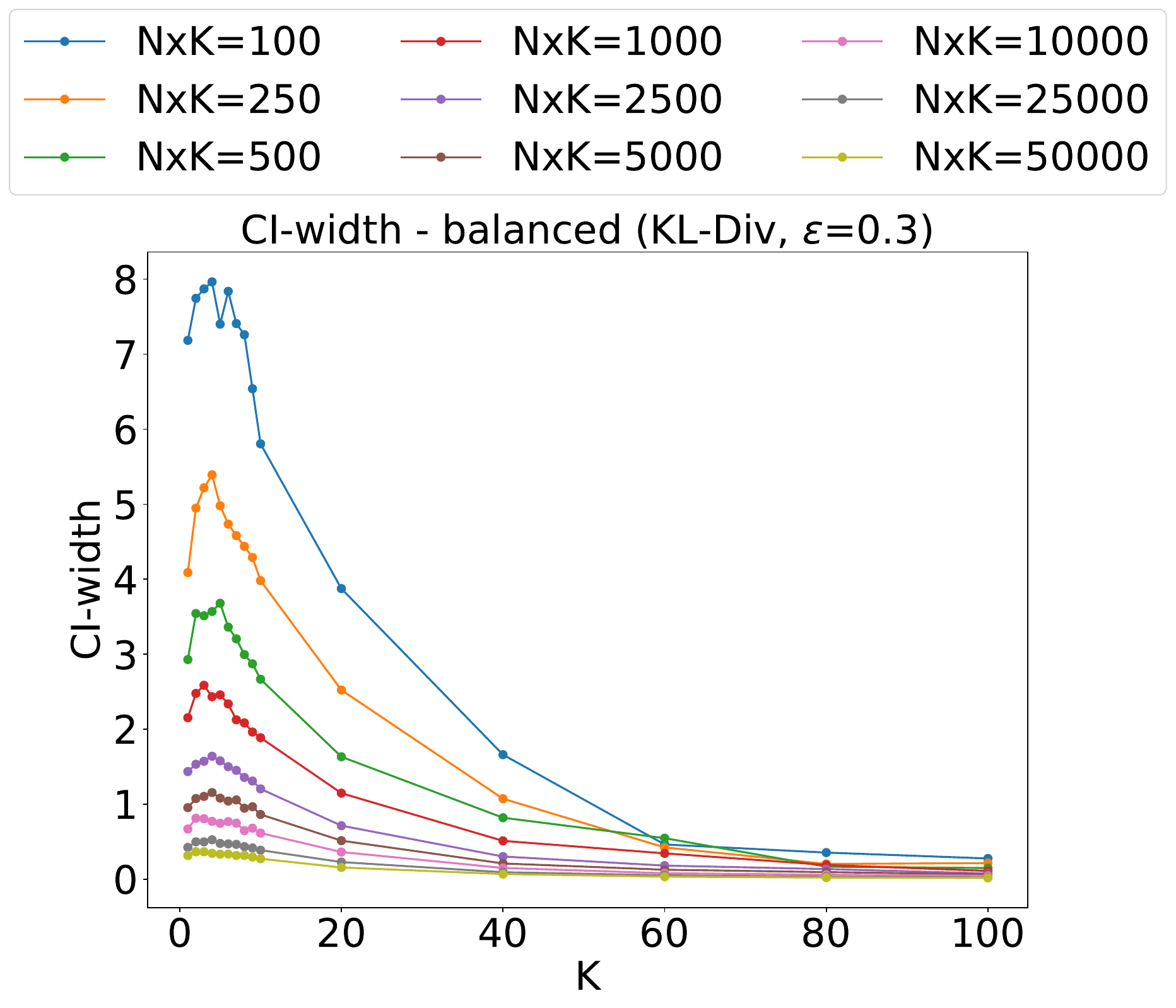}
    \caption{$\epsilon = 0.3$}
    \label{fig:uniform_ci_kl_cat4_e03}
  \end{subfigure} \hfill
  \begin{subfigure}[b]{0.24\linewidth}
    \centering
    \includegraphics[width=\linewidth]{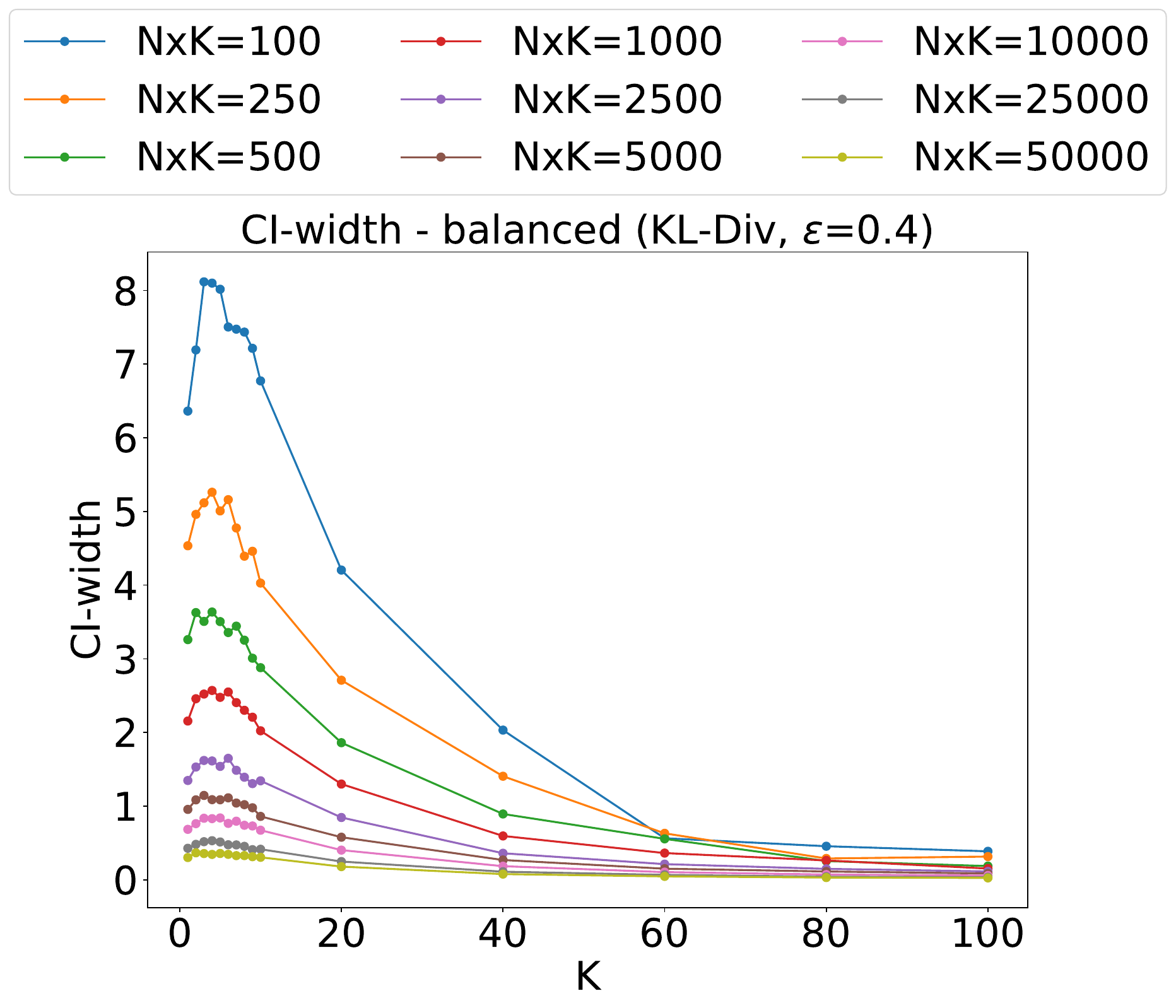}
    \caption{$\epsilon = 0.4$}
    \label{fig:uniform_ci_kl_cat4_e04}
  \end{subfigure}
  \caption{CI-width plots for balanced alphas with KL-divergence as the metric ($M=4$)}
  \label{fig:uniform_ci_kl_cat4}
\end{figure*}

\begin{figure*}
  \centering
  \begin{subfigure}[b]{0.24\linewidth}
    \centering
    \includegraphics[width=\linewidth]{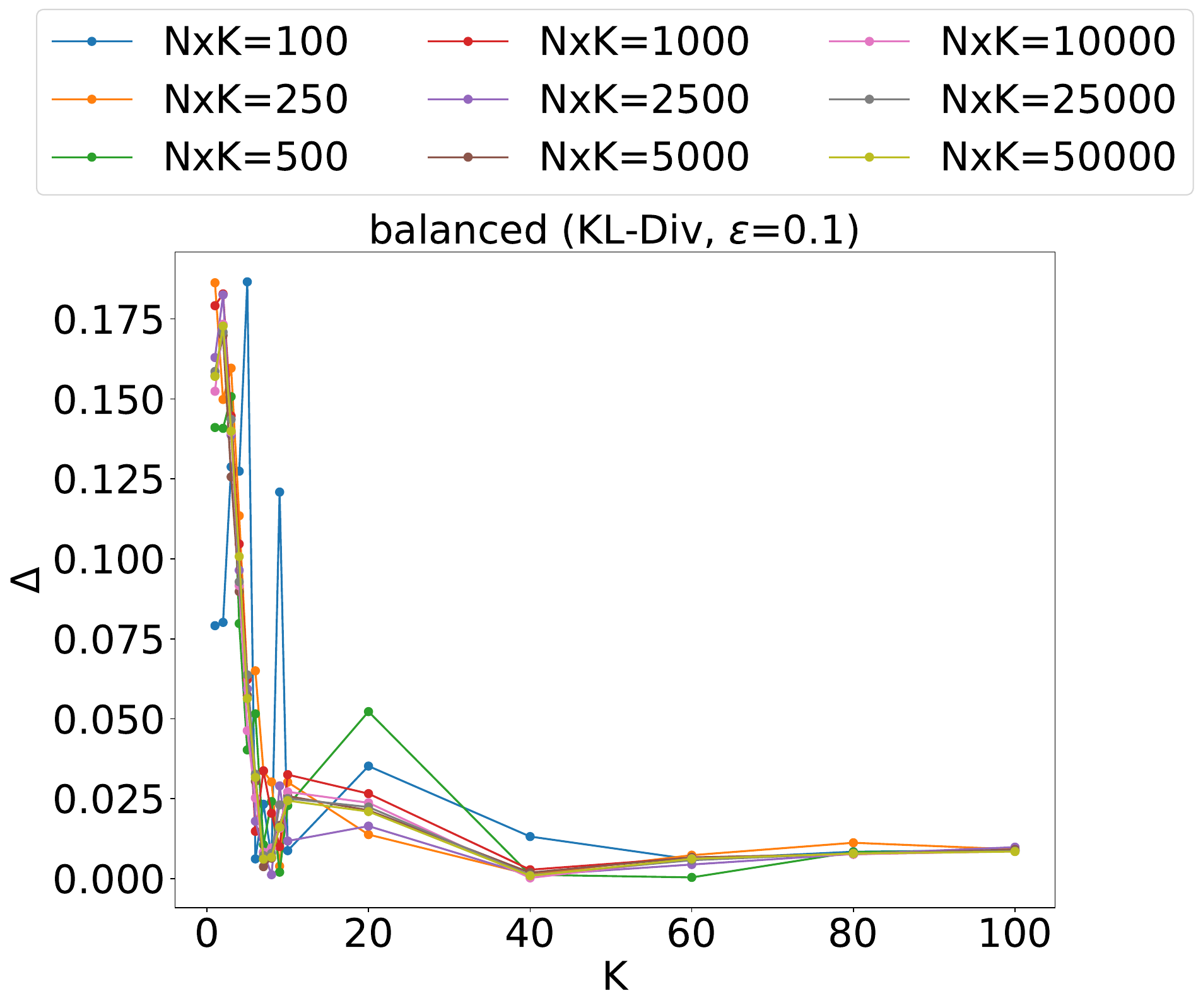}
    \caption{$\epsilon = 0.1$}
    \label{fig:uniform_delta_kl_cat4_e01}
  \end{subfigure} \hfill
  \begin{subfigure}[b]{0.24\linewidth}
    \centering
    \includegraphics[width=\linewidth]{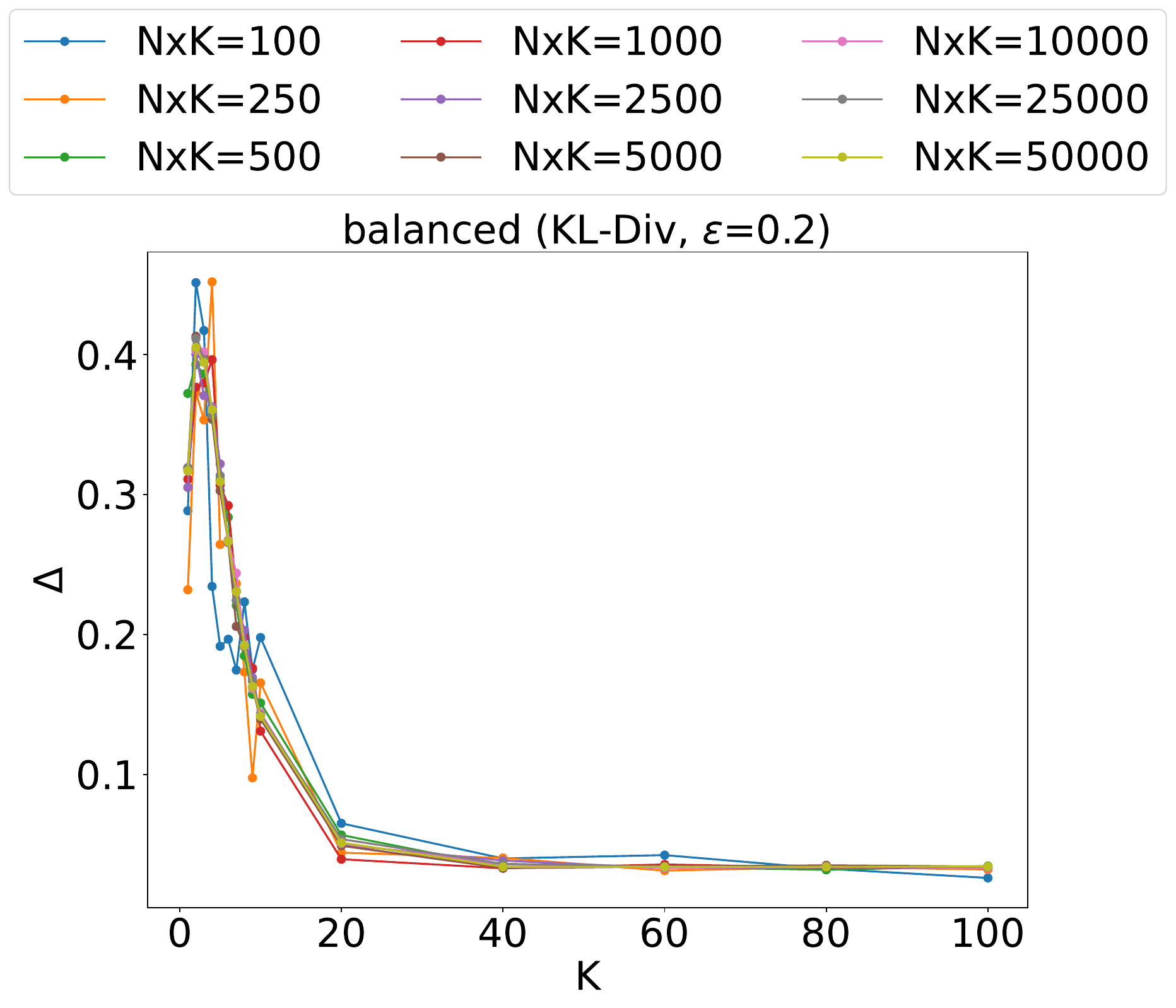}
    \caption{$\epsilon = 0.2$}
    \label{fig:uniform_delta_kl_cat4_e02}
  \end{subfigure} \hfill
  \begin{subfigure}[b]{0.24\linewidth}
    \centering
    \includegraphics[width=\linewidth]{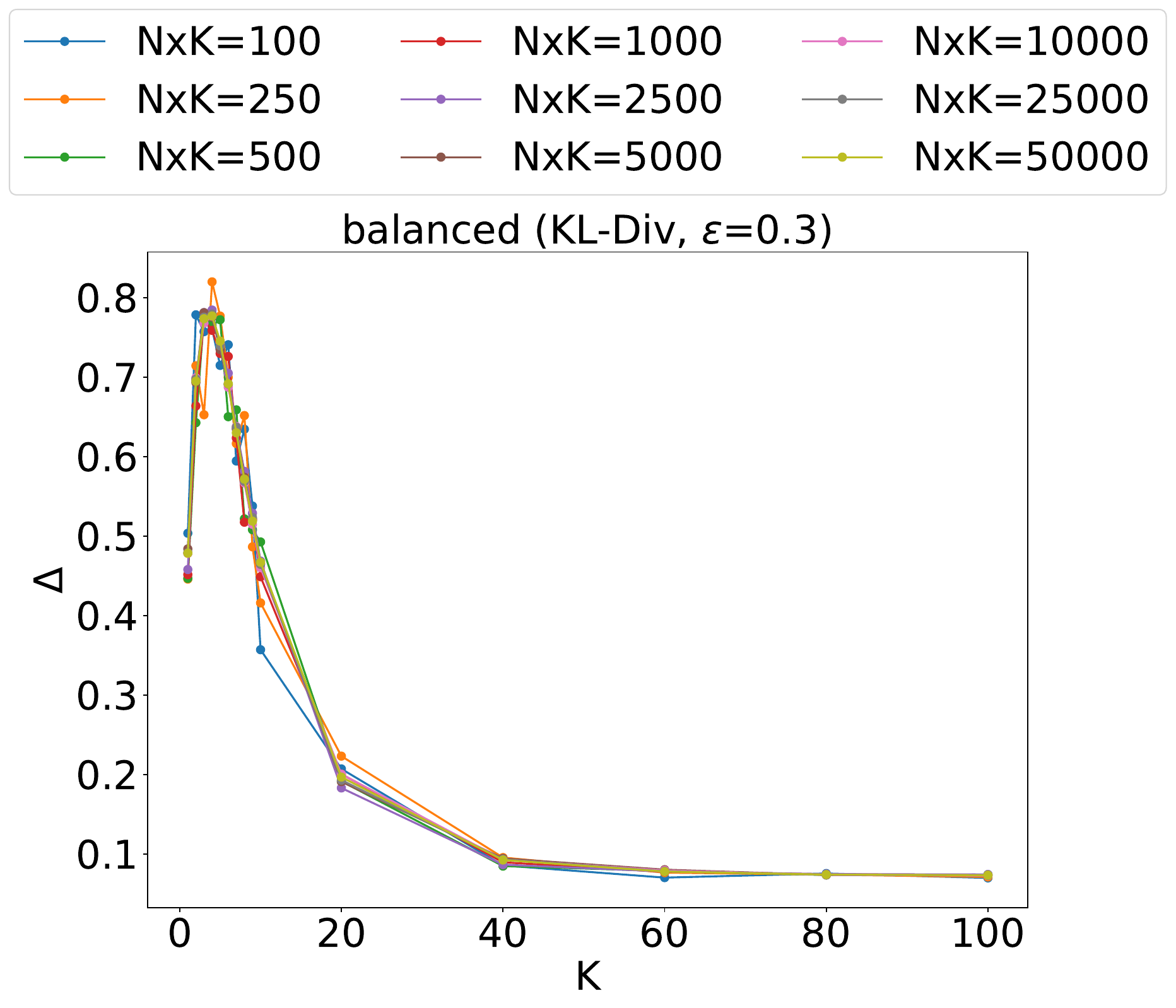}
    \caption{$\epsilon = 0.3$}
    \label{fig:uniform_delta_kl_cat4_e03}
  \end{subfigure} \hfill
  \begin{subfigure}[b]{0.24\linewidth}
    \centering
    \includegraphics[width=\linewidth]{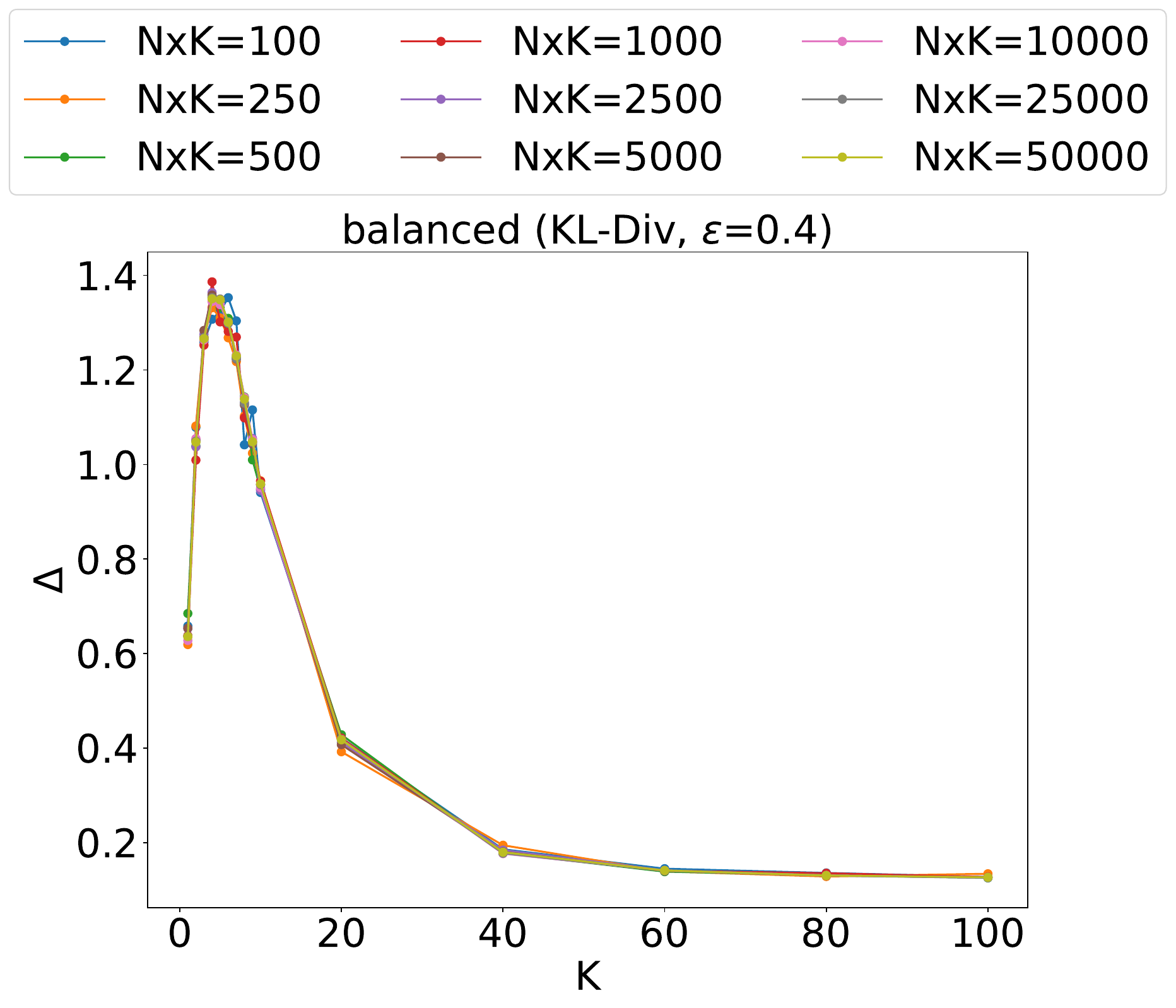}
    \caption{$\epsilon = 0.4$}
    \label{fig:uniform_delta_kl_cat4_e04}
  \end{subfigure}
  \caption{Effect sizes ($\Delta$) for balanced alphas with KL-divergence as the metric ($M=4$)}
  \label{fig:uniform_delta_kl_cat4}
\end{figure*}

\begin{figure*}
  \centering
  \begin{subfigure}[b]{0.24\linewidth}
    \centering
    \includegraphics[width=\linewidth]{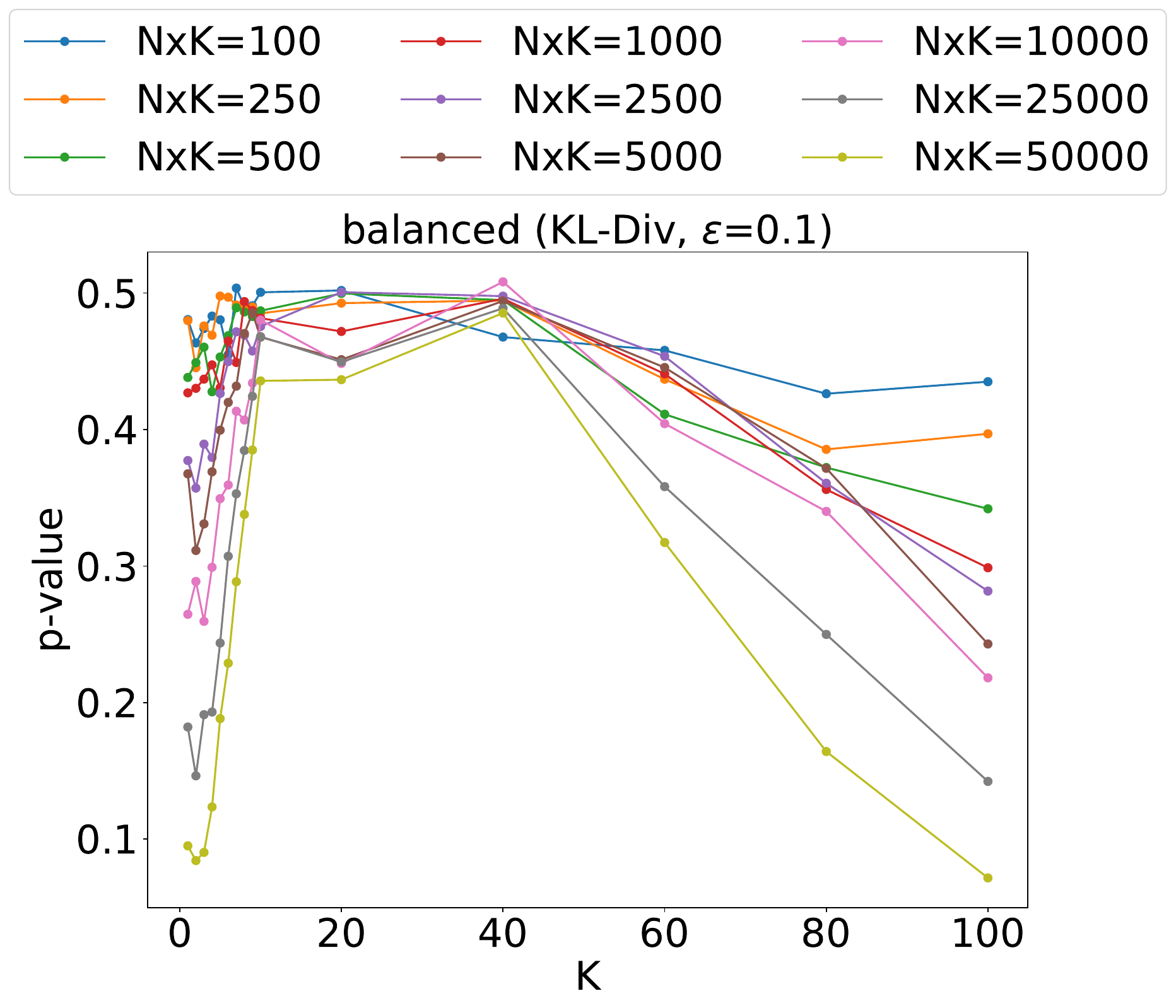}
    \caption{$\epsilon = 0.1$}
    \label{fig:uniform_kl_cat5_e01}
  \end{subfigure} \hfill
  \begin{subfigure}[b]{0.24\linewidth}
    \centering
    \includegraphics[width=\linewidth]{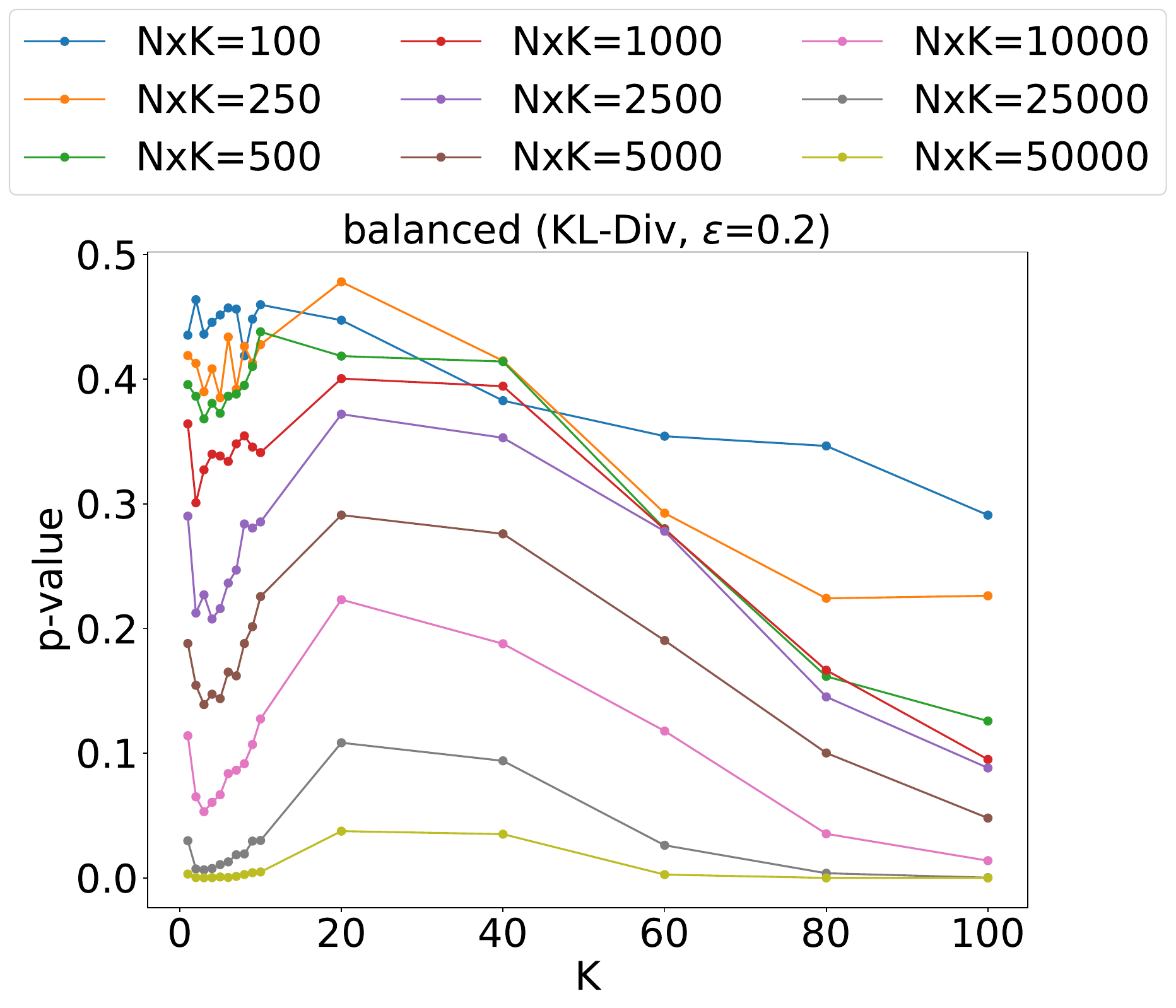}
    \caption{$\epsilon = 0.2$}
    \label{fig:uniform_kl_cat5_e02}
  \end{subfigure} \hfill
  \begin{subfigure}[b]{0.24\linewidth}
    \centering
    \includegraphics[width=\linewidth]{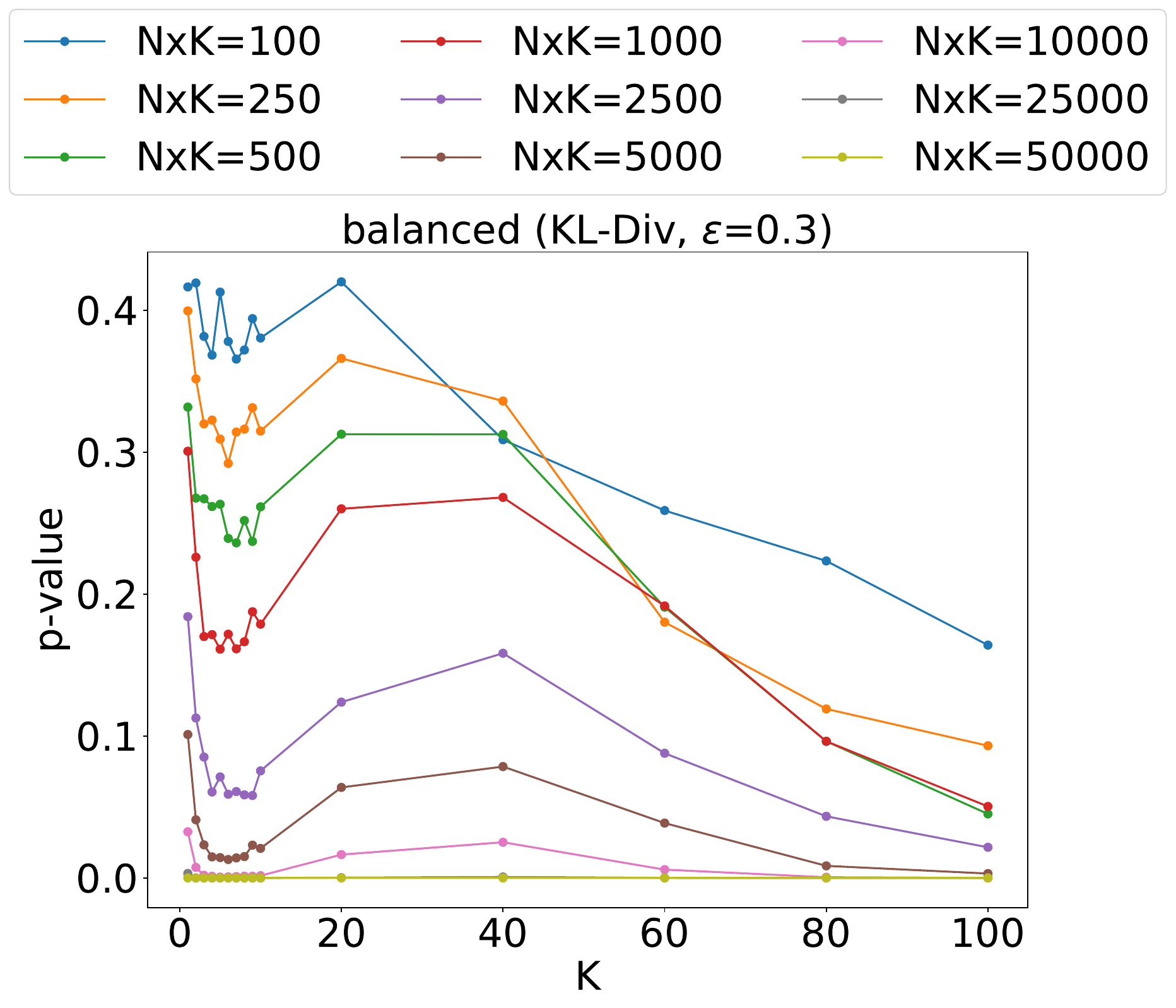}
    \caption{$\epsilon = 0.3$}
    \label{fig:uniform_kl_cat5_e03}
  \end{subfigure} \hfill
  \begin{subfigure}[b]{0.24\linewidth}
    \centering
    \includegraphics[width=\linewidth]{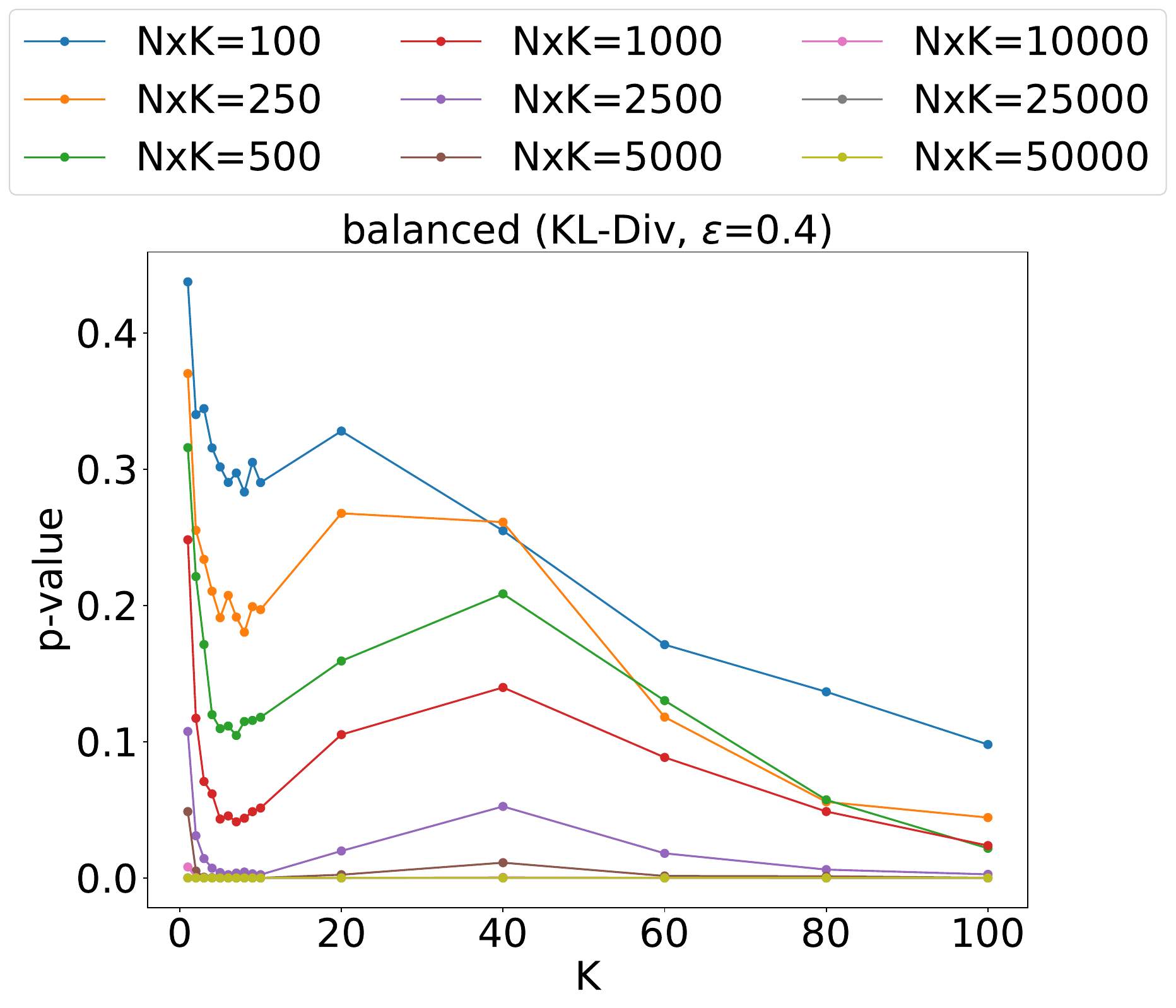}
    \caption{$\epsilon = 0.4$}
    \label{fig:uniform_kl_cat5_e04}
  \end{subfigure}
  \caption{P-value plots for balanced alphas with KL-divergence as the metric ($M=5$)}
  \label{fig:uniform_kl_cat5}
\end{figure*}

\begin{figure*}
  \centering
  \begin{subfigure}[b]{0.24\linewidth}
    \centering
    \includegraphics[width=\linewidth]{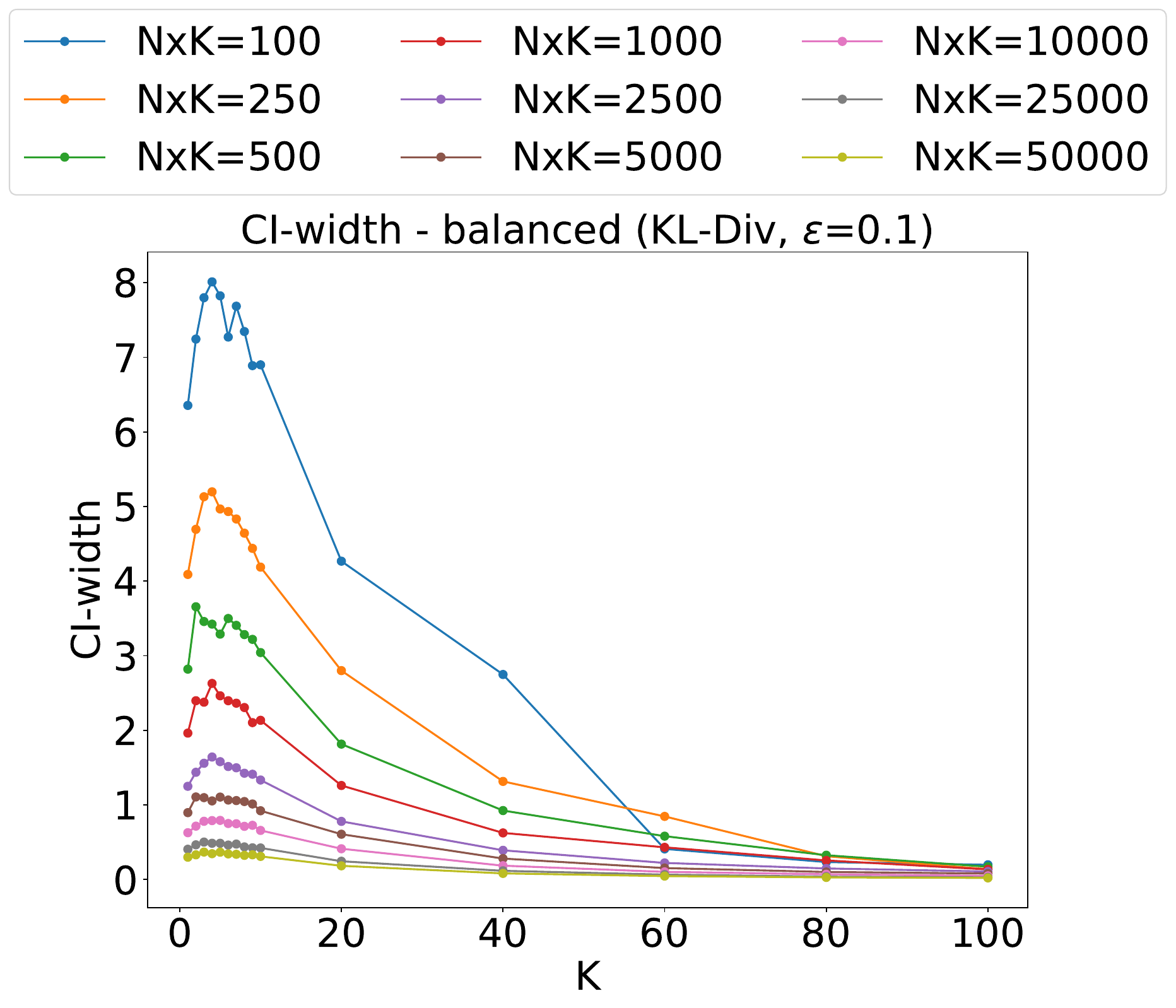}
    \caption{$\epsilon = 0.1$}
    \label{fig:uniform_ci_kl_cat5_e01}
  \end{subfigure} \hfill
  \begin{subfigure}[b]{0.24\linewidth}
    \centering
    \includegraphics[width=\linewidth]{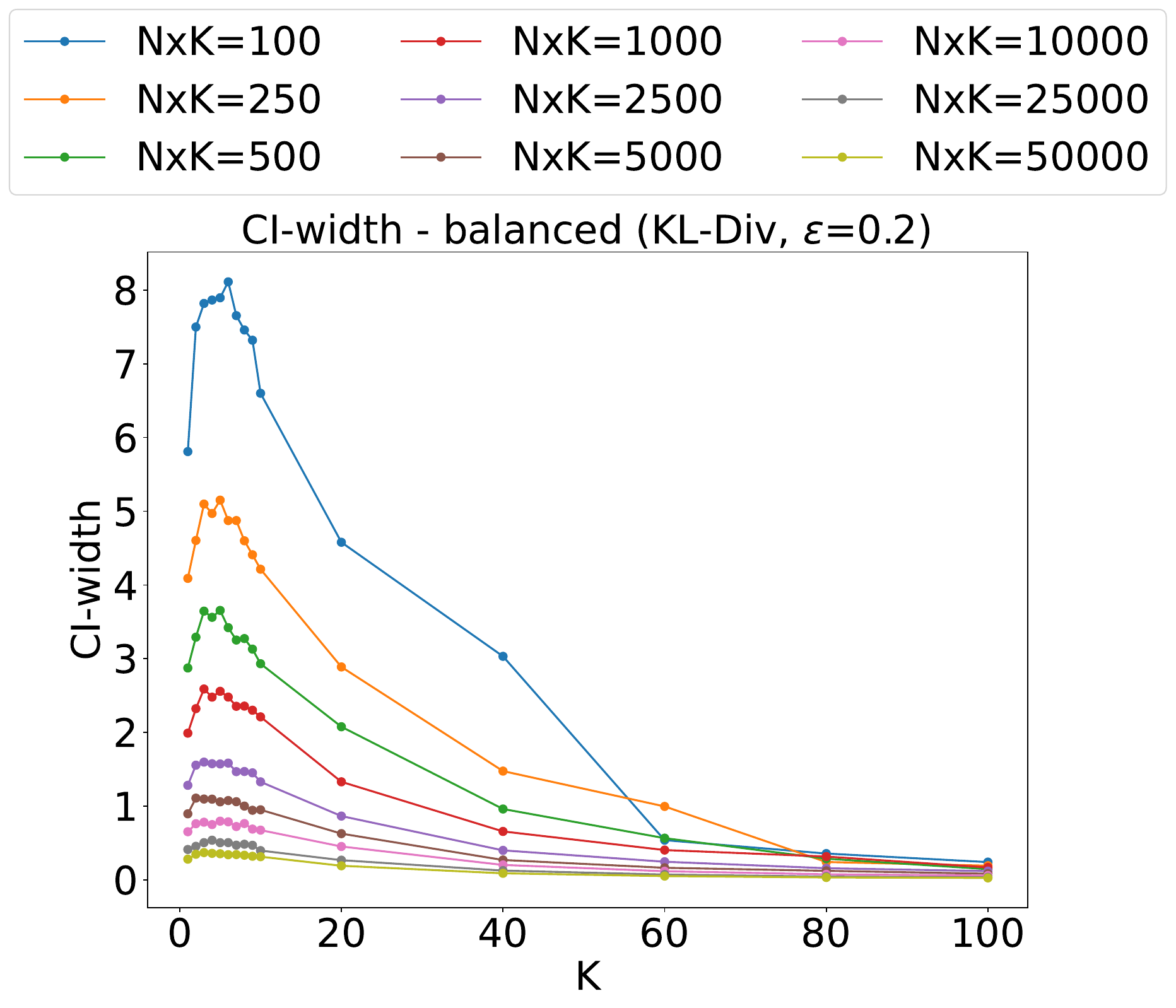}
    \caption{$\epsilon = 0.2$}
    \label{fig:uniform_ci_kl_cat5_e02}
  \end{subfigure} \hfill
  \begin{subfigure}[b]{0.24\linewidth}
    \centering
    \includegraphics[width=\linewidth]{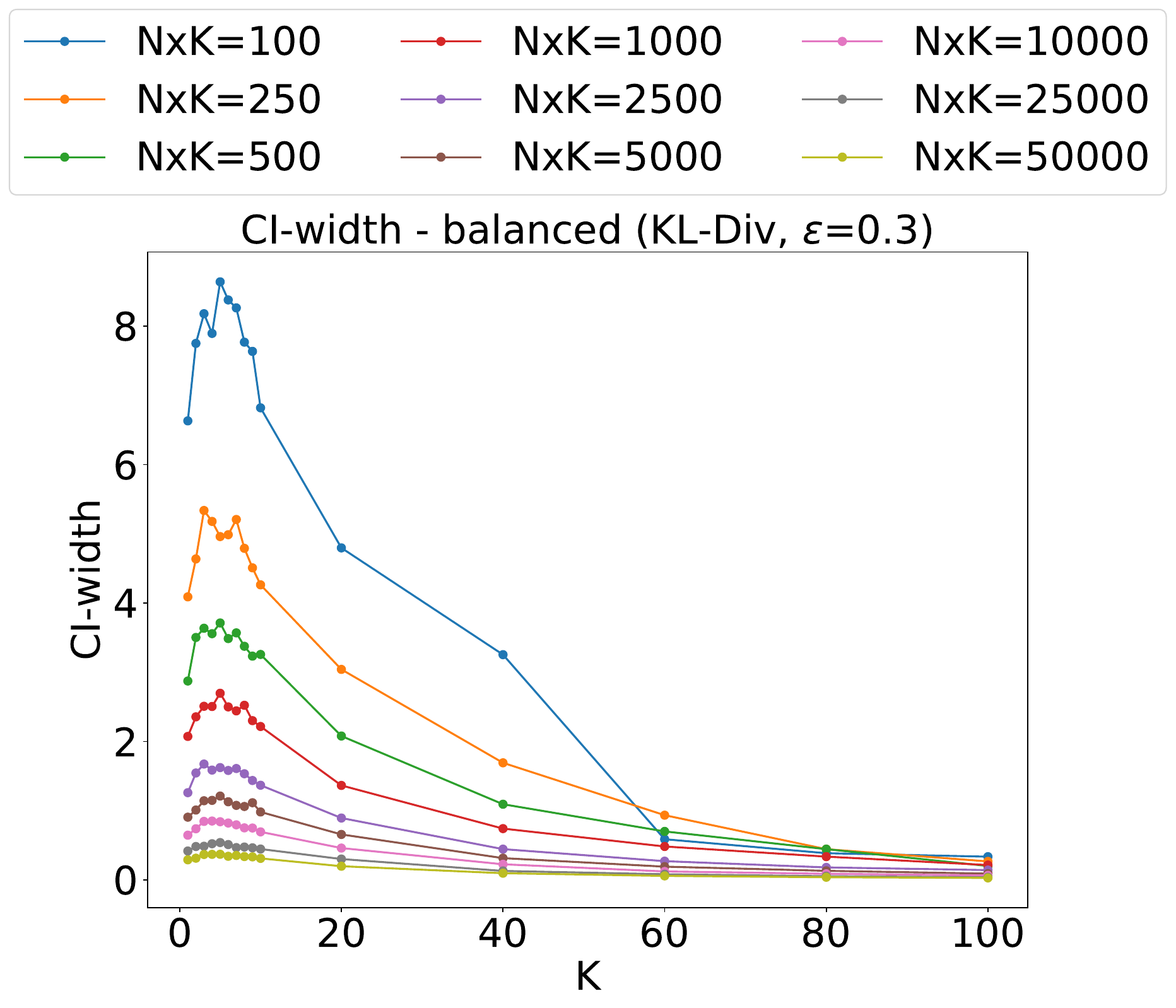}
    \caption{$\epsilon = 0.3$}
    \label{fig:uniform_ci_kl_cat5_e03}
  \end{subfigure} \hfill
  \begin{subfigure}[b]{0.24\linewidth}
    \centering
    \includegraphics[width=\linewidth]{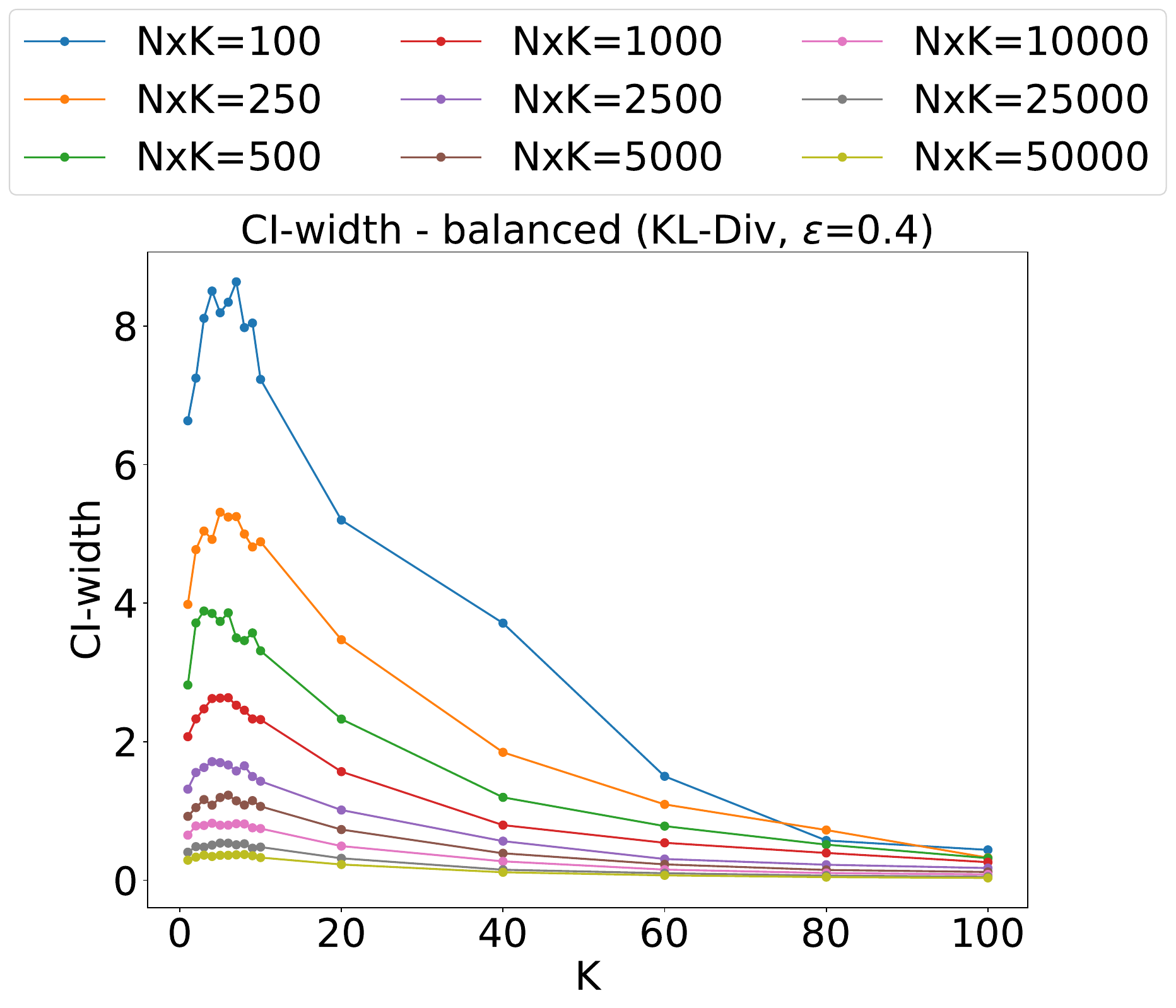}
    \caption{$\epsilon = 0.4$}
    \label{fig:uniform_ci_kl_cat5_e04}
  \end{subfigure}
  \caption{CI-width plots for balanced alphas with KL-divergence as the metric ($M=5$)}
  \label{fig:uniform_ci_kl_cat5}
\end{figure*}

\begin{figure*}
  \centering
  \begin{subfigure}[b]{0.24\linewidth}
    \centering
    \includegraphics[width=\linewidth]{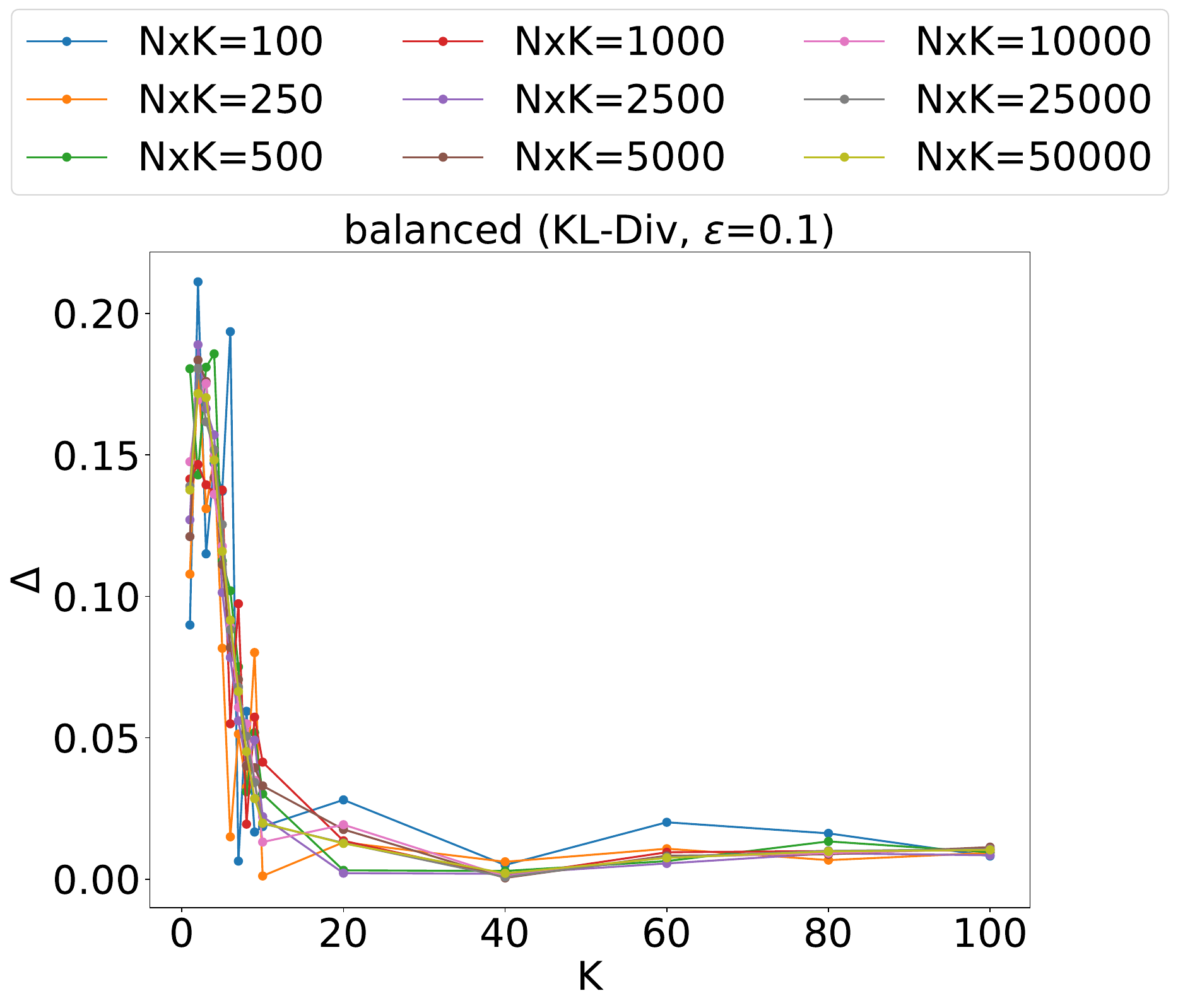}
    \caption{$\epsilon = 0.1$}
    \label{fig:uniform_delta_kl_cat5_e01}
  \end{subfigure} \hfill
  \begin{subfigure}[b]{0.24\linewidth}
    \centering
    \includegraphics[width=\linewidth]{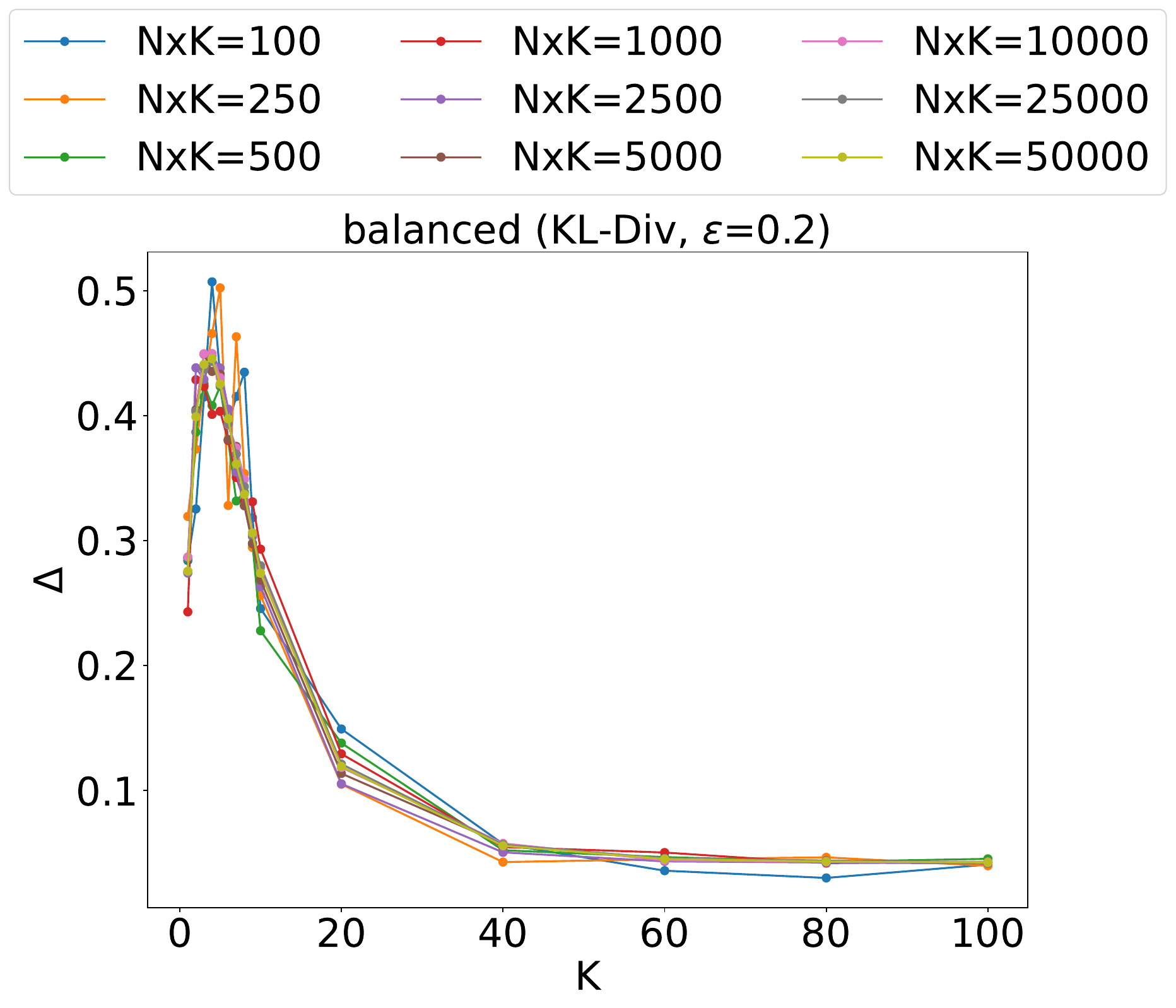}
    \caption{$\epsilon = 0.2$}
    \label{fig:uniform_delta_kl_cat5_e02}
  \end{subfigure} \hfill
  \begin{subfigure}[b]{0.24\linewidth}
    \centering
    \includegraphics[width=\linewidth]{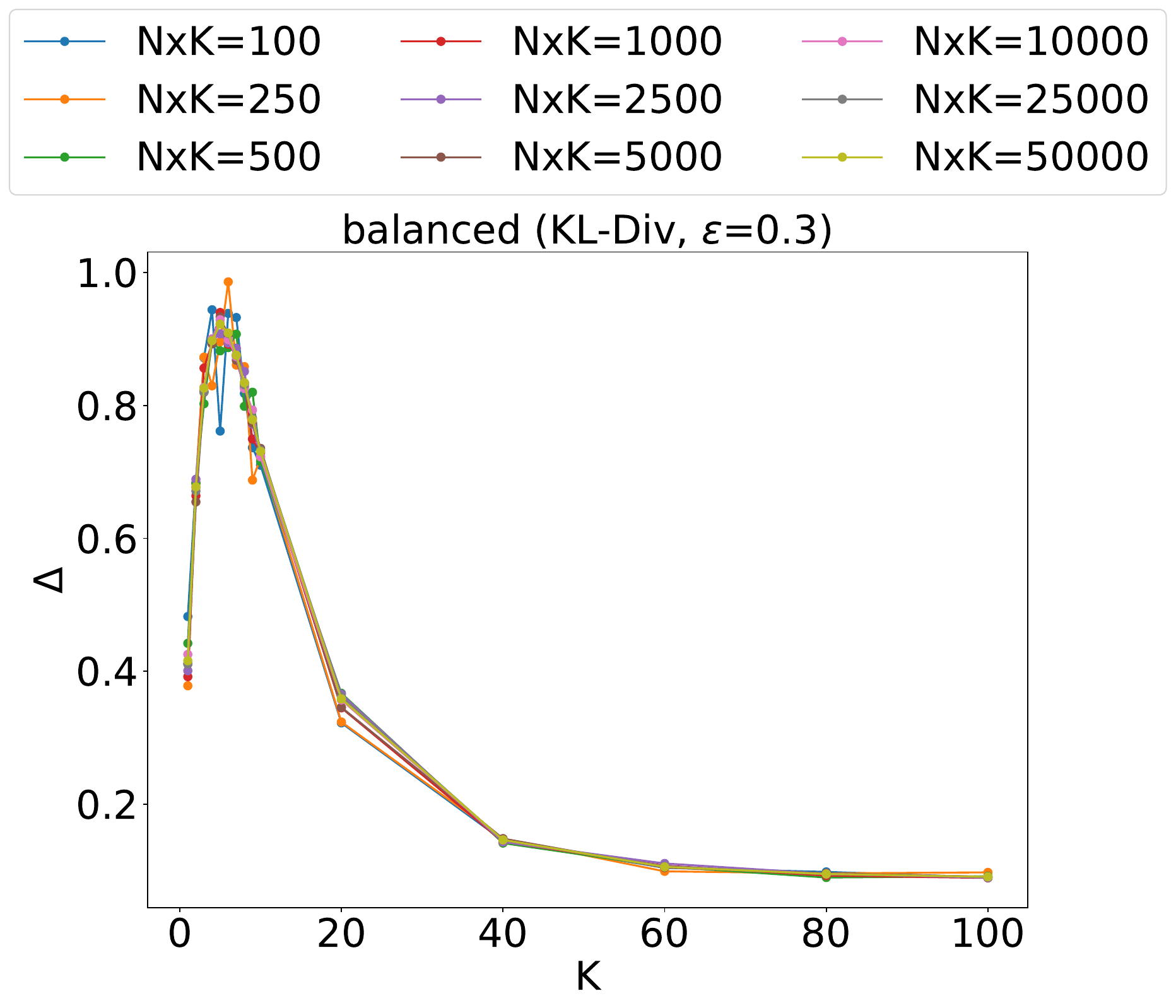}
    \caption{$\epsilon = 0.3$}
    \label{fig:uniform_delta_kl_cat5_e03}
  \end{subfigure} \hfill
  \begin{subfigure}[b]{0.24\linewidth}
    \centering
    \includegraphics[width=\linewidth]{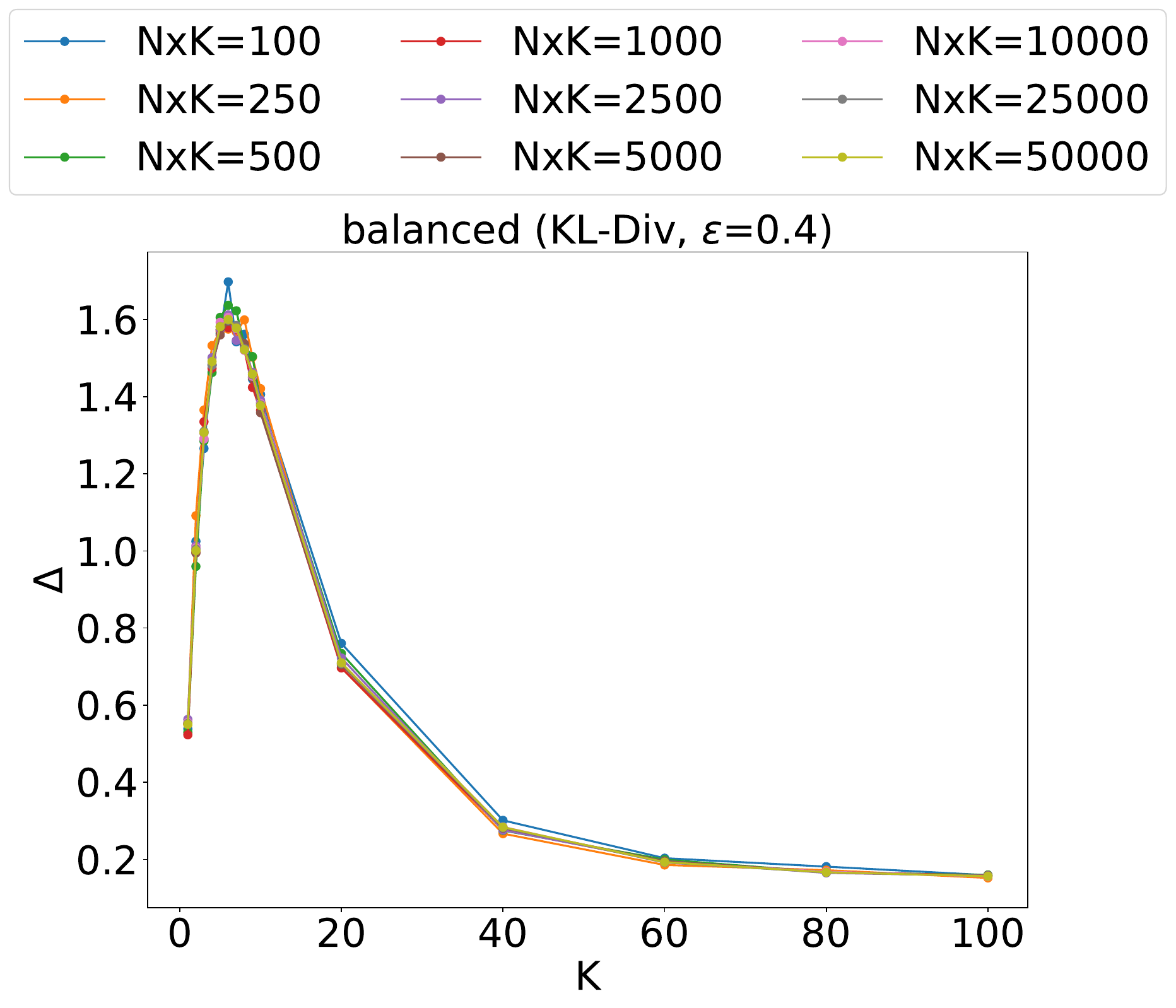}
    \caption{$\epsilon = 0.4$}
    \label{fig:uniform_delta_kl_cat5_e04}
  \end{subfigure}
  \caption{Effect sizes ($\Delta$) for balanced alphas with KL-divergence as the metric ($M=5$)}
  \label{fig:uniform_delta_kl_cat5}
\end{figure*}

\begin{figure*}
  \centering
  \begin{subfigure}[b]{0.24\linewidth}
    \centering
    \includegraphics[width=\linewidth]{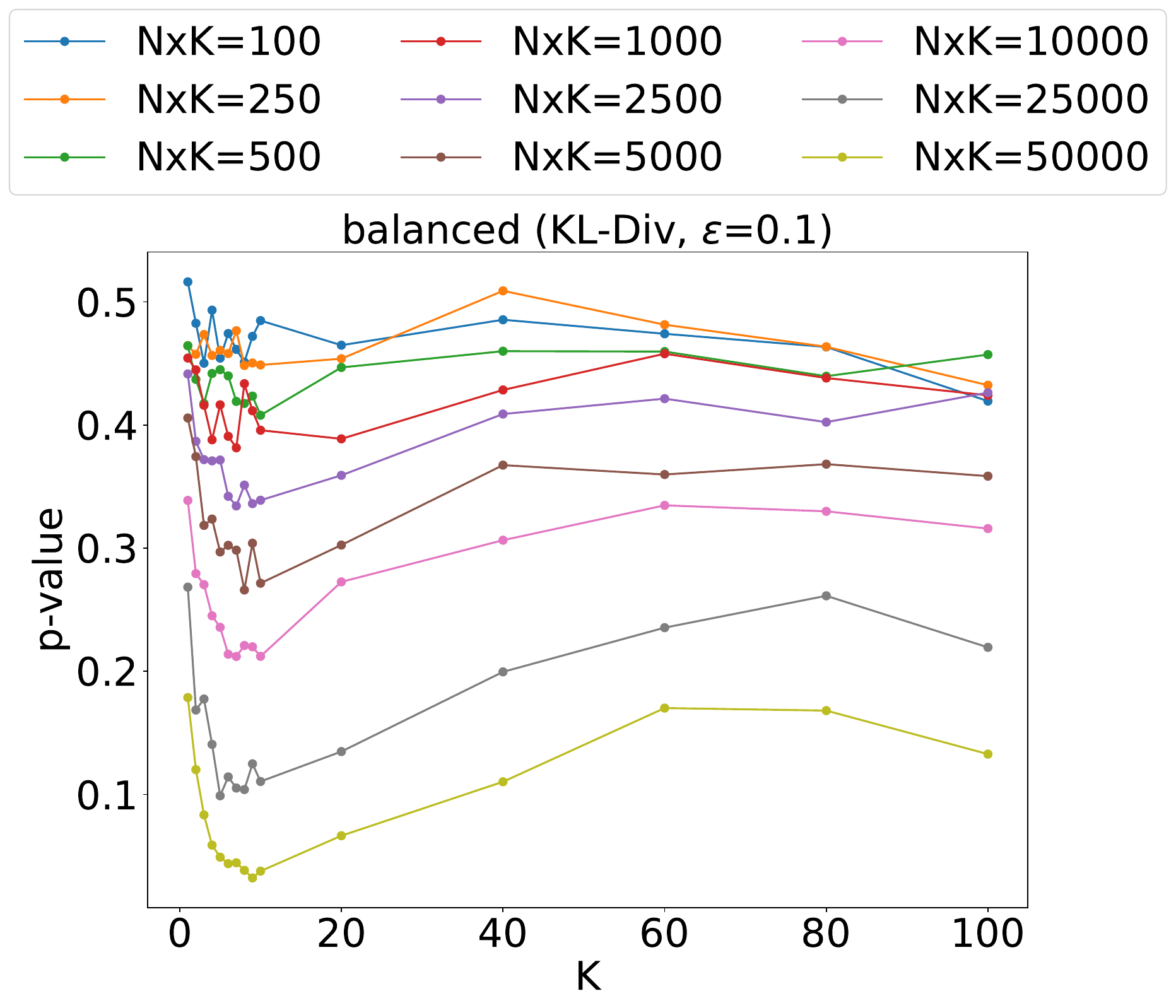}
    \caption{$\epsilon = 0.1$}
    \label{fig:uniform_kl_cat12_e01}
  \end{subfigure} \hfill
  \begin{subfigure}[b]{0.24\linewidth}
    \centering
    \includegraphics[width=\linewidth]{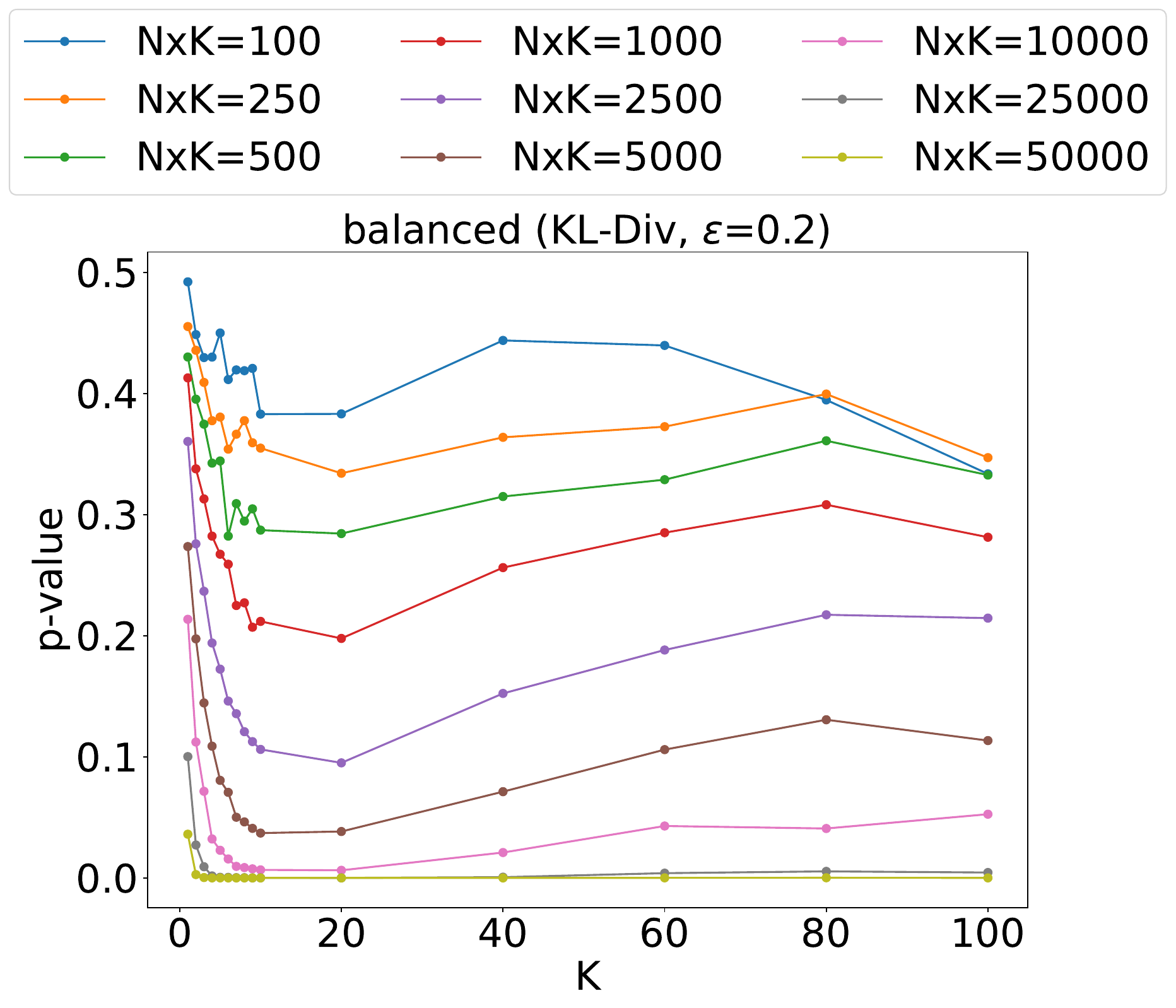}
    \caption{$\epsilon = 0.2$}
    \label{fig:uniform_kl_cat12_e02}
  \end{subfigure} \hfill
  \begin{subfigure}[b]{0.24\linewidth}
    \centering
    \includegraphics[width=\linewidth]{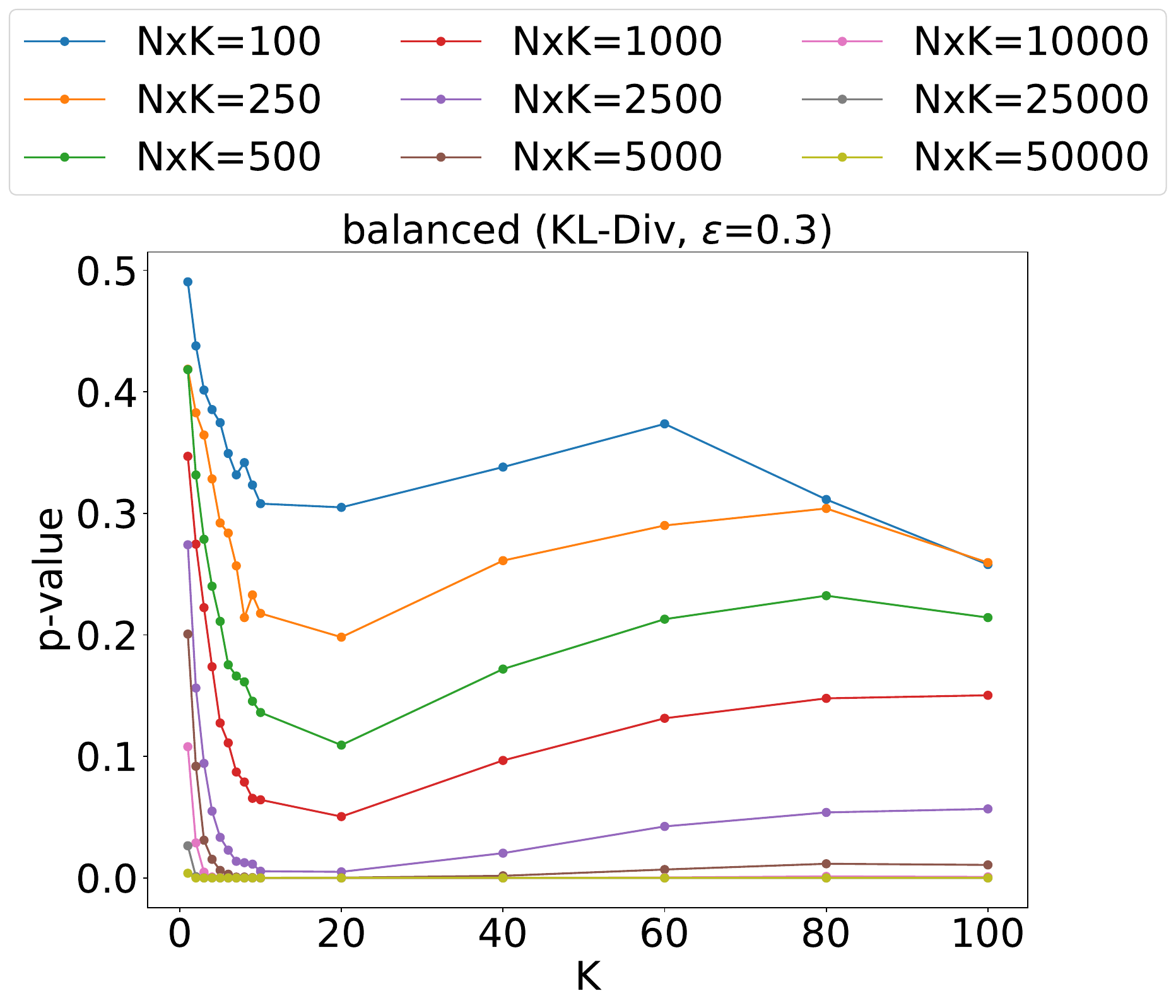}
    \caption{$\epsilon = 0.3$}
    \label{fig:uniform_kl_cat12_e03}
  \end{subfigure} \hfill
  \begin{subfigure}[b]{0.24\linewidth}
    \centering
    \includegraphics[width=\linewidth]{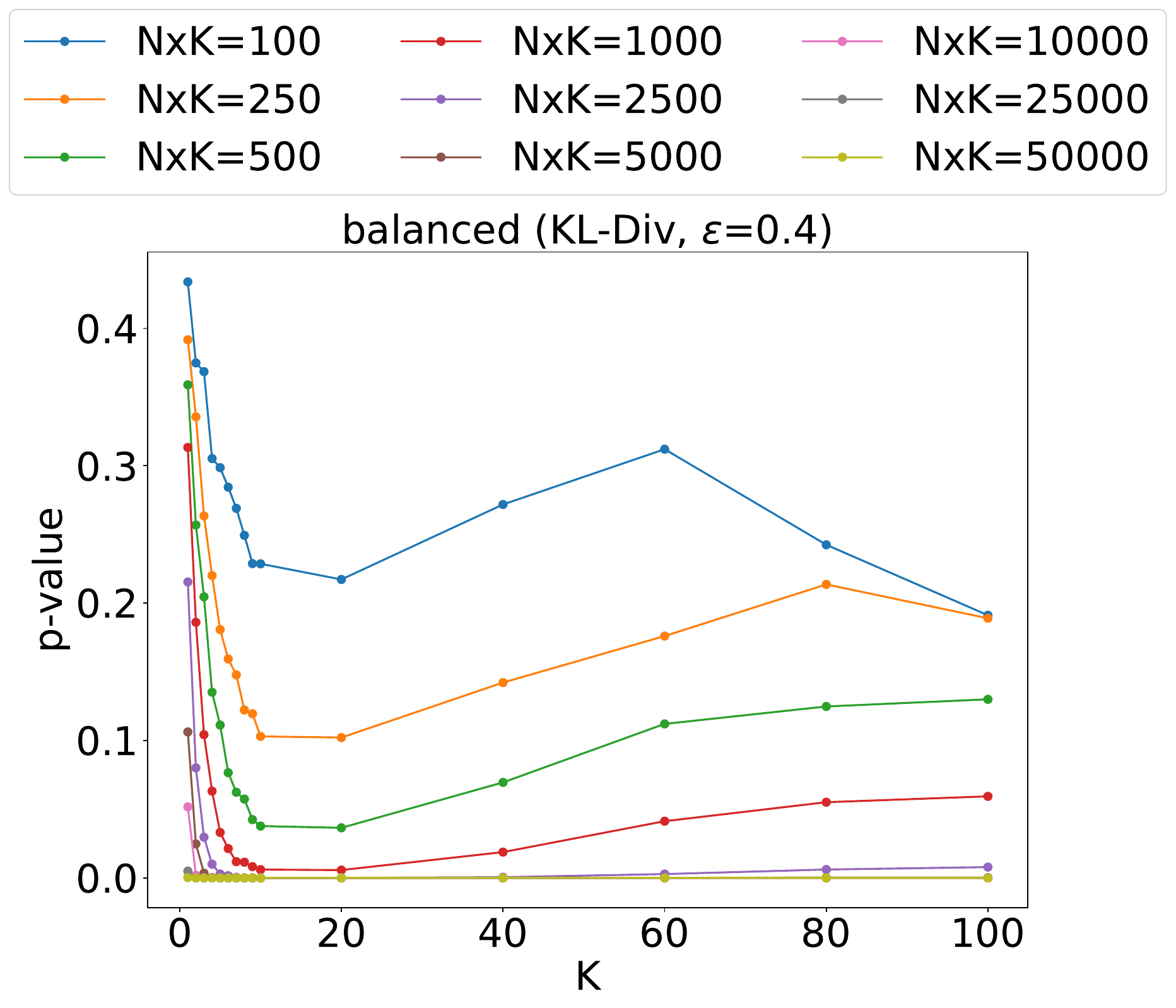}
    \caption{$\epsilon = 0.4$}
    \label{fig:uniform_kl_cat12_e04}
  \end{subfigure}
  \caption{P-value plots for balanced alphas with KL-divergence as the metric ($M=12$)}
  \label{fig:uniform_kl_cat12}
\end{figure*}

\begin{figure*}
  \centering
  \begin{subfigure}[b]{0.24\linewidth}
    \centering
    \includegraphics[width=\linewidth]{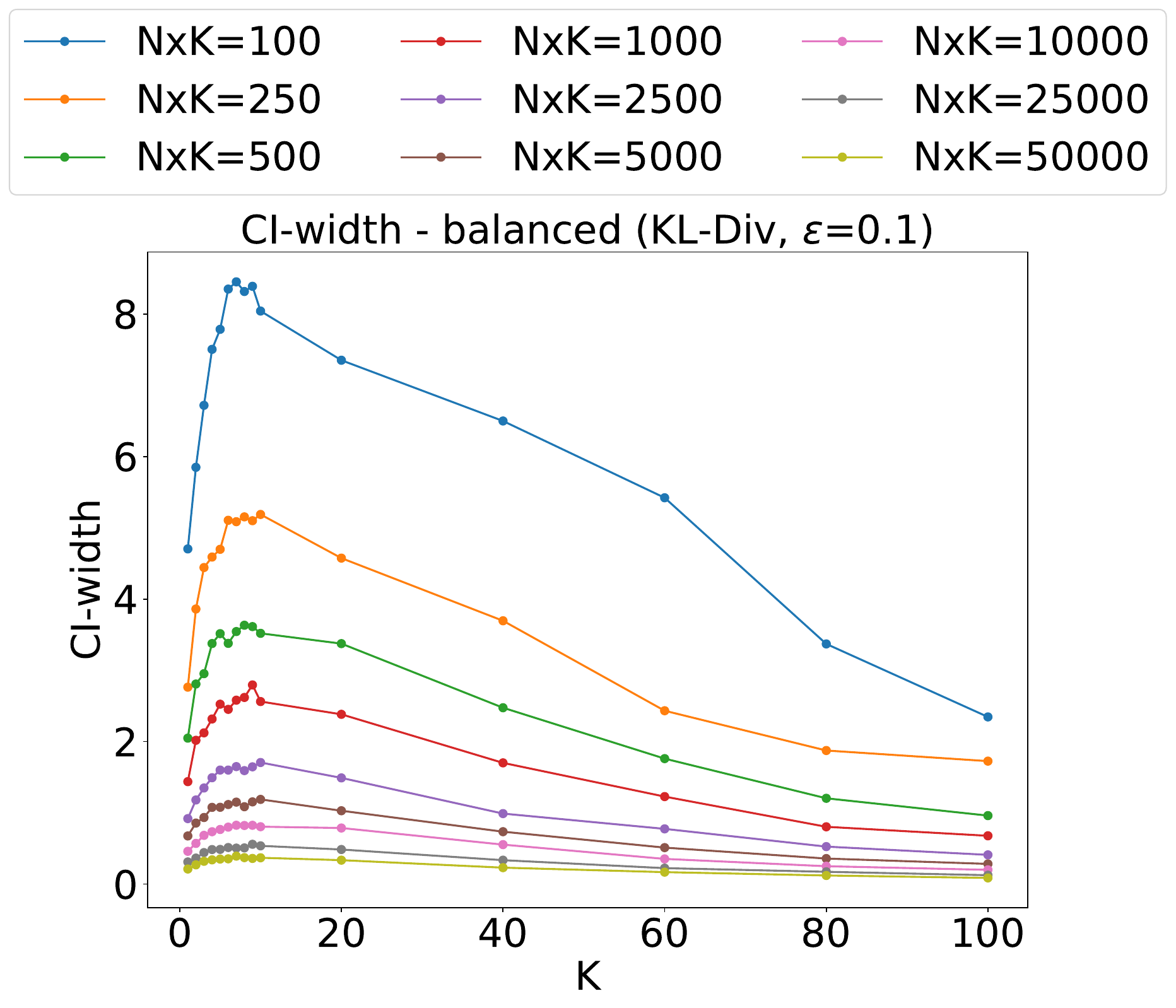}
    \caption{$\epsilon = 0.1$}
    \label{fig:uniform_ci_kl_cat12_e01}
  \end{subfigure} \hfill
  \begin{subfigure}[b]{0.24\linewidth}
    \centering
    \includegraphics[width=\linewidth]{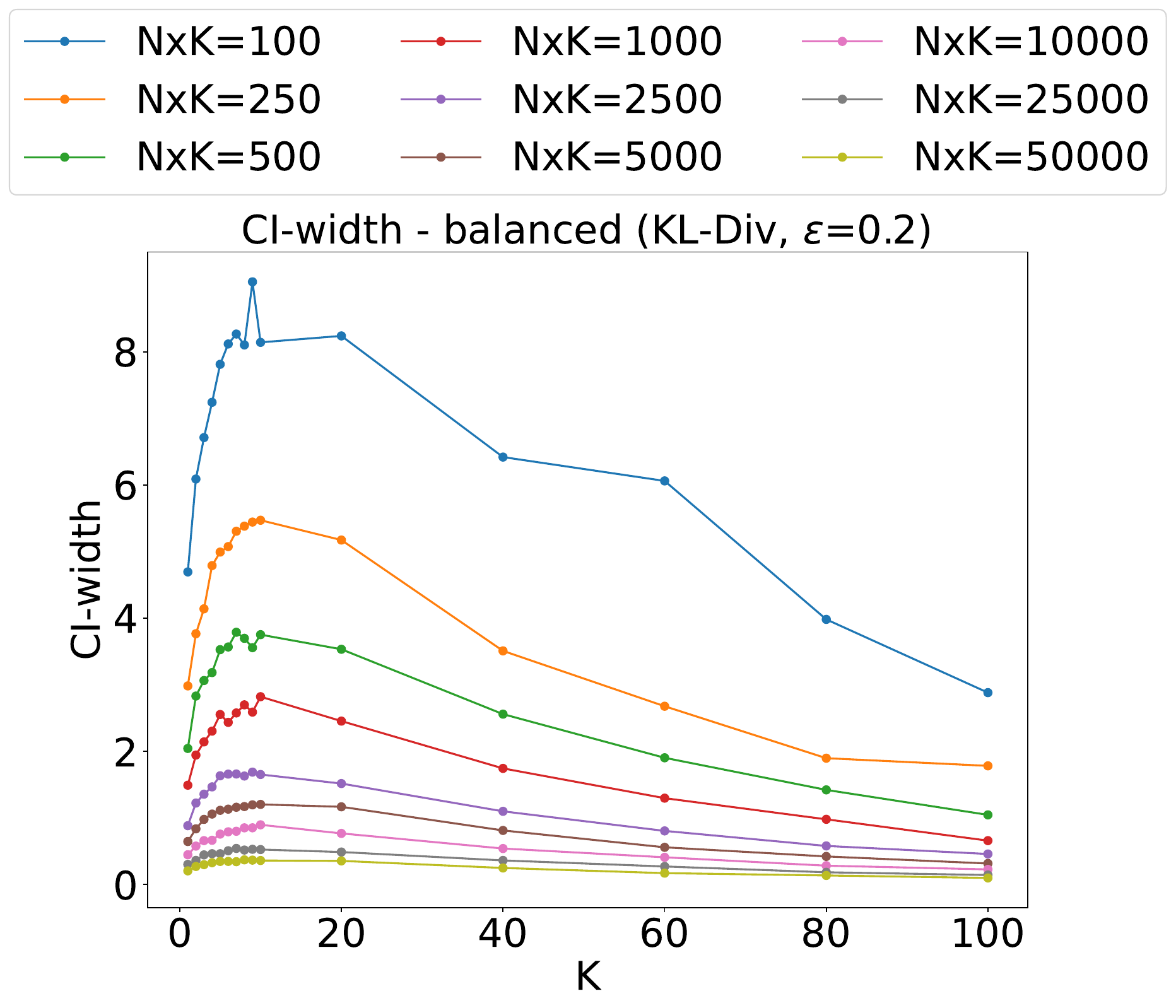}
    \caption{$\epsilon = 0.2$}
    \label{fig:uniform_ci_kl_cat12_e02}
  \end{subfigure} \hfill
  \begin{subfigure}[b]{0.24\linewidth}
    \centering
    \includegraphics[width=\linewidth]{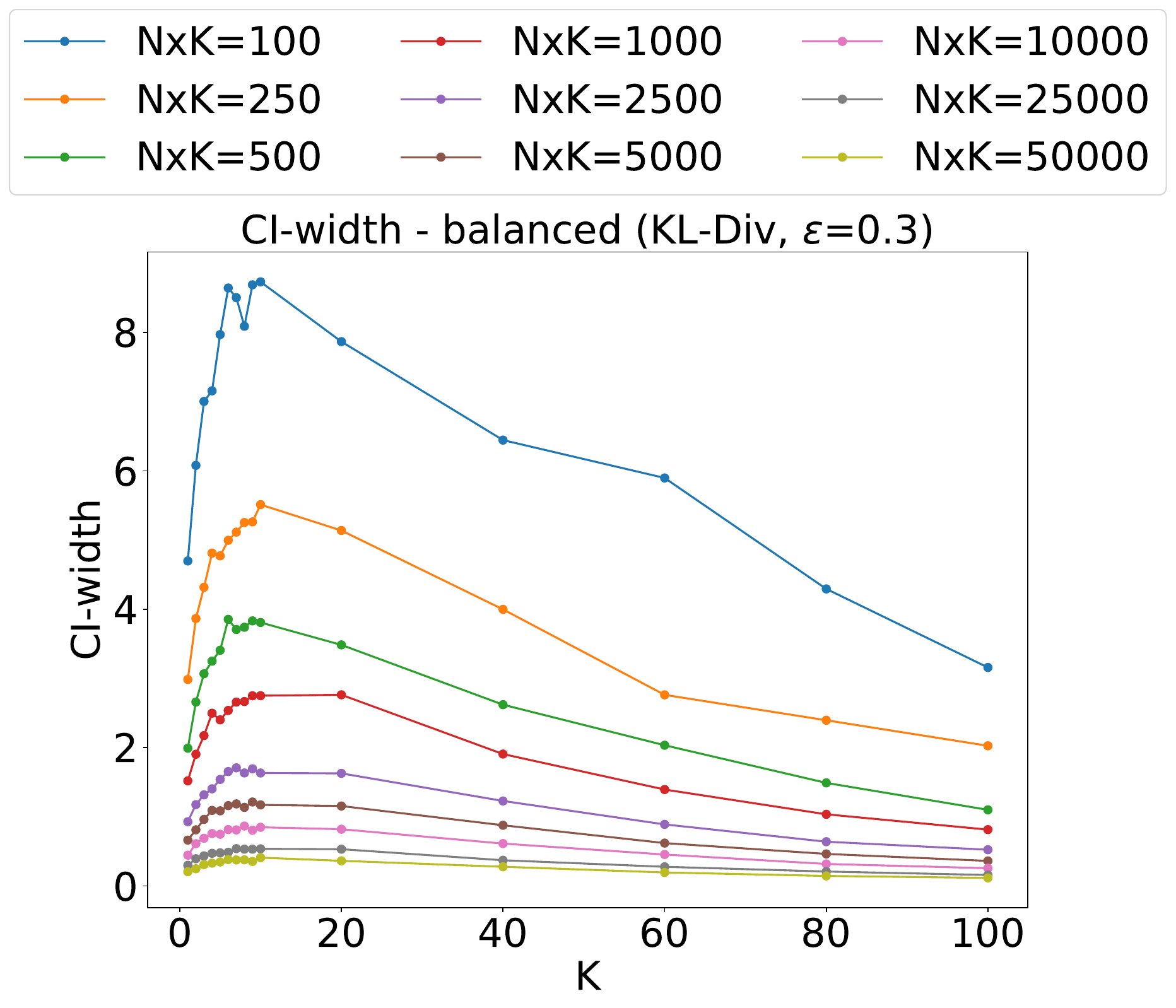}
    \caption{$\epsilon = 0.3$}
    \label{fig:uniform_ci_kl_cat12_e03}
  \end{subfigure} \hfill
  \begin{subfigure}[b]{0.24\linewidth}
    \centering
    \includegraphics[width=\linewidth]{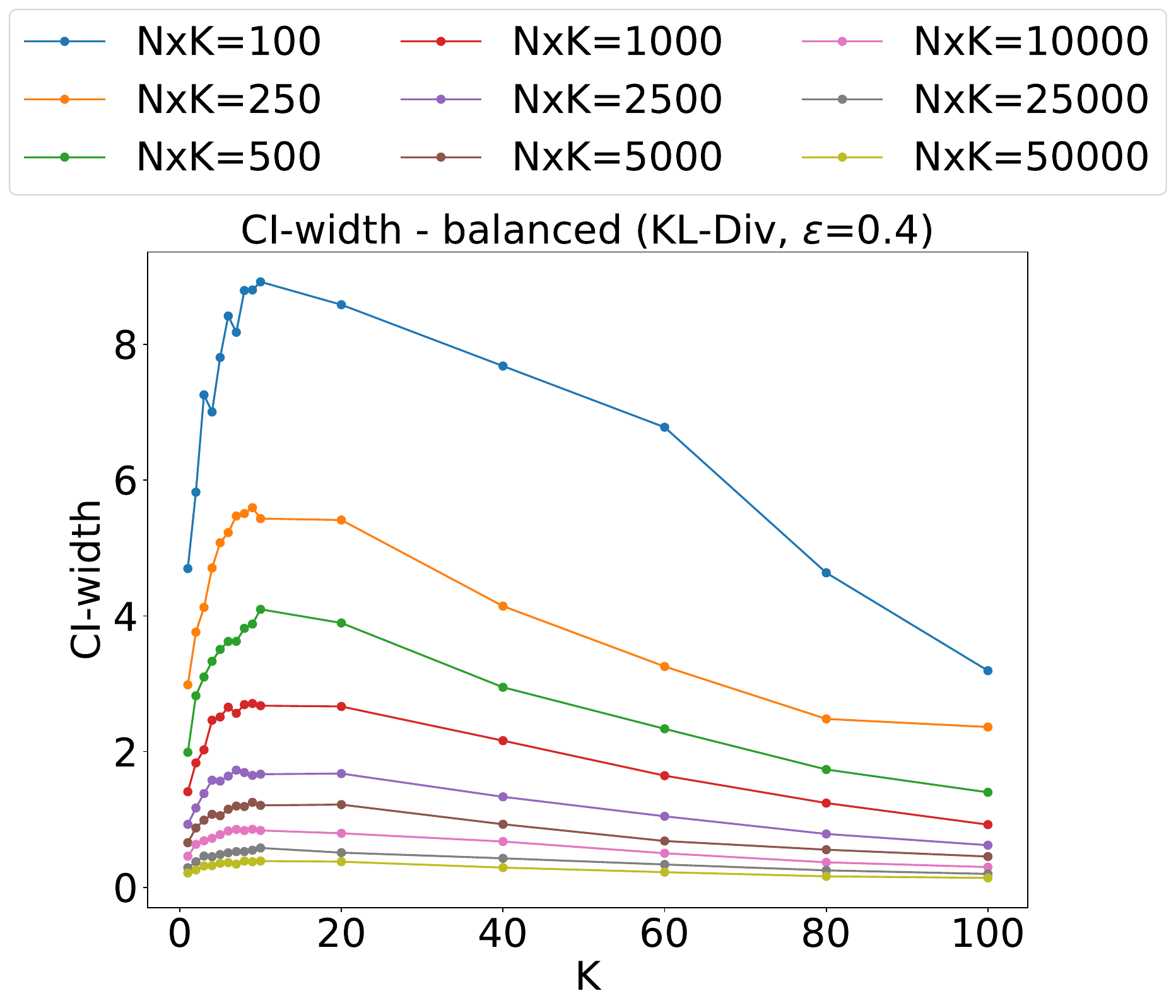}
    \caption{$\epsilon = 0.4$}
    \label{fig:uniform_ci_kl_cat12_e04}
  \end{subfigure}
  \caption{CI-width plots for balanced alphas with KL-divergence as the metric ($M=12$)}
  \label{fig:uniform_ci_kl_cat12}
\end{figure*}

\begin{figure*}
  \centering
  \begin{subfigure}[b]{0.24\linewidth}
    \centering
    \includegraphics[width=\linewidth]{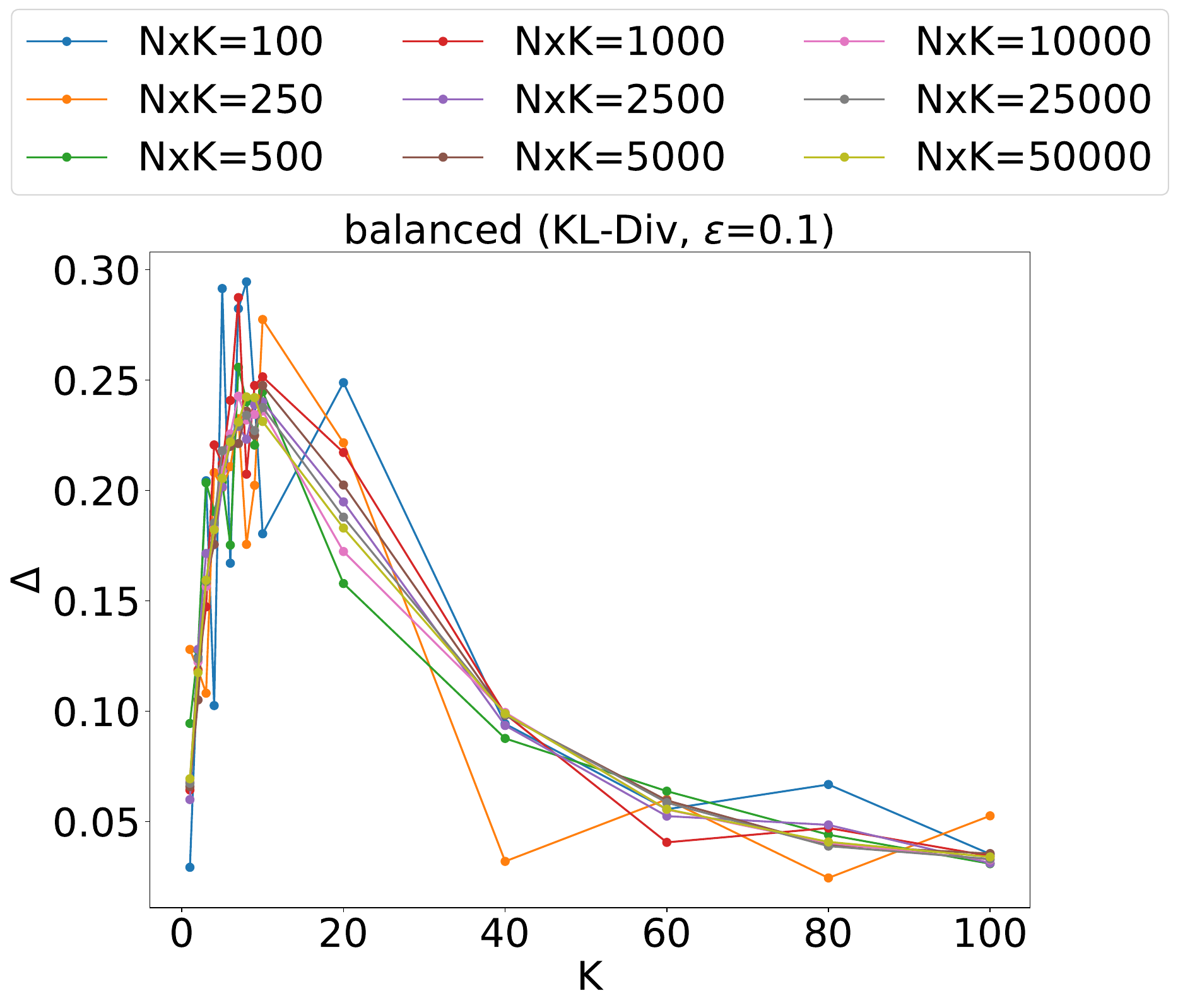}
    \caption{$\epsilon = 0.1$}
    \label{fig:uniform_delta_kl_cat12_e01}
  \end{subfigure} \hfill
  \begin{subfigure}[b]{0.24\linewidth}
    \centering
    \includegraphics[width=\linewidth]{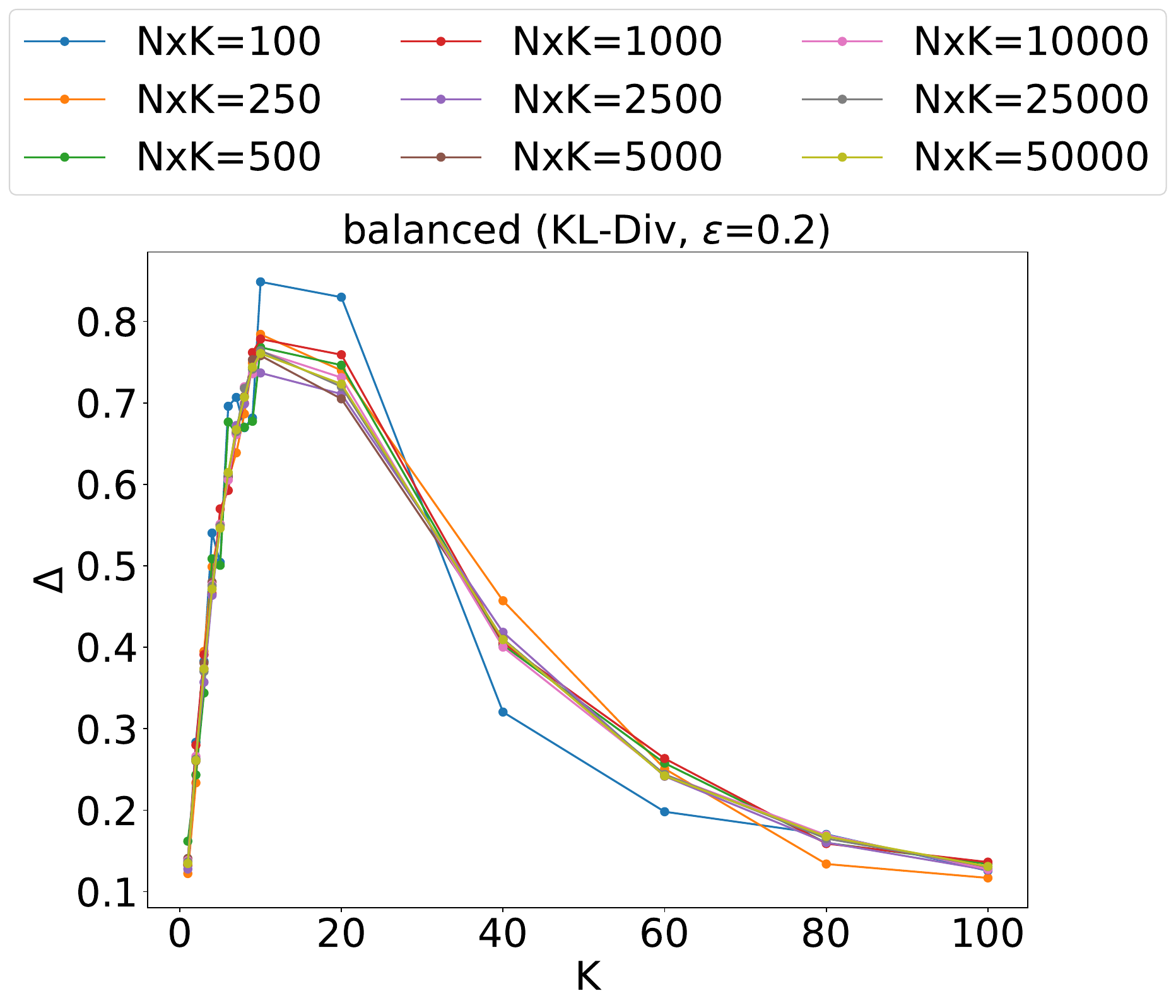}
    \caption{$\epsilon = 0.2$}
    \label{fig:uniform_delta_kl_cat12_e02}
  \end{subfigure} \hfill
  \begin{subfigure}[b]{0.24\linewidth}
    \centering
    \includegraphics[width=\linewidth]{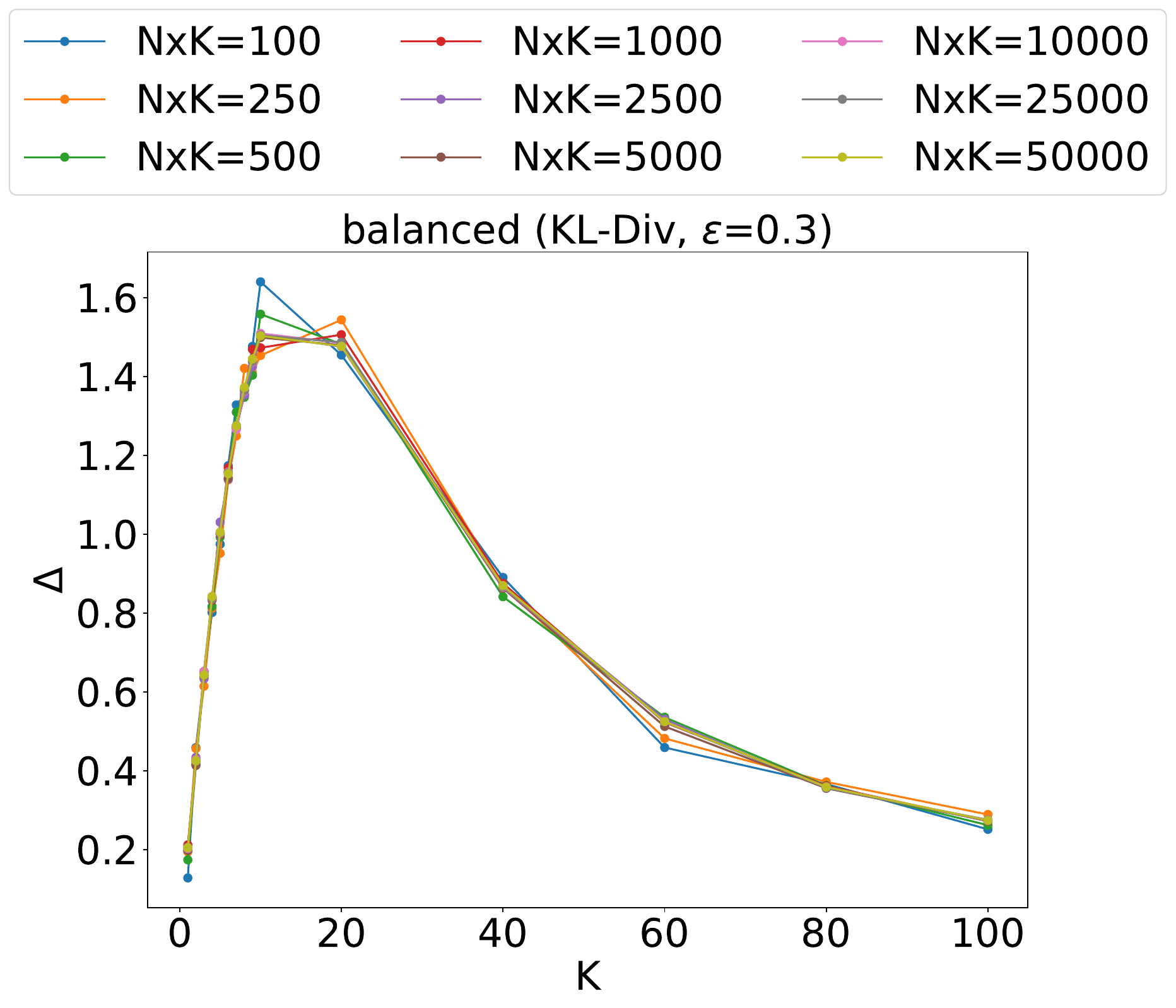}
    \caption{$\epsilon = 0.3$}
    \label{fig:uniform_delta_kl_cat12_e03}
  \end{subfigure} \hfill
  \begin{subfigure}[b]{0.24\linewidth}
    \centering
    \includegraphics[width=\linewidth]{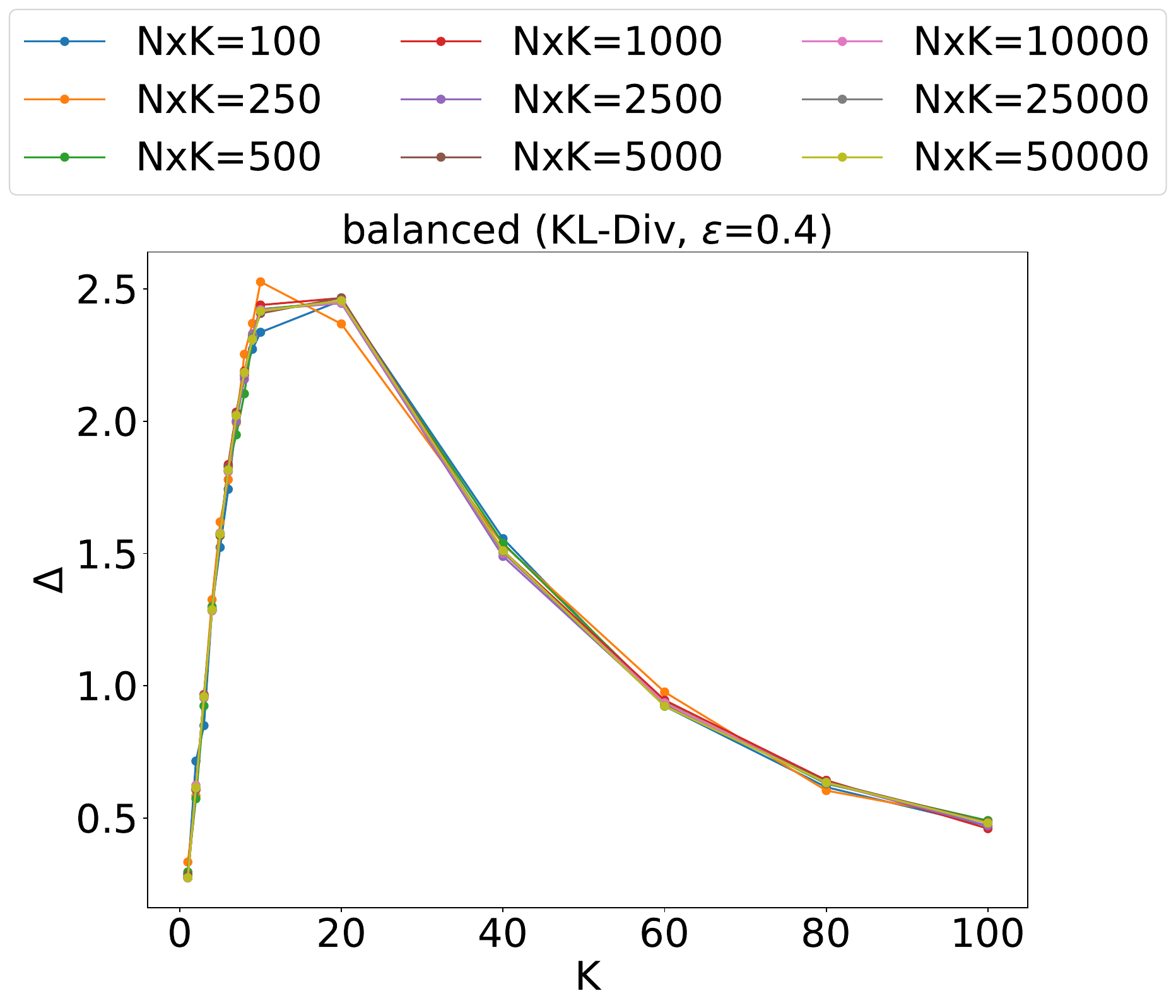}
    \caption{$\epsilon = 0.4$}
    \label{fig:uniform_delta_kl_cat12_e04}
  \end{subfigure}
  \caption{Effect sizes ($\Delta$) for balanced alphas with KL-divergence as the metric ($M=12$)}
  \label{fig:uniform_delta_kl_cat12}
\end{figure*}

Figures \ref{fig:uniform_kl_cat2}--\ref{fig:uniform_delta_kl_cat12} show results for \pv s, confidence intervals, and effect sizes of KL-divergence for balanced priors on parameter $\alpha$ with different values of $\epsilon$ and increasing $K$ for various $N \times K$. For simulations with balanced categories, \pv s\ exhibit a single or double peak initially but settle down with higher $K$. CI-width and effect sizes decrease with increasing $K$; however start to have one peak as $M$ increases.

\subsubsection{Artificial Distribution with Unbalanced Priors}

\begin{figure*}
  \centering
  \begin{subfigure}[b]{0.24\linewidth}
    \centering
    \includegraphics[width=\linewidth]{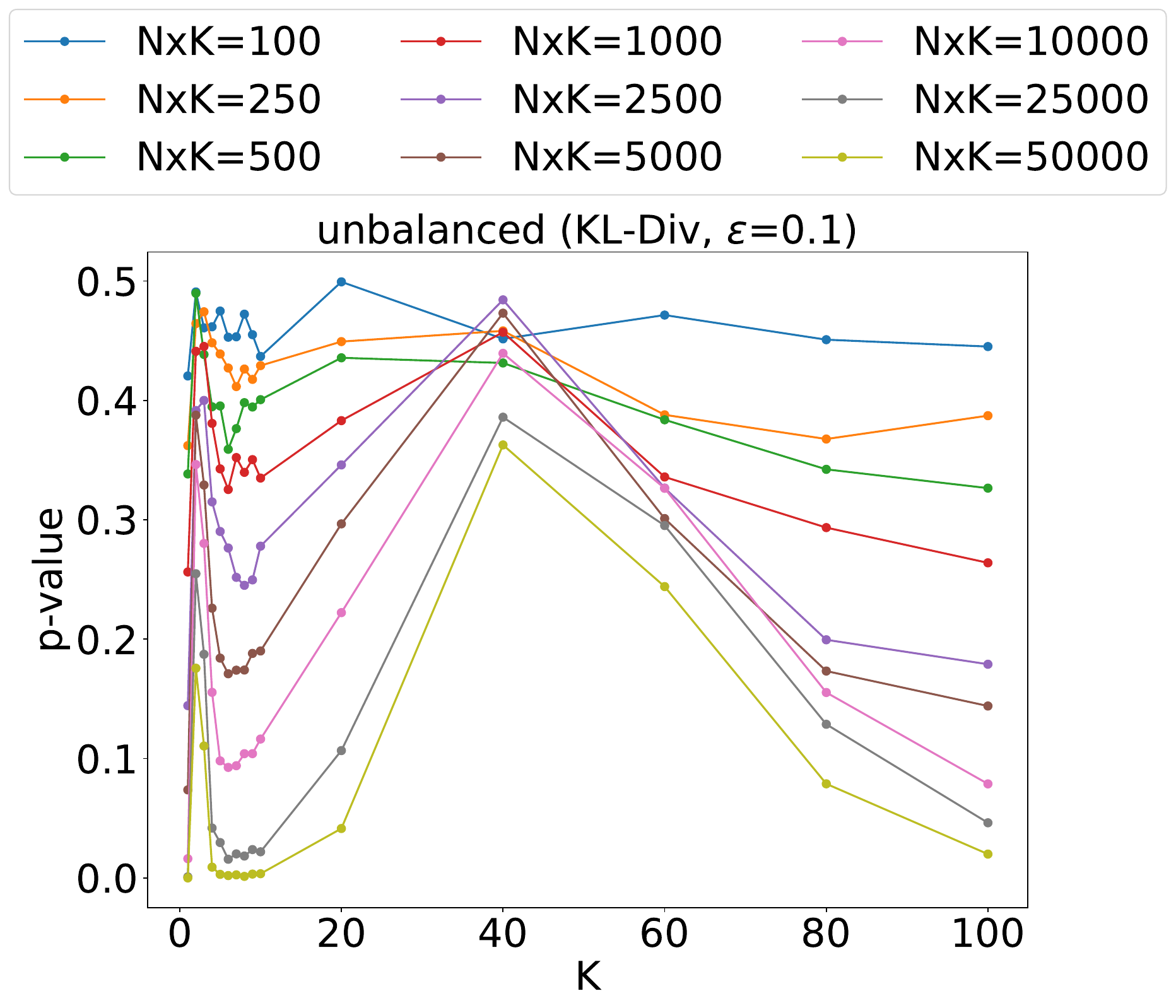}
    \caption{$\epsilon = 0.1$}
    \label{fig:gamma_kl_cat2_e01}
  \end{subfigure} \hfill
  \begin{subfigure}[b]{0.24\linewidth}
    \centering
    \includegraphics[width=\linewidth]{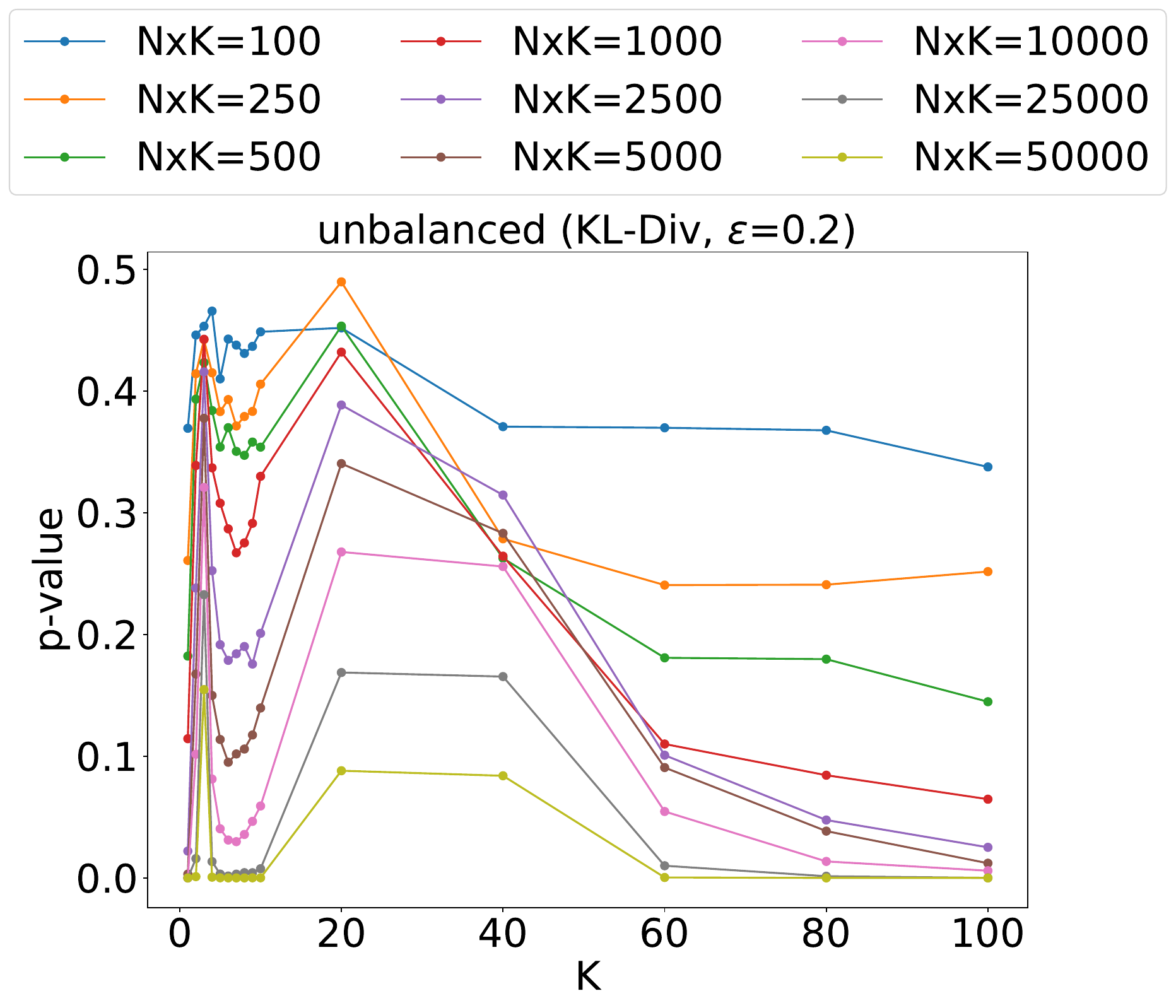}
    \caption{$\epsilon = 0.2$}
    \label{fig:gamma_kl_cat2_e02}
  \end{subfigure} \hfill
  \begin{subfigure}[b]{0.24\linewidth}
    \centering
    \includegraphics[width=\linewidth]{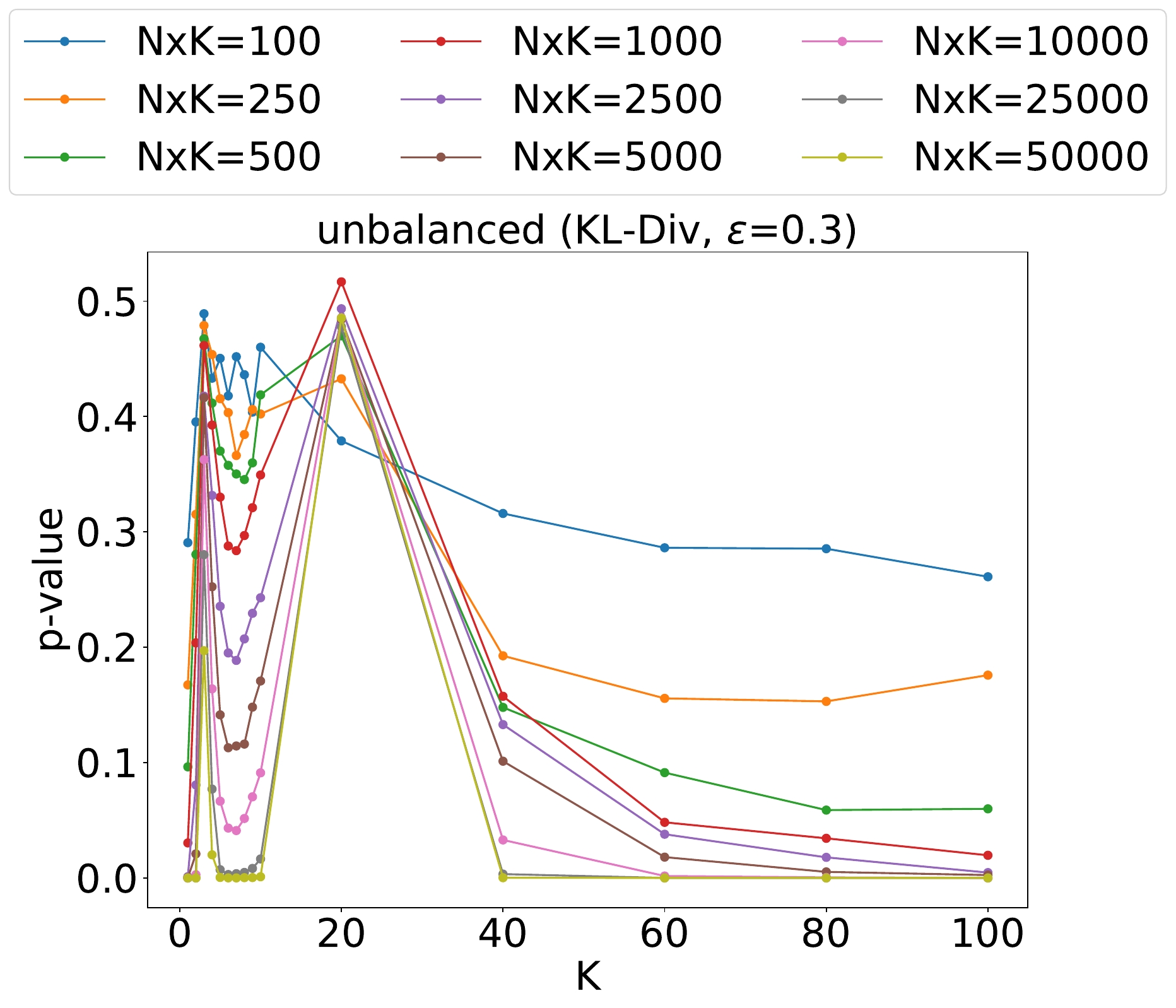}
    \caption{$\epsilon = 0.3$}
    \label{fig:gamma_kl_cat2_e03}
  \end{subfigure} \hfill
  \begin{subfigure}[b]{0.24\linewidth}
    \centering
    \includegraphics[width=\linewidth]{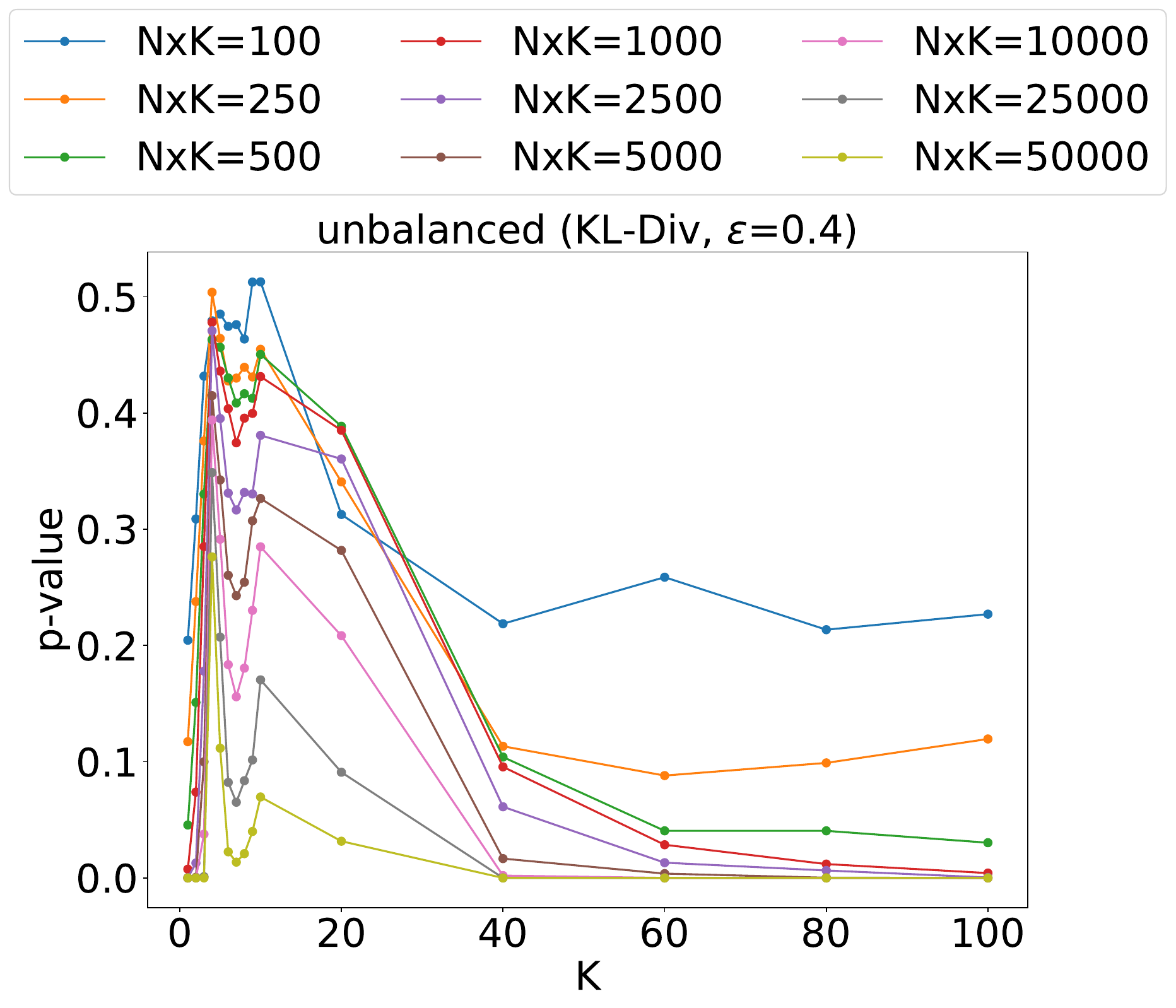}
    \caption{$\epsilon = 0.4$}
    \label{fig:gamma_kl_cat2_e04}
  \end{subfigure}
  \caption{P-value plots for unbalanced alphas with KL-divergence as the metric ($M=2$)}
  \label{fig:gamma_kl_cat2}
\end{figure*}

\begin{figure*}
  \centering
  \begin{subfigure}[b]{0.24\linewidth}
    \centering
    \includegraphics[width=\linewidth]{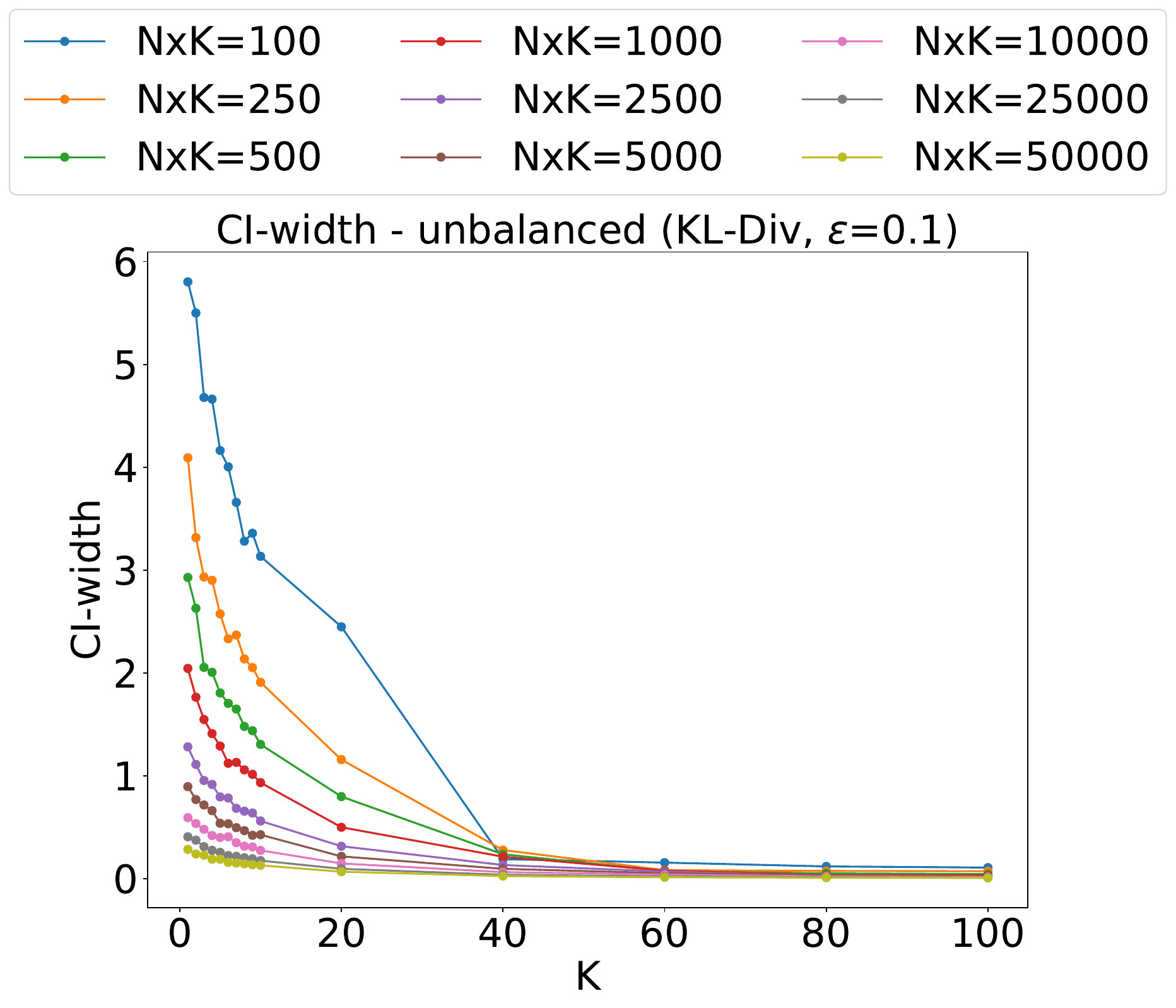}
    \caption{$\epsilon = 0.1$}
    \label{fig:gamma_ci_kl_cat2_e01}
  \end{subfigure} \hfill
  \begin{subfigure}[b]{0.24\linewidth}
    \centering
    \includegraphics[width=\linewidth]{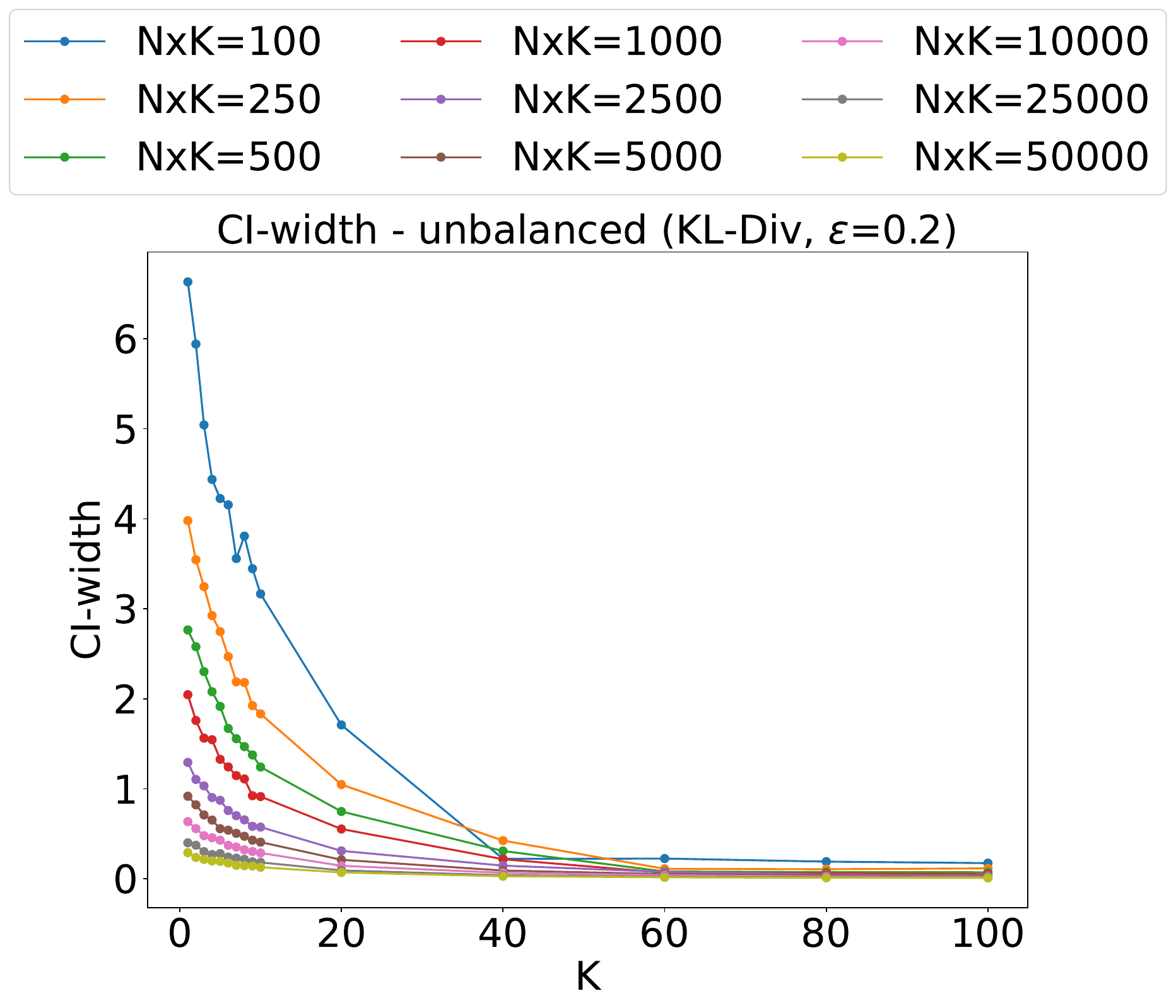}
    \caption{$\epsilon = 0.2$}
    \label{fig:gamma_ci_kl_cat2_e02}
  \end{subfigure} \hfill
  \begin{subfigure}[b]{0.24\linewidth}
    \centering
    \includegraphics[width=\linewidth]{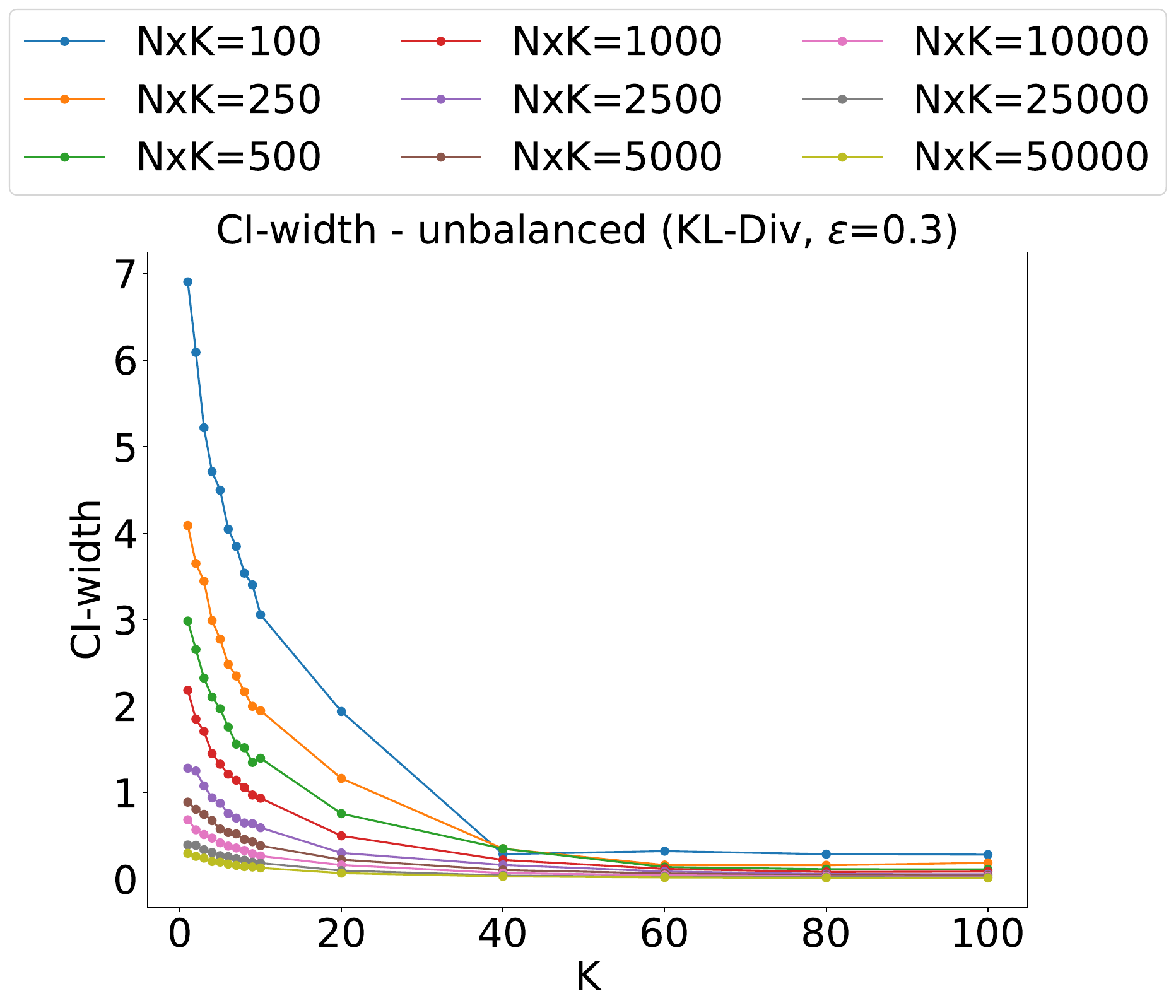}
    \caption{$\epsilon = 0.3$}
    \label{fig:gamma_ci_kl_cat2_e03}
  \end{subfigure} \hfill
  \begin{subfigure}[b]{0.24\linewidth}
    \centering
    \includegraphics[width=\linewidth]{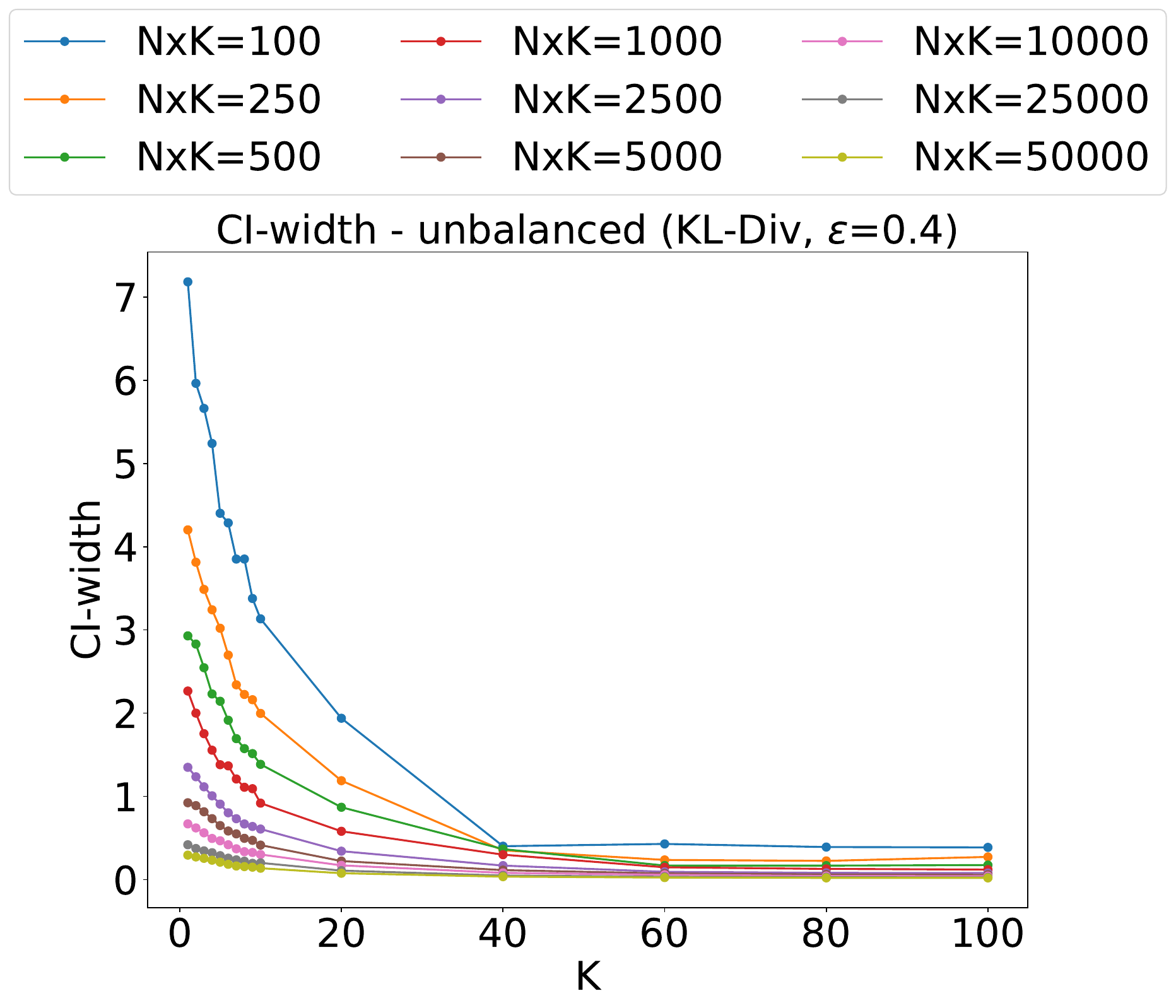}
    \caption{$\epsilon = 0.4$}
    \label{fig:gamma_ci_kl_cat2_e04}
  \end{subfigure}
  \caption{CI-width plots for unbalanced alphas with KL-divergence as the metric ($M=2$)}
  \label{fig:gamma_ci_kl_cat2}
\end{figure*}

\begin{figure*}
  \centering
  \begin{subfigure}[b]{0.24\linewidth}
    \centering
    \includegraphics[width=\linewidth]{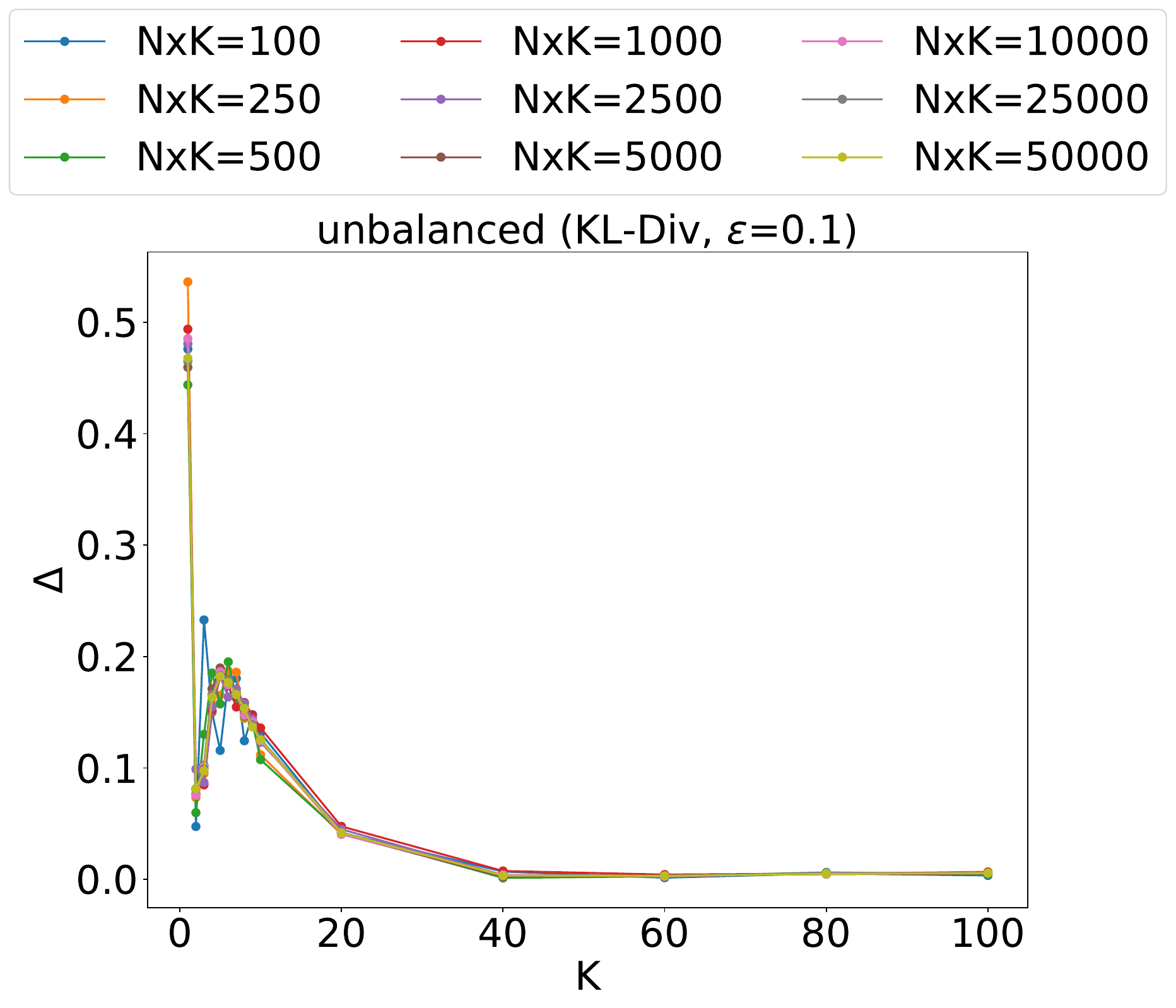}
    \caption{$\epsilon = 0.1$}
    \label{fig:gamma_delta_kl_cat2_e01}
  \end{subfigure} \hfill
  \begin{subfigure}[b]{0.24\linewidth}
    \centering
    \includegraphics[width=\linewidth]{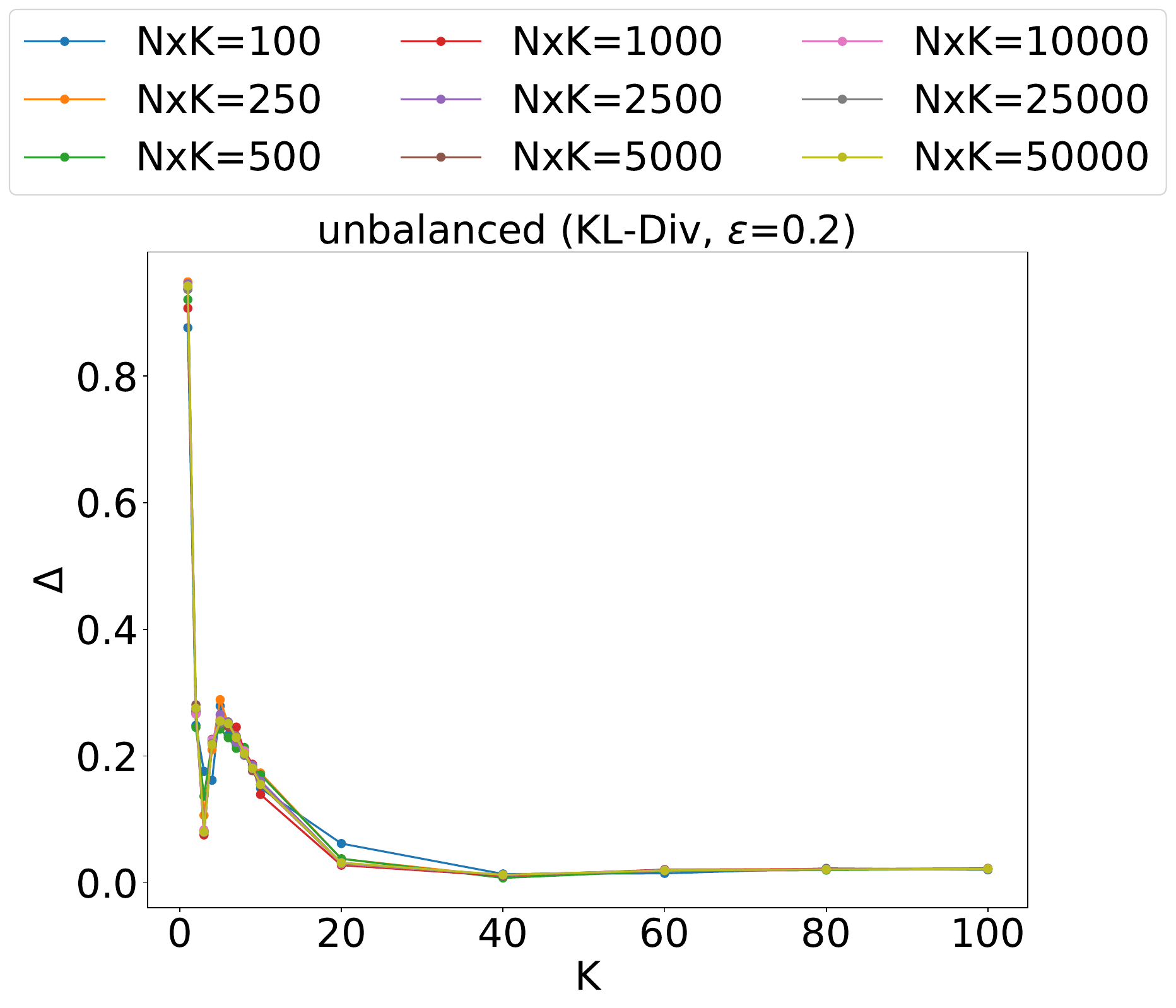}
    \caption{$\epsilon = 0.2$}
    \label{fig:gamma_delta_kl_cat2_e02}
  \end{subfigure} \hfill
  \begin{subfigure}[b]{0.24\linewidth}
    \centering
    \includegraphics[width=\linewidth]{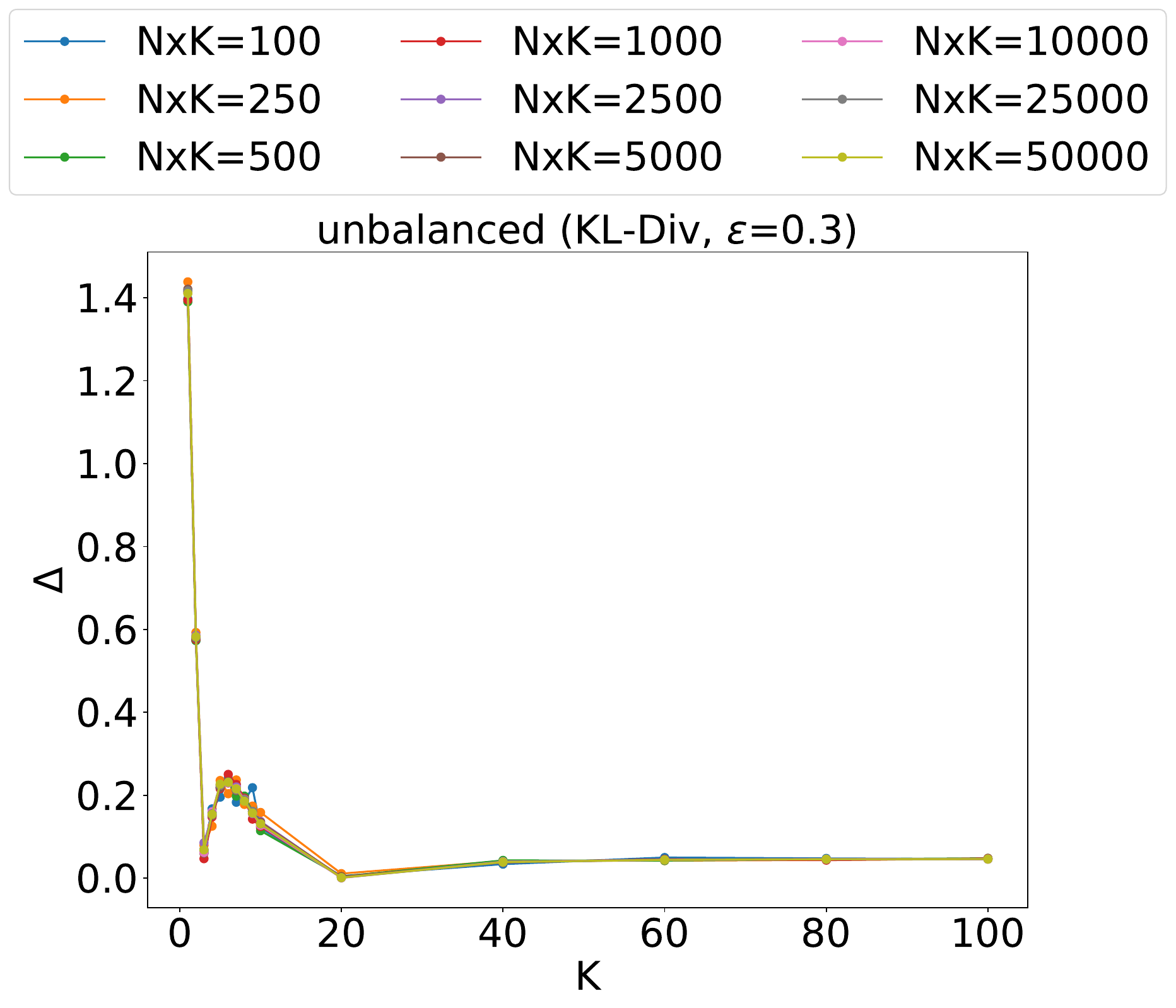}
    \caption{$\epsilon = 0.3$}
    \label{fig:gamma_delta_kl_cat2_e03}
  \end{subfigure} \hfill
  \begin{subfigure}[b]{0.24\linewidth}
    \centering
    \includegraphics[width=\linewidth]{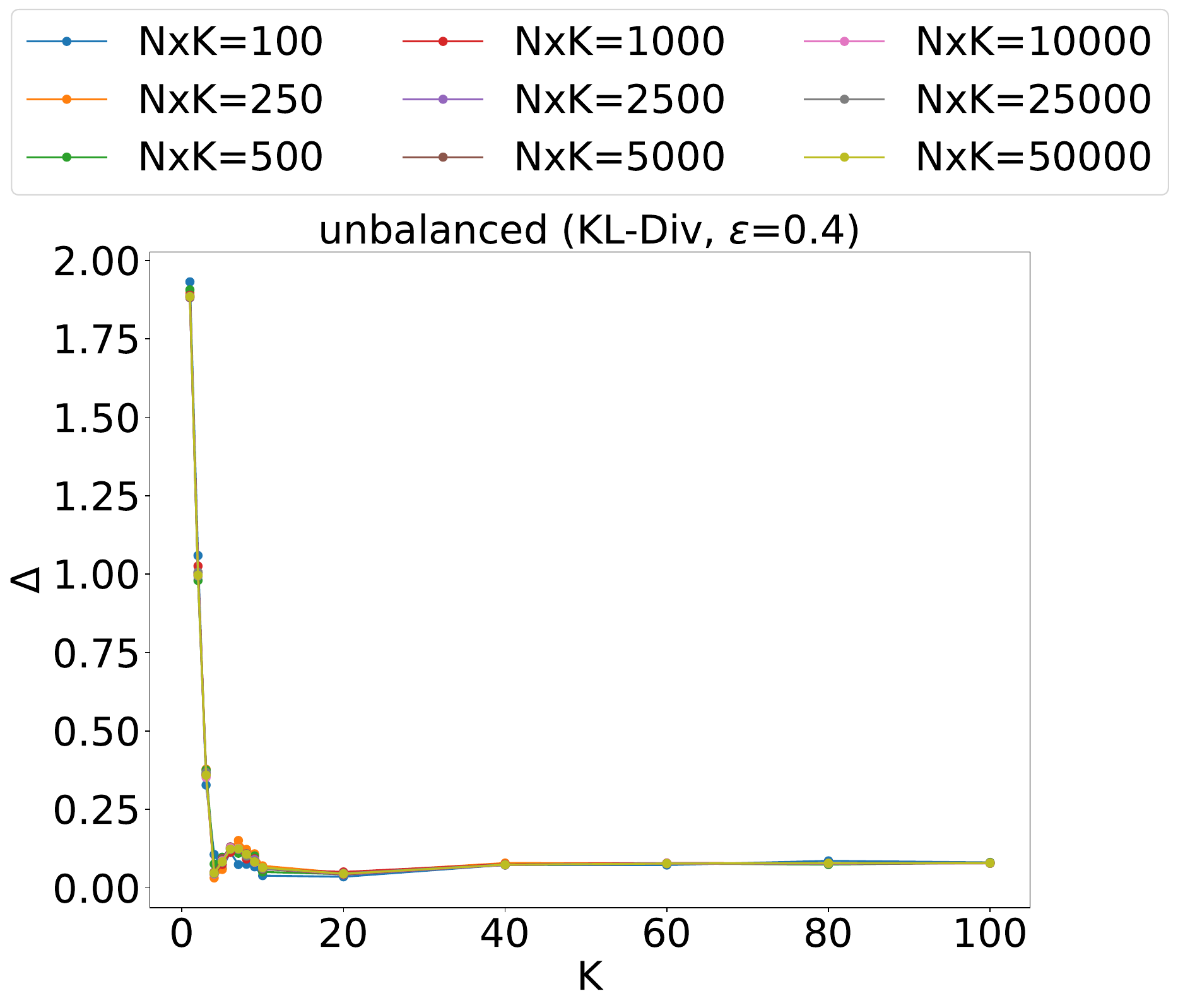}
    \caption{$\epsilon = 0.4$}
    \label{fig:gamma_delta_kl_cat2_e04}
  \end{subfigure}
  \caption{Effect sizes ($\Delta$) for unbalanced alphas with KL-divergence as the metric ($M=2$)}
  \label{fig:gamma_delta_kl_cat2}
\end{figure*}

\begin{figure*}
  \centering
  \begin{subfigure}[b]{0.24\linewidth}
    \centering
    \includegraphics[width=\linewidth]{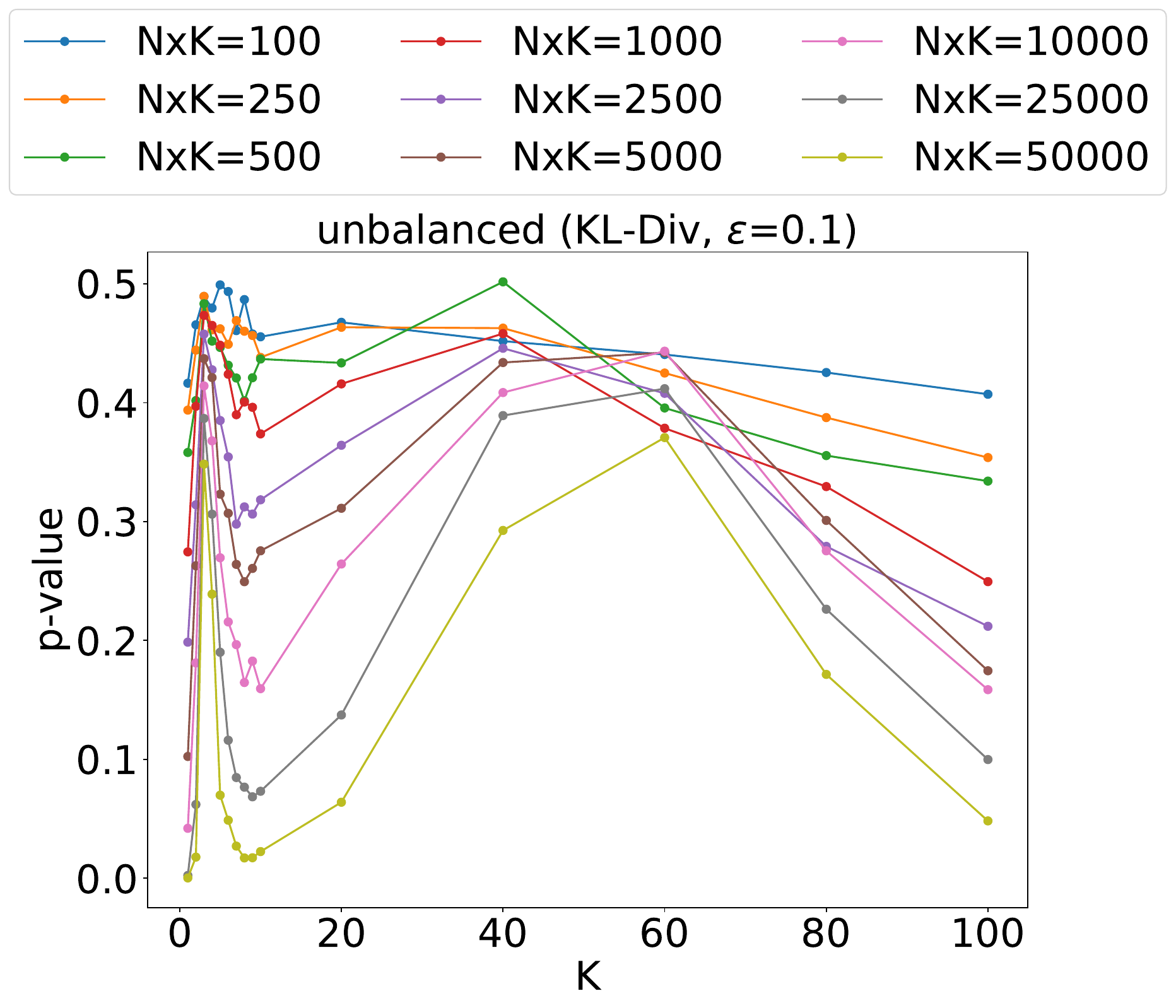}
    \caption{$\epsilon = 0.1$}
    \label{fig:gamma_kl_cat3_e01}
  \end{subfigure} \hfill
  \begin{subfigure}[b]{0.24\linewidth}
    \centering
    \includegraphics[width=\linewidth]{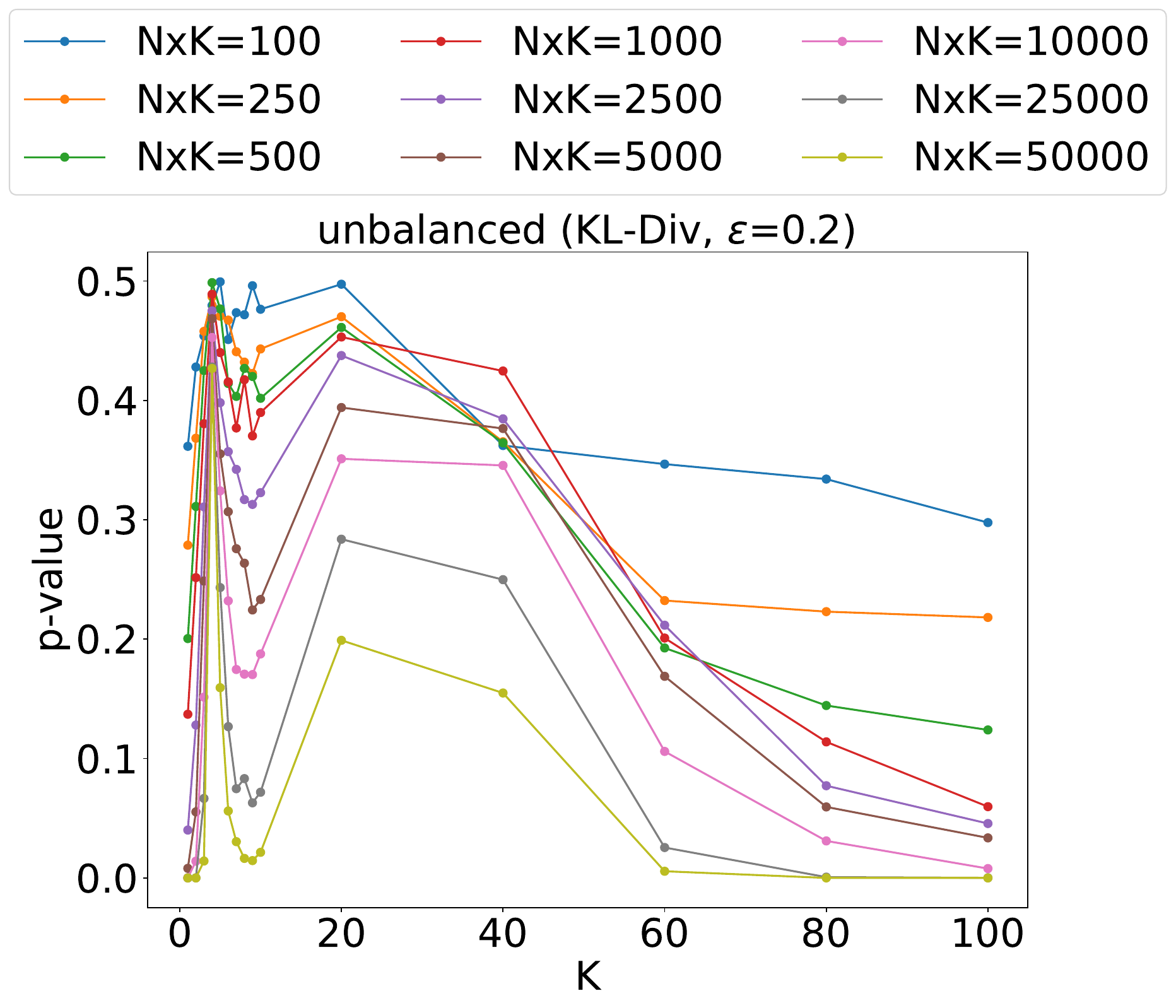}
    \caption{$\epsilon = 0.2$}
    \label{fig:gamma_kl_cat3_e02}
  \end{subfigure} \hfill
  \begin{subfigure}[b]{0.24\linewidth}
    \centering
    \includegraphics[width=\linewidth]{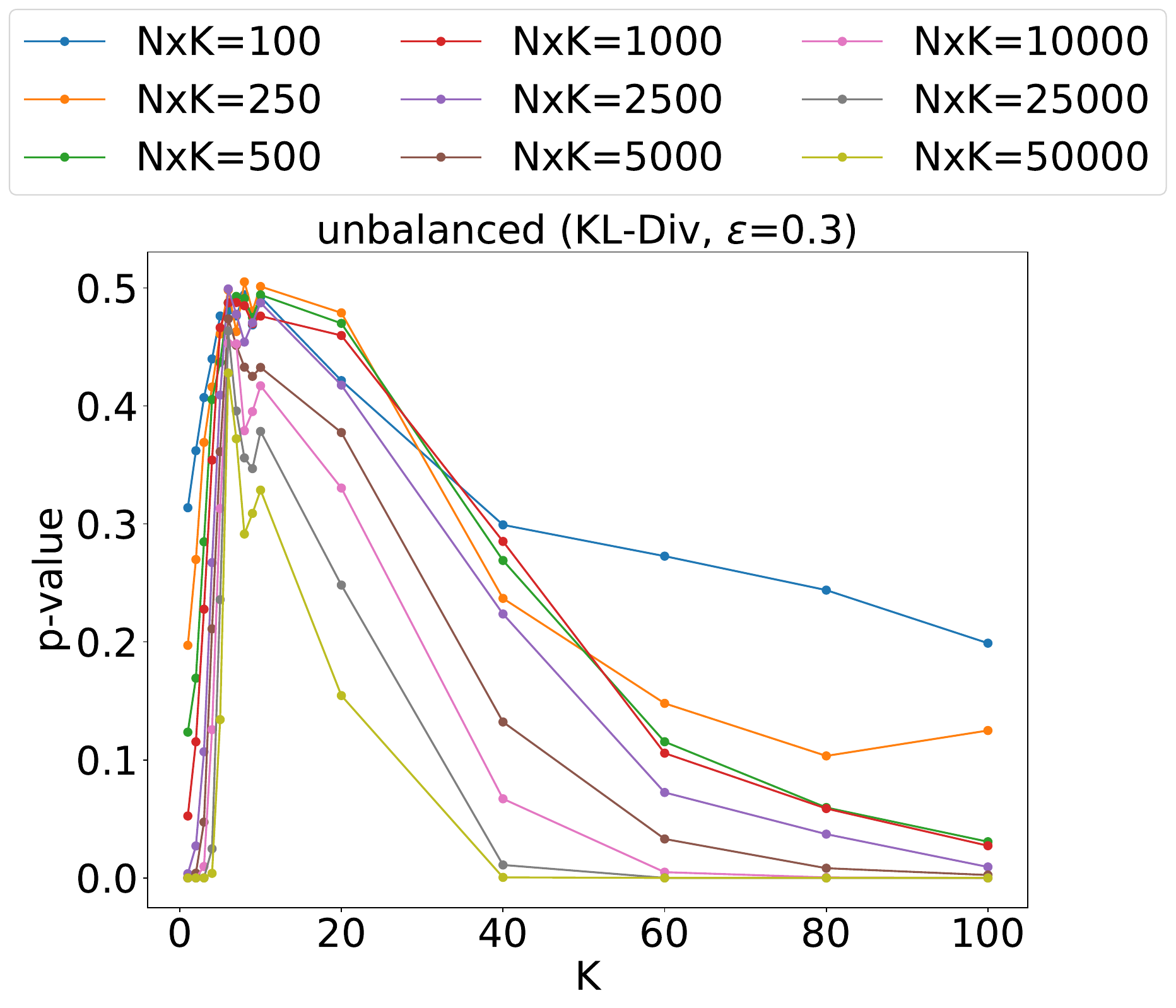}
    \caption{$\epsilon = 0.3$}
    \label{fig:gamma_kl_cat3_e03}
  \end{subfigure} \hfill
  \begin{subfigure}[b]{0.24\linewidth}
    \centering
    \includegraphics[width=\linewidth]{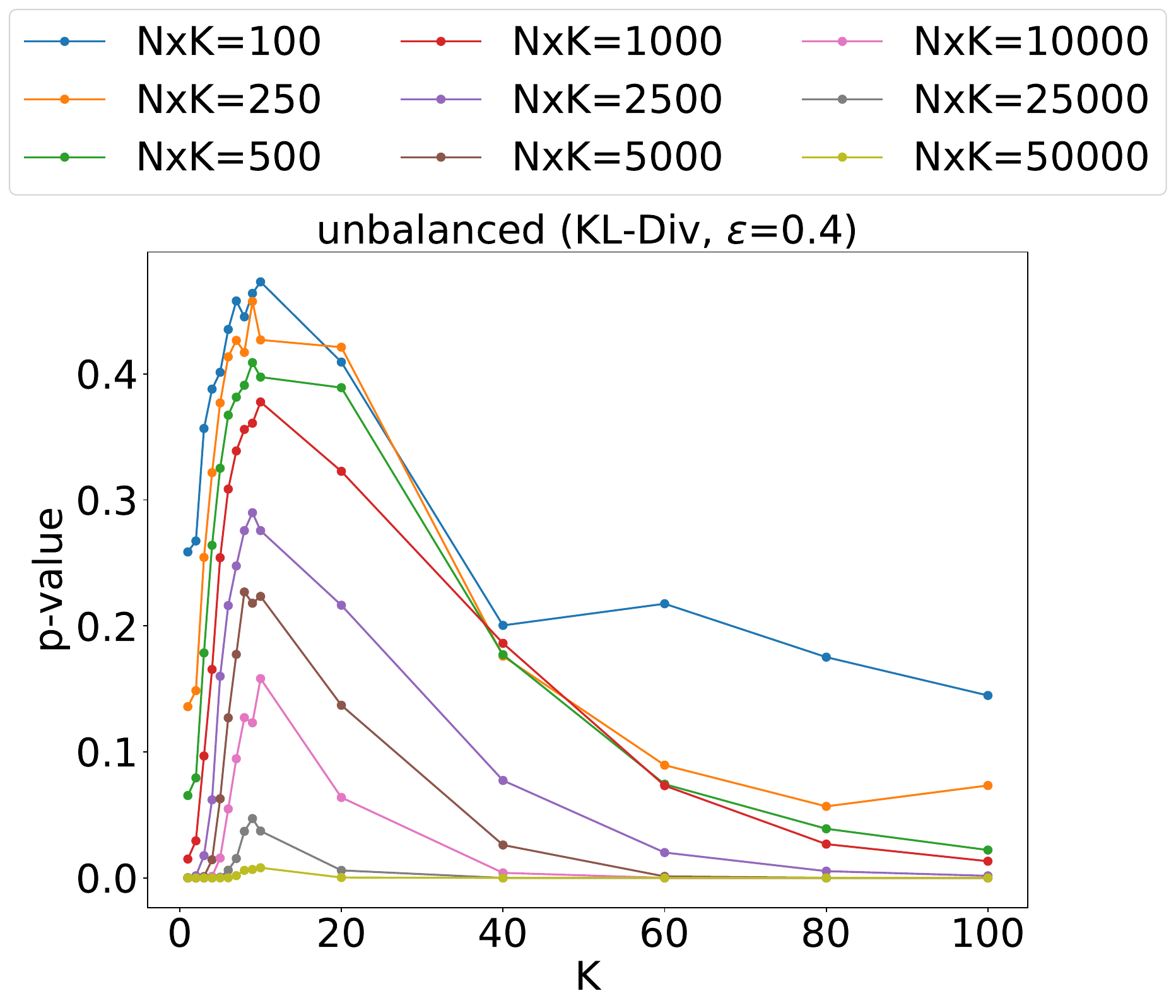}
    \caption{$\epsilon = 0.4$}
    \label{fig:gamma_kl_cat3_e04}
  \end{subfigure}
  \caption{P-value plots for unbalanced alphas with KL-divergence as the metric ($M=3$)}
  \label{fig:gamma_kl_cat3}
\end{figure*}

\begin{figure*}
  \centering
  \begin{subfigure}[b]{0.24\linewidth}
    \centering
    \includegraphics[width=\linewidth]{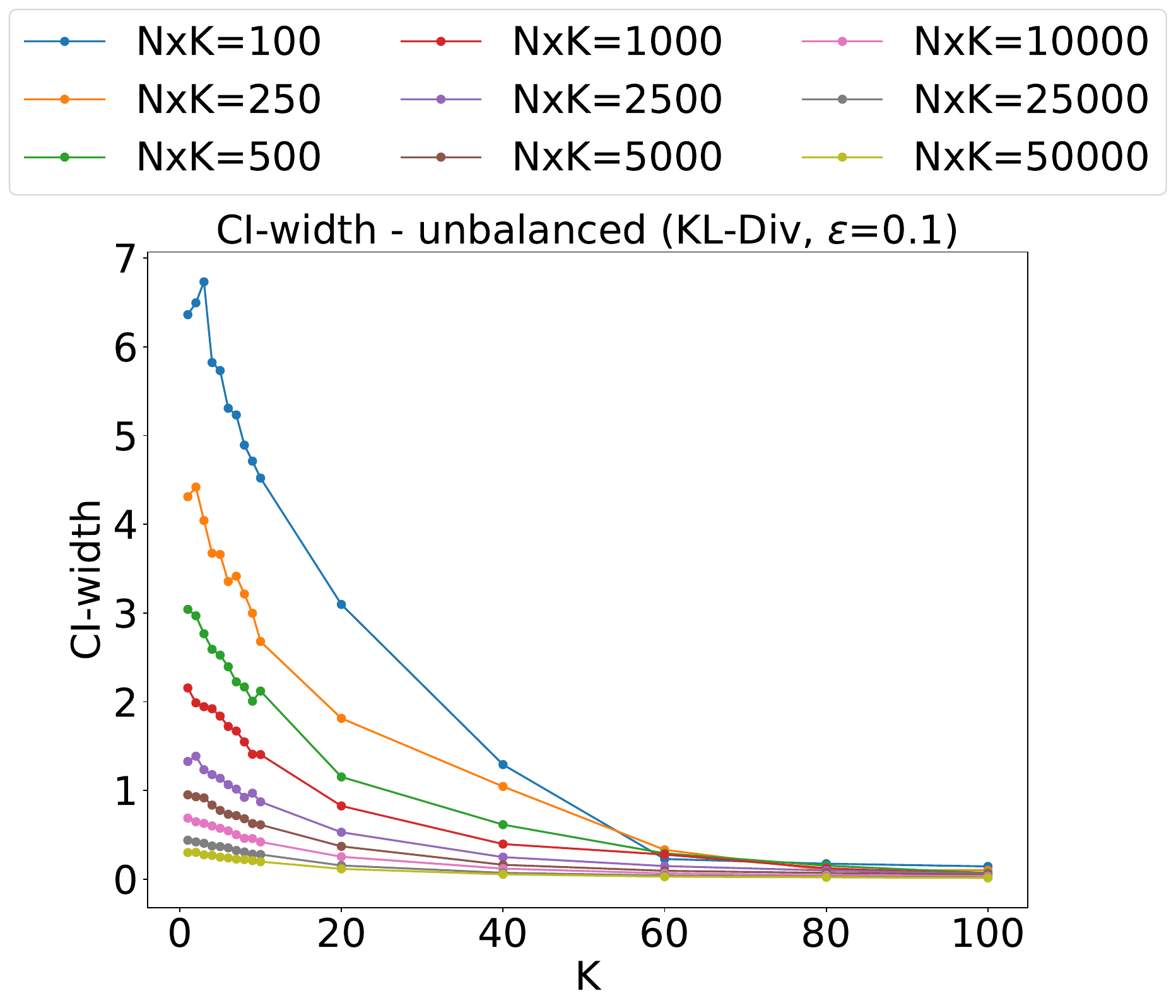}
    \caption{$\epsilon = 0.1$}
    \label{fig:gamma_ci_kl_cat3_e01}
  \end{subfigure} \hfill
  \begin{subfigure}[b]{0.24\linewidth}
    \centering
    \includegraphics[width=\linewidth]{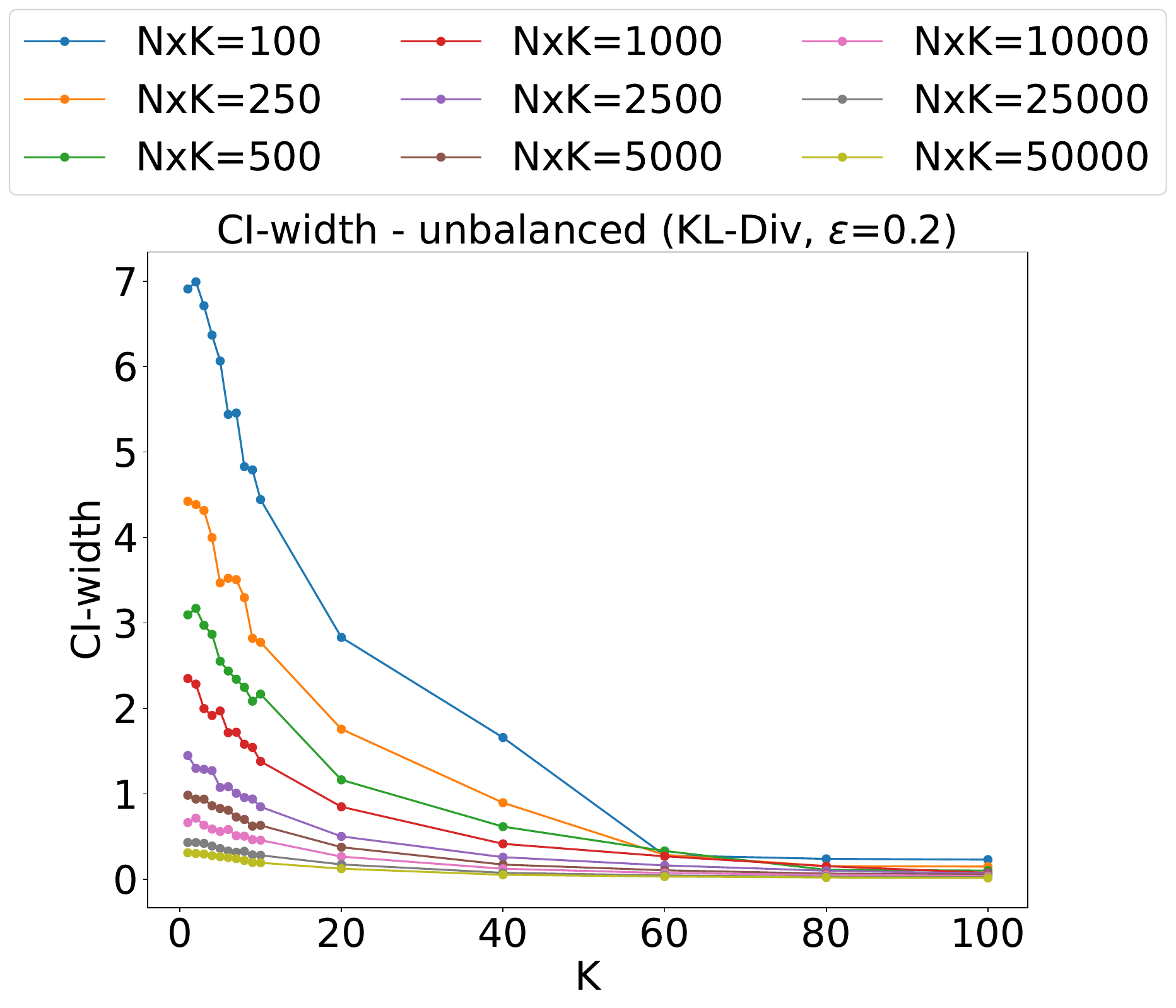}
    \caption{$\epsilon = 0.2$}
    \label{fig:gamma_ci_kl_cat3_e02}
  \end{subfigure} \hfill
  \begin{subfigure}[b]{0.24\linewidth}
    \centering
    \includegraphics[width=\linewidth]{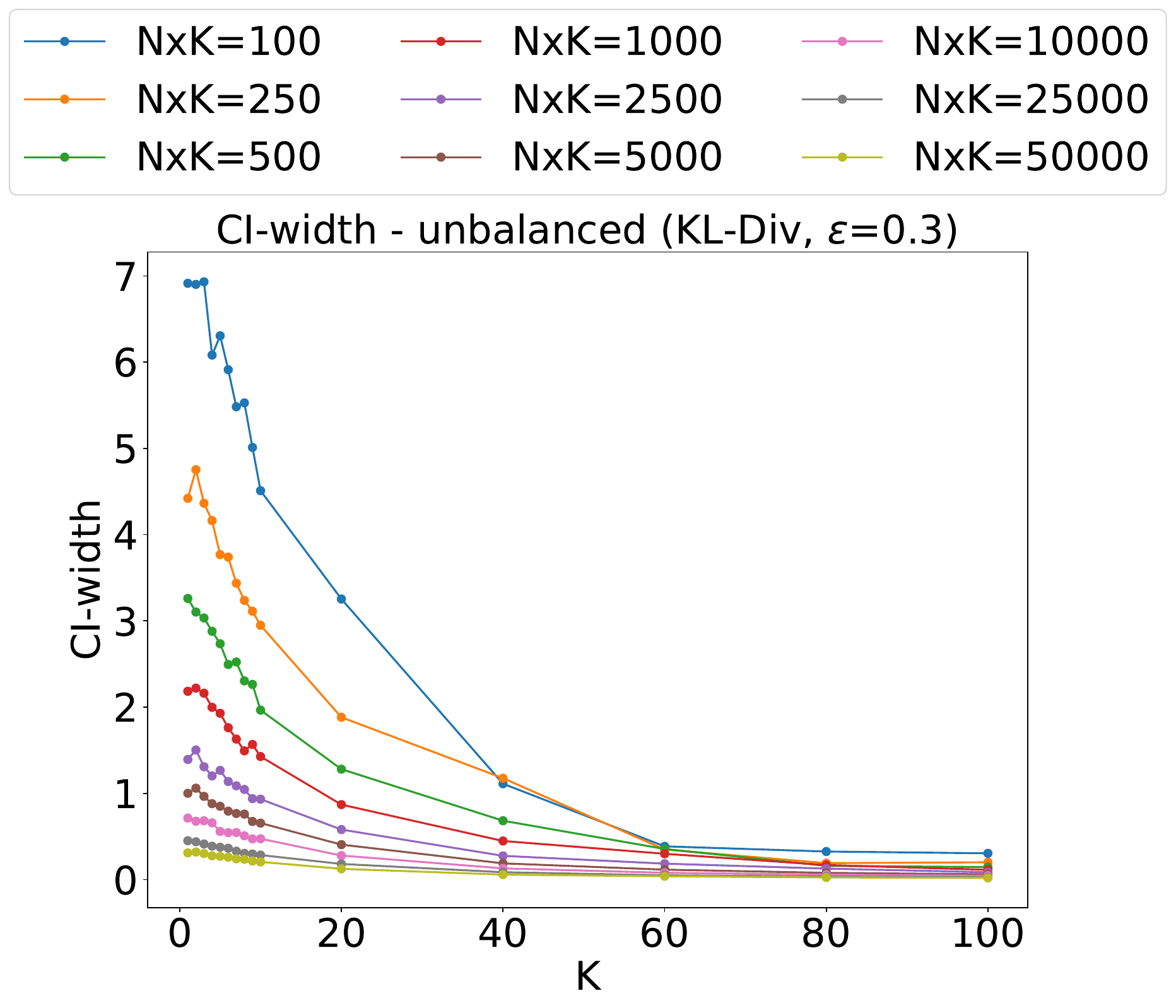}
    \caption{$\epsilon = 0.3$}
    \label{fig:gamma_ci_kl_cat3_e03}
  \end{subfigure} \hfill
  \begin{subfigure}[b]{0.24\linewidth}
    \centering
    \includegraphics[width=\linewidth]{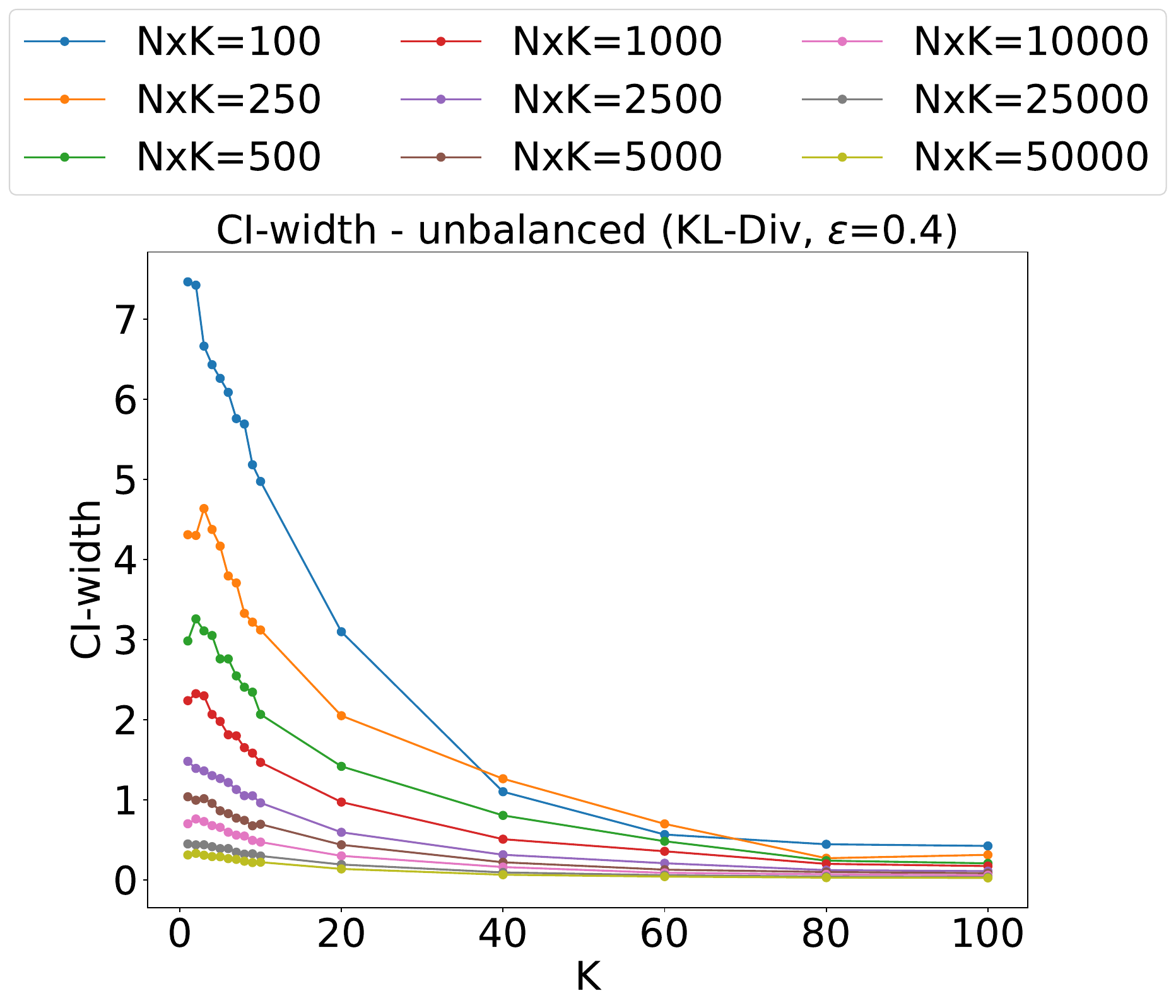}
    \caption{$\epsilon = 0.4$}
    \label{fig:gamma_ci_kl_cat3_e04}
  \end{subfigure}
  \caption{CI-width plots for unbalanced alphas with KL-divergence as the metric ($M=3$)}
  \label{fig:gamma_ci_kl_cat3}
\end{figure*}

\begin{figure*}
  \centering
  \begin{subfigure}[b]{0.24\linewidth}
    \centering
    \includegraphics[width=\linewidth]{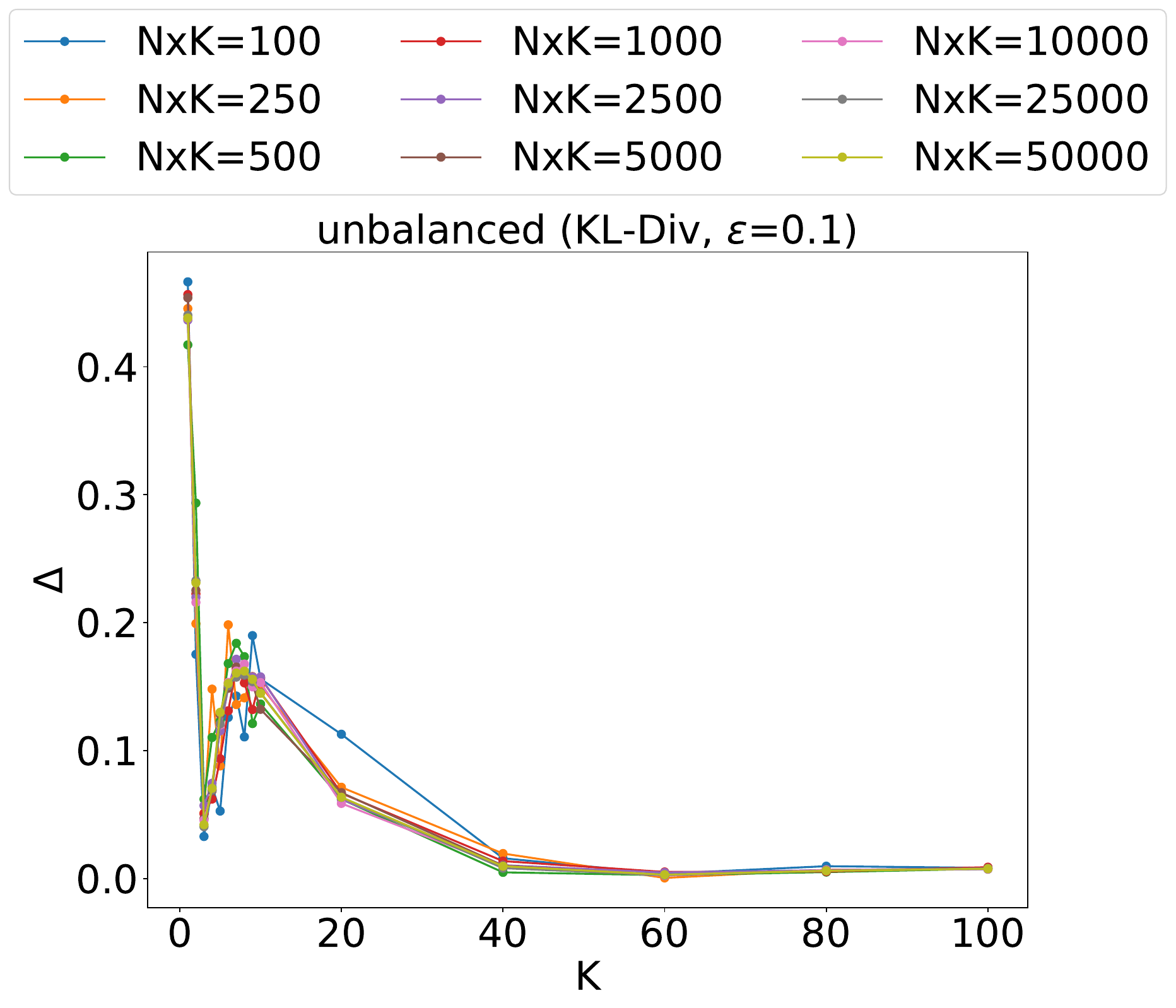}
    \caption{$\epsilon = 0.1$}
    \label{fig:gamma_delta_kl_cat3_e01}
  \end{subfigure} \hfill
  \begin{subfigure}[b]{0.24\linewidth}
    \centering
    \includegraphics[width=\linewidth]{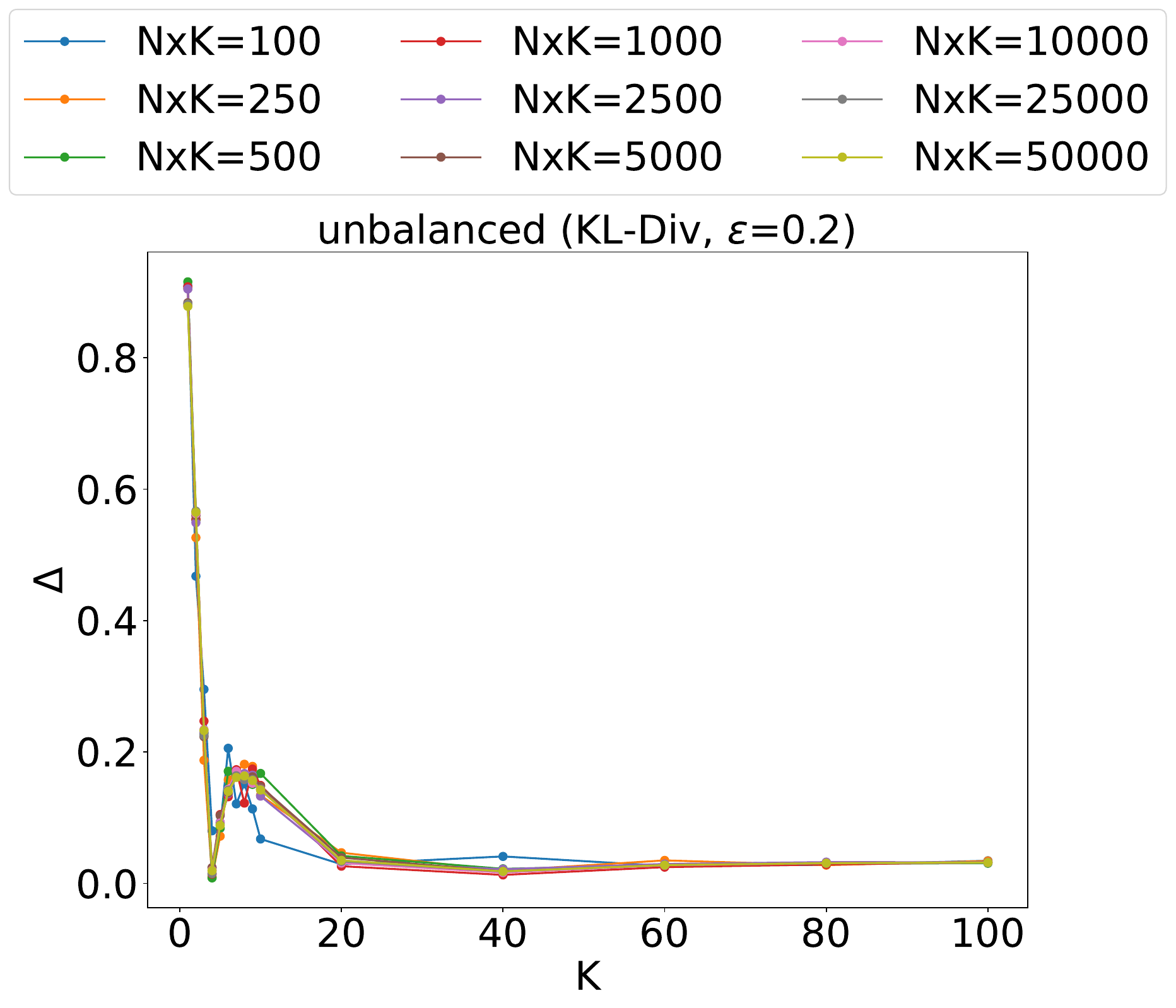}
    \caption{$\epsilon = 0.2$}
    \label{fig:gamma_delta_kl_cat3_e02}
  \end{subfigure} \hfill
  \begin{subfigure}[b]{0.24\linewidth}
    \centering
    \includegraphics[width=\linewidth]{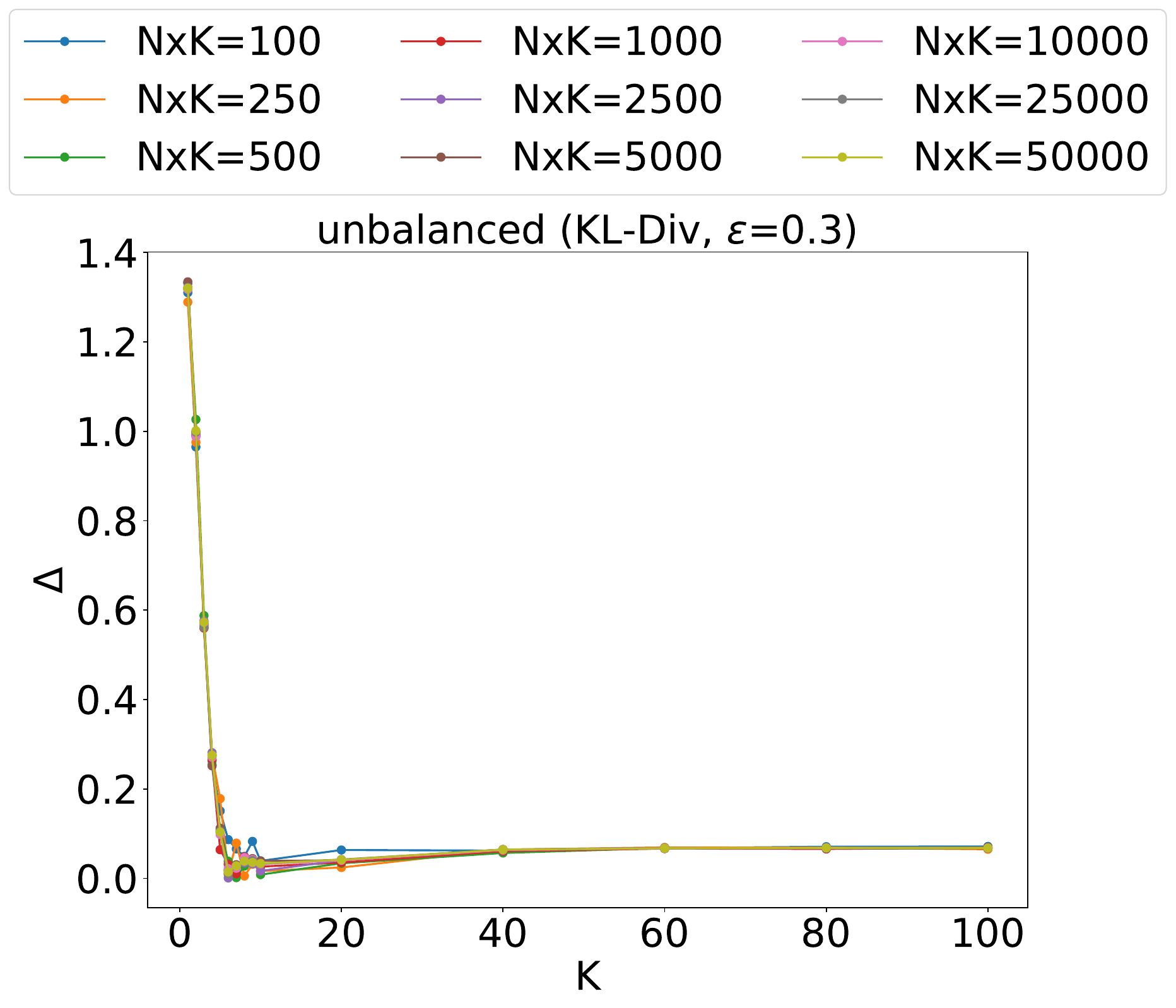}
    \caption{$\epsilon = 0.3$}
    \label{fig:gamma_delta_kl_cat3_e03}
  \end{subfigure} \hfill
  \begin{subfigure}[b]{0.24\linewidth}
    \centering
    \includegraphics[width=\linewidth]{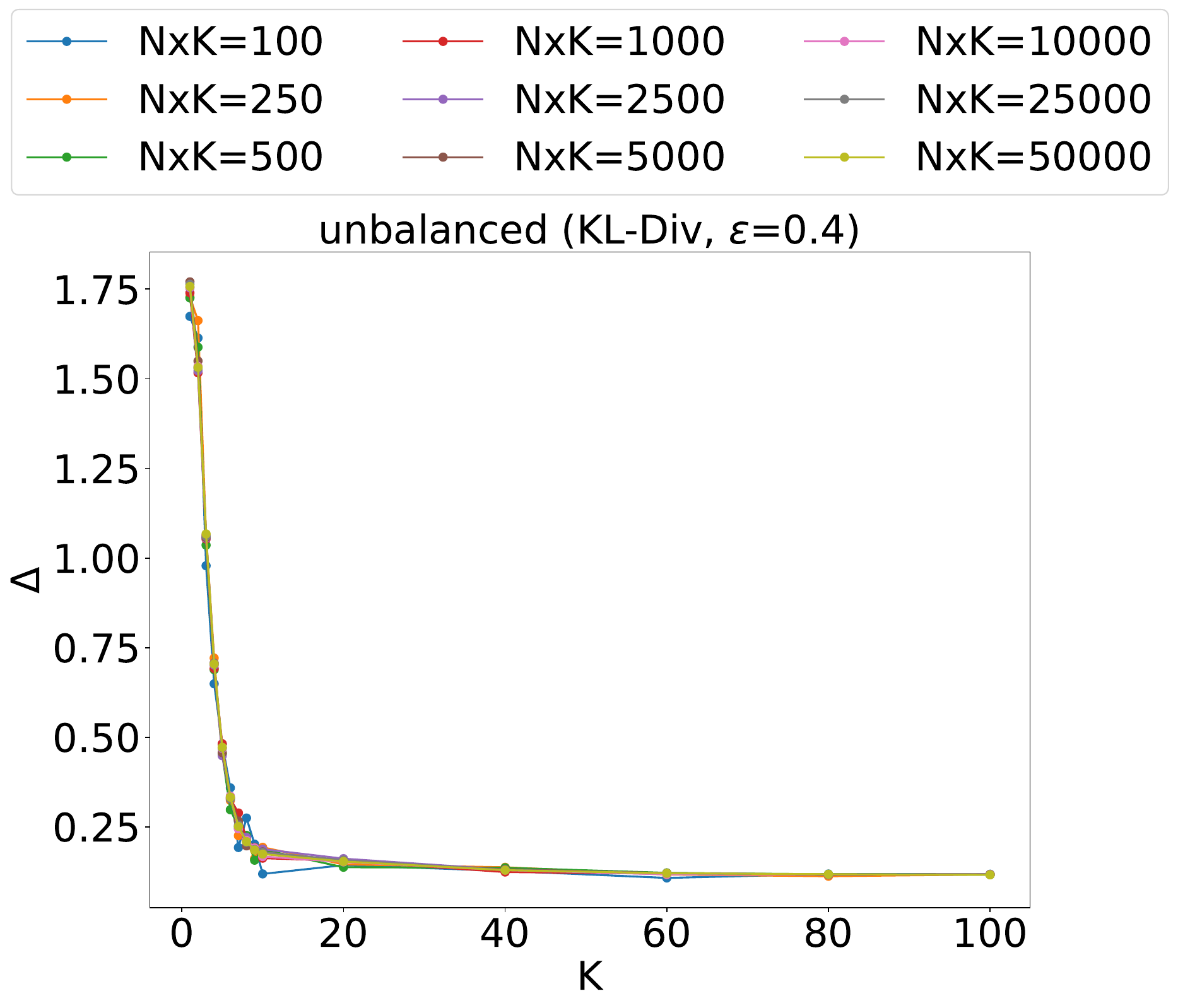}
    \caption{$\epsilon = 0.4$}
    \label{fig:gamma_delta_kl_cat3_e04}
  \end{subfigure}
  \caption{Effect sizes ($\Delta$) for unbalanced alphas with KL-divergence as the metric ($M=3$)}
  \label{fig:gamma_delta_kl_cat3}
\end{figure*}

\begin{figure*}
  \centering
  \begin{subfigure}[b]{0.24\linewidth}
    \centering
    \includegraphics[width=\linewidth]{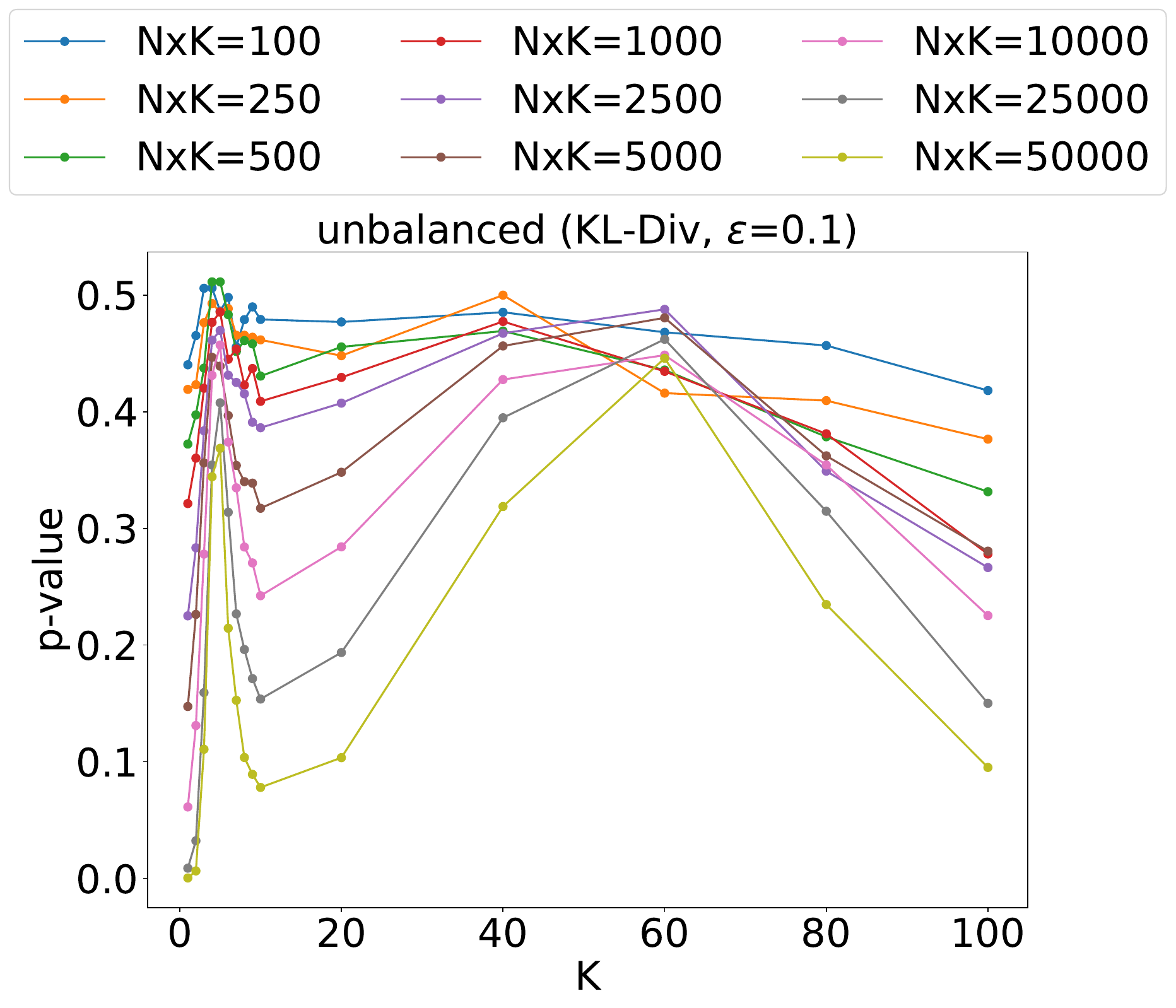}
    \caption{$\epsilon = 0.1$}
    \label{fig:gamma_kl_cat4_e01}
  \end{subfigure} \hfill
  \begin{subfigure}[b]{0.24\linewidth}
    \centering
    \includegraphics[width=\linewidth]{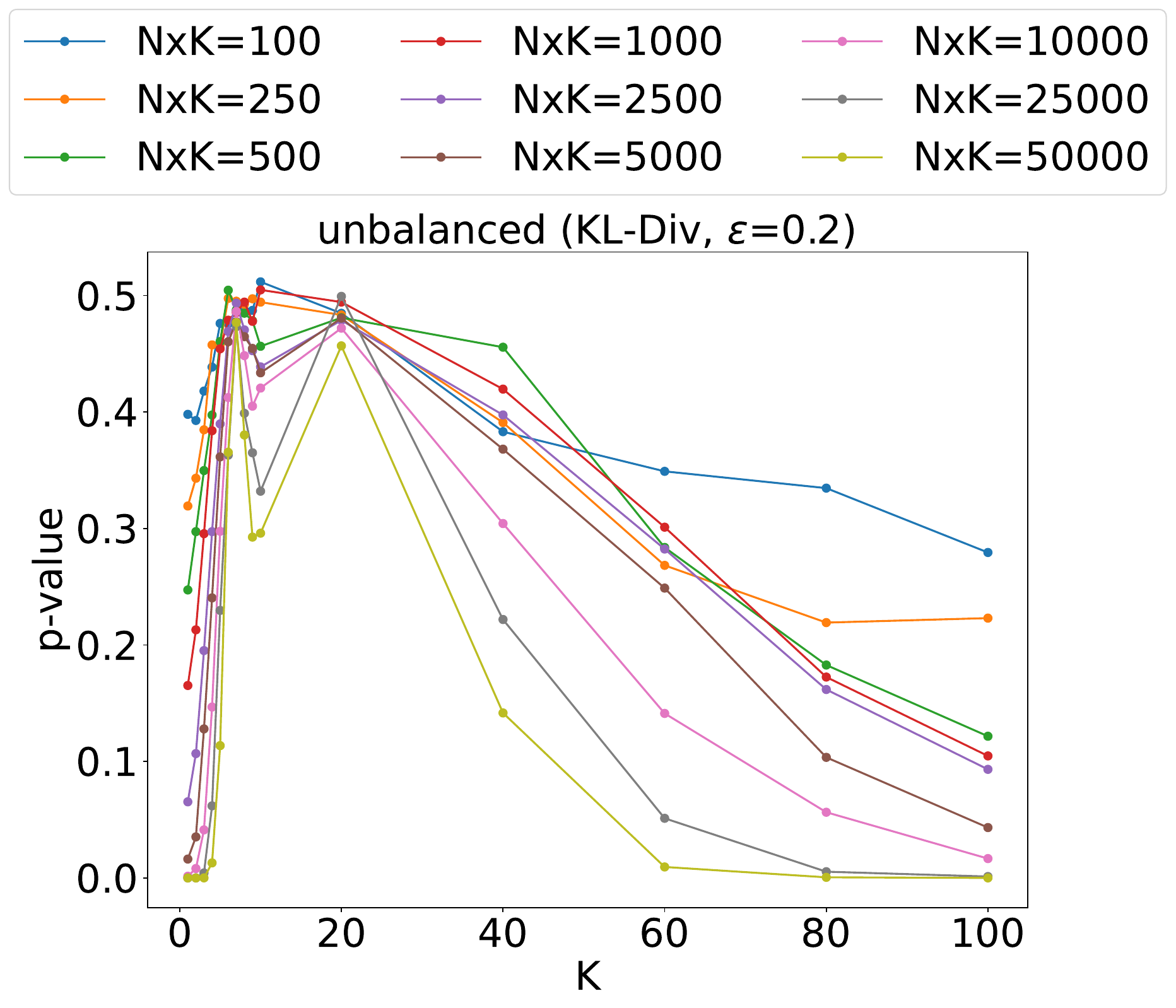}
    \caption{$\epsilon = 0.2$}
    \label{fig:gamma_kl_cat4_e02}
  \end{subfigure} \hfill
  \begin{subfigure}[b]{0.24\linewidth}
    \centering
    \includegraphics[width=\linewidth]{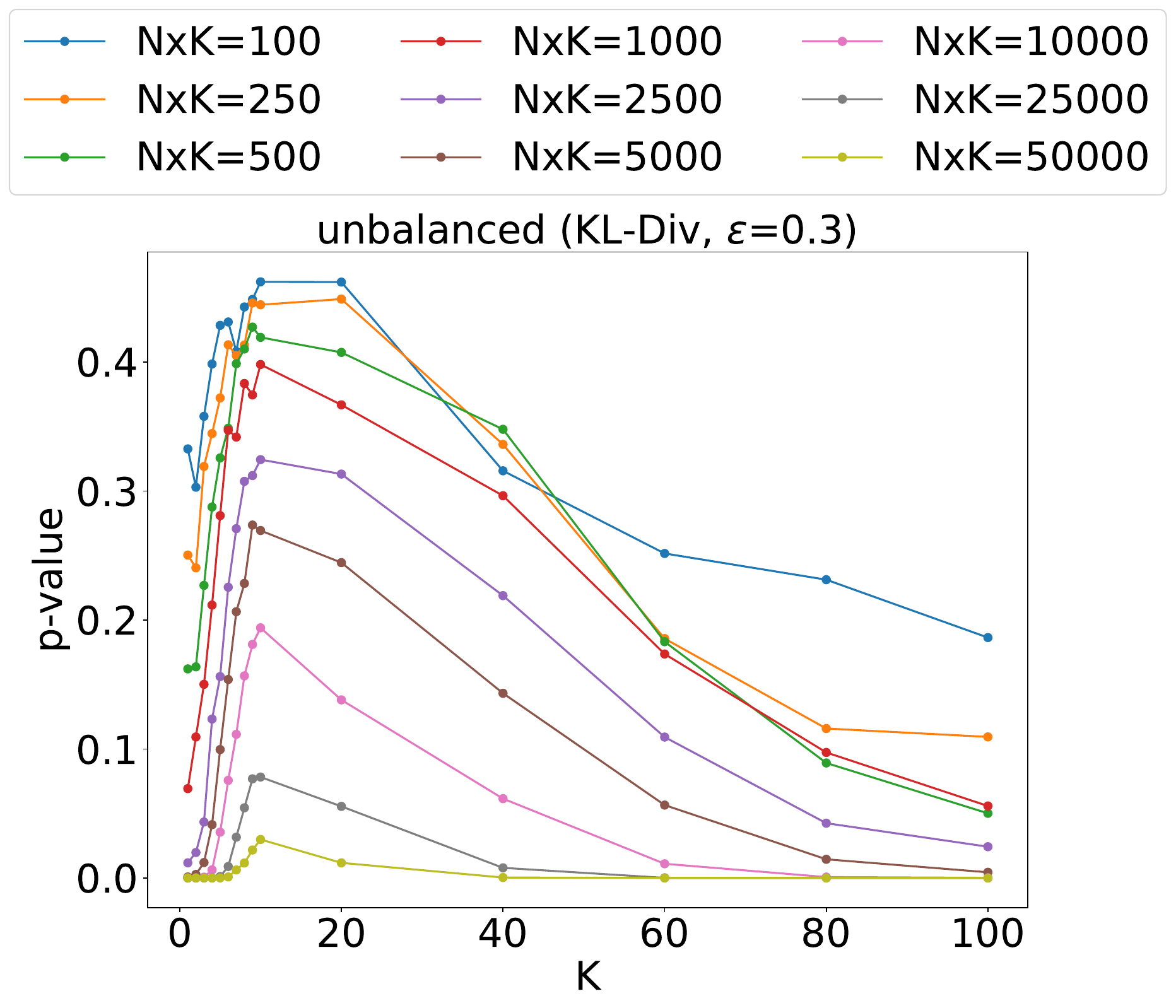}
    \caption{$\epsilon = 0.3$}
    \label{fig:gamma_kl_cat4_e03}
  \end{subfigure} \hfill
  \begin{subfigure}[b]{0.24\linewidth}
    \centering
    \includegraphics[width=\linewidth]{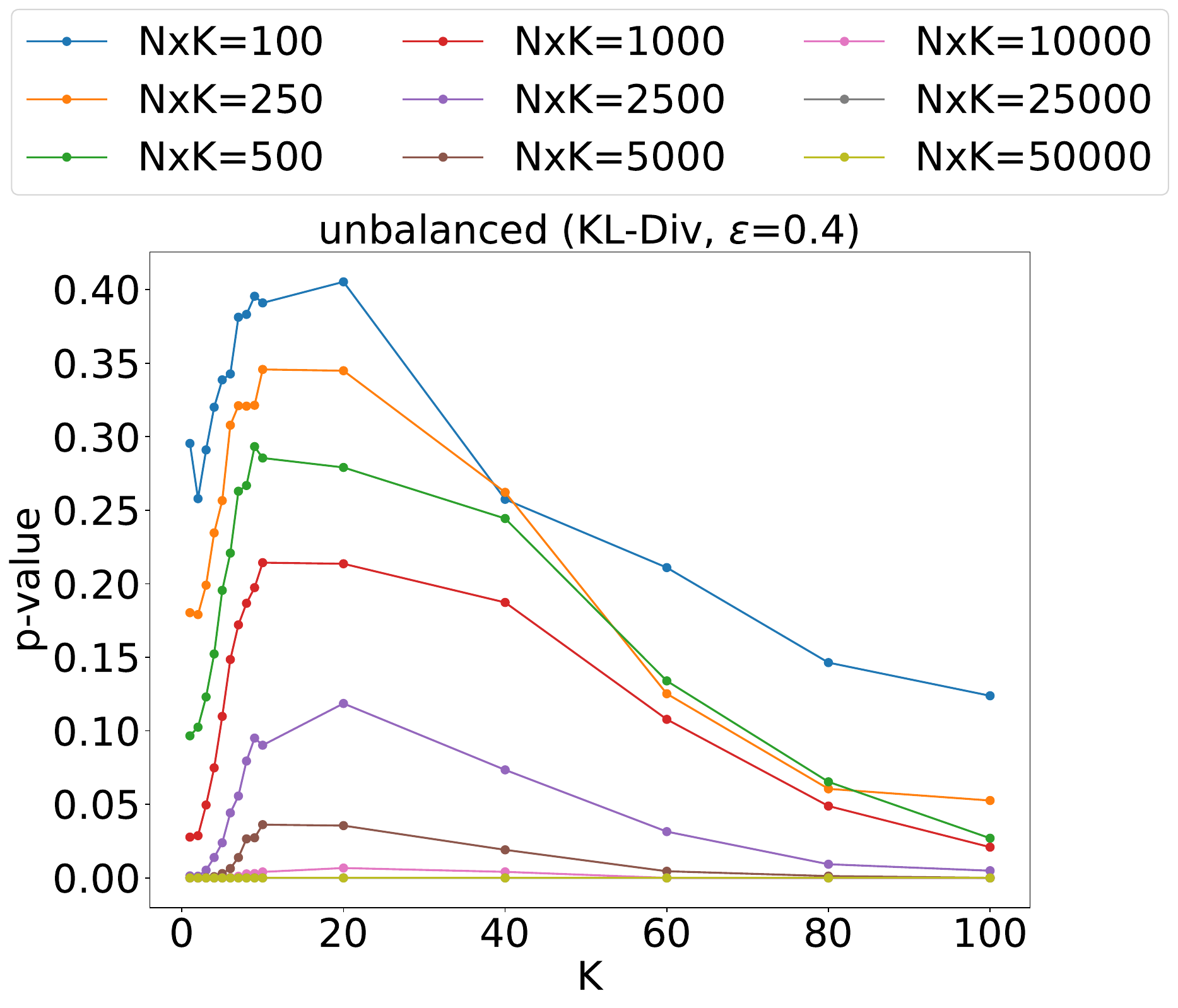}
    \caption{$\epsilon = 0.4$}
    \label{fig:gamma_kl_cat4_e04}
  \end{subfigure}
  \caption{P-value plots for unbalanced alphas with KL-divergence as the metric ($M=4$)}
  \label{fig:gamma_kl_cat4}
\end{figure*}

\begin{figure*}
  \centering
  \begin{subfigure}[b]{0.24\linewidth}
    \centering
    \includegraphics[width=\linewidth]{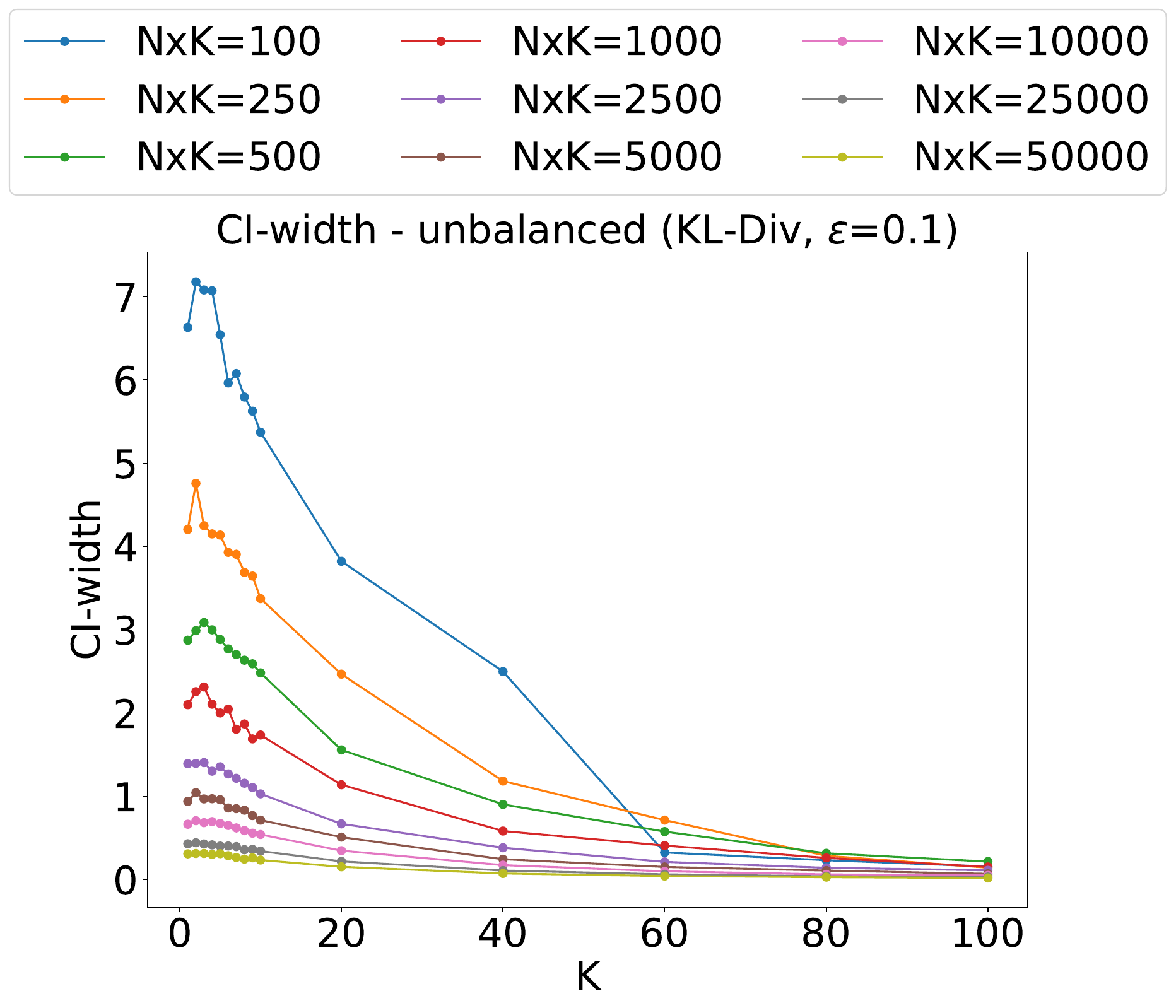}
    \caption{$\epsilon = 0.1$}
    \label{fig:gamma_ci_kl_cat4_e01}
  \end{subfigure} \hfill
  \begin{subfigure}[b]{0.24\linewidth}
    \centering
    \includegraphics[width=\linewidth]{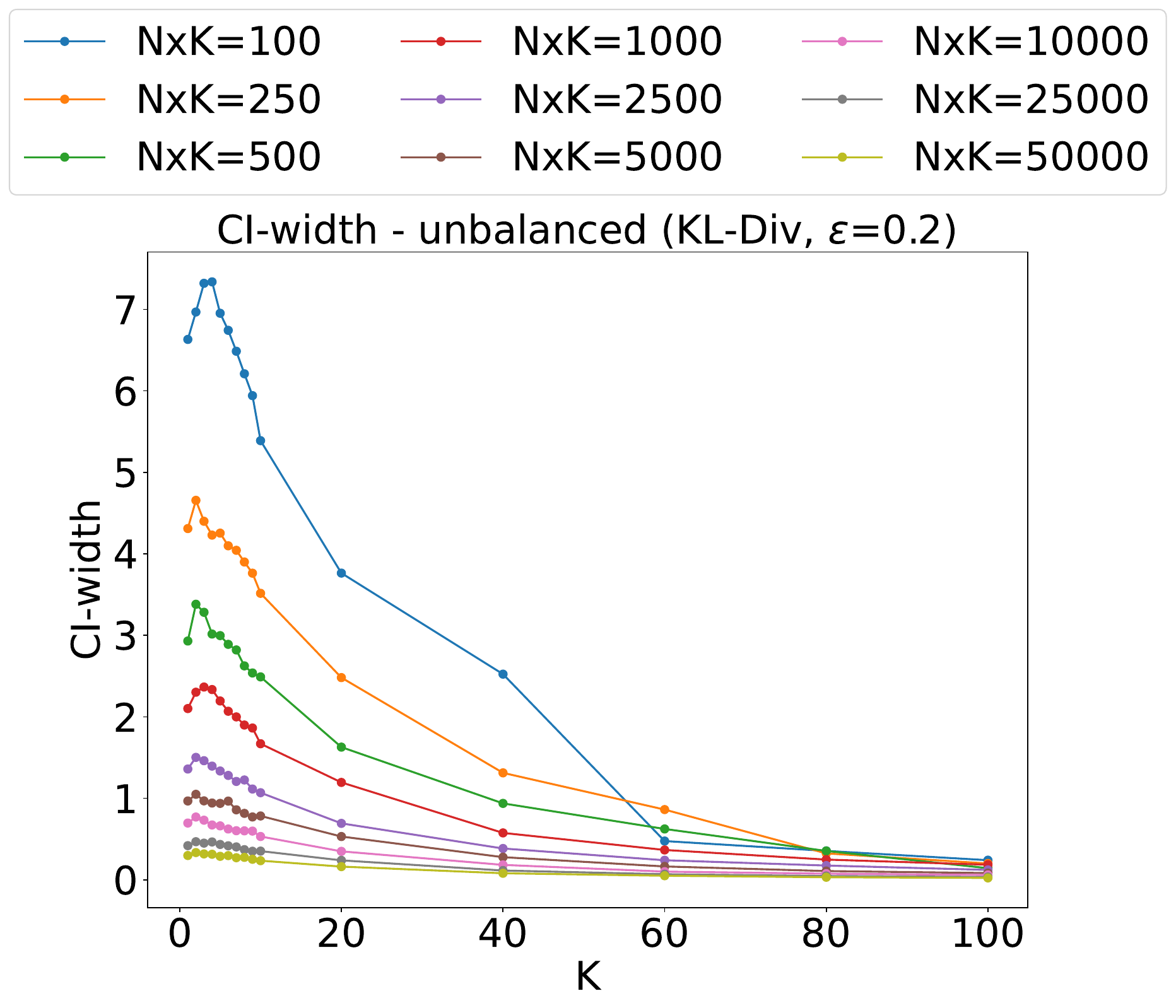}
    \caption{$\epsilon = 0.2$}
    \label{fig:gamma_ci_kl_cat4_e02}
  \end{subfigure} \hfill
  \begin{subfigure}[b]{0.24\linewidth}
    \centering
    \includegraphics[width=\linewidth]{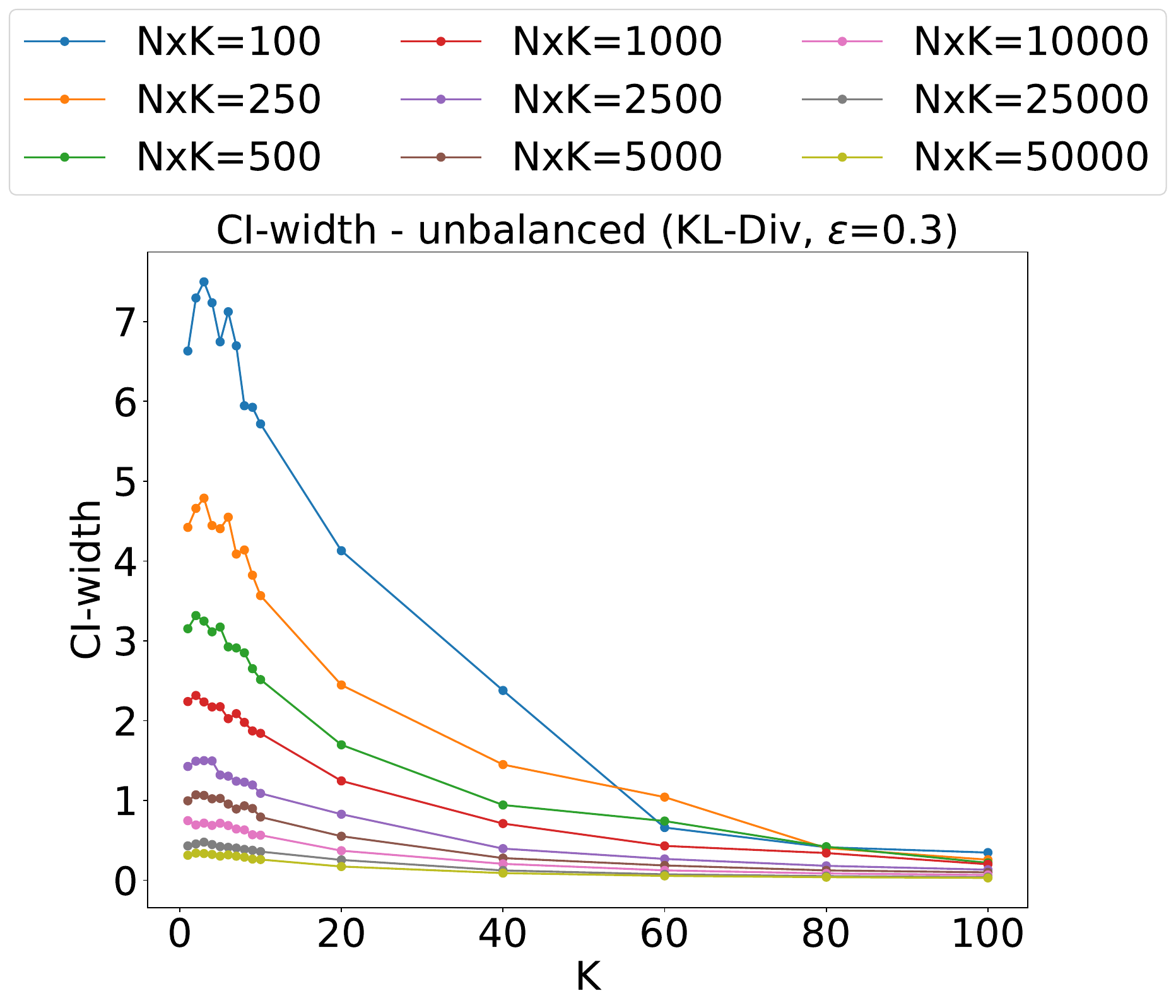}
    \caption{$\epsilon = 0.3$}
    \label{fig:gamma_ci_kl_cat4_e03}
  \end{subfigure} \hfill
  \begin{subfigure}[b]{0.24\linewidth}
    \centering
    \includegraphics[width=\linewidth]{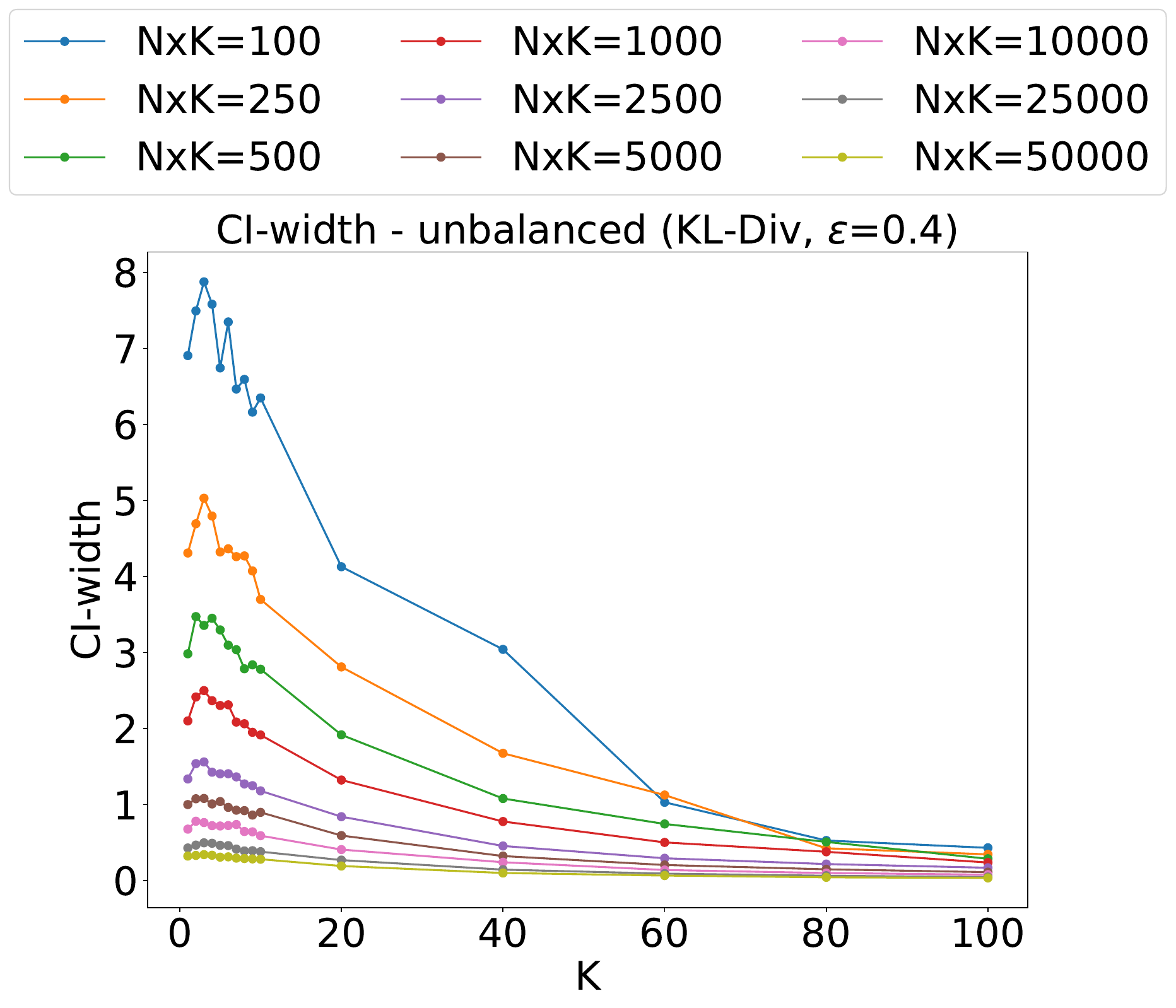}
    \caption{$\epsilon = 0.4$}
    \label{fig:gamma_ci_kl_cat4_e04}
  \end{subfigure}
  \caption{CI-width plots for unbalanced alphas with KL-divergence as the metric ($M=4$)}
  \label{fig:gamma_ci_kl_cat4}
\end{figure*}

\begin{figure*}
  \centering
  \begin{subfigure}[b]{0.24\linewidth}
    \centering
    \includegraphics[width=\linewidth]{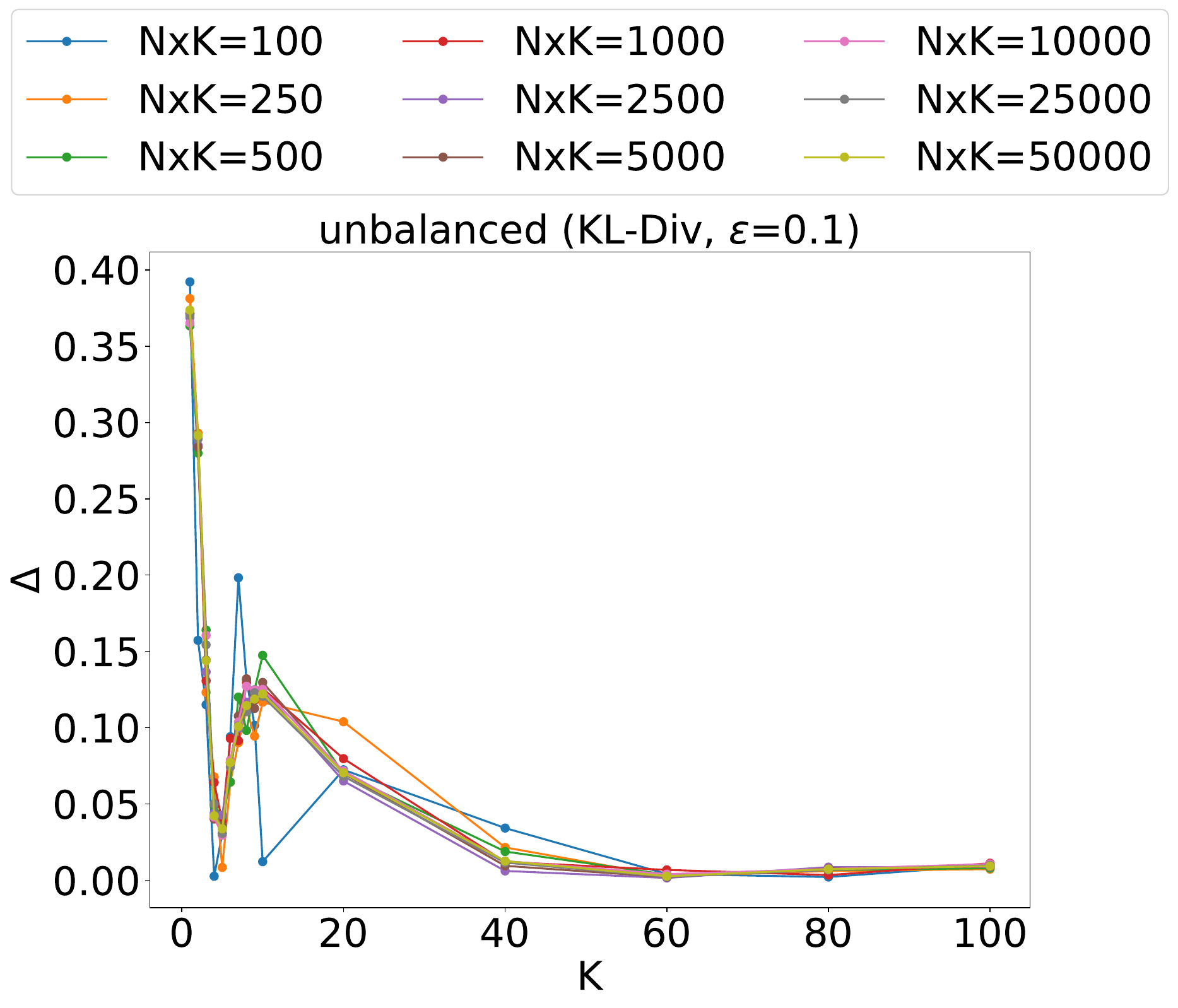}
    \caption{$\epsilon = 0.1$}
    \label{fig:gamma_delta_kl_cat4_e01}
  \end{subfigure} \hfill
  \begin{subfigure}[b]{0.24\linewidth}
    \centering
    \includegraphics[width=\linewidth]{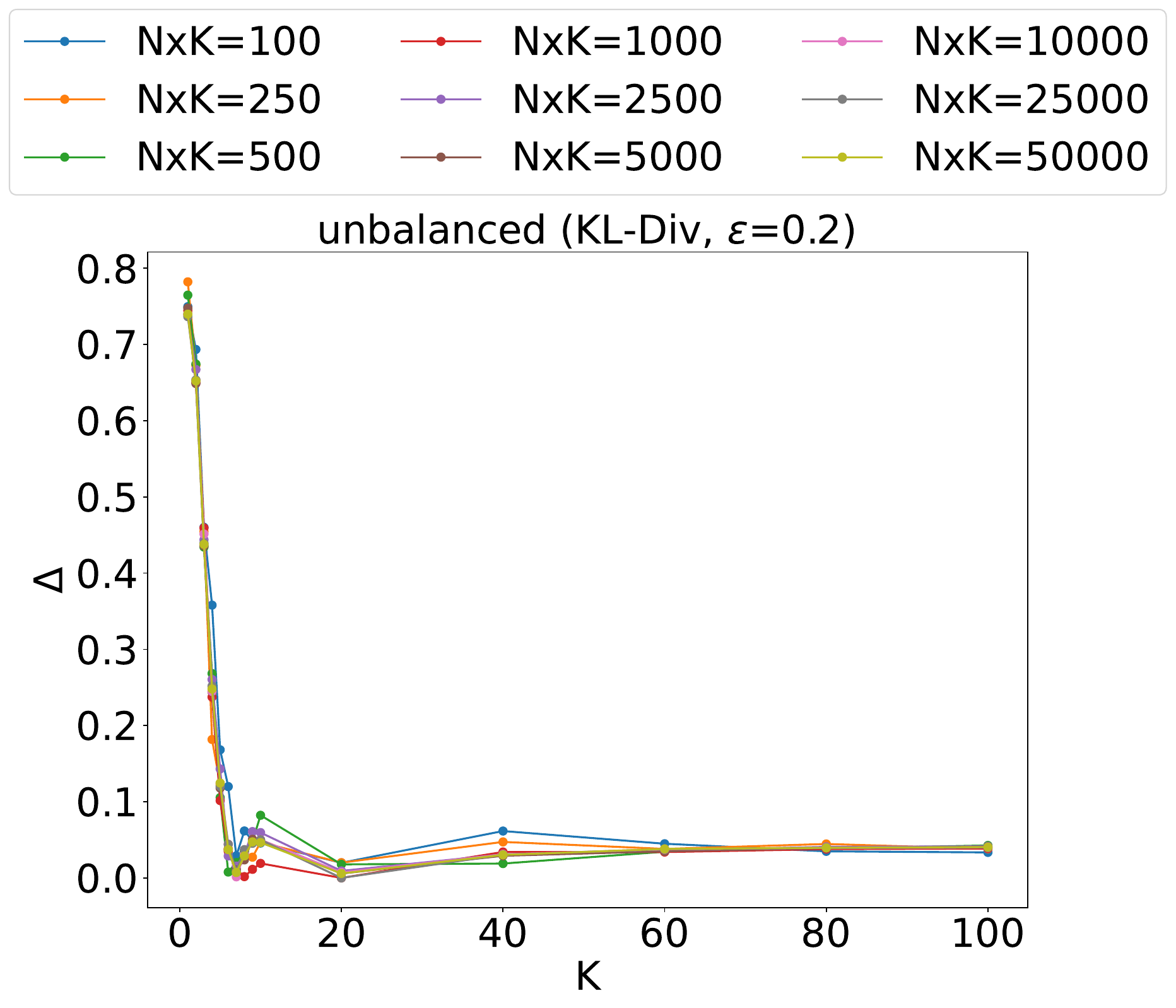}
    \caption{$\epsilon = 0.2$}
    \label{fig:gamma_delta_kl_cat4_e02}
  \end{subfigure} \hfill
  \begin{subfigure}[b]{0.24\linewidth}
    \centering
    \includegraphics[width=\linewidth]{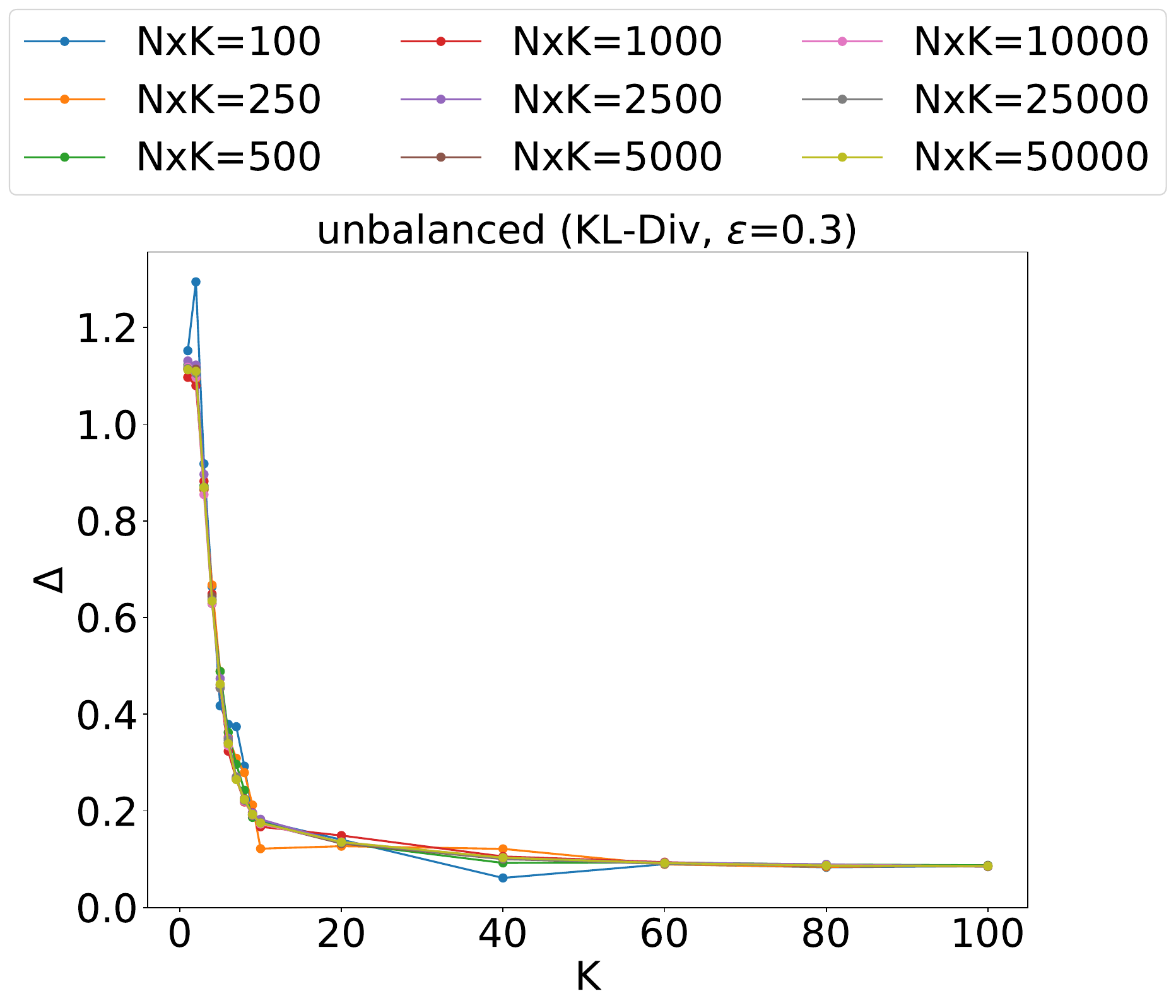}
    \caption{$\epsilon = 0.3$}
    \label{fig:gamma_delta_kl_cat4_e03}
  \end{subfigure} \hfill
  \begin{subfigure}[b]{0.24\linewidth}
    \centering
    \includegraphics[width=\linewidth]{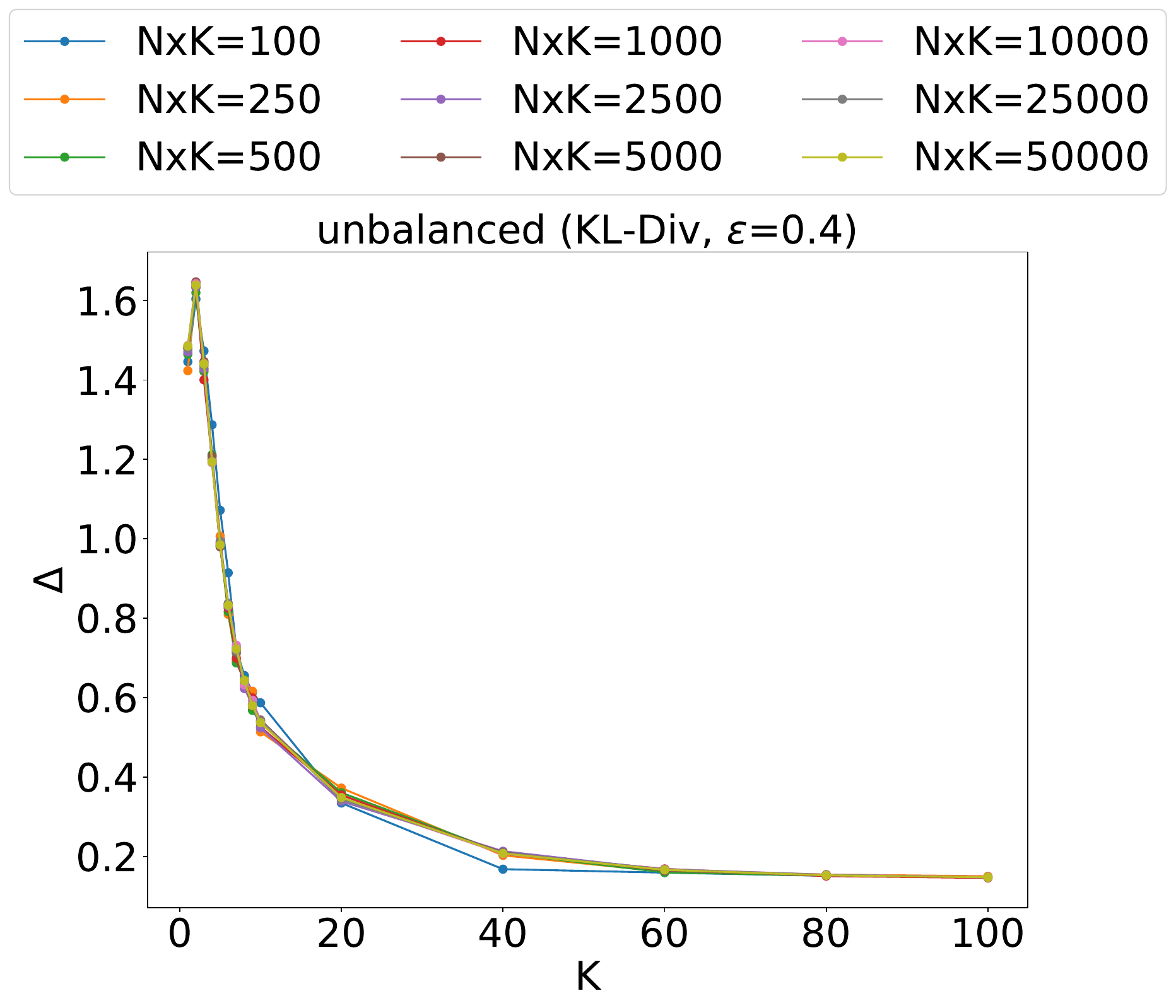}
    \caption{$\epsilon = 0.4$}
    \label{fig:gamma_delta_kl_cat4_e04}
  \end{subfigure}
  \caption{Effect sizes ($\Delta$) for unbalanced alphas with KL-divergence as the metric ($M=4$)}
  \label{fig:gamma_delta_kl_cat4}
\end{figure*}

\begin{figure*}
  \centering
  \begin{subfigure}[b]{0.24\linewidth}
    \centering
    \includegraphics[width=\linewidth]{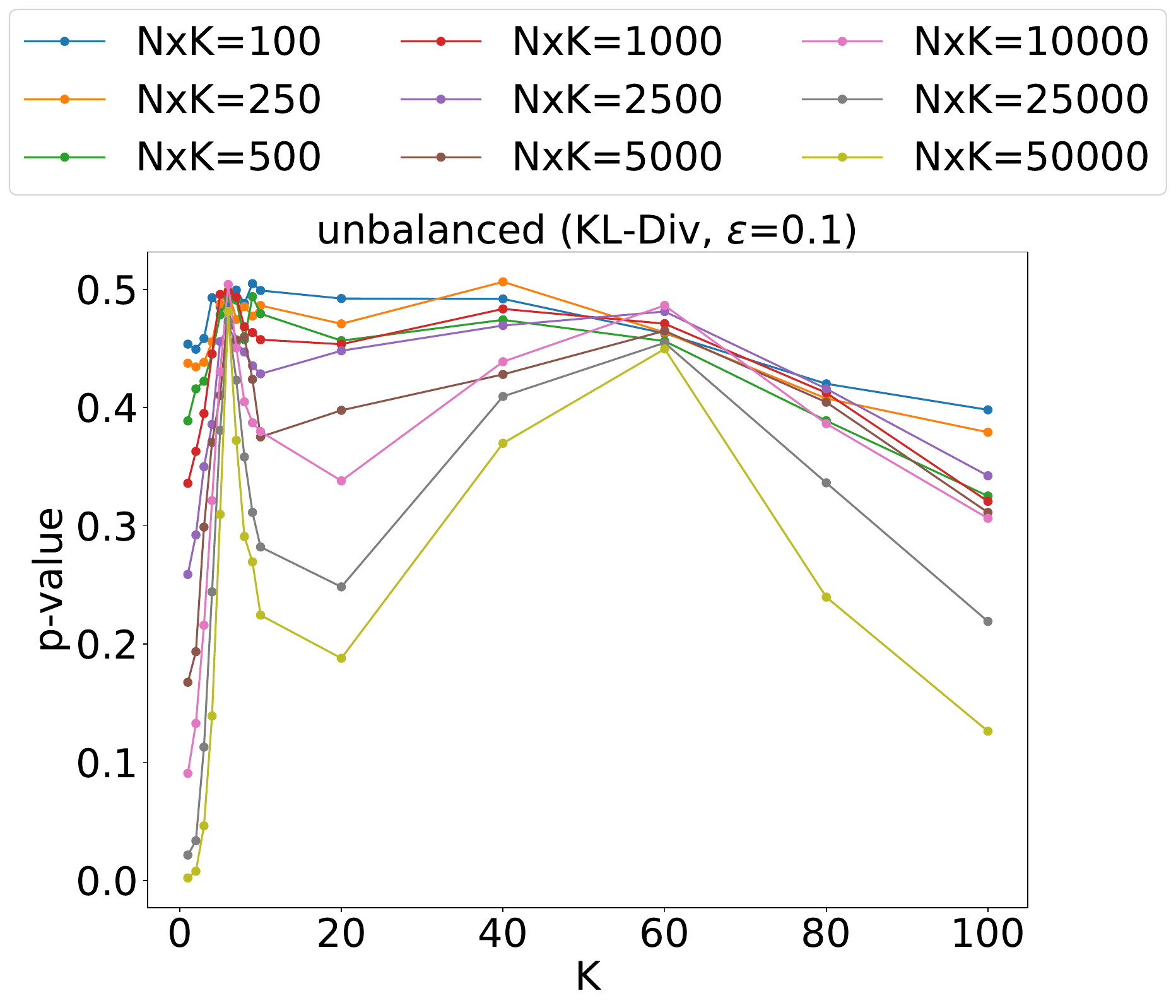}
    \caption{$\epsilon = 0.1$}
    \label{fig:gamma_kl_cat5_e01}
  \end{subfigure} \hfill
  \begin{subfigure}[b]{0.24\linewidth}
    \centering
    \includegraphics[width=\linewidth]{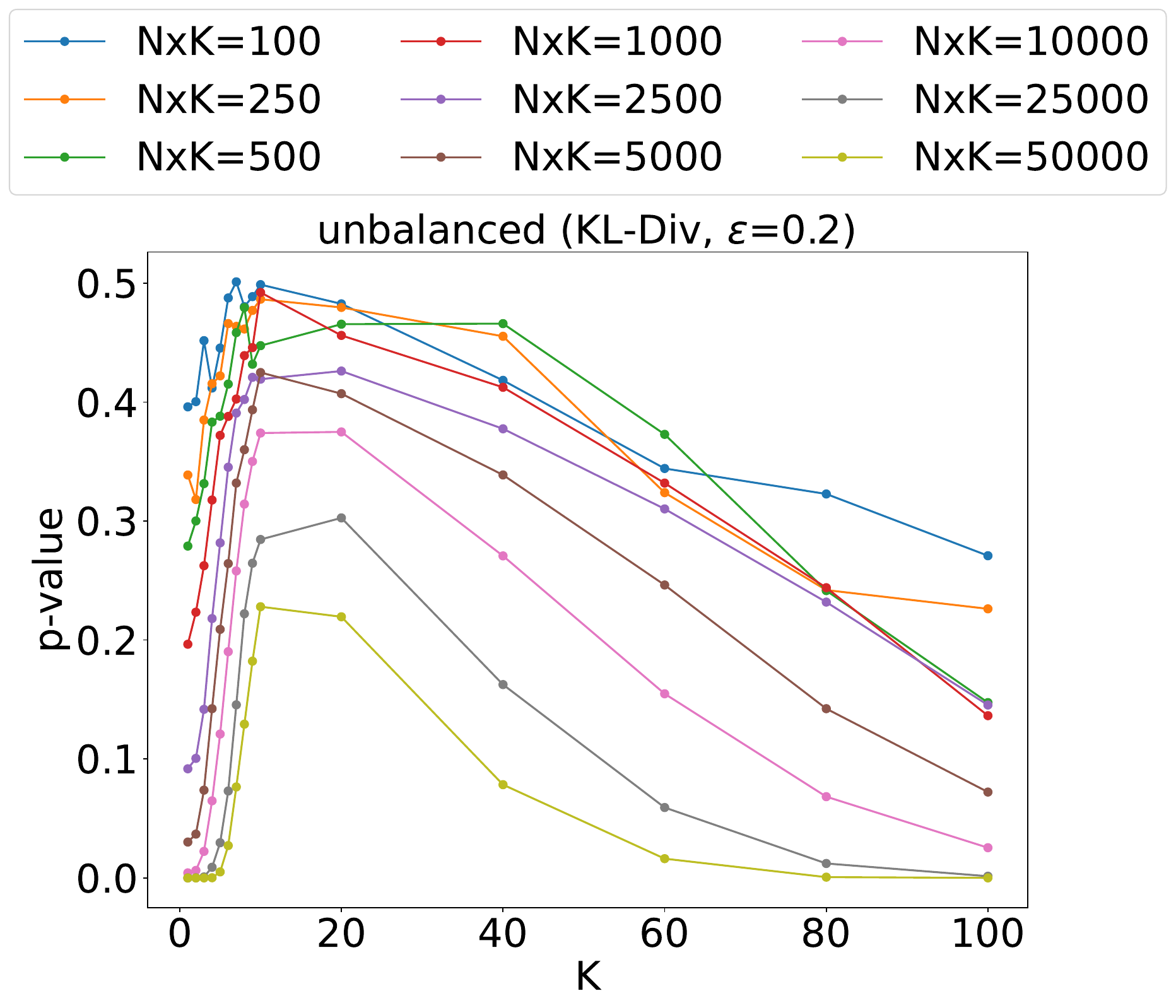}
    \caption{$\epsilon = 0.2$}
    \label{fig:gamma_kl_cat5_e02}
  \end{subfigure} \hfill
  \begin{subfigure}[b]{0.24\linewidth}
    \centering
    \includegraphics[width=\linewidth]{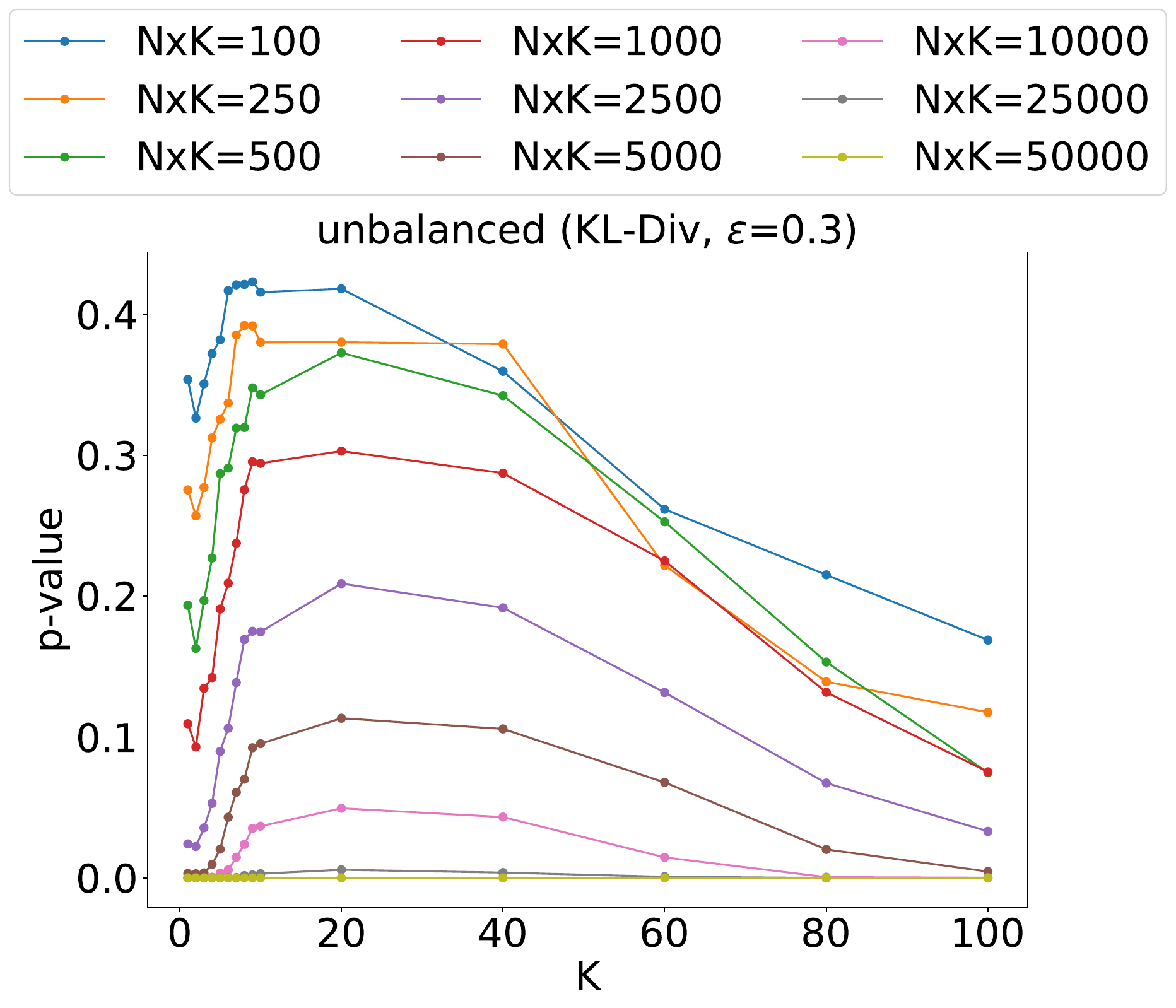}
    \caption{$\epsilon = 0.3$}
    \label{fig:gamma_kl_cat5_e03}
  \end{subfigure} \hfill
  \begin{subfigure}[b]{0.24\linewidth}
    \centering
    \includegraphics[width=\linewidth]{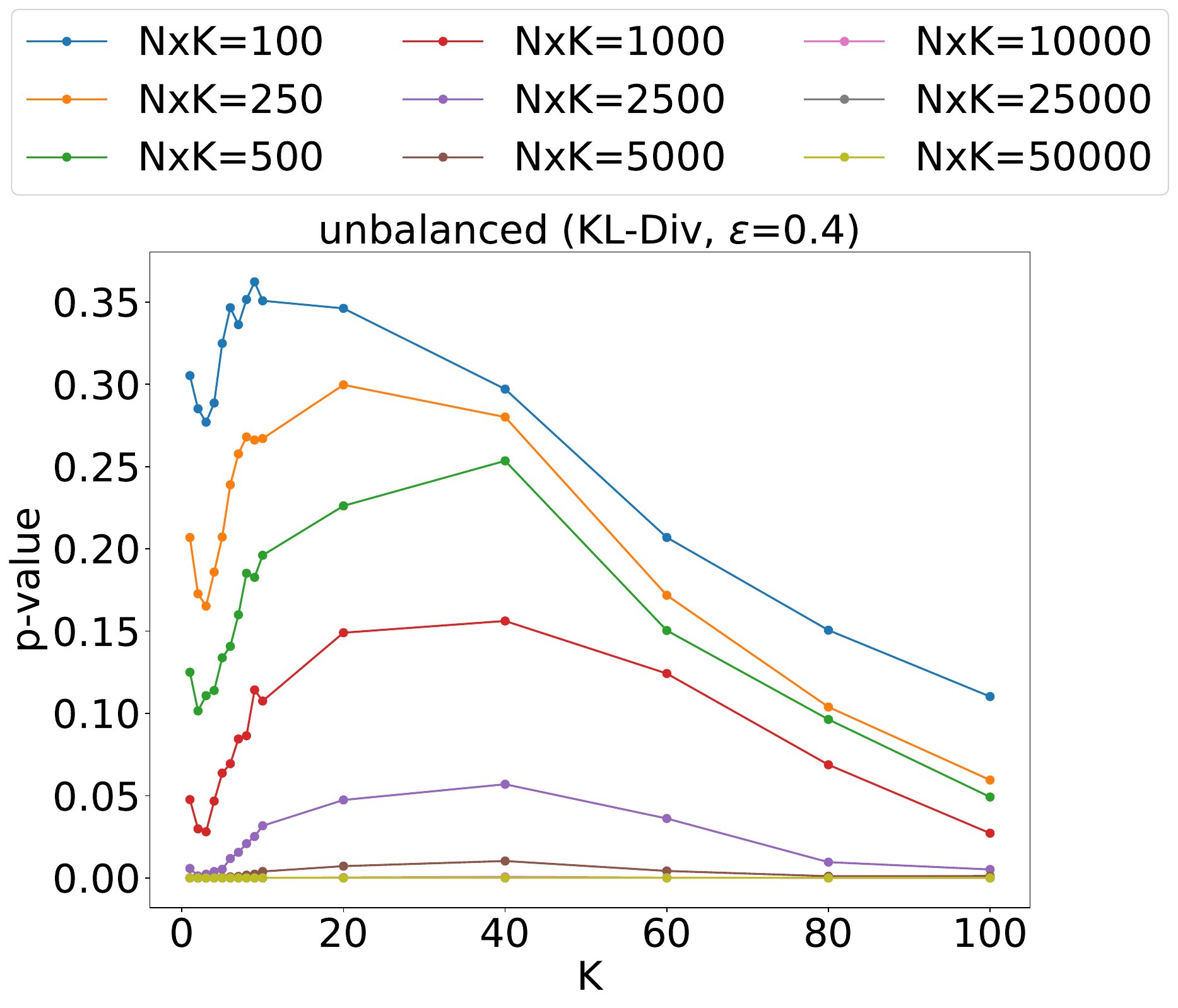}
    \caption{$\epsilon = 0.4$}
    \label{fig:gamma_kl_cat5_e04}
  \end{subfigure}
  \caption{P-value plots for unbalanced alphas with KL-divergence as the metric ($M=5$)}
  \label{fig:gamma_kl_cat5}
\end{figure*}

\begin{figure*}
  \centering
  \begin{subfigure}[b]{0.24\linewidth}
    \centering
    \includegraphics[width=\linewidth]{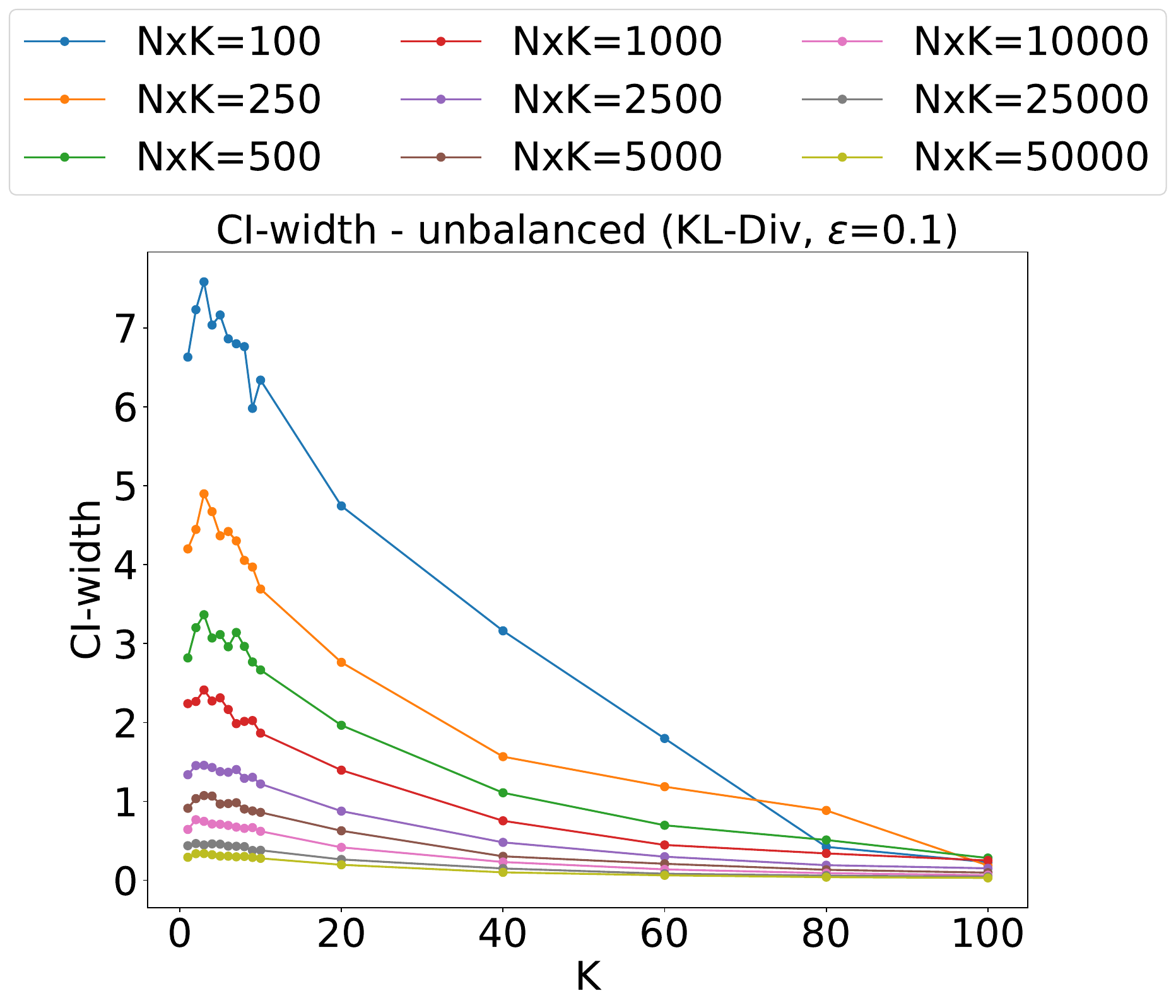}
    \caption{$\epsilon = 0.1$}
    \label{fig:gamma_ci_kl_cat5_e01}
  \end{subfigure} \hfill
  \begin{subfigure}[b]{0.24\linewidth}
    \centering
    \includegraphics[width=\linewidth]{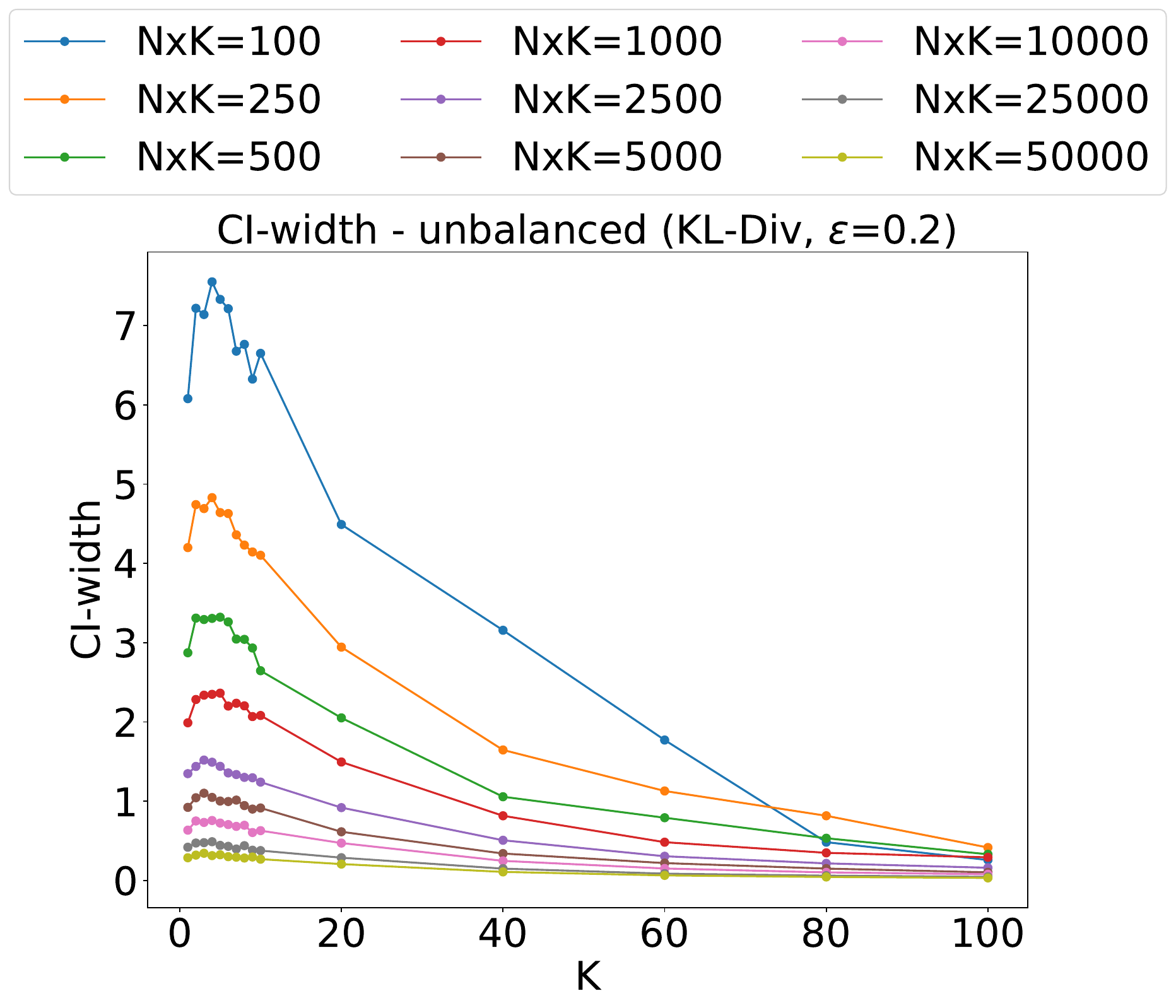}
    \caption{$\epsilon = 0.2$}
    \label{fig:gamma_ci_kl_cat5_e02}
  \end{subfigure} \hfill
  \begin{subfigure}[b]{0.24\linewidth}
    \centering
    \includegraphics[width=\linewidth]{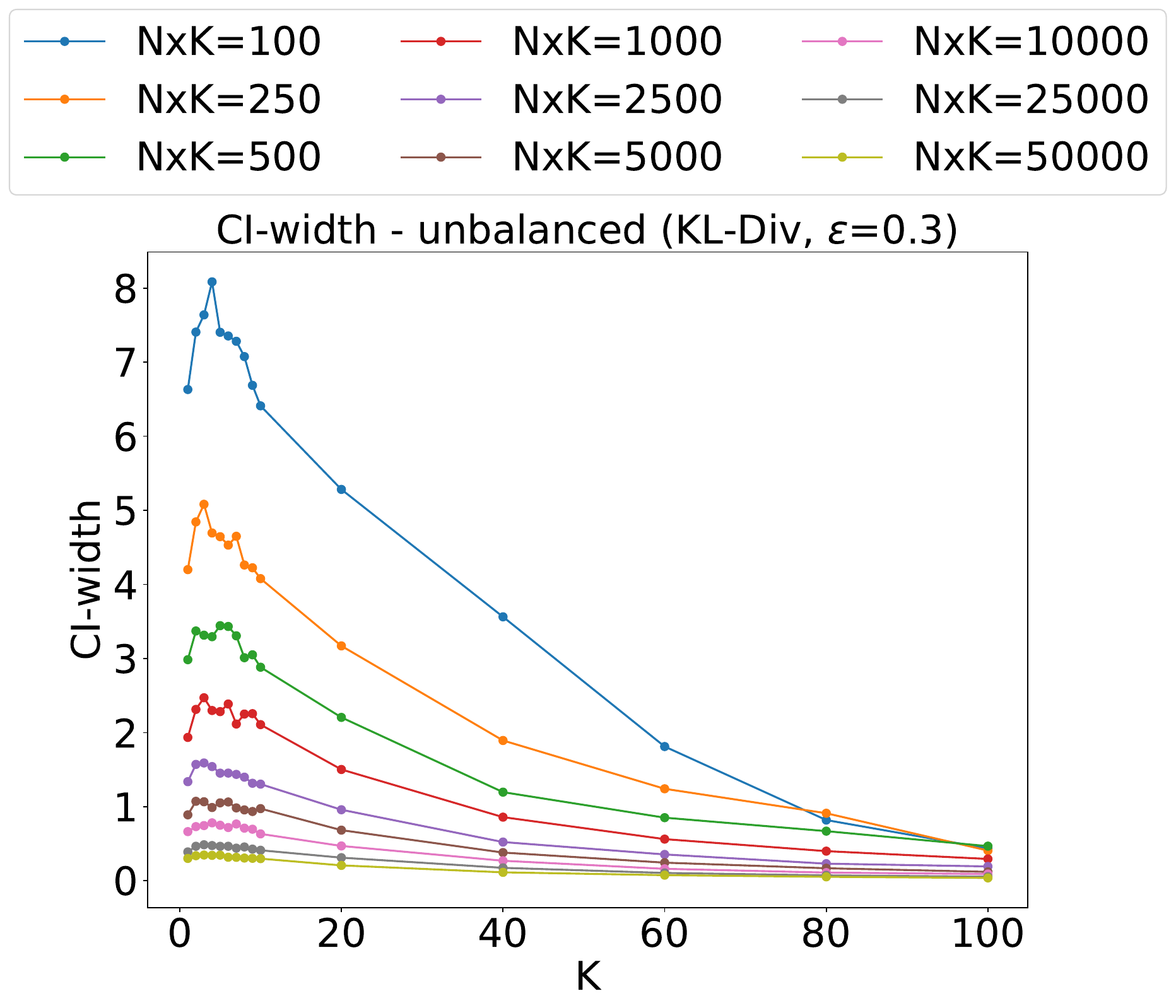}
    \caption{$\epsilon = 0.3$}
    \label{fig:gamma_ci_kl_cat5_e03}
  \end{subfigure} \hfill
  \begin{subfigure}[b]{0.24\linewidth}
    \centering
    \includegraphics[width=\linewidth]{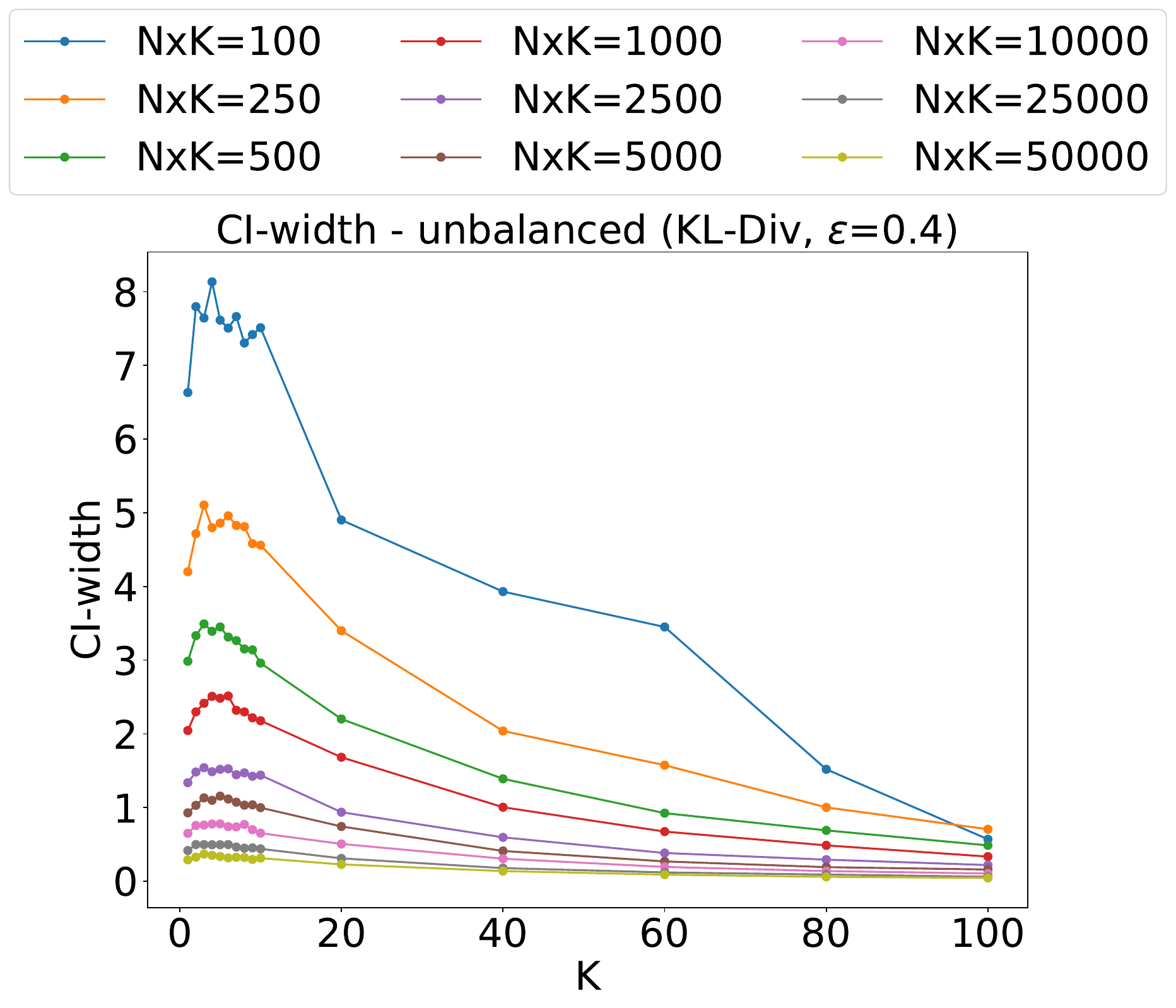}
    \caption{$\epsilon = 0.4$}
    \label{fig:gamma_ci_kl_cat5_e04}
  \end{subfigure}
  \caption{CI-width plots for unbalanced alphas with KL-divergence as the metric ($M=5$)}
  \label{fig:gamma_ci_kl_cat5}
\end{figure*}

\begin{figure*}
  \centering
  \begin{subfigure}[b]{0.24\linewidth}
    \centering
    \includegraphics[width=\linewidth]{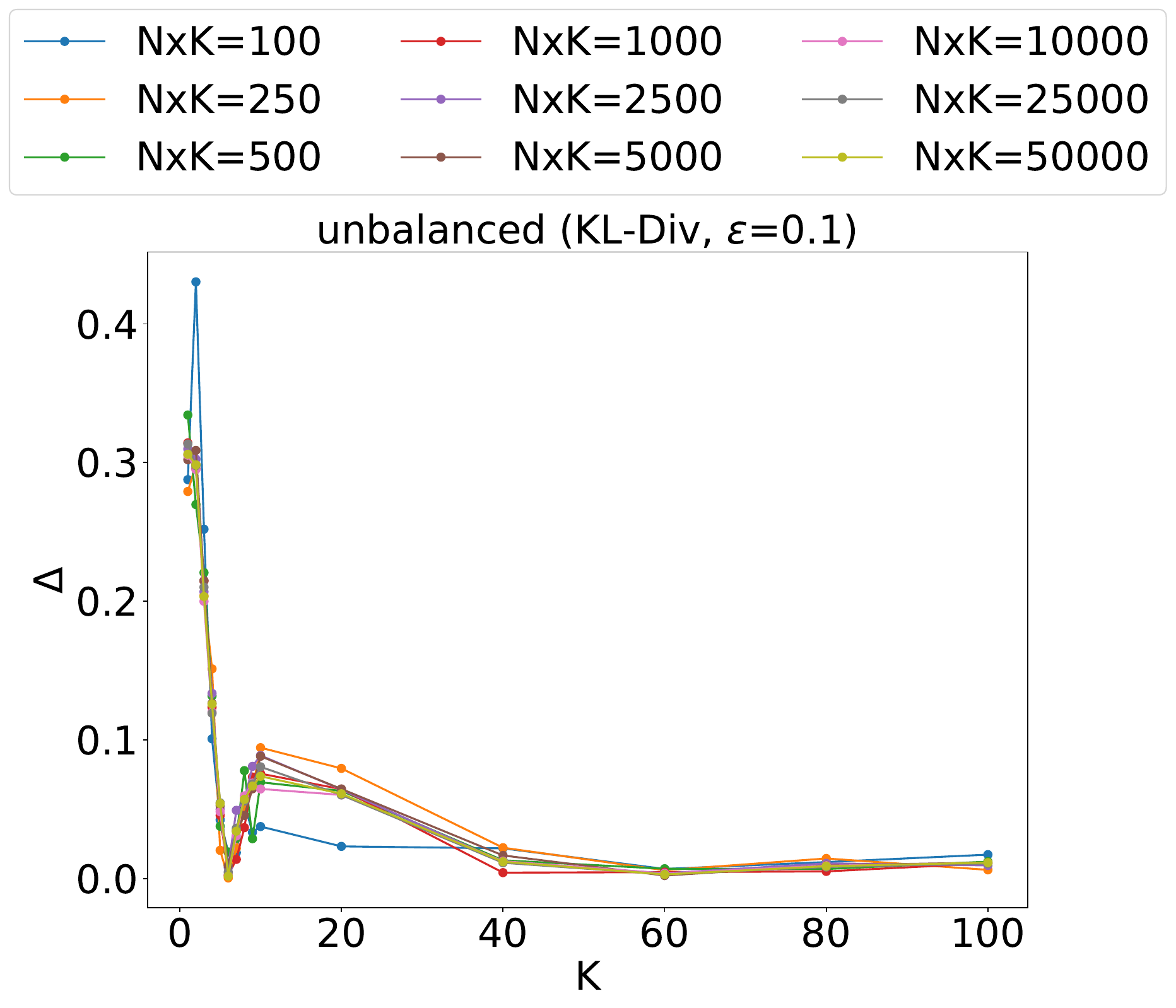}
    \caption{$\epsilon = 0.1$}
    \label{fig:gamma_delta_kl_cat5_e01}
  \end{subfigure} \hfill
  \begin{subfigure}[b]{0.24\linewidth}
    \centering
    \includegraphics[width=\linewidth]{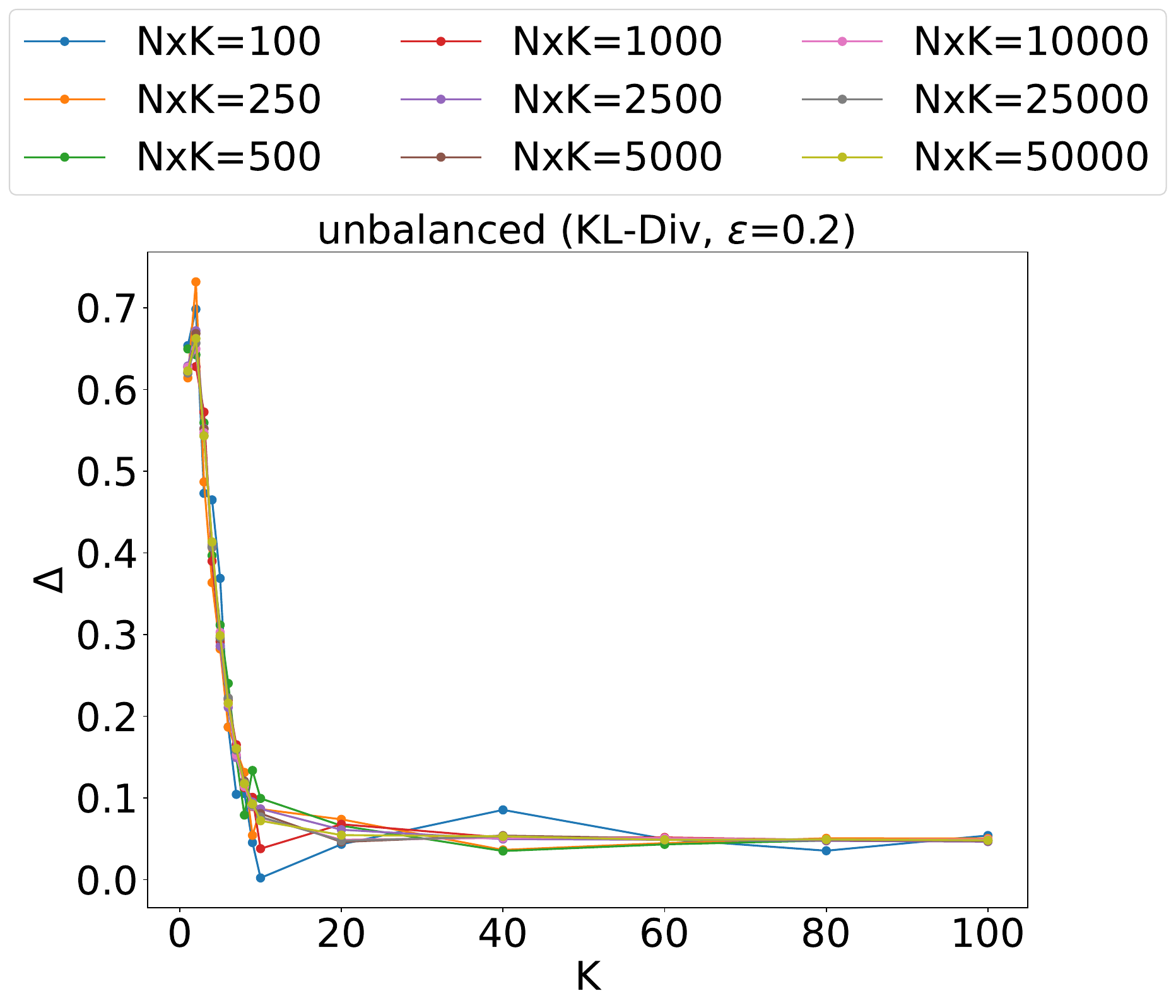}
    \caption{$\epsilon = 0.2$}
    \label{fig:gamma_delta_kl_cat5_e02}
  \end{subfigure} \hfill
  \begin{subfigure}[b]{0.24\linewidth}
    \centering
    \includegraphics[width=\linewidth]{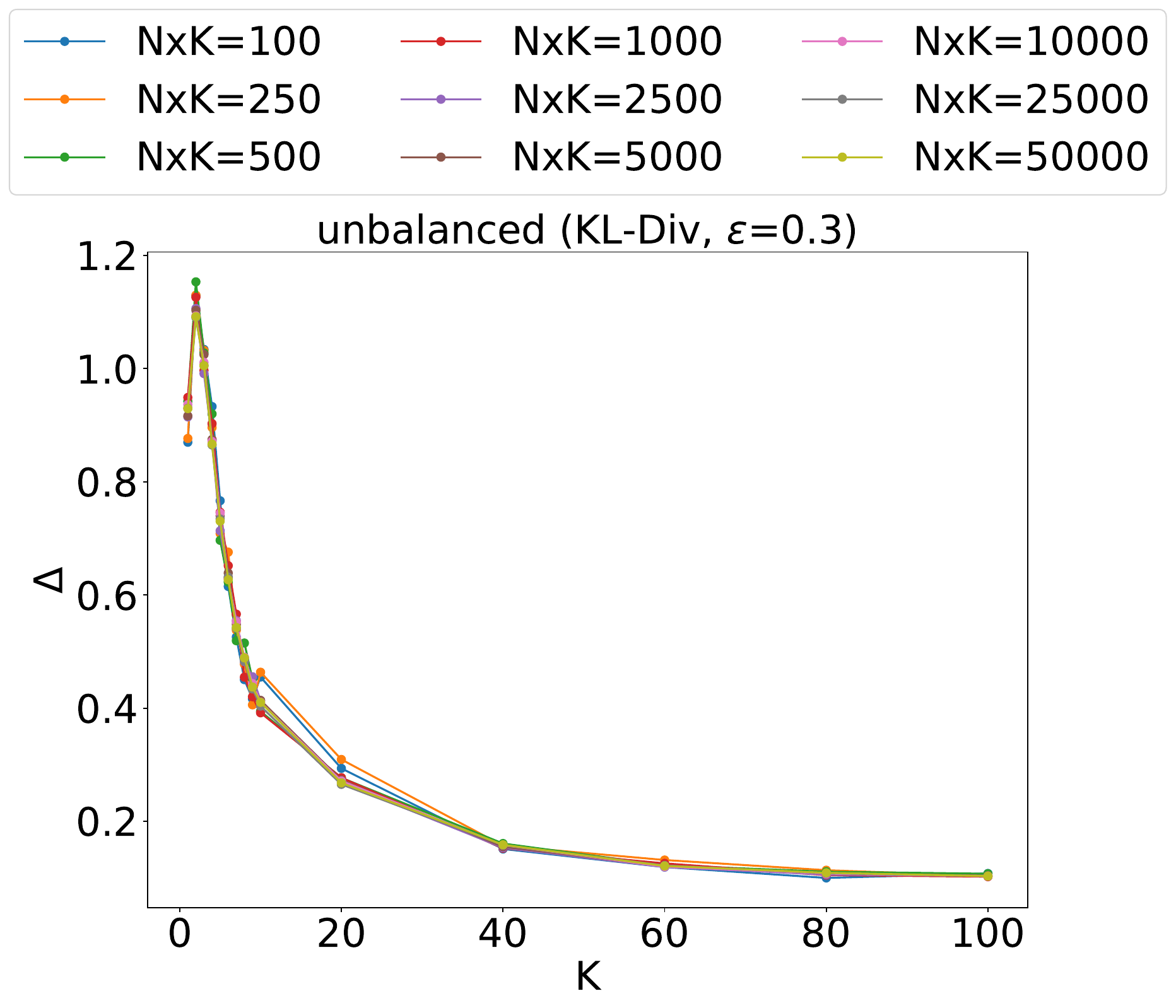}
    \caption{$\epsilon = 0.3$}
    \label{fig:gamma_delta_kl_cat5_e03}
  \end{subfigure} \hfill
  \begin{subfigure}[b]{0.24\linewidth}
    \centering
    \includegraphics[width=\linewidth]{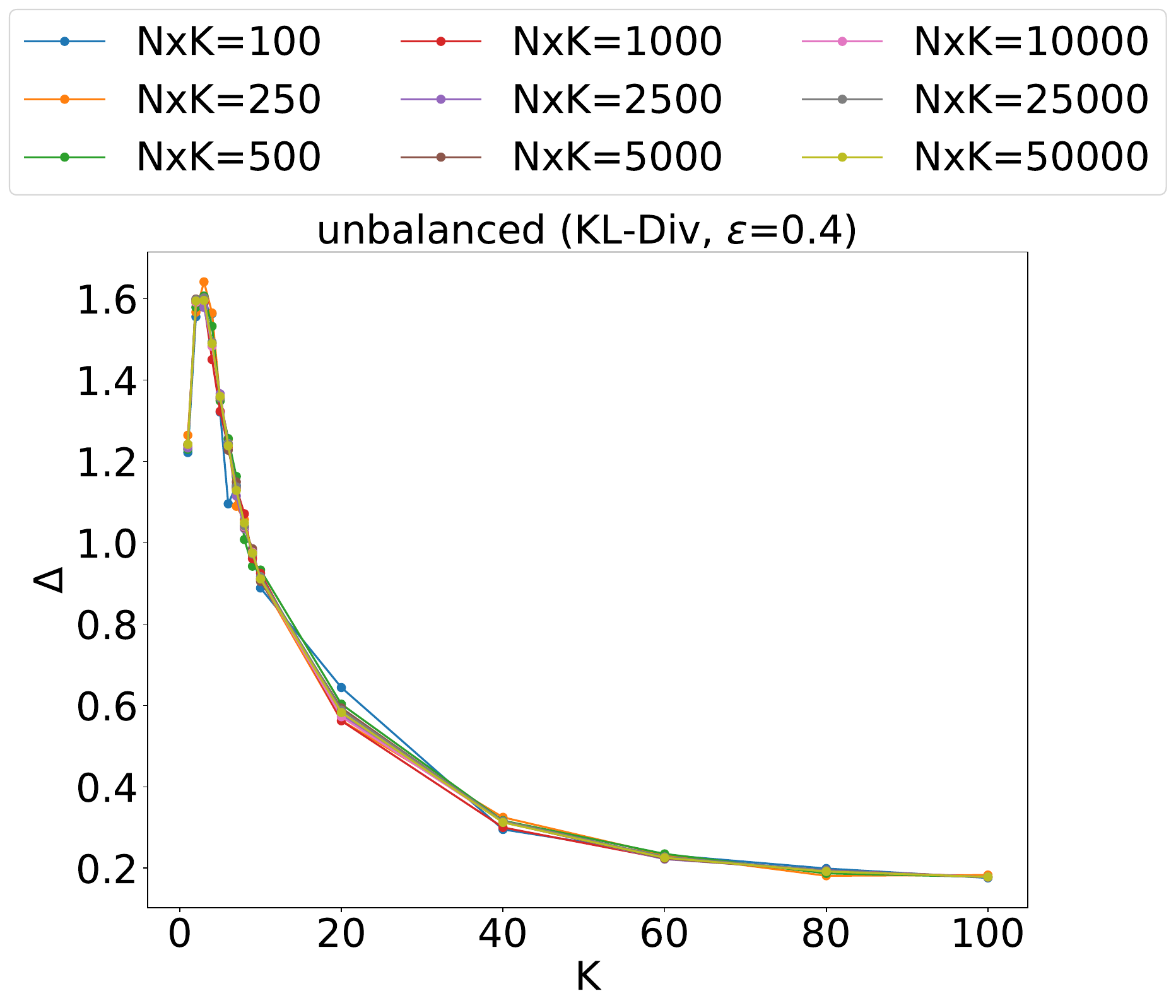}
    \caption{$\epsilon = 0.4$}
    \label{fig:gamma_delta_kl_cat5_e04}
  \end{subfigure}
  \caption{Effect sizes ($\Delta$) for unbalanced alphas with KL-divergence as the metric ($M=5$)}
  \label{fig:gamma_delta_kl_cat5}
\end{figure*}

\begin{figure*}
  \centering
  \begin{subfigure}[b]{0.24\linewidth}
    \centering
    \includegraphics[width=\linewidth]{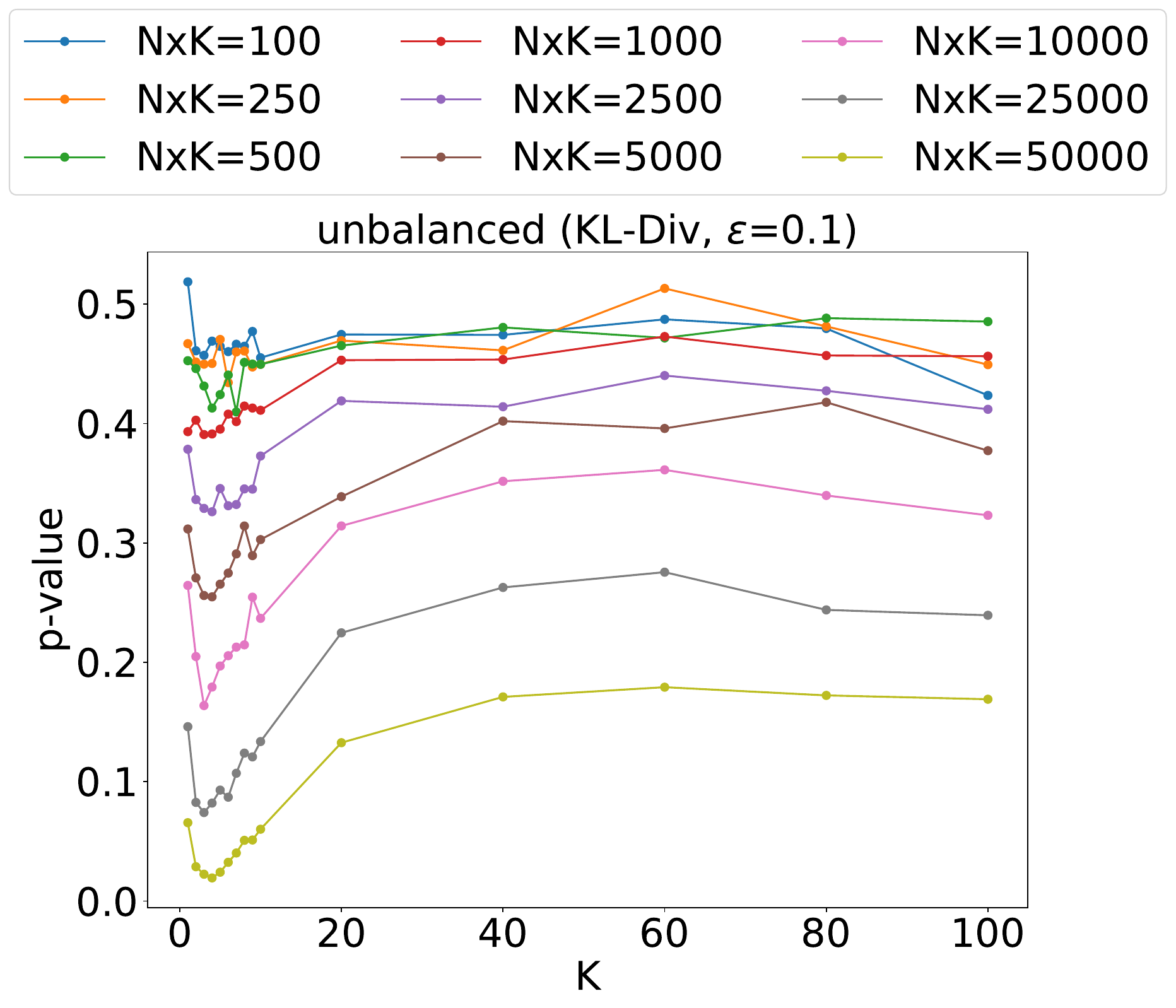}
    \caption{$\epsilon = 0.1$}
    \label{fig:gamma_kl_cat12_e01}
  \end{subfigure} \hfill
  \begin{subfigure}[b]{0.24\linewidth}
    \centering
    \includegraphics[width=\linewidth]{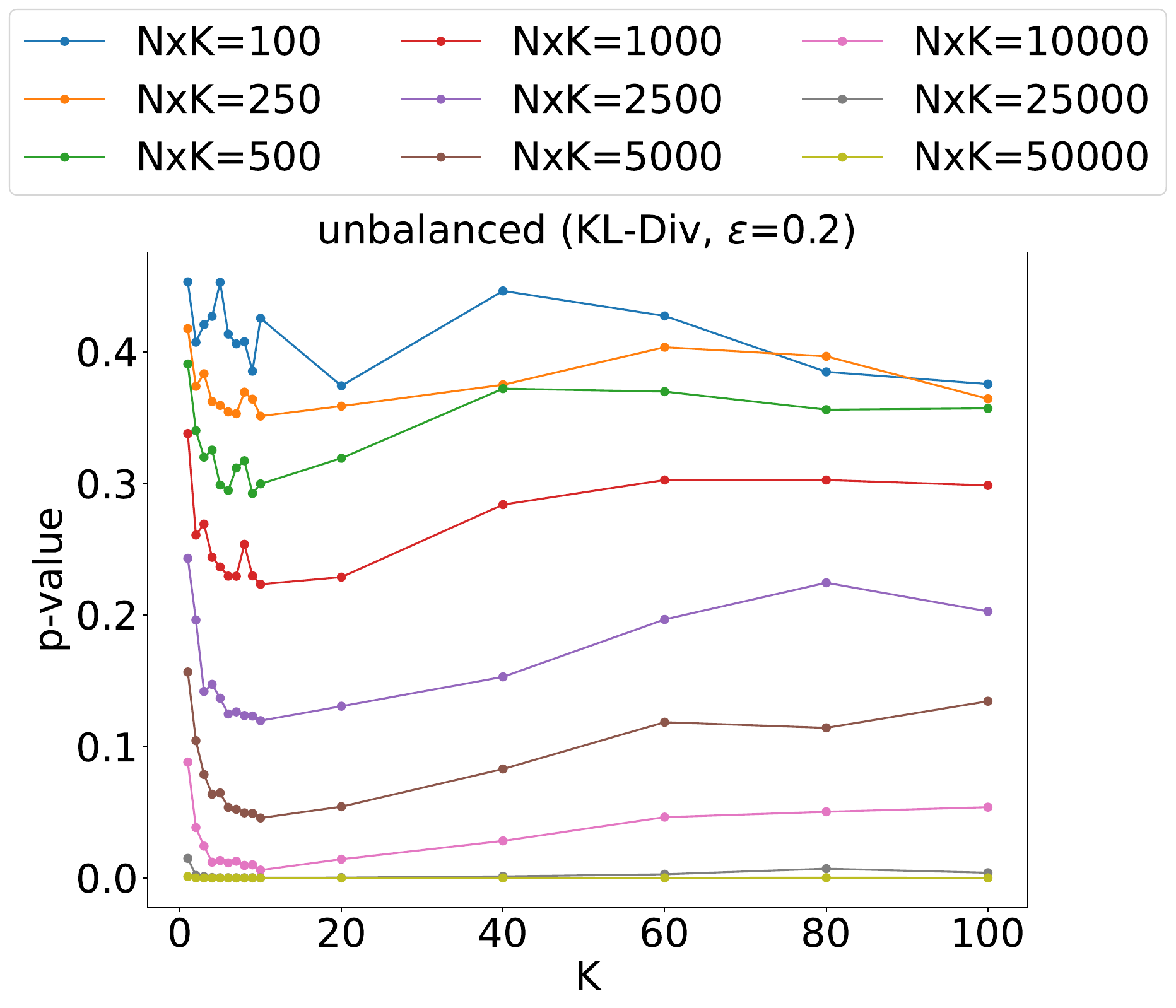}
    \caption{$\epsilon = 0.2$}
    \label{fig:gamma_kl_cat12_e02}
  \end{subfigure} \hfill
  \begin{subfigure}[b]{0.24\linewidth}
    \centering
    \includegraphics[width=\linewidth]{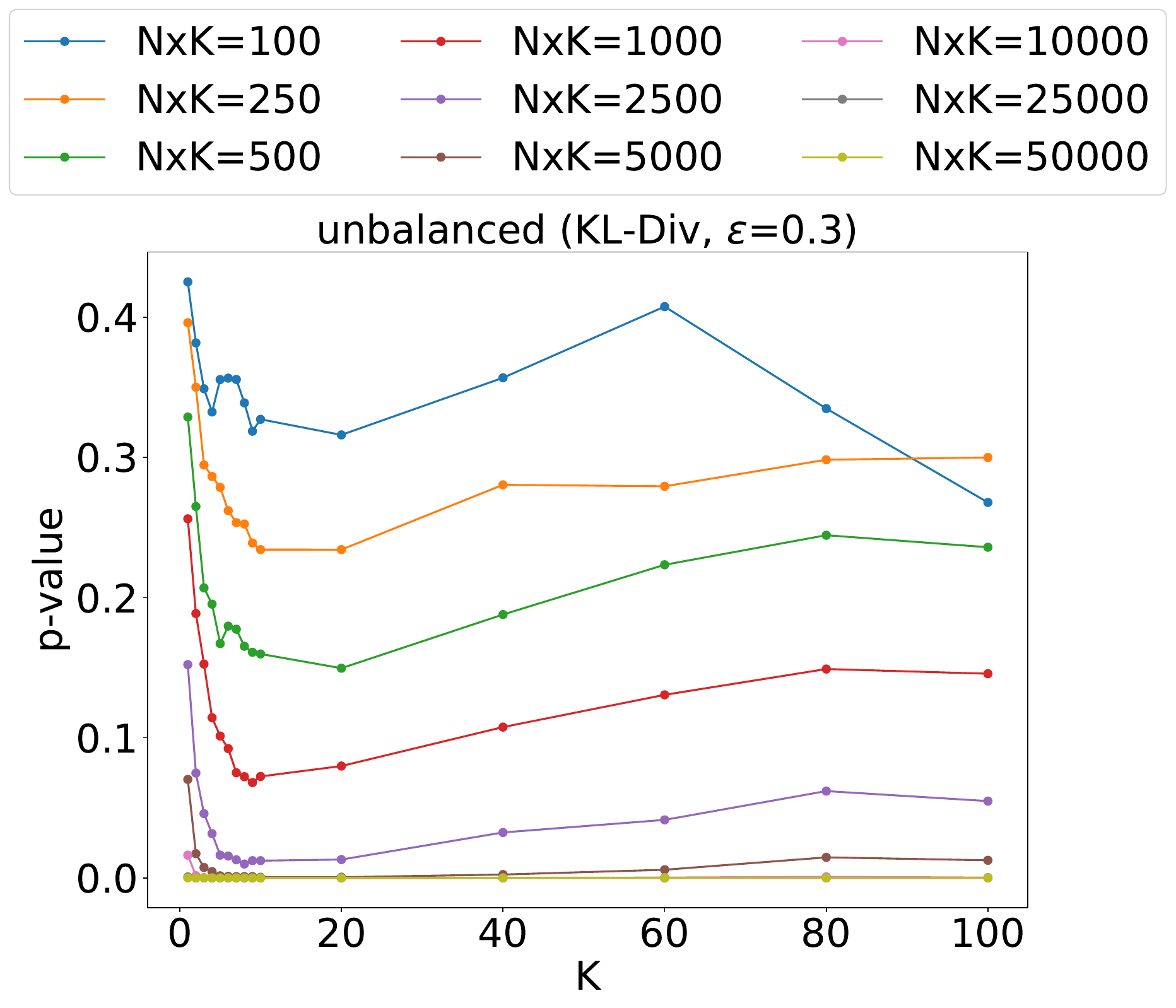}
    \caption{$\epsilon = 0.3$}
    \label{fig:gamma_kl_cat12_e03}
  \end{subfigure} \hfill
  \begin{subfigure}[b]{0.24\linewidth}
    \centering
    \includegraphics[width=\linewidth]{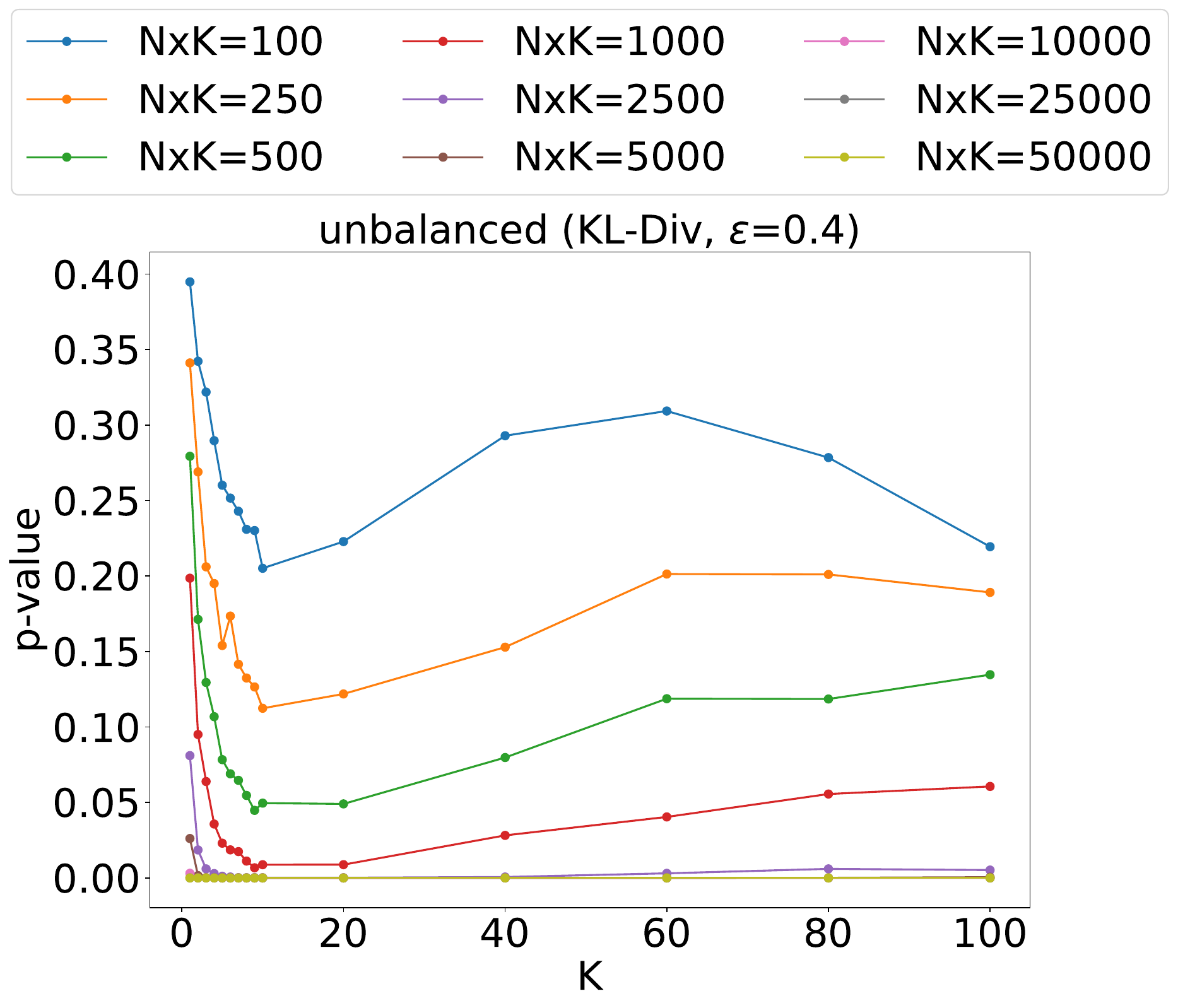}
    \caption{$\epsilon = 0.4$}
    \label{fig:gamma_kl_cat12_e04}
  \end{subfigure}
  \caption{P-value plots for unbalanced alphas with KL-divergence as the metric ($M=12$)}
  \label{fig:gamma_kl_cat12}
\end{figure*}

\begin{figure*}
  \centering
  \begin{subfigure}[b]{0.24\linewidth}
    \centering
    \includegraphics[width=\linewidth]{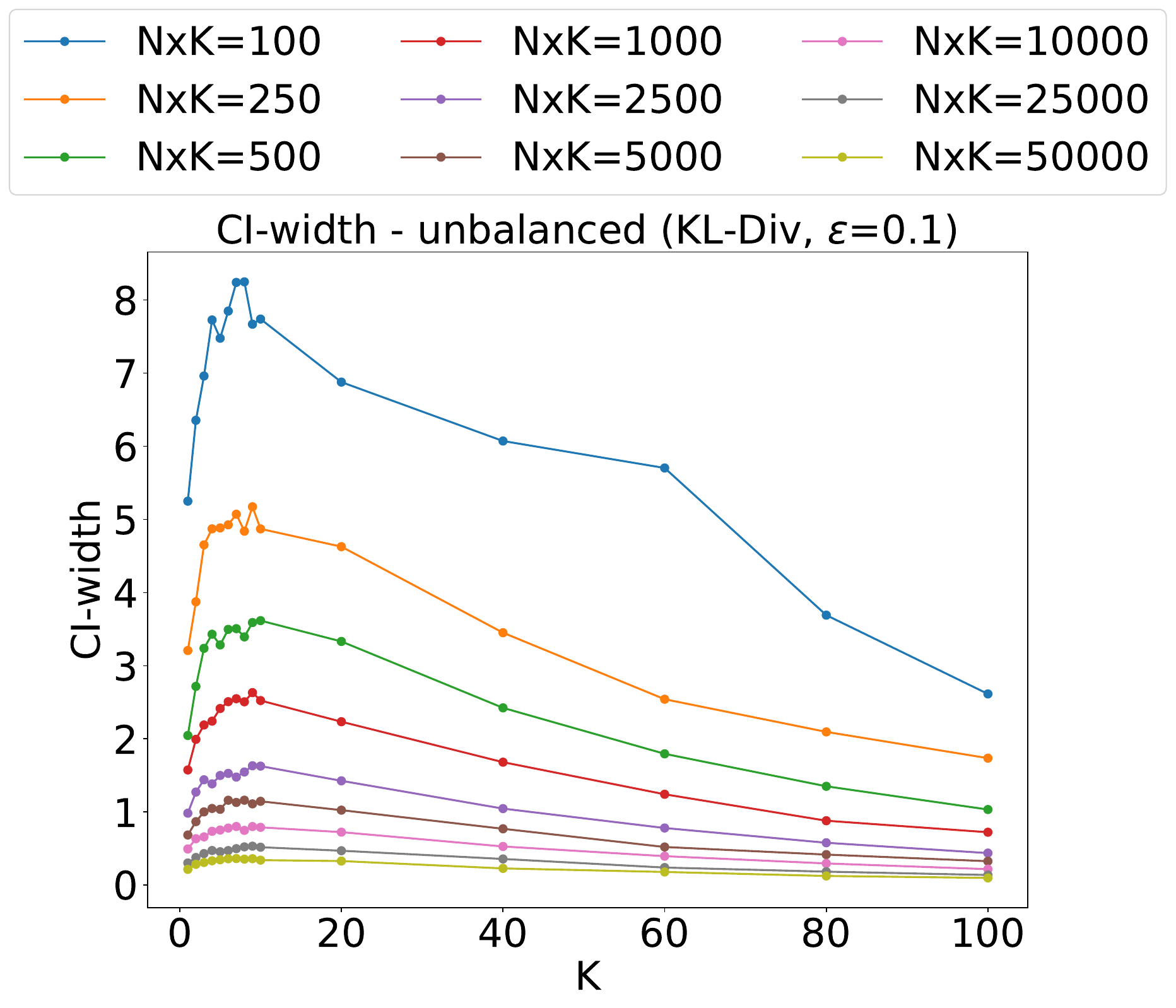}
    \caption{$\epsilon = 0.1$}
    \label{fig:gamma_ci_kl_cat12_e01}
  \end{subfigure} \hfill
  \begin{subfigure}[b]{0.24\linewidth}
    \centering
    \includegraphics[width=\linewidth]{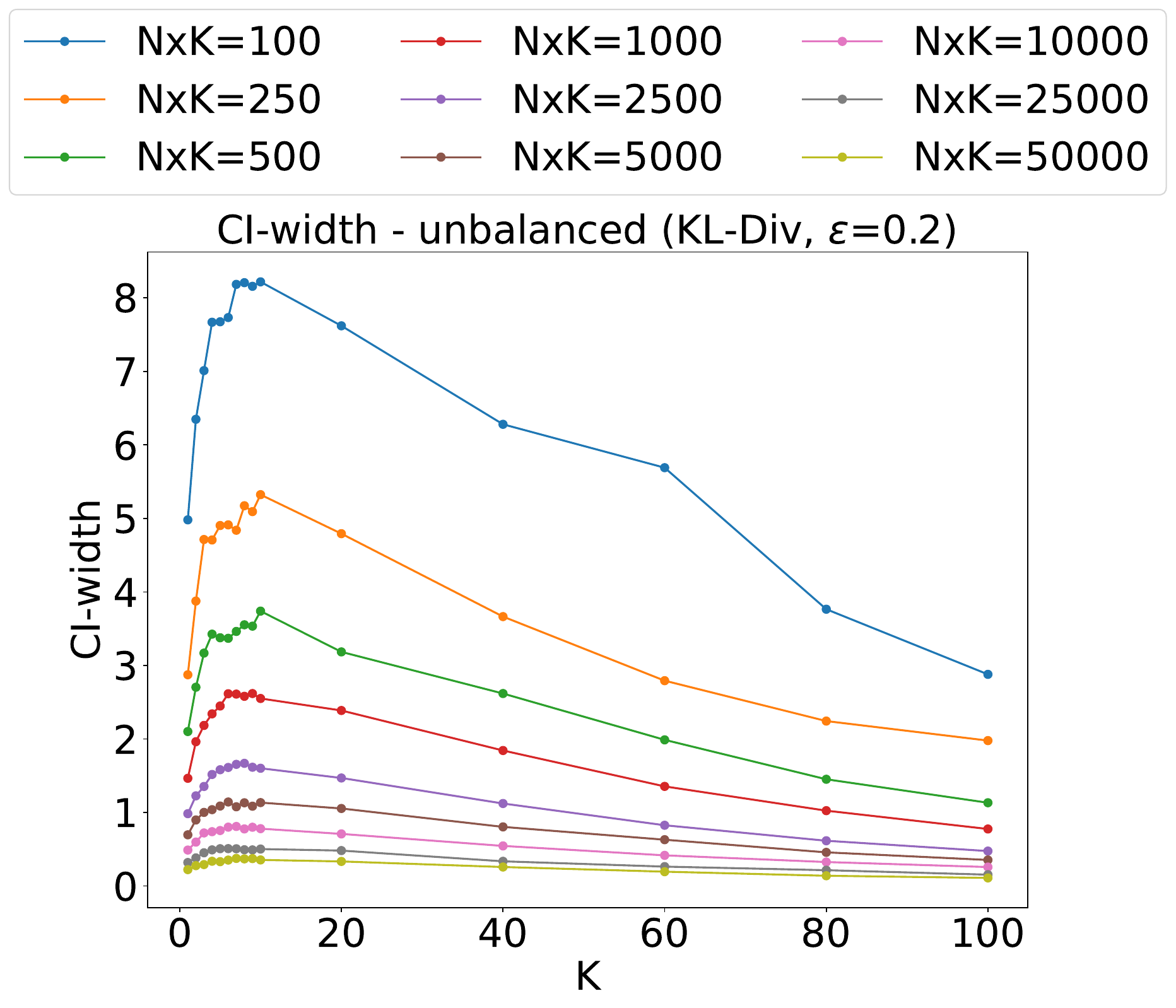}
    \caption{$\epsilon = 0.2$}
    \label{fig:gamma_ci_kl_cat12_e02}
  \end{subfigure} \hfill
  \begin{subfigure}[b]{0.24\linewidth}
    \centering
    \includegraphics[width=\linewidth]{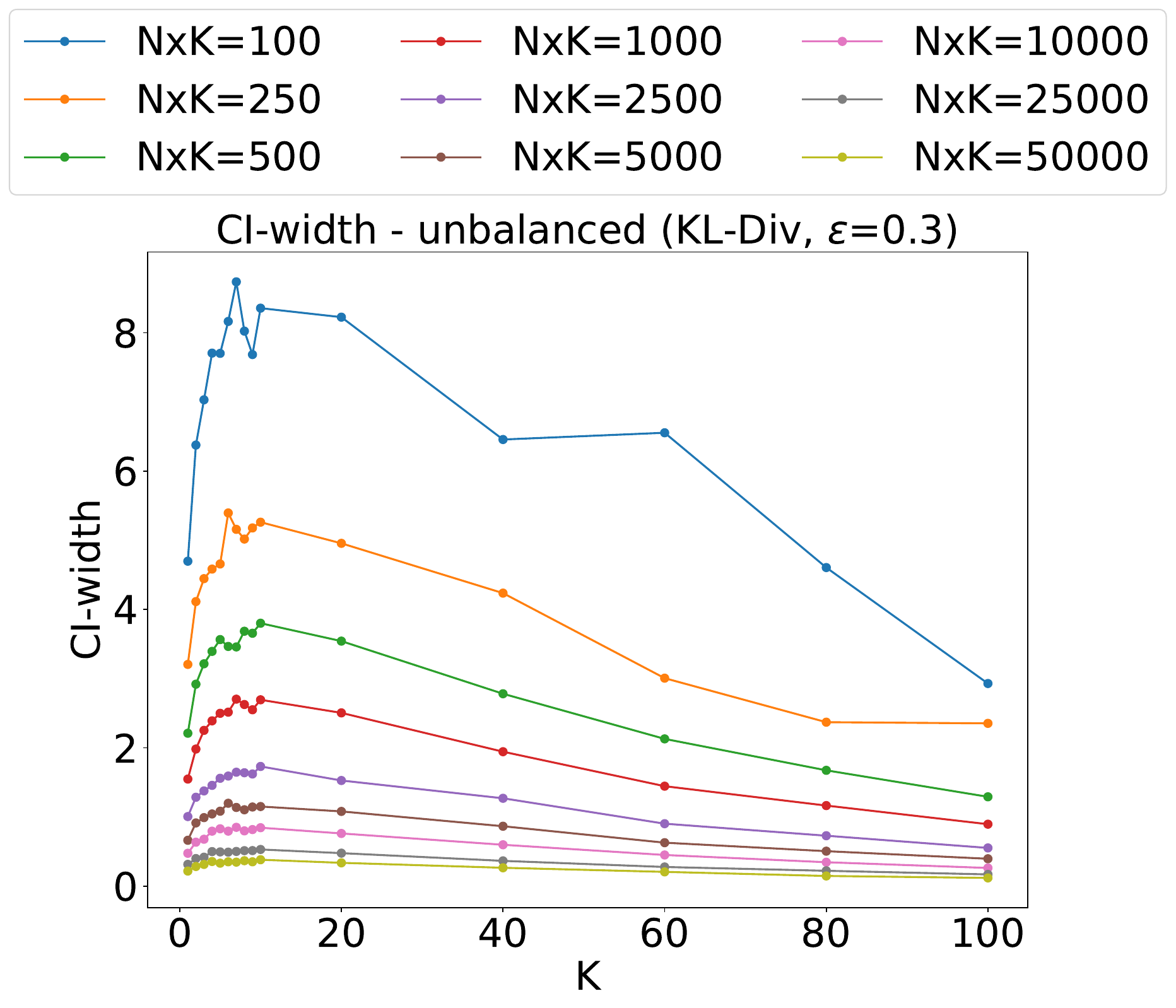}
    \caption{$\epsilon = 0.3$}
    \label{fig:gamma_ci_kl_cat12_e03}
  \end{subfigure} \hfill
  \begin{subfigure}[b]{0.24\linewidth}
    \centering
    \includegraphics[width=\linewidth]{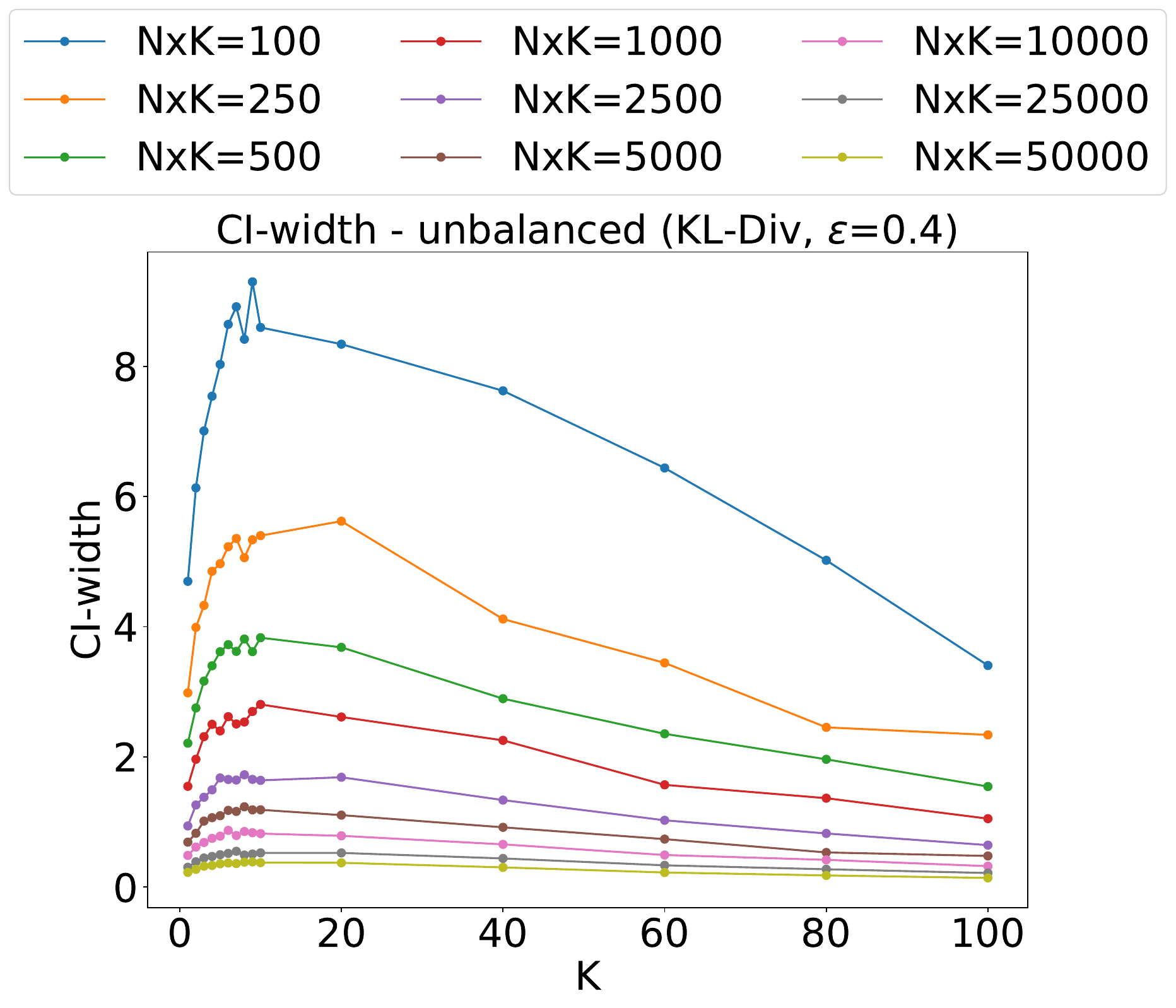}
    \caption{$\epsilon = 0.4$}
    \label{fig:gamma_ci_kl_cat12_e04}
  \end{subfigure}
  \caption{CI-width plots for unbalanced alphas with KL-divergence as the metric ($M=12$)}
  \label{fig:gamma_ci_kl_cat12}
\end{figure*}

\begin{figure*}
  \centering
  \begin{subfigure}[b]{0.24\linewidth}
    \centering
    \includegraphics[width=\linewidth]{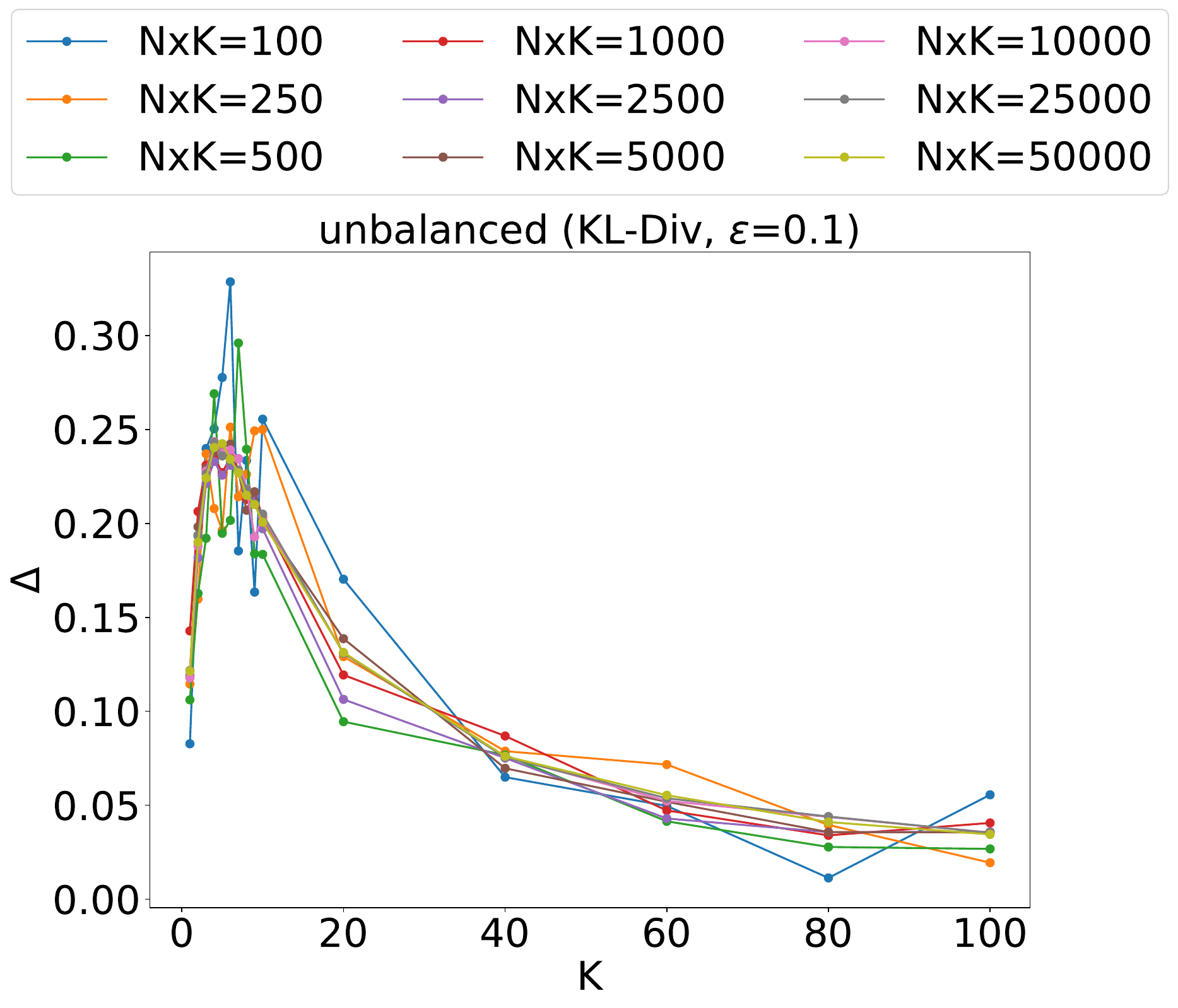}
    \caption{$\epsilon = 0.1$}
    \label{fig:gamma_delta_kl_cat12_e01}
  \end{subfigure} \hfill
  \begin{subfigure}[b]{0.24\linewidth}
    \centering
    \includegraphics[width=\linewidth]{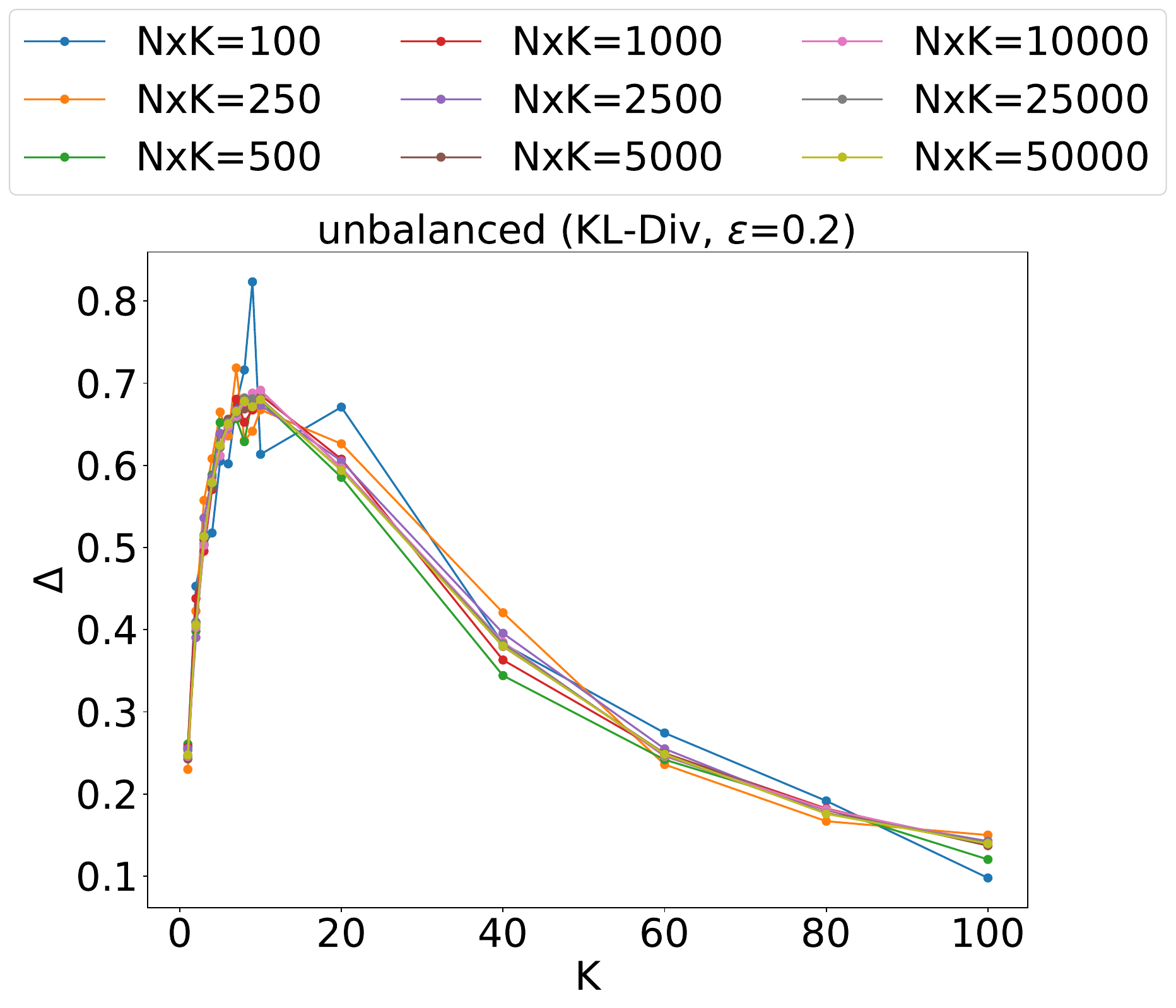}
    \caption{$\epsilon = 0.2$}
    \label{fig:gamma_delta_kl_cat12_e02}
  \end{subfigure} \hfill
  \begin{subfigure}[b]{0.24\linewidth}
    \centering
    \includegraphics[width=\linewidth]{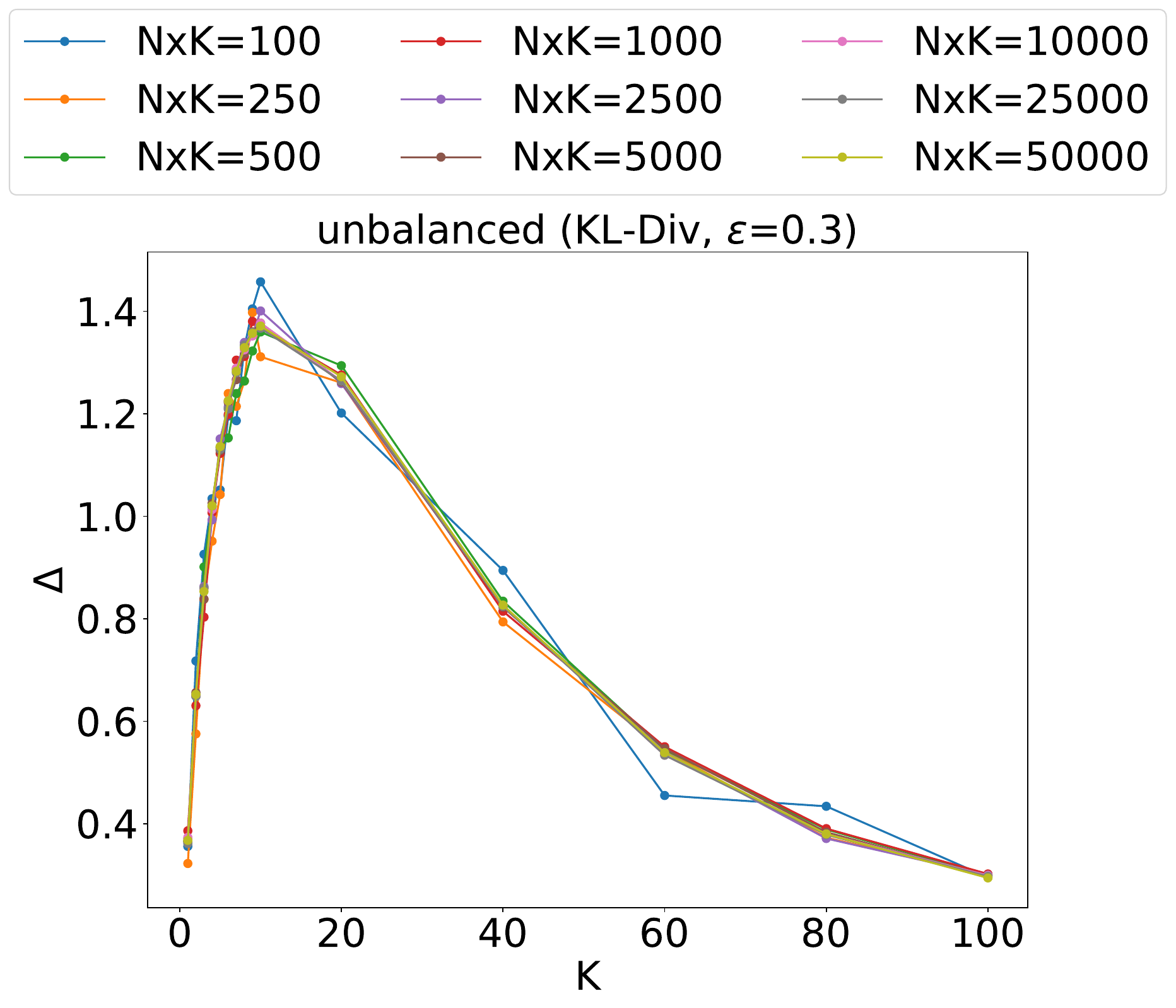}
    \caption{$\epsilon = 0.3$}
    \label{fig:gamma_delta_kl_cat12_e03}
  \end{subfigure} \hfill
  \begin{subfigure}[b]{0.24\linewidth}
    \centering
    \includegraphics[width=\linewidth]{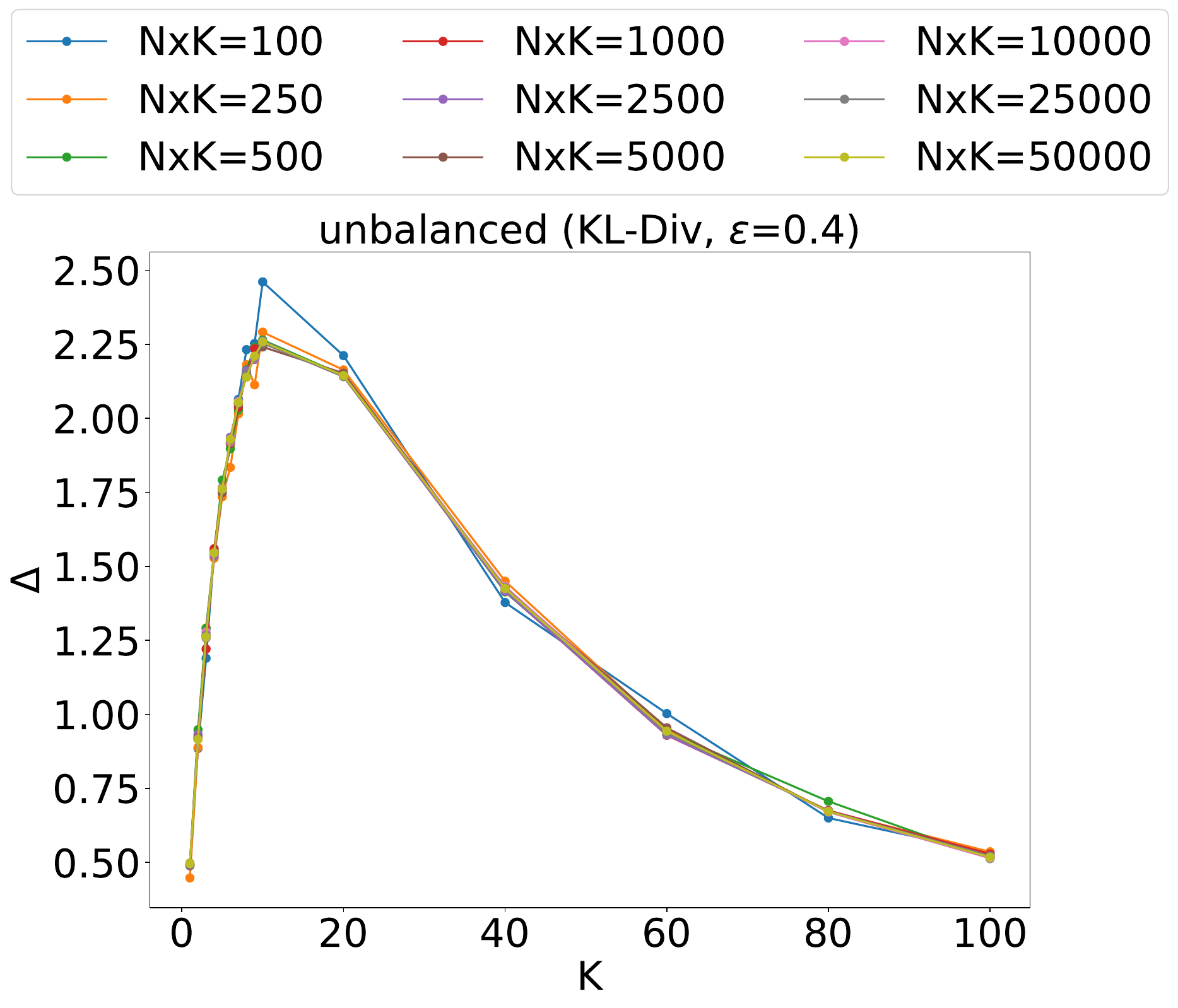}
    \caption{$\epsilon = 0.4$}
    \label{fig:gamma_delta_kl_cat12_e04}
  \end{subfigure}
  \caption{Effect sizes ($\Delta$) for unbalanced alphas with KL-divergence as the metric ($M=12$)}
  \label{fig:gamma_delta_kl_cat12}
\end{figure*}

Figures \ref{fig:gamma_kl_cat2}--\ref{fig:gamma_delta_kl_cat12} show results for \pv s, confidence intervals, and effect sizes of KL-divergence for unbalanced priors on parameter $\alpha$ with different values of $\epsilon$ and increasing $K$ for various $N \times K$. For simulations with unbalanced categories, \pv s\, CI-width, and effect sizes show similar behavior to balanced categories.


\section{Additional Results}

Tables \ref{tab:toxicity_accuracy_nk}--\ref{tab:jobsq3_accuracy_nk} show results for \pv s\, effect sizes, scores of machine $A$, and scores for machine $B$ for different $N$ and $K$ for $N \times K = 2500, \epsilon=0.2$. Looking at the metric scores, it is worth noting that raw scores generally improve as $K$ increases. Taking accuracy as an example, this is easy to see why: the larger $K$ is, the more likely it is that the most likely response is the most common one. So as long as the machine and gold have the same most frequent response, accuracy increases. However, accuracy increases for \emph{both} machines, and this, under our error model, allows for machine B to catch up. And as $N$ decreases, we would expect there to be more variance between individual samples.

\begin{table*}[h]
\centering
\begin{tabular}{cccccc}
\toprule
N & K & Accuracy\_p-value & Accuracy\_$\Delta$ & Accuracy\_A & Accuracy\_B \\
\midrule
2500 & 1 & 0.0671 & 0.0269 & 0.6348 & 0.6079 \\
1250 & 2 & 0.1584 & 0.0223 & 0.7137 & 0.6915 \\
833 & 3 & 0.1189 & 0.0335 & 0.7086 & 0.6752 \\
625 & 4 & 0.1393 & 0.0324 & 0.7446 & 0.7122 \\
500 & 5 & 0.1469 & 0.0354 & 0.7517 & 0.7163 \\
416 & 6 & 0.1769 & 0.0336 & 0.7708 & 0.7373 \\
357 & 7 & 0.1603 & 0.0365 & 0.7797 & 0.7432 \\
312 & 8 & 0.1709 & 0.0366 & 0.7947 & 0.7580 \\
277 & 9 & 0.1927 & 0.0367 & 0.7998 & 0.7631 \\
250 & 10 & 0.2042 & 0.0364 & 0.8096 & 0.7732 \\
125 & 20 & 0.2421 & 0.0380 & 0.8596 & 0.8216 \\
62 & 40 & 0.3131 & 0.0388 & 0.8965 & 0.8577 \\
41 & 60 & 0.3141 & 0.0450 & 0.9130 & 0.8681 \\
31 & 80 & 0.3186 & 0.0485 & 0.9254 & 0.8769 \\
25 & 100 & 0.3703 & 0.0466 & 0.9330 & 0.8864 \\
\bottomrule
\end{tabular}
\caption{Toxicity Accuracy ($N\times K=2500, \epsilon=0.2$).}
\label{tab:toxicity_accuracy_nk}
\end{table*}

\begin{table*}[h]
\centering
\begin{tabular}{cccccc}
\toprule
N & K & Accuracy\_p-value & Accuracy\_$\Delta$ & Accuracy\_A  & Accuracy\_B  \\
\midrule
2500 & 1 & 0.0292 & 0.0357 & 0.5114 & 0.4757 \\
1250 & 2 & 0.0469 & 0.0352 & 0.7096 & 0.6745 \\
833 & 3 & 0.1237 & 0.0354 & 0.6334 & 0.5980 \\
625 & 4 & 0.0767 & 0.0455 & 0.6980 & 0.6525 \\
500 & 5 & 0.1441 & 0.0393 & 0.6803 & 0.6410 \\
416 & 6 & 0.1180 & 0.0438 & 0.7281 & 0.6843 \\
357 & 7 & 0.1463 & 0.0435 & 0.7117 & 0.6682 \\
312 & 8 & 0.1407 & 0.0445 & 0.7457 & 0.7012 \\
277 & 9 & 0.1717 & 0.0423 & 0.7410 & 0.6986 \\
250 & 10 & 0.1712 & 0.0452 & 0.7598 & 0.7146 \\
125 & 20 & 0.2259 & 0.0459 & 0.8158 & 0.7699 \\
62 & 40 & 0.2867 & 0.0480 & 0.8628 & 0.8148 \\
41 & 60 & 0.3004 & 0.0502 & 0.8866 & 0.8363 \\
31 & 80 & 0.3481 & 0.0457 & 0.8972 & 0.8515 \\
25 & 100 & 0.3553 & 0.0540 & 0.9082 & 0.8542 \\
\bottomrule
\end{tabular}
\caption{DICES Accuracy ($N\times K=2500, \epsilon=0.2$).}
\label{tab:dices_accuracy_nk}
\end{table*}

\begin{table*}[h]
\centering
\begin{tabular}{cccccc}
\toprule
N & K & Accuracy\_p-value & Accuracy\_$\Delta$ & Accuracy\_A  & Accuracy\_B  \\
\midrule
2500 & 1 & 0.1314 & 0.0213 & 0.6074 & 0.5860 \\
1250 & 2 & 0.1462 & 0.0194 & 0.8044 & 0.7850 \\
833 & 3 & 0.1450 & 0.0316 & 0.6818 & 0.6502 \\
625 & 4 & 0.1644 & 0.0254 & 0.8009 & 0.7755 \\
500 & 5 & 0.1460 & 0.0365 & 0.7277 & 0.6912 \\
416 & 6 & 0.1883 & 0.0289 & 0.8106 & 0.7817 \\
357 & 7 & 0.1612 & 0.0383 & 0.7585 & 0.7203 \\
312 & 8 & 0.2064 & 0.0316 & 0.8223 & 0.7907 \\
277 & 9 & 0.1713 & 0.0396 & 0.7840 & 0.7444 \\
250 & 10 & 0.2151 & 0.0330 & 0.8330 & 0.8000 \\
125 & 20 & 0.2517 & 0.0375 & 0.8679 & 0.8304 \\
62 & 40 & 0.2920 & 0.0423 & 0.8994 & 0.8571 \\
41 & 60 & 0.3279 & 0.0448 & 0.9155 & 0.8708 \\
31 & 80 & 0.3069 & 0.0489 & 0.9268 & 0.8780 \\
25 & 100 & 0.3294 & 0.0508 & 0.9365 & 0.8857 \\
\bottomrule
\end{tabular}
\caption{D3code Accuracy ($N\times K=2500, \epsilon=0.2$).}
\label{tab:d3code_accuracy_nk}
\end{table*}

\begin{table*}[h]
\centering
\begin{tabular}{cccccc}
\toprule
N & K & Accuracy\_p-value & Accuracy\_$\Delta$ & Accuracy\_A  & Accuracy\_B  \\
\midrule
2500 & 1 & 0 & 0.0700 & 0.5478 & 0.4778 \\
1250 & 2 & 0 & 0.0647 & 0.8399 & 0.7752 \\
833 & 3 & 0.0146 & 0.0534 & 0.7801 & 0.7268 \\
625 & 4 & 09 & 0.0741 & 0.8679 & 0.7938 \\
500 & 5 & 0.0192 & 0.0571 & 0.8601 & 0.8030 \\
416 & 6 & 0.0126 & 0.0593 & 0.9065 & 0.8472 \\
357 & 7 & 0.0238 & 0.0547 & 0.9083 & 0.8536 \\
312 & 8 & 0.0277 & 0.0522 & 0.9333 & 0.8811 \\
277 & 9 & 0.0435 & 0.0477 & 0.9381 & 0.8904 \\
250 & 10 & 0.0430 & 0.0458 & 0.9521 & 0.9063 \\
125 & 20 & 0.1611 & 0.0242 & 0.9916 & 0.9674 \\
62 & 40 & 0.5841 & 0.0083 & 0.9996 & 0.9913 \\
41 & 60 & 0.8571 & 0.0036 & 1 & 0.9964 \\
31 & 80 & 0.9551 & 0.0015 & 1 & 0.9985 \\
25 & 100 & 0.9730 & 0.0011 & 1 & 0.9989 \\
\bottomrule
\end{tabular}
\caption{JobsQ1 Accuracy ($N\times K=2500, \epsilon=0.2$).}
\label{tab:jobsq1_accuracy_nk}
\end{table*}

\begin{table*}[h]
\centering
\begin{tabular}{cccccc}
\toprule
N & K & Accuracy\_p-value & Accuracy\_$\Delta$ & Accuracy\_A  & Accuracy\_B  \\
\midrule
2500 & 1 & 0.1749 & 0.0134 & 0.1492 & 0.1358 \\
1250 & 2 & 0.0845 & 0.0293 & 0.2316 & 0.2023 \\
833 & 3 & 0.0847 & 0.0379 & 0.2626 & 0.2246 \\
625 & 4 & 0.0927 & 0.0408 & 0.2603 & 0.2195 \\
500 & 5 & 0.1109 & 0.0447 & 0.2683 & 0.2237 \\
416 & 6 & 0.1007 & 0.0517 & 0.2964 & 0.2446 \\
357 & 7 & 0.0773 & 0.0593 & 0.3294 & 0.2701 \\
312 & 8 & 0.0742 & 0.0693 & 0.3607 & 0.2914 \\
277 & 9 & 0.0746 & 0.0755 & 0.3791 & 0.3036 \\
250 & 10 & 0.0692 & 0.0807 & 0.3951 & 0.3144 \\
125 & 20 & 0.0472 & 0.1294 & 0.5464 & 0.4170 \\
62 & 40 & 0.0251 & 0.2049 & 0.7391 & 0.5342 \\
41 & 60 & 0.0192 & 0.2455 & 0.8483 & 0.6027 \\
31 & 80 & 0.0182 & 0.2633 & 0.9099 & 0.6466 \\
25 & 100 & 0.0161 & 0.2752 & 0.9477 & 0.6725 \\
\bottomrule
\end{tabular}
\caption{JobsQ3 Accuracy ($N\times K=2500, \epsilon=0.2$).}
\label{tab:jobsq3_accuracy_nk}
\end{table*}

\end{document}